\documentclass{report}

\usepackage[final]{neurips_2021_nonotice}

\usepackage[utf8]{inputenc} 
\usepackage[T1]{fontenc}    
\usepackage[backref=page]{hyperref}       
\usepackage{url}            
\usepackage{booktabs}       
\usepackage{amsfonts}       
\usepackage{nicefrac}       
\usepackage{microtype}      
\usepackage{xcolor}         

\title{The Nonlinearity Coefficient - \\ A Practical Guide to Neural Architecture Design}

\author{%
  George Philipp \\
  School of Computer Science\\
  Carnegie Mellon University\\
  Pittsburgh, PA 15213 \\
  \texttt{george.philipp@email.de} \\
}

\usepackage{natbib}
\usepackage{amssymb}
\usepackage{amsmath}
\usepackage{amsthm}
\usepackage{color}
\usepackage{bbm}
\usepackage{graphicx}
\usepackage[export]{adjustbox}
\usepackage{tablefootnote}
\usepackage{wrapfig}
\usepackage[linesnumbered]{algorithm2e}
\usepackage{euscript}
\usepackage{float}
\usepackage{enumitem}
\usepackage{longtable}
\usepackage{array}
\usepackage{booktabs}
\usepackage{pifont}

\newcommand*\rot{\rotatebox{90}}
\newcommand{\xmark}{\ding{55}}

\makeatletter
\renewcommand\@makefntext[1]{\leftskip=2em\hskip-2em\@makefnmark#1}
\makeatother

\newcommand{\finding}[1]{#1}

\newcommand{\contrib}{{\textcircled +}}
\newcommand{\backgr}{{\textcircled b}}

\makeatletter
\newtheorem*{rep@theorem}{\rep@title}
\newcommand{\newreptheorem}[2]{%
\newenvironment{rep#1}[1]{%
 \def\rep@title{#2 \ref{##1}}%
 \begin{rep@theorem}}%
 {\end{rep@theorem}}}
\makeatother

\usepackage{amsmath,amsfonts,bm}

\def\eqref#1{equation~\ref{#1}}

\def\1{\bm{1}}

\DeclareMathAlphabet{\mathsfit}{\encodingdefault}{\sfdefault}{m}{sl}
\SetMathAlphabet{\mathsfit}{bold}{\encodingdefault}{\sfdefault}{bx}{n}

\newcommand{\Cov}{\mathrm{Cov}}

\DeclareMathOperator{\sign}{sign}
\DeclareMathOperator{\Tr}{Tr}

\DeclareMathOperator{\midOp}{mid}

\DeclareMathOperator{\clipin}{clipin}
\DeclareMathOperator{\clipout}{clipout}

\allowdisplaybreaks
\newcommand{\vece}{\vec{\epsilon}}
\newcommand{\limd}{\lim_{d_\text{MF}\text{ a.s.}}}
\newcommand{\lime}{\lim_{\vec{\epsilon},N}}
\newcommand{\limb}{\lim_{B}}
\newcommand{\limbn}{\lim_{B,N}}
\newcommand{\limbas}{\lim_{B\text{ a.s.}}}
\newcommand{\limep}{\lim_{\vec{\epsilon}}}

\theoremstyle{plain}
\newtheorem{theorem}{Theorem}
\newreptheorem{theorem}{Theorem}
\newtheorem{proposition}{Proposition}
\newreptheorem{proposition}{Proposition}
\newtheorem{lemma}{Lemma}
\newreptheorem{lemma}{Lemma}

\newreptheorem{corollary}{Corollary}
\newtheorem{backgroundtheorem}{Background Theorem}
\newtheorem{backgroundcorollary}{Background Corollary}

\theoremstyle{definition}
\newtheorem{definition}{Definition}
\newtheorem{metricDefinition}{Metric definition}
\newtheorem{assumption}{Assumption}
\newreptheorem{assumption}{Assumption}
\newtheorem{npc}{ZSAD guideline definition}
\newreptheorem{npc}{ZSAD guideline definition}

\usepackage[nottoc,notlot,notlof]{tocbibind}

\begin{document}

\maketitle

\begin{abstract}

In essence, a neural network is an arbitrary differentiable, parametrized function. Choosing a neural network architecture for any task is as complex as searching the space of those functions. For the last few years, `neural architecture design' has been largely synonymous with `neural architecture search' (NAS), i.e. brute-force, large-scale search. NAS has yielded significant gains on practical tasks. However, NAS methods end up searching for a local optimum in architecture space in a small neighborhood around architectures that often go back decades, based on CNN or LSTM.

In this work, we present a different and complementary approach to architecture design, which we term {\it zero-shot architecture design} (ZSAD). We develop methods that can predict, without any training, whether an architecture will achieve a relatively high test or training error on a task after training. We then go on to explain the error in terms of the architecture definition itself and develop tools for modifying the architecture based on this explanation. This confers an unprecedented level of control on the deep learning practitioner. They can make informed design decisions before the first line of code is written, even for tasks for which no prior art exists.

Our first major contribution is to show that the {\it degree of nonlinearity} of a neural architecture is a key causal driver behind its performance, and a primary aspect of the architecture's {\it model complexity}. We introduce the {\it nonlinearity coefficient} (NLC), a scalar metric for measuring nonlinearity. Via extensive empirical study, we show that the value of the NLC in the architecture's randomly initialized state {\it before} training is a powerful predictor of test error {\it after} training and that attaining a right-sized NLC is essential for attaining an optimal test error. The NLC is also conceptually simple, well-defined for any feedforward network, easy and cheap to compute, has extensive theoretical, empirical and conceptual grounding, follows instructively from the architecture definition, and can be easily controlled via our {\it nonlinearity normalization} algorithm. We argue that the NLC is the most powerful scalar statistic for architecture design specifically and neural network analysis in general. Our analysis is fueled by mean field theory, which we use to uncover the {\it meta-distribution} of layers.

Beyond the NLC, we uncover and flesh out a range of metrics and properties that have a significant explanatory influence on test and training error. We go on to explain the majority of the error variation across a wide range of randomly generated architectures with these metrics and properties. We compile our insights into a practical guide for architecture designers, which we argue can significantly shorten the trial-and-error phase of deep learning deployment.

Our results are grounded in an experimental protocol that exceeds that of the vast majority of other deep learning studies in terms of carefulness and rigor. We study the impact of e.g. dataset, learning rate, floating-point precision, loss function, statistical estimation error and batch inter-dependency on performance and other key properties. We promote research practices that we believe can significantly accelerate progress in architecture design research.

\end{abstract}





\renewcommand{\baselinestretch}{0.75}\normalsize
\tableofcontents
\renewcommand{\baselinestretch}{1.0}\normalsize

\chapter{Introduction} \label{introductionChapter}

Neural networks have been highly successful on a range of machine learning tasks recently. To a large degree, this success is based on the power of gradient methods. As long as we specify an architecture and a parameter initialization scheme, we can use algorithms like stochastic gradient descent to set millions of parameter components simultaneously and efficiently. Unfortunately, there exists no known neural architecture that when trained with gradient methods performs well on all machine learning tasks. Therefore, we are faced with the challenge of choosing a high-performing neural architecture for any given task; this challenge is broadly known as `neural architecture design'.

Despite the ubiquity of deep learning in modern research, there is a lack of robust and general methods that, given only the definition of an architecture, can predict and explain its performance after training. In this work, we develop methods that can predict performance directly from the architecture definition without conducting training. We term this approach {\it zero-shot architecture design (ZSAD)}. It has immediate practical utility, as architectures that can be predicted to perform poorly can be discarded before training and do not need to be considered further.

While giving a complete theory of the foundations of neural architecture design is beyond the scope of this work, we compile an unprecedented repository for understanding (i) what drives architecture performance, (ii) why popular neural architectures are designed the way they are and (iii) how to choose or design neural architectures for a novel task in any domain, even if no prior art exists for that task, as well as improve existing architectures.

While this work contains many individual contributions, they can be grouped into the following overarching contributions.

\begin{enumerate}
\item Establish \colorbox{green}{the {\it nonlinearity coefficient} (NLC)} as one of the most important metrics for neural architecture design, as the standard practical measure of the {\it degree of nonlinearity} and {\it expressivity} of a neural architecture, and hence as a primary measure of the {\it model complexity} of a neural architecture and a primary tool for neural network analysis in general. It (i) is well-defined for any network with a single input and output, (ii) is easy and cheap to compute, (iii) is capable of predicting the test error of an architecture without training, (iv) can be instructively and accurately estimated from the architecture definition, (v) can be easily controlled by the designer via our nonlinearity normalization algorithm, (vi) is conceptually simple, (vii) can be shown to be a measure of a network's degree of nonlinearity through many lenses, as well as a measure of expressivity, noise sensitivity and model complexity, (viii) is grounded in mean field theory and has many unique theoretical properties, (ix) can be controlled independently of other core performance drivers like width, parameter dimensionality and orthogonality and (x) is robust to a wide range of confounders such as data distribution, initial parameter draw and change of scale. Via the NLC, we show that, contrary to prior belief, high model complexity is harmful to generalization but not trainability in neural networks. Main chapter: \ref{nlcChapter}
\item Create \colorbox{green}{a practical guide to neural architecture design} by enumerating {\it what can go wrong} when building an architecture. We assemble an arsenal of guidelines that (i) explains the majority of test and training error variation across a wide range of randomly generated architectures without training, (ii) explains much of the utility of some of the most popular architecture building blocks, like ReLU, batch normalization and skip connections and (iii) makes at least designing fully-connected architectures relatively foolproof. We both introduce novel guidelines and flesh out previously known guidelines. Main chapters: \ref{beyondNlcChapter}, \ref{surveyChapter}
\item Develop \colorbox{green}{a mean field theory of meta-distributions}. We prove and empirically show that, under practical conditions, the value of a fully-connected layer in the randomly initialized state is {\it meta-Gaussian meta-distributed}, i.e. the distribution of each neuron as induced by the data is Gaussian and has a Gaussian random mean that is induced by the parameter initialization scheme. We use this to derive simple recursive calculation rules for the mean field limits of many properties including layer value magnitude, biasedness of neurons and the NLC in popular fully-connected architectures. These rules reveal that the NLC of such an architecture can be estimated simply and instructively from the nonlinearity of its activation functions without ever evaluating the architecture. We uncover the property of {\it Gaussian stability}, a key prerequisite for the practical predictiveness of mean field theory in general and a driver of architecture performance. Main chapter: \ref{meanFieldNnaChapter}
\item Introduce \colorbox{green}{{\it nonlinearity normalization} (nlnorm)}, a simple algorithm that allows the architecture designer to (i) control the architecture's NLC by tuning it like a hyperparameter, (ii) ensure that neuron values are not too large or small (thereby ensuring `scale stability') and (iii) prevent the means of neuron distributions from deviating too far from zero (thereby avoiding `neuron bias'). This achieves three of our core design guidelines. We show that nlnorm can greatly reduce the test error of suboptimal architectures with minimal modification. Thereby, nlnorm can capture a large fraction of the performance boost provided by many of the most popular building blocks, like batch normalization, skip connections, and specific activation functions. It captures several aspects of their utility. Main chapter: \ref{nlnormChapter}
\item \colorbox{green}{A more scientific approach to architecture design research based on well-defined metrics}. \linebreak A key distinguishing feature of some of our guidelines, and especially the NLC, from much prior work is that they are based on well-defined, quantitative {\it metrics} that can be evaluated for any network. This stands in contrast with ill-defined guidelines like ``avoid exploding gradients'', ``choose an architecture on the edge of chaos'' and ``use an appropriate depth''. Based on our analysis, we argue that metrics have several key advantages. (i) Generalizability: Metrics automatically generalize to any network for which they are mathematically valid. (ii) Lack of ambiguity: Ill-defined guidelines can behave very differently depending on how they are quantified. Metrics are already quantitative. (iii) Accountability: It is possible to determine unambiguously when a metric fails to predict performance. (iv) Improvability: Metrics can be replaced with better metrics via transparent comparison. (v) Democratization: Metrics open up machine learning research to non-experts. Main chapter: \ref{relatedWorkChapter}
\item Show the value of \colorbox{green}{a more scientific approach to experimental protocols}. At a high level, we advocate (i) controlling confounders, such as parameter dimensionality and loss function, (ii) independently and exhaustively tuning key hyperparameters, such as learning rate and the NLC, (iii) covering a broad and unpredictable range of deep learning pipelines and (iv) managing computational errors such as rounding error and estimation error. We derive specific recommendations and show how our meticulous experimental protocol underpins the results of our empirical studies. We believe these empirical studies can serve as a guide for the design of analytical deep learning studies in general. Main chapter: \ref{empiricalStudiesChapter}
\end{enumerate}

\section{Reader's guide}

The results presented in this work are highly interdependent. While we gave the main chapter(s) for each overarching contribution above, all chapters contribute to (almost) all overarching contributions. The drawback is that in order to derive maximum utility from any part of this work, it is necessary to read the rest. The advantage is that the overall utility is maximized.

In general, we order our material to minimize the need for understanding results presented on later pages in order to fully appreciate results that come before. Chapter \ref{backgroundChapter} can be regarded as the logical starting point of this work, where we develop foundational concepts like `neural networks' and `gradient-based training', define layer operations and training algorithms, and introduce a range of notation and terminology. Building from this, in section \ref{designOverviewSection}, we discuss the historical and current state of deep learning and neural architecture design; and motivate the need for an approach to architecture design that is different from and complementary to neural architecture search (NAS). Then in section \ref{zeroShotDesignSection}, we introduce our approach: zero-shot architecture design (ZSAD). Motivated by the intricate challenges of neural architecture design, we detail the meticulous design of our empirical studies in chapter \ref{empiricalStudiesChapter}. These empirical studies underpin our analysis, which is presented in the ``main matter'' from chapter \ref{nlcChapter} through \ref{relatedWorkChapter}. Finally, we prove the theoretical results given in chapter \ref{nlcChapter} in chapter \ref{finiteNetChapter} and the theoretical results given in chapter \ref{meanFieldNnaChapter} in chapter \ref{mfntChapter}. For novice readers, we recommend proceeding in this order.

While we believe that reading this work in its entirety is valuable to almost any reader, we expect that most readers will want to continue with the general summary, overview and outlook in section \ref{thesisOverviewSection}. Among other things, this section lists all key contributions of this work. We place a very large number of cross-references not just in that section, but throughout this work. This should allow readers to find and jump to those parts they are most interested in. We provide the most salient notation, terminology and conventions from chapter \ref{backgroundChapter} and sections \ref{designOverviewSection} and \ref{zeroShotDesignSection} in summary section \ref{notationSummarySection}. We provide the most salient information about the design and presentation of our empirical studies from chapter \ref{empiricalStudiesChapter} in summary section \ref{metricsSummarySection}. We recommend reading at least sections \ref{notationSummarySection} and \ref{metricsSummarySection} before any later chapters. Information from those sections, while sometimes repeated later where applicable, is often assumed implicitly.

\section{General summary, overview and outlook} \label{thesisOverviewSection}

We motivate our work (section \ref{designOverviewSection}) based on the limitations of the historical processes which brought about the current state-of-the-art in neural architectures design, as well as the limitations of automated design based on training many architectures, which is known as neural architecture search (NAS). Our work is also motivated by the opportunity presented by the emergence of deep learning across the machine learning spectrum. Because neural architectures representing black-box, parametrized, differentiable functions are applied across a wide range of machine learning domains and settings, novel insights in model design can be replicated across the entire spectrum, which massively boosts their utility. We formalize deep learning as the {\it functional-gradient paradigm} in figure \ref{boxFunctionalLearning}.

The central goal of this work is to formalize and advocate a different and complementary approach to neural architecture design based on general, predictive, explanatory and, ideally, well-defined principles that can be applied without conducting any training. We term it `zero-shot architecture design' (ZSAD; section \ref{zeroShotDesignSection}), and each design principle a `ZSAD guideline'. Please see figure \ref{boxZSAD} for our full definition. The recent success of deep learning stands in contrast with a lack of general, meaningful and well-defined ZSAD guidelines (section \ref{darkArtSection}, chapter \ref{relatedWorkChapter}).

The modus operandi of this work is to find and flesh out ZSAD guidelines (section \ref{zsadGuidelineSection}) which encapsulate properties of (i) the definition of an architecture or (ii) an architecture's randomly initialized state. This makes it possible to determine whether an architecture follows one of these guidelines without conducting training. In contrast, a ZSAD guideline must not encapsulate a property of the final state after training, as we do not have access to such information during the manual design stage (section \ref{generalizationMeasuresSection}). 

We focus specifically on guidelines that apply across datasets and task domains while being themselves as data-independent as possible in their formulation (section \ref{moduloDataSection}, \ref{guidelineDataSection}). We also focus on guidelines that are capable of predicting test error and / or training error after training. These are the most important aspects of architecture performance in academic ML. Hence, our guidelines allow us to decide which architectures to train or study further. All of our analysis flows from the search for ZSAD guidelines, and is highly interdependent.

\begin{itemize}
\item We introduce novel ZSAD guidelines and flesh out existing ones.
\item We demonstrate a wide range of properties for our guidelines, including their ability to predict error after training, along the lines of the utility criteria given in figure \ref{boxNPM}. 
\item We uncover the deeper meaning behind the predictive power of our guidelines and thereby advance the fundamental understanding of neural architectures, e.g. of their model complexity.
\item We extend mean field theory to ground our guidelines theoretically and to understand and predict which architectures follow our guidelines without conducting forward propagation or backpropagation.
\item We develop tools for controlling whether an architecture follows our guidelines.
\item We explain popular design strategies and building blocks such as ReLU, batch normalization or skip connections based on how they enable architectures to follow our guidelines.
\item We develop research practices to make our guidelines more scientifically rigorous and advocate the general application of those practices.
\end{itemize}


Each of the next 6 subsections is dedicated to one of the overarching contributions given at the beginning of this chapter. Throughout this section, when we reference an individual contribution based on original concepts, theory or empirical analysis, we mark it with \contrib. When we reference a previously known concept or result (``background''), we use \backgr. When we discuss or interpret a result, we do not use a marker. (The distinction between the three cases can be somewhat fluid and subjective.) Finally, we summarize limitations of our work in subsection \ref{limitationsSummarySection} and opportunities for future work in subsection \ref{futureWorkSection}.

We validate most of our individual contributions empirically by using architectures trained with a meticulous training protocol. The architectures we trained can largely be grouped into two empirical studies: study A, based on fully-connected architectures trained on CIFAR10, MNIST and waveform-noise; and study B, based on convolutional architectures trained on CIFAR10. We detail our studies in chapter \ref{empiricalStudiesChapter}, summarize them in section \ref{metricsSummarySection}, and discuss their most important aspects in section \ref{experimentsSummarySection}. We detail limitations in section \ref{limitationsSection} and summarize them in section \ref{limitationsSummarySection}. We also refer to these empirical studies as ``our studies'' for short.

\subsection{The nonlinearity coefficient} \label{nlcSummarySection}


We introduce the nonlinearity coefficient (NLC; section \ref{nlcDefinitionSection}) \contrib.

$$NLC(f, \mathcal{D}) = \sqrt{\frac{\mathbb{E}_x\Tr(\mathcal{J}(x)\Cov_x\mathcal{J}^T(x))}{\Tr(\Cov_f)}}$$

Here, $f : \mathbb{R}^{d_\text{in}} \rightarrow \mathbb{R}^{d_\text{out}}$ is the network, $\mathcal{D}$ is the input distribution, $x \sim \mathcal{D}$ is the row input vector, $\mathcal{J}(x) = \frac{df(x)}{dx} \in \mathbb{R}^{d_\text{out} \times d_\text{in}}$ is the Jacobian, $\Cov_x = \mathbb{E}_x(x-\bar{x})^T(x-\bar{x})$ is the input covariance matrix, $\Cov_f = \mathbb{E}_x(f(x)-\bar{f})^T(f(x)-\bar{f})$ is the output covariance matrix and $\Tr$ is the trace.

We also introduce the accompanying ZSAD guideline (section \ref{nlcPredictiveSection}).

\begin{repnpc}{npcNLC}
`Use an appropriate NLC' requires $1 \le NLC \le 5$. \contrib
\end{repnpc}

That is, given an input distribution $\mathcal{D}$, we generally recommend choosing an architecture such that $1 \le NLC(f, \mathcal{D}) \le 5$ holds, where $f$ is the architecture's randomly initialized state. This ensures that the `degree of nonlinearity' of the architecture is not excessive. When we refer to the value of a metric not based on error or loss for an architecture, such as the NLC value, we generally mean the value in the initial state.

We now summarize the properties of the NLC by the utility criteria of figure \ref{boxNPM}. To our knowledge, the NLC fulfills these criteria at least to the degree of any other ZSAD guideline. It is important to note that many of the properties we discuss rely on `Gaussian stability' (section \ref{meanFieldSummarySection} below) and hold especially in an architecture's randomly initialized state, which is of primary importance (sections \ref{zsadGuidelineFramingSection}, \ref{generalizationMeasuresSection}).

\paragraph{Criterion \ref{criterionWellDefined}: Well-definedness} We formalize the concept of the {\it functional-gradient paradigm} (section \ref{functionalGradientSection} and figure \ref{boxFunctionalLearning}) \contrib, which is an abstract definition of deep learning as it exists in the year 2020. It implies that a neural architecture can be an arbitrary function with a trainable parameter for which an accurate local linear approximation around a given parameter value can be found. We point out that the NLC makes no additional structural assumptions about the network (sections \ref{nlcFunctionalGradientSection}, \ref{nlcComputeSection}) \contrib, such as the presence of neurons or layers, and thus builds directly upon the functional-gradient paradigm. It only makes a few technical assumptions, which we show to be mild (section \ref{nlcDefinitionSection}) \contrib. Hence, the NLC is valid for effectively any network with a single data input and output. It can be formally generalized to networks that use batch normalization, and hence do not represent a function of a single input (section \ref{nlcComputeSection}) \contrib.

\paragraph{Criterion \ref{criterionComputable}: Computability (section \ref{nlcComputeSection})} The NLC can be implemented in a few lines of code using a deep learning software framework \contrib. The denominator requires only forward-propagation of regular inputs from the dataset and taking neuron standard deviations \contrib. By ``piggybacking'' on the forward propagation that already takes place during training and when computing error, the need for additional forward propagation can be eliminated and the computational cost further reduced \contrib. In addition to the forward propagation of inputs, the numerator requires only backpropagating Gaussian noise in place of the gradient of the loss function \contrib. This can be easily achieved via automatic differentiation built into popular deep learning software frameworks. 

Because the NLC formally depends on the input distribution, it has to be computed via statistical estimation. Because it uses expectation and standard deviation operators, the canonical statistical estimator is stable even for small datasets (sections \ref{nlcRobustDataSection}, \ref{metricComputationSection}, \ref{nlcComputeSection}) \contrib. The NLC does not suffer from floating-point rounding error as long as the network output itself does not suffer from rounding error that exceeds its mathematical variation \contrib. The NLC can be applied to networks with batch normalization without modifying the program used to compute it or significantly compromising statistical efficiency \contrib. 

\paragraph{Criterion \ref{criterionPredictive}: Predictiveness} The NLC, evaluated in the architecture's randomly initialized state, is predictive of test error after training and attaining a right-sized NLC is essential for minimizing test error (section \ref{nlcPredictiveSection}) \contrib. When controlling for neuron bias, Gaussian stability and noise stability (summarized in sections \ref{practicalGuideSection}, \ref{meanFieldSummarySection} below), the NLC explains the vast majority of test error variation across our architectures (section \ref{beyondNlcSummarySection}) \contrib.

We observed that architectures can only attain an optimal test error with an initial NLC between 1 and 5, though this does not guarantee close-to-optimal or even better-than-random test error (section \ref{nlcPredictiveSection}) \contrib. Our analysis suggests that this range is widely applicable, as we further discuss below in section \ref{initialFinalSummarySection}. Unfortunately, we were unable to narrow the recommended NLC range beyond [1,5], as different types of architectures appear to require different NLC values and no specific value, either before or after training, leads to close-to-optimal test error for all types (section \ref{nlnormBestNLCSection}) \contrib.

While architectures with NLC greater $10^5$ attain a random test error \contrib, architectures with much larger initial and final NLC can attain a zero training error, though, again, a small NLC does not guarantee a better-than-random training error (section \ref{nlcPredictiveSection}) \contrib. We found a trend that an NLC close to 1 can be associated with elevated, though better-than-random, training and test error, and hence underfitting (section \ref{nlcPredictiveSection}) \contrib.

\paragraph{Criterion \ref{criterionPredictable}: Predictability} We use mean field theory to derive simple, recursive formulas that allow the calculation of the value of the NLC in the infinite width limit of popular fully-connected architectures given only the architecture definition (chapter \ref{meanFieldNnaChapter}). We extensively demonstrate the practical predictiveness of that estimate. The estimate allows us to explain the NLC of an architecture instructively from its definition, specifically in terms of the nonlinearity of the activation functions used. We summarize these contributions in section \ref{meanFieldSummarySection} below.


\paragraph{Criterion \ref{criterionControllable}: Controllability} We introduce the simple nonlinearity normalization algorithm that is effective at controlling the NLC in the design stage and turning it into a tunable hyperparameter (chapter \ref{nlnormChapter}). We summarize it in section \ref{nlnormSummarySection} below.

\paragraph{Criterion \ref{criterionSimple}: Simplicity} Fundamentally, the simplicity of the NLC follows from its definition. We also show that at least in the initial state of fully-connected architectures, the NLC may be replaceable by even simpler metrics. For example, the Jacobian may be replaceable by a ratio of loss gradients (sections \ref{nlcSimpleMetricsSection}, \ref{nlcExplainSection}) \contrib.

\paragraph{Criterion \ref{criterionMeaningful}: Meaningfulness} See section \ref{nlcMeaningSection} below.

\paragraph{Criterion \ref{criterionTheoreticalGrounding}: Theoretical grounding} The NLC is theoretically grounded through its relationship with mean field theory, which we summarize in section \ref{meanFieldSummarySection} below. We have additional theoretical results in sections \ref{nlcDefinitionSection}, \ref{nlcLinearApproximationSection} and \ref{nlcNoiseSection}, which we summarize under ``well-definedness'' above and in section \ref{nlcMeaningSection} below.

\paragraph{Criterion \ref{criterionSynergy}: Synergy} The NLC is especially synergistic with the guidelines of `avoid neuron bias', `use an appropriate width / parameter dimensionality' and `ensure orthogonality', as we summarize in section \ref{practicalGuideSection} below.

\paragraph{Criterion \ref{criterionGeneral}: Generality} The generality of the NLC is grounded in its well-definedness and predictiveness as summarized above. While the NLC technically depends on an input distribution $\mathcal{D}$ in addition to a network, it depends on the latter largely through either mean and variance (fully-connected networks) or mean and covariance (convolutional networks), and these dependencies are conceptually necessary (section \ref{nlcRobustDataSection}, \ref{nlcSimpleMetricsSection}) \contrib. Further, the NLC of practical architectures does not vary much from one draw of the random parameter initialization scheme to the next (section \ref{nlcRandomInit}) \contrib. Rather, it depends largely on the random initialization scheme itself, which we consider part of the architecture definition and subject to direct control. Hence, the NLC can be fundamentally regarded as an architecture property and is suitable for data-agnostic ZSAD (section \ref{moduloDataSection}), especially since mean- and variance-normalization are common data processing practices. For example, we could view the ``NLC of an architecture'' as its NLC on unit Gaussian input in the initial state.

Beyond this, the NLC is robust to changes of width as long as the initial weight variance co-varies according to the LeCun initialization (section \ref{widthInvarianceSection}, \ref{vanishWidthSection}) \contrib. It is robust to decomposition into the sum-product of NLCs of individual layers (section \ref{nlcDecomposableSection}) \contrib, to scaling layers and inputs (sections \ref{linearEquivarianceSection}, \ref{vanishWhyBadSection}) \contrib \ and to shifting layers and inputs (sections \ref{chaosCorrelationConfoundingSection}, \ref{vanishWhyBadSection}) \contrib. The NLC changes smoothly and regularly from layer to layer (section \ref{nlcBackpropSection}, \ref{vanishWhyBadSection}) \contrib, and tends to be somewhat invariant during training (section \ref{nlcEvolutionSection}) \contrib. We explain many of the above properties via mean field theory (section \ref{nlcExplainSection}) \contrib.

\subsubsection{The meaning of the NLC: nonlinearity, expressivity and model complexity} \label{nlcMeaningSection}

\paragraph{The NLC is a measure of nonlinearity: elementary properties} We give results that establish the NLC as a measure of the degree of nonlinearity of a network through several different lenses. We derive the NLC from insights about the nonlinearity of 1-dimensional functions (section \ref{whatIsNonlinearitySection}) \contrib. We prove the NLC is equal to 1 for linear networks (section \ref{nlcNonlinearityMetricSection}) \contrib.

\begin{repproposition}{finiteNetNlcEquals1}
Let $f$ be linear. Then $NLC(f,\mathcal{D}) = 1$.
\end{repproposition}

We prove the NLC is invariant to certain linear transformations of network input and output (section \ref{nlcNonlinearityMetricSection}) \contrib \ and that the NLC is greater than 1 for non-linear networks with Gaussian inputs (section \ref{nlcNonlinearityMetricSection}) \contrib.

\begin{reptheorem}{finiteNetNlcGreater1}
Let $\mathcal{D}$ be Gaussian. Then $NLC(f, \mathcal{D}) \ge 1$, where equality holds if and only if $f$ is linear.
\end{reptheorem}

We further demonstrate that, at least in the initial state, the value of the NLC on the datasets we study is very close to its value on Gaussian inputs (section \ref{nlcRobustDataSection}) \contrib. Also on Gaussian input, we prove the NLC is at least $\sqrt{2}$ for networks that do not have a linear component (section \ref{nlcLinearApproximationSection}) \contrib. 

\begin{reptheorem}{finiteNetTilde}
Let $\mathcal{D}$ be Gaussian and let the least squares linear approximation to $f$ under $\mathcal{D}$ be the zero function. Then we have $$NLC(f,\mathcal{D}) \ge \sqrt{2}$$
\end{reptheorem}

We prove that the square of the NLC is proportional to the $L2$ linear approximation error of the network (section \ref{nlcLinearApproximationSection}) \contrib.

\begin{reptheorem}{finiteNetLAR}
Let $\mathcal{D}$ be Gaussian and assume $f$ is not linear. Let $xA_f + b_f$ be the least squares linear approximation to $f$ under $x \sim \mathcal{D}$ and let $\tilde{f} = f - xA_f - b_f$ be the residual. Then we have $$NLC(f,\mathcal{D})^2 = 1 + \frac{\mathbb{E}_x||\tilde{f}(x)||_2^2}{\mathbb{E}_x||\tilde{f}(x)||_2^2 + \mathbb{E}_x||(x-\bar{x})A_f||_2^2}(NLC(\tilde{f},\mathcal{D})^2 - 1)$$
\end{reptheorem}

By depicting the network function corresponding to architectures in the initial state, we demonstrate that, at least for some examples, the NLC makes visual and intuitive sense as a measure of nonlinearity and expressivity (section \ref{eyeTestSection}) \contrib.  We show the NLC is empirically and conceptually related to the diameter of the regions in input space in which the gradient-based local linear approximations are accurate (section \ref{nlcSensiSection}) \contrib.

\paragraph{The NLC is related to underfitting} Because the NLC is proportional to the $L2$ linear approximation error of the network (see theorem \ref{finiteNetLAR} above), where the strength of the relationship $NLC(\tilde{f},\mathcal{D})^2 - 1$ is at least 1 (see theorem \ref{finiteNetTilde} above), the closer the NLC is to 1, the closer the network has to be to a linear function in an $L2$ sense. Hence, if a linear architecture underfits on a task, so do low-NLC architectures, at least as long as nonlinearity does not increase during training. We observe underfitting for convolutional architectures (section \ref{nlcPredictiveSection}) \contrib.

\paragraph{The NLC is related to noise sensitivity and overfitting} We prove the NLC can be considered the first-order approximation of the sensitivity of the network output to white noise added to the input (section \ref{nlcNoiseSection}) \contrib.

\begin{repproposition}{finiteNetNoiseSensitivity}
Let $\delta_\text{in} \sim \mathcal{N}(0,\Cov_x)$ and $\delta_\text{out} \sim \mathcal{N}(0,\Cov_f)$ be row vectors. Assume $NLC > 0$. Then $$\mathbb{E}_{x,\delta_\text{in}}||\frac{\delta_\text{in}}{NLC}\mathcal{J}(x)^T||_2^2 = \mathbb{E}_{\delta_\text{out}}||\delta_\text{out}||_2^2$$
\end{repproposition}

As a rule of thumb, a random input perturbation of relative magnitude $\frac{1}{NLC}$ will corrupt the network output. For example, if $NLC=1000$, a relative input perturbation of 0.1\% is sufficient. If the inputs are e.g. images, a random change of 0.1\% to the intensity of each pixel is almost certainly imperceptible. This explains the association of the NLC with overfitting. If the network is sensitive to noise significantly smaller than the distance between training and test inputs in input space, generalization fails. We demonstrate this mechanism empirically in detail (section \ref{nlcNoiseSection}) \contrib. In the context of classification, high-NLC networks assign inputs that are very close together to different classes. This is very reminiscent of the ``shattering'' property of VC dimension.

\paragraph{The NLC is a measure of model complexity} We connect the NLC to model complexity via the notion of kernel bandwidth, which is a primary measure of model complexity in the field of kernel methods. We utilize the connection of wide networks with Gaussian processes, which is a core feature of mean field theory. The network function in the initial state is captured by the `covariance kernel' (section \ref{covarianceKernelSection}) \backgr \ and the course of training by the `neural tangent kernel' (section \ref{meanFieldNTKsection}) \backgr. We show a strong empirical relationship between the NLC and the neural tangent kernel, which we capture via the TTNTK metric (section \ref{nlcKernelSection}) \contrib.

$$TTNTK(f,\theta,\ell,D_\text{train},D_\text{test}) = \frac{(\mathbb{E}_{(x,y) \in D_\text{train}}\frac{d\ell(f(\theta,x),y)}{d\theta})(\mathbb{E}_{(x,y) \in D_\text{test}}\frac{d\ell(f(\theta,x),y)}{d\theta})^T}{(\mathbb{E}_{(x,y) \in D_\text{train}}\frac{d\ell(f(\theta,x),y)}{d\theta})(\mathbb{E}_{(x,y) \in D_\text{test}}\frac{d\ell(f(\theta,x),y)}{d\theta})^T}$$

Here, $\ell$ is the loss function, $\theta$ is the trainable parameter, $D_\text{train}$ is the training set, $D_\text{test}$ is the test set and $(x,y)$ is a datapoint. It is worth noting that TTNTK is an interesting metric worth studying as a performance predictor in its own right. We go on to prove that in popular fully-connected architectures, the mean field limit of the NLC is the first-order approximation of the bandwidth of the covariance kernel, as summarized in section \ref{meanFieldSummarySection} below.

\paragraph{The NLC is the best practical measure of expressivity}  The notion of expressivity has been accepted as an aspect of model complexity in neural networks and associated with many concepts in deep learning. Out of these concepts, ``exploding / vanishing gradients'', ``order / chaos'' and ``depth'' are the most prominent ones. We argue that the NLC, together with the concept of neuron bias, largely supersedes these concepts, at least in the context of designing feedforward architectures, as we argue throughout chapter \ref{relatedWorkChapter} and summarize in section \ref{relatedWorkSummarySection} below. In a similar vein, we show that the Hessian, a traditional measure of function nonlinearity, is ineffective at characterizing the network when meaningful second-order information is not available. This happens when e.g. the second derivative of activation functions used is not meaningful, which is common for neural networks (e.g. ReLU, section \ref{hessianSection}) \contrib. Other metrics that could be considered for expressivity that arise in our work are MGLLA (section \ref{nlcSensiSection}), which is based on the diameter of linearly approximable regions in input space, and TTNTK (section \ref{nlcKernelSection}), which is based on the neural tangent kernel.

\paragraph{Additional properties} A high NLC tends to induce low effective depth (section \ref{effectiveDepthSection}) \contrib. Effective depth was the subject of one of our earlier works \citep{expl} and was introduced by \citet{residualEnsembles}.

\paragraph{Summary} We argue for the NLC as the best practical measure of expressivity and a primary measure of model complexity because (i) it captures the notion of kernel bandwidth in neural networks via its connection to neural tangent kernel and covariance kernel, (ii) it is related to underfitting via its connection to the $L2$ approximation error, (iii) it is related to overfitting via its connection to noise sensitivity, (iv) it is advantageous relative to prior measures of expressivity, (v) as opposed to esoteric properties, like whether the architecture can represent high-order polynomials, the NLC has been shown to be able to efficiently determine the suitability of practical architectures (see ``predictiveness'' above) (vi) of the intuitive connection between nonlinearity and expressivity and (vii) its overall utility as demonstrated in this work.

The universal nature of the NLC is underscored by the fact that it has strong ties with and adds understanding and meaning to every other ZSAD guideline we cover in this work, as we summarize in sections \ref{practicalGuideSection} and \ref{relatedWorkSummarySection} below.

\subsubsection{Understanding expressivity via the NLC} \label{expressivitySection}

Expressivity is a fuzzy and ill-defined architecture property that is strongly associated with the notion of model complexity in neural networks. At a high level, the expressivity of an architecture refers to its ability to represent network functions that are considered complex as the parameter varies. The most common approach to argue for the expressivity of an architecture is to define a specific class of functions and show that the architecture can represent them with specific parameter values. This has been the primary strategy for arguing for the importance of depth in neural networks. Deep networks can be shown to be able to represent certain function classes with a moderate number of neurons whereas shallow architectures, usually of depth 2, would require an enormous number of neurons. \citet{depth9} provides a recent, detailed overview. We further discuss this point in section \ref{depthSection}. The notion of expressivity has also been formally associated with e.g. the order / chaos / edge of chaos concept (section \ref{orderChaosSection}), as well as the number of linear regions in the input space of a ReLU architecture \citep{trajectoryTransitions}. Informally, expressivity has been linked to exploding / vanishing gradients and associated concepts like Jacobian eigenvalues and the Lipschitz constant (section \ref{vanishingGradientSection}). The idea here is that the gradient of the network dictates how similar the network output has to be for nearby inputs, and that functions where this similarity is high are of low complexity. We also explain this in the context of kernel bandwidth (section \ref{nlcKernelSection}). We argued for the NLC as a measure of expressivity above.

It can be viewed as a drawback that the NLC only measures the complexity of a given network function, but not the (maximal) complexity of any network function representable by an architecture for an arbitrary parameter value. As our work is about architecture design, we focus on properties of the architecture in the initial state, and therefore on the properties of networks obtained by drawing typical parameter values from the random initialization scheme. We bridge the initial-final state gap by observing that gradient-based training tends to approximately preserve many key properties, including expressivity, as we further discuss below. Much of the expressivity literature performs a {\it best-case analysis} by considering arbitrary parameter values, whereas we perform a {\it typical-case analysis} by considering typical parameter values. We are not aware of much practical value derived from any theoretical result under the best-case approach. As always in this work, we take the practical route.

Having established the NLC as an expressivity measure, we can now study the notion of expressivity via the NLC. We show that test error is suboptimal when expressivity is either too large or too small, and that there is an intermediate ``sweet spot'' of expressivity that is ideal. In this way, the concept of expressivity in deep learning corresponds to the concept of model complexity in classical machine learning. Especially in the context of depth, expressivity has often been regarded as strictly beneficial, i.e. an increase in expressivity should lead to an increase in performance. {\it The more expressivity, the better.} Our work refutes that view. This opens up a large unanswered question. In what contexts is great depth truly beneficial and why?

In contrast to the concept of depth, the exploding / vanishing gradient problem has been viewed through the lens of the ``sweet spot''. The classical view is that it is suboptimal when gradients grow too large or too small during backpropagation; and that a stable gradient is ideal. We show that this is only true in specific contexts, which we clearly define (section \ref{vanishMultipleSection}) \contrib, and that the ``sweet spot of stable gradients'' does not in fact correspond to the ``sweet spot of desirable expressivity'' and may in fact be suboptimal (section \ref{vanishDichotomySection}) \contrib. We show the same for the ``order / chaos problem'' (section \ref{chaosGoodSection}) \contrib.

We can achieve arbitrary NLC values at any depth by e.g. using activation functions of differing degrees of nonlinearity (sections \ref{depthSection}, \ref{pathEquationSection}) \contrib. Hence, we argue that we can fundamentally achieve any expressivity level at any depth. This further underscores the need for new explanations of the value of depth in neural networks. Among our fully-connected architectures, depth is positively correlated with test error and we observe no benefits from increasing depth beyond around 10 (section \ref{depthSection}) \contrib.

While low-expressivity architectures can have elevated training error due to underfitting, we show that high-expressivity architectures can still be trained to zero or near-zero training error (section \ref{nlcPredictiveSection}) \contrib. To our knowledge, we are the first to explicitly demonstrate the trainability of ultra-high expressivity architectures. For example, \citet{depthScalesMeanField} and \citet{meanFieldCNN} previously argued that this was impossible. We uncover that there are a number of requirements for successful training: small learning rate (section \ref{learningRateSection}), lack of noise-inducing building blocks like batch normalization or data augmentation, and high floating-point precision (sections \ref{noiseStabilitySection}, \ref{beyondNlcSummarySection}). When those requirements hold, we attain a low training error in our empirical studies in all but one case, where we obtain a moderate training error.

Should expressivity be viewed as equivalent to model complexity in neural networks? Via mean field theory, we show that the NLC can be viewed as largely independent of the width, and hence the parameter dimensionality of the architecture (section \ref{meanFieldPracticalSection}, \ref{nlcExplainSection}) \contrib. Parameter dimensionality is another classical, statistical measure of model complexity that has been widely shown to also be a key driver of neural architecture performance. Hence, the NLC and width / parameter dimensionality emerge as measures of two independent axes of model complexity, which are represented by the notions of expressivity and `capacity'. One is based on the complexity of the kind of functions expressed by the architecture, while the other is based on how many different functions it can represent. It would be very interesting to study specifically the differences, similarities and interactions of these two axes of model complexity.

\subsubsection{Initial NLC vs the complexity of the true input-label function} \label{initialFinalSummarySection}

While we find that the initial NLC must lie in a specific range for an architecture to achieve better-than-random performance, the NLC after training must lie in a much narrower range, which is close to 1 and depends on the dataset (section \ref{nlcPredictiveSection}) \contrib. Specifically, we identify two types of architectures. The first type has an initial NLC of less than around $10^5$ before training and tends to have a much smaller NLC after training. In fact, there is an approximately linear relationship in log space between initial and final NLC across those architectures with slope less than 1. The second type has almost exactly the same NLC after training as before. It does not achieve a better-than-random test error but can still achieve a zero training error (section \ref{nlcEvolutionSection}) \contrib. We note that we found a very small number of architectures that succeed with relatively large learning rates that can drastically and unpredictably change their properties during training (sections \ref{nlcPredictiveSection}, \ref{nlcEvolutionSection}, \ref{learningRateSection}) \contrib.

We introduce the distribution-valued PNLCD metric for measuring the apparent nonlinearity of the true input-label function inherent in the dataset (section \ref{bestNlcSection}) \contrib. (We use the word ``apparent'' because of the crudeness of the measure.) PNLCD is based on the gradient of line segments between datapoints. According to this metric, our datasets CIFAR10, MNIST and waveform-noise only appear slightly nonlinear as the probability mass of PNLCD concentrates around 1 (section \ref{bestNlcSection}) \contrib. By constructing artificial datasets with very high PNLCD sample values, we show that the peak of the PNLCD distribution is closely associated with the NLC value at which architectures attain the best test error values for a given dataset (section \ref{bestNlcSection}) \contrib. Further, we argue that the reason PNLCD concentrates around 1 for our datasets is because they are part of a large class of datasets we term {\it neural regular} as we further summarize in section \ref{meanFieldSummarySection} below. The existence of such a class would make the NLC as data-agnostic a ZSAD guideline as possible (section \ref{moduloDataSection}). The reason for our recommended [1,5] range for the NLC, beyond our raw performance values, is that it is close to the PNLCD peak at 1. Some of our artificial datasets required initial NLC values of up to $10^4$.

We interpret these observations as follows. The test error of an architecture will be lower the closer the network function is to the true input-label function after training. Specifically, it must mirror the degree of nonlinearity. The architecture indeed tries to adopt an ideal NLC through training by decreasing its initial NLC if it is too large. However, only if the initial NLC is less than $10^5$ does the architecture achieve a final state with better-than-random test error, and only if the initial NLC is less than 5 does the architecture achieve a final state with close-to-optimal test error, assuming the dataset is neural regular and has its PNLCD peak close to 1. If the initial NLC is too large, the architecture can only achieve a better-than-random training error by memorizing the training set. Solidifying this view is an interesting direction for future work.

This analysis reveals a general aspect of zero-shot architecture design. Since we study the properties of an architecture's definition and initial state, no matter how harmful the property is to e.g. generalization or trainability, it is usually technically possible that the architecture loses that property early in training and achieves a low training and test error from that point forward. Throughout this work, we find that many key properties we study, including NLC, Gaussian stability and neuron bias, persist between initial and final state to a large degree, assuming that training was successful, and that the best strategy is to design an architecture that has desirable properties in the initial state, rather than relying on change during training. While it may be unsatisfying that we do not have theoretical results that make final-state guarantees about e.g. NLC or error based on e.g. initial NLC, we note that making guarantees of this kind that are also practical has been nearly impossible throughout the history of deep learning. As always, we take the practical route.

\subsection{A practical guide to neural architecture design} \label{practicalGuideSection}

\begin{center}
{\centering ``Anything that can go wrong, will go wrong.''} - Murphy's Law
\end{center}

The space of neural architectures is as large as the space of parametrized, differentiable functions. But, even if we are content to build a neural architecture from only popular building blocks, there are still a large number of choices. How many layers should there be? How wide should layers be? What activation functions and normalization layers should be used? What layer connectivity and parameter initialization scheme should be used? With many choices, many things can go wrong. And the number of pitfalls further increases the more the requirements of a given ML task differ from extensively studied benchmark tasks.

As summarized at the beginning of section \ref{practicalGuideSection}, we formalize the concept of zero-shot architecture design (ZSAD; section \ref{zeroShotDesignSection}) as well as that of a ZSAD guideline (section \ref{zsadGuidelineSection}) \contrib. We frame the notion of architecture performance in such a way that ZSAD guidelines encapsulating properties of the randomly initialized state can be considered performance predictors of the final state across training algorithms, datasets and random parameter draws (section \ref{zsadGuidelineSection}, \ref{architecturePerformanceSection}) \contrib. In this work, we present ZSAD guidelines for preventing training and / or generalization failure. In fact, we present what we believe is the most comprehensive repository of ``things that can go wrong'' in neural architecture design, alongside ways to prevent these things from going wrong. We believe that we have made at least designing fully-connected architectures relatively foolproof, and have substantially alleviated uncertainty regarding whether a given architecture leaves large amounts of performance on the table. 

\sloppy As mentioned at the beginning of section \ref{thesisOverviewSection}, we also focus on guidelines that are as data-independent as possible. We demonstrate this in detail for the NLC (see ``generality'' above) and briefly discuss it for other guidelines in section \ref{guidelineDataSection} and section \ref{meanFieldSummarySection} below.

In figure \ref{beyondSummary}, we plot the NLC vs test and training error across our range of randomly generated architectures (sections \ref{studyAArchitecturesSection}, \ref{studyBArchitecturesSection}), and we use colors to indicate whether architectures follow or do not follow certain ZSAD guidelines as given below. In that figure, we explain the majority of test and training error variation across these architectures with our ZSAD guidelines \contrib. This figure reveals that test and training error is largely {\it all-or-nothing}. To ensure low error, we must follow all ZSAD guidelines. To succeed, nothing must go wrong. But, by Murphy's law, this means that we must ensure that nothing can go wrong.

We provide an overview of the ZSAD guidelines we discuss in this work in table \ref{surveySummary}, along with building blocks that can cause or prevent an architecture from following these guidelines, as discussed throughout chapter \ref{surveyChapter}. We argue that the degree to which building blocks help an architecture follow our covered guidelines largely explains their popularity. Now we summarize our guidelines one by one.

\paragraph{Use an appropriate NLC}

\begin{repnpc}{npcNLC}
`Use an appropriate NLC' requires $1 \le NLC \le 5$. \contrib
\end{repnpc}

We have summarized the NLC and associated guideline in section \ref{nlcSummarySection} above.

An effective strategy for controlling the NLC is nlnorm, as summarized in section \ref{nlnormSummarySection} below. The NLC can be reduced by skip connections, especially when normalization layers are not placed between successive residual units (section \ref{skipConnectionsSection}) \contrib. It can also be reduced by using relatively linear activation functions, like SELU or softplus (sections \ref{plainArchitectureSection}, \ref{surveyDepthSection}). The NLC can be estimated via mean field theory (section \ref{meanFieldSummarySection} below).

\paragraph{Ensure Gaussian stability} Main sections: \ref{meanFieldDistributionSection}, \ref{gaussianStabilityExplanationSection}, \ref{gaussianStabilitySection}

\begin{repnpc}{npcGaussian}
`Ensure Gaussian stability' requires that a network exhibits Gaussian stability as defined in section \ref{gaussianStabilityExplanationSection}. \contrib
\end{repnpc}

Gaussian instability is likely to lead to high test error (sections \ref{gaussianStabilitySection}, \ref{beyondNlcSummarySection}) \contrib. We summarize this guideline further in section \ref{meanFieldSummarySection} below.

An effective strategy for inducing Gaussian stability is using activation functions that exhibit mean field Gaussian stability (section \ref{gaussianStabilityExplanationSection}) \contrib, which is at least approximately achieved by all popular activation functions (section \ref{gaussianStabilityExplanationSection}) \contrib. At least in fully-connected architectures, layer normalization is also sufficient (sections \ref{gaussianStabilityExplanationSection}, \ref{surveyBNsection}) \contrib.

\paragraph{Ensure scale stability} Main section: \ref{forwardStabilitySection}

\begin{repnpc}{npcForward}
`Ensure scale stability' requires that at each layer in a network $f$, the overall magnitude of neuron values across inputs and neurons is not excessively large or small \backgr. Specifically, we advocate measuring this via $LSCALE_l \approx 1$ at each layer $f_l$, where $LSCALE_l(f,\mathcal{D}) = \sqrt{\frac{1}{d_l}\mathbb{E}_x||f_l(x)||^2_2}$ is the quadratic expectation of layer quadratic means. \contrib
\end{repnpc}

Here, $f_l : \mathbb{R}^{d_\text{in}} \rightarrow \mathbb{R}^{d_l}$ is the layer as a function of the input, $l$ is the layer index, $\mathcal{D}$ is the input distribution and $x\sim\mathcal{D}$ is the input vector.

As we show, scale stability is based on the insight that popular deep learning pipeline building blocks, such as activation functions, the softmax+cross-entropy loss function and basic training algorithms using a single learning rate for all layers, are designed to work best when the overall neuron value magnitude at each layer is around 1 (section \ref{forwardStabilitySection}) \contrib. For example, the nonlinearity of activation functions can grow very small or large when the inputs to those activation functions grow small or large, which can lead to an excessive or insufficient NLC, and hence underfitting or overfitting (section \ref{actFunNLCsection}) \contrib. However, when the properties of the building blocks of the deep learning pipeline are understood and adjusted accordingly, scale stability becomes irrelevant (section \ref{linearEquivarianceSection}) \contrib.

Effective strategies for inducing scale stability include nlnorm (section \ref{nlnormPerformanceSection}) \contrib, normalization layers (section \ref{surveyBNsection}), careful weight initialization or simply multiplying layers with fixed constants recursively (section \ref{nlnormRelatedWorkSection}) \backgr. Scale stability has a clear meaning in mean field theory based on the mean field estimate of $LSCALE_l$ (section \ref{meanFieldSummarySection} below). Via this estimate, we introduce `mean field scale stability' (section \ref{forwardStabilitySection}) \contrib.

\paragraph{Ensure training stability} Main section: \ref{covariateShiftSection}

\begin{repnpc}{npcTraining}
`Ensure training stability' requires that the value of metrics or the presence of properties that either directly or indirectly influence architecture performance, such as LSCALE, LCV, NLC, LBIAS, Gaussian stability or noise stability, are not prone to change in an uncontrolled or harmful way during training for an architecture $f$. \contrib
\end{repnpc}

In order to achieve better-than-random performance, both NLC (section \ref{nlcPredictiveSection}, section \ref{initialFinalSummarySection} above) and LBIAS (section \ref{outputBiasSection} and below) need to be in a narrow range after training. Hence, we must ensure that these metrics do not increase drastically during training. We show how the nonlinearity of many activation functions, such as tanh, increase as their inputs grow (sections \ref{covariateShiftSection}, \ref{actFunNLCsection}) \contrib. The parameter value tends to grow during training from gradient updates, though to wildly varying degrees (section \ref{meanFieldPracticalEmpiricalSection}) \contrib. This can lead to neuron value growth, a loss of scale stability, then to activation function nonlinearity growth and thus, finally, NLC growth (sections \ref{meanFieldPracticalEmpiricalSection}, \ref{forwardStabilitySection}) \contrib.

We suspect that it is desirable to preserve all our ZSAD guidelines more or less during training.

Effective strategies for improving training stability are the use of ReLU and normalization layers, especially batch normalization (section \ref{surveyBNsection}) \backgr \ /  \contrib.

\paragraph{Avoid neuron bias} Main section: \ref{outputBiasSection}

\begin{repnpc}{npcBias}
`Avoid neuron bias' requires that absolute neuron expectations are not excessively large \backgr. Specifically, we advocate measuring this via $LBIAS_l \approx 0$ at each layer $f_l$, where $LBIAS_l(f, \mathcal{D}) =\frac{\sqrt{\mathbb{E}_x||f_l(x)||_2^2}}{||\mathbb{S}_xf_l(x)||_2}$ is the layer bias. We especially recommend ensuring $LBIAS_L \approx 0$. \contrib 
\end{repnpc}

Here, $\mathbb{S}_x$ represents the standard deviation operator. We write LBIAS short for $LBIAS_l$ evaluated at the output layer, i.e. $LBIAS_L$. We find that neuron bias is close to its maximum across layers at the output layer in our architectures' initial state (section \ref{outputBiasSection}) \contrib.

We show that LBIAS, evaluated in the architecture's randomly initialized state, is predictive of test error after training. We also find that an LBIAS less than around 10 in the initial state is essential for minimizing test error (section \ref{outputBiasSection}) \contrib.

This is largely explained by two things: (i) Assuming that the class frequencies in the dataset are relatively balanced, in order for the architecture to model the true input-label function, the network function must be relatively unbiased in the final state. Indeed, we find that all our architectures that achieve a better-than-random error also achieve an LBIAS value close to 1 in the final state (section \ref{outputBiasSection}) \contrib. Like for the NLC, it appears that this can only be achieved through training if the initial LBIAS value is already somewhat close. (ii) Neuron biases negatively affect the training of linear layers. 

We demonstrate the impact of those two factors by introducing two methods: `output debiasing' and the `debiased gradient descent' training algorithm (section \ref{outputBiasSection}) \contrib. The first ``takes care'' of the bias of the network function. The second removes the impact of the bias on linear layer training. Architectures with very high neuron bias levels can successfully train and generalize with those two methods (section \ref{outputBiasSection}) \contrib. Hence, we show that architectures that exhibit extreme vanishing gradients and `order' can both train and generalize with a slightly modified training protocol \contrib. We suspect that further investigation may reveal that when deep learning pipeline building blocks are adjusted to bias, this guideline becomes irrelevant in a similar manner as scale stability, though this adjustment may be significantly more cumbersome.

Among plain architectures with scale stability as defined in sections \ref{actFunLengthKernelSection} and \ref{actFunCovKernelSection}, there exists a trichotomy where either (i) LBIAS grows exponentially with depth, which also corresponds to vanishing gradients and `order', (ii) both NLC and LBIAS grow sub-exponentially, which also corresponds to `edge of chaos' or (iii) the NLC grows exponentially with depth, which also corresponds to exploding gradients and `chaos' (sections \ref{actFunNLCsection}, \ref{plainArchitectureSection}, \ref{vanishMultipleSection}, \ref{orderChaosLimitedSection}) \contrib. Hence, NLC and LBIAS represent a fundamental tradeoff, at least in the context of plain architectures. We even find that this tradeoff largely applies across the entire range of architectures we study empirically (sections \ref{plainArchitectureSection}, \ref{vanishMultipleSection}) \contrib. Contrary to popular belief, we argue that case (iii) is most desirable (section \ref{plainArchitectureSection}).

Effective strategies for avoiding neuron bias are nlnorm (section \ref{nlnormSummarySection} below), batch normalization (section \ref{surveyBNsection}) or unbiased activation functions like tanh or SELU (section \ref{plainArchitectureSection}) \backgr. Neuron bias can be reduced by skip connections, especially when normalization layers are not placed between residual units (section \ref{skipConnectionsSection}) \contrib. Neuron bias can be estimated via mean field theory (section \ref{meanFieldSummarySection} below).

\paragraph{Ensure noise stability} Main section: \ref{noiseStabilitySection}

\begin{repnpc}{npcNoise}
`Ensure noise stability' requires that the random choices made when evaluating the network function do not induce output variation of a magnitude greater than the range of predictions that can be considered accurate for a given task. In this context, floating-point rounding error can also be viewed as noise. \contrib
\end{repnpc}

Noise stability is a very general guideline, and we study it using the examples of floating-point rounding error and batch selection noise induced by batch normalization (BN). We show that the noise induced by using 32-bit rather than 64-bit floating-point precision and / or BN make architectures with a large NLC untrainable (section \ref{noiseStabilitySection}) \contrib. The NLC is a crucial mediator for the guideline of noise stability because it drives the noise sensitivity of the architecture (section \ref{nlcMeaningSection} above). It is interesting to note that, while floating-point rounding error behaves like white noise as studied in section \ref{nlcNoiseSection}, batch selection noise has a less severe impact on performance than its magnitude would suggest (section \ref{noiseStabilitySection}) \contrib.

Noise stability can be ensured by eliminating sources of significant noise, like low floating-point precision, batch normalization with a small batch size, or dropout with a high dropout rate. Crucially, architectures that can perform well according to the NLC are also relatively robust to noise.

\paragraph{Ensure orthogonality} Main section: \ref{orthogonalitySection}

\begin{repnpc}{npcOrthogonality}
`Ensure orthogonality' requires that the absolute singular values of characteristic matrices of the network should be relatively similar. These matrices include the Jacobian, Gram matrix, covariance matrix, Hessian and Fisher matrix. \backgr
\end{repnpc}

We discuss guidelines \ref{npcOrthogonality} through \ref{npcWidth} mainly in the context of related work and results we obtained for other guidelines. Their impact on performance is not studied explicitly in this work. 

We show that mean field theory, which assumes Gaussian initialized weights, accurately estimates values of e.g. NLC, LBIAS and LSCALE when weights are orthogonally initialized (section \ref{meanFieldPracticalEmpiricalSection}) \contrib. Hence, we can improve orthogonality via orthogonal initialization without impacting those metric values.

In our discussion (section \ref{orthogonalitySection}), we further point out that orthogonality is a natural complement to the NLC in at least two other ways: (i) The NLC is related to the overall magnitude of singular values of the Jacobian via the numerator, whereas orthogonality deals with the relative variation of singular values. (ii) Next to expressivity, orthogonality is the second key ingredient for high-performing arbitrarily deep fully-connected architectures \backgr.

Orthogonality can be ensured to a significant degree by orthogonally initializing linear layers. We suspect that this is generally sufficient in practice when also following other ZSAD guidelines.

\paragraph{Avoid pseudo-linearity} Main section: \ref{pseudoLinearitySection}

\begin{repnpc}{npcPseudoLinearity}
`Avoid pseudo-linearity' requires that the network does not contain an activation layer that is effectively identical to a linear function across the inputs it receives from its parent in the layer graph. \backgr
\end{repnpc}

In our discussion (section \ref{pseudoLinearitySection}), we point out the relationships between pseudo-linearity, which was introduced in one of our prior works \citep{expl}, and other guidelines. (i) Pseudo-linearity tends to arise automatically in very deep networks with a right-sized NLC. This calls into question the value of extreme depth. (ii) Pseudo-linearity tends to arise automatically when $LBIAS_l$ values are high. This underscores the need to avoid neuron bias. (iii) Pseudo-linearity tends to be minimized when nonlinearity is spread evenly throughout the network, which corresponds to gradient explosion (under certain definitions) and `chaos'. Hence, those phenomena may actually be desirable (sections \ref{vanishGoodSection}, \ref{chaosGoodSection}). (iv) Pseudo-linearity is related to scale stability. For example, tanh is almost identical to the identity function when its inputs are small.

\paragraph{Use an appropriate width / parameter dimensionality} Main section: \ref{widthSection}

\begin{repnpc}{npcWidth}
`Use an appropriate width / parameter dimensionality' requires that the network (i) has sufficient capacity to absorb information from the dataset, (ii) is sufficiently wide to prevent harmful information loss and (iii) does not have so much capacity and does not retain so much information that it overfits. \backgr
\end{repnpc}

The NLC can be viewed as largely independent of width in the initial state (section \ref{meanFieldPracticalSection}, \ref{nlcExplainSection}). Hence, width / parameter dimensionality and NLC form two relatively independent axes of model complexity, as summarized in section \ref{expressivitySection} above. Similarly, width can be viewed as largely independent of any metric that has a mean field limit, such as $LBIAS_l$ and $LSCALE_l$ (sections \ref{widthSection}, \ref{meanFieldPracticalSection}) \contrib.

\paragraph{Perform exhaustive learning rate tuning} Main section: \ref{learningRateSection}

This is not technically a ZSAD strategy, as the learning rate is not part of the architecture definition. However, we cover it in this section (as we cover it in chapter \ref{beyondNlcChapter}) because it is another core choice that must be made within the deep learning pipeline that is crucial for optimal performance. Case in point, we show that for each dataset we study in this work, the range of best starting learning rates (SLRs) across all our architectures for minimizing validation / test error has width around $10^7$ in log space. Across all tasks, it has width around $10^{10}$. When minimizing training error, we find the range of best SLRs spans 30 orders of magnitude (section \ref{learningRateSection}) \contrib.

Because of the width of that range, we investigate several strategies for predicting the best SLR, though this investigation is somewhat preliminary. Evidence suggests the following. (i) The NLC does not help significantly with choosing SLR, at least when choosing an SLR to minimize validation / test error (section \ref{learningRateSection}) \contrib. However, the best SLR scales as the inverse square NLC when minimizing training error for high-NLC architectures. (section \ref{learningRateSection}) \contrib. (ii) Architectures which obtain the very lowest test error values have a smaller range of best SLRs (section \ref{learningRateSection}) \contrib. Hence, if we are content not to minimize the test error of architectures that cannot perform as well as other architectures to begin with, our tuning might not have to be as extensive. (iii) The best-performing architectures in particular have the property that the initial parameter vector has a similar length to the change of the parameter vector during training (section \ref{learningRateSection}) \contrib. This is only achieved by a relatively narrow range of SLRs and is an intriguing thread for further investigation. (iv) Especially the best-performing architectures have the property that the initial outputs have a similar magnitude to the change of the outputs during the first gradient update (section \ref{learningRateSection}) \contrib. Again, this is only achieved by some SLRs and is an intriguing thread for further investigation as well. This result is also reminiscent of line search. (v) Architectures that are identical except for differing slightly in the nonlinearity of the activation functions used can still have drastically different best SLRs (section \ref{nlnormLearningRateSection}) \contrib. (vi) Architectures that have been tuned to have an optimal NLC still have a relatively wide range of best SLRs (section \ref{nlnormLearningRateSection}) \contrib.

\subsection{A mean field theory of meta-distributions} \label{meanFieldSummarySection}

\paragraph{Background} (section \ref{meanFieldBackgroundSection}) \backgr \ Mean field theory studies the behavior of architectures in the randomly initialized state in the theoretical limit as the width of layers converges to infinity. In this limit, the value of many important quantities converges with probability 1, where randomness is induced by the parameter initialization scheme.

Our results are grounded in the most well-studied quantities with mean field limits - (i) the `square mean' of a layer value $\mathbb{E}_if_l(x^{(1)})[i]^2$, whose limit we denote by $\mathfrak{q}^{(1)}_l$ and (ii) the `co-mean' of two layer values $\mathbb{E}_if(x^{(1)})[i]f(x^{(2)})[i]$, whose limit we denote by $\mathfrak{c}^{(1,2)}_l$. Here, $f(x)[i]$ denotes the value of the $i$'th component of layer $f_l : \mathbb{R}^{d_\text{in}} \rightarrow \mathbb{R}^{d_l}$ when evaluated on input $x$. $\mathbb{E}_i$ denotes the mean over components.

We base our analysis on the work of \citet{meanFieldNetsorGP}. They give rules for calculating $\mathfrak{q}^{(1)}_l$, $\mathfrak{q}^{(2)}_l$ (corresponding to $x^{(2)}$) and $\mathfrak{c}^{(1,2)}_l$ for each layer recursively from input layer to output layer, given the input square means $q^{(1)} = \mathbb{E}_ix^{(1)}[i]^2$ and $q^{(2)}  = \mathbb{E}_ix^{(2)}[i]^2$ and the input co-mean $c^{(1,2)} = \mathbb{E}_ix^{(1)}[i]x^{(2)}[i]$. In essence, the limits forward-propagate deterministically through the architecture, starting from the corresponding input quantities. In our work, we focus on the case $q^{(1)} = q^{(2)}$. Since $\mathfrak{q}^{(1)}_l$ does not depend on $q^{(2)}$ for any $l$ and vice versa, we obtain $\mathfrak{q}^{(1)}_l = \mathfrak{q}^{(2)}_l$. For the purpose of this summary, we write $q = q^{(1)} = q^{(2)}$, $c = c^{(1,2)}$, $\mathfrak{q}_l = \mathfrak{q}^{(1)}_l = \mathfrak{q}^{(2)}_l$ and $\mathfrak{c}_l = \mathfrak{c}^{(1,2)}_l$.

The recursive calculation of the limits is chiefly driven by the activation functions used in the architecture. Let $\tau$ be an activation function used by some layer, which we term an `activation layer'. Then the key property of $\tau$ that controls the limit calculation is its `covariance kernel'.

$$\mathfrak{C}_\tau(q,c) = \mathbb{E}_{s,t\sim \mathcal{N}(\mu,\Sigma)} \tau(s)\tau(t) \text{ where } \mu = \begin{pmatrix} 0 \\ 0 \end{pmatrix} \text{, } \Sigma = \begin{pmatrix} q & c \\ c & q \end{pmatrix} \text{ and } q \ge c \ge -q$$

The values of $\mathfrak{q}_l$ and $\mathfrak{c}_l$ at the activation layer $f_l$ using activation function $\tau_l$ are then $\mathfrak{q}_l = \mathfrak{C}_{\tau_l}(\mathfrak{q}_k,\mathfrak{q}_k)$ and $\mathfrak{c}_l = \mathfrak{C}_{\tau_l}(\mathfrak{q}_k,\mathfrak{c}_k)$, where $\mathfrak{q}_k$ and $\mathfrak{c}_k$ are the limits at the parent layer of $f_l$ in the layer graph. Our definition of $\mathfrak{C}_\tau$ is simplified because of the $q^{(1)} = q^{(2)}$ assumption. When $c=0$, the definition simplifies to $\mathfrak{C}_\tau(q,0) = (\mathbb{E}_{s\sim\mathcal{N}(0,q)}\tau(s))^2$ and when $c=q$, the definition simplifies to $\mathfrak{C}_\tau(q,q) = \mathbb{E}_{s\sim\mathcal{N}(0,q)}\tau(s)^2$.

Finally, we obtain values $\mathfrak{q}_L$ and $\mathfrak{c}_L$, where the $L$ subscript indicates the output layer. Just like for activation functions, the `covariance kernel' of the architecture $\mathfrak{C}(q,c)$ is the function that yields $\mathfrak{q}_L = \mathfrak{C}(q,q)$ and $\mathfrak{c}_L = \mathfrak{C}(q,c)$, where $q$ and $c$ now refer to the input square mean and co-mean.

\paragraph{Elementwise meta-Gaussian meta-distributions} The most important metrics in this work are those that represent or measure ZSAD guidelines: NLC, $LBIAS_l$ (for neuron bias), $LSCALE_l$ (for scale stability). NLC and $LBIAS_l$ are based on layer distributions via $\Cov_f$ / $\mathbb{S}_x$ respectively. These distributions are induced by the input distribution $\mathcal{D}$, but {\it not} by the parameter initialization scheme. In other words, the metrics are computed as the input varies according to $\mathcal{D}$, but after the trainable parameter has been drawn and is fixed. However, mean field theory deals with a random parameter. Hence, we must investigate the distribution over ``layer distributions induced by $\mathcal{D}$'' as the parameter varies on a meta-level. We call the layer distribution induced by $\mathcal{D}$ with a fixed parameter the `base distribution' and the distribution over base distributions induced by the initialization scheme the `meta-distribution'.

We prove that the meta-distribution of a fixed-width fully-connected layer, as the width of all layers between that layer and the input layer converges to infinity, converges to an {\it elementwise distribution generated by a meta-Gaussian} (section \ref{meanFieldDistributionSection}) \contrib. This means the following. (i) In each base distribution, each neuron is independent of all other neurons. (ii) In each base distribution, each neuron is Gaussian. (iii) The standard deviation of the neurons is constant across the layer and across base distributions. (iv) Drawing from the meta-distribution corresponds to drawing each base distribution neuron mean independently from a Gaussian distribution with mean zero and a standard deviation that is fixed across neurons. This meta-distribution is determined by two parameters: the variance of the neuron base distributions and the variance of the means of the neuron base distributions.

Our theorem depends critically on a condition that was not previously present in mean field theory, which we term `elem-like$(q,c)$'. It requires that all inputs have the same square mean $q$ and all pairs of inputs have the same co-mean $c$. While this is technically impossible, it holds approximately for neural regular datasets, as we discuss below. Under this assumption, the two parameters of the meta-distribution described above turn out to be exactly $\mathfrak{q}_l - \mathfrak{c}_l$ and $\mathfrak{c}_l$ respectively (section \ref{meanFieldDistributionSection}) \contrib.

We conduct a very large number of experiments to verify that this theoretical result is predictive for fully-connected layers in our fully-connected architectures in the initial state, but to some degree also in the final state (section \ref{meanFieldPracticalSection}) \contrib. In a nutshell, we verify each property (i) through (iv) of the meta-distribution given above in turn. We measure independence via correlation and Gaussianity via excess kurtosis and a comparison with the Gaussian cumulative distribution function. Only architectures that exhibit Gaussian instability, as we further summarize below, fail to exhibit meta-Gaussian fully-connected layers.

\paragraph{Neural regular data} While our investigation in this regard is preliminary, it suggests that there is a class of datasets which we introduce as `neural regular' (section \ref{elemLikeSection}) \contrib. A neural regular dataset has the following properties: (i) The square means and co-means of its inputs have similar respective values. (ii) Input components are relatively independent. (iii) The degree of nonlinearity of the true input-label function is close to 1.  In a nutshell, these datasets have inputs that are spread out relatively far from each other across the input domain. Neural regularity enables both ZSAD guideline \ref{npcNLC} (section \ref{initialFinalSummarySection} above) and the mean field theory of meta-distributions. We show that properties (i) through (iii) are theoretically related (section \ref{elemLikeSection}) \contrib. Our datasets CIFAR10, MNIST and waveform-noise (sections \ref{studyADataSection}, \ref{studyBTrainingSection}) are neural regular (sections \ref{bestNlcSection},\ref{elemLikeSection}) \contrib. We further argue that a large fraction of practical deep learning datasets is neural regular (section \ref{elemLikeSection}). We suspect that neural regular datasets exhibit similar behavior in many contexts. Investigating them as a class is an interesting direction for future work.

\paragraph{Mean field limits of practical architectures} Using \citet{meanFieldNetsorGP}, it is possible to obtain mean field limits at each layer with recursive calculation rules. It is, however, cumbersome, as practical architectures have to be re-cast in terms of the abstract layer operations in which they frame their analysis. We prove a set of practical calculation rules for a somewhat general class of fully-connected architectures built using popular building blocks and design strategies, which we call `A-architectures' (section \ref{metricLimitSection}) \contrib. This includes the limit of the square mean of the layer gradient with respect to the input, which we denote by $\mathfrak{g}_l$. See especially figure \ref{tableNLCPropagation}.

\paragraph{Novel limits: NLC, neuron bias, scale stability} We then go on to prove mean field limits for metrics that depend on the base distribution (section \ref{metricLimitSection}) \contrib. This specifically yields:

\begin{eqnarray*}
\lim NLC &=& \sqrt{\frac{\mathfrak{g}_L(\mathfrak{q}_0 - \mathfrak{c}_0)}{\mathfrak{g}_0(\mathfrak{q}_L - \mathfrak{c}_L)}}\\
\lim LBIAS_l &=& \sqrt{\frac{\mathfrak{q}_l}{\mathfrak{q}_l - \mathfrak{c}_l}}\\
\lim LSCALE_l &=& \sqrt{\mathfrak{q}_l}\\
\end{eqnarray*}

It turns out that the mean field limit of our metrics can be expressed in terms of the basic limits $\mathfrak{q}$, $\mathfrak{c}$ and $\mathfrak{g}$. Hence, the recursive calculation rules can be re-used. Note that not only $LBIAS_l$ and $LSCALE_l$ have such simple mean field limits, but also other metrics that could be reasonably used to measure neuron bias and scale stability respectively. We term the limit of the NLC `mean field NLC' and write $\mathfrak{n} = \sqrt{\frac{\mathfrak{g}_L(\mathfrak{q}_0 - \mathfrak{c}_0)}{\mathfrak{g}_0(\mathfrak{q}_L - \mathfrak{c}_L)}}$.

As before, we assume that the input distribution is elem-like$(q,c)$. As before, the calculation rules depend on the data only through $q$ and $c$. Hence, any metric with a mean field limit, and hence any ZSAD guideline based on such a metric, is largely independent of data, as long as mean field theory is predictive. Further, the calculation rules depend on the parameter only via its initialization scheme. Therefore, mean field limits can be considered properties of the architecture definition.

We empirically validate the limits of both simple and more advanced metrics across our fully-connected architectures (section \ref{meanFieldPracticalEmpiricalSection}) \contrib. We do this by slightly modifying the definition of the limit quantities to enable us to compute them for architectures that are not Gaussian initialized or not in the initial state. Indeed, we find that $LSCALE_l$ in particular is still close to this surrogate estimate in the final state (section \ref{meanFieldPracticalEmpiricalSection}) \contrib. Again, practical predictiveness breaks down for architectures that do not exhibit Gaussian stability.

\paragraph{The nonlinearity path equation} Using the calculation rules of figure \ref{tableNLCPropagation}, we derive the nonlinearity path equation (NPE) for A-architectures, which we give in a simplified form below (section \ref{pathEquationSection}) \contrib.

$$\mathfrak{n}(f,q,c) = \sqrt{\sum_{p \in P} w(p) \prod_{\tau_l \in p} \mathfrak{n}_{\tau_l}(\mathfrak{q}_k,\mathfrak{c}_k)^2}$$

Here, $f$ represents the architecture definition including parameter initialization scheme and $q$ and $c$ stem from the `elem-like$(q,c)$' condition. $P$ is the set of directed paths through the layer graph from input to output layer and $p$ is a path in that set. $w(p)$ is the weight of the path, which is proportional to the fraction of the signal flowing through the network that flows through that path. We have $\sum_{p \in P} w(p) = 1$. Finally, $\mathfrak{n}_{\tau_l}(\mathfrak{q}_k,\mathfrak{c}_k)$ is the `activation function NLC' of $\tau_l$ used at activation layer $f_l$ with parent $f_k$, which we show can be viewed as the value of the NLC of $\tau_l$ with respect to meta-Gaussian input (section \ref{actFunMetaGaussianSection}) \contrib, which is the kind of input one would expect stemming from a fully-connected layer.

In plain words, the NPE states the following: The square of the mean field NLC of the architecture is the weighted average of the mean field NLCs of the directed paths through the layer graph. The mean field NLC of each path is the product of the activation function NLCs on that path. The path is weighted according to the fraction of the total signal that flows through it.

The nonlinearity path equation instructively explains the value of the initial NLC in terms of the activation function NLCs, which can themselves be regarded as measures of degree of nonlinearity and expressivity. If the layer graph consists only of a single path, the mean field NLC is the product of all activation function NLCs. Hence, expressivity compounds exponentially with depth in certain plain architectures, which has been widely observed in the past. The nonlinearity path equation immediately yields a wide range of other properties given in sections \ref{pathEquationSection} and \ref{nlcExplainSection}, as well as in later chapters as well as throughout this and the previous two subsections.

\paragraph{A mean field theory of activation functions} We further investigate the activation function covariance kernel and prove a range of results (sections \ref{meanFieldActivationSection}, \ref{meanFieldPracticalSection}) \contrib. We provide an extensive repository of key information about all activation functions used in this work, including many popular activation functions (sections \ref{meanFieldActivationSection}, \ref{nlcLinearApproximationSection}) \contrib. We then use this, together with the recursive calculation rules and the NPE, to develop a taxonomy of the behaviors of activation functions and of plain architectures containing those activation functions (sections \ref{meanFieldActivationSection}, \ref{gaussianStabilityExplanationSection}, \ref{plainArchitectureSection}, \ref{surveySummarySection}, \ref{vanishMultipleSection}, \ref{orderChaosLimitedSection}) \contrib. We give the highlights below.

We prove that under mild conditions, the mean field NLC of an activation function depends simply on its covariance kernel and is greater or equal to 1 (sections \ref{actFunMetaGaussianSection}, \ref{actFunCovKernelSection}, \ref{actFunNLCsection}) \contrib. 

$$\mathfrak{n}_\tau(c) = \sqrt{\frac{\mathfrak{C}_\tau'(1)(1-c)}{\mathfrak{C}_\tau(1)-\mathfrak{C}_\tau(c)}} \ge 1$$

Here, $\mathfrak{C}_\tau(1,c)$ is shortened to $\mathfrak{C}_\tau(c)$ and $\mathfrak{n}_\tau(1,c)$ is shortened to $\mathfrak{n}_\tau(c)$. We focus on the case $q=1$ for brevity. We then use this result to prove the equivalent statement for A-architectures (section \ref{networkCovKerSection}) \contrib.

$$\mathfrak{n}(f,q,c) = \sqrt{\frac{\frac{d}{dq}\mathfrak{C}(q,q')|_{q'=q}(q - c)}{\mathfrak{C}(q,q) - \mathfrak{C}(q,c)}}$$

The term on the right-hand side represents the normalized first-order approximation of the bandwidth of the covariance kernel, which ties the NLC to model complexity (section \ref{nlcSummarySection} above).

We study the behavior of plain architectures with various activation functions. A plain architecture is an architecture with alternating fully-connected and activation layers, where the fully-connected and the activation layers are identical respectively, as defined in section \ref{actFunLengthKernelSection}.

The sigmoid activation function is a bad choice for a deep plain architecture. If weights are initialized to attain mean field scale stability, mean field LBIAS grows exponentially with depth, which makes the architecture untrainable with standard methods (section \ref{plainArchitectureSection}, \ref{outputBiasSection}) \contrib. The underlying property of sigmoid is $\frac{\mathfrak{C}_\tau(1)'}{\mathfrak{C}_\tau(1)} < 1$, which is equivalent to $\mathbb{E}_{s\sim\mathcal{N}(0,1)}\tau'(s)^2 < \mathbb{E}_{s\sim\mathcal{N}(0,1)}\tau(s)^2$.

The tanh activation function does not have this issue, because $\mathbb{E}_{s\sim\mathcal{N}(0,q)}\tau(s) = 0$ for all $q$. Hence, if mean field $LBIAS_l$ is zero at the input layer, it is also zero throughout the architecture. Instead, the mean field NLC grows exponentially from layer to layer if scale stability holds (section \ref{plainArchitectureSection}) \contrib. The underlying property is $\frac{\mathfrak{C}_\tau(1)'}{\mathfrak{C}_\tau(1)} > 1$, or, equivalently, $\mathbb{E}_{s\sim\mathcal{N}(0,1)}\tau'(s)^2 > \mathbb{E}_{s\sim\mathcal{N}(0,1)}\tau(s)^2$. However, if the initial weight variance is too small, then mean field $LSCALE_l$ collapses exponentially to zero from layer to layer (section \ref{plainArchitectureSection}) \contrib. Conversely, if the weight magnitude is very large, the gradient becomes unstable to the point of exploding in expectation but vanishing in probability, as we informally explain (section \ref{vanishTanhSection}). The same effect can happen when weights grow during training, which means tanh architectures have low training stability (section \ref{covariateShiftSection}) \contrib.

The ReLU activation function requires individual weights to be initialized with a variance of 2 over the architecture width to attain mean field scale stability, which is known as He initialization (section \ref{plainArchitectureSection}) \contrib. However, a ReLU architecture is still relatively forgiving to a lack of scale stability because $c\tau_\text{ReLU}(s) = \tau_\text{ReLU}(cs)$ for all $c>0$, which also implies higher training stability (sections \ref{forwardStabilitySection}, \ref{covariateShiftSection}) \contrib. While mean field gradients are stable in a He-initialized ReLU architecture, both mean field NLC and LBIAS grow linearly with depth. The underlying property is $\frac{\mathfrak{C}_\tau(1)'}{\mathfrak{C}_\tau(1)} = 1$, or, equivalently, $\mathbb{E}_{s\sim\mathcal{N}(0,1)}\tau'(s)^2 = \mathbb{E}_{s\sim\mathcal{N}(0,1)}\tau(s)^2$. Hence, ReLU still turns out to be a suboptimal choice for deep plain architectures.

The SELU activation function is the best choice for deep plain architectures out of the activation functions we consider. A SELU architecture exhibits mean field scale stability when the initial weight variance is 1 over width (section \ref{plainArchitectureSection}) \backgr, which can be considered the default choice. In that case, mean field LBIAS converges to zero (section \ref{plainArchitectureSection}) \contrib. As with tanh, mean field NLC grows exponentially with depth, but now the growth rate is slow. The underlying property is $\mathfrak{n}_{\tau_\text{SELU}}(0,1) = 1.035$. Hence, by the NPE, depth can be as high as $\log_{1.035}5 \approx 46$ and $\mathfrak{n} < 5$ will still hold in accordance with ZSAD guideline \ref{npcNLC} (section \ref{vanishGoodSection}) \contrib. For comparison, $\mathfrak{n}_{\tau_\text{tanh}}(0,1) = 1.085$.

Based on the value of $\frac{\mathfrak{C}_\tau(1)'}{\mathfrak{C}_\tau(1)}$, we find that there is a trichotomy in terms of how activation functions behave in scale-stable plain architectures (see the segment on `neuron bias' in section \ref{practicalGuideSection} above).

The above insights about popular activation functions were largely informally known to the community. The observations we make are similar to those of e.g. \citet{correlationLimit}. However, to our knowledge, we are the first to derive all these insights rigorously out of a theoretical framework. For brevity, we do not give our main theorems \ref{covkerLregular} and \ref{covkerCregular} here. See sections \ref{actFunLengthKernelSection} and \ref{actFunCovKernelSection}.

\paragraph{Gaussian stability} We discover a class of architectures among those we study that does not have a significant fraction of the properties we demonstrate in this work. We introduce this ``meta-property'' as `Gaussian stability' (section \ref{gaussianStabilityExplanationSection}) \contrib. The name stems from the fact that these architectures do not have meta-Gaussian linear layers in the initial state as summarized above. Unfortunately, we have not reached the point of defining Gaussian stability in terms of a well-defined metric. Hence, diagnosing which of our architectures exhibits Gaussian stability is similarly not quite precise. Further study is required to determine the ``essence'' of this emergent phenomenon. Note that we use ``Gaussian instability'' synonymously with ``absence of Gaussian stability''.

Almost all of our architectures that can be said to exhibit Gaussian instability use either the square ($\tau_\text{square}(s) = s^2$) or odd square ($\tau_\text{odd square}(s) = s*|s|$) activation function (section \ref{metricTerminologySection}) \contrib. We uncover the underlying property, which we term `mean field Gaussian instability' (section \ref{gaussianStabilityExplanationSection}) \contrib.

$$\frac{q\frac{d}{dq}\mathfrak{C}_\tau(q,q)}{\mathfrak{C}_\tau(q,q)} > 1$$ 

Specifically, for square and odd square, this value is always equal to 2. We note that this value is equal to 1 for ReLU, which can cause architectures based on ReLU to exhibit a mild degree of Gaussian instability, a property which we term `Gaussian edge' (sections \ref{meanFieldDistributionEmpiricalSection}, \ref{gaussianStabilityExplanationSection}) \contrib.

Mean field Gaussian instability causes $\mathfrak{C}_\tau$ to be unstable, i.e. to grow small perturbations. This can cause small deviations of practical metric values from their mean field limit to explode exponentially or even super-exponentially during forward propagation.

We show that many properties depend, to one degree or another, on Gaussian stability \contrib. They are as follows. (i) Fully-connected layers are meta-Gaussian meta-distributed (section \ref{meanFieldDistributionSection}). (ii) Mean field limits in A-architectures are practically predictive (sections \ref{meanFieldPracticalSection}, \ref{meanFieldCNNsection}). (iii) The statistical estimator of the NLC is stable (section \ref{nlcRobustDataSection}). (iv) The NLC is robust to changes in data distribution (section \ref{nlcRobustDataSection}). (v) The NLC is relatively invariant from one draw of the random initialization scheme to another (section \ref{nlcRobustDataSection}). (vi) The NLC can be approximated by even simpler metrics (section \ref{nlcSimpleMetricsSection}). (vii) The NLC is robust to decomposition into the sum-product of NLCs of individual layers (section \ref{nlcDecomposableSection}). (viii) The NLC is robust to width change (section \ref{widthInvarianceSection}). (ix) The NLC changes smoothly and regularly from layer to layer (section \ref{nlcBackpropSection}).

In summary, Gaussian stability modulates the predictiveness of mean field theory and properties that are directly tied to mean field theory (section \ref{nlcExplainSection}). We note that Gaussian instability does not affect the technical validity of mean field theory but ``only'' its practical predictiveness. Gaussian stability as a crucial prerequisite for mean field predictiveness was, to our knowledge, hitherto unknown. As we pointed out, Gaussian stability is also critical for architecture performance. Of course, an architecture cannot even hope to have Gaussian stability if it does not contain e.g. fully-connected or convolutional layers. Hence, this concept may hold the key to explaining the power and necessity of linear layers for deep learning in general. This is an exciting research direction.

\paragraph{Application to CNNs} Due to time and space limitations our mean field analysis in chapter \ref{meanFieldNnaChapter} is, unfortunately, largely focused on fully-connected architectures. In mean field theory, a convolutional layer is modeled as a collection of fully-connected layers. Hence, our theoretical results technically hold for certain convolutional architectures, though interpreting them is much more difficult. We outline this interpretation process and provide initial empirical evidence for practical predictiveness (section \ref{meanFieldCNNsection}) \contrib. The behavior of e.g. activation functions, normalization layers and skip connections remain largely the same for CNNs, though there are a few differences. For example, layer normalization is capable of ensuring Gaussian stability in fully-connected networks, but not in convolutional networks (section \ref{surveyBNsection}) \contrib.

We validate the nonlinearity normalization algorithm (section \ref{nlnormSummarySection} below), which is grounded in mean field theory and specifically the nonlinearity path equation, on convolutional architectures. This further indicates the generality of the underlying principles.

\subsection{Nonlinearity normalization} \label{nlnormSummarySection}

We introduce the `nonlinearity normalization' (nlnorm) algorithm as a design strategy for controlling the architecture's initial NLC, $LBIAS_l$ and $LSCALE_l$ values to achieve ZSAD guidelines \ref{npcNLC} (appropriate NLC), \ref{npcForward} (scale stability) and \ref{npcBias} (avoiding neuron bias) by minimally modifying the architecture's activation functions. We give its form for simple architectures in figure \ref{boxnlnorm} along with a wider discussion (section \ref{nlnormDefinitionSection}) \contrib.

nlnorm is based on the insight derived from the nonlinearity path equation (section \ref{meanFieldSummarySection} above) that changing the degree of nonlinearity of a network's activation functions changes the degree of nonlinearity of the network itself. Specifically, the mean field NLC $\mathfrak{n}$ is monotonic in the activation function NLCs $\mathfrak{n}_{\tau_l}$. Hence, replacing an activation function $\tau_l^{(1)}$ with another $\tau_l^{(2)}$ such that $\mathfrak{n}_{\tau_l^{(2)}}(\mathfrak{q}_k,\mathfrak{c}_k) < \mathfrak{n}_{\tau_l^{(1)}}(\mathfrak{q}_k,\mathfrak{c}_k)$ is guaranteed to reduce $\mathfrak{n}$, assuming there are no knock-on effects on downstream $\mathfrak{q}$ and $\mathfrak{c}$ values.

nlnorm replaces each instance of an activation function $\tau(s)$ that occurs in the architecture by a modified activation function $c\ddot{\tau}(l,s) + b$, where $l$ is a tunable hyperparameter we term the `linearization parameter' that controls the degree of nonlinearity. A basic choice is $\ddot{\tau}(l,s) = \tau(s) + ls$. Because $\mathfrak{n}_{\tau(s) + ls}(1,0)^2$ is proportional to the $L2$ linear approximation error of $\tau(s)+ls$ (section \ref{nlcMeaningSection} above), the $ls$ term is guaranteed to control the activation function NLC with respect to the unit Gaussian (section \ref{linearizationSection}) \contrib. Indeed, we find that the $L2$ linear approximation error of practical activation functions explains their degree of nonlinearity to a significant degree (section \ref{nlcLinearApproximationSection}) \contrib. For a given $l$, we jointly set $b$ and $c$ to attain $\mathbb{E}_{s \sim \mathcal{N}(0,1)}c\ddot{\tau}(l,s)+b=0$ and $\mathbb{E}_{s \sim \mathcal{N}(0,1)}(c\ddot{\tau}(l,s)+b)^2=1$, which correspond to avoiding neuron bias and ensuring scale stability respectively. nlnorm uses ``normalization by recursion'', where layers closer to the output are normalized based on assumptions derived from normalizing layers closer to the input. See section \ref{nlnormDefinitionSection} for details.

We show that nlnorm is indeed effective at controlling the NLC as intended (section \ref{nlnormControlSection}) \contrib \ and that it is consistently able to find good NLC levels (section \ref{nlnormWorksSection}) \contrib. Deploying nlnorm for an architecture that has suboptimal NLC or LBIAS levels leads to massive reduction in test error (section \ref{nlnormPerformanceSection}) \contrib, which implies that the NLC and neuron bias are causally related to architecture performance.

By way of $l$, the NLC of the architecture becomes a tunable hyperparameter. This turns model expressivity itself into a hyperparameter. Thus, we argue that tuning the NLC is of similar importance to tuning key hyperparameters like learning rate and width. A drawback of turning the NLC into a hyperparameter is that it does require tuning. While we suggest the $[1,5]$ range for the initial NLC, there is still a lot of room in that range. We find that different types of architectures perform optimally with different initial NLCs, including NLCs larger than 5 for certain suboptimal architecture types (section \ref{nlnormBestNLCSection}) \contrib. Even for tuned architectures, the NLC appears to change chaotically from initial to final state within a limited range (section \ref{nlnormBestNLCSection}) \contrib. More work is needed to make more specific NLC recommendations.

Using nlnorm, we attain close-to-optimal test error values using a range of activation functions, including activation functions that we designed ad hoc and which induce random error values when nlnorm is not used (section \ref{nlnormActFunDesignSection}) \contrib. This result suggests that the degree of nonlinearity of an activation function is more important for performance than its overall shape. Hence, the success of many activation functions proposed over the years, such as SELU \citep{selu}, PReLU \citep{heInit}, ELU \citep{elu} or Swish \citep{swish}, has likely more to do with their nonlinearity than with other properties. However, by improving arbitrary activation functions, nlnorm also opens up activation function design to new possibilities and reduces the need to use specific activation functions, such as ReLU.

We present evidence that nlnorm may significantly reduce the need for normalization layers, like batch or layer normalization (section \ref{nlnormvsbnsection}) \contrib. This is because normalization layers also control the architecture's scale stability and neuron bias (section \ref{surveyBNsection}). We also present evidence that nlnorm may significantly reduce the need for skip connections (section \ref{nlnormvsbnsection}) \contrib. This is because skip connections also tend to reduce the NLC (section \ref{skipConnectionsSection}). Note that replacing $\tau(s)$ with $\tau(s) + ls$ corresponds to bypassing the activation layer with a skip connection of strength $l$. We do not claim that nlnorm entirely replaces these popular building blocks. For example, normalization layers also increase training stability (section \ref{practicalGuideSection} above) and batch normalization has been linked to orthogonality \citep{bnOrthogonality}. In our experiments, skip connections made an independent contribution to test error reduction which is not explained by any of our ZSAD guidelines from section \ref{practicalGuideSection} (section \ref{skipConnectionsSection}) \contrib.
 
\subsection{Better practices I: well-defined metrics} \label{relatedWorkSummarySection}

In section \ref{nlcSummarySection} above, we summarized the properties that make the NLC well-defined, meaningful and general. In section \ref{practicalGuideSection}, we introduced $LBIAS_l$ and $LSCALE_l$ as measures for neuron bias and scale stability respectively. While more work is necessary to determine the best measure(s) for Gaussian stability, we suggest several (section \ref{gaussianStabilityExplanationSection}). We also define mean field Gaussian stability (section \ref{gaussianStabilityExplanationSection}).

When it comes to understanding neural architectures, contemporary research often runs on the principle of ``word associations''. Studies label experimental observations with terms. These terms later get reused by other authors when they are subjectively reminded of a study that previously used the term. Over time, ideas get diluted and terms drift, obtain multiple meanings or lose meaning altogether. Of course, neural networks have achieved impressive practical success under this regime, so it is not without value. This work also picks up on some ambiguous terms like expressivity. Nevertheless, a core objective of this work is to serve as a first step towards a template for a different and complementary approach to understanding neural architectures based on well-defined, widely-applicable and extensively-validated metrics. We hope that ZSAD can be an emblem for this approach, just like NAS is an emblem for large-scale search.

In this work, we document some of the pitfalls associated with ill-defined ZSAD guidelines. Specifically, we focus on what we believe are three of the most popular and fleshed-out guidelines. Our high-level criticisms are of a general nature and not specific to these three guidelines.

\begin{itemize}
\item ``avoid exploding / vanishing gradients'' (section \ref{vanishingGradientSection}; e.g. \citet{RNNvanishingGradient,hochreiterThesis,heInit, depthScalesMeanField,resNetMeanField, normalizedInitialization,orthogonalInitialization, RecurrentNetsPascanu,eigenspectrum} and many more)
\item ``choose an architecture on the edge of chaos'', ``ensure correlation preservation'' and ``ensure signal propagation'' (section \ref{orderChaosSection}; e.g. \citet{correlationLimit,depthScalesMeanField,resNetMeanField, meanFieldCNN,meanFieldRNN,meanFieldDeepField,meanFieldBN})
\item ``use an appropriate depth'' (section \ref{depthSection}; e.g. \citet{depth8,depth10}; \citet{depth9} provides a recent overview)
\end{itemize}

The concepts order / chaos / edge of chaos, correlation preservation and signal propagation co-occur in the same set of studies and are used somewhat interchangeably. Hence, we study them jointly. While the phrase ``correlation preservation'' is never used explicitly in prior work, we use it to refer to the implied desideratum that the correlations of inputs should be preserved from layer to layer during forward propagation. Throughout this subsection as in chapter \ref{relatedWorkChapter}, we shorten exploding / vanishing gradients to EVG and order / chaos / edge of chaos to OCE.

While we are critical of the high-level discourse around architecture design, we recognize the many invaluable contributions of the works cited above. Without, for example, mean field theory, this work would not have been possible in its current form.

We now cover what we view as the key advantages of well-defined metrics.

\paragraph{Generalizability} There is no agreed-upon way to determine whether a given network exhibits EVG (section \ref{vanishWellDefinedSection}). There are some measures for EVG that are valid for general networks, such as average length of the gradient of the loss with respect to the input. However, it is not clear to what degree any specific measure captures the EVG concept, because any property demonstrated for EVG may only hold for whatever measure was used by the study that demonstrated it. OCE and correlation preservation can be considered as well-defined in limited contexts via the depth scale / timescale metric (section \ref{chaosDepthScaleSection}). Regardless, determining whether a network exhibits order or chaos is generally a question requiring expert judgment. This issue gets worse when architectures are non-homogeneous, e.g. when they use different initial weight variances or activation functions in different layers (section \ref{orderChaosLimitedSection}). Signal propagation is never quantified (section \ref{chaosSignalSection}). The notion of depth is inherently ill-defined (section \ref{neuralNetworksSection}). While there are agreed-upon definitions in limited contexts, like plain feedforward networks, they do not easily generalize (section \ref{depthSection}). 

Hence, applying EVG, OCE and depth to networks or architectures remains a case-by-case judgment. Therefore, it is challenging for a non-expert to apply these concepts to novel situations. Applying a metric is automatic as long as it is mathematically valid and there are no computational issues. Further, once properties have been demonstrated for a concrete metric in one context, it is possible to determine their validity in another context. Checking the validity of the NLC's properties from section \ref{nlcPropertiesSection} not only provides information about the value of the NLC in a novel situation, but also about the challenges and idiosyncrasies of that situation in general. Since the ``properties of EVG'' are informal beliefs, they are not subject to automated verification.

\paragraph{Lack of ambiguity} Based on how an ill-defined concept is quantified, we obtain very different results. For example, EVG can be measured based on the length of the gradient of the loss with respect to layers or based on the quadratic mean of gradient components. While both measures are similar, they behave very differently when the width of layers varies (section \ref{vanishWidthSection}) \contrib. Some very prominent papers deal with the interaction of EVG and width \citep{normalizedInitialization,heInit}. However, we show that some of their recommendations are an artifact stemming from their quantification of EVG and not reflective of a real problem that requires fixing (section \ref{vanishWidthSection}) \contrib. Because metrics do not need to be quantified every time they are used, results are unambiguous.

EVG is commonly applied to both recurrent networks and feedforward networks. We explain that EVG has very different underlying meanings in both contexts, and that EVG has yet another meaning for sigmoid / tanh networks (section \ref{vanishMultipleSection}) \contrib. While there is no metric that precisely captures all these situations at once, they get amalgamated under a single term because they appear similar at a very high level. In contrast, the use of metrics can lead situations that require different metrics to be recognized as different.

\paragraph{Accountability} Given a claim that a metric predicts performance, it is possible to study its limitations by pointing to contexts where it fails to predict performance. We do this for our own guidelines. We show that scale stability, as measured by $LSCALE_l$, can be irrelevant depending on the design process (section \ref{linearEquivarianceSection}) \contrib. We show that neuron bias, as measured by $LBIAS_l$, can be mitigated with a bespoke training algorithm, and its importance appears limited to the output layer and the parents of linear layers (section \ref{outputBiasSection}) \contrib. We explain how our empirical results are dependent on our choice of architectures (section \ref{architectureSensitivitySection}). We generate artificial datasets where our recommended range for the initial NLC of [1,5] is not applicable (section \ref{bestNlcSection}) \contrib.

We point out significant issues with specific ways of measuring EVG and correlation preservation. If EVG is measured via the length of the gradient vector, it becomes a superficial property of the network that can be easily confounded by manipulating irrelevant aspects, such as the scale of the input in a network with batch normalization (BN), the scale of weights in a ReLU or BN network, or the scale of the loss function (sections \ref{linearEquivarianceSection}, \ref{vanishWhyBadSection}) \contrib. We show that the correlation of different inputs or layer values can be manipulated to arbitrary degrees by e.g. shifting the inputs or using intermediate bias layers (section \ref{chaosCorrelationMeaningSection}) \contrib. In each case, EVG and correlation preservation do not capture a robust pathology and therefore lack meaning.

``Exploding gradients are worse than stable gradients'', ``chaos is worse than edge of chaos'' and ``depth is beneficial'' are general mantras. We show that it is just as easy to argue for their opposites. When the property ``exploding gradients'' is interpreted as ``exponentially growing gradients'', it can actually be desirable (section \ref{vanishGoodSection}) \contrib, whereas networks with stable gradients can exhibit both an excessive NLC and excessive neuron bias (section \ref{vanishDichotomySection}) \contrib. Chaos is often preferable to the edge of chaos for the same reason (section \ref{chaosGoodSection}) \contrib. OCE has little meaning for architectures that behave differently than their infinite depth limit (section \ref{chaosLimitSection}) \contrib. In our empirical studies, we find that depth is negatively associated with test error (section \ref{depthSection}) \contrib \ and OCE does not drive test error (section \ref{chaosGoodSection}) \contrib.

Ultimately, it is difficult to criticize general mantras because they can be re-interpreted ad hoc to excuse any shortcoming. When they fail to predict performance, the criticism can be deflected by claiming that the guidelines were somehow not understood or were not applied correctly. They encourage the use of the ``no true Scotsman'' fallacy.

\paragraph{Improvability} We document contexts in which EVG / OCE do predict performance. In fact, exploding gradients can correspond to excessive NLC and vanishing gradients can correspond to excessive neuron bias (section \ref{vanishMultipleSection}) \contrib. Similarly, there is a chaos-NLC and order-neuron bias correspondence (section \ref{chaosOppositeSection}) \contrib, as well as a correlation change-neuron bias correspondence (section \ref{chaosCorrelationMeaningSection}) \contrib. However, we argue that considering NLC and LBIAS directly, at least in the context of feedforward networks, is preferable because it isolates the differences between the NLC and neuron bias pathologies, rather than lumping both together under a single guideline. Ultimately, architectures that seem ideal under EVG / OCE actually tend to suffer from both excessive NLC and LBIAS, as mentioned above. The EVG / OCE terminology suggests that exploding and vanishing gradients and order and chaos, respectively, are opposites, when they really correspond to different pathologies (sections \ref{vanishMultipleSection}, \ref{chaosOppositeSection}) \contrib. Throughout this work, we document the robustness and practicality of the NLC. The concept of depth is questionable in this regard (section \ref{depthSection}) \contrib.

We argue that the NLC, along with neuron bias, largely supersedes EVG / OCE / depth, at least as a ZSAD guideline for feedforward architectures. However, we recognize that this is a difficult argument to make, because ill-defined concepts are loaded with connotations and intuitions. Given a set of known properties of e.g. the gradient length metric, it could be rapidly shown that another metric like the NLC has better properties. However, if ZSAD guidelines are simply viewed as subjective lenses for judging architectures, there is no scientific argument that any one is more effective than any other. Hence, progress stalls.

\paragraph{Democratization} In order to take part in machine learning research, one must use the terms of the machine learning community. However, it is difficult to conceptualize exploding gradients, depth or signal propagation in the same way as an established group of researchers does. Replicating how researchers talk about vague terms becomes a barrier to entry. Common attitudes, rather than being testable hypotheses, become linguistic gatekeepers.

However, when one actually drills into the details, one finds that even famous papers make statements about EVG, OCE or signal propagation that are, at least, misleading or overly general (sections \ref{vanishPapersSection}, \ref{chaosGoodSection}, \ref{chaosSignalSection}) \contrib.

\subsection{Better practices II: careful experimental protocol} \label{experimentsSummarySection}

One of our goals in this work is to elevate the standard of architecture design research specifically and deep learning research in general by deploying and promoting a careful experimental protocol. In our opinion, protocol deficiency is by far the most pervasive and significant shortcoming among deep learning studies.

We detail our protocol in chapter \ref{empiricalStudiesChapter}, which we consider a contribution in its own right \contrib. Most of our experimental results are derived from two large studies: study A using fully-connected architectures (section \ref{studyASection}), and study B using convolutional architectures (section \ref{studyBSection}). We also conducted some additional experiments using fully-connected architectures and the same training protocol as study A (section \ref{additionalExperimentsSection}). Of course, our protocol also still has limitations, which we detail in section \ref{limitationsSection} and summarize in subsection \ref{limitationsSummarySection}.

Of course, not every deep learning study requires the same protocol. However, a significant fraction of studies does the following.

\begin{itemize}
\item Compare different instances of a certain ``target aspect'' of a deep learning pipeline, e.g. different activation functions, different learning rate schedules or different objective functions.
\item Choose one or more deep learning pipelines to embed these instances into in order to compare them by executing the entire pipeline and evaluating its performance with each of the instances. ``Performance'' can refer broadly to any behavior of interest. For example, different activation functions may be compared via the test error or adversarial robustness attained by the pipeline into which they are embedded.
\end{itemize}

Large portions of our work also fall under this ``comparative study'' paradigm.

Below, we give standards that we advocate for this paradigm. We detail how we followed the standard in this work, give examples of how following the standard impacted our results and make specific recommendations.

\paragraph{Control confounders: parameter dimensionality, scale stability and loss function}

Changing a target aspect of a deep learning pipeline can have side-effects on properties that are known to be performance drivers. For example, changing depth can change parameter dimensionality, which is a driver of test error. If this occurs, any observed performance difference may be caused by the property that is varying indirectly, rather than the explicitly varied target aspect. For example, a changing test error as depth varies may not be caused by the change in depth, but by the change in parameter dimensionality. One way to counteract this is to change other aspects of the pipeline jointly with the target aspect to keep the property that is varying indirectly approximately constant. For example, one may change width along with depth to control parameter dimensionality.

In study A, we do just this. When varying depth, we also varied width to compensate and kept parameter dimensionality around 1 million. In study B, we did not vary width or depth. Parameter dimensionality is arguably the most fundamental performance driver in machine learning in general (section \ref{widthSection}) \backgr, and is controlled in many deep learning studies.

In both study A and B, we controlled scale stability to a significant degree. We did not choose weight initialization schemes that differed significantly from the LeCun initialization, where the initial weight variance is 1 over the number of multiply-adds that contribute to a neuron in a linear layer. All of our activation functions $\tau$ were the result of scaling with a constant to achieve $\mathbb{E}_{s\sim\mathcal{N}(0,1)}\tau(s)^2 = 1$. Scale stability is known to significantly impact performance (\backgr \ and section \ref{practicalGuideSection} above). Because we co-vary initial weight variance with width, we were able to show how e.g. test error and NLC are invariant in certain contexts as width changes (sections \ref{widthInvarianceSection}, \ref{vanishWidthSection}).

In study A, we controlled the magnitude of the values that are fed into the softmax+cross-entropy loss function. Specifically, we used an augmented version of the standard softmax+cross-entropy loss function that scales the network outputs before applying softmax+cross-entropy (section \ref{studyATrainingSection}). We knew that softmax+cross-entropy has a strong preference for values with an overall magnitude around 1, as we also demonstrate (section \ref{forwardStabilitySection}) \contrib. On one occasion, we studied the behavior of an architecture as the magnitude of the input varied, which caused the magnitude of the output to change as well. We were only able to observe a consistent test error value as input magnitude decreased because we used the augmented loss function (section \ref{nlcRobustDataSection}). Since we were interested in studying architecture performance, and we did not consider the loss function to be part of the architecture, we wanted to minimize the impact of the loss function as much as possible. In general, we recommend using this augmented loss function when comparing architectures that return outputs of differing magnitudes for classification. We know of studies that did not and were likely subject to confounding.

In study B, we controlled both scale stability and neuron bias when varying the linearization parameter $l$ of the activation function $c\ddot{\tau}(l,s)+b$ by also setting parameters $b$ and $c$ to achieve $\mathbb{E}_{s\sim\mathcal{N}(0,1)}(c\ddot{\tau}(l,s)+b)^2 = 1$ and $\mathbb{E}_{s\sim\mathcal{N}(0,1)}c\ddot{\tau}(l,s)+b = 0$ (section \ref{nlnormSummarySection} above).

In study B, we fixed the depth of our convolutional architectures to control the spatial frequency composition of the output, as studied by \citet{meanFieldCNN}.

In general, when attempting to discover new performance drivers, e.g. the NLC, that are intended to be synergistic and complementary to known drivers, such as those mentioned above, it can be valuable to control those known drivers. This avoids re-discovery and diluted results (section \ref{architectureSensitivitySection}). Also consider that ZSAD guidelines tend to behave in an all-or-nothing manner when predicting test or training error (section \ref{practicalGuideSection} above). Hence, unless a significant number of studied architectures follow all guidelines to a significant degree, a large fraction of them may exhibit random error, which would leave little performance variation to explain.

Of course, the caveat of controlling pipeline properties is that it becomes necessary to understand the interaction of the target aspect with the properties that were controlled (e.g. sections \ref{widthInvarianceSection}, \ref{forwardStabilitySection}, also section \ref{practicalGuideSection} above) to understand the limits of generality (sections \ref{linearEquivarianceSection}, \ref{chaosCorrelationConfoundingSection}, also section \ref{relatedWorkSummarySection} above). It is also necessary to know about potential confounders. Hence, eliminating confounders is an iterative process of discovery.

\paragraph{Independently and exhaustively tune key hyperparameters: learning rate and NLC}

When changing the target aspect of a deep learning pipeline, another property of that pipeline can ``fit'' differently with different instances of the target aspect. If this occurs, then any observed performance difference may be caused by the difference in fit with that property, rather than the target aspect directly. In that case, it is necessary to exhaustively search over values of that property (``hyperparameter'') and pair each value with each instance of the target aspect to fairly assess performance.

The key example of such a hyperparameter that is universal to nearly all pipelines is learning rate. We train each study A architecture independently with 40 starting learning rates (SLRs) and we train each study B architecture independently with 20 SLRs. We divided the learning rate 10 times by 3 (study A) / 3 times by 10 (study B) during each training run. We show that indeed, the best SLR varies enormously across our architectures (section \ref{learningRateSection}) \contrib, making this tuning essential for a fair comparison, as we do not consider any specific SLR value as privileged over any other. Some of our architectures could not generalize but were able to achieve close-to-zero training error when (i) we considered an even wider range of 60 SLRs that spanned close to 30 orders of magnitude and (ii) early stopping based on validation error was not used and architectures were allowed to train for as long as training error kept improving. We re-trained many of our study A architectures with these protocol changes and successfully achieved low training errors for almost all architectures for which this was possible based on our ZSAD guidelines. Without exhaustive tuning, we would have obtained that high-NLC, and hence high-expressivity, architectures are untrainable (section \ref{expressivitySection} above). On multiple occasions in this work, we find that test error is invariant when a specific hyperparameter value varies (weight variance scaling factor: sections \ref{forwardStabilitySection} and \ref{linearEquivarianceSection}; loss scaling factor: section \ref{linearEquivarianceSection}). These discoveries depend on allowing the learning rate to co-vary with that hyperparameter. Further, we find on one occasion that test error is minimized by architectures of different NLC levels on different artificial datasets where the true input-label function has different degrees of nonlinearity. Again, pairing high-NLC architectures with smaller learning rates was essential for this discovery (section \ref{bestNlcSection}).

While not all deep learning studies will require considering 60 different SLRs, we advocate finding the ``best'' learning rate for each individual pipeline configuration in any comparative study. The failure to do this has likely confounded many studies. Developing efficient methods to make this computationally feasible is an interesting topic of study. We make a guess at what range of SLRs might be ideal for an architecture as part of our experimental protocol, and we base that guess on the magnitude of the parameter gradient (section \ref{studyATrainingSection}), with some success (section \ref{learningRateSection}). We analyze our results with respect to learning rate as summarized above in section \ref{practicalGuideSection}.

In this work, we choose an SLR for each architecture based on which of them yields the lowest validation, test or training error, depending on context. This is a reasonable choice when comparing architectures based on error. However, it can be problematic when comparing other architecture properties. Specifically, there exist architectures that do not attain a better-than-random error across any reasonable set of SLRs. Choosing an SLR based on lowest error then leads to an essentially random choice. However, SLR still has a massive, non-random impact on properties of an architecture's final state after training, such as NLC and scale stability (section \ref{nlcEvolutionSection}, \ref{meanFieldPracticalEmpiricalSection}, \ref{covariateShiftSection}) \contrib. With a random SLR, we cannot meaningfully associate properties of the final state with the architecture itself. Hence, throughout this work, when plotting the value of any metric not based on error, we discard all architectures that did not attain a better-than-random validation / test / training error when SLR was selected based on validation / test / training error respectively (section \ref{metricTerminologySection}).
 
While learning rate is the classical example of a key hyperparameter in deep learning, we uncover the NLC as another such hyperparameter. Our nlnorm algorithm makes the NLC, and thus model expressivity, tunable. When the NLC is controlled for via exhaustive tuning, the test error impact of some of the most popular building blocks, like skip connections, normalization layers and various activation functions, is very different than when the nonlinearity is ignored (section \ref{nlnormSummarySection} above). While we do not claim that any one building block becomes obsolete with nlnorm, we argue that tuning the NLC when comparing deep learning pipelines (i) would not only likely have simplified and accelerated the progression of deep learning research in the past, but (ii) will be important going forward in order to avoid ``discovering'' ever more methods that reduce error indirectly by simply reducing NLC, while being believed to succeed for various quirky reasons.

A major challenge when tuning hyperparameters for deep learning is that the architecture's final state is often extremely sensitive to very small perturbations in the value of hyperparameters that affect the initial state or the training procedure. Hence, it is usually impossible to find the global minimum of a performance measure like error across a continuous hyperparameter space, such as the space of all SLRs or the space of all linearization parameters when nlnorm is used. When using e.g. grid or random search or even Bayesian optimization, we have to be content with choosing the hyperparameter configuration from a discrete set of sample points without a guarantee about the performance of other configurations. We further discuss this point in section \ref{sharpValleySection}.

\paragraph{Cover a broad and unknown range of pipelines} Above, we discussed controlling known performance drivers. In addition to those, many other pipeline properties likely have a small or unpredictable impact on behavior. Across our architectures, we varied: depth; activation function; initial weight variance; type of weight initialization scheme; presence and initialization of bias and elementwise multiplication layers; normalization layer; presence, type and strength of skip connections; data processing; presence and type of pooling; presence of data augmentation. This is a wider range than the vast majority of deep learning studies. Taking into account learning rate tuning, we conducted nearly 300,000 independent training runs on fully-connected architectures and nearly 12,000 independent training runs on convolutional architectures.

Of course, no matter how broad the set of pipelines considered, the outcome of a comparative study will ultimately depend on the set of pipelines chosen (section \ref{architectureSensitivitySection}). We argue that a key to making a comparative study more scientifically valid is to vary pipeline properties where the performance impact is not known. When choosing our architecture space, we had little understanding of how the properties we varied would ultimately influence the NLC and performance. For example, we were not aware of the nonlinearity path equation. We designed and conducted study B at a later time than study A, when we already had a good understanding of the NLC. Hence, we decided to vary properties like pooling and data augmentation that we did not consider in study A.

If we had not ``invented'' the square and odd square activation functions (figure \ref{actFunIllu}), we would not have discovered Gaussian stability, which is a key condition for many properties of the NLC (section \ref{gaussianStabilityExplanationSection}). Case in point, as far as we know, no mean field study has yet considered an activation function with mean field Gaussian instability. If we had not considered high depths, we would have discovered neither the impact of noise stability (section \ref{noiseStabilitySection}) nor the limits of the computability of the NLC at a given level of floating-point precision (section \ref{nlcComputeSection}). If we had not used architectures that fall under each of the three types of the NLC-LBIAS trichotomy (section \ref{practicalGuideSection} above), we would not have discovered that trichotomy, which reveals how even simple architectures with a small NLC can fail.

While we generally consider only a single training algorithm per architecture (SGD for study A and momentum for study B), we do verify that Adam behaves equivalently to SGD, at least with regards to the predictiveness of the NLC (section \ref{nlcPredictiveSection}) \contrib. While we generally consider only three different datasets, we note that almost all of our results were highly consistent across them, as we document throughout the work \contrib.

\paragraph{Managing computational errors: estimation error, floating-point rounding error, sample independence and batch normalization} It is very rare that deep learning studies explicitly discuss computational errors, i.e. situations where the output produced by a computation differs significantly and systematically from the value of the mathematical function that this computation represents. To be clear, we believe that most studies do not suffer from significant computational errors because they do not consider pipelines or architectures that significantly differ from popular pipelines or architectures. Moreover, one of the factors that can cause an architecture to become popular is robustness to computational errors. However, we advocate that studies that explicitly consider experimental or randomly generated architectures, as we do, for the purpose of a wide comparison explicitly consider and discuss computational properties.

We conducted all computation associated with study A with 64-bit floating-point precision (section \ref{metricUnderflowSection}). This was critical for both training and metric computation. We show that the guideline of noise stability requires high precision when training networks with very high NLCs (section \ref{noiseStabilitySection}) \contrib. Without high precision, we would have obtained that high-NLC, and hence high-expressivity, architectures are untrainable (section \ref{expressivitySection} above). Further, we show how computing metrics such as NLC (sections \ref{nlcComputeSection}, \ref{noiseStabilitySection}), LBIAS (section \ref{outputBiasSection}) and especially MGLLA (section \ref{nlcSensiSection}) requires high numerical precision for high-NLC / high-LBIAS architectures \contrib. While we were unable to use 64-bit precision in study B, we found that it did not contain architectures with NLC or LBIAS values that were as large as those found in study A (sections \ref{nlcPredictiveSection}, \ref{outputBiasSection}).

We defined many of our metrics via an abstract input distribution $\mathcal{D}$. Hence, they need to be computed via statistical estimation (section \ref{metricEstimationSection}). Since our metrics depend on $\mathcal{D}$ via basic probabilistic operators like expectation, standard deviation and median, we can utilize the canonical estimators corresponding to those operators, which are mostly stable. However, this stability can break down in the absence of Gaussian stability (section \ref{nlcRobustDataSection}) \contrib. As summarized above in section \ref{meanFieldSummarySection}, Gaussian instability leads layer square means to diverge from their mean field limit. Because this divergence happens to different degrees for different datapoints, the layer distribution as well as gradient distribution induced by $\mathcal{D}$ can become very heavy-tailed, which in turn increases the sample complexity of e.g. the sample mean estimator.

When defining metrics via the input distribution $\mathcal{D}$, we must also ensure that the data samples we use to compute those metrics are sufficiently independent from the data samples used for e.g. training, data processing or architecture initialization (section \ref{statisticalChallengesSection}). It is especially important not to use the training set to compute metric values after training, except for the purpose of targeted investigation. We follow this rule in this work. It is interesting to note that, especially for our convolutional architectures, we found that the final NLC evaluated on the training set is close to the final NLC evaluated on the test set (section \ref{nlcRobustDataSection}) \contrib.

Batch normalization (BN), while unreasonably powerful, can be a nuisance for analysis. With BN, a feedforward network does not represent a function that maps an individual input to an individual output (section \ref{metricBNsection}). We show that there exists a natural generalization of the NLC for networks with BN such that (i) applying the generalization to networks without BN yields the original definition, (ii) the exact same program can be used in both cases and (iii) statistical efficiency is not significantly compromised (section \ref{nlcComputeSection}) \contrib. While we generally do not discuss this explicitly throughout the work, we go through the same process of deriving such a generalization for the BN case and of ensuring that our program properly captures that generalization for all metrics. We derive the generalization by considering the network as a function that maps batches of inputs to batches of outputs (section \ref{metricBNsection}). In general, we advocate investigating and discussing the process of generalizing concepts to BN networks when appropriate.

\subsection{Limitations and assumptions} \label{limitationsSummarySection}

The following limitations are specific to our work.

\begin{itemize}
\item Due to limited access to code and computational resources, we did not have the chance to replicate in study B all the experimental measurements we made in study A. Hence, many of our results (though not our most important ones) are only validated on fully-connected architectures. There were also some mostly minor limitations associated with the training protocol used for study B relative to study A (section \ref{codeLimitationsSection}). We also did not have the chance to investigate more than one dataset in study B.
\item We only explicitly study feedforward networks that have a static layer graph and represent deterministic functions that take a single input (or batch in the BN case) and return a single output (section \ref{rnnLimitationSection}).
\item We do not validate our results beyond the supervised classification / empirical risk minimization setting (section \ref{settingLimitationSection}).
\item We only validate our results on architectures that resemble popular architectures (section \ref{architectureLimitationsSection}).
\item We do not validate our results on datasets of large size or architectures of large width (section \ref{largeScaleLimitationSection}).
\item We only validate nlnorm on study B and hence convolutional architectures / CIFAR10, as study A was not designed for this purpose (chapter \ref{nlnormChapter}).
\item We do not conduct hyperparameter tuning beyond starting learning rate, though we do tune the NLC in the context of chapter \ref{nlnormChapter}. Specifically, we generally only consider a single training algorithm and loss function per architecture.
\item We implicitly assume that neural networks are differentiable throughout much of our discussion and explicitly assume it in our definition of e.g. the NLC and our theoretical results of chapter \ref{nlcChapter} (section \ref{nonDifferentiableSection}).
\item Many of our theoretical results from chapter \ref{meanFieldNnaChapter} are technically restricted to fully-connected architectures that resemble popular architectures, where activation functions are also twice differentiable (introductions of chapters \ref{meanFieldNnaChapter}, \ref{surveyChapter}).
\item We do not know the range of possible convergence behaviors of the mean field NLC of a plain stable architecture with increasing depth when (i) the normalized covariance kernel of the activation function used has a derivative less than 1 at 1 and is not twice differentiable at 1, or when (ii) the activation function is not piecewise 5-differentiable as assumed in theorem \ref{covkerCregular} (section \ref{actFunNLCsection}).
\end{itemize}

The following limitations are inherent in the type of the empirical studies we conduct.

\begin{itemize}
\item Experimental results that aggregate across architectures are dependent on the precise set of architectures considered (section \ref{architectureSensitivitySection}).
\item Datapoints from our datasets are reused for the statistical estimation of quantities that sequentially depend on each other (section \ref{statisticalChallengesSection}).
\item It is not possible to find the global error minimum across continuous hyperparameter spaces (section \ref{sharpValleySection}).
\end{itemize}

\subsection{Future work} \label{futureWorkSection}

\paragraph{Straightforward extensions} Like in most deep learning studies, many of our results are limited in scope for no other reason than a want for time / computational resources to state, prove and / or validate them in a more general fashion. Below, we give a list of opportunities for generalizing results in a way that we suspect is more or less straightforward. That is, the generalized result should be essentially analogous to the original result. These opportunities roughly map onto the limitations given in the previous subsection.

\begin{itemize}
\item Validate the results of this work that have been validated on only either fully-connected or convolutional architectures on both (section \ref{codeLimitationsSection}, chapter \ref{nlnormChapter}).
\item Validate the results of this work on datasets of large size and / or architectures of large width (section \ref{largeScaleLimitationSection}).
\item Validate the results of this work by conducting hyperparameter tuning across a wider range of training protocols, e.g. across a number of training algorithms and loss functions (section \ref{hyperparameterTuningSection}).
\item Validate the results of this work across a range of architectures not built using popular building blocks and design strategies (section \ref{architectureLimitationsSection}).
\item Validate the results of this work across a wider range of task settings, such as regression, reinforcement learning, image generation or noisy label prediction (section \ref{settingLimitationSection}).
\item Extend the results of this work to other architecture types, such as non-deterministic architectures, RNNs and memory networks (section \ref{rnnLimitationSection}).
\item Give an explicit generalization of nlnorm that ensures scale stability and avoids neuron bias for more general classes of architectures, such as those containing skip connections (section \ref{nlnormDefinitionSection}).
\item Investigate the neural regularity properties of a larger number of practical datasets (section \ref{elemLikeSection}).
\item Explicitly cover one or more cases of non-everywhere-differentiable networks, such as directionally differentiable networks, in the definition of metrics like the NLC and the theory of chapter \ref{nlcChapter} (section \ref{nonDifferentiableSection}).
\item Extend the theory of chapter \ref{meanFieldNnaChapter} beyond A-architectures to e.g. convolutional architectures. Find instructive patterns in the mean field nonlinearity of CNNs analogous to the nonlinearity path equation (chapter \ref{meanFieldNnaChapter}).
\item Determine the convergence behavior of the mean field NLC of a plain stable architecture with increasing depth when the normalized covariance kernel of the activation function used has a derivative less than 1 at 1 and is not twice differentiable at 1 (section \ref{actFunNLCsection}). More fundamentally, extend theorem \ref{covkerCregular} beyond piecewise 5-differentiable activation functions.
\item Prove that the NLC is the first-order approximation of the kernel bandwidth for arbitrary Gaussian processes (section \ref{networkCovKerSection}).
\end{itemize}

\paragraph{Continuing investigations} We uncovered a number of intriguing threads throughout this work that we only followed to a limited degree. Below, we give opportunities for continuing investigations for which this work already gives some results which are sufficient to envision fully-fledged solutions.

\begin{itemize}
\item Continue the formalization of ZSAD guidelines. We do not definitively associate our non-NLC guidelines with concrete metrics as more analysis is needed to determine the best ones. For the same reasons we stress the importance of the well-definedness of the NLC (section \ref{wellDefinedSection}, \ref{nlcSummarySection}, \ref{relatedWorkSummarySection}), we believe it is valuable to frame other guidelines as metrics. We give suggestions for scale stability (section \ref{forwardStabilitySection}) and neuron bias (section \ref{outputBiasSection}) and give a range of candidates for Gaussian stability (section \ref{gaussianStabilityExplanationSection}).
\item Continue the investigation of pseudo-linearity as a ZSAD guideline. We discuss its relationship with scale stability, neuron bias and the nonlinearity of deep networks (section \ref{pseudoLinearitySection}). We believe it may be possible to further develop this guideline into a ``nonlinearity uniformity'' guideline, which would state that nonlinearity should be spread evenly throughout the architecture. For example, we show that the most popular type of residual architecture contains more nonlinearity in earlier than later residual units (section \ref{skipConnectionsSection}). We conducted some preliminary experiments not presented in this work which suggest this may be suboptimal.
\item Continue the investigation of orthogonality as a ZSAD guideline. We briefly discuss the complementary nature of NLC and orthogonality as guidelines (section \ref{orthogonalitySection}). Orthogonality, especially under the term of dynamical isometry, has been discussed alongside expressivity (e.g. \citet{meanFieldCNN,meanFieldRNN}). Its mean field theory has been developed \citep{meanFieldNetsorMatrix,meanFieldOrtho}. We believe that integrating our analysis of the NLC with that of orthogonality may strengthen both approaches.
\item Continue the investigation of how the NLC modulates noise stability with respect to other deep learning concepts like quantization, noise-based regularization and adversarial robustness. This could build on our investigation of batch selection noise and floating-point rounding error (section \ref{noiseStabilitySection}). Relating the NLC to other specific sources of noise could further widen its scope.
\item Continue developing methods for determining the best NLC level for a given dataset. We predict the best NLC by examining the apparent nonlinearity of the dataset via PNLCD (section \ref{bestNlcSection}). It may be worth validating this method on further datasets, determining how to apply it when PNLCD doesn't have a clear peak, or developing better methods altogether. 
\item Continue the investigation of the properties of neural regular data. We found that there was a relationship between (i) the apparent nonlinearity of a dataset measured via line segments between datapoints, (ii) input component independence and (iii) consistency of square means and co-means across inputs (section \ref{elemLikeSection}). We believe that further solidifying these relationships theoretically and empirically across a wide range of datasets, as well as adding new properties and relationships, could greatly enhance our understanding of the kind of data on which deep learning and mean field theory do well.
\item Continue the investigation of setting the starting learning rate. We found some intriguing patterns when measuring the shift of the parameter as well as the shift of the network output during the first iteration when using the best starting learning rate. We believe that it may be possible to derive methods that reliably find the best starting learning rate, at least for high-performing architectures, within a few guesses (section \ref{learningRateSection}). Our experiments suggest that line search, at least during the first iteration, might be a valuable strategy for neural network training.
\item Continue the investigation of the evolution of the NLC during training. Our experiments suggest that the architecture tries to attain a more optimal NLC during training when its initial NLC is not too far away (section \ref{nlcEvolutionSection}). However, within a certain range, the NLC reached after training appears unpredictable (section \ref{nlnormBestNLCSection}). Since the NLC ultimately matters in the final state, when the network must match the true input-label function, but ZSAD decisions are made in the initial state, we believe further bridging this gap is valuable.
\item Continue the comparison of NLC with the MGLLA and TTNTK metrics (sections \ref{nlcSensiSection}, \ref{nlcKernelSection}). Solidifying the understanding of the other two metrics should enhance the understanding of the NLC and expressivity in general. It may turn out that MGLLA or TTNTK are either superior to NLC for performance prediction or contain important complementary information. TTNTK is similar to the OSGR metric from \citet{nnntkMetric}, which was developed independently.
\item Continue the theoretical investigation of low-NLC networks. It may be difficult to make guarantees about high-NLC networks, as they can behave like low-NLC networks, except with an enormous gradient on an irrelevantly tiny subset of the input space. Conversely, we show that low-NLC networks are close to linear functions in an $L2$ sense on Gaussian inputs (section \ref{nlcLinearApproximationSection}). There may be opportunities to develop this further to provide theoretical error guarantees in practical situations.
\item Continue the investigation of the exploding / vanishing gradient problem in tanh and sigmoid networks. This problem is widely known and discussed. We explained how, in these networks, gradients can explode in expectation but vanish with high probability (section \ref{vanishTanhSection}). When fleshed out in the manner of our other guidelines, this phenomenon could yield a valuable complementary piece as well as shed light on the infamous ``cliffs'' of the recurrent network objective landscape.
\item Continue the investigation of the value of bias and elementwise multiplication layers. Initializing both to the identity prevents them from inducing scale instability or neuron bias (sections \ref{surveyBiasSection}, \ref{surveyScalingSection}). Intuitively, controlling expectation and variance of neurons allows the network to alter e.g. the NLC during training. However, as far as we know, unique benefits of those layer operations have thus far not been isolated.
\end{itemize}

\paragraph{Major investigations} In various places, our work hints at opportunities for other major investigations that would go well beyond the scope of this work and would likely relate to several other ongoing strands of investigation in the deep learning community. These are given below.

\begin{itemize}
\item Use the ZSAD guidelines of this work to build novel state-of-the-art architectures or improve existing ones. One of the challenges of ZSAD in this stage of development is that it is more likely to explain current designs than to suggest new ones. For example, applying nlnorm in study B only leads to significant test error reduction if the original architecture was suboptimal. It may be worth scanning the deep learning landscape for state-of-the-art architectures that do not follow guidelines to an ideal degree and thus present an opportunity for improvement. For example, it may be worth spreading out nonlinearity evenly in residual or other multi-path networks as mentioned above.
\item Integrate ZSAD with NAS. Currently, NAS algorithms are initialized by specifying ranges for each hyperparameter and architecture property over which to optimize. Guidelines can act as a prior for the ``architecture definition-to-error function'' or as a filter that prevents architectures suggested by NAS from being trained. This may reduce runtime and enable the search over wider hyperparameter ranges. \citet{nasRejection} recently employed a filter based on the activation function and normalization layers used.
\item Investigate the benefit of using (very) deep architectures when expressivity is controlled via e.g. NLC. We argue that NLC is a better measure of expressivity than depth (section \ref{depthSection}). However, the current main argument in favor of depth is that it enables expressivity. When depth is not needed to control expressivity, new sources of utility need to be found.
\item Investigate the benefit of skip connections and multi-path architectures in general when expressivity is controlled via e.g. NLC. While there has been work on explaining the value of skip connections, we are not aware of work that specifically factors out overall expressivity, which is not fundamentally related to skip connections. Our experiments in this work (section \ref{skipConnectionsSection}) and general experience suggest that there is indeed additional value. Also, in the context of multi-path architectures, it may be worth controlling the NLC via the addition weights associated with different paths in the architecture (e.g. skip connection strength) instead of nlnorm.
\item Investigate the infinite-depth, finite-width, finite-NLC limit. We suspect that, as the depth of e.g. plain architectures converges to infinity while activation function NLC co-varies to keep architecture NLC constant, the distribution over network functions and many key metrics induced by the parameter initialization scheme converges in a similar fashion to mean field theory. This investigation might yield another highly predictive theoretical framework that complements mean field theory and may illuminate behavior at moderate to high depth just like mean field theory illuminates the behavior at moderate to high width.
\item Investigate the frequency spectrum of networks and activation functions. We suspect that the NLC is equal to the square root of the mean of frequencies present in a network, where each frequency is weighted by its magnitude. For example, we verified that the NLC of the the component of $\tau(s) = s^n$ that is orthogonal to all lower powers on unit Gaussian input is equal to $\sqrt{n}$ for $n \le 10$. By considering the full frequency spectrum, we may be able to design custom activation functions based on the frequencies in a dataset or design activation functions that are universally high-performing across neural regular datasets.
\item Investigate the relationship of NLC and expressivity to sample complexity, training complexity and training time.
\item Formulate a theory of model complexity for neural architectures where expressivity, as measured by e.g. NLC, and capacity, as measured by e.g. parameter dimensionality, act as two near-independent dimensions as briefly discussed in section \ref{widthSection}.
\item Investigate the importance of Gaussian neuron distributions. We demonstrated that Gaussian stability is key for architecture performance (section \ref{gaussianStabilitySection}) and the predictiveness of mean field theory (sections \ref{meanFieldDistributionEmpiricalSection}, \ref{meanFieldPracticalEmpiricalSection}). However, the underlying mechanisms of these observations, especially the former, are largely unexplored. There may be significant utility in determining the scope of the set of Gaussian stable architectures and further determining and explaining the properties they have as a class. This may ultimately reveal a path to transcending this class of architectures.
\item Investigate scope, origin and importance of the neural network property that the gradient-based local linear approximation tends to be accurate across a large fraction of the codomain. Specifically, in section \ref{nlcSensiSection} we found evidence that neural networks behave like a sine curve, in that the tangent hyperplane stays close to the network function across the range of outputs (typically) returned by the network. We have not found this property discussed in literature. It seems, however, intuitively necessary for trainability, in particular the use of ``large'' learning rates, as well as several properties of the NLC.
\item Use the zero-shot architecture design approach, including the use of well-defined metrics (section \ref{relatedWorkSummarySection}) and careful protocols (section \ref{metricsSummarySection}), to frame the investigation of aspects of neural architecture performance other than training and test error, such as computational efficiency, privacy or adversarial robustness.
\end{itemize}

\section{Problem setup: an overview of neural architecture design} \label{designOverviewSection}

\subsection{From biological imitation to probabilistic inference to function optimization} \label{historySection}

To understand contemporary neural architectures, we must understand the history of neural networks.

Models called `neural networks' have existed for at least 75 years \citep{neuralNetOrig}, which far predates the field of machine learning. Their original purpose was to be an abstract representation of neurons in the brain, which form networks through synaptic connections, and to explain how those neurons learn. Neurons were commonly modeled as computational units that first linearly aggregate the outputs of other neurons they are connected to and then apply a binary threshold gate to this aggregated value \citep{sigmoid3,binary}. The reason for this binary thresholding was that neurons in the brain were regarded to have a binary state - either ``firing'' or ``not firing''. The thresholding function became known as the ``activation function'', as it was meant to determine which incoming signals lead to the neuron emitting an electrical impulse.

For at least 55 years, neural networks have been trained with gradient methods \citep{backpropagation}. This turned out to be a far more efficient way of setting the free parameters in a neural network than learning algorithms based on imitating brain function \citep{hebbianLearning}. Of course, gradient methods require that the computation performed by a neuron be differentiable, which is not the case for the binary activation function. The sigmoid activation function (table \ref{actFunIllu}) became the new standard as it was considered the closest differentiable approximation of the binary function \citep{sigmoid1,sigmoid2,sigmoid3,sigmoid4}. Using gradient methods, neural networks became a practical tool for tasks such as digit recognition \citep{cnnOrig}. What made neural networks attractive to machine learning was their ability to compose a large number of generic computational units which can together learn complex relationships in data while reducing the need for model or data engineering.

In the early 2000's, neural networks had fallen somewhat out of favor in the machine learning community. While there is not a clear single reason for this development, it is said that there was limited success in building ``deep'' models when utilizing the sigmoid activation function. We will revisit the difficulty of utilizing the sigmoid in chapter \ref{surveyChapter}. During this period, the dominant machine learning model for learning complex relationships in data was the probabilistic graphical model (PGM) \citep{pgm1,pgm2,pgm3}. The central idea of PGMs is to model the data as a set of random variables called `observed variables'. Each datapoint in the dataset is considered an independent sample of these variables. Then, we introduce additional random variables, called `latent variables', which are not observed as part of the data. This allows us to define intricate, hand-crafted joint or conditional distributions over both sets of variables. Finally, we derive properties of the data via probabilistic inference.

The transition to the deep learning era began in 2006 with the introduction of the Deep Belief Net (DBN) \citep{DBN2006details}. The DBN is a hybrid between PGMs and neural networks as it (i) defines a probability distribution over observed and latent variables and utilizes probabilistic inference, while also (ii) using a large number of latent variables with a generic distribution. The difference between `neuron' and `latent variable' is erased. Finally, (iii) DBNs stack layers of latent variables in a ``deep'' fashion. Those layers were trained using a complex and intricate two-stage process involving the contrastive divergence algorithm \citep{ContrastiveDivergenceLearning,CDlong}. Around 2010, these hybrid models were still the driving force in the emerging deep learning field \citep{pfm1,pfm2}.

The deep learning era arrived in 2012 when \citet{ImageNetAlexNet} trained a 7-layer convolutional network on the ImageNet dataset, eviscerating all previous benchmarks. What was remarkable at the time was that an architecture of significant depth was trained with a ``pure'' gradient method from a random initial parameter value using only empirical risk minimization. It eschewed both contrastive divergence and its non-probabilistic equivalent, the autoencoder \citep{GoogleBrain,ContractiveAutoEncoder,denoisingAutoencoders,autoencoder1}. Both of these predict not just the class label, but the input distribution itself.

\subsection{The functional-gradient paradigm} \label{functionalGradientSection}

The formalism used by \citet{ImageNetAlexNet} for classification can be paraphrased as follows. (For simplicity, we assume datasets contain only one datapoint in this section.)

$$\theta^{(t)} = \theta^{(t-1)} - \alpha \frac{d\ell(f(\theta^{(t-1)}, x),y)}{d\theta}$$

Here, $f$ is the neural architecture, $x$ is the input, $\theta$ is the trainable parameter, $\ell$ is the loss function, $\alpha$ is the learning rate and $(t)$ is the iteration counter. What is striking about this formalism is that the neural network is reduced to a function that could, in principle, be almost anything. The only requirements are that the function takes in and returns values of the correct type, that it has a trainable parameter, and that it is differentiable, though in practice it is enough that we can efficiently compute a local linear approximation to the function that is sufficiently accurate in a sufficiently large neighborhood of the current parameter value, as explained in sections \ref{howDifferentiableSection} and \ref{nonDifferentiableSection}. Compared to the complexity that can be rolled up in the function $f$, the formalism itself is very simple. 

It was this recipe of combining a simple formalism with black-box functions that became dominant and led deep learning from success to success \citep{GAN,transformer,graphNetOrig,alphaZeroNature}. Consider, for example, generative adversarial networks \citep{GAN}.

$$\min_{\theta_G} \max_{\theta_D}\log f_D(\theta_D,x) +  \mathbb{E}_{z} \log (1 - f_D(\theta_D,f_G(\theta_G,z)))$$

Again, the majority of the complexity is rolled up in the arbitrary functions $f_D$ and $f_G$. Updates to their trainable parameters are made via the gradient of the objective.

Finally, consider REINFORCE, a staple building block in deep reinforcement learning \citep{reinforceOrig}.

$$\theta^{(t)} = \theta^{(t-1)} +\alpha v_t\frac{d\log f(\theta^{(t-1)}, a_t|s_t)}{d\theta}$$

In this algorithm, it is not immediately apparent what objective is being optimized. Ignoring $\alpha$, the second term can be written as $\frac{v_t}{f}\frac{df}{d\theta}$. This is a product between the gradient of $f$ and a term that can be interpreted as the change in the output of $f$ that the update is desired to induce. We can also rewrite the classification and GAN updates in this way, where the desired change to the output of $f$ is the gradient of the objective with respect to $f$, which, in the case of simple classification, is the gradient of the loss function.

\begin{figure}
\makebox[\textwidth][c]{
   \fbox{\begin{minipage}{0.8\textwidth}
\centerline{\bf The functional-gradient learning paradigm}  
\bigskip
\begin{itemize}
\item {\it Black-box functions}: Complex relationships in data are modeled by functions $f$. While some functions may lead to far superior learning outcomes than other functions, the learning formalism itself requires of a function only (i) that its input(s) and output(s) have certain data types, (ii) that it has a free parameter $\theta$ and (iii) that a sufficiently accurate local linear approximation in a sufficiently large neighborhood of a given parameter value can be found.
\item {\it Gradient updates}: $\theta$ is updated based on the product of the desired change to the output of $f$ and the gradient of the local linear approximation of $f$.
\item {\it Simple formalisms}: The way $f$ is used within the learning formalism is simplified as much as possible. The complexity of model and algorithm is pushed into the function. The training algorithm is not much more complicated than gradient descent \citep{nesterovOrig,Adam}. If an objective containing $f$ is used, it is largely a vehicle for specifying a loss and not for encoding structure or intricate priors.
\end{itemize}

    \end{minipage}}}
\caption{Description of the functional-gradient learning paradigm.} \label{boxFunctionalLearning}
\end{figure}

We are now ready to formalize what we call the `functional-gradient learning paradigm' in figure \ref{boxFunctionalLearning}. It is a close approximation of how the term `deep learning' is used as of 2020. The term `neural architecture' then approximately corresponds to the phrase ``function $f$ as it may occur in a functional-gradient learning system'' and `neural network' approximately corresponds to the phrase ``function-parameter pair $(f,\theta)$ as it may occur in a functional-gradient learning system''. In essence, a neural network is any model to which gradient methods can be applied, and that is how we define it in section \ref{neuralNetworksSection}. Of course, deep learning comes with a very large amount of associated concepts and popular conventions, the most important of which we detail throughout chapter \ref{backgroundChapter}. However, none of them are essential from the point of view of the formalism. Also, nearly all core concepts in the deep learning field are fuzzy, subjective and not without exception. 

Note that while we focus this work on feedforward networks, which have a single input and output, multi-input and multi-output functions, as represented by e.g. recurrent networks, are also used within the functional-gradient paradigm.

Interestingly, the functional-gradient paradigm can even be used for probabilistic modeling. In the GAN objective above, $z$ is a random variable with a simple (generally Gaussian) distribution which is transformed into a distribution over (fake) inputs. The complexity of the conditional distribution $z|x$ is fully contained in $f_G$. The same strategy is applied in e.g. variational autoencoders \citep{VAE}. The convergence of a significant fraction of machine learning around the functional-gradient paradigm is epitomized by the development of functional learning software frameworks like TensorFlow and PyTorch. They are so popular that the choices made in their development have effectively become soft constraints on the trajectory of machine learning research.

\subsection{The blessing and curse of neural architecture design}

While gradient methods are immensely powerful, they are not quite powerful enough to make choosing an architecture $f$ obsolete, i.e. no architecture is known that performs near-optimally on all tasks. Hence, for any given task, we are still left with choosing an architecture $f$, as we also explain in section \ref{modelSelectionSection}. We use the term `neural architecture design' in a broad sense to encompass any process that contributes to the choosing of a neural architecture for a task. We use `architecture design strategy' to refer to any piece of information that contributes to the choosing of an architecture. The actual choice of architecture is specified via what we refer to as the `architecture definition', which is simply all the information required to uniquely specify an architecture. It is the information given to e.g. a functional learning framework like TensorFlow to instantiate the architecture in memory.

Fortunately, the functional-gradient paradigm has two massive advantages that are unprecedented in the field of machine learning. First, because the learning formalism admits arbitrary functions, there is an almost limitless capacity for research and innovation in the field of neural architecture design. Second, because functions can be used, to a large degree, for an arbitrary task, any novel strategy may be applicable across a large fraction of the machine learning spectrum. These advantages were the primary motivators behind us conducting this work. 

Unfortunately, the blessing of flexibility is also a curse. While we have the power to choose an arbitrary differentiable, parametrized function, neural architecture design is also as complex as searching the space of those functions. And, even if we are content to build an architecture from only popular design strategies and building blocks, there are still a large number of choices. How many layers should there be? How wide should layers be? What activation functions and normalization layers should be used? Even beyond the architecture, what training algorithm and learning rate should be used?

In the next three subsections, we discuss what we argue are the main drivers behind neural architecture design in the era of the functional-gradient paradigm, followed by a discussion of automated design, which has become popular recently.

\subsection{Historical bias in neural architecture design} \label{historicalBiasSection}

Many of the architectures we use today closely resemble architectures that were developed decades ago, such as CNN \citep{cnnOrig} and LSTM \citep{LSTM}. At that time, the functional-gradient paradigm did not exist in the way it does today and the objectives of architecture design were somewhat different. Consider a simple architecture with alternating linear layers and activation layers. From a historical perspective, the linear layer originated from the additive aggregation of incoming electrical potentials in biological neurons, as discussed above. The activation layer originated from the response of those biological neurons to incoming potentials. The connectivity structure resembles cascades of biological neurons. These architectures also resemble popular PGMs, such as the graphical Lasso \citep{gLasso} or restricted Boltzmann machine \citep{ContrastiveDivergenceLearning}. 

Historical influences are the first major driver in neural architecture design. In fact, the strength of these historical influences suggests that popular conventions may be keeping architecture design stuck in a ``local optimum'' to some degree. While proving or disproving this idea goes beyond the scope of this work, it is worth noting that much of the recent progress in architecture design has come by addressing new data types, such as graph data \citep{graphNetOrig,graph1,graph2}, set data \citep{setNetOrig,set1,set2}, sphere data \citep{sphereOrig,sphere} or mesh data \citep{mesh1,mesh2}, or has been domain-specific. Out of the thousands of architecture design strategies and building blocks that have been proposed in the last decade, only a handful have attained widespread use across domains and data types, including batch normalization \citep{batchNormalization}, skip connections \citep{resNet} and attention \citep{attention}. Yet even the usefulness of those strategies has not been fully explained. 

\subsection{Programs = Functions, and the importance of computational efficiency} \label{programFunctionSection}

Machine learning is a computational discipline. The practical goal is to create a pipeline of programs that transforms raw data into useful information. Each program in this pipeline consumes some kind of input and produces some kind of output. Hence, these programs can be abstracted as mathematical functions. All we need to do is replace floating-point data with real values. The output of the program, up to rounding error, is then the same as the output of the function. Because of this duality, oftentimes machine learning models and training algorithms are developed as mathematical constructs first and then translated into programs. This can have a major drawback. The space of functions we can reason about mathematically is very different from the space of programs that can be tractably executed on a machine. PGMs are a prime example of this. Because they originate from probability theory, the mathematical formalisms involved in PGMs often involve, for example, expectation operators over intractable continuous distributions. In the heyday of PGMs, creating programs that sufficiently approximate those operators was a massive research effort \linebreak \citep{inference1,inference2}. Another example is a machine learning field called spectral algorithms, which involves eigenspectrum operators applied to potentially enormous matrices. In the functional-gradient paradigm, the tension between the mathematical and computational is largely resolved by construction. Because formalisms in this paradigm can use arbitrary functions, we have the ability to choose functions that trivially correspond to programs that can be efficiently executed on a machine. This ability is a core reason for the success of the paradigm.

Computational efficiency is the second main driver in neural architecture design. Linear operations, such as matrix multiplication and convolution as well as elementwise operations, can be efficiently implemented using specialized hardware such as GPUs \citep{DNNGPUOrig,gpu1}. Highly optimized implementations exist for those operations \citep{directConvolution}. Composing simple operations, such as addition, multiplication, maximum / minimum or exponentiation, enables efficient gradient computation with a computational complexity that is generally less than thrice the computational complexity of the network evaluation itself through a process called automatic differentiation \citep{autoDiffLong,autoDiff2}. Thus, these compositional architectures make full use of the extraordinary power of gradient methods. At a cost of less than three evaluations, we can approximate the value of $f$ in an entire region around the current parameter value with the local linear approximation defined by the gradient.

\subsection{``Designing neural networks is a dark art.''} \label{darkArtSection}

The statement in the heading of this subsection was a popular sentiment that was echoed in the machine learning community until not many years ago. It was motivated by the large and seemingly unpredictable variations in architecture performance based on the design choices that were made. Hence, these choices were made based on experience by deep learning practitioners.

There have been a few general design guidelines: avoid vanishing / exploding gradients \citep{RNNvanishingGradient,heInit,normalizedInitialization,RecurrentNetsPascanu,depthScalesMeanField}, use an appropriate depth / width \citep{depth10,depth9,width1,width4}, and, recently, choose a network ``on the edge of chaos'' \citep{correlationLimit,depthScalesMeanField,meanFieldCNN, meanFieldRNN,meanFieldBN}. The gist behind these guidelines is often that an architecture should have an appropriate model complexity so that it neither underfits nor overfits nor becomes untrainable. For example, width is associated with the dimensionality of the parameter, which is a traditional measure of model complexity. These guidelines have remained somewhat vague and circumstantial because (i) there are no well-defined, agreed-upon metrics that measure concepts like exploding gradients and overall width in practical situations; and (ii) while prior research presented evidence that these guidelines lead to success in certain situations, their generality is unclear. For example, empirical validation of research in neural network analysis often occurs on only a single activation function (e.g. \citet{normalizedInitialization,gradientCorrelation,meanFieldCNN,depthScalesMeanField, resNetMeanField,reluFiniteWidthEffects,meanFieldKernelPerformance1}). Informal concepts like covariate shift \citep{batchNormalization} or signal propagation \citep{correlationLimit} often remain undefined.

``Trial and error'' is the third main driver in neural architecture design. The success of deep learning can be significantly attributed to the growth of the number of neural architectures that have been found to perform well through trial and error and that exist in the public domain. The prevailing wisdom among practitioners is to fit a novel task to an existing architecture as much as possible.

\subsection{The strengths and limitations of neural architecture search (NAS)} \label{nasSection}

A major recent development in neural architecture design is the emergence of `neural architecture search' (NAS). The basic and original formalism for NAS is as follows. (i) A number of architectures are trained on a training set and their error is evaluated on a validation set. (ii) A meta-model, trained via e.g. Bayesian optimization or reinforcement learning, attempts to predict the validation error of an architecture from its definition. (iii) The meta-model is used to suggest new architectures to train, taking into account both exploration and exploitation. Finally, steps (ii) and (iii) are repeated, ever-expanding the pool of trained architectures \citep{spearmintHyperBayesian,DNNintoGaussianHyper,bnnHyper,reinforceHyper1, reinforceHyper2}. Of course, this formalism is very expensive. Generating a single datapoint for the meta-model involves training an entire architecture. Recent efforts have increased the efficiency of NAS in various ways: parallelizing architecture training \citep{parallelHyperopt,BObatchSelection}; sharing parameter components between different architectures \citep{mixtureOfConvLayers,enas,darts,nas2,nas3,nas9}; terminating training runs that do not appear promising early \citep{hyperband}; taking into account performance on similar datasets \citep{autosklearn}; predicting the learning curve \citep{learningCurvePrediction}; learning an embedding for the architecture definition \citep{nas7,nas1,nas6}; training the meta-model with labels of varying fidelity \citep{nas4}; using the meta-model itself to generate ``pseudo-data'' \citep{nas5}; using architectures trained in parallel to supervise each other \citep{nas8} - just to name a few.

NAS has had a significant positive impact on the performance of practical machine learning systems. There has been a migration away from architectures that were entirely manually designed (i.e. without computation) or found through random search or grid search \citep{randomSearchHyper}. The success of NAS is exemplified by the fact that the terms `neural architecture design' and `neural architecture search' have recently become somewhat synonymous in the community. However, conceptually, NAS is neither built for nor capable of exploring the space of all differentiable functions. NAS conducts a search for a local optimum in architecture space in a small neighborhood around the manually designed architectures that were used before the advent of NAS. The need for manual design is therefore not greatly alleviated by NAS. (This does not even take into account the need to manually design a NAS algorithm.) A strength of NAS is that it is built on top of the functional-gradient paradigm and can handle arbitrary network functions. The downside of treating networks as black boxes is that NAS yields few qualitative insights that feed back into the manual design process. The popularity of black-box approaches also underscores the lack of practical guidelines that could be used to inform the search. Finally, we note that there is an inherent tension between the efficiency and flexibility of NAS. Infusing the search with gradient information \citep{ADhyperGBrestart,optConditionHyperDanish,dLdWinitial,darts} increases its power but also restricts it to well-behaved, continuous, finite-dimensional search spaces, as opposed to the heterogeneous spaces that can be explored by basic NAS \citep{randomForestHyper}. Sharing parameter components between different architectures requires them to be sufficiently similar so that such sharing is meaningful \citep{nas2}.

\subsection{Summary}

Neural architecture design harbors great possibility but also great challenge. The success of neural networks, which is based on gradient methods, computational simplicity, design flexibility and cross-task generalization, stands in contrast to the lack of explanatory, well-defined design guidelines as well as a general lack of understanding of {\it why} certain architectures perform the way they do. Hence, deep learning practitioners often fall back on anecdotal experience, convenience and black-box search when choosing architectures.

The more abstract neural networks became, the more success they had. Therefore, we believe that by reducing the focus on the historical view of neural networks as cascades of computational units and increasing the focus on the purely functional view, we can build towards a more powerful theory of architecture design. We believe this work is a small step in that direction. Our flagship design guideline, the NLC, is based entirely on the network function and is independent of the layer representation. As a consequence, it can be applied to arbitrary feedforward networks. At the same time, it is a meaningful architecture property that explains its generalization behavior. While much of our work is still rooted in the concepts of layers, weight matrices and activation functions, throughout this work we empirically investigate a wider range of architectures than the vast majority of prior work. See chapter \ref{empiricalStudiesChapter}. We put a similar emphasis on generality in our theoretical work. Hence, we believe that a large fraction of our results convey valuable lessons about the general class of feedforward networks and beyond, rather than only specific designs.

\section{Our approach: zero-shot architecture design (ZSAD)} \label{zeroShotDesignSection}

Because the search space of possible architectures is enormous, we cannot train them all. We cannot even evaluate them on a single input for a single parameter value, or even specify them in code. All we can hope to do is to {\it think} about a modest number of architectures, and to use our understanding of deep learning to select an even smaller set for further analysis. Hence, inevitably, the manual stage of architecture design is a critical stage. The goal of this work is to develop and advocate an approach to manual design that is as sound, principled and well-reasoned as possible. Similarly, we advocate such an approach to the process of architecture design research itself.

\begin{figure}
\makebox[\textwidth][c]{
   \fbox{\begin{minipage}{0.8\textwidth}
\centerline{\bf Zero-shot architecture design}
\bigskip

\begin{definition}
`Zero-shot architecture design' (ZSAD) is any process that contributes to the choosing of a neural architecture for a given task based on general, predictive, explanatory and ideally well-defined principles that go beyond the imitation of specific designs or the use of specific building blocks that have exhibited high performance in the past; and that does so without training architectures, without utilizing the properties of specific previously trained networks and without heavy computation that mimics the effect of training. We term such a general, predictive, explanatory and ideally well-defined principle a `ZSAD guideline'.
\end{definition}

\end{minipage}}}
\caption{Definition of zero-shot architecture design and ZSAD guideline.} \label{boxZSAD}
\end{figure}

Once something has a name, it takes on a different dynamic. So we give our approach a name: `zero-shot architecture design' (ZSAD). We define it in detail in figure \ref{boxZSAD} above.

Of course, this definition is not meant to imply that we should choose a final architecture without NAS for any given task. However, even if we do not manually choose an architecture, it is critical that we choose the right NAS algorithm, and that we choose the right architecture space to search with NAS. Before we write even a single line of code towards any deep learning system, we must make sweeping decisions about what kinds of architectures or training protocols to consider. ZSAD represents a framework for making these decisions not just based on ``what has worked before'', but also based on ``general, predictive, explanatory and ideally well-defined principles''. Of course, the boundary between ZSAD, NAS and other types of architecture design is fuzzy and subjective.

The ideal outcome of ZSAD can be viewed as being able to predict the performance of an architecture after training, based only on the definition of the architecture and task, with minimal computation or side information, in a way that yields qualitative insights that improve the efficacy of the manual design stage. Of course, this outcome is more of an ideal than a practical milestone, and so we must be content with attempting to come as close to it as possible, and to put up with a certain amount of subjectivity in our assessment of how close we have actually come.

When we say that ZSAD aims to predict `performance', we include any desiderata that may exist for the machine learning task in question under the term performance. For example, we include computational efficiency, privacy or adversarial robustness under this umbrella. Reasoning about all those properties is valuable. In this work, we focus on supervised classification. The core aspect of performance in this setting is generalization, which is usually measured via test error in a research context. In this work, we focus on predicting test error after training, and sometimes on predicting training error after training. Going forward, we will generally use the term performance in this more concrete and limited sense, and correspondingly use ZSAD to refer to principled architecture design for the purpose of generalization and trainability. Correspondingly, we also restrict our discussion of existing guidelines to those objectives.

\subsection{What is architecture performance?} \label{architecturePerformanceSection}

\subsubsection{Architecture performance modulo training algorithm} \label{moduloAlgorithmSection}

We can already see a major problem with the above plan. A neural architecture does not have an inherent performance after training. Because training a machine learning model is an intricate process, almost all training runs lead to a network that has not learned anything. At the other extreme, among the set of all possible training algorithms that could be dreamt up, there is probably one that would find the global optimum in parameter space with a single iteration by sheer chance, and we know that these global optima tend to perform much better than networks commonly found with gradient methods \citep{distillationOriginal,distillation}.

The only practical way to measure post-training performance of an architecture that can be seen as ``inherent'' to the architecture is the following three-step process. (i) Choose a set of training algorithms that are representative of the state of the art in neural network training. (ii) Conduct an exhaustive search over key hyperparameters of those algorithms via independent training runs, obtaining validation error values in the process. (iii) Evaluate the test error after the training run that attained the least validation error. (Here we assume that performance corresponds to test error.) Luckily, the functional-gradient paradigm uses training algorithms that are either SGD or a simple variant of SGD, and architectures tend to attain somewhat similar error levels across these algorithms (see also section \ref{nlcPredictiveSection}). The only critical hyperparameter that these algorithms have which has no agreed-upon value is the learning rate. However, the importance of performing an exhaustive search over the learning rate cannot be overstated. The lack of independent, exhaustive tuning of hyperparameters like learning rate has been a significant hindrance to making robust comparisons between different architectures and studies in the deep learning community. While it is impossible to quantify the extent of this problem exactly, via correspondence with authors and via replication, we know of two well-known analytical deep learning papers whose results have been significantly impacted by this problem. We note that there has been a recent push towards comparability in the NAS community \citep{nasCompare1,nasCompare2,nasCompare3,nasCompare4}.

In this work, we ensure exhaustive learning rate tuning for all experiments. We investigate the importance of learning rate in section \ref{learningRateSection} and discuss it in section \ref{experimentsSummarySection}. Throughout this work, when we reference ``architecture performance'', we usually speak of the test or training error attained when training with SGD or momentum using the learning rate that yielded the best validation, test or training error depending on context.

It is important to note that even conducting an exhaustive search over learning rate or other hyperparameters does not allow us to estimate the actual global error minimum across a continuous hyperparameter space. This is due to the sharp valley problem that we discuss in section \ref{sharpValleySection}. In the case of learning rate, we search over a geometric sequence of values with spacing factor between 3 and $\frac{10}{3}$. This kind of spacing factor has generally been found to yield a robust and representative estimate of the error minimum with high probability.

\subsubsection{Architecture performance modulo data} \label{moduloDataSection}

In the same vein as the training algorithm, the data is a major driver behind performance. Because signal is rare in the real world, any architecture trained with any algorithm is likely to not generalize at all. Even among datasets that are well-understood, we might reasonably expect very different performance levels from one dataset to another. When distinguishing cats from planes, we might desire near-perfect performance. When predicting stock price movements using only information that is publicly available to market participants, a 49\% test classification error might be outstanding. Finally, no machine learning model performs equally well on all tasks.

In this work, we aim to develop principles for ZSAD that apply across datasets and task domains while being themselves as data-independent as possible in their formulation. At face value, this appears to be a contradiction. Without reasoning about the data, how can a guideline account for the influence of that data on performance? To reconcile this, we have to adopt a ``relative'' view of performance for the purpose of ZSAD. For example, we can assess performance relative to the performance attained by the best architecture for a task that is known, relative to the performance of random guessing, or we can consider the performance ranking of all architectures trained on a task. Using this approach, almost all results we obtain in this work are highly consistent across three different datasets. In section \ref{bestNlcSection} and throughout chapter \ref{meanFieldNnaChapter}, we perform an analysis that provides an explanation for why it is possible to find strong patterns in architecture performance that span across datasets. In fact, we identify several conditions that define a class of datasets on which consistent patterns of architecture performance can be expected in section \ref{elemLikeSection}. Throughout this work, when considering architecture performance without reference to a specific dataset, we mean the relative performance attained on the three datasets we studied in particular and the relative performance attained on the class of datasets described in section \ref{elemLikeSection} in general.

Of course, we do not claim that data-dependent ZSAD is of lesser importance than data-independent ZSAD. For example, explicitly infusing a model with information about known invariances present in a given dataset is clearly one of the most important requirements for attaining success with deep learning. We argue that those questions are somewhat orthogonal to the questions we study in this work, which build on statistical properties of datasets that are common across domains and tasks.

\subsubsection{Architecture performance modulo initial parameter value} \label{moduloParameterSection}

Another important driver of performance after training is the value taken by the parameter before training. This initial value is a gradient method hyperparameter and not specified by the training algorithm. In practice, the initial parameter value is drawn from a distribution known as the parameter initialization scheme. A key property of popular initialization schemes, as we also discover throughout this work, is that they cause the network to behave in almost exactly the same way no matter the specific initial parameter value drawn from them. Hence, when considering performance, we can largely abstract away that specific value and focus on the initialization scheme only.

We then perform what is essentially a verbal sleight of hand. We simply consider the initialization scheme as part of the architecture definition. When we discuss the performance of an architecture, we already have a specific initialization scheme in mind. In practice, the initialization scheme is generally specified together with other architecture properties like depth or activation function. Hence, thinking of the initialization scheme as just another architecture property makes practical sense.

\subsection{ZSAD guidelines} \label{zsadGuidelineSection}

\subsubsection{Framing ZSAD guidelines: encapsulated properties of the randomly initialized state} \label{zsadGuidelineFramingSection}

\begin{figure}
\makebox[\textwidth][c]{
   \fbox{\begin{minipage}{0.8\textwidth}
\centerline{\bf ZSAD guideline: framing and utility criteria}
\bigskip

We frame a ZSAD guideline as postulating that a certain property of an architecture or an architecture's randomly initialized state is related to its performance after training. We assess the guideline's utility for ZSAD based on the degree to which that property fulfills the criteria below.

\begin{enumerate}
\item Well-definedness: The property is a concrete metric which yields a specific value, such as a real scalar or truth value, given a network or architecture. \label{criterionWellDefined}
\item Computability: The property can be determined in a way that is easy and cheap to compute, and easy to implement in code. \label{criterionComputable}
\item Predictiveness: The property is predictive of architecture performance after training, where architecture performance is conceptualized as in section \ref{architecturePerformanceSection}. \label{criterionPredictive}
\item Predictability: The property can be simply and instructively determined or estimated from the architecture definition. \label{criterionPredictable}
\item Controllability: It is possible to minimally modify a given architecture to change the property without changing other important properties. This in turn changes performance, which establishes the property as causal for performance. \label{criterionControllable}
\item Simplicity: The property is conceptually simple and easy to understand. \label{criterionSimple}
\item Meaningfulness: The property is deep and meaningful. It does not just predict performance, but also explains it. \label{criterionMeaningful}
\item Theoretical grounding: The property is closely related to important theoretical frameworks of deep learning in a way that enhances the understanding of both the property and that theoretical framework, and facilitates further analysis. \label{criterionTheoreticalGrounding}
\item Synergy: The aspects of architecture performance captured by the property are independent of other guidelines. It is possible to consider guidelines jointly to predict performance even more accurately. \label{criterionSynergy}
\item Generality: The above criteria hold across a wide range of architectures and tasks. Further, the property as well as the validity of the above criteria cannot be easily changed by manipulating irrelevant or superficial aspects of the architecture or wider learning pipeline.\label{criterionGeneral}
\end{enumerate}

\end{minipage}}}
\caption{Our framing of ZSAD guidelines and our criteria for assessing the utility of a guideline.} \label{boxNPM}
\end{figure}

We defined ZSAD to be based on ``general, predictive, explanatory and ideally well-defined principles''. We refer to such a principle as a `ZSAD guideline', as we also define in figure \ref{boxZSAD}. In this work specifically, and often in architecture design research in general, ZSAD guidelines can be boiled down to an architecture or network property that is (said to be) influencing performance. Consider guidelines referenced in section \ref{darkArtSection}, e.g. ``avoid vanishing / exploding gradients'', ``choose an architecture on the edge of chaos'' or ``use an appropriate depth''. Here, ``vanishing / exploding gradients'', ``edge of chaos'' and ``depth'' refer to more or less well-defined properties of an architecture or network. Further examples include ``minimize dying ReLUs'', ``avoid local minima in the loss surface'', ``ensure feature diversity'' and ``maximize classification margin''. The properties are ``dying ReLUs'', ``local minima in the loss surface'', ``feature diversity'' and ``classification margin''. The words ``minimize'', ``avoid'', ``ensure'' and ``maximize''  indicate that the former two properties are negative / harmful / that their value or degree of presence should be minimized / that their presence should be prevented, and that the latter two properties are positive / helpful / that their value or degree of presence should be maximized / that their presence should be induced. When we merely want to express that the value of a property is important, we say e.g. ``use an appropriate width''.

We base our ZSAD guidelines on properties throughout this work, and we refer to the guideline and the encapsulated property interchangeably. For example, ``the guideline of `avoid (the property of) neuron bias' '' means the same as ``the neuron bias guideline''.

Because ZSAD does not involve training, the property encapsulated by a guideline must not be a property of a trained network. For example, ``avoid exploding gradients {\it after} training'' is not a ZSAD guideline, but ``avoid exploding gradients {\it before} training'' is. ``Avoid dying ReLUs during training'' is not a ZSAD guideline, but ``avoid architectures that are prone to having their ReLUs die during training'' might be, depending on whether it can be measured in a meaningful way before training. In this work, we consider properties for ZSAD that are either defined directly in terms of the architecture definition, or in terms of the architecture's initial state. In the latter case, we relate the property back to the architecture itself via relative invariance to the random parameter draw as explained in section \ref{moduloParameterSection}. We further discuss pre- vs post-training properties below in section \ref{generalizationMeasuresSection}.

\subsubsection{Data-independence of guidelines} \label{guidelineDataSection}

In section \ref{moduloDataSection}, we outlined how it is possible to conceptualize architecture performance in a data-independent way. We also stated that we aim to develop ZSAD guidelines that are as independent of data as possible. Concretely, this means that the property encapsulated by the guideline should ideally depend only on the architecture definition or the initial state. In general, we do not reach this level of independence. In fact, many of our encapsulated properties are defined explicitly in terms of an input distribution in addition to a network, including the NLC (e.g. section \ref{nlcDefinitionSection}, \ref{forwardStabilitySection}, \ref{outputBiasSection}). Other properties, while not explicitly referencing an input distribution, cannot be evaluated without one (e.g. section \ref{gaussianStabilitySection}, \ref{noiseStabilitySection}). It turns out that this ``problem'' is not severe in practice, because our encapsulated properties are largely invariant to structure present in practical input distributions, being sensitive only to their expectation and covariance, as well as being invariant to the sample used for statistical estimation, at least in the initial state. We show this empirically for the NLC in section \ref{nlcRobustDataSection} and theoretically for the NLC and many other guidelines via mean field theory in sections \ref{meanFieldPracticalSection} and \ref{nlcExplainSection}.

\subsubsection{The utility of a ZSAD guideline}

The ZSAD concept is somewhat vague. This vagueness is especially suboptimal for a concept that is introduced specifically to combat the vagueness of architecture design research. To mitigate this issue as much as possible, we formulated a list of 10 `utility criteria' for ZSAD guidelines, and more specifically for the properties encapsulated by guidelines. They are given in figure \ref{boxNPM}. Of course, while these criteria add detail, they are themselves somewhat fuzzy and subjective, and they can hold to varying degrees. We use the criteria not just to validate ZSAD guidelines, but to determine whether any given architecture design strategy meets the threshold of being a ZSAD guideline. Again, unfortunately, this is a subjective judgment.

Our flagship guideline, `ensure $1 \le NLC \le 5$', based on the nonlinearity coefficient, fulfills all criteria to a significant degree. Hence, we argue that it is state-of-the-art among ZSAD guidelines. We summarize our results for the NLC in terms of the utility criteria in section \ref{nlcSummarySection}.

\subsubsection{Determining the value of network properties - towards well-definedness} \label{wellDefinedSection}

To be able to apply a ZSAD guideline, we must determine the property that the guideline encapsulates. We must give it a `value'.  For example, consider the property `parameter dimensionality'. If the parameter is indeed a vector, its dimensionality can be unambiguously determined. If we believed, say, that an architecture with parameter dimensionality over 1 million exhibits high performance, we can in fact determine the value of the parameter dimensionality of an architecture and then apply the guideline. In contrast, say we believed that a ``chaotic'' architecture exhibits high performance. If we cannot determine whether a given architecture is chaotic, chaos is not a helpful property. The ability to determine the property is implicit in our discussion of ZSAD guidelines (see e.g. utility criteria \ref{criterionComputable}, \ref{criterionPredictive}, \ref{criterionPredictable} and \ref{criterionControllable}). When we say e.g. that a ``property can be changed'', we mean that its value can be changed. That value can be e.g. binary when indicating whether the property is present (e.g. ``uses ReLU''), real-valued (e.g. Lipschitz constant) or integer-valued (e.g. parameter dimensionality).

Ideally, a property already comes with an inherent recipe for determining its value given an architecture. Again, parameter dimensionality, Lipschitz constant and ``uses ReLU'' have unambiguous and unique values for typical architectures. We call such properties `well-defined', which corresponds to utility criterion \ref{criterionWellDefined}. In general, a well-defined property can be viewed as a function of an object like an architecture or dataset that assigns the object a value. We call this function a `metric'.

Most existing ZSAD guidelines are not well-defined. While properties like ``exploding gradients'' have a quantitative ring to them, there is no agreed-upon way to determine whether a network has exploding gradients. Other properties like depth are thought of as specific values in certain contexts, like simple feedforward networks with alternating linear and activation layers, but do not have a meaningful general definition as we explain in section \ref{neuronDepthSection}.

Of course, when a ZSAD guideline is applied, some value(s) must be chosen for it. In many research studies, these values are made up on the fly by authors. The impact that a specific recipe for determining the property has on the results of a scientific study is often not discussed. Worse, the recipe is often not even mentioned. This leads to massive problems, as we discuss in chapter \ref{relatedWorkChapter} and summarize in section \ref{relatedWorkSummarySection}. Evangelizing well-definedness as a standard in neural architecture design and analysis is one of the core goals of this work.

Of course, reaching the standard of well-definedness is challenging. In order for a metric to become a universally-accepted guideline, it must be shown to have truly broad applicability. A fuzzy concept like ``exploding gradients'' can be re-interpreted from situation to situation to make it appear relevant. A concrete metric, on the other hand, is less ``flexible''. It can actually be shown to fail to predict performance in a given situation. While this is essential for scientific validity, it also means that careful and extensive study is required to find ``the right metric'' for capturing a phenomenon. We argue that this work establishes the NLC as a core measure of model complexity in deep learning and a key causal, explanatory driver of architecture performance. The fact that this work is so long, and such a large fraction of it is dedicated to one guideline, the NLC, underscores the effort involved in finding a metric. It took us several iterations of analysis to arrive at the present definition of the NLC.

While a large fraction of this work is dedicated to the NLC, we also introduce other ZSAD guidelines and flesh out existing ones. For none of them do we get to the point of defining them entirely in terms of a metric. However, for some guidelines (e.g. neuron bias), we make recommendations for what we consider are advantageous ways to measure them (e.g. LBIAS). We hope that future work can firmly establish those measures or develop better ones.

Even if we do not define a ZSAD strategy via a concrete metric, we at least ensure that an explicit definition is given, which goes beyond many other works.

\subsection{ZSAD is not a ``generalization measure'' - on the pre-eminence of the initial state}  \label{generalizationMeasuresSection}

As explained at the end of section \ref{zsadGuidelineFramingSection}, it is crucial that the property encapsulated by a ZSAD guideline is not a property of a trained network. As mentioned, we consider properties of the architecture as well as properties of its randomly initialized state. Of course, the property we ultimately care about, which is performance, is a property of the architecture's final state after training. One of the key challenges of ZSAD is to bridge the gap between the trained and untrained network. This is often simply done by observing that certain properties, like the NLC, are unlikely to change dramatically during training, and changes that do occur tend to be uncontrolled. As a general trend, we observe that it is advantageous for properties that we require after training to also hold before training so that a better trained network can be found by gradient methods. Of course, some properties, like parameter dimensionality, are fixed throughout training by definition.

We find it important to distinguish zero-shot architecture design from a sub-field of deep learning research that might be termed ``generalization measures''. \citet{generalizationOverview,generalizationDysfunction} are recent overview papers. Studies in this field estimate or bound neural network generalization either empirically, via metrics based on e.g. Jacobian singular values or classification margin, or theoretically, through frameworks like VC-dimension or PAC-Bayes. Crucially, they do this {\it for networks that were already trained}, and they {\it make use of the final parameter value}. In ZSAD, we attempt to predict the performance of an architecture after training {\it across} training algorithms and potentially {\it across} datasets {\it given access only to the architecture definition}, but {\it without conducting training or knowing the results of training}. (See section \ref{architecturePerformanceSection} for details.) The problem we study is harder and more practical. ZSAD can directly inform the manual architecture design stage. In contrast, having access to a theoretical estimate of generalization is much less meaningful after training, at which point it is possible to actually compute test error.

Going forward, when we refer to `predicting' a final state property such as test error, we always imply that the prediction was made before training.

There is some similarity in terms of study design between the generalization measures field and this work. For example, \citet{marginPrediction,explosionGeneralization} compare metrics that have some similarity to the NLC, evaluated in the final state, against performance for a range of architectures and training hyperparameters. Showing that metrics that have clear conceptual ties to error and loss actually empirically correlate with error and loss when evaluated on the very same network with the very same parameter value is not entirely satisfying. For example, error is clearly related to loss (else gradient-based training wouldn't work) and loss is clearly related to gradient magnitude (imagine scaling the loss function with a constant). Hence, showing that loss gradient-based metrics are related to error in the final state is a very different thing from showing that Jacobian-based metrics like the NLC, when evaluated in the initial state, are predictive of error.

\subsection{Summary}

Zero-shot architecture design (ZSAD) is an approach and a framework for architecture design based on general, predictive, explanatory and ideally well-defined principles. It empowers deep learning practitioners to make informed design decisions before any code is written, as architectures that can be predicted to perform badly can be eliminated from consideration a priori, which can significantly shorten the trial-and-error and NAS phases of deep learning deployment. We focus specifically on discovering patterns in test and training error that do not strongly depend on the data or the task domain. With this work, we hope to evangelize research practices like expressing ZSAD guidelines via well-defined metrics; addressing a range of utility criteria that are important for practical relevance; independently and exhaustively tuning key hyperparameters like learning rate to enable robust comparison; and empirically investigating a wide range of architectures.

\chapter{Background, notation, terminology and conventions} \label{backgroundChapter}

In this chapter, we build the core concepts that underpin this work from first principles: gradient-based training, neural networks, architecture design, etc. We define the building blocks of neural architectures and deep learning pipelines that feature in this work and are most popular in practice. We explain core terminology and define key notation that is used throughout this work.

We believe that reading this chapter enables a better appreciation of some of the nuances that arise throughout this work and in architecture design research in general. As such, while this chapter is primarily geared towards deep learning novices, we believe there is value for most readers. We provide a summary of the most important notation, terminology and conventions from chapters \ref{introductionChapter} and \ref{backgroundChapter} in section \ref{notationSummarySection} that should enable advanced readers to easily follow the technical material presented in later chapters as well as chapter \ref{introductionChapter}. This summary contains those definitions from chapters \ref{introductionChapter} and \ref{backgroundChapter} that (i) may be used in later chapters and chapter \ref{introductionChapter} without further explanation and (ii) are not widely agreed-upon in the machine learning community.

\section{Machine learning} \label{machineLearningSection}

On a high level, `machine learning' is about extracting useful information from raw data. It encompasses a wide range of tasks and settings where a wide range of questions are answered based on a wide range of data.

In this work, we focus on the simplest, most common and most well-studied setting: `supervised prediction'. In the prediction setting, the goal is to use a `model' $f$ to compute a `prediction' or `output' $f(x)$ from an `input' $x$ from a set $\mathbb{X}$. $f(x)$ should be close or equal to the `label' $y$ from a set $\mathbb{Y}$, which is the true value of some desired piece of information about the input. $f$ can be an arbitrary function defined on $\mathbb{X}$. (Note that throughout this work, we use the term `input' to refer to $x$ specifically as well as to a generic function or program input. We use the term `output' to refer to $f(x)$ specifically as well as to a generic function or program output. We hope this is sufficiently clear.)

{\bf Example.} In `image classification', we are given images depicting an object of a certain type, such as a cat, car or tree. The goal is to predict the object type from the image.

\begin{itemize}
\item[$x$] The image
\item[$\mathbb{X}$] The space of images considered, e.g. $\{0,1\}^{64*3S_1S_2}$ for 64-bit images with $S_1 \times S_2$ pixels with 3 color channels
\item[$y$] The true type of the object present in the image
\item[$\mathbb{Y}$] The set of object types considered, e.g. \{cat, dog, tree, plane, car\}
\item[$f$] A function taking an image as input and returning an object type or a distribution over object types as output
\item[$f(x)$] The type assigned to $x$ by $f$, or the distribution over types assigned to $x$ by $f$
\end{itemize}

{\bf Example.} In `sentiment analysis', we are given a piece of text describing a service or product. The goal is to predict how positive or negative the author's attitude towards that service or product is based on this piece of text.

\begin{itemize}
\item[$x$] The text segment
\item[$\mathbb{X}$] The space of text segments considered, e.g. $\{0,1,2,..,255\}^{100}$ for ASCII text segments of 100 characters
\item[$y$] The true sentiment score, e.g. +1 for entirely positive sentiment and -1 for entirely negative sentiment
\item[$\mathbb{Y}$] The set of sentiment scores considered, e.g. [-1, 1]
\item[$f$] A function taking a piece of text as input and returning a sentiment score as output.
\item[$f(x)$] The sentiment score assigned to $x$ by $f$
\end{itemize}

In general, we desire a model $f$ that exhibits `high performance'. We use this term in a broad sense. Depending on the task, it can cover anything from computational efficiency and privacy to adversarial robustness. The core aspect of performance for a prediction model is whether the model returns a prediction close to or equal to the label for a large number of inputs. A model exhibiting high performance is a function that reliably returns useful information, like the sentiment of text segments, which can then be used for a downstream application.

However, the automation of the prediction process is not yet considered machine learning. In classical artificial intelligence, decision rules that lead to accurate predictions were explicitly and manually specified. For machine learning to occur, the generation of the model itself must also be automated. Specifically, machine learning requires the running of an algorithm called the `training algorithm'. This algorithm takes in data and attempts to find a high-performing model, which it returns as its output. We denote the training algorithm by $\mathcal{A}$. The process of running the training algorithm is called `training' or `learning'.

In supervised prediction, we are given a `dataset' $D$ of `datapoints' $(x,y)$. Each datapoint consists of an input and label. In the image classification example, a dataset consists of images together with their respective true object type labels. In the sentiment analysis example, the dataset consists of text pieces together with their respective true sentiment scores. Because we can evaluate the performance of a model $f$ on $D$, we can search for a model that exhibits high performance on $D$. The rationale behind this is the assumption that if $f$ makes accurate predictions on $D$, it will also make accurate predictions on other inputs from $\mathbb{X}$. In general, this is of course not true. There is no inherent guarantee that what works on $D$ specifically works on $\mathbb{X}$ in general. For a counterexample, we can simply imagine a situation where the label has no statistical relationship at all with the input. For example, if we were to try to predict the value of an element in a sequence of independent random digits from its position in the sequence, no matter what (spurious) patterns we might find from studying a finite number of example digits, extrapolation would inevitably fail. 

However, in certain situations, the input and label have a statistical relationship that enables extrapolation. The simplest and by far the most common such relationship is continuity: If two inputs $x^{(1)}$ and $x^{(2)}$ are similar, as measured by e.g. Euclidean distance, then so tend to be their labels $y^{(1)}$ an $y^{(2)}$. For example, changing a single pixel in an image does not tend to change the object type depicted. Changing a single character in a text does not tend to change the sentiment. Virtually every practical model $f$ makes use of this principle to some degree and would indeed return similar outputs for inputs that are sufficiently similar. In general, whatever explicit or implicit strategy a model uses for extrapolation is called its `inductive bias'. It is the responsibility of the machine learning practitioner to select a training algorithm that generates models that have an inductive bias that matches the statistical relationships present in the data at hand. The model has the property of `generalization' to the degree to which the accuracy of the predictions translates from $D$ to $\mathbb{X}$. Conversely, the model has the property of `overfitting' to the degree to which the accuracy of the predictions does not translate from $D$ to $\mathbb{X}$. Finally, the model has the property of `underfitting' if the accuracy does translate, but was not high to begin with.

When we say that there is a statistical relationship between input and label, we imply that, for a given input, some labels are more likely to occur than others. This idea can be encapsulated by a function that maps inputs to distributions over labels which mimics the natural process that generates the data. For example, if the input is a gene sequence, we might conceptualize the distribution over phenotypes that arises from each gene sequence as a function from gene sequences to distributions over phenotypes. We term this construct the `true input-label function'. Commonly, it is also reasonable to assume that this function is deterministic, i.e. that it maps inputs to labels. In many applications in artificial intelligence, such as image classification and sentiment analysis, situations where the label is ambiguous are rare. Going forward, we will assume determinism.

In this work, we focus on `classification', the most common type of supervised prediction. Here, $\mathbb{Y}$ is a discrete, finite set and inputs which share a label constitute a `class'. The most popular example is image classification, as described above. Sentiment analysis can also be cast as a classification task if we consider reviews with positive or negative sentiment as the respective classes. In contrast, the setting where $\mathbb{Y}$ is a subset of a real vector space is called `regression'.

By far the most popular strategy for training in supervised prediction is `empirical risk minimization' (ERM), which we focus on in this work. Here, a `loss function' $\ell$ over the prediction and the label is defined. The training algorithm then corresponds to an optimization algorithm that searches for a model $\hat{f}$ from a `hypothesis space' $\mathcal{H}$ for which the `loss' is as low as possible.

\begin{equation*}
\hat{f} = \mathcal{A}(\mathcal{H},\ell,D) \approx \arg \min_{f \in \mathcal{H}} \frac{1}{|D|}\sum_{(x,y)\in D}\ell(f(x),y)
\end{equation*}

We use $\approx$ loosely here to denote, among other things, that the training algorithm cannot usually solve the minimization problem exactly, that there may be no minimum, that there may be more than one minimum or that there may be additional terms in the objective function (e.g. regularization).

The idea behind ERM is that we gain some (negative) utility from each prediction made by the model on $D$: if the prediction is accurate, we suffer little to no negative utility; if the prediction is inaccurate, we suffer large negative utility. ERM attempts to minimize the total negative utility suffered. Note that $\ell$ is often chosen not as the best estimate of the actual negative utility suffered (if such a thing even exists), but to make optimization easier / possible. For example, in many cases, an accurate metric of the negative utility suffered in practice is 0 when $f(x) = y$ and 1 when $f(x) \neq y$. However, this metric is generally discontinuous as a function of $f(x)$, which makes optimization difficult. Hence, a surrogate continuous loss function is often used instead. In contrast to the loss function, the metric that represents the best estimate of the actual negative utility suffered is called the `error function' $e$, which is often discontinuous. Of course, $\ell$ and $e$ are also identical in many situations.

We can estimate whether a given model generalizes well if we make the additional assumption that the end goal is to minimize negative utility on datapoints drawn from some `data distribution' $\mathcal{D}$ with support in $(\mathbb{X} \times \mathbb{Y})$ as well as the assumption that the datapoints in $D$ are drawn IID from $\mathcal{D}$. We call the marginal of $\mathcal{D}$ over inputs $x$ the `input distribution'. We also use the symbol $\mathcal{D}$ to denote the input distribution, as it is clear from context which distribution $\mathcal{D}$ refers to. (If the expression containing $\mathcal{D}$ also contains $y$, it refers to the data distribution. If not, it refers to the input distribution.) In discussion, we use the term `data distribution' more generally to refer to either or both constructs.

The `true error' or `generalization error' $E_\text{true}$ is then defined as follows.

\begin{equation*}
E_\text{true}(f, e, \mathcal{D}) = \mathbb{E}_{(x,y) \sim \mathcal{D}} e(f(x),y)
\end{equation*}

We will make the assumption of the existence of $\mathcal{D}$ from which the dataset is drawn throughout this work. The simplest and most popular way to exploit this assumption is to split the dataset into two shards in a uniformly random fashion. One of the shards is then used for ERM. That shard is called the `training set'. The other shard is used to estimate the true error and is called the `test set'. Specifically, the true error can be estimated via the `test error', which is simply the error obtained on the test set.

\begin{equation*}
E_\text{test}(f, e, D_\text{test}) = \frac{1}{|D_\text{test}|}\sum_{(x,y)\in D_\text{test}}e(f(x),y)
\end{equation*}

By making weak assumptions about the distribution of $e(f(x),y)$ induced by $\mathcal{D}$, we can place the true error within a frequentist confidence interval around the test error. By assumption / construction, the test set is independent of the training set and so the test error is an unbiased estimate of the true error. In this way, it is possible to compare different models, which were perhaps obtained from different training algorithms, using a single training and test set. In a practical research context, in the supervised prediction setting, generalization is usually measured via test error.

However, if we were to choose the model with the lowest test error as our final model, we are again left without an unbiased estimate of its true error. This is because if we choose between a large number of models in this way, the resulting model will be biased towards having a test error that is lower than its true error. To overcome this problem, a third shard is often carved out of the dataset, called the `validation set'. Given multiple models, we can choose the best one by comparing the `validation error' $E_\text{valid}$, which is evaluated on the validation set analogously to the test error. Then, finally, we can use the test error as an unbiased estimate of the true error of the chosen model. We use the term `data shard' to refer to any subset of the dataset, including the dataset itself, but most often to refer to either the training set, validation set or test set. Below, we give a complete set of definitions.

\begin{metricDefinition}
The `true error' $E_\text{true}$, `test error' $E_\text{test}$, `validation error' $E_\text{valid}$, `training error' $E_\text{train}$, `true loss' $L_\text{true}$, `test loss' $L_\text{test}$, `validation loss' $L_\text{valid}$ and `training loss' $L_\text{train}$ are

\begin{eqnarray*}
E_\text{true}(f, e, \mathcal{D}) &=& \mathbb{E}_{(x,y) \sim \mathcal{D}} e(f(x),y)\\
E_\text{test}(f, e, D) &=& \mathbb{E}_{(x,y) \in D_\text{test}} e(f(x),y)\\
E_\text{valid}(f, e, D) &=& \mathbb{E}_{(x,y) \in D_\text{valid}} e(f(x),y)\\
E_\text{train}(f, e, D) &=& \mathbb{E}_{(x,y) \in D_\text{train}} e(f(x),y)\\
L_\text{true}(f, \ell, \mathcal{D}) &=& \mathbb{E}_{(x,y) \sim \mathcal{D}} \ell(f(x),y)\\
L_\text{test}(f, \ell, D) &=& \mathbb{E}_{(x,y) \in D_\text{test}} \ell(f(x),y)\\
L_\text{valid}(f, \ell, D) &=& \mathbb{E}_{(x,y) \in D_\text{valid}} \ell(f(x),y)\\
L_\text{train}(f, \ell, D) &=& \mathbb{E}_{(x,y) \in D_\text{train}} \ell(f(x),y)
\end{eqnarray*}

\end{metricDefinition}

Here, and throughout this work, we denote the mean over a finite set with the expectation operator. A desideratum for loss functions is that they cause training to lead to low true error.

Throughout this work, we follow the formalism of using a training set for ERM and a test set for evaluating the final model. Therefore, we use the term `performance' largely as a synonym for test error and sometimes as a synonym for training error, depending on context. For the purpose of greater understanding, we are often also interested in training error.

Throughout this section, we have introduced several idealizing assumptions. In practice, labels are not ``true values'', but the outcome of some imperfect labeling process, such as human annotation. Hence, by assuming these labels are correct, we are susceptible to learning models that mimic the shortcomings of the labeling process. In general, datapoints contained in $D$ do not have the same distribution as datapoints the model will ultimately be deployed on. For example, the scenes that appear in images and the words that appear in text change over time and from situation to situation. Also, there may be no such thing as a single correct label. For example, more than one object might be equally prominent in an image. The author of a product review might have a neutral stance. Finally, some datapoints may be partially missing from $D$. There is ample work investigating the relaxation of these idealizing assumptions. Undoubtedly, many results we present in this work would still be meaningful or could be easily generalized to these relaxed settings. However, such an investigation goes beyond the scope of this work.

\subsection{Program / function overloading} \label{programFunctionOverloadingSection}

We have introduced a prediction model $f$ as a function that takes inputs $x$ and returns predictions $f(x)$. As we further explain in sections \ref{programFunctionSection} and \ref{neuralNetworkNotationSection}, there is a duality between programs and functions in machine learning in general. In addition to being viewed as a mathematical function that maps elements of the set $\mathbb{X}$ to elements of the set $\mathbb{Y}$, $f$ can also be viewed as a computer program that consumes inputs of data type $\mathbb{X}$ and produces outputs of data type $\mathbb{Y}$. Similarly, the test error $E_\text{test}$ can be viewed as the mathematical average of values $e(f(x),y)$, or it can be viewed as the floating-point computation that produces such an (approximate) average. Similarly, the dataset $D$ can be viewed as a finite subset of $\mathbb{X}$ or as a physical data collection stored on a machine. Because conceptual development necessarily takes place in the realm of mathematics and practical deployment necessarily takes place in the realm of computation, we are always confronted with this duality.

Throughout this work, we will introduce a large number of concepts that exhibit this duality. We will cope with this by overloading our notation and terminology. For example, when we write ``model'' or $f$, we refer to both the mathematical and computational object at the same time. We endow $f$ with computational properties, such as runtime or memory requirement, as well as mathematical properties, such as gradient or convexity. For this to make sense in practice, we have to assume that the program is a sufficiently close approximation of the mathematical object. This requires, for example, that the rounding error induced by floating-point computation is not too large and that expectations over distributions are computed over a sufficiently large number of samples. In general, we will implicitly ensure this throughout the work. However, we will also explicitly discuss this issue whenever the need arises, such as in sections  \ref{metricsSection} and \ref{nlcComputeSection}.

\section{Training with gradient methods} \label{gradientMethodSection}

Over the decades, a type of training algorithm has emerged dominant: `gradient methods'. These algorithms can be used to (approximately) solve general optimization problems of the form $\arg \min_{\theta \in \mathbb{T}} F(\theta)$, where $F$ is a differentiable function of the `parameter' $\theta$ called the `objective function'. $\theta$ is an element of the `parameter space' $\mathbb{T}$, which is generally a subset of a real vector space. The algorithm starts with an `initial parameter value' $\theta^{(0)}$. It then proceeds iteratively. At `iteration' $t$, it takes the gradient of the objective function with respect to the parameter at its current value $\frac{dF(\theta^{(t-1)})}{d\theta}$. It then uses this gradient to determine the `update' $\delta \theta^{(t)}$, and then adds it to the current parameter value to obtain the new value $\theta^{(t)} = \theta^{(t-1)} + \delta \theta^{(t)}$. After $T$ iterations, we obtain the `final parameter value' $\theta^{(T)}$, which can be viewed as the best effort solution to the optimization problem. 

The simplest gradient method is `gradient descent'. There, the update is a negative scalar multiple of the gradient. Since $F$ is differentiable, if this multiple has small enough length, we are guaranteed $F(\theta^{(t)}) < F(\theta^{(t-1)})$. We are further guaranteed to converge to the globally optimal $\theta$ if, for example, $F$ is strongly convex and update lengths are chosen appropriately. More complex gradient methods might combine the current gradient with past gradients (e.g. Nesterov's accelerated gradient \citep{nesterovOrig}) or second-order information such as the Hessian $\frac{d^2F}{d\theta^2}$ (e.g. Newton's method \citep{NumericalOptimization}).

To use gradient methods specifically as training algorithms that return models, we augment $f$ with a parameter $\theta$, which is generally a real-valued vector. Instead of having the training algorithm choose $f$ directly, we fix $f$ and have the training algorithm choose only $\theta$. In the context of empirical risk minimization, this yields

$$\hat{\theta} = \theta^{(T)}= \mathcal{A}(f,\theta^{(0)},
\ell,D_\text{train}) \approx \arg \min_{\theta \in \mathbb{T} \subseteq \mathbb{R}^{\text{dim}(\theta)}} \frac{1}{|D_\text{train}|} \sum_{(x,y) \in D_\text{train}} \ell(f(\theta,x),y)$$

Now, the hypothesis space $\mathcal{H}$ becomes $\mathbb{T}$, the objective function $F$ becomes $L_\text{train}$, and the prediction is simply $f(\theta,x)$. Training a model with a gradient method is also called `gradient-based training'. The degree to which it is possible for a model to achieve low training error by gradient-based training is called its `trainability'.

If $|D_\text{train}|$ is large, computing the gradient is expensive. Instead of computing the exact gradient, we can approximate it stochastically by replacing $D_\text{train}$ with a random subset. This subset is known as the `batch' $B$. If $B^{(t)} \subseteq D_\text{train}$ is the batch used at iteration $t$, then the gradient used at iteration $t$ is $\frac{d}{d\theta}\frac{1}{|B^{(t)}|} \sum_{(x,y) \in B^{(t)}} \ell(f(\theta^{(t-1)},x),y)$. Gradient methods using this approximation instead of the exact gradient are called `stochastic gradient methods'. The batch is usually drawn (approximately) uniformly at random, which yields an (approximately) unbiased approximation of the gradient. The stochastic equivalent of gradient descent is called `stochastic gradient descent' (SGD). In the case of non-stochastic methods, for consistency of notation, we write $B^{(t)} = D_\text{train}$ and say that that the batch equals the training set. If there is a sequence of iterations $t, t+1, .., t'$ such that $|B^{(t)}| + |B^{(t+1)}| + .. + |B^{(t')}| \approx |D_\text{train}|$, then this sequence of iterations is called an `epoch', indicating that each datapoint contributed approximately once on average to a gradient computation during that sequence of iterations. If the size of the batch is equal to a fixed value $|B|$ in each iteration, and if that batch size divides the training set size, then we say an epoch contains exactly $\frac{|D_\text{train}|}{|B|}$ iterations.

Instead of a specific update, many gradient methods merely generate an `update proposal' $\delta_\text{prop}\theta$ at each iteration. The final update $\delta_\text{prop} \theta$ is then some positive scalar multiple $\alpha\delta_\text{prop} \theta$ of the proposal. $\alpha$ is called the `learning rate'. In the case of gradient descent, the update is a negative scalar multiple of the gradient, and hence the negative gradient is the proposal. While there are strategies for finding a good learning rate for gradient descent, the algorithm itself does not specify it. It is a `hyperparameter'. Oftentimes, whether a gradient method succeeds in solving the optimization problem depends critically on using not just one effective value for $\alpha$, but an effective value at every iteration. 

\subsection{How differentiable does $f$ need to be?} \label{howDifferentiableSection}

The gradient method formalism, at first glance, requires $f$ to be differentiable with respect to $\theta$ for each value $\theta$ can take. In practice, many models $f$ that are used with gradient methods are not differentiable everywhere. A key strength of gradient methods is that even when batches are used, it is not necessary to compute the exact gradient over the batch, but only a (further) approximation. If the value of $\frac{d\ell(f(\theta^{(t-1)},x),y)}{d\theta}$ for some $(\theta^{(t-1)},x,y)$ is required during the running of the gradient method, then it is enough to find a linear function $\mathsf{lin}(\theta)$ such that $\mathsf{lin}(\theta)$ approximates $f(\theta,x)$ as a function of $\theta$ sufficiently accurately in a sufficiently large neighborhood of $\theta = \theta^{(t-1)}$. $\mathsf{lin}$ can depend on $(\theta^{(t-1)},x,y)$. $\mathsf{lin}$ is a `local linear approximation' of $f$. The gradient of $\mathsf{lin}$ can then be used as a surrogate gradient for the purpose of the gradient method. Using the exact gradient for the gradient method, $\mathsf{lin}$ becomes the tangent hyperplane of $f(\theta,x)$ at $\theta^{(t-1)}$ and so we have $\mathsf{lin}(\theta) = f(\theta^{(t-1)}, x) + \frac{d\ell(f(\theta^{(t-1)},x),y)}{d\theta}(\theta - \theta^{(t-1)})^T$. Of course, there is no guarantee that the tangent hyperplane is a sufficiently accurate approximation of $f$ in a sufficiently large neighborhood of $\theta^{(t-1)}$ that meaningful progress can be made by the gradient method, nor is there a guarantee that the tangent is the best local linear approximation for this purpose. We refer to the tangent hyperplane as the `gradient-based local linear approximation'.

Throughout this work, we will use the notation and terminology of differentiability while understanding that our analysis generalizes smoothly to certain common non-differentiable contexts. We discuss this issue in detail in section \ref{nonDifferentiableSection}.

\subsection{Model Selection} \label{modelSelectionSection}

The price we pay for using gradient methods is that we can only consider hypothesis spaces that can be cast as a parameter vector space and a single model which is differentiable with respect to the parameter. A search over such a space with a gradient method is often not sufficient in practice. For example, consider the linear model $f_\text{lin}(\theta,x) = \theta^Tx$ and the Gaussian model $f_\text{gauss}(\theta,x) = \frac{1}{\sqrt{2\pi}}e^{-\frac{1}{2}||x-\theta||^2_2}$. There is no clear way to cast the union of both parameter spaces as a single parameter space that can be searched meaningfully by a gradient method. Therefore, in practice, we are left with the challenge of selecting a model on top of selecting a parameter. This process is aptly called  `model selection'. The simplest strategy would be to enumerate a finite number of parametrized models, run a gradient method on each of them, and then select the best-performing model-parameter pair via validation error. Of course, there are also much more complex model selection methods that e.g. search over the model and parameter jointly, using a gradient method as a sub-routine to update the parameter.

\subsection{Hyperparameter tuning} \label{hyperparameterTuningSection}

In practice, there are often a plethora of hyperparameters that need to be set a priori, before the training algorithm can begin. We have already encountered several of them. In the case of gradient methods, we generally need to choose a model as well as a learning rate. We also need to choose $\ell$ and $\theta^{(0)}$. Further hyperparameters arise due to the simple fact we can always embellish training algorithms with additional arguments, no matter how general or complex that training algorithm already is. The machine learning community is always hard at work inventing new arguments.

Hyperparameters are often set manually. Any automated process for choosing hyperparameter values is known as `hyperparameter tuning'. Model selection is a special case of hyperparameter tuning that is concerned with choosing a model to supply to a gradient method. Of course, hyperparameter tuning can be a bottomless pit. There is no reason that an algorithm for hyperparameter tuning should not itself have hyperparameters that need to be tuned by another algorithm and so on. Eventually, we must manually conduct the highest level of tuning. (This fact alone shows that there is no such thing as fully automated machine learning.) Furthermore, where does training end and hyperparameter tuning begin? After all, both processes ultimately exist for the purpose of making automated predictions. In general, if a gradient method is used in the process of training, any computation associated with that gradient method is considered part of the training algorithm and any computation beyond the gradient method is considered part of hyperparameter tuning.

We discuss our protocol for hyperparameter tuning in chapter \ref{empiricalStudiesChapter}. We largely focus on exhaustive learning rate tuning.

\section{Neural networks} \label{neuralNetworksSection}

The term `neural network' is an elusive one and has been applied to seemingly disparate corners of the machine learning spectrum. It originally stems from the neuroscientific concept of the same name, but quickly took on its own meaning within the machine learning community. In section \ref{historySection}, we present a brief overview of the history of neural networks. We argue that the most useful, if imperfect, definition of a neural network in modern machine learning is as follows.

\begin{definition}
A `neural network' is any model to which gradient methods can be applied.
\end{definition}

We detail and argue this definition in section \ref{functionalGradientSection}. Note that, in general, gradient methods can be applied to constructs more complex than simple functions. They can be applied to functions with multiple inputs and outputs, a variable number of inputs and outputs, functions with ``memory'', etc. As stated above, in this work we focus on supervised prediction, where the network generally corresponds to a simple function where the single input, the single output and the parameter are real vectors of fixed dimensionality. Such a network is termed a `feedforward network'. Throughout this work, we often use the terms `neural network' and `feedforward network' interchangeably, especially since the term `feedforward network' does not have an entirely agreed-upon definition in the community. Feedforward networks include fully-connected networks (section \ref{architectureDesignParadigmsSection}) and convolutional networks (section \ref{architectureDesignParadigmsSection}), among others.

\subsection{Neurons and depth} \label{neuronDepthSection}

Beyond being a function, a neural network generally has additional structure. Specifically, it is composed of a large number of simple operational units called `neurons' that successively extract useful information from the input. These neurons take as input (parts of) the network input as well as outputs of other neurons. Each input to a neuron is a scalar and the output of a neuron is a scalar. If neuron A directly takes in the output of neuron B, we say neuron B is a `dependency' of neuron A. Hence, the neurons form a directed, acyclic graph in which neuron B is the parent of neuron A if it is a dependency of neuron A. This graph is termed the `neuron graph' and literally represents a ``neural network''.

A key property of neural networks is that neurons often form long `dependency chains', i.e. some neuron takes in the output of another neuron, which takes in the output of another neuron, which takes in the output of another neuron and so on. This property is known as `depth', and the length of the longest dependency chain in a given network is known as the `depth of the network'. Networks that are particularly deep are also known as `deep neural networks', although the meaning of the phrase ``particularly deep'' is highly subjective and situational. The strategy of applying a deep neural network (or, really, any neural network) to a machine learning task is known as `deep learning'.

In feedforward networks, the neuron graph is static and each neuron is evaluated exactly once, as soon as all its dependencies are evaluated. In order to cast such a neuron graph as a model $f$, it must take in an input $x$ and return an output $f(x)$. A network generally has a dedicated set of `input neurons', which do not have parents in the neuron graph. To make a prediction for an input vector $x$, each component of $x$ is assigned to its corresponding input neuron. The network is then `evaluated', i.e. each neuron in the neuron graph is evaluated individually when all its parents have been evaluated. The evaluation of input neurons can be considered trivial. A network generally also has a dedicated set of `output neurons', which do not have children in the neuron graph. The output $f(x)$ is then a vector where each component is equal to its corresponding output neuron.

A simple network is given below.

\begin{eqnarray*}
z_1 &=& -\mathsf{in}^2\\
z_2 &=& e^{z_1}\\
z_3 &=& \cos(z_1) + \mathsf{in}\\
\mathsf{out} &=& z_1 + z_2*z_3
\end{eqnarray*}

$\{\mathsf{in},z_1,z_2,z_3,\mathsf{out}\}$ are the neurons. The longest dependency chain here is $(\mathsf{in}\rightarrow z_1,z_1 \rightarrow z_3,z_3\rightarrow \mathsf{out})$, so this network has depth 3. However, we can already see the highly volatile nature of the concept of depth, as we could simply define a different network as follows.

\begin{eqnarray*}
z_1 &=& -\mathsf{in}^2\\
z_2 &=& e^{z_1}\\
z_3 &=& \cos(z_1)\\
z_4 &=& z_3 + \mathsf{in}\\
\mathsf{out} &=& z_1 + z_2*z_4
\end{eqnarray*}

Compared to the previous network, the depth has increased to 4 but the network function has remained the same. Further, the compiler might transform the programs implementing both networks into the exact same machine code. Hence, the concrete value of depth often depends on choices which are quite arbitrary.

\subsection{Training neural networks} \label{trainingNeuralNetworksSection}

In a modern context, neural networks are trained with gradient methods, though alternative algorithms have been used widely in the past (section \ref{historySection}) and are being developed at present (e.g. \citet{directFeedbackAlignment,nonGradientTraining1, nonGradientTraining2,nonGradientTraining3,nonGradientTraining4}). To use gradient methods, we augment the network with a parameter $\theta$. $f$ is then referred to as the `neural architecture' or simply `architecture', though the terms architecture and network are interchangeable in many situations. Training then yields the network $f(\theta^{(T)},x)$. In general, the initial parameter value $\theta^{(0)}$ is drawn from a distribution called the `parameter initialization scheme', and that distribution is usually specified with and considered part of the architecture. Throughout this work, when we refer to choosing an architecture, we include in that choice the parameter initialization scheme. The network $f(\theta^{(0)},x)$, as it exists at the beginning of training, is known as the `randomly initialized state' or simply `initial state' of the architecture and $f(\theta^{(T)},x)$ is known as the `final state'. We refer to a singular execution of a training algorithm that transforms an initial state network into a final state network as a `training run'.

\subsection{Layers} \label{layersSection}

In addition to having neurons, neural networks have `layers'. A layer is a group of neurons. The layers themselves form a directed, acyclic graph called the `layer graph'. In it, layer A is a parent of layer B if there exists a neuron in A that is the parent of a neuron in B. By convention, the following properties hold: (i) each neuron is grouped into exactly one layer and (ii) neurons in the same layer do not depend on each other. This ensures that neurons in the same layer can be evaluated in parallel, as soon as all parent layers have been evaluated. Feedforward networks generally have a single layer that consists of the input neurons (`input layer') and a single layer that consists of the output neurons (`output layer'). The longest directed path in the layer graph is then an upper bound on the depth of the network, and is generally equal to the depth of the network. Layers return vectors of dimensionality equal to the number of neurons they contain. Each component of that vector corresponds to the scalar returned by the corresponding neuron. The number of neurons in a layer is called its `width', which is synonymous with `dimensionality'. A layer takes as input the output vectors of its dependencies in the layer graph.

In practice, the operations of neurons in the same layer are identical or near-identical. Hence, the layer concept encourages the utilization of large numbers of similar neurons that are evaluated in parallel. This has obvious benefits. Identical parallel operations can be evaluated efficiently on specialized hardware such as GPUs. It is much easier to specify a layer graph than a neuron graph if the number of neurons is very large. In fact, when working with layers, a neuron simply becomes a vector component. The neuron concept is then generally no longer explicitly considered. 

A significant fraction of this work is grounded in the layer concept. This fact in itself provides an implicit explanation for the power of layers. We make this explanation explicit in chapter \ref{surveyChapter}.

\subsection{Neural network notation and terminology} \label{neuralNetworkNotationSection}

We write a network $f$ as a succession of layers $f_l$, $0\le l\le L$.  Each layer returns a real-valued vector of dimensionality $d_l$. Each of the $d_l$ components of $f_l$ is a neuron. While we assume the width is fixed for each layer, it can vary between layers. Each layer $f_l$ has zero or more other layers as dependencies, and it takes as input the outputs of its dependencies. Without loss of generality, assume that the index of a dependency of a layer is lower than the index of the layer itself. So for example, we cannot have $f_4$ be the dependency of $f_2$. $f_0$ is the input layer to which $x$, the input of the network, is assigned. $x$ is a real-valued vector of fixed dimensionality $d_\text{in}$, which implies $d_\text{in}=d_0$. Each of the $d_\text{in}$ components of $x$ is a `feature'. The `output layer' $f_L$ returns the output of the network, a vector of fixed dimensionality $d_\text{out} = d_L$.

To use gradient methods, each layer is augmented with a real-valued `parameter sub-vector' $\theta_l$ with a fixed dimensionality that is possibly zero. The $\theta_l$ collectively make up the parameter vector $\theta = (\theta_1,..,\theta_L)$. $f$ then becomes an architecture and $f$ together with a specific value of $\theta$ is a network. $f_l$ in isolation becomes an architecture layer and $f_l$ together with a specific value of $\theta_l$ is a network layer.

$k_l$ denotes the vector of the indices of the dependencies of $f_l$, and $K_l$ denotes the dimensionality of $k_l$. Hence for any $1 \le l \le L$ we have

$$f_l(\theta_l,f_{k_l[1]},f_{k_l[2]}, .., f_{k_l[K_l]}): \mathbb{R}^{\text{dim}(\theta_l)} \times \mathbb{R}^{d_{k_l[1]}} \times \mathbb{R}^{d_{k_l[2]}} \times .. \times \mathbb{R}^{d_{k_l[K_l]}} \rightarrow \mathbb{R}^{d_l}$$

This definition implies that each layer is well-defined for any set of vector inputs of a given dimensionality, which is generally the case in practice and which we assume throughout this work. Similarly, we consider $\mathbb{X}$ as $\mathbb{R}^{d_\text{in}}$, the network $f$ as a function from $\mathbb{R}^{d_\text{in}}$ to $\mathbb{R}^{d_\text{out}}$ and the architecture $f$ as a function from $\mathbb{R}^{\text{dim}(\theta)} \times \mathbb{R}^{d_\text{in}}$ to $\mathbb{R}^{d_\text{out}}$. Throughout this work, square brackets $[]$ are reserved for indexing tensors in addition to denoting closed intervals. If layer $f_l$ has a single dependency, we often omit the square brackets and write $f_{k_l}$ or simply $f_k$ for the dependency. Throughout this work, $k$ subscripts of quantities indicate that this quantity corresponds to the dependency or dependencies of $f_l$. In contrast, we use an $m$ subscript just like an $l$ subscript to refer to arbitrary layers.

Throughout a large fraction of this work, including parts of this section, we do not consider the parameter $\theta$ explicitly. For brevity and readability, we then discuss neural networks and layers as if they did not have parameters. Adding the parameter to our notation and terminology in those cases is always straightforward. (For example, when we say a network ``has a Jacobian'', we ignore the parameter.)

\paragraph{5-fold overloading: program / function / graph / vector / distribution} In section \ref{programFunctionOverloadingSection}, we explained how we overload our notation and terminology to simultaneously refer to mathematical functions and computer programs whenever a concept exhibits this duality. For neural networks, things ``get worse''. In addition to being a function and a program, a neural network is also a neuron graph, a layer graph and can be viewed as the vector value represented by its output. Given an input distribution, it can be viewed as a distribution. As opposed to the layer graph, we never consider the neuron graph explicitly. However, we will further overload our notation and endow it with both graph and vector properties.

When we use, for example, $f_l$ to refer specifically to the layer {\it function}, we also refer to it as the `layer operation'. $f_l$ where $l > 0$ takes $K_l$ vectors as input as discussed above. The function that is $f$ is called simply the `network function'.

As a {\it program}, $f_l$ where $l > 0$ consumes $K_l$ floating-point arrays with $d_{k_l[1]}, .., d_{k_l[K_l]}$ entries respectively and is referred to as the `layer program'. The input-output mapping represented by the program is identical to the function up to floating-point rounding error. Popular layer operations correspond to programs that do not experience underflow or overflow if the input arrays do not take extremely small or large values. However, when composing many layers or training with e.g. high learning rates, underflow or overflow can occur. We discuss this further in section \ref{architectureDesignParadigmsSection} under `scale stability'. When considering the entire `network program' $f$, layers also take on the meaning of variables in that program.

As a {\it graph node}, $f_l$ has properties such as children (layers that depend on $f_l$), parents (dependencies of $f_l$), ancestors (layers from which there exists a directed path to $f_l$) and descendents (layers to which there exists a directed path from $f_l$). If every directed path from $f_0$ to $f_l$ contains $f_m$, we say $f_m$ is a `bottleneck' for $f_l$.

When we use $f_l$ to denote the output {\it vector} of a layer, we also refer to it as the `layer value'. In this context, $f_l$ has properties like components, length and norm. Specifically, we write $f_l[i_l]$ to refer to the $i_l$'th component of $f_l$, where $0 \le i_l < d_l$. (Note that our layer component indices are zero-based.) Often, we further shorten this to $f_l[i]$. We also alternatively write $f_l[j_l]$. Each component corresponds to a neuron, and the scalar output of that neuron is the `neuron value'. The output of the network is the `network value'. When we use layers or networks to form mathematical expressions, we generally consider their vector values. For example, $f_l=f_m$ indicates that the value returned by both layers is equal, not that they are the same graph node or that the layer operations are equals.

Often, we associate a network with an input distribution $\mathcal{D}$.  Then, $f_l$ can denote a {\it distribution}. We use the terms `output distribution', `layer distribution' and `neuron distribution' respectively. Drawing from that distribution is equivalent to drawing an input from $\mathcal{D}$ and then forward-propagating that input. As a distribution, $f_l$ has properties like expectation and (co)variance. 

5-fold overloading leads to compact and consistent notation at the price of a certain amount of ambiguity. We endeavor to limit this ambiguity throughout this work.

\paragraph{Omitting function inputs} Throughout this work, for brevity and readability, we often omit inputs of functions in our notation. For example, if we consider a network without reference to the parameter value or input, we may simply write $f$ instead of $f(\theta,x)$. If we consider a layer of a network or architecture without reference to a parameter value, we may simply write $f_l$ or $f_l(f_{k_l[1]}, .., f_{k_l[K_l]})$ or $f_l(f_k)$. Further, we sometimes consider a layer not as a function of its parents, but of one or more ancestors. For example, we write $f_l(x)$ to denote the value of $f_l$ as $x$ varies. Hence, $f(x)$ and $f_L(x)$ denote the same function. If $f_m$ is a bottleneck of $f_l$, we write $f_l(f_m)$ to denote the value of $f_l$ as $f_m$ varies. In this expression, $f_l$ is a function but $f_m$ is a vector. In general, we pick and choose which inputs we denote explicitly. The same is true for subscripts and superscripts.

\paragraph{Further concepts} Let $\mathcal{J}_{l,m}(\theta, x, y)$ be the Jacobian $\frac{df_l}{df_m}$ of $f_l$ with respect to $f_m$ evaluated with parameter $\theta$ at $(x,y)$, where $0\le m\le l\le L$. We shorten the Jacobian of the output $\mathcal{J}_{L,l}$ to $\mathcal{J}_l$ and we further shorten $\mathcal{J}_{L,0}$ to $\mathcal{J}$, i.e. $\mathcal{J} = \frac{df}{dx}$.

We consider the loss function $\ell$ to take as inputs the network output $f(x)$ as well as the label $y$. We denote the gradient of the loss function $\frac{d\ell(f,y)}{df_l}$ by $g_l$.

We denote the layer distribution of $f_l$ by $f_l(\mathcal{D})$. In general, when we use some distribution $\mathsf{dist}$ like a variable in any expression, e.g. $\mathsf{dist} + v$ or $\mathsf{dist}[i]$, then that expression also denotes a distribution where drawing from that distribution is equivalent to drawing from $\mathsf{dist}$ and then evaluating the expression.

When we reference the distribution of a datapoint $(x,y)$, it is the data distribution $\mathcal{D}$ by default. When we reference the distribution of an input $x$, it is the input distribution $\mathcal{D}$ by default. Hence, we write e.g. $\mathbb{E}_{(x,y)}$ or even $\mathbb{E}$ short for $\mathbb{E}_{(x,y)\sim\mathcal{D}}$, and $\mathbb{E}_x$ or even $\mathbb{E}$ short for $\mathbb{E}_{x\sim\mathcal{D}}$. Conversely, we write e.g. $\mathbb{E}_{i_l}f_l[i_l]$ or $\mathbb{E}_{i}f_l[i]$ for the finite mean of neurons in a layer. A bar on top of an expression indicates its expectation with respect to $\mathcal{D}$. So we write e.g. $\bar{f}$ short for $\mathbb{E}_{x\sim\mathcal{D}}f(x)$. Correspondingly, we write $\Cov_\mathsf{vec}$ for the covariance matrix of the vector $\mathsf{vec}$ with respect to $\mathcal{D}$.

Throughout this work, we use the term ``expectation'' specifically to refer to the $\mathbb{E}_{x\sim \mathcal{D}}$ and $\mathbb{E}_{(x,y)\sim \mathcal{D}}$ operations, while we use the term ``mean'' to refer to unweighted averages of finite sets, such as $\mathbb{E}_{i}f_l(x)[i]$ and $\mathbb{E}_{(x,y)\in D}$, as well as the mean parameter of Gaussian distributions. While verbally distinguishing between e.g. $\mathbb{E}_xf_l(x)$ and $\mathbb{E}_if_l(x)[i]$ can be tricky at times, we hope this convention will improve readability.

The same notational conventions that apply to the $\mathbb{E}$ operator apply to other probabilistic operators, such as the standard deviation $\mathbb{S}$, throughout this work.

\paragraph{Our vectors are row vectors} One can either consider the vectors that have been defined as row vectors or column vectors. Depending on this choice, mathematical expressions such as the inner product, while equivalent, look somewhat different. Throughout this work, we employ the following conventions. Vectors like $f_l$ and $\theta_l$ are row vectors. Correspondingly, Jacobians have left dimension equal to the output dimensionality and right dimension equal to the input dimensionality. For example, $\mathcal{J}_{l,m}$ has size $d_l \times d_m$.

\subsection{Forward propagation and backpropagation} \label{backpropagationSection}

\begin{algorithm}[t]
\CommentSty{\color{blue}}
\KwIn{$f_l$ for $0 \le l \le L$: values of all layers obtained during forward propagation}
\KwOut{$\phi_l$ for $1 \le l \le L$: gradients of the loss function $\ell$ with respect to parameter sub-vectors $\theta_l$}
\For{$l=0$ \KwTo $L$}
{
	$g_l := 0$\;
}
$g_{L} := \frac{d\ell(f,y)}{df}$\;
\For{$l=L$ \KwTo $1$\label{backpropagationAlgorithms:outerLoop}}
{
	\For{$\kappa_l=1$ \KwTo $K_l$\label{backpropagationAlgorithms:innerLoop}}
	{
		$g_{k_l[\kappa_l]} = g_{k_l[\kappa_l]} + g_l\frac{df_l(\theta_l,f_{k_l[1]}, .., f_{k_l[K_l]})}{df_{k_l[\kappa_l]}}$\;
	}
	$\phi_l := g_l\frac{df_l(\theta_l,f_{k_l[1]}, .., f_{k_l[K_l]})}{d\theta_l}$\;
}
\caption{The backpropagation algorithm for a single datapoint. Note that the loop on line \ref{backpropagationAlgorithms:innerLoop} can be fully parallelized and the loop on line \ref{backpropagationAlgorithms:outerLoop} can be parallelized to some degree as long as dependency conflicts are avoided. \label{backpropagationAlgorithms}} 
\end{algorithm}

At each iteration of a gradient method, we must compute $\frac{d\ell}{d\theta}$ for a batch of datapoints. This is done by first evaluating each layer in the network whenever all its dependencies have been evaluated. This is known as the `forward pass' or `forward propagation'. Then, $\frac{d\ell}{df_l}$ and ultimately $\frac{d\ell}{d\theta_l}$ is computed for each layer via what is known as the `backpropagation algorithm' given in algorithm \ref{backpropagationAlgorithms}. In that algorithm, the gradient with respect to each layer $f_l$ is computed over time as the in-flowing gradient from each child is added to the current value. When all in-flowing gradients have been added, we can then compute the gradient of $f_l$ with respect to its parents using the chain rule. We can compute the gradient of a layer as soon as the gradients of all its children in the layer graph have been evaluated. Hence, we traverse the layer graph in reverse. This process is called the `backward pass' or `backpropagation'. Crucially, because the layer values from the forward pass are retained, they are available when the Jacobians of the individual layers need to be computed. Note that practical implementations of backpropagation do not compute those Jacobians explicitly, but instead use highly optimized programs that are specific to individual layer operations. 

\subsection{Batching} \label{jointPropagationSection}

At each iteration of a gradient method, we must evaluate $f$ for a batch of inputs in the forward pass. This process is embarrassingly parallel across individual inputs. (An exception to this is batch normalization as described in section \ref{batchWiseLayersSection}, which is parallel across neurons instead.) This opens up the possibility for great computational efficiency. Each layer may be evaluated in parallel across datapoints. Popular layer operations generally have the property that this parallel evaluation is indeed highly efficient. For example, assume we have $f_l = f_kA$ for some matrix $A$. Then computing multiple values of $f_l$ from multiple values of $f_{l-1}$ is tantamount to matrix multiplication, an operation for which there exist highly optimized implementations on CPUs and GPUs. The backward pass can be parallelized in a similar fashion.

We further discuss the desideratum of efficiency in section \ref{programFunctionSection}.

\section{Neural architecture design} \label{statusQuoSection}

In this work, we use the term `neural architecture design' in a broad sense as it is used, to the best of our knowledge, in the community. It encompasses any process that contributes to the choosing of a neural architecture for a task, or of the neural architecture that is eventually deployed. The choice of architecture is specified via what we refer to as the `architecture definition', which is simply all the information required to uniquely specify an architecture. It is the information given to e.g. a functional learning framework like TensorFlow to instantiate the architecture in memory. Under the umbrella of architecture design falls `architecture selection', which is the use of algorithms to choose an architecture automatically. It is simply model selection as described in section \ref{modelSelectionSection} applied to neural networks. Under the umbrella of architecture selection falls `neural architecture search' (NAS) as detailed in section \ref{nasSection}, which is the training of a meta-model that predicts validation error from the architecture definition based on the known validation error values of a set of trained architectures.

If automated design is one side of the coin, the other is manual design. In machine learning, we strive for automation, but there are obvious limitations to this. Even if we automate a given step, we must manually specify that which automates, as we explained in section \ref{hyperparameterTuningSection}. We discuss the limitations of NAS in section \ref{nasSection}. Manual design strategies that are truly general, that transcend individual layer operations, task domains and data types, and that are explanatory in nature, have received far less attention than automated ones, which is epitomized by the fact that there exists no term for this kind of manual design. In section \ref{zeroShotDesignSection}, we introduce the term `zero-shot neural architecture design' (ZSAD), which refers to the process of using general, predictive, explanatory and ideally well-defined principles and a minimal amount of computation to either choose an architecture or choose an algorithm for choosing an architecture, or something in between. We refer to such a principle as a `ZSAD guideline'. Unlike NAS, ZSAD does not conduct any training or rely on the properties of any specific trained network.

In contrast to the term `ZSAD guideline', we use the term `(architecture) design strategy', like `neural architecture design', in a very broad sense for any piece of information that contributes to the choosing of an architecture. It includes anything from the use of neurons, layers and depth to specific layer operations like ReLU or convolution. One of the challenges of this work is that we have to draw an inevitably arbitrary and subjective boundary between the broad design strategy category and the narrow ZSAD guideline category. We hope this distinction will become clear as we go on, helped especially by chapter \ref{surveyChapter} and table \ref{surveySummary}, where we explicitly contrast the two.

Over many decades of neural network research, a core arsenal of design strategies has emerged. We outline this arsenal throughout the remainder of this section. In subsection \ref{layerTypesSection}, we discuss popular layer operations. In subsection \ref{architectureDesignParadigmsSection}, we discuss more high-level strategies which are about configuring and connecting layers that use these operations. The key drivers behind the popularity of these strategies are (i) historical momentum as described in section \ref{historicalBiasSection}, (ii) computational efficiency as described in section \ref{programFunctionSection} and (iii) trial-and-error. A full explanation for why they lead to good performance relative to other strategies that are no longer popular does not exist. However, recently, explanations have begun to emerge in the community, though they have not always been made explicit and precise. In chapter \ref{surveyChapter}, we provide substantial explanations for the performance induced by many of the strategies discussed in this section, and we discuss how they relate to and expand upon existing explanations.

For more information on the basic concepts laid out in this section, see an introductory deep learning text such as \citet{dnnBook}.

\subsection{Layer operations} \label{layerTypesSection}

In this subsection, we introduce and define the layer operations that are most popular, especially in feedforward architectures, and relevant for the remainder of the work. There are a number of minor variations from study to study and from system to system in the way these operations are defined. Here, we present the simplest and most frequently used definitions, which we also use in this work.

As explained in section \ref{neuralNetworksSection}, breaking down the computation of a network into individual neurons and layers is a somewhat arbitrary and subjective process. In fact, there are two competing conventions for this in the deep learning community. The first convention is to refer to an instance of each of the operations defined in this section as a layer. This is the convention we employ in this work. The second convention is to refer to an instance of a group of operations that occur together frequently as a layer. We refer to the instance of such a group as a `macro-layer' as detailed in section \ref{architectureDesignParadigmsSection}. This distinction is also described on page 336 of \citet{dnnBook}.

Layers are named after their operation. For example, a layer that uses the fully-connection operation is a `fully-connected layer' and a layer that uses the activation operation is an `activation layer'.

\paragraph{Fully-connected (FC) operation}

\begin{equation*}
f_l = f_kW_l
\end{equation*}

The fully-connected operation $f_l$ has a single dependency $f_k$ which is multiplied by a dense matrix $W_l$ of size $d_k \times d_l$ called the `weight matrix'. The entries of the weight matrix are called `weights' and correspond to the components of the trainable parameter sub-vector $\theta_l$, which is a vector of dimensionality $d_kd_l$.

\paragraph{Bias operation}

\begin{equation*}
f_l = f_k + \beta_l
\end{equation*}

The bias operation $f_l$ has a single dependency $f_k$ to which a vector $\beta_l$ of dimensionality $d_k$ is added. We must have $d_k = d_l$. $\beta_l$ is called the `bias vector' and its components correspond to the components of $\theta_l$, which is also a vector of dimensionality $d_l$. The bias operation changes slightly in the context of convolutional networks. See section \ref{tensorLayerSection}.

\paragraph{Elementwise multiplication operation}

\begin{equation*}
f_l = \gamma_l . f_k
\end{equation*}

The elementwise multiplication operation $f_l$ has a single dependency $f_k$ to which a vector $\gamma_l$ of dimensionality $d_k$ is multiplied elementwise. We denote elementwise multiplication of $\mathsf{expr1}$ and $\mathsf{expr2}$ by $\mathsf{expr1}.\mathsf{expr2}$. We must have $d_k = d_l$. We call $\gamma_l$ the `scaling vector' and its components correspond to the components of $\theta_l$, which is also a vector of dimensionality $d_l$. The elementwise multiplication operation changes slightly in the context of convolutional networks. See section \ref{tensorLayerSection}.

\paragraph{Activation operation}

\begin{table}
{
\centering
\begin{tabular}{lcccccc}
Act. fun. &ReLU&SELU&tanh&sigmoid&Swish&softplus\\ \hline\hline
Formula&$\max(s,0)$&{\bf \textdagger}&$\tanh(s)$&$\frac{1}{1+e^{-s}}$&$\frac{s}{1+e^{-s}}$&$\log(1+e^{-s})$\\
Illustration&\includegraphics[scale=0.135,valign=c]{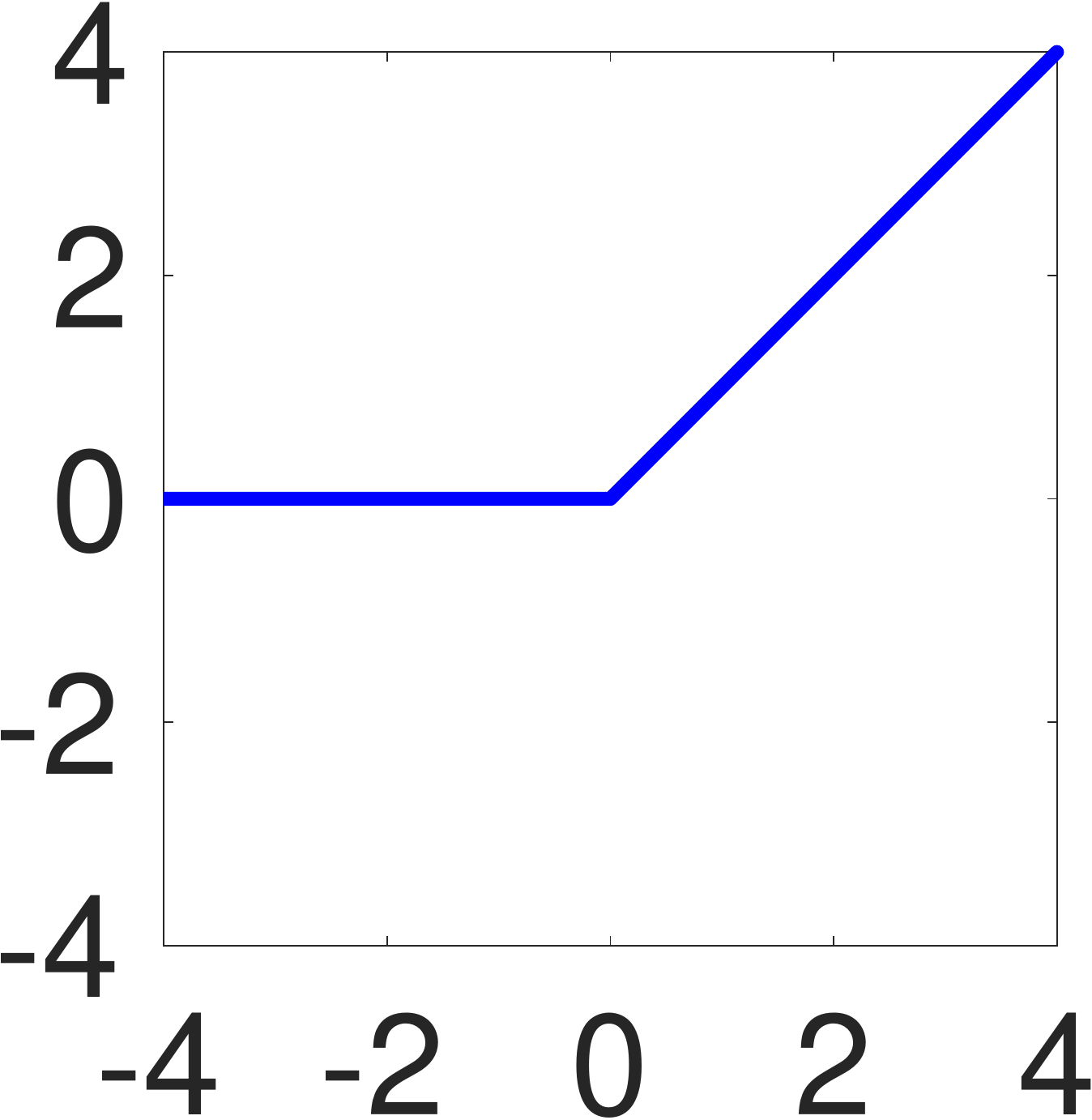}&\includegraphics[scale=0.135,valign=c]{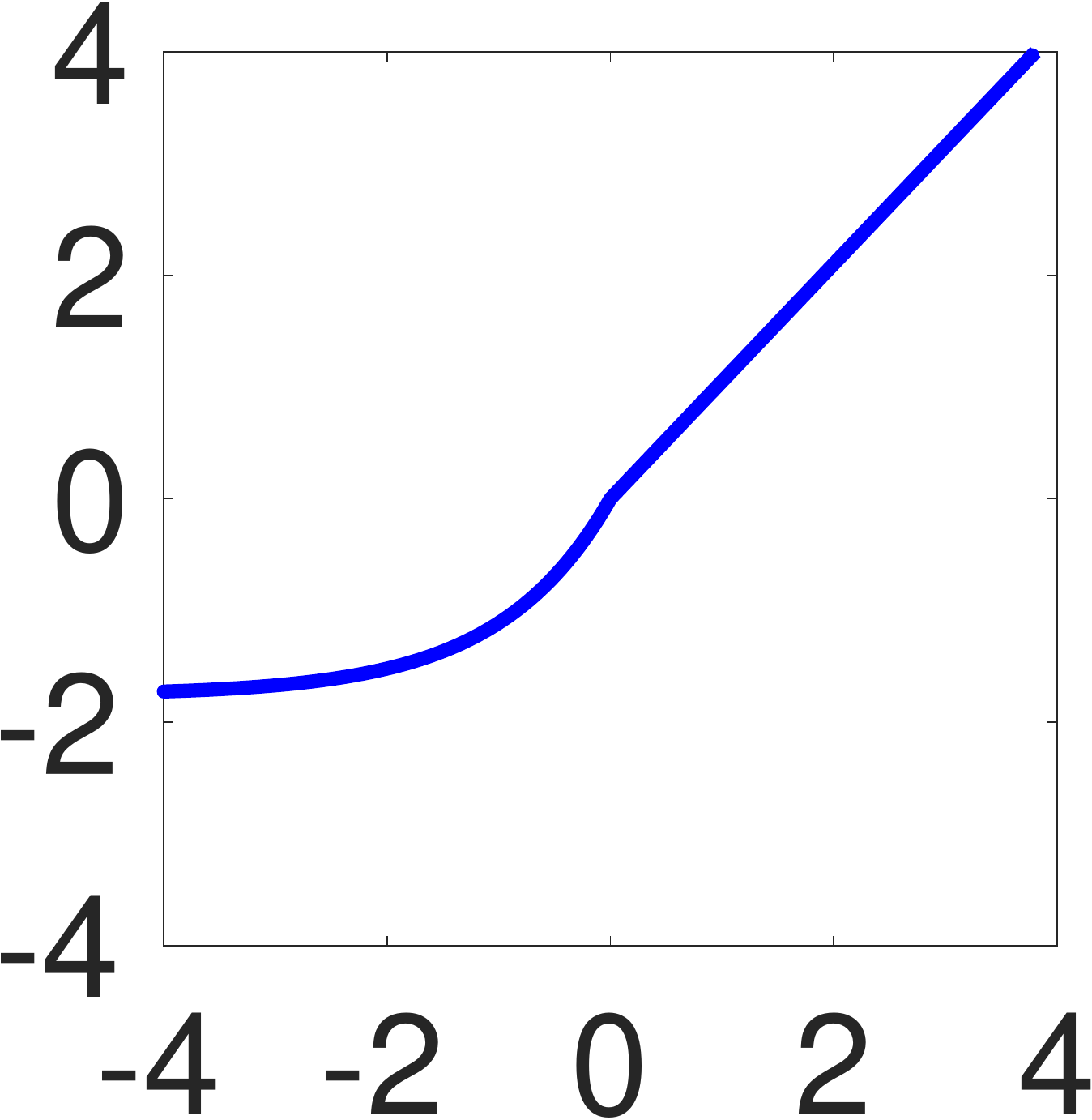}&\includegraphics[scale=0.135,valign=c]{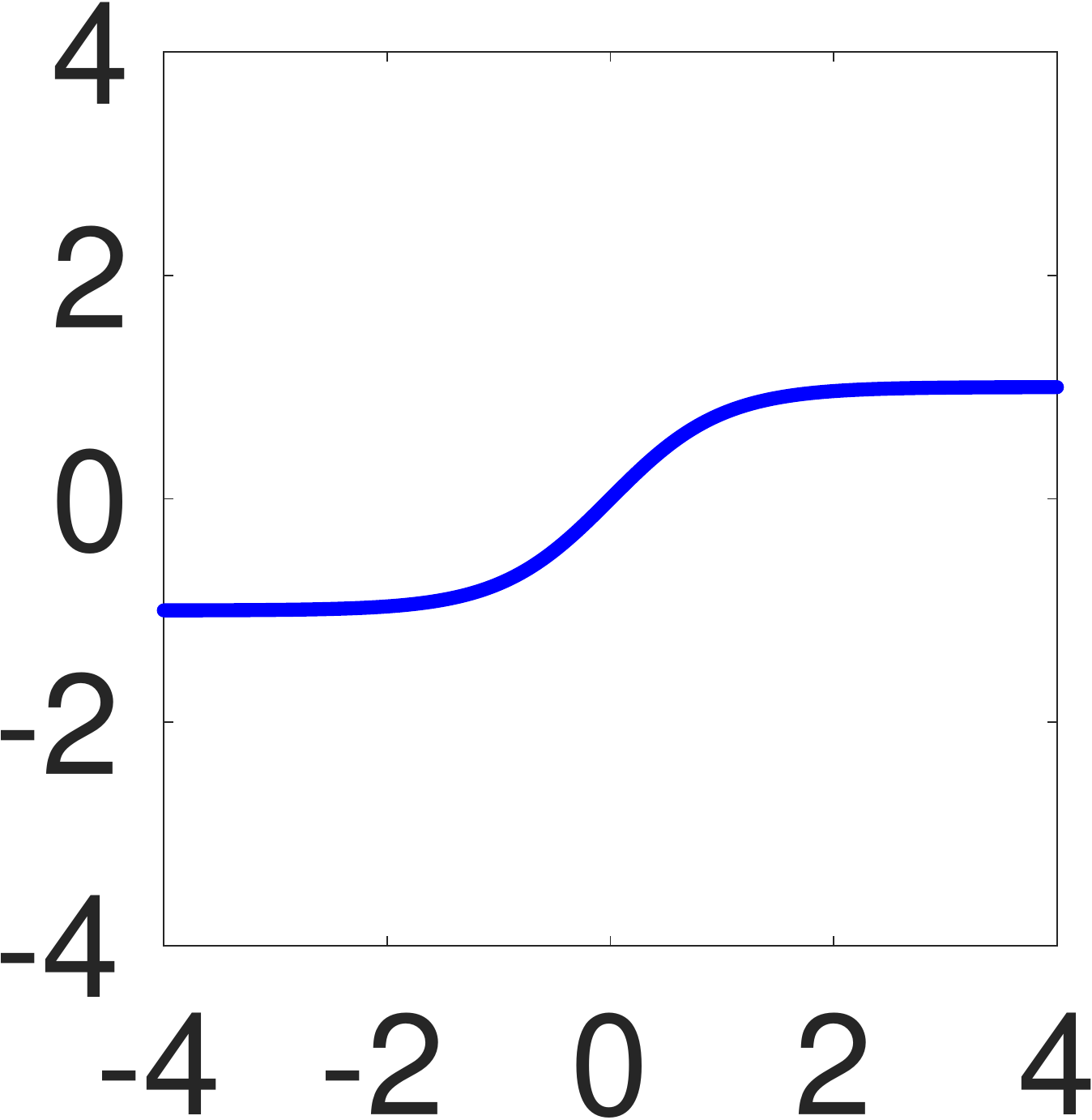}&\includegraphics[scale=0.135,valign=c]{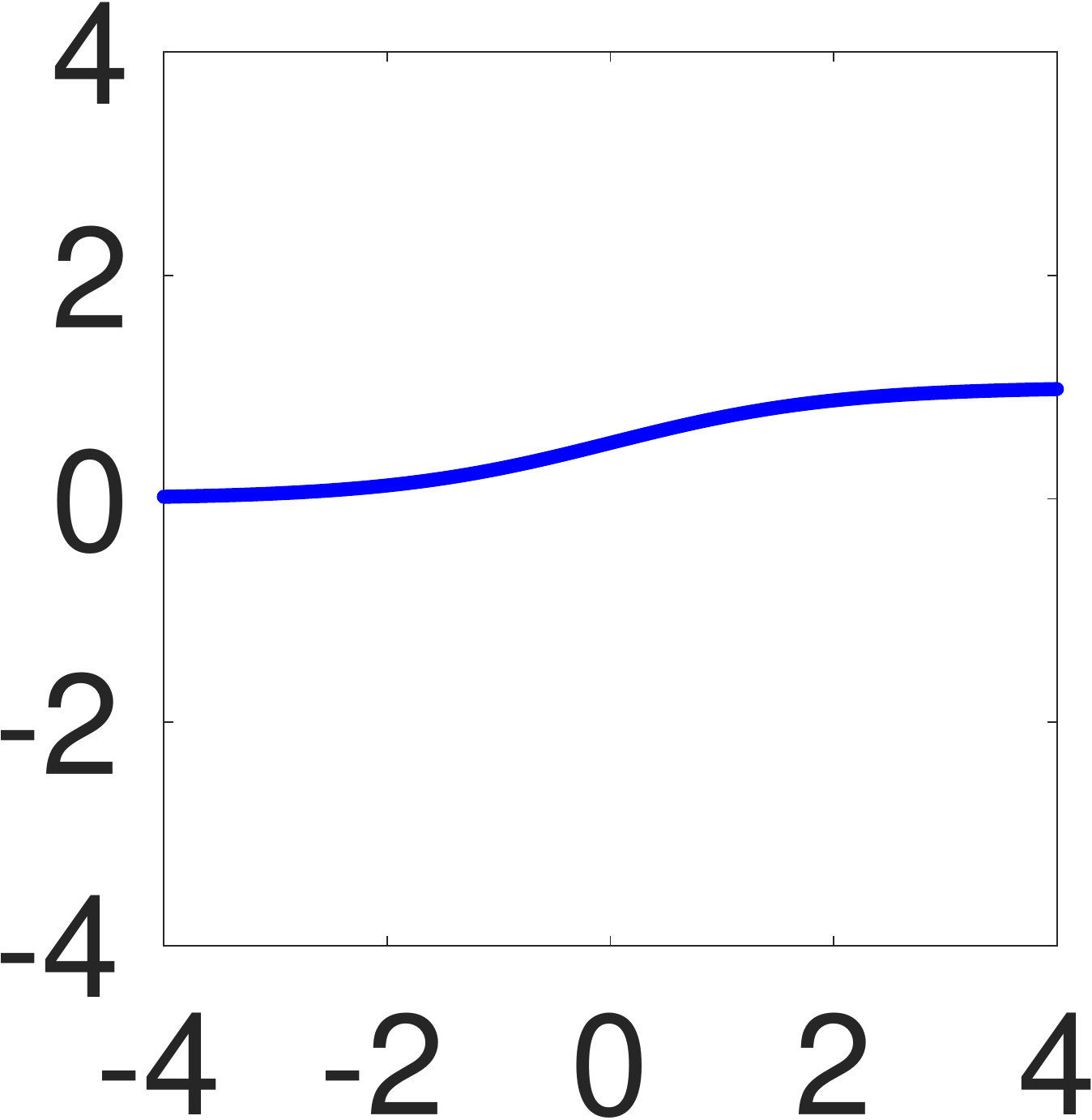}&\includegraphics[scale=0.135,valign=c]{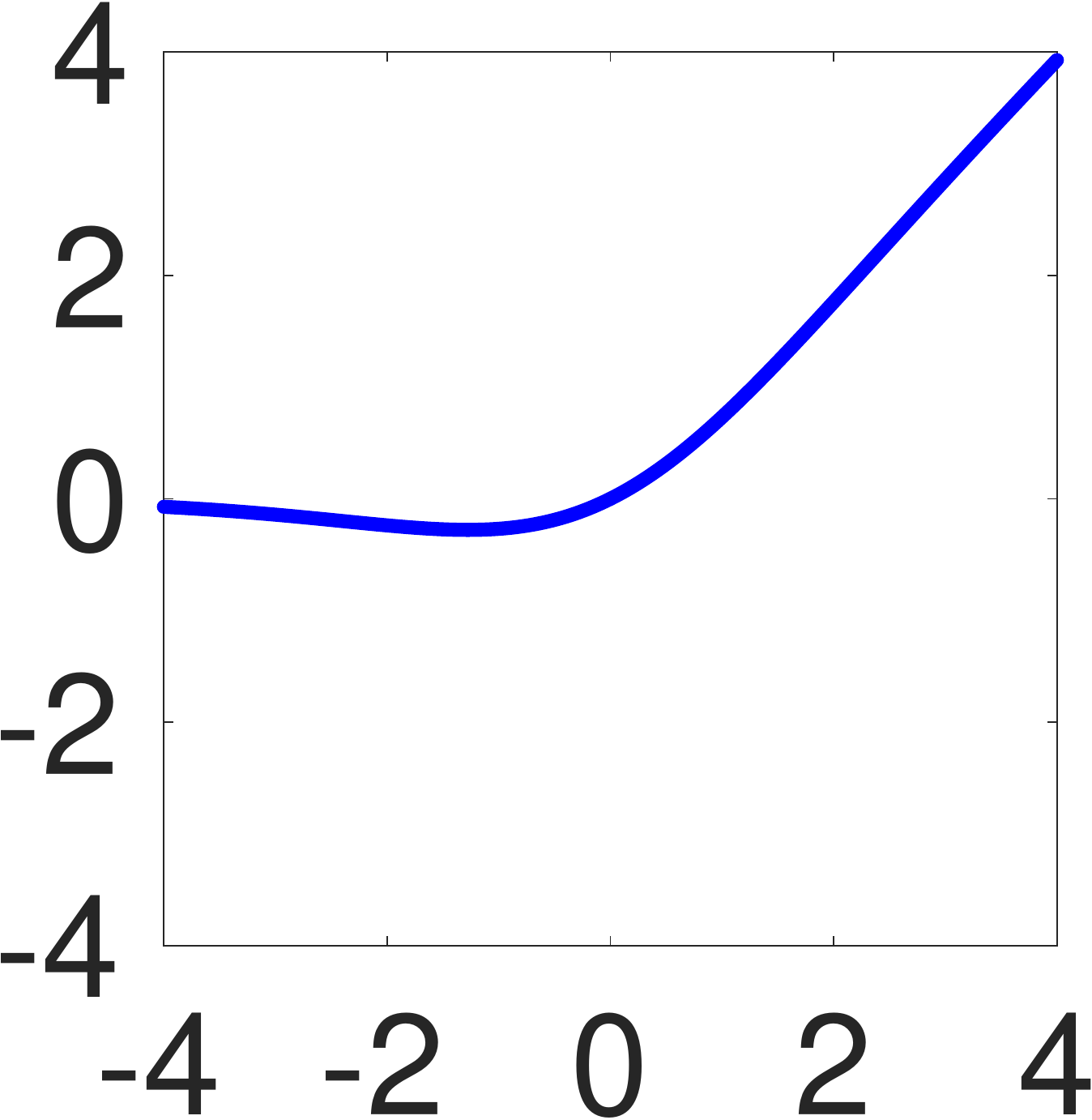}&\includegraphics[scale=0.135,valign=c]{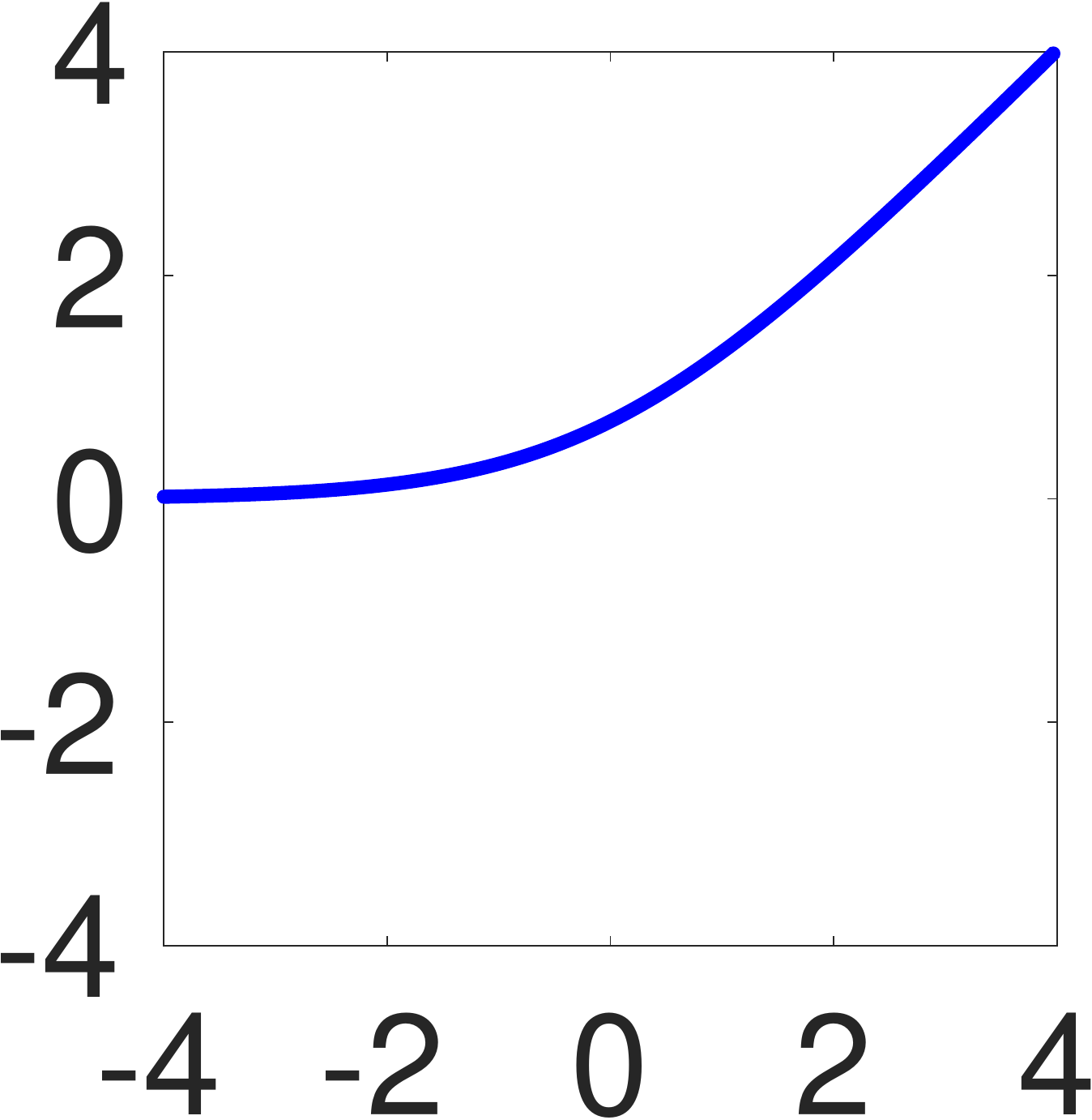}\\
\\
\end{tabular}
\begin{tabular}{lcccccc}
Act. fun.&abs. val.&even tanh&Gaussian&odd square&square&sawtooth\\ \hline\hline
Formula&$|s|$&$|\tanh(s)|$&$\frac{1}{\sqrt{2\pi}}e^{-\frac{s^2}{2}}$&$s*|s|$&$s^2$&\textdaggerdbl\\
Illustration&\includegraphics[scale=0.135,valign=c]{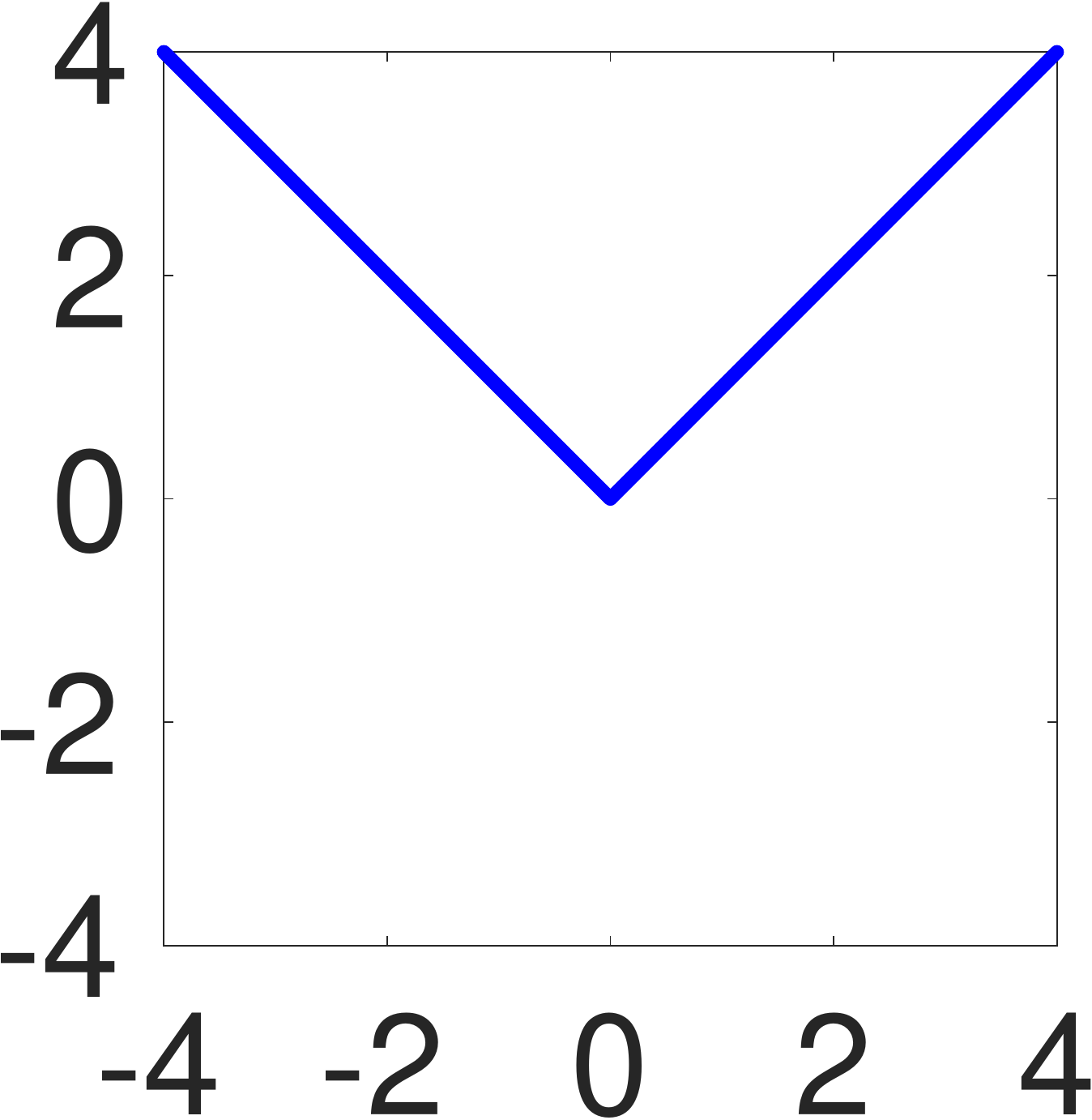}&\includegraphics[scale=0.135,valign=c]{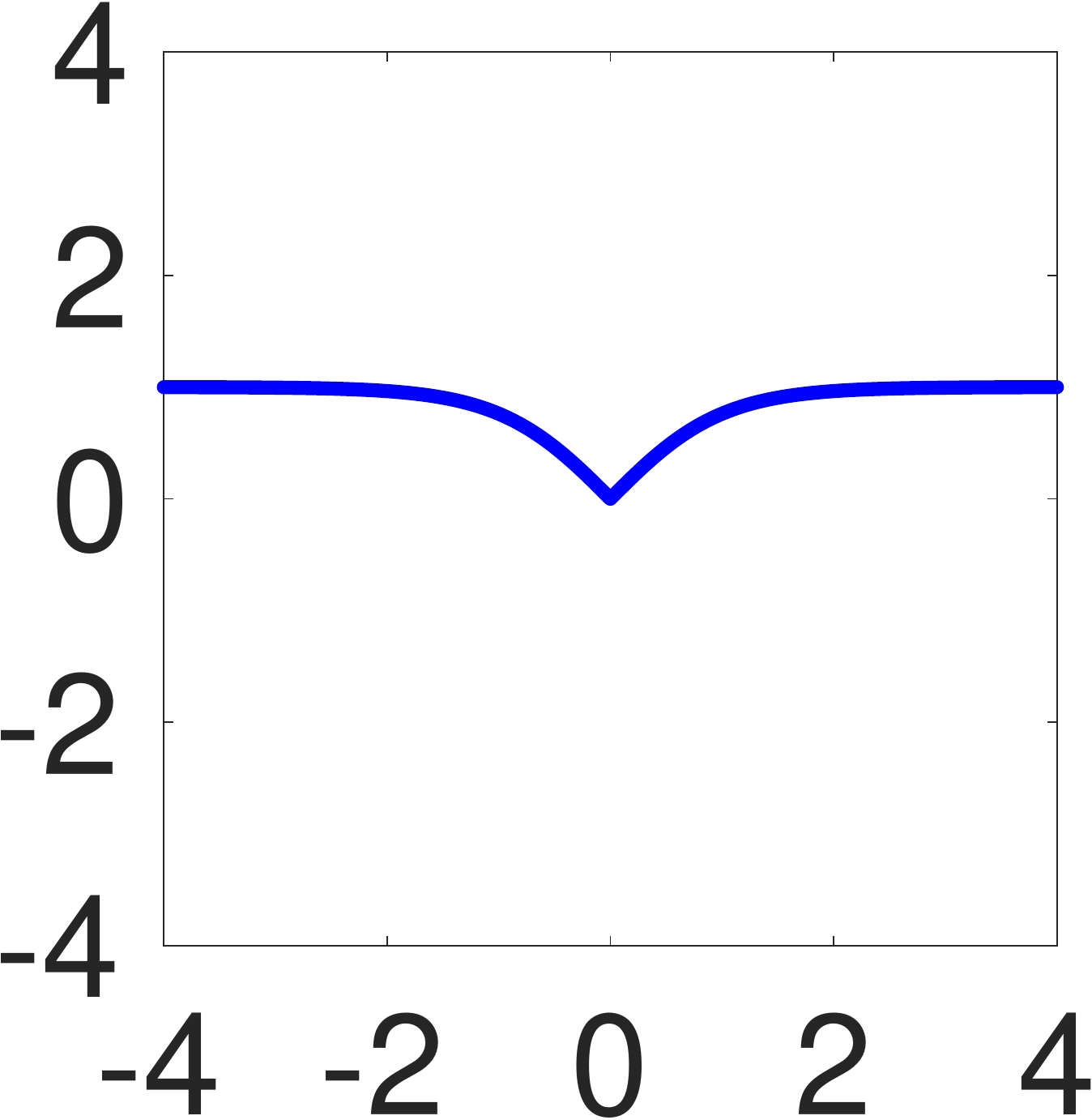}&\includegraphics[scale=0.135,valign=c]{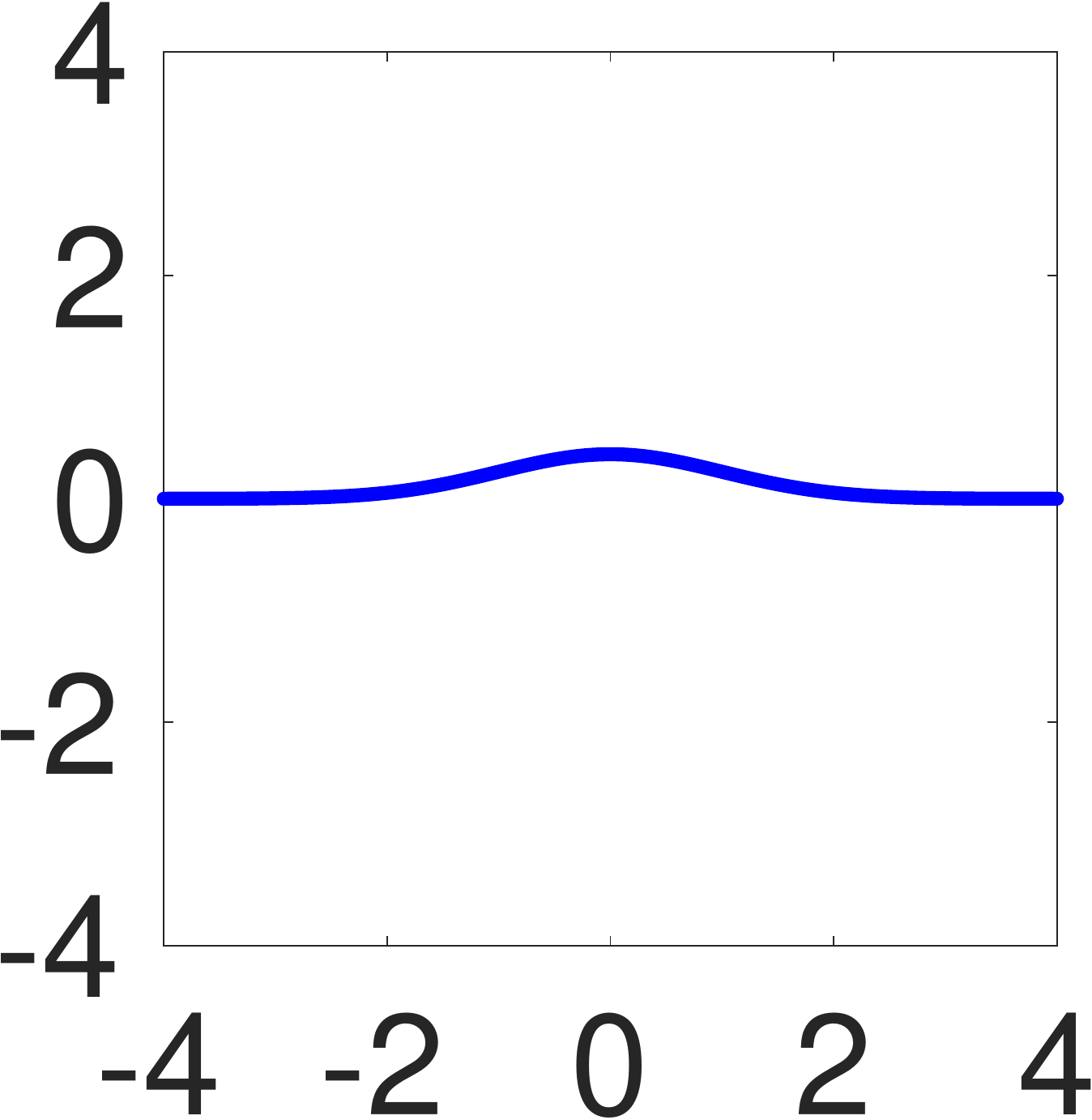}&\includegraphics[scale=0.135,valign=c]{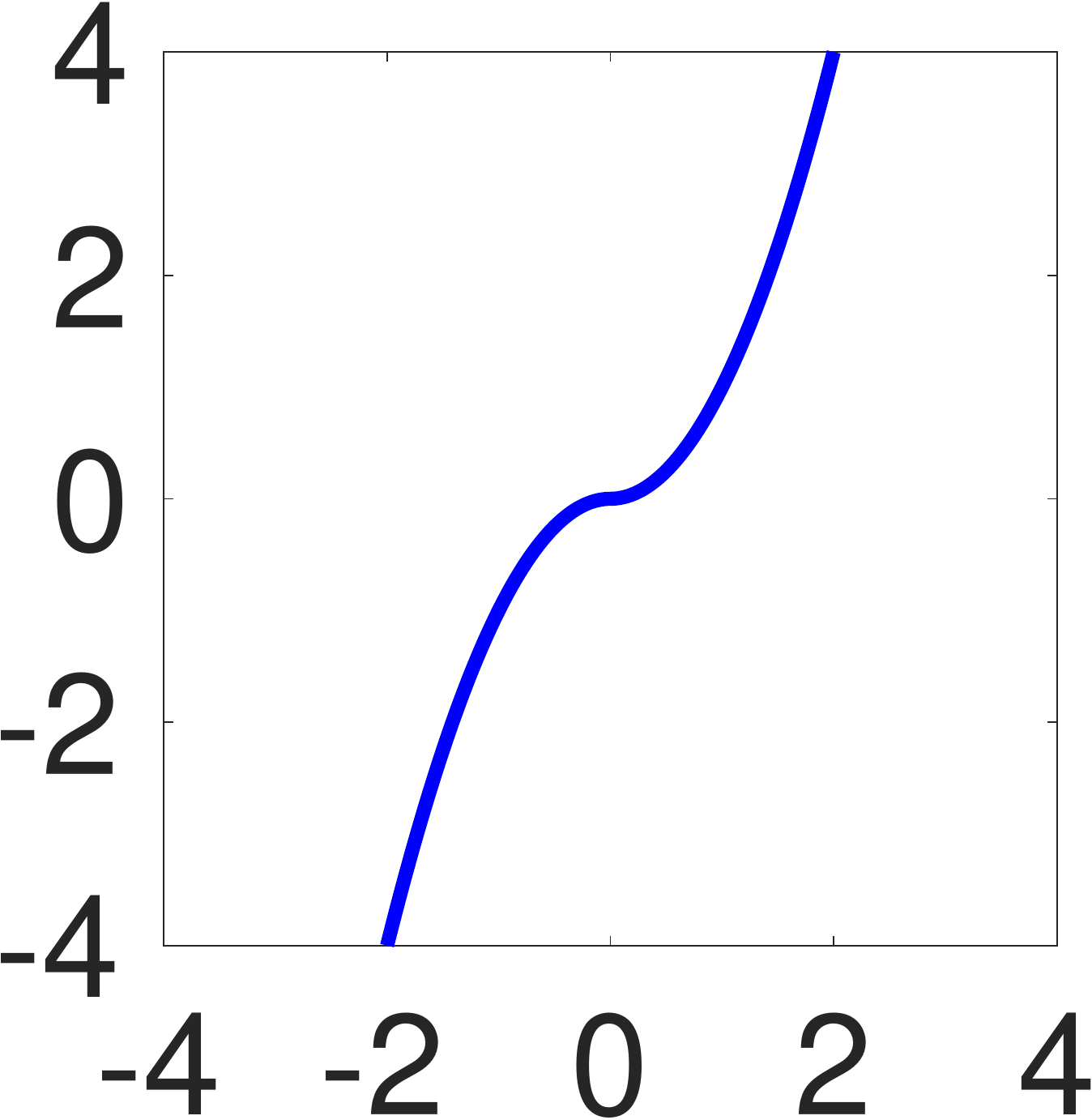}&\includegraphics[scale=0.135,valign=c]{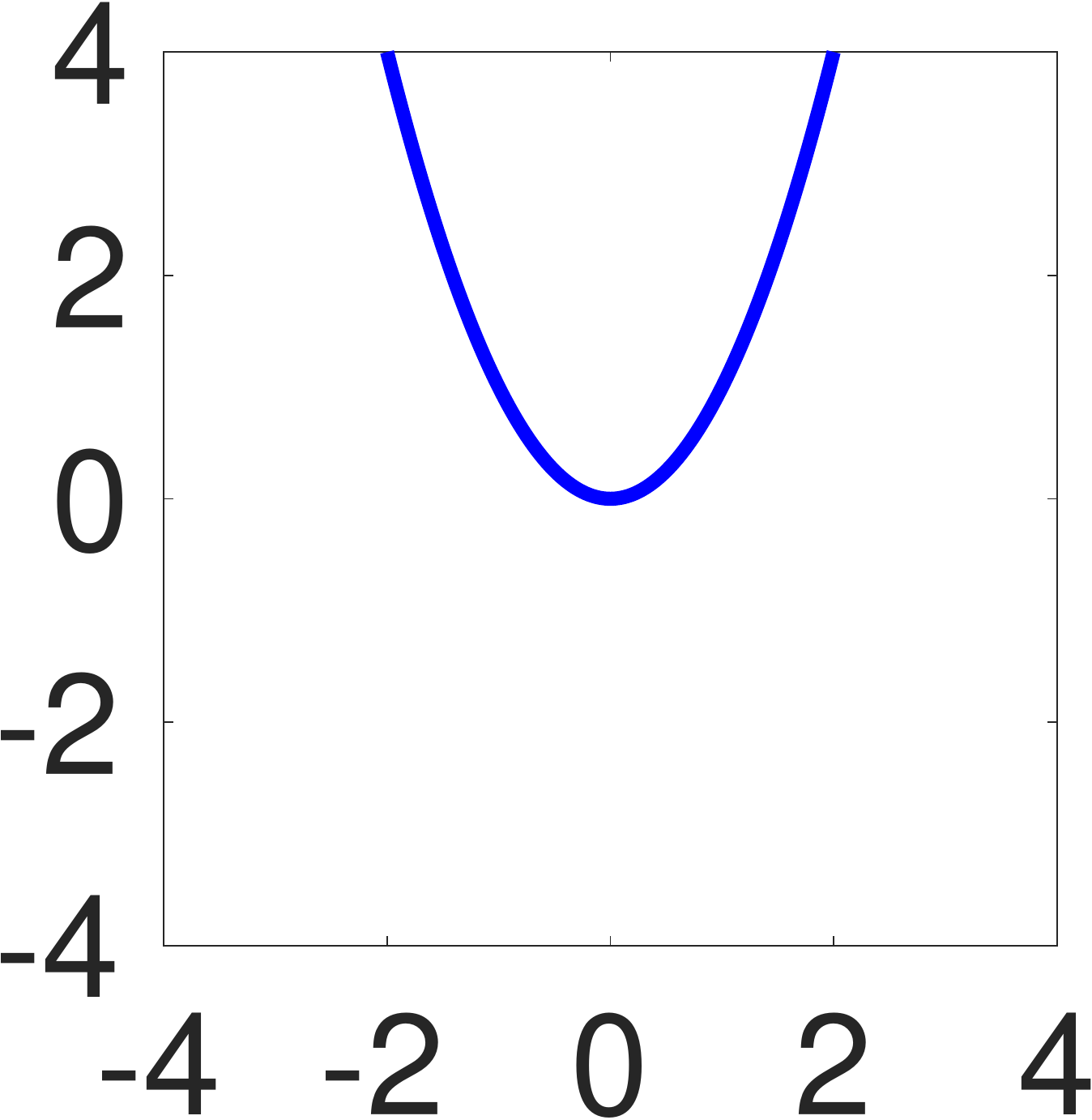}&\includegraphics[scale=0.135,valign=c]{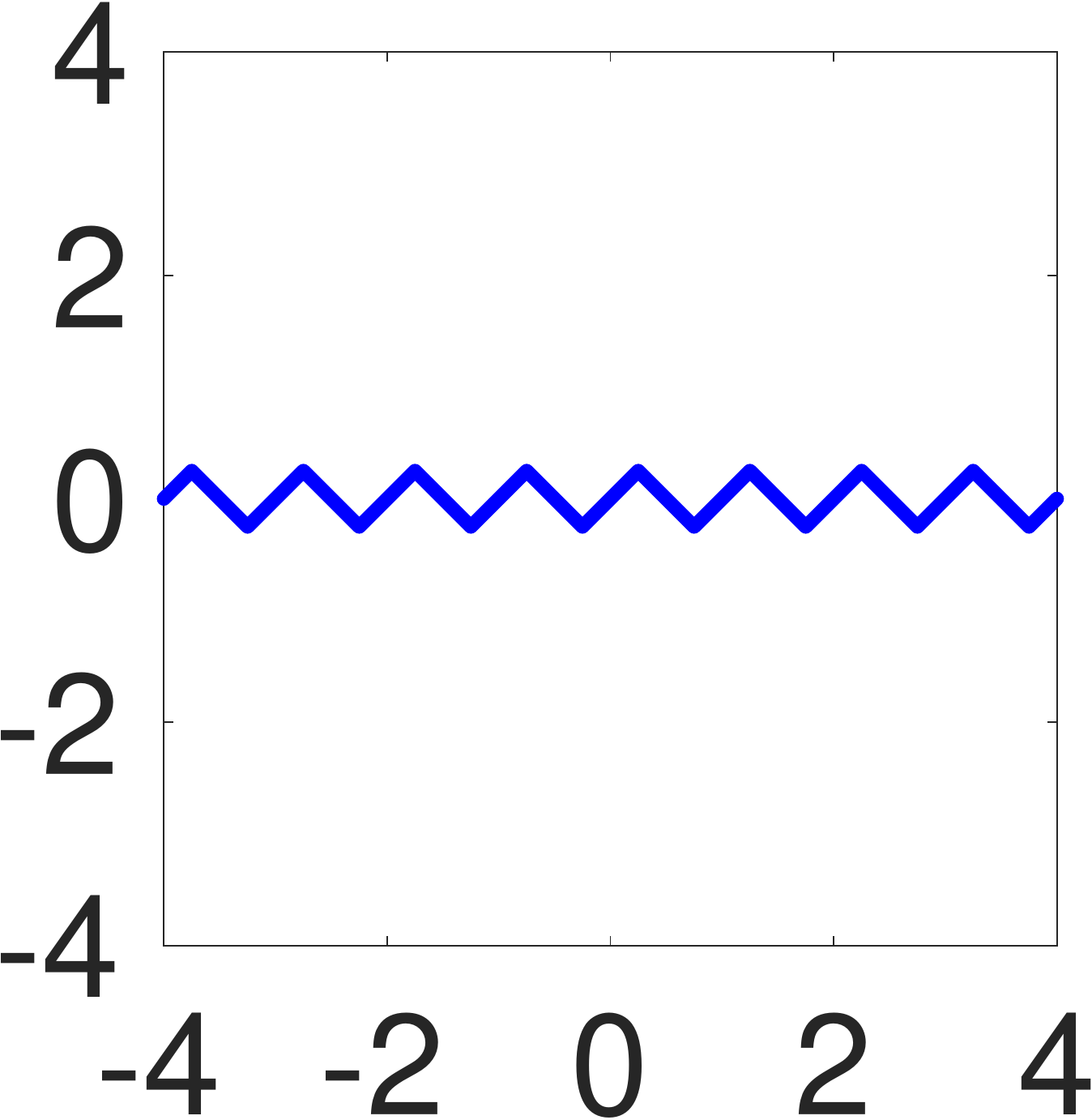}\\
\end{tabular}
\\
}
{\scriptsize \textdagger$\mathbbm{1}_{s > 0}1.0507s + \mathbbm{1}_{s < 0}1.75814(e^s-1)$\\
\textdaggerdbl $\tau(s) = s - \lfloor s\rfloor$ if $s - \lfloor s\rfloor < 0.25$; $\tau(s) = s - \lfloor s\rfloor - 1$ if $s - \lfloor s\rfloor > 0.75$; $\tau(s) = 0.5 - s + \lfloor s\rfloor$ else}
\caption{Activation functions used in this work. Top row: popular activation functions. Bottom row: activation functions created ad hoc for this work.}
\label{actFunIllu}
\end{table}

\begin{equation*}
f_l = \tau_l.(f_k)
\end{equation*}

$\tau_l : \mathbb{R} \rightarrow \mathbb{R}$ is applied elementwise to the single dependency, as denoted by $.()$\ . $\tau$ is called the `activation function' or the `nonlinearity'. We will use exclusively the former term to avoid confusion with the concept of `degree of nonlinearity' that is a core subject of this work. Usually, activation operations have no trainable parameter sub-vector, i.e. $\theta_l$ is empty, i.e. it has dimensionality zero. This holds throughout this work. We generally denote a scalar input to an activation function by $s$.

By far the most popular activation function today is the rectified linear unit (ReLU; \citet{relu}). It and other popular activation functions are given at the top of table \ref{actFunIllu}. SELU was introduced by \citet{selu} and became subsequently popular \citep{seluCNN,seluGAN,seluRL,seluAuto,seluMLP}. Swish was ``found'' through automated search by \citet{swish}. Sigmoid and its variant tanh are the ``original'' activation functions as discussed in section \ref{historySection}.

Activation functions that we created in an ``ad hoc'' fashion specifically for study in this work are given at the bottom of table \ref{actFunIllu}.

\paragraph{Layer normalization (LN) operation \citep{layerNormalization}}

\begin{equation*}
f_l = \frac{f_k - \mathbb{E}_{i_k} f_k[i_k]}{\sqrt{\mathbb{E}_{i_k} f_k[i_k]^2 - (\mathbb{E}_{i_k} f_k[i_k])^2 + \epsilon_l}}
\end{equation*}

The single dependency is normalized by the mean and standard deviation of its components. We must have $d_l = d_k$. $\theta_l$ is empty, though layer normalization is often composed with an elementwise multiplication and / or bias operation to compensate for this fact. $\epsilon_l$ is a small fixed regularizer that is used to ensure $f_l$ is well-defined for all inputs. In practice, the square root in the denominator of LN is virtually never zero, so we can omit $\epsilon_l$ in most implementations of LN. To maintain mathematical consistency, we can pretend $\epsilon_l$ is so small that its floating-point representation is zero.

A `normalization operation' refers to one of several operations that normalize their dependency. In this work, we focus on batch normalization (see below) and layer normalization, which are the most popular and second most popular normalization operations respectively.

\paragraph{Addition operation}

\begin{equation*}
f_l = \sum_{\kappa_l=1}^{K_l} w_{l,\kappa_l}f_{k_l[\kappa_l]}
\end{equation*}

A weighted sum over the dependencies is taken. The scalars $w_{l,\kappa_l}$ can either be considered fixed constants or components of $\theta_l$. Usually, they are fixed to be 1. We must have $d_l = d_{k_l[\kappa_l]}$ for all $1 \le \kappa_l \le K_l$. We often shorten $w_{l,\kappa_l}$ to $w_{\kappa_l}$.

\subsubsection{Batch-wise layer operations} \label{batchWiseLayersSection}

As described in section \ref{jointPropagationSection}, the nature of gradient methods and computational realities encourage the joint forward propagation of inputs and the joint backpropagation of the corresponding gradients. This fact can be harnessed for architecture design. Specifically, there are layer operations whose value $f_l(x)$ for some input $x$ does not only depend on the value of their dependencies $f_{k_l[\kappa_l]}(x)$ for the same input $x$, but also on the values of $f_{k_l[\kappa_l]}$ for inputs that are forward propagated jointly with $x$. However, there is only one popular such operation, and we detail it below.

\paragraph{Batch normalization (BN) operation \citep{batchNormalization}}

\begin{equation*}
f_l[i_l] = \frac{f_k[i_l] - \mathbb{E}_b f_k[i_l]}{\sqrt{\mathbb{E}_b f_k[i_l]^2 - (\mathbb{E}_b f_k[i_l])^2 + \epsilon_l}}
\end{equation*}

As in layer normalization, we normalize the single dependency. However, instead of using the mean and standard deviation over components of $f_k$, we normalize each individual component with its mean and standard deviation over inputs that are jointly propagated. $\mathbb{E}_b$ denotes the mean over those inputs. We must have $d_l = d_k$. $\theta_l$ is empty, though BN is often composed with an elementwise multiplication and / or bias operation to compensate. As in LN, $\epsilon_l$ is a small fixed regularizer that is used to ensure that $f_l$ is valid for all inputs. As for LN, it is usually set to zero in practice, and in theory we can set it small enough that its floating-point representation is zero. The BN operation changes slightly in the context of convolutional networks. See section \ref{tensorLayerSection}.

What is to be gained from introducing a cross-input dependency? Like with other layer operations listed in this subsection, networks that employ BN have exhibited high performance in the past, which has led to BN becoming popular. There exists no widely proven strategy to replicate all aspects of this high performance without cross-input dependencies. Like with other operations, a full explanation for its performance does not exist, but we provide a substantial explanation in chapter \ref{surveyChapter}.

When using BN, it is important to decide which inputs propagate jointly. When using stochastic gradient methods, there is an opportunity to jointly propagate inputs in the batch $B^{(t)}$ at each iteration $t$. This is generally done in practice if this joint propagation is not too resource-intensive. If the batch $B^{(t)}$ is very large or computational devices used for training are weak, the batch is split up into slices which are forward-propagated either on separate devices or sequentially on the same device. The mean and standard deviation for BN are either taken independently for each slice or a single mean and standard deviation is taken across slices by communicating the required statistics across devices. In this work, we always propagate all inputs in a single batch jointly, and hence we use the term `batch' interchangeably for (i) a set of datapoints or inputs considered at an iteration of a gradient method, (ii) a set of datapoints or inputs considered for the moments of batch normalization and (iii) a set of datapoints or inputs that are jointly propagated on a single device along with their corresponding gradients.

As the batch size increases, the mean and standard deviation over the batch converges to the mean and standard deviation over the entire training set. Therefore, when a trained network is deployed in practice, the batch moments are simply replaced by the analogous moments taken over the entire training set, thereby eliminating the dependence of the prediction on the batch. Since taking the mean and standard deviation over the entire training set is expensive, it cannot feasibly be done at each iteration of stochastic gradient-based training. From this point of view, it appears as if the batch moments are simply noisy proxies for the training set moments. However, it turns out that this noise can actually be beneficial for training as long as it is not too large. We further analyze this point in section \ref{noiseStabilitySection}, though fully explaining the benefits of this noise goes beyond the scope of this work.

Even if we decide to use the batch for taking moments during training and the training set for taking moments when deploying the network, this still leaves open the question of how to take moments in all other contexts, and many different contexts appear throughout this work. For example, consider a statement like ``in network $f$, we observe that the first and second derivative have similar magnitude''. If $f$ contains BN, the question always arises whether batch or training set moments or different moments are used. Further, if batch moments are used, how are batches constructed? Throughout this work, we adopt a uniform policy of always using batch moments, never training set moments, for all quantities and metrics, even test error. For each architecture, the same batch size is used whenever that architecture is evaluated, and batches are always drawn uniformly at random without replacement. Because the training stage is the most important stage of our (simple) machine learning pipelines, we decided that it is best to have all of our analysis reflect the behavior of the network in that critical stage. We also avoid the additional complexity of having to analyze what are essentially two different networks: one batch-dependent and one batch-independent.

The presence of BN is an unfortunate reality that forces us into one of two bad options. The first option is to write this work with the implicit assumption that a neural network is a function that maps single inputs $x$ to single outputs. If we go down this route, this work is not immediately applicable to architectures with BN. The second option is to always consider the potential dependency of the output of $f$ on the batch explicitly. This would make the work unreadable while conferring minimal practical benefit, as the generalization of our methods and analysis to the BN case is almost always trivial and boring, as becomes clear e.g. in section \ref{metricBNsection}. We choose the former option and present our work in terms of single input-single output networks. The incorporation of BN is done ``under the hood'' and is rarely brought up explicitly, though we do cover the BN case in detail in sections \ref{metricBNsection}, \ref{nlcComputeSection} and \ref{meanFieldBNsection}.

\subsubsection{Tensor operations} \label{tensorLayerSection}

In many practical architectures, the majority of layers, including the input layer, are specified as multi-dimensional tensors. This enables us to more easily specify certain operations that are naturally defined in terms of tensors. If a layer is specified as a tensor, we say it has $T_l$ dimensions and its size is $C_l \times S_{l, 1} \times S_{l, 2} \times .. \times S_{l, T_l - 1}$. We write $f_l[c_l,s_{l,1},s_{l,2}, .., s_{l,T-1}]$ for entries in that tensor where $0 \le c_l < C_l$ and $0 \le s_{l,t_l} < S_{l,t_l}$ for all $1 \le t_l < T_l$. Each $T-1$-dimensional slice of $f_l$ defined by a value of $c_l$ is called a `channel' and each 1-dimensional slice of $f_l$ defined by a tuple $(s_1,..,s_{T_l-1})$ is called a `spatial location'. The first dimension of $f_l$ is called the `channel dimension' and the other dimensions are called `spatial dimensions'.

Of course, networks with tensor layers are still equivalent to networks with vector layers as defined in section \ref{neuralNetworkNotationSection}, as we can simply re-specify all operations in terms of vectors, though this may be much less compact. Throughout this work, we almost always use vector notation, but by doing so we simultaneously cover the tensor case. For detailed information about the usage of and motivation behind tensor layers and operations, see chapter 9 of \citet{dnnBook}.

\paragraph{Convolutional operation}

\begin{eqnarray*}
&&f_l[c_l, s_{l,1}, s_{l,2}, .., s_{l,T_l-1}]\\
&=& \sum_{c_k,h_{l,1},h_{l,2},..,h_{l,T_l-1}} f_k[c_k,s_{l,1}+h_{l,1}-o_{l,1},s_{l,2}+h_{l,2}-o_{l,2},.., s_{l,T_l-1}+h_{l,T_l-1}-o_{l,T_l-1}]\\
&&*W_l[c_k,c_l,h_{l,1},h_{l,2},..,h_{l,T_l-1}]
\end{eqnarray*}

The convolutional operation $f_l$ has a single dependency $f_k$, which is convolved with the tensor $W_l$, known as the `weight tensor', as given in the formula above. In that formula, $c_l$, $c_k$ and the $s_{l,t_l}$ have canonical ranges. We must have $T_l = T_k$. $W_l$ has $T_l+1$ dimensions and size $C_k \times C_l \times H_{l,1} \times H_{l,2} \times .. \times H_{l,T_l-1}$. The entries of $W_l$ are called `weights' and correspond to the components of $\theta_l$. $h_{l,t_l}$ ranges from 0 to $H_{l,t_l}-1$. The `offsets' $o_{l,t_l}$ are fixed constants. In practice, we almost always have $H_{l,t_l}$ odd and $2o_{l,t_l} + 1 = H_{l,t_l}$. Since \citet{vggNet}, most architectures use $H_{l,t_l} = 3$ or $H_{l,t_l} = 1$ for the majority of $(l, t_l)$ pairs.

For the above formula to be well-defined, we need to have $o_{l,t_l} \le 0$ and $H_{l,t_l} + S_{l,t_l} - o_{l,t_l} - 1 \le S_{k,t_l}$. This is generally not true in practice. Therefore, we need to handle the case where $f_k$ is indexed at a position that is out of range. The most popular convention for this is called `zero padding', and we use it in this work. When the index $s_{l,t_l} + h_{l,t_l} - o_{l,t_l}$ is out of range, then $f_k[..,s_{l,t_l} + h_{l,t_l} - o_{l,t_l},..]$ is replaced by zero in the above formula.

Entries of a weight tensor $W_l$ are known as weights, just like weight matrix entries. A `linear layer' refers to a layer that either uses the fully-connected operation, the convolutional operation or a similar linear operation that is not covered in this work. Note that linear layers that have multiple dependencies are sometimes used in practice. However, they can be broken down into single-dependency linear layers and an addition layer. Hence, despite not explicitly considering multi-dependency linear layers, we do not lose representational power, but we do simplify notation throughout the work.

\paragraph{Subsampling operation}

\begin{equation*}
f_l[c_l, s_{l,1}, s_{l,2}, .., s_{l,T_l-1}]= f_k[c_l,r_{l,1}s_{l,1}, r_{l,2}s_{l,2}, .., r_{l,T_l-1}s_{l,T_l-1}]
\end{equation*}

Each spatial dimension $t_l$ of the single dependency is sub-sampled with `stride' $r_{l,t_l}$. We must have $T_k = T_l$ and $S_{k,t_l} = r_{l,t_l}S_{l,t_l} $ for all $1 \le t_l \le T_l - 1$. $\theta_l$ is empty.

\paragraph{Pooling operation}

\begin{eqnarray*}
&&f_l[c_l, s_{l,1}, s_{l,2}, .., s_{l,T_l-1}]\\
&=& p_l(\{f_k[c_l,s_{k,1}, s_{k,2}, .., s_{k,T_l-1}] \text{ : } r_{l,t_l}s_{l,t_l} \le s_{k,t_l}  < r_{l,t_l}(s_{l,t_l}+1)  \})
\end{eqnarray*}

The single dependency is broken up into sub-tensors along its spatial dimensions with strides $r_{l,t_l}$. Each sub-tensor is then replaced by a single-entry tensor according to the `pooling function' $p_l$ that takes in all entries in that sub-tensor. The most popular pooling functions are `average pooling', which takes the average value over the sub-tensor, and `max pooling', which takes the maximum value over it. The pooling operation is a generalization of the subsampling operation. Usually, $\theta_l$ is empty. `Global pooling' refers to the choice $r_{l,t_l}=S_{k,t_l}$ for all $t_l$, which induces $S_{l,t_l}=1$.

\paragraph{Miscellaneous channel-wise operations}

If a layer is specified as a tensor, some of the operations described earlier change their definition slightly. Bias vectors then have $C_l$ elements, and the same component is added to all entries of $f_k$ that are in the same channel. The same goes for scaling vectors. For BN, the mean and standard deviation are taken over all entries in the same channel as well as inputs in the batch. These modified operations are similar to the convolutional operation in that they apply the same transformation at each spatial location. This is key for attaining high performance in situations where tensor layers are used. Throughout this work, we use these modified operations in all our architectures that utilize tensor layers.

\subsection{High-level design strategies} \label{architectureDesignParadigmsSection}

In this section, we explain some of the most popular and ubiquitous high-level design strategies that are also relevant for the remainder of the work.

\paragraph{Layers and width} The layer concept itself can be regarded as the most important architecture design strategy. Neural networks are firstly gradient-trainable functions and secondly neuron graphs. Additionally having a layer graph is not a conceptual necessity. In section \ref{layersSection}, we outlined the clear computational and representational benefits of layers. It is important to note that fully attaining those benefits requires that, in a layer operation, every neuron has an identical or near-identical operation. Technically, we could define layer operations where each neuron is completely different. When we say that a network ``uses layers'', we implicitly assume neuron operation homogeneity. The use of layers ties in with the ZSAD guideline of ``using an appropriate width'', as the concept of width itself is defined in terms of layers. See section \ref{widthSection}. Mean field theory, which we cover in chapter \ref{meanFieldNnaChapter}, is also grounded in the layer concept.

\paragraph{Depth} The concept of depth was detailed in section \ref{neuronDepthSection}. It refers to the utilization of long neuron or layer dependency chains. The property of depth refers to the length of the longest dependency chain. As we explained, the length of that chain as well as what is considered ``particularly deep'' is subjective and situational. It is a popular view that the recent success of deep learning is based to a significant degree on ``going deeper'', i.e. on using longer and longer dependency chains. The use of deep networks gives rise to the ZSAD guideline of ``using an appropriate depth''. See section \ref{depthSection}.

\paragraph{Macro-layers} Layers performing certain operations tend to occur as a group, and popular architectures are often built by repeatedly composing the same group. We refer to such a group as a `macro-layer'. Macro-layers usually contain at least a linear layer and an activation layer. The linear layer provides the parameter sub-vector which can be trained efficiently using gradient methods. The activation layer makes the network nonlinear, a desideratum which we explore in detail in this work. Beyond this two-layer group, a more complex and even more popular configuration consists of a linear layer, a normalization layer, an elementwise multiplication layer, a bias layer and an activation layer. While some macro-layers have multiple activation layers utilizing different activation functions (e.g. LSTM \citep{LSTM}), activation functions generally don't vary between macro-layers unless the architecture is highly bespoke.

In section \ref{neuralNetworksSection}, we explained how the neural network property of depth is subjective because the neuron and layer concepts are subjective. The most popular convention for measuring depth is to measure the length of the longest directed path in the ``macro-layer graph''. We will use this convention for the remainder of this work and denote this depth value by $M$. Of course, this convention can only apply if networks are meaningfully specified in terms of macro-layers, as we do for the architectures we study empirically in chapter \ref{empiricalStudiesChapter}. This is not possible for all architectures.

\paragraph{Random initialization} The initial value of weight matrices and tensors is generally drawn from a distribution. Most often, each individual weight is drawn independently from a distribution with mean zero and a variance that is equal for all weights in the same layer but not necessarily for weights in different layers. The variance of these distributions, which we term `initial weight variance', is often related to dimensionality. For fully-connected layers, the most popular variance value is $\frac{1}{d_k}$. For convolutional layers, the most popular variance value is $(C_k\prod_{t_l=1}^{T_l-1}H_{l,t_l})^{-1}$. In both cases, the variance is 1 over the number of multiply-adds that contribute to a neuron in the linear layer. The strategy of using this initial weight variance is sometimes termed {\bf LeCun initialization} \citep{1998LeCunTricks} and sometimes Xavier or Glorot initialization. The latter two names, however, are technically inaccurate as we explain in section \ref{vanishWidthSection}. We refer to the variance value itself as `LeCun variance'.

The distribution of each weight is usually either Gaussian or uniform. This is called `Gaussian initialization' or `uniform initialization' respectively. Note that attaining some specific variance $\sigma^2$ requires the uniform distribution to have support $[-\sqrt{3}\sigma, \sqrt{3}\sigma]$, {\it not} $[-\sigma, \sigma]$. Inexplicably, for example, PyTorch uses uniform initialization with support $[-\frac{1}{d_k}, \frac{1}{d_k}]$ for weight matrices by default. We are not aware of any studies demonstrating the superiority of either uniform or Gaussian initialization over the other.

Weight matrices are also often initialized as scalar multiples of sub-matrices of $\max(d_k,d_l) \times \max(d_k,d_l)$ uniformly random orthogonal matrices. This is called {\bf orthogonal initialization}. The scalar multiple is chosen to control the initial weight variance, which now refers to the variance of the marginal distribution of each weight. Again, the most popular value is the LeCun variance. Orthogonal initialization is related to the ZSAD guideline of `orthogonality' as discussed in section \ref{orthogonalitySection}.

\paragraph{Bias vector initialization} Components of bias vectors are generally initialized to zero. This causes bias layers to correspond to the identity operation in the initial state.

\paragraph{Scaling vector initialization} Components of scaling vectors are generally initialized to 1. This causes elementwise multiplication layers to correspond to the identity operation in the initial state.

\paragraph{Scale stability} The reason for the popularity of the LeCun initialization as described above is that it causes the linear layer to approximately preserve the overall magnitude of neuron values in the initial state. For example, assume that for some fully-connected layer $f_l$ we have $\frac{1}{d_k}||f_k(x)||_2^2 = c$ for some $c$. If the weights of $f_l$ are initialized using e.g. Gaussian or orthogonal initialization with the LeCun variance, it is easy to check that $\mathbb{E}_{W_l}\frac{1}{d_l}||f_l(x)||_2^2 = c$. If $d_l$ is not too small, then $\frac{1}{d_l}||f_l(x)||_2^2 \approx c$ with high probability. If an architecture were only composed of a single dependency chain of linear layers, then choosing any initial weight variance other than the LeCun variance would cause the overall magnitude of neuron values to decay or grow exponentially from layer to layer in expectation. Floating-point computation makes such numerical instability undesirable. This gives rise to the following.

\begin{definition}
We say a neural network $f$ exhibits `scale stability' if at each layer, the overall magnitude of neuron values across inputs and neurons is not excessively large or small.
\end{definition}

Since the vast majority of macro-layers in feedforward architectures used in practice do contain a linear layer, LeCun initialization often ends up being reasonable for ensuring scale stability in the initial state. One notable exception to this is architectures composed of macro-layers composed of a linear layer and a ReLU activation layer. It is easy to check that LeCun initialization would yield $\mathbb{E}_{W_l}\frac{1}{d_l}||f_l(x)||_2^2 = \frac{1}{2d_k}||f_k(x)||_2^2$. Thus, the overall magnitude of neuron values would be halved by each macro-layer. Hence, the `He initialization' \citep{heInit} is generally used instead, which doubles the initial weight variance of the LeCun initialization. We refer to this doubled variance value as the `He variance'.

`Ensure scale stability' is the first design strategy we encounter in this chapter that we also consider a ZSAD guideline. It is one of the most widely known and well-understood design principles \citep{1998LeCunTricks,normalizedInitialization,heInit,selu,normProp,saneInit}. 
Unfortunately, there is no widely-accepted term or explicit definition for this guideline, and hence we coin the term `scale stability'. The lack of terminology is a testament to the underdevelopment of ZSAD in the deep learning community in general. The term `scale stability' hints at the fact that an excessive growth or decay of neuron magnitude generally happens gradually from layer to layer. However, our specific definition of `scale stability' is based on the neuron value magnitudes at individual layers, without reference to how those magnitudes came about. We define `scale instability' to mean the absence of scale stability.

Concerns about numerical overflow or underflow are not the only reasons behind the importance of scale stability. In a nutshell, constructs such as layer operations and loss functions are designed, explicitly or implicitly, to perform optimally when their dependencies have neuron values that are not too large or too small overall. We further explore scale stability in section \ref{forwardStabilitySection}. In section \ref{linearEquivarianceSection}, we show that the generality of scale stability as a ZSAD guideline is also limited in important ways. We approximately follow scale stability in the architectures used in our empirical studies as detailed in chapter \ref{empiricalStudiesChapter} by using approximately the LeCun variance.

\paragraph{(Avoiding) neuron bias} In addition to specifying an effective value for the initial weight variance, \citet{1998LeCunTricks} also advocated for neuron values to be unbiased, i.e. to have $\mathbb{E}_xf_l[i_l] \approx 0$ for each $(l,i_l)$ pair. This is another ZSAD guideline. It is less well-known and less consistently defined than scale stability, but can be seen in the work of e.g. \citet{normProp,selu}. Those two papers introduced a strategy we term {\bf activation function debiasing}, which is to choose an activation function such that $\mathbb{E}_{s\sim\mathcal{N}(0,1)}\tau(s) = 0$, which can be achieved by e.g. developing a new activation function or adding a constant to the output of a given activation function. Modeling the input to $\tau$ as a Gaussian is justified by mean field theory and is further expounded upon in section \ref{nlnormDefinitionSection}. We employ activation function debiasing in some of the architectures used in our empirical studies as detailed in chapter \ref{empiricalStudiesChapter}. We further explore neuron bias in detail in section \ref{outputBiasSection}.

\paragraph{Fully-connected network} A `fully-connected network' (FCN) is (approximately) a feedforward network that uses fully-connected layers but no convolutional layers or other linear layers. This is a relatively popular type of network. In general, it is used when there is no specific structure to the data, like tensor or graph structure, that would suggest the use of other linear layers. A key property of fully-connected layers is that they are symmetric with respect to the components of the dependency. This is appropriate when those components appear exchangeable.

In practice, the term `fully-connected network' comes with the expectation that the architecture at least resembles an architecture built from popular building blocks and using popular design strategies, like the building blocks and strategies described in this section and like the architectures we use in our empirical studies as detailed in section \ref{studyAArchitecturesSection}.

\paragraph{Convolutional network} A `convolutional network' (CNN) is a network where linear layers are predominantly convolutional layers as defined in section \ref{tensorLayerSection}, or a variant thereof. Almost always, the input layer is specified as a tensor to represent the natural properties of the data. It is then followed by layers such as convolutional layers and pooling layers that preserve the tensor structure. This is arguably the most popular type of network.

Again, in practice, the term `convolutional network' comes with the expectation that the architecture at least resembles an architecture built from popular building blocks and using popular design strategies, like the building blocks and strategies described in this section and like the architectures we use in our empirical studies as detailed in section \ref{studyBArchitecturesSection}.

\paragraph{Residual network \citep{resNet}} A `residual network' (ResNet) is a network that utilizes building blocks of the form $f_+(f_s(f_a),f_r(f_a))$, where $f_+(.,.)$ is an addition layer with two dependencies, and $f_s(f_a)$ and $f_r(f_a)$ are each chains of single-dependency layers that begin at the same layer $f_a$. $f_r(f_a)$ is referred to as the `residual block' and contains one or more activation layers. Usually, it contains two linear layers, two activation layers and two normalization layers \citep{resNetTrueIdentity,wideResNet}. $f_s(f_a)$ is referred to as the `skip block' or `skip connection' and contains no activation layers, and is often just equal to $f_a$. We refer to the entire group of layers $f_+(f_s(f_a),f_r(f_a))$ as a `residual unit'. The addition weights applied to $f_s$ and $f_r$ by the addition layer are generally fixed throughout training and are usually set to 1.

\section{Deep learning pipeline} \label{pipelineSection}

Beyond the neural architecture, there are other components that comprise a deep learning pipeline. In this section, we define the most popular choices for the core components that also feature prominently in this work.

\subsection{Training algorithms} \label{trainingAlgorithmsSection}

Below, we give the formulas that the most popular gradient methods for neural networks use to derive the update proposal $\delta_\text{prop} \theta$ from the loss gradient. The parameter update is the product of the proposal with the learning rate, as described in section \ref{gradientMethodSection}. Note that in practice, there are slight variations in the algorithms presented below from study to study. We use $\frac{d L}{d \theta}^{(t)}$ to denote the gradient of the loss with respect to the parameter obtained at iteration $t$. The batch remains implicit.

It is noteworthy that none of the popular algorithms explicitly use second-order information. The conjugate gradient and L-BFGS algorithms, which try to infer second-order information from gradients, used to be somewhat popular (e.g. \citet{CodingOptimization}) but have fallen by the wayside. Earlier versions of the algorithms presented below include AdaGrad \citep{adaGrad} and AdaDelta \citep{adadelta}. Newer versions, such as AMSGrad \citep{amsGrad} and AdamW \citep{adamW}, continue to be developed.

\paragraph{Gradient Descent} 

\begin{equation*}
\delta_\text{prop} \theta^{(t)} = -\frac{d L}{d \theta}^{(t)}
\end{equation*}

\paragraph{Momentum}

\begin{eqnarray*}
m_1^{(t)} &=& \beta_1 m_1^{(t-1)} + (1-\beta_1)\frac{d L}{d \theta}^{(t)} \\
\delta_\text{prop} \theta^{(t)} &=& -m_1^{(t)}
\end{eqnarray*}

In the momentum algorithm, the parameter is updated with an exponential moving average of gradients, which is persisted between iterations. $\beta_1$ is a scalar referred to as the `decay rate'. It is generally set to $0.9$. The idea here is to reduce the variation of the update proposal by ``spreading out'' each gradient over multiple updates. The variation reduction is especially evident when small batches are used. $m_1$ is initialized to the zero vector.

\paragraph{Nesterov accelerated gradient \citep{nesterovOrig}}

\begin{eqnarray*}
m_1^{(t)} &=& \beta_1 m_1^{(t-1)} + (1-\beta_1)\frac{d L}{d \theta}^{(t)} \\
\delta_\text{prop} \theta^{(t)} &=& -\Big(\beta_1 m_1^{(t)} + (1-\beta_1)\frac{d L}{d \theta}^{(t)}\Big)
\end{eqnarray*}

This is similar to momentum, except that the current gradient receives a higher weighting in the update.

\paragraph{RMSprop \citep{rmsprop}}

\begin{eqnarray*}
c_2^{(t)} &=& \beta_2 c_2^{(t-1)} + (1-\beta_2)\\
m_2^{(t)} &=&\beta_2 m_2^{(t-1)} + (1-\beta_2)\Big(\frac{d L}{d \theta}^{(t)}\Big)^2 \\
\delta_\text{prop} \theta^{(t)} &=& \frac{-\frac{d L}{d \theta}^{(t)}}{\sqrt{\frac{m_2^{(t)}}{c_2^{(t)}} + \epsilon}}
\end{eqnarray*}

Here, we normalize the gradient using the exponential moving average of its square, which is persisted between iterations. Note that both vector squaring and division in the above formulas are elementwise. $\beta_2$ is generally set to a value between 0.99 and 0.999. Because of this, we must normalize $m_2$ itself by the total capacity $c_2$ of the exponential moving average, as we would otherwise divide by unreasonably small values during the first few iterations. $m_2$ and $c_2$ are initialized to the zero vector. $\epsilon$ is a regularizer introduced for numerical stability and is generally set to a very small value, between $10^{-5}$ and $10^{-10}$.

The idea here is to remove the influence of the overall magnitude of individual gradient components over time on training. This overall magnitude can vary wildly between components, which can cause some components to receive significantly more training than others.

\paragraph{Adam \citep{Adam}}

\begin{eqnarray*}
c_1^{(t)} &=& \beta_1 c_1^{(t-1)} + (1-\beta_1)\\
m_1^{(t)} &=&\beta_1 m_1^{(t-1)} + (1-\beta_1)\frac{d L}{d \theta}^{(t)} \\
c_2^{(t)} &=& \beta_2 c_2^{(t-1)} + (1-\beta_2)\\
m_2^{(t)} &=&\beta_2 m_2^{(t-1)} + (1-\beta_2)\Big(\frac{d L}{d \theta}^{(t)}\Big)^2\\
\delta_\text{prop} \theta^{(t)} &=& \frac{-\frac{m_1^{(t)}}{c_1^{(t)}}}{\sqrt{\frac{m_2^{(t)}}{c_2^{(t)}} + \epsilon}}
\end{eqnarray*}

Finally, Adam combines momentum and RMSprop.

\subsection{Loss functions} \label{lossFunctionSection}

\paragraph{Softmax+Cross-entropy} 

\begin{equation*}
\ell(f,y) = \log \Big(\sum_{i_L} e^{f[i_L]}\Big) - f[y]
\end{equation*}

This is by far the most popular loss function for classification. Here, $f$ returns a vector of dimensionality $|\mathbb{Y}|$, which is interpreted as a scoring function indicating how much of a fit the input is to each class. $f[y]$ denotes the degree of fit of the input to the $y$'th class. The loss decreases as $f[y]$ increases and the loss increases as other components of $f$ increase.

\paragraph{$L2$ loss}

\begin{equation*}
\ell(f,y) = \frac{1}{2}||f - y||_2^2
\end{equation*}

This is by far the most popular loss (and error) function for regression. Here, $f$ returns a vector of the same dimensionality as $y$. Each component is scored based on how close it is to the corresponding component of $y$.

\subsection{Error functions} \label{errorFunctionsSection}

\paragraph{Classification error}

\begin{equation*}
e(f,y) = \mathbbm{1}_{y \neq \arg \max_{i_L} f[i_L]}
\end{equation*}

This is by far the most popular error function for classification. Again, $f$ returns a vector of dimensionality $|\mathbb{Y}|$, which is interpreted as a scoring function indicating how much of a fit the input is to each class. The idea here is that the class we ultimately assign to $x$ based on $f$ corresponds to the maximum score contained in $f(x)$. If that class matches the label, the error, i.e. the negative utility, is 0. Otherwise, it is 1.

\subsection{Data processing} \label{dataProcessingSection}

`Data processing' refers to the process of taking raw data obtained in the real world and converting it into a dataset to be used for machine learning. Data processing is a wide field that includes methods for e.g. converting categorical to real-valued inputs, imputing missing vector components and standardizing dimensionality across inputs. In general, these objectives must be met in the context of feedforward networks. However, they are not the subject of this work. Like most deep learning studies, we use datasets where inputs come as fixed-dimensional, fully specified real vectors. For these datasets, there are only two highly popular processing methods, which are both simple.

\paragraph{Componentwise normalization}

$$x'[i] = \frac{x[i] - \mathbb{E}_{(x,y) \in D}x[i]}{\sqrt{\mathbb{E}_{(x,y)  \in D}x[i]^2 - (\mathbb{E}_{(x,y)  \in D}x[i])^2}}$$

where $x$ is the raw value and $x'$ is the processed value. This normalization method mirrors batch normalization.

\paragraph{Pointwise normalization}

$$x' = \frac{x - \mathbb{E}_{i}x[i]}{\sqrt{\mathbb{E}_{i}x[i]^2 - (\mathbb{E}_{i}x[i])^2}}$$

where $x$ is the raw value and $x'$ is the processed value. This normalization method mirrors layer normalization.

\subsection{Data augmentation} \label{dataAugmentationSection}

Once a dataset is processed and ready to be used for learning, there are additional strategies for modifying the data in order to increase generalization. In section \ref{machineLearningSection}, we explained that supervised prediction always relies on the assumption that patterns found in the dataset extrapolate to other points within $\mathbb{X}$. Oftentimes, we know a high-performing model should extrapolate in certain ways. For example, if we flip a single pixel in the context of image classification or a single character in the context of sentiment analysis, we expect the label to remain the same. If we rotate or reflect an image, we expect the label to remain the same. Therefore, in addition to considering the input / label pairs in the training set, we can generate additional pairs by applying a transformation to the input that (with high probability) does not change the label. This effectively increases the amount of data available for training, which tends to improve generalization. `Data augmentation' refers to any post-processing made specifically to encode desired ways of extrapolation for the purpose of increasing generalization.

A naive way of performing augmentation would be to apply all possible transformations to each input in the training set and store the resulting enlarged training set. Many data augmentation strategies involve random choices. If the number of possible choices is very large, this would lead to storing very large training sets. To circumvent this, instead of performing augmentation as a separate stage before training begins, it is performed during training. Whenever a datapoint from the un-augmented training set is chosen for a batch, a random `augmentation function' is applied on the fly before forward propagation begins. If the datapoint and augmentation are chosen uniformly and independently at random, this corresponds to sampling uniformly from the hypothetical enlarged training set.

In this work, we consider two data augmentation methods that are popular for images, and for the CIFAR10 dataset in particular.

\paragraph{Cropping} Let the raw value $x$ be specified as a tensor as defined in section \ref{tensorLayerSection}. Then we set

$$x'[c,s_1,..,s_{T-1}] = x[c,s_1+o_1,..,s_{T-1}+o_{T-1}]$$

where $x'$ is the augmented value. The $o_t$ are chosen IID from the set of integers between $-O$ and $O$ for some fixed $O$. If the index of $x$ is out of range, the corresponding component of $x'$ is set to zero. For CIFAR10, $O$ is generally set to 2.

\paragraph{Horizontal flipping} Assume the number of dimensions of the input tensor is equal to 3, which is usually the case for images, and let the second spatial dimension correspond to the horizontal image dimension. With 50\% probability, we set $x'=x$. With 50\% probability, we set 

$$x'[c,s_1,s_2] = x[c,s_1,S_2-s_2-1]$$

\section{Differentiability and integrability}

\subsection{Handling non-differentiable networks} \label{nonDifferentiableSection}

Throughout this chapter, we have made extensive use of the gradient of neural networks and individual layers to define key concepts such as backpropagation (section \ref{backpropagationSection}) or specific gradient methods (section \ref{trainingAlgorithmsSection}). We will continue to use the gradient throughout this work. Unfortunately, many popular layer operations, and hence many networks, are not differentiable everywhere with respect to input and / or parameter. This leads to a host of practical, conceptual and theoretical issues. There are practical issues because we need to write programs that represent algorithms that require the gradient at inputs where it is not defined. There are conceptual issues because concepts that rely on gradients are not well-defined when the underlying object is not differentiable. There are theoretical issues because theoretical results that assume differentiability do not technically hold for these objects.

However, these issues generally turn out to be benign. Since neural networks are trained with gradient methods, even when a network is not differentiable everywhere, it must come with a recipe for computing a surrogate gradient, i.e. the gradient of a local linear approximation, which can be supplied to the training algorithm, as explained in section \ref{howDifferentiableSection}. If these surrogate gradients work in the context of gradient methods, they tend to work in other contexts as well. The ability to compute at least a surrogate gradient is one of the foundations of neural networks as defined in section \ref{neuralNetworksSection} and the functional-gradient paradigm as defined in section \ref{functionalGradientSection}.

Unfortunately, there does not exist a general solution for obtaining a surrogate gradient that works for all types of non-everywhere-differentiable networks. Therefore, in general, we have to place the burden on the reader to generalize the contents of this work to whatever non-everywhere-differentiable case they are interested in and to judge to what degree our results apply to that case. Note that reliance on gradients is standard procedure in deep learning literature. Simply ``pretending'' that an architecture is differentiable just ``works out'', both in theory and in practice. The price that would be paid in terms of increased presentational complexity and naked word count to treat these non-differentiable cases explicitly can be prohibitively high. In this work, we choose not to pay this price and stick with the differentiable framework whenever the need for gradients arises.

In this work, we include two sources of non-everywhere-differentiability in the architectures we study empirically, as detailed in chapter \ref{empiricalStudiesChapter}: (i) non-everywhere-differentiable activation functions, given in table \ref{actFunIllu}, and (ii) max pooling. In both cases, we have a layer operation that is differentiable almost everywhere and directionally differentiable everywhere. Below, we describe how we cope with such layer operations on a practical, conceptual and theoretical level.

On a practical level, both the left and right derivatives provide a reasonable local linear approximation for our activation functions. Firstly, it is easy to see that the approximation provided by the directional derivative at points where the derivative is not available (e.g. the zero point for ReLU) is ``almost as good'' as the approximation provided by the derivative at nearby points. Secondly, the probability of the event that an input to an activation function is at a non-differentiable point is very small, even in the context of floating-point computation. Thirdly, even if we were to consider the directional derivative at such points as random noise, that noise is of a small overall magnitude if the network contains many neurons in activation layers. We know gradient methods are robust to small amounts of noise. Hence, when computing the gradient via the backpropagation algorithm, we can replace the derivative of the activation function with either the left or right derivative without significant downside. Similarly, any directional derivative works for the max pooling function.

On a conceptual and theoretical level, we have ensured that all our mathematical constructs, such as definitions and proofs, are fundamentally applicable and somewhat easily extensible to the directionally differentiable case. One way to understand this is as follows. We can approximate each of our non-everywhere-differentiable activation / pooling functions with an everywhere-differentiable activation / pooling function that is arbitrarily close to it such that all quantities we care about are preserved to an arbitrary degree of accuracy. For example, consider $\tau_\text{ReLU}(s) = \max(s,0)$. We could replace it with $\frac{1}{c}\log(1 + e^{cs})$ for some enormous value of $c$. The difference would be negligible, especially considering that floating-point computation is inexact to begin with. Regardless, all our concepts and theory would apply directly.

\subsection{Assuming integrability} \label{integrabilitySection}

The flipside of differentiability is integrability. Integrals that are implicit in, e.g., expectation operators over quantities with continuous distribution are ubiquitous in this work and in deep learning literature in general. For those integrals to be valid, the respective quantities must be Lebesgue integrable.

Integrability has two aspects. First, integrals over bounded sets must be valid. This holds in any reasonable, practical machine learning situation. Functions that violate this condition, such as the scalar function that returns 1 for rational inputs and 0 for other inputs, are only a curiosity. Second, integrals must be finite over unbounded sets. This condition is almost as universal. While it is technically possible to devise e.g. activation functions that do not yield finite expectations with respect to distribution we care about, this would only matter in practice if those activation functions would be evaluated on arbitrarily large inputs, which is impossible with floating-point computation, which can only represent a bounded set of values anyway.

Going forward, we use the term ``integrable'' as follows.

\begin{definition}
When we say a function $F$ is ``integrable'' with respect to a measure $d\mu$, we mean that the Lebesgue integral of $F$ with respect to $d\mu$ is valid over any bounded set, and that the integral of the absolute value $|F|$ is finite over any unbounded set. We use ``integrable with respect to a distribution $\mathsf{dist}$'' interchangeably with ``integrable with respect to the measure corresponding to $\mathsf{dist}$''.
\end{definition}

By including finiteness of the integral of $|F|$ in addition to $F$, we eliminate any ambiguity with regards to how the limit of bounded sets towards a given unbounded set is taken in the definition of the improper integral.

In the majority of even theory-focused deep learning papers, integrability is implicitly assumed for simple expressions involving standard objects like networks, layers, gradients, data distributions and Gaussian distributions. We do the same throughout this work, as we also reiterate in e.g. sections \ref{nlcDefinitionSection}, \ref{finiteNetAssumptionsSection} and \ref{covkerAssumptionsSection}.

\section{Summary of notation, terminology and conventions} 
\label{notationSummarySection}

In this section, we provide a summary of the most important notation, terminology and conventions from chapters \ref{introductionChapter} and \ref{backgroundChapter} that should enable advanced readers to easily follow the technical material presented in later chapters. This summary contains those definitions from chapters \ref{introductionChapter} and \ref{backgroundChapter} that (i) may be used in later chapters and chapter \ref{introductionChapter} without further explanation and (ii) are not widely agreed-upon in the machine learning community. Specifically, we do not repeat here most of the definitions of layer operations, deep learning pipeline building blocks and design strategies given in bold letters in sections \ref{statusQuoSection} and \ref{pipelineSection}.

\subsection{Core terminology}

\begin{itemize}
\item Neural network: Any model to which gradient methods can be applied.
\item Neural architecture: A neural network associated with an unspecified parameter value that must be set via training.
\item Functional-gradient paradigm: Our (approximate) framing of the deep learning field as of the year 2020 based on black-box functions, gradient updates and simple formalisms. See figure \ref{boxFunctionalLearning}.
\item Neural architecture design: Any process that contributes to the choosing of a neural architecture for a task, or of the neural architecture that is eventually deployed.
\item Architecture design strategy: Any piece of information that contributes to the choosing of an architecture.
\item Architecture definition: All the information required to uniquely specify an architecture. It is the information given to e.g. a functional learning framework like TensorFlow to instantiate the architecture in memory. While this technically conflicts with our definition of `neural architecture', we also consider the parameter initialization scheme as part of the architecture definition.
\item Zero-shot architecture design (ZSAD): See figure \ref{boxZSAD}.
\item ZSAD guideline: Any general, predictive, explanatory and ideally well-defined principle that contributes to ZSAD. In this work, we frame a ZSAD guideline as a postulate that a certain property of an architecture or an architecture's randomly initialized state is related to its performance after training. We assess the guideline's utility for ZSAD based on the degree to which that property fulfills the criteria in figure \ref{boxNPM}.
\item Well-defined property: A property of an object like an architecture or dataset that comes with an inherent recipe for determining its unambiguous and unique value. For example, `parameter dimensionality' is well-defined but ``has exploding gradients'' is not.
\item Metric: For a well-defined property, a function that assigns to an object like an architecture or dataset the value of that property.
\end{itemize}

\subsection{Focus of this work}

\begin{itemize}
\item We focus on feedforward networks, including fully-connected and convolutional networks. We often use the terms `neural network' and `feedforward network' interchangeably.
\item We focus on the supervised classification setting, and specifically on the empirical risk minimization approach.
\item We assume that our datapoints are drawn IID from a data distribution, and that the input and label have a deterministic relationship within that distribution which is captured by the true input-label function.
\item We focus on ZSAD guidelines for predicting test error, and sometimes training error, after training. While we view the performance of an architecture as including anything from computational efficiency and privacy to adversarial robustness, we will generally use the term `performance' in this more concrete and limited sense going forward.
\item We focus on developing ZSAD guidelines that, while being as independent as possible of the dataset and task domain in their formulation, are applicable as widely as possible across datasets and task domains.
\end{itemize}

\subsection{Notation}

\begin{itemize}
\item $x \in \mathbb{R}^{d_\text{in}}$: input
\item $d_\text{in}$: dimensionality of the input
\item $y \in \mathbb{Y}$: label
\item $\mathcal{D}$: data distribution over $(x,y)$ pairs or input distribution over $x$ (in discussion, we use the term `data distribution' more generally to refer to either or both constructs)
\item $D$: finite dataset of $(x,y)$ pairs
\item $D_\text{train} \subseteq D$, $D_\text{valid} \subseteq D$, $D_\text{test} \subseteq D$: finite training set, validation set and test set of $(x,y)$ pairs respectively
\item $f : \mathbb{R}^{d_\text{in}} \rightarrow \mathbb{R}^{d_\text{out}}$: neural network; the input of $f$ is $x$
\item $f_0$: input layer of the neural network $f$ to which $x$ is assigned
\item $L$: number of non-input layers in a network
\item $f_L$: output layer of the network $f$ which also returns the output of $f$
\item $d_\text{out}$: dimensionality of the output
\item $\ell : \mathbb{R}^{d_\text{out}} \times \mathbb{Y} \rightarrow \mathbb{R}$ : loss function; the inputs of $\ell$ are $f(x)$ and $y$
\item $0 \le l \le L$ as well as $0 \le m \le L$ when used as a subscript: layer index
\item $d_l$: width / dimensionality of the $l$'th layer
\item $K_l$: number of parents of the $l$'th layer in the layer graph
\item $k_l$: 1-based vector of layer indices that correspond to the parents of the $l$'th layer in the layer graph; treated as a scalar when there is only one parent
\item $K$, $k$: shortened version of $K_l$,  $k_l$ respectively when $l$ is clear from context
\item $f_l: \mathbb{R}^{d_{k_l[1]}} \times \mathbb{R}^{d_{k_l[2]}} \times .. \times \mathbb{R}^{d_{k_l[K_l]}} \rightarrow \mathbb{R}^{d_l}$: $l$'th neural network layer; the inputs to $f_l$ are $f_{k_l[1]}$, .., $f_{k_l[K_l]}$
\item $\mathcal{J}_{l,m} = \frac{df_l}{df_m}$
\item $\mathcal{J}_l = \mathcal{J}_{L,l} = \frac{df}{df_l}$
\item $\mathcal{J} = \mathcal{J}_{L,0} = \frac{df}{dx}$
\item $g_l = \frac{d\ell(f,y)}{df_l}$
\item $\tau : \mathbb{R} \rightarrow \mathbb{R}$: activation function
\item $\tau_l : \mathbb{R} \rightarrow \mathbb{R}$: activation function used by $f_l$ if it is an activation layer
\item $W_l \in \mathbb{R}^{d_k} \times \mathbb{R}^{d_l}$: weight matrix used by layer $f_l$ if it is a fully-connected layer
\item $W_l$ (overloaded): weight tensor used by layer $f_l$ if it is a convolutional layer
\item $[]$: square brackets; used exclusively for tensor indexing and denoting closed intervals
\item $0 \le i_l < d_l$ as well as $0 \le j_l < d_l$: zero-based layer component index
\item $i$, $j$: shortened version of $i_l$, $j_l$ respectively when $l$ is clear from context
\item $\mathbb{E}_{(x,y)}$: shortened version of $\mathbb{E}_{(x,y)\sim\mathcal{D}}$
\item $\mathbb{E}_x$ as well as $\mathbb{E}$: shortened version of $\mathbb{E}_{x\sim\mathcal{D}}$
\item $\mathbb{E}_\mathsf{var}\mathsf{expr}$ where $\mathsf{var}$ has a finite set of values: mean of $\mathsf{expr}$ over $\mathsf{var}$
\item $\overline{\mathsf{expr}}$: shortened version of $\mathbb{E}_x\mathsf{expr}$
\item $\Cov_\mathsf{vec}$: the covariance matrix of $\mathsf{vec}$ with respect to $\mathcal{D}$
\item $\theta \in \mathbb{R}^{\text{dim}(\theta)}$: trainable parameter
\item $f : \mathbb{R}^{\text{dim}(\theta)} \times \mathbb{R}^{d_\text{in}} \rightarrow \mathbb{R}^{d_\text{out}}$ (overloaded) : neural architecture; the inputs to $f$ are $\theta$ and $x$
\item $\theta_l$ where $\theta = (\theta_1, .., \theta_L)$: trainable parameter sub-vector of the $l$'th layer which may be empty
\item $f_l : \mathbb{R}^{\text{dim}(\theta)} \times \mathbb{R}^{d_{k_l[1]}} \times \mathbb{R}^{d_{k_l[2]}} \times .. \times \mathbb{R}^{d_{k_l[K_l]}} \rightarrow \mathbb{R}^{d_l}$ (overloaded) : $l$'th neural architecture layer; the inputs to $f_l$ are $\theta_l$, $f_{k_l[1]}$, .., $f_{k_l[K_l]}$
\item $T$: number of training iterations
\item $1 \le t \le T$: training iteration counter
\item $\theta^{(t)}$: value of the parameter after iteration $t$.
\item $\theta^{(0)}$: initial parameter value
\item $\theta^{(T)}$: final parameter value
\item $B \subseteq D_{\text{train}}$: batch of datapoints
\item $|B|$: batch size
\item $B^{(t)}$: batch used at iteration $t$
\end{itemize}

\subsection{Technical conventions}

\begin{itemize}
\item {\it Network vs architecture:} We use two different conventions for the symbol $f$ and related concepts. In one case, we use $f$ to denote a function that maps an input $x$ to an output $f(x)$ without reference to a parameter value. In that case, $f$ is a neural network. In the other, we use $f$ to denote a function that maps a parameter $\theta$ and an input $x$ to an output $f(\theta,x)$. In that case, $f$ is a neural architecture, and an $(f,\theta)$ pair with a specific $\theta$ is a neural network. We switch between the conventions based on whether the context calls for an explicit treatment of $\theta$.
\item {\it Overloading:} We overload our neural network notation and terminology to refer to mathematical functions, computer programs, graph nodes and vector values at the same time, as well as distributions when $f$ is associated with $\mathcal{D}$. $f_l$ has, for example, a gradient, a runtime, ancestors, vector components and an expectation respectively. The mathematical function provides the abstract definition and its output is a vector. The computer program implements the mathematical function. Each layer is a graph node in the layer graph. We use e.g. the terms `layer operation' and `network function' for the function aspect, `layer program' for the program aspect, `layer value' for the vector aspect, and `layer distribution' and `output distribution' for the distribution aspect. When we use layers or networks to form mathematical expressions, we generally consider their vector values. For example, $f_l=f_m$ indicates that the value returned by both layers is equal, not that they are the same graph node or that the layer operations are equal. Finally, we add tensor structure to a layer if it is part of a convolutional network. However, this does not invalidate our vector-based notation.
\item {\it Omitting inputs:} We often omit inputs of functions in our notation, i.e. we may simply write $f$ instead of $f(\theta,x)$. We sometimes consider a layer not as a function of its parents, but of one or more ancestors. For example, we write $f_l(x)$ to denote the value of $f_l$ as $x$ varies. In general, we pick and choose which inputs to denote explicitly. The same is true for subscripts and superscripts.
\item {\it Expectation vs mean}: We use the term ``expectation'' specifically to refer to the $\mathbb{E}_{x\sim \mathcal{D}}$ and $\mathbb{E}_{(x,y)\sim \mathcal{D}}$ operations, while we use the term ``mean'' to refer to unweighted averages of finite sets, such as $\mathbb{E}_{i}f_l(x)[i]$ and $\mathbb{E}_{(x,y)\in D}$, as well as the mean parameter of Gaussian distributions. While verbally distinguishing between e.g. $\mathbb{E}_xf_l(x)$ and $\mathbb{E}_if_l(x)[i]$ can be tricky at times, we hope this convention will improve readability.
\item {\it Assuming differentiability:}  Throughout large parts of this work, we implicitly assume differentiability on a conceptual level by e.g. using the Jacobian $\mathcal{J}$. All of our concepts and theoretical results are fundamentally applicable and somewhat easily generalizable to practical non-differentiable cases, such as architectures that use ReLU. We also validate our empirical results on such architectures.
\item {\it Assuming integrability:} Throughout this work, we implicitly assume integrability on a conceptual level by e.g. using the expectation operator $\mathbb{E}$ over continuous distributions. This can be considered to hold in all practical cases. Please see section \ref{integrabilitySection} for our precise usage of the term ``integrable''.
\item {\it Batch normalization:} We present our work in terms of single-input, single-output networks $f$. Oftentimes, this does not directly apply to networks with batch normalization. For brevity and readability, the generalization to the BN case often remains implicit. It is explicitly discussed in e.g. sections \ref{nlcComputeSection} and \ref{meanFieldBNsection}.
\item {\it Probabilistic operators:} The same notational conventions that apply to the $\mathbb{E}$ operator apply to other probabilistic operators, such as the standard deviation $\mathbb{S}$.
\item {\it Row vectors:} Vector-valued concepts associated with neural networks such as $f_l$ and $\theta$ are row vectors by default. Jacobians $\mathcal{J}_{l,m}$ have left dimension $d_l$ and right dimension $d_m$.
\item {\it Distribution transformations:} When a distribution $\mathsf{dist}$ is used in expression $\mathsf{expr}$ as if it was a value, then $\mathsf{expr}$ denotes a distribution. Drawing from that distribution is equivalent to drawing a value from $\mathsf{dist}$ and then evaluating $\mathsf{expr}$ with that value.
\item {\it Ordered layers:} Without loss of generality, if $l > m$, then $f_l$ is not an ancestor of $f_m$ in the layer graph.
\end{itemize}

\subsection{Technical terminology}

\begin{itemize}
\item Input: used both in a general sense for a function or program input and in a specific sense for the network input $x$
\item Output: used both in a general sense for a function or program output and in a specific sense for the network output $f(x)$
\item Parameter: used both in a general sense and in a specific sense for the trainable parameter of a neural architecture
\item Layer: used both in a general sense for an arbitrary network node and in a specific sense for an instance of one of the operations defined in section \ref{layerTypesSection}, such as the fully-connected or activation operation
\item Macro-layer: This refers to an instance of a group of operations that occur together frequently, such as a fully-connected operation followed by an activation operation, or a fully-connected operation followed by a normalization operation, an elementwise multiplication operation, a bias operation and an activation operation. Many publications use the term `layer' in the way we use `macro-layer'.
\item Initial weight variance: the variance of an entry of a weight tensor under the parameter initialization scheme
\item LeCun variance: This is 1 over the number of multiply-adds that contribute to a neuron in a linear layer. For example, for fully-connected layers $f_l$, this refers to the value $\frac{1}{d_k}$.
\item LeCun initialization: any parameter initialization scheme where the initial weight variance is the LeCun variance
\item He variance: twice the LeCun variance
\item He initialization: any parameter initialization scheme where the initial weight variance is the He variance
\item Initial state / randomly initialized state of an architecture $f$: the neural network $(f, \theta^{(0)})$
\item Final state of an architecture $f$: the neural network $(f, \theta^{(T)})$
\item Training run: a singular execution of a training algorithm that transforms an initial state network into a final state network
\item Residual unit: the segment in the layer graph from the beginning to the end of a skip connection
\item Skip connection / skip block: the layer graph path within the residual unit that generally does not contain an activation layer
\item Residual block: the layer graph path within the residual unit that generally does contain an activation layer
\item Bottleneck: $f_m$ is a bottleneck for $f_l$ if every directed path from $f_0$ to $f_l$ contains $f_m$.
\item Data shard: any subset of the dataset such as the training or test set, including the dataset itself
\item Predicting a property: When we refer to predicting a final state property, such as test error, we always imply that the prediction was made before training.

\end{itemize}

\chapter{Empirical study design} \label{empiricalStudiesChapter}

In many chapters of this work, we refer to experiments we conducted for the purpose of empirical analysis and validation. The scope of these experiments is a key distinguishing feature of this work. To our knowledge, we go beyond the vast majority of prior work that analyzes ZSAD guidelines, such as the works referenced in section \ref{darkArtSection}. We believe our empirical studies can serve as a guide for designing analytical deep learning studies in general. In this chapter, we detail these studies. Our experiments stand out particularly in three ways: (i) in terms of the variety of architectures studied, (ii) in terms of the carefulness of training and (iii) in terms of the carefulness of metric computation. We conducted exhaustive learning rate tuning independently for each architecture we studied, and we trained for a large number of iterations at many different learning rate levels. We conducted nearly 300,000 independent training runs on fully-connected architectures and nearly 12,000 independent training runs on convolutional architectures, which consumed around 50,000 GPU-hours. For our fully-connected architectures, we conducted all computation using 64-bit floating-point precision. These choices had a large impact on our results, as we summarize in section \ref{experimentsSummarySection}.

Our experiments can roughly be grouped into three buckets, which we discuss in the next three subsections respectively. (i) In study A, we trained 750 fully-connected architectures on three datasets (section \ref{studyASection}). (ii) In study B, we trained 552 convolutional architectures on CIFAR10 (section \ref{studyBSection}). (iii) We conducted further experiments which were similar to study A, but where we changed the architecture or specific aspects of the training protocol (section \ref{additionalExperimentsSection}). 

We computed metrics on the architectures we trained, both in their initial and final state. We also computed metrics on further architectures in their initial state as well as metrics on datasets and a range of activation functions and datasets. Computing these metrics can be trickier than it appears at first glance. We discuss them in section \ref{metricsSection}. In section \ref{metricTerminologySection}, we explain how we present the results of our experiments visually throughout this work. In section \ref{limitationsSection}, we point out limitations to our empirical analysis and discuss our attempts to mitigate those limitations. Some of these limitations are intrinsic to the type of analysis we conducted. Others are specific to the conditions under which this work was conducted.

We repeat the most salient information of this chapter in section \ref{metricsSummarySection}. There, we summarize (i) the choices made in the design of our empirical studies that were most responsible for obtaining the results we cover in this work and (ii) the terminology and conventions that are most important for understanding and interpreting our figures, tables and discussion of experimental measurements. 

\paragraph{Background from prior chapters} Throughout this chapter, we use the terminology, notation and conventions of section \ref{notationSummarySection}.

\section{Study A: Fully-connected networks} \label{studyASection}

We studied a total of 750 neural architectures, 250 for each of three datasets: CIFAR10, MNIST and waveform-noise. In section \ref{studyAArchitecturesSection}, we describe the procedure we used to generate the architectures. In section \ref{studyATrainingSection}, we give the protocol we used for training. In section \ref{studyADataSection}, we detail how we chose our three datasets and what data processing occurred before training began.

We designed the original version of this study without any strong beliefs or expectations about the results we would obtain. After observing some initial results, we only made minor alterations to the study to increase the diversity of observed architecture behaviors without compromising the statistical validity of our results. Crucially, we did not design this study specifically to obtain any result we cover in this work.

\subsection{Architectures used} \label{studyAArchitecturesSection}

\paragraph{Summary} We generated a total of 750 neural architectures at random. These architectures relatively closely resemble popular architectures built using popular design strategies. See section \ref{statusQuoSection} for the definition of the layer operations and design strategies we use. Of course, in this work we aim to develop ZSAD guidelines that reach beyond the space of familiar architectures. We further discuss this point in section \ref{architectureLimitationsSection}. In current practice, when choosing an architecture, certain properties, like depth or the activation function(s) present, are regarded as key to attaining high performance. We validate this view in chapter \ref{surveyChapter}. When generating our architectures, we randomly vary properties that are considered important and that vary between practical architectures: depth, width, weight matrix and bias vector initialization, the activation function used, the normalization operation used, whether the architecture is residual, where the skip connections are located, and the addition weights used by the skip connections. Each property was chosen independently for each architecture, and independently from other properties, with a few exceptions as detailed below.

The full list of architectures is given in the appendix in section \ref{fullListA}. That list should be interpreted in light of the remainder of this section.

\begin{figure}
\centering
\adjincludegraphics[scale=1]{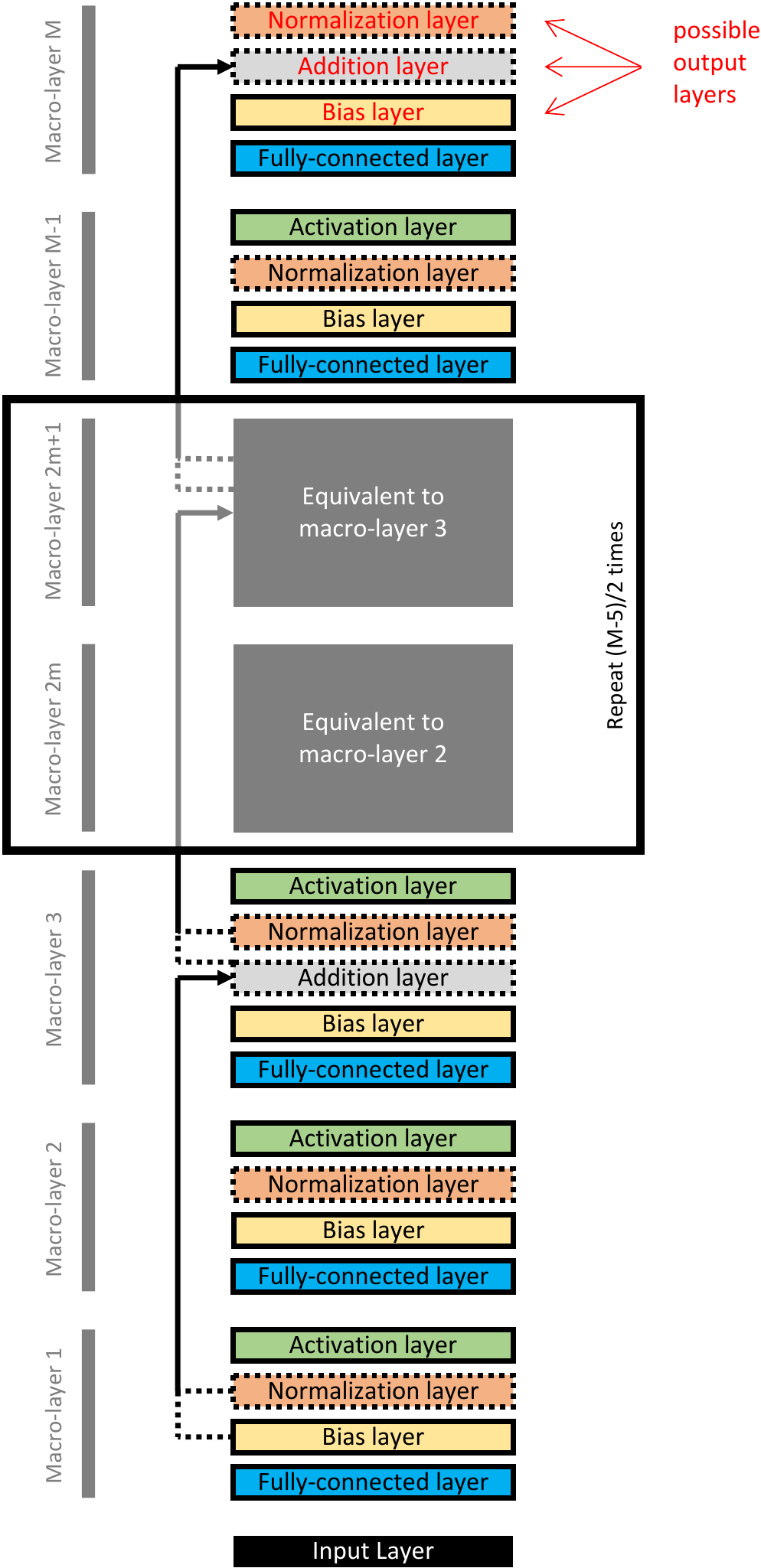}
\caption{The layer graph template for the fully-connected architectures used in study A. Layers with dotted boundary are present in some but not all architectures. Each layer in the sequence directly depends on the preceding layer. Addition layers also depend on exactly one of the layers they are connected to with an arrow. If $M=3$, ``Macro-layer M-1'' is directly preceded by ``Macro-layer 1''.} \label{fcArchIllu}
\end{figure}

\paragraph{Layer graph} The template for the layer graphs of our architectures is depicted in figure \ref{fcArchIllu}. Each architecture is composed of an input layer followed by $M$ macro-layers. Each macro-layer contains at least a fully-connected layer, a bias layer and an activation layer, except the last macro-layer which does not contain an activation layer. Some architectures also have a normalization layer in each macro-layer. In macro-layer $m$, the normalization layer precedes the activation layer if $m < M$ and it is the last layer if $m=M$. In all macro-layers $m$ with $3 \le m \le M$ and $m$ odd, residual architectures have an addition layer directly after the bias layer that adds that bias layer to either the normalization layer or addition layer from macro-layer $m-2$ if $m \ge 5$, or the normalization layer or bias layer from macro-layer 1 if $m=3$. The output layer is simply the last layer in macro-layer $M$ which, depending on the architecture, can be a bias layer, addition layer or normalization layer.

\paragraph{Depth} In our architectures, we define depth as the number of macro-layers $M$, as is common practice for the kind of architectures we study. See section \ref{architectureDesignParadigmsSection}. The depth was chosen uniformly from the set of odd integers between 3 and 49, i.e. $\{3, 5, 7, .., 47, 49\}$. Odd depths are required for residual architectures built according to our layer graph template.

\paragraph{Width} In each of our architectures, almost all layers have the same width, and we call this simply the `width' of the architecture. This width is set automatically as a function of depth so that the parameter has dimensionality approximately equal to 1 million. In section \ref{darkArtSection}, we mentioned the ZSAD guideline of ``using an appropriate width''. While it is unclear how to choose the ``correct width'', the fact that the dimensionality of the parameter has some impact on performance is a foundational observation in machine learning and statistics. We wanted to exclude this source of performance variation for this study. We discuss this choice further in sections \ref{architectureSensitivitySection}, \ref{widthSection} and \ref{experimentsSummarySection}. Setting almost all layer widths equal is the most popular type of width configuration for fully-connected architectures.

The width of the input layer was set to the input dimensionality of the dataset. This is 810 for CIFAR10, 334 for MNIST and 40 for waveform-noise after data processing. See section \ref{studyADataSection} for how we arrived at these numbers. The first fully-connected layer has width equal to the overall architecture width. All layers from then on have the same width until the last fully-connected layer. Finally, all layers from the last fully-connected layer to the output layer have width equal to the number of classes present in the dataset, which is 10 for CIFAR10 and MNIST and 3 for waveform-noise.

\paragraph{Activation function} In each of our architectures, all activation layers use the same activation function, as is overwhelmingly popular for simple fully-connected architectures. This activation function is of form $\tau(s) = c\dot{\tau}(ds+h)+b$, where $\dot{\tau}$ is an activation function from table \ref{actFunIllu} and $d,h,c,b$ are fixed constants. ReLU, SELU and Gaussian are selected as $\dot{\tau}$ with probability $\frac{2}{11}$ each and tanh, even tanh, sigmoid, square and odd square with probability $\frac{1}{11}$ each. We reduced the probabilities of tanh, even tanh and sigmoid as we considered them similar. The same holds for square and odd square.  $d$ is 1 with a 50\% probability, 1.2 with a 25\% probability and 0.8 with a 25\% probability. $h$ is 0 with a 50\% probability, 0.2 with a 25\% probability and -0.2 with a 25\% probability. Finally, we set $(b,c)$ jointly using two constraints. With a 50\% probability, the first constraint is $b=0$ and with a 50\% probability, it is $\mathbb{E}_{s\sim \mathcal{N}(0,1)}c\dot{\tau}(ds+h)+b = 0$. The second constraint is always $\mathbb{E}_{s\sim \mathcal{N}(0,1)}(c\dot{\tau}(ds+h)+b)^2 = 1$.

The primary reason behind introducing these random variations was to increase the diversity of architectures studied. Setting $\mathbb{E}_{s\sim \mathcal{N}(0,1)}\tau(s)^2 = 1$ follows the ZSAD guideline of scale stability as explained in section \ref{architectureDesignParadigmsSection}. If the overall magnitude of neuron values at the dependency of the activation layer is around 1, so it is in the activation layer itself. We wanted to exclude the potential of performance variation due to scale instability in this study. Ensuring $\mathbb{E}_{s\sim \mathcal{N}(0,1)}\tau(s) = 0$ is activation function debiasing, which follows the ZSAD guideline of avoiding neuron bias, as explained in section \ref{architectureDesignParadigmsSection}. We wanted to explicitly study the impact of following or not following that ZSAD guideline.

\paragraph{Weight matrix and bias vector initialization} We use orthogonal LeCun initialization. With a probability of 50\%, we initialize the bias vectors in all bias layers as zero vectors and, with a probability of 50\%, we initialize their components as independent zero mean Gaussians with a variance of 0.05. We took the 0.05 value from \citet{depthScalesMeanField}. If the bias vectors are initialized as nonzero, we scale the weight matrices with a factor of $\sqrt{0.95}$ to attempt to approximately preserve the overall magnitude of neuron values after applying both the fully-connected and bias operations. Finally, with a 25\% probability, we then additionally multiply all weight matrices and bias vectors jointly by 0.9 and with a 25\% probability, we multiply them by 1.1. We did not deviate too much from the LeCun variance to, again, exclude the source of performance variation that is a potential lack of scale stability for this study. Orthogonal initialization has been shown to be superior to Gaussian and uniform initialization for fully-connected layers \citep{orthogonalInitialization,eigenspectrumGram,orthRNN,fastfoodUnitaryRNN,meanFieldCNN,eigenspectrum}.  See also section \ref{orthogonalitySection}.

\paragraph{Residual architectures} With a 50\% probability, an architecture is residual. It then contains addition layers as described above. The addition weights are fixed throughout training. The addition weights associated with a residual block are always 1. With a 50\% probability, all addition weights associated with a skip connection are 1. With a 50\% probability, all addition weights associated with a skip connection are set to the same random scalar that is sampled uniformly from the interval $[0,1]$. This can be considered an ``interpolation'' between a residual and non-residual architecture. With a 50\% probability, all skip connections start at the previous addition layer. (The first skip connection would start at the first bias layer.) This is the most popular choice. With a 50\% probability, skip connections start at a normalization layer, which are always present in our residual architectures (see below). We introduced these variations to obtain a more diverse range of nonlinearity levels among residual architectures. Note that normalizing between successive residual units rather than only within the residual block increases the nonlinearity for reasons explained in section \ref{skipConnectionsSection}.

The last skip connection in the architecture, which starts at macro-layer $M-2$ and ends at macro-layer $M$, adds two layers of different width. Therefore, we modify the addition layer to first multiply the skip connection with a fixed orthogonally LeCun initialized matrix that transforms the vector dimensionality as required, before adding it to the residual block.

\paragraph{Normalization layers} A normalization layer is always used if the architecture is residual. This is necessary to ensure scale stability as explained in section \ref{skipConnectionsSection}. A normalization layer is also always used if the architecture is based on the square or odd square activation function (table \ref{actFunIllu}). This is because if normalization layers were not used,  those architectures would exhibit numerical overflow and underflow during forward propagation even in the initial state due to Gaussian instability as explained in e.g. section \ref{gaussianStabilityExplanationSection}. We would obtain ``nan'' values at the output layer even when using 64-bit precision. This is because repeated squaring can lead to an extremely fast growth or decay from macro-layer to macro-layer.

In all other cases, a normalization layer is used with a 50\% probability. If normalization layers are used, they are batch normalization with a 50\% probability and layer normalization with a 50\% probability. Overall, we obtain the following aggregate frequencies: no normalization layers - 20.4\%, BN - 39.8\%, LN - 39.8\%. The regularizer of the normalization layers is set to zero for the purpose of floating-point computation.

\subsection{Training protocol}\label{studyATrainingSection}

\paragraph{Summary} We trained each of the 750 architectures with stochastic gradient descent 40 times with different starting learning rates and selected the best starting learning rate, independently for each architecture, based on the error on a held-out validation set. During each of the 40 training runs, we reduced the learning rate 10 times by a factor of 3 based on when the validation error stopped improving. All training was conducted with 64-bit precision floating-point computation. We then re-trained the 500 architectures belonging to the CIFAR10 and waveform-noise datasets without early stopping based on validation error to minimize training error, this time using 60 runs to select the best starting learning rate. See section \ref{pipelineSection} for the definition of some of the building blocks referenced in this subsection.

\paragraph{Data shards} CIFAR10 and MNIST come as a pre-specified training set of size 50,000 / 60,000 respectively and a pre-specified test set of size 10,000. We further extracted a validation set of size 10,000 from the pre-specified training set that was drawn uniformly at random, so that our training set had size 40,000 / 50,000 respectively. waveform-noise comes as a single dataset of size 5,000. We extracted a test and validation set of size 1,000 each, drawn uniformly at random, so that we were left with a training set of size 3,000. For each dataset, we used the same training, validation and test set throughout this study.

\paragraph{Parameter initialization} For each architecture, we considered a single random initialization, i.e. a single draw from the random initialization scheme. This is the initial state of the architecture. Given a limited computational budget, we considered it more important to study as many different architectures as possible rather than multiple initializations of the same architecture. 

\paragraph{Training algorithm} We trained each architecture with SGD applied to the training set. We used batches of size 250, which were drawn uniformly at random without replacement. 

\paragraph{Learning rate decay and early stopping} We trained with the starting learning rate (SLR) until the validation error (VE) had not decreased for 10 epochs. We evaluated the VE at the end of each epoch for the purpose of making this determination. Then we re-set the parameter to the value it had 10 epochs prior, when the least VE had been attained. Then we divided the learning rate by 3 and continued training until the VE had not decreased for 5 epochs. We divided the learning rate by 3 again, re-set the parameter and continued training until the VE had not decreased for 5 epochs again. This process continued until the learning rate had been divided by 3 ten times. When the VE had again not decreased for 5 epochs, we re-set one last time and then terminated training. `Early stopping' refers to the common practice of stopping training when the VE no longer decreases, even if the training error may still be decreasing.

\paragraph{Starting learning rate tuning} To ensure that there is no bias with regards to SLR which may skew our results, we tuned the SLR independently for each architecture by training each architecture 40 times. All 40 training runs were fully independent of each other. Each training run used a different SLR. The 40 SLRs formed a geometric series with spacing factor 3. For each training run, the architecture attains its lowest measured VE at the end of that run due to the early stopping procedure described above. We selected as the best SLR the one that yielded the lowest VE and also did not cause overflow at any time during training. Those SLRs are given in section \ref{fullListA}. (Overflow happens when any value involved in the training computation grows beyond the largest value representable by floating-point computation.)

For each architecture, the smallest SLR considered was determined as follows. We ran SGD for 1 epoch with a learning rate of 1 without actually applying the updates to the parameter. For the weight matrix in each macro-layer, we thus obtained one update per iteration. Let $\delta W_m^{(t)}$ denote the update obtained for the weight matrix in macro-layer $m$ at iteration $t$ and let $W_m^{(0)}$ denote the initial value of the weight matrix in macro-layer $m$. The smallest SLR was then chosen to be $10^{-8}\Big(\sum_{m=1}^M\frac{\sqrt{\mathbb{E}_{t}||\delta W_m^{(t)}||^2_F}}{||W_m^{(0)}||_F}\Big)^{-1}$. The reasoning behind this choice is that individual weight matrix updates obtained with the smallest SLR should not perturb weight matrices by more than approximately $10^{-8}$ relative to the norm of the weight matrix. We chose the $10^{-8}$ factor so that our smallest SLR would be less than the smallest learning rate that can be meaningfully used under 32-bit precision floating-point computation. Of course, this choice of smallest SLR is merely a heuristic. The goal of this heuristic is to ensure that the best SLR that would be found by an unbounded grid search is, with very high probability, within the range of SLRs we consider. (Note that even the SLR found by an unbounded grid search is not the true best SLR, as further explained in section \ref{sharpValleySection}.) We validated this heuristic by checking that no architecture that attained a non-random VE for any SLR attained its lowest VE with either one of the smallest five or largest five SLRs considered. This condition was fulfilled for all architectures. See section \ref{learningRateSection} for further analysis on this point.

\paragraph{Hyperparameter tuning} We did not tune any hyperparameters beyond the SLR. We stress the conceptual importance of tuning the learning rate exhaustively in section \ref{moduloAlgorithmSection}. We analyze the importance empirically in section \ref{learningRateSection}. As described above, we also train exhaustively with a large number of learning rates within a single training run.

We defined the concept of hyperparameter tuning in section \ref{hyperparameterTuningSection}. Conducting and analyzing 40 training runs would be considered hyperparameter tuning. After the hyperparameter tuning stage, technically, we would then have to begin the ``actual training'' stage, where we train with the selected hyperparameter values. Of course, at that point, training with the selected SLR value has already been conducted during the tuning stage and does not have to be repeated. Hence, hyperparameter tuning and ``actual training'' are part of the same procedure in our pipeline.

\paragraph{Floating-point precision} All training was conducted with 64-bit precision floating-point computation. This was essential for a significant number of architectures to attain a less-than-random error. See section \ref{noiseStabilitySection} for further analysis on this point.

\paragraph{Loss function} We used an augmented version of softmax+cross-entropy as the loss function. After initializing each architecture, we evaluated the quadratic mean of the output layer neuron values on the training set: $\sqrt{\frac{1}{d_\text{out}}\mathbb{E}_{(x,y) \in D_\text{train}} ||f(\theta^{(0)},x)||_2^2}$. We then had the loss function divide the network outputs by this scalar value before feeding them into the softmax+cross-entropy operation. softmax+cross-entropy yields very different behaviors and levels of performance for networks that return outputs of different overall magnitude, as is known and as we show in section \ref{forwardStabilitySection}. We did not want this fact to confound the results of our study. In essence, we ensure scale stability for the network output. We believe that the preference of softmax+cross-entropy for network outputs of a certain magnitude has confounded the results of studies in the past. The value of this scalar multiplier remained fixed throughout training.

\paragraph{Error function} We used classification error, as is overwhelmingly popular practice for simple classification tasks.

\paragraph{Re-training for training error minimization} We re-trained some architectures to minimize their training error. We refer to this as ``training error minimization''. We made two changes to the protocol. (i) We divided the learning rate by a factor of 3 only once the training error had not decreased for 10 / 5 epochs respectively. We evaluated the full training error at the end of each epoch for the purpose of making this determination. (ii) We considered 60 different SLRs which formed a geometric series with spacing factor 3 and lowest value $10^{-16}\Big(\sum_{m=1}^M\frac{\sqrt{\mathbb{E}_{t}||\delta W_m^{(t)}||^2_F}}{||W_m^{(0)}||_F}\Big)^{-1}$. Therefore, we considered even the smallest learning rate that is meaningful under 64-bit precision floating-point computation. For each training run, the architecture attained its lowest training error at the end of that run. We selected as the best SLR the one that yielded the lowest training error and also did not cause overflow at any time during training. We also found that only one architecture that attained a non-random training error attained its lowest training error with one of either the smallest or largest five SLRs. Note that if we had used the original set of 40 SLRs, we would have been unable to train several architectures that required a tiny SLR, as we explain in section \ref{learningRateSection}.

The reason we did not re-train our 250 MNIST architectures in this way was the very long training times and considerable computational expense incurred when not using early stopping based on validation error. We preferred studying a slightly smaller number of architectures / datasets to compromising our training protocol.

\subsection{Data} \label{studyADataSection}

\paragraph{Selection} We wanted to conduct experiments on three different datasets. First, we chose CIFAR10 and MNIST as they are the two most popular datasets for evaluating deep neural networks. They are also small enough for us to conduct a very large number of training runs with the computational resources we had available. We decided to choose our third dataset from the UCI repository of machine learning datasets. \citet{selu} validated the SELU activation function, which has become somewhat popular, on a large number of datasets from this repository. We wanted to choose a dataset that \citet{selu} also used. To decide upon the specific dataset, we applied the following requirements.

\begin{itemize}
\item The dataset is a classification dataset.
\item The frequency of no class is more than 50\% larger than the average frequency of all classes.
\item The size of the dataset is between 1,000 and 100,000.
\item The dimensionality of the input is at least 10.
\item The dataset does not contain images. (We already study 2 image datasets in CIFAR10 and MNIST.)
\item No component of the input vector is very sparse across the dataset.
\item We are actually able to find the dataset on the repository website.
\end{itemize}

Only two datasets fulfilled all those requirements: waveform and waveform-noise. They are very similar. We chose the latter because of the greater dimensionality of its input. 

\paragraph{Description} The CIFAR10 dataset is composed of 32 by 32 color images and a class label for each image. Each image depicts an object that is of one of 10 types, and the class label corresponds to that type. (citation: \citet{CIFAR10}). Images have three 8-bit color channels, which means each pixel is represented by three integers between 0 and 255. Hence, the entire image is represented by $3*32*32=3072$ integers. For the purpose of applying deep learning, we treat these integers as real values. This makes sense given that their underlying meaning (color intensity) is a continuous concept. In study A, we used fully-connected networks. Hence, we treat the input as a 3072-dimensional real vector. Our data processing, as described below, further reduced the dimensionality to 810.

The MNIST dataset is composed of 28 by 28 grayscale images of handwritten digits associated with a digit label that takes values 0 through 9 (citation: \citet{MNIST}). Because the images are grayscale, they are represented by $1*28*28 = 784$-dimensional vectors, which we consider real-valued. Data processing further reduced the dimensionality to 334.

The inputs of the waveform-noise dataset are 40-dimensional real vectors consisting of wave attributes. Each input is associated with one of three class labels based on the wave type (citation: \citet{waveform-noise}).

CIFAR10 and MNIST come as a pre-specified training set of size 50,000 / 60,000 respectively and a pre-specified test set of size 10,000. waveform-noise comes as a single dataset of size 5,000. In each dataset, all classes are approximately equally frequent.

\paragraph{Processing} We processed CIFAR10 and MNIST via the following sequential procedure.

\begin{enumerate}
\item We perform pointwise normalization as defined in section \ref{dataProcessingSection}.
\item We perform componentwise normalization as defined in section \ref{dataProcessingSection} where only the mean was subtracted.
\item Via PCA, we determine the dimensionality $\mathsf{dim}$ of the input subspace that holds 99\% of the variance of the input. 
\item We multiply all inputs with a fixed $\mathsf{orig} \times \mathsf{dim}$ randomly orthogonally initialized matrix, where $\mathsf{orig}$ is the input dimensionality of the unprocessed dataset.
\item We multiply all inputs with a single scalar constant so that the mean of squares across all components of all inputs becomes 1.
\end{enumerate}

For CIFAR10, $\mathsf{orig} = 3072$ and $\mathsf{dim} = 810$. For MNIST, $\mathsf{orig} = 784$ and $\mathsf{dim} = 334$. Hence, the input layer width of our CIFAR10 architectures was 810 and the input layer width of our MNIST architectures was 334.

In preliminary experiments, we found that this processing scheme led to faster training and lower error values compared to using componentwise normalization only. The reason we developed this scheme was to reduce input dimensionality, so as to reduce the computational complexity of the first fully-connected layer as well as the amount of memory required to train and store the dataset. This allowed us to train more architectures at a given budget.

For waveform-noise, we performed componentwise normalization only. Hence, the input layer width of our waveform-noise architectures was 40.

We note that we did not use the test set ``actively'' in data processing. Our data processing schemes do not apply independently to each input. Rather, they involve taking means over the entire dataset as well as PCA. Both means and PCA used the union of training and validation set, but not the test set. If we treat the value of the means, as well as $\mathsf{dim}$, as fixed constants, then our data processing schemes do apply independently to each input, i.e. they can be viewed as fixed processing functions. These functions were ultimately applied to the test set for consistency. Therefore, if the test set is an IID draw from the original data distribution, then the processed test set is an IID draw from the processed data distribution, i.e. it is an IID draw from the image of the original data distribution under the processing function. The test error remains an unbiased estimate of the true error after processing, though the training and validation sets are no longer IID draws from the data distribution. See section \ref{statisticalChallengesSection} for further discussion on this point. See section \ref{studyATrainingSection} above for how we split the dataset into training, validation and test set.

\section{Study B: Convolutional networks} \label{studyBSection}

We studied a total of 552 architectures on CIFAR10. In section \ref{studyBArchitecturesSection}, we describe the procedure we used to generate the architectures. In section \ref{studyBTrainingSection}, we give the protocol we used for training. See section \ref{studyADataSection} above for information on the CIFAR10 dataset. For study B, the data processing scheme varies as described in section \ref{studyBArchitecturesSection}. The CIFAR10 inputs are naturally cast as 3-dimensional tensors with 3 channels and 2 spatial dimensions of size 32.

While we designed study A without significant prior knowledge about the results we would obtain, we had a strong understanding of the subject matter when designing study B. To mitigate the damage of ``overfitting'' our study design by infusing prior knowledge, we chose some different properties to randomly vary between architectures compared to study A - properties we did not have as much experience with. We also used a different gradient method. Once we determined the study design, we made no further changes after observing results.

\subsection{Architectures used} \label{studyBArchitecturesSection}

\paragraph{Summary} We generated a total of 552 architectures, partially at random. As in study A, we varied properties that are considered important. As stated above, we wanted to choose some different properties to vary compared to study A in order to (i) learn more about these properties and (ii) reduce the contamination of our study design with prior knowledge. Additionally, we wanted to vary only properties for which there does not exist a consensus standard value in the community. For example, we always used the LeCun variance for the purpose of weight initialization in this study.  Another design constraint was that the study needed to be suitable to validate nonlinearity normalization, the algorithm we present in chapter \ref{nlnormChapter}. The reason behind some of the choices made will become clear in that chapter.

We deterministically varied the activation function used, the normalization operation used, and whether the architecture was residual. We additionally varied the following at random: weight initialization scheme, whether bias and elementwise multiplication layers were used, data processing scheme, whether data augmentation was used, the type of pooling layer used, and whether a global average pooling layer was used. (For convenience, we consider data processing and data augmentation as part of the architecture definition in this study, as they were varied randomly along with architecture properties.) Each of these properties was sampled independently of the others.

The full list of architectures is given in the appendix in section \ref{fullListB}. That list should be interpreted in light of the remainder of this section. Some of the building blocks we use are defined in section \ref{statusQuoSection} and some are defined below.

\begin{figure}
\centering
\adjincludegraphics[scale=1]{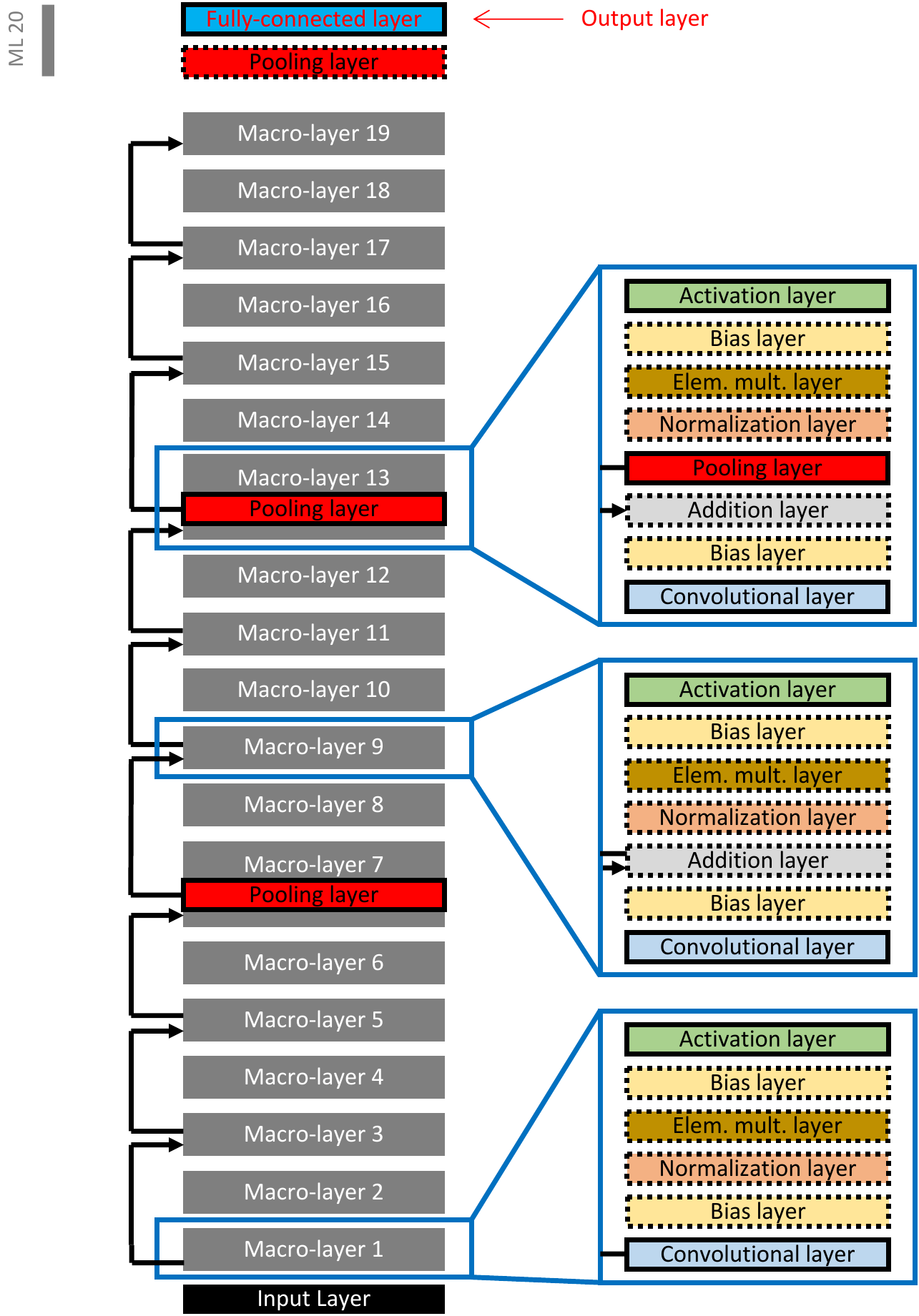}
\caption{The layer graph template for the convolutional architectures used in study B. Layers with a dotted boundary are present in some, but not all, architectures. Each layer in the sequence directly depends on the preceding layer. Addition layers also depend on the layer they are connected to with an arrow. Even macro-layers other than macro-layer 20 mirror macro-layer 1. Odd macro-layers other than 1, 7 and 13 mirror macro-layer 9. Macro-layer 7 mirrors macro-layer 13.} \label{cnnArchIllu}
\end{figure}

\paragraph{Layer graph} The template for the layer graph of our architectures is depicted in figure \ref{cnnArchIllu}. Each architecture is composed of an input layer followed by 20 macro-layers. We begin by discussing macro-layers 1 through 19, which are similar. Each of these macro-layers is composed of some of the layers listed below. All layers present in a macro-layer also appear in the order in which they are listed below.

\begin{itemize}
\item Convolutional layer: always present
\item (First) bias layer: present if the architecture does not use normalization layers
\item Addition layer: present in macro-layers $m$, where $3 \le m \le 19$ and $m$ odd, if the architecture is residual
\item Pooling layer: present in macro-layers 7 and 13
\item Normalization layer: present in some architectures
\item Elementwise multiplication layer: present if randomly chosen to be so (see below)
\item (Second) bias layer: present if the elementwise multiplication layer is present
\item Activation layer: always present
\end{itemize}

Addition layers in macro-layer $m$ add the preceding layer to the addition layer from macro-layer $m-2$ if $m \in \{5, 7, 11, 13, 17, 19\}$, to the pooling layer from macro-layer $m-2$ if  $m \in \{9, 15\}$ and to the convolutional layer from macro-layer 1 if $m=3$.

In contrast to macro-layers 1 through 19, macro-layer 20 contains only a single fully-connected layer which may be preceded by a global average pooling layer. The fully-connected layer is the output layer.

\paragraph{Depth} The depth is set to 20, i.e. there are 20 macro-layers. We did not vary depth because, in convolutional networks, depth has an intricate interaction with the spatial frequency composition of the output, as studied by \citet{meanFieldCNN}. Therefore, depth has a significant indirect influence on performance. We wanted to exclude this source of performance variation for this study. The depth of 20 was derived from \citet{vggNet}. This paper represented the state-of-the art in convolutional architectures before residual architectures were introduced. To our knowledge, 20 is approximately the largest depth at which simple convolutional architectures were observed to have high performance.

\paragraph{Layer size} All layers except the output layer are cast as 3-dimensional tensors with 1 channel dimension and 2 spatial dimensions. See section \ref{tensorLayerSection} for notation and terminology. The final fully-connected layer treats its dependency as a simple vector and returns an output that is not cast as a tensor.

The input layer has size $3 \times 32 \times 32$ according to the natural layout of CIFAR10 inputs. The first pooling layer changes the spatial dimensions to 16. The second pooling layer changes them to 8. The pooling layer in the last macro-layer, if present, changes them to 1. No other layer that is cast as a tensor changes the spatial dimensions.

The convolutional layer in macro-layer 1 changes the size of the channel dimension to 16. The convolutional layer in macro-layer 8 changes the size of the channel dimension to 32. The convolutional layer in macro-layer 14 changes the size of the channel dimension to 64. The addition layers in macro-layers 9 and 15, if present, would then add layers of different channel dimension. Therefore, before the addition, we multiply each spatial location in the skip connection with a random projection matrix of size $16 \times 32$ (layer 9) / $32 \times 64$ (layer 15) before adding it to the residual block. The matrix is the same for all spatial locations and is fixed throughout training. It is LeCun Gaussian initialized unless the convolutional layers in the architecture are delta-initialized with an orthogonal slice (see below). In that case, the matrix is LeCun orthogonally initialized. Apart from the convolutional layers and addition layers just discussed, no layer that is cast as a tensor changes the size of the channel dimension.

Finally, the output layer has width 10 as CIFAR10 has 10 classes. The pattern of layer sizes follows the smallest architecture studied in the landmark paper by \citet{wideResNet}. For the same reasons as in study A, we did not vary the dimensionality of the parameter between architectures.

\begin{table}
{
\centering
\begin{tabular}{lcccc}
AFLM &ReLU-interpolation&ReLU-shift&softplus&Swish\\ \hline\hline
Formula&$\max(s,0)+ls$&$\max(s+l,0)$&$\frac{1}{l}\ln(1+e^{ls})$&$\frac{s}{1+e^{-ls}}$\\
Illustration&\includegraphics[scale=0.208,valign=c]{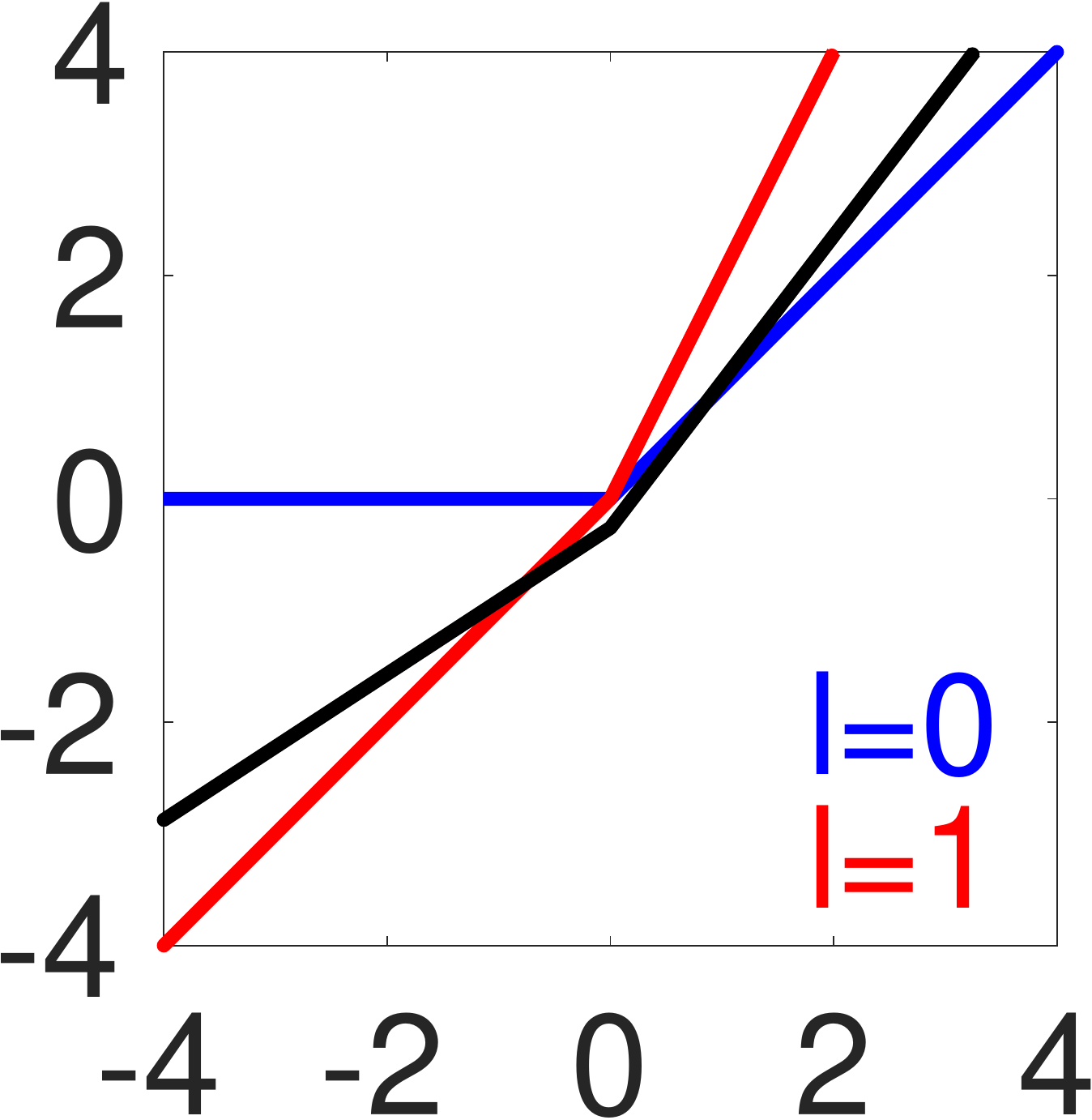}&\includegraphics[scale=0.208,valign=c]{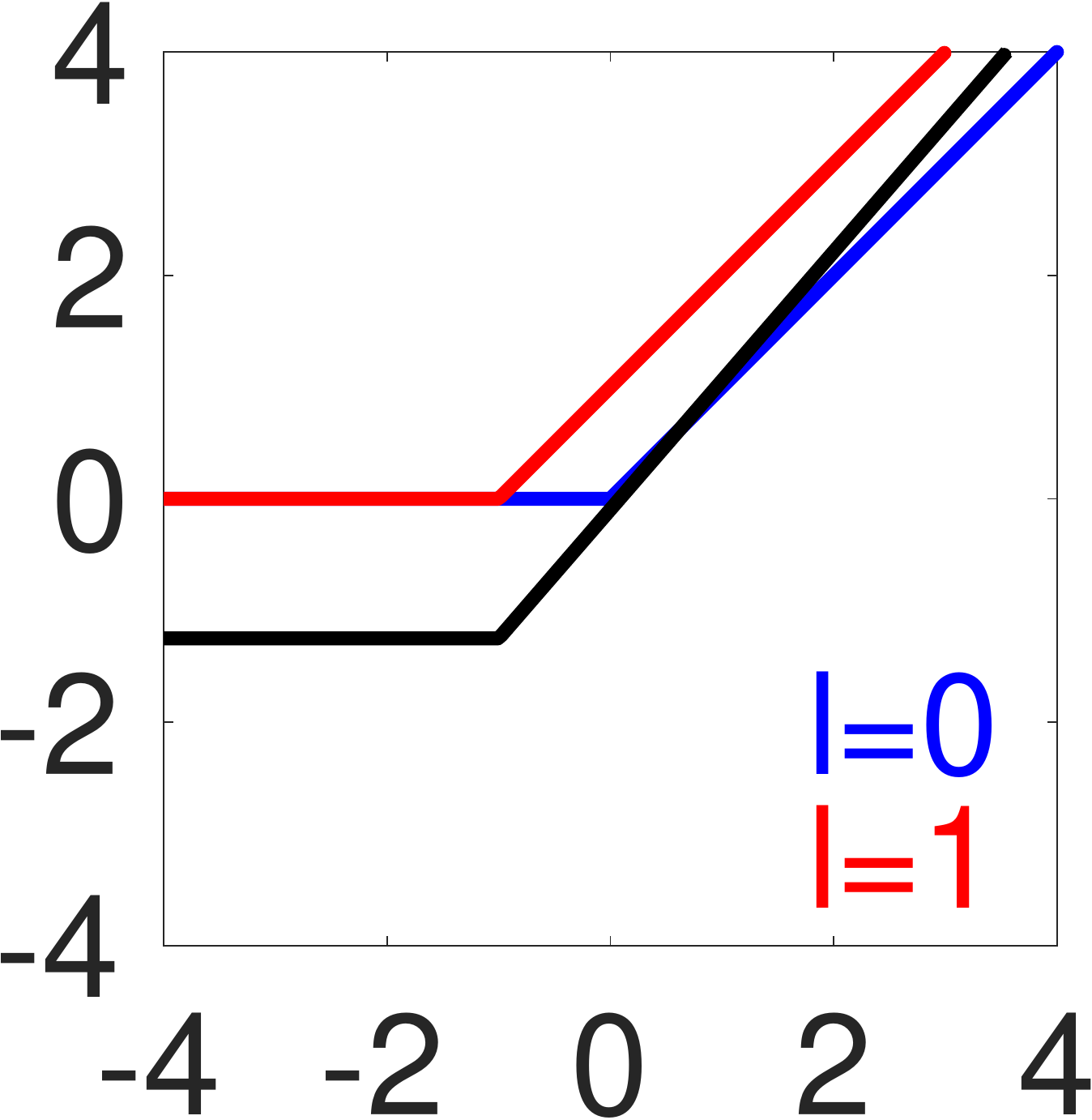}&\includegraphics[scale=0.208,valign=c]{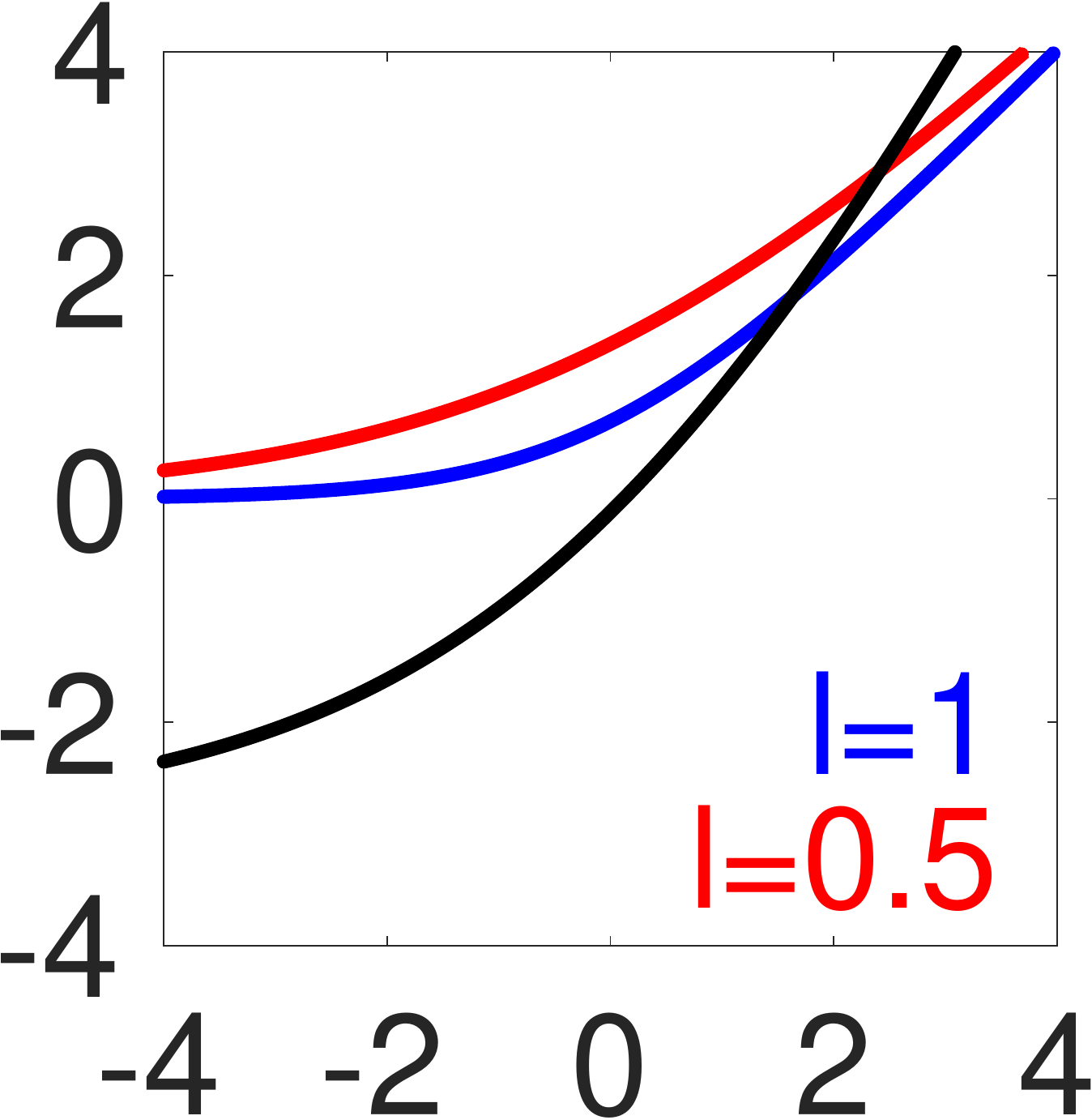}&\includegraphics[scale=0.208,valign=c]{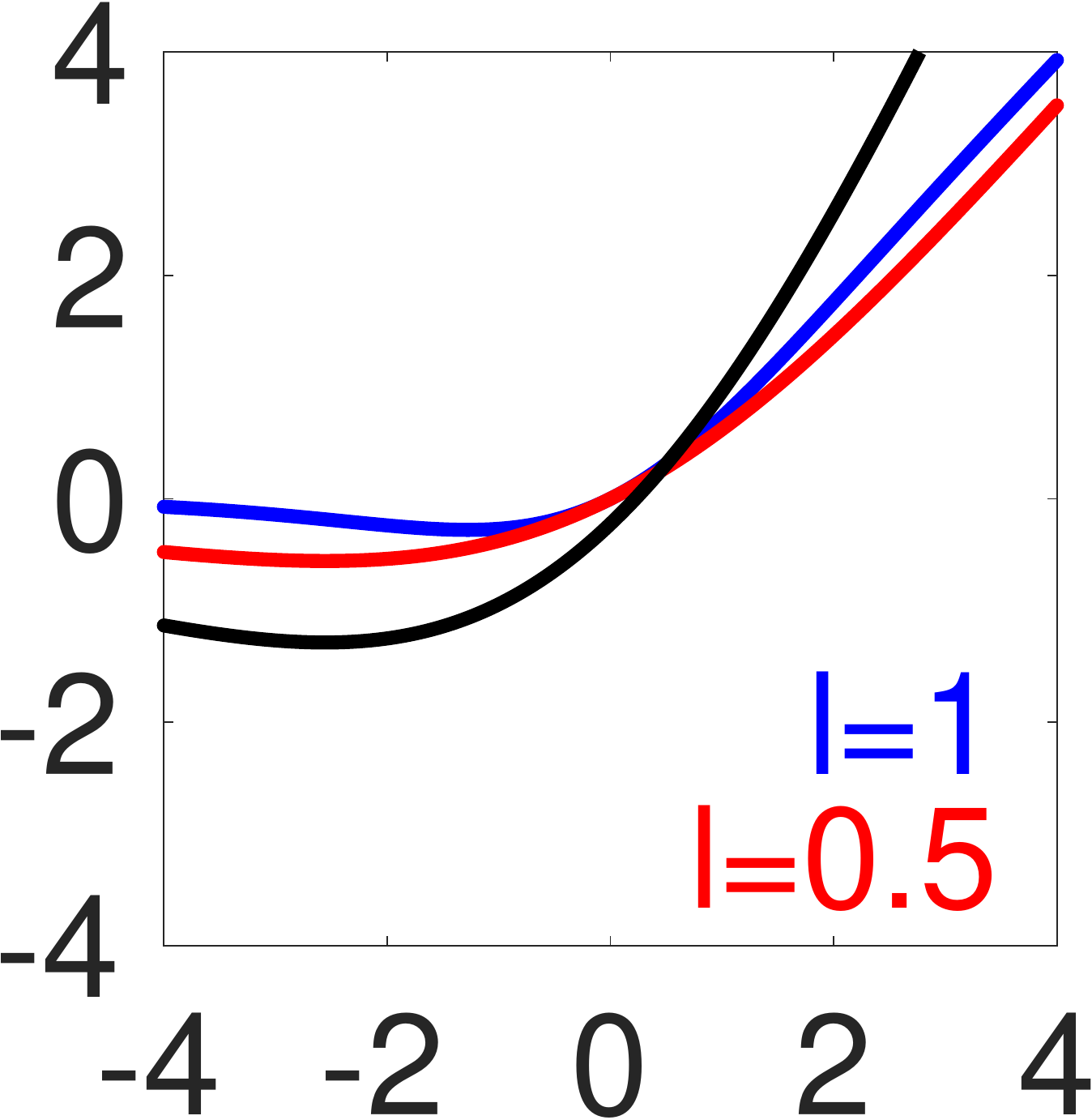}\\
\\
\end{tabular}
\begin{tabular}{lcccc}
AFLM &abs. val.&SELU&tanh-interpolation&tanh-dilate\\ \hline\hline
Formula&$|s+l|$&$\tau_\text{SELU}(s)+ls$&$\tanh(s)+ls$&$\tanh(ls)$\\
Illustration&\includegraphics[scale=0.208,valign=c]{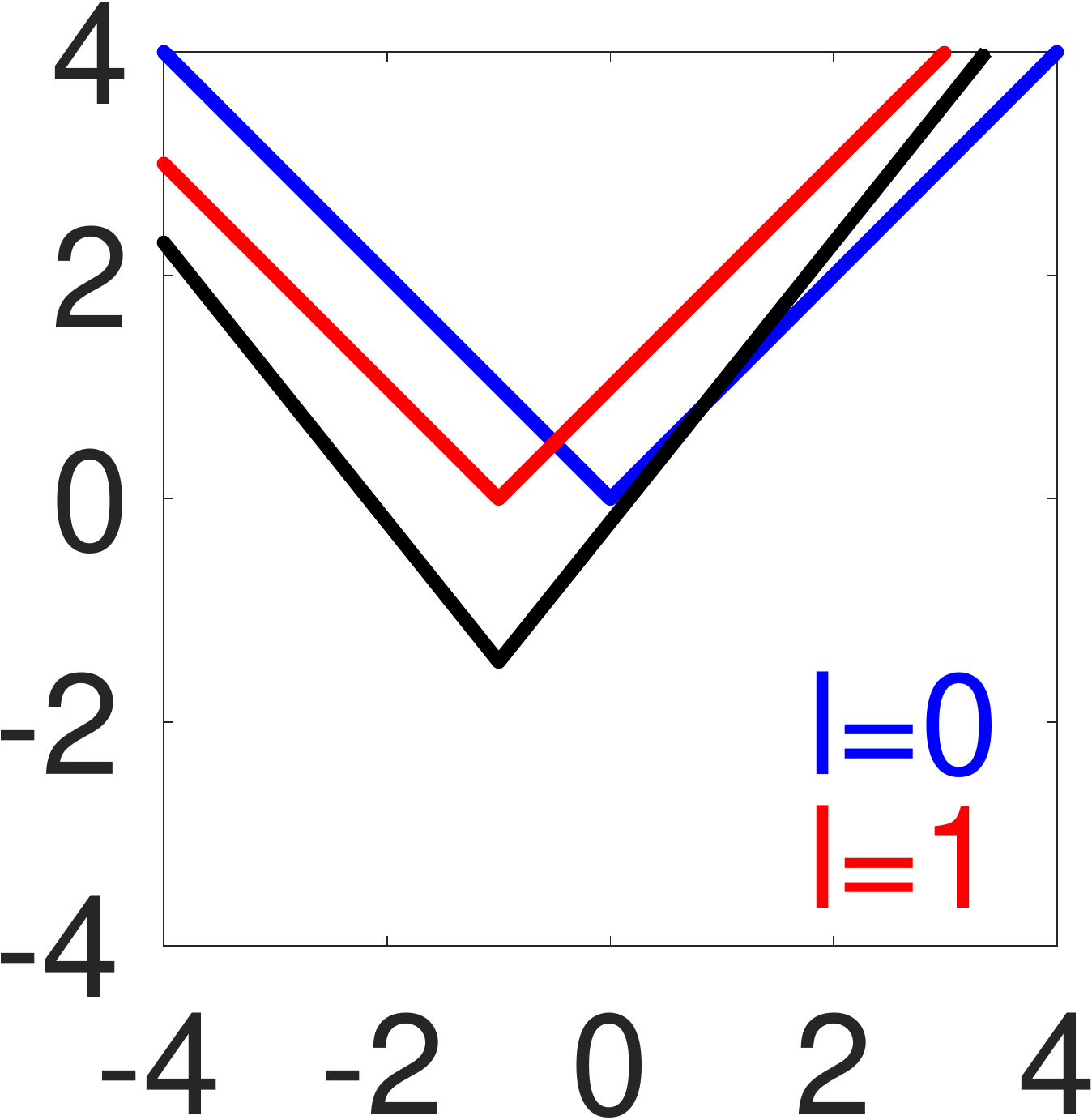}&\includegraphics[scale=0.208,valign=c]{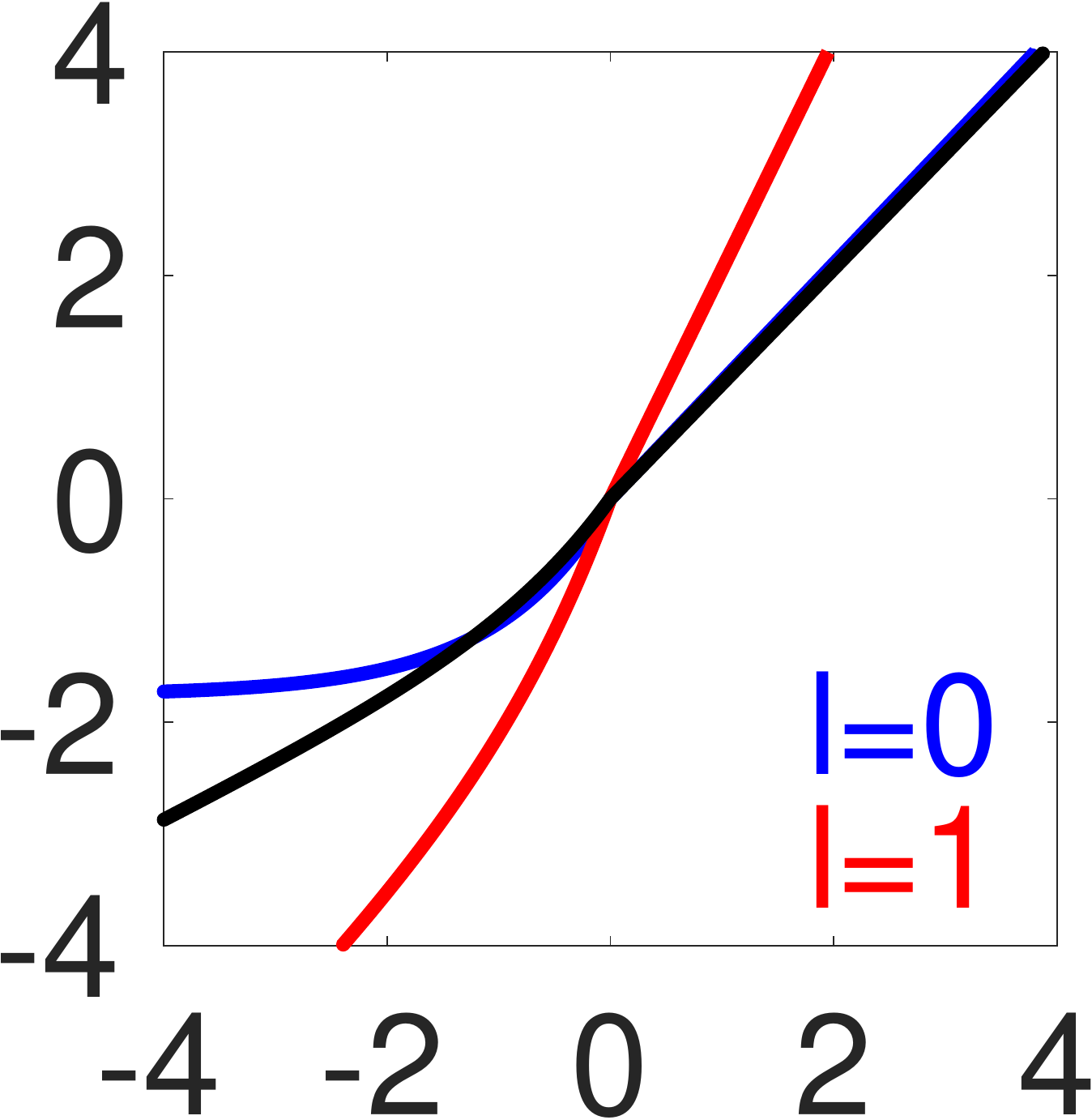}&\includegraphics[scale=0.208,valign=c]{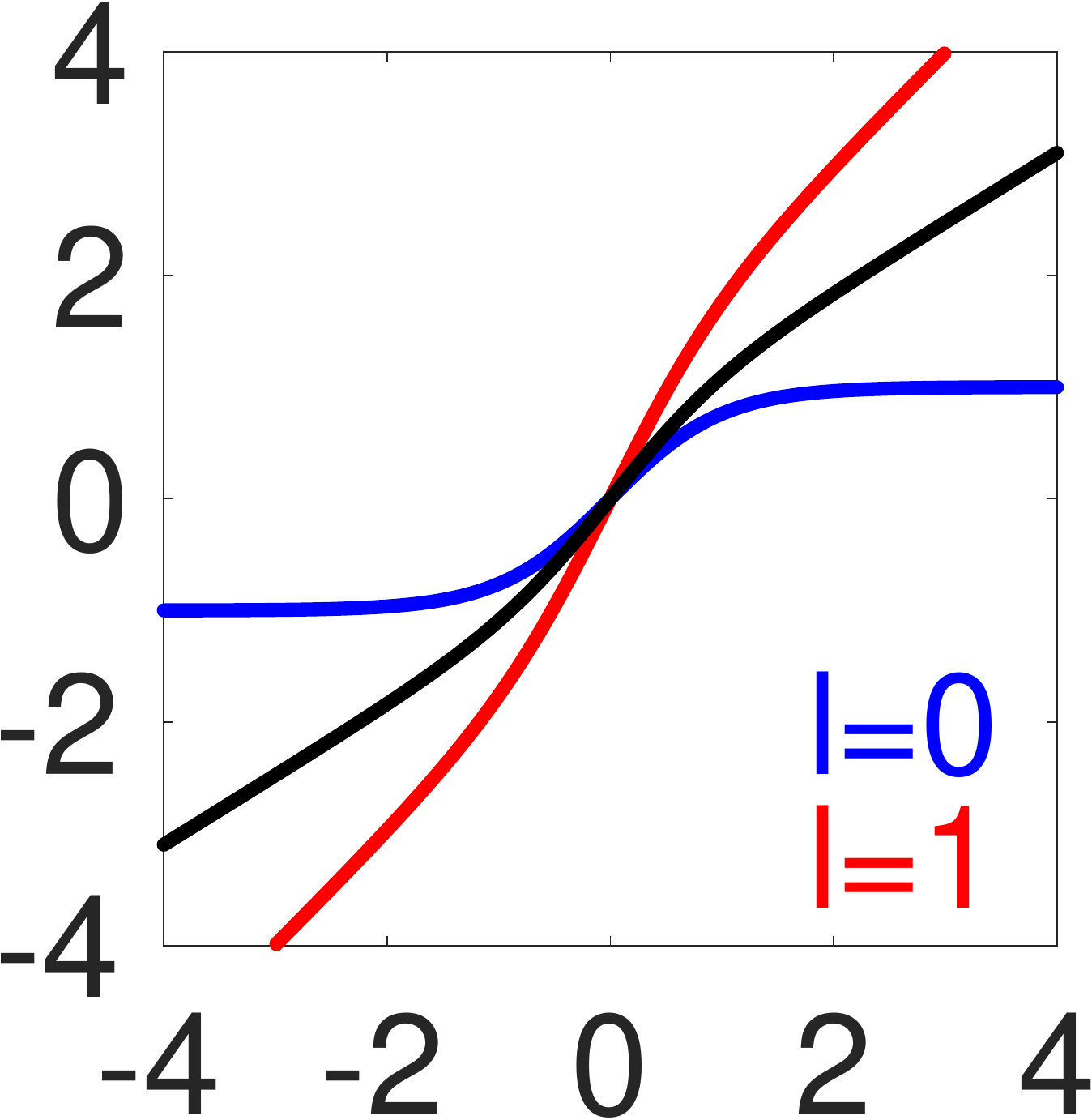}&\includegraphics[scale=0.208,valign=c]{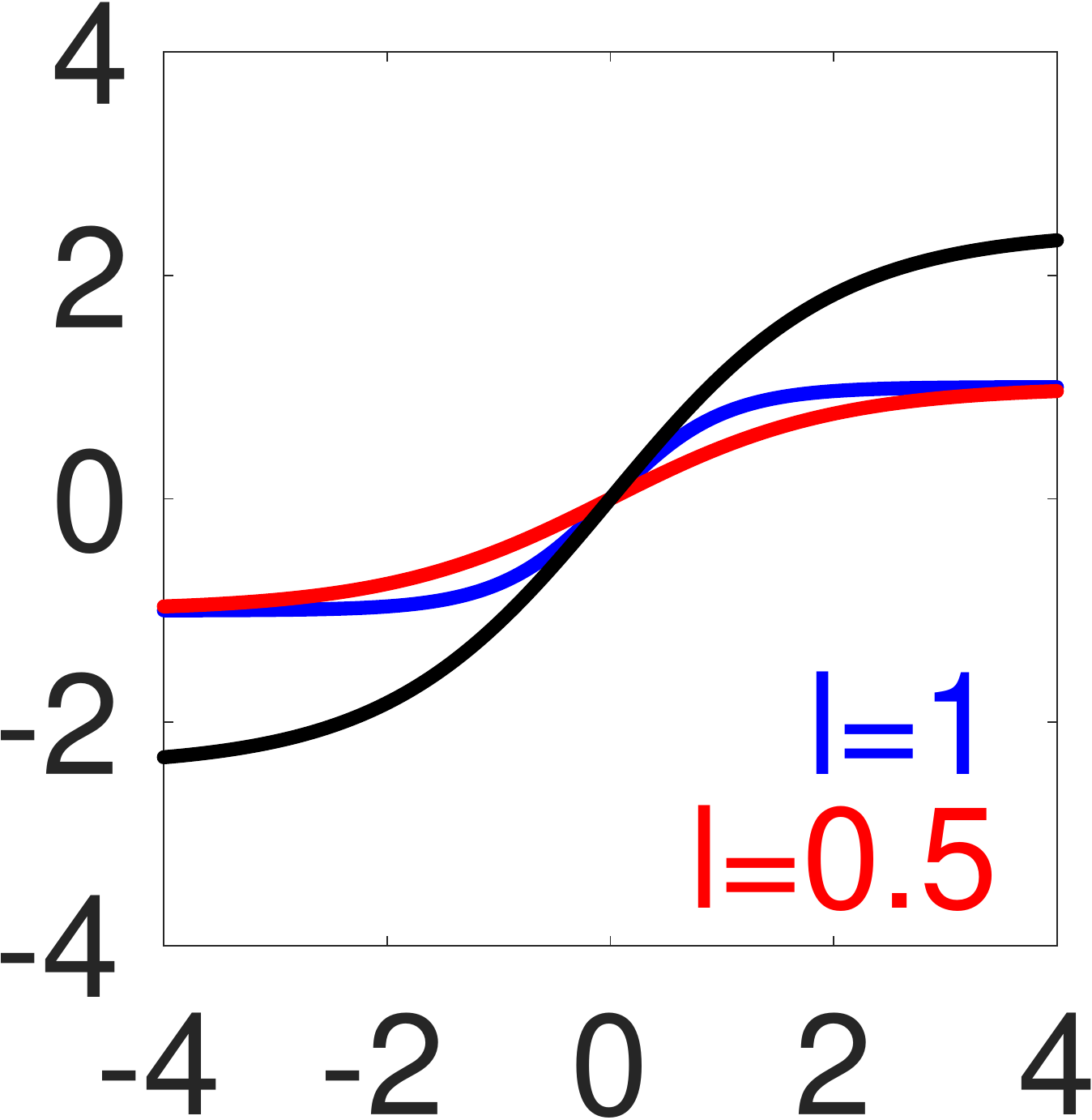}\\
\\
\end{tabular}
\begin{tabular}{lcccc}
AFLM&even tanh&Gaussian&odd square&square\\ \hline\hline
Formula&$|\tanh(s)|+ls$&$\frac{1}{\sqrt{2\pi}}e^{-\frac{s^2}{2}}+ls$&$s*|s|+ls$&$s^2+ls$\\
Illustration&\includegraphics[scale=0.208,valign=c]{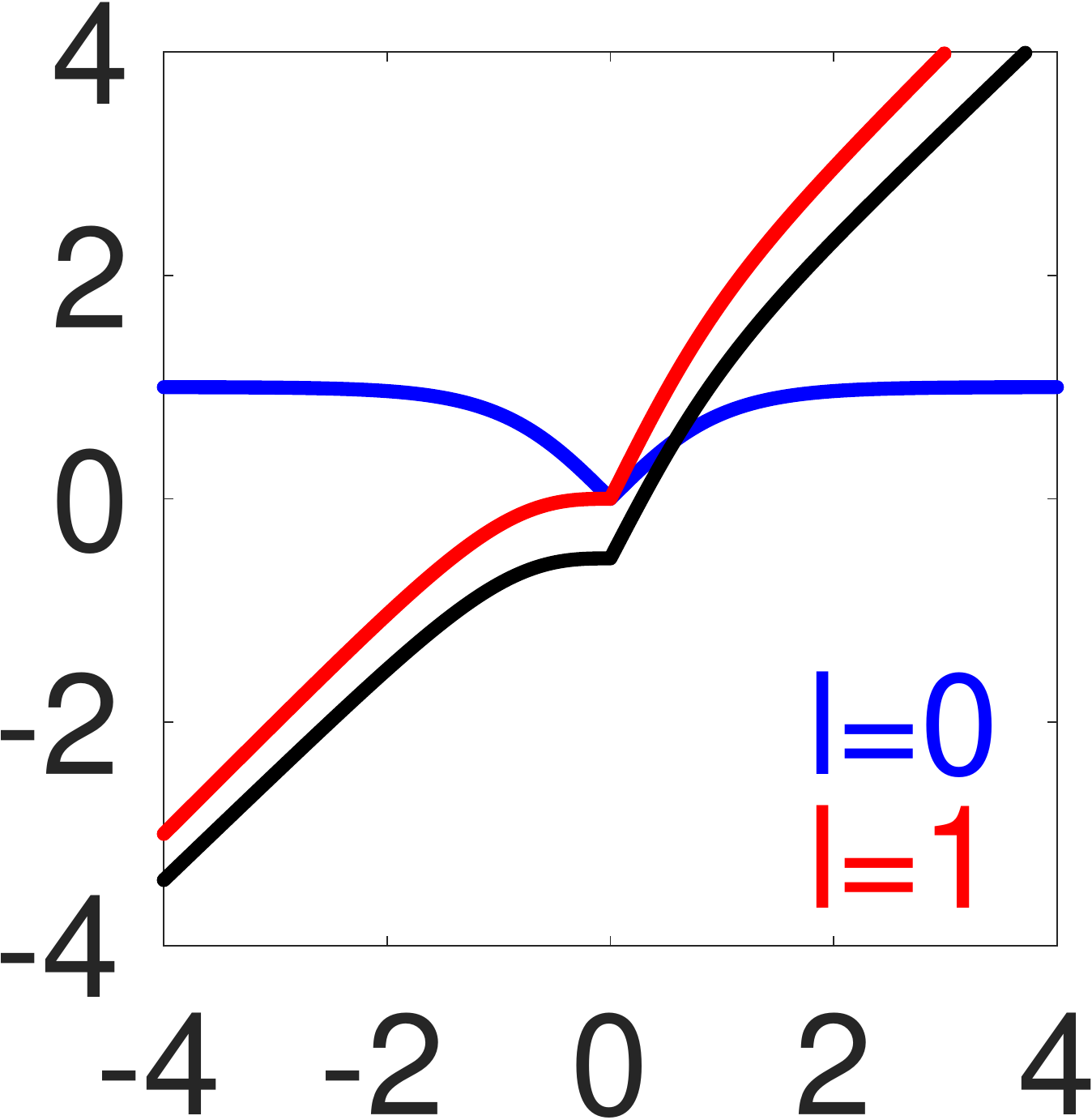}&\includegraphics[scale=0.208,valign=c]{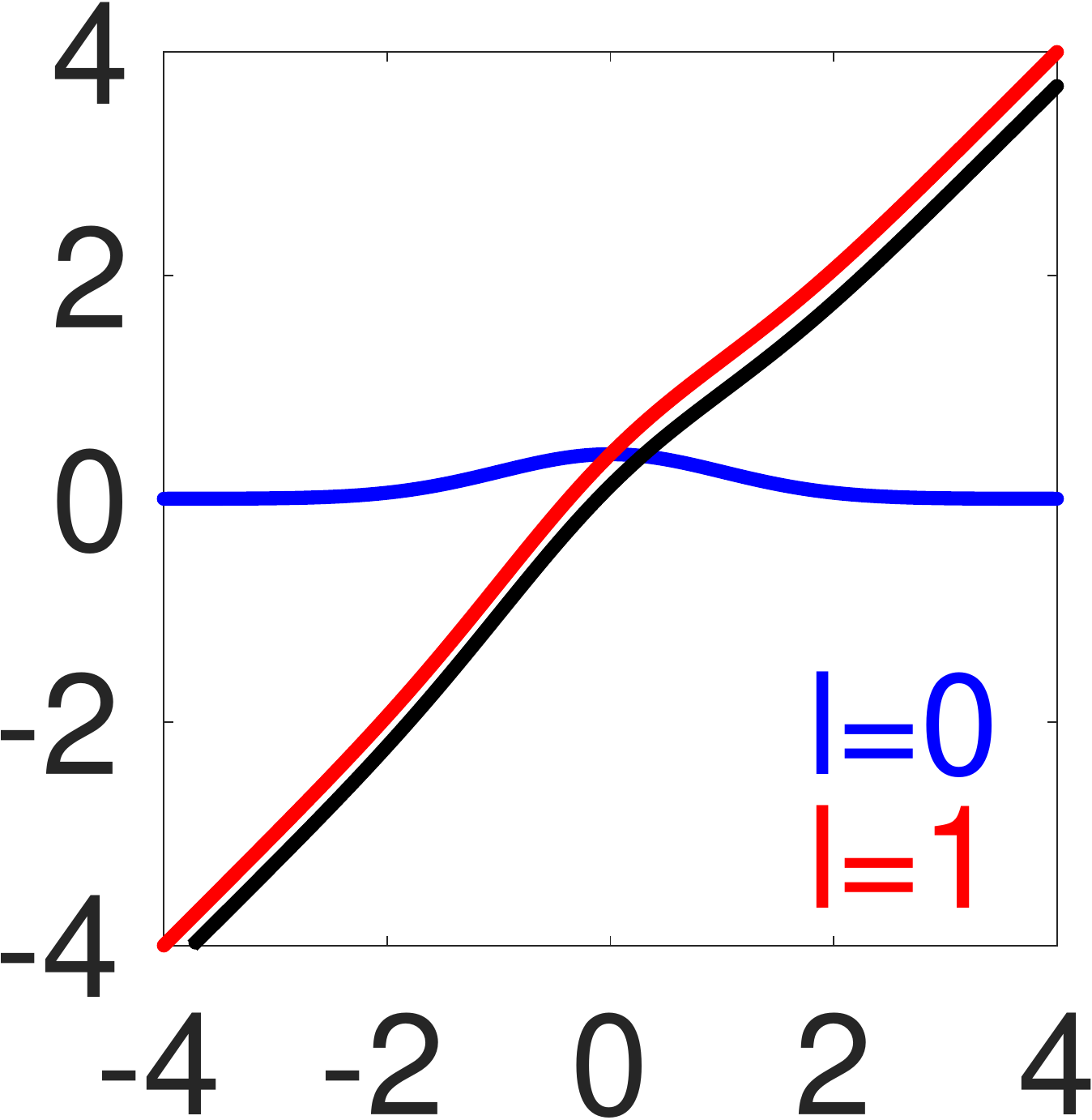}&\includegraphics[scale=0.208,valign=c]{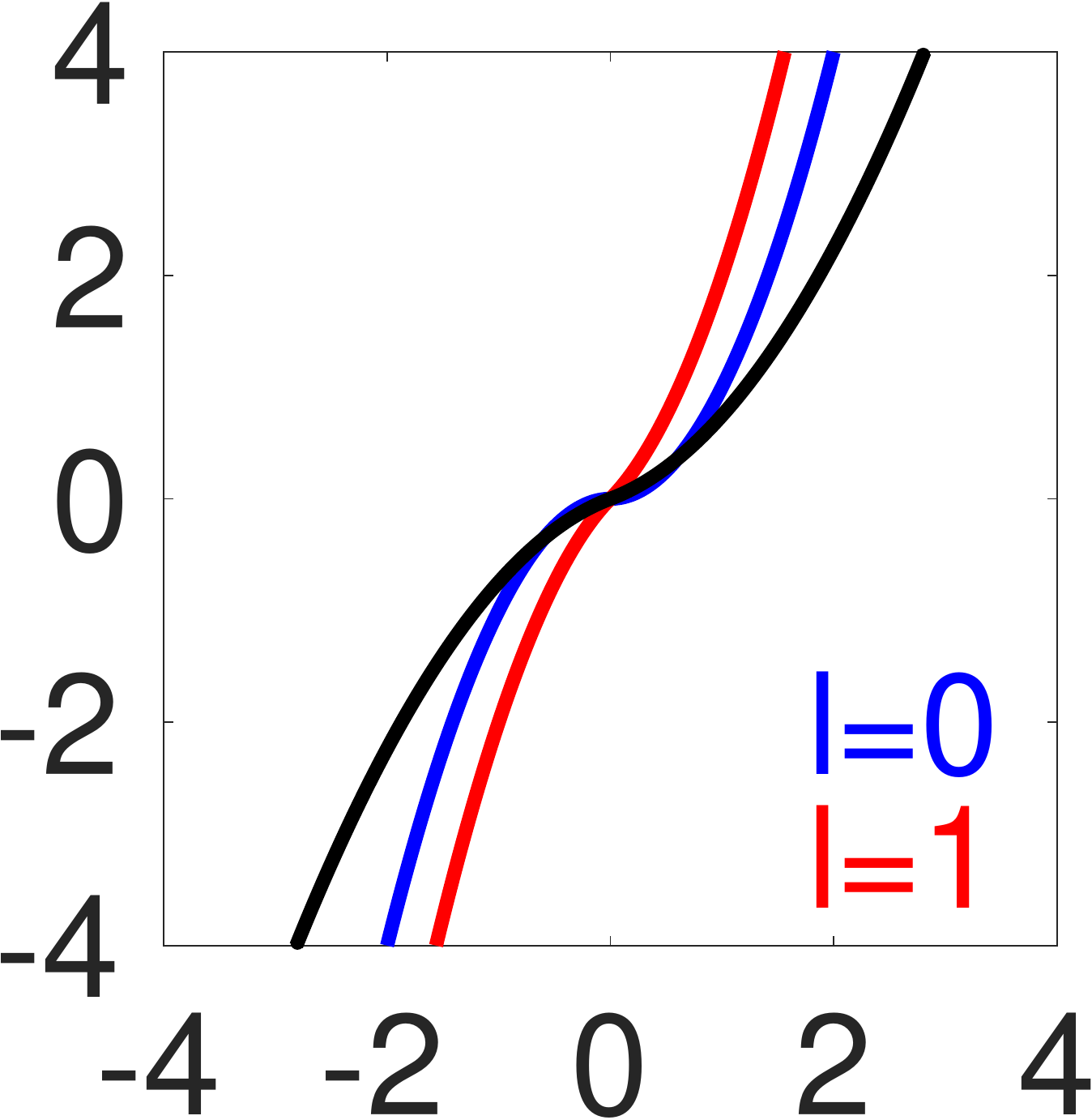}&\includegraphics[scale=0.208,valign=c]{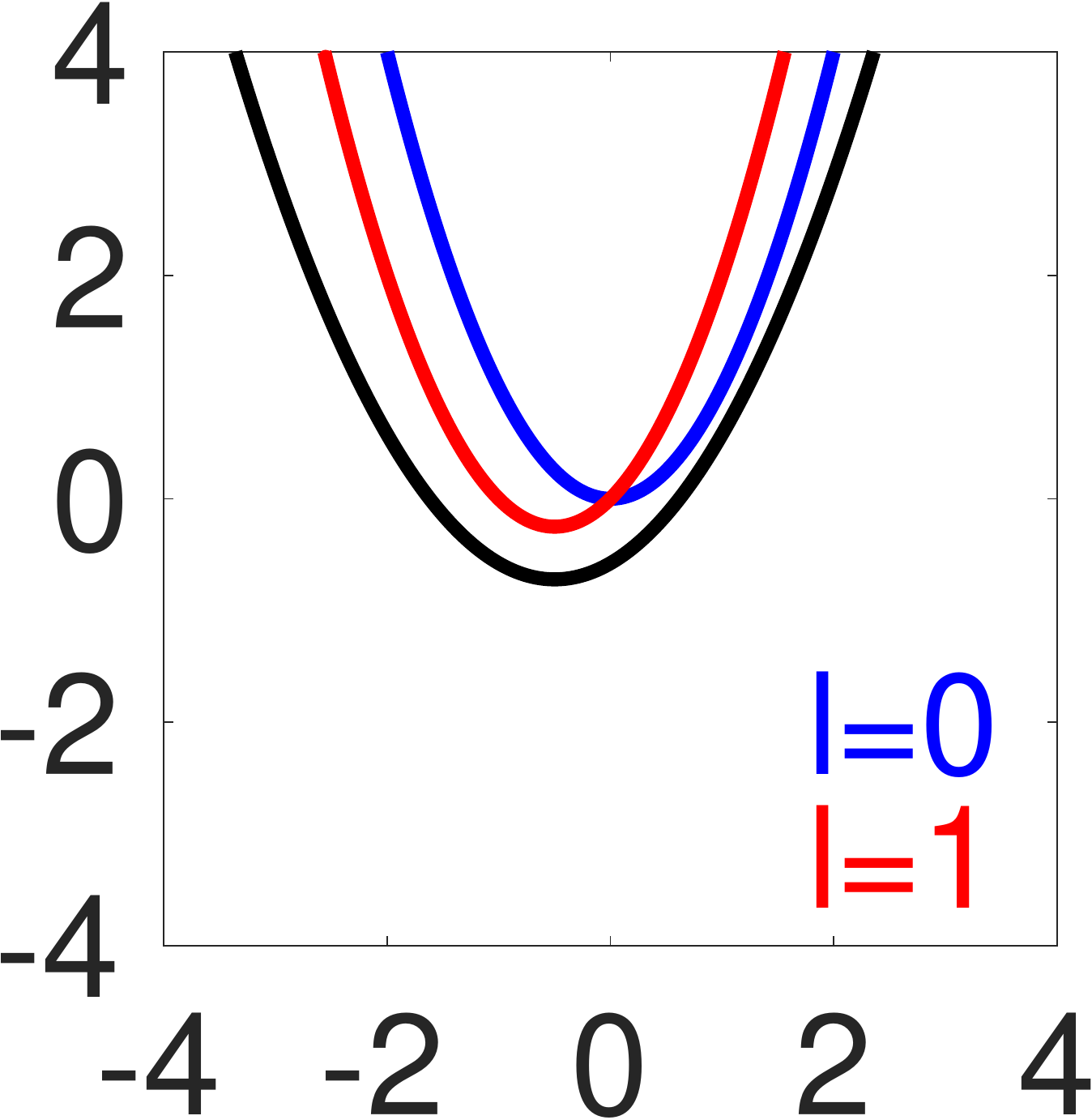}\\
\end{tabular}
}
\caption{Activation functions augmented with linearization methods $\ddot{\tau}(l,s)$ used in study B. The illustrations depict $\ddot{\tau}$ with the default value of $l$ in blue. That activation function is identical to one in figure \ref{actFunIllu}. In red, they depict a linearized version of that activation function. In black, they depict $c\ddot{\tau}(l,s)+b$, where $c$ and $b$ are chosen to achieve $\mathbb{E}_{s \sim \mathcal{N}(0,1)}(c\ddot{\tau}(l,s)+b)=0$ and $\mathbb{E}_{s \sim \mathcal{N}(0,1)}(c\ddot{\tau}(l,s)+b)^2=1$, and where $l$ has the same value as for the red curve.}
\label{AFLMillu}
\end{table}

\paragraph{Activation function} All activation layers in a given architecture use the same activation function, as is overwhelmingly popular for simple convolutional networks. This activation function is of form $\tau(s) = c\ddot{\tau}(l,s) + b$, where $l,c,b$ are fixed constants. (Note that when we don't use the letter $l$ as a subscript, it generally does not denote a layer index as it does in the expression $f_l$.) $\ddot{\tau}(l,s)$ is an activation function augmented with a `linearization method'. A linearization method is a method for interpolating an activation function with a linear function. We abbreviate the concept of an (activation function, linearization method) pair as `AFLM'. In table \ref{AFLMillu}, we depict all AFLMs used in the study. The `linearization parameter' $l$ indicates how close to a linear function $\ddot{\tau}$ is. We will formalize this intuitive concept in later chapters. $l$ always has a `default value' that makes the AFLM revert to the basic activation function from table \ref{actFunIllu}.

Each of the 12 AFLMs in table \ref{AFLMillu} is used with 12 different triplets of values $(l,c,b)$ in the study. See section \ref{fullListB} for details. In the first triplet, $l$ is set to the default value, $b$ is set to 0 and $c$ is set to achieve $\mathbb{E}_{s \in \mathcal{N}(0,1)}\tau(s)^2 = 1$. Across the other 11 triplets, $l$ varies. This causes $\tau$ to be more or less linear. Roughly, these 11 triplets correspond to 0\% linearization, 10\% linearization, etc. up to 100\% linearization. $(b,c)$ are set jointly to achieve $\mathbb{E}_{s \in \mathcal{N}(0,1)}\tau(s)^2 = 1$ and $\mathbb{E}_{s \in \mathcal{N}(0,1)}\tau(s) = 0$. Both strategies for setting $b$ and $c$ are familiar from and explained in section \ref{studyAArchitecturesSection} above.

Combining 12 AFLMs with 12 triplets each yields $12*12 = 144$ activation functions. Each activation function is used in exactly four architectures, except those based on the square and odd square AFLMs, which are used in exactly three architectures. 

Finally, note that the sigmoid activation function is equivalent to tanh-dilate with $l = 0.5$, assuming $b$ and $c$ are set as above. Therefore, we did not use an AFLM that was explicitly based on sigmoid.

\paragraph{Normalization layers and residual architectures} All activation functions not based on the square or odd square AFLM are used in four architectures as given below.

\begin{itemize}
\item a non-residual architecture not using normalization layers
\item a non-residual architecture using batch normalization
\item a non-residual architecture using layer normalization
\item a residual architecture using batch normalization
\end{itemize}

The residual architecture contains addition layers as described above. Addition weights are fixed to 1. The regularizer of the normalization layers is set to zero for the purpose of floating-point computation.

The 24 activation functions based on the square and odd square AFLMs are used in three architectures, which are the second, third and fourth listed above. We always combine those activation functions with normalization layers for the same reason as in study A.

All randomly chosen properties discussed throughout the remainder of this subsection always take the same value for all 36 / 48 architectures associated with the same AFLM. These values are assigned uniformly at random to AFLMs.

\paragraph{Convolutional layers and weight initialization} The spatial dimensions of the weight tensor are always set to 3 with an offset of 1, as established in \citet{wideResNet}. Zero padding is used, as is overwhelmingly popular practice. 

6 AFLMs use LeCun initialization and 6 use delta initialization as proposed by \citet{meanFieldCNN}. In delta initialization, only weight tensor entries that connect each spatial location in the convolutional layer with the same spatial location in the dependency are initialized at random, with all other weight tensor entries initialized to zero. The non-zero components correspond to a 2-dimensional tensor slice where one dimension has size equal to the number of channels in the convolutional layer and one has size equal to the number of channels in the dependency. For 3 AFLMs, that slice is Gaussian initialized, and for 3 AFLMs, it is orthogonally initialized. The variance of the non-zero entries in delta initialization is set to $3*3 = 9$ times the variance used for regular LeCun initialization. This can be considered the LeCun variance for delta initialization, because it ensures scale stability by the same mechanism that we described in section \ref{architectureDesignParadigmsSection}.

\paragraph{Bias and elementwise multiplication layers} Bias vectors are initialized to zero and scaling vectors are initialized to 1. When normalization layers are not used, a bias layer always follows each convolutional layer. When normalization layers are used, for 6 AFLMs, the normalization layer is followed by an elementwise multiplication layer and a bias layer. For the other 6 AFLMs, no bias or elementwise multiplication layers are used in that spot.

\paragraph{Pooling} The pooling layers in macro-layers 7 and 13 use the same pooling function. For 6 AFLMs, subsampling is used, which is a special case of pooling. For 3 AFLMs, average pooling is used. For 3 AFLMs, max pooling is used. The stride is always 2.

The pooling layer in macro-layer 20 is present for 6 AFLMs and not present for the other 6. It is always average pooling, where the average is taken over all spatial locations.

\paragraph{Data Processing} For 3 AFLMs, we used componentwise normalization as defined in section \ref{dataProcessingSection}, where mean and standard deviation were taken over the training set. For 3 AFLMs, we used pointwise normalization. For 3 AFLMs, we used pointwise normalization followed by componentwise normalization. For 3 AFLMs, we used `global normalization', i.e. we normalized with the single scalar mean and standard deviation across all components of all training inputs. As in study A, even though the test set was not used for the means and standard deviations, the same processing function was applied to both training and test set for consistency.

\paragraph{Data Augmentation} For 6 AFLMs, data augmentation was not used. For 6 AFLMs, we used cropping by up to 2 pixels and flipping. Both are defined in section \ref{dataAugmentationSection}. This is the standard data augmentation strategy for CIFAR10 \citep{wideResNet}. Data augmentation was applied on the training set, but not on the test set. Data augmentation was applied after data processing so that the same processing function can be meaningfully applied to both training and test set. Data augmentation was not just applied when training the architecture, but whenever the network was evaluated on a training input, such as during metric computation. See section \ref{metricEstimationSection} for more information on this point.

\subsection{Training protocol}\label{studyBTrainingSection}

\paragraph{Summary} We trained each of the 552 architectures with momentum 20 times with different starting learning rates and selected the best SLR via the test set, independently for each architecture. Each training run lasted for 100,000 iterations, which approximately corresponds to 256 epochs. The learning rate was reduced by a factor of 10 at iteration 40,000, 60,000 and 80,000. All training was conducted with 32-bit precision floating-point computation. See section \ref{pipelineSection} for the definition of some of the building blocks referenced in this subsection.

Some of the choices made in this protocol, especially those that differed from study A, were necessary because we only had a limited amount of time to work with the code base and the compute cluster which we needed to conduct this study. We had to pick and choose which functionalities to implement in code in this limited time frame. We also had to use a lot of pre-written code that imposed constraints on the code we were able to add. We discuss this point further in section \ref{codeLimitationsSection}.

\paragraph{Data shards} CIFAR10 comes as a pre-specified training set of size 50,000 and a pre-specified test set of size 10,000. In contrast to study A, we did not extract a validation set and so the training set we used had size 50,000.

\paragraph{Training algorithm} We trained each architecture with momentum applied to the training set. The decay rate of momentum was set to 0.9, as is overwhelmingly popular practice. We used batches of size 128, which were drawn uniformly at random without replacement. 

\paragraph{Data augmentation} Data augmentation was applied on the fly after a batch was sampled from the un-augmented training set, as described in section \ref{dataAugmentationSection}. We can view the training set under data augmentation as the set consisting of all augmented versions of the ``original'' training datapoints, except that batches cannot contain multiple augmented datapoints that stem from the same original datapoint, as batches are sampled without replacement.

\paragraph{Learning rate decay and number of iterations} We conducted each training run by first training with the SLR for 40,000 iterations. Then we divided the learning rate by 10 and continued training for another 20,000 iterations. We repeated this process twice more until we reached 100,000 iterations, which approximately corresponds to 256 epochs. Then training is terminated. We used a mild version of early stopping based on training error. Specifically, we terminated training after 10,000 iterations if the training error had not fallen below 0.8 until then. Because CIFAR10 has 10 classes of equal frequency, we consider an error above 0.8 as approximately random, i.e. close to the kind of error attained by random guessing.

\paragraph{Starting learning rate tuning} To ensure that there is no bias with regards to SLR which may skew our results, we tuned the SLR independently for each architecture by training each architecture 20 times. Due to limitations in computational budget, in contrast to study A, we only considered 20 SLRs per architecture instead of 40. We also did not adjust those SLRs from architecture to architecture but used the same values for every architecture: $3, 1, 0.3, .., 3*10^{-9}, 1*10^{-9}$. In contrast to study A, it was not entirely clear that a wider grid would not have yielded a better SLR for some of the architectures considered, though definitely not for more than a few. See section \ref{learningRateSection} for further analysis on this point. We selected as the best SLR the one that yielded the lowest test error after training and also did not cause overflow at any time during training. Note that because we did not use early stopping based on test or validation error, the test error after training could, in theory, be significantly higher than the test error at some intermediate iteration. We did not observe such behavior in preliminary experiments. All selected SLRs are given in section \ref{fullListB}.

\paragraph{Hyperparameter tuning} As in study A, we did not tune any hyperparameters beyond the SLR. As in study A, conducting 20 training runs is simultaneously hyperparameter tuning and ``actual training'', because once the value of the SLR hyperparameter is selected, training with that SLR has already been conducted. In contrast to study A, we used the test set, not the validation set, to select the SLR. This means that the test error attained with the selected SLR is no longer an unbiased estimate of the true error. However, our analysis in section  \ref{nlnormBestNLCSection} suggests that this is not a significant problem.

\paragraph{Parameter initialization and random number generation} In contrast to study A, we used a different random initialization for each of the 20 training runs, i.e. a different draw from the random initialization scheme. This was because the deep learning software framework we used for study B did not have strong support for synchronizing random number generation across different processes. In fact, we had to change the random number sequence for all computations associated with the training run. This affected not only parameter initialization, but also batch selection and random choices associated with metric computation (see section \ref{metricsSection}).

\paragraph{Floating-point precision} All training was conducted with 32-bit precision floating-point computation. It is possible that this impacted the performance of some architectures. See sections \ref{noiseStabilitySection} and \ref{beyondNlcSummarySection} for further analysis on this point.

\paragraph{Loss function} We used softmax+cross-entropy as the loss function, as is overwhelmingly popular practice for simple classification. In contrast to study A, we did not have the loss function normalize the network output.

\paragraph{Error function} We used classification error, as is overwhelmingly popular practice for simple classification.

\paragraph{Training error minimization} At various points in this work, we are interested in the training error of our architectures after training (e.g. figure \ref{nlcPredTrainInit}). For those situations, we selected as the best SLR the one that yielded the lowest training error after training, without overflow. In contrast to study A, since early stopping based on test or validation error was not used to begin with, we did not conduct any re-training. Again, we refer to this as ``training error minimization''.

\paragraph{Re-training for result replication} For some architectures, after conducting 20 training runs as described above, we conducted additional training runs. We used the SLR selected from the original runs and trained the architecture another 10 times with different random number sequences. The aim was to replicate the results from the original run. Results from these additional runs are presented in chapter \ref{nlnormChapter}.

We note that some of those reruns exhibited numerical overflow. When presenting the results of study B reruns, we ignore every individual run that overflowed. From each set of 10 reruns, at least 5 did not overflow.

\section{Additional experiments} \label{additionalExperimentsSection}

A large fraction of our analysis was validated through studies A and B. However, we also conducted additional training runs which utilize a different architecture and / or training protocol. We also analyze architectures that do not belong to study A or B in their initial state, without training them. Results from experiments that do not fall entirely under study A or B appear in tables \ref{discillu} and \ref{dsetfusion}, as well as figures \ref{nlcAdam}, \ref{nlcMeanVar}, \ref{nlcRandomInit}, \ref{sensiDist}, \ref{nlcWidth}, \ref{mfMetaGaussiankurt}A through \ref{mfMetaGaussianfq}A, \ref{beyondForward}, \ref{beyondFirst}, \ref{beyondDGD}, \ref{beyondNoiseTest}, \ref{beyondNoiseTrain}, \ref{beyondSummary}, \ref{relatedConf} and \ref{relatedOscillation}. When referring to those experiments, we are usually quite brief. For example, we will write: ``We evaluated the Hessian of a tanh-BN residual architecture of depth 10 and width 100 in its initial state.'' Such brief statements imply that there are a lot of ``default settings'' for non-study A/B experiments. We will specify those default settings, which hold unless stated otherwise, in this section.

\paragraph{Architectures used} Architectures are fully-connected. They follow the same layer graph template as architectures from study A. See section \ref{studyAArchitecturesSection}. The following defaults apply.

\begin{itemize}
\item Width: As in study A, all layers from the first fully-connected layer (inclusive) until the last fully-connected layer (exclusive) have equal width. We refer to the width of those layers as simply the `width' of the architecture. By default, that width is 100. As in study A, the width of the input layer and last few layers is 810 / 10 respectively if CIFAR10 was used, is 334 / 10 when MNIST was used, and is 40 / 3 when waveform-noise was used. It defaults to 810 / 10 otherwise.
\item Activation function: All activation layers use the same activation function. By default, that activation function is from table \ref{actFunIllu}. Any modifications are specified explicitly.
\item Weight matrix initialization: When the activation function is not ReLU, we use Gaussian LeCun initialization. When the activation function is ReLU, we use Gaussian He initialization.
\item Bias layers: By default, there are no bias layers. If there are, bias vectors are initialized to zero.
\item Normalization layers: Normalization layers are not used.
\item Residual architectures: By default, the architecture is not residual, i.e. does not contain addition layers. If the architecture is residual, the addition weights are 1 and the skip connections are configured as they are in study A when they do not start at a normalization layer.
\end{itemize}

\paragraph{Training protocol} The default training protocol was the one used in study A (section \ref{studyATrainingSection}). As in study A, we always verified that the best starting learning rate was not among the largest or smallest 5 considered.

As in studies A and B, in some instances, we were interested in minimizing training error instead of validation / test error. The ``re-training protocol'' from study A was used for training error minimization. See the end of section \ref{studyATrainingSection}.

\paragraph{Datasets and data processing} CIFAR10, MNIST and waveform-noise were processed as in study A. See section \ref{studyADataSection}. Note that the training / validation split depends on the random seed as described below.

\paragraph{Random seeds} In study A, we considered only a single random initialization of each architecture. In fact, we used a single random seed for all computations associated with a given architecture. In other experiments, we allowed the random seed to change. If we vary the random seed, this affects all computations unless stated otherwise.  The random seed controls training set / validation set split, parameter initialization, batch selection and random choices associated with computing metrics (see below). In contrast to the validation set, the test set was specified a priori and remained fixed. Also, when comparing architectures that differ in ways that do not affect the meaning of the random number sequence, such as in figures \ref{beyondForward} or \ref{relatedConf}A-C, we use the same random number sequence(s) for all architectures. This also implies that the initial parameter value does not vary, except possibly in its length as in figure \ref{relatedConf}C.

\section{Metrics} \label{metricsSection}

Almost all of our empirical analysis in this work is based on `metrics'. We use this term loosely in accordance with section \ref{notationSummarySection} to refer to functions of one or more of the following: neural architecture, neural network, layer, parameter, data distribution, input distribution, dataset, data shard, activation function, input, label, loss function, error function and / or layer component index. A metric represents a property of those constructs, or a measure for an ill-defined property.

As an example, consider the true error metric defined in section \ref{machineLearningSection}. It is a function of a network, an error function and a data distribution. In the context of prediction, the true error is arguably the most important measure of performance, which is arguably the most important property of a network. As another example, consider the nonlinearity coefficient, which is the most important metric we study apart from error / loss. It is defined in section \ref{nlcDefinitionSection}. It is a measure of the `degree of nonlinearity' as introduced in section \ref{whatIsNonlinearitySection}. It is a function of network and input distribution.

In section \ref{metricComputationSection}, we discuss a range of challenges that arise when computing metric values. In section \ref{metricTerminologySection}, we introduce key terminology and conventions which we use throughout this work when discussing and presenting metric values and associated concepts. We also describe the visual tools we use for presenting results based on metrics. 

\subsection{The challenges of computing metric values} \label{metricComputationSection}

In sections \ref{programFunctionOverloadingSection} and \ref{neuralNetworkNotationSection}, we explained in detail how a large number of concepts in machine learning have a dual nature as both mathematical functions and programs that compute approximations to those mathematical functions. The same is true for metrics. We introduce them as mathematical functions, but for the purpose of empirical analysis we must compute values for them. In section \ref{programFunctionSection}, we explained that a core advantage of the functional-gradient paradigm is that we get to use machine learning models which are relatively easy to evaluate on a computer. Unfortunately, computing a value for a metric of a network is often not as easy as it is to evaluate the network or conduct backpropagation. Of course, we design metrics in this work to be as easy to compute as possible while capturing the properties we care about. However, it is inevitable that some challenges arise when computing metric values. Below, we discuss those challenges.

Specifically, the three core challenges we faced when computing metric values for this work were (i) metrics that are defined in terms of a probability distribution, (ii) floating-point rounding error and (iii) computing metric values for architectures utilizing batch normalization. We discuss these challenges in the next three subsubsections respectively. Because of the limited space we have available, these discussions must remain somewhat high-level. Presenting an exhaustive discussion of the computational challenges surrounding each individual metric we consider, unfortunately, goes beyond the scope of this work. For the NLC, we do give an exhaustive discussion in section \ref{nlcComputeSection}. We hope this can serve as a blueprint for readers to reason about other metrics we use or other metrics they are interested in. In general, we urge the reader to carefully consider the issues we discuss here.

\subsubsection{Probability distributions in metric definitions} \label{metricEstimationSection}

In this subsubsection, we explain how we compute values for metrics that are defined in terms of probability distributions. We define two types of such metrics. (i) We define metrics that are ``distribution-valued'', i.e. where the output of the metric function is itself a distribution. These metrics are ``distribution transformations'', i.e. drawing from their output distribution is equivalent to drawing from other distribution(s) and then applying the transformation represented by the metric. (ii) We define metrics that are scalar-valued. These metrics apply a probabilistic operator over a distribution.

\paragraph{Distribution-valued metrics} A small number of our metrics are functions that output a distribution. In all cases, our aim is simply to plot that distribution for various metric inputs. This is done in figures \ref{sensiDist}, \ref{groundtruth}, \ref{dsetfusion} and \ref{mfElemLike}. As long as the sample drawn from the distribution(s) that are being transformed is large enough, the plots are informative. We describe below how the sample is selected.

\paragraph{Probabilistic operators} The majority of our metrics that are defined in terms of distribution(s) utilize probabilistic operators. The most frequent such operator is the expectation operator $\mathbb{E}$. The others are defined below.

\begin{itemize}
\item Standard deviation operator: $$\mathbb{S}\textsf{var} = \sqrt{\mathbb{E}\textsf{var}^2 - (\mathbb{E}\textsf{var})^2}$$
\item Median operator: $$\mathbb{M}\textsf{var} = \textsf{val} \text{ where } P(\textsf{var}\le \textsf{val}) = 0.5$$
\item Covariance operator: $$\mathbb{C}(\textsf{var}_1,\textsf{var}_2) =\mathbb{E}\textsf{var}_1\textsf{var}_2 - (\mathbb{E}\textsf{var}_1)(\mathbb{E}\textsf{var}_2)$$
\item Excess kurtosis operator: $$\mathbb{K}\textsf{var} = \frac{\mathbb{E}(\textsf{var} - \mathbb{E}\textsf{var})^4}{\big(\mathbb{E}(\textsf{var} - \mathbb{E}\textsf{var})^2\big)^2} - 3$$
\end{itemize}

$\textsf{var}$ simply denotes an arbitrary random variable. If $\textsf{var}$ is multi-dimensional, the operators are applied elementwise.

We now face the challenge of devising programs that can compute sufficiently accurate approximate values for these probabilistic operators. We employ two different strategies for this: numerical integration and statistical estimation. We discuss them in turn.

\paragraph{Numerical integration} Some of our metrics are defined in terms of expectations over 1-dimensional or 2-dimensional Gaussian distributions for which the mean and (co)variance is known. These expectations always involve activation functions. For example, consider $\mathbb{E}_{s\sim\mathcal{N}(0,1)}\tau(s)$ for some activation function $\tau$. To compute an approximation, we can either draw a Gaussian sample and compute the empirical mean of $\tau$ over that sample, or we can compute $\int_s\tau(s)n(s)ds$ using numerical integration methods, where $n$ here is the unit Gaussian density. It turns out that the latter method is far more accurate, given that $n$ is smooth and all our $\tau$ are also well-behaved. Hence, this is the strategy we employ. For maximum accuracy, it is important to handle points where $\tau$ is not smooth properly.

Earlier in this chapter, we have already encountered several `Gaussian expectations'. In sections \ref{studyAArchitecturesSection} and \ref{studyBArchitecturesSection}, we used them to calibrate our activation functions.  We will continue to use this kind of calibration in e.g. section \ref{nlnormDefinitionSection}. We always compute these Gaussian expectations to within an error of less than $10^{-7}$. In general, calibrating our activation functions to high accuracy is essential, because errors can compound from layer to layer. The same compounding issue arises in mean field metrics as defined in section \ref{meanFieldPracticalSection}, which also utilize Gaussian expectations.

\paragraph{Statistical estimation} The majority of the time, our metrics are defined in terms of distributions that are high-dimensional and / or not explicitly known. In that case, we must use statistical estimation. Unless stated otherwise, we use the following basic estimators for the probabilistic operators we defined above.

$$\mathbb{E}\textsf{var} \text{ becomes } \mathbb{E}_S \textsf{var}$$

$$\mathbb{S}\textsf{var} \text{ becomes } \sqrt{\frac{|S|}{|S|-1}\Big(\mathbb{E}_S \textsf{var}^2 - (\mathbb{E}_S \textsf{var})^2\Big)}$$

$$\mathbb{M}\textsf{var} \text{ becomes } \mathbb{M}_S \textsf{var}$$

$$\mathbb{C}\textsf{var}_1\textsf{var}_2 \text{ becomes } \sqrt{\frac{|S|}{|S|-1}\Big(\mathbb{E}_S\textsf{var}_1\textsf{var}_2 - (\mathbb{E}_S\textsf{var}_1)(\mathbb{E}_S\textsf{var}_2)\Big)}$$

$$\mathbb{K}\textsf{var} \text{ becomes } \frac{(|S|+1)(|S|-1)\mathbb{E}_S(\textsf{var} - (\mathbb{E}_S\textsf{var}))^4}{(|S|-2)(|S|-3)(\mathbb{E}_S(\textsf{var} - (\mathbb{E}_S\textsf{var}))^2)^2} - 3\frac{(|S|-1)^2}{(|S|-2)(|S|-3)}$$

$S$ denotes a sample of the distribution over which the operator is taken. $\mathsf{var}$ becomes a function of an element of $S$. $\mathbb{E}_S$ denotes the mean over the finite sample. $\mathbb{M}_S$ denotes the median over the finite sample. All the basic estimators above are canonical. Again, consider the true error as an example. It is defined in terms of an expectation over datapoints $(x,y)$, which are distributed according to the data distribution $\mathcal{D}$. We estimate it via the mean over a finite sample of datapoints drawn from $\mathcal{D}$. This is precisely what e.g. the test error does.

By default, we define estimators for all our metrics in a canonical fashion by building on the basic estimators given above. For example, to estimate the hypothetical metric $\frac{\mathbb{S}_{x\sim \mathcal{D}}||x||_2}{\mathbb{E}_{x\sim \mathcal{D}}||x||_2}$, we would use the basic estimator for both the standard deviation and expectation, and then divide the values we obtain. We give our metric definitions in this work specifically in such a way as to make the translation from mathematical definition to estimator obvious. The exception to this rule is the expression $\mathbb{E}_x\Tr(\mathcal{J}(x)\Cov_x\mathcal{J}^T(x))$, which appears in the metrics NLC, NLCNUM and LNLC. We detail the estimation of this expression in section \ref{nlcComputeSection}.

\paragraph{Choosing a sample for a data distribution} Many of our metrics are defined in terms of a data distribution or input distribution $\mathcal{D}$. These constructs are defined in section \ref{machineLearningSection}. In our metrics, data distributions always arise as inputs that can be freely chosen along with other inputs. In general, we want to evaluate metrics on practical data distributions. Of course, data distributions are hypothetical constructs. We simply assume that datapoints in each of our datasets $D$ are drawn IID from such a distribution. In general, the only concrete piece of knowledge we have about $\mathcal{D}$ is the (assumed) sample that is $D$. So, in order to compute an approximation of a probabilistic operator over $\mathcal{D}$, we have to use estimation.

The clearest choice for the sample $S$ is the dataset $D$ itself. In practice, we use a shard of $D$ as the sample. The most common choices are training set, validation set and test set. Sometimes we further randomly subsample or bootstrap those shards. It is important to note that even if we assume that the datasets as originally downloaded from the internet are composed of independent draws from the data distribution, there are factors internal to our pipeline that render this assumption untrue for the samples we actually feed into our estimators. We further discuss this point below and in section \ref{statisticalChallengesSection}.

\paragraph{Choosing a sample from high-dimensional Gaussian or uniform distributions} Some of our metrics are defined in terms of high-dimensional Gaussian and / or uniform random variables. Their distribution is specified as part of the metric definition and is not a free input. Metrics that utilize such random variables also always have a data or input distribution input. The limited amount of data available for the data distribution is always the statistical bottleneck in these cases. We can always draw a sufficiently large sample for our Gaussian or uniform random variables to perform sufficiently accurate estimation.

Sometimes, we wish to evaluate metrics that have as an input an input distribution by supplying a Gaussian distribution as that input distribution. See e.g. section \ref{nlcRobustDataSection}. In that case, it is sometimes possible to derive an exact, closed-form value for the metric. However, for the sake of consistency, we always use estimation as we do for practical input distributions. In fact, we drew three fixed samples of 10,000 points each from unit Gaussian distributions for use throughout this work. The size of 10,000 is comparable to the sample sizes we get from our data shards. Effectively, we pretend that our 10,000 points are draws from an unknown distribution. Points in these three samples have dimensionality 810, 334, and 40 respectively. For any given architecture, we use the sample whose dimensionality matches the width of the architecture's input layer.

\paragraph{Stability of our estimators} In general, throughout this work, it is not obvious how stable our estimators are, i.e. to what degree the value they yield depends on the specific sample chosen. The choice of defining metrics in terms of a data or input distribution in the first place implies that we expect that different samples from a practical dataset, as well as hypothetical samples drawn from the corresponding data distribution, all yield approximately the same value, as long as those samples are about as large as the data shards we have available. While conducting our empirical analysis, we spent a significant amount of time and effort confirming the stability of our estimators both theoretically and empirically. For example, we recomputed various metric values with different random seeds and different data shards.

However, there does exist one key condition behind the stability of our estimators for many metrics: Gaussian stability. This is a concept that features prominently throughout this work and is defined in section \ref{gaussianStabilityExplanationSection}. It is a foundation of mean field theory, which we cover in chapter \ref{meanFieldNnaChapter}. In a nutshell, Gaussian stability ensures that, in the majority of architectures we study, the neuron distributions induced by practical input distributions at linear layers are approximately Gaussian. Therefore, neuron value distributions at all layers have favorable statistical properties, including that e.g. their expectation can be estimated accurately from a small sample.

In section \ref{meanFieldPracticalSection}, we derive highly accurate theoretical approximations of metrics based on mean field theory. The metrics depend on the input distribution, but the theoretical approximations do not. Therefore, the influence of the sample on the metric value must be small.

Throughout this work, we demonstrate very strong associations between the values computed for various different metrics. If these values are noisy, then this can at most mean that the associations between the true metric values are even stronger than the associations we demonstrate.

\paragraph{Incorporating data processing and data augmentation}

We used the same data processing that we used for training also for all other computations involving the dataset. We describe our data processing schemes in sections \ref{studyADataSection} and \ref{studyBArchitecturesSection}. On a theoretical level, we assume that the data distribution of a dataset we study is actually the distribution of the processed datapoints. Drawing from this distribution is equivalent to drawing from the original data distribution and then applying the processing function. As mentioned in section \ref{studyADataSection} and further discussed in section \ref{statisticalChallengesSection}, the training and validation set points are then no longer independent of each other because these data shards are used to compute the processing function itself. On a practical level, we simply use the processed data shards as the sample.

We use the same strategy for incorporating data augmentation. When data augmentation is used, on a theoretical level, we assume that drawing from the data distribution is equivalent to drawing from the un-augmented data distribution and then applying the random augmentation function. Of course, we only use data augmentation on the training set in study B. Hence, in practice, only when we use the training set or a subset / bootstrap thereof for estimating a metric in the context of study B, we apply data augmentation to the datapoints, and we do so on the fly before feeding them into the estimator. Hence, the training and test set samples take a slightly different form in the context of study B. This is not a significant issue because data augmentation schemes are designed to capture invariances of the data distribution.

\subsubsection{Floating-point rounding error} \label{metricUnderflowSection}

For each of our metrics, we had to carefully verify whether computing them for specific inputs was naively possible given the presence of rounding error. Thankfully, we were able to compute the vast majority of values accurately with 64-bit precision floating-point computation. We cannot overstate how important this is. If we had been restricted to 32-bit precision throughout our empirical studies, we would have either had to incur an enormous coding overhead by implementing custom routines that reproduce 64-bit computations with 32-bits, or we would have had to severely restrict the space of architectures we study. When conducting study B (section \ref{studyBSection}), we were indeed restricted to 32-bit computation, which did not impact metric computation but did potentially affect training. We discuss this point further in sections \ref{noiseStabilitySection} and \ref{beyondNlcSummarySection}.

There were three situations where catastrophic floating-point error actually arose for us in practice. (i) The standard deviation operator, as defined in the previous subsubsection, was sometimes applied to a quantity with a much lower standard deviation than expectation. This can lead the rounding error to exceed the standard deviation. This occurs for the NLC and related metrics as discussed in section \ref{nlcComputeSection} and LBIAS as discussed in section \ref{outputBiasSection}. (ii) Some metrics require that we evaluate a network for inputs that are only an infinitesimal distance apart. If the actual distance that is chosen when computing a value for these metrics is too large, we do not capture the true behavior of networks over an infinitesimal distance. However, if this distance is chosen too small, then the rounding error introduced at any stage of the network evaluation can corrupt the result. This occurs for our GLLAD, MGLLA, MES and MGLLAHE metrics as discussed in sections \ref{nlcSensiSection}, \ref{nlcKernelSection} and \ref{hessianSection}. (iii) Some metrics require an iterative computation at an unstable fixed point. Even a tiny rounding error can explode over some number of iterations. This occurs for our mean field metrics as discussed in section \ref{meanFieldPracticalEmpiricalSection}. 

There are other potential sources of catastrophic rounding error. For example, when summing a large number of values in 32-bit precision, we have to ensure not to add values to the sum one after the other in a naive fashion. In study A, we did not have to worry about this as we used 64-bit precision. In study B, we used Tensorflow. While we don't assume that Tensorflow actually caused additional cases of catastrophic rounding due to suboptimal implementation when e.g. summing the entries of a tensor, we weren't able to verify this fact. We had to trust in Tensorflow.

\subsubsection{Metrics and batch normalization} \label{metricBNsection}

An additional layer of complexity arises because of batch normalization. This is because BN violates our basic definition of networks as functions mapping single inputs to single outputs. This assumption is baked into the $f(x)$ notation we employ. Consider again the true error $E_\text{true}(f, e, \mathcal{D}) = \mathbb{E}_{(x,y) \sim \mathcal{D}} e(f(x),y)$. If $f$ depends not just on a single input, but also on other inputs in the same batch, this definition does not technically apply. We discussed BN and its surrounding issues in significant detail in section \ref{batchWiseLayersSection}. One possible way around the problem of batch dependency that we discussed in that section is to replace the mean and standard deviation over the batch in the definition of BN with the mean and standard deviation over the training set. This is commonly done when e.g. computing test error. As we explained in section \ref{batchWiseLayersSection}, we do not do this because we want our metrics to capture how our architectures behave during training. 

It turns out that there is a trick that allows us to incorporate BN relatively seamlessly. Let $\mathcal{D}^\text{batch}$ be the distribution over batches of given size, where each point in the batch is drawn from $\mathcal{D}$ independently of the other points. If we view $f$ as a function mapping batches of inputs to batches of outputs, and we view the loss and error functions as the mean of their pointwise values across the batch, all metrics we define throughout this work can be applied directly to the BN case. In practice, we do generalize each metric using this trick, though we also make additional modifications. All of our generalized definitions have the property that if they are applied to an architecture not containing BN, they revert to the original definition for the BN-free case that is given explicitly.

For the purpose of statistical estimation, we can easily obtain a sample point for $\mathcal{D}^\text{batch}$ by drawing a batch from any data shard without replacement. Of course, the downside is that, if we draw multiple batches, those batches may contain the same datapoint and thus they may not be independent draws from $\mathcal{D}^\text{batch}$. This is a benign issue.

Because we also use batches for the purpose of forward propagation in the BN-free case, it turns out that we do not have to alter any of our programs that compute metric values in the BN-free case for the purpose of accommodating BN. Such is the power of the $\mathcal{D}^\text{batch}$ trick.

Finally, we note that we use the $\mathcal{D}^\text{batch}$ trick not just to enable metric value computations for architectures with BN, but other computations as well. For example, it enables us to extend the loss function we use for study A (section \ref{studyATrainingSection}) to the BN case.

\subsection{Metric terminology, conventions and presentation} \label{metricTerminologySection}

\paragraph{4-fold overloading: function / estimator / program / value} In section \ref{neuralNetworkNotationSection}, we explained how the concept of a neural network $f$ has five different aspects. It is a mathematical function, a program that implements a close approximation to that mathematical function, a graph over layers, the vector value that it outputs, and the distribution over that vector value when associated with an input distribution. We explained how we overload our notation and terminology to simultaneously refer to these five aspects.

In the same fashion, any of our metrics has (up to) four different aspects. It is (i) a mathematical function as we define it in the text, (ii) (if one of its inputs is a data or input distribution) a statistical estimator for that function, (iii) a program implementing the function or estimator and (iv) a scalar or distribution value. As with networks, we overload our notation and terminology to simultaneously refer to all four aspects.

Throughout the remainder of this work, the use of statistical estimation when computing metric values remains implicit, except when we state which data shard the sample for the estimator was taken from. We will say ``The value of the metric evaluated on the training set was ...''. We discussed how we conduct estimation in detail in section \ref{metricEstimationSection}. We use canonical programs to implement our metrics and estimators. We discuss these programs only whenever our results were affected by catastrophic floating-point rounding error as discussed in section \ref{metricUnderflowSection}. When metrics have a network or architecture as input, we define and discuss the metric under the assumption that the network maps single inputs to single outputs. When we want to evaluate the metric on a network or architecture with BN, we generalize the metric as described in section \ref{metricBNsection}. Except for the NLC, this generalization remains implicit. We re-iterate that values for these generalized metrics are computed using the exact same programs as values for the corresponding un-generalized metrics, and that the generalized definition reverts to the original definition for networks or architectures that do not use BN. We discuss all the above issues in detail for the NLC, our most important metric besides error / loss, in section \ref{nlcComputeSection}.

\paragraph{Shorthands and defaults} Many of our metrics have a network or architecture as input. Whenever a metric is a function of a network, we can equivalently define it as a function of an architecture and parameter. While we generally use the former option for brevity, the two are used interchangeably throughout this work.

Often, we will use shorthands to refer to values of metrics that have a network or architecture input. We will say things like ``The value of the NLC of $f$ is ...''. When we do this, the following conventions and defaults apply for the other inputs of the metric.

\begin{itemize}
\item The error function $e$ defaults to classification error.
\item The loss function $\ell$ defaults to regular softmax+cross-entropy for study B and our augmented version of softmax+cross-entropy used for study A otherwise.
\item The data distribution $\mathcal{D}$ defaults to the hypothetical data distribution of the dataset used to train the architecture on which the metric was computed. If the architecture falls outside of study A or B and was not trained, the default is the data distribution of CIFAR10. By default, the sample used for statistical estimation is taken from the training set if the metric is computed on a randomly initialized parameter value, and from the test set otherwise.
\item The parameter value defaults as follows. When we reference the metric value for an architecture in the ``initial state'' or ``before training'', we mean that the parameter value was drawn from the initialization scheme. Unless otherwise stated, the specific parameter value used was as follows. For study A architectures, we use the parameter value that was used to begin each training run for that architecture. For study B architectures, we use the parameter value that was used to start the training run that used the starting learning rate chosen by hyperparameter tuning as described in section \ref{studyBTrainingSection}. For experiments outside of studies A and B, if the architecture was trained, we use the same initial parameter value used for training. If the architecture was not trained, we simply sample a fresh parameter value. If we sample multiple parameter values in this way, we make it explicit. 

When we reference the metric value for an architecture in the ``final state'' or ``after training'', the parameter value is the one obtained after training with the best starting learning rate chosen by hyperparameter tuning as described in sections \ref{studyATrainingSection} and \ref{studyBTrainingSection}. Hence, metric values for the initial and final state always use parameter values that belong to the same training run.

In the context of training error minimization, metric values in the initial and final state use the parameter values from the training run that yielded the least training error. If the architecture was fully-connected, we used a completely different set of training runs for training error minimization as described at the end of section \ref{studyATrainingSection}. Whenever we give results from training error minimization in any graph, we display the {\bf train-opt} marker above it. See e.g. figure \ref{nlcPredTrainInit}.
\end{itemize}

For example, when we quote a value for ``the NLC of $f$ before training'' we imply the default data distribution, sample and random parameter value as described above, and we imply that the value was computed via the estimators and program as described earlier in this section. When we deviate from the defaults, we say e.g. ``the NLC of $f$ evaluated on the validation set'' if the sample stems from the validation set. 

\paragraph{Metric values before and after training} Throughout this work, we focus mainly on analyzing the properties of architectures in the initial state, rather than the final state. This is because one of our core goals is to improve architecture design without the need for training. We want to enable readers to understand and predict the performance and behavior of architectures by examining the initial state, not the final state.

For some architectures, no training run achieved an error that we deemed better-than-random. The threshold for this was 0.5 for waveform-noise and 0.8 for CIFAR10 and MNIST. Many of our metrics depend critically on starting learning rate when evaluated after training. If no starting learning rate yielded better-than-random performance, there was no meaningful way to choose the best starting learning rate. Hence, we cannot meaningfully choose a final state for evaluating metrics. Hence, whenever we reference values for any metric not based on error in the final state, we restrict ourselves to architectures that achieved a better-than-random error on at least one training run. If we selected the starting learning rate based on validation error, we determined better-than-random-ness also based on validation error. The same goes for training error / test error.

\paragraph{Conventions of notation} Throughout this work, we define metrics using the {\bf metric definition} environment. We have used this environment already to define error and loss in section \ref{machineLearningSection}. Metrics are denoted by ``labels'' consisting of (usually more than one) capital letter. Their inputs are given in parentheses. An exception to this is when those inputs are layer indices or activation functions. In that case, they are denoted as subscripts. As an example, when a metric $MET$ takes as input an architecture $f$, a parameter $\theta$ and a layer index $l$, we write $MET_l(f,\theta)$. When we first introduce and define a metric, we always denote all its inputs explicitly. However, later on, we pick and choose which inputs we denote explicitly based on the situation. All of this is equivalent to how we use $f$ to denote a network or architecture function. See section \ref{neuralNetworkNotationSection} for details. When we do not state metric inputs explicitly, defaults apply as given above.

\paragraph{Scatter plots} We present our results most often in the form of scatter plots, where each marker generally corresponds to a single architecture and each axis depicts the value of a metric of that architecture. We sometimes give the correlation value as well as the statistical significance of the correlation at the top of the graph. If a metric is depicted in log scale, the logarithm of that metric is also used to evaluate this correlation and significance. Each axis is labeled with the metric depicted on it. Above the graph, we state which architectures are depicted and which dataset is used for training and metric computation. `CIFAR10 - FC' refers to study A architectures trained on CIFAR10. `MNIST' / `waveform-noise' refers to study A architectures trained on the respective dataset. `CIFAR10 - Conv' refers to study B architectures. We use the {\bf train-opt} marker above a graph when the initial and / or final state considered stems from training error minimization.

When metric values evaluated on architectures in the final state are depicted, such as test error, by default, training is conducted using the protocols given in sections \ref{studyATrainingSection}, \ref{studyBTrainingSection} and \ref{additionalExperimentsSection}. When those protocols are altered, the specific alteration is given in parentheses above the graph.

Sometimes, we plot a metric value corresponding to one architecture on the x-axis and a metric value corresponding to another architecture on the y-axis. Or, we plot metric values corresponding to two final states obtained from two different training protocols. In those cases, we use a label of form ``(X vs. Y)'' above the graph, where X and Y describe the difference in architecture / protocol. In this context, ``original'' refers to the default architecture / protocol as given in this chapter. Further, on each axis, we give the architecture / protocol used for that axis after a colon.

Sometimes, we plot the difference or ratio of a metric value between two architectures or two training protocols on one axis. Again, we use a label of form ``(X vs. Y)'' above the graph. On the axis in question, we specify which difference / ratio is used after a colon.

Whenever a metric value was unavailable due to rounding error, we often simply do not plot that value. However, for the NLC and LBIAS metrics, it was possible to derive reasonable approximations for missing values. Oftentimes, we do plot the approximations in our scatter plots. See section \ref{nlcComputeSection} for more information.

Sometimes, we found that multiple markers in our scatter plots were in (almost) the exact same location. When it is important that these markers be individually identifiable for the purpose of interpreting the graph, we added tiny random perturbations to these markers. The following things were always true about situations where we found this to be necessary: (i) The graphs depicted the training or test error of the architectures represented by the markers on the x- or y-axis. (ii) The architectures were always study B architectures trained on CIFAR10. (iii) The overlapping markers always corresponded to architectures with an error close or equal to 0.9, which corresponds to random performance. (iv) We added random perturbations to any axis value of the overlapping markers that depicted error.

In general, we vary the range of the x- and y-axes in our scatter plots so that as much of the plot area as possible is utilized. However, we also strive to keep the axis ranges the same for similar plots, to enhance comparability. Please check the x- and y-axis ranges, especially when comparing plots.

\paragraph{Gaussian stability} Throughout this work, we study the concept of Gaussian stability, which is a major driver of (lack of) performance and many other behaviors. See e.g. sections \ref{meanFieldDistributionSection}, \ref{gaussianStabilityExplanationSection} and \ref{gaussianStabilitySection}. We do not define Gaussian stability via a concrete metric. For the purpose of our empirical analysis, we say that an architecture is a `Gaussian unstable architecture' (GUA) if it falls in one of the following categories. (i) The architecture is fully-connected, based on the square or odd square activation function, and uses BN. (ii) The architecture is convolutional and uses an activation function based on square or odd square. (iii) The architecture is convolutional, uses an activation function based on abs. val., and $l < 0.8$ holds for the linearization parameter. Further, we say that an architecture is a `Gaussian edge architecture' (GEA) if it is fully-connected, is based on ReLU and does not use LN. We explain these choices in section \ref{gaussianStabilityExplanationSection} and give the full list of GUAs and GEAs in the appendix in chapter \ref{fullListChapter}. We stress that our designation of GUAs and GEAs is merely for the purpose of simplicity and empirical evaluation. We do not advocate for criteria (i) through (iii) above to be the definition of the actual phenomenon of Gaussian stability / instability. 

In many figures in this work, markers that correspond to GUAs are displayed in green and markers that correspond to GEAs are displayed in red (e.g. figures \ref{nlcDataGaussInit}, \ref{mfPredfq}). In all but two figures, which are figures \ref{nlcDecomposableInit} and \ref{nlcDecomposableFinal}, GUAs and GEAs are also ``displayed in the foreground''. Whenever a green or red marker corresponding to a GUA or GEA overlaps with a (usually black) marker corresponding to another architecture, the green or red marker is fully visible while the other marker is partially or fully occluded. It is important to note that not all of our figures use green or red markers. A lack of such markers does not mean the GUAs or GEAs are excluded from the figure. It simply means that they do not behave differently from other architectures, so there is no need to visually distinguish them.

\section{Limitations of empirical studies} \label{limitationsSection}

There were a number of limitations that affected our empirical studies, which we discuss in this section. Some of these limitations are intrinsic to the type of analysis we conducted. Others are specific to the conditions under which this work was conducted, i.e. limited time, code and computational resources. Each of the following subsections discusses a broad limitation. Discussions about specific limitations that affect only a small number of metric values are dispersed throughout the work.

\subsection{Limitations of architecture choice: breadth vs relevance} \label{architectureLimitationsSection}

Throughout chapter \ref{introductionChapter}, we stressed that one of our core objectives is to develop guidelines that are universal and reach beyond popular designs. Yet, in section \ref{studyAArchitecturesSection} and \ref{studyBArchitecturesSection}, we stated that we conducted our empirical studies specifically on architectures that relatively closely resemble popular architectures. How do we reconcile this apparent contradiction?

The simple answer is that, like any study, we only had access to a finite amount of computational resources. Hence, we had to validate our results on a set of architectures that would inevitably be a tiny subset of the set of all possible architectures. If we can only validate results empirically on {\it some} set of architectures, we might as well do it on a set of architectures built from popular design strategies. Since there are a large number of factors influencing architecture performance, it makes sense to begin by focusing on those that are most relevant to popular architectures. We believe there is great value in explaining the majority of performance variation for architectures built from popular strategies, as we do in chapter \ref{beyondNlcChapter}. Finally, we note that the vast majority of all neural architectures perform like random guessing. Generating architectures that have a good chance of succeeding is the best way to ensure that performance variation is observed at all.

The key to generalizing our empirical results beyond popular architectures is to understand {\it why} our results hold on the architectures we study and {\it what} properties of those architectures cause the observed behavior, which corresponds to utility criterion \ref{criterionMeaningful}. We investigate these questions in great detail in this work. By building a scientific theory on top of our individual results, our results explain and reinforce each other. This yields a degree of certainty that extends to architectures on which we did not explicitly conduct experiments.

Despite the limitations on the breadth of architectures we were able to study, we stress that this same breadth went far beyond the vast majority of related studies, and indeed far beyond that of most deep learning studies, period.

\subsection{On the sensitivity of our results to the architecture space} \label{architectureSensitivitySection}

Throughout this work, we investigate questions like ``Is there an association between some metric A and another metric B?''. For example, we investigate whether the NLC before training is associated with test error after training. The empirical validity of such associations is always, perhaps unfortunately, highly dependent on the specific set of architectures across which it is measured. There is absolutely nothing we can do about this dependence.

For example, consider study A. It contains 250 randomly sampled architectures that were trained on CIFAR10. Let's say that we observe that the value of metric A is correlated at a level of 0.99 with the value of metric B across those 250 architectures. While this is certainly a strong signal, we have to bear in mind that by manipulating even just the frequency of properties like activation function and normalization operation among those 250 architectures, we can change these correlation levels to almost arbitrary degrees. For example, we will find that, throughout this work, architectures that exhibit Gaussian instability represent outliers with respect to many trends. By increasing or decreasing the frequency of certain architecture properties that induce Gaussian instability, we can therefore control the strength of these trends.

Just as in section \ref{architectureLimitationsSection}, the most important remedy to this problem is to understand {\it why} our results hold for the architectures we study and {\it what} properties of those architectures cause the observed behavior. When we can predict what our results would look like on different sets of architectures based on our understanding of deep learning, we are no longer dependent on specific empirical values. A second remedy is to design our random architecture generators without an agenda to elicit specific results. When we developed the experimental protocol for study A as described in section \ref{studyASection}, we did so certainly with some understanding of the field of deep learning, but without a strong expectation about the results we would obtain regarding the concepts we study in this work. For example, the fact that two of our activation functions in study A sometimes cause Gaussian instability while the others do not, and the fact that those activation functions account for approximately $\frac{2}{11}$ of our architectures, happened by pure coincidence. When we designed study B, we did have a strong understanding of concepts, such as nonlinearity, and of architecture properties, such as activation functions. Therefore, we chose some different properties, like data processing, to vary in study B relative to study A.

When designing our random architecture generators, we also decided to exclude some known sources of performance variation. For example, we did not vary the dimensionality of the parameter significantly within either study A or B. We also ensured that all our architectures exhibited scale stability to a significant degree. We always had our skip connections skip 2 macro-layers, which is the popular standard \citep{resNetTrueIdentity,wideResNet}. The reasoning here is that we did not want to ``rediscover'' these factors. Of course, in some sense, this ``inflated our numbers''. For example, if we had studied architectures with a parameter dimensionality of 100 alongside architectures with a parameter dimensionality of 1 million, the NLC would explain a lesser fraction of test error variation since parameter dimensionality would likely explain a large fraction. Again, we stress that we are not primarily interested in empirical values on some specific set of architectures, but in universal conceptual understanding.

\subsection{Limited hyperparameter tuning beyond learning rate} \label{hyperparameterLimitationSection}

While we conduct exhaustive learning rate tuning and emphasize its importance (sections \ref{moduloAlgorithmSection}, \ref{learningRateSection}), we do not generally tune other training hyperparameters. The next most important such hyperparameters are (i) the choice of training algorithm itself and (ii) the loss function. 

We generally consider a single training algorithm per architecture. Without a deeper reason, we use SGD for fully-connected architectures and momentum with a decay rate of 0.9 for convolutional architectures. An exception to this is section \ref{nlcPredictiveSection}, where we also validate our results with Adam. Ultimately, as in section \ref{nlcPredictiveSection}, we do not expect the choice of algorithm, as long as it is one of the few popular algorithms, to significantly impact our results.

For study A, we construct a special augmented version of the softmax+cross-entropy loss function that normalizes the overall magnitude of network outputs to eliminate the confounding impact of softmax+cross-entropy on architecture comparisons, as we mention in section \ref{studyATrainingSection} and investigate in section \ref{forwardStabilitySection}. In study B, we use regular softmax+cross-entropy. Ultimately, we did not further tune the choice of loss function as this is the overwhelming default choice for simple classification, and because we were able to successfully explain our results without reference to the loss function.

\subsection{Lack of ``large-scale'' experiments} \label{largeScaleLimitationSection}

Some strengths of our studies, including the wide range of architectures, the carefulness of selecting the best starting learning rate, and the large number of training iterations, came at the cost of high computational expense. This meant that our experiments were not ``large-scale'', i.e. we had to limit dataset size and layer width. 

Throughout this work, we observe that dataset size and layer width do not impact our results. We provide deep explanations for our results that do not depend on whether experiments are small-scale or large-scale. Therefore, we do not believe that increasing dataset size or layer width further would provide significant benefit to our analysis.

\subsection{Limitations of architecture type} \label{rnnLimitationSection}

We restrict our empirical analysis to deterministic feedforward networks. We do not study recurrent networks, memory networks or dropout, for example. While we do not doubt that our core results extend to those network types and that there are interesting and important extensions to be found, that goes beyond the scope of this work.

\subsection{Limitations of task setting} \label{settingLimitationSection}

We also restrict our empirical analysis to the supervised classification setting and the empirical risk minimization approach. While it is the most popular setting in deep learning, a core feature of the functional-gradient paradigm is that architectures can work in any setting. It would have been nice to verify that the relative performance of our architectures is roughly the same when they are applied to e.g. regression, reinforcement learning, image generation or noisy label predictions. We note that, while we frame our notation and terminology in terms of the prediction setting for brevity and readability, our analysis does not suggest that our results are fundamentally restricted to that setting.

\subsection{Code base limitations} \label{codeLimitationsSection}

Because study B used convolutional networks, its computational cost exceeded that of study A by over an order of magnitude. We are grateful that we had access to a massive amount of computing resources for a certain time period in order to conduct study B. Without those resources, study B would not have been possible. However, we only had access to those resources for a limited time, and we did not have access to the code we used to conduct study A during that time. Hence, we could only implement a limited amount of code to use for study B and had to rely significantly on pre-written code. Because of this, we were only able to compute values for our most important metrics, such as the NLC. While study B was massive in terms of the number of training runs conducted, we did not obtain as many measurements as we would have liked from each training run. Because we also lost access to the code used for study B once study B was completed, we did not have the chance to make further measurements at a later date. Hence, a significant fraction of our results are validated empirically only for fully-connected networks.

Beyond obtaining fewer metric values, as mentioned at the beginning of section \ref{studyBTrainingSection}, we also had to ``cut corners'' in the experimental protocol of study B itself. Since we had to use pre-written code, we had to abide by choices that were baked into that code. Many of these choices, which we view as shortcomings relative to standards we are hoping to set, are already mentioned in section \ref{studyBSection}. They are also referenced when necessary throughout this work. Below is a list of the most important shortcomings.

\begin{itemize}
\item The single most important shortcoming is that all computation was conducted with 32-bit floating-point precision, which potentially hurt training for some architectures. See sections \ref{noiseStabilitySection} and \ref{beyondNlcSummarySection}.
\item We did not have access to a validation set. We had to select the best starting learning rate based on test error rather than validation error. We study the impact of this in section \ref{nlnormBestNLCSection}.
\item We could not use the same parameter value to start all 20 training runs for each architecture. Rather, we sampled a different value from the initialization scheme for each run. Therefore, the performance difference between runs was not entirely due to the difference in starting learning rate. We rely on the high degree of consistency that architectures exhibit from one draw of the initialization scheme to the next for the purpose of our analysis.
\item We had to use a fixed number of training iterations and a fixed number of iterations for each learning rate level. We could not increase the number of iterations dynamically based on whether training was still making progress. We were not able to stop training at the parameter value at which we measured the lowest error.
\item We had to use Tensorflow. Therefore, we were not always able to verify whether catastrophic floating-point rounding error occurred. We had to assume that Tensorflow makes ``smart'' choices, for example when summing a large number of floating-point values, that do not exacerbate rounding error.
\end{itemize}

\subsection{Lack of independent samples for statistical estimation} \label{statisticalChallengesSection}

One of the challenges of this work is to compute accurate values for metrics when we have to rely on statistical estimation. We cover this challenge in detail in section \ref{metricEstimationSection}. Many of our metrics are defined in terms of a data or input distribution. The only option we have for constructing samples for estimating those metrics is to use the datapoints from our datasets. Because our datasets are not large, we have to reuse our datapoints both during the computation of individual metric values, and to conduct multiple computations that depend on each other sequentially. Every time we use a datapoint, information ``leaks'' into the computation. The next time we use that datapoint, it no longer constitutes an independent draw from $\mathcal{D}$ relative to the whole pipeline. Hence, even if we assume that the datasets as originally downloaded from the internet are composed of independent draws from a data distribution, there are factors internal to our pipeline that render this assumption untrue for the samples we actually feed into our estimators. We list various forms of information leakage below.

Overall, we believe the impact of this sample non-independence is ultimately negligible. The types of datapoint reuse we engage in are common in the deep learning community and rarely mentioned.

\begin{itemize}
\item In all our experiments, we utilize data processing. In many cases, we use statistics computed on the training and validation set for this. This causes each datapoint to be normalized based on the value of other datapoints, so inter-dependency is introduced. We at least make sure that datapoints in the test set are not used to normalize other datapoints.
\item When an architecture is trained, the datapoints in the training set are infused into the parameter value. If a metric utilizes the trained parameter value, the training set is no longer an independent sample. Information from the validation set is also infused if early stopping is used. Hence, we never implicitly use any datapoints other than test set datapoints for computing metrics on trained architectures.
\item Some of our metrics, such as the NLC, utilize multiple probabilistic operators over the data or input distribution. In general, we (slightly improperly) use overlapping samples for each sub-estimator. 
\item When computing metrics on architectures that use BN, we do not ensure that batches never re-use the same datapoint. Therefore, batches are not independent draws from $\mathcal{D}^\text{batch}$ as defined in section \ref{metricBNsection}.
\item When we apply data augmentation, we do so on the fly after the batch is drawn from the un-augmented training set without replacement. This introduces some inter-dependency as our batches can never contain two augmented datapoints that stem from the same un-augmented datapoint.
\item Sometimes, we generate a sample by bootstrapping the data shard. This can obviously re-use the same datapoint.
\end{itemize}

\subsection{Hyperparameters and the sharp valley problem} \label{sharpValleySection}

In many places in this work, we wish to find the ``best hyperparameter value''. For example, we wish to determine the least test error attained by an architecture under SGD for any starting learning rate (SLR). In deep learning, computing the minimal value of a metric such as error in the final state, as the initial parameter value, architecture or training protocol varies continuously, is impossible. This is because the gradient $\frac{d\theta^{(T)}}{d\theta^{(0)}}$ is often enormous, as it requires backpropagation through the entire training procedure, which often leads it to explode \citep{initialFinalGradient}. Therefore, the sensitivity of the final error with respect to a hyperparameter that affects either $\theta^{(0)}$, the way $\theta^{(0)}$ is used by the architecture, or the progress of training starting from $\theta^{(0)}$, is also often enormous. While the impact of large changes to the initial condition, like changing every activation function in an architecture or doubling the number of layers, is often predictable, the impact of very small changes is chaotic. Changing the initial parameter value in the 10th significant digit may change the final error in the second significant digit. This means that no matter how many values of e.g. SLR we ``try out'', we cannot estimate how close the lowest error we have observed is to the actual lowest error. If an SLR of 0.7465 induces an error of 0.103 and an SLR of 0.7466 induces an error of 0.104, we cannot guarantee that an SLR of 0.74655 does not induce an error of 0.07, for example. We term this the `sharp valley problem'. Further, even if we were able to compute the true lowest error at great computational expense for a small architecture, that value would be of limited practical utility because it would likely not be feasible to compute the corresponding value for large, real-world architectures.

Thus, in practice, when choosing a grid of hyperparameter values for which to conduct full training runs, we must choose a grid fine enough so that we capture ``meaningful macro-effects'' but not so fine that we end up optimizing over ``chaotic micro-effects''. In the case of SLR, considering a geometric sequence of values with spacing factor between 2 and 10 is a popular choice. This is reflected in our study design. We then choose the best value from this sequence. Throughout this work, we refer to this chosen value as the ``best starting learning rate'' (best SLR) while acknowledging that this phrasing is not technically accurate.

Beyond calibrating the grid, the best way to mitigate the sharp valley problem is independent validation. For example, in study A, when selecting the best SLR for minimizing test error, we actually choose the SLR that yields the least validation error. So if we were ``unluckily lucky'' and found an SLR value that yields a validation error that is not reflective of the validation error in the ``general vicinity'' of that SLR value, there is still a good chance that the test error is not inside a sharp valley.

\section{Summary of empirical studies} \label{metricsSummarySection}

In this chapter, we detailed the experiments used throughout this work for the purpose of empirical analysis and validation. In this section, we summarize (i) the choices made in the design of our empirical studies that were most responsible for obtaining the results we cover in this work; and (ii) the terminology and conventions that are most important for understanding and interpreting our figures, tables and discussion of experimental measurements. We also repeat specific pieces of information from this summary in later chapters where applicable.

\paragraph{Study A (section \ref{studyASection})} We trained 750 randomly generated architectures with SGD and exhaustive learning rate tuning. We trained 250 architectures for each of three datasets: CIFAR10, MNIST and waveform-noise. waveform-noise is from the UCI repository of datasets popular for evaluating fully-connected architectures \citep{selu}. 

\begin{itemize}
\item {\it Architectures}: We generated 750 fully-connected architectures by randomly and largely independently varying depth, weight matrix and bias vector initialization, the activation function used, the normalization operation used, whether the architecture is residual, where the skip connections are located, and the addition weights used by the skip connections. The full list of architectures is given in section \ref{fullListA}.
\item {\it Activation functions:} An architecture's activation function is of form $\tau(s) = c\dot{\tau}(ds+h)+b$, where $\dot{\tau}$ is an activation function from table \ref{actFunIllu} and $d,h,c,b$ are fixed constants. $d$ is set to 1, 1.2 or 0.8. $h$ is set to 0, 0.2 or -0.2. Finally, we set $(b,c)$ jointly using two constraints. With a 50\% probability, the first constraint is $b=0$ and with a 50\% probability, it is $\mathbb{E}_{s\sim \mathcal{N}(0,1)}c\dot{\tau}(ds+h)+b = 0$. The second constraint is always $\mathbb{E}_{s\sim \mathcal{N}(0,1)}(c\dot{\tau}(ds+h)+b)^2 = 1$. Using $\mathbb{E}_{s\sim \mathcal{N}(0,1)}c\dot{\tau}(ds+h)+b = 0$ corresponds to `activation function debiasing'.
\item {\it Weight initialization:} Weight matrices were LeCun orthogonally initialized and then further scaled by a factor that was 1.0, 0.9 or 1.1. By not deviating too far from the LeCun variance and by using the constraint $\mathbb{E}_{s\sim \mathcal{N}(0,1)}(c\dot{\tau}(ds+h)+b)^2 = 1$, we controlled scale stability (section \ref{architectureDesignParadigmsSection}).
\item {\it Depth}: We define depth as the number of macro-layers $M$. This was chosen uniformly from the set of odd integers between 3 and 49, i.e. $\{3, 5, 7, .., 47, 49\}$.
\item {\it Width:} Width was chosen deterministically as a function of depth so that the dimensionality of the trainable parameter was around 1 million. This controls the performance driver that is parameter dimensionality (section \ref{widthSection}).
\item {\it Normalization layers:} Architectures used either layer normalization, batch normalization or no normalization layers. When skip connections or an activation function based on square or odd square is used, we always use normalization layers to control scale stability. 
\item {\it Learning rate:} We tuned the starting learning rate by independently training each architecture 40 times with different starting learning rates from the same initial state. The smallest starting learning rate considered varies between architectures. It was approximately inversely proportional to the magnitude of the parameter gradient. The ``best starting learning rate'' was selected based on the error on a held-out validation set. During each training run, we decreased the learning rate 10 times by a factor of 3 and then terminated the run whenever the validation error stopped improving. Hence, we used `early stopping'. We study the importance of learning rate tuning in section \ref{learningRateSection}.
\item {\it Floating-point precision}: We conducted all computation associated with study A in 64-bit floating-point precision. We analyze the importance of this in section \ref{noiseStabilitySection}.
\item {\it Loss function:} We used an augmented version of softmax+cross-entropy as the loss function. After initializing each architecture, we evaluated the quadratic mean of the output layer neuron values on the training set: $\sqrt{\frac{1}{d_\text{out}}\mathbb{E}_{(x,y) \in D_\text{train}} ||f(\theta^{(0)},x)||_2^2}$. We then had the loss function divide the network outputs by this fixed scalar value before feeding them into the softmax+cross-entropy operation. This controls the impact of the loss function on performance via the output magnitude (section \ref{forwardStabilitySection}).
\item {\it Re-training / training error minimization:} We re-trained the CIFAR10 and waveform-noise architectures without early stopping based on validation error, with the aim of determining the least training error that could be achieved. We refer to this as ``training error minimization''. We tuned the starting learning rate by independently training each architecture 60 times with different starting learning rates from the same initial state as before; and then selecting based on the final training error.
\end{itemize}

\paragraph{Study B (section \ref{studyBSection})} We trained 552 partially randomly generated architectures with momentum and exhaustive learning rate tuning on CIFAR10. Many of the differences compared to study A arose because we only had a limited time to work with the code base and compute cluster used to conduct study B (section \ref{codeLimitationsSection}). We did not have the chance to replicate many experimental results obtained for fully-connected architectures on convolutional architectures.

\begin{itemize}
\item {\it Architectures:} We deterministically varied activation function, the normalization operation used, and whether the architecture was residual. We additionally varied the following at random: weight initialization scheme, whether bias and elementwise multiplication layers were used, data processing scheme, whether data augmentation was used, the type of pooling layer used, and whether a global average pooling layer was used. (For convenience, we consider data processing and data augmentation as part of the architecture definition in this study, as they were varied randomly along with architecture properties.) Each of these properties was sampled independently of the others. The full list of architectures is given in section \ref{fullListB}.
\item {\it Activation function:} An architecture's activation function is of form $\tau(s) = c\ddot{\tau}(l,s) + b$, where $l,c,b$ are fixed constants. (Note that when we don't use the letter $l$ as a subscript, it generally does not denote a layer index as it does in the expression $f_l$.) $\ddot{\tau}(l,s)$ is an activation function augmented with a `linearization method'. A linearization method is a method for interpolating an activation function with a linear function. We abbreviate the concept of (activation function, linearization method) pair as `AFLM'. In table \ref{AFLMillu}, we depict all AFLMs used in the study. The `linearization parameter' $l$ indicates how close to a linear function $\ddot{\tau}$ is. We will formalize this intuitive concept in later chapters. $l$ always has a `default value' that makes the AFLM revert to a basic activation function from table \ref{actFunIllu}.

Each of the 12 AFLMs in table \ref{AFLMillu} is used with 12 different triplets of values $(l,c,b)$ in the study. See section \ref{fullListB} for details. In the first triplet, $l$ is set to the default value, $b$ is set to 0 and $c$ is set to achieve $\mathbb{E}_{s \in \mathcal{N}(0,1)}\tau(s)^2 = 1$. Across the other 11 triplets, $l$ varies. This causes $\tau$ to be more or less linear. Roughly, these 11 triplets correspond to 0\% linearization, 10\% linearization, etc. up to 100\% linearization. $(b,c)$ are set jointly to achieve $\mathbb{E}_{s \in \mathcal{N}(0,1)}\tau(s)^2 = 1$ and $\mathbb{E}_{s \in \mathcal{N}(0,1)}\tau(s) = 0$. As before, $\mathbb{E}_{s \in \mathcal{N}(0,1)}\tau(s) = 0$ corresponds to activation function debiasing.
\item {\it Weight initialization:} As in study A, we choose the initial weight variance along with $\mathbb{E}_{s \in \mathcal{N}(0,1)}\tau(s)^2 = 1$ to achieve scale stability.
\item {\it Depth}: Depth was fixed to 20 and not varied in order to control the spatial frequency composition of the output \citep{meanFieldCNN}.
\item {\it Layer size:} Layer size was fixed to control parameter dimensionality.
\item {\it Normalization layers:} All activation functions not based on the square or odd square AFLM are used in four architectures: (i) a non-residual architecture not using normalization layers, (ii) a non-residual architecture using batch normalization, (iii) a non-residual architecture using layer normalization and (iv) a residual architecture using batch normalization. The 24 activation functions based on the square and odd square AFLMs are used only in the three architectures (ii) through (iv) to prevent scale instability. All randomly chosen architecture properties always take the same value for all 36 / 48 architectures associated with the same AFLM.
\item {\it Learning rate:} We tuned the starting learning rate by independently training each architecture 20 times with different starting learning rates from different initial states. The 20 values were $3, 1, 0.3, .., 3*10^{-9}, 1*10^{-9}$. The ``best starting learning rate'' was selected based on the lowest test error achieved after training. We study the impact of not having access to a validation set in section \ref{nlnormOverfitSection}. Each training run lasted 100,000 iterations, which is approximately 256 epochs, and the learning rate was divided by 10 after 40,000, 60,000 and 80,000 iterations. If the training error had not decreased below 0.8 after 10,000 iterations, training was terminated.
\item {\it Floating-point precision}: We conducted all computation associated with study B in 32-bit floating-point precision.
\item {\it Loss function:} We used regular softmax+cross-entropy.
\item {\it Training error minimization:} Because early stopping based on validation or test error was not used to begin with, to determine the least training error that could be achieved, we simply selected the best starting learning rate from our original 20 training runs based on the training error after training.
\item {\it Re-training:} For some architectures, after conducting 20 training runs as described above, we conducted additional training runs. We used the starting learning rate selected from the original runs and trained the architecture another 10 times with different random number sequences. Results from this re-training are presented in chapter \ref{nlnormChapter}.
\end{itemize}

\paragraph{Additional experiments (section \ref{additionalExperimentsSection})} A large fraction of our analysis was validated through studies A and B. However, we also conducted additional training runs which utilize a different architecture and / or training protocol. We also analyze some architectures that do not belong to study A or B in their initial state, without training them. Results from experiments that do not fall entirely under study A or B appear in tables \ref{discillu} and \ref{dsetfusion}, as well as figures \ref{nlcAdam}, \ref{nlcMeanVar}, \ref{nlcRandomInit}, \ref{sensiDist}, \ref{nlcWidth}, \ref{mfMetaGaussiankurt}A through \ref{mfMetaGaussianfq}A, \ref{beyondForward}, \ref{beyondFirst}, \ref{beyondDGD}, \ref{beyondNoiseTest}, \ref{beyondNoiseTrain}, \ref{beyondSummary}, \ref{relatedConf} and \ref{relatedOscillation}.

Architectures that do not fall under study A or B are always fully-connected and of a layout similar to study A. We used the same training protocol as in study A, including 64-bit precision, learning rate tuning and augmented loss function, unless stated otherwise. We also used the ``re-training protocol'' for training error minimization.

\paragraph{Metrics, terminology, convention and presentation (section \ref{metricsSection})} Almost all of our empirical analysis in this work is based on `metrics'. We use this term loosely in accordance with section \ref{notationSummarySection} to refer to functions of one or more of the following: neural architecture, neural network, layer, parameter, data distribution, input distribution, dataset, data shard, activation function, input, label, loss function, error function, layer component index. A metric represents a property of those constructs, or a measure for an ill-defined property.

\begin{itemize}
\item {\it Statistical estimation:} Many of our metrics are defined in terms of input or data distributions. We use statistical estimation to compute values for those metrics, where samples are taken from our data shards.
\item {\it Overloading:} Any of our metrics has (up to) four different aspects. It is (i) a mathematical function as we define it in the text, (ii) (if one of its inputs is a data or input distribution) a statistical estimator for that function, (iii) a program implementing the function or estimator and (iv) a scalar or distribution value. We overload our notation and terminology to simultaneously refer to all four aspects. We do the same with neural networks and associated concepts as described in section \ref{neuralNetworkNotationSection} / \ref{notationSummarySection}.
\item {\it Initial state}: When we reference the metric value for an architecture in the ``initial state'' or ``before training'', we imply that the parameter value was drawn from the initialization scheme. When we reference the metric value for an architecture without qualifier, we generally mean the value in the initial state, unless the metric is based on error. Unless otherwise stated, the specific parameter value used is as follows. For study A architectures, we use the parameter value that was used to begin each training run for that architecture. For study B architectures, we use the parameter value which was used to start the training run that used the starting learning rate chosen by hyperparameter tuning. For experiments outside of studies A and B, if the architecture was trained, we use the same initial parameter value used for training. If the architecture was not trained, we simply sample a fresh parameter value. If we sample multiple parameter values in this way, we make it explicit. By default, the sample used for statistical estimation in the initial state is taken from the training set.
\item {\it Final state:} When we reference the metric value for an architecture in the ``final state'' or ``after training'', we imply that the parameter value is the one obtained after training with the best starting learning rate. Hence, metric values for the initial and final state always use parameter values that belong to the same training run. When we reference the value of a metric based on error for an architecture without qualifier, we generally mean the value in the final state. By default, the sample used for statistical estimation in the final state is taken from the test set.
\item {\it Training error minimization:} Under training error minimization, metric values in the initial and final state use the parameter values from the training run that yielded the least training error, as described above. If the architecture was fully-connected, re-training applies. Whenever we display results from training error minimization in any graph, we display the {\bf train-opt} marker above it. See e.g. figure \ref{nlcPredTrainInit}.
\item {\it Discarding randomly performing architectures:} For some architectures, no training run achieved an error that we deemed better-than-random. The threshold for this was 0.5 for waveform-noise and 0.8 for CIFAR10 and MNIST. Many of our metrics depend critically on starting learning rate when evaluated after training. If no starting learning rate yielded better-than-random performance, there was no meaningful way to choose the best starting learning rate. Hence, we cannot meaningfully choose a final state for evaluating metrics. Hence, whenever we reference values for any metric not based on error in the final state, we restrict ourselves to architectures that achieved a better-than-random error on at least one training run. If we selected the starting learning rate based on validation error, we determined better-than-random-ness also based on validation error. The same goes for training error / test error.
\item {\it Scatter plots:} We present our results most often in the form of scatter plots, where each marker usually corresponds to a single architecture and each axis depicts the value of a metric of that architecture. We sometimes give the correlation value as well as the statistical significance of the correlation at the top of the graph. If a metric is depicted in log scale, the logarithm of that metric is also used to evaluate this correlation and significance. Each axis is labeled with the metric depicted on it. Above the graph, we state which architectures are depicted and which dataset is used for training and metric computation. `CIFAR10 - FC' refers to study A architectures trained on CIFAR10. `MNIST' / `waveform-noise' refers to study A architectures trained on the respective dataset. `CIFAR10 - Conv' refers to study B architectures.
\item {\it Gaussian stability:} Throughout this work, we study the concept of Gaussian stability, which is a major driver of (lack of) performance and many other behaviors. See e.g. sections \ref{meanFieldDistributionSection}, \ref{gaussianStabilityExplanationSection} and \ref{gaussianStabilitySection}. We do not define Gaussian stability via a concrete metric. For the purpose of our empirical analysis, we say that an architecture is a `Gaussian unstable architecture' (GUA) if it falls in one of the following categories. (i) The architecture is fully-connected, based on the square or odd square activation function and uses BN. (ii) The architecture is convolutional and uses an activation function based on square or odd square. (iii) The architecture is convolutional, uses an activation function based on abs. val. and $l < 0.8$ holds for the linearization parameter. Further, we say that an architecture is a `Gaussian edge architecture' (GEA) if it is fully-connected, is based on ReLU and does not use LN. We explain these choices in section \ref{gaussianStabilityExplanationSection} and give the full list of GUAs and GEAs in the appendix in chapter \ref{fullListChapter}. We stress that our designation of GUAs and GEAs was merely for the purpose of simplicity and empirical evaluation. We do not advocate for criteria (i) through (iii) above to be the definition of the actual phenomenon of Gaussian stability / instability. 

In many figures in this work, markers that correspond to GUAs are displayed in green and markers that correspond to GEAs are displayed in red (e.g. figures \ref{nlcDataGaussInit}, \ref{mfPredfq}). In all but two figures, which are figures \ref{nlcDecomposableInit} and \ref{nlcDecomposableFinal}, GUAs and GEAs are also ``displayed in the foreground''. Whenever a green or red marker corresponding to a GUA or GEA overlaps with a (usually black) marker corresponding to another architecture, the green or red marker is fully visible but the other marker is partially or fully occluded. It is important to note that not all of our figures use green or red markers. A lack of such markers does not mean the GUAs or GEAs are excluded from the figure. It simply means that they do not behave differently from other architectures, so there was no need to visually distinguish them.

\end{itemize}

\chapter{The Nonlinearity Coefficient (NLC)} \label{nlcChapter}

In this chapter, we begin the process of establishing the nonlinearity coefficient (NLC) as a core metric for neural architecture design and a primary tool for neural network analysis in general. The NLC is a function of a network that, when evaluated in an architecture's randomly initialized state, is predictive of the architecture's test error after training, and that fulfills many other utility criteria as postulated in figure \ref{boxNPM} to a significant degree. Hence, we establish the utility of the NLC for zero-shot architecture design (ZSAD; figure \ref{boxZSAD}). Please see section \ref{nlcSummarySection} for a detailed overview of this chapter as well as related results.

Analyzing the NLC is a process that spans all chapters and the majority of sections of this work from this point forward. In each chapter, we take a slightly different approach. In this chapter, we take a ``one-by-one approach'', i.e. we investigate properties of the NLC in isolation, one after the other, section by section and subsection by subsection. We use conceptually simple experiments and theory to validate results individually. Of course, designing the right experimental protocol and proving the theorems is far from simple. In fact, in order to keep this chapter manageable, we outsource our presentation and discussion of our experimental protocol to chapter \ref{empiricalStudiesChapter} and we provide proofs for the propositions and theorems of this chapter in chapter \ref{finiteNetChapter}. The strength of this chapter is its simplicity, its weakness is that, at least in some parts, we largely rely on empirical results, which depend to some degree on our choice of which neural architectures to study. We discuss this tradeoff in section \ref{architectureSensitivitySection}. Later chapters provide context and explanation for the results of this chapter. For example, in section \ref{nlcExplainSection}, we explain results using mean field theory. In section \ref{beyondNlcSummarySection}, we explain architectures that represent outliers in our figures of section \ref{nlcPredictiveSection}.

The NLC is a measure of the property `degree of nonlinearity' of a neural network, which is intuitively related to the notion of the complexity of a mathematical function. We begin this chapter by asking the fundamental question, ``What is nonlinearity?'' in section \ref{whatIsNonlinearitySection}. In section \ref{nlcDefinitionSection}, we derive the definition of the NLC, and give a theorem that establishes the NLC as a nonlinearity measure. In section \ref{eyeTestSection}, we provide a visual illustration of the meaning of nonlinearity and its relation to the NLC. In section \ref{nlcPropertiesSection}, we demonstrate a range of properties of the NLC, where each subsection corresponds to a different property which is given in the title. In section \ref{bestNlcSection}, we investigate the interplay between the dataset and performance prediction using the NLC. As discussed in section \ref{moduloDataSection}, we aim to develop ZSAD guidelines that are as data-agnostic as possible.

\paragraph{Background from prior chapters} Throughout this chapter, we use the terminology, notation and conventions of section \ref{notationSummarySection}.

\section{What is nonlinearity?} \label{whatIsNonlinearitySection}

To define nonlinearity, we must first understand linearity. Linear functions are a fundamental building block of mathematics, as well as a fundamental building block of machine learning. In the context of prediction, having the output be a linear transformation of the input is highly popular. For example, combining a linear model with a softmax+cross-entropy loss function (section \ref{lossFunctionSection}) leads to logistic regression. Combining a linear model with the $L2$ loss function (section \ref{lossFunctionSection}) leads to linear regression. Combining a linear model with the hinge loss function leads to a support vector machine.

In the first instance, nonlinear simply means not linear. Almost all functions are not linear. Consider functions $F : \mathbb{R}^{d_\text{in}} \rightarrow \mathbb{R}^{d_\text{out}}$. Linearity implies that there exists some $d_\text{in} \times d_\text{out}$ matrix $A$ and $d_\text{out}$-dimensional vector $b$ such that $F(\chi)$ can be written for all $\chi\in \mathbb{R}^{d_\text{in}}$ as $\chi A + b$. $A$ and $b$ only have a combined $(d_\text{in}+1)d_\text{out}$ degrees of freedom. To recognize just how restrictive linearity is, consider that for linear $F$ we have $F(\chi') = F(\chi) + (\chi' - \chi)\frac{dF(\chi)}{d\chi}^T$ for all $\chi ,\chi' \in \mathbb{R}^{d_\text{in}}$. This means that the value of $F$ everywhere is determined by a single function value $F(\chi)$ and a single Jacobian $\frac{dF(\chi)}{d\chi}$. $F$ is determined by its value in an arbitrarily small neighborhood around any point. 

Neural networks are generally not linear, and for good reason. No matter how fancy a network may appear, as long as it is a linear function of the input $x$, it can simply be represented as $xA + b$. To train a linear model, it is sufficient to consider $A$ and $b$ as trainable parameters and apply e.g. a gradient method directly to $xA + b$. Because the expression $xA + b$ is convex in both $A$ and $b$, this is the best way to train a linear model. While linear networks are interesting constructs for theoretical study (e.g. \citet{orthogonalInitialization,eigenspectrum}), they are not practically useful.

In this work, of course, we want to go beyond the binary distinction of ``linear'' and ``not linear''. We are interested in studying the degree of nonlinearity of a network. This concept is not mathematically well-defined. We must therefore come up with a measure that captures this intuitive notion.

\begin{table}[t]
\centering
{
\begin{tabular}{lcccccccc}
Function&(a)&(b)&(c)&(d)&(e)&(f)&(g)&(h)\\ \hline\hline
Illustration&\includegraphics[scale=0.2,valign=c]{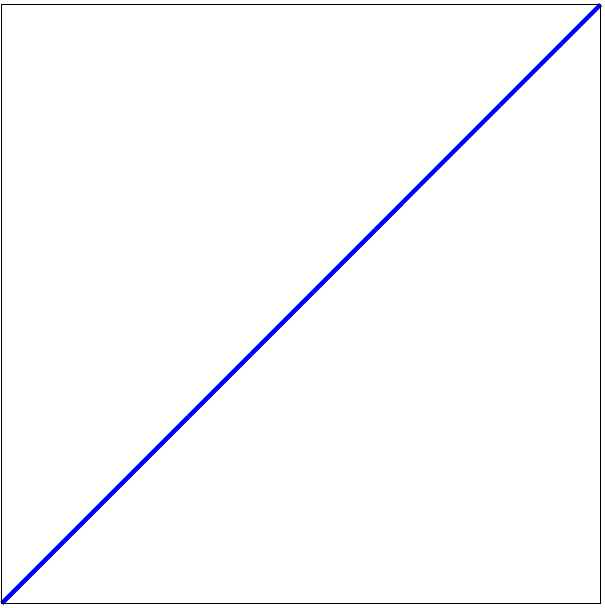}&\includegraphics[scale=0.2,valign=c]{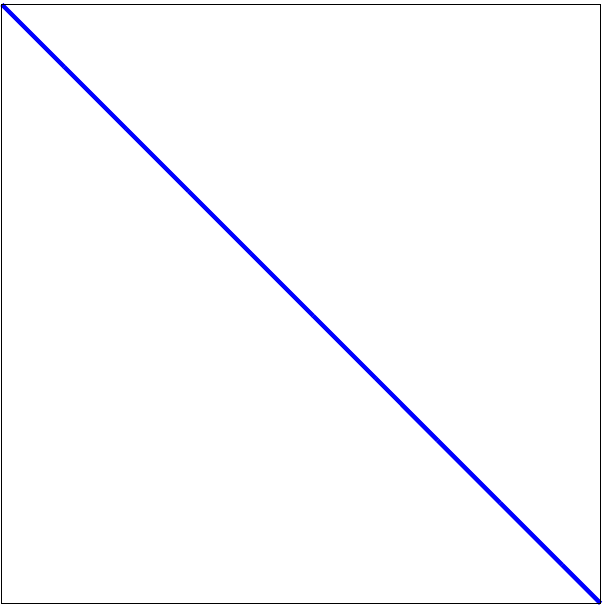}&\includegraphics[scale=0.2,valign=c]{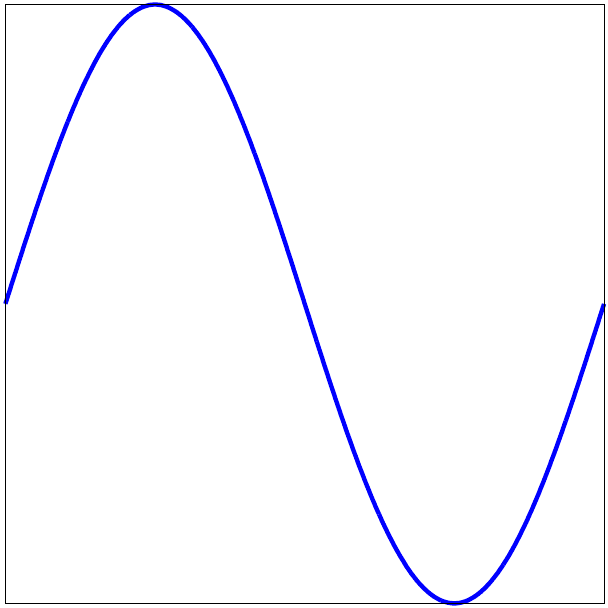}&\includegraphics[scale=0.2,valign=c]{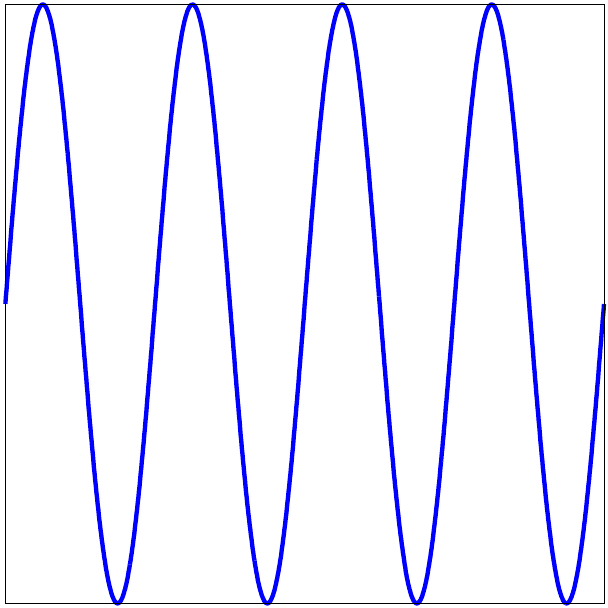}&\includegraphics[scale=0.2,valign=c]{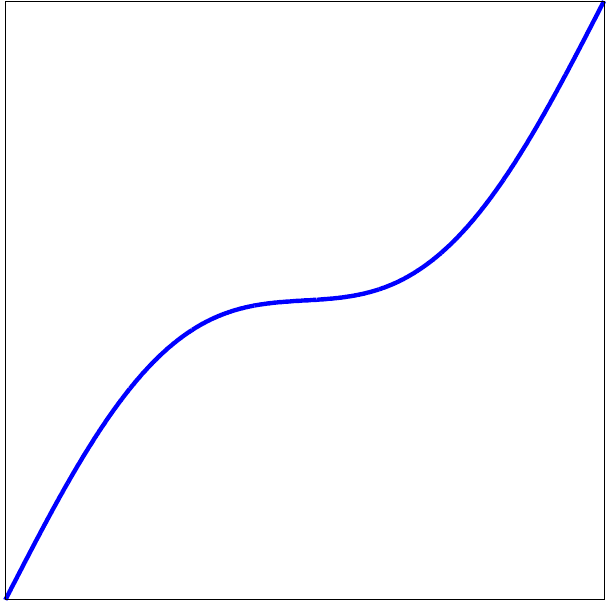}&\includegraphics[scale=0.2,valign=c]{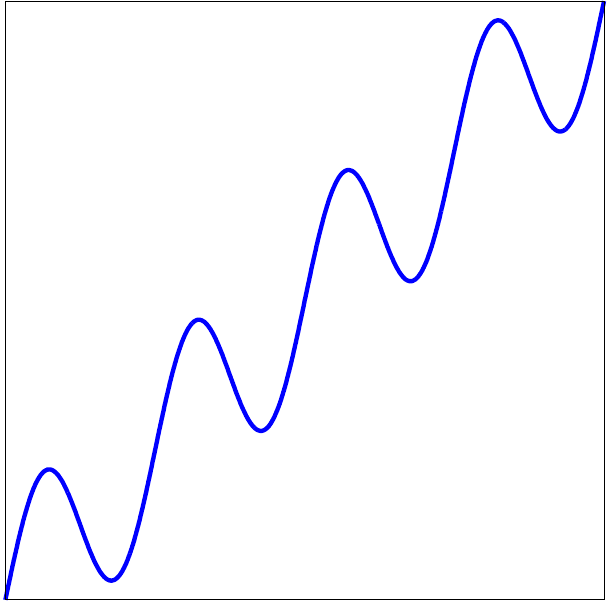}&\includegraphics[scale=0.2,valign=c]{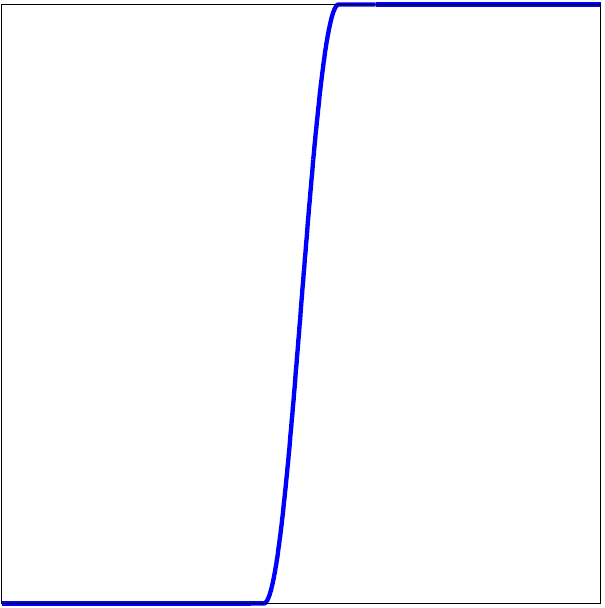}&\includegraphics[scale=0.2,valign=c]{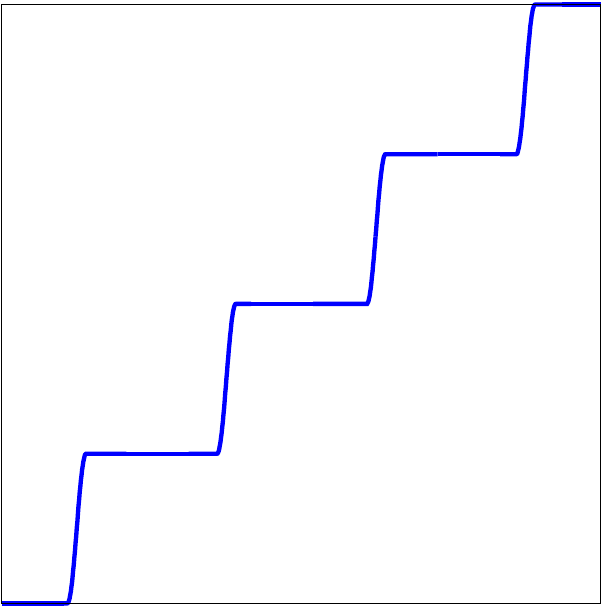}\\
Is linear? & yes & yes & no & no & no & no & no & no\\
$\frac{|\mathsf{dom}|}{|\mathsf{co}|}\mathbb{E}|F'(s)|$&1&1&2&8&1&2.48&1&1\\
$\frac{|\mathsf{dom}|}{|\mathsf{co}|}\sqrt{\mathbb{E}F'(s)^2}$&1&1&2.22&8.89&1.20&2.85&3.14&3.14\\
\end{tabular}
}
\caption{Comparison of scalar functions with regards to nonlinearity. $\mathsf{dom}$ refers to the domain, which spans the range of the x-axis. $\mathsf{co}$ refers to the codomain, which spans the range of the y-axis. $|.|$ applied to an interval denotes interval length. {\it Conclusion:} The expected square derivative is more suitable for measuring nonlinearity than the expected absolute derivative.}
\label{linillu}
\end{table}

Let's begin by investigating the 1-dimensional case. In table \ref{linillu}, we depict eight different scalar functions. It is clear that functions (a) and (b) are linear, while functions (c) through (h) are not linear. However, it is unclear exactly how different each of the functions (c) through (h) are from a linear function. For example, comparing function (f) and (g), we can say that function (g) can be closely approximated by three line segments, whereas function (f) requires many line segments. On the other hand, function (g) changes direction very drastically in two places, whereas function (f) is at least somewhat locally linear. Comparing functions (g) and (h), we can say that (h) can be approximated better by a single line, in the mean square error sense, relative to (g). On the other hand, we can say that the largest interval where the function is exactly linear is larger for (g) than for (h).

We can already see that there is no right answer to the question ``What is nonlinearity?''. If we were to come up with a real-valued metric to measure the degree of nonlinearity of not just scalar functions, but practical neural networks, at best, we could hope to have it (i) be intuitively reasonable and (ii) fulfill the utility criteria of figure \ref{boxNPM} as much as possible. An aspect of the meaningfulness of the metric (criterion \ref{criterionMeaningful}) would then be that it is a measure of nonlinearity.

\section{Deriving and defining the NLC} \label{nlcDefinitionSection}

The NLC is based on the simple insight that linear networks have a constant gradient. In a nutshell, we will say that the more a network's gradient changes direction and magnitude, the more nonlinear it is.

Let's again look at the 1-dimensional case. Let the closed and bounded real interval $\mathsf{dom}$ denote the domain of the differentiable scalar function $F$. Let $\mathsf{co}$ denote the codomain, which we define as the set of values taken by $F$. Let $|.|$ applied to an interval denote interval length. One way to express that linear $F$ have a constant derivative is via the formula $|F'(s)| = \frac{|\mathsf{co}|}{|\mathsf{dom}|}$, which holds for all $s \in \mathsf{dom}$. Hence, we have

$$\frac{|\mathsf{dom}|}{|\mathsf{co}|}\mathbb{E}|F'(s)| = 1$$

We take $s$ to be uniformly distributed over the domain. (For now, we ignore the possibility of dividing by zero.) Further, for any $F$ we have

$$\frac{|\mathsf{dom}|}{|\mathsf{co}|}\mathbb{E}|F'(s)| \ge 1$$

We can see this by considering that equality holds if $F$ is monotonic, due to the fundamental theorem of calculus. The left-hand side effectively measures the number of times the function traverses the codomain.

Now we have a functional, a metric, namely $\frac{|\mathsf{dom}|}{|\mathsf{co}|}\mathbb{E}|F'(s)|$, that is minimized by any linear function. Therefore, it is a potential candidate for measuring the nonlinearity of scalar functions. In table \ref{linillu}, we give the value of this metric for functions (a) through (h). Indeed, all values are at least 1. Unfortunately, as mentioned above, the metric is equal to 1 for all monotonic functions even if they are nonlinear, e.g. functions (e), (g) and (h). This is a fatal flaw. In essence, the metric captures changes in direction of the derivative, but not changes in magnitude. To overcome the flaw, we note that by Jensen's inequality we have

$$p^{-1}(\mathbb{E}p(|F'(s)|)) \ge \mathbb{E}|F'(s)|$$

for any convex, strictly increasing $p : \mathbb{R}_{\ge 0} \rightarrow \mathbb{R}_{\ge 0}$. Further, if $p$ is strictly convex, equality holds if and only if $|F'(s)|$ is constant.  If $F$ is also monotonic, equality holds if and only if $F$ is linear. And the gap between the left- and right-hand side increases the greater the variation in derivative magnitude. 

The simplest and most obvious choice for $p(s)$ is $s^2$. Thus we obtain

$$\frac{|\mathsf{dom}|}{|\mathsf{co}|}\sqrt{\mathbb{E}F'(s)^2} \ge 1$$

where equality holds if and only if $F$ is linear, and the gap increases the greater the variation in derivative magnitude and direction. This gives us another metric, $\frac{|\mathsf{dom}|}{|\mathsf{co}|}\sqrt{\mathbb{E}F'(s)^2}$. Let's call this metric NL1D. As before, we give the value of NL1D in table \ref{linillu}. It is larger than 1 for functions (c) through (h), as desired. All values are indeed intuitively reasonable. We can also immediately observe several intuitively reasonable properties in the table.

\begin{itemize}
\item NL1D is invariant under shifting and scaling. Specifically, the value of the metric for some $F(s)$ on $[L,R]$ is equal to its value for $aF(cs+d)+b$ on $[\frac{1}{c}L - \frac{d}{c},\frac{1}{c}R - \frac{d}{c}]$ for scalars $a,b,c,d$ with $a,c\neq 0$. This is reasonable because shifting and scaling preserves linearity. This invariance is why we left the range of the x- and y-axes undefined in table \ref{linillu}. These ranges are irrelevant for the value of our metrics.
\item NL1D is proportional to the frequency of a periodic function such as a sine curve. For example, compare functions (c) and (d), which differ in frequency by a factor of 4.
\item NL1D is unchanged when a monotonic function is repeated. For example, compare functions (g) and (h).
\end{itemize}

So far, so good. But how could we apply NL1D to neural networks? There are several issues. 

\begin{itemize}
\item NL1D is only defined for scalar functions. Neural networks can have multi-dimensional inputs and outputs.
\item NL1D assumes there is a well-defined bounded domain and codomain. There is no equivalent assumption in the case of neural networks.
\item In NL1D, $s$ is sampled uniformly from the domain. In practice, the input distribution $\mathcal{D}$ may have the majority of its probability mass concentrated in certain regions of input space. We may want our metric to preferentially sample the behavior of $f$ in those regions. 
\end{itemize}

We are now ready to define the NLC as a solution to these issues.

\begin{metricDefinition}

Let $f$ be a neural network and let $x$ be drawn from some input distribution $\mathcal{D}$. Then the `nonlinearity coefficient' (NLC) of $f$ with respect to $\mathcal{D}$ is

$$NLC(f, \mathcal{D}) = \sqrt{\frac{\mathbb{E}_x\Tr(\mathcal{J}(x)\Cov_x\mathcal{J}^T(x))}{\Tr(\Cov_f)}}$$

where $\Cov_x[i,i'] = \mathbb{C}_x(x[i],x[i'])$ is the covariance matrix of the input, \linebreak $\Cov_f[j,j'] = \mathbb{C}_x(f(x)[j],f(x)[j'])$ is the covariance matrix of the output and $\mathcal{J}(x) = \frac{df}{dx}$ is the Jacobian. $\mathbb{C}$ is the standard covariance operator as defined in section \ref{metricEstimationSection}. $\Tr$ is the trace. This follows our standard notation from section \ref{notationSummarySection}.
\end{metricDefinition}

The NLC can be viewed as a multi-dimensional generalization of NL1D. $\sqrt{\Tr(\Cov_f)}$, which is also equal to the square root of the sum of eigenvalues of $\Cov_f$, can be viewed as the radius of the (ill-defined) codomain. Similarly, $\sqrt{\Tr(\Cov_x)}$ can be viewed as the radius of the domain. $\sqrt{\mathcal{J}\mathcal{J}^T} = ||\mathcal{J}||_F$ is the direct generalization of $\sqrt{F'^2}$. Combining $\mathcal{J}$ and $\Cov_x$ in the numerator means that we only consider the gradient of $f$ in directions where $x$ actually varies, which seems desirable.

\paragraph{Technical considerations} From a mathematical standpoint, we need to make the following assumptions for the NLC to be valid.

\begin{assumption} \label{assumptionDifferentiable}
$f$ is differentiable everywhere.
\end{assumption}

This is often not true in practice. For example, networks using the ReLU activation function are at most directionally differentiable everywhere. However, this assumption still turns out to be acceptable, as we argue extensively in section \ref{nonDifferentiableSection}. In practice, we can replace the Jacobian in the NLC with the Jacobian of a local linear approximation of $f$, which is defined in the same way as the gradient of the local linear approximation that is used for training, as we explain in section \ref{howDifferentiableSection}. Such a gradient must be computable for training to be possible. See also section \ref{nlcComputeSection}.

\begin{assumption} \label{assumptionIntegrable}
All expectation operators involved in the definition of the NLC are valid and finite, i.e. several simple expressions involving $x$, $f$ and $\mathcal{J}$ need to be integrable with respect to the input distribution.
\end{assumption}

This is a mild assumption, as we argue in section \ref{integrabilitySection}.

\begin{assumption} \label{assumptionPositive} 
$\Tr(\Cov_f) > 0$
\end{assumption}

This assumption is also very mild, as shown by the proposition below.

\begin{proposition} \label{finiteNetPositiveDenominator}
Assume there exists an open set $S$ where $\mathcal{D}$ has a continuous, positive density function and $f$ is not constant on $S$. Then $\Tr(\Cov_f)> 0$.
\end{proposition}

Finally, we make another assumption that will also greatly simplify notational bookkeeping.

\begin{assumption} \label{assumptionNonSingular}
$\Cov_x$ is non-singular.
\end{assumption}

This assumption is also very mild, as shown by the proposition below. Given any practical input or output distribution with singular covariance, we can simply attain non-singularity via linear projection / change of coordinates.

\begin{proposition} \label{finiteNetNonSingular}
Assume there exists a non-empty open set $S$ where $\mathcal{D}$ has a continuous, positive density function. Then $\Cov_x$ is non-singular.
\end{proposition}

As a general rule, going forward, we will not address technical issues, such as integrability, differentiability, dividing by zero and singularities, in the predominantly empirical parts of this work, as these issues generally work out in a straightforward manner. We explain this with respect to differentiability in section \ref{nonDifferentiableSection} and with respect to integrability in section \ref{integrabilitySection}. We will reserve technical discussions for the theoretical chapters \ref{meanFieldNnaChapter}, \ref{finiteNetChapter} and \ref{mfntChapter}.

\subsection{The NLC builds upon the functional-gradient paradigm} \label{nlcFunctionalGradientSection}

The mildness of the assumptions required for the NLC is one of its core strengths. It underpins utility criteria \ref{criterionWellDefined} and \ref{criterionGeneral}. The strongest assumption is differentiability.

In section \ref{functionalGradientSection}, we outlined the functional-gradient paradigm, which has emerged as dominant in machine learning. It is based on black-box functions with three key properties as given in figure \ref{boxFunctionalLearning}. One of these properties is that a local linear approximation in a region around the parameter value can be found which is suitable for gradient updates. But this is precisely what is necessary to evaluate the NLC, as explained above. Hence, the NLC builds directly upon the functional-gradient paradigm and thereby inherits all of its advantages, including its full generality.

\subsection{The NLC is a nonlinearity measure} \label{nlcNonlinearityMetricSection}

What makes NL1D a nonlinearity measure is that it attains its minimum exactly for linear functions. It turns out the NLC generalizes this property to multi-dimensional networks with Gaussian inputs.

\begin{theorem} \label{finiteNetNlcGreater1}
Let $\mathcal{D}$ be Gaussian. Then $NLC(f, \mathcal{D}) \ge 1$, where equality holds if and only if $f$ is linear.
\end{theorem}

The downside of this theorem is that the NLC can be as small as zero if $\mathcal{D}$ is not Gaussian. This happens, for example, when the support of $\mathcal{D}$ has several connected components and $f$ is constant on each component, but takes different values on different components. In general, we might say that a downside of the NLC is that it is technically not a nonlinearity measure for networks $f$, but for pairs $(f,\mathcal{D})$. However, in section \ref{nlcRobustDataSection}, and then further in chapter \ref{meanFieldNnaChapter}, we generate highly accurate approximations for the NLC using Gaussian inputs that mimic only the expectation and covariance of practical input distributions $\mathcal{D}$. In section \ref{nlcRobustDataSection} we also show that the dependency on expectation and covariance is necessary for any reasonable nonlinearity measure.

Even in the non-Gaussian regime, we have several valuable properties.

\begin{proposition} \label{finiteNetNlcEquals1}
Let $f$ be linear. Then $NLC(f,\mathcal{D}) = 1$.
\end{proposition}

\begin{proposition} \label{finiteNetNlcOrthoMatrix}
Let $A$ be an orthogonal matrix of size $d_\text{out} \times d_\text{out}$ and $b$ be a $d_\text{out}$-dimensional vector. Then $NLC(fA + b,\mathcal{D}) = NLC(f,\mathcal{D})$.
\end{proposition}

\begin{proposition} \label{finiteNetNlcAnyMatrix}
Let $A$ be a matrix of size $d_\text{in} \times d_\text{in}$ and $b$ be a $d_\text{in}$-dimensional vector. Assume $\Tr(\Cov_{f(xA+b)}) > 0$. Then $NLC(f(xA+b),\mathcal{D}) = NLC(f(x),\mathcal{D}A+b)$, where drawing $x$ from $\mathcal{D}A+b$ is equivalent to drawing $x$ from $\mathcal{D}$ and applying $xA + b$.
\end{proposition}

We develop the mathematics behind the theorems and propositions of this chapter and provide proofs in chapter \ref{finiteNetChapter}.

\section{An eye test} \label{eyeTestSection}

\begin{table*}[t]
\centering
{
\begin{tabular}{lcccc}
depth & 1 & 10 & 25 & 50\\
 \hline\hline
NLC&1.0&1.4&2.7&7.6\\
Illustration&\includegraphics[width=0.2\textwidth,valign=c]{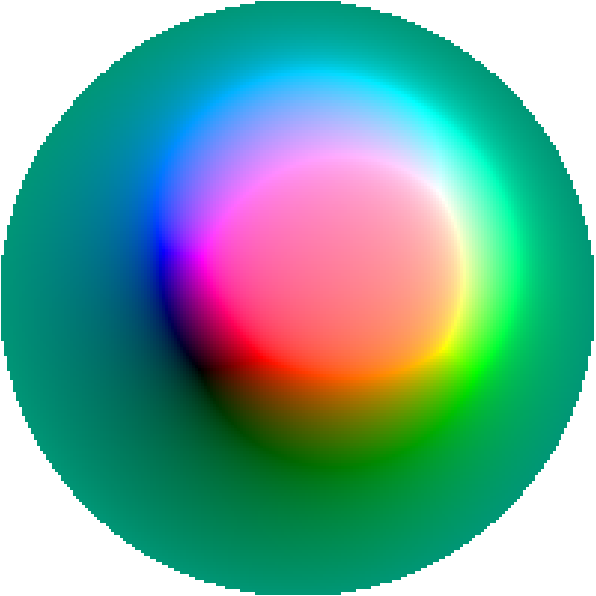}&\includegraphics[width=0.2\textwidth,valign=c]{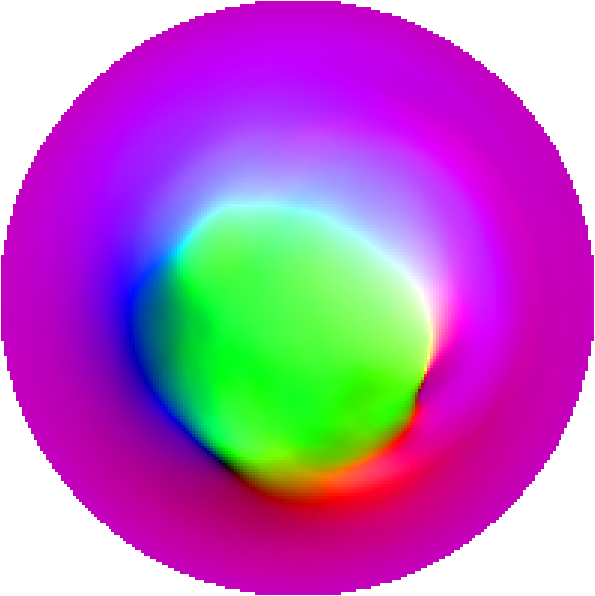}&\includegraphics[width=0.2\textwidth,valign=c]{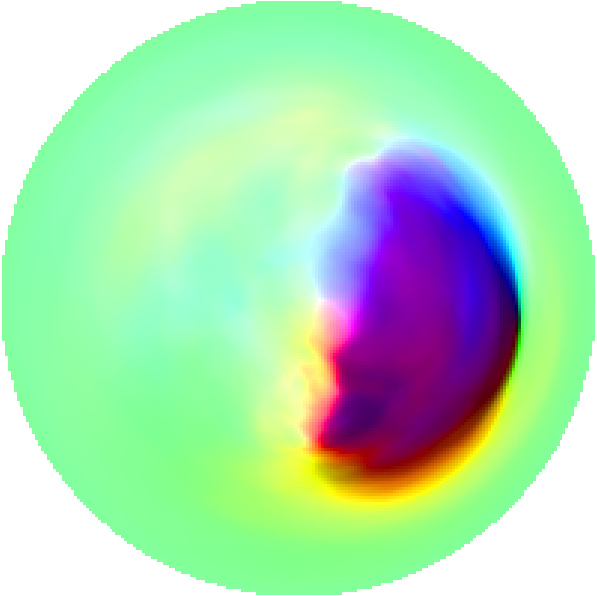}&\includegraphics[width=0.2\textwidth,valign=c]{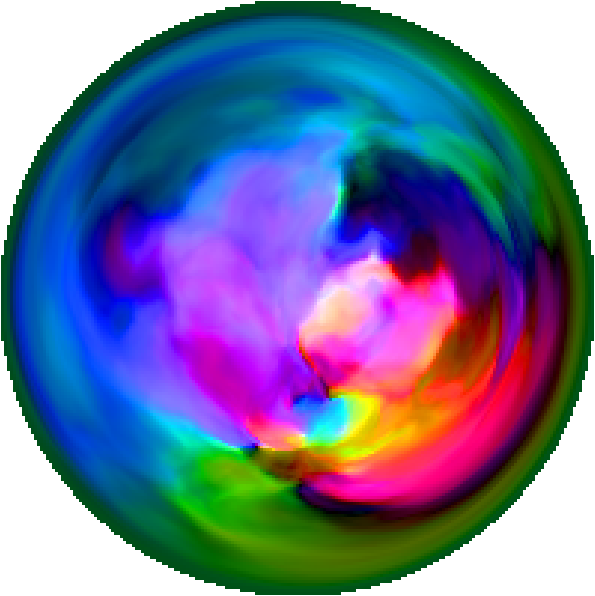}
\end{tabular}
\begin{tabular}{lcccc}
\\
depth & 75 & 100 & 150 & 200\\
 \hline\hline
NLC&1.7e1&4.9e1&3.7e2&2.2e3\\
Illustration&\includegraphics[width=0.2\textwidth,valign=c]{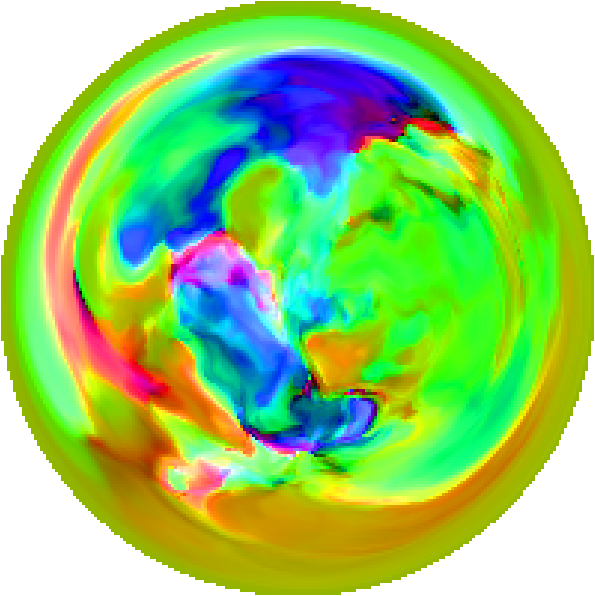}&\includegraphics[width=0.2\textwidth,valign=c]{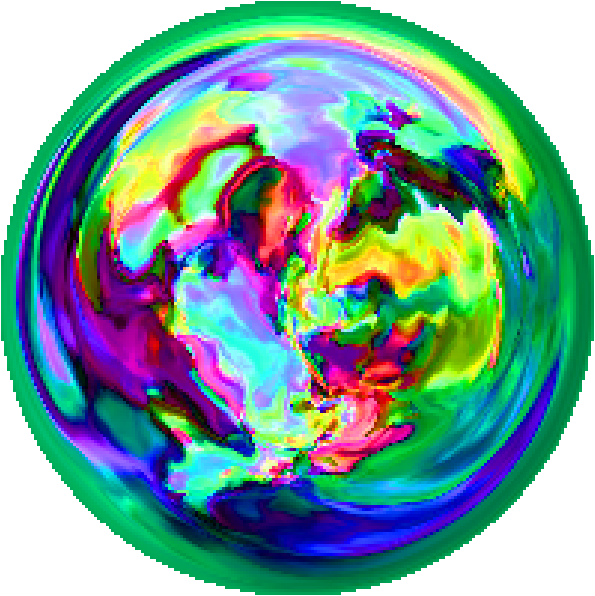}&\includegraphics[width=0.2\textwidth,valign=c]{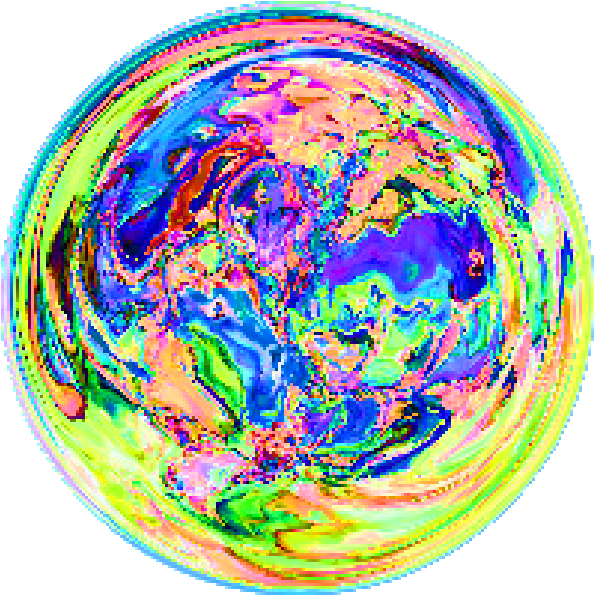}&\includegraphics[width=0.2\textwidth,valign=c]{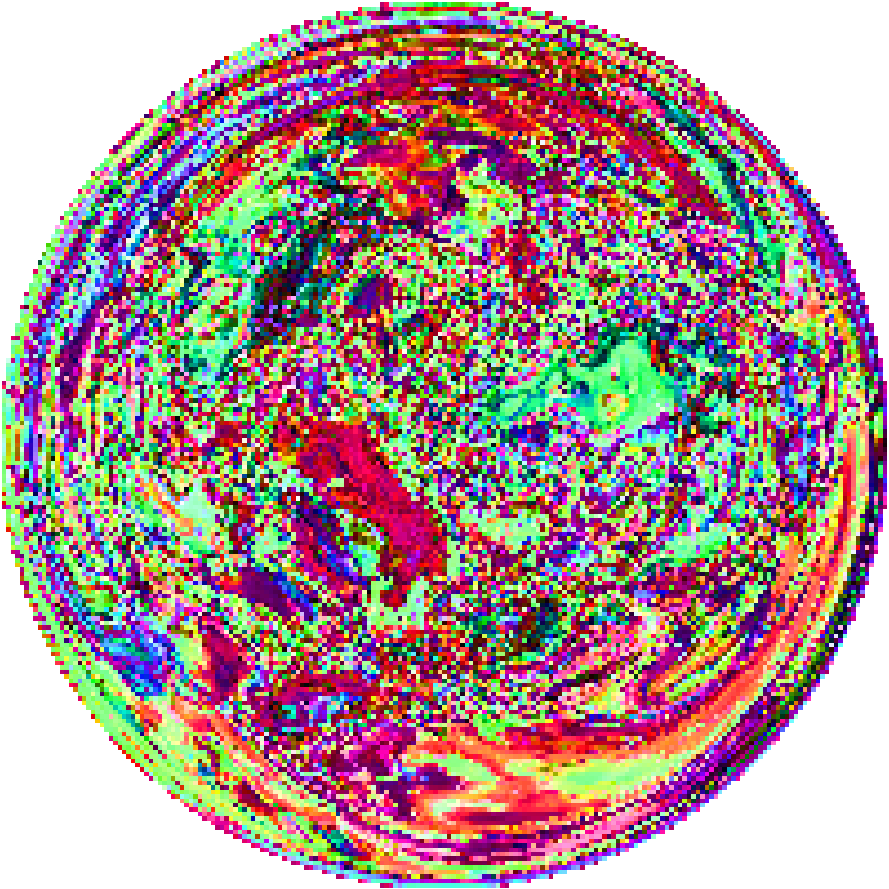}
\end{tabular}
}
\caption{Illustration of the network functions of fully-connected architectures with the SELU activation function at different depths in the initial state. Each disc depicts a  spherical 2D subspace of the input space as an azimuthal projection, and each color corresponds to a different region of the output space. The NLC was evaluated on unit Gaussian input. {\it Conclusion:} The more nonlinear the network appears, the higher the NLC.}
\label{discillu}
\end{table*}

Is the NLC intuitively reasonable as a measure of nonlinearity? Since this question is non-quantitative by nature, it is impossible to verify it on a broad scale. Notwithstanding, in this section we check whether the NLC passes an eye test.

In table \ref{linillu}, we saw that the NL1D metric corresponds reasonably well to the intuitive notion of nonlinearity across eight scalar functions. In table \ref{discillu}, we depict the network functions of fully-connected architectures with SELU activation function, $d_\text{out}=3$ and different depths in their initial state. See section \ref{additionalExperimentsSection} for architectural and other details.

\begin{wrapfigure}[16]{r}{5.5cm}
\includegraphics[width=5.4cm]{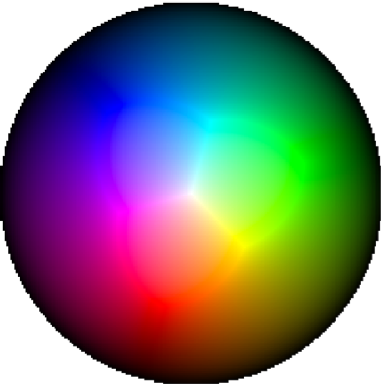}
\caption{Coloring of the output sphere used for the illustrations in table \ref{discillu}, depicted as an azimuthal projection.}\label{outputSphere}
\end{wrapfigure}

We drew three inputs $x^{(1)}$, $x^{(2)}$ and $x^{(3)}$ from the unit Gaussian distribution. We associated each point $(a,b,c)$ that lies on the unit sphere in $\mathbb{R}^3$, i.e. that has $a^2+b^2+c^2=1$, with the input $ax^{(1)} + bx^{(2)} + cx^{(3)}$. We call the sphere of points $(a,b,c)$ associated with these inputs the ``input sphere''. We evaluated the network on a dense grid of those inputs. For each input, we obtained a 3-dimensional output, which we divided by its length. Now the output lies on the unit sphere in $\mathbb{R}^3$. Each point on that ``output sphere'' was associated with a color as depicted in figure \ref{outputSphere}. Finally, we colored each point on the input sphere according to its respective color on the output sphere. 

The RGB values of colors were chosen so that the R component is largest whenever the value of the first output neuron is largest, the G component is largest whenever the value of the second output neuron is largest and the B component is largest whenever the value of the third output neuron is largest. If we imagine that the output is fed into a softmax+cross-entropy loss function for 3-class classification, then ``purer'' colors correspond to more confident predictions.

The colored input spheres are given in table \ref{discillu}. We also give the NLC of those same networks taken on unit Gaussian input. \finding{As desired, we find that the more nonlinear a network appears, the higher the NLC.} While we do not have space for more colored input spheres, we did verify that this relationship also holds across e.g. different activation functions and widths.

\section{Properties of the NLC} \label{nlcPropertiesSection}

In this long section, we demonstrate a range of properties of the NLC, where each subsection corresponds to a different property which is given in its title. 

\paragraph{Background from prior chapters} The majority of subsections rely on empirical evidence, which is based on our empirical studies. Our experiments can roughly be grouped into three buckets. (i) In study A, we trained 750 randomly generated fully-connected architectures, 250 each on CIFAR10, MNIST and waveform-noise. (ii) In study B, we trained 552 convolutional architectures on CIFAR10. (iii) We conducted further experiments which were similar to study A, but where we changed the architecture or specific aspects of the training protocol. Our studies are laid out in detail in chapter \ref{empiricalStudiesChapter} and summarized in section \ref{metricsSummarySection}, along with the terminology and conventions that are most important for understanding and interpreting our figures, tables and discussion of experimental measurements. While we recommend reading at least the summary section, we also hope to present our results in this work in such a way that reading chapter \ref{empiricalStudiesChapter} is not necessary beyond the summary or not necessary at all.

The scope of our empirical studies, especially in terms of the variety of architectures considered, the carefulness of training and the carefulness of metric computation, is a key distinguishing feature of this work relative to prior work. We conducted nearly 300,000 independent training runs on fully-connected architectures and nearly 12,000 independent training runs on convolutional architectures. Hence, we believe chapter \ref{empiricalStudiesChapter} has much value in its own right for informing the design of analytical deep learning studies. 

\subsection{The NLC is predictive of test error and to some degree training error} \label{nlcPredictiveSection}

In this subsection, we establish the NLC's most important property. The NLC of an architecture, when evaluated in the initial state before training, is a powerful predictor of test error after training, and attaining a right-sized NLC is essential for attaining an optimal test error. This is ultimately the motivation behind caring about nonlinearity. To our knowledge, no single metric has ever been shown to have the same predictive power across such a wide range of architectures as the NLC. (Of course, concepts that are related to the NLC, such as gradient magnitude, automatically inherit some of its predictiveness, as we will see in section \ref{vanishMultipleSection}.)

\begin{figure}[H]
\centering
\includegraphics[width=0.98\textwidth]{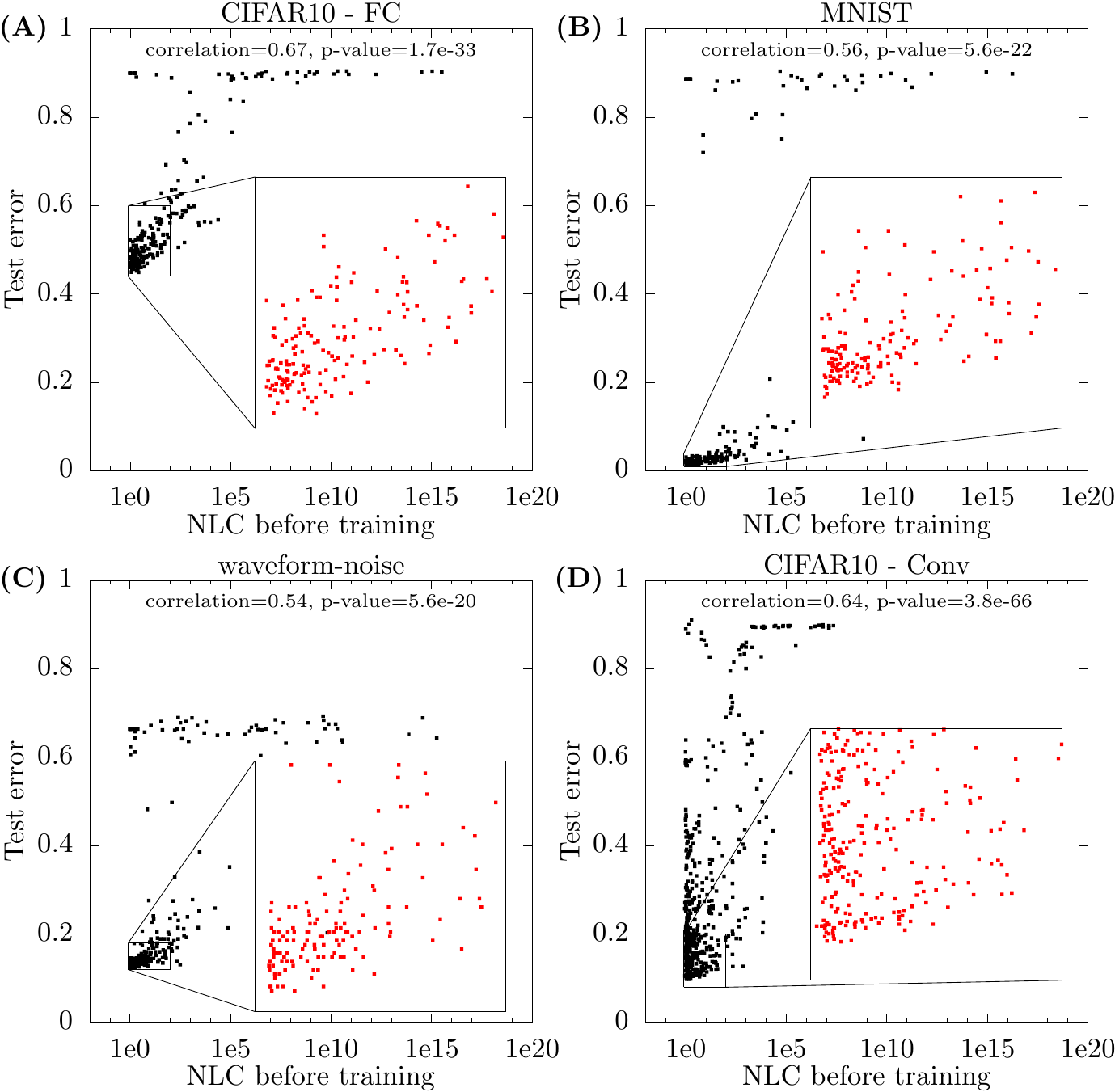}
\caption{The NLC evaluated in the architecture's initial state vs the test error evaluated in the architecture's final state, for all architectures in studies A and B. Graphs A, B and C depict results from study A. Graph D depicts results from study B. Each point in the scatter plots corresponds to a single architecture. The correlation of the metrics plotted on the x- and y- axes, as well as the statistical significance of that correlation, is given at the top of the graph. If a metric is depicted in log scale, then the log of that metric is used to evaluate the correlation. The dataset used is given above the figure. For CIFAR10, we also note whether the architectures used were fully-connected (study A) or convolutional (study B). Inset graphs in the bottom right are magnifications of the region $0.8 < NLC < 100$. Note that one black point in graph C is ``hidden'' among red points. {\it Conclusion:} The NLC of an architecture, when evaluated in the initial state before training, is a powerful predictor of test error after training, and attaining a right-sized NLC is essential for attaining an optimal test error.} \label{nlcPredTestInit}
\end{figure}

\paragraph{Initial NLC predicts test error} In figure \ref{nlcPredTestInit}, we plot the NLC in the initial state vs test error for all architectures in studies A and B. Figure \ref{nlcPredTestInit}, like many figures in this work, depicts scatter plots where each marker corresponds to a single randomly generated architecture. Throughout this work, results are depicted separately for each dataset used in study A, thereby underscoring the consistency of our results across datasets.

In figure \ref{nlcPredTestInit}, we find that \finding{there is a strong association between initial NLC and test error}. Further, \finding{architectures that achieve the lowest test error values for a given dataset / study all have NLCs that lie in a narrow range, approximately between 1 and 5}. Hence, we formulate the following ZSAD guideline.

\begin{npc} \label{npcNLC}
`Use an appropriate NLC' requires $1 \le NLC \le 5$.
\end{npc}

This has immediate practical utility as architectures with an excessive initial NLC can be discarded before training and do not need to be considered further, either conceptually or within the context of NAS. \citet{nasRejection} provides an example for filtering out undesirable architectures.

We depict a magnification of the region $0.8 < NLC < 100$ in each of the graphs in figure \ref{nlcPredTestInit}. \finding{Even within that relatively narrow range, architectures with smaller NLCs tended to perform better (graphs A/B/C) or at least seem capable of performing better (graph D)}. \finding{While a right-sized NLC was necessary to attain good performance, it was not sufficient}. \finding{There were a significant number of architectures with NLCs in the 1-to-5 range that did not perform optimally or did not even attain a better-than-random test error}. We explore the reasons behind the failure of these architectures in detail in chapter \ref{beyondNlcChapter}. Just as for close-to-optimal performance, \finding{there is a range that the initial NLC must fall in for better-than-random performance, approximately between 1 and $10^5$. One architecture was an exception to this rule. It was trained on the waveform-noise dataset with initial NLC around $10^{10}$ and achieved a relatively low error.} We further discuss this architecture in section \ref{nlcEvolutionSection}. 

In section \ref{nlcNonlinearityMetricSection}, we showed that the NLC of linear networks is 1 and that the NLC of nonlinear networks is greater than 1 if the input is Gaussian. We find that \finding{the lower bound of 1 holds for all NLCs in figure \ref{nlcPredTestInit} up to estimation error, and thus for architectures in the initial state}.

In figure \ref{nlcPredTestInit}, like in other figures in this work, we give the correlation of the two metrics plotted on the x- and y-axes at the top. If any metric is depicted in log scale, we also use its logarithm for evaluating the correlation. \finding{The strength of the association between NLC and test error is underscored by their correlation ranging from 0.54 to 0.67 across studies / datasets.} Note that those specific correlation values should be taken with a grain of salt, as they depend heavily on the choices we made in generating the random architectures. In chapter \ref{beyondNlcChapter}, we detail a range of factors that influence performance other than the NLC. As a general rule, the more these other factors vary among architectures considered, the less the fraction of the performance variation explained by the NLC. We further discuss this point in section \ref{architectureSensitivitySection}. Note that we did not generate our architectures with the goal of achieving specific correlation values. We use correlation as a ``quick and dirty'' tool to statistically demonstrate metric associations. We are not implying that the metrics are linearly related. In fact, it is clear in figure \ref{nlcPredTestInit} that a nonlinear predictor from NLC to test error would be more accurate and would explain an even greater fraction of performance variation. 

We give the initial NLC and test error of all our architectures in the appendix in chapter \ref{fullListChapter}.

\begin{figure}[H]
\centering
\includegraphics[width=0.98\textwidth]{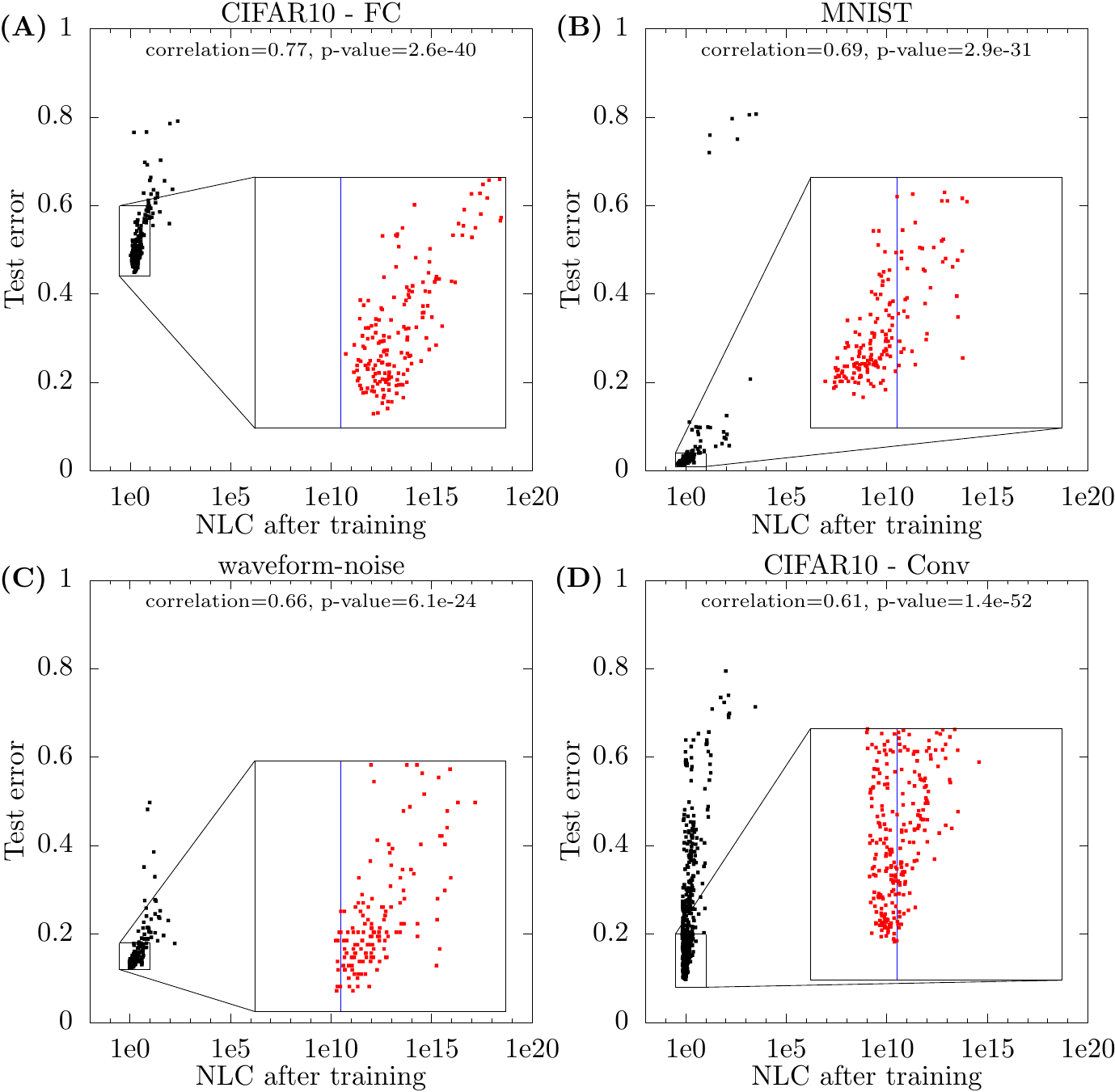}
\caption{NLC vs test error in the final state, for all architectures in studies A and B that achieved a better-than-random validation error (study A) / test error (study B). Inset graphs in the bottom right are magnifications of the region $0.3 < NLC < 10$. Blue lines indicate $NLC = 1$. {\it Conclusion:} The final NLC is associated with test error, and optimal / better-than-random performance requires the NLC to lie in a narrow range.}\label{nlcPredTestFinal}
\end{figure}

\paragraph{Final NLC is associated with test error} In figure \ref{nlcPredTestFinal}, we plot the NLC evaluated in the final state vs test error. Note that whenever we evaluate any metric not based on error in the final state, we automatically discard all architectures for which no starting learning rate achieved better-than-random validation error (study A) or test error (study B). This is because metric values in the final state often depend greatly on the learning rate. Without a meaningful way to choose the best starting learning rate, there is no meaningful final state metric value. Because architectures that achieve better-than-random validation or test error must have relatively small NLCs, the range of NLCs depicted in figures such as \ref{nlcPredTestFinal} is much narrower compared to e.g. figure \ref{nlcPredTestInit}. (Of course, this also affects correlation values.)

We find that \finding{final NLC is also strongly associated with test error}, with \finding{correlation values comparable to those of the initial NLC}. \finding{Architectures that exhibit close-to-optimal performance still have their NLC lie in a narrow range, though this range changes based on dataset / study.} \finding{Some architectures now have an NLC less than 1}. \finding{For better-than-random performance, the final NLC must lie in a range narrower than the initial NLC, approximately between 1 and 100}. However, again, \finding{a close-to-optimal NLC is no guarantee of close-to-optimal performance}.

\begin{figure}[H]
\centering
\includegraphics[width=0.98\textwidth]{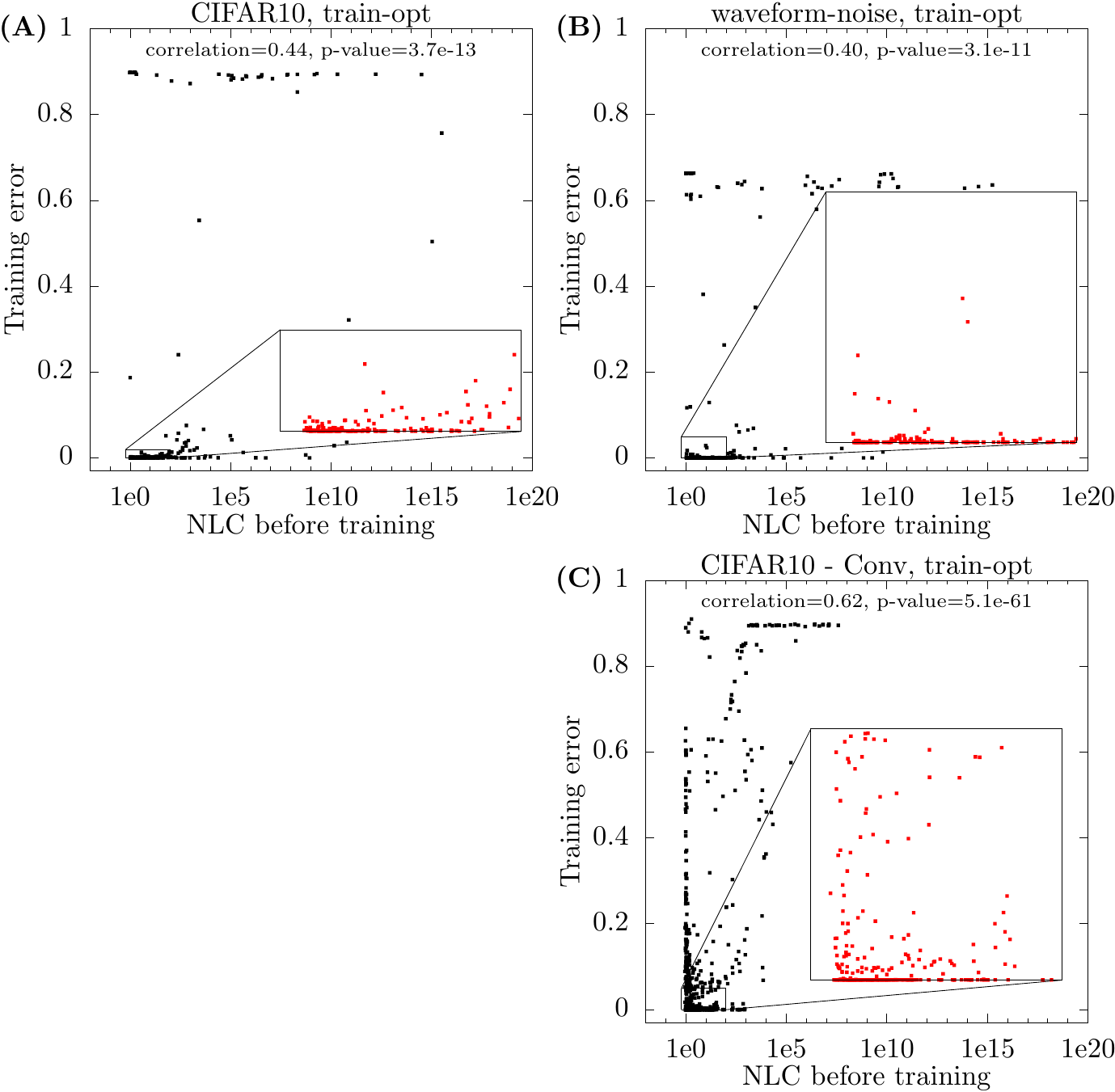}
\caption{Initial NLC vs training error. In graph A, we depict CIFAR10 architectures from study A. In graph B, we depict waveform-noise architectures from study A. In graph C, we depict architectures from study B. Both training error and NLC were evaluated on the training run which yielded the lowest training error after training, as signified by the {\bf train-opt} marker. Study A architectures were re-trained without early stopping based on validation error. Note that the y-axis extends below the zero point for improved visibility. Inset graphs in the bottom right are magnifications of the region $0.6 < NLC < 100$. {\it Conclusion:} The initial NLC is somewhat predictive of training error, especially when it comes to underfitting, but some high-NLC architectures are also trainable, at least when they are fully-connected.} \label{nlcPredTrainInit}
\end{figure}

\begin{figure}[H]
\centering
\includegraphics[width=0.98\textwidth]{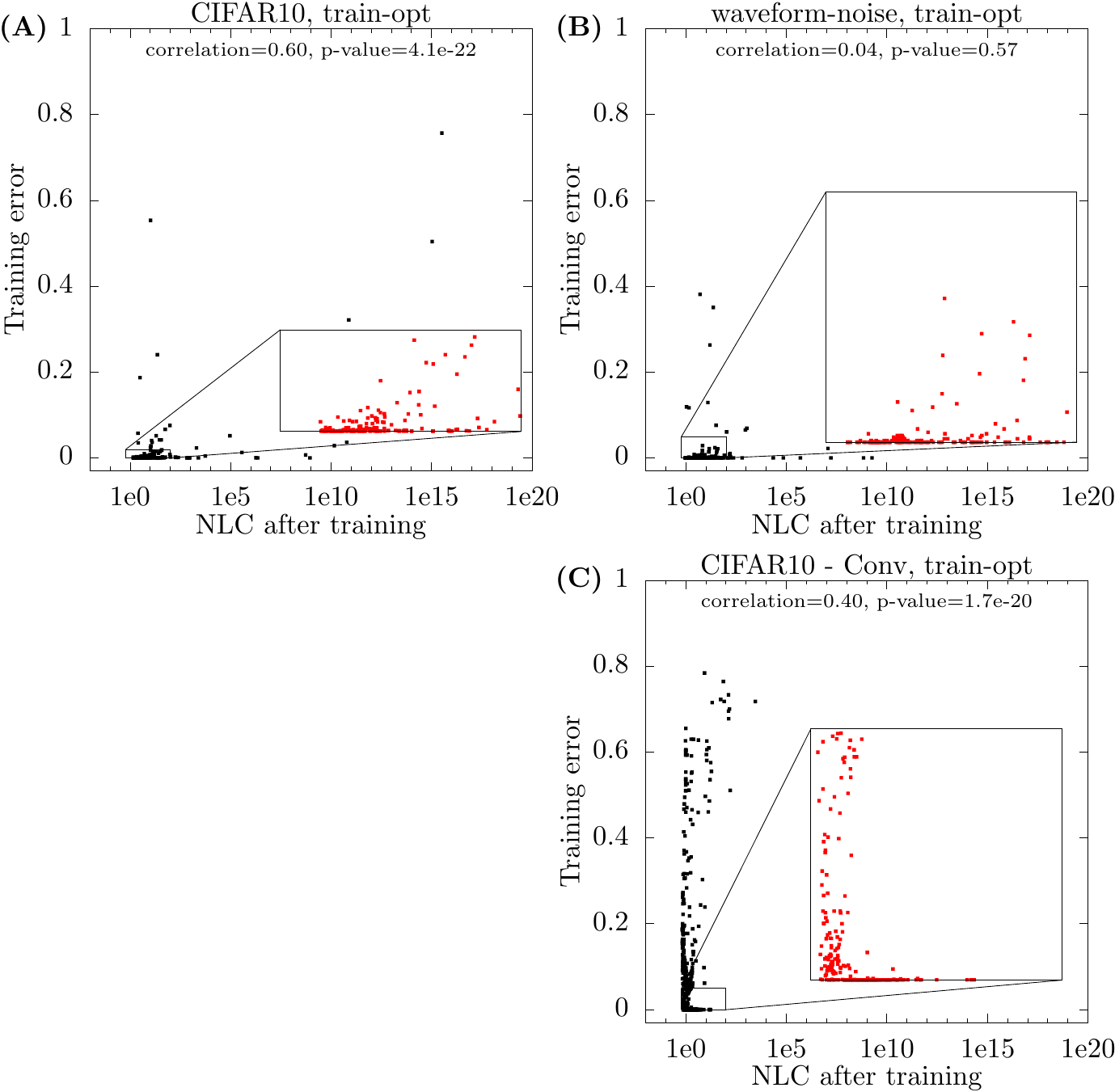}
\caption{NLC vs training error in the final state. In graph A, we depict CIFAR10 architectures from study A. In graph B, we depict waveform-noise architectures from study A. In graph C, we depict architectures from study B. Only architectures that achieved a better-than-random training error are depicted. Inset graphs in the bottom right are magnifications of the region $0.6 < NLC < 100$. {\it Conclusion:} Successfully trained architectures can have very high NLC after training, at least when they are fully-connected. A small or moderate NLC after training does not guarantee low training error.} \label{nlcPredTrainFinal}
\end{figure}

\paragraph{NLC predicts training error in the case of underfitting} In figure \ref{nlcPredTrainInit} and \ref{nlcPredTrainFinal}, we plot the initial and final NLC vs training error. Let's first consider the study A architectures (graphs A/B). We find that, while \finding{at least the initial NLC is correlated with training error}, \finding{there exist several architectures with very high NLC both before and after training that achieve zero or close to zero training error}. This behavior differs from test error. As one would expect, \finding{overall error levels are also much lower}. Just as some high-NLC architectures trained well, we find that \finding{some low-NLC architectures did not train}.

While the NLC is less predictive of training error than test error, the trainability of high-NLC architectures is an important finding in its own right, as we further detail in e.g. sections \ref{nlcKernelSection}, \ref{beyondNlcSummarySection} and \ref{expressivitySection}, as it demonstrates the trainability of ultra-high complexity architectures.

The values in figures \ref{nlcPredTrainInit}A/B and \ref{nlcPredTrainFinal}A/B stem from our re-training of study A architectures, where we tune starting learning rate by considering 60 different values and where we do not use early stopping based on validation error. Hence, the final state differs from e.g. figure \ref{nlcPredTestFinal}. While exhaustive learning rate tuning was essential for a fair comparison of architectures in figure \ref{nlcPredTrainInit}, this even more exhaustive protocol was essential to be able to train high-NLC architectures at all. See section \ref{learningRateSection} for details. Using 64-bit floating-point precision was also essential (section \ref{noiseStabilitySection}). Whenever we give results from training conducted to minimize training error, we always use the {\bf train-opt} marker above our graphs. Whenever such graphs depict metrics not based on error in the final state, such as figure \ref{nlcPredTrainFinal}A/B, we depict only architectures that attained a better-than-random training error. 

In figures \ref{nlcPredTrainInit}C and \ref{nlcPredTrainFinal}C, we plot the NLC vs training error for study B. Here, we find that \finding{no architecture with high NLC was trainable}. However, as we will go on to show in sections \ref{noiseStabilitySection} and \ref{beyondNlcSummarySection}, this was due to all our high-NLC architectures using batch normalization or data augmentation. \finding{There are a large number of architectures with a very small NLC that attain a training error significantly above 0 but below 0.6. In contrast, the vast majority of architectures in graphs A/B either have a near-zero or near-random training error.} This difference was caused by how we generated the architectures in studies A and B. In study B, we used activation functions that were interpolations between standard activation functions and linear functions. See section \ref{studyBArchitecturesSection} and table \ref{AFLMillu}. This meant that many architectures had activation functions that were close to linear functions and therefore many architectures were themselves close to linear functions and thus also had low NLCs. It turns out that logistic regression, which is based on a linear model, attains around a 0.6 training and test error on CIFAR10 due to underfitting. Hence, our results indicate that many architectures in study B also exhibit underfitting. See sections \ref{nlcLinearApproximationSection} and \ref{nlnormWorksSection} for further analysis on this point. In contrast, in e.g. sections \ref{nlcNoiseSection} and \ref{nlcKernelSection}, we show that overfitting explains why a high NLC leads to high test error. The underfitting phenomenon also caused the greater diversity of test error values in figure \ref{nlcPredTestInit}D versus \ref{nlcPredTestInit}A-C.

We did not re-train our convolutional architectures for the purpose of figures \ref{nlcPredTrainInit}C and \ref{nlcPredTrainFinal}C, as we used a fixed number of iterations and did not use early stopping based on validation or test error to begin with. However, we did re-select the starting learning rate and therefore the training run based on the lowest training error after training. Again, we signify this with the {\bf train-opt} marker.

We give the initial NLC and training error of our architectures in the appendix in chapter \ref{fullListChapter}. Note that we did not conduct re-training for our MNIST architectures due to limitations in computational budget. Hence, we do not consider those architectures throughout this work when it comes to training error.

\begin{figure}[H]
\centering
\includegraphics[width=0.98\textwidth]{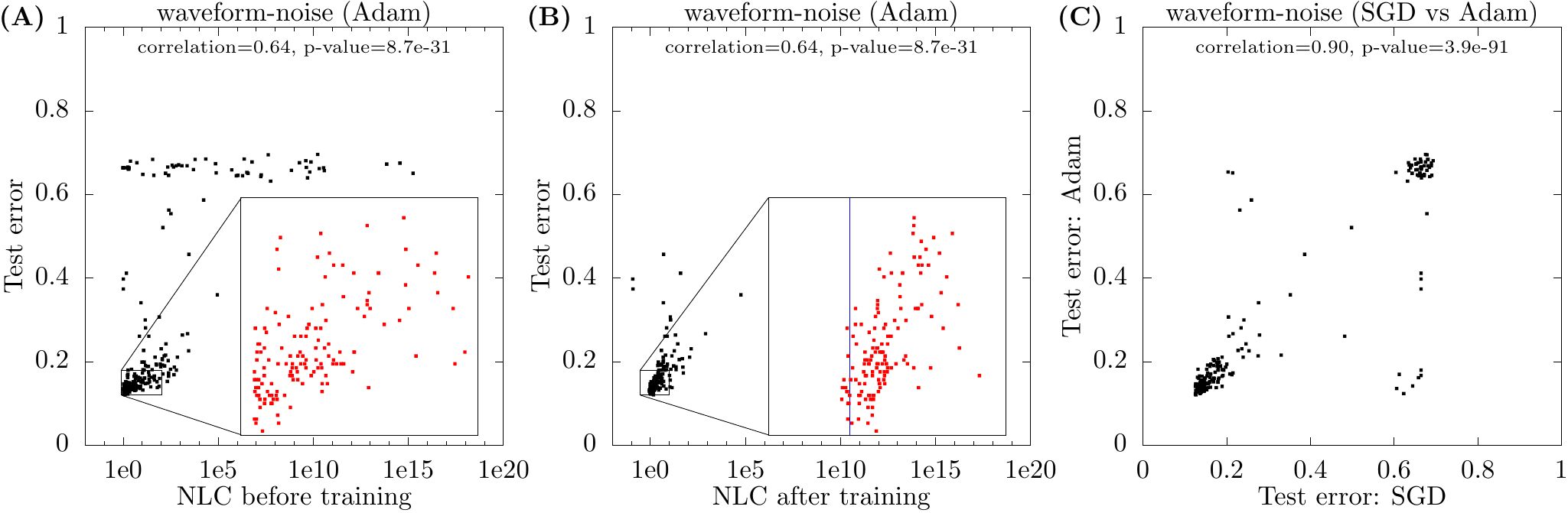}
\caption{Comparing NLC and test error for study A waveform-noise architectures, trained with Adam and SGD. Inset graphs in the bottom right are magnifications of the region $0.8 < NLC < 100$ in graph A and $0.3 < NLC < 10$ in graph B. The blue line indicates $NLC=1$. {\it Conclusion:} Adam behaves very similarly to SGD.} \label{nlcAdam}
\end{figure}

\paragraph{We obtain the same findings with Adam} Of course, the error achieved by an architecture depends to some extent on the training algorithm used. We trained all architectures in study A with SGD. To a significant degree, we eliminated the confounding effect of the training algorithm by conducting extensive and independent learning rate tuning for each architecture. The learning rate hyperparameter is the most important determinant of performance other than architecture. See sections \ref{moduloAlgorithmSection} and \ref{learningRateSection} for further information. To further validate our results, we retrained our waveform-noise architectures using the Adam algorithm. We otherwise used the same careful training protocol as in study A. Both training algorithms used the same initial state, so the NLC before training was also the same. In figure \ref{nlcAdam}, we plot the initial and final NLC vs test error after training with Adam (graphs A/B), and we compare the test error achieved with SGD and Adam directly (graph C). \finding{Graphs A and B appear very similar to figures \ref{nlcPredTestInit}C and \ref{nlcPredTestFinal}C respectively, yielding the same findings}, as desired. The only difference is that \finding{there were a small number of architectures that attained an NLC significantly smaller than 1 after training with Adam}. We did not previously observe this for waveform-noise, but we did for MNIST (see figure \ref{nlcPredTestFinal}B/C). In figure \ref{nlcAdam}C, we find that \finding{the error achieved by both algorithms does differ somewhat, but there were only a few architectures that attained a better-than-random test error with one algorithm but not the other. Adam was slightly superior in this regard}.

\newpage

\newpage

\begin{figure}[H]
\centering
\includegraphics[width=0.98\textwidth]{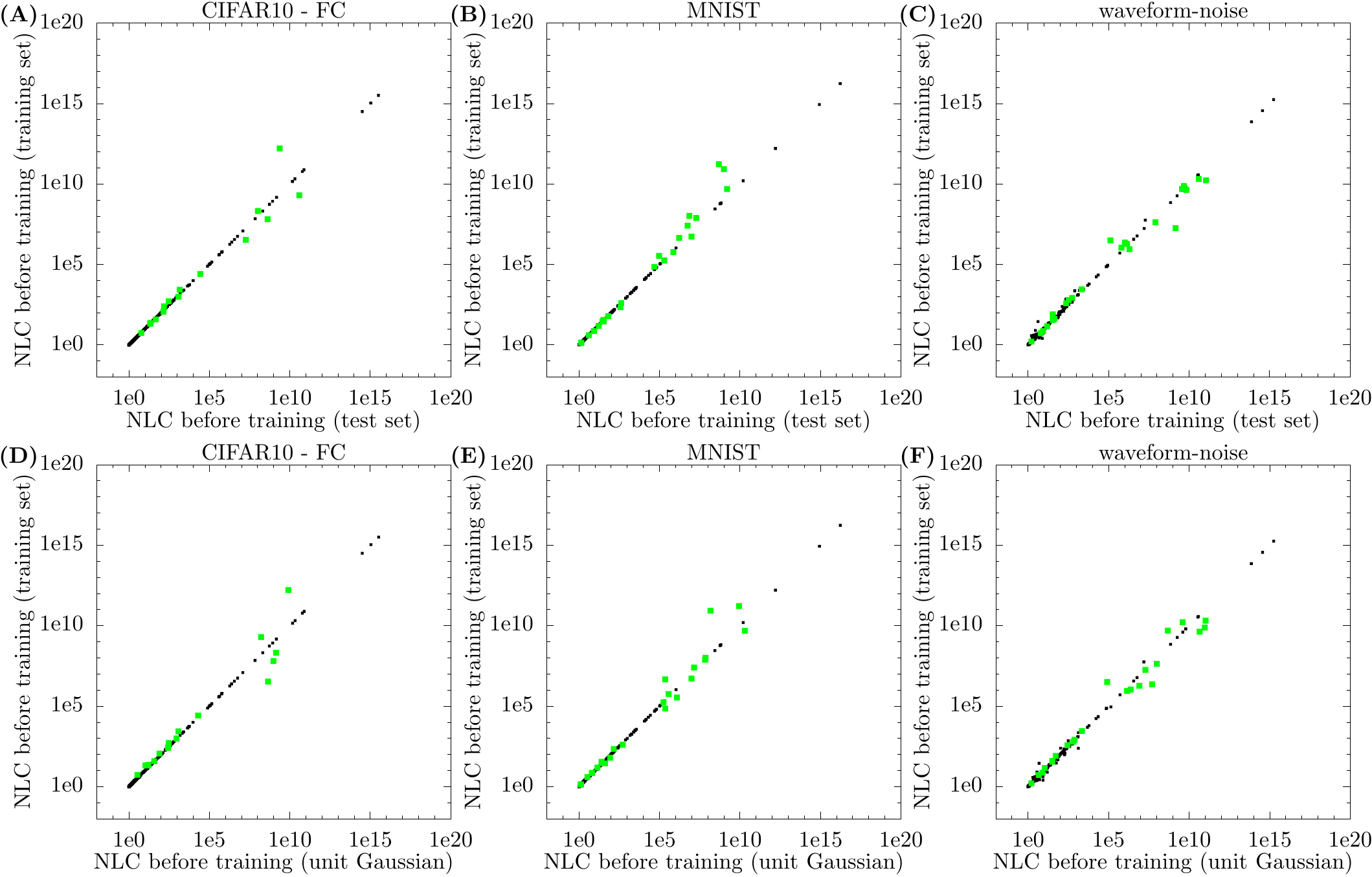}
\caption{Initial NLC evaluated on the training set vs test set and unit Gaussian input for study A architectures. Green markers correspond to Gaussian unstable architectures (GUAs), and they are displayed in the foreground. All correlation values are close to 1. {\it Conclusion:} The estimator of the NLC is stable for our samples. The NLC on practical input distributions matches the NLC on Gaussian inputs.} \label{nlcDataGaussInit}
\end{figure}

\subsection{The NLC is robust to data distribution and data sample} \label{nlcRobustDataSection}

The NLC does not, strictly speaking, measure the nonlinearity of networks, but the nonlinearity of a network with respect to a distribution. In this subsection, we show that the NLC can nonetheless be fundamentally regarded as a property of the network as it is invariant to structure present in practical input distributions, being sensitive only to their expectation and covariance, as well as being insensitive to the sample used for estimation, at least in the initial state.

\paragraph{The NLC estimator is stable} The NLC is defined in section \ref{nlcDefinitionSection} via probabilistic operators over inputs drawn from the input distribution. The data distribution is only known via the dataset, which is assumed to be an IID sample. Hence, we must compute the NLC using statistical estimation. We discuss the computation of the NLC in section \ref{nlcComputeSection} and the computation of our metrics in general in section \ref{metricComputationSection}.

This raises the question of whether our datasets are large enough for our estimators to yield stable values. In figure \ref{nlcDataGaussInit}A-C, for study A architectures, we plot the value of the initial NLC evaluated on a sample that stems from the training set vs the value of the initial NLC evaluated on a sample that stems from the test set. Assuming that both shards are drawn from $\mathcal{D}$, we would desire these values to be as close as possible. Indeed, \finding{we find a very close match for most architectures}. As one would expect based on dataset size, \finding{the difference between both values is slightly larger for waveform-noise than for CIFAR10 / MNIST}.

However, there are also a few architectures for which we observe a more pronounced difference between both estimates. The short and shallow explanation for this is that all these architectures are based on either the square or odd square activation function and they use batch normalization. The deep explanation is that this combination of layer operations can lead to a pathology we term `Gaussian instability', which we investigate in detail later in this work, such as sections \ref{meanFieldDistributionSection}, \ref{gaussianStabilityExplanationSection} and \ref{gaussianStabilitySection}. Since we do not provide an exact definition for Gaussian instability, for the purpose of our empirical analysis, we say an architecture is a `Gaussian unstable architecture' (GUA) if it is built from specific layer operations given at the end of section \ref{metricTerminologySection} and section \ref{metricsSummarySection}. (Note that we do not advocate for our designation of GUA as the definition of the actual phenomenon of Gaussian stability.) The full list of GUAs is given in the appendix in chapter \ref{fullListChapter}. If a figure in this work contains green markers, such as figure \ref{nlcDataGaussInit}, then those markers correspond to the GUAs unless otherwise stated. By default, these markers are also displayed in the foreground, i.e. they fully or partially occlude other markers if they are in the same or in highly similar locations in the graph. GUAs defy, to one degree or another, many valuable properties we uncover throughout this work. As we will find in section \ref{gaussianStabilitySection}, perhaps fortunately, GUAs also tend to exhibit high test error. It is important to note that not all of our figures use green markers. A lack of green markers does not mean the GUAs are excluded from the figure. It simply means that they do not behave differently from other architectures, so there was no need to visually distinguish them.

\begin{figure}
\centering
\includegraphics[width=0.98\textwidth]{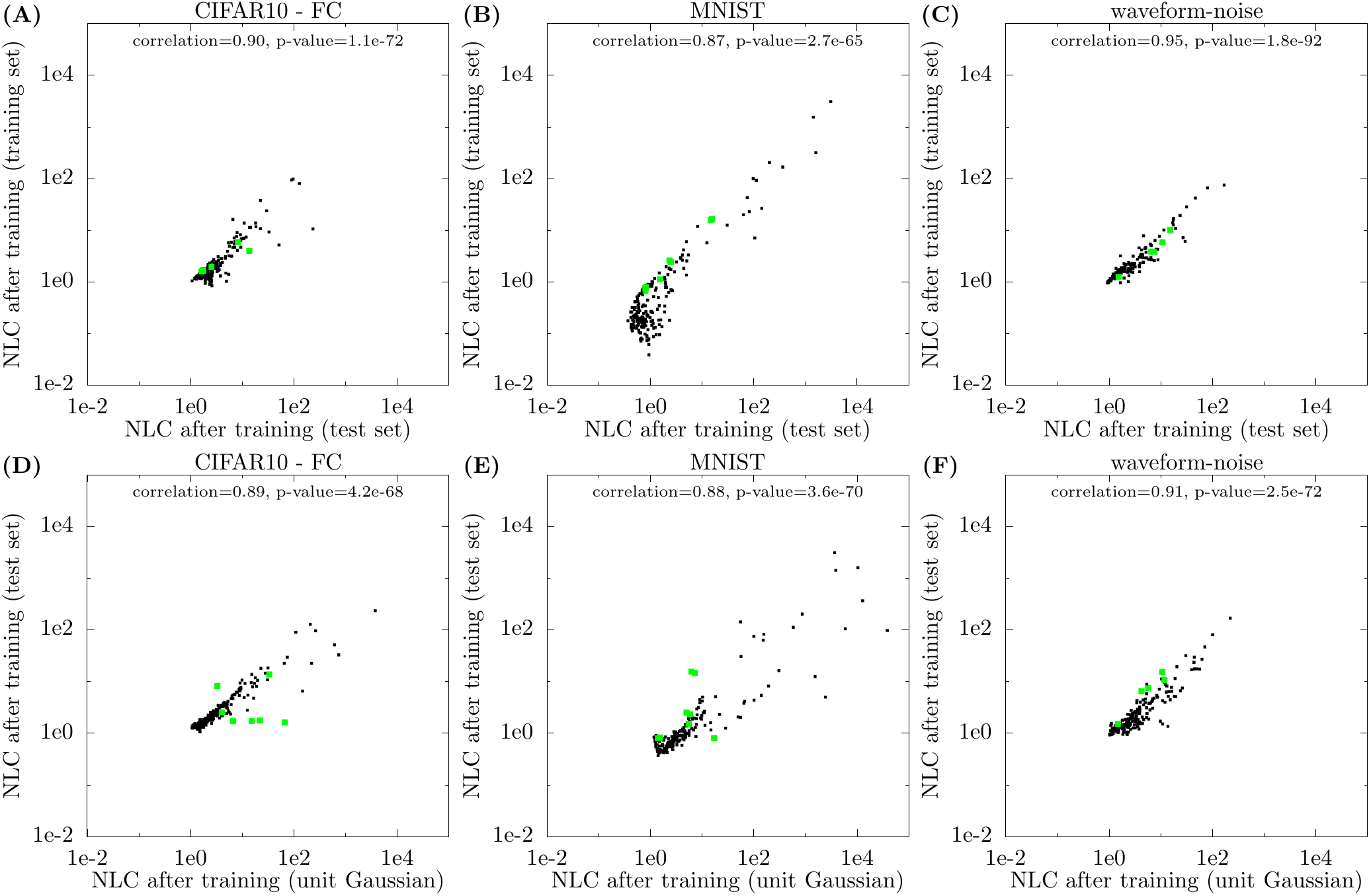}
\caption{Final NLC evaluated on the test set vs training set and unit Gaussian input for study A architectures. Note that the x- and y-axis ranges differ significantly from figure \ref{nlcDataGaussInit}. Throughout this work, axis ranges can vary when related, but not identical, metric values are depicted. However, we also strive to keep axis ranges the same for comparability when practical. {\it Conclusion:} The final NLC is still relatively consistent across data shards and distributions.} \label{nlcDataGaussFinal}
\end{figure}

In figure \ref{nlcDataGaussInit}, we find that \finding{for GUAs, the NLC evaluated on the training and test set is not necessarily close-to-equal}, which means that the samples are not large enough to ensure stability of the NLC estimator. This fact should be kept in mind when interpreting graphs throughout this work.

\begin{wrapfigure}[20]{r}{5.5cm}
\includegraphics[width=5.4cm]{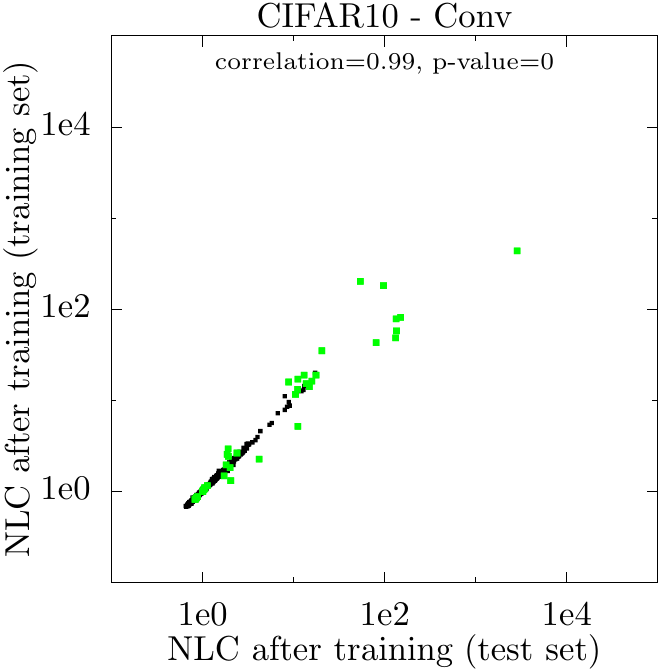}
\caption{Final NLC evaluated on the training vs test set for study B architectures. {\it Conclusion:} Both values are close to equal, especially when excluding GUAs.}\label{nlcDataGaussConv}
\end{wrapfigure}

\paragraph{Training and test set NLC are similar even after training} Previously, we compared the NLC evaluated on the training and test set in the initial state with the clear expectation that both values should be close to equal. In the final state, this is not necessarily the case because the training set was used to run the training algorithm, and is thus far from an unbiased sample for the purpose of estimating the NLC. In figure \ref{nlcDataGaussFinal}A-C, we make the same comparison as in figure \ref{nlcDataGaussInit}A-C, but in the final state. We find that \finding{the NLC values taken on training and test set are no longer near-identical, but the correlation is still high}. In figure \ref{nlcDataGaussConv}, we make the same comparison for study B architectures. Here, we find that \finding{the correlation is actually still close to 1} even after training, and even when including GUAs. This is a rare occasion where we find convolutional architectures to be better-behaved than fully-connected architectures. Note that some of our convolutional architectures, as opposed to fully-connected architectures, employ data augmentation on the training set. If this is the case, then training and test inputs cannot even be regarded to be from the same distribution, as we explain in section \ref{metricEstimationSection}.

This consistency between training and test set is noteworthy considering the nature of the NLC and gradient methods. The NLC is based on $\frac{df}{dx}$, and gradient methods are designed specifically to minimize $\frac{dL}{d\theta}$ on the training set. Therefore, one might expect that $\frac{df}{dx}$ values are also significantly smaller on the training set compared to the test set. While this is certainly \finding{true for some architectures, especially when trained on MNIST, it does not hold for the majority of architectures, especially convolutional architectures}.

\paragraph{The NLC on practical datasets mirrors the NLC on Gaussian distributions} We evaluated the NLC of all architectures in study A in the initial and final state on unit Gaussian input. For this purpose, we generated sets of 10,000 points where each component of each point was independently drawn from $\mathcal{N}(0,1)$. We generated three such sets of different dimensionality to match the width of the input layer of architectures generated for CIFAR10, MNIST and waveform-noise respectively. Samples for estimating the NLC were then taken from those sets.

In figure \ref{nlcDataGaussInit}D-F, we plot the initial NLC taken on the training set vs the NLC on unit Gaussian input.  \finding{The match is as good as between training and test set values}. This means that any structure present in our input distributions is completely ignored by the NLC, at least in the initial state. In figure \ref{nlcDataGaussFinal}D-F, we make the same comparison in the final state. As with training and test set, \finding{the match is not as good as before training, but the correlation still hovers around 0.9}.

In theorem \ref{finiteNetNlcGreater1}, we showed that the NLC is at least 1 on Gaussian input, for any network $f$. \finding{This theoretical finding is confirmed in figures \ref{nlcDataGaussInit} and \ref{nlcDataGaussFinal}, up to estimation error.} The close match between Gaussian and practical NLC values makes this bound very meaningful.

As a general rule, throughout this work we will find that many properties hold almost exactly in the initial state, but deteriorate in the final state. The well-behaved nature of randomly initialized architectures can be largely explained by mean field theory, which we cover in chapter \ref{meanFieldNnaChapter}. This theory directly predicts the behavior we observe in this subsection, and many of the following subsections, for architectures in their initial state. In general, we place greater emphasis on the behavior of architectures in the initial state, because a core goal of this work is to improve architecture design without the need for training. We want to enable readers to understand and predict the performance and behavior of architectures by examining the initial state, not the final state.

The results reported here are corroborated by section \ref{nlcDecomposableSection}.

\paragraph{The NLC takes into account input expectation and variance} One commonality between the unit Gaussian distribution and the three datasets from study A is that each input feature has mean zero and the average variance across features is 1. This is critical. If these two moments vary significantly between input distributions, then the NLC is also likely to vary significantly. Consider a simple fully-connected network of depth 2 with the sawtooth activation function (table \ref{actFunIllu}). If, instead of using a dataset where inputs have mean zero and variance one, we vary the variance by multiplying the inputs with a fixed scaling factor, then we obtain a wide range of NLC values. Applying a scaling factor to the inputs means that the layer values in the dependency of the sawtooth layer are also multiplied with this factor. This effectively increases the size of the ``domain'' of values fed into the sawtooth layer, which in turn increases the frequency of the activation function relative to the domain. Remember table \ref{linillu}. We found that increasing the frequency of a periodic scalar function proportionally increases its nonlinearity as measured by NL1D. Therefore, we would expect increasing the standard deviation of the input to a 2-layer sawtooth network would also proportionally increase the NLC.

\begin{wrapfigure}[22]{r}{5.5cm}
\includegraphics[width=5.4cm]{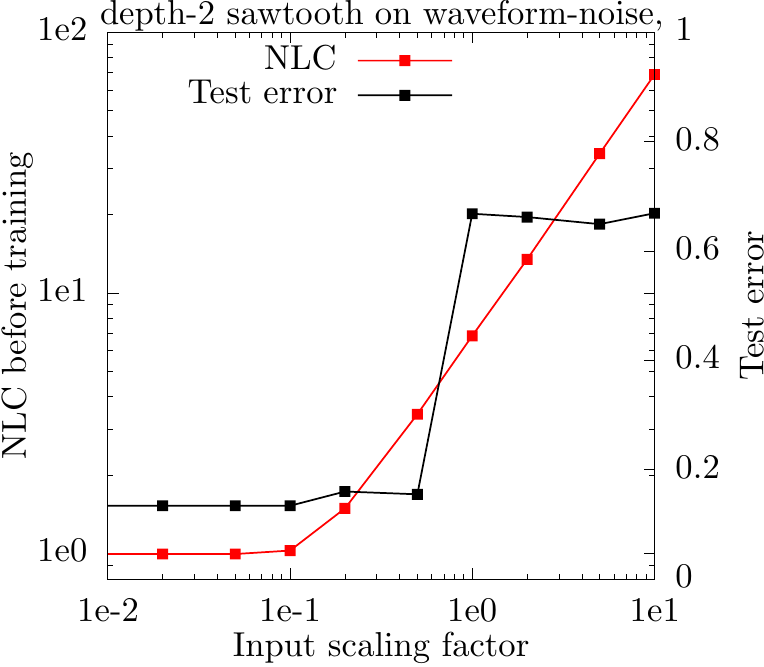}
\caption{Initial NLC and test error as the magnitude of the inputs is varied by multiplying with a fixed scaling factor, for depth-2 fully-connected sawtooth architectures on waveform-noise. {\it Conclusion:} Data variance impacts both NLC and test error.}\label{nlcMeanVar}
\end{wrapfigure}

This is exactly what we find in figure \ref{nlcMeanVar}. There, we plot the initial NLC of 2-layer fully-connected sawtooth architectures on the waveform-noise dataset as a function of the input scaling factor. When the scaling factor is small, almost all values fed into a sawtooth neuron are contained within its linear segment around 0. \finding{When this happens, sawtooth, and hence the entire network, is approximately linear, and we find the NLC is close to 1}. We say the activation function is `pseudo-linear'. However, \finding{as the scaling factor increases and multiple periods of a sawtooth neuron are covered by its inputs, the NLC increases proportionally with the scaling factor}.

While the relationship between input variance and NLC can be intuitively understood for 2-layer sawtooth networks, in general, there is a complex interplay between input variance and NLC, as well as many of our other metrics. In section \ref{meanFieldPracticalSection}, we will find that the input expectation is equally important.

We might find this dependence on input expectation and variance undesirable as it means the NLC is not a pure network property. However, it is important to note that changes to the moments of the data also impact performance. For the NLC to be predictive of performance, it must therefore be sensitive to changes in moments. Given our insights from section \ref{nlcPredictiveSection}, we expect a high initial NLC to be associated with high test error. In figure \ref{nlcMeanVar}, we also plot the test error attained by our 2-layer sawtooth architectures when trained on waveform-noise. We use the same careful training protocol as in study A (section \ref{studyATrainingSection} / \ref{metricsSummarySection}). The results are as expected. \finding{If the scaling factor, and hence the NLC, is below a certain level, test error is low. If it is above that level, it is close-to-random}. Two aspects of our careful protocol are critical for obtaining this result. Learning rate tuning allows the pairing of small scaling factors with small learning rates that keep the inputs to the activation function small. Using our augmented loss function, which normalizes the network output before applying softmax+cross-entropy, is important to maintain close-to-optimal performance with small scaling factors that also cause the network output to decrease in magnitude (section \ref{forwardStabilitySection}).

\paragraph{On the NLC and input covariance} So far, we have compared the value of the NLC on practical data shards and unit Gaussian input. While in both cases $x$ has the same mean and variance, the covariance matrix is not equal. Practical datasets tend to have sparse input spectra, whereas the unit Gaussian only has unit eigenvalues. (Though of course, the spectra of our 10,000-point samples are not nearly as uniform.) We found that nonetheless, in the initial state, the NLC values match for fully-connected study A architectures.

While we do not have NLC values for our convolutional architectures on Gaussian input, we find in section \ref{nlcSimpleMetricsSection} that not just the variance of the data, but the entire covariance structure matters even in the initial state. This is expected, as convolutional nets are built to take advantage of the covariance structure of images and similar types of data.

\paragraph{Summary} We found that the NLC depends on the input distribution largely through its expectation and covariance. An expectation of zero and unit covariance can be viewed as the default setting. Hence, it is justified to view the NLC as a network property.

\newpage

\begin{figure}[H]
\centering
\includegraphics[width=0.98\textwidth]{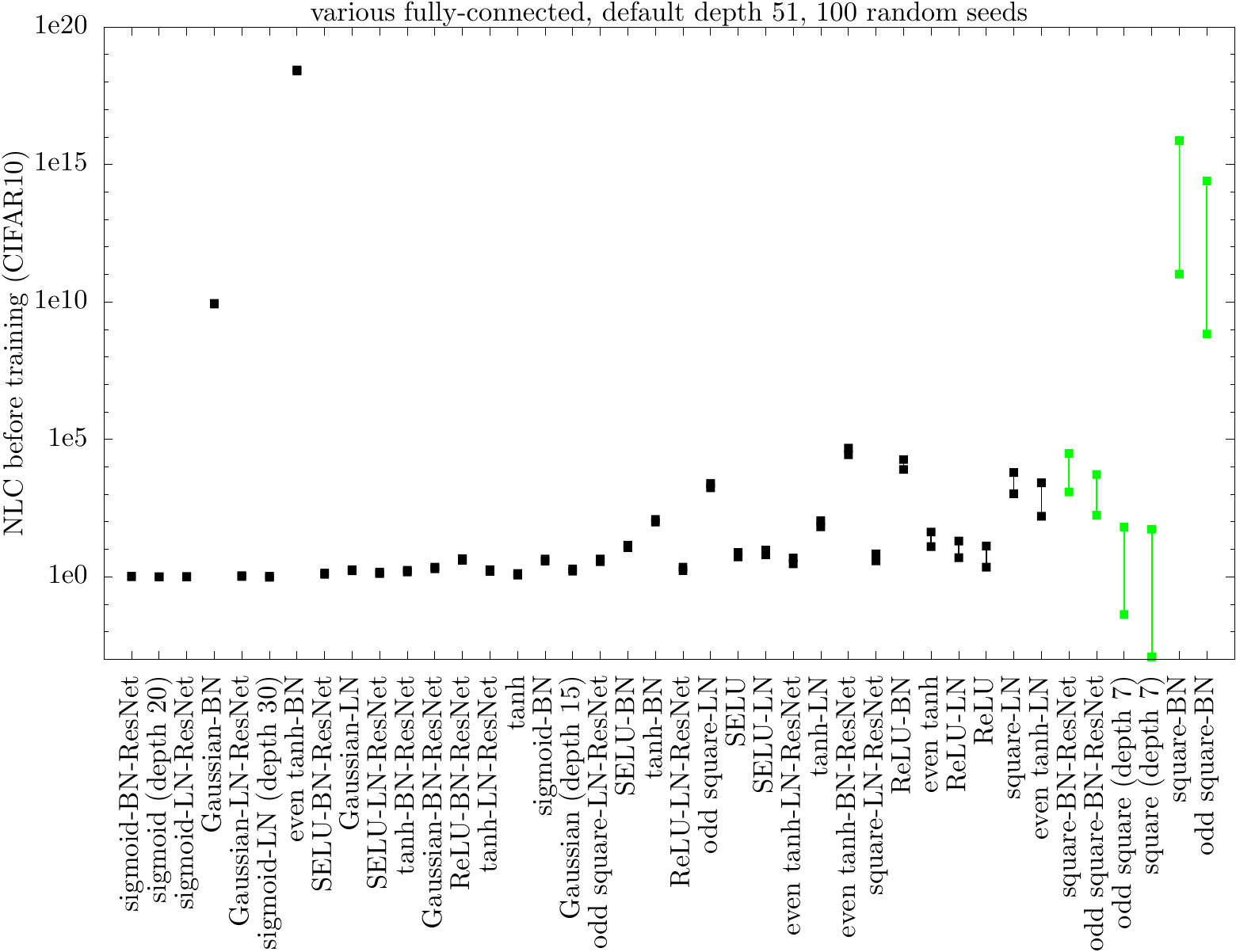}
\caption{Initial NLC across 100 different random seeds for 40 simple fully-connected architectures. Green markers correspond to GUAs. The highest point of each interval indicates the largest of the 100 NLCs and the lowest point of each interval indicates the smallest of the 100 NLCs. The architectures are sorted on the x-axis by the width of the interval in log space. {\it Conclusion:} The NLC does not vary much from one random initialization to the next.} \label{nlcRandomInit}
\end{figure}

\subsection{The NLC is robust to random initialization} \label{nlcRandomInitSection}

In the previous subsection, we showed that the NLC depends on practical data distributions largely through their input expectation and covariance. In this section, we will further show that the NLC in the initial state is largely a property of the architecture rather than the network, meaning that the NLC does not vary significantly from one draw of the parameter initialization scheme to the next. In figure \ref{nlcRandomInit}, we depict this variation for 40 different fully-connected architectures, where we vary activation function, normalization operation and whether the architecture is residual. Most architectures use a depth of 51, though some use a lower depth, as indicated on the x-axis. Lower depths are due to overflow caused by Gaussian instability or because the NLC was not computable due to floating-point rounding error (see section \ref{nlcComputeSection}). We always chose the depth as large as possible while still avoiding those issues.

We find that, indeed, \finding{the NLC does not vary much from one random initialization to the next, except for GUAs}. Hence, we can meaningfully refer to the NLC of an architecture in the initial state on unit Gaussian input as the ``NLC of the architecture''. Note that when generating the results for figure \ref{nlcRandomInit}, we not only varied the value of the parameter, but all random choices controlled by the random seed, including training / validation set split (the NLC was evaluated on the training set), batch selection and random noise used in the estimator of the NLC as described in section \ref{nlcComputeSection}.

\subsection{The NLC is simple and cheap to compute} \label{nlcComputeSection}

As we have established the NLC as a predictive measure of architecture nonlinearity, we now turn our attention towards its computability. We show the following.

\finding{
\begin{itemize}
\item The NLC can be implemented in a few lines of code.
\item The NLC can be estimated accurately even on small datasets via sample means and sample standard deviations.
\item The NLC requires little computation beyond standard forward propagation and backpropagation of $f$. Further, the NLC can ``piggyback'' on the forward propagation that already takes place during training and when computing error.
\item The NLC does not suffer from floating-point rounding error as long as the network output itself does not suffer from rounding error that exceeds its mathematical variation.
\item The NLC can be trivially generalized to a network for which no gradient or directional derivative is available, as long as a local linear approximation can be found for that network that is suitable for gradient methods. (While we do not investigate this explicitly, we suspect that our analysis fully applies even to those networks.)
\item The NLC can be applied to networks with batch normalization without modifying the program used to compute it, while maintaining mathematical and statistical consistency.
\end{itemize}
}

In this section, we discuss the computation of the NLC and surrounding issues both abstractly and with regards to how they manifested in our empirical studies. This section is a follow-up to section \ref{metricComputationSection}.

\paragraph{Computation via estimation} The NLC, like a large fraction of metrics used in this work, is defined in terms of an input distribution $\mathcal{D}$. Of course, $\mathcal{D}$ is hypothetical. The only information we have are the inputs in the dataset $D$, which is assumed to be an IID sample from $\mathcal{D}$. Hence, we have to compute the NLC using statistical estimation and a sample of the distribution. In our empirical studies, we take our samples from one of the three main data shards - training set, validation set and test set - or from a set of points drawn from the unit Gaussian as in section \ref{nlcRobustDataSection}.

\paragraph{Estimating the numerator} We present two different estimators for the numerator of the NLC, which we call `backward estimator' and `forward estimator' respectively. Of course, these are simply suggestions. There are many other options which we leave to the reader to explore.

Letting $x, x' \sim \mathcal{D}$, we have

\begin{eqnarray*}
&&\mathbb{E}_x\Tr(\mathcal{J}(x)\Cov_x\mathcal{J}(x)^T)\\
&=&\mathbb{E}_x\Tr(\mathcal{J}(x)(\mathbb{E}_{x'}(x'-\bar{x})^T(x'-\bar{x}))\mathcal{J}(x)^T)\\
&=&\mathbb{E}_{x,x'}||\mathcal{J}(x)(x'-\bar{x})^T||_2^2\\
&=&\mathbb{E}_{x,x',u\sim\mathcal{N}(0,I)}(u\mathcal{J}(x)(x'-\bar{x})^T)^2\\
\end{eqnarray*}

Backward estimation is a 2-step process utilizing the last line of the above derivation. In the first step, it estimates $\bar{x} = \mathbb{E}_xx$ via the sample mean. In the second step, it estimates the outer expectation. We require a sample $S$ of triplets $(x,x',u)$. The estimator is

$$\frac{|S|}{|S|-1}\mathbb{E}_{(x,x',u)\in S}(u\mathcal{J}(x)(x'-\bar{x})^T)^2$$

As usual, $\mathbb{E}$ applied to a finite set denotes the mean. The estimator uses a correction based on the sample size as the difference between $x'$ and the mean can be regarded as a variance. We use backward estimation in our empirical studies.

Forward estimation utilizes the penultimate line of the above derivation. It also first estimates $\bar{x}$ via the sample mean. Then it estimates the outer expectation using a sample $S$ of pairs $(x,x')$ as below.

 $$\frac{|S|}{|S|-1}\mathbb{E}_{(x,x')\in S}||\mathcal{J}(x)(x'-\bar{x})^T||_2^2$$

\paragraph{Computing and implementing the numerator} $\bar{x}$ can simply be computed once for all networks for a sample of inputs as it does not depend on $f$. Of course, oftentimes, data shards are processed to have mean zero, which allows skipping this step entirely.

For backward estimation, for a given $(x,x',u)$ triplet, the value $u\mathcal{J}(x)$ can be computed by forward-propagating $x$, then backpropagating $u$ as if it were the gradient of the loss function with respect to the network output. Computing $(u\mathcal{J}(x)(x'-\bar{x})^T)^2$ is then straightforward. This is incredibly convenient. Forward-propagating inputs and backpropagating gradients are already part of the standard deep learning pipeline. All that needs to be added is the ability to backpropagate Gaussian noise, which can be done in a few lines of code in most deep learning software frameworks.

Evaluating $(u\mathcal{J}(x)(x'-\bar{x})^T)^2$ for batches of triplets at a time instead of individual triplets works naturally. In our studies, we uniformly sample individual batches of $x$ and $x'$ without replacement one at a time from the data shard, as we do during training. Every batch is sampled independently. Effectively, we bootstrap the data shard except that a single batch cannot contain the same input multiple times. For each pair of an $x$ and $x'$ batch, we draw a $u$ batch of independent unit Gaussian values.

For forward estimation, for a given $(x,x')$ pair, the value $\mathcal{J}(x)(x'-\bar{x})^T$ can be computed by forward-propagating $x$, then {\it forward-propagating $x'-\bar{x}$ as the gradient} using forward-mode automatic differentiation. Forward estimation is slightly less noisy due to the absence of $u$. However, we did not use it because we did not have access to forward-mode AD in our code, and because it is not straightforward to generalize to the case where $f$ contains batch normalization (see below).

\paragraph{Estimating the denominator} Again, we will give two different estimators which we call `exact estimator' and `convenient estimator'.

We have 

\begin{eqnarray*}
&&\Tr(\Cov_f)\\
&=&\sum_{j=0}^{d_\text{out}-1}\Cov_f[j,j]\\
&=&\sum_{j=0}^{d_\text{out}-1}\mathbb{C}_x(f(x)[j],f(x)[j])\\
&=& \sum_{j=0}^{d_\text{out}-1}(\mathbb{S}_xf(x)[j])^2\\
&=&||\mathbb{S}_xf(x)||_2^2
\end{eqnarray*}

As defined in section \ref{metricEstimationSection}, $\mathbb{S}$ denotes the standard deviation operator and $\mathbb{C}$ denotes the covariance operator. Exact estimation simply replaces $\mathbb{S}$ on the last line above with the sample standard deviation. This is also equivalent to replacing $\mathbb{C}$ above with the sample covariance. We always use exact estimation in our studies, which we call ``exact'' because it exactly takes the standard deviation over the sample.

Convenient estimation works explicitly on a batch-by-batch basis. It divides the sample into batches and takes the sample standard deviation over each batch. The final value is then the mean of squares of the estimates for each batch and component.

\paragraph{Computing and implementing the denominator} For exact estimation, we recommend first forward-propagating the sample in batches and accumulating the mean. Then, forward-propagate the sample again. This time, after propagating each batch, subtract the previously computed sample mean from the output, which allows the direct accumulation of the mean of residual squares. We use this method in our studies. It can be superior to computing the standard deviation in a single pass by accumulating mean and mean of squares, as we detail below. We use the whole data shard as the sample.

For convenient estimation, it is possible to accumulate the residual sum of squares in a single pass over the sample. This is why it is ``convenient''. The flipside is that the random batching process introduces some noise.

\paragraph{Floating-point rounding error} When implemented correctly, the NLC does not suffer from catastrophic floating-point rounding error, unless the network outputs are so close together that the rounding error induced by the network evaluation exceeds their variation. Then, it is impossible to compute their standard deviation without modifying the neural network implementation. Fortunately, in our studies, as we suspect in most practical situations, these cases are easy to detect as we simply have to check whether the output values vary at least somewhat more than the width of the floating-point grid. One could argue that this ``output collapse'' is actually a case of the network experiencing catastrophic rounding error, not the NLC.

It turns out that this collapse precisely corresponds to a severe case of output bias, a pathology we investigate in section \ref{outputBiasSection} and beyond. We will argue that this collapse is undesirable from a performance standpoint.

At this point, we want to caution the reader against introducing catastrophic rounding error unnecessarily via a suboptimal implementation. Some pitfalls that apply to the NLC also apply to the majority of other metrics we define in this work. For example, when summing a large number of floating-point values, we have to ensure not to naively add these values one by one to a running sum. If we assume these values are random, IID and their standard deviation does not greatly exceed their absolute expectation, we suffer significant rounding error when summing more than $\approx 10^5$ values in 32-bit precision in this way. 

Some pitfalls are specific to the NLC. When computing the denominator, the simple and naive way would be to accumulate the mean as well as the mean of squares of the output during a single pass over the sample, and then to subtract the square of the accumulated mean from the accumulated mean of squares. It turns out that this incurs catastrophic rounding error when $||\mathbb{S}f(x)||_2 \lessapprox \sqrt{\epsilon}||\mathbb{E}f(x)||_2$, where $\epsilon$ denotes the relative floating-point grid spacing, i.e. around $10^{-7}$ for 32-bit and $10^{-16}$ for 64-bit. In other words, the naive method can only compute the standard deviation if the output varies in the first half of its significant digits. That is, in 32-bit, the output has to vary in the first 4 significant digits. In 64-bit, it has to vary in the first 8 significant digits. Using either of the two options for implementing the denominator that we recommended above circumvents this issue.

\paragraph{Handling missing values} In study A, there are four architectures for which we were unable to compute an NLC in the initial state due to output collapse. However, we could easily estimate that the true NLC value was close to 1 based on our analysis from section \ref{meanFieldPracticalSection}. See section \ref{plainArchitectureSection} for further information on why output collapse is associated with a small NLC. Hence, whenever we plot the NLC against a metric value that we were able to compute for study A architectures, we impute an NLC value of 1 for those four architectures.

\paragraph{Estimator stability} The NLC can be estimated from sample means and sample standard deviations as described above. Both of these basic estimators are well-behaved and well-understood. They are accurate even when the sample size is small as long as the underlying distribution is not too heavy-tailed. As we discover in sections \ref{meanFieldDistributionSection} and \ref{gaussianStabilityExplanationSection}, Gaussian stability ensures that this is not the case in the majority of our architectures. In e.g. sections \ref{nlcRobustDataSection} and \ref{meanFieldPracticalSection}, we confirm the stability of our NLC estimator empirically.

\paragraph{Sample independence} It is possible to reuse inputs from the data shard during sample generation. During bootstrapping, the same input can be drawn multiple times. When generating $(x,x',u)$ or $(x,x')$ tuples, an input can show up as $x$ and $x'$ in different tuples, or even the same tuple. Finally, the same input can be used both when estimating the denominator and the numerator. All these things occur in our studies. While this compromises sample independence, we think this is a benign and negligible issue, especially relative to the issue of having access to a limited amount of data to begin with.

\paragraph{Incorporating data augmentation} There are two possibilities for interpreting the definition of the NLC in the presence of data augmentation. First, we can consider the augmentation function to be part of the network and absorb it into the Jacobian. Second, we can consider the augmentation function to apply before the network and absorb it into the input distribution. This would mean evaluating the NLC with respect to $\mathcal{D}^\text{aug}$, which is the input distribution that arises when drawing from $\mathcal{D}$ and then applying (possibly random) data augmentation to the drawn input $x$. In general, we don't think that there is a significant difference between both choices. In our studies, we choose the latter option as it allows us to not explicitly deal with non-deterministic networks. See section \ref{metricEstimationSection}.

Given an IID sample from $\mathcal{D}$, applying data augmentation to each input independently yields an IID sample from $\mathcal{D}^\text{aug}$. Whenever we reuse a datapoint for estimation as described above in the context of data augmentation, we re-apply the random augmentation function.

\paragraph{Computing the NLC while training} Computing the NLC involves forward-propagating inputs and backpropagating noise. Forward propagation of inputs is conducted naturally during training, so it is possible to piggyback on this to further reduce the computational overhead of computing the NLC throughout training. When accumulating moments for our estimators, we can use exponential moving averages to obtain an ``average of recent NLC values''. It is important to keep in mind never to use these EMAs to compute moments over the current batch. For example, do not use the moving average to compute the residual sum of squares over the outputs in the current batch as part of an estimate of the output standard deviation. The convenient estimator for the denominator requires only a single scalar EMA. The same is true for both estimators we recommended for the numerator. Note that the training set may not be ideal for the NLC computation once training has started, as it is no longer an independent sample of $\mathcal{D}$ relative to the parameter value. See section \ref{nlcRobustDataSection} for further discussion of this point.

\paragraph{Computing the NLC while computing error or loss} It is even easier to integrate the NLC computation with the computation of e.g. training loss or test error. Again, we can piggyback on the forward propagation of inputs, and now we can use simple averages instead of EMAs.

\paragraph{On computing the NLC without gradients} In sections \ref{nonDifferentiableSection} and \ref{nlcFunctionalGradientSection}, we explained how to extend the NLC even to networks for which no gradient is available. Non-differentiable networks, such as quantized networks, generally come with a recipe for computing the gradient of a local linear approximation of the network that can then be supplied to the training algorithm in lieu of a gradient. Above, we showed how we can compute the NLC numerator by applying standard backpropagation to the Gaussian noise vector $u$, rather than computing e.g. the Jacobian explicitly. Hence, for a non-differentiable network, at the point where we backpropagate $u$, we can simply substitute whatever method ``comes with the network'' for computing the surrogate gradient. We do not need to develop any additional algorithms.

\subsubsection{The NLC under batch normalization} \label{nlcBnsection}

\paragraph{Generalizing the definition of the NLC} If the network contains BN, it is no longer a function mapping single inputs to single outputs. Hence, the definition of the NLC from section \ref{nlcDefinitionSection} does not directly apply. To generalize the definition, we use a trick we introduced in section \ref{metricBNsection}. Namely, we re-define $f$ as a function of batches of inputs that returns batches of outputs. The input distribution $\mathcal{D}^\text{batch}$ is then over vectors of dimensionality $|B|d_\text{in}$, where $|B|$ is the batch size and each ``segment'' of dimensionality $d_\text{in}$ is drawn independently from $\mathcal{D}$. $f$ then returns outputs of dimensionality $d_\text{in}|B|$, which are obtained by forward-propagating each segment of the input through $f$ while taking batch moments across segments. For the remainder of this subsubsection, let $x$ and $f$ be defined in this way and let $u$ also be a unit Gaussian noise vector of dimensionality $d_\text{out}|B|$. Note that the network function then also depends on the batch size, and reasonably so.

\paragraph{Generating batch samples} Generating samples of batches is straightforward. Given any IID sample from $\mathcal{D}$, uniformly dividing that sample into batches of the proper size yields an IID sample from $\mathcal{D}^\text{batch}$. In our studies, we generate samples of batches in the BN case in the exact same way as we generate batches of samples in the BN-free case. Since we reuse inputs during estimation in the way described above, our batches can overlap, though each one is sampled without replacement. In the BN case, we do have to take care that we use the same batch size we use during training if we are interested in capturing the network's behavior during training exactly. In our studies, we use a batch size of 250 for fully-connected networks and 128 for convolutional networks for all network evaluations.

\paragraph{Generalizing the estimators of the numerator} When generalizing estimators from the BN-free case to the BN case, the following three criteria apply.

\begin{itemize}
\item[(i)] When the generalized estimator is applied to a network without BN, it should return the same value as the original estimator for any batch size.
\item[(ii)] The program that is used for computing the original estimator should also be usable for the generalized estimator so that little to no additional code is needed.
\item[(iii)] The generalized estimator should not have significantly less statistical power than the original estimator.
\end{itemize}

Both the forward and backward estimator of the numerator given above can be generalized naively by letting $S$ be a set of tuples of batches rather than a set of tuples of individual inputs. This works seamlessly for the forward estimator in that all three criteria above are fulfilled. However, for the backward estimator this naive approach fulfills none of the three criteria. Specifically, the reason why the naive approach lacks statistical power is that obtaining a single value of $(u\mathcal{J}(x)(x'-\bar{x})^T)^2$ for the sample mean would require propagating an entire batch instead of just propagating an individual input as in the original backward estimator. We would like to take the sample mean over a set of size $|S||B|$ instead of just $|S|$ so that the number of sample points relative to our propagation effort remains the same as in the original estimator.

Luckily, this can be achieved as follows. As for the original estimator, we begin by estimating $\bar{x}$ via the sample mean, which corresponds to $|B|$ concatenated copies of its original value. Using $r$ as a shorthand for $x' - \bar{x}$, we have

\begin{eqnarray*}
&&\mathbb{E}_x\Tr(\mathcal{J}(x)\Cov_x\mathcal{J}(x)^T)\\
&=&\mathbb{E}_x\Tr(\mathcal{J}(x)(\mathbb{E}_{x'}(x'-\bar{x})^T(x'-\bar{x}))\mathcal{J}(x)^T)\\
&=&\mathbb{E}_{x,x'}\Tr(\mathcal{J}r^Tr\mathcal{J}^T)\\
&=&\mathbb{E}_{x,x'}\sum_{i,b,j,p,i',b'=0}^{d_\text{in}, |B|, d_\text{out}, |B|, d_\text{in}, |B|}\mathcal{J}[pd_\text{out}+j,bd_\text{in}+i]r[bd_\text{in}+i]r[b'd_\text{in}+i']\mathcal{J}[pd_\text{out}+j,b'd_\text{in}+i']\\
&=&\mathbb{E}_{x,x'}\sum_{i,b,j,p,i'}\mathcal{J}[pd_\text{out}+j,bd_\text{in}+i]r[bd_\text{in}+i]r[bd_\text{in}+i']\mathcal{J}[pd_\text{out}+j,bd_\text{in}+i']\\
&=&\mathbb{E}_{x,x',u}\sum_{i,b,j,p,i'}\mathcal{J}[pd_\text{out}+j,bd_\text{in}+i]r[bd_\text{in}+i]r[bd_\text{in}+i']\mathcal{J}[pd_\text{out}+j,bd_\text{in}+i']u[pd_\text{out}+j]^2\\
&=&\mathbb{E}_{x,x',u}\Big(\sum_{i,b,j,p,i',j'p'}\mathcal{J}[pd_\text{out}+j,bd_\text{in}+i]r[bd_\text{in}+i]r[bd_\text{in}+i']\mathcal{J}[p'd_\text{out}+j',bd_\text{in}+i']\\
&&u[pd_\text{out}+j]u[p'd_\text{out}+j']\Big)\\
&=&\mathbb{E}_{x,x',u}\sum_b\Big(\big(\sum_{i,j,p}\mathcal{J}[pd_\text{out}+j,bd_\text{in}+i]r[bd_\text{in}+i]u[pd_\text{out}+j]\big)\\
&&\big(\sum_{i',j',p'}\mathcal{J}[p'd_\text{out}+j',bd_\text{in}+i']r[bd_\text{in}+i']u[p'd_\text{out}+j']\big)\Big)\\
&=&\mathbb{E}_{x,x',u}\sum_b\Big(\sum_{i,j,p}\mathcal{J}[pd_\text{out}+j,bd_\text{in}+i]r[bd_\text{in}+i]u[pd_\text{out}+j]\Big)^2\\
&=&|B|\mathbb{E}_{x,x',u,b}\Big(\sum_i(u\mathcal{J}(x))[bd_\text{in}+i](x' - \bar{x})[bd_\text{in}+i]\Big)^2\\
\end{eqnarray*}

The key insight here is that, because each segment of $x$ is drawn independently from $\mathcal{D}$, we have $\mathbb{E}_{x'} r[bd_\text{in}+i]r[b'd_\text{in}+i'] = 0$ when $b \neq b'$. Hence, we can eliminate this case from the sum above. In the last line, we give $b$ the uniform distribution over $\{0, .., |B|-1\}$. The generalized backward estimator then becomes

$$|B|\frac{|B||S|}{|B||S|-1}\mathbb{E}_{(x,x',u)\in S,b}\Big(\sum_i(u\mathcal{J}(x))[bd_\text{in}+i](x' - \bar{x})[bd_\text{in}+i]\Big)^2$$

Now, given a sample $S$ of triplets $(x,x',u)$, we can average over $|S||B|$ values by also varying $b$ as desired. And, we still only need to propagate $|S|$ batches in total. Unfortunately, the $|S||B|$ values cannot be considered independent draws from a distribution because groups of $|B|$ values derive from the same network evaluation. Because the cross-batch dependency is actually mild when $|B|$ is not too small, the generalized backward estimator still retains most of the statistical power of the original backward estimator from the BN-free case, which yields criterion (iii). It turns out that criteria (i) and (ii) are also fulfilled. The only minor difference between the generalized and original estimator is that we now have an additional factor of $|B|$, which will, however, cancel out with a similar factor that will pop up in the denominator.

\paragraph{Generalizing the estimator of the denominator} As with the numerator, we can try to generalize our estimators naively. This breaks down completely for the convenient estimator, because each batch now only yields a single value for each output component. For the exact estimator, the naively generalized version is statistically workable, though we would not have criteria (i) or (ii). As before, we can obtain much better estimators by co-varying $b$ and the sample point. We have

\begin{eqnarray*}
&&\Tr(\Cov_f)\\
&=&\sum_{j=0}^{d_\text{out}-1}\sum_{b=0}^{|B|-1}(\mathbb{S}_xf(x)[bd_\text{out}+j])^2\\
&=&|B|\sum_{j=0}^{d_\text{out}-1} (\mathbb{S}_{x,b}f(x)[bd_\text{out}+j])^2
\end{eqnarray*}

To obtain the last expression, we use the fact that the marginal distribution of $f(x)[bd_\text{out}+j]$ is independent of $b$. This last expression now gives us $|B|$ values for the sample standard deviation per batch as desired.

Taking the sample standard deviation over all $(x,b)$ pairs yields the generalized exact estimator. Criteria (i) and (ii) are fulfilled. Again, we recommend accumulating the mean in a first pass over the sample and then the residual mean of squares in a second pass for maximum numerical precision. As with the backward estimator of the numerator, the generalized exact estimator has a tiny loss of statistical power as groups of $|B|$ values for the standard deviation again originate from the same network evaluation.

We obtain the generalized convenient estimator by taking sample standard deviations over the groups of $|B|$ values. Again, we have (i) and (ii). However, the loss of statistical power may be more significant because each sample standard deviation is now based on only a single network evaluation. Since we do not use the convenient estimator in our studies, we did not study this point further.

\subsection{The NLC can (sometimes) be proxied by even simpler metrics} \label{nlcSimpleMetricsSection}

In the previous subsection, we found that the denominator of the NLC is simply the average variance of output neurons. Comparatively, the numerator is somewhat more complex. In this subsection, we find that, at least in the initial state of fully-connected networks, it is possible to proxy the NLC with other metrics that have a simpler numerator. Conversely, for convolutional networks we find it necessary to use the NLC itself.

One question that arises in this work is what we gain by studying nonlinearity over ``gradient vanishing / explosion''. In this subsection, we look at when the Jacobian-covariance product in the NLC numerator can be accurately approximated by a ratio of loss gradients. See chapter \ref{relatedWorkChapter} for a detailed analysis.

\paragraph{It is enough to use a single batch of inputs at a time when estimating the numerator} In the previous subsection, we showed that the NLC numerator equals $\mathbb{E}_{x,x',u\sim\mathcal{N}(0,I)}(u\mathcal{J}(x)(x'-\bar{x})^T)^2$. Obtaining a single value for the expression inside the expectation requires sampling two inputs $x$ and $x'$. When computing the NLC in our empirical studies, we sampled two batches of inputs, one for $x$ and one for $x'$, to obtain a single batch of values for the sample mean.

We suspected that it would be sufficient to instead sample a single batch of inputs and simply shuffle it for the purpose of pairing up individual $x$ and $x'$ values. Of course, the probability of pairing up an input with itself is at least $\frac{1}{|B|}$ in this case, so sample independence deteriorates. The upshot is that the NLC computation can more easily piggyback on the training or error computation as described in section \ref{nlcComputeSection}, which generally only requests a single batch at a time from the data loading component of the deep learning pipeline. We use $NLC^\text{single}$ to refer to this slightly less statistically sound way of computing the NLC.

\begin{figure}[H]
\centering
\includegraphics[width=0.98\textwidth]{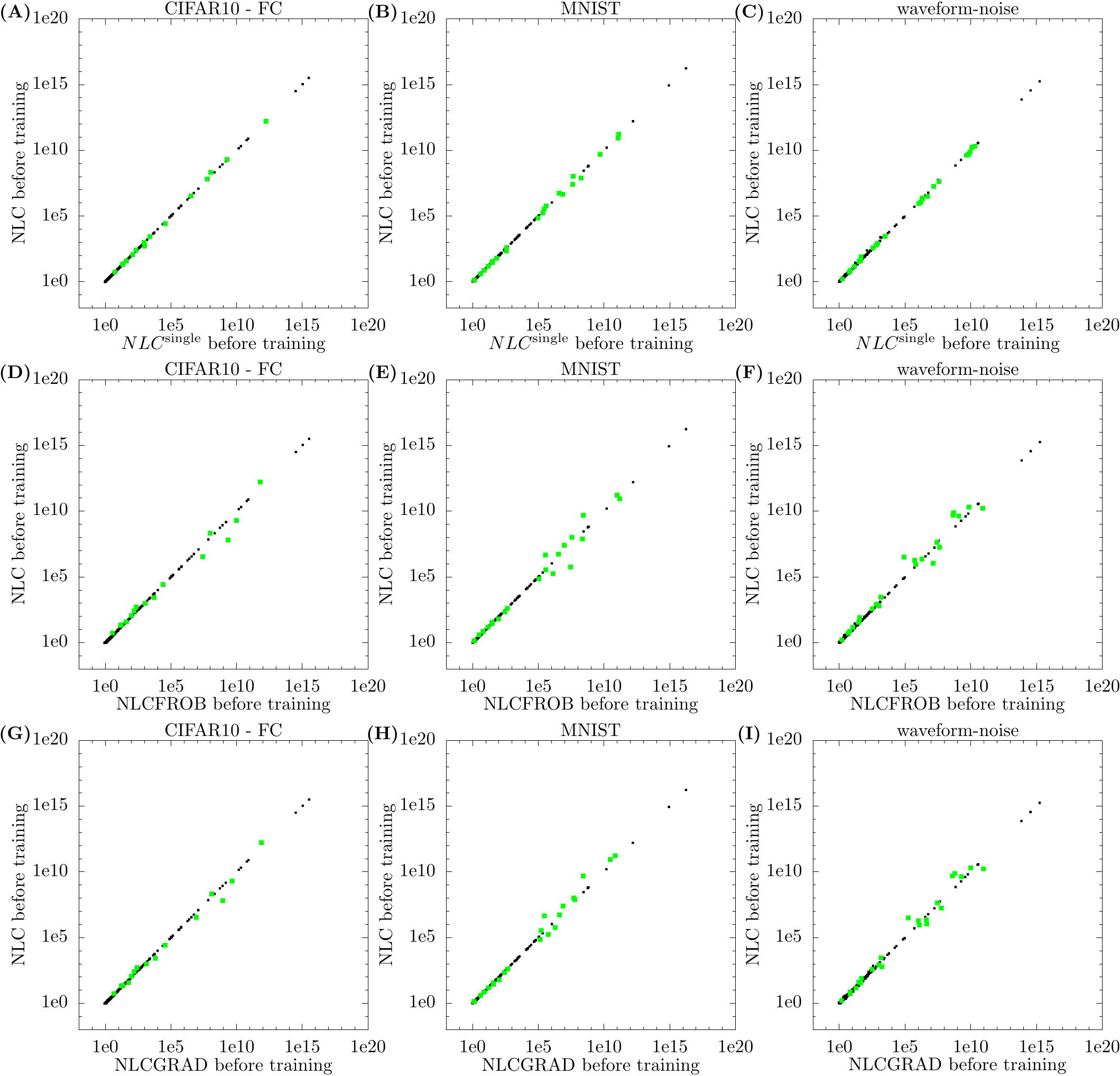}
\caption{NLC vs simpler metrics / estimators for study A architectures in the initial state. Green markers correspond to GUAs and are displayed in the foreground, as always. All correlation values are close to 1. {\it Conclusion:} The NLC can be proxied by simpler metrics / estimators for fully-connected architectures in the initial state.} \label{nlcSimpleMetricsInitFull}
\end{figure}

In figure \ref{nlcSimpleMetricsInitFull}(A-C), we plot the value of $NLC^\text{single}$ vs the NLC in the initial state for study A architectures. \finding{We obtain a very close match, even for GUAs.} This was possible because we made sure that all random processes (batches, $u$ vectors ...) used the same random number sequence for both $NLC^\text{single}$ and NLC.

One concern we had in this subsection, and throughout this work, was that some results only hold because of special properties of the data distributions corresponding to our datasets. For this reason, we investigated some properties of the NLC on a more diverse range of input distributions by using the following trick. We designated a layer in the middle of the network, about halfway between input and output, as a surrogate input layer. We propagated the data shard forward to that layer and then computed the NLC on the remainder of the network by taking our sample from those propagated values. In other words, if $f_l$ is the surrogate input layer, we computed $NLC(f_L(f_l),f_l(\mathcal{D}))$. Informally, we say that we computed the NLC on the ``second half'' of the network. Of course, the input distribution to the second half then depends on the first half of the network, which drastically increases input distribution diversity. The layer at which the second half begins is an addition layer for residual networks, thus ensuring that this layer is a bottleneck, and a linear layer otherwise.

\begin{figure}[H]
\centering
\includegraphics[width=0.98\textwidth]{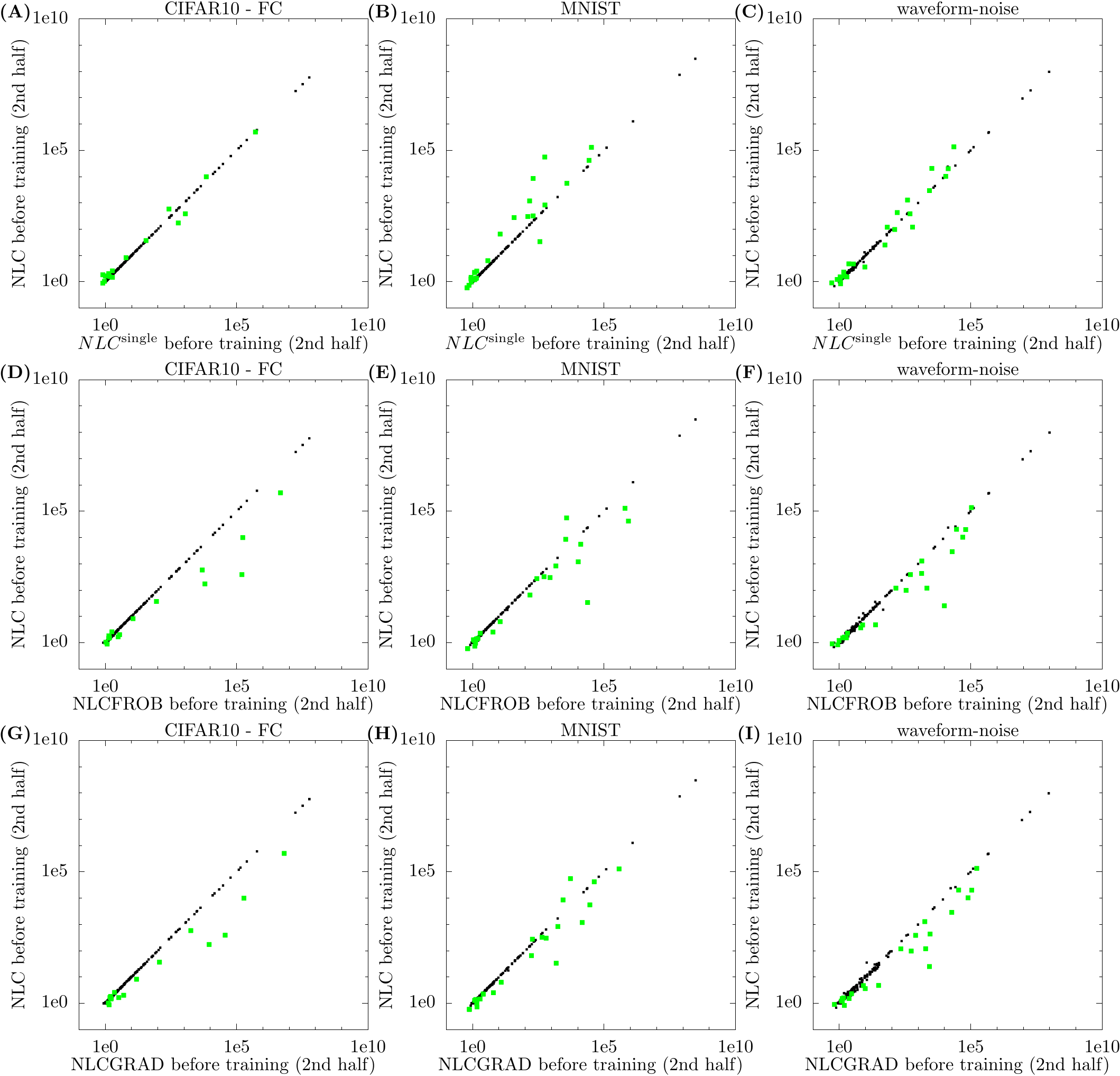}
\caption{NLC vs simpler metrics / estimators for the second half of study A architectures in the initial state. All correlation values are close to 1. {\it Conclusion:} The NLC can be proxied by simpler metrics / estimators for the second half of fully-connected architectures in the initial state.} \label{nlcSimpleMetricsInitHalf}
\end{figure}

\begin{figure}[H]
\centering
\includegraphics[width=0.98\textwidth]{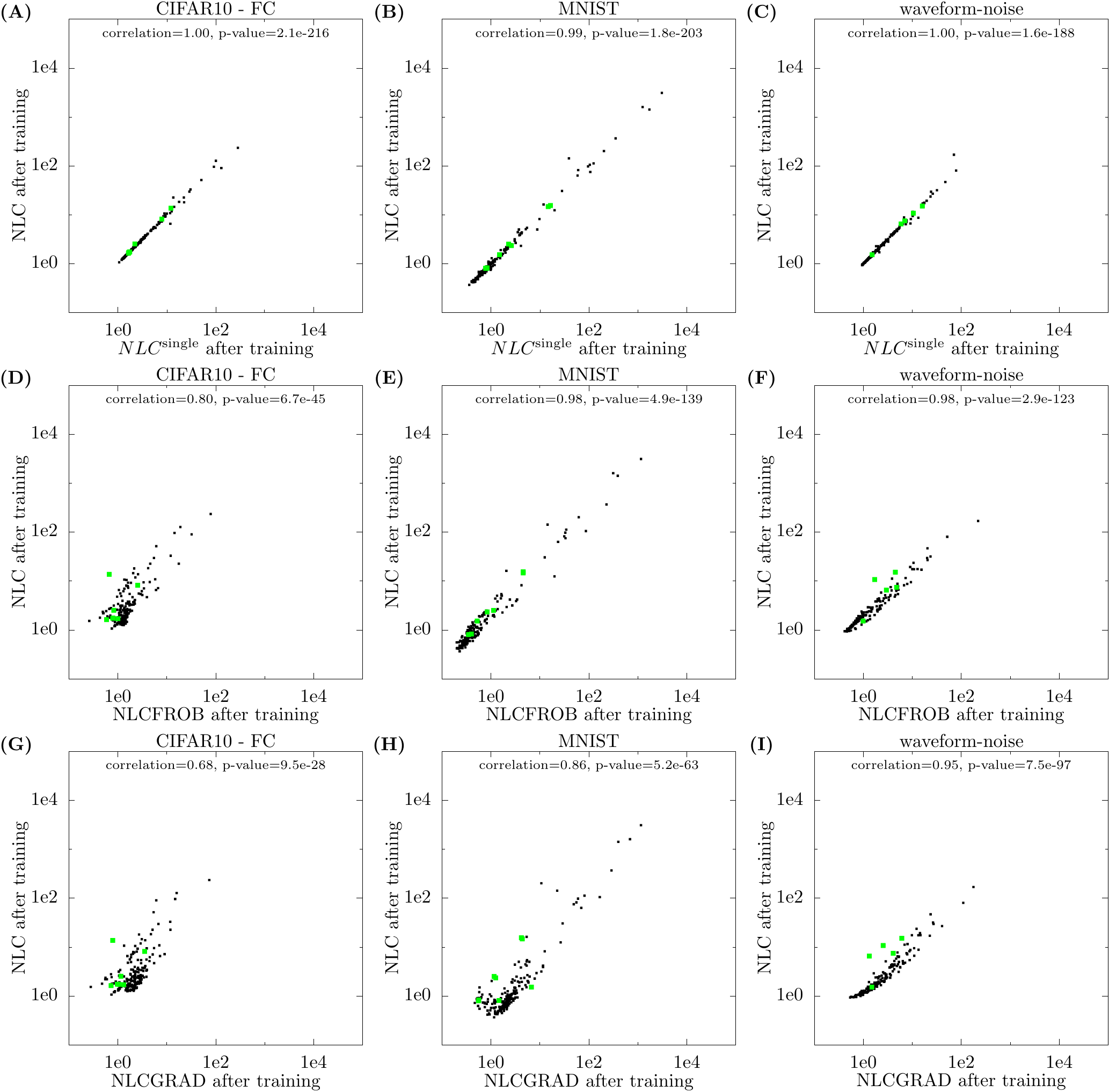}
\caption{NLC vs simpler metrics / estimators for study A architectures in the final state. {\it Conclusion:} The more we simplify, the more the approximation deteriorates.} \label{nlcSimpleMetricsFinalFull}
\end{figure}

\begin{figure}[H]
\centering
\includegraphics[width=0.98\textwidth]{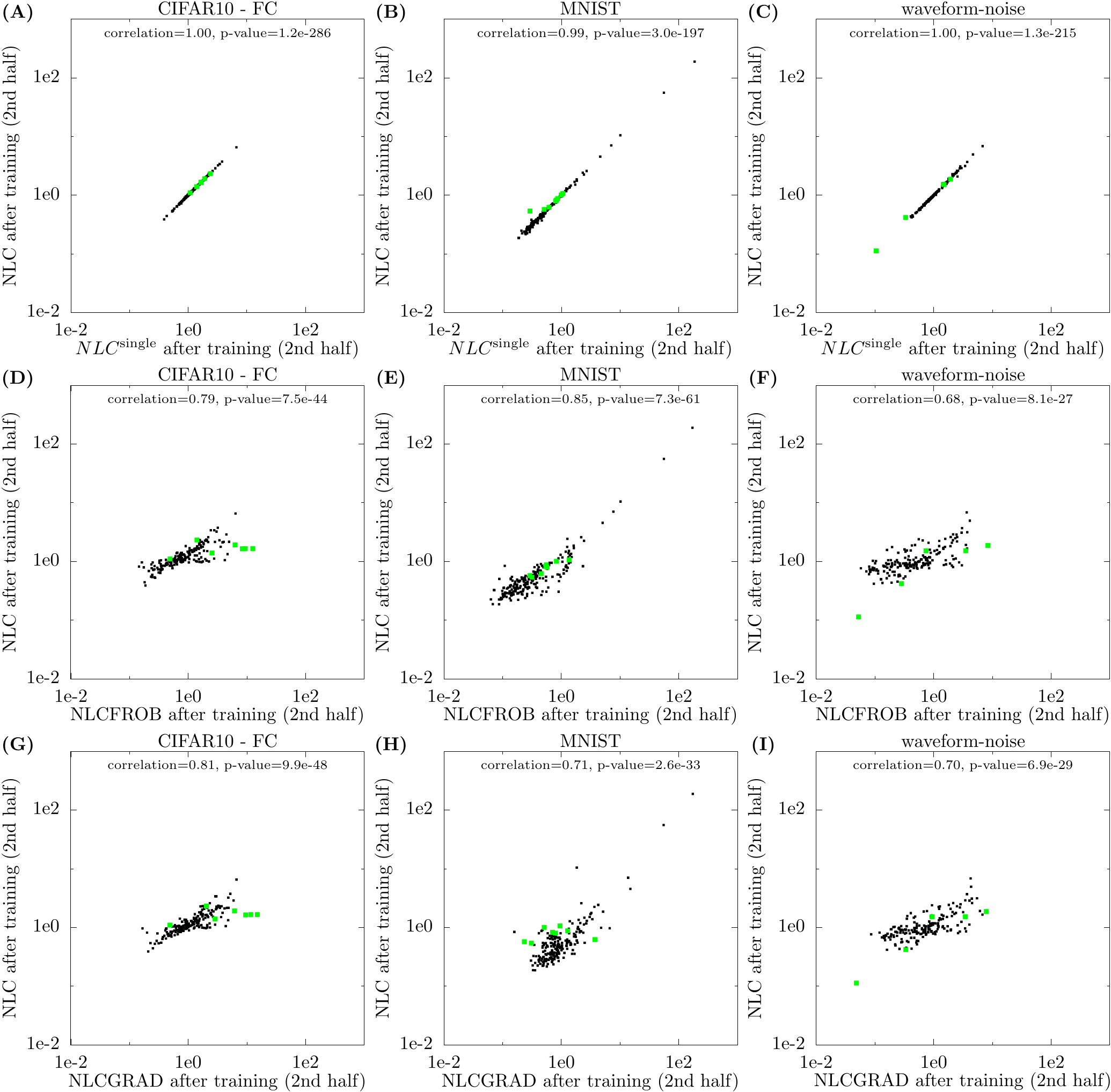}
\caption{The NLC vs simpler metrics / estimators for the second half of study A architectures in the final state. {\it Conclusion:} Simplifying the estimator works well, but simplifying the metric leads to significant deterioration.} \label{nlcSimpleMetricsFinalHalf}
\end{figure}

In figure \ref{nlcSimpleMetricsInitHalf}A-C, we plot $NLC^\text{single}$ vs the NLC in the initial state for the second half of study A architectures. We find that \finding{the match is again near-perfect except for GUAs, where it is still decent}. In figure \ref{nlcSimpleMetricsFinalFull}A-C and \ref{nlcSimpleMetricsFinalHalf}A-C, we make the comparison in the final state. \finding{The approximation deteriorates only slightly.}

\paragraph{For fully-connected networks, we can decouple the Jacobian and input covariance} The ``next level'' of simplification is to decouple the Jacobian from the input covariance and approximate $\mathbb{E}_{x,x',u\sim\mathcal{N}(0,I)}(u\mathcal{J}(x)(x'-\bar{x})^T)^2 \approx \frac{1}{d_\text{in}}\mathbb{E}_{x,u\sim\mathcal{N}(0,I)}||u\mathcal{J}(x)||^2_2\mathbb{E}_x||x-\bar{x}||_2^2$. 

\begin{metricDefinition}

$$NLCFROB(f,\mathcal{D}) = \sqrt{\frac{(\mathbb{E}_x\Tr(\mathcal{J}(x)\mathcal{J}(x)^T))\Tr(\Cov_x)}{d_\text{in}\Tr(\Cov_f)}} = \sqrt{\frac{\mathbb{E}_x||\mathcal{J}(x)||_F^2\Tr(\Cov_x)}{d_\text{in}\Tr(\Cov_f)}}$$

\end{metricDefinition}

We compute $\mathbb{E}_x||\mathcal{J}(x)||_F^2=\mathbb{E}_{x,u\sim\mathcal{N}(0,I)}||u\mathcal{J}(x)||^2_2$ similarly to the backward estimator of the NLC numerator in section \ref{nlcComputeSection}. We compute the traces like we compute the NLC denominator in section \ref{nlcComputeSection}.

\begin{figure}[H]
\centering
\includegraphics[width=0.98\textwidth]{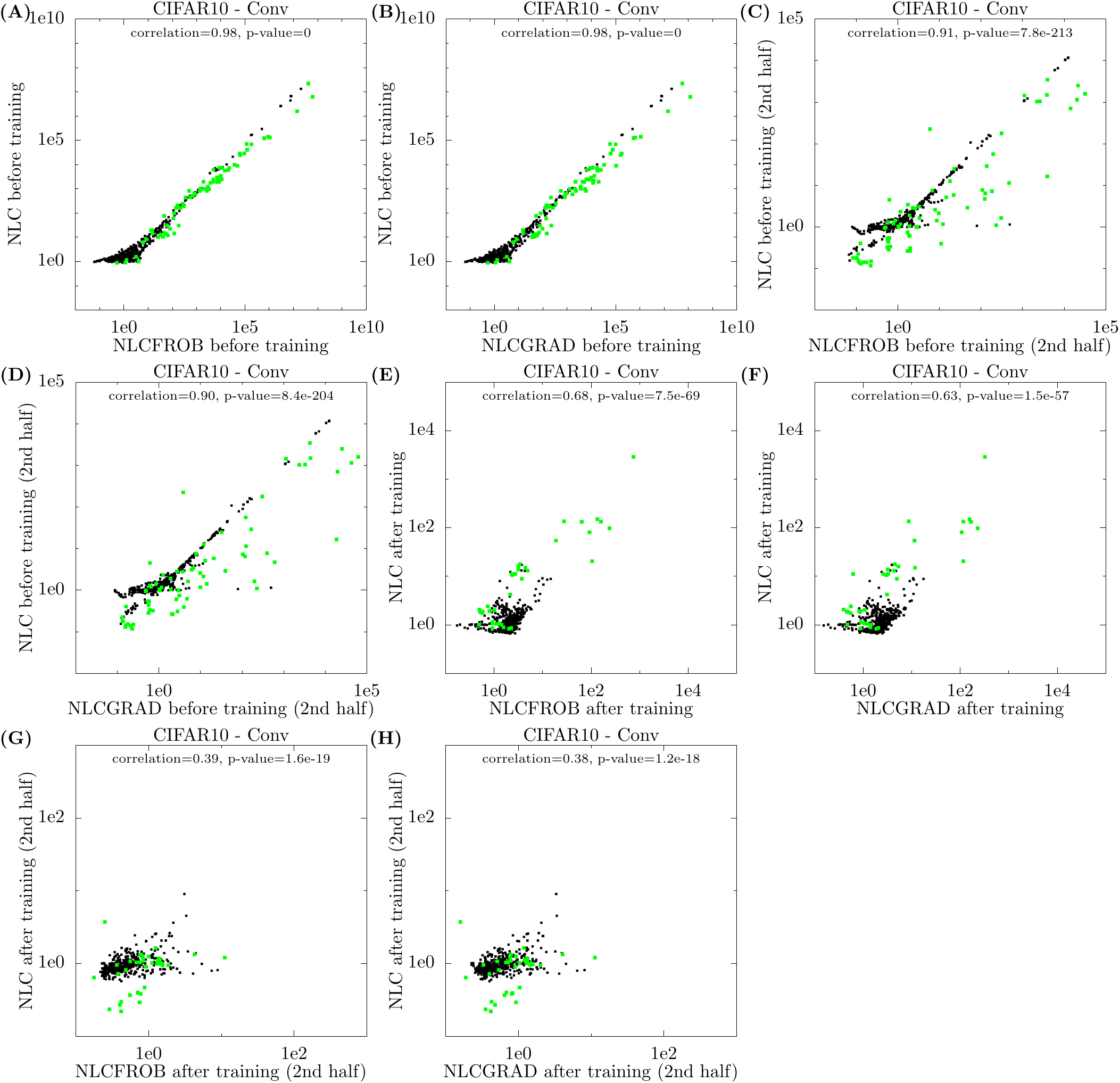}
\caption{The NLC vs simpler metrics for study B architectures. {\it Conclusion:} The simpler metrics are insufficient approximations, especially considering the lower bound of 1 no longer holds in the initial state.} \label{nlcSimpleMetricsConv}
\end{figure}

In figures \ref{nlcSimpleMetricsInitFull}D-F and \ref{nlcSimpleMetricsInitHalf}D-F, we find that \finding{this approximation still works almost perfectly in the initial state} except for GUAs. In figure \ref{nlcSimpleMetricsConv}A-C, we find that \finding{this is not true for convolutional networks from study B}. This is somewhat expected as convolutional networks take advantage of the covariance structure of images and similar types of data. Specifically, neighboring spatial locations tend to be highly correlated. In section \ref{nlcRobustDataSection}, we showed that, for fully-connected architectures, the NLC was preserved when the input was replaced by unit Gaussian noise. Here, we show that decoupling the covariance in the NLC does not work for convolutional networks, not even in the initial state. Interestingly, we also find that \finding{$NLCFROB \ge 1$ does not hold}.

After training, \finding{while the correlation between NLCFROB and the NLC is still substantial in many cases, the approximation deteriorates significantly} (figure \ref{nlcSimpleMetricsFinalFull}D-F \ref{nlcSimpleMetricsFinalHalf}D-F,  \ref{nlcSimpleMetricsConv}E/G). This follows the general trend we outlined in section \ref{nlcRobustDataSection}.

\paragraph{For fully-connected networks, we can further replace the Jacobian with a ratio of gradients} To estimate metrics involving the Jacobian, we generally need to take additional computational steps such as backpropagating Gaussian noise $u$, as described in section \ref{nlcComputeSection}. If we do not want to incur additional computational cost beyond backpropagating the gradient of the loss function, which is necessary for training, we can make the following further approximation.

\begin{metricDefinition}

$$NLCGRAD(f,\ell,\mathcal{D}) = \sqrt{\frac{d_\text{out}(\mathbb{E}_x||g_0||^2_2)\Tr(\Cov_x)}{d_\text{in}(\mathbb{E}_x||g_L||^2_2)\Tr(\Cov_f)}}$$

\end{metricDefinition}

$g_0 = \frac{d\ell}{dx}$ is obtained from $g_L = \frac{d\ell}{dx}$ through multiplication with the Jacobian, so we can proxy the magnitude of the Jacobian via its impact on the length of $g_L$. If $g_L$ was unit Gaussian distributed, this approximation would be exact and we would obtain $g_0 = u\mathcal{J}(x)$, which is an expression that we are by now familiar with.

Throughout figures \ref{nlcSimpleMetricsInitFull}, \ref{nlcSimpleMetricsInitHalf}, \ref{nlcSimpleMetricsFinalFull} and \ref{nlcSimpleMetricsFinalHalf}, we find that \finding{the approximation obtained from NLCGRAD is nearly identical in quality to NLCFROB}. A caveat is that we only consider a single loss function.

\newpage

\newpage

\subsection{The NLC is related to the size of linearly approximable regions} \label{nlcSensiSection}

We began this chapter with a discussion of nonlinearity. We established the NLC as a measure of nonlinearity through its relationship with NL1D, theoretical results and an eye test. In this subsection and the next, we uncover yet more ways in which the NLC measures network nonlinearity. Thus, we further establish the NLC as a meaningful and deep metric, in the vein of utility criterion \ref{criterionMeaningful}. In this subsection, we show that the NLC is closely related to the size of linearly approximable regions of a network in input space.

As mentioned in section \ref{whatIsNonlinearitySection}, a key property of linear functions $F : \mathbb{R}^{d_\text{in}} \rightarrow \mathbb{R}^{d_\text{out}}$ is that the `gradient-based local linear approximation' (GLLA) $F(\chi) + (\chi' - \chi)\frac{dF(\chi)}{d\chi}^T$ taken at any $\chi\in \mathbb{R}^{d_\text{in}}$ is equal to $F(\chi')$ everywhere. To the degree to which this is true for an arbitrary differentiable function $F$, we can say it is more or less nonlinear. This makes sense especially in the context of neural networks trained with gradient methods. They rely on the fact that an accurate local linear approximation around the parameter value is available. The findings of this subsection might also be valuable for understanding the network training process.

To develop a metric to capture the property of ``local linear approximability'', like in section \ref{nlcDefinitionSection}, we first turn to the case of a function $F$ with a bounded domain. Let us also assume that this domain is convex. Consider some $\chi$ in that domain and some $\chi_\text{bound}$ on the boundary of that domain. Then we can say that the greater the fraction of the distance from $\chi$ to $\chi_\text{bound}$ the GLLA at $\chi$ remains accurate, the more linearly approximable $F$ is at $\chi$ in the direction of $\chi_\text{bound}$. We can measure this accuracy in terms of whether the GLLA remains within some tolerance of $F$ when projected onto a certain direction in output space. Formally, given some tolerance $T$ and (row) vector $\delta_\text{out} \in \mathbb{R}^{d_\text{out}}$, we can measure the local linear approximablity of $F$ with respect to $(\chi,\chi_\text{bound},\delta_\text{out},T)$ via the largest $C \le 1$ such that for all $c \le C$ we have $$\frac{c}{T}\delta_\text{out}\frac{dF(\chi)}{d\chi}(\chi_\text{bound} - \chi)^T \le \delta_\text{out}(F(\chi + c(\chi_\text{bound} - \chi)) - F(\chi))^T \le cT\delta_\text{out}\frac{dF(\chi)}{d\chi}(\chi_\text{bound} - \chi)^T$$

In order to turn this into a practical metric for neural networks, we have to (i) replace the bounded, convex domain with the input distribution $\mathcal{D}$ and (ii) eliminate the need to manually specify $\chi$, $\chi_\text{bound}$ and $\delta_\text{out}$. Just as with the NLC, we will draw the input from $\mathcal{D}$ and model the domain as a Gaussian with covariance $\Cov_x$. Specifically, we draw $\chi_\text{bound} - \chi$ from $\mathcal{U}\Cov_x^{\frac{1}{2}}$, where $\mathcal{U}$ is the uniform distribution over vectors of dimensionality $d_\text{in}$ and length $\sqrt{d_\text{in}}$. This distribution has the same covariance as $\mathcal{D}$. It can be viewed as the uniform distribution over ``1-standard deviation offsets in $\mathcal{D}$'', or as the uniform distribution over radii of $\mathcal{D}$. Finally, (i) we draw $\delta_\text{out}$ from the unit Gaussian $\mathcal{N}(0,I_{d_\text{out}})$, (ii) we consider batches instead of individual inputs and (iii) we invert the value of $C$ to arrive at the metric below.

\begin{metricDefinition}
The `gradient-based local linear approximability distribution' (GLLAD) of a network $f$ with respect to $\mathcal{D}$, tolerance $T$ and batch size $|B|$ is the distribution over $C$, where $C$ is the smallest value greater or equal to 1 such that for all $c \le \frac{1}{C}$ we have $$\frac{c}{T}\sum_{b=1}^{|B|}\delta_\text{out}^{(b)}\mathcal{J}(x^{(b)})\delta_\text{in}^{(b)T} \le \sum_{b=1}^{|B|}\delta_\text{out}^{(b)}(f(x^{(b)} + c\delta_\text{in}^{(b)}) - f(x^{(b)}))^T \le cT\sum_{b=1}^{|B|}\delta_\text{out}^{(b)}\mathcal{J}(x^{(b)})\delta_\text{in}^{(b)T}$$ The distribution over $C$ is induced by $x^{(b)} \sim \mathcal{D}$, $\delta_\text{in}^{(b)} \sim \mathcal{U}\Cov_x^{\frac{1}{2}}$ and $\delta_\text{out}^{(b)} \sim \mathcal{N}(0, I_{d_\text{out}})$, drawn independently for $1 \le b \le |B|$. $\mathcal{U}$ is the uniform distribution over vectors of dimensionality $d_\text{in}$ and length $\sqrt{d_\text{in}}$.
\end{metricDefinition}

Linear networks achieve $C=1$ with probability 1. In general, $C \ge 1$. Note that $C=1$ can hold for nonlinear networks, even with probability 1. We define GLLAD over batches instead of individual inputs so we can easily generalize it to networks with batch normalization. We invert the value of $C$ so that larger $C$ corresponds to larger nonlinearity, just as with the NLC. Note that GLLAD effectively measures the radius of linearly approximable regions relative to the radius of the domain, not the volume of linearly approximable regions relative to the volume of the domain. Informally speaking, the number of linearly approximable regions required to cover the domain behaves as $GLLAD^{d_\text{in}}$.

\begin{figure}[t]
\includegraphics[width=0.98\textwidth]{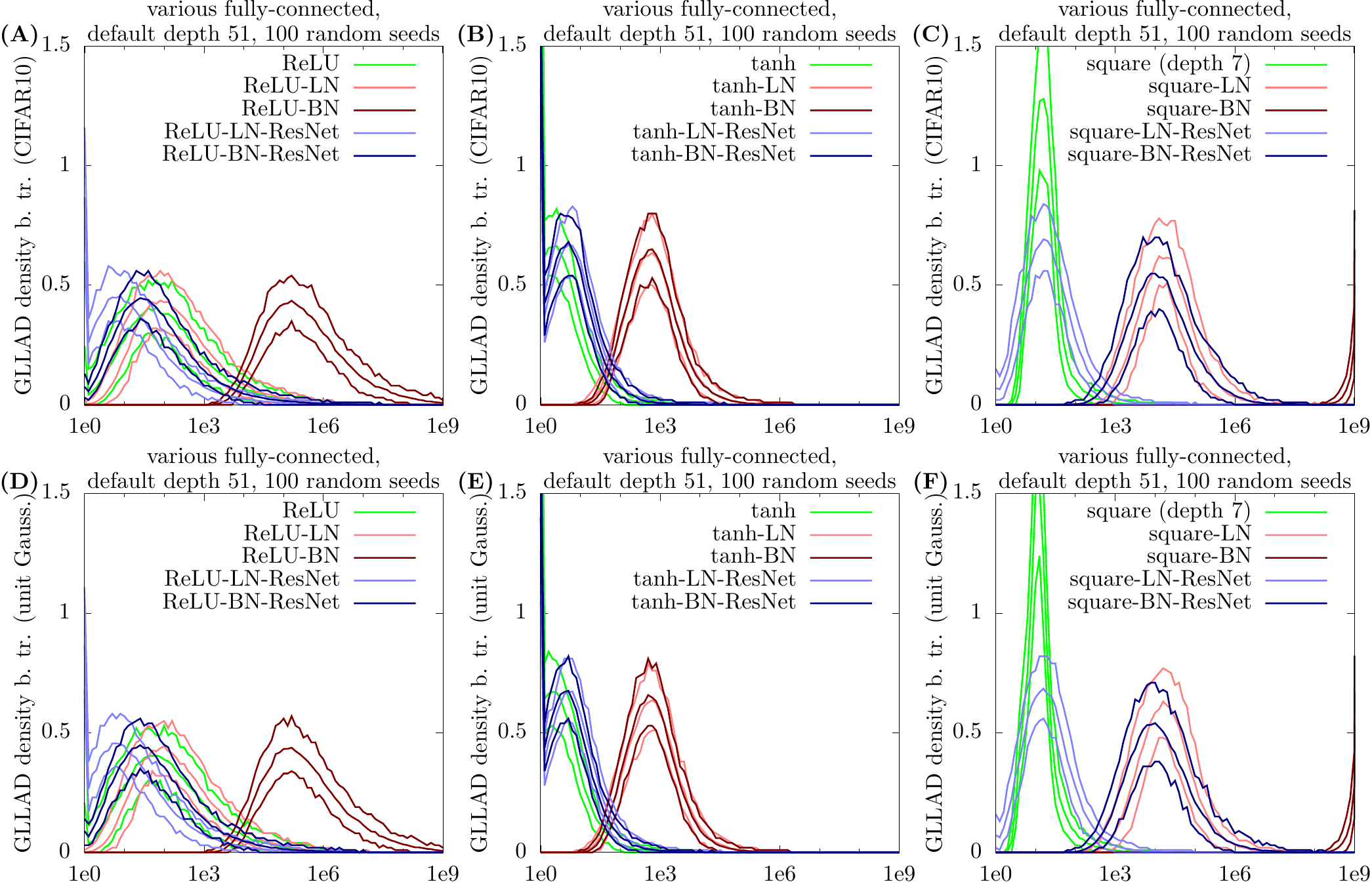}
\caption{GLLAD for 15 simple fully-connected architectures of depth 51 (unless otherwise specified) in their initial state on CIFAR10 (top row) and unit Gaussian input (bottom row). For each architecture, the lower curve denotes the 5th percentile over 100 random seeds. The middle curve denotes the mean over those seeds. The upper curve denotes the 95th percentile. The density is defined over the base 10 logarithm. We set $T=2$. $|B|=250$ as always for fully-connected architectures. Note that we could not compute the density in the region greater $10^9$, due to limitations associated with floating-point computation. {\it Conclusion:} All distributions are well-behaved and can be summarized by e.g. their median. They are also relatively robust to random seed change. Both CIFAR10 and Gaussian inputs yield the same results.} \label{sensiDist}
\end{figure}

To build an understanding of GLLAD, we plot it for 15 simple fully-connected architectures in the initial state on CIFAR10 in figure \ref{sensiDist}(A-C). For each architecture, we consider 100 different random seeds. Random seed controls the parameter value and other things specified in section \ref{additionalExperimentsSection}. We plot the mean, the 5th percentile and the 95th percentile of the 100 distributions in log space. We find that \finding{all distributions are well-behaved and have a clear, single peak that is relatively consistent as the random seed changes. This is true even for architectures `square' and `square-BN-ResNet', which are GUAs}. Therefore, we summarize the GLLAD with a single value, its median.

\begin{metricDefinition}
The `median gradient-based local linear approximability' (MGLLA) is the median of the GLLAD defined above, i.e. $\mathbb{M}_{x^{(b)},\delta_\text{in}^{(b)},\delta_\text{out}^{(b)}\text{ for }1\le b \le |B|}GLLAD$.
\end{metricDefinition}

In figure \ref{sensiDist}(D-F), we plot GLLAD for the same architectures as in (A-C), but evaluated on unit Gaussian input analogously to sections \ref{nlcRobustDataSection} and figure \ref{discillu}. \finding{The plots are near-identical to (A-C), even for GUAs.} Again, nonlinearity proves robust to changes in input distribution.

\begin{figure}[t]
\includegraphics[width=0.98\textwidth]{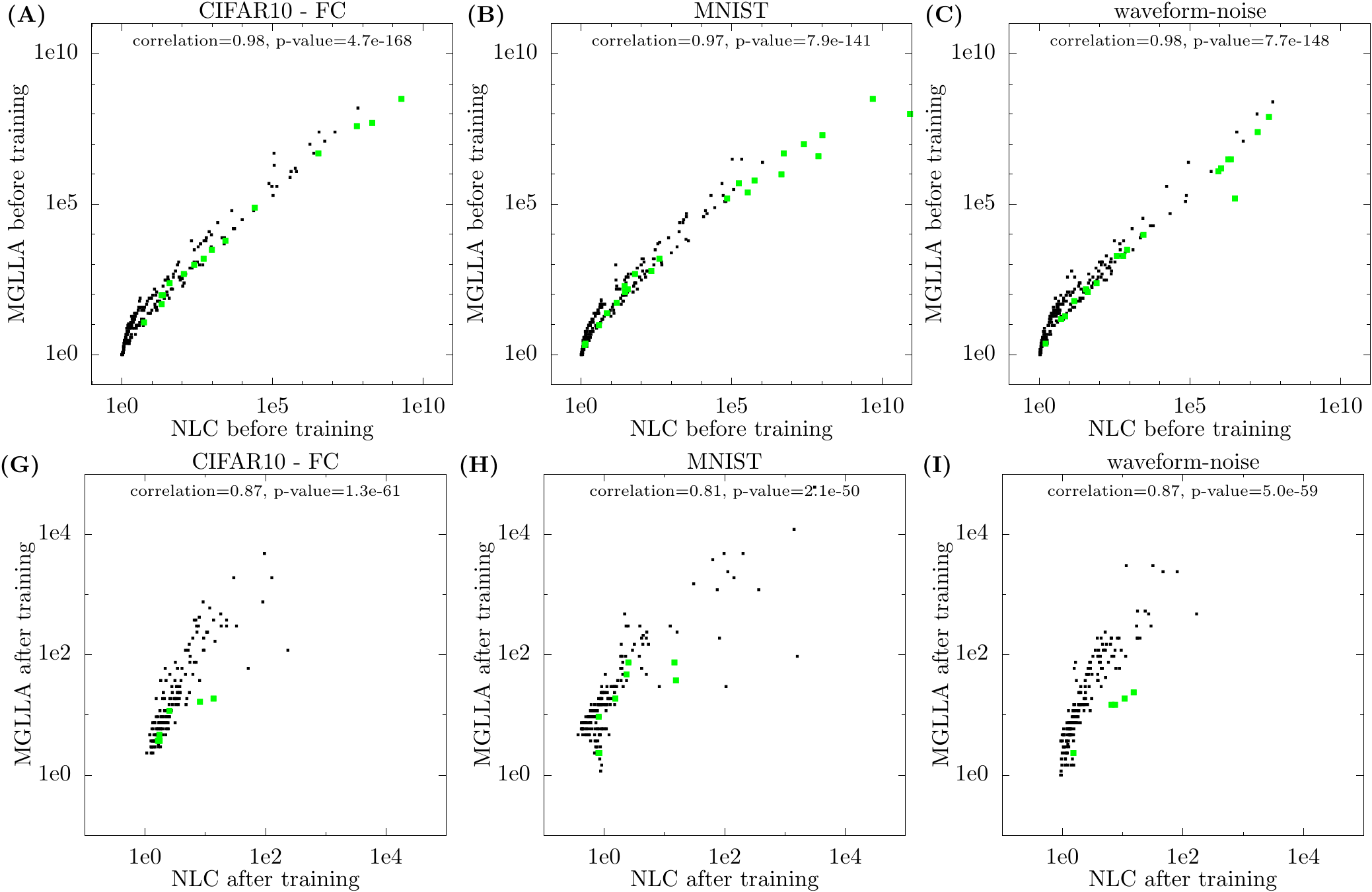}
\caption{NLC vs MGLLA for study A architectures. We set $T=2$. Only architectures for which $MGLLA < 10^9$ are depicted, due to limitations associated with floating-point computation. {\it Conclusion:} Both metrics are highly associated, especially in the initial state.} \label{nlcSensiGrad}
\end{figure}

\paragraph{NLC vs MGLLA: empirical relationship} Now we are ready to investigate whether the NLC is a measurement of nonlinearity by way of measuring local linear approximability. In figure \ref{nlcSensiGrad}, we plot the NLC vs MGLLA for all study A architectures in the initial and final state. We find that \finding{both values are highly correlated, especially in the initial state}. \finding{In the initial state, there are no outliers, not even GUAs}. \finding{The NLC somewhat underestimates MGLLA, especially in the final state}.

\begin{wrapfigure}[17]{r}{5.5cm}
\includegraphics[width=5.4cm]{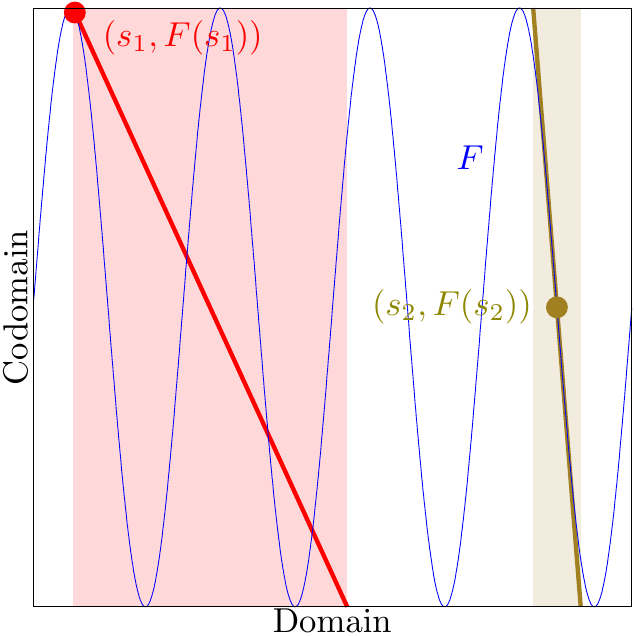}
\caption{1d pictorial illustration of the connection between NLC and MGLLA.}\label{sinillu}
\end{wrapfigure}

\paragraph{NLC vs MGLLA: intuitive connection} In figure \ref{sinillu}, we illustrate the intuitive connection between NLC and GLLA with a simple 1D example, similarly to section \ref{nlcDefinitionSection}. Assume $F$ is a sine curve, depicted in blue. $s_1$ and $s_2$ are two scalars. We plot the location of $(s_1,F(s_1))$ in red and the location of $(s_2,F(s_2))$ in olive. The thick red and olive lines correspond to the GLLA of $F$ at $s_1$ and $s_2$ respectively, which is simply the tangent line of the blue curve. The shaded olive and red regions correspond to the intervals in which the tangent falls inside the codomain, which is defined to be the set of values taken by $F$ over the domain.

It is easy to check that the proportion of the domain covered by the red interval and olive interval is $\frac{|\mathsf{co}|}{|F'(s_1)||\mathsf{dom}|}$ and $\frac{|\mathsf{co}|}{|F'(s_2)||\mathsf{dom}|}$ respectively. The key insight is that both tangents can only be close to $F$ while they remain inside the codomain, and therefore within their respective shaded area. This is evidently true in both cases, as both tangent lines quickly move away from $F$ outside the shaded region. In the case of $s_2$, this bound is also tight as the tangent tracks $F$ closely everywhere in the olive region. However, in the case of $s_1$, the bound is loose, as the red line completely decouples from $F$ throughout a large part of the red region. 

So, we can view $\frac{|\mathsf{co}|}{|F'(s)||\mathsf{dom}|}$ as an upper bound on gradient-based local linear approximability, so we can view $\frac{\sqrt{\mathbb{E}_sF'(s)^2}|\mathsf{dom}|}{|\mathsf{co}|}$ as a lower bound on MGLLA. But we can view the NLC as a generalization of $\frac{\sqrt{\mathbb{E}_sF'(s)^2}|\mathsf{dom}|}{|\mathsf{co}|}$ to multi-dimensional networks, where $F'$ becomes the Jacobian, the diameter of the domain is measured by $2\sqrt{\Tr(\Cov_x)}$ and the diameter of the codomain is measured by $2\sqrt{\Tr(\Cov_f)}$. This is the exact same generalization process as in section \ref{nlcDefinitionSection}. In summary, we would expect the NLC to be a tight lower bound of MGLLA, and this is exactly what we find in figure \ref{nlcSensiGrad}.

From the perspective of training neural networks, it is interesting that we found such a close match between MGLLA and NLC. This implies that GLLAs of practical neural networks are close to the true value of $f$ as long as the value of the GLLA remains inside the range of values taken by $f$. So GLLAs of practical networks tend to behave like the olive tangent in figure \ref{sinillu}, and not like the red tangent. If a similar property holds for the GLLA around the parameter, then a single gradient update can shift the network output to a desired value for a given input $x$ without the update having to cross the boundary of the approximable region. This seems to be a highly valuable property for the purpose of trainability that we believe warrants further investigation.

\paragraph{Computing MGLLA} We could not (fully) compute GLLAD / MGLLA for some architectures, such as square-BN in figure \ref{sensiDist}. Computing these metrics is challenging in general. To obtain the approximate value of $C$ for some batch of triplets $(x,\delta_\text{in}, \delta_\text{out})$, we need to start with an infinitesimal value of $c$, verify that the GLLA is accurate, and then gradually increase $c$ while retesting the GLLA until it is no longer accurate. This causes a dilemma. If the first tested value of $c$ is too large, it will already exceed $\frac{1}{C}$, and the first test will fail. Conversely, if the first value of $c$ is too small, then we may experience catastrophic rounding error during forward propagation. In section \ref{nlcComputeSection}, we explained how we couldn't compute the NLC when network outputs were indistinguishable because of rounding error. When inputs to a network are only a tiny distance apart, that problem is correspondingly exacerbated. 

We determined that, by using 64-bit precision, we were able to compute values of $C$ as long as they were at most $10^9$. Hence, both figures \ref{sensiDist} and \ref{nlcSensiGrad} are cut off at $GLLAD = 10^9$ and $MGLLA = 10^9$ respectively. We see no reasons why the patterns we detect should not hold for more nonlinear networks.

Another challenge for computing MGLLA is that we need to compute the full $\Cov_x$ matrix, instead of just its trace like in the NLC. This is both computationally more expensive and statistically more uncertain, as it is not clear how the estimation errors that occur for each entry of $\Cov_x$ compound in the context of the MGLLA metric. This challenge applies to several other metrics defined later that also require $\Cov_x$. $\Cov_x$ is estimated via elementwise sample covariance as described in section \ref{metricEstimationSection}.

We considered investigating the MGLLA metric as our primary metric and base this work on MGLLA instead of NLC. We even developed a mean field theory of MGLLA. The deciding factors were that the NLC is much more computable and takes a much simpler form.

\subsection{The NLC is related to $L2$ linear approximation error and underfitting} \label{nlcLinearApproximationSection}

In the previous subsection, we showed a strong relationship between the NLC and local linear approximability. In this subsection, we show a relationship between the NLC and {\it global} linear approximability in an $L2$ sense. We do this via two theorems that we illustrate using our activation functions. We also begin explaining {\it why} the NLC is related to test error, by connecting the NLC to underfitting.

We begin by introducing a simple decomposition of a network $f$.

\begin{definition}
Denote the least squares linear fit to $f$ under some input distribution $\mathcal{D}$ by $xA_f + b_f$.
\begin{itemize}
\item The `linear component' of $f$ is $(x-\bar{x})A_f$.
\item The `constant component' of $f$ is $b_f+\bar{x}A_f$.
\item The `nonlinear component' of $f$ is $\tilde{f} = f - xA_f - b_f$.
\item The `nonlinear basis' of $f$ is $$\frac{\tilde{f}}{\mathbb{E}||\tilde{f}||_2^2}$$
\end{itemize}
\end{definition}
\begin{metricDefinition} Let $A_f$, $b_f$ and $\tilde{f}$ be defined as above.
\begin{itemize}
\item The `linear approximation ratio' (LAR) is $$LAR(f,\mathcal{D}) = \frac{\mathbb{E}||(x-\bar{x})A_f||_2^2}{\mathbb{E}||f||_2^2}$$
\item The `constant approximation ratio' (CAR) is $$CAR(f,\mathcal{D}) = \frac{||b_f+\bar{x}A_f||_2^2}{\mathbb{E}||f||_2^2}$$
\item The `nonlinear approximation ratio' (NAR) is $$NAR(f,\mathcal{D}) = \frac{\mathbb{E}||\tilde{f}||_2^2}{\mathbb{E}||f||_2^2}$$
\end{itemize}
\end{metricDefinition}

Based on assumption \ref{assumptionPositive}, we have $\mathbb{E}||f||_2^2 = \Tr(\Cov_f) + ||\mathbb{E}f||_2^2 > 0$. The name `nonlinear basis' expresses that $((x-\bar{x})A_f,b_f+\bar{x}A_f,\tilde{f})$ is a decomposition of $f$ into three functions which can be viewed as a basis in function space. The three functions are orthogonal and thus we have $CAR + LAR + NAR = 1$. Further, the nonlinear component is orthogonal to arbitrary linear functions.

\begin{proposition} \label{finiteNetBasisOrtho}
We have
\begin{enumerate}
\item $\mathbb{E}F^T\tilde{f} = 0$ and $\mathbb{E}F\tilde{f}^T = 0$ for any linear function $F : \mathbb{R}^{d_\text{in}} \rightarrow \mathbb{R}^{d_\text{out}}$
\item $\mathbb{E}((x-\bar{x})A_f)\tilde{f}^T = 0$, $\mathbb{E}(b_f+\bar{x}A_f)\tilde{f}^T = 0$ and $\mathbb{E}((x-\bar{x})A_f)(b_f+\bar{x}A_f)^T = 0$
\item LAR + CAR + NAR = 1
\end{enumerate}
\end{proposition}

$NLC(\tilde{f},\mathcal{D})$ is valid as long as $f$ is not linear.

\begin{proposition} \label{finiteNetPositiveBasis}
Assume there exists an open set $S$ where $\mathcal{D}$ has a continuous, positive density function and $f$ is not linear on $S$. Then $\Tr(\Cov_{\tilde{f}})> 0$.
\end{proposition}

Now we can state the two key theorems of this subsection.

\begin{theorem} \label{finiteNetTilde}
Let $\mathcal{D}$ be Gaussian and let the linear component of $f$ be the zero function. Then we have $$NLC(f,\mathcal{D}) \ge \sqrt{2}$$
\end{theorem}

\begin{theorem} \label{finiteNetLAR}
Let $\mathcal{D}$ be Gaussian and assume $f$ is not linear. Then we have $$NLC(f,\mathcal{D})^2 = \frac{LAR}{NAR + LAR} + \frac{NAR}{NAR + LAR}NLC(\tilde{f},\mathcal{D})^2$$
\end{theorem}
 
These theorems significantly extend theorem \ref{finiteNetNlcGreater1}, which originally motivated the NLC as a nonlinearity measure. Again, we assume that $\mathcal{D}$ is Gaussian. Theorem \ref{finiteNetTilde} states that networks that do not have a linear component have their NLC bounded away from 1, at $\sqrt{2}$. Theorem \ref{finiteNetLAR} states that the square NLC of a network that has a nonlinear component is the weighted average of the square NLC of the nonlinear component, which is $NLC(\tilde{f},\mathcal{D})^2$, and the square NLC of the linear component, which is 1. Each squared NLC is weighted by the relative power of the respective component. In other words, the square NLC of a network is proportional to the $L2$ error incurred when approximating that function with a linear function, relative to the overall magnitude of the function. This is yet another avenue for connecting NLC and nonlinearity. Note that the presence and power of the constant component is irrelevant for theorem \ref{finiteNetLAR}.

Finally, we can connect both theorems with the observation that $NLC(\tilde{f},\mathcal{D}) \ge \sqrt{2}$ because the linear component of $\tilde{f}$ is the zero function.

\begin{proposition} \label{finiteNetTheoremConnector}
Let $xA_{\tilde{f}}+b_{\tilde{f}}$ be the least squares linear fit to $\tilde{f}$. Then $A_{\tilde{f}}= 0$ and $b_{\tilde{f}}=0$.
\end{proposition}

Another important consequence of these theorems is that they provide a first explanation for the NLCs relationship to test error. In the initial state, when the NLC of an architecture on the dataset mirrors the NLC on Gaussian input, an NLC close to 1 implies that the network is close to a linear function in an $L2$ sense. Hence, if a linear model underfits on a given dataset, the network is likely to underfit as well. This is what we find in figure \ref{nlcPredTrainInit}C. While it is possible for the properties of a network to change during training, relying on such a change might not be desirable. We investigate the change of the NLC during training in e.g. sections \ref{nlcEvolutionSection} and \ref{meanFieldPracticalSection}.

It is difficult to compute LAR, CAR and NAR on practical networks with respect to practical input distributions. Therefore, we restrict our empirical investigation in this section to activation functions and $\mathcal{N}(0,1)$. The properties of this function-distribution pair will be important, especially in chapters \ref{meanFieldNnaChapter} and \ref{nlnormChapter}. The standard formula for the least squares linear fit yields $A_\tau = \mathbb{E}_{s\sim\mathcal{N}(0,1)}s\tau(s)$ and $b_\tau = \mathbb{E}_{s\sim\mathcal{N}(0,1)}\tau(s)$, where both $A_\tau$ and $b_\tau$ are scalars.

\begin{table}
{
\centering
\begin{tabular}{lcccccc}
Act. fun. &ReLU&SELU&softplus&Swish&abs. val.&tanh\\ \hline\hline
Formula&$\max(s,0)$&{\bf \textdagger}&$\log(1+e^{-s})$&$\frac{s}{1+e^{-s}}$&$|s|$&$\tanh(s)$\\
Illustration&\includegraphics[scale=0.13,valign=c]{graphsFinal/actFunillu/relu.pdf}&\includegraphics[scale=0.13,valign=c]{graphsFinal/actFunillu/selu.pdf}&\includegraphics[scale=0.13,valign=c]{graphsFinal/actFunillu/softplus.pdf}&\includegraphics[scale=0.13,valign=c]{graphsFinal/actFunillu/swish.pdf}&\includegraphics[scale=0.13,valign=c]{graphsFinal/actFunillu/abs.pdf}&\includegraphics[scale=0.13,valign=c]{graphsFinal/actFunillu/tanh.pdf}\\
\begin{tabular}[x]{@{}l@{}}Nonlinear\\basis\end{tabular}&\includegraphics[scale=0.13,valign=c]{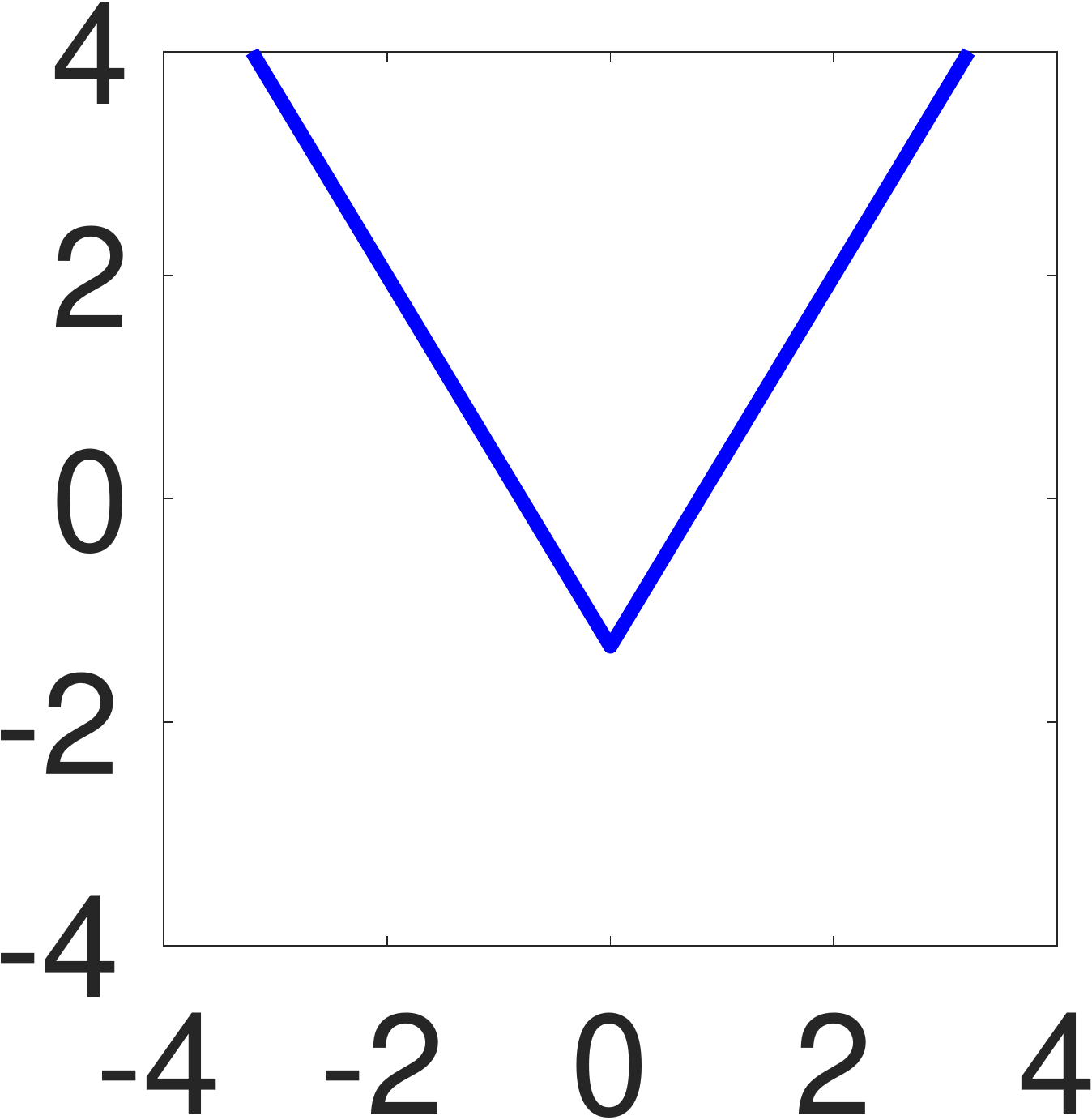}&\includegraphics[scale=0.13,valign=c]{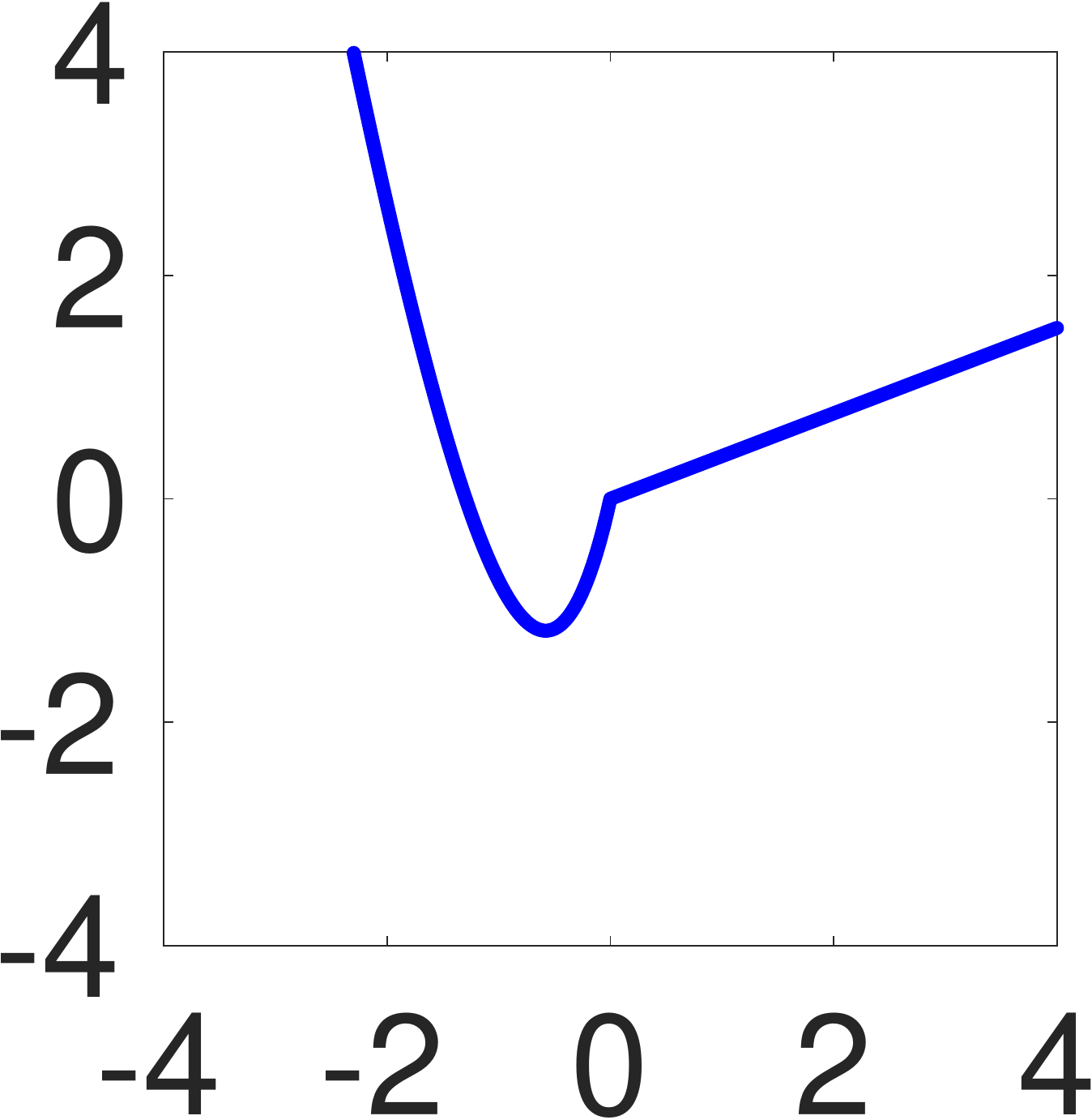}&\includegraphics[scale=0.13,valign=c]{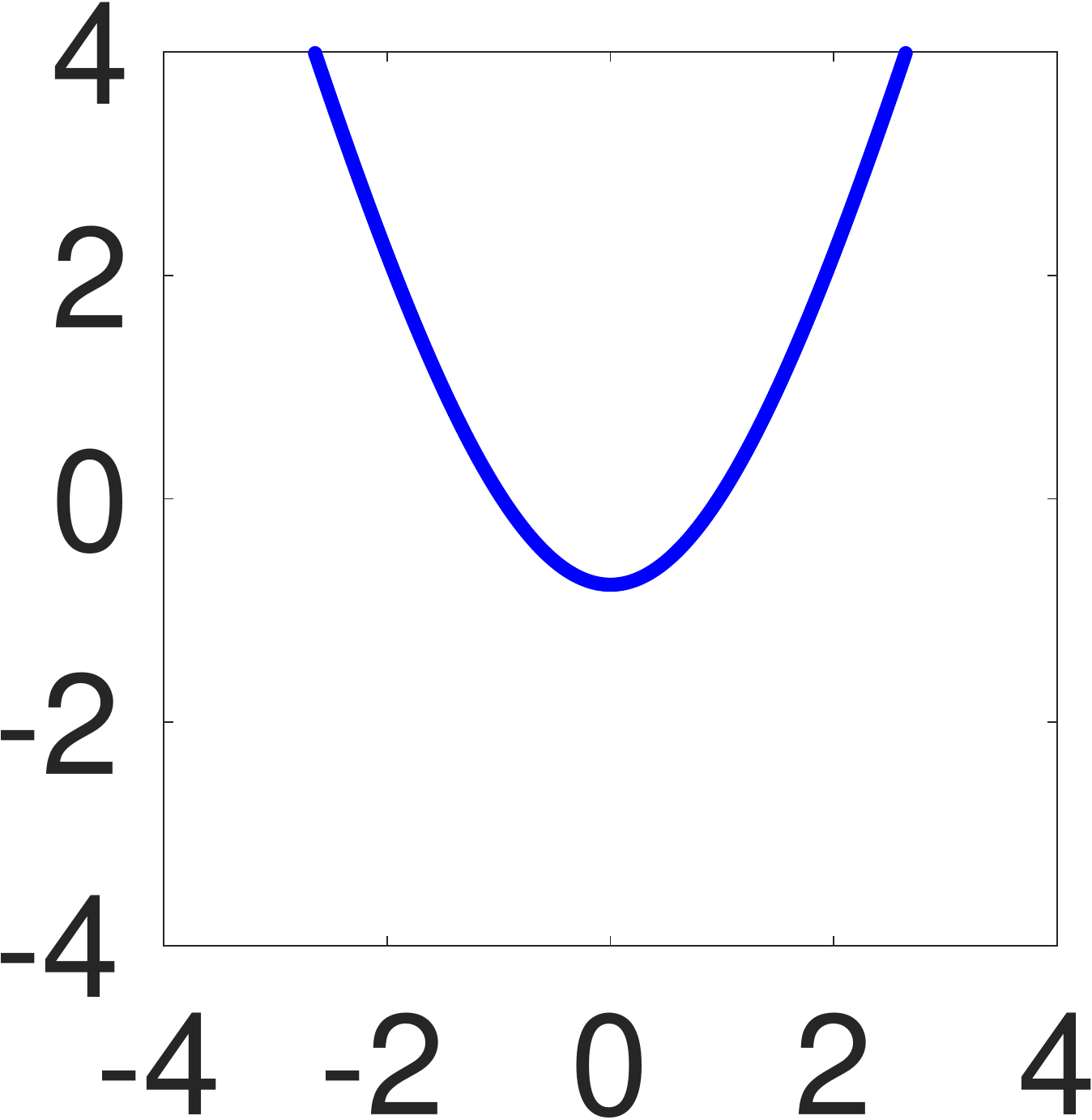}&\includegraphics[scale=0.13,valign=c]{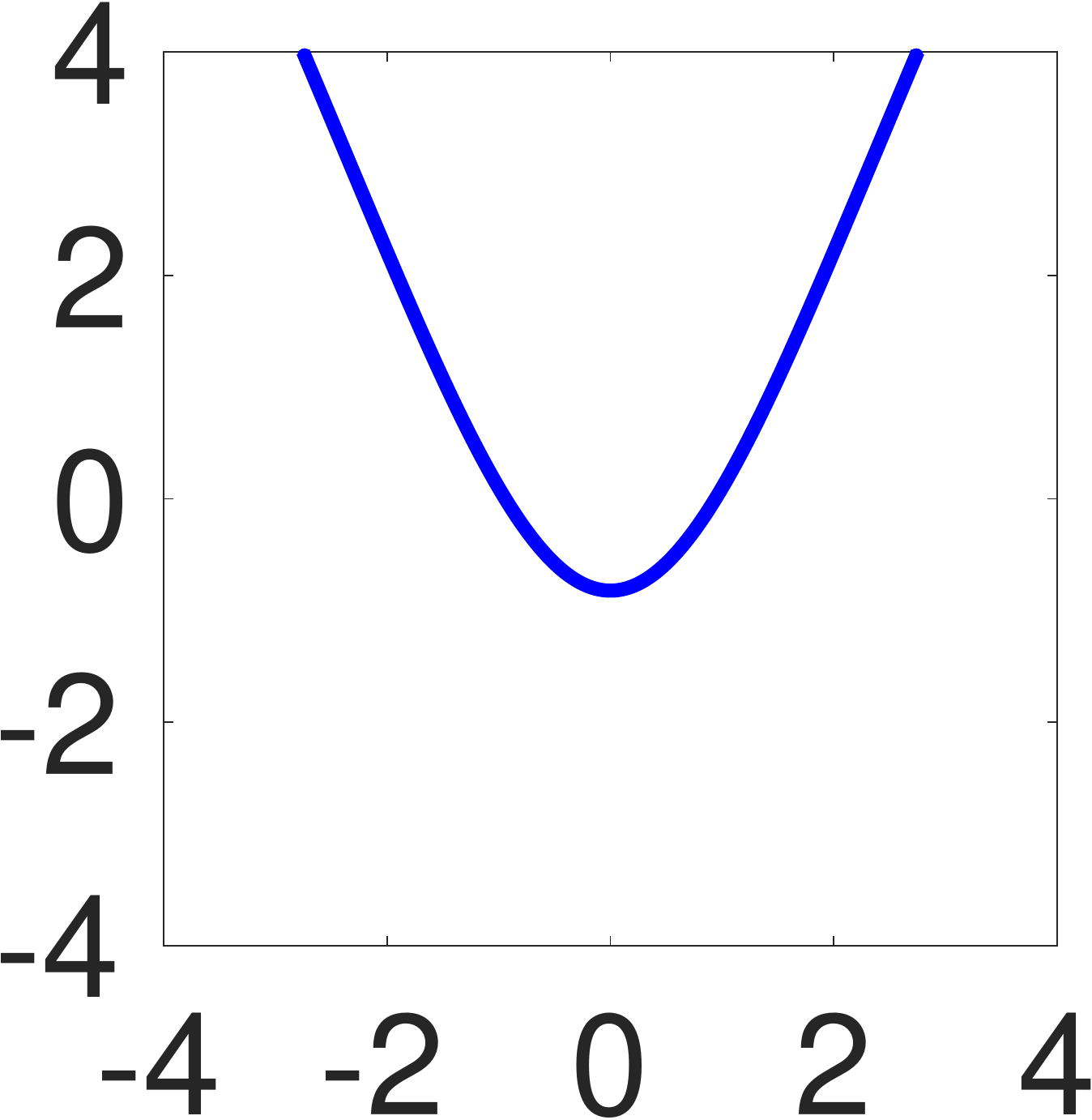}&\includegraphics[scale=0.13,valign=c]{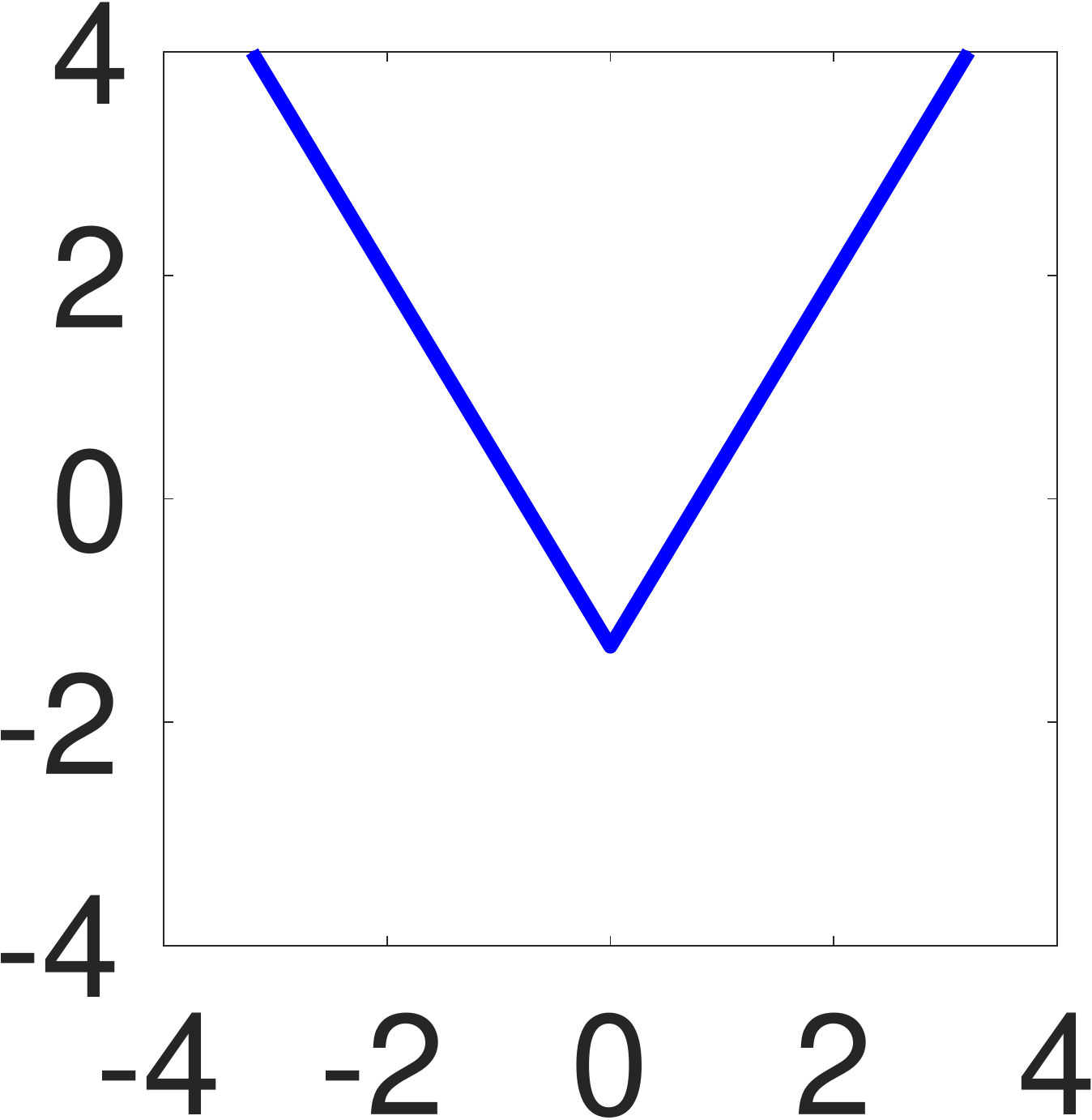}&\includegraphics[scale=0.13,valign=c]{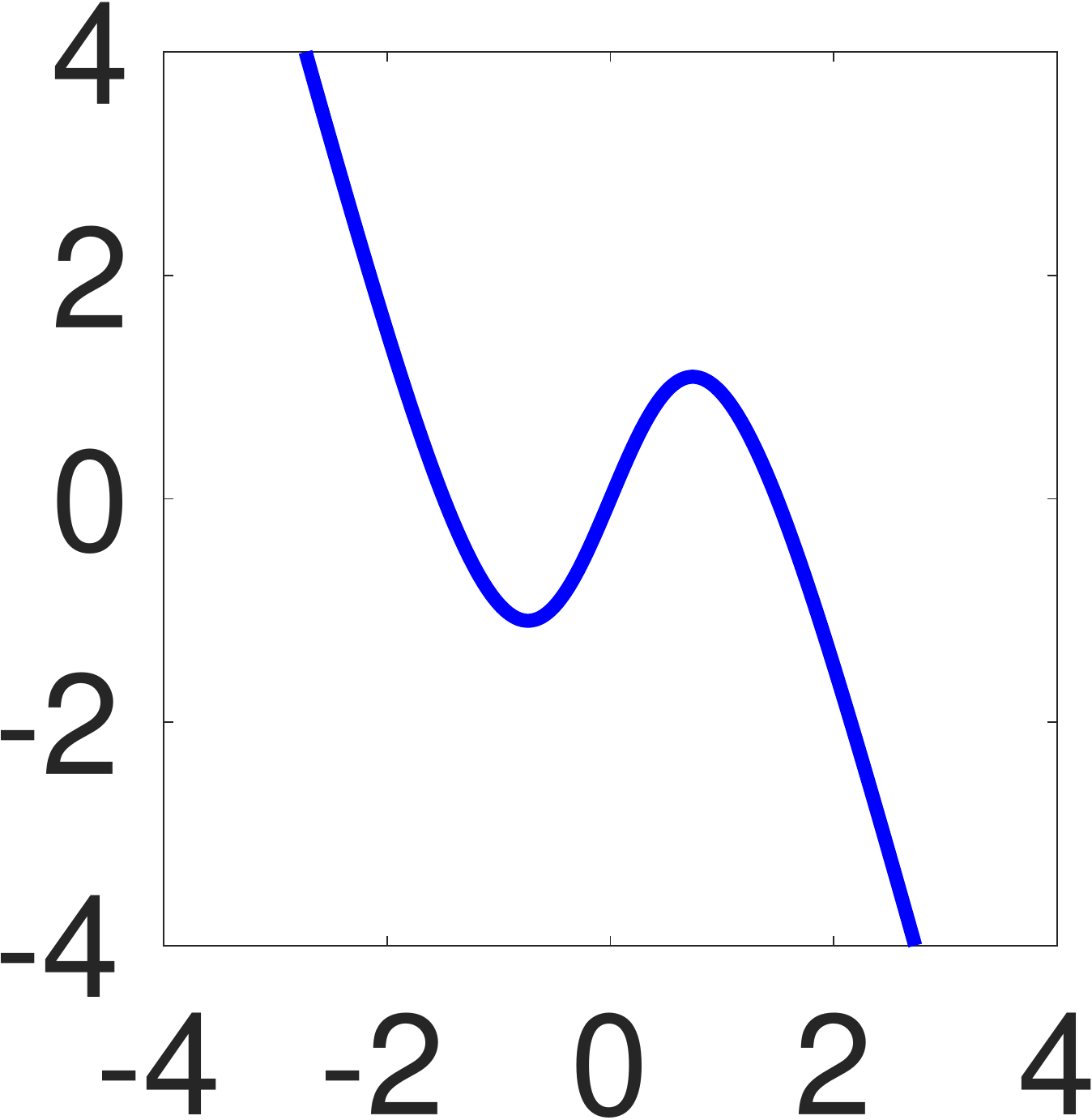}\\
LAR&0.5&0.971&0.271&0.703&0&0.930\\
CAR&0.318&0.000&0.705&0.120&0.637&0\\
NAR&0.182&0.029&0.023&0.177&0.363&0.070\\
NLC&1.211&1.035&1.039&1.101&1.659&1.085\\
$NLC(\tilde{f})$&1.659&1.855&1.420&1.433&1.659&1.886\\
\\
\end{tabular}
\begin{tabular}{lcccccc}
Act. fun.&sigmoid&even tanh&Gaussian&odd square&square&sawtooth\\ \hline\hline
Formula&$\frac{1}{1+e^{-s}}$&$|\tanh(s)|$&$\frac{1}{\sqrt{2\pi}}e^{-\frac{s^2}{2}}$&$s*|s|$&$s^2$&\textdaggerdbl\\
Illustration&\includegraphics[scale=0.13,valign=c]{graphsFinal/actFunillu/sigmoid.pdf}&\includegraphics[scale=0.13,valign=c]{graphsFinal/actFunillu/abstanh.pdf}&\includegraphics[scale=0.13,valign=c]{graphsFinal/actFunillu/gaussian.pdf}&\includegraphics[scale=0.13,valign=c]{graphsFinal/actFunillu/square.pdf}&\includegraphics[scale=0.13,valign=c]{graphsFinal/actFunillu/abssquare.pdf}&\includegraphics[scale=0.13,valign=c]{graphsFinal/actFunillu/saw.pdf}\\
\begin{tabular}[x]{@{}l@{}}Nonlinear\\basis\end{tabular}&\includegraphics[scale=0.13,valign=c]{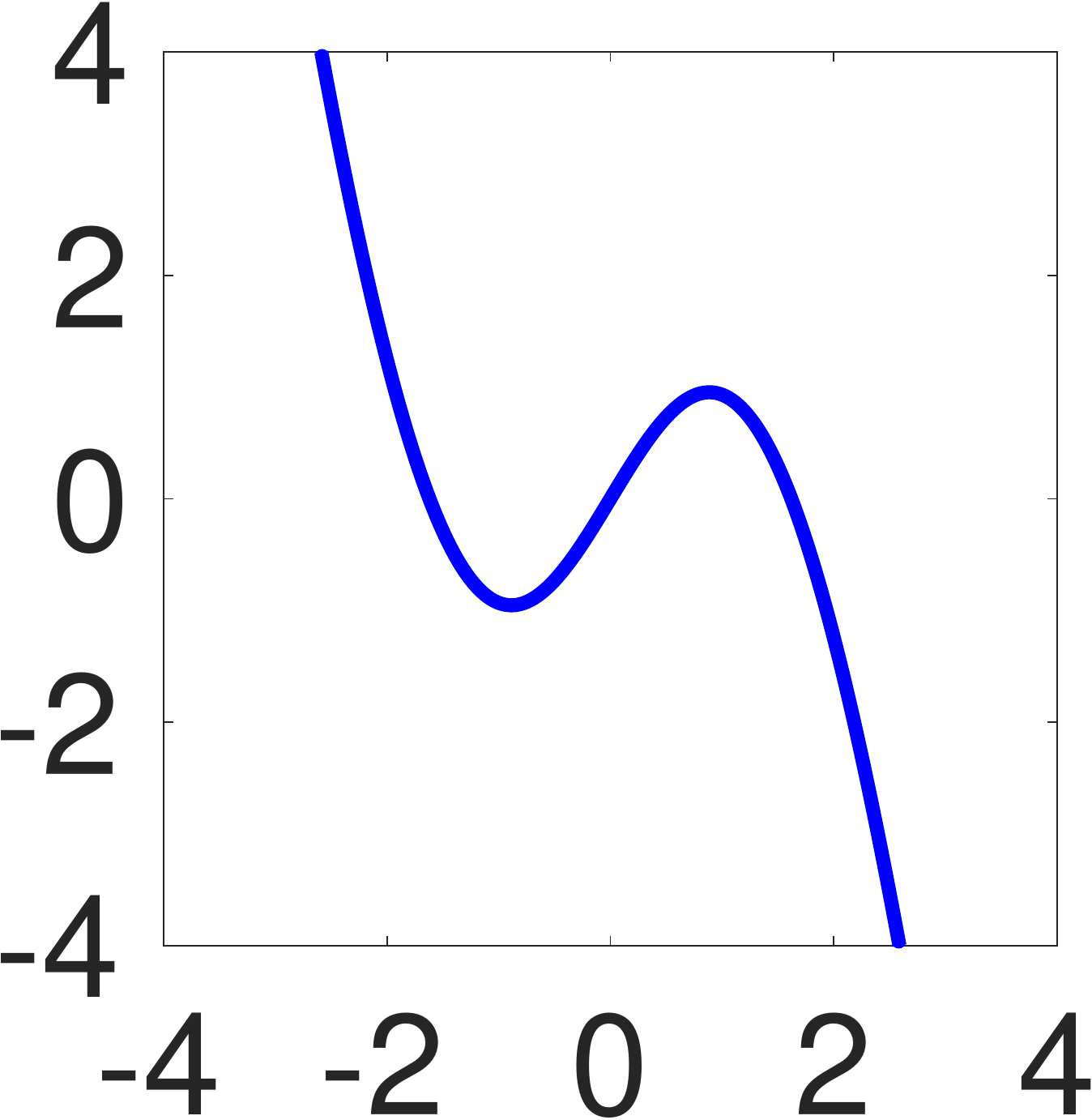}&\includegraphics[scale=0.13,valign=c]{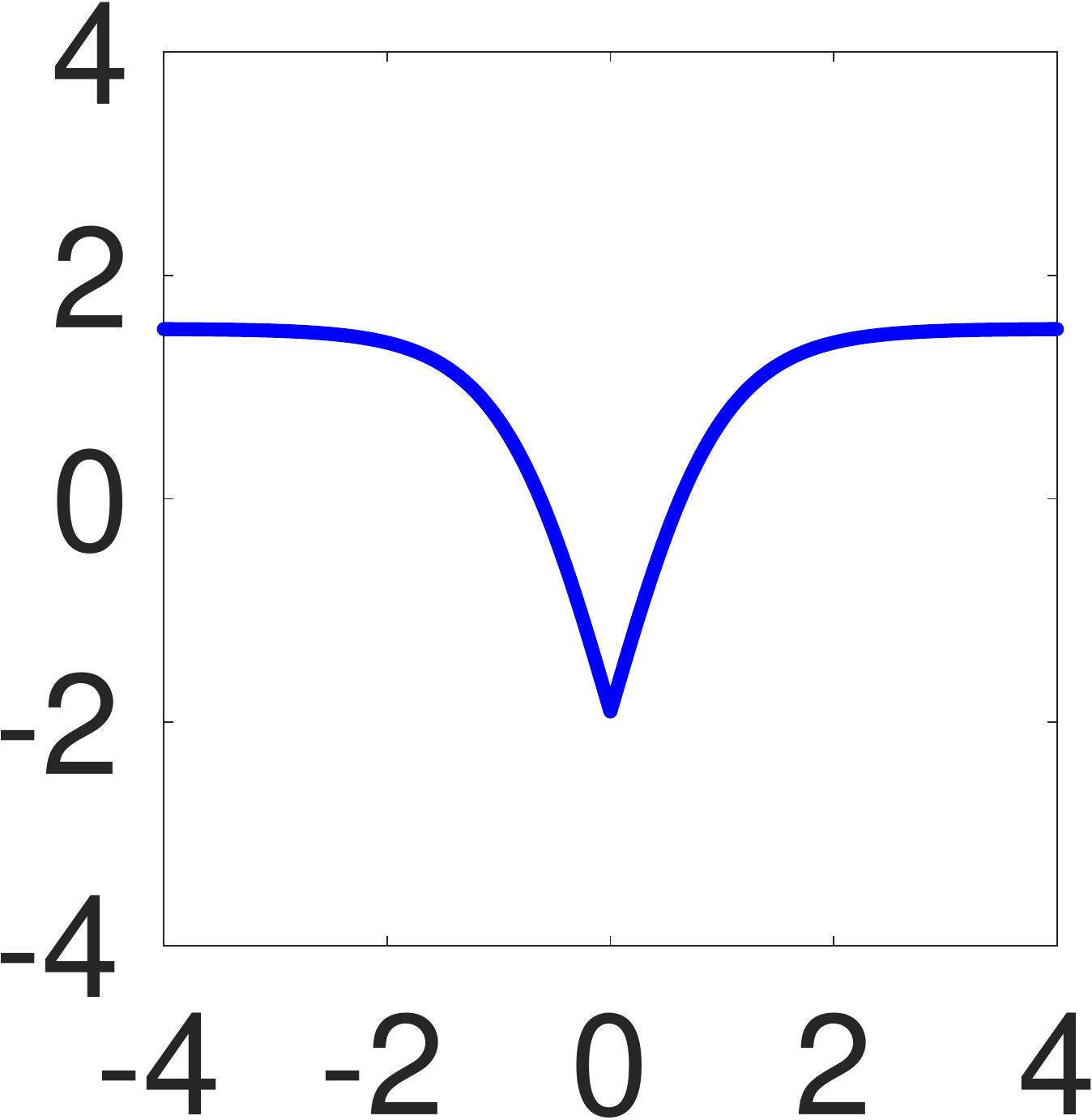}&\includegraphics[scale=0.13,valign=c]{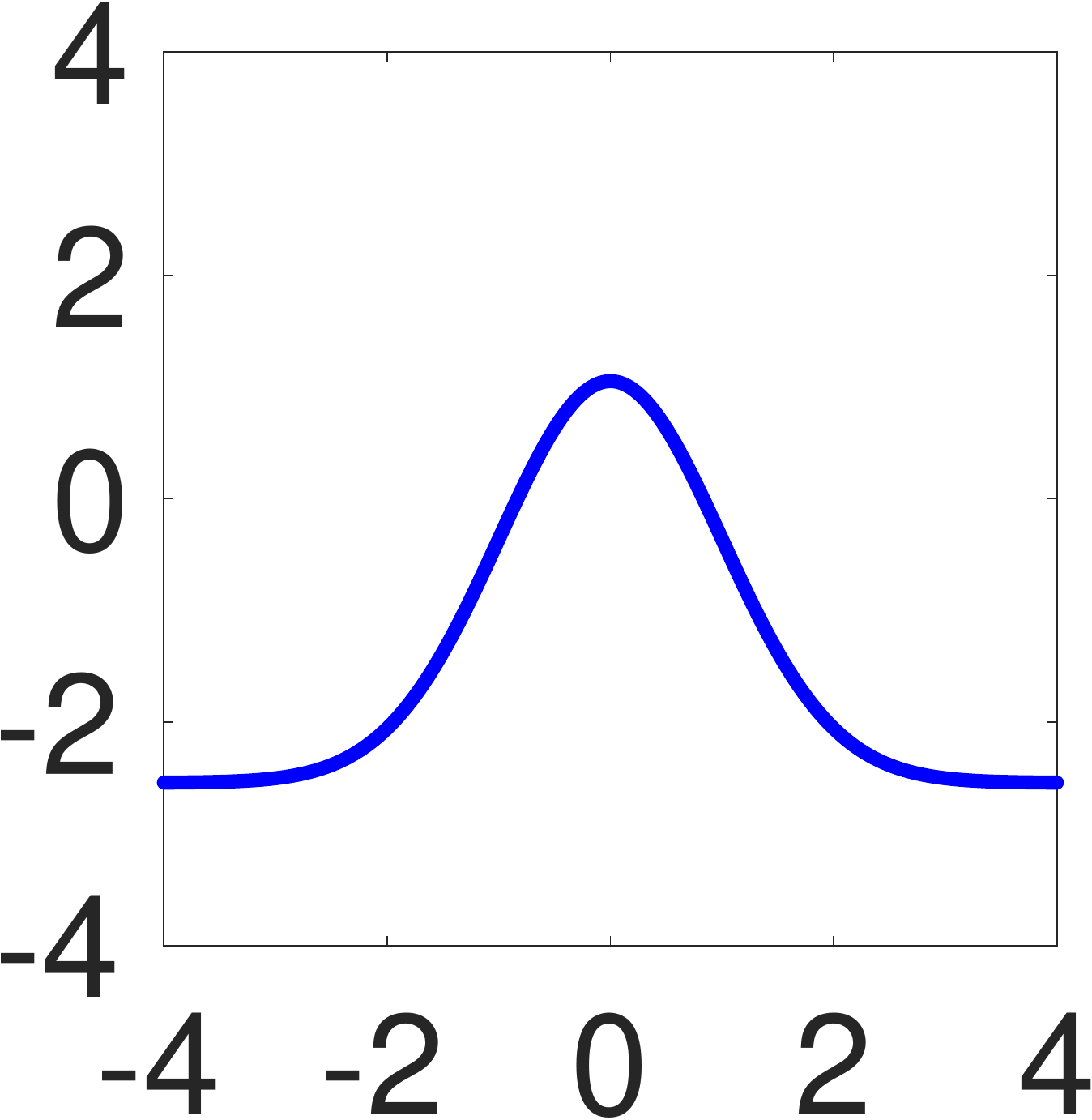}&\includegraphics[scale=0.13,valign=c]{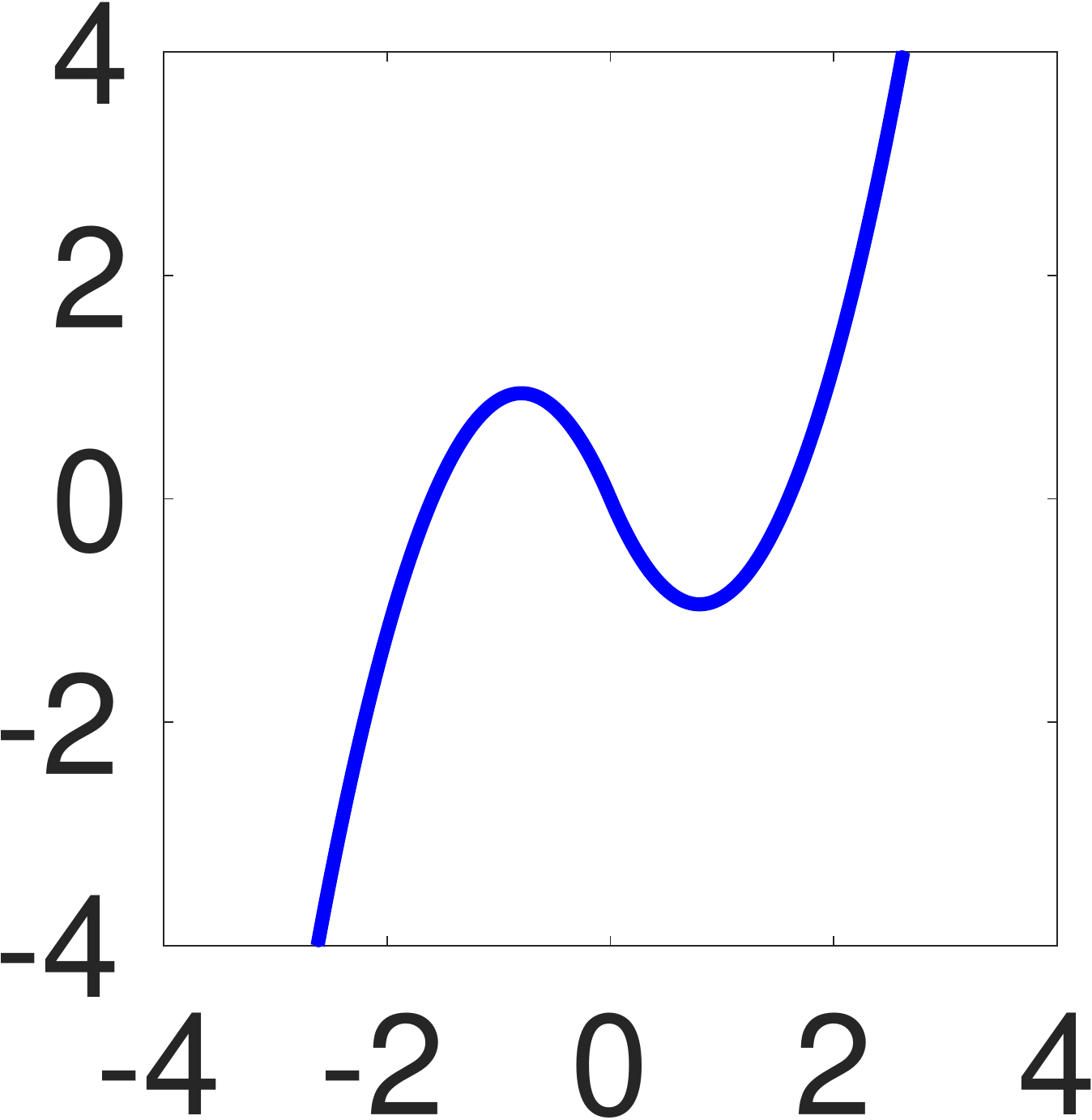}&\includegraphics[scale=0.13,valign=c]{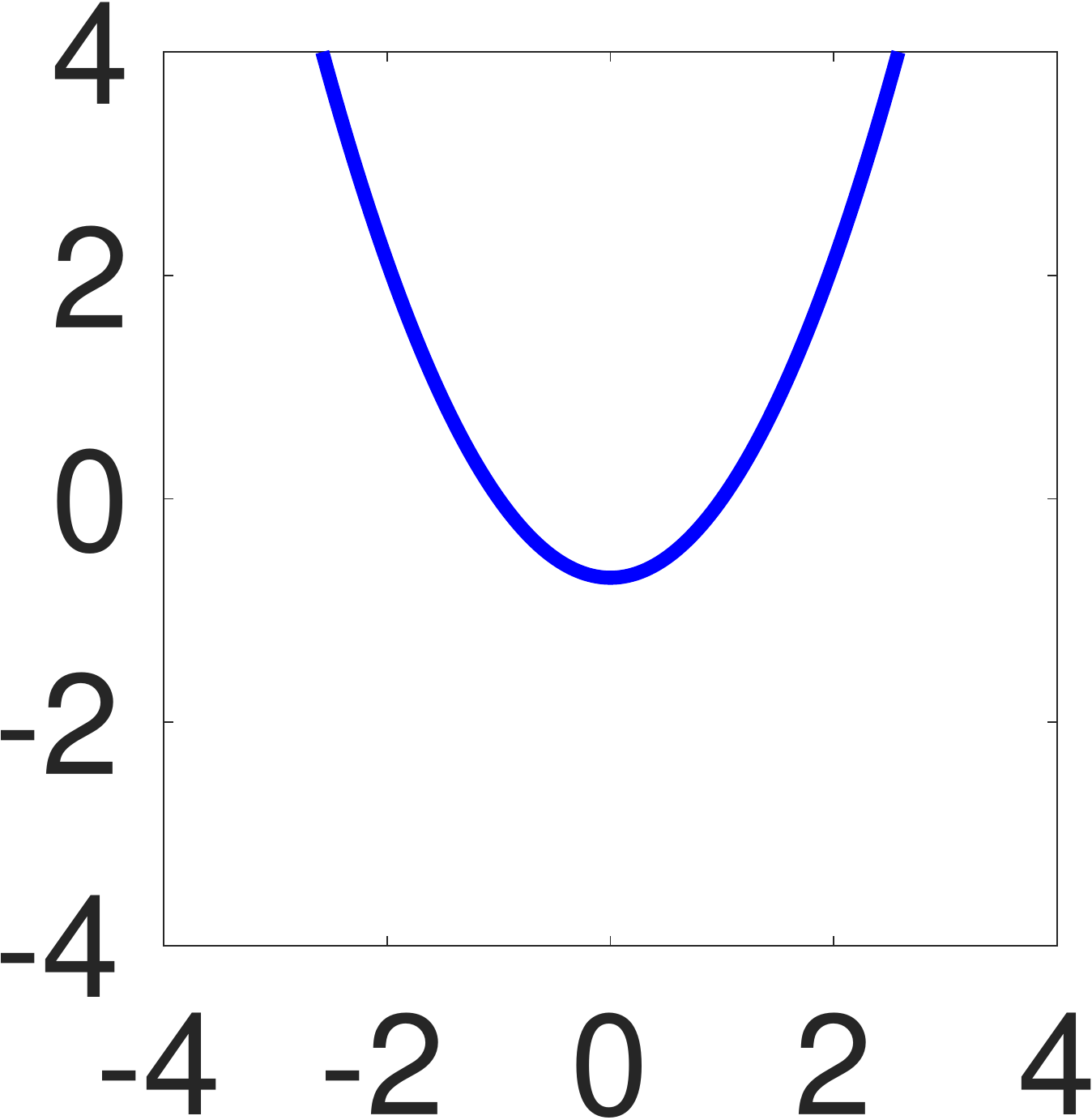}&\includegraphics[scale=0.13,valign=c]{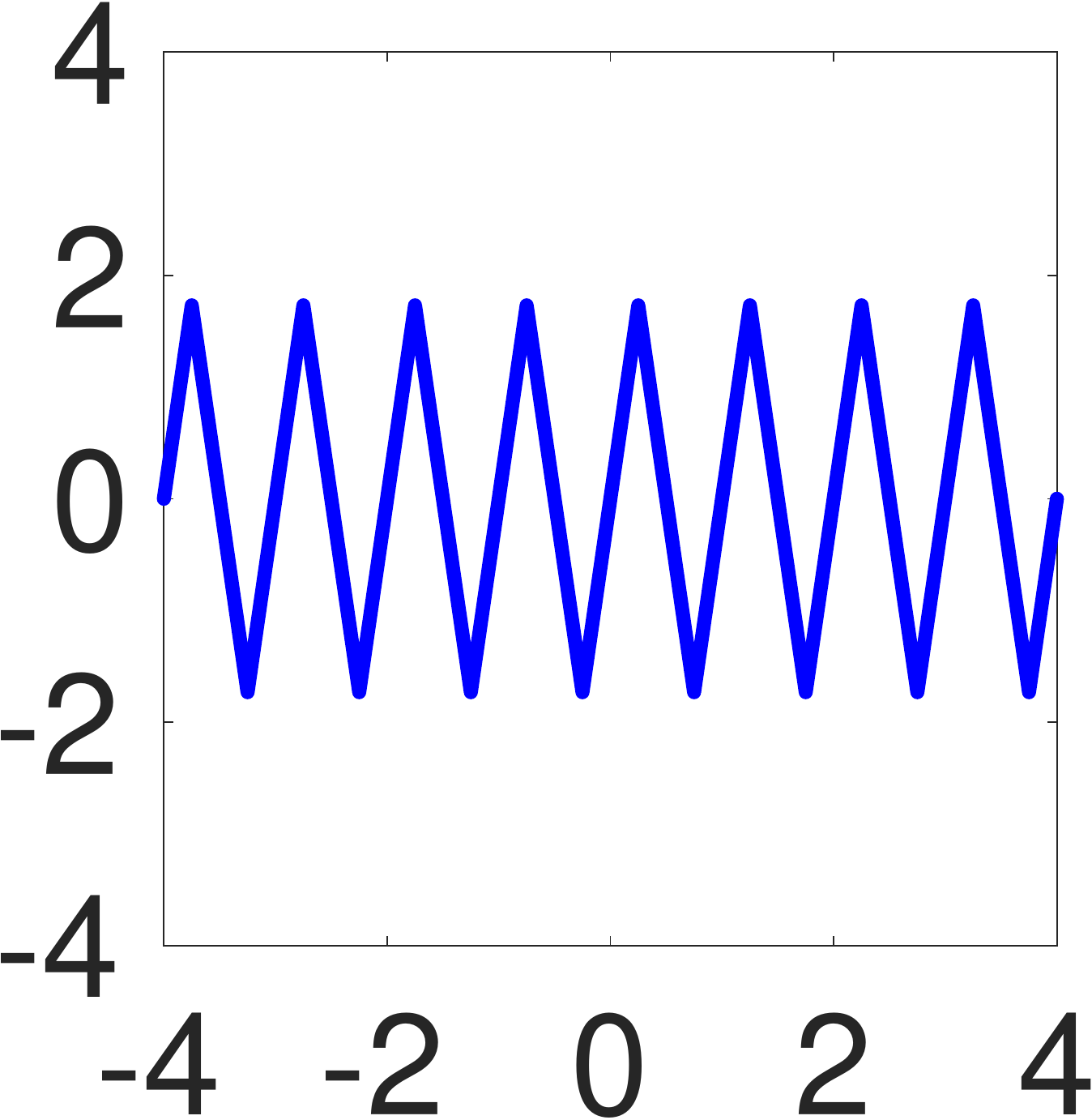}\\
LAR&0.146&0&0&0.849&0&0\\
CAR&0.852&0.784&0.866&0&0.333&0\\
NAR&0.002&0.216&0.134&0.151&0.667&1\\
NLC&1.017&2.335&1.577&1.155&1.414&6.928\\
$NLC(\tilde{f})$&1.767&2.335&1.577&1.790&1.414&6.928\\
\end{tabular}
\\
}
{\scriptsize \textdagger$\mathbbm{1}_{s > 0}1.0507s + \mathbbm{1}_{s < 0}1.75814(e^s-1)$\\
\textdaggerdbl $\tau(s) = s - \lfloor s\rfloor$ if $s - \lfloor s\rfloor < 0.25$; $\tau(s) = s - \lfloor s\rfloor - 1$ if $s - \lfloor s\rfloor > 0.75$; $\tau(s) = 0.5 - s + \lfloor s\rfloor$ else}
\caption{Activation functions with key metrics computed on unit Gaussian input. {\it Conclusion:} Theorems \ref{finiteNetTilde} and \ref{finiteNetLAR}, as well as proposition \ref{finiteNetBasisOrtho}, hold.}
\label{basisillu}
\end{table}

In table \ref{basisillu}, we give a range of metric values. \finding{It is easy to check that theorems \ref{finiteNetTilde} and \ref{finiteNetLAR}, as well as the last statement of proposition \ref{finiteNetBasisOrtho}, hold for all activation functions. There are a significant number of activation functions that do not have a constant and / or linear component. Also, some activation functions have very small nonlinear components, such as sigmoid, SeLU and softplus, and are very close to a linear function in an $L2$ sense. Ranking activation functions by their NLC is comparable to ranking them by their NAR, which shows that the presence of linear components significantly dictates the NLC. When the linear component is removed, the NLC ranking changes drastically. For example, the least possible $NLC(\tilde{f})$, $\sqrt{2}$, is achieved by the square activation function, which does not have a small NLC. ReLU and abs. val. have the same nonlinear basis. Softplus, Swish and square have similar nonlinear bases, which is exemplified by their similar $NLC(\tilde{f})$ values.}

\subsection{The NLC is related to noise sensitivity and overfitting} \label{nlcNoiseSection}

In the last subsection, we began the project of explaining the relationship between NLC and performance, and we did so in terms of underfitting. In this subsection, we continue the project by relating NLC and overfitting, and specifically by relating the NLC and the sensitivity of the network output to input perturbation. We show that when test inputs are {\it too far away} from training inputs of the same class relative to the NLC, networks fail to generalize.

We begin by giving a simple proposition that explicates the relationship between NLC, input noise and output change.

\begin{proposition} \label{finiteNetNoiseSensitivity}
Let $\delta_\text{in} \sim \mathcal{N}(0,\Cov_x)$ and $\delta_\text{out} \sim \mathcal{N}(0,\Cov_f)$ be row vectors. Assume $NLC > 0$. Then $$\mathbb{E}_{x,\delta_\text{in}}||\frac{\delta_\text{in}}{NLC}\mathcal{J}(x)^T||_2^2 = \mathbb{E}_{\delta_\text{out}}||\delta_\text{out}||_2^2$$
\end{proposition}

Informally, the NLC measures what fraction of the domain of $f$ we need to traverse in a random direction in order to traverse the codomain of $f$, assuming the input is propagated forward through the gradient-based local linear approximation (GLLA) given by $\mathcal{J}(x)$. The NLC measures the required magnitude of random noise {\it relative} to the variation of the input that is capable of significantly corrupting the network output. The assumption that the input is propagated through the GLLA is justified by section \ref{nlcSensiSection}, where we showed that the GLLA tends to be accurate throughout a large fraction of the codomain.

\begin{figure}
\centering
\includegraphics[scale=0.8]{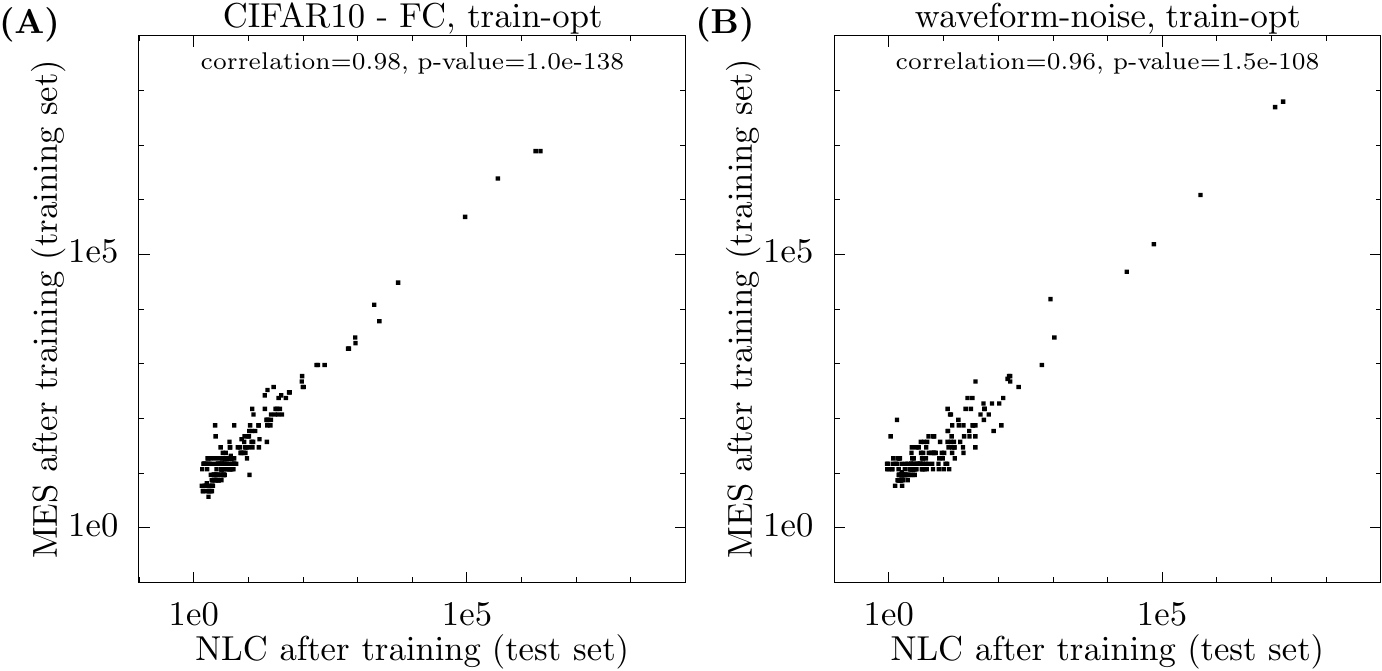}
\caption{NLC vs MES for study A architectures after training error minimization. We set $T=1.05$. Only architectures for which $MES < 10^9$ are depicted, due to limitations associated with floating-point computation. {\it Conclusion:} The NLC predicts the influence of input noise on error.}\label{nlcKernelSensi}
\end{figure}

The magnitude of noise relative to the variation of the input, or to the variation of the value of intermediate layers, shows up all over the place in deep learning. The field of adversarial examples studies imperceptible noise \citep{adversarialOrig,adversarial1,adversarial2,adversarial3}, which implies that the magnitude of the noise is small relative to the magnitude of the input. Specifically, it studies noise that can, in the context of classification, flip the class prediction. ``Flipping the class prediction'', ``traversing the codomain'' and ``corrupting the output'' can be regarded as roughly equivalent. Having an output that is robust to noise is important for a range of strategies such as batch normalization, dropout \citep{dropout} and quantization \citep{quantization2,quantization3,quantization4}. Each of these three strategies introduces noise at intermediate layers that has a roughly constant magnitude relative to the variation of the values at that layer, assuming that batch size / dropout rate / number of quantization buckets is constant respectively. We further investigate the relationship of noise sensitivity and architecture performance in the context of batch normalization and floating-point rounding error in section \ref{noiseStabilitySection}.

Based on this discussion, we can already see how an excessive NLC could be harmful to generalization. In the vast majority of practical datasets that are used for deep learning, a change to the input that is very small relative to $\sqrt{\Tr(\Cov_x)}$ almost never affects the label. This is a fundamental property of the true input-label function which we investigate further in section \ref{bestNlcSection}. For example, flipping a single pixel in an image almost never changes the type of the object depicted. Of course, the smaller the number of pixels in an image, the correspondingly smaller the number of pixels that need to be changed to change the type. Flipping a single character in a text of many characters almost never changes the sentiment of the text in the context of sentiment analysis. A function that experiences a large relative output change in response to a small relative input change cannot be a close match to the true input-label function.

We began our empirical investigation by verifying that input noise indeed corrupts network output, as suggested by proposition \ref{finiteNetNoiseSensitivity}. We did this by measuring the change in error while the inputs are being perturbed.

\begin{metricDefinition}
The `noise-corrupted error' (NCE) of a network $f$ with respect to error function $e$, data distribution $\mathcal{D}$ and scalar noise weight $w$ is

$$NCE(f, e, \mathcal{D}, w) = \mathbb{E}_{(x,y),u} e(f(wu + (1-w)x), y)$$

where $u \sim \mathcal{N}(0, \Cov_x)$ independently of $(x,y)$.
\end{metricDefinition} 

\begin{figure}[H]
\centering
\includegraphics[width=0.98\textwidth]{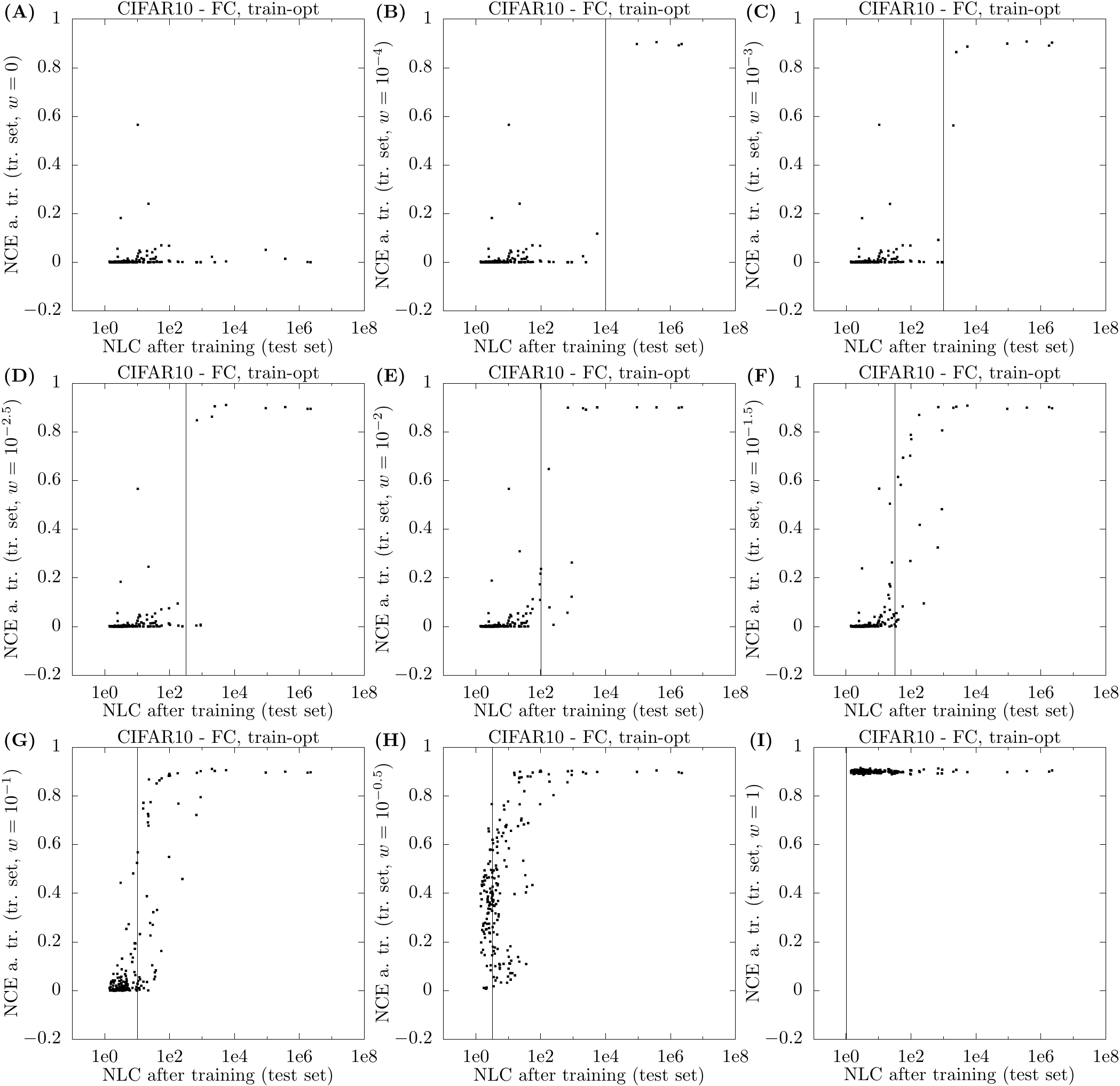}
\caption{NLC vs NCE for study A CIFAR10 architectures after training error minimization. The noise weight $w$ is specified on the y-axis. The black line indicates where the NLC equals $\frac{1}{w}$. Architectures with final NLC greater $10^8$ are not depicted to improve the visibility of low-NLC architectures. However, they follow the same trends as other high-NLC architectures. {\it Conclusion:} The NLC predicts the noise level at which the error increases.} \label{nlcKernelInterpGaussianCifar}
\end{figure}

\begin{figure}[H]
\centering
\includegraphics[width=0.98\textwidth]{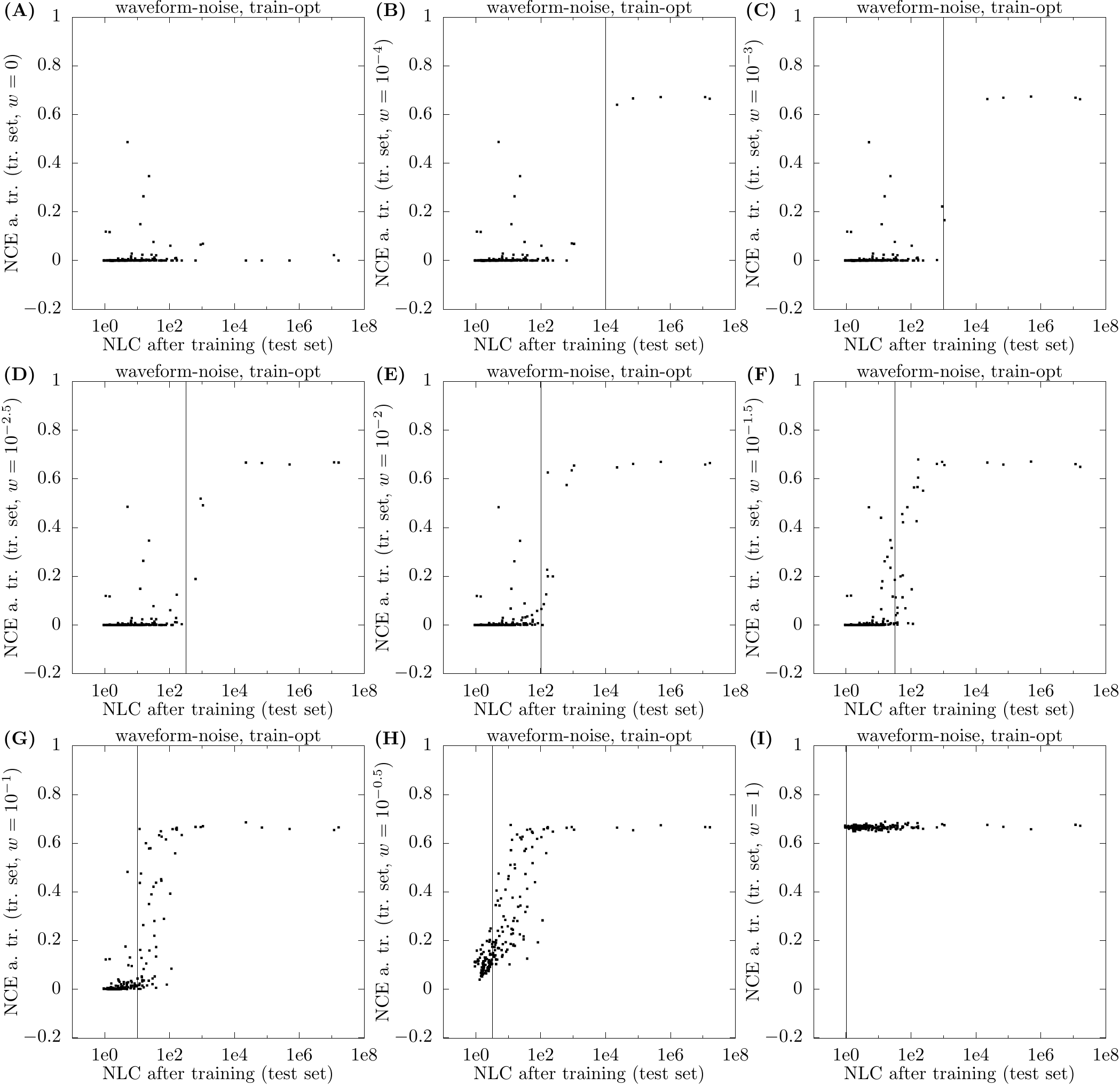}
\caption{NLC vs NCE for study A waveform-noise architectures after training error minimization. The figure and its conclusion are analogous to the previous figure.} \label{nlcKernelInterpGaussianWave}
\end{figure}

Of course, the value of NCE, like regular error, is only interesting after training. And, it is only interesting when the regular error is better-than-random. Because of this, we computed NCE on our CIFAR10 and waveform-noise architectures on the training set after they were (re-)trained to minimize training error, because this is the context where even high-NLC architectures can achieve low error, as shown in figures \ref{nlcPredTrainInit} and \ref{nlcPredTrainFinal}. In figures \ref{nlcKernelInterpGaussianCifar} and \ref{nlcKernelInterpGaussianWave}, we plot NCE vs NLC in the final state. \finding{We find that there are two distinct regimes. For some architectures, adding noise weighted by $w$ massively increases the error to the point of random performance, while for other architectures there is no change in error. For each value of $w$, we find that the transition point between both regimes is around the point $w = \frac{1}{NLC}$.}

We wanted to go further and investigate specifically at what noise level the error started to increase significantly for each architecture. To develop a metric that captures this property, we follow a similar strategy as with MGLLA in section \ref{nlcSensiSection}. Given some input $x$ and offset in input space $\delta_\text{in}$, we are interested in how large a scalar $c$ can be such that $f$ still makes the correct prediction on the entire line segment from $x$ to $x + c\delta_\text{in}$. 

\begin{metricDefinition}
The `median error sensitivity' (MES) of $f$ with respect to error function $e$, data distribution $\mathcal{D}$, tolerance $T$ and batch size $|B|$ is the median of the distribution over $C$, where $C$ is the smallest value greater or equal to 1 such that

$$\sum_{b=1}^{|B|} \max_{0 \le c < \frac{1}{C}} e(f(x^{(b)}+c\delta_\text{in}^{(b)}),y^{(b)}) \le \sum_{b=1}^{|B|} e(f(x^{(b)}),y^{(b)})(1+T)$$

The distribution over $C$ is induced by $(x^{(b)},y^{(b)}) \sim \mathcal{D}$ and $\delta_\text{in}^{(b)} \sim \mathcal{U}\Cov_x^{\frac{1}{2}}$, drawn independently for $1 \le b \le |B|$. $\mathcal{U}$ is the uniform distribution over vectors of dimensionality $d_\text{in}$ and length $\sqrt{d_\text{in}}$.
\end{metricDefinition}

In figure \ref{nlcKernelSensi}, we plot the NLC vs MES, again after training error minimization. \finding{We find a strong association.} Note that the same computational challenges apply to MES as apply to MGLLA, as discussed at the end of section \ref{nlcSensiSection}.

Figures \ref{nlcKernelInterpGaussianCifar}, \ref{nlcKernelInterpGaussianWave} and \ref{nlcKernelSensi} provide strong evidence for the practical predictiveness of proposition \ref{finiteNetNoiseSensitivity}. Under classification error, we can tolerate a change to the output that is a roughly a fixed fraction of the codomain diameter before the correct class prediction on an input flips. The NLC accurately predicts the magnitude of input noise necessary for this to occur. It is worth understanding just how sensitive high-NLC networks are. If $NLC=1000$, a relative input corruption of 0.1\% is sufficient to corrupt the output. If the inputs are e.g. images, a random change of 0.1\% to the intensity of each pixel is almost certainly imperceptible.

\begin{figure}[H]
\centering
\includegraphics[width=0.98\textwidth]{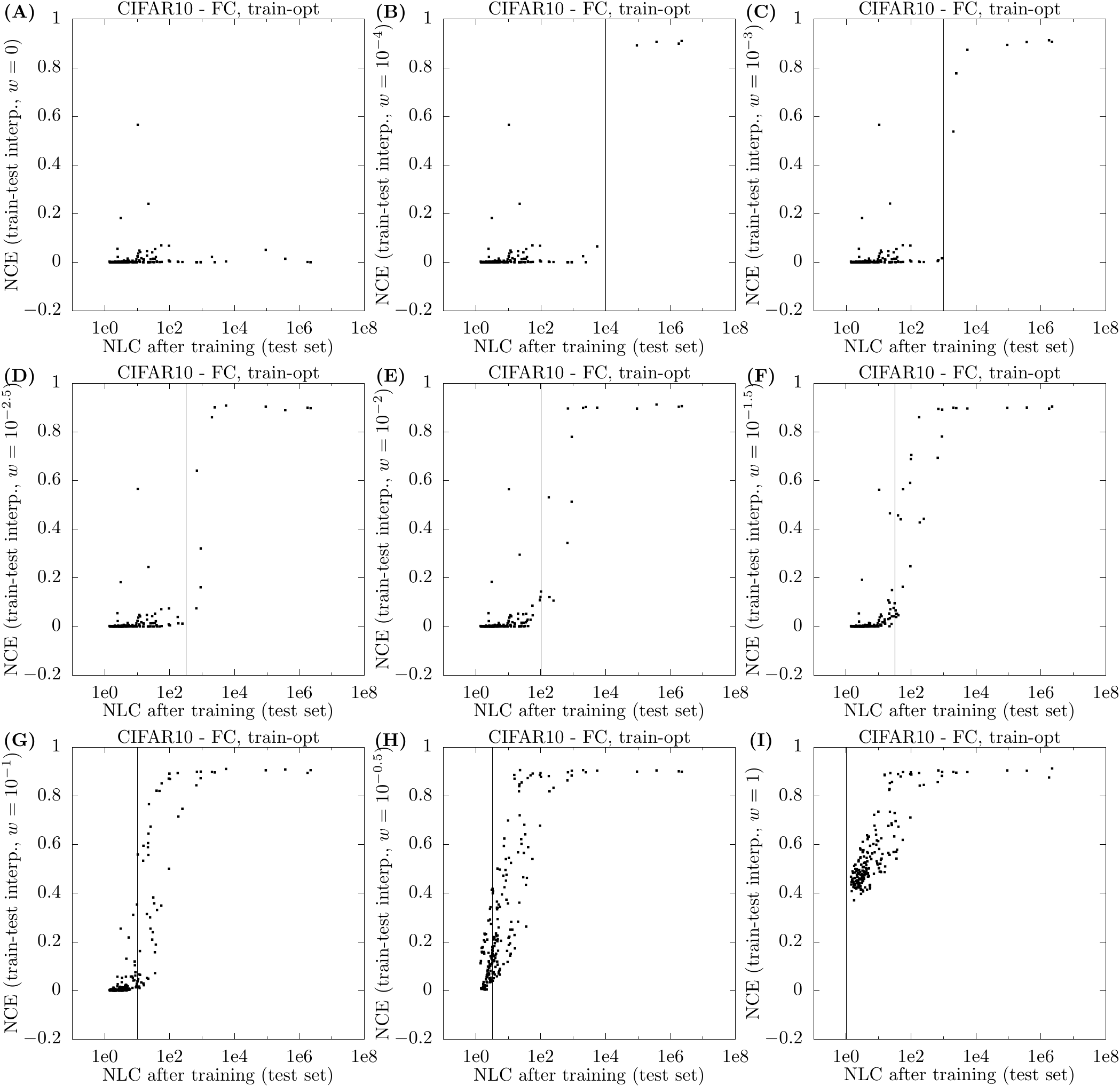}
\caption{NLC vs a modified NCE where training set inputs are interpolated with their closest test input, for study A CIFAR10 architectures after training error minimization. The noise weight $w$ is specified on the y-axis. The black line indicates where the NLC equals $\frac{1}{w}$. Architectures with final NLC greater $10^8$ are not depicted to improve the visibility of low-NLC architectures. However, they follow the same trends as other high-NLC architectures. {\it Conclusion:} Generalization is closely related to sensitivity to random noise, and that sensitivity is predicted by the NLC.} \label{nlcKernelInterpTestCifar}
\end{figure}

\begin{figure}[H]
\centering
\includegraphics[width=0.98\textwidth]{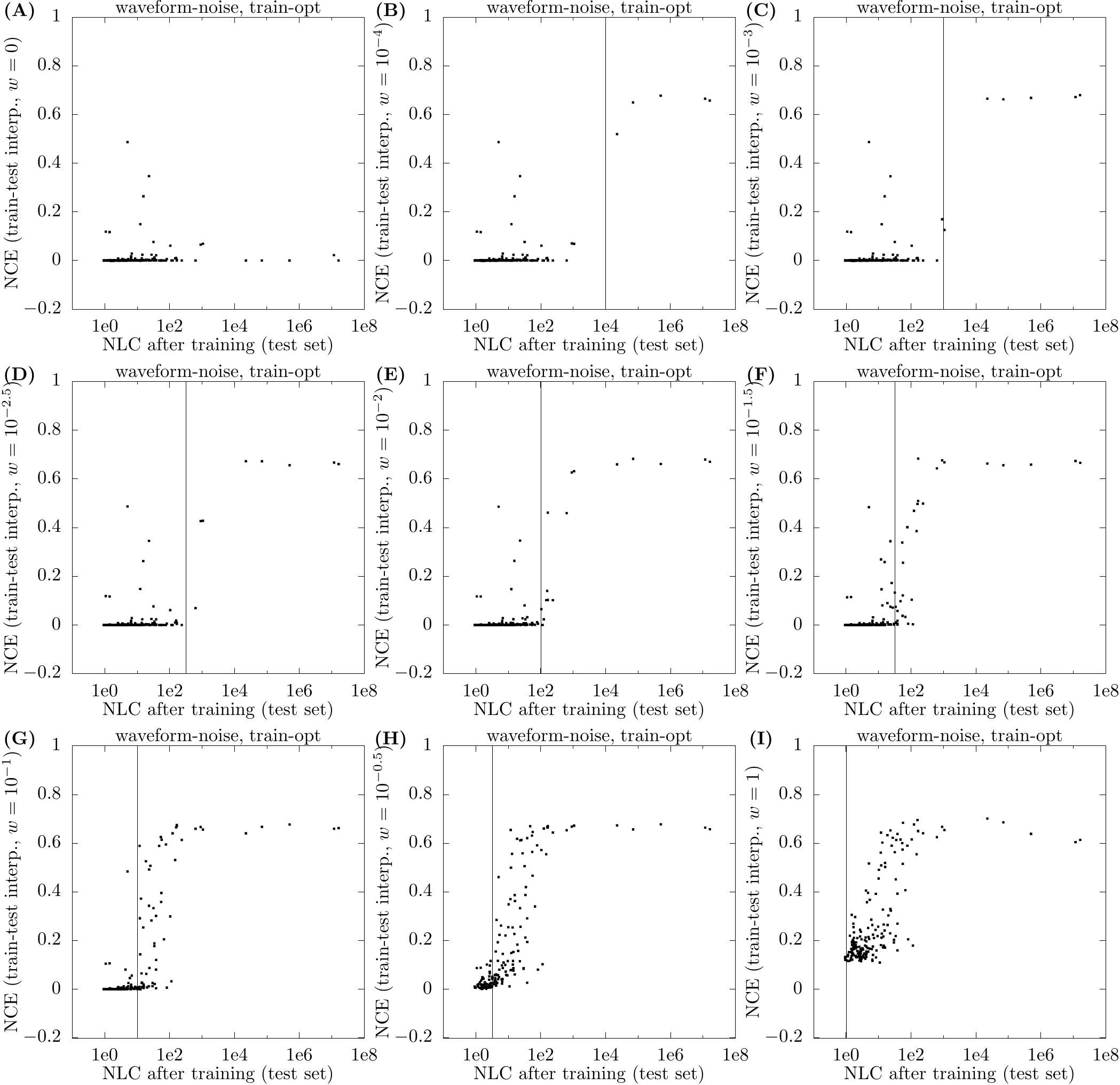}
\caption{NLC vs a modified NCE where training set inputs are interpolated with their closest test input, for study A waveform-noise architectures after training error minimization. The figure and its conclusion are analogous to the previous figure.} \label{nlcKernelInterpTestWave}
\end{figure}

We now have a clear hypothesis for how the NLC predicts overfitting. If we view the offset between an input in a training set and an input in the test set that is part of the same class as noise, then the NLC must be small enough so that adding that noise vector to the training input does not flip the class prediction. To verify this, we repeated our experiment with the NCE metric with a slight change. Instead of letting $u$ be a noise vector, we let it be the test input that is closest to the respective training input by Euclidean distance. Then, evaluating NCE with $w=0$ corresponds to a stochastic version of training error, whereas NCE with $w=1$ corresponds to a stochastic version of test error. \finding{In figures \ref{nlcKernelInterpTestCifar} and \ref{nlcKernelInterpTestWave}, we find virtually the same pattern as in figures \ref{nlcKernelInterpGaussianCifar} and \ref{nlcKernelInterpGaussianWave}. Switching from training inputs to interpolated inputs tends to cause no or a massive error increase, where the NLC controls the transition between both regimes. There is a slight departure from the previous figures in that the error increase appears slightly lower for a given value of $w$.} This is partially explained by the fact that the distance from training inputs to their closest test input is on average only 65\% of the distance to a Gaussian vector. Therefore, the same value of $w$ corresponds to a lower level of corruption. While it does appear that (as one would hope) networks generalize better to test inputs than to Gaussian noise, it is clear that the NLC after training has to be close to 1 for better-than-random generalization. This is what we found in figure \ref{nlcPredTestFinal}.

In summary, networks with a large NLC overfit because the output is susceptible to relatively small input changes. In section \ref{bestNlcSection}, we further verify that the ideal NLC for a dataset is determined by the distance between training and test inputs. In section \ref{machineLearningSection}, we explained that the common denominator between almost all machine learning models is that they extrapolate from seen datapoints to unseen datapoints by being continuous, i.e. assigning similar outputs to similar inputs. The NLC is a measure of the degree to which this principle holds for a neural network. The particular brand of continuity we study in this subsection, where the class assigned by the network does not change on the line segment between inputs of the same actual class, is also the basis for the popular data augmentation method mixup \citep{mixup}, which feeds linear combinations of training set inputs to the network during training.

\newpage

\newpage

\subsection{The NLC is related to kernel methods and model complexity} \label{nlcKernelSection}

In this subsection, we will further explain why the NLC is predictive of test error, via a relationship between the NLC, model complexity and kernel methods. Neural architectures can be viewed as kernel methods via mean field theory, specifically via the covariance kernel and neural tangent kernel. The NLC can be viewed as a measure of the {\it bandwidth} of those kernels, and bandwidth is a popular measure of model complexity in kernel methods.

Model complexity is one of the core concepts in machine learning and statistics. Like many popular concepts we encounter in this work, it is not well-defined. We must view it through the lens of popular measures such as VC dimension, Rademacher complexity and kernel bandwidth. We can immediately relate these measures to the insights gained in the previous subsection. In the context of classification, all of these measures depend on a model's ability to assign nearby inputs to different classes. We showed that the NLC predicts a neural network's ability to do this.

In the machine learning field of kernel methods, predictions made by models $f$ depend on the `kernel function' $K(x,x')$ that measures the similarity of input pairs $(x,x')$. The larger the value of $K(x,x')$, the more the model believes the inputs $x$ and $x'$, and hence the corresponding outputs $f(x)$ and $f(x')$, should be similar. Large values of $K$ across the dataset indicate that individual inputs are considered similar to many other inputs. In that case, individual predictions are averages of many labels in the dataset, leading to a function $f$ that is relatively smooth, unresponsive to input changes, and hence of ``low complexity'', which can lead to underfitting. Conversely, small values of $K$ across the dataset indicate that individual inputs are considered similar to only a small number of other inputs. In that case, individual predictions are averages of few labels in the dataset, leading to a function $f$ that is relatively erratic, highly responsive to input changes, and hence of ``high complexity'', which can lead to overfitting. The degree to which a kernel function assigns large values to input pairs is called its `bandwidth' $h$. This concept is defined more specifically in the common example of the Gaussian kernel function $K_\text{gauss}(x,x') = e^{-\frac{||x-x'||_2^2}{h}}$. Here, $h$ controls the Euclidean distance at which $K$ still assigns non-negligible values to input pairs.

The major theoretical framework that connects neural networks to kernel methods is mean field theory, which we cover in chapter \ref{meanFieldNnaChapter}. Mean field theory states (among other things) that, in the limit as the width of architecture layers converges to infinity, (i) the initial state is equivalent to a Gaussian process when the parameter is viewed as a random variable drawn from the initialization scheme. The kernel of that GP is called the `covariance kernel'. For fully-connected architectures with length-normalized inputs, it is a scalar function of the input co-mean, i.e. $\mathbb{E}_ix[i]x'[i]$. And (ii) the course of training is controlled by the neural tangent kernel (NTK). In the limit of fully-connected architectures with length-normalized inputs, the NTK is also a scalar function of $\mathbb{E}_ix[i]x'[i]$.

The NLC has a key meaning in the context of both of those results. With regards to (i), the NLC is the first-order approximation of the bandwidth of the covariance kernel. We will defer this result to chapter \ref{meanFieldNnaChapter} and specifically section \ref{networkCovKerSection}, when the necessary theoretical machinery has been introduced. For now, we will focus on the relationship between the NLC and the NTK, which we investigate empirically.

\begin{figure}
\centering
\includegraphics[width=0.98\textwidth]{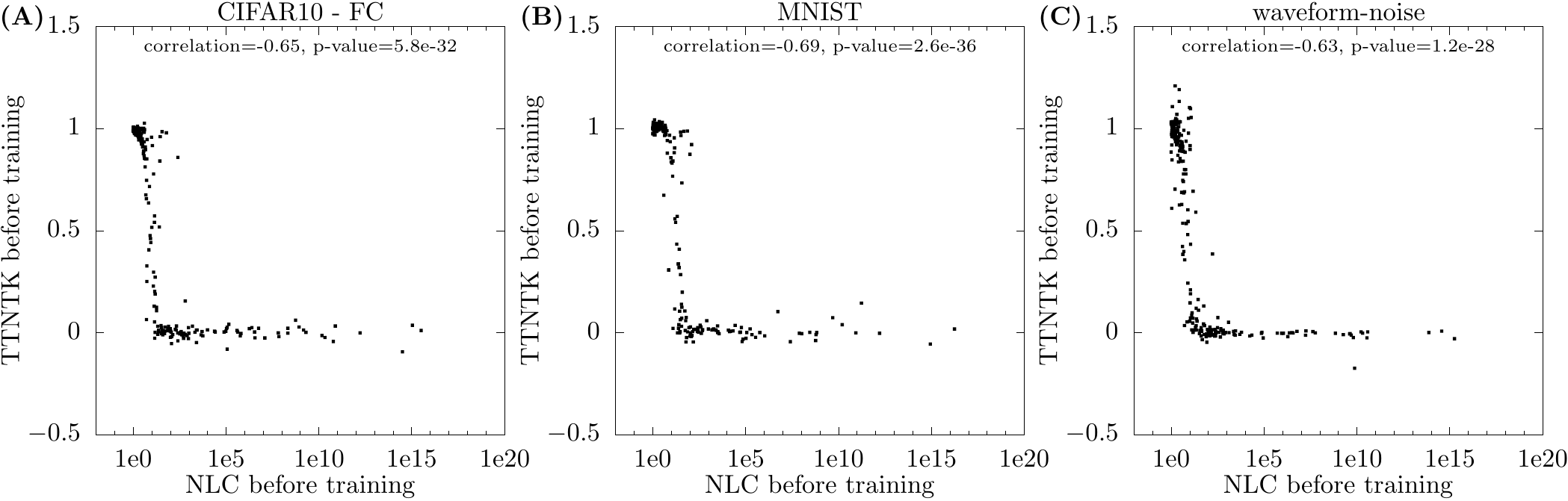}
\caption{NLC vs TTNTK for study A architectures in the initial state. {\it Conclusion:} The NLC is related to the neural tangent kernel, and thus to kernel bandwidth and the surrounding conceptual and theoretical machinery.} \label{nlcKernelNTK}
\end{figure}

The NTK is defined to be the outer product of the parameter Jacobians $\frac{df}{d\theta}$, i.e. we have $$K_\text{NTK}(x,x') = \frac{df(\theta, x)}{d\theta}\frac{df(\theta, x')}{d\theta}^T$$ This is a matrix of size $d_\text{out} \times d_\text{out}$. It can be interpreted as follows. If we make an update using SGD based on a batch consisting only of a single datapoint $(x,y)$, then the update will be $-\alpha g_L(\theta,x,y)\frac{df(\theta,x)}{d\theta}$, where $\alpha$ is the learning rate. Under the gradient-based local linear approximation, the change to the loss value at $(x,y)$ is then $-\alpha g_L(\theta,x,y)\frac{df(\theta,x)}{d\theta}\frac{df(\theta,x)}{d\theta}^Tg_L(\theta,x,y)^T=-\alpha g_L(\theta,x,y)K_\text{NTK}(x,x')g_L(\theta,x,y)^T$, and the change to the loss value at $(x',y')$ is $-\alpha g_L(\theta,x,y)K_\text{NTK}(x,x')g_L(\theta,x',y)^T$. Therefore, if $K_\text{NTK}(x,x')$ for $x \neq x'$ is large relative to $K_\text{NTK}(x,x)$, then we would expect a gradient update based on $(x,y)$ to have a strong impact on the loss value at $(x',y')$. This implies a high bandwidth and is ``low complexity behavior''. Conversely, if $K_\text{NTK}(x,x')$ for $x \neq x'$ is small relative to $K_\text{NTK}(x,x)$, then we would expect a gradient update based on $(x,y)$ to have a small impact on the loss value at $(x',y')$. This implies a low bandwidth and is ``high complexity behavior''.

For the purpose of empirical investigation, as usual, we would like to define a scalar metric that captures the property in question, i.e. the NTK, for a given network and dataset. We do this by considering the gradient update obtained when the entire training set is used as the batch. Specifically, we measure the impact of that update on the loss value of the entire test set, relative to the impact on the loss value of the entire training set.

\begin{metricDefinition}
The `train-test neural tangent kernel' (TTNTK) is 

$$TTNTK(f,\theta,\ell,D_\text{train},D_\text{test}) = \frac{(\mathbb{E}_{(x,y) \in D_\text{train}}\frac{d\ell(f(\theta,x),y)}{d\theta})(\mathbb{E}_{(x,y) \in D_\text{test}}\frac{d\ell(f(\theta,x),y)}{d\theta})^T}{(\mathbb{E}_{(x,y) \in D_\text{train}}\frac{d\ell(f(\theta,x),y)}{d\theta})(\mathbb{E}_{(x,y) \in D_\text{test}}\frac{d\ell(f(\theta,x),y)}{d\theta})^T}$$

\end{metricDefinition}

In figure \ref{nlcKernelNTK}, we plot the NLC vs TTNTK in the initial state. We find that \finding{TTNTK is around 1 when the NLC is around 1. As the NLC increases, TTNTK decreases roughly proportionally with the log of the NLC until we have $NLC \approx 30$, and then remains around 0}. The two metrics are strongly associated, though not linearly across the entire range of observed NLC values. Hence, the correlation values depicted are a significant underestimation of the strength of the association.

Assuming the batch is the entire training set and the learning rate is sufficiently small, a small TTNTK in the initial state implies that the first SGD update does not reduce the loss on the test set significantly. Conversely, a TTNTK close to 1 indicates that the test loss reduction is similar to the training loss reduction. Therefore, TTNTK is the first-order approximation of generalization in the initial state. The relationship between TTNTK and NLC is thus another explanation for the NLC predicting overfitting, and establishes a relationship between the NLC and bandwidth. We find TTNTK to be a highly interesting metric that may deserve significant study in its own right. TTNTK is similar to the OSGR metric from \citet{nnntkMetric}, which was developed independently.

Given the emergence of the NLC as a measure of model complexity, figures \ref{nlcPredTrainInit} and \ref{nlcPredTrainFinal} take on a new significance. There, we showed the successful training of architectures that have a very high NLC in both initial and final state. To our knowledge, we are first to note the trainability of ultra-high complexity neural architectures in general. For example, \citet{depthScalesMeanField} and \citet{meanFieldCNN} previously argued this was impossible. We discuss this point more in e.g. sections \ref{beyondNlcSummarySection}, \ref{chaosGoodSection} and \ref{expressivitySection}.

\subsection{The NLC is related to effective depth} \label{effectiveDepthSection}

\begin{figure}[h]
\centering
\includegraphics[scale=0.8]{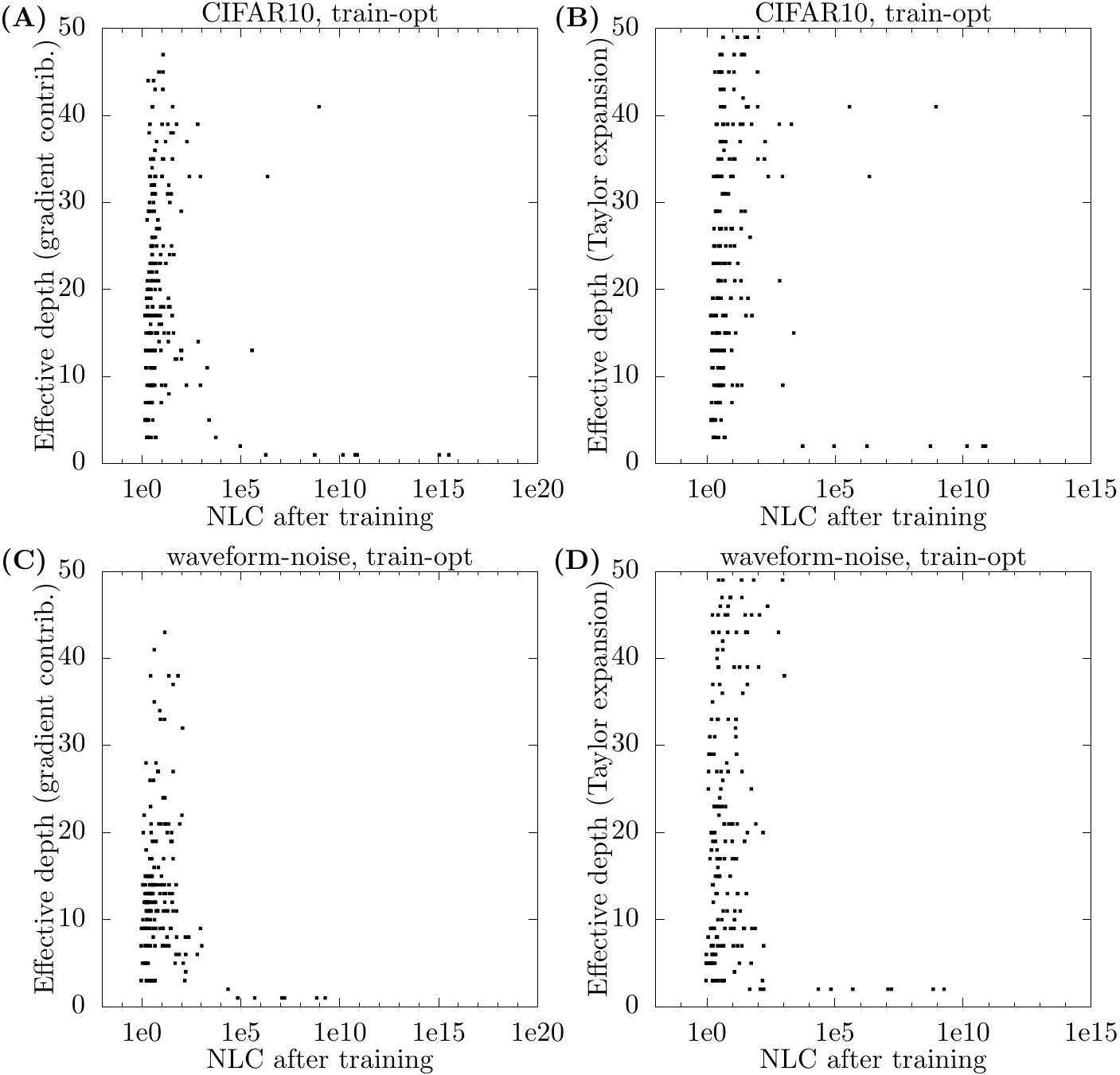}
\caption{NLC vs 2 measures of effective depth for study A architectures after training error minimization. Each graph corresponds to one of two measures discussed in \citet{expl}. Correlation values are not significant because of the non-linearity of the relationship of x- and y-axis values. In graphs B/D, we omit architectures with $NLC > 10^{15}$ as floating-point rounding error makes comparing the Taylor expansion with the original network impossible. {\it Conclusion:} A high NLC usually implies a low effective depth, and hence a non-attainment of the benefits associated with depth.} \label{nlcEffectiveDepth}
\end{figure}

The concept of effective depth was introduced in \citet{residualEnsembles} and expanded upon in one of our prior works \citep{expl}. The idea is that if a deep network function can be approximated by a much shallower network, then we do not attain the benefits of having a deep network to begin with, and we say the deep network has a low `effective depth'.

\citet{expl} presents two methods for measuring effective depth. One method uses the magnitude of gradients that flow through the parameter updates versus the initial parameter values, and the other method uses the Taylor expansion of the network around the initial parameter value. Here, we will not explain how those methods work. We will refer the reader to \citet{expl} for their definition and for an extended discussion of effective depth. We use the same metric definitions and implementations as \citet{expl} to compute effective depth based on gradients. We compute effective depth based on the Taylor expansion by considering the shallowest network obtainable by replacing layers of the network successively with their first-order Taylor expansion around the initial parameter value without corrupting the network output by more than 1\%. The Taylor expansion is defined and implemented as in \citet{expl}.

Effective depth is only interesting after training and when the error is better-than-random. Hence, in figure \ref{nlcEffectiveDepth}, we plot the NLC vs both measures of effective depth after training error minimization, similar to section \ref{nlcNoiseSection}. We find that \finding{whenever the NLC was greater than around $10^4$ after training, the effective depth of networks was always 1 or 2 for both measures for waveform-noise, and more likely than not to be 1 or 2 for CIFAR10. Those networks can be approximated by networks where any directed path in the macro-layer graph has at most 2 macro-layers with trainable parameters}. Since we hope to train networks of significantly greater depth than 2 in practice, we find that high-NLC architectures are generally unsuitable.

\citet{expl} shows how the effective depth of a network is related to the magnitude of the parameter updates during training relative to the magnitude of the initial parameter value, and hence to the difference between initial and final parameter value. In section \ref{learningRateSection}, we show that most of our high-NLC architectures indeed require small parameter updates to train, and do not exhibit large parameter changes. There, we also show that the few high-NLC architectures that did attain a high effective depth were those that we found trainable with a large learning rate.

\subsection{The NLC is decomposable into NLCs of individual layers} \label{nlcDecomposableSection}
 
We finish section \ref{nlcPropertiesSection} with four properties that fall under the rubric of robustness, like the properties of subsections \ref{nlcRobustDataSection} and \ref{nlcRandomInitSection}. In the following subsections, we show the NLC is robust to changing layer width, somewhat robust to training, and robust to adding a single additional layer to the network. In this subsection, we show that, in the initial state, the NLC is robust to breaking the network down into its individual layers and computing the NLC on a layer-by-layer basis. Assuming the network is composed of a single layer dependency chain, the NLC is the product of layerwise NLCs.

Assume we have a network $f$ that is composed of a single layer dependency chain $f_L(f_{L-1}(..(f_1(x))..))$. Then we have

\begin{eqnarray*}
&&NLC(f,\mathcal{D})\\
&=&\Big(\frac{\mathbb{E}\Tr(\mathcal{J}_{L,0}\Cov_{f_0}\mathcal{J}_{L,0}^T)}{\Tr(\Cov_{f_L})}\Big)^{\frac{1}{2}}\\
&=&\Big(\frac{\mathbb{E}\Tr(\mathcal{J}_{L,0}\Cov_{f_0}\mathcal{J}_{L,0}^T)}{\mathbb{E}\Tr(\mathcal{J}_{L,L}\Cov_{f_L}\mathcal{J}_{L,L}^T)}\Big)^{\frac{1}{2}}\\
&=&\Big(\prod_{l=0}^{L-1}\frac{\mathbb{E}\Tr(\mathcal{J}_{L,l}\Cov_{f_l}\mathcal{J}_{L,l}^T)}{\mathbb{E}\Tr(\mathcal{J}_{L,{l+1}}\Cov_{f_{l+1}}\mathcal{J}_{L,{l+1}}^T)}\Big)^{\frac{1}{2}}\\
&=&\Big(\prod_{l=0}^{L-1}\frac{\mathbb{E}\Tr(\mathcal{J}_{L,l+1}\mathcal{J}_{l+1,l}\Cov_{f_l}\mathcal{J}_{l+1,l}^T\mathcal{J}_{L,l+1}^T)}{\mathbb{E}\Tr(\mathcal{J}_{L,{l+1}}\Cov_{f_{l+1}}\mathcal{J}_{L,{l+1}}^T)}\Big)^{\frac{1}{2}}\\
\end{eqnarray*}

The expression on the last line almost looks like $NLC(f_{l+1},f_l(\mathcal{D}))$, except that the matrix inside the trace in both the numerator and denominator is multiplied on each side by $\mathcal{J}_{L,{l+1}}$. If we could simply cancel out those matrices, we would indeed obtain that the NLC of the network is the product of the NLC of each layer. We use the above derivation as a motivation to hypothesize that the NLC of the network might be of similar value to the product of the NLCs of individual layers.

What about residual networks? There, we encounter addition layers of the form $f_l = f_s(f_a) + f_r(f_a)$, where $f_s$ is the skip block and $f_r$ is the residual block as defined at the end of section \ref{architectureDesignParadigmsSection}. We can use the above derivation to motivate decomposing the NLC into the product of NLCs of residual units and of NLCs of individual layers not contained in a residual unit. Beyond this, we further want to decompose the NLC of each residual unit. We have

\begin{eqnarray*}
&&\Big(\frac{\mathbb{E}\Tr(\mathcal{J}_{l,a}\Cov_{f_a}\mathcal{J}_{l,a}^T)}{\Tr(\Cov_{f_l})}\Big)^{\frac{1}{2}}\\
&=&\Big(\frac{\mathbb{E}\Tr((\mathcal{J}_{s,a}+\mathcal{J}_{r,a})\Cov_{f_a}(\mathcal{J}_{s,a}+\mathcal{J}_{r,a})^T)}{\Tr(\mathbb{E}(f_r + f_s - \bar{f_r} - \bar{f_s})^T(f_r + f_s - \bar{f_r} - \bar{f_s}))}\Big)^{\frac{1}{2}}\\
&=&\Big(\frac{\mathbb{E}\Tr(\mathcal{J}_{s,a}\Cov_{f_a}\mathcal{J}_{s,a}^T+\mathcal{J}_{r,a}\Cov_{f_a}\mathcal{J}_{s,a}^T+ \mathcal{J}_{s,a}\Cov_{f_a}\mathcal{J}_{r,a}^T+\mathcal{J}_{r,a}\Cov_{f_a}\mathcal{J}_{r,a}^T)}{\text{\scalebox{0.88}{$\mathbb{E}\Tr((f_r - \bar{f}_r)^T(f_r - \bar{f}_r)+(f_r - \bar{f}_r)^T(f_s - \bar{f}_s)+(f_s - \bar{f}_s)^T(f_r - \bar{f}_r)+(f_s - \bar{f}_s)^T(f_s - \bar{f}_s))$}}}\Big)^{\frac{1}{2}}
\end{eqnarray*}

Both numerator and denominator contain the sum of four terms, two of which depend on both $f_r$ and $f_s$. We will now make the assumption that the expectation of these mixed terms is approximately 0. This assumption is based on the analysis from one of our prior works \citep{expl} and further justified by mean field theory in section \ref{meanFieldPracticalSection}. The above formula then becomes

$$\Big(\frac{\Tr(\Cov_{f_s})}{\Tr(\Cov_{f_s}) + \Tr(\Cov_{f_r})}NLC(f_s,f_a(\mathcal{D}))^2 +\frac{\Tr(\Cov_{f_r})}{\Tr(\Cov_{f_s}) + \Tr(\Cov_{f_r})}NLC(f_r,f_a(\mathcal{D}))^2 \Big)^{\frac{1}{2}}$$

So the squared NLC of the residual unit becomes approximately equal to the weighted average of the squared NLC of both blocks, where the weight of each block is proportional to the squared radius of the codomain. Note that in some residual networks, the addition layer multiplies the skip block and / or the residual block with an addition weight $w_s$ / $w_r$ before adding them together. In that case, the squares of those weights are simply multiplied to the $\Tr(\Cov_s)$ and $\Tr(\Cov_r)$ terms respectively in the above formula.

Based on all the above derivations, we define a metric to capture the decomposed NLC.

\begin{figure}
\centering
\includegraphics[width=0.98\textwidth]{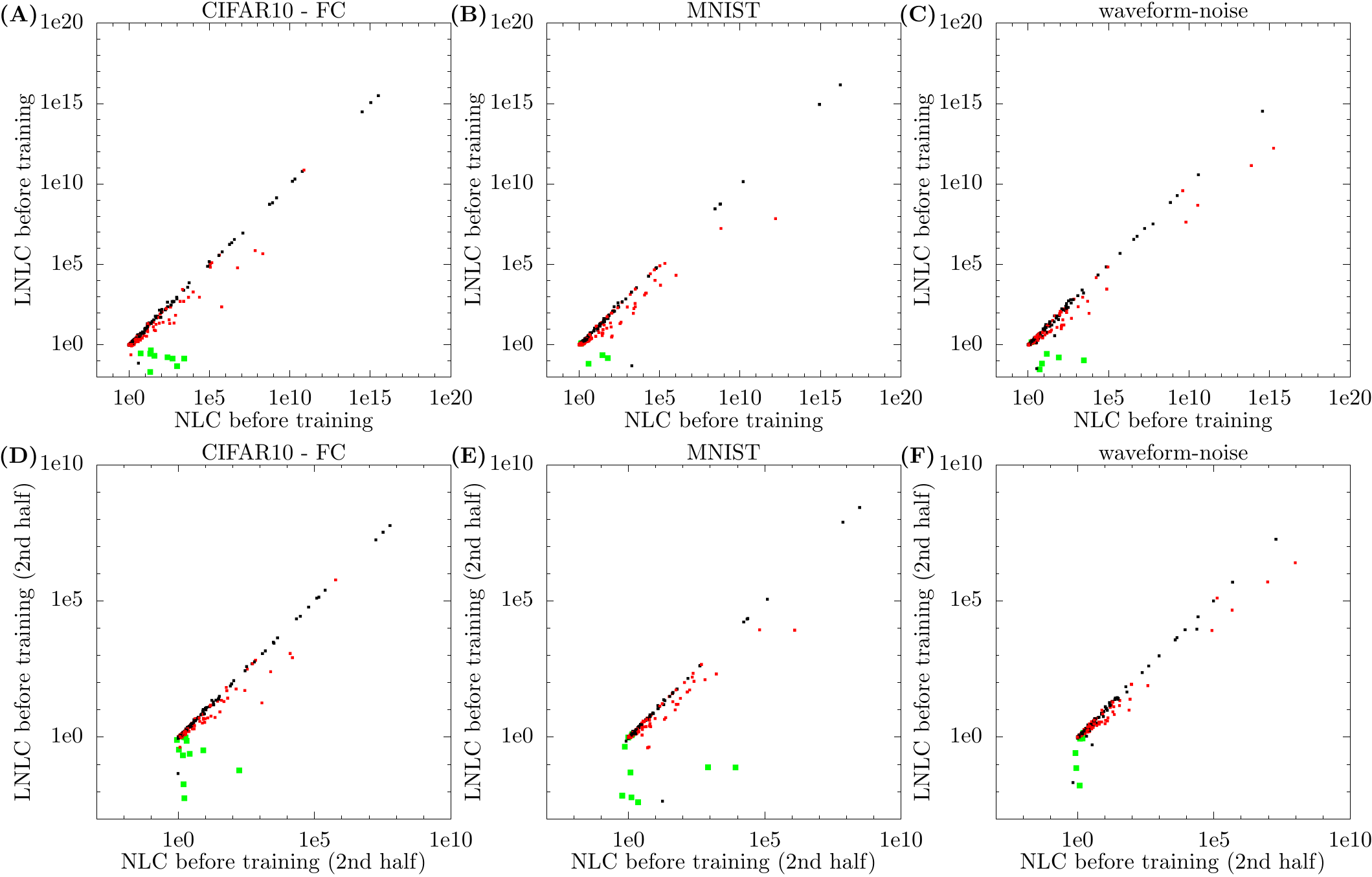}
\caption{NLC vs LNLC for study A architectures in the initial state. Green markers correspond to GUAs. Red markers correspond to residual architectures that are not GUAs. Note that in this figure, GUAs are actually displayed in the background so that black and red markers that represent outliers are visible. We omit correlation values due to the presence of extreme outliers. The top row gives results from decomposing the NLC of the whole network. The bottom row gives results from decomposing the second half of the network. There were also GUAs that exhibited LNLC values of as low as $10^{-30}$. They are not depicted in the graphs in order to improve visibility for non-GUAs. {\it Conclusion:} The NLC of fully-connected networks can usually be decomposed into the NLC of individual layers in the initial state, except for GUAs.} \label{nlcDecomposableInit}
\end{figure}

\begin{metricDefinition}
The `layerwise NLC' (LNLC) of a network $f$ with respect to the input distribution $\mathcal{D}$ is as follows. If $f$ is a single dependency chain of layers, then 

$$LNLC(f, \mathcal{D}) = \prod_{l=0}^{L-1} NLC(f_{l+1}(f_l), f_l(\mathcal{D}))$$

If $f$ contains residual units, we take the product over NLCs of residual units and layers not contained in a residual unit. We then replace the NLC of a residual unit $w_sf_s(f_a) + w_rf_r(f_a)$ by

{\fontsize{8.5}{1} $$\sqrt{\frac{w_s^2\Tr(\Cov_{f_s})}{w_s^2\Tr(\Cov_{f_s}) + w_r^2\Tr(\Cov_{f_r})}NLC(f_s,f_a(\mathcal{D}))^2 +\frac{w_r^2\Tr(\Cov_{f_r})}{w_s^2\Tr(\Cov_{f_s}) + w_r^2\Tr(\Cov_{f_r})}NLC(f_r,f_a(\mathcal{D}))^2}$$}

Finally, we replace the $NLC(f_s,f_a(\mathcal{D}))$ and the $NLC(f_r,f_a(\mathcal{D}))$ by the product of the NLCs of the layers contained in them as above. We do not apply LNLC to networks that are not of the types covered above.
\end{metricDefinition}

In figure \ref{nlcDecomposableInit}A-C, we plot NLC vs LNLC in the initial state for our study A architectures. We find that \finding{for non-residual architectures that are not GUAs, depicted in black, the match is very close, except for a small number of outliers. Those few outliers can attain an LNLC significantly less than 1. For residual architectures that are not GUAs, depicted in red, LNLC somewhat underestimates the NLC}. Of course, residual networks have a more complicated LNLC involving weighted sums. Finally, \finding{GUAs, depicted in green as always, have a complete mismatch between both quantities}. In fact, figure \ref{nlcDecomposableInit} does not tell the full story. Some GUAs have an LNLC of less than $10^{-30}$, which is not visible in the figure. In figure \ref{nlcDecomposableInit}D-F, we plot the NLC vs LNLC on just the second half of the network, i.e. we consider a layer halfway through the network as the surrogate input layer. We previously did this in section \ref{nlcSimpleMetricsSection} for the purpose of validating on a wider range of (surrogate) input distributions. \finding{We obtain the same findings}.

\begin{figure}
\centering
\includegraphics[width=0.98\textwidth]{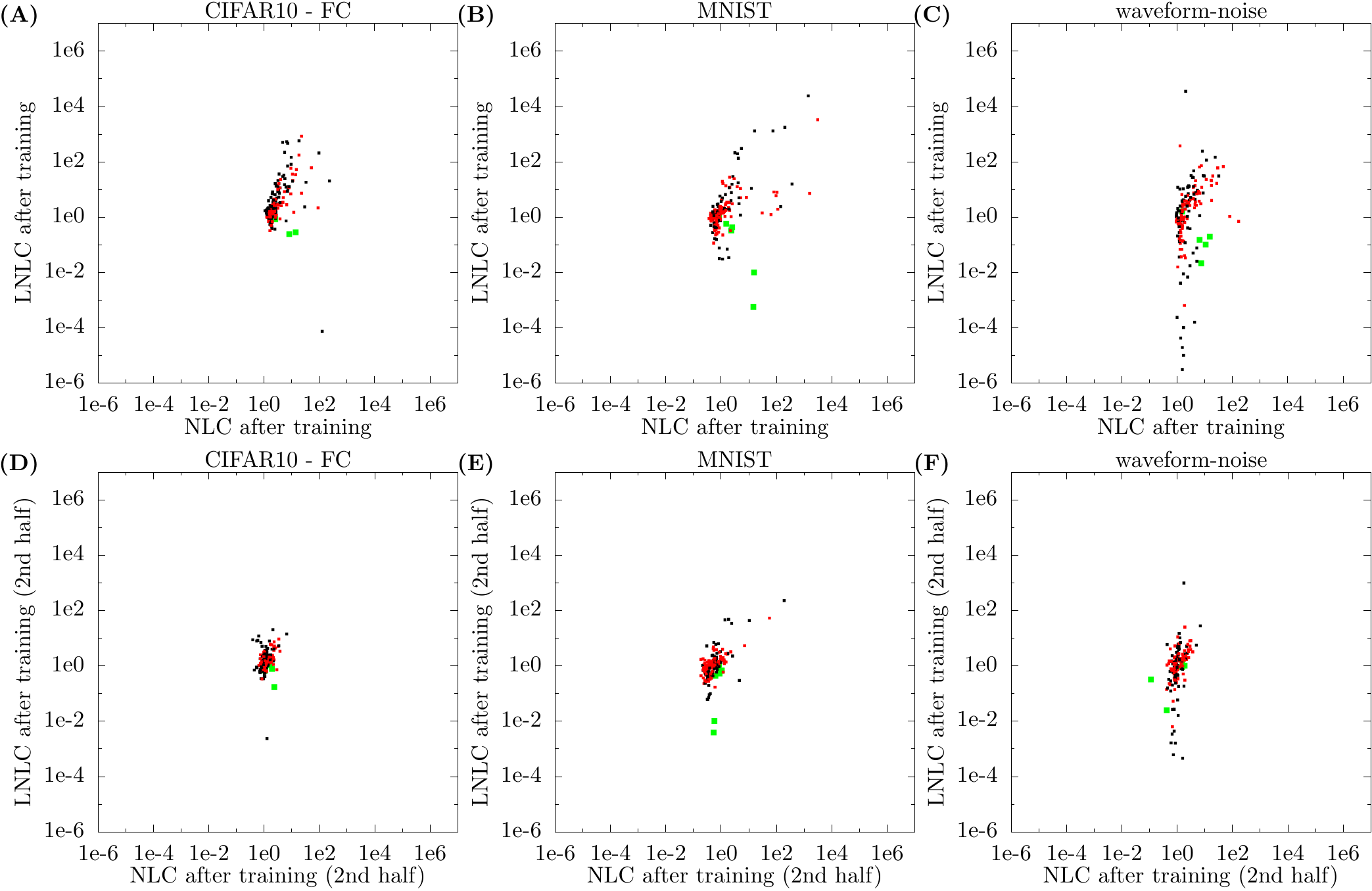}
\caption{NLC vs LNLC for study A architectures in the final state. Graphs are analogous to the previous figure. {\it Conclusion:} There is only a weak association between the NLC and its decomposition in the final state.} \label{nlcDecomposableFinal}
\end{figure}

Our findings bolster our findings from section \ref{nlcRobustDataSection}. The NLC of an individual layer uses an input distribution that can be very different from the network's input distribution. The fact that this variation does not impact decomposability is noteworthy.

In figure \ref{nlcDecomposableFinal}, we plot NLC vs LNLC in the final state. We find that \finding{there is still an association between both metrics, but it is significantly weaker than in the initial state. For waveform-noise, a large number of architectures have an LNLC much less than 1. The degradation of the association from initial to final state appears greater than in previous subsections.}

Finally, we note that when computing LNLC, we do not compute the NLC of linear layers, but set them to 1 according to proposition \ref{finiteNetNlcEquals1}.

\subsection{The NLC is robust to width change} \label{widthInvarianceSection}

\begin{figure}
\centering
\includegraphics[width=0.98\textwidth]{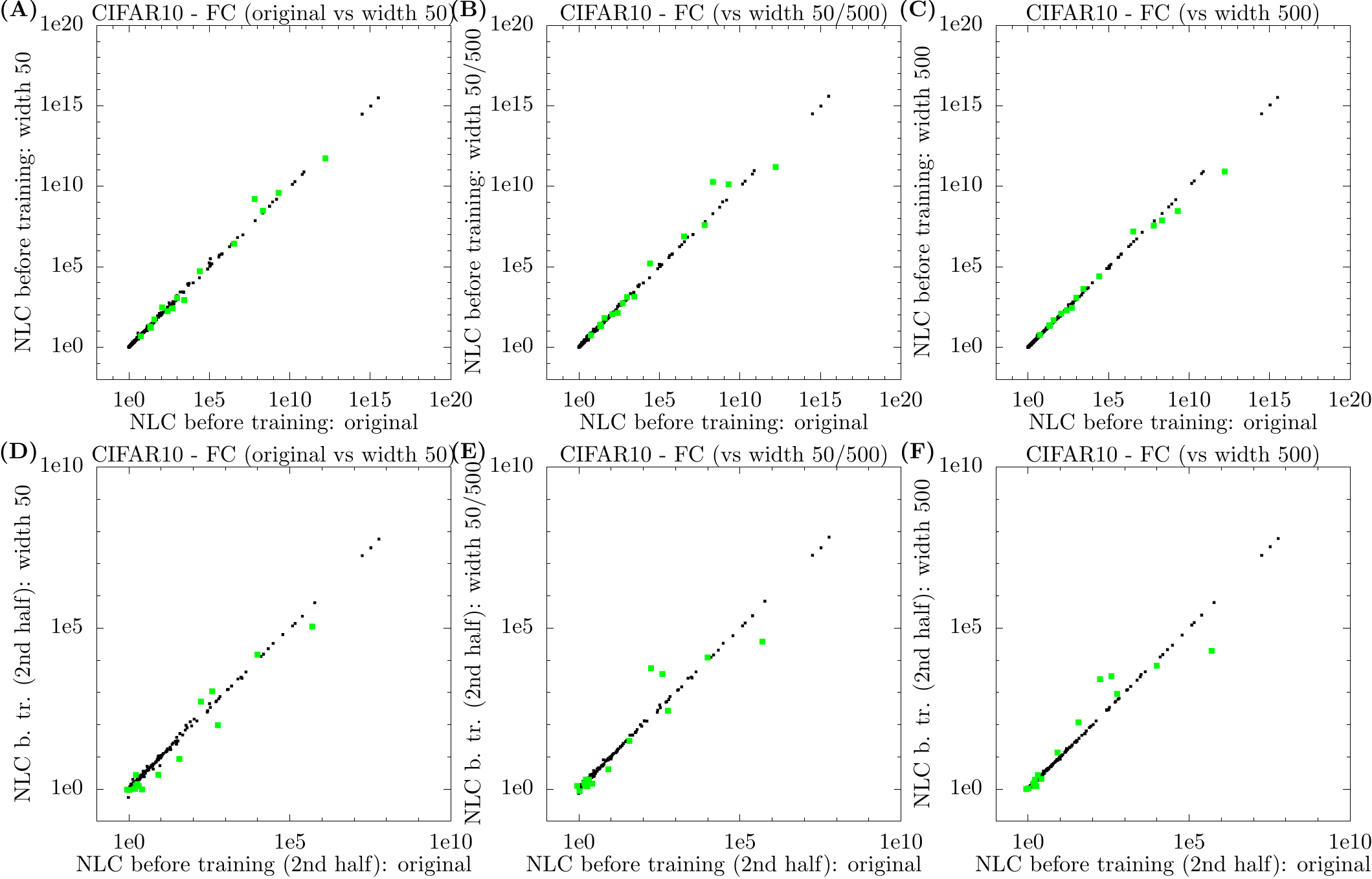}
\caption{Initial NLC for study A CIFAR 10 architectures vs equivalent architectures with altered width. Correlation values are close to 1. {\it Conclusion:} The NLC is robust to width change.} \label{nlcWidth}
\end{figure}

Changing layer widths does not impact the NLC significantly, as long as those layers do not become unreasonably narrow. We compared the NLC of each of our CIFAR10 architectures from study A in the initial state vs the NLC of corresponding architectures with different layer widths. In figure \ref{nlcWidth}A, we change the architecture width to 50. In figure \ref{nlcWidth}B, we change layer widths so that the fully-connected layers oscillate in width between 50 and 500. In figure \ref{nlcWidth}C, we change the architecture width to 500. In all cases, we find \finding{the NLC was preserved almost perfectly, except for GUAs}.

Of course, in all these cases, the width of the input and output layer remained constant, as dictated by the dataset. In figure \ref{nlcWidth}D-F, we evaluate the NLC only on the second half of the network, as in section \ref{nlcSimpleMetricsSection}. The dimensionality of the surrogate input layer to the second half of the network varies both between architectures and before vs after width change. We find that \finding{this does not alter the strength of the association}.

Two things are worth keeping in mind. First, altering width also changes initial weight variance as dictated by LeCun initialization (section \ref{statusQuoSection}, section \ref{additionalExperimentsSection}). This is necessary for obtaining robustness to width change. Second, changing width causes a random re-initialization of the weights, as in section \ref{nlcRandomInitSection}, but does not impact any other processes controlled by the random seed.

\subsection{The initial NLC predicts the final NLC} \label{nlcEvolutionSection}

\begin{figure}
\centering
\includegraphics[width=0.98\textwidth]{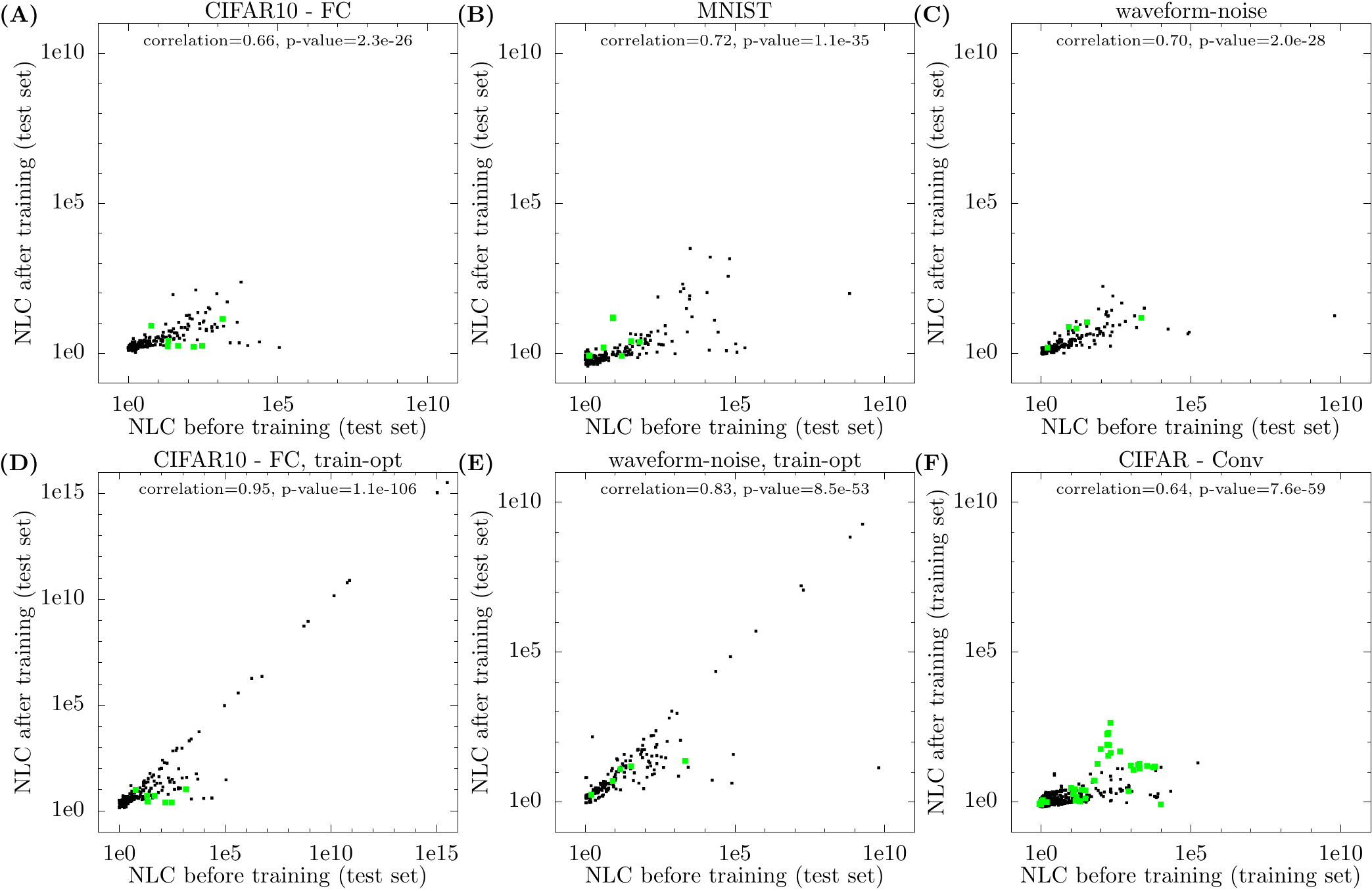}
\caption{Initial NLC vs final NLC for study A and B architectures. {\it Conclusion:} The initial NLC is associated with the final NLC.} \label{nlcEvolution}
\end{figure}

In section \ref{nlcPredictiveSection}, we showed that the NLC is associated with test error, both before and after training. In this section, we show that the NLC before training is also predictive of the NLC after training. In figure \ref{nlcEvolution}, we plot both NLCs vs each other, for both study A and B architectures. Throughout this chapter, we have evaluated NLCs on the training set before training and on the test set after training, unless otherwise stated. To enable a direct comparison, we evaluated both NLCs on the test set for fully-connected architectures, and we evaluated both NLCs on the training set for convolutional architectures in this subsection. For study B, the initial NLC was not available on the test set due to code base limitations. By our results in figure \ref{nlcDataGaussInit} and \ref{nlcDataGaussConv}, this should not be a big issue.

We find that \finding{while the NLC decreases overall, there appears to be a linear relationship between the logarithm of both quantities, with a slope of less than 1}. Graphs D and E are especially interesting. There, we plot the NLC before vs after training when the architectures were trained to minimize training error. As we showed in figures \ref{nlcPredTrainInit} and \ref{nlcPredTrainFinal}, some architectures with high NLC achieved a better-than-random training error. It turns out that \finding{those high-NLC architectures preserved their NLC value almost exactly. This is in contrast with low-NLC architectures, which tend to experience a decrease}. This points towards high-NLC architectures absorbing the information from the training set differently, perhaps via what is termed ``memorization'' in the community (e.g. \citet{memorization,memorization2}).

We interpret these results in conjunction with the results in section \ref{nlcPredictiveSection} as follows. When the NLC is sufficiently small, then the architecture can learn to adopt a suitable NLC, which is close to 1, and achieve better-than-random test error. However, only if the initial NLC is already ideal can a truly optimal performance be reached. If the NLC is too large, the architecture can, if it is trainable, memorize the training data but will only achieve a random test error. Further investigating the evolution of the NLC during training is an interesting topic for future work.

There is one significant outlier to the trends described above. \finding{One waveform-noise architecture that started with an NLC of around $10^{10}$ ended up with an NLC of around $10$}. In subsection \ref{nlcPredictiveSection}, we found this architecture to be the only architecture with a very large NLC that achieved a better-than-random test error. This drastic change in NLC is what enabled this relatively low error. We further investigated this architecture and found that \finding{the learning rate that led to this change led the parameter update during the first iteration of training to be several orders of magnitude greater than the initial parameter value}. This immediately changed the fundamental properties of the network function, including its nonlinearity. In general, large learning rates have the power to cause such changes. Still, we found that it is necessary to start with a good NLC to attain optimal and reliable performance. We further investigate the relationship between parameter and update magnitude and NLC in e.g. sections \ref{meanFieldPracticalEmpiricalSection} and \ref{learningRateSection}.

\subsection{The NLC is continuous from layer to layer} \label{nlcBackpropSection}

We end section \ref{nlcPropertiesSection} by demonstrating that the NLC changes gradually, and often with great regularity, from layer to layer as the Jacobian is backpropagated. That is, when we write a network as a dependency chain $f_L(f_{L-1}(..(f_1(x))..))$, then $NLC(f_L(f_l),f_l(\mathcal{D}))$ changes smoothly as $l$ decreases from $L$ to 0.

\begin{figure}[H]
\centering
\includegraphics[width=0.98\textwidth]{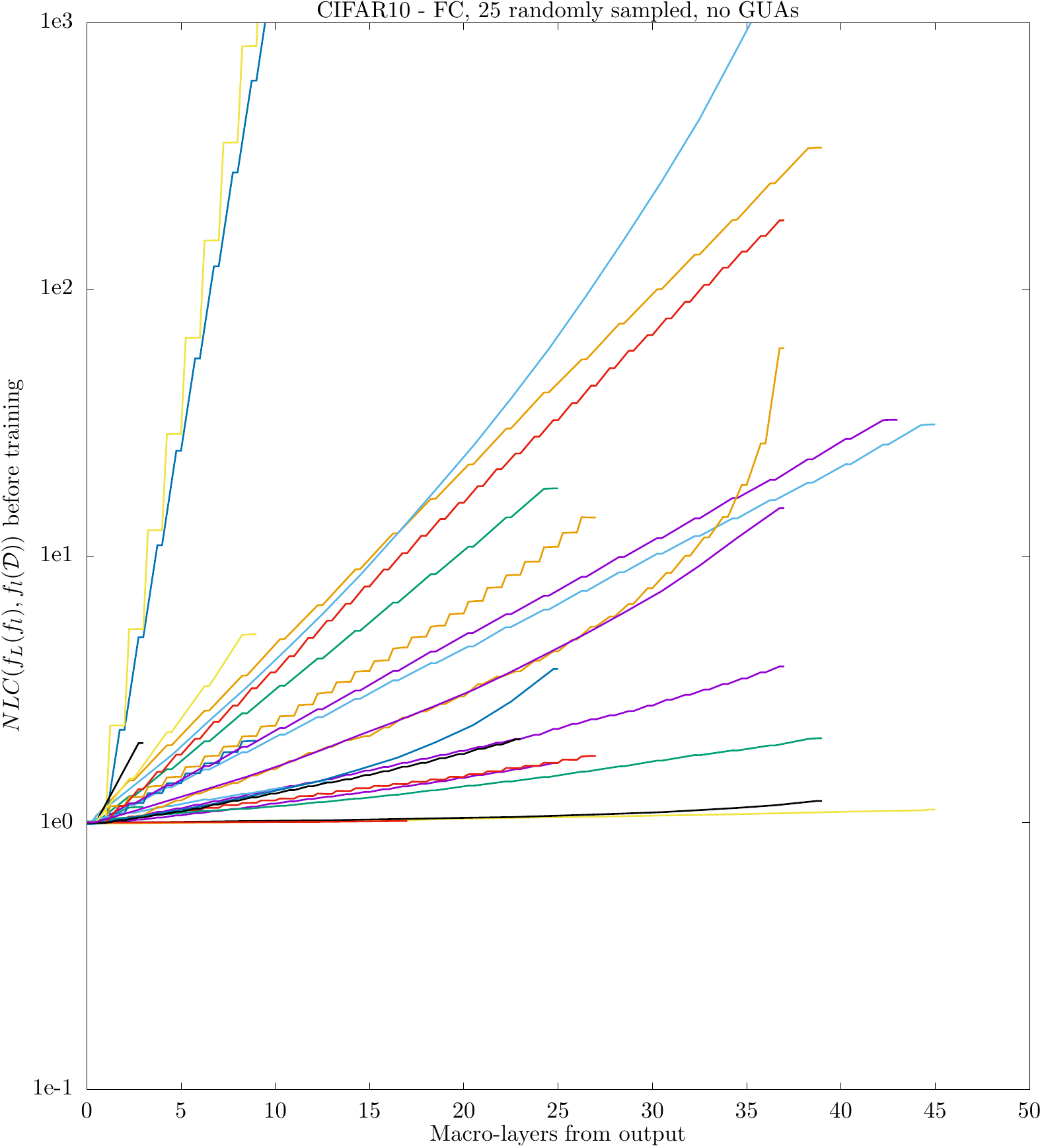}
\caption{$NLC(f_L(f_l),f_l(\mathcal{D}))$ at different layers $f_l$ for 25 randomly selected study A CIFAR10 architectures that are not GUAs in the initial state. We plot the value for a layer $f_l$ if it is a fully-connected layer, a normalization layer, an activation layer or an addition layer, unless that layer is bypassed by a skip connection. All these layers are placed on the x-axis according to their distance from the output layer, measured in the (possibly fractional) number of macro-layers. The curves arise by connecting points corresponding to neighboring layers. {\it Conclusion:} The NLC increases smoothly from layer to layer. Often, this change is linear in log space from macro-layer to macro-layer. Often, there is a jitter for every macro-layer.} \label{nlcBackpropCifar}
\end{figure}

In figure \ref{nlcBackpropCifar}, we plot $NLC(f_L(f_l),f_l(\mathcal{D}))$ for 25 randomly selected non-GUA study A CIFAR10 architectures in the initial state. We place each linear layer, activation layer, normalization layer and addition layer on the x-axis according to its distance from the output layer, measured in the (possibly fractional) number of macro-layers. So, for example, if an architecture has $M$ macro-layers, then the value of the NLC of the output as a function of the normalization layer in macro-layer $M-2$ is plotted at x-coordinate 2.25, and the value of the NLC of the output as a function of the linear layer in macro-layer $M-20$ is plotted at x-coordinate 20.75, assuming $M > 20$. The x-coordinate of all layers is ordered according to their ordering in the network. See section \ref{studyAArchitecturesSection}. Of course, if a given architecture has, say, no normalization layers, nothing is plotted for those layers. The curves arise from connecting neighboring plotted points for each architecture. If an architecture is residual, we omit all layers that are bypassed by skip connections as they are not bottlenecks.

There are a number of interesting findings from figure \ref{nlcBackpropCifar}. First, we confirm that \finding{the NLC changes smoothly from layer to layer}. We also find that \finding{the NLC increases from layer to layer. For most architectures, that increase is linear from macro-layer to macro-layer in log space. Also, for many architectures, there is a small jitter for every macro-layer}. The jitter occurs because the change to the NLC depends on the type of layer that is being newly included in the section of the network on which the NLC is evaluated. While it is not directly apparent from figure \ref{nlcBackpropCifar}, we found that \finding{whenever we include another activation layer, the NLC increases. Conversely, when we include another normalization or linear layer, the NLC is stable}. It turns out that including an additional linear layer always preserves the NLC exactly, by proposition \ref{finiteNetNlcAnyMatrix}. (Note that the value of the estimator of the NLC can and does still change slightly, due to estimation error.) Including an additional bias layer also preserves the NLC, as well as preserving the estimator. (Therefore, in figure \ref{nlcBackpropCifar}, we do not plot bias layers.) Note that \finding{we do not see jitters for residual architectures}, because we do not plot every layer.

These results are somewhat expected, given our results from section \ref{nlcDecomposableSection}. We found that the NLC in the initial state could be replicated by the (sum-)product of layerwise NLCs. By theorem \ref{finiteNetNlcGreater1} and section \ref{nlcRobustDataSection}, we would expect those NLCs to be least 1. Hence, we would expect the curves in figure \ref{nlcBackpropCifar} to approximately arise from successively multiplying values that are at least 1, up to estimation error.

We make one final observation with regards to figure \ref{nlcBackpropCifar}. While the curves increase roughly linearly for a majority of architectures, \finding{they are bending upwards for others and stay close to the y-coordinate of 1 for others}. We explain those different behaviors in detail in chapter \ref{surveyChapter}.

\begin{figure}[H]
\centering
\includegraphics[width=0.98\textwidth]{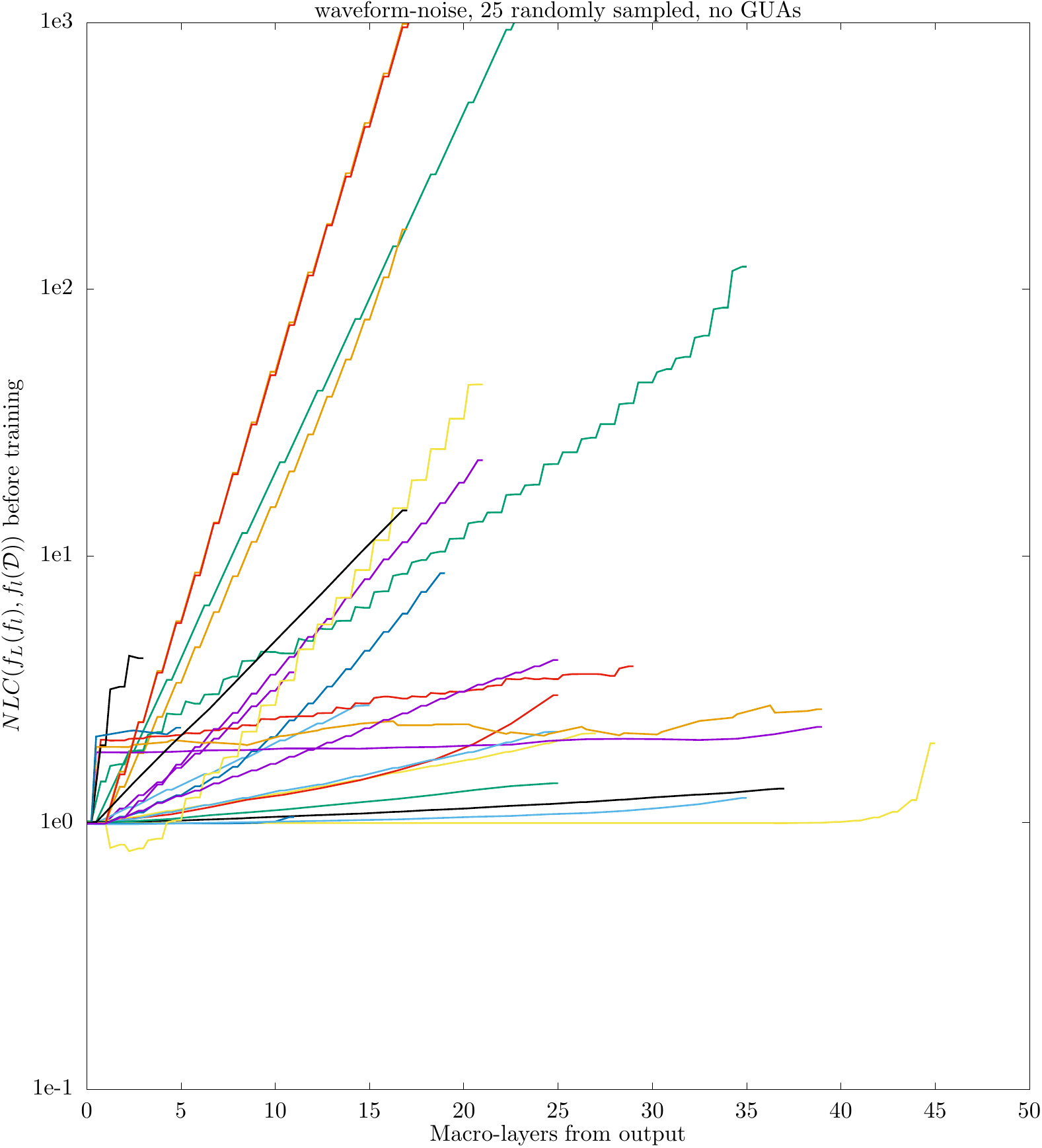}
\caption{$NLC(f_L(f_l),f_l(\mathcal{D}))$ at different layers $f_l$ for 25 randomly selected study A waveform-noise architectures that are not GUAs in the initial state. The graph is analogous to figure \ref{nlcBackpropCifar}. {\it Conclusion:} The NLC still changes smoothly from layer to layer, but the curves are less regular than in figure \ref{nlcBackpropCifar} and not always increasing.} \label{nlcBackpropWave}
\end{figure}

In figure \ref{nlcBackpropWave}, we give the same NLC curves as before, but for 25 randomly selected non-GUA waveform-noise architectures. Here, \finding{the curves appear somewhat less regular}. Specifically, we find that \finding{one curve dips below a value of 1. Several curves increase quickly at first but then become flat. Some linearly increasing curves are also not perfectly regular}.

\begin{figure}[H]
\centering
\includegraphics[width=0.98\textwidth]{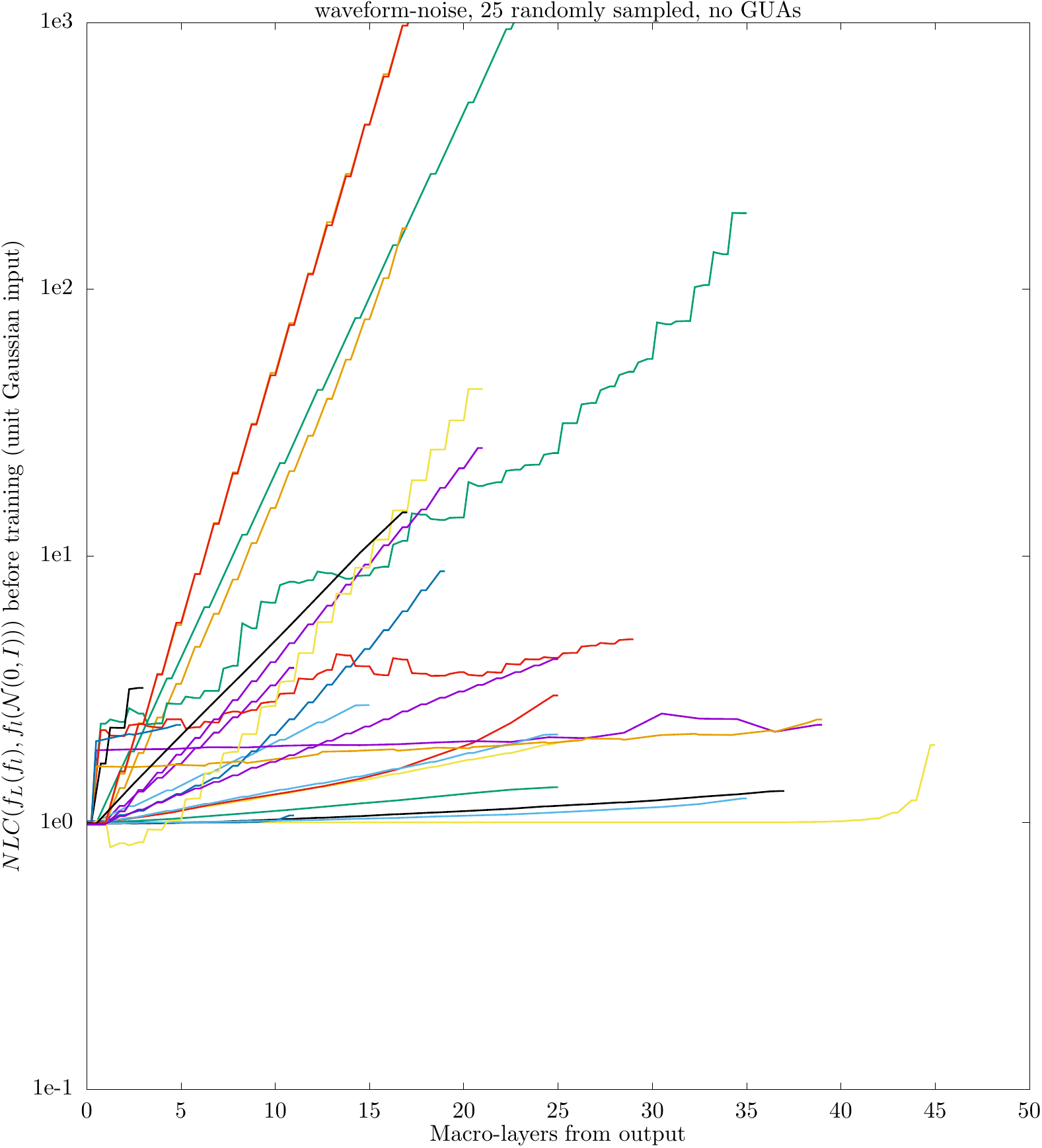}
\caption{$NLC(f_L(f_l),f_l(\mathcal{N}(0,I)))$ at different layers $f_l$ for the same 25 architectures as in figure \ref{nlcBackpropWave} in the initial state. The graph is analogous to figure \ref{nlcBackpropCifar}. {\it Conclusion:} The curves closely track those of figure \ref{nlcBackpropWave}.} \label{nlcBackpropGaussian}
\end{figure}

We were interested in the cause behind the degradation of the patterns we found for CIFAR10. In figure \ref{nlcBackpropGaussian}, we plot the same curves for the same architectures as in figure \ref{nlcBackpropWave}, but we replace $\mathcal{D}$ with the unit Gaussian, as done in e.g. section \ref{nlcRobustDataSection}. \finding{While those curves are surprisingly even less regular than those in figure \ref{nlcBackpropWave}, they still track those in figure \ref{nlcBackpropWave} pretty closely.} This shows that the irregular behavior relative to figure \ref{nlcBackpropCifar} is not caused by the input distribution. We replicated the curves from figure \ref{nlcBackpropCifar} with unit Gaussian input and found that \finding{they are an exact match to figure \ref{nlcBackpropCifar}}. So, CIFAR10 and waveform-noise architectures show different behavior, even on the same inputs. Both groups of architectures were drawn from the same distribution over architectures, with one exception. CIFAR10 architectures have $d_\text{in} = 810$ and $d_\text{out} = 10$. waveform-noise architectures have $d_\text{in} = 40$ and $d_\text{out} = 3$. So it is their narrow input and / or output layer that make waveform-noise architectures irregular in these experiments.

Mean field theory (chapter \ref{meanFieldNnaChapter}) relies on architectures having at least a moderate width. Here we have an example where low width when exhibited by only two layers harms the regularity of networks.

\begin{figure}[H]
\centering
\includegraphics[width=0.98\textwidth]{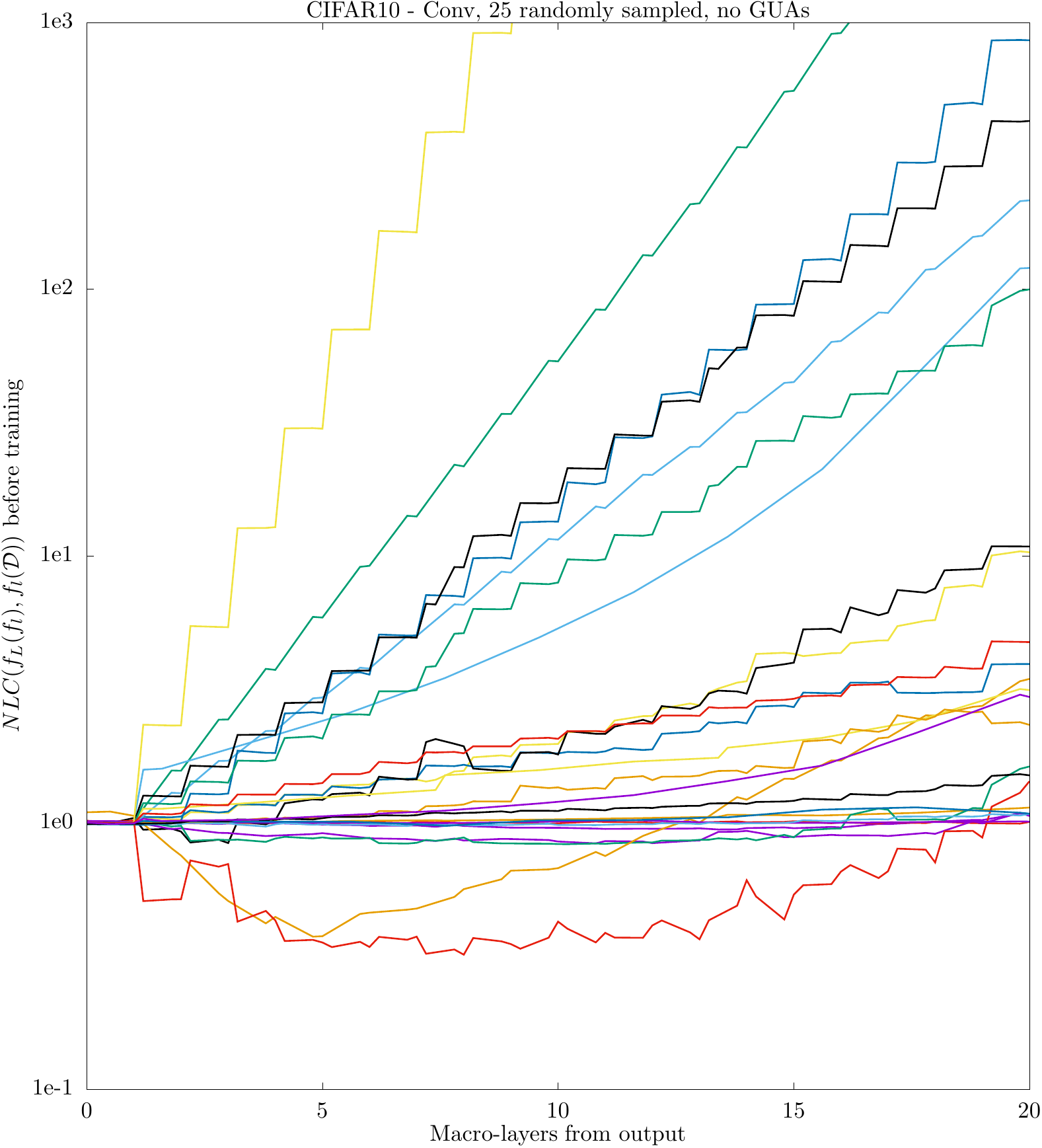}
\caption{$NLC(f_L(f_l),f_l(\mathcal{D}))$ at different layers $f_l$ for 25 randomly selected study B architectures that are not GUAs in the initial state. We plot the value for layer $f_l$ if it is a linear layer, a normalization layer, an activation layer, an addition layer or a pooling layer, unless that layer is bypassed by a skip connection.  All these layers are placed on the x-axis according to their distance from the output layer, measured in the (possibly fractional) number of macro-layers. The curves arise by connecting points corresponding to neighboring layers. {\it Conclusion:} The NLC changes smoothly from layer to layer, but the curves are less regular than in figure \ref{nlcBackpropCifar}.} \label{nlcBackpropConvi0}
\end{figure}

In figure \ref{nlcBackpropConvi0}, we plot the same curves for 25 randomly selected non-GUA study B convolutional architectures in the initial state. Compared to fully-connected architectures, \finding{while we still see many of the same patterns, the regularity is degraded}. \finding{The NLC dips significantly below zero for some layers of 2 of the 25 architectures.} We note that the number of channels in the convolutional architectures we studied was considerably less than the width of layers in the fully-connected architectures we studied, as is usually the case in practice. For this and further reasons explained in section \ref{meanFieldCNNsection}, we would expect less regularity.

\begin{figure}[H]
\centering
\includegraphics[width=0.49\textwidth]{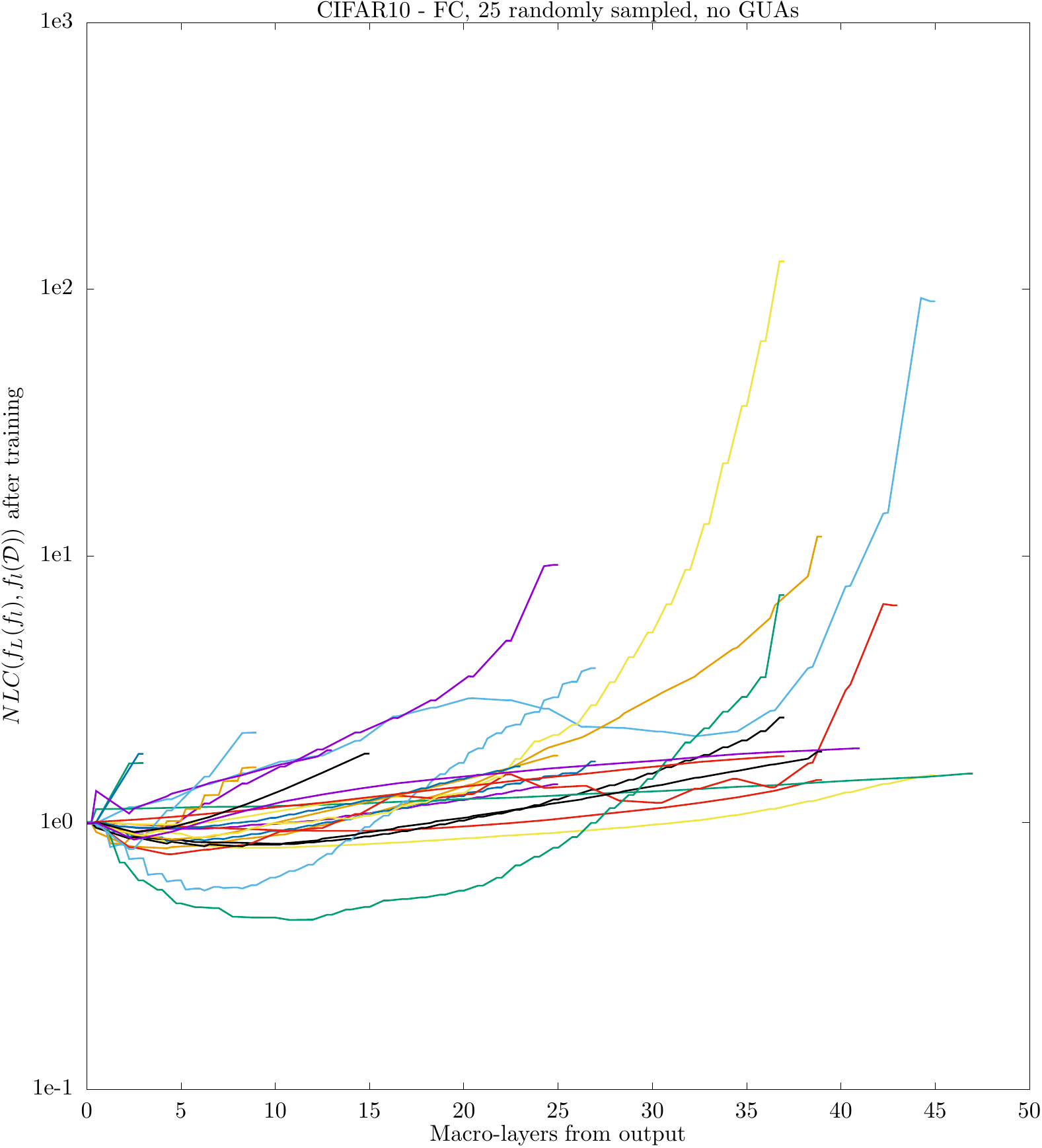}
\includegraphics[width=0.49\textwidth]{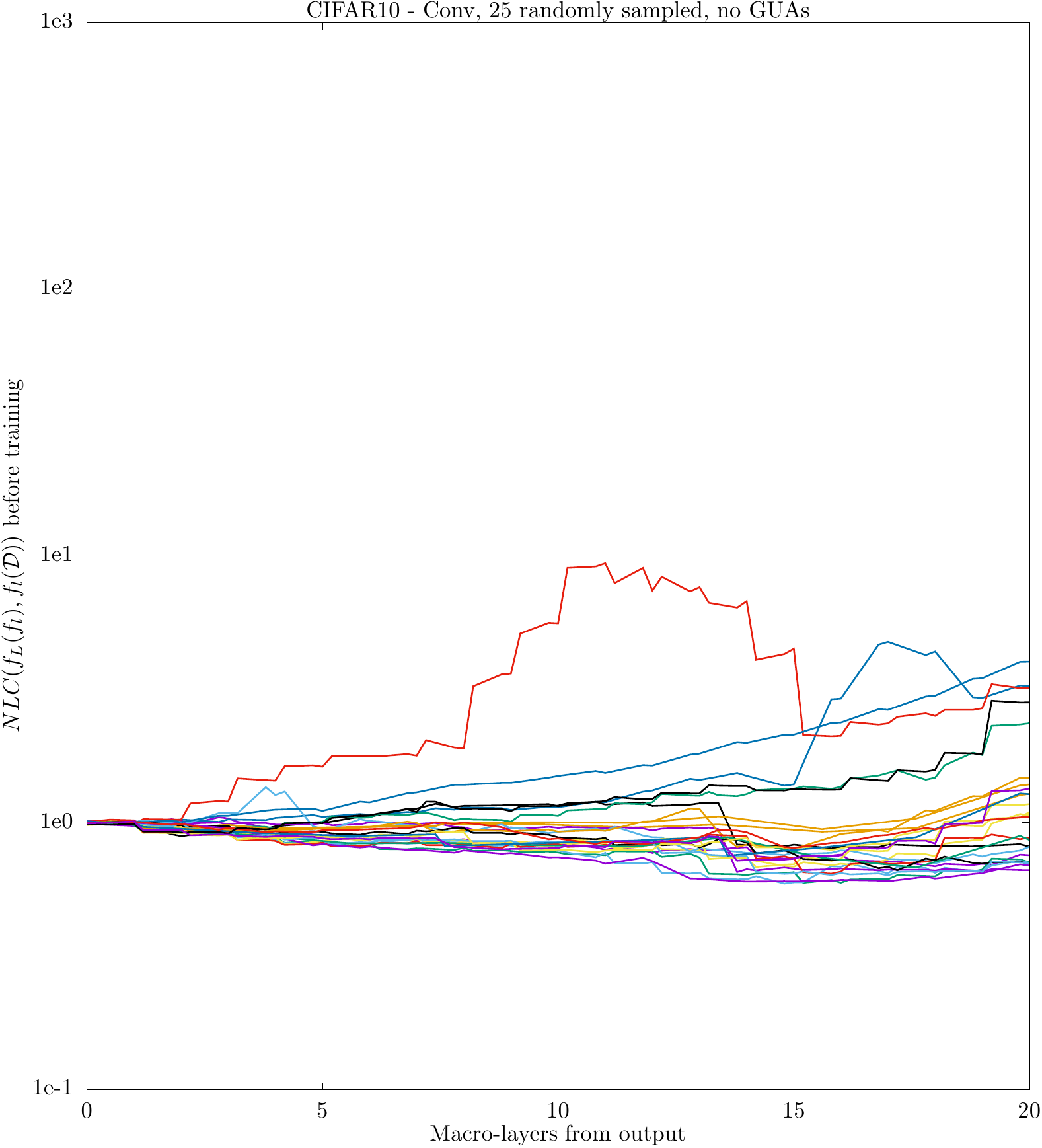}
\includegraphics[width=0.49\textwidth]{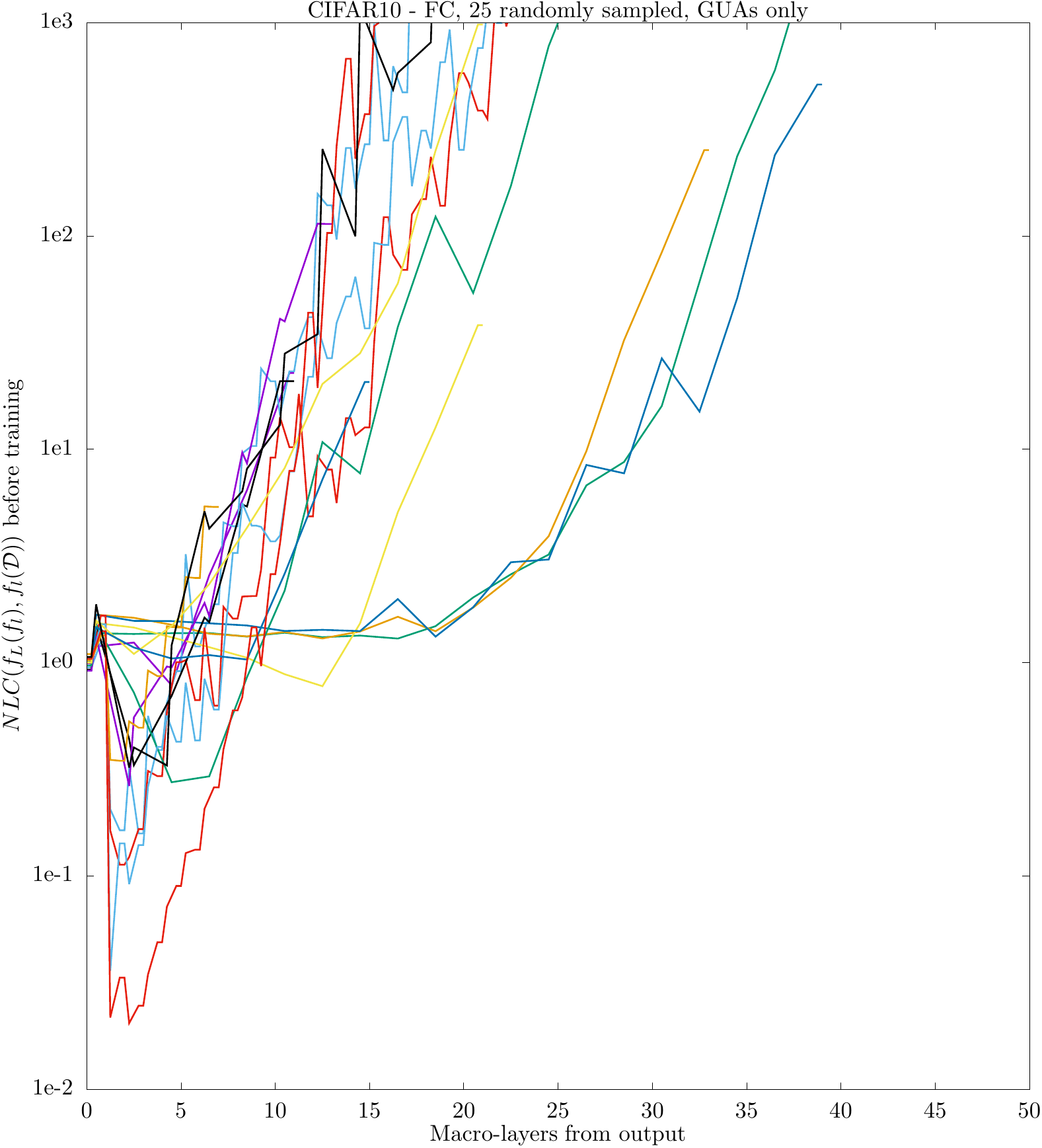}
\includegraphics[width=0.49\textwidth]{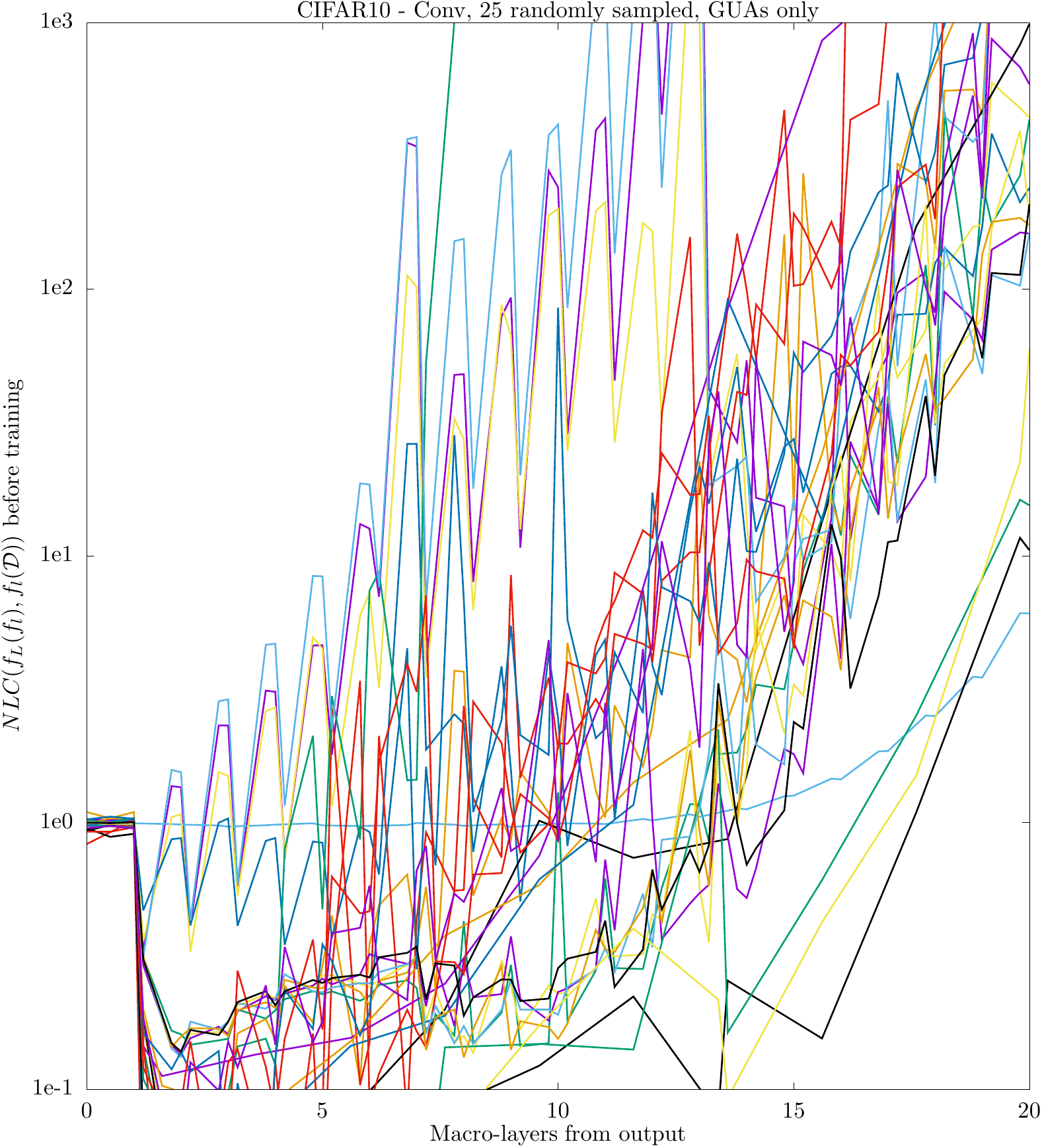}
\caption{$NLC(f_L(f_l),f_l(\mathcal{D}))$ at different layers $f_l$ for architectures from figure \ref{nlcBackpropCifar} in the final state (top left); for architectures from figure \ref{nlcBackpropConvi0} in the final state (top right); for 25 randomly selected study A CIFAR10 GUAs in the initial state (bottom left); for 25 randomly selected study B GUAs in the initial state (bottom right). Graphs are analogous to figure \ref{nlcBackpropCifar} for study A and \ref{nlcBackpropConvi0} for study B. {\it Conclusion:} The patterns of previous figures degrade significantly in the final state, and completely for GUAs.} \label{nlcBackpropCombo}
\end{figure}

As usual, we find that \finding{the regularity further degrades when we look at the final state}. In figure \ref{nlcBackpropCombo}(top), we depict the architectures from figures \ref{nlcBackpropCifar} and \ref{nlcBackpropConvi0} in their final state. \finding{While the NLC is still relatively smooth from layer to layer, the curves are neither reliably linear nor reliably increasing}. Finally, in figure \ref{nlcBackpropCombo}(bottom), we depict GUAs in the initial state. \finding{Chaos reigns!}

\newpage

\newpage

\section{What is the best NLC for a dataset?} \label{bestNlcSection}

In subsection \ref{nlcPredictiveSection}, we showed that the NLC in the initial state predicts test error. For all three datasets we studied, architectures that attained close-to-optimal test error had an initial NLC between 1 and 5. In this section, we show that the optimal range is not universal across all datasets, but argue that it should be highly similar for most practical datasets.

\begin{table}
{
\centering
\begin{tabular}{lccc}
Dataset&CIFAR10&MNIST&waveform-noise \\ \hline\hline 
\\
\begin{tabular}[x]{@{}l@{}}PNLCD\\density\end{tabular}&\includegraphics[scale=0.302,valign=c]{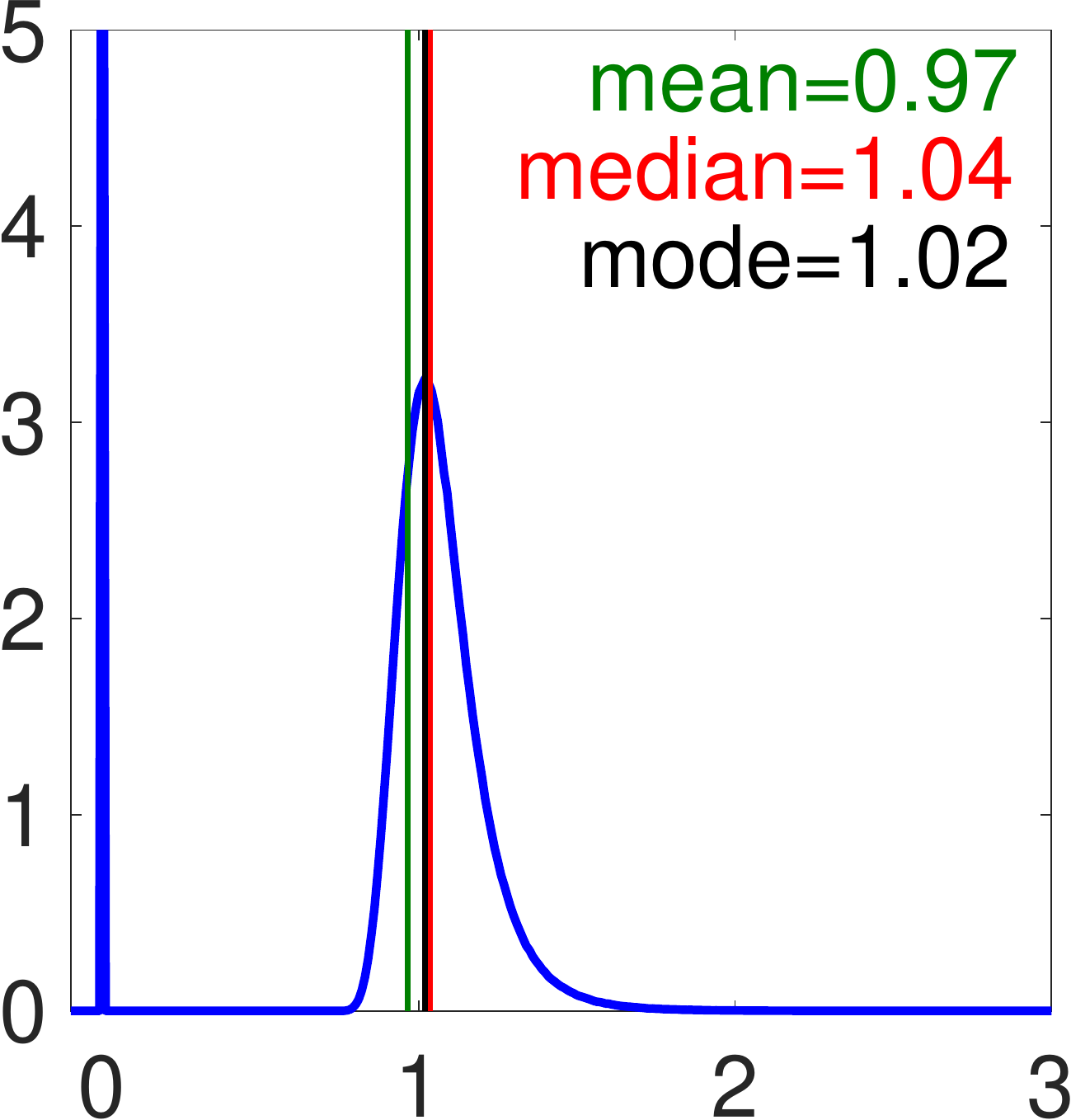}&\includegraphics[scale=0.302,valign=c]{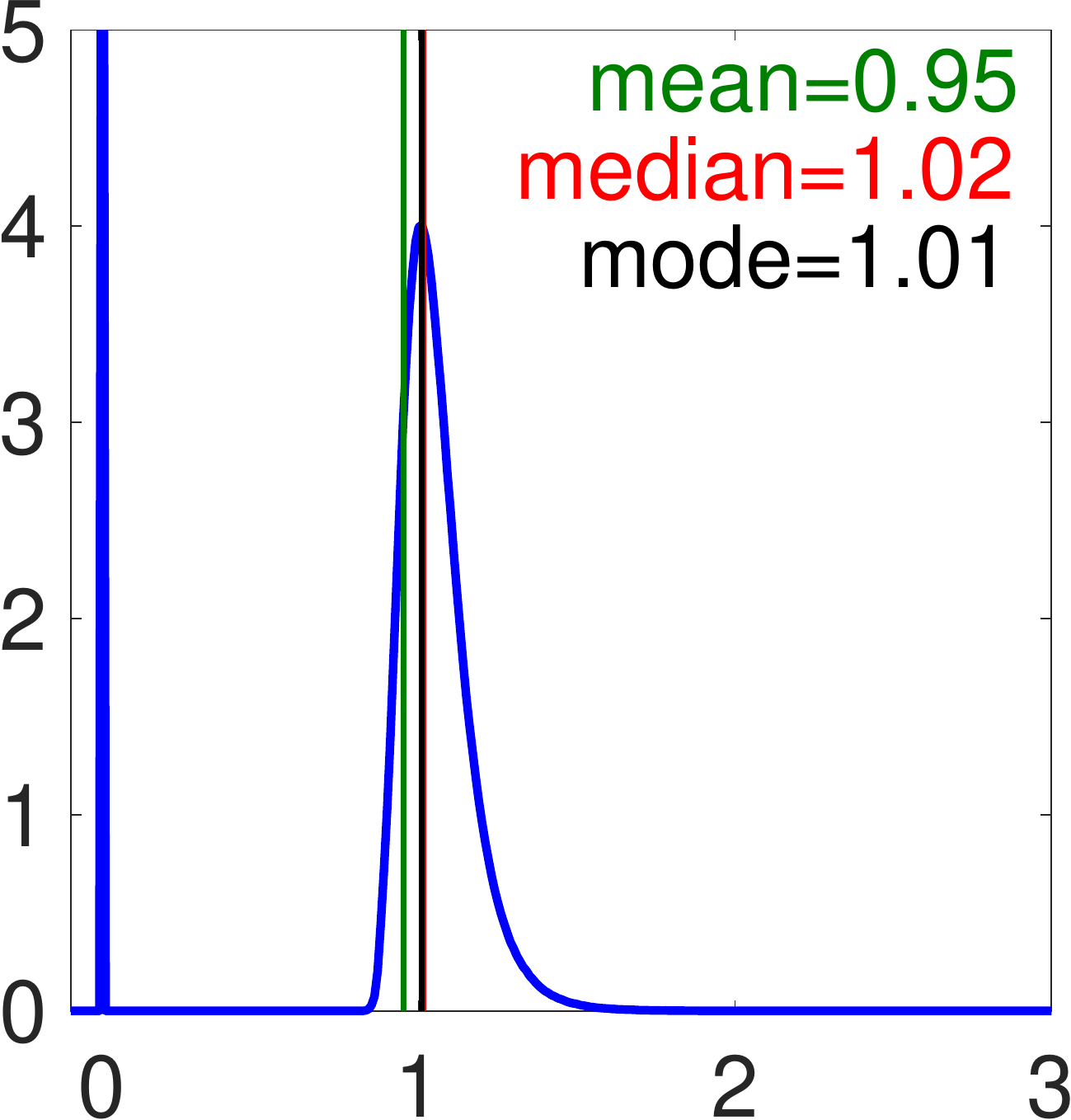}&\includegraphics[scale=0.302,valign=c]{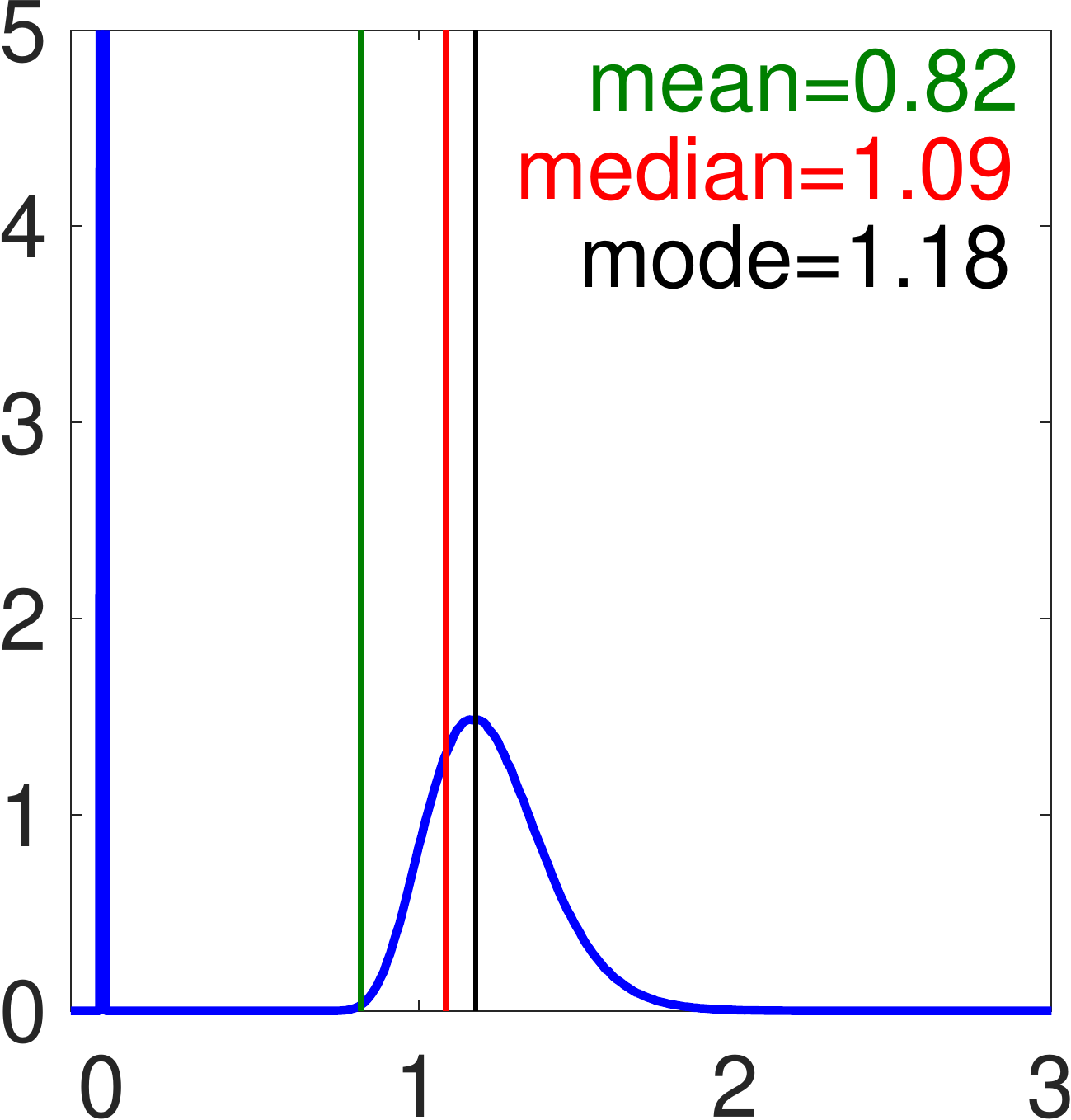}
\end{tabular}
\caption{PNLCD evaluated on the union of training and validation set for study A datasets. The mode ignores the $PNLCD = 0$ point. {\it Conclusion:} The true input-label functions of our datasets appear to be only slightly nonlinear.}
\label{groundtruth}
}
\end{table}

In sections \ref{nlcLinearApproximationSection}, \ref{nlcNoiseSection} and \ref{nlcKernelSection}, we showed how the NLC is related to underfitting and overfitting. If the output of the network $f$ is too sensitive to input perturbation, then it cannot generalize from training to test points that are a certain distance apart. Conversely, if the network is too close to a linear function, its performance does not exceed that of a linear model. Overall, we suspect that an ideal network has a degree of nonlinearity that matches the nonlinearity of the true input-label function. So, to choose an optimal NLC to choose an optimal architecture, we would like to estimate that nonlinearity. In this section, we give a very simple method for (crudely) doing so.

Consider two datapoints $(x,y)$ and $(x',y')$ drawn from $\mathcal{D}$. We can estimate the fraction of the domain that lies between the two inputs as $\frac{||x-x'||_2}{2\sqrt{\Tr(\Cov_x)}}$ and the fraction of the codomain that lies between the two outputs as $\frac{||y-y'||_2}{2\sqrt{\Tr(\Cov_y)}}$. We proxy the diameter of the hypothetical domain and codomain with twice the trace, as in sections \ref{nlcDefinitionSection} and \ref{nlcSensiSection}. In the case of classification, where $y$ represents a class label, we cast $y$ as a `one-hot vector', i.e. a vector of dimensionality equal to the number of classes that has a 0 at each component except at the component corresponding to the $y$'th class, where it has a 1. Therefore, for this pair of points. we can estimate the fraction of the codomain traversed between them relative to the fraction of the domain traversed as $\sqrt{\frac{||y-y'||_2^2\Tr(\Cov_x)}{||x-x'||_2^2\Tr(\Cov_y)}}$. 

\begin{metricDefinition}
The `pairwise nonlinearity coefficient distribution' (PNLCD) is

$$PNLCD(\mathcal{D}) = \sqrt{\frac{||y-y'||_2^2\Tr(\Cov_x)}{||x-x'||_2^2\Tr(\Cov_y)}} \text{ where } (x,y),(x',y') \sim \mathcal{D}$$

Here, $y$ is cast as a one-hot vector in the case of classification.

\end{metricDefinition}

PNLCD is similar to the NLC as $\Cov_y$ is similar to $\Cov_f$, and $\frac{||y-y'||_2}{||x-x'||_2}$ can be viewed as the derivative along the line segment from $(x,y)$ to $(x',y')$. Note that we do not summarize PNLCD as a scalar as it is not as well-behaved as we observed e.g. GLLAD to be in section \ref{nlcSensiSection}. In table \ref{groundtruth}, we depict PNLCD for our 3 datasets evaluated on the union of training and validation set. We also give the expectation, median and mode. Note that the mode ignores the $PNLCD = 0$ point.

We find that \finding{each distribution has a singular peak at zero}. This corresponds to pairs of datapoints that have the same label. \finding{The remaining probability mass of PNLCD concentrates around 1, with the median and mode being slightly larger than 1 in all cases}. Therefore, according to PNLCD, the input-label function appears to be only very slightly nonlinear. While this may seem surprising at first glance, it is actually expected. If both $x$ and $y$ were drawn IID from high-dimensional unit Gaussian distributions, PNLCD would concentrate around 1. In this idealized context, obtaining even a single large sample value from PNLCD would require sampling a number of datapoint pairs that is exponential in the number of dimensions. If anything, we would expect a practical input-label function to appear more linear than Gaussian noise. Unless we have specific reason to believe that there are clusters of inputs with varying labels that concentrate in regions of input space that are small relative to the domain itself, then we should expect PNLCD to concentrate around 1.

Now we demonstrate that PNLCD predicts the ideal NLC for an architecture for a dataset. Unfortunately, we are not familiar with any specific practical dataset with yields PNLCD samples that include many large values, for the reasons given above. Hence, we need to study the relationship between NLC and PNLCD on artificial datasets.

We generated artificial datasets as follows. We drew two 40-dimensional vectors from the unit Gaussian distribution. Then we assigned each datapoint in the waveform-noise dataset to either one of these two noise vectors at random. We replaced each input in the waveform-noise dataset with the weighted average of itself and the assigned noise vector. This yielded a modified dataset. Inputs that were assigned to different noise vectors cause $\Tr(\Cov_x)$ to not become too small for the modified dataset, whereas points assigned to the same noise vector cause some $||x-x'||_2^2$ values to become very small. We generated 6 datasets in this way. We used the same 2 noise vectors for each of them, but weighted those vectors differently when taking the average between them and the waveform-noise inputs. If the weight associated with the noise vector was $w$, then each input in the artificial dataset $x'$ was generated as $x' = wv + (1-w)x$, where $x$ was the corresponding input from waveform-noise and $v$ is the assigned noise vector. The labels remained the same.

In table \ref{dsetfusion}, we give PNLCD for each of the artificial datasets. \finding{As expected, PNLCD sample values corresponding to pairs of datapoints assigned to the same noise vector that also have different labels increase proportionally with the noise weight. PNLCD sample values corresponding to pairs of datapoints that are assigned to different noise vectors and have different labels remain around 1}. Note that we do not depict the peak at $PNLCD=0$ corresponding to pairs with the same label because the density is depicted in log scale. 

We then trained 19 depth-2 fully-connected sawtooth architectures with layer normalization on each of the 6 datasets. We used the same careful protocol as in study A (section \ref{studyATrainingSection} / \ref{metricsSummarySection}). In each architecture, we dilated the sawtooth activation function $\tau(s)$ (table \ref{actFunIllu}) with a different fixed constant, i.e. we replaced $\tau(s)$ with $\tau(ds)$ for some $d$ between $10^{-2}$ and $10^7$. This has essentially the same effect as scaling the inputs, which we did in section \ref{nlcRobustDataSection}. The larger the value of $d$, the higher the frequency of the activation function, and the higher the NLC. In table \ref{dsetfusion}, we give the initial NLC and test error attained by these architectures. As expected, \finding{the NLC is roughly proportional to $d$ and the ideal NLC for a dataset is roughly proportional to the expectation of PNLCD. When the noise weight is 0.9, an architecture that is almost linear in the initial state is still capable of adapting to the shortened distances between inputs. When the noise weight is at 0.99 or higher, then those architectures always underfit drastically}. We also note that \finding{the best achieved test error is higher for noise weights 0.99 or above}. We suspect that this is because these noise weights force the architecture to effectively learn to make correct predictions on two different datasets, one dataset per noise vector, with only half as much data available per dataset. 

Two aspects of our careful training protocol are critical for obtaining the test error values of figure \ref{dsetfusion}. Learning rate tuning allows the use of very small learning rates, which are required for high-NLC architectures (section \ref{learningRateSection}). 64-bit precision enables us to tease apart inputs that are very close together and to use small parameter updates that stem from small learning rates (section \ref{noiseStabilitySection}).

In light of this section, let us revisit section \ref{moduloDataSection}. There, we stated that we are interested in developing ZSAD guidelines that are data-agnostic, i.e. that apply across a large class of datasets that spans across task domains. Almost all of our empirical results from this chapter held for all 3 datasets, just not always to quite the same degree. In this section, we took a first significant step towards identifying a class of datasets across which architectures may show consistent behavior and a universal range of ``good'' NLC values, namely $[1,5]$. These are datasets where inputs are somewhat evenly distributed across the domain, like Gaussian noise, and where PNLCD concentrates around 1, at least for inputs with different labels. We extend this discussion in section \ref{elemLikeSection}.

We end with a few remarks. It is worth keeping in mind that even though the output of a linear function changes slowly when the input is perturbed with random noise, it can change extremely quickly when the input perturbation is in the direction of the gradient. In fact, the sensitivity to gradient perturbation scales with $\sqrt{d_\text{in}}$. Hence, even a linear network is capable of representing extremely quick change in {\it some} direction.

Another effect of high dimensionality is that even a network that is close to linear has an enormous degree of freedom. Imagine that a network is defined over the unit cube of dimensionality $d_\text{in}$ and side length 1, and assume that the cube is partitioned (roughly!) into subcubes of side length $1-\epsilon$ on which the network has to be linear. Then because the total number of subcubes would be $\frac{1}{1-\epsilon}^{d_\text{in}}$, the network would still have a number of degrees of freedom that is exponential in $d_\text{in}$! As long as the parameter and input are high-dimensional enough, this may provide an explanation of why we observe many networks with NLC very close to 1 still substantially outperform linear models.

One practical case of a dataset with inputs that are very non-uniform might be a dataset containing adversarial inputs and ``regular'' inputs, where the label denotes whether the input was generated via an adversarial perturbation (e.g. \citet{adversarialDetection}). These datasets might warrant further study in the context of nonlinearity.

\newpage

\begin{table}[H]
{
\centering
\begin{tabular}{lccc}
Noise weight&0&0.9&0.99 \\ \hline\hline 
\\
\begin{tabular}[x]{@{}l@{}}PNLCD\\density\end{tabular}&\includegraphics[scale=0.27,valign=c]{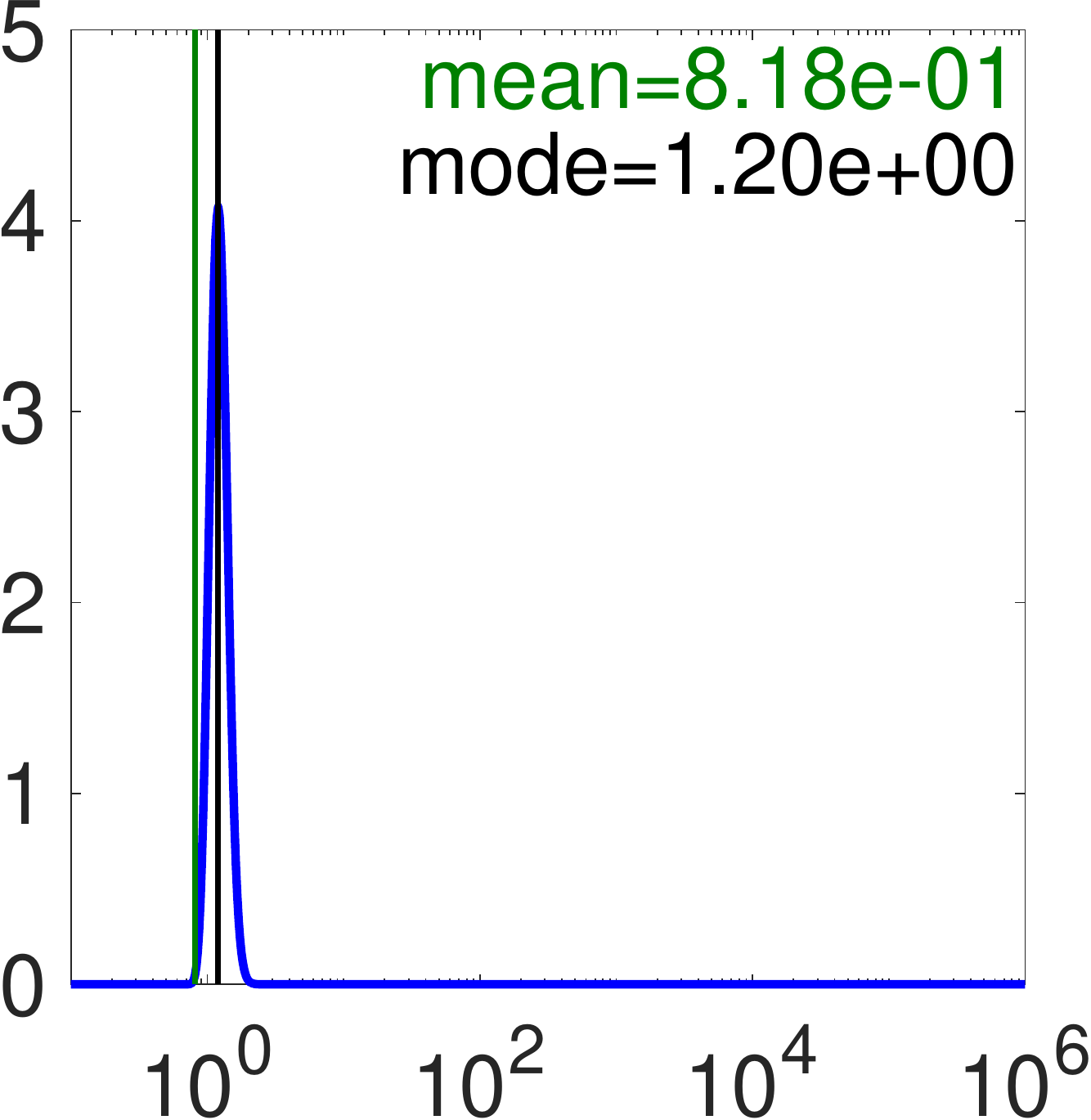}&\includegraphics[scale=0.27,valign=c]{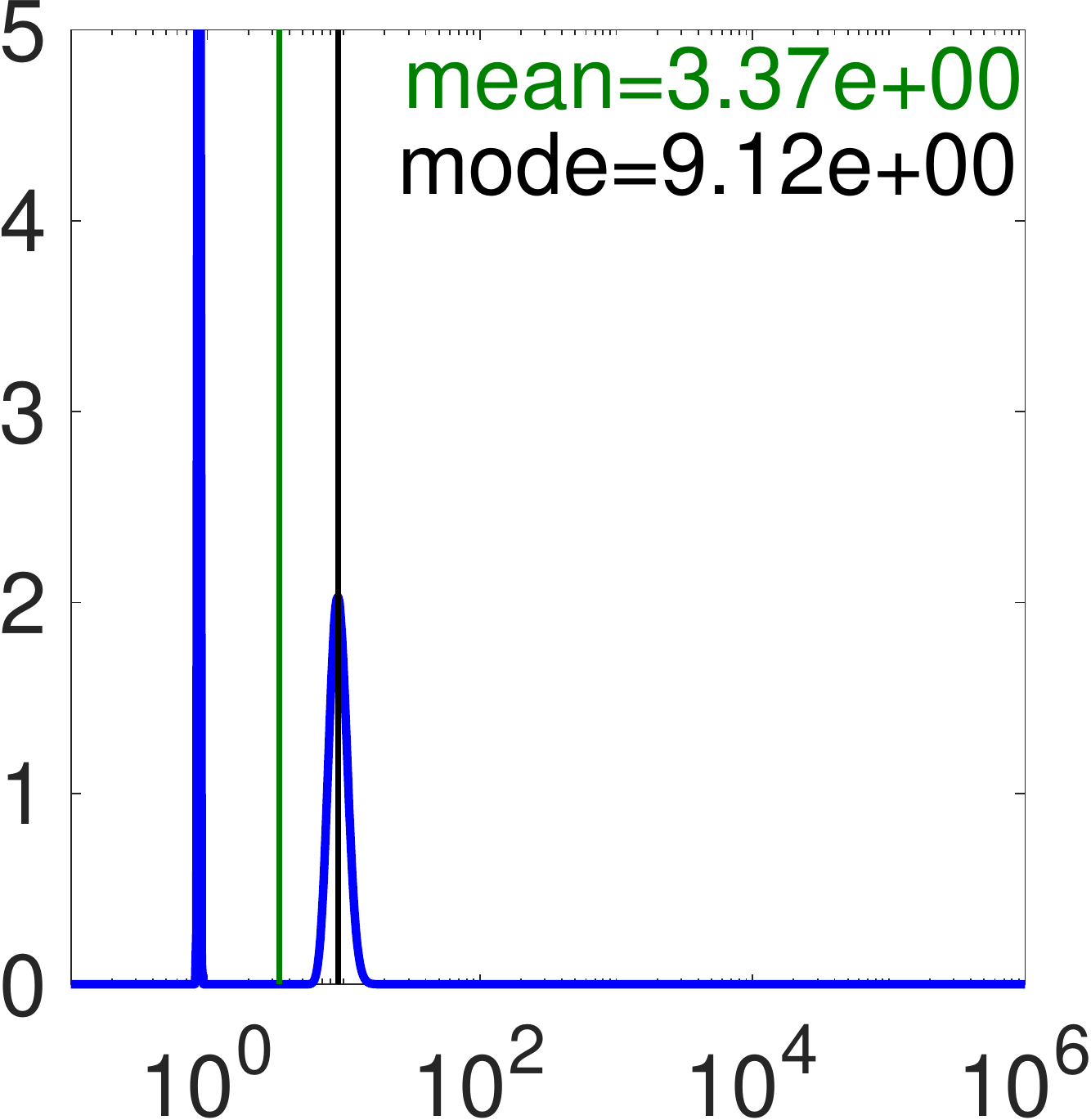}&\includegraphics[scale=0.27,valign=c]{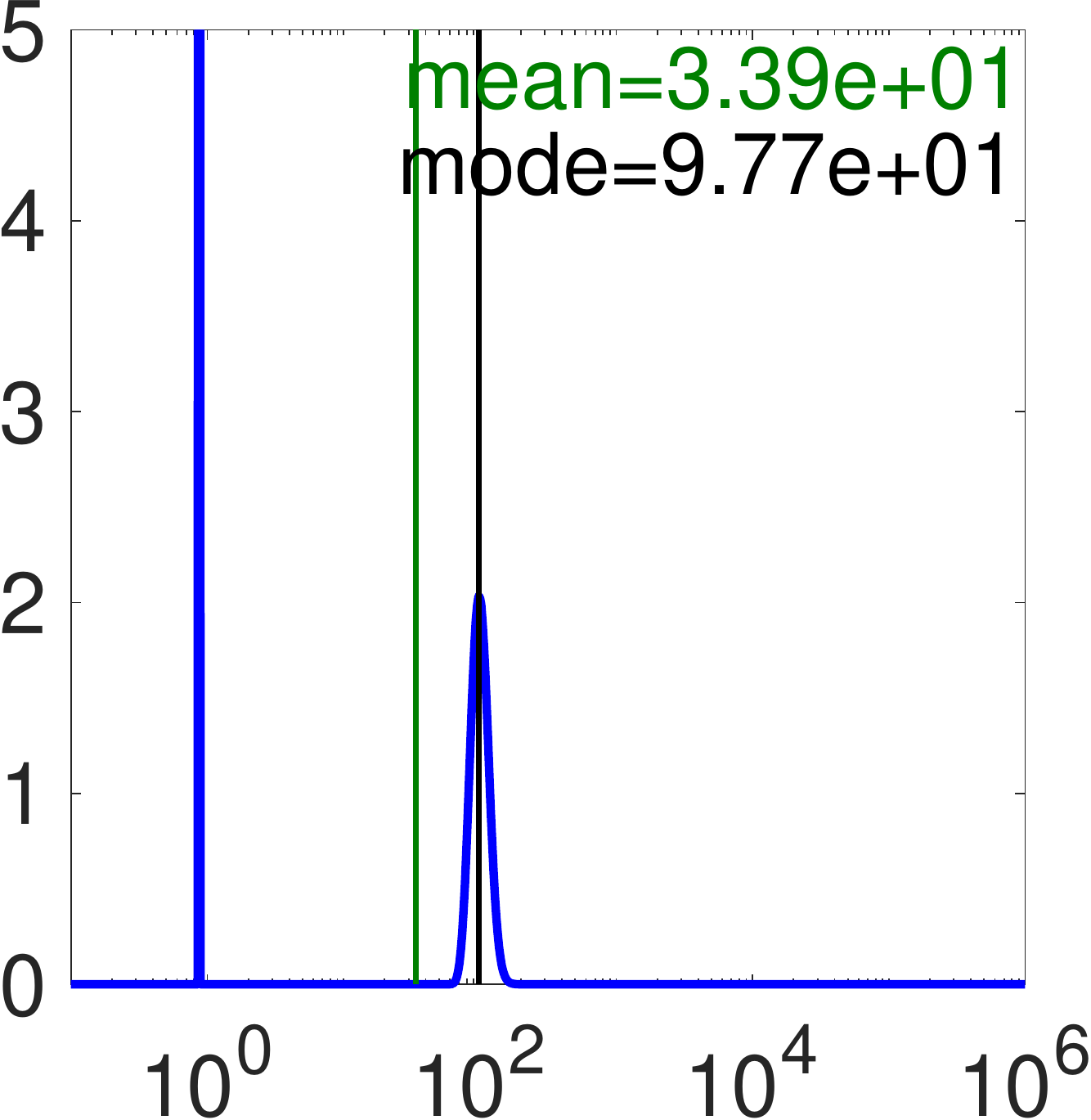}\\
\begin{tabular}[x]{@{}l@{}}Training\\results\end{tabular}&\includegraphics[scale=0.48,valign=c]{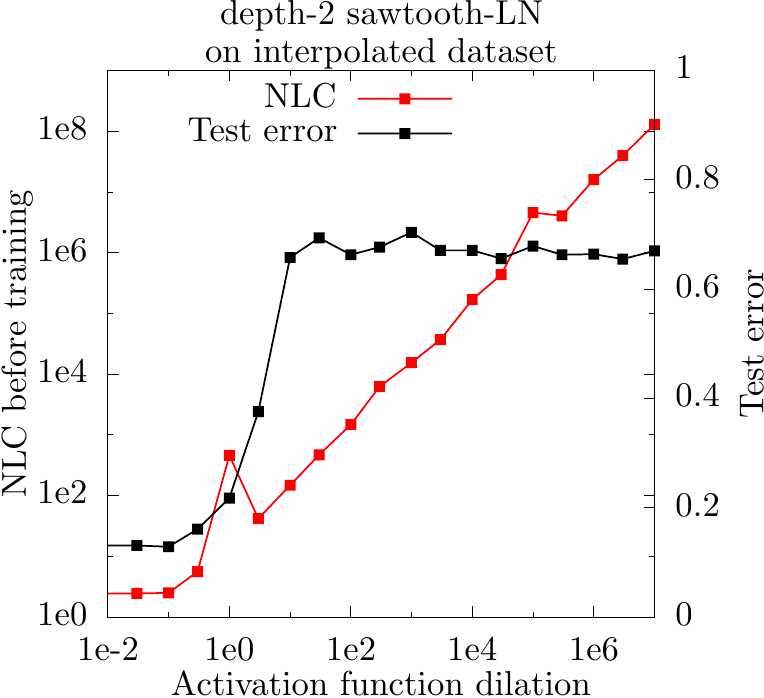}&\includegraphics[scale=0.48,valign=c]{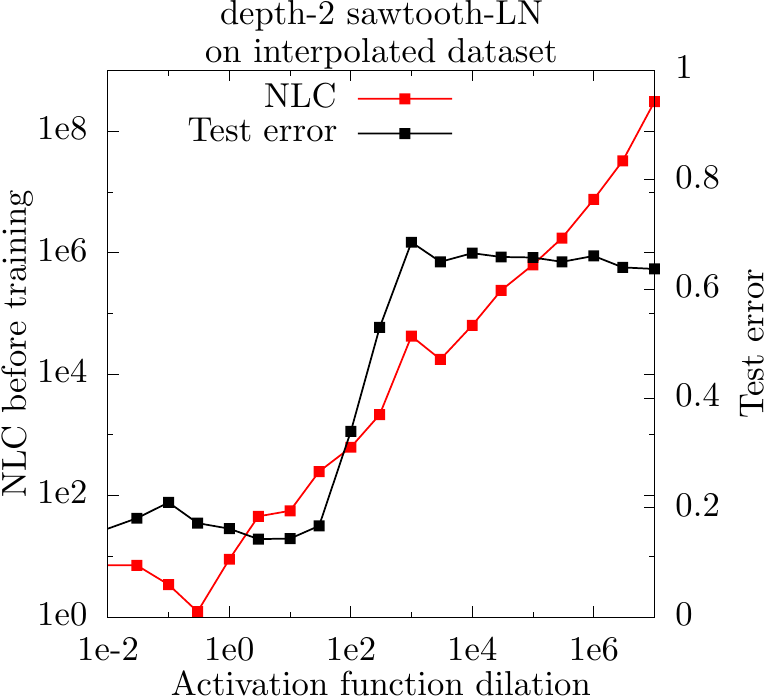}&\includegraphics[scale=0.48,valign=c]{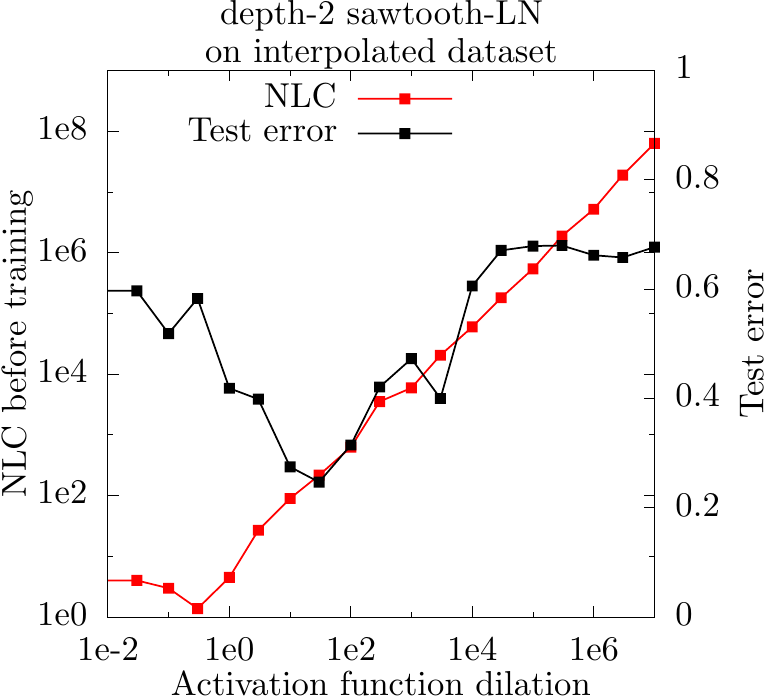}\\
\\
\end{tabular}
\\
\begin{tabular}{lccc}
Noise weight&0.999&0.9999&0.99999 \\ \hline\hline 
\\
\begin{tabular}[x]{@{}l@{}}PNLCD\\density\end{tabular}&\includegraphics[scale=0.27,valign=c]{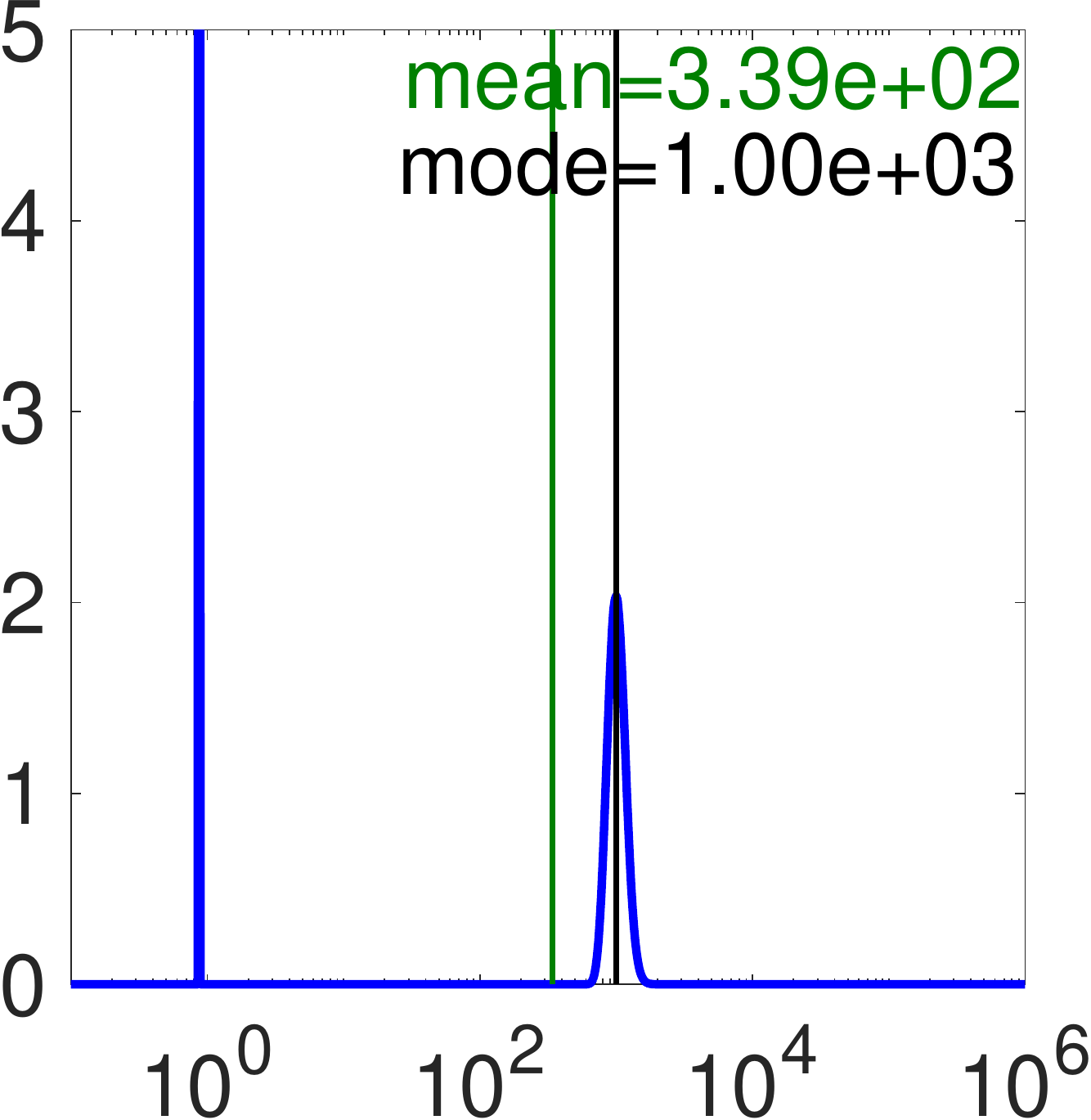}&\includegraphics[scale=0.27,valign=c]{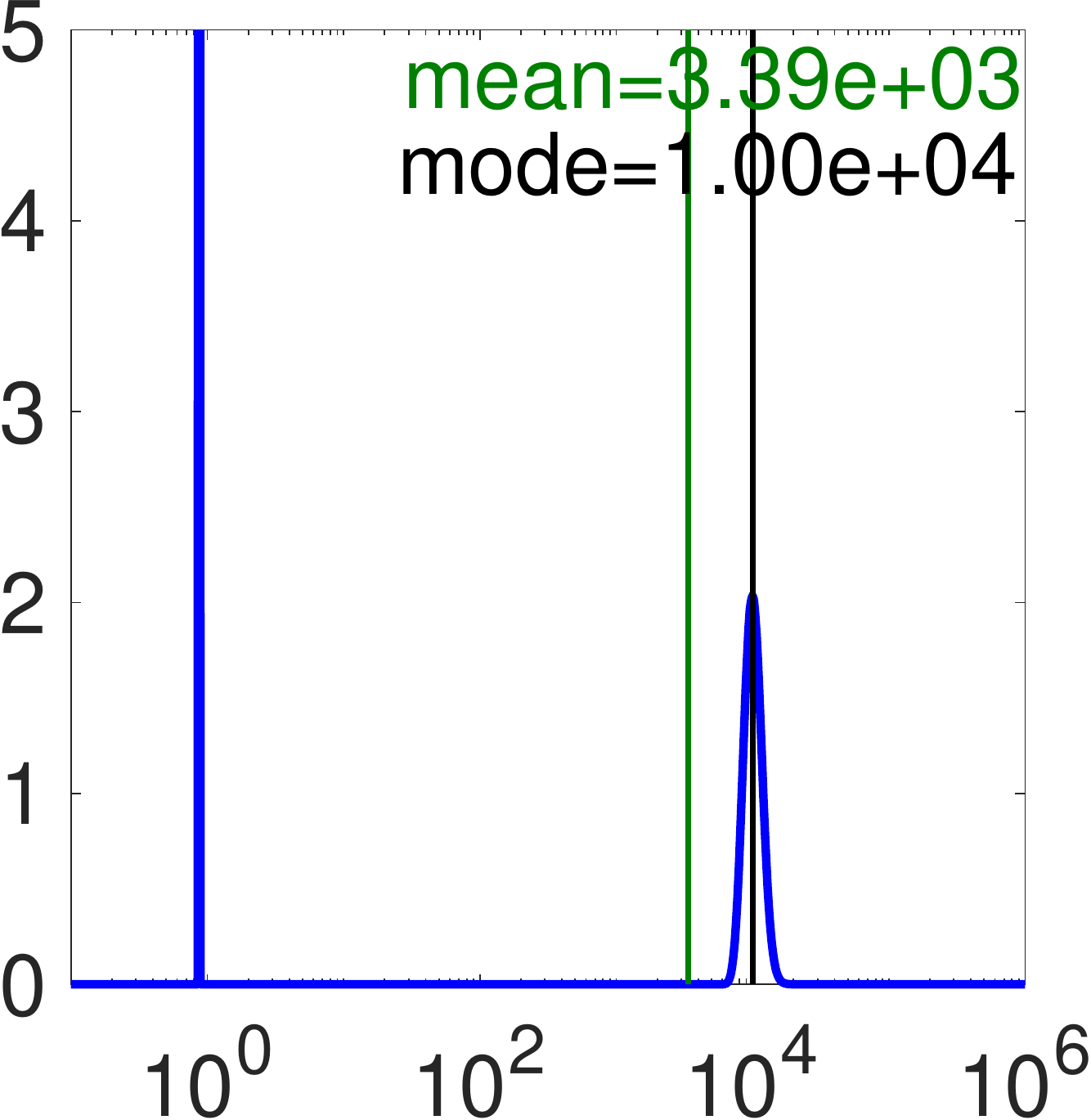}&\includegraphics[scale=0.27,valign=c]{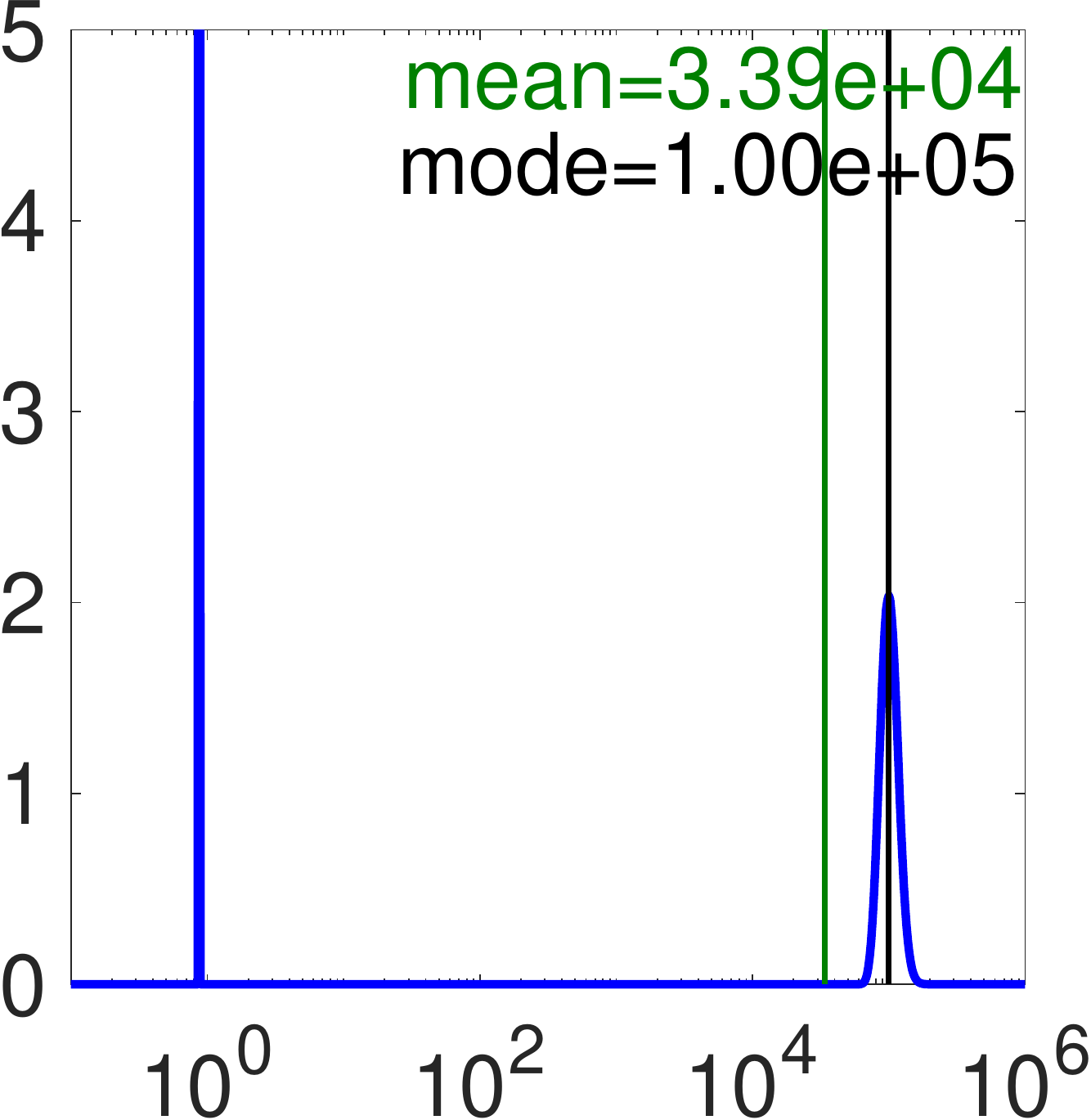}\\
\begin{tabular}[x]{@{}l@{}}Training\\results\end{tabular}&\includegraphics[scale=0.48,valign=c]{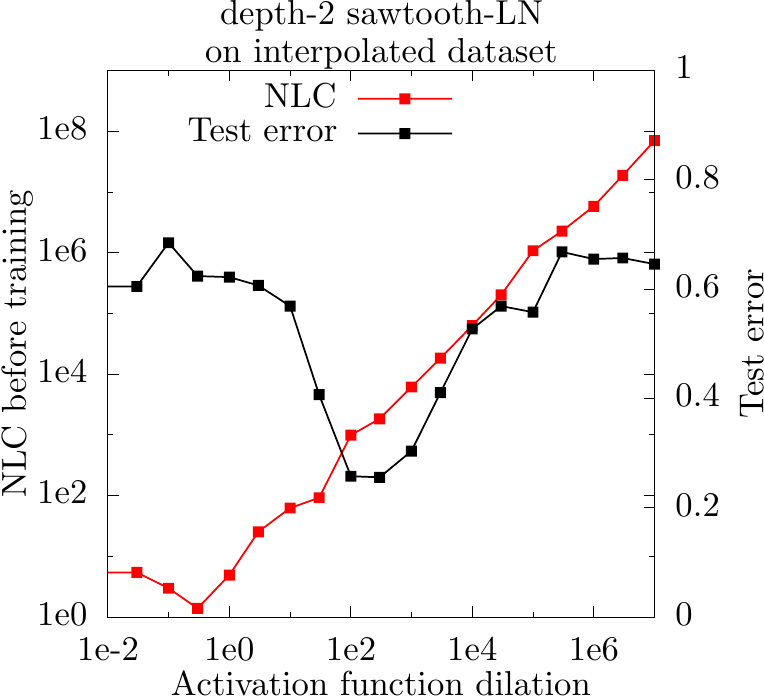}&\includegraphics[scale=0.48,valign=c]{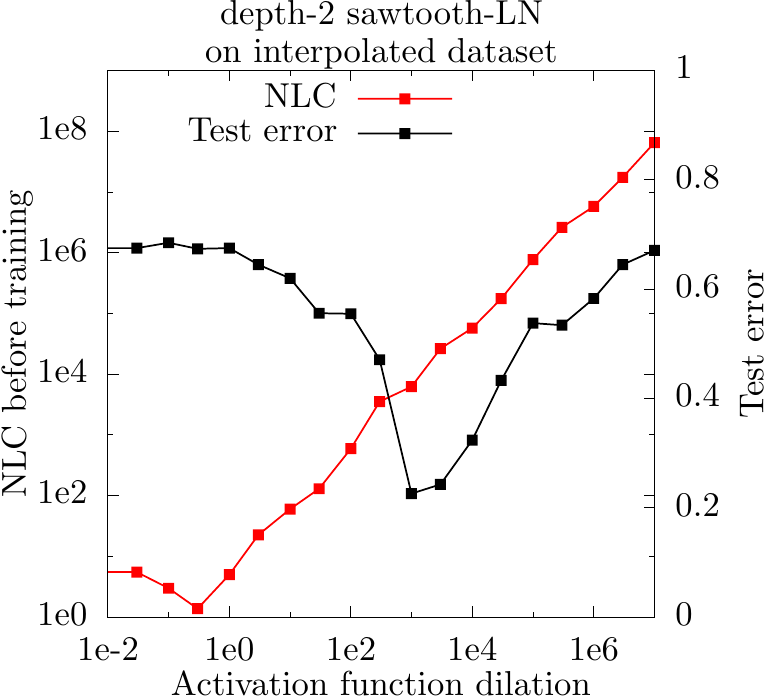}&\includegraphics[scale=0.48,valign=c]{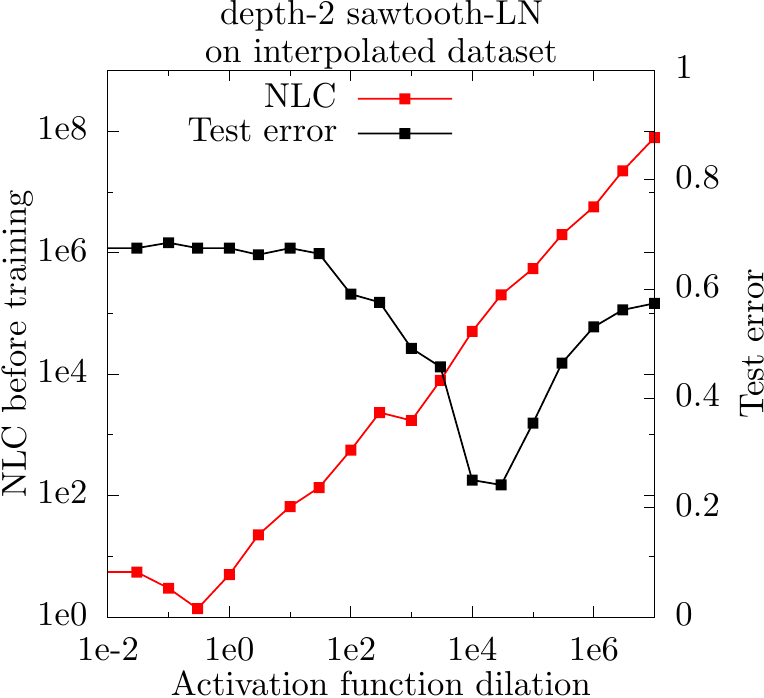}
\end{tabular}
\caption{Metric values for 6 datasets generated from interpolating waveform-noise inputs with 2 Gaussian noise vectors. Top rows: PNLCD on the union of training and validation set. Bottom rows: Initial NLC and test error of depth-2 fully-connected sawtooth-LN architectures with dilated activation functions. Note that PNLCD is zero for pairs of datapoints that share a label. We do not depict this in our graphs as we use log scale. Note that the mode we give for PNLCD ignores the peaks at $PNLCD=0$ and $PNLCD=1$. {\it Conclusion:} The noise weight used to interpolate waveform-noise inputs with the Gaussian noise vectors predicts the PNLCD peak above 1. The location of the peak predicts the NLC that leads to minimal test error.}
\label{dsetfusion}
}
\end{table}

\chapter{Mean field nonlinearity analysis} \label{meanFieldNnaChapter}

In chapter \ref{nlcChapter}, we showed {\it what} properties the NLC has. In this chapter, we further explain {\it why} it has those properties and {\it why} a given network has the NLC value that it has. We extend `mean field theory' to link an architecture's initial NLC and other metric values to its definition in terms of the layer graph. The highlight of this chapter is corollary \ref{mfntNPE}, which gives an embarrassingly simple and instructive formula, which we term the `nonlinearity path equation', for estimating and understanding an architecture's initial NLC without any computation, for fully-connected architectures like those in study A. Many of the properties that were shown empirically in chapter \ref{nlcChapter} are trivially explained by that formula. Another highlight is proposition \ref{mfntCNLC}, which shows the mean field limit of the NLC is the first-order approximation of the bandwidth of the covariance kernel. This is the most important of our results that establish the NLC as a primary measure of model complexity in deep learning.

Mean field theory studies neural architectures in their randomly initialized state, as the width of layers converges to infinity. In a nutshell, we specify architectures where the width of some or all layers is undetermined. The definition of individual layers, along with the random initialization scheme of their parameter sub-vectors, is then specified as a function of the width of that layer as well as the widths of the dependencies of that layer. We can then consider the value of metrics such as the NLC. For each layer width configuration, we obtain what is generally a distribution over values for that metric induced by the parameter initialization scheme and / or input distribution. The payoff comes when the layer widths jointly converge to infinity. For many important metrics, their distribution will then converge almost surely to a deterministic value that only depends on the architecture definition in an instructive manner. When those metrics are then computed on practical architectures with fixed, finite layer widths, their values are often close to these theoretical values obtained from the large-width limit.

Our core theoretical innovation is to extend mean field theory to cover not just distributions of layer values, but {\it meta-}distributions, which are obtained by having input and parameter vary in two hierarchical stages. Specifically, we show that the value of a fully-connected layer is {\it meta-Gaussian meta-distributed}, i.e. the distribution of each neuron as induced by the data is Gaussian and has a Gaussian random mean that is induced by the parameter initialization scheme. While this does require stronger assumptions than previous results in mean field theory, we demonstrate the approximate empirical validity of these assumptions and the empirical predictiveness of the resulting theory in painstaking detail. Another important contribution is that we document not just the successes, but the {\it failures} of mean field theory. Empirical validation of mean field studies is almost always conducted using well-behaved architectures using the tanh or ReLU activation function, or one of its derivatives such as hard tanh (e.g. \citet{correlationLimit,meanFieldCNN,meanFieldKernelPerformance1}). In our opinion, this leads the generality of mean field theory to be somewhat implicitly overstated. Failure analysis gives rise to the notion of `Gaussian stability', which is as, if not more, important to architecture performance than nonlinearity.

While mean field theory bolsters and explains the results we have presented so far, it also lays the groundwork for results in later chapters. For example, it directly motivates nonlinearity normalization in chapter \ref{nlnormChapter}. It enables the estimation of LBIAS (section \ref{outputBiasSection}) from the architecture definition along with the NLC. It helps explain the popularity of many architecture designs in chapter \ref{surveyChapter}. It facilitates the comparison of the NLC to prior work in chapter \ref{relatedWorkChapter}.

\paragraph{Chapter overview} In section \ref{meanFieldBackgroundSection}, we give the most important prior results of mean field theory as they apply to this work. Unfortunately, this section is quite dense as mean field theory relies on many high-level abstractions. We attempt to minimize the number of direct references made to this section throughout the rest of the chapter. In section \ref{meanFieldDistributionSection}, we extend mean field theory to obtain the layer and output meta-distributions of an architecture in the infinite-width limit, which leads to NLC estimates and is a valuable theoretical innovation in its own right. In section \ref{meanFieldPracticalSection}, we give a theorem that specifies a list of very simple rules for calculating the limit of metrics like the NLC from only the layer operations and layer graph that define an architecture similar to those used in study A (section \ref{studyAArchitecturesSection}). In section \ref{meanFieldActivationSection}, we dive deeper into how the activation functions used in an architecture influence its mean field properties, which culminates in establishing the NLC as a measure of the kernel bandwidth of neural networks. In section \ref{nlcExplainSection}, we explain many of the properties of the NLC that we demonstrated empirically in chapter \ref{nlcChapter} with mean field theory. In section \ref{gaussianStabilityExplanationSection}, we document the phenomenon of Gaussian stability and discuss its implications for many aspects of this work and deep learning in general. Finally, in section \ref{meanFieldCNNsection}, we give an outlook of how to extend our results to convolutional architectures and provide experimental evidence that this is likely to work.

We give a more detailed overview of our mean field-related results, which roughly correspond to the results in this chapter, in section \ref{meanFieldSummarySection}. 

We prove theoretical results from this chapter in chapter \ref{mfntChapter}.

\paragraph{Background from prior chapters} We invite readers to review our core notation, terminology and conventions given in section \ref{notationSummarySection}, which will be used throughout this chapter. In some places in this chapter, we deviate slightly from the standards laid out specifically in section \ref{neuralNetworkNotationSection} with regards to neural architectures, but we always point out explicitly when this happens. We also sometimes overload notation, but we try to limit this to individual sections.

We use the empirical studies laid out in chapter \ref{empiricalStudiesChapter} for validation. We recommend reading at least summary section \ref{metricsSummarySection} before proceeding. Chapter \ref{nlcChapter} is largely not required as background, except for motivating the continued study of the NLC.

\paragraph{Technical considerations} The only technical assumption we make implicitly throughout this chapter is that all integrals implicit in the probabilistic operators we use (e.g. $\mathbb{E}$) are valid and finite whenever we apply them. This is analogous to assumption \ref{assumptionIntegrable} from section \ref{nlcDefinitionSection}. See section \ref{integrabilitySection} for details. While we sometimes assume e.g. differentiability and positive denominators for the purpose of discussion, we always make these kinds of assumptions explicit in the theoretical results of this chapter.

\paragraph{Limitations} The biggest shortcoming of this chapter is that our main theorem \ref{mfntPropagation}, propositions \ref{mfntNPE} and \ref{mfntCNLC}, and hence some of the discussion in later parts of this chapter, are technically restricted to fully-connected architectures using popular building blocks and design strategies as we use in study A, where activation functions are also twice differentiable. Like almost all results in this work, we expect these results to not be fundamentally limited to fully-connected architectures on a conceptual level. Like most theoretical approaches to deep learning, mean field theory is a ``one building block at a time'' approach, where proofs have to be generalized incrementally over time. Nonetheless, the empirical predictiveness of mean field theory has been outstanding even on complex architectures (e.g. \citet{meanFieldNetsorGP}).

\section{Background} \label{meanFieldBackgroundSection}

Over the last few years, a range of papers have investigated mean field theory (e.g. \citet{correlationLimit,depthScalesMeanField,eigenspectrum, eigenspectrumGram,resNetMeanField,neuralTangentKernel, meanFieldCNN,meanFieldRNN,meanFieldDeepField,meanFieldGP, meanFieldBN,meanFieldCNNGP,meanFieldNetsorGP, meanFieldNetsorNTK,ntkDoesntWork,meanField1,meanField2, meanField3,meanField4,meanField5,meanFieldOrtho} and many more). \citet{meanFieldKernelPerformance1} provides a very long list of references. The popularity of this approach is perhaps exemplified by the existence of a mean field software framework \citep{meanFieldSoftware}. Recently, \citet{meanFieldNetsorGP}, \citet{meanFieldNetsorNTK} and \citet{meanFieldNetsorMatrix} presented what we consider are the most important insights of the theory in an instructive fashion. In this section, we will review primarily the results from \citet{meanFieldNetsorGP}, on which we base this chapter. Note that we use terminology and notation already established earlier in this work as opposed to notation and terminology used in \citet{meanFieldNetsorGP}. While it may be cumbersome for readers familiar with \citet{meanFieldNetsorGP} or the follow-up papers to adjust to new notation, we believe it would ultimately be more cumbersome to use two sets of notation and terminology in this work. We also do not present all results from \citet{meanFieldNetsorGP} in their full generality. We limit ourselves to what is relevant for this work to keep this background section manageable. (For example, we do not cover RNNs.)

\subsection{Mean field architectures} \label{meanFieldArchitectureSection}

\citet{meanFieldNetsorGP,meanFieldNetsorNTK,meanFieldNetsorMatrix} present their theory for architectures having specific properties, which we introduce in this subsection. We term these architectures `mean field architectures'. In section \ref{meanFieldPracticalSection}, we will go on to express practical architectures as mean field architectures.

Mean field architectures are built from only two layer operations. One operation is the fully-connected operation defined as in section \ref{layerTypesSection}. The other operation is the abstract elementwise operation.

\paragraph{Elementwise operation}

\begin{equation*}
f_l = \rho_l.(f_{\dot{k}_l[1]},f_{\dot{k}_l[2]},..,f_{\dot{k}_l[\dot{K}_l]},\mathbb{E}_if_{\hat{k}_l[1]}[i],\mathbb{E}_if_{\hat{k}_l[2]}[i],..,\mathbb{E}_if_{\hat{k}_l[\hat{K}_l]}[i])
\end{equation*}

Here, as throughout this chapter, each $i$ ranges over the width of the respective layer. This is a generalization of the activation operation from section \ref{layerTypesSection}. It can have multiple dependencies $f_{\dot{k}_l[1]}$ through $f_{\dot{k}_l[\dot{K}_l]}$ to which it is applied elementwise. We call these `elementwise dependencies'. It can also have additional dependencies $f_{\hat{k}_l[1]}$ through $f_{\hat{k}_l[\hat{K}_l]}$ from which it uses the layer mean. We call these `mean dependencies'. Hence, $\rho_l$ is a function from $\mathbb{R}^{\dot{K}_l+\hat{K}_l}$ to $\mathbb{R}$. We term it `multi-activation function' and say it has `elementwise inputs' and `mean inputs'. The trainable parameter sub-vector of an elementwise layer is empty.

In their most general result, \citet{meanFieldNetsorGP} allow their elementwise operation, which they term `Nonlin+', to depend on general scalar functions of previous layers, not just layer means. However, \citet{meanFieldNetsorGP,meanFieldNetsorNTK,meanFieldNetsorMatrix}  focus largely on layer means. For ease of presentation, we focus entirely on layer means. \citet{meanFieldNetsorGP} also use the addition operation to build their architectures, which they term `LinComb'. As they point out, this operation can be absorbed into the elementwise operation without losing representational power. This is what we do.

In contrast to regular feedforward architectures, mean field architectures may have multiple input and / or output layers. Input layers do not have parents in the layer graph and do not have an operation. Output layers do not have children in the layer graph.

Mean field theory considers the properties of architectures as the width of layers converges to infinity. Since architectures generally have more than one layer, there are multiple ways in which this convergence can occur. While \citet{meanFieldNetsorGP} deals with arbitrary forms of convergence, for ease of presentation, we will focus on the case where the ratio of layer widths is constant as convergence occurs, similar to what is done in \citet{meanFieldNetsorNTK}. We denote the width of $f_l$ by $d_ld_\text{MF}$, where $d_l$ remains fixed as usual and $d_\text{MF}$ can be taken to infinity. Since the limit of infinite width is a purely theoretical construction to begin with, the exact style of convergence does not matter as long as practically predictive theorems can be proven with it.

To a significant degree, mean field theory focuses on the properties of architectures in the initial state where no training has been conducted. Of course, it is still necessary to initialize the parameter, which, in mean field architectures, is composed only of the weight matrices of fully-connected layers. Here, it is critical to use a fixed multiple of Gaussian LeCun initialization. For each fully-connected layer $f_l$, we choose a fixed constant $\sigma_l^2$ we term the `variance parameter' and initialize each entry of the weight matrix as a Gaussian of mean zero and variance $\frac{\sigma_l^2}{d_kd_\text{MF}}$, where a $k$ subscript indicates the dependency of layer $f_l$ as always. Another important aspect of mean field architectures is that they allow what is called `weight sharing'. Different fully-connected layers can use the same weight matrix, i.e. the set of all fully-connected layers is partitioned into subsets, which can be of size 1, and fully-connected layers in the same subset use the same weight matrix which is randomly initialized once for the subset. Of course, within each subset the $\sigma_l^2$, $d_l$ and $d_k$ values must be equal.

Finally, as in most of \citet{meanFieldNetsorGP,meanFieldNetsorNTK,meanFieldNetsorMatrix}, for ease of presentation, we make the assumption that elementwise dependencies of elementwise layers can only be fully-connected layers or input layers. This does not limit representational power, because any elementwise dependencies that are themselves elementwise layers could simply be recursively replaced by their own dependencies to eliminate such a configuration.

We now summarize our definition of mean field architectures.

\begin{definition}
We say a `mean field architecture' $f$ is a neural architecture similar to that defined in section \ref{neuralNetworkNotationSection} with the following properties. Properties marked with (*) differ from our ``regular architectures'' as defined in section \ref{neuralNetworkNotationSection}.

\begin{itemize}
\item The width of a layer $f_l$ is equal to $d_ld_\text{MF}$, where $d_l$ is fixed but $d_\text{MF}$ is variable. (*)
\item There can be multiple input layers and output layers. (*)
\item Each layer $f_l$ is either a fully-connected layer with initial weight variance $\mathcal{N}(0,\frac{\sigma_l^2}{d_ld_\text{MF}})$ for some `variance parameter' $\sigma_l^2$, or it is an elementwise layer where each elementwise dependency is either an input layer or a fully-connected layer.
\item Different fully-connected layers can share the same weight matrix, which is initialized once for all layers that use it. (*)
\end{itemize}

Our definition of `mean field architecture' mirrors the definition of `NETSOR program' from \citet{meanFieldNetsorGP,meanFieldNetsorNTK,meanFieldNetsorMatrix}.
\end{definition}

If the widths of input layers change as $d_\text{MF}$ changes, then the inputs must also change. Of course, practical inputs have fixed dimensionality. We show how to reconcile this in section \ref{covarianceKernelSection} with readin layers. For now, we use a surrogate input distribution.

\begin{definition}
We say a distribution $\mathsf{dist}$ over vectors $\chi$ of fixed dimensionality $d$ is `elementwise' if there exists a distribution $\mathcal{G}$ over scalars such that drawing a vector from $\mathsf{dist}$ is equivalent to drawing each component from $\mathcal{G}$ independently.

Further, we say a distribution $\mathsf{dist}$ over a tuple of $N$ vectors $(\chi^{(1)}, .., \chi^{(N)})$ of fixed dimensionality $d$ is `elementwise' if there exists an $N$-dimensional distribution $\mathcal{G}$ such that drawing $(\chi^{(1)}, .., \chi^{(N)})$ from $\mathsf{dist}$ is equivalent to drawing $(\chi^{(1)}[i], .., \chi^{(N)}[i])$ independently for all $0 \le i < d$ from $\mathcal{G}$.

In both cases, we term $\mathcal{G}$ the `generator distribution' or `generator' for short.

Finally, we say a distribution $\mathsf{dist}$ over a tuple of $N$ vectors $(\chi^{(1)}, .., \chi^{(N)})$ that do not necessarily have the same dimensionality is `product elementwise' if the set of vectors can be partitioned into (one or more) subsets of vectors of equal dimensionality such that $\mathsf{dist}$ is elementwise on each subset and the subsets are independent under $\mathsf{dist}$. We say the generator distribution of $\mathsf{dist}$ is the product of generator distributions of the subsets.
\end{definition}

Going forward, we use a product elementwise distribution as the surrogate input distribution for a mean field architecture. If the architecture has $N$ input layers, the product elementwise distribution must be over $N$ vectors. Each vector is assigned to one input layer. The generator of the surrogate input distribution is over $\mathbb{R}^N$. Now we see how this construction allows us to consider architectures of variable input dimensionality. We associate a mean field architecture with a generator $\mathcal{G}$. Then as $d_\text{MF}$ varies, we draw the input layers from the product elementwise distribution that corresponds to $\mathcal{G}$, $d_\text{MF}$ and the $d_l$ for the various inputs layers $f_l$. While the surrogate input distribution varies, $\mathcal{G}$ stays fixed. Note that under this construction, components of input layers are all assumed to be independent except components that share the same index $i$ and are in layers of equal dimensionality.

\subsection{Master theorem} \label{masterTheoremSection}

\citet{meanFieldNetsorGP} state (several versions of) a theorem which they term ``master theorem''. Indeed, this result is highly general and many core insights from mean field theory can be derived from it (as well as the master theorems of \citet{meanFieldNetsorNTK,meanFieldNetsorMatrix}) without further low-level analysis. Our work very much relies on this. In a nutshell, the master theorem states that as $d_\text{MF}$ converges to infinity, the metrics $\mathbb{E}_i f_l[i]$ and $\mathbb{E}_i f_l[i]f_m[i]$ with $d_l = d_m$ converge to deterministic values almost surely as parameter and surrogate input are random. We can then use these core metrics to build towards more complex metrics as we do in section \ref{meanFieldPracticalSection}.

We now state a few more definitions and then the master theorem.

\begin{definition} For any vector object, we use the term `mean' to refer to the unweighted average of its components. For example, the `layer mean' is the mean of layer components and the `input mean' is the mean of input components. Analogously, we say the `square mean' of a vector is the mean of its square components; and we say the `co-mean' of two vectors of equal dimensionality is the mean of the components of their elementwise product.
\end{definition}

As stated in section \ref{notationSummarySection}, we use the term ``mean'' to refer to finite averages, like the ones defined above, as well as mean parameters of Gaussian distributions. We use the term ``expectation'' for the $\mathbb{E}_{x\sim\mathcal{D}}$ and $\mathbb{E}_{(x,y)\sim\mathcal{D}}$ operations.

\begin{definition}
Let $f$ be a mean field architecture and $\mathcal{G}$ a generator distribution. Let $S$ be a set of elementwise layers in $f$ such that there exists a weight matrix such that for each member of $S$ there exists a fully-connected layer using that weight matrix and depending on that member. Let $A_S$ be the $|S| \times |S|$ matrix where each entry is the layer co-mean of the layers corresponding to the row and column indices of that entry. We say that $(f,\mathcal{G})$ has `rank stability' if the following statement holds for each possible $S$. If $A_S$ converges almost surely to some $A_S^\infty$ as $d_\text{MF} \rightarrow \infty$, then the probability that the rank of $A_S$ equals the rank of $A_S^\infty$ converges to 1 as $d_\text{MF} \rightarrow \infty$. The randomness of $A_S$ is induced jointly by $\theta$ and $\mathcal{G}$.
\end{definition}

\begin{table}
{
\centering
\begin{tabular}{lccc}
Value & $f_l$ & $f_m$ & Formula\\ \hline\hline
$\mathfrak{m}_l$& input & - & $\mathbb{E}\mathcal{G}[n]$ where $f_l$ is the $n$'th input layer\\
$\mathfrak{m}_l$& FC & - & $0$\\
$\mathfrak{m}_l$& elem & - & $\mathbb{E}_e\rho_l(e,\mathfrak{m})$\\
$\mathfrak{c}_{l;m}$& input & input & $\mathbb{E}\mathcal{G}[n]\mathcal{G}[n']$ where $f_l$ / $f_m$ is the $n$'th / $n'$'th input layer\\
$\mathfrak{c}_{l;m}$& input & FC & $0$ \\
$\mathfrak{c}_{l;m}$& FC & FC & $\sigma_l^2\mathfrak{c}_{k_lk_m}$ if $f_l$ and $f_m$ share their weight matrix else $0$\\
$\mathfrak{c}_{l;m}$& elem & elem & $\mathbb{E}_e\rho_l(e,\mathfrak{m})\rho_m(e,\mathfrak{m})$\\
$\mathfrak{c}_{l;m}$& elem & input & $\mathbb{E}_e\rho_l(e,\mathfrak{m})e_m$\\
$\mathfrak{c}_{l;m}$& elem & FC & $\mathbb{E}_e\rho_l(e,\mathfrak{m})e_m$\\
\end{tabular}
\\
}
\caption{Rules for calculating the mean field limit of layer means and co-means. $\rho_l(e,\mathfrak{m})$ is short for $\rho_l(e_{\dot{k}_l[1]}, .., e_{\dot{k}_l[\dot{K}_l]}, \mathfrak{m}_{\hat{k}_l[1]}, ..,  \mathfrak{m}_{\hat{k}_l[\hat{K}_l]})$. $\rho_m(e,\mathfrak{m})$ is an equivalent abbreviation. $(e_0, e_1, .., e_L)$ forms a multi-variate Gaussian vector with moments given by $\mathbb{E}e_l = \mathfrak{m}_l$ and $\mathbb{E}e_le_m = \mathfrak{c}_{l;m}$. $\mathfrak{c}_{l;m}$ is only defined when $d_l=d_m$. The $\mathfrak{m}$ values and moments of $e$ used in any given rule are obtained by recursion.}
\label{tableBackgroundPropagation}
\end{table}

\begin{definition}
Let $F : \mathbb{R}^{d_\text{in}} \rightarrow \mathbb{R}$ be an arbitrary function with scalar output. Let $\mathcal{C}$ be an arbitrary class of functions from $\mathbb{R}^{d_\text{in}}$ to $\mathbb{R}$. We say $F$ is `controlled' by $\mathcal{C}$ if there exists a function in $\mathcal{C}$ that is greater or equal to $|F|$ everywhere.
\end{definition}

\begin{definition}
Let a multi-activation function be denoted by $\rho(e,m)$, where $e \in \mathbb{R}^{\dot{K}}$ and $m \in \mathbb{R}^{\hat{K}}$ represent the elementwise and mean inputs respectively. We say $\rho$ is `parameter-controlled' at $m^* \in \mathbb{R}^{\hat{K}}$ by function class $\mathcal{C}$ if there exist functions $\rho^\text{elem} : \mathbb{R}^{\dot{K}} \rightarrow \mathbb{R}$ and $\rho^\text{mean} : \mathbb{R}^{\hat{K}} \rightarrow \mathbb{R}_{\ge 0} \cup \{\infty\}$ such that

\begin{itemize}
\item $\rho(e,m^*)$ is controlled by $\mathcal{C}$ as a function of $e$
\item $\rho^\text{mean}(m^*) = 0$
\item $\rho^\text{mean}$ is continuous at $m^*$
\item $\rho^\text{elem}$ is controlled by $\mathcal{C}$
\item $|\rho(e,m) - \rho(e,m^*)| \le \rho^\text{elem}(e)\rho^\text{mean}(m)$ for all $e$ and $m$, where the right-hand side may be infinite
\end{itemize}

\end{definition}

\begin{definition}
Let $\mathcal{C}^\text{E2}$ be the class of functions $\mathbb{R}^d \rightarrow \mathbb{R}$ of form $e^{c||\chi||_2^{2-c'}+c''}$ where $c$, $c'$ and $c''$ are positive constants with $c'\le 2$, $d$ is a positive integer, and $\chi \in \mathbb{R}^d$ is the function input.
\end{definition}

\begin{backgroundtheorem}[\citep{meanFieldNetsorGP}] \label{backgroundMaster}
Let $f$ be a mean field architecture and $\mathcal{G}$ a generator distribution. Assume:

\begin{itemize}
\item $(f,\mathcal{G})$ has rank stability.
\item $\mathcal{G}$ is a Gaussian which may have a singular or even zero covariance matrix.
\item All multi-activation functions in $f$ are parameter-controlled by $\mathcal{C}^\text{E2}$ at $(\mathfrak{m}_{\hat{k}[1]}, .., \mathfrak{m}_{\hat{k}[\hat{K}]})$.
\end{itemize}

Then for all $0 \le l,m \le L$ with $d_l = d_m$ we have

\begin{eqnarray*}
\lim_{d_\text{MF} \rightarrow \infty}  \mathbb{E}_if_l(x)[i] &=& \mathfrak{m}_l \text{ a.s.}\\
\lim_{d_\text{MF} \rightarrow \infty}  \mathbb{E}_if_l(x)[i]f_m(x)[i] &=& \mathfrak{c}_{l;m} \text{ a.s.}
\end{eqnarray*}

where the values of $\mathfrak{m}_l$ and $\mathfrak{c}_{l;m}$ are defined recursively as in table \ref{tableBackgroundPropagation}. a.s. stands for `almost surely'. The randomness is induced jointly by $\theta$ and $\mathcal{G}$.

\end{backgroundtheorem}

Note that the theorem itself recursively yields that every $A_S$ that arises in the definition of rank stability above indeed converges to some limit $A_S^\infty$ a.s. as defined by the corresponding $\mathfrak{c}_{l;m}$ values. Also, the theorem itself recursively yields the $\mathfrak{m}$ values used in the parameter-control condition. In \citet{meanFieldNetsorGP}, this version of the master theorem is called theorem C.11, the ``Self-Parameterized NETSOR+ Master Theorem''.

\subsection{Fully-connected layers are Gaussian distributed} \label{meanFieldGPsection}

Examining table \ref{tableBackgroundPropagation}, we can informally describe forward propagation in an infinitely wide architecture as follows. The output of a fully-connected layer is independent of all other fully-connected layers that do not share its weight matrix. The outputs of sets of fully-connected layers that do share a weight matrix are jointly elementwise distributed. Each layer by itself is elementwise distributed. Each of these insights is powerful in its own right.

\citet{meanFieldNetsorGP} go on to formalize this informal description with a corollary based on the master theorem. To do this, they use a construct called `readout weights', which we translate as `readout layer'.

\begin{definition}
We say $f$ is a `mean field architecture with readout layers' if $f$ is a valid mean field architecture except that some or all output layers $f_l$ have fixed dimensionality $d_l$ which does not change as $d_\text{MF}$ varies, just like in our ``original architectures'' in section \ref{neuralNetworkNotationSection}. These output layers are called `readout layers'. Readout layers are fully-connected and may share weights with each other but not other fully-connected layers. A readout layer $f_l$ is Gaussian initialized with initial weight variance $\frac{\sigma_l^2}{d_kd_\text{MF}}$, where $\sigma_l^2$ is the variance parameter as before.  We write $\clipout(f)$ for the architecture obtained by removing the readout layers. (This may create new output layers.) If $f$ has exactly one readout layer, we call the value returned by it the `output' of $f$ and denote its dimensionality by $d_\text{out}$.
\end{definition}

\begin{backgroundcorollary}[\citep{meanFieldNetsorGP}] \label{backgroundMFGP}
Let $f$ be a mean field architecture with readout layers and $\mathcal{G}$ be a generator distribution. Assume the conditions of background theorem \ref{backgroundMaster} hold for $(\clipout(f), \mathcal{G})$. Then as $d_\text{MF}$ converges to infinity, the components of all readout layers jointly converge in distribution to a Gaussian. This Gaussian is product elementwise where the generator has mean zero and the covariance corresponding to layers $f_l$ and $f_m$ is $\mathfrak{c}_{l;m}$ as defined in table \ref{tableBackgroundPropagation} when $d_l=d_m$ and zero otherwise. The randomness is induced jointly by $\theta$ and $\mathcal{G}$.
\end{backgroundcorollary}

This corollary is explicitly concerned with the readout layers of $f$. Of course, we can also use it to obtain the Gaussianity of intermediate layers as well if we restrict the layer graph of the architecture to a sub-graph and redefine which layers are readout layers. The key requirement is that readout layers must use fresh weights. The formulation of the above corollary roughly corresponds to corollary C.13 in \citet{meanFieldNetsorGP}.

\subsection{The covariance kernel} \label{covarianceKernelSection}

So far, we have considered mean field architectures that can have multiple input and output layers and share weights. The reason for this is not that we are especially strongly interested in such architectures, but that it allows us to build ``trick architectures''. We can often prove a result for an architecture by transforming or extending that architecture and then applying background theorem \ref{backgroundMaster} or background corollary \ref{backgroundMFGP}. Such a transformation may add additional input or output layers. This process is a recurrent theme in prior work, in this chapter and especially chapter \ref{mfntChapter}.

The first example of this process is given in this subsection. Here, we use background corollary \ref{backgroundMFGP} to determine the outputs returned by architectures on fixed, finite-dimensional datasets. First, we introduce finite-dimensional input layers in addition to the finite-dimensional output layers we introduced in section \ref{meanFieldGPsection}.

\begin{definition}
We say $f$ is a `mean field architecture with finite input layers' if $f$ is a valid mean field architecture except that some or all input layers $f_l$ have a fixed dimensionality $d_l$ which does not change as $d_\text{MF}$ varies, just like in our ``original architectures'' in section \ref{neuralNetworkNotationSection}. We call these layers `finite input layers'. Layers that depend on finite input layers must be fully-connected layers. Those fully-connected layers are called `readin layers'. A readin layer $f_l$ is Gaussian initialized with initial weight variance $\frac{\sigma_l^2}{d_k}$, where $\sigma_l^2$ is the variance parameter as before. Readin layers may share weights with each other but not other fully-connected layers. We write $\clipin(f)$ for the architecture obtained by removing the finite input layers and turning the readin layers into input layers. If $f$ has exactly one finite input layer, we call the value $x$ assigned to it its `input' and denote its dimensionality by $d_\text{in}$. Finally, if $f$ has exactly one finite input layer and one readout layer, we write $f(x)$ for the value returned by the readout layer when $x$ is assigned to the finite input layer. A readout layer cannot be a readin layer.
\end{definition}

The way we build our trick architecture in this subsection is by duplicating the original architecture. The very long definition below formalizes architecture duplication where copies share weights.

\begin{definition}
Let $f$ be a mean field architecture that may have readout layers. We say $f^N$ is the `$N$-duplex' of $f$ if $f^N$ is composed of $N(L+1)$ layers which can be divided into $N$ groups of size $L+1$ such that there exists a map from the layers of $f^N$ to the layers of $f$ such that the following holds.

\begin{itemize}
\item Layers in a group can only depend on layers in the same group.
\item The map is a bijection when restricted to any group.
\item Each layer in $f^N$ has the same definition as its map image.
\item The dependencies of the map image of any layer in $f^N$ are the map images of the dependencies of that layer.
\item Fully-connected layers in $f^N$ share weights if and only if their map images are identical or share weights in $f$.
\end{itemize}

Further, let $\mathcal{G}$ be a generator distribution. Then we say $\mathcal{G}^N$ is the `$N$-duplex' of $\mathcal{G}$ if having the input layers of $f^N$ generated by $\mathcal{G}^N$ is equivalent to generating the input layers of $f$ by $\mathcal{G}$ and then assigning to each input layer in $f^N$ the vector that was assigned to its map image. (This leads to each layer group in $f^N$ having the same values assigned to its input layers as all other groups.)

Finally, let $f$ be a mean field architecture with one or more readin layers, and possibly readout layers, and $A$ some symmetric, positive semi-definite $N \times N$ matrix. Then we write $\mathcal{G}(A)$ for the generator that, when associated with the input layers of $\clipin(f)^N$ that stem from the readin layers in $f$ has the following properties. It is Gaussian, has mean zero and covariance matrix with entries 

$$\sigma_l\sigma_mA[n,n']\mathbbm{1}_\text{$l=m$ or $f_l$ and $f_m$ share weights}$$

$\mathcal{G}(A)$ has to generate $IN$ input layers, where $I$ is the number of readin layers in $f$. Hence, each row of the covariance matrix of $\mathcal{G}(A)$ corresponds to a readin layer $f_l$ and duplex index $n$ and each column also corresponds to a readin layer $f_m$ and duplex index $n'$. The entries of the covariance matrix are defined in terms of $l,m,n,n'$ above.

\end{definition}

Let $f$ be a mean field architecture with a single finite input layer and a single readout layer, and let $\mathcal{G}$ be a generator distribution associated with its non-finite input layers. Let $x^{(1)}, .., x^{(N)}$ be the inputs in a dataset $D$. Let $K_\text{in}$ be the $N \times N$ kernel matrix of the co-means of the inputs, i.e. $K_\text{in}[n,n'] = \mathbb{E}_{0 \le i < d_\text{in}} x^{(n)}[i]x^{(n')}[i]$. Consider the readin layers. Because all their weights are Gaussian, the joint distribution over all readin layer component values in response to all $N$ inputs is Gaussian. It is easy to check that this joint distribution is in fact product elementwise and generated by $\mathcal{G}(K_\text{in})$ as defined above, except that the duplex index is now replaced with the input index. Now consider $\clipin(f)^N$, the $N$-duplex of $\clipin(f)$. It has $N(I+I')$ input layers. $NI$ of these stem from the $I$ readin layers in $f$ and $NI'$ stem from the $I'$ non-finite input layers in $f$. For these input layers of $\clipin(f)^N$, consider the generator $\mathcal{G}(K_\text{in}) \times \mathcal{G}^N$, where the two factors are used for the two types of input layers. It is clear that the input layers of $\clipin(f)^N$ then have the same joint distribution as the non-finite input layers and readin layers of $f$ in response to the $N$ inputs, where again the duplex index is replaced by the input index, assuming that the non-finite input layers are fixed as the finite-dimensional input varies. Also note that $\mathcal{G}$ is independent of the weights in the readin layers of $f$. Hence, the output of $\clipin(f)^N$ also has the same distribution as the output of $f$ across the $N$ inputs. So background corollary \ref{backgroundMFGP} yields that the joint distribution of the $N$ output vectors returned by $f$ for the $N$ inputs converges in distribution to a Gaussian that is elementwise with a generator that has mean zero and some covariance $K_\text{out}$. Assuming that $f$ and $\mathcal{G}$ are fixed, applying table \ref{tableBackgroundPropagation} yields that the diagonal entries of $K_\text{out}$ are a function of only the corresponding entry of $K_\text{in}$, whereas the off-diagonal elements depend on the corresponding entry as well as the two diagonal entries in the same row / column of $K_\text{in}$.

\begin{backgroundcorollary} \label{backgroundKernel}
Let $f$ be a mean field architecture with a single finite input layer and a single readout layer, and let $\mathcal{G}$ be a generator distribution associated with its non-finite input layers. Let $x^{(1)}, .., x^{(N)}$ be the inputs in a dataset $D$ and let $K_\text{in}$ be the co-mean kernel matrix of these inputs. Assume $(\clipout(\clipin(f))^N, \mathcal{G}(K_\text{in}) \times \mathcal{G}^N)$ fulfills the conditions of background theorem \ref{backgroundMaster}. Then there exists a function $\mathfrak{C}$ such that the following holds. As $d_\text{MF}$ converges to infinity, $(f(x^{(1)}), .., f(x^{(N)}))$ converges in distribution to a Gaussian. This Gaussian is elementwise over output vectors where the generator has mean zero and covariance matrix $K_\text{out}$, where

$$K_\text{out}[n,n'] = \mathfrak{C}(K_\text{in}[n,n], K_\text{in}[n',n'], K_\text{in}[n,n']) \text{, } 1 \le n,n' \le N$$ 

$\mathfrak{C}$ is derived via table \ref{tableBackgroundPropagation} applied to $(\clipin(f)^N,\mathcal{G}(K_\text{in}) \times \mathcal{G}^N)$. The randomness is induced by $\theta$ and the surrogate input generated by $\mathcal{G}$, both of which are identical across inputs $x^{(n)}$.
\end{backgroundcorollary}

\begin{definition}
The `covariance kernel' $\mathfrak{C}(q,q',c)$ is the function that arises in the above corollary. We write $\mathfrak{C}(q,c)$ short for $\mathfrak{C}(q,q,c)$ and $\mathfrak{C}(c)$ short for $\mathfrak{C}(1,1,c)$. We also refer to both these shortened versions as the covariance kernel. Note that $\mathfrak{C}(q,q',c)$ is only valid when $q,q'\ge0$ and $|c| \le \sqrt{qq'}$.
\end{definition}

This corollary states that, in the limit, $\mathfrak{C}(q,q',c)$ completely determines the joint distribution of all outputs of $f$ on finite sets of inputs where randomness is induced by $\theta$ and $\mathcal{G}$. Further, we find that the scalar function $\mathfrak{C}(c)$ is sufficient when inputs are length-normalized. Throughout this chapter, we will focus on the case where inputs indeed have the same square mean for ease of presentation. Hence, we focus on $\mathfrak{C}(q,c)$ and $\mathfrak{C}(c)$. This is sufficient to study the majority of interesting behaviors that are observed in this work and in prior work. In practice, inputs are generally normalized in some way.

It is possible to generalize the above background corollary to multiple finite input and readout layers. As throughout the remainder of this work, we focus on the single input-single output case. The background corollary can be considered a more formal version of corollary 5.6 in \citet{meanFieldNetsorGP}. The notion of mapping two input square means and an input co-mean to output square mean / co-mean is taken from e.g. \citet{neuralTangentKernel}. 

\subsection{Wide networks are Gaussian processes}

Given some `mean function' $\mu : \mathbb{R}^{d_\text{in}} \rightarrow \mathbb{R}$ and `covariance function' $\nu : \mathbb{R}^{d_\text{in}} \times \mathbb{R}^{d_\text{in}} \rightarrow \mathbb{R}$, a `Gaussian process', as also defined in \citet{meanFieldNetsorGP} is a stochastic process $\mathbb{R}^{d_\text{in}} \rightarrow \mathbb{R}$ that maps any finite set of inputs $x^{(1)}, .., x^{(N)}$ to outputs that are jointly Gaussian distributed, where the $n$'th component of the mean vector is $\mu(x^{(n)})$ and the $(n,n')$'th entry of the covariance matrix is $\nu(x^{(n)}, x^{(n')})$. By background corollary \ref{backgroundKernel}, each output neuron of $f$ converges to a Gaussian process as $d_\text{MF}\rightarrow \infty$ in the sense that for any finite set of inputs $x^{(1)}, .., x^{(N)}$, the joint distribution of $(f(x^{(1)})[i], .., f(x^{(N)})[i])$ converges in distribution to a Gaussian where the mean vector is zero and the $(n,n')$'th entry of the covarience matrix is 

{\fontsize{9.7}{1} $$\nu_f(x^{(n)},x^{(n')}) = \mathfrak{C}(K_\text{in}[n,n], K_\text{in}[n',n'], K_\text{in}[n,n']) = \mathfrak{C}(\mathbb{E}_ix^{(n)}[i]^2,\mathbb{E}_ix^{(n')}[i]^2,\mathbb{E}_ix^{(n)}[i]x^{(n')}[i])$$}

Formally, under the conditions of background corollary \ref{backgroundKernel}, each output neuron of $f$ converges in finite distribution to a Gaussian process. Since we can only ever apply $f$ to a finite set of inputs in practice, it is reasonable to say that ``wide networks are Gaussian processes'' \citep{meanFieldNetsorGP}. The covariance kernel is so named because it corresponds to the covariance function of a Gaussian process. The equivalence between wide networks and Gaussian processes was first described by \citet{nngpOrig1} and expanded upon by e.g. \citet{nngpOrig2,meanFieldGP,meanFieldCNNGP,meanFieldNetsorGP}.

\subsection{Neural tangent kernel} \label{meanFieldNTKsection}

So far, we have cast $\theta$ as a random variable. Hence, the Gaussian process equivalence holds only in the architecture's randomly initialized state. So, while the covariance kernel determines the initial state of wide mean field architectures, it does not necessarily predict the course of training.

In their seminal paper, \citet{neuralTangentKernel} showed that the `neural tangent kernel' actually determines the course of training in the infinite width limit, assuming SGD and infinitesimal step sizes are used. As defined in section \ref{nlcKernelSection}, we have $K_\text{NTK}(x,x') = \frac{df(\theta, x)}{d\theta}\frac{df(\theta, x')}{d\theta}^T$. While this is a $d_\text{out} \times d_\text{out}$ matrix, for mean field architectures with a readin and readout layer, it can be shown that its limit when $d_\text{MF} \rightarrow \infty$ is a scalar multiple of the identity matrix, where the multiple can be calculated similarly to background corollary \ref{backgroundKernel} from input square means and co-means. While \citet{meanFieldNetsorGP} focus on the Gaussian process equivalence, \citet{meanFieldNetsorNTK} focus on the NTK.

The theoretical power of the NTK rests on the observation that it does not change during training in the infinite width limit. \citet{meanFieldNTK1,meanFieldNTK2} study the evolution of the NTK in practical, finite-width architectures. The NTK has been generalized for training using natural gradient descent \citep{meanFieldOrtho}, weight decay \citep{meanFieldNTK3} and when labels are involved \citep{meanFieldNTK4}. Recently, a number of works have linked great width to the existence of global minima of deep learning optimization landscapes and their reachability with gradient-based training (e.g.  \citet{meanFieldLandscape1,meanFieldLandscape3}).

A flipside of the theoretical power of the NTK is that it is not as predictive for practical, finite-width architectures trained with practical step sizes as e.g. background theorem \ref{backgroundMaster}. NTK theory predicts that in the infinite width limit, only the last linear layer learns. This phenomenon is termed `lazy training'. However, practical architectures often exceed the test error obtained from lazy training, even at very high width \citep{ntkDoesntWork,ntkDoesntWork2,ntkDoesntWork3, meanFieldCNNGP,meanFieldKernelPerformance1}. While we investigate the NTK empirically in section \ref{nlcKernelSection}, we do not further consider it in this chapter.

\section{Mean field theory of meta-distributions} \label{meanFieldDistributionSection}

Throughout section \ref{meanFieldBackgroundSection}, we studied the distribution of layer and network quantities where randomness was induced by the parameter $\theta$ and the surrogate input for the non-finite input layers generated by $\mathcal{G}$. In section \ref{masterTheoremSection}, we studied layer means and co-means of mean field architectures. In section \ref{meanFieldGPsection}, we examined entire distributions of layer values. In section \ref{covarianceKernelSection}, we introduced a finite number of fixed inputs and examined the joint distribution of layer values across these inputs.

In this section, we extend mean field theory to cover not just distributions of layer values, but meta-distributions. Let $\mathcal{D}$ be an input distribution for a single finite input layer. For a fixed value of $\theta$ and surrogate input, $\mathcal{D}$ induces a distribution at each layer. If we now additionally allow $\theta$ and surrogate input to vary at random on top of this process, we obtain a `meta-distribution', i.e. a distribution over distributions over layer values. The first level of randomness stems from $x \sim \mathcal{D}$, and the second level of randomness stems from $\theta$ and surrogate input, which corresponds to the entire randomness from section \ref{meanFieldBackgroundSection}. As given in section \ref{notationSummarySection}, we use the term `layer distribution' to refer to the distribution over layer values at some layer $f_l$ induced by $\mathcal{D}$ for a fixed $\theta$ and surrogate input, i.e. $f_l(\mathcal{D})$. We use e.g. `output distribution' and `neuron distribution' similarly.

There are at least three reasons why investigating the meta-distribution is interesting. First, we want to derive the limit of the NLC, which is defined in terms of $\mathcal{D}$ for a fixed $\theta$. Hence, we need to understand how e.g. $f(\mathcal{D})$ varies as a distribution as $\theta$ varies. Second, we want to investigate batch normalization when batch size converges to infinity. While BN with a fixed, finite batch size is covered by \citet{meanFieldNetsorGP}, it is a rather cumbersome operation. When $|B| \rightarrow \infty$, we observe simple and nice behaviors for BN, similar to layer normalization when width converges to infinity. Third, it is instructive to understand the types of layer distributions we can expect for a fixed $\theta$. When we are dealing with a practical architecture in the initial state, we are dealing with a specific parameter value, not with parameter values that are drawn independently for each input. A practical layer distribution is a draw from the meta-distribution, and those draws can differ significantly from the distribution obtained when varying $x$ and $\theta$ jointly. 

Our strategy for deriving the meta-distribution is to discretize $\mathcal{D}$ and then apply background corollary \ref{backgroundKernel}. Specifically, instead of considering $\mathcal{D}$ directly, we consider the uniform discrete distribution over an independent sample of size $N$ drawn from $\mathcal{D}$. A draw from the meta-distribution of any other layer is then also a uniform discrete distribution over $N$ fixed values. We then take the limit of that ``discrete meta-distribution'' as $d_\text{MF}$ converges to infinity. Finally, we ``unify'' these limits across all possible integer values of $N$. We formalize this process in a way that is intuitive, suitable for our analysis and similar to the notion of convergence in finite distribution used by \citet{meanFieldNetsorGP}.

\begin{definition}
For any meta-distribution $\mathcal{M}$, let the `$N$-expansion' of $\mathcal{M}$ be the distribution over $N$ values that is equivalent to drawing a distribution from $\mathcal{M}$ and then drawing $N$ values from that distribution. Further, we say a sequence of meta-distributions $(\mathcal{M})$ `expansion-converges' to a meta-distribution $\mathcal{M}^\text{lim}$ if the sequence of $N$-expansions of the $\mathcal{M}$ converges in distribution to the $N$-expansion of $\mathcal{M}^\text{lim}$ for all $N\ge 1$.
\end{definition}

For any $N$, $\mathcal{D}$ induces a distribution over $N \times N$ input co-mean kernel matrices $K_\text{in}$ when the $x^{(n)}$ are drawn from $\mathcal{D}$. In the setup of background corollary \ref{backgroundKernel}, as $d_\text{MF} \rightarrow \infty$, $K_\text{in}$ is mapped deterministically to the $N \times N$ output co-mean matrix $K_\text{out}$ via the covariance kernel function, which then generates the elementwise Gaussian output. So in the limit, we can say that $\mathcal{D}$ induces a distribution over $K_\text{out}$. To prove a theorem based on expansion-convergence, we need two things. We need (i) a $\mathcal{D}$ that induces a ``nice and manageable'' distribution over $K_\text{out}$ for each $N$ and (ii) a meta-distribution that yields elementwise $N$-expansions with generators that are (potentially infinite) mixtures of Gaussians with mean zero, where the covariance matrix is distributed as $K_\text{out}$ in (i). It turns out that what we call the `elem-like' distribution and the `meta-Gaussian' meta-distribution fulfill (i) and (ii) respectively. We now provide several definitions and then the main theorem of this section.

\begin{definition}
We say a meta-distribution $\mathsf{metadist}$ over distributions $\mathsf{dist}_\text{vec}$ over vectors of dimensionality $d$ is `elementwise' if there exists a meta-distribution $\mathsf{metadist}_\text{gen}$ over distributions $\mathsf{dist}_\text{sca}$ over scalars such that drawing a value of $\mathsf{dist}_\text{vec}$ from $\mathsf{metadist}$ is equivalent to drawing $d$ values of $\mathsf{dist}_\text{sca}$ from $\mathsf{metadist}_\text{gen}$ and then taking the product of those $d$ values. $\mathsf{metadist}_\text{gen}$ is called the `generator meta-distribution' or `generator' for short.
\end{definition}

This mirrors the definition from section \ref{masterTheoremSection}. It implies that any distribution drawn from an elementwise meta-distribution has independent components. It also implies that the distribution over vectors obtained by drawing from a draw from the meta-distribution is elementwise. 

\begin{definition}
We say a meta-distribution over distributions over scalars is `meta-Gaussian' $\mathcal{MN}(q,c)$ with $q \ge c \ge 0$ if a distribution drawn from it is Gaussian with variance $q - c$ and mean drawn from another Gaussian with mean zero and variance $c$.
\end{definition}

\begin{definition}
We say an input distribution $\mathcal{D}$ is `elem-like$(q,c)$' with $q \ge c \ge 0$ if $x,x'\sim \mathcal{D}$ implies $\mathbb{E}_i x[i]^2 = \mathbb{E}_i x'[i]^2 = q$ and $\mathbb{E}_i x[i]x'[i] = c$ with probability 1. We write $K_\text{in}(N,q,c)$ for the $N \times N$ matrix with diagonal entries equal to $q$ and off-diagonal entries equal to $c$, which is the co-mean kernel matrix for a sample of size $N$ drawn from an elem-like$(q,c)$ input distribution $\mathcal{D}$ with probability 1.
\end{definition}

\begin{theorem} \label{mfntMetaGaussian}
Let $f$ be a mean field architecture with a single finite input layer and a single readout layer. Let $\mathcal{G}$ be a generator distribution associated with the non-finite input layers of $f$. Let $\mathcal{D}$ be an input distribution associated with the finite input layer of $f$. Assume:

\begin{itemize}
\item $\mathcal{D}$ is elem-like$(q,c)$.
\item $(\clipout(\clipin(f))^N, \mathcal{G}(K(N,q,c)) \times \mathcal{G}^N)$ has rank stability for all $N$.
\item $\mathcal{G}$ is Gaussian.
\item All multi-activation functions used in $f$ are parameter-controlled by $\mathcal{C}^\text{E2}$ at $(\mathfrak{m}_{\hat{k}[1]}, .., \mathfrak{m}_{\hat{k}[\hat{K}]})$.
\end{itemize}

Then as $d_\text{MF}$ converges to infinity, the meta-distribution of the readout layer expansion-converges to the elementwise meta-distribution with generator $\mathcal{MN}(\mathfrak{C}(q,q),\mathfrak{C}(q,c))$. $\mathfrak{C}$ and the $\mathfrak{m}_{\hat{k}[\hat{\kappa}]}$ are derived via table \ref{tableBackgroundPropagation} applied to $(\clipin(f)^N, \mathcal{G}(K(N,q,c)) \times \mathcal{G}^N)$. The first level of randomness is induced by $\mathcal{D}$. The second level of randomness is induced by $\theta$ and $\mathcal{G}$.
\end{theorem}

In practical terms, theorem \ref{mfntMetaGaussian} states that neuron distributions in fully-connected layers are Gaussians with a standard deviation that is fixed across the layer and an expectation drawn from another zero mean Gaussian that is induced by the parameter. While we can glean this practical meaning, the statement of the theorem is still somewhat opaque. Specifically, it may not be clear how to cast a regular, practical architecture as a mean field architecture and generator $\mathcal{G}$. In a nutshell, we must re-cast segments of the layer graph between successive fully-connected layers of the regular architecture as a single elementwise layer in the mean field architecture. Then, we must re-cast the trainable parameter sub-vectors in non-fully-connected layers as a surrogate input generated by $\mathcal{G}$. Hence, $\theta$ in a regular context maps onto $\theta$ and $\mathcal{G}$ in a mean field context. (Hence, we usually skip over $\mathcal{G}$ in our non-technical discussions.) We demonstrate how to do this in section \ref{meanFieldPracticalSection} and in greater detail in section \ref{mfntPropagationsection}.

The major condition for theorem \ref{mfntMetaGaussian} is that $\mathcal{D}$ is elem-like. We can immediately see that this condition cannot be achieved exactly by any input distribution over finite-dimensional vectors. For general $(q,c)$, for a set of $N$ vectors of dimensionality $d$ to have co-mean Kernel matrix $K(N,q,c)$, we must have $N \le d$! A strong requirement indeed. The elem-like condition is also not just a convenience. To see this, consider a $\mathcal{D}$ that is itself a discrete distribution over a finite set of values. If this is the case, the output distribution is clearly neither Gaussian nor elementwise, even in the limit $d_\text{MF} \rightarrow \infty$. In the limit, while the distribution of a neuron value in a fully-connected layer induced by $\theta$ for a fixed $x$ is Gaussian under very general conditions, the neuron distribution induced by $x \sim \mathcal{D}$ for a fixed $\theta$ is not. We further analyze the elem-like condition in section \ref{elemLikeSection}.

Finally, note that in our definition of `elem-like$(q,c)$', we require the $c$ parameter to be non-negative. This is because a negative $c$ would imply that every vector drawn from $\mathcal{D}$ is negatively correlated with every other vector drawn from $\mathcal{D}$, which doesn't even approximately make sense. $c$ also cannot be negative in proposition \ref{mfntElemLike}.

\subsection{Empirical analysis} \label{meanFieldDistributionEmpiricalSection}

Like for most concepts in this work, the ultimate justification for theorem \ref{mfntMetaGaussian} is its strong practical predictiveness, which we now demonstrate. Specifically, we investigate layer distributions of our fully-connected architectures from study A (section \ref{studyASection}). We also investigate layer distributions of 40 ``simple architectures'' which have been randomly initialized 100 times using different random seeds (section \ref{additionalExperimentsSection}). These architectures and initializations are the same as those used in e.g. section \ref{nlcRandomInitSection}.

For each architecture, we investigate a single layer that is roughly halfway between input and output layer, which is the same layer used as the surrogate input layer for the ``second half'' of the network in sections \ref{nlcSimpleMetricsSection}, \ref{nlcDecomposableSection} and \ref{widthInvarianceSection}. This layer is either a fully-connected layer in the case of non-residual architectures, or it is an addition layer in the case of residual architectures. Note that in our residual architectures, addition layers are effectively the sum of (possibly normalized) fully-connected layers. Therefore, we expect them to have the same meta-Gaussian properties as the layers they sum over. The reason for studying intermediate layers instead of output layers is that output layers in our architectures had width 10 or less, which makes it more difficult to assess e.g. whether neuron expectations in that layer appear to be Gaussian distributed across the layer. We denote the intermediate layer by $f_{\frac{1}{2}}$.

Our goal is to determine whether the layer meta-distribution is approximately the elementwise distribution that is generated by $\mathcal{MN}(\mathfrak{C}(q,q),\mathfrak{C}(q,c))$. We are not aware of a consensus protocol for empirically identifying a meta-distribution. However, it turns out that verifying that some $f_l$ indeed has the above meta-distribution is equivalent to verifying the following sub-properties.

\begin{enumerate}
\item For each $\theta$ and $i$, $f_l(\mathcal{D})[i]$ is Gaussian.
\item $\bar{f}_l$ is elementwise and generated by $\mathcal{N}(0,\mathfrak{C}(q,c))$ when $\theta$ is random.
\item For each $\theta$ and $i$, $\mathbb{S}_xf_l(x)[i] = \mathfrak{C}(q,q) - \mathfrak{C}(q,c)$.
\item For each $\theta$ and $i \neq i'$, $f_l(\mathcal{D})[i]$ is independent of $f_l(\mathcal{D})[i']$.
\end{enumerate}

Of course, these properties, which hold exactly in the limit, can only hold approximately and with high probability over $\theta$ in practice. $\mathbb{S}$ here is the standard deviation operator as defined in section \ref{metricEstimationSection}.

A convenient aspect of properties 1, 3 and 4 is that they are defined in terms of a fixed value of $\theta$. Hence, we can verify them on neural networks using fixed $\theta$. We do not have to compare different values of $\theta$. (In practice, we will use a trick to achieve the same thing for property 2.) Below, we investigate all four properties in turn. Note that we will defer some of the investigation to section \ref{meanFieldPracticalEmpiricalSection}, when we have derived rules for calculating $\mathfrak{C}(q,q)$ and $\mathfrak{C}(q,c)$ for study A architectures.

\begin{figure}[H]
\centering
\includegraphics[width=0.98\textwidth]{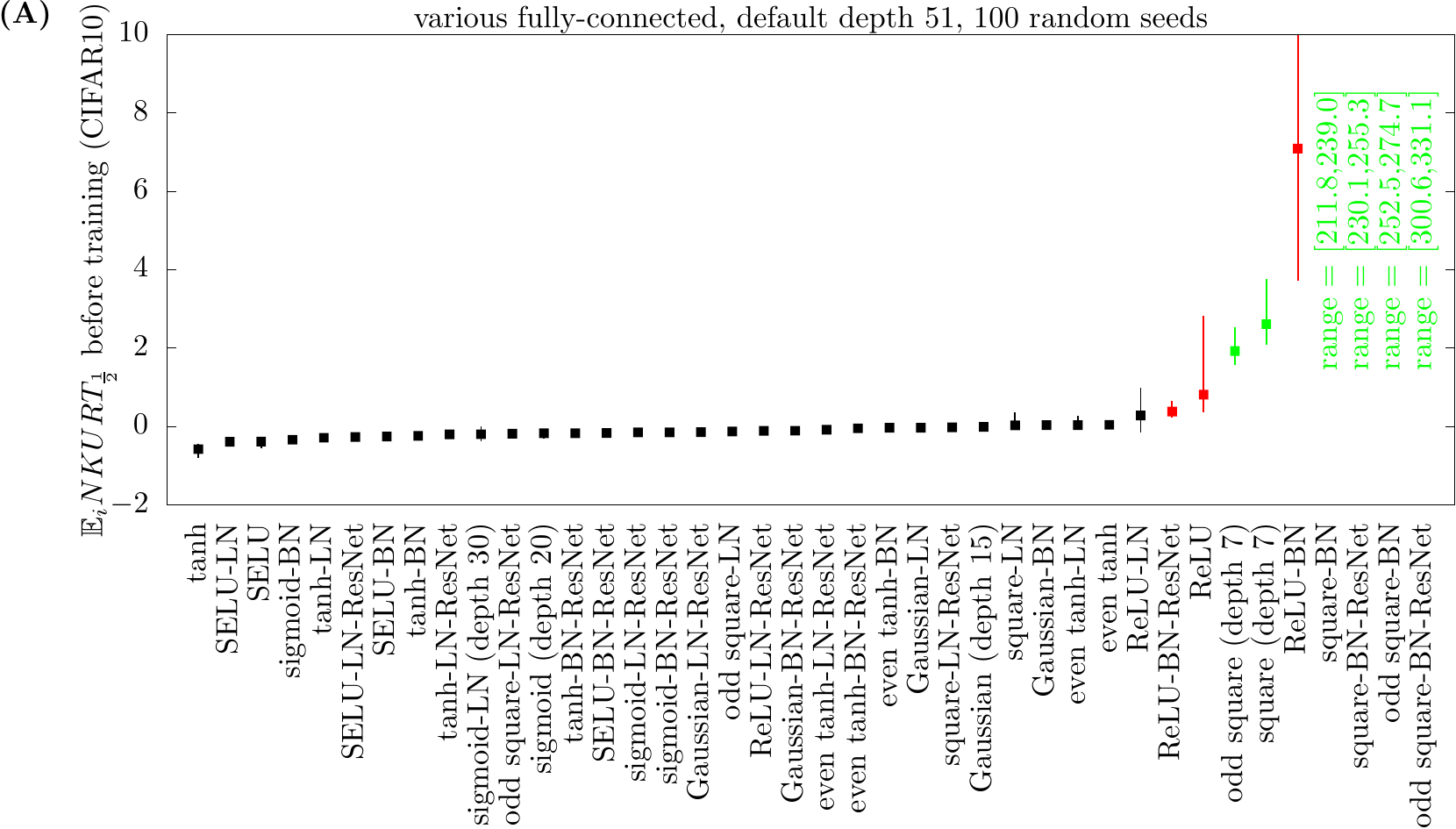}
\includegraphics[width=0.98\textwidth]{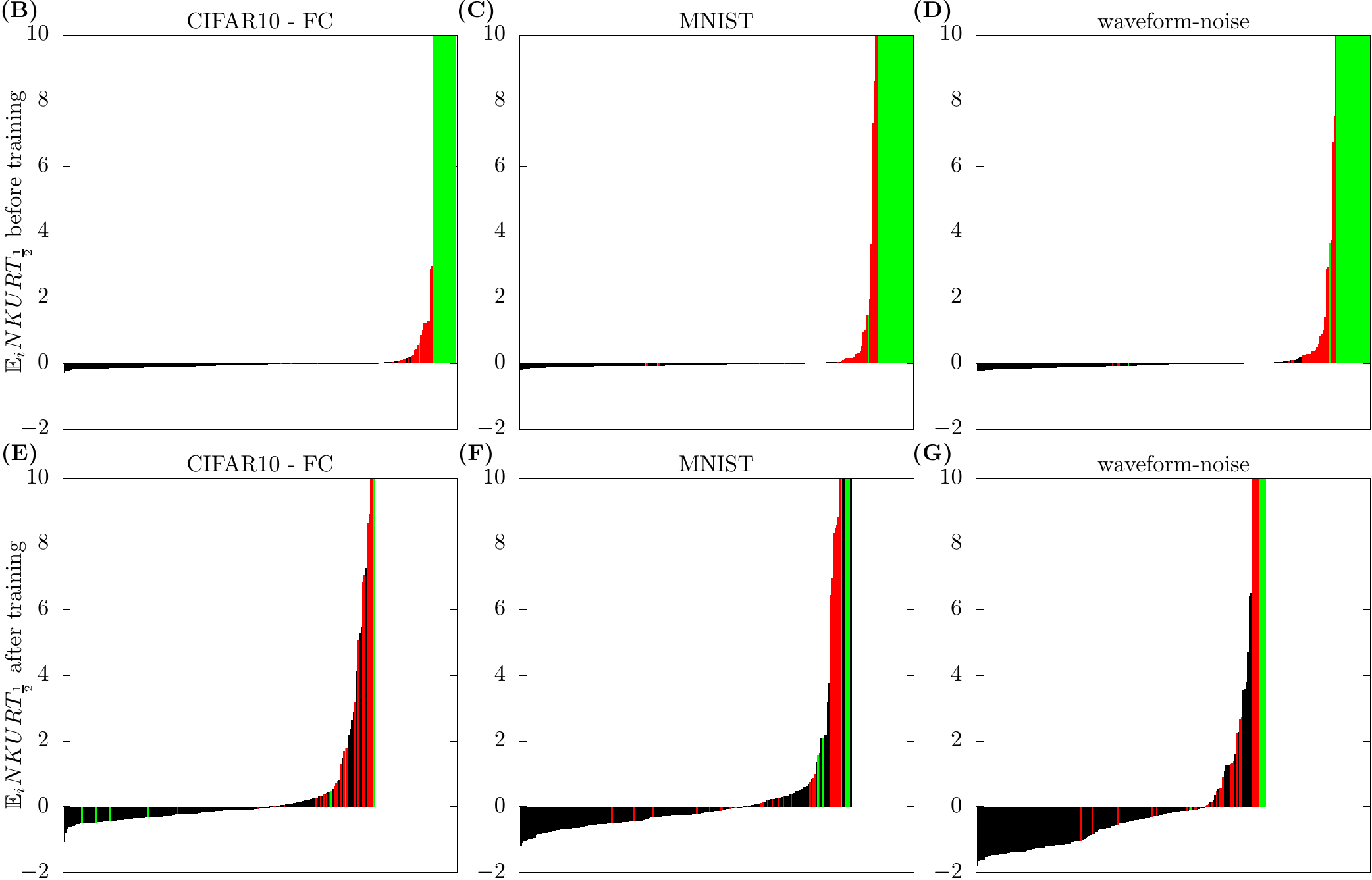}
\caption{NKURT averaged across an intermediate fully-connected or addition layer. In graph A, we depict the range across 100 random seeds and therefore 100 random initializations for 40 simple fully-connected architectures on CIFAR10 with a default depth of 51. The interval depicts the range of values across the random seeds and the filled square depicts the mean. In graphs B-G, we depict the value for study A architectures, before and after training. We place architectures on the x-axis in ascending order. Green markers correspond to GUAs and red markers correspond to GEAs. Some values fall outside the range of the y-axis. In graph A, we specify those values in the graph. {\it Conclusion:} Neuron distributions in FC layers of stable architectures in the initial state are approximately Gaussian with respect to excess kurtosis. This is not true for GUAs / GEAs.} \label{mfMetaGaussiankurt}
\end{figure}

\begin{figure}[H]
\centering
\includegraphics[width=0.98\textwidth]{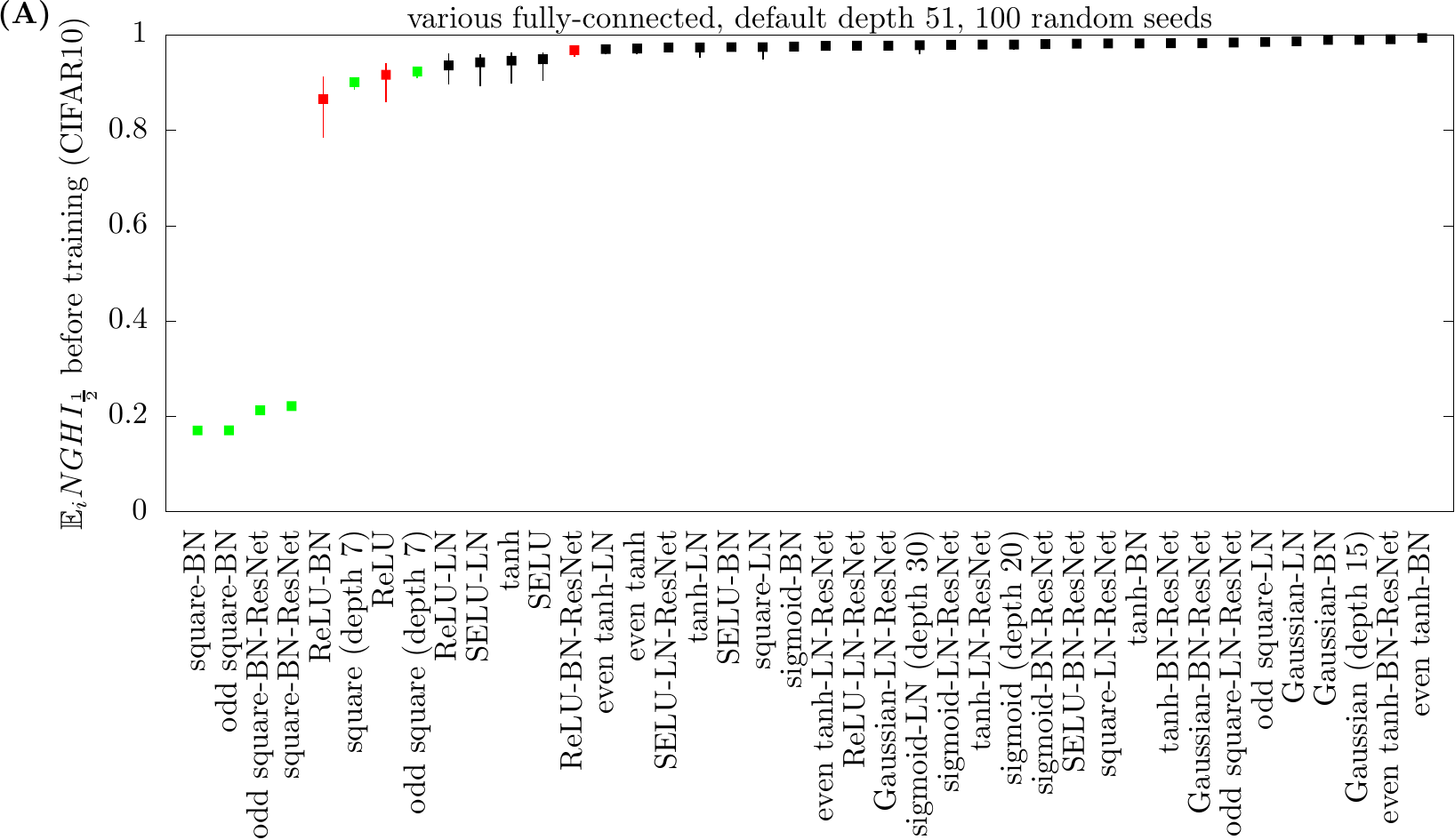}
\includegraphics[width=0.98\textwidth]{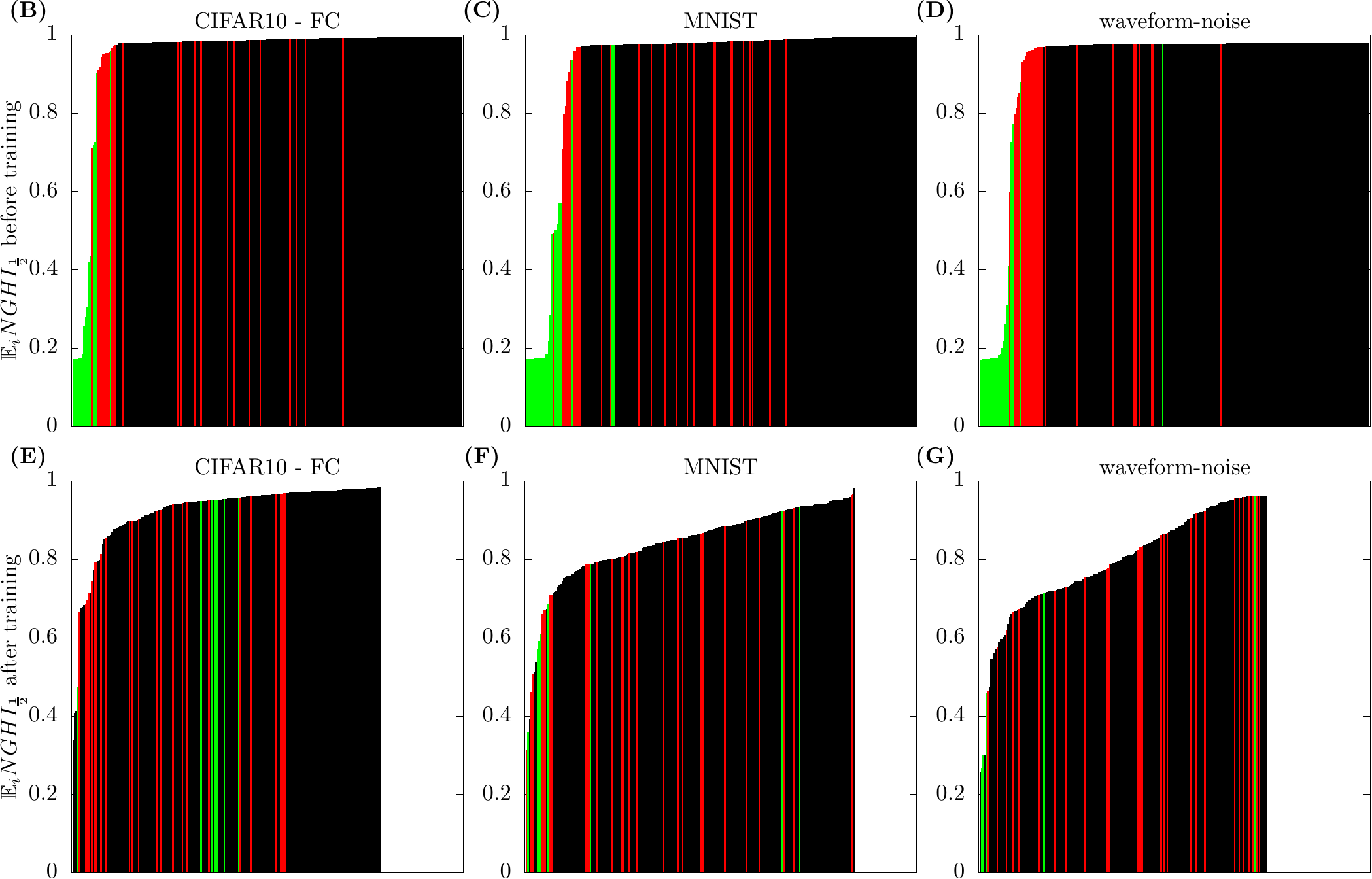}
\caption{NGHI averaged across an intermediate fully-connected or addition layer. Graphs are analogous to figure \ref{mfMetaGaussiankurt}. {\it Conclusion:} Neuron distributions in FC layers of stable architectures in the initial state are approximately Gaussian in their cumulative distribution function. This is not necessarily true for GEAs and especially GUAs.} \label{mfMetaGaussianiou}
\end{figure}

\begin{figure}[H]
\centering
\includegraphics[width=0.98\textwidth]{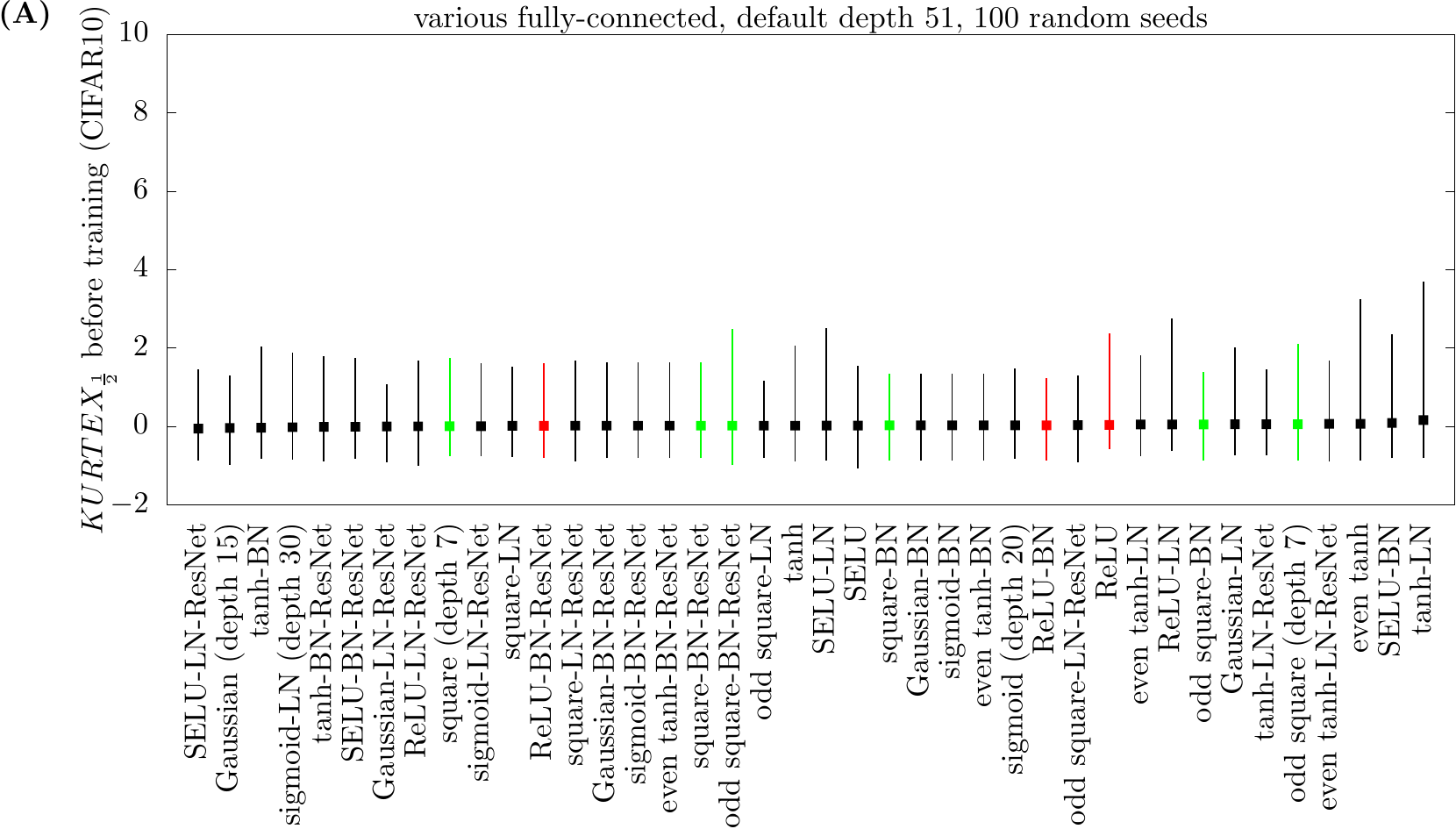}
\includegraphics[width=0.98\textwidth]{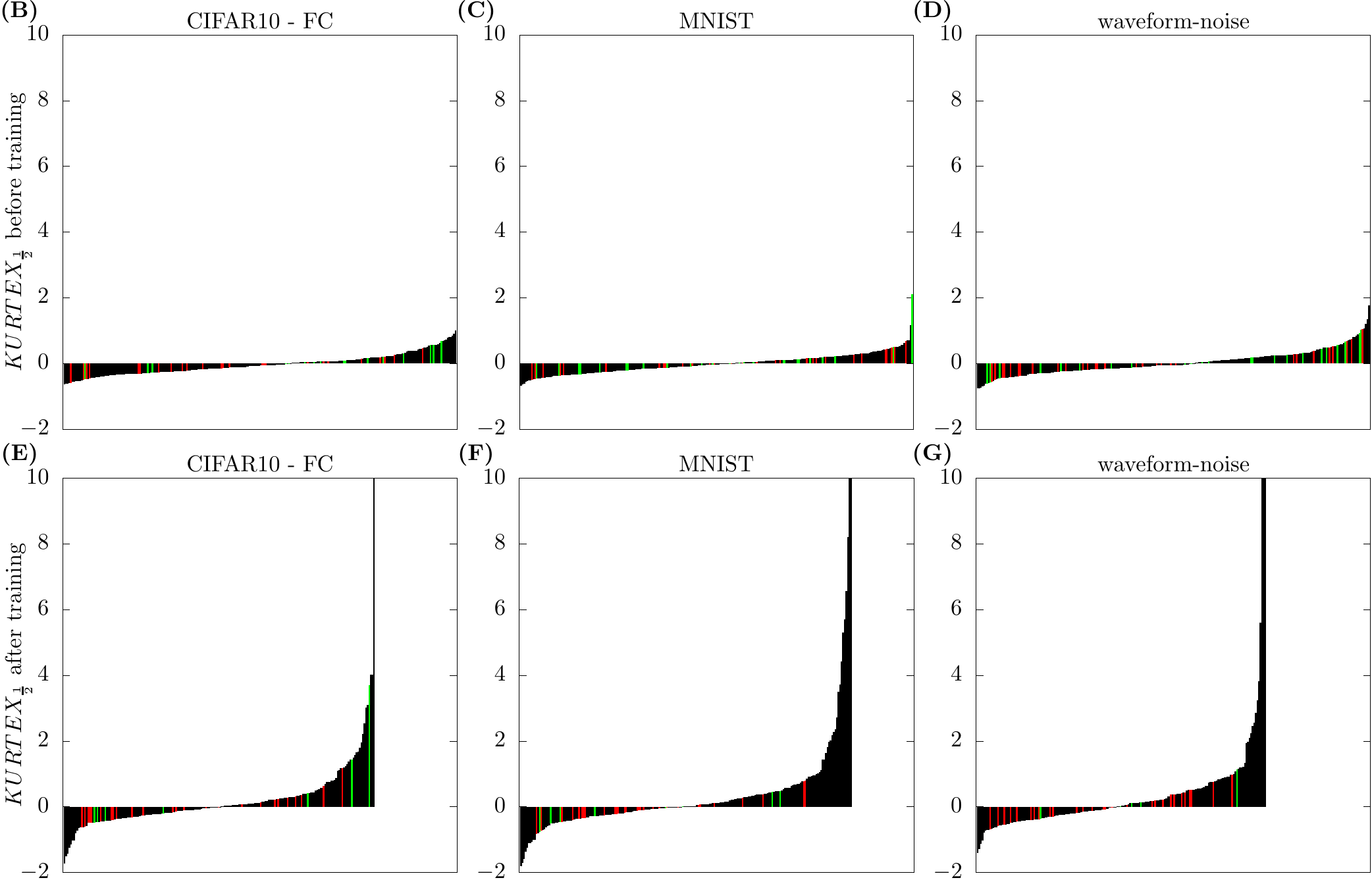}
\caption{KURTEX at an intermediate fully-connected or addition layer. Graphs are analogous to previous figures. A very small number of values fall outside the range of the y-axis in graphs E- G. {\it Conclusion:} Neuron expectations appear very Gaussian across an FC layer, especially before training.} \label{mfMetaGaussiankmu}
\end{figure}

\begin{figure}[H]
\centering
\includegraphics[width=0.98\textwidth]{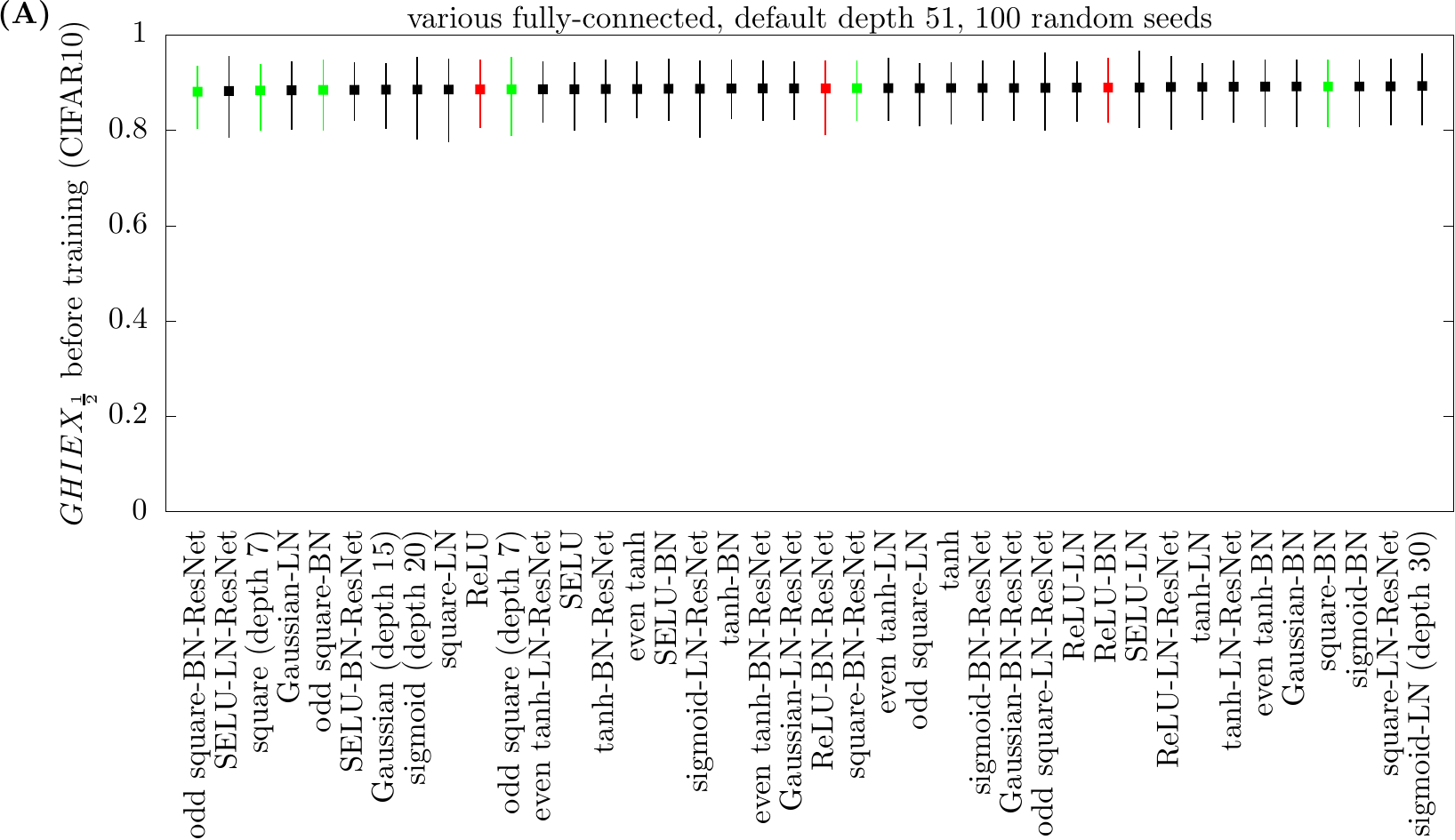}
\includegraphics[width=0.98\textwidth]{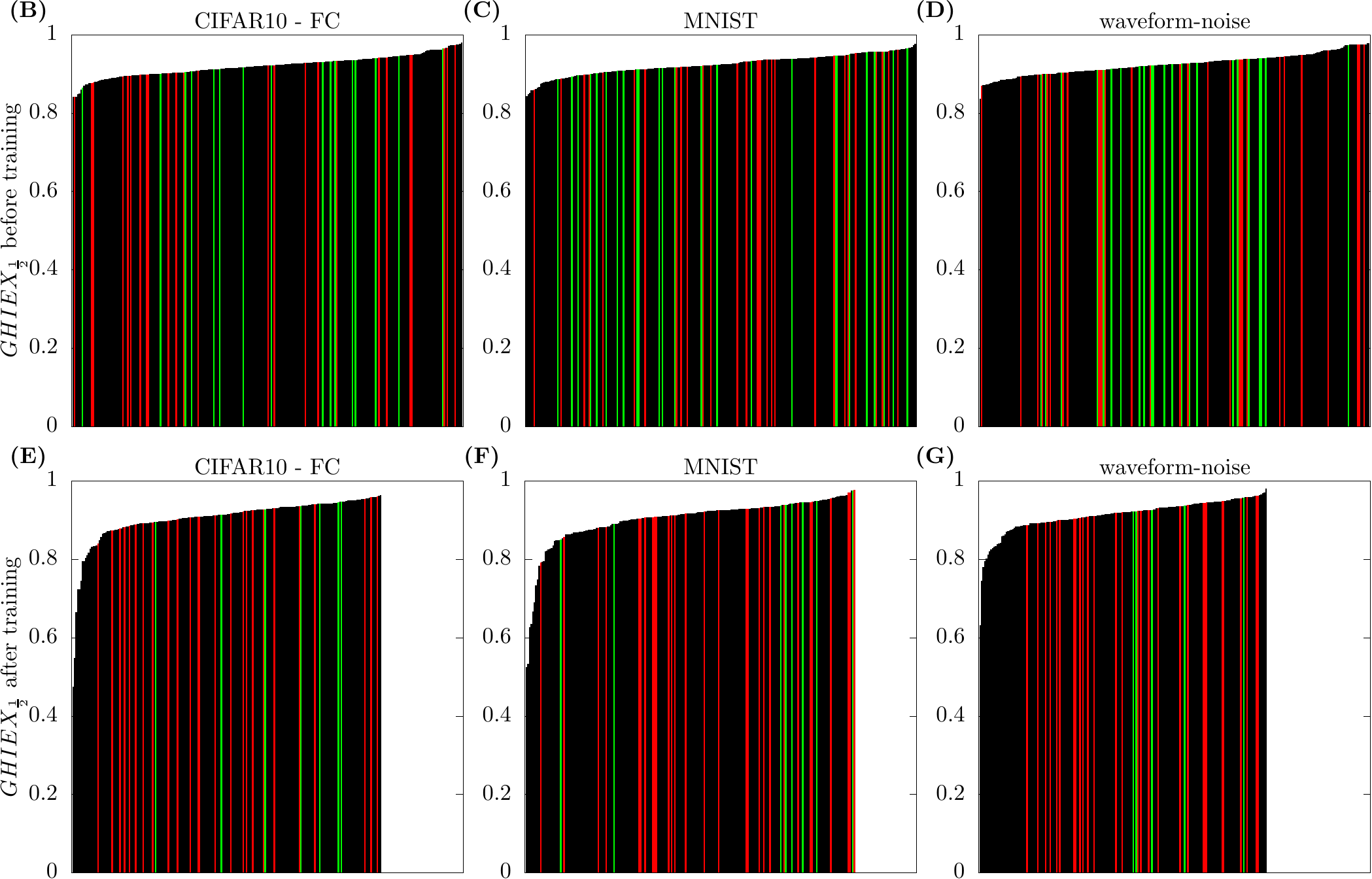}
\caption{GHIEX at an intermediate fully-connected or addition layer. Graphs are analogous to previous figures. {\it Conclusion:} Neuron expectations appear very Gaussian across an FC layer, especially before training.} \label{mfMetaGaussianimu}
\end{figure}

\begin{figure}[H]
\centering
\includegraphics[width=0.98\textwidth]{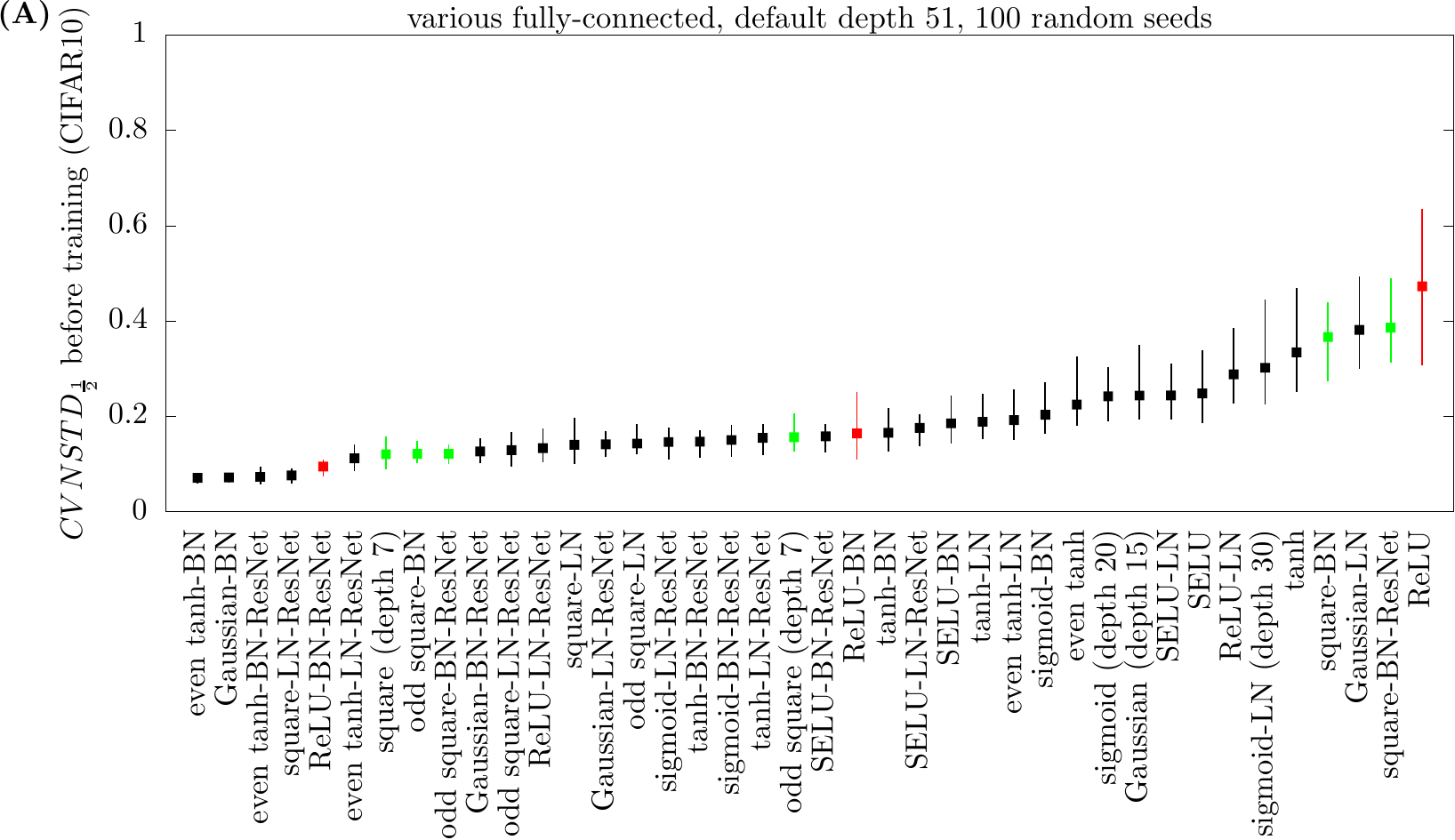}
\includegraphics[width=0.98\textwidth]{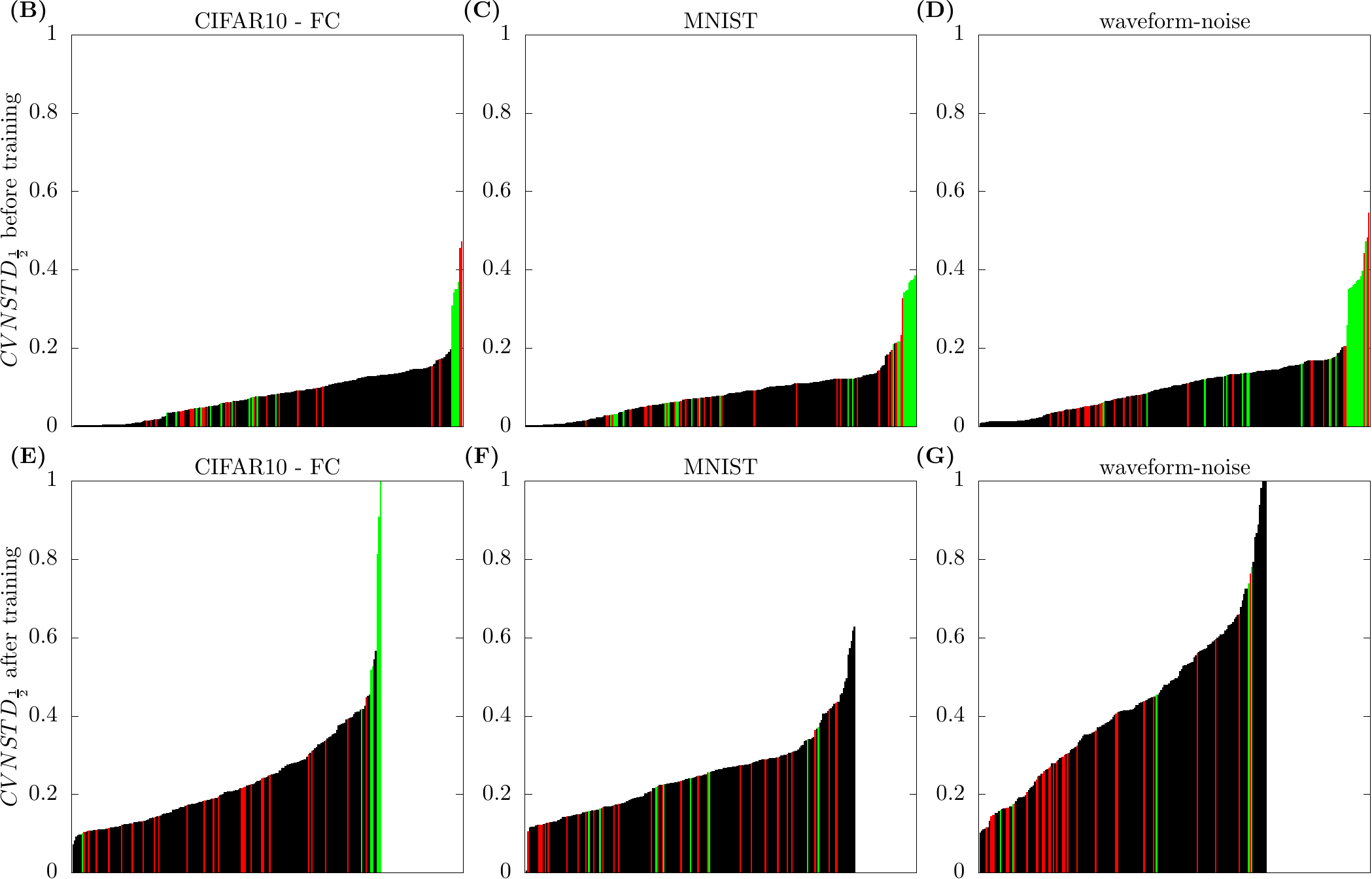}
\caption{CVNSTD at an intermediate fully-connected or addition layer. Graphs are analogous to previous figures. A very small number of values fall outside the range of the y-axis in graphs E and G. {\it Conclusion:} Standard deviations are relatively constant across an FC layer, especially for stable architectures before training.} \label{mfMetaGaussianstd}
\end{figure}

\begin{figure}[H]
\centering
\includegraphics[width=0.98\textwidth]{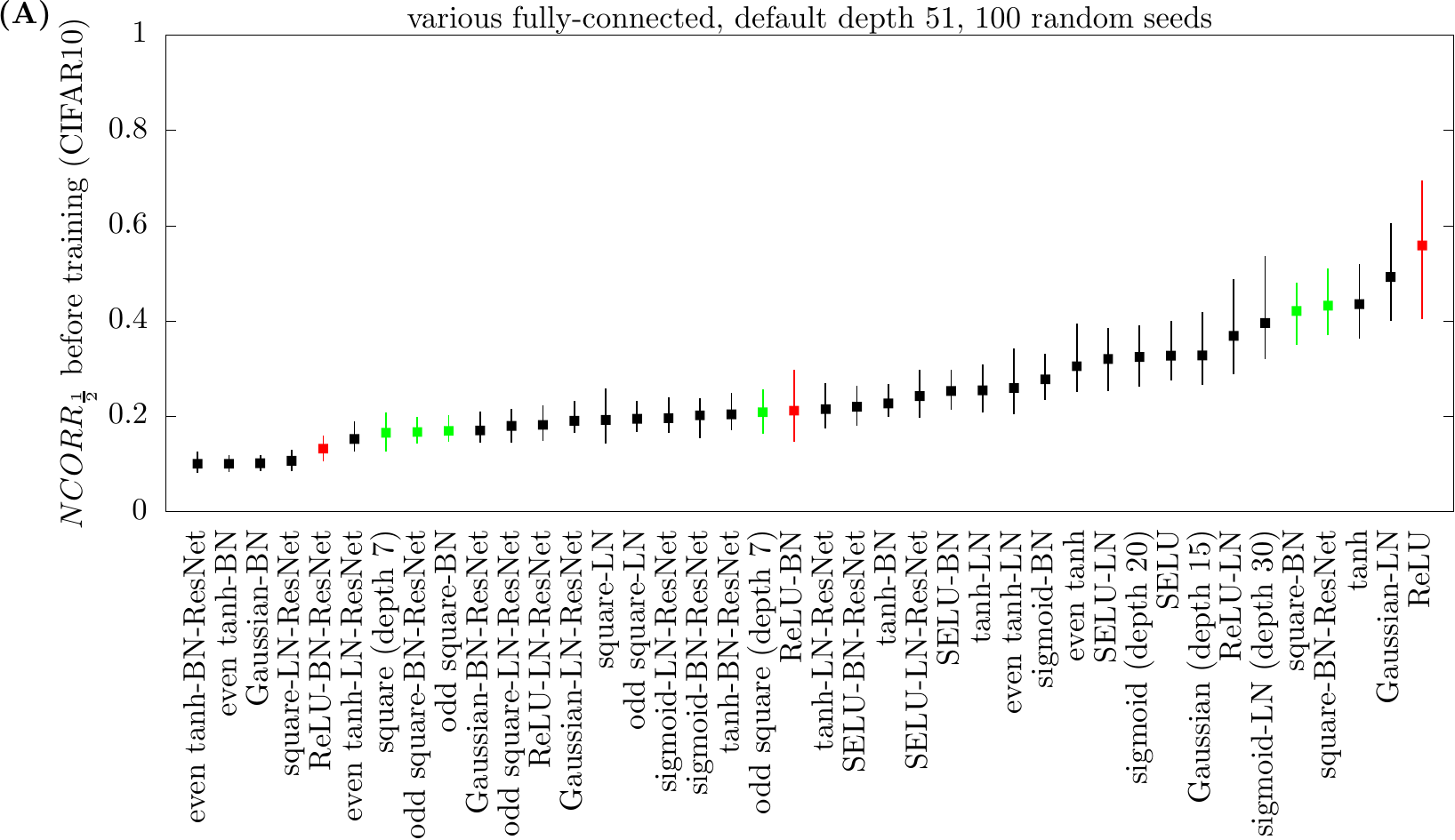}
\includegraphics[width=0.98\textwidth]{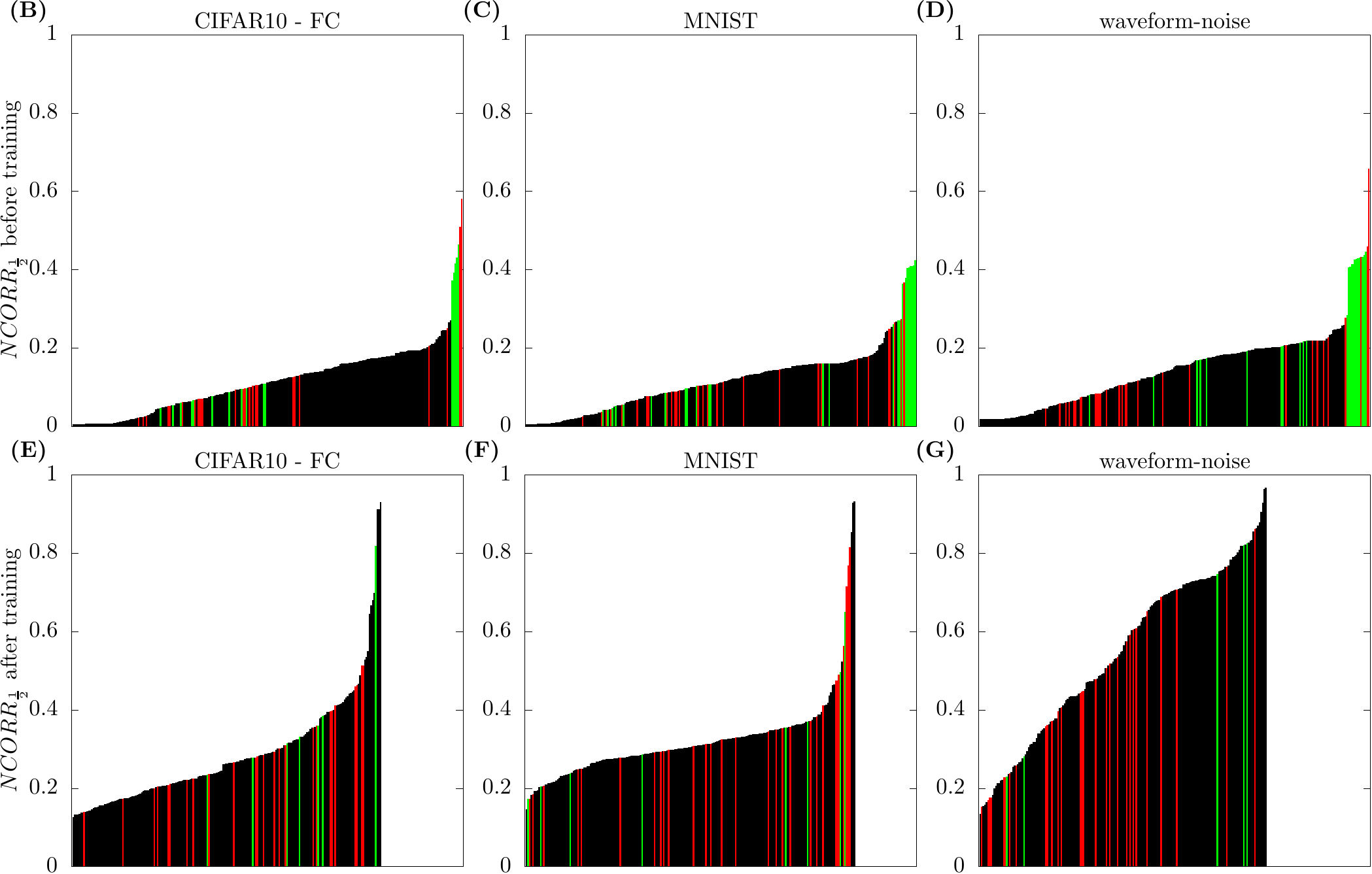}
\caption{NCORR at an intermediate fully-connected or addition layer. Graphs are analogous to previous figures. {\it Conclusion:} Some correlation between neurons does arise, especially when Gaussian initialization (graph A) is used over orthogonal initialization (graphs B-F).} \label{mfMetaGaussiancorr}
\end{figure}

\begin{figure}[H]
\centering
\includegraphics[width=0.98\textwidth]{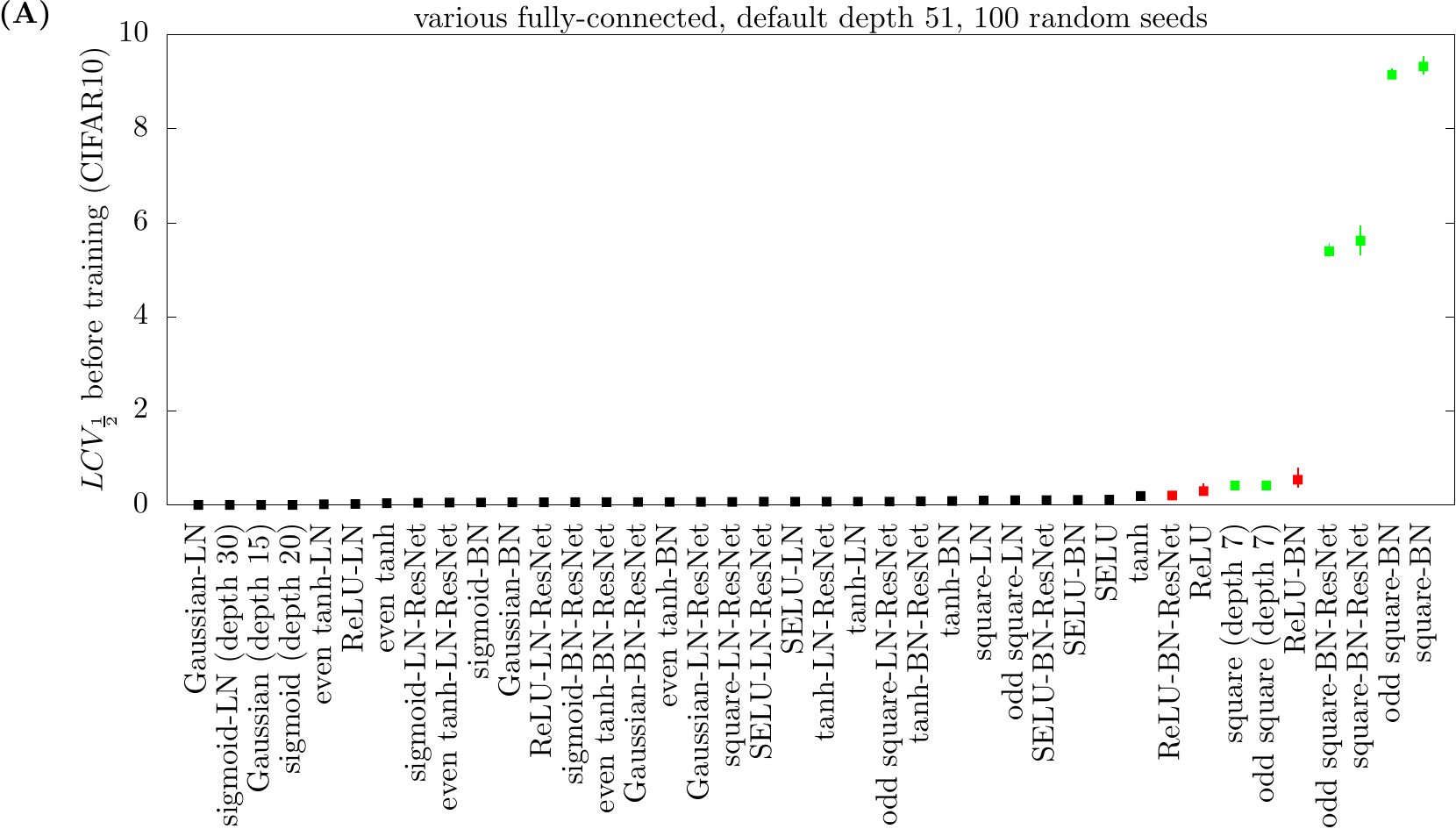}
\includegraphics[width=0.98\textwidth]{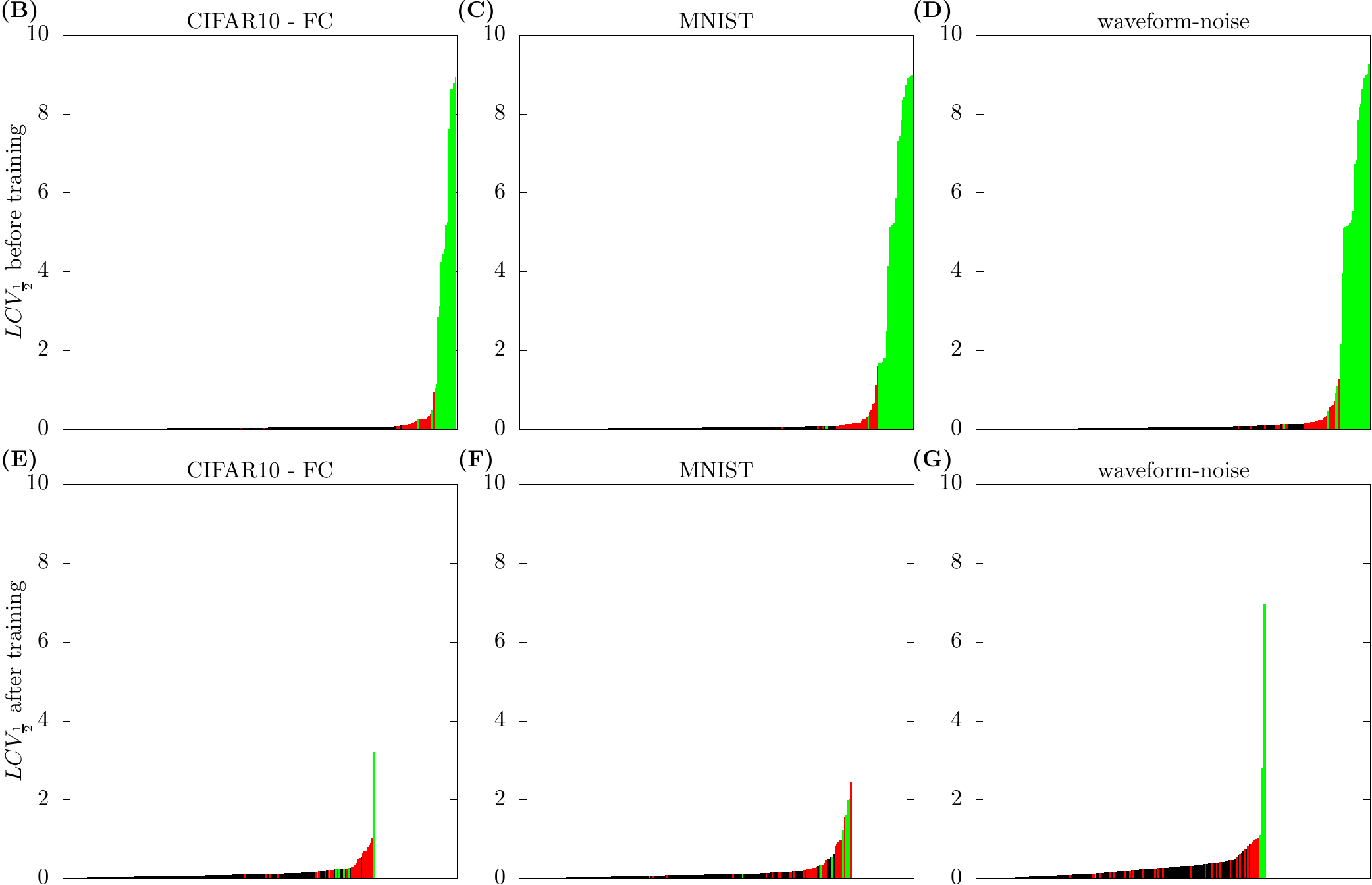}
\caption{LCV at an intermediate fully-connected or addition layer. Graphs are analogous to previous figures. {\it Conclusion:} Stable architectures have near constant layer lengths, especially before training, whereas GUAs and GEAs often have wildly diverging lengths.} \label{mfMetaGaussianfq}
\end{figure}

In this chapter, as in section \ref{nlcPropertiesSection}, we will define metrics that capture the properties we investigate and plot the metric values for our architectures. Figures \ref{mfMetaGaussiankurt} through \ref{mfMetaGaussianfq} are all laid out in the same way. In graph (A), we plot the metric values for our 40 simple architectures on CIFAR10 in the initial state. The interval depicted gives the range of the metric value across the 100 random seeds and the filled square depicts the mean. In graphs (B-D), we plot the metric values for our study A architectures in the initial state. In graphs (E-G), we plot the metric values for our study A architectures in the final state. While our theoretical framework applies directly only to the initial state, throughout this chapter, we also investigate how well it predicts the final state. In this chapter in general, as in section \ref{nlcPropertiesSection}, we find that its predictiveness degrades significantly but not completely. As in section \ref{nlcPropertiesSection}, whenever we quote the value of any metric not based on error in the final state, we exclude all architectures that did not attain a better-than-random validation error for any starting learning rate. If there is no meaningful way to select a starting learning rate, then the final state of the architecture is not meaningfully defined for the purpose of metric computation. Hence, the total number of architectures depicted in graphs E-G is somewhat lower than in graphs B-D in the following figures.

In this chapter, as in section \ref{nlcPropertiesSection}, Gaussian unstable architectures (GUAs) play a prominent role as they again defy many trends that hold for other architectures. As always, they are depicted in graphs by green markers. (In figures \ref{mfPredfq} and \ref{mfPredjac}, they are depicted as blue points with green confidence intervals for visibility.) They were first discussed in section \ref{nlcRobustDataSection}. As described at the end of section \ref{metricTerminologySection}, we say one of our fully-connected architectures is a GUA if (i) it uses either the square or odd square activation function and (ii) it does not use layer normalization. In this and the next section, we add the color red for what we term `Gaussian edge architectures' (GEAs), which use ReLU and also do not use LN. We give a full explanation of these phenomena in section \ref{gaussianStabilityExplanationSection}. For now, we notice that throughout this section, we find that GUAs and GEAs are ``not (meta-) Gaussian''. Throughout this section and section \ref{meanFieldPracticalEmpiricalSection} only, we use the term `stable' to refer to architectures that are neither GUAs nor GEAs.

Now we turn to our four properties as given above.

\paragraph{Property 1: Gaussianity of individual neurons} We begin by investigating whether $f_\frac{1}{2}(\mathcal{D})[i]$ is Gaussian for fixed $\theta$. A staple metric for measuring ``degree of Gaussianity'' is excess kurtosis.

\begin{metricDefinition}
The `neuron excess kurtosis' (NKURT) of a network $f$ with respect to an input distribution $\mathcal{D}$ at the $i$'th neuron in layer $l$ is
$$NKURT_l(f, \mathcal{D},i) = \mathbb{K}_xf_l(x)[i]$$
\end{metricDefinition}

The excess kurtosis operator $\mathbb{K}$ as well as its estimator are given in section \ref{metricEstimationSection}. The excess kurtosis of a Gaussian distribution is 0. Hence, we interpret NKURT being close to zero as Gaussian behavior. Specifically, in order to assess Gaussianity, excess kurtosis focuses on outliers, i.e. it measures whether a distribution is heavy-tailed or light-tailed relative to a Gaussian.

In figure \ref{mfMetaGaussiankurt}, we plot the value of NKURT averaged over all neurons in $f_{\frac{1}{2}}$, i.e. $\mathbb{E}_iNKURT_{\frac{1}{2}}$. In figure \ref{mfMetaGaussiankurt}(A-D), we find that \finding{for stable architectures, depicted in black, excess kurtosis is very close to zero, and actually more likely to be below than above zero}, indicating Gaussianity. On the other hand, \finding{for GEAs and GUAs, depicted in red and green respectively, excess kurtosis is significant to enormous}, indicating that neuron distributions are very non-Gaussian. \finding{GUAs have especially non-Gaussian neuron distributions.} Specifically, before training, stable architectures tend to have neuron distributions in fully-connected (FC) layers that are as or more light-tailed than a Gaussian, whereas GEAs and especially GUAs have neuron distributions in FC layers that are heavy-tailed. In figure \ref{mfMetaGaussiankurt}(E-G), we find that \finding{neuron distributions are often very light-tailed after training, but that there are also a few stable architectures that have large NKURT values}. 

In addition to $\mathbb{E}_iNKURT_{\frac{1}{2}}$, we investigated the standard deviation of excess kurtosis values across the layer, i.e. $\mathbb{S}_iNKURT_{\frac{1}{2}}$. We found that \finding{the standard deviation never significantly exceeded the absolute value of the mean and was often significantly lower}. Hence, we do not display those values here as they provide little additional information. However, this does confirm that the neuron kurtosis is a characteristic value of the network and layer and does not vary too much from neuron to neuron in an FC layer.

In addition to capturing heavy-tailedness via NKURT, we wanted to examine the neuron distribution with another metric that focuses on whether the cumulative distribution function at each neuron is close to the cumulative distribution function of a Gaussian between, say, -2 and +2 standard deviations. For this purpose, we devised the `Gaussian histogram intersection' metric. We divide the real line into a small number of buckets and obtain a histogram from both the unit Gaussian and the expectation and variance normalized neuron distribution. We then sum the bucket-wise minimum of the two histograms to obtain the distribution overlap. A value of 1 then constitutes a perfect match, whereas a value close to 0 indicates dissimilarity.

\begin{metricDefinition}

The `neuron Gaussian histogram intersection' (NGHI) is

\begin{eqnarray*}
&&NGHI_l(f,\mathcal{D},i,E)\\
&=&\sum_{e=1}^{|E|-1} \min\Big(N(E[e+1]) - N(E[e]), \mathbb{E}_x\{E[e] \le \frac{f_l[i]-\bar{f}_l[i]}{\sqrt{\mathbb{E}_x(f_l[i]-\bar{f}_l[i])^2}} < E[e+1]\}\Big)
\end{eqnarray*}

where $N()$ is the unit Gaussian CDF, $E$ is the edge vector that defines the buckets and curly braces denote an indicator variable.
\end{metricDefinition}

In figure \ref{mfMetaGaussianiou}, we plot $\mathbb{E}_iNGHI_{\frac{1}{2}}$ with 

$$E=(-\infty,-1.8,-1.4,-1,-0.6,-0.2,0.2, 0.6, 1, 1.4, 1.8, \infty)$$ 

We obtain very similar results as for NKURT. \finding{Before training, stable architectures have an NGHI value above 0.9, and often above 0.95}, indicating that their neuron distributions have CDFs comparable to a Gaussian. \finding{GUAs have NGHI values close to 0, with GEAs still having NGHI values mostly above 0.8}. After training, \finding{NGHI is significantly lower overall and the architecture type is less predictive}.

\paragraph{Property 2: Distribution of neuron expectations} To show that $\bar{f}_\frac{1}{2}$ is elementwise and generated by $\mathcal{N}(0,\mathfrak{C}(q,c))$ as $\theta$ is random, we need to evaluate the covariance kernel. We calculate the kernel for study A architectures in the next section. For now, we will show that the neuron expectations across an FC layer for a given (typical) $\theta$ appear like a Gaussian sample. We use similar metrics as above.

\begin{metricDefinition}

$$KURTEX_l(f,\mathcal{D}) = \mathbb{K}_i\bar{f}[i]$$

\end{metricDefinition}

\begin{metricDefinition}

\begin{eqnarray*}
&&GHIEX_l(f,\mathcal{D}, E)\\
&=& \sum_e \min\Big(N(E[e+1]) - N(E[e]), \mathbb{E}_i\{E[e] \le \frac{\bar{f}_l[i] - \mathbb{E}_{i'}\bar{f}_l[i']}{\sqrt{\mathbb{E}_{i'}(\bar{f}_l[i'] - \mathbb{E}_{i''}\bar{f}_l[i''])^2}} < E[e+1]\}\Big)
\end{eqnarray*}

where $N()$ is the unit Gaussian CDF, $E$ is the edge vector that defines the buckets and curly braces denote an indicator variable.

\end{metricDefinition}

We give results in figures \ref{mfMetaGaussiankmu} and \ref{mfMetaGaussianimu} with the same edge vector as before. Before training, \finding{KURTEX is close to zero and GHIEX is close to 1 for all architectures including GUAs and GEAs}, indicating that neuron expectations appear highly Gaussian across the layer.

This is to be expected. If $f_l$ is an FC layer, we have $f_l = f_kW_l$, so we have $\bar{f}_l = \bar{f}_kW_l$. $\bar{f}_k$ can be viewed as a random vector that depends on the parameter sub-vectors of layers $f_1$ through $f_k$. Drawing $\bar{f}_l$ from the meta-distribution corresponds to drawing both $\bar{f}_k$ and $W_l$ independently. Because of the symmetry of $W_l$ when it is either Gaussian or orthogonally initialized, we have that the orientation of $\bar{f}_l$ is independent of $\bar{f}_k$. Further, since KURTEX and GHIEX are independent of the length of $\bar{f}_l$, both metrics are entirely independent of $\bar{f}_k$. But for a fixed $\bar{f}_k$, the distribution of $\bar{f}_l$ is either Gaussian when $W_l$ is Gaussian initialized or indistinguishably close to Gaussian when $W_l$ is orthogonally initialized. Hence, for any FC layer $f_l$ in any architecture with respect to any input distribution, the distribution of KURTEX and GHIEX induced by $\theta$ is the distribution those metrics attain on a Gaussian sample with size equal to the width of $f_l$. In plain terms, the distribution of neuron expectations across an FC layer appears as Gaussian as a sample from an actual Gaussian.

Because of this, any deviation we observe from $KURTEX=0$ and $GHIEX=1$ in figures \ref{mfMetaGaussiankmu} and \ref{mfMetaGaussianimu} in the initial state can be regarded as sampling noise caused by the finite layer width, at least for non-residual architectures where $f_\frac{1}{2}$ is actually FC. In graph (A), architectures have width 100 and in graphs (B-D), architectures have width between 130 and 1000. This causes neuron expectations to appear more Gaussian for our study A architectures. 
This above insight also provides further context for figures \ref{mfMetaGaussiankurt} and \ref{mfMetaGaussianiou}. For our stable architectures, the individual neuron distributions in $f_\frac{1}{2}$ in the initial state appear significantly more Gaussian than samples from a Gaussian distribution with size between 100 and 1000.

\finding{After training, we find that neuron expectations across $f_\frac{1}{2}$ tend to become less Gaussian, though not much.}

While we have shown that neuron expectations across an FC layer appear Gaussian for a fixed $\theta$, this does not necessarily imply that they are mean zero elementwise Gaussian as $\theta$ is random. However, we would obtain this from the above analysis if $\bar{f}_k$ has fixed square mean. In section \ref{meanFieldPracticalSection}, we show that this is true as $d_\text{MF} \rightarrow \infty$ under the conditions of theorem \ref{mfntPropagation} and approximately true empirically. Finally, in that section, we will calculate $\mathfrak{C}(q,c)$ and show that it empirically predicts the standard deviation of neuron expectations in FC layers. This completes the analysis of property 2.

\paragraph{Property 3: Standard deviation of individual neurons}

Again, we will not calculate the value $\mathfrak{C}(q,q) - \mathfrak{C}(q,c)$ as it arises in property 3 for now. We will focus on demonstrating that the standard deviations of neuron distributions are approximately constant across the layer. Hence, we use the coefficient of variation of standard deviations.

\begin{metricDefinition}

The `coefficient of variation of neuron standard deviations' (CVNSTD) is

$$CVNSTD_l(f,\mathcal{D}) = \frac{\mathbb{S}_i(\mathbb{S}_xf_l[i])}{\mathbb{E}_i(\mathbb{S}_xf_l[i])}$$

\end{metricDefinition}

A low value of CVNSTD indicates that the mean of standard deviations is much larger than the standard deviation of standard deviations, which indicates that the standard deviations are relatively constant across the layer as desired. \finding{This is what we find in figure \ref{mfMetaGaussianstd}}. \finding{The largest CVNSTD values before training are around 0.5  and are largely observed on GUAs and GEAs. However, the ``advantage'' of stable architectures is not as drastic as e.g. in figures \ref{mfMetaGaussiankurt} and \ref{mfMetaGaussianiou}. We also find that overall, CVNSTD values are much smaller for study A architectures in graphs (B-D) than for the simple architectures in graph (A)}. This is because study A uses orthogonal initialization for weight matrices, whereas the simple architectures use Gaussian initialization. Given some fixed value of $\Cov_{f_k}$, it is easy to see that minimizing $CVNSTD_l$ requires that the columns of $W_l$ have fixed and equal length, which happens under orthogonal initialization when $d_k \ge d_l$, but not under Gaussian initialization. Further, the spectrum of $\Cov_{f_k}$ itself tends to be better conditioned under orthogonal initialization \citep{eigenspectrum,orthogonalInitialization}.

\finding{After training, while CVNSTD values are much larger, standard deviations of neuron distributions are still relatively constant across the layer.}

In section \ref{meanFieldPracticalSection}, we prove that  $\frac{1}{d_ld_\text{MF}}||\mathbb{S}_xf_l||_2^2$ converges to $\mathfrak{C}(q,q) - \mathfrak{C}(q,c)$ a.s. as $d_\text{MF} \rightarrow \infty$. Demonstrating this empirically for FC layers in section \ref{meanFieldPracticalEmpiricalSection} will complete the analysis of property 3.

\paragraph{Property 4: Neuron independence} Finally, we examine to what degree the neuron distributions are independent. A staple metric for assessing independence is correlation.

\begin{metricDefinition}
The `neuron correlation' (NCORR) is

$$NCORR_l(f, \mathcal{D}) = \sqrt{\mathbb{E}_{i,i'}\Big(\frac{\mathbb{C}_x(f_l[i],f_l[i']))}{\mathbb{S}_xf_l[i]\mathbb{S}_xf_l[i']}\Big)^2}$$

\end{metricDefinition}

The covariance operator $\mathbb{C}$ and its estimator are defined in section \ref{metricComputationSection}. It is worth noting that we consider the quadratic mean of pairwise correlations rather than the arithmetic mean. This is necessary because a positive and negative correlation with the same absolute value is equally likely due to the symmetry of $W_l$.

In figure \ref{mfMetaGaussiancorr}, we plot $NCORR_{\frac{1}{2}}$. \finding{We find that even before training, most architectures exhibit some degree of correlation, and that the largest correlations tend to occur in GUAs and GEAs}. Some amount of correlation can be explained by Gaussian initialization itself. Non-zero correlations between columns of $W_l$ lead to non-zero correlations between neurons. In fact, even if $\Cov_{f_k}$ is the identity matrix, we would expect $NCORR_\frac{1}{2} \approx \frac{1}{\sqrt{d_\frac{1}{2}}}$. Since architectures in figure \ref{mfMetaGaussiancorr}(A) have $d_\frac{1}{2} = 100$, it is unsurprising that we do not observe $NCORR_{\frac{1}{2}}$ values below $\frac{1}{\sqrt{100}}$. As expected, \finding{neuron correlations are smaller for study A architectures than the simple architectures, as they are orthogonally initialized}.

It is also interesting to note that \finding{$NCORR_0$ is 0.19 / 0.16 / 0.17 when evaluated for CIFAR10 / MNIST / waveform-noise respectively}. Hence, \finding{the majority of stable architectures with orthogonal initialization actually reduce neuron correlation during forward propagation in the initial state}. Since we would expect an orthogonal transformation to approximately preserve neuron correlation, this reduction is caused by the nonlinear layers. \citet{eigenspectrumGram} studied the decorrelating effect of activation layers.

\finding{After training, NCORR rises significantly and can attain values close to 1.}

If neuron distributions are independent across a layer, we expect the square mean of layer values to be approximately constant across $\mathcal{D}$. See e.g. proposition \ref{mfntElemLike} in the next subsection. Hence, another way to measure independence is with the metric below.

\begin{metricDefinition}

The `layer coefficient of variation' (LCV) is

 $$LCV_l(f,\mathcal{D}) = \frac{\mathbb{S}_x||f_l||_2}{\mathbb{E}_x||f_l||_2}$$
 
 \end{metricDefinition}
 
In figure \ref{mfMetaGaussianfq}, we plot $LCV_\frac{1}{2}$. \finding{The results are more drastic than for the $NCORR$ metric. Stable architectures have an almost perfectly constant length, whereas many GUAs and GEAs exhibit severe variation.} To appreciate the extent of this variation, consider an architecture with batch normalization. The largest possible value of LCV at a BN layer is attained when a single layer value in the batch has far greater length than the others, and then $LCV \approx \sqrt{|B|}$. Since we use a batch size of 250, many GUAs actually come close to this theoretical maximum. \finding{Length variation remains very small for stable architectures after training.}

It turns out that studying the degree of variation among layer lengths is a key to explaining Gaussian instability in section \ref{gaussianStabilityExplanationSection}.

\newpage

\newpage

\subsection{Neural regular data} \label{elemLikeSection}

\begin{table}
{
\centering
\begin{tabular}{lccc}
Dataset&CIFAR10&MNIST&waveform-noise \\ \hline\hline 
\\
INSQD density&\includegraphics[scale=0.27,valign=c]{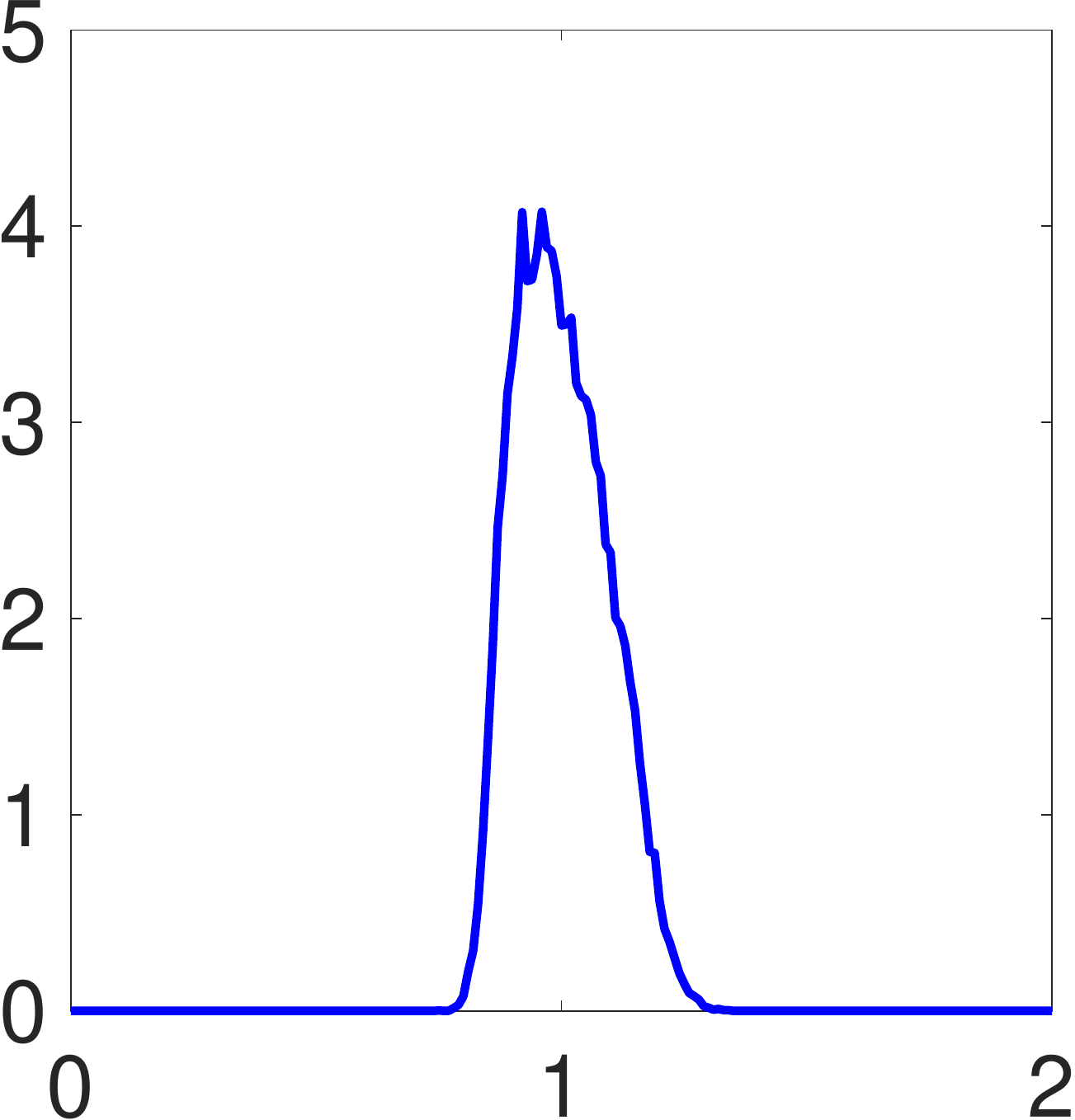}&\includegraphics[scale=0.27,valign=c]{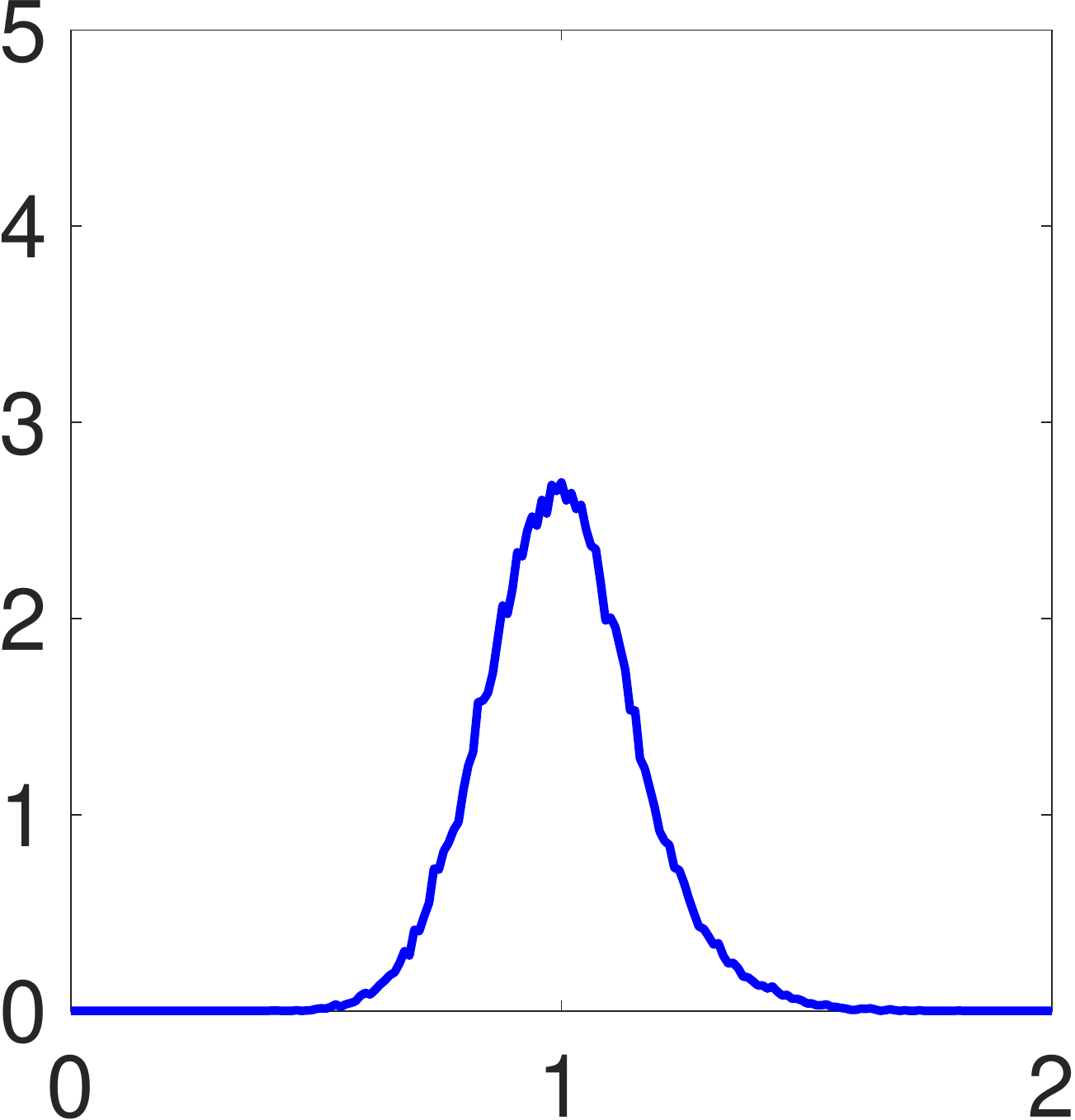}&\includegraphics[scale=0.27,valign=c]{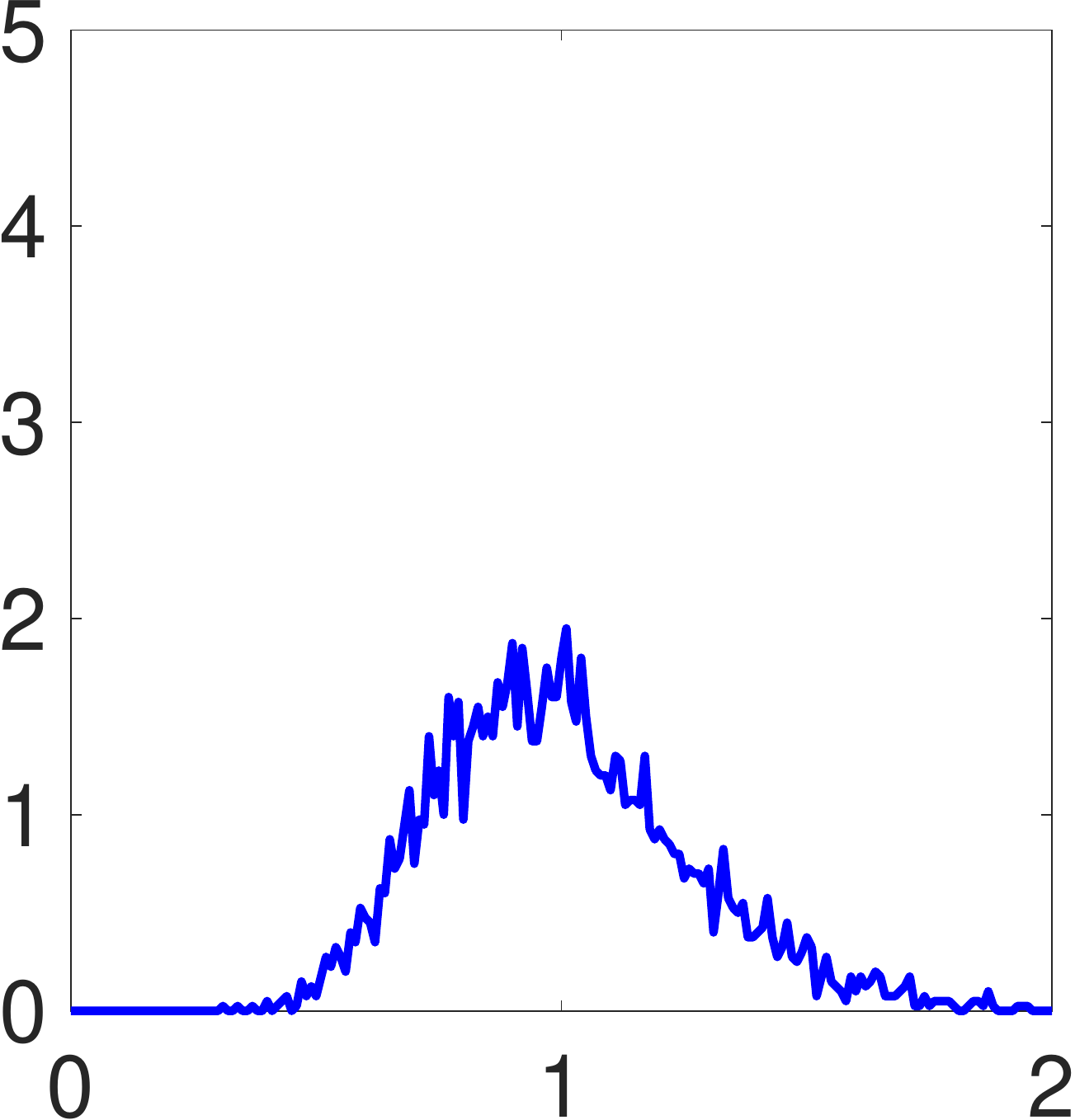}\\
INCMD density&\includegraphics[scale=0.27,valign=c]{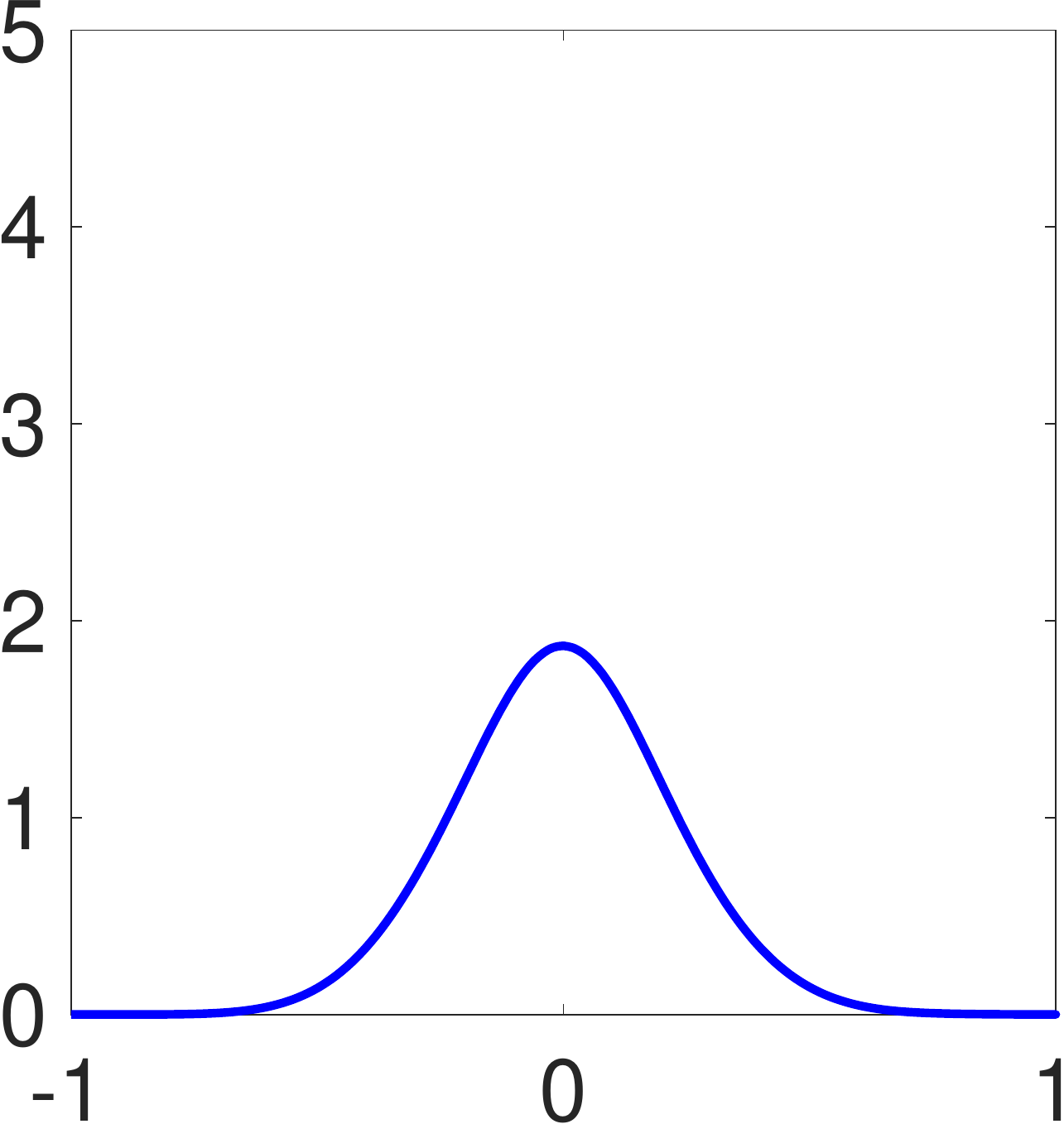}&\includegraphics[scale=0.27,valign=c]{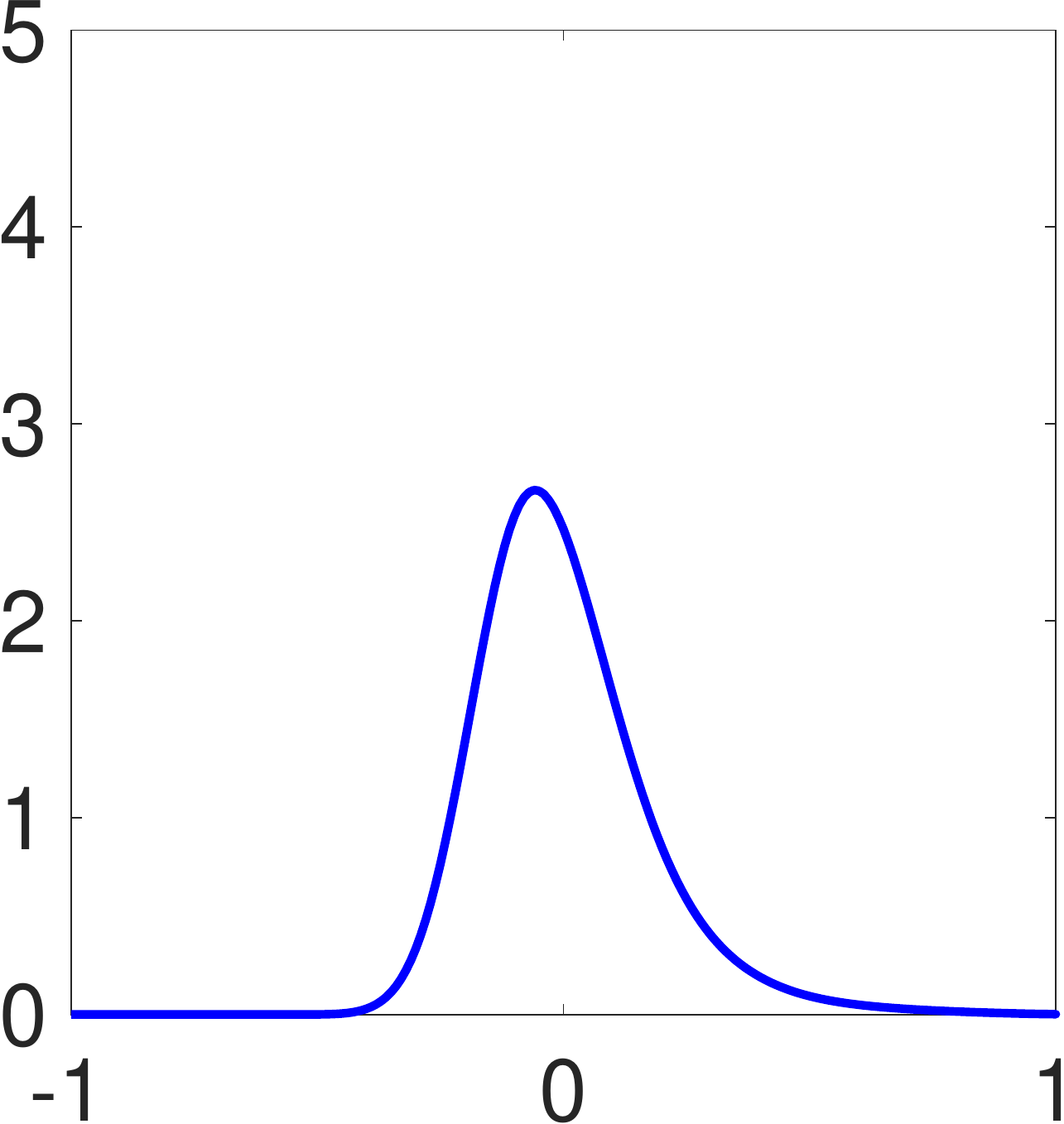}&\includegraphics[scale=0.27,valign=c]{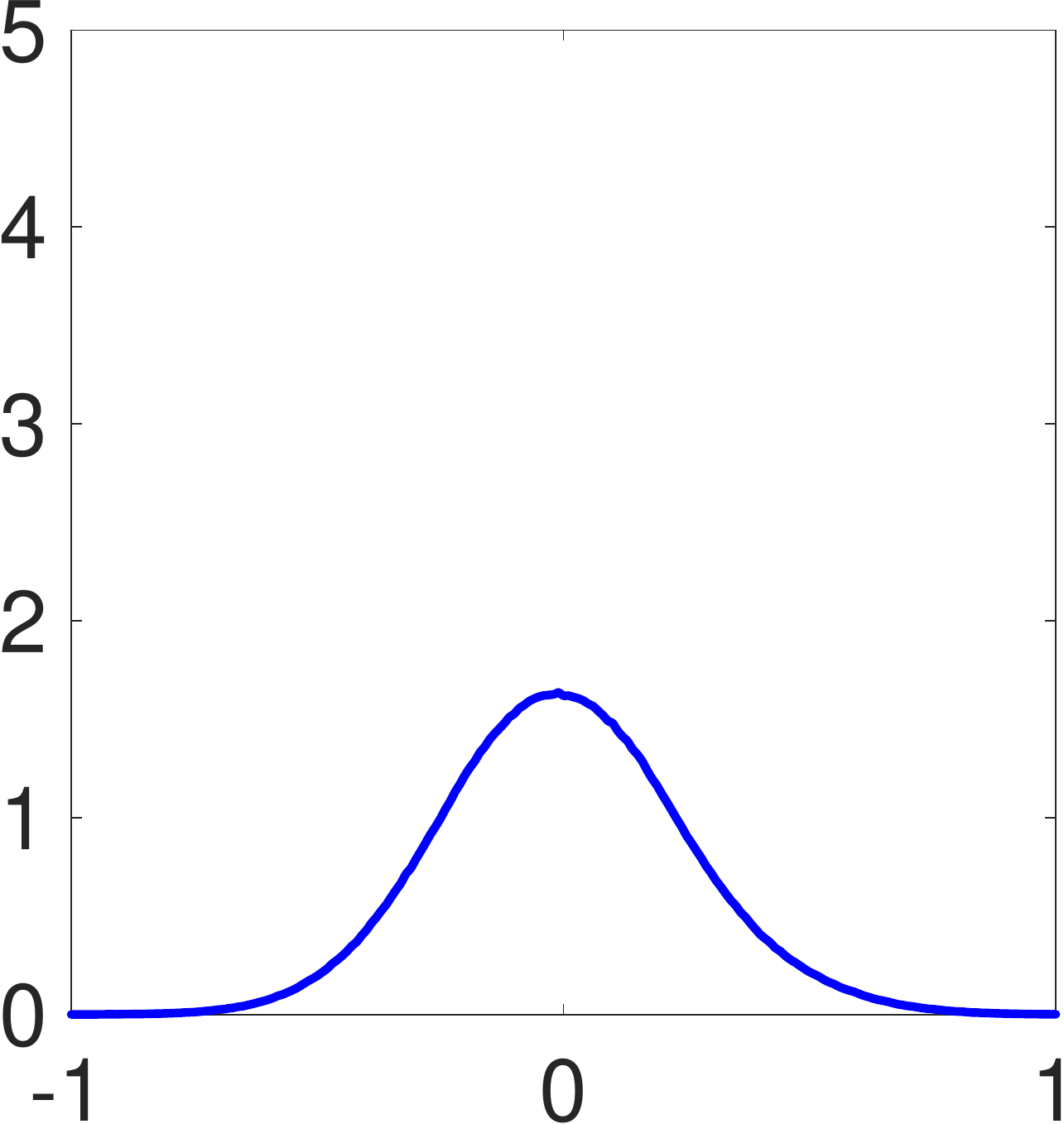}\\
\end{tabular}
\caption{INSQD and INCMD taken on the union of training and validation set for study A datasets. Note that INSQD for waveform-noise appears jittery because the sample size is smaller. {\it Conclusion:} Our datasets can be viewed as approximately `elem-like(1,0)'.}
\label{mfElemLike}
}
\end{table}

In theorem \ref{mfntMetaGaussian}, we used the condition that the input distribution $\mathcal{D}$ is elem-like$(q,c)$. We explained that this condition cannot be fulfilled exactly. In this subsection, we further analyze it. We begin by simply examining the distribution of the input square mean and co-mean.

\begin{metricDefinition}
The `input square mean distribution' (INSQD) and `input co-mean distribution' (INCMD) are

\begin{eqnarray*}
INSQD(\mathcal{D}) &=& \mathbb{E}_ix[i]^2 \text{ where } x \sim \mathcal{D}\\
INCMD(\mathcal{D}) &=& \mathbb{E}_ix[i]x'[i] \text{ where } x,x' \sim \mathcal{D}\\
\end{eqnarray*} 

\end{metricDefinition}

We plot the density functions of INSQD and INCMD in table \ref{mfElemLike}. They were taken on the union of training and validation set after data processing as always. We find that \finding{$INSQD \approx 1$ and $INCMD \approx 0$.} Of course, for a hypothetical exact elem-like(1,0) distribution, INSQD would have all its probability mass concentrated at 1 and INCMD would have all its probability mass concentrated at 0. This suggests that our datasets are approximately elem-like$(1,0)$ and hence suitable for our mean field theory of meta-distributions.

Note that $INSQD=1$ can simply be achieved by normalizing the length of inputs and $\mathbb{E}_{x,x'}INCMD=0$ can simply be achieved by normalizing the expectation of input components. See pointwise and componentwise normalization in section \ref{dataProcessingSection} respectively. $\mathbb{E}_xINSQD=1$ and $\mathbb{E}_{x,x'}INCMD=0$ indeed holds on the union of training and validation set of our datasets due to data processing according to section \ref{studyADataSection}, and hence in figure \ref{mfElemLike}. (Of course, this means that training and validation set are no longer strictly an independent sample for evaluating INSQD and INCMD.)

\subsubsection{Elem-like $\approx$ elementwise}

It turns out that an elementwise distribution becomes more and more elem-like as dimensionality increases. This explains our choice of the name `elem-like'.

\begin{proposition} \label{mfntElemLike}
Let $\mathcal{M}_1$ be a meta-distribution over distributions $\mathcal{X}_1$ over scalars $s$. For each $d > 0$, let $\mathcal{M}_d$ be the elementwise meta-distribution over distributions $\mathcal{X}_d$ over vectors $\chi_d$ of dimensionality $d$ that is generated by $\mathcal{M}_1$. Let $q = \mathbb{E}_{\mathcal{X}_1\sim\mathcal{M}_1,s\sim\mathcal{X}_1}s^2$ and $c =  \mathbb{E}_{\mathcal{X}_1\sim\mathcal{M}_1}(\mathbb{E}_{s\sim\mathcal{X}_1}s)^2$. Let $\chi_d$ and $\chi'_d$ be drawn from the same $\mathcal{X}_d$, which is drawn from $\mathcal{M}_d$. Then 

$$\lim_{d\rightarrow \infty} \mathbb{E}_i\chi_d[i]^2 = q \text{ a.s.}$$ 
$$\lim_{d\rightarrow \infty} \mathbb{E}_i\chi_d[i]\chi'_d[i] = c \text{ a.s.}$$ 
\end{proposition}

Note that if $\mathcal{M}_1$ is $\mathcal{MN}(q,c)$ in the above proposition, then the $q$ and $c$ values we obtain are exactly the parameters of the meta-Gaussian. This was the reason behind parametrizing the meta-Gaussian in the way we did. Of course, the proposition also holds when $\mathcal{M}_1$ has all its probability mass on a single fixed distribution $\mathcal{X}_1$, which implies that each component of $\chi_d$ is drawn IID from $\mathcal{X}_1$.

In light of the proposition, the elem-like property can be interpreted as requiring that $\mathcal{D}$ mimics an elementwise distribution, which happens to be the type of distribution that allowed us to set up mean field theory in section \ref{meanFieldBackgroundSection} in the first place. More generally, we can glean from proposition \ref{mfntElemLike} that the law of large numbers will lead to approximate elem-like-ness if input components are ``relatively independent'', i.e. sufficiently independent that an average over all input components tends to exhibit a significant amount of convergence towards its mean. As mentioned in section \ref{meanFieldDistributionEmpiricalSection}, the $NCORR_0$ values obtained from our datasets are 0.19 for CIFAR10, 0.16 for MNIST and 0.17 for waveform-noise, i.e. they are low.

\subsubsection{Elem-like $\approx$ use a small NLC}

In section \ref{bestNlcSection}, we showed how the apparent nonlinearity of the true input-label function, as measured by the  PNLCD metric, is predictive of the ideal architecture NLC for a dataset. PNLCD is based on comparing the distance of labels relative to the size of the codomain to the distance of inputs relative to the size of the domain. Our evidence suggested that PNLCD concentrating around 1 would lead the dataset to also require a small initial NLC between 1 and 5. We described how inputs drawn from the unit Gaussian distribution, which is elementwise, would lead to PNLCD concentrating around 1. We showed that datasets must have ``tight input clusters'' to attain larger PNLCD sample values. We showed that our three study A datasets largely do not have such clusters. Now, we relate that analysis back to this section.

\begin{proposition} \label{mfntPNLCD}
Let $\mathcal{D}$ be a data distribution using a finite set of labels in the form of one-hot vectors. Let $p_1, .., p_C > 0$ be the probabilities of each of the $C > 1$ classes occurring. Let the input of $\mathcal{D}$ be elem-like$(q,c)$ for some $q > c \ge 0$.

Then $PNLCD(\mathcal{D})=\frac{1}{\sqrt{1 - \sum_{\zeta=1}^Cp_\zeta^2}}$ with probability $1 - \sum_{\zeta=1}^Cp_\zeta^2$ and 0 with probability $\sum_{\zeta=1}^Cp_\zeta^2$.
\end{proposition}

In practical terms, if the inputs are elem-like and the class frequencies are not too unbalanced, PNLCD is concentrated at a value close to 1 for datapoints of different classes.

\subsubsection{Discussion}

We now have three different views of dataset regularity - (i) via its apparent nonlinearity, ideal NLC and the PNLCD metric, (ii) via the mean field theory of meta-distributions, elem-like-ness and the INSQD and INCMD metrics and (iii) via surrogate input distributions, elementwise-ness and the NCORR metric. These are all related. Below, we state a definition that unifies the three views.

\begin{definition}
A data distribution is `neural regular' if (i) datapoints with relatively different labels tend to have inputs that are relatively distant within their domain, (ii) input components are relatively independent and (iii) the distributions of input square means and co-means are relatively concentrated around fixed values.
\end{definition}

Neural regularity, unfortunately, is not well-defined at this point. However, as also argued in section \ref{bestNlcSection}, we suspect a large fraction of practical deep learning datasets, especially after data processing, to be sufficiently neural regular to produce the results obtained in this work, as well as other important results. This can serve as a foundation for data-agnostic ZSAD, as outlined in section \ref{moduloDataSection}. 

We note that the NLC is not just useful for neural regular datasets. In section \ref{bestNlcSection} we showed that the NLC is also predictive of test error when $PNLCD \approx 1$ is severely violated for datapoints of different classes. In that case, we have to estimate the ideal NLC range based on e.g. examining PNLCD.

Of course, we are not claiming that the precise requirements for the predictiveness of the NLC and the predictiveness of our mean field theory of meta-distributions will turn out to be exactly the same. There may be ``imperfectly neural regular'' datasets that are more suitable for one concept than the other. Investigating this beyond the analysis already performed is a task for future work. However, it is noteworthy that the {\it predictiveness} of the NLC, as modulated by the true input-label function, is closely related to the {\it predictability} of the NLC using mean field theory in the sense of figure \ref{boxNPM}.

\section{Mean field theory of practical metrics and architectures} \label{meanFieldPracticalSection}

\subsection{A-architectures} \label{aArchitectureSection}

We now apply the abstract results from previous sections to derive even more practical insights. Specifically, we calculate the limit of metrics such as the NLC as width converges to infinity. In subsection \ref{pathEquationSection}, our analysis results in a simple and instructive formula for the limit of the NLC.

In section \ref{studyAArchitecturesSection}, we detailed the types of fully-connected architectures we used for study A, which are some of the simplest types of practical architectures. We will use our study A architectures as a blueprint in this section.

\begin{definition}
We say an architecture $f$ is an `A-architecture' (after study A) if it has the following properties. A-architectures follow our standard framework from section \ref{neuralNetworkNotationSection}, except for a single property marked with (*).

\begin{enumerate}[label=22.\arabic*,leftmargin=1.5cm]
\item It is composed of layers using the following operations: fully-connected, activation, bias, layer normalization, batch normalization and addition. See section \ref{layerTypesSection} for how we define these operations. \label{aprop1}
\item Activation functions used are non-constant, twice differentiable, and both they and their derivatives are controlled by $\mathcal{C}^\text{E2}$.  \label{aprop2}
\item Addition layers have fixed, non-zero addition weights.  \label{aprop3}
\item The input layer has fixed width $d_0=d_\text{in}$. Non-input, non-output layers $f_l$ have width $d_ld_\text{MF}$, where $d_l$ is fixed and $d_\text{MF}$ can vary. There is at least one non-input, non-output layer. The output layer may have variable width $d_Ld_\text{MF}$ or be a fully-connected `readout layer' of fixed width $d_L=d_\text{out}$. We also use $d_l^\text{MF}$ to denote the width of a layer, which can be $d_l$ or $d_ld_\text{MF}$ depending on the layer. (*)  \label{aprop4}
\item Components of weight matrices $W_l$ are initialized as independent Gaussians with mean zero and variance $\frac{\sigma_l^2}{d_k^\text{MF}}$ for some `variance parameter' $\sigma_l^2 > 0$ that is fixed as $d_\text{MF}$ varies.  \label{aprop5}
\item Components of bias vectors $\beta_l$ are initialized as independent Gaussians with mean zero and variance $\sigma_l^2$ for some fixed `variance parameter' $\sigma_l^2$.  \label{aprop6}
\item If $f_l$ is a normalization layer, the regularizer $\epsilon_l$ is positive. \label{apropX1}
\item $f$ uses either batch normalization or layer normalization layers, but not both.  \label{aprop7}
\item In the layer graph, there do not exist two directed paths that begin at distinct activation layers, end at the same layer, and do not contain any fully-connected layers.  \label{aprop8}
\item In the layer graph, there does not exist a directed path that begins at an activation layer, ends at a different activation or normalization layer, and does not contain a fully-connected layer.  \label{aprop9}
\item In the layer graph, there do not exist two distinct directed paths that begin and end at the same layer and contain no fully-connected layer outside of their starting point.  \label{aprop10}
\end{enumerate}

\end{definition}

A-architectures follow the framework of section \ref{neuralNetworkNotationSection}, except that layers have variable width. Variable width stems from mean field architectures defined in section \ref{meanFieldArchitectureSection}. In contrast to mean field architectures, A-architectures have a single input and output layer and there is no weight sharing.

Our architectures from study A are A-architectures when $d_\text{MF} = 1$, except (i) weight matrices are orthogonally initialized, (ii) some activation functions are only directionally differentiable instead of twice differentiable everywhere and (iii) the last addition layer in residual architectures multiplies the skip connection with a fixed orthogonally initialized matrix. These are minor differences in this context. Large orthogonally initialized matrices have approximately Gaussian entries with very mild cross-entry dependency. Our directionally differentiable activation functions are also very close to smooth activation functions. For example, ReLU is approximately $\frac{1}{c}\log(1+e^{cx})$ for some very large $c$. See section \ref{nonDifferentiableSection} for further discussion on assuming differentiability in neural network analysis.

A-architectures turn out to be a specialization of mean field architectures. This insight is behind the results given below. While A-architectures allow more than the two layer operations used in mean field architectures, each possible combination of these operations can nonetheless be expressed in terms of those two. \citet{meanFieldNetsorGP} explained how to cast operations such as layer normalization and bias as elementwise layers. In this section, we cast entire architectures made up of up to five different operations as mean field architectures, which requires groups of layers to be re-cast jointly. This then enables the utilization of e.g. background theorem \ref{backgroundMaster} to prove the results below. Despite re-casting groups of layers as one, we show that we can nonetheless calculate a wide range of key metrics via recursion from one A-architecture layer to the next.

We note that many of the properties of A-architectures given in the definition, which are restrictive in nature, are not ``strictly necessary''. One can derive results similar to the ones given below for more general classes of architectures, including those admitting convolutional layers, as we further discuss at the end of section \ref{meanFieldPracticalInterpretationsection} and in section \ref{meanFieldCNNsection}. Because of space and time limitations, we decided to focus on a class of architectures that would yield calculation rules that are as simple and instructive as possible. We hope that our presentation is itself instructive and will allow readers to generalize our results to whatever architecture class they are interested in with the techniques we employ.

\subsection{Metric limits in A-architectures} \label{metricLimitSection}

We now show how to calculate the limits of metrics as width converges to infinity in A-architectures. After a definition, we state the main theorem of this section.

\begin{definition}
The `covariance kernel' $\mathfrak{C}_\tau$ of an activation function $\tau$ and scalars $q \ge c \ge -q$ is

$$\mathfrak{C}_\tau(q,c) = \mathbb{E}_{s,t\sim \mathcal{N}(\mu,\Sigma)} \tau(s)\tau(t) \text{ where } \mu = \begin{pmatrix} 0 \\ 0 \end{pmatrix} \text{, } \Sigma = \begin{pmatrix} q & c \\ c & q \end{pmatrix}$$

We also write $\mathfrak{C}_\tau(c)$ with $1 \ge c \ge -1$ short for $\mathfrak{C}_\tau(1,c)$.

\end{definition}

This definition mirrors that of the covariance kernel of an architecture from section \ref{covarianceKernelSection}. When $c=0$, the definition simplifies to $\mathfrak{C}_\tau(q,0) = (\mathbb{E}_{s\sim\mathcal{N}(0,q)}\tau(s))^2$ and when $c=q$, the definition simplifies to $\mathfrak{C}_\tau(q,q) = \mathbb{E}_{s\sim\mathcal{N}(0,q)}\tau(s)^2$.

\begin{theorem} \label{mfntPropagation}
Let ...

\begin{itemize}
\item ... $f$ be an A-architecture with an output layer of variable width.
\item ... $\mathcal{D}$ be an input distribution.
\item ... $x^{(1)},..,x^{(N)}$ be a sample from $\mathcal{D}$ of size $N \ge 2$ and let $\mathcal{D}^{(N)}$ be the discrete uniform distribution over that sample.
\item ... $\vec{\epsilon}$ be the vector of all regularizers used by normalization layers in $f$.
\end{itemize}

Assume:

\begin{itemize}
\item $\mathcal{D}$ is elem-like$(q,c)$.
\item $q > c$
\end{itemize}

If $f$ does not contain batch normalization but can contain layer normalization layers, we have

\begin{eqnarray}
\lim\mathbb{E}_if_l(x^{(1)})[i] &=& \mathfrak{m}_l\label{eqn5p0}\\
\lim\mathbb{E}_if_l(x^{(1)})[i]^2 &=& \mathfrak{q}_l\label{eqn5p1}\\
\lim\mathbb{E}_if_l(x^{(1)})[i]f_l(x^{(2)})[i] &=& \mathfrak{c}_l \label{eqn5p2}\\
\lim\frac{1}{d_l^\text{MF}}||\mathcal{J}_{l,m}(x^{(1)})||_F^2 &=& \frac{\mathfrak{g}_l}{\mathfrak{g}_m} \label{eqn5p4}\\
\lim\frac{1}{d_l^\text{MF}}\mathbb{E}_x||f_l(x)||^2_2 &=& \mathfrak{q}_l\label{eqn5p5}\\
\lim\frac{1}{d_l^\text{MF}}||\mathbb{E}_xf_l(x)||^2_2 &=& \mathfrak{c}_l \label{eqn5p6}\\
\lim\frac{1}{d_l^\text{MF}}||\mathbb{S}_xf_l(x)||^2_2 &=& \mathfrak{q}_l - \mathfrak{c}_l \label{eqn5p7}\\
\lim\frac{1}{d_l^\text{MF}}\mathbb{E}_x||\mathcal{J}_{l,m}(x)||_F^2 &=& \frac{\mathfrak{g}_l}{\mathfrak{g}_m} \label{eqn5p8}\\
\lim\frac{1}{d_l^\text{MF}}\mathbb{E}_x\Tr(\mathcal{J}_{l,m}(x)\Cov_{f_m}\mathcal{J}_{l,m}(x)^T) &=& \frac{\mathfrak{g}_l(\mathfrak{q}_m - \mathfrak{c}_m)}{\mathfrak{g}_m}\label{eqn5p9}\\
\lim NLC(f_l(f_m),f_m(\mathcal{D}^{(N)})) &=& \sqrt{\frac{\mathfrak{g}_l(\mathfrak{q}_m - \mathfrak{c}_m)}{\mathfrak{g}_m(\mathfrak{q}_l - \mathfrak{c}_l)}}\label{eqn5p10}
\end{eqnarray}

where

\begin{itemize}
\item ... $x \sim \mathcal{D}^{(N)}$
\item ... $\lim$ stands for $\lim_{\vec{\epsilon}\rightarrow 0,N\rightarrow\infty}\big(\lim_{d_\text{MF}\rightarrow\infty} ... \text{a.s.}\big)$. The inner limit takes $d_\text{MF}$ to infinity and the outer limit takes $N$ to infinity as well as $\vec{\epsilon}$ to zero. The inner limit is an ``almost sure'' limit.
\item ... $\mathfrak{m}_l$, $\mathfrak{q}_l$, $\mathfrak{c}_l$ and $\mathfrak{g}_l$ are calculated via table \ref{tableNLCPropagation}. 
\item ... $f_m$ is a bottleneck for $f_l$.
\item ... $0 \le m \le l \le L$
\item ... randomness is induced by $\theta$ and the $x^{(n)}$.
\end{itemize}

Further, if the output layer is instead a readout layer of fixed dimensionality $d_\text{out}$, the above limits do not necessarily hold when $l=L$. Instead, (a) as $d_\text{MF} \rightarrow \infty$, the meta-distribution of the output layer expansion-converges to the elementwise meta-distribution with generator $\mathcal{MN}(\mathsf{parm1}, \mathsf{parm2})$, where $\lim_{\vec{\epsilon}\rightarrow 0}\mathsf{parm1}=\mathfrak{q}_L$ and $\lim_{\vec{\epsilon}\rightarrow 0}\mathsf{parm2}=\mathfrak{c}_L$. The first level of randomness is induced by $\mathcal{D}$ and the second level by $\theta$. And, (b) when $N$ is fixed, for almost all samples, as $d_\text{MF} \rightarrow \infty$, $(f(x^{(1)}), .., f(x^{(N)}))$ converges in distribution to a Gaussian that is elementwise over output vectors where the generator has mean zero and covariance matrix $K(N, \mathsf{parm1}, \mathsf{parm2})$, where again $\lim_{\vec{\epsilon}\rightarrow 0}\mathsf{parm1}=\mathfrak{q}_L$ and $\lim_{\vec{\epsilon}\rightarrow 0}\mathsf{parm2}=\mathfrak{c}_L$. Randomness for each sample is induced by $\theta$.

If $f$ does not contain layer normalization but can contain batch normalization layers, analogous statements hold. See section \ref{meanFieldBNsection} for details.
\end{theorem}

Because this theorem is already very long, we decided not to give the BN case explicitly here, but instead defer it to section \ref{meanFieldBNsection}. Nonetheless, we include BN in table \ref{tableNLCPropagation} and in the discussion below. As in section \ref{nlcComputeSection}, we generalize to the BN case by letting $f$ take batches of inputs instead of individual inputs.

\subsubsection{Interpreting theorem \ref{mfntPropagation}} \label{meanFieldPracticalInterpretationsection}

\paragraph{Assumptions} As in theorem \ref{mfntMetaGaussian}, the most severe assumption is that $\mathcal{D}$ is elem-like. See section \ref{meanFieldDistributionSection} for a discussion of elem-like-ness, which holds approximately for our study A datasets.

$q > c$ is a mild assumption. $q = c$ would effectively imply that all inputs to $f$ are equal, and therefore that each layer takes a single fixed value. This would e.g. make the NLC invalid and BN degenerate as $\vec{\epsilon}$ converges to zero. Since the definition of `elem-like$(q,c)$' also requires $c \ge 0$, we have $q > c \ge 0$.

Note that we no longer explicitly require rank stability or parameter-controlled multi-activation functions in theorem \ref{mfntPropagation} the way we did throughout sections \ref{meanFieldBackgroundSection} and \ref{meanFieldDistributionSection}. Now, we obtain these things indirectly from our other assumptions and the definition of A-architectures.

\begin{table}
{
\centering
\begin{tabular}{lcc}
Value & $f_l$ & Formula\\ \hline\hline
$\mathfrak{q}_l$&input&$q$ parameter of inputs\\
$\mathfrak{q}_l$&FC&$\sigma_l^2\mathfrak{q}_k$\\
$\mathfrak{q}_l$&activation&$\mathfrak{C}_{\tau_l}(\mathfrak{q}_k,\mathfrak{q}_k)$\\
$\mathfrak{q}_l$&bias&$\mathfrak{q}_k + \sigma_l^2$\\
$\mathfrak{q}_l$&LN&$1$\\
$\mathfrak{q}_l$&BN&$1$\\
$\mathfrak{q}_l$&addition&$\sum_{\kappa_l=1}^{K_l} w_{l,\kappa_l}^2\mathfrak{q}_{k_l[\kappa_l]}$\\
$\mathfrak{c}_l$&input&$c$ parameter of inputs\\
$\mathfrak{c}_l$&FC&$\sigma_l^2\mathfrak{c}_k$\\
$\mathfrak{c}_l$&activation&$\mathfrak{C}_{\tau_l}(\mathfrak{q}_k,\mathfrak{c}_k)$\\
$\mathfrak{c}_l$&bias&$\mathfrak{c}_k + \sigma_l^2$\\
$\mathfrak{c}_l$&LN&$\frac{\mathfrak{c}_k}{\mathfrak{q}_k}$\\
$\mathfrak{c}_l$&BN&$0$\\
$\mathfrak{c}_l$&addition&$\sum_{\kappa_l=1}^{K_l} w_{l,\kappa_l}^2\mathfrak{c}_{k_l[\kappa_l]}$\\
$\mathfrak{m}_l$&input&undetermined\\
$\mathfrak{m}_l$&FC&$0$\\
$\mathfrak{m}_l$&activation&$\mathbb{E}_{s \sim \mathcal{N}(0,\mathfrak{q}_k)}\tau_l(s)$\\
$\mathfrak{m}_l$&bias&$\mathfrak{m}_k$\\
$\mathfrak{m}_l$&LN&$0$\\
$\mathfrak{m}_l$&BN&$0$\\
$\mathfrak{m}_l$&addition&$\sum_{\kappa_l=1}^{K_l} w_{l,\kappa_l}\mathfrak{m}_{k_l[\kappa_l]}$\\
$\mathfrak{g}_l$&input&1\\
$\mathfrak{g}_l$&FC&$\sigma_l^2\mathfrak{g}_k$\\
$\mathfrak{g}_l$&activation&$\mathfrak{C}_{\tau_l'}(\mathfrak{q}_k,\mathfrak{q}_k)\mathfrak{g}_k$\\
$\mathfrak{g}_l$&bias&$\mathfrak{g}_k$\\
$\mathfrak{g}_l$&LN&$\frac{\mathfrak{g}_k}{\mathfrak{q}_k}$\\
$\mathfrak{g}_l$&BN&$\frac{\mathfrak{g}_k}{\mathfrak{q}_k-\mathfrak{c}_k}$\\
$\mathfrak{g}_l$&addition&$\sum_{\kappa_l=1}^{K_l} w_{l,\kappa_l}^2\mathfrak{g}_{k_l[\kappa_l]}$\\
\end{tabular}
\\
}
\caption{Recursive rules for calculating mean field limits in an A-architecture $f$.}
\label{tableNLCPropagation}
\end{table}

\paragraph{Limits} It is important that we take the limit over $d_\text{MF}$ first, before we take the limit over $N$. Therefore, we are not technically making statements about metric values with respect to $\mathcal{D}$ as $d_\text{MF}$ converges to infinity, but only about the metric values with respect to arbitrarily large samples of $\mathcal{D}$ as $d_\text{MF}$ converges to infinity. This is not entirely satisfying from a mathematical standpoint. However, it is practically sufficient because we can only ever apply a neural network to a finite number of inputs. \citet{meanFieldNetsorGP} also focused on arbitrarily large, finite sets of inputs. The reason for this ordering of the limit is because we follow a proof strategy as outlined in section \ref{covarianceKernelSection}, where we propagate the sample forward using the $N$-duplex of $f$. We can then evaluate means and co-means of that $N$-duplex to obtain the metric values with respect to the sample of size $N$. Exchanging the limits is likely possible, but would require a more powerful version of background theorem \ref{backgroundMaster}, which goes beyond the scope of this work.

The limit over $\vec{\epsilon}$ is of minor importance. We need positive regularizers to eliminate the singularity of the normalization operations, which can be considered a ``nuisance'' for theoretical analysis. If $\vec{\epsilon}$ is very small, as it is in practice, it has no significant impact on any observed phenomena.

As mentioned at the beginning of section \ref{meanFieldDistributionSection}, batch normalization with a finite batch size is a cumbersome operation from a mean field perspective. When theorem \ref{mfntPropagation} is applied to architectures with BN, we take the limit of batch size to infinity while taking the limit of $N$ to infinity. When neural networks with BN are deployed after training, the batch moments are generally replaced by training set moments, as we described in section \ref{batchWiseLayersSection}. This is similar to taking the batch size towards infinity. Unless very small batches are used during training, infinite batches capture the mean field behavior of the architecture adequately.

\paragraph{Limit quantities} $\mathfrak{m}_l$, $\mathfrak{q}_l$ and $\mathfrak{c}_l$ can be easily interpreted as layer mean, square mean and co-mean based on statements (\ref{eqn5p0}), (\ref{eqn5p1}) and (\ref{eqn5p2}). $\mathfrak{g}_l$ can be understood in light of statement (\ref{eqn5p4}). This yields $\lim\frac{1}{d_l^\text{MF}}||\mathcal{J}_{l,0}(x^{(1)})||_F^2 = \mathfrak{g}_l$. We further have $||\mathcal{J}_{l,0}(x^{(1)})||_F^2 = \mathbb{E}_u||\mathcal{J}_{l,0}(x^{(1)})u||_2^2$ when $u$ is unit Gaussian. Hence, $\mathfrak{g}_l$ can be regarded as the limit of the square mean of the gradient as it is {\it forward-propagated} from the input. Using forward-mode propagation instead of backpropagation for the gradient enables us to base theorem \ref{mfntPropagation} on background theorem \ref{backgroundMaster} instead of the master theorem of \citet{meanFieldNetsorNTK}. \citet{meanFieldNetsorNTK} requires more complex mean field architectures and stronger assumptions to model gradient backpropagation. This strategy also allows us to bypass the gradient of the loss function $\frac{d\ell}{df}$ which is not part of the mean field framework. \citet{meanFieldNetsorNTK} have to use a Gaussian approximation for the gradient of the loss function. We suspect that using forward-mode propagation, the tools and weaker assumptions from \citet{meanFieldNetsorGP} may actually be sufficient to obtain the results of \citet{meanFieldNetsorNTK}.

\paragraph{Statements} We will interpret the statements one by one.

\begin{itemize}[leftmargin=3cm]
\item[(\ref{eqn5p0})] This is equivalent to the first statement of background theorem \ref{backgroundMaster}.
\item[(\ref{eqn5p1}) and (\ref{eqn5p2})] This is equivalent to the second statement of background theorem \ref{backgroundMaster}. $\mathfrak{q}_l$ is the layer square mean and corresponds to $\mathfrak{c}_{l;l}$. $\mathfrak{c}_l$ is the layer co-mean and is obtained by propagating $x^{(1)}$ and $x^{(2)}$ jointly through the 2-duplex of $f$. $\mathfrak{c}_l$ corresponds to $\mathfrak{c}_{l;l'}$, where $f_l$ and $f_{l'}$ form a pair in the duplex.
\item[(\ref{eqn5p4})] As explained above, $\mathfrak{g}_l$ can be regarded as the square mean of the forward-propagated gradient. It ultimately derives from the second statement of background theorem \ref{backgroundMaster} when gradient propagation is cast as ``regular'' forward propagation in a mean field architecture (section \ref{mfntPropagationsection}). By definition, $\mathcal{J}_{l,0} = \mathcal{J}_{l,m}\mathcal{J}_{m,0}$ because $f_m$ is a bottleneck for $f_l$. Statement (\ref{eqn5p4}) yields $||\mathcal{J}_{l,0}(x^{(1)})||_F^2 \approx \mathfrak{g}_l$, $||\mathcal{J}_{m,0}(x^{(1)})||_F^2 \approx \mathfrak{g}_m$ and $||\mathcal{J}_{l,m}(x^{(1)})||_F^2 \approx \frac{\mathfrak{g}_l}{\mathfrak{g}_m}$. So statement (\ref{eqn5p4}) can be interpreted as saying that the Frobenius norm of the Jacobian is decomposable, similar to section \ref{nlcDecomposableSection}.
\item[(\ref{eqn5p5})] This is similar to (\ref{eqn5p1}), except that we take the expectation over $x$ in addition to the mean over $i$. Since $x$ is drawn from a finite sample, we simply take the mean over that sample, which works as expected.
\item[(\ref{eqn5p6})] This is an interesting statement that relates the co-mean of layer values to the magnitude of neuron expectations. It can be understood in light of proposition \ref{mfntElemLike}, which relates the co-mean of two vectors to the square mean of their componentwise expectations.
\item[(\ref{eqn5p7})] This follows directly from the previous two statements as $||\mathbb{S}_xf_l||_2^2 = \mathbb{E}_x||f_l||_2^2 - ||\mathbb{E}_xf_l||_2^2$.
\item[(\ref{eqn5p8})] This follows from (\ref{eqn5p4}) by taking the mean over a finite sample.
\item[(\ref{eqn5p9})] First, we observe $\Tr(\Cov_{f_m}) = ||\mathbb{S}_xf_m(x)||_2^2$ and hence $\lim \frac{1}{d_m^\text{MF}}\Tr(\Cov_{f_m}) = (\mathfrak{q}_m - \mathfrak{c}_m)$, according to statement (\ref{eqn5p6}). Second, we observe $\frac{1}{d_l^\text{MF}}\mathbb{E}_x\Tr(\mathcal{J}_{l,m}(x)\mathcal{J}_{l,m}(x)^T)=\frac{1}{d_l^\text{MF}}\mathbb{E}_x||\mathcal{J}_{l,m}||_F^2 \approx (\frac{1}{d_l^\text{MF}}\mathbb{E}_x||\mathcal{J}_{l,0}||_F^2)(\frac{1}{d_m^\text{MF}}\mathbb{E}_x||\mathcal{J}_{m,0}||_F^2)^{-1}$ if $\mathcal{J}_{m,0}$ is assumed to be uniformly random. Statement (\ref{eqn5p9}) is the amalgamation of both observations.
\item[(\ref{eqn5p10})] This is the square root of the ratio of (\ref{eqn5p9}) and (\ref{eqn5p7}).
\end{itemize}
 
\paragraph{Expansion-convergence and convergence in distribution results} These are straightforward applications of theorem \ref{mfntMetaGaussian} and background corollary \ref{backgroundKernel} respectively, where the abstract $\mathfrak{C}$ based on table \ref{tableBackgroundPropagation} has been replaced by concrete values based on table \ref{tableNLCPropagation}.

\paragraph{Table \ref{tableNLCPropagation}} We will begin with some high-level observations. 

$\mathfrak{g}_l$ always depends multiplicatively on $\mathfrak{g}_k$. This is because gradient propagation multiplies the gradient of each layer to the in-flowing gradient. If the magnitude of the in-flowing gradient changes by some factor, so does the magnitude of the out-flowing gradient. 

The recursion of $\mathfrak{q}$ does not depend on $\mathfrak{c}$. This is because $\mathfrak{q}$ is the limit of quantities that can be obtained by forward-propagating only a single input without regard to its relationship to other inputs, as long as $f$ does not use BN. And the BN operation sets $\mathfrak{q}$ to a fixed value regardless of $\mathfrak{c}$. Conversely, the recursion of $\mathfrak{c}$ does depend on $\mathfrak{q}$. This is because operations such as activation and LN depend upon the square mean of the dependency in a nonlinear fashion.

The recursion of $\mathfrak{q}$ and $\mathfrak{c}$ does not depend on $\mathfrak{m}$. This is because we chose the definition of A-architectures intentionally to eliminate this dependency. $\mathfrak{q}$ and $\mathfrak{c}$ are required to calculate the NLC limit, but $\mathfrak{m}$ is not. So A-architectures allow us to calculate the NLC with the simplest possible recursion. Both $\mathfrak{q}$ and $\mathfrak{c}$ would depend on $\mathfrak{m}$ if we allowed, say, activation layers followed by normalization layers or addition layers that depend on multiple activation layers.

In order for all statements of theorem \ref{mfntPropagation} to be valid, we require $\mathfrak{q}_l > \mathfrak{c}_l \ge 0$ and $\mathfrak{g}_l > 0$ for all $l$. Since we assume $q > c \ge 0$, we have $\mathfrak{q}_0 > \mathfrak{c}_0 \ge 0$ and table \ref{tableNLCPropagation} yields $\mathfrak{g}_0 =1 > 0$. It turns out that the recursion preserves this property from layer to layer. This is easy to see for non-activation layers. For activation layers, we have the following.

\begin{proposition} \label{covkerPositive}
Assume $\tau$ is continuous, non-constant and $q > c > -q$. Then $$\mathfrak{C}_{\tau}(q,q) > \mathfrak{C}_{\tau}(q,c)$$
\end{proposition}

\begin{proposition} \label{covkerPositive2}
For any $\tau$ and $q \ge c \ge 0$, we have $$\mathfrak{C}_{\tau}(q,c) \ge 0$$
\end{proposition}

\begin{proposition} \label{covkerPositive3}
Assume $\tau$ is continuous, not the zero function and $q > 0$. Then $$\mathfrak{C}_{\tau}(q,q) > 0$$
\end{proposition}

So, putting things together we have the following.

\begin{proposition} \label{mfntPositive}
Let $f$ be an A-architecture. Let $\mathfrak{q}_l$, $\mathfrak{c}_l$ and $\mathfrak{g}_l$ be defined according to table \ref{tableNLCPropagation} where $q > c \ge 0$. Then $\mathfrak{q}_l > \mathfrak{c}_l \ge 0$ and $\mathfrak{g}_l > 0$ for all $l$.
\end{proposition}

In general, the rules for $\mathfrak{q}$, $\mathfrak{c}$ and $\mathfrak{g}$ look similar to the definition of the operation itself, and similar to each other. This will become more apparent below, where we discuss the rules for each individual operation.

\begin{itemize}[leftmargin=2cm]
\item[input] The conditions on $\mathcal{D}$ were chosen specifically to yield a fixed $\mathfrak{q}$ and $\mathfrak{c}$ value.
\item[FC] Both the operation and its gradient apply a multiplicative transformation to a vector, where the scale of that transformation is controlled by $\sigma_l^2$. This is reflected in the recursive calculation rules. Because each weight matrix entry has mean zero, we obtain an $\mathfrak{m}$ value of zero.
\item[activation] We effectively replace $\tau$ with $\mathfrak{C}_\tau$ and $\tau'$ with $\mathfrak{C}_{\tau'}$ during input / gradient propagation respectively. In this way, the recursion mimics the layer operation. $\mathfrak{C}_\tau$ is defined in terms of an expectation over a Gaussian distribution, similar to $\mathbb{E}_e$ in table \ref{tableBackgroundPropagation}. We will dive deeper into this in section \ref{meanFieldActivationSection}.
\item[bias] Because the same bias vector is added to both $f_k(x^{(1)})$ and $f_k(x^{(2)})$, we experience the same increase in both the square mean and co-mean. Bias vector addition has a unit gradient, so there is no effect on $\mathfrak{g}_k$. Since bias vector components have mean zero, there is no effect on $\mathfrak{m}_k$.
\item[LN] Note that one property of A-architectures is that normalization layers cannot follow activation layers before another FC layer. Therefore, we have $\mathfrak{m}_k = 0$. So the mean subtraction part of LN has no effect on the limit and we obtain $\mathfrak{m}_l = 0$ regardless. Dividing the dependency by the standard deviation has the effect of dividing $\mathfrak{q}_k$, $\mathfrak{c}_k$ and $\mathfrak{g}_k$ by $\mathfrak{q}_k$.
\item[BN] The key difference with LN is that BN removes the neuron mean across inputs in a batch, whereas LN removes the layer mean for each input. So the subtraction part of BN affects $\mathfrak{c}_k$ instead of $\mathfrak{m}_k$, which may be nonzero. So we subtract $\mathfrak{c}_k$ from both $\mathfrak{q}_k$ and $\mathfrak{c}_k$ and then divide by $\mathfrak{q}_k - \mathfrak{c}_k$, which also affects the gradient.
\item[addition] We apply weighted addition to all limit quantities, where quantities that represent a square mean or co-mean require squaring the weights.
\end{itemize}

\paragraph{Scope} As discussed in section \ref{aArchitectureSection}, the proof of theorem \ref{mfntPropagation} requires us to re-cast segments of the layer graph of the A-architecture between successive fully-connected layers as a single elementwise layer in a mean field architecture. This process is highly dependent on the form that this segment takes. Properties \ref{aprop1}, \ref{aprop7}, \ref{aprop8}, \ref{aprop9} and \ref{aprop10} of A-architectures limit the range of segments that can occur in an A-architecture. While the complexity of segments for which we prove mean field limits, to our knowledge, goes beyond that of related work, we need to draw the line somewhere. As stated, we orient ourselves on the architectures we use for study A. Limiting the scope of the class of A-architectures also keeps the calculation rules of table \ref{tableNLCPropagation} simple and enables e.g. proposition \ref{mfntNPE}. 

If we did not have property \ref{aprop9}, it would be possible to have consecutive activation layers in an A-architecture. The first activation layers could induce non-Gaussian neuron distributions. Then it would no longer be possible to model the dependency of the second activation layer as Gaussian, as is done by the covariance kernel $\mathfrak{C}_\tau$. In table \ref{tableNLCPropagation}, the recursion for $\mathfrak{q}$, $\mathfrak{c}$, and $\mathfrak{g}$ do not depend on $\mathfrak{m}$. Without property \ref{aprop8}, it would be possible to have an addition layer that adds together two activation layers, which can have non-zero $\mathfrak{m}$ values. Then, depending on whether those $\mathfrak{m}$ values e.g. have the same or different signs, they could cancel out or amplify in the addition layer, which would have a knock-on effect on e.g. $\mathfrak{q}$. The calculation rules for addition layers in table \ref{tableNLCPropagation} assume that the weighted sum is taken over essentially independent layers. Without property \ref{aprop10}, this would not be the case as there could be an addition layer that e.g. adds a layer to itself.

Properties \ref{aprop4}, \ref{aprop5} and \ref{aprop6} are basic requirements for applying mean field theory. Property \ref{apropX1} is technical and ensures that normalization operations do not diverge for e.g. zero input vectors. We suspect property \ref{aprop7} is unnecessary in that theorem \ref{mfntPropagation} as stated would hold without it. However, BN and LN layers require somewhat different proof strategies. We did not consider them jointly to keep the complexity and length of the proof at its current level. Property \ref{aprop3} is a convenience. Any layer graph edge corresponding to a zero addition weight can simply be removed.

The aspect of property \ref{aprop2} that requires differentiable activation functions is necessary to state theorem \ref{mfntPropagation} in terms of gradients. We use the second derivative of activation functions to prove parameter-control for layer graph segments containing LN and activation layers. A weaker assumption, such as a locally Lipschitz derivative, might also suffice. Finally, assuming control by $\mathcal{C}^\text{E2}$ is necessary to apply background theorem \ref{backgroundMaster} and its corollaries.

To fully appreciate the relevance of each of the properties of an A-architecture for theorem \ref{mfntPropagation}, it is necessary to study the proof (sections \ref{mfntPropagationsection}, \ref{mfntPropagationBNsection}).

The requirement that $f_m$ is a bottleneck for $f_l$ can likely be eliminated if we set all $\mathfrak{g}$ values corresponding to layers that are not descendants of $f_m$ to zero and apply all other calculation rules as normal.

\subsection{Empirical analysis} \label{meanFieldPracticalEmpiricalSection}

\begin{table}
{
\centering \small
\begin{tabular}{p{3.2cm}cccc}
Metric name & Notation & Definition & Statement & Limit\\ \hline\hline
Layer quadratic mean & $LQM_l(f,x)$ & $\sqrt{\mathbb{E}_if_l(x)[i]^2}$ & (\ref{eqn5p1})& $\sqrt{\mathfrak{q}_l}$\\
Jacobian Frobenius norm & $JACF_{l,m}(f,x)$ & $\frac{1}{\sqrt{d_l}}||\mathcal{J}_{l,m}(x)||_F$ & (\ref{eqn5p4})& $\sqrt{\frac{\mathfrak{g}_l}{\mathfrak{g}_m}}$\\
Layer scale & $LSCALE_l(f,\mathcal{D})$ & $\sqrt{\frac{1}{d_l}\mathbb{E}_x||f_l||^2_2}$ & (\ref{eqn5p5})& $\sqrt{\mathfrak{q}_l}$\\
Quadratic mean of neuron expectations & $QMNEX_l(f,\mathcal{D})$ & $\frac{1}{\sqrt{d_l}}||\mathbb{E}_xf_l||_2$ & (\ref{eqn5p6})& $\sqrt{\mathfrak{c}_l}$\\
Quadratic mean of neuron standard deviations & $QMNSTD_l(f,\mathcal{D})$ & $\frac{1}{\sqrt{d_l}}||\mathbb{S}_xf_l||_2$ & (\ref{eqn5p7})& $\sqrt{\mathfrak{q}_l - \mathfrak{c}_l}$\\
Nonlinearity coefficient numerator & $NLCNUM_{l,m}(f,\mathcal{D})$ & $\sqrt{\frac{1}{d_l}\mathbb{E}_x\Tr(\mathcal{J}_{l,m}\Cov_{f_m}\mathcal{J}_{l,m}^T)}$ & (\ref{eqn5p9})& $\sqrt{\frac{\mathfrak{g}_l(\mathfrak{q}_m - \mathfrak{c}_m)}{\mathfrak{g}_m}}$\\
Nonlinearity coefficient & $NLC_{l,m}(f,\mathcal{D})$ & $\sqrt{\frac{\mathbb{E}_x\Tr(\mathcal{J}_{l,m}\Cov_{f_m}\mathcal{J}_{l,m}^T)}{\Tr(\Cov_{f_l})}}$ & (\ref{eqn5p10})& $\sqrt{\frac{\mathfrak{g}_l(\mathfrak{q}_m - \mathfrak{c}_m)}{\mathfrak{g}_m(\mathfrak{q}_l - \mathfrak{c}_l)}}$\\
\end{tabular}
\\
}
\caption{Metrics used to determine the practical predictiveness of theorem \ref{mfntPropagation}. For each metric, we give the statement in theorem \ref{mfntPropagation} it corresponds to as well as its mean field limit.}
\label{tableMetricsMfEmpirical}
\end{table}

In this subsection, we validate theorem \ref{mfntPropagation} empirically. Specifically, we focus on statements (\ref{eqn5p1}), (\ref{eqn5p4}) through (\ref{eqn5p7}), (\ref{eqn5p9}) and (\ref{eqn5p10}). Combining those statements with the analysis done in section \ref{meanFieldDistributionEmpiricalSection} validates the expansion-covergence clause at the end of theorem \ref{mfntPropagation}. Each statement (\ref{eqn5p0}) through (\ref{eqn5p10}) is an equality statement with a left- and right-hand side. The left-hand side is a limit of a metric defined in terms of propagation through a network of variable but finite width. The right-hand side contains expressions of ``limit quantities'', which are defined via table \ref{tableNLCPropagation} and do not require any network or layer evaluation. Below, we discuss how we compute a representative value for each side. Then, we present results.

\paragraph{Computing the finite width metrics} To compute values for the left-hand side of the theorem statements, we simply ignore the limit. This requires choosing values for the quantities over which the limit is taken. We replace $\mathcal{D}^{(N)}$ with $\mathcal{D}$ and set $d_\text{MF} = 1$. We use a batch size of 250 and set $\vec{\epsilon}$ small enough so that its floating-point representation is zero, as usual. This turns the left-hand side into metrics of fixed width networks under the framework of section \ref{neuralNetworkNotationSection} as we computed them throughout this work. The metrics we compute are given in table \ref{tableMetricsMfEmpirical}.

\begin{metricDefinition}
See table \ref{tableMetricsMfEmpirical}.
\end{metricDefinition}

Each metric corresponds to one of the statements from theorem \ref{mfntPropagation}. In table \ref{tableMetricsMfEmpirical}, we include additional square roots. Taking the root on both sides of a theorem statement does not alter its validity, but makes the quantities a bit more intuitive. (In general, we prefer means over squares of means.)

\paragraph{Computing mean field metrics} To compute values for the right-hand side of the theorem statements, we have to use table \ref{tableNLCPropagation}. A few issues arise. (i) Our datasets are not exactly elem-like and thus do not come with $q$ and $c$ parameters. Because our datasets are normalized to have component means of 0 and an average neuron variance of 1 across the union of training and validation set, we set $q=1$ and $c=0$ in table \ref{tableNLCPropagation}. See also section \ref{elemLikeSection}. (ii) As in section \ref{meanFieldDistributionEmpiricalSection}, we want to conduct experiments in the initial state, when weight matrices are not Gaussian initialized, and in the final state. Hence, we do not generally have explicit access to the variance parameters as strictly defined in section \ref{aArchitectureSection}. Hence, we replace $\sigma_l^2$ in table \ref{tableNLCPropagation} with its finite-width estimate, i.e. $\frac{1}{d_l}||W_l||^2_F$ for FC layers and $\frac{1}{d_l}||\beta_l||^2_2$ for bias layers. For Gaussian initialized bias and especially FC layers, this estimate would be very close to $\sigma_l^2$ for practical widths, due to the law of large numbers.

\begin{metricDefinition}
The `mean field metrics' $\hat{\mathfrak{m}}_l(f,\theta,q,c)$, $\hat{\mathfrak{q}}_l(f,\theta,q,c)$, $\hat{\mathfrak{c}}_l(f,\theta,q,c)$ and $\hat{\mathfrak{g}}_l(f,\theta,q,c)$ of an architecture $f$, parameter value $\theta$ and scalars $q > c \ge 0$ are calculated via table \ref{tableNLCPropagation}, where we replace $\sigma_l^2$ with $\frac{1}{d_l}||W_l||^2_F$ for FC layers and $\frac{1}{d_l}||\beta_l||^2_2$ for bias layers. For all these metrics, the default value of $q$ is 1 and the default value of $c$ is zero. $f$ must be composed of the layer operations in table \ref{tableNLCPropagation} for the metrics to be valid. Finally, using the same function arguments on both sides, we write

$$\hat{\mathfrak{n}}_{l,m}(f,\theta,q,c) = \sqrt{\frac{\hat{\mathfrak{g}}_l(\hat{\mathfrak{q}}_m - \hat{\mathfrak{c}}_m)}{\hat{\mathfrak{g}}_m(\hat{\mathfrak{q}}_l - \hat{\mathfrak{c}}_l)}}$$
\end{metricDefinition}

Mean field metrics are different from the metrics we defined previously in this work. They depend not on the network $f$ via function evaluation, but on the architecture $f$ via recursive calculation rules that utilize the architecture definition. Computing these metrics poses somewhat different challenges. Instead of statistical estimation, we have to use numerical integration for the Gaussian expectation inside $\mathfrak{C}_\tau$. While this works well enough for a single evaluation of $\mathfrak{C}_\tau$, it can create problems when we iterate many times for a deep network. Under Gaussian instability, even 64-bit floating-point rounding errors can grow to the point of overflow, though this does not happen in our experiments. We further discuss this in section \ref{actFunLengthKernelSection}. See also section \ref{metricEstimationSection}.

\begin{figure}[H]
\centering
\includegraphics[width=0.98\textwidth]{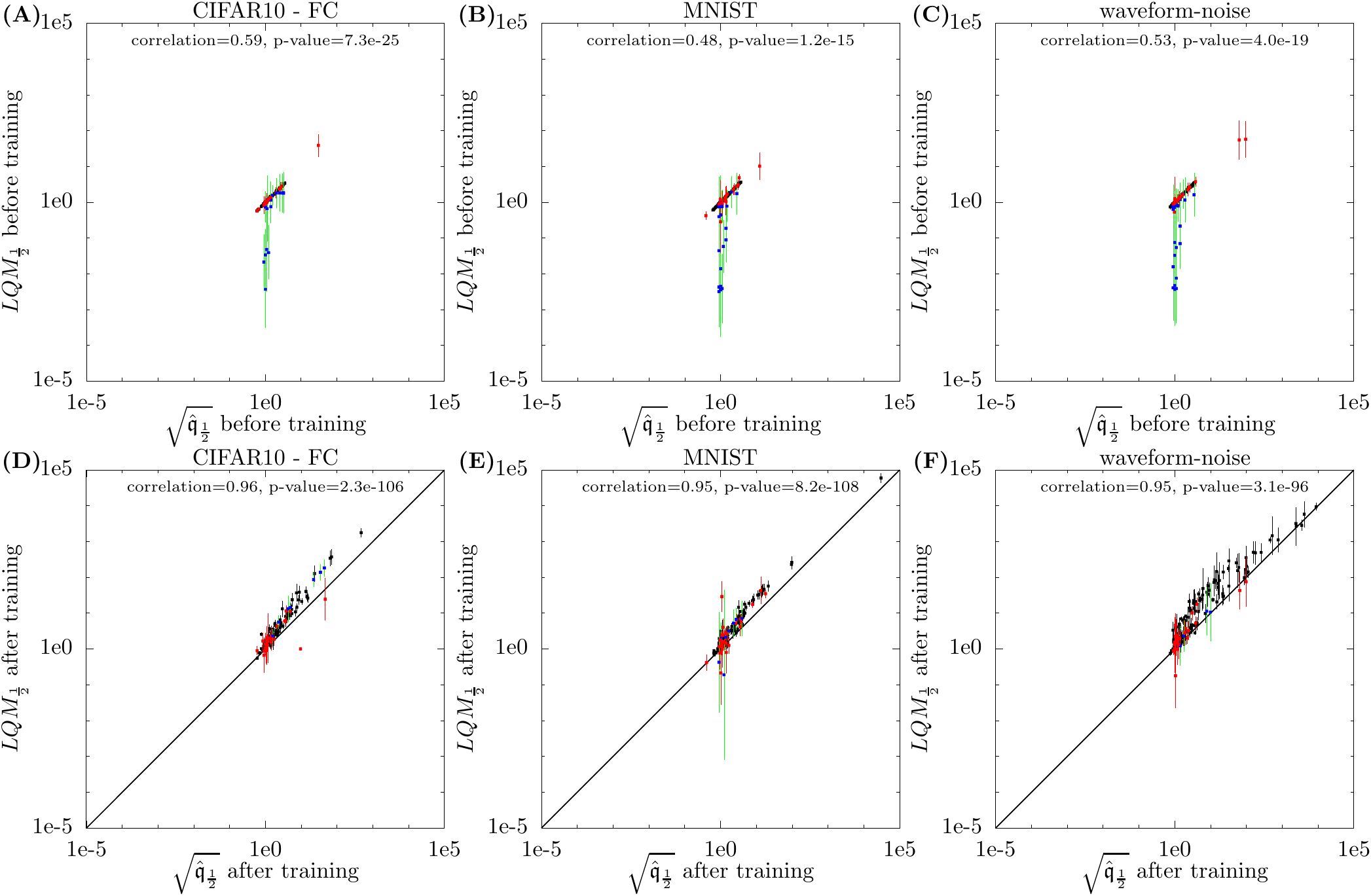}
\caption{LQM vs its mean field estimate for study A architectures. The metrics are evaluated at a fully-connected or addition layer halfway through the network, as in figures in section \ref{meanFieldDistributionEmpiricalSection}, because of the narrowness of the output layer. Vertical lines correspond to confidence intervals of 2 standard deviations. GUAs are depicted as blue points with green lines. GEAs are depicted as red points with red lines. Both GUAs and GEAs are displayed in the foreground as throughout chapter \ref{nlcChapter}, i.e. their markers fully or partially occlude the markers of stable architectures when they overlap. Some graph diagonals are given in black as a visual aid. Note that all x- and y-axis ranges of graphs across this figure, as well as across each of the figures below, are identical to enable easier comparison. {\it Conclusion:} The mean field estimate is highly accurate for stable architectures and GEAs before training and still relatively accurate after training.} \label{mfPredfq}
\end{figure}

\begin{figure}[H]
\centering
\includegraphics[width=0.98\textwidth]{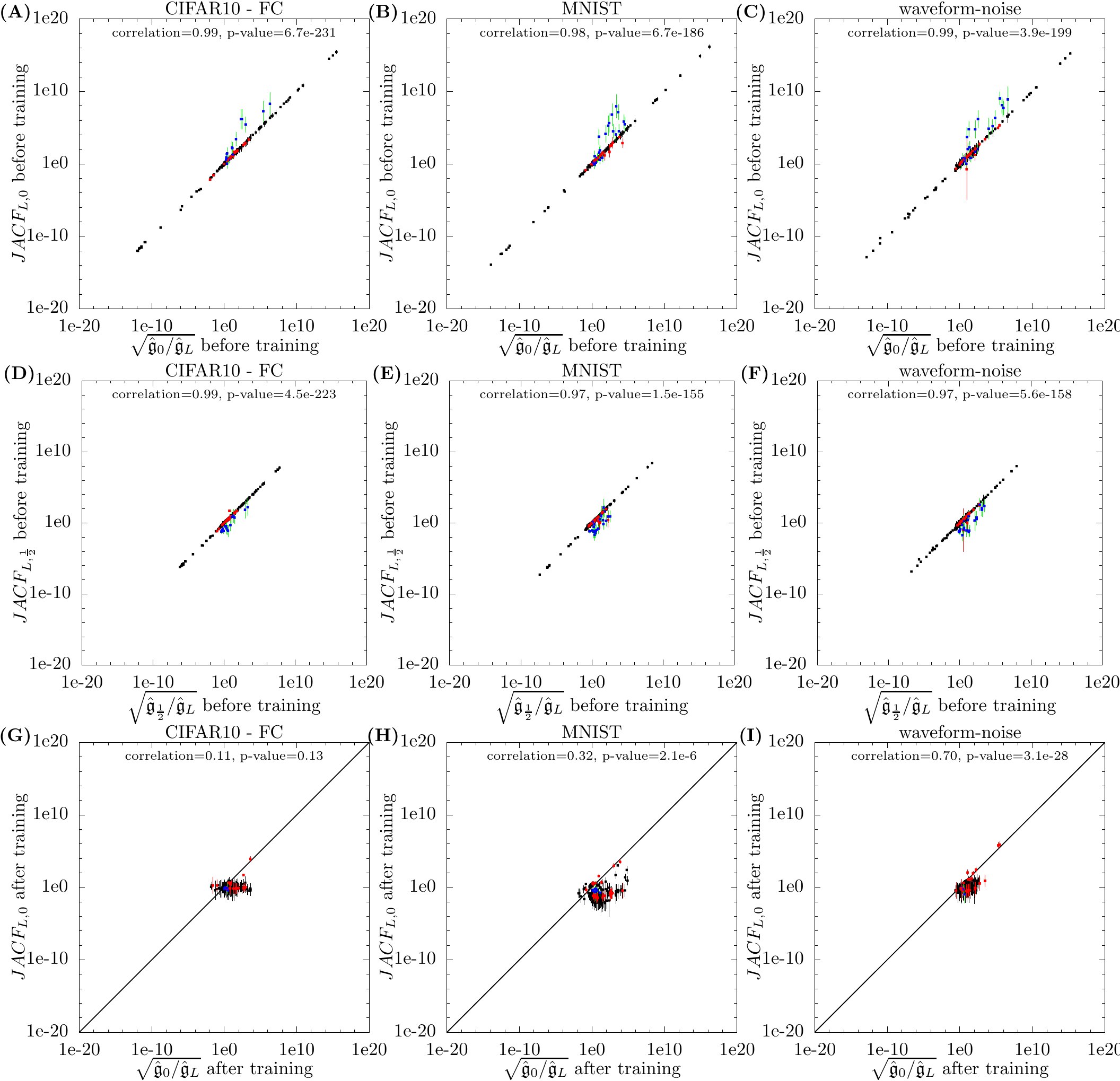}
\caption{JACF vs its mean field estimate at an intermediate and at the input layer. Graphs are analogous to figure \ref{mfPredfq}. {\it Conclusion:} The mean field estimate is highly accurate for stable architectures and GEAs before training, but not after training.} \label{mfPredjac}
\end{figure}

\begin{figure}[H]
\centering
\includegraphics[width=0.98\textwidth]{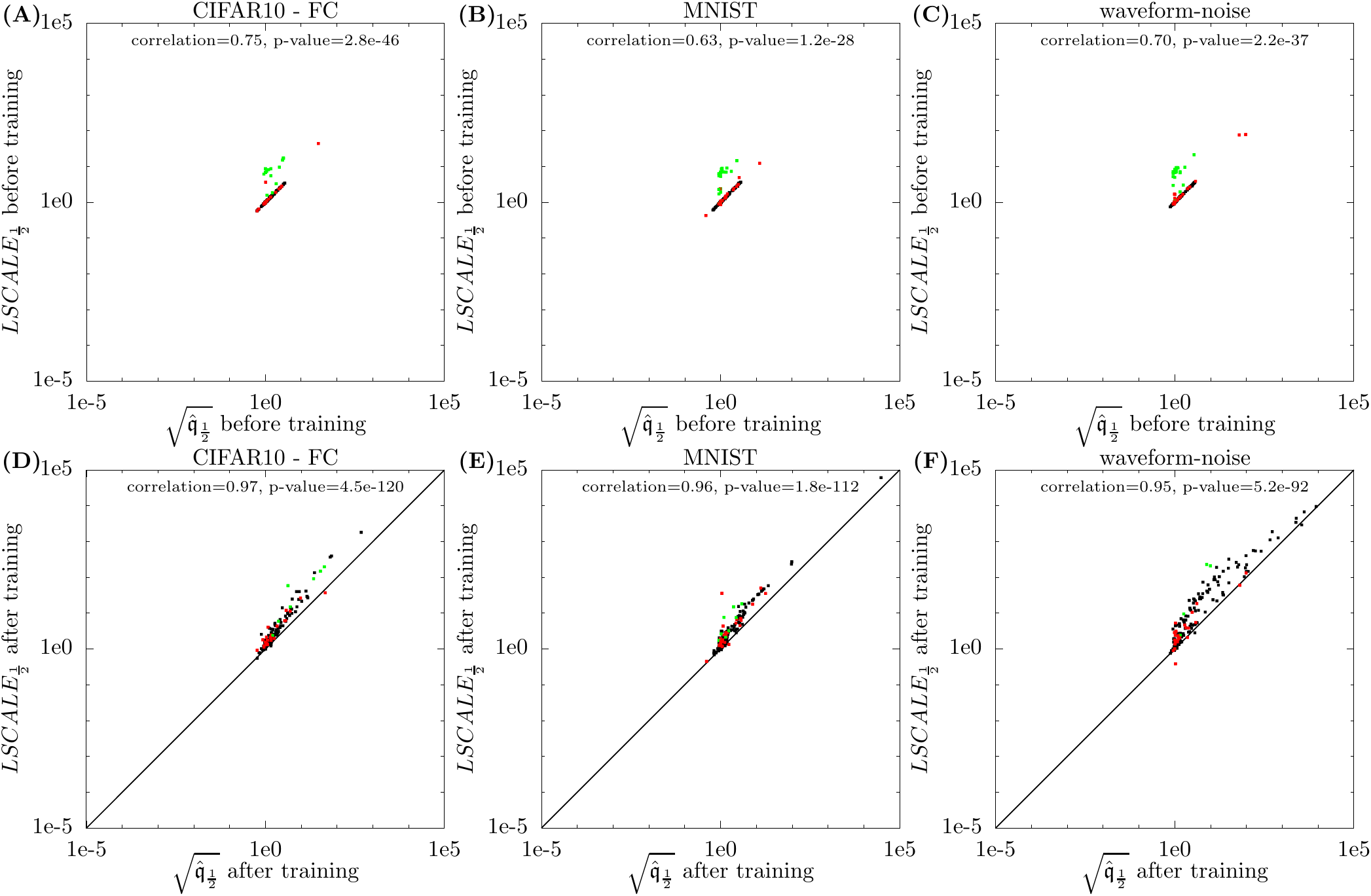}
\caption{LSCALE vs its mean field estimate at an intermediate layer. GUAs are depicted in green. GEAs are depicted in red. Both are displayed in the foreground. There are no confidence intervals. Graphs are otherwise analogous to previous figures. {\it Conclusion:} The mean field estimate is highly accurate for stable architectures and almost all GEAs before training, and still relatively accurate after training.} \label{mfPredqf}
\end{figure}

\begin{figure}[H]
\centering
\includegraphics[width=0.98\textwidth]{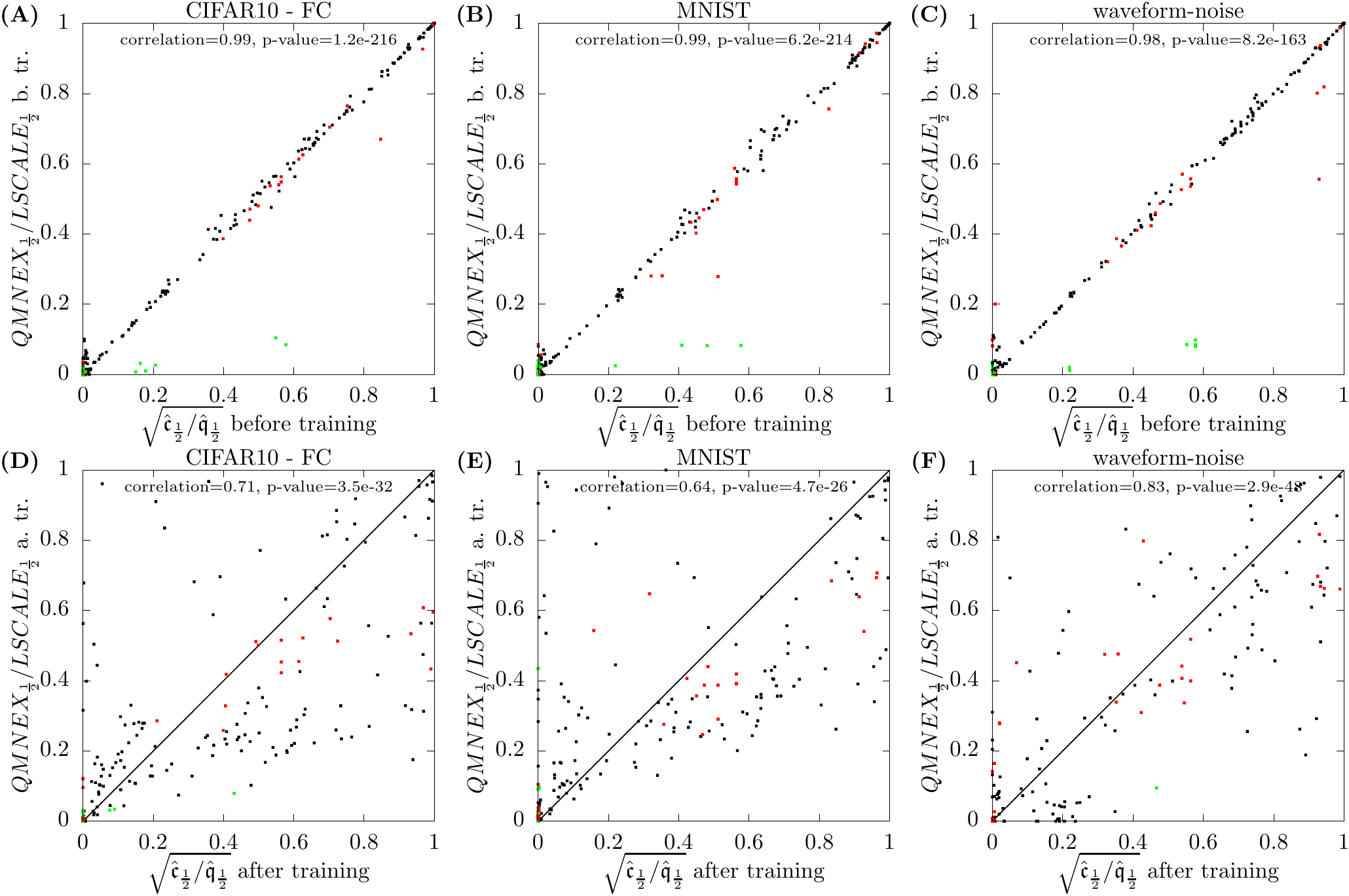}
\caption{$\frac{QMNEX}{LSCALE}$ vs its mean field estimate at an intermediate layer. Graphs are analogous to previous figures. {\it Conclusion:} The mean field estimate is highly accurate for stable architectures and fairly accurate for GEAs before training, but not after training.} \label{mfPredqmu}
\end{figure}

\begin{figure}[H]
\centering
\includegraphics[width=0.98\textwidth]{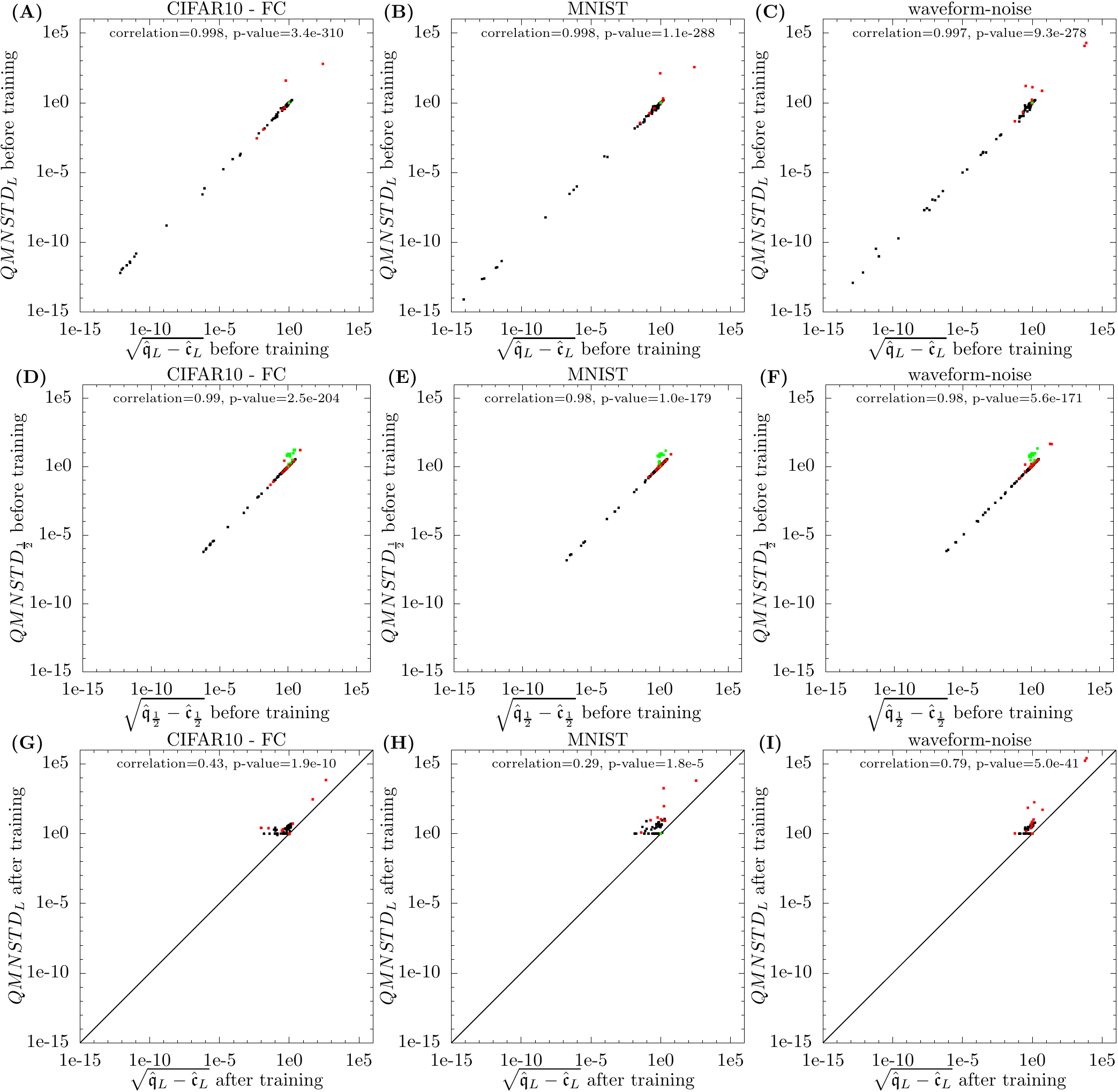}
\caption{QMNSTD vs its mean field estimate at an intermediate layer and at the output later. Graphs are analogous to previous figures. {\it Conclusion:} The mean field estimate is highly accurate for stable architectures and almost all GEAs before training. After training, the range of QMNSTD values is small, so it is difficult to assess accuracy.} \label{mfPredqsigma}
\end{figure}

\begin{figure}[H]
\centering
\includegraphics[width=0.98\textwidth]{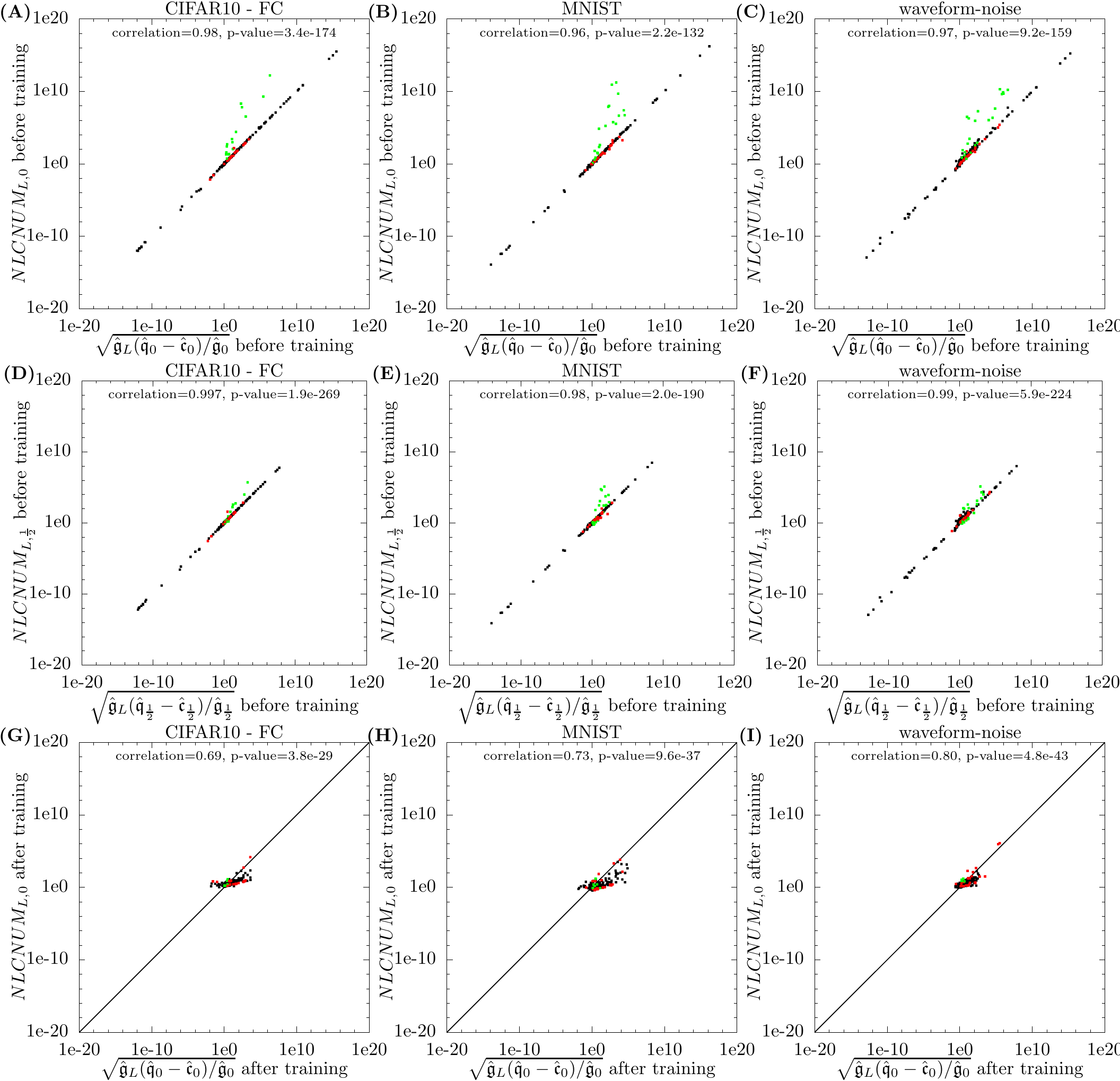}
\caption{NLCNUM vs its mean field estimate, for the whole network and for the second half of the network. Graphs are analogous to previous figures. {\it Conclusion:} The mean field estimate is highly accurate for stable architectures and GEAs before training. After training, the mean field estimate tends to overestimate NLCNUM, but is still highly correlated with it.} \label{mfPredjastq}
\end{figure}

\begin{figure}[H]
\centering
\includegraphics[width=0.98\textwidth]{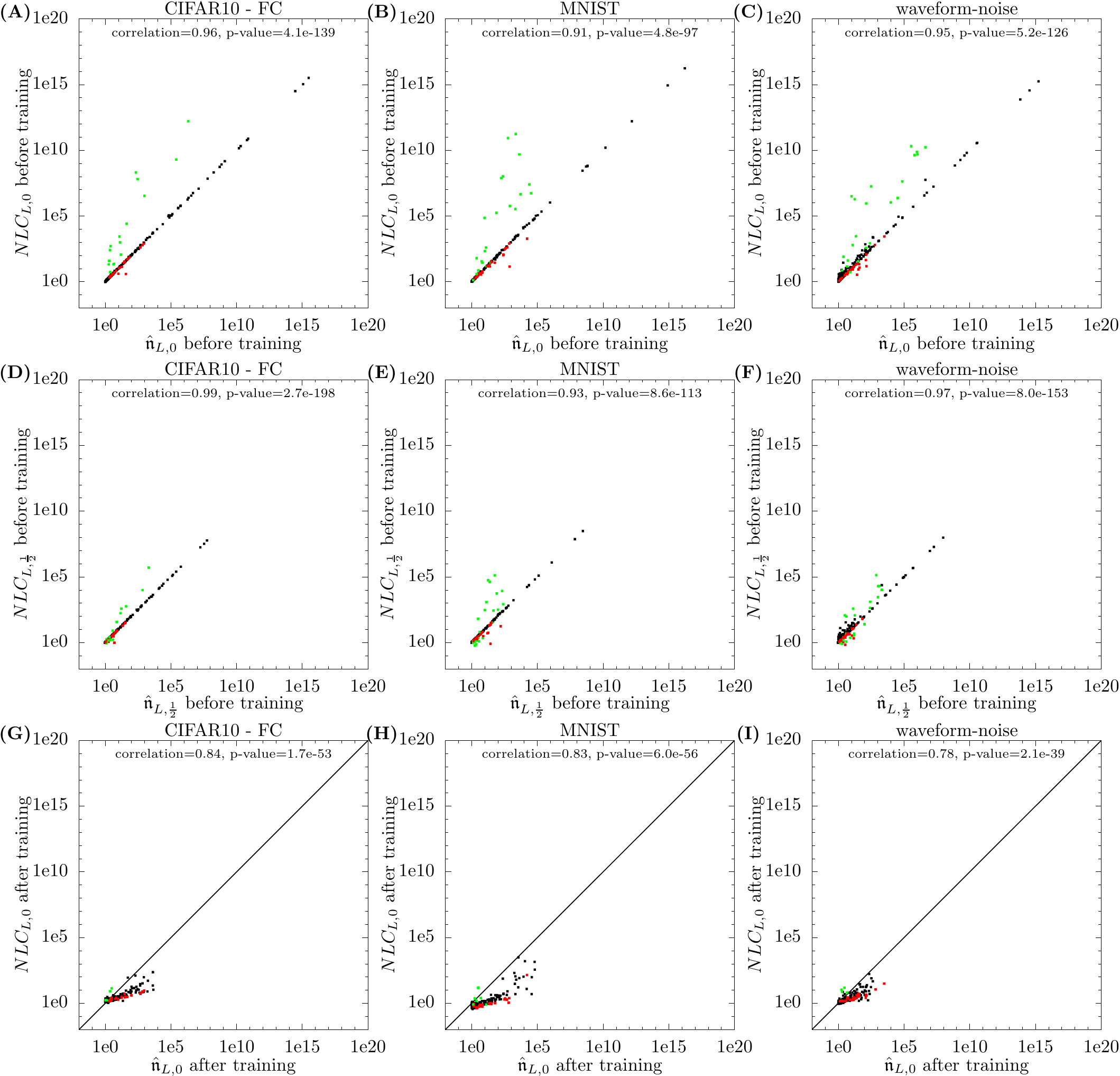}
\caption{NLC vs its mean field estimate, for the whole network and for the second half of the network. Graphs are analogous to previous figures. {\it Conclusion:} The mean field estimate is highly accurate for stable architectures and GEAs before training. After training, the mean field estimate usually overestimates the NLC, but it is still highly correlated with it.} \label{mfPredNLC}
\end{figure}

\paragraph{Mean field theory predicts finite width metrics} We plot the finite width metrics from table \ref{tableMetricsMfEmpirical} vs their estimate based on mean field metrics in figures \ref{mfPredfq}, \ref{mfPredjac}, \ref{mfPredqf}, \ref{mfPredqmu}, \ref{mfPredqsigma}, \ref{mfPredjastq} and \ref{mfPredNLC}. In all cases, we depict the estimate on the x-axis and the finite width metric on the y-axis. Each figure corresponds to one finite width metric. Note that we do not present the value of QMNEX explicitly, but via the ratio $\frac{QMNEX}{LSCALE}$. This is because computing the neuron expectation incurs an estimation error proportional to the neuron standard deviation. Therefore when $QMNEX << QMNSTD$ we cannot hope to compute QMNEX accurately in absolute terms, but only relative to QMNSTD or LSCALE.
 
We find that \finding{all mean field estimates are highly accurate when three conditions hold. (i) The architecture is in the initial state, (ii) we use CIFAR-10 or MNIST and (iii) the architecture is stable, i.e. depicted in black}. Consider especially figure \ref{mfPredjac}(A-B). There, we plot the network Jacobian Frobenius norm along with a confidence interval of +/-2 standard deviations. We find that \finding{even at +/-2 standard deviations, JACF is very close to the mean field estimate, even though the range of JACF values obtained for different architectures spans 30 orders of magnitude!} Consider also figure \ref{mfPredNLC}(A-B).  We find that \finding{mean field theory predicts the NLC of stable architectures highly accurately in the initial state, and still reasonably accurately for GEAs. It only fails for GUAs},  which do not use popular activation functions. This means that we can use table \ref{tableNLCPropagation} as a neural architecture design guide. This is the central insight of this chapter.

Let's look at the three factors we mentioned above that cause the degradation of mean field predictiveness. (i) We find that \finding{estimates for waveform-noise are somewhat less accurate than estimates for the other datasets for most metrics, including the NLC}. This is likely explained by the lower input and output dimensionality, which caused irregular behavior in e.g. section \ref{nlcBackpropSection}, and / or the lower dataset size which induced noise e.g. in section \ref{nlcRobustDataSection}. (ii) As usual, \finding{GUAs and GEAs are less well-behaved}. Of course, in table \ref{tableNLCPropagation} the limits of activation layers model the neurons in the dependency as Gaussian. Therefore, if they are less Gaussian in practice, as we showed in section \ref{meanFieldDistributionEmpiricalSection}, we can expect the theory to be less predictive. \finding{GEAs, depicted in red, are still estimated relatively accurately in general}. Note, however, that \finding{GEAs have some of the biggest confidence intervals in figures \ref{mfPredfq} and \ref{mfPredjac}}. (iii) As usual, architectures are not as well-behaved after training. However, it is worth noting that \finding{the accuracy of mean field estimates after training does differ significantly from metric to metric}. Specifically, \finding{LQM (figure \ref{mfPredfq}) and LSCALE (figure \ref{mfPredqf}) are still estimated quite well}. We conjecture that this is because the architecture does not have an incentive to modify layer quadratic means during training. The architecture does have an incentive to attain a low NLC after training for reasons discussed in sections \ref{nlcEvolutionSection} and \ref{bestNlcSection}. This conjecture is supported by figure \ref{mfPredNLC}, where we find that \finding{mean field theory overestimates the NLC after training}. More generally, we find that \finding{gradient-related metrics (figures \ref{mfPredjac}, \ref{mfPredjastq}) are overestimated and QMNSTD (figure \ref{mfPredqsigma}), which is the denominator of the NLC, is underestimated}.

We include correlation values in the figures. As usual, the caveat applies that the exact correlation values are significantly influenced by the frequency of GUAs and GEAs among architectures, as well as by only considering architectures that attained a non-random validation error after training in the bottom row of each figure.

\begin{figure}[H]
\centering
\includegraphics[width=0.98\textwidth]{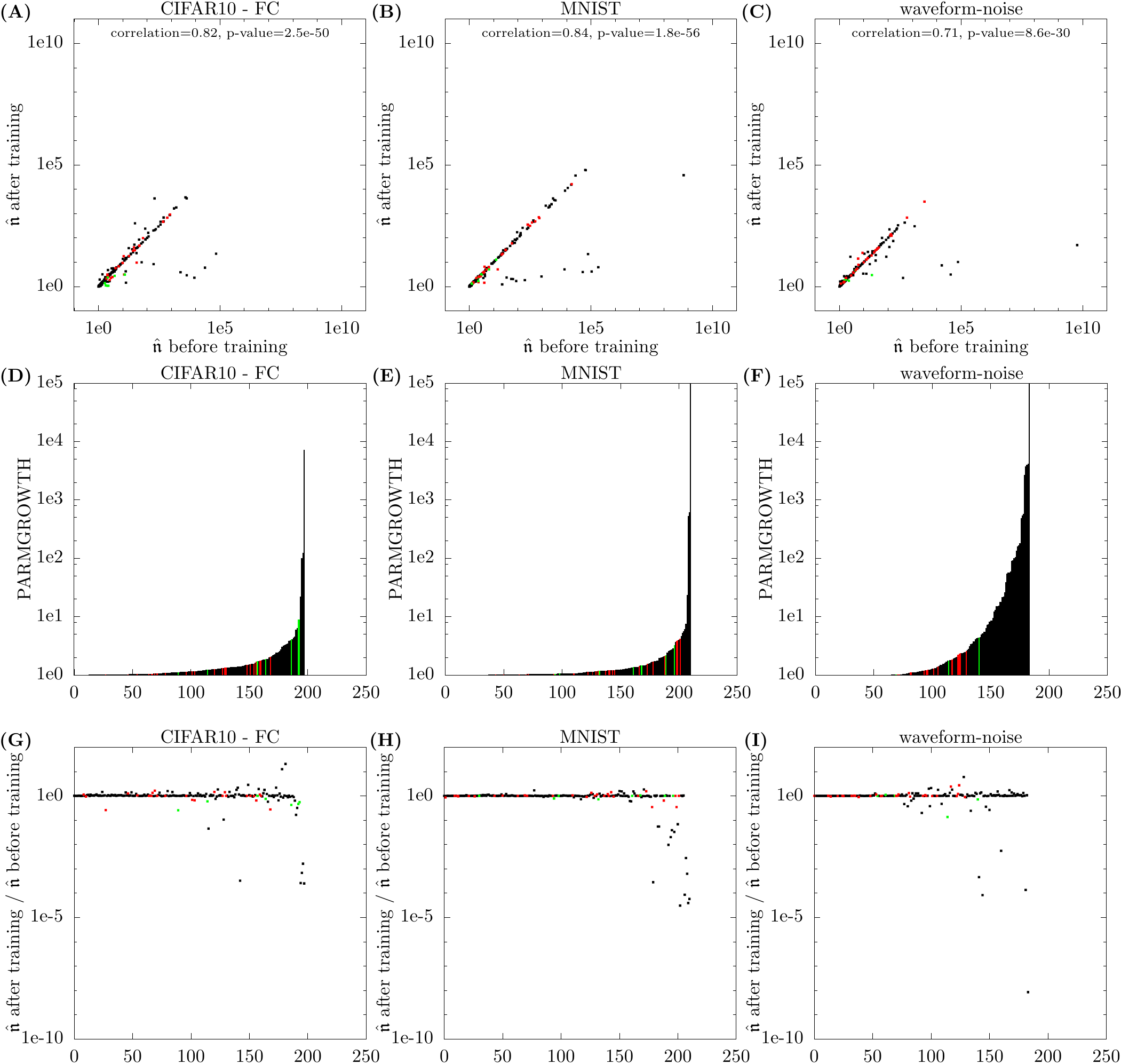}
\caption{Metric values for study A architectures. In graphs A-C, we plot the initial vs final $\tilde{\mathfrak{n}}$. In graphs D-F, we plot PARMGROWTH for study A architectures, sorted on the x-axis in ascending order. In graphs G-I, we plot the ratio of initial over final $\tilde{\mathfrak{n}}$. Architectures are placed on the x-axis in the same order as in graphs D-F. {\it Conclusion:} $\tilde{\mathfrak{n}}$ is stable for most architectures, but changes drastically for some. This change is generally associated with significant PARMGROWTH.} \label{mfEvolution}
\end{figure}

\paragraph{The mean field estimate of the NLC usually stays relatively constant during training, but sometimes changes drastically} In figure \ref{mfEvolution}(A-C), we plot the mean field estimate of the NLC before vs after training. We find that \finding{the value changes very little for most architectures}. Of course, the only way in which training can impact the mean field estimate is via a change of weight matrix and / or bias vector magnitude. 

\begin{metricDefinition}
The `parameter growth' (PARMGROWTH) of a final parameter value $\theta^{(T)}$ relative to an initial parameter value $\theta^{(0)}$ is

$$PARMGROWTH(\theta^{(0)}, \theta^{(T)}) = \frac{||\theta^{(T)}||_2}{||\theta^{(0)}||_2}$$

\end{metricDefinition}

In figure \ref{mfEvolution}(D-F), we plot PARMGROWTH. We find that \finding{the parameter length never decreases more than a tiny amount, and it stays approximately constant for a majority of architectures}. However, \finding{for some architectures it increases by orders of magnitude}. In figure \ref{mfEvolution}(G-I), we confirm that \finding{significant changes to the mean field estimate of the NLC only happen when the parameter length undergoes a large relative change, except for a very small number of architectures}. Even \finding{when the parameter length changes drastically, the mean field estimate often does not}. It is worth noting that the mean field estimate of the NLC is actually independent of the parameter in study A architectures when batch normalization is used. Also, PARMGROWTH is highly dependent on the learning rate. While we chose the starting learning rate in a very systematic fashion (section \ref{studyATrainingSection}), there were sometimes wide ranges of starting learning rates that yielded very similar performance, such that choosing from within this range was essentially random. However, this choice still affects PARMGROWTH enormously.

Overall, the inaccuracy of the mean field estimate of the NLC after training is not caused by a change in the value of the estimate, as it is approximately constant for most architectures.

\newpage

\newpage

\newpage

\subsection{The mean field NLC and nonlinearity path equation} \label{pathEquationSection}

While each recursion rule in table \ref{tableNLCPropagation} makes sense by itself, taken together they still appear somewhat ``black-box'', i.e. it is not clear whether there is a simple relationship between specific qualitative architecture properties, such as depth, activation functions or skip connections, and the mean field limits of important metrics. In this section, we show that we can indeed express the mean field estimate of the NLC in an incredibly instructive way. This allows us to ultimately explain the nonlinearity of an architecture using mean field theory and the architecture definition.

\begin{metricDefinition}
The `mean field NLC' $\mathfrak{n}_{l,m}$ of an A-architecture $f$ and scalars $q > c \ge 0$ is

$$\mathfrak{n}_{l,m}(f,q,c) = \sqrt{\frac{\mathfrak{g}_l(\mathfrak{q}_m - \mathfrak{c}_m)}{\mathfrak{g}_m(\mathfrak{q}_l - \mathfrak{c}_l)}}$$

where the limit quantities on the right-hand side are calculated according to table \ref{tableNLCPropagation}. The variance parameters are considered part of the architecture definition as usual. We write $\mathfrak{n}$ short for $\mathfrak{n}_{L,0}$ for the ``mean field NLC of $f$''.
\end{metricDefinition}

Going forward, we will use the term `mean field NLC' interchangeably for $\mathfrak{n}_{l,m}$ and $\hat{\mathfrak{n}}_{l,m}$, as defined in the previous subsection, as both have nearly identical values for practical A-architectures in the initial state without severe Gaussian instability.

We can immediately observe that $\mathfrak{n}_{l,m}$ is decomposable in the sense of section \ref{nlcDecomposableSection}, i.e. if we break down the section of the architecture between $f_m$ and $f_l$ into a series of segments where the endpoints of those segments are bottlenecks, $\mathfrak{n}_{l,m}$ is the product of mean field NLCs of these segments. Specifically, given a chain of layers in which each has a single dependency, the mean field NLC of that chain is equal to the product of mean field NLCs of each individual layer.

We also observe that the mean field NLC of individual layers has a simple form. For FC, bias, LN and BN layers, it is equal to 1. For addition layers with a single dependency, it is equal to 1. This is expected for operations that are linear functions of the dependency, like FC, bias and addition, because of proposition \ref{finiteNetNlcEquals1}. Informally, we can say that LN and BN layers are approximately linear for large width and batch sizes. To see this, consider that subtracting the mean of a vector is linear. Dividing by the standard deviation projects the values of the dependency onto a lower-dimensional space which is curved in only a single dimension per neuron (BN) / input (LN). Therefore, the nonlinearity of that space becomes increasingly negligible as width / batch size increases.

If $f_l$ is an activation layer, we have

$$\mathfrak{n}_{l,k} = \sqrt{\frac{(\mathfrak{q}_k - \mathfrak{c}_k)\mathfrak{C}_{\tau_l'}(\mathfrak{q}_k,\mathfrak{q}_k)}{\mathfrak{C}_{\tau_l}(\mathfrak{q}_k,\mathfrak{q}_k) - \mathfrak{C}_{\tau_l}(\mathfrak{q}_k,\mathfrak{c}_k)}}$$

Hence, we define the following.

\begin{definition}
The `activation function NLC' of an activation function $\tau$ and scalars $q > c \ge 0$ is

$$\mathfrak{n}_\tau(q,c) =  \sqrt{\frac{(q - c)\mathfrak{C}_{\tau'}(q,q)}{\mathfrak{C}_{\tau}(q,q) - \mathfrak{C}_{\tau}(q,c)}}$$

We write $\mathfrak{n}_\tau(c)$ short for $\mathfrak{n}_\tau(1,c)$. When $c=0$, the formula simplifies to

$$\mathfrak{n}_\tau(q,0) =  \sqrt{\frac{q\mathbb{E}_{s\sim\mathcal{N}(0,q)}\tau'(s)^2}{\mathbb{E}_{s\sim\mathcal{N}(0,q)}\tau(s)^2 - (\mathbb{E}_{s\sim\mathcal{N}(0,q)}\tau(s))^2}}$$

\end{definition}

We will investigate $\mathfrak{n}_\tau$ in detail in section \ref{meanFieldActivationSection}. For now, note that $\mathfrak{n}_\tau$ is valid for any $q > c > -q$ for any differentiable and non-constant $\tau$ due to propositions \ref{covkerPositive} and \ref{covkerPositive2}. Note also that the $q$ and $c$ arguments in $\mathfrak{n}_\tau(q,c)$ are somewhat different from those in $\mathfrak{n}_{m,l}(f,q,c)$ in that they capture the properties of the input to $\tau$, instead of the properties of $\mathcal{D}$.

In A-architectures, the only layer operation that can have multiple dependencies is the addition operation. In other words, if $p$ and $p'$ are two directed paths through the layer graph, then those paths can only join at an addition layer.

\begin{definition}
Define the `path-weight' $w(p)$ of a directed path in the layer graph of an A-architecture as follows. For each addition layer contained in the path, consider the ratio $\frac{(\mathfrak{q}_k-\mathfrak{c}_k)w^2}{\mathfrak{q}_l-\mathfrak{c}_l}$, where $f_l$ is the addition layer, $f_k$ is the dependency contained in the path, and $w$ is the addition weight corresponding to $f_k$. $w(p)$ is then the product of all such ratios for addition layers contained in the path that are not its starting point.
\end{definition}

Considering the calculation rule for the $\mathfrak{q}$ and $\mathfrak{c}$ values of addition layers in table \ref{tableNLCPropagation}, it is easy to see that the sum of path weights over all directed paths from input to output layer in a given layer graph equals 1. Informally, if we consider $\mathfrak{q}_L - \mathfrak{c}_L$ as the variance of the network output, the path weight indicates the fraction of that variance that is contributed by that particular path. The path representation now enables the following proposition, which can be proven simply using induction.

\begin{proposition}[`Nonlinearity path equation'] \label{mfntNPE}
Let $\mathfrak{q}_l$, $\mathfrak{c}_l$ and $\mathfrak{g}_l$ be defined according to table \ref{tableNLCPropagation}. Let $0 \le m \le l \le L$ and let $f_m$ be a bottleneck for $f_l$ in the A-architecture $f$. Then 

$$\mathfrak{n}_{l,m}(f,q,c) = \sqrt{\sum_{p \in P_{m,l}} w(p) \prod_{f_{l'}\in p} \mathfrak{n}_{\tau_{l'}}(\mathfrak{q}_{k'},\mathfrak{c}_{k'})^2}$$

where $P_{m,l}$ is the set of directed paths from $f_m$ to $f_l$, the product is over all activation layers $f_{l'}$ on path $p$ excluding its starting point and $f_{k'}$ is the dependency of $f_{l'}$.
\end{proposition}

When $P_{m,l}$ only contains a single path / dependency chain $p_{m,l}$, this simplifies to

$$\mathfrak{n}_{l,m}(f,q,c) = \prod_{f_{l'}\in p_{m,l}} \mathfrak{n}_{\tau_{l'}}(\mathfrak{q}_{k'},\mathfrak{c}_{k'})$$

In plain words, the mean field NLC of an A-architecture is the weighted quadratic mean of the mean field NLC of each directed path in the layer graph, where each path is weighted according to the fraction of the output variance it contributes. The mean field NLC of a path is the product of the activation function NLCs on that path.

This is the instructive relationship between $\mathfrak{n}$ and the architecture definition we have been looking for. It allows us to make manual design decisions while controlling the initial NLC. We will use the nonlinearity path equation throughout the remainder of this work as a reference point for compiling design insights. For now, we will give some first observations.

\finding{
\begin{itemize}
\item $\mathfrak{n}$ is independent of width as long as the $\sigma_l^2$ are fixed.
\item $\mathfrak{n}$ decreases with the introduction of linear skip connections, proportionally with the square of the skip connection strength, assuming there are no knock-on effects on downstream $\mathfrak{q}$ and $\mathfrak{c}$ values.
\item $\mathfrak{n}$ does not directly depend on normalization layers used.
\item Replacing an activation function with another of higher $\mathfrak{n}_{\tau}$ value increases $\mathfrak{n}$, assuming there are no knock-on effects on downstream $\mathfrak{q}$ and $\mathfrak{c}$ values.
\item Composition of activation functions corresponds to multiplying nonlinearity, a principle that is familiar from the chain rule of calculus, the exploding / vanishing gradient problem and the notion of neural network expressivity.
\end{itemize}
}

\section{Mean field theory of activation functions} \label{meanFieldActivationSection}

The key properties of an activation function from a mean field perspective are $\mathfrak{C}_\tau$, $\mathfrak{C}_{\tau'}$ and the derived quantity $\mathfrak{n}_\tau$. In this section, we study these constructs as activation function metrics. We will establish a plethora of useful properties that will be utilized later in this work. This effort culminates in subsection \ref{networkCovKerSection}, where we show that the mean field NLC is the first-order approximation of the bandwidth of the covariance kernel. This is the most important of our results that establish the NLC as a primary measure of model complexity in deep learning.

\begin{metricDefinition} $\mathfrak{C}_\tau(q,c)$ and $\mathfrak{n}_\tau(q,c)$ are defined as in the beginning of section \ref{metricLimitSection} and section \ref{pathEquationSection} respectively. 
\end{metricDefinition}

As stated in section \ref{pathEquationSection}, $\mathfrak{n}_\tau$ is valid for any $q > c > -q$ and differentiable, non-constant $\tau$. We will implicitly assume this in our discussion, but not in our theoretical results. In practical terms, activation function metrics are computed using numerical integration techniques as described in section \ref{metricComputationSection}. The discussion of section \ref{nonDifferentiableSection} applies to the issue of non-differentiable activation functions such as ReLU. However, in this section we do also give results for $\tau$ that are only directionally differentiable, so that we have at least some theory in this work that explicitly includes this case.

\subsection{The NLC of activation functions with meta-Gaussian input} \label{actFunMetaGaussianSection}

We begin by showing that $\mathfrak{n}_\tau$ is not just an abstract quantity arising from table \ref{tableNLCPropagation}, but a direct measure of activation function nonlinearity.

Let the distribution $\mathcal{T}$ over vectors $\chi$ of dimensionality $d_\tau$ be drawn from an elementwise meta-distribution with generator $\mathcal{MN}(q,c)$, $q > c \ge 0$. Then

$$NLC(\tau,\mathcal{T}) = \sqrt{\frac{(q-c)\mathbb{E}_{i,s\sim\mathcal{N}(0,q-c)}\tau'(\bar{\chi}[i] + s)^2}{\mathbb{E}_{i,s\sim\mathcal{N}(0,q-c)}\tau(\bar{\chi}[i] + s)^2 - \mathbb{E}_i(\mathbb{E}_{s\sim\mathcal{N}(0,q-c)}\tau(\bar{\chi}[i] + s))^2}}$$

Here, the component means $\bar{\chi}[i]$ are drawn on the meta-level from a Gaussian with mean zero and variance $c$, and $\mathbb{E}_i$ is a finite mean over $d_\tau$ values. The term $NLC(\tau,\mathcal{T})$ implies that the scalar function $\tau$ is applied elementwise to vectors drawn from $\mathcal{T}$. Given that the sample mean converges almost surely to the distribution expectation, we obtain

$$\lim_{d_\tau \rightarrow \infty}NLC(\tau,\mathcal{T}) = \sqrt{\frac{(q-c)\mathbb{E}_{s\sim\mathcal{N}(0,q)}\tau'(s)^2}{\mathbb{E}_{s\sim\mathcal{N}(0,q)}\tau(s)^2 - \mathbb{E}_{t\sim\mathcal{N}(0,c)}(\mathbb{E}_{s\sim\mathcal{N}(0,q-c)}\tau(t + s))^2}} \text{ a.s.}$$

We can cast the Gaussian expectations in the above formula as a metric of $\tau$, $q$ and $c$, like we did with the covariance kernel.

\begin{metricDefinition}
The `bias kernel' $\mathfrak{B}_\tau$ of an activation function $\tau$ and scalars $q \ge c \ge 0$ is $$\mathfrak{B}_\tau(q,c) = \mathbb{E}_{s\sim\mathcal{N}(0,c)}(\mathbb{E}_{t\sim\mathcal{N}(0,q-c)}\tau(s+t))^2$$
\end{metricDefinition}

Then we obtain

$$\lim_{d_\tau \rightarrow \infty}NLC(\tau,\mathcal{T}) = \sqrt{\frac{(q-c)\mathfrak{B}_{\tau'}(q,q)}{\mathfrak{B}_{\tau}(q,q) - \mathfrak{B}_{\tau}(q,c)}} \text{ a.s.}$$

The right-hand side is exactly $\mathfrak{n}_\tau(q,c)$, except that the covariance kernel is replaced by the bias kernel. We close the loop with the following result.

\begin{proposition} \label{covkerBisC}
For any $\tau$ and $q \ge c \ge 0$, we have $$\mathfrak{B}_\tau(q,c) = \mathfrak{C}_\tau(q,c)$$
\end{proposition}

Hence, we have 

$$\lim_{d_\tau \rightarrow \infty}NLC(\tau,\mathcal{T}) = \mathfrak{n}_\tau(q,c) \text{ a.s.}$$

So the activation function NLC is the limit of the NLC under meta-Gaussian input. Of course, this is not unexpected given theorem \ref{mfntMetaGaussian}, which establishes that fully-connected layers are meta-Gaussian meta-distributed. Proposition \ref{mfntElemLike} also shows that these meta-Gaussians lead to the square means and co-means that arise in table \ref{tableNLCPropagation}. The above result also justifies the name `activation function NLC'. Finally, we obtain that the nonlinearity path equation relates the mean field NLC of the architecture to the actual nonlinearity of its activation functions.  

\subsection{The length kernel of activation functions} \label{actFunLengthKernelSection}

The term $\mathfrak{C}_\tau(q,q)$ has appeared repeatedly in this chapter. In this subsection, we investigate it further. We begin by giving it its own definition.

\begin{metricDefinition}
The `length kernel' $\mathfrak{L}_\tau$ of an activation function $\tau$ and scalar $\lambda \ge 0$ is $$\mathfrak{L}_\tau(\lambda) = \sqrt{\mathfrak{C}_\tau(\lambda^2,\lambda^2)} = \sqrt{\mathbb{E}_{s\sim\mathcal{N}(0,\lambda^2)}\tau(s)^2}$$
\end{metricDefinition}

The length kernel represents the infinite width limit of the quadratic mean of an activation layer when its dependency is distributed according to $\mathcal{N}(0,\lambda^2 I)$. We believe that the quadratic mean is a more intuitive quantity to deal with than the square mean. Hence, we define $\mathfrak{l}_l = \sqrt{\mathfrak{q}_l}$. Table \ref{tableNLCPropagation} then yields $\mathfrak{l}_l = \mathfrak{L}_{\tau_l}(\mathfrak{l}_k)$ for activation layers $f_l$. Since a deep network generally composes many activation layers, $\mathfrak{l}_L$ depends on $\mathfrak{l}_0$ via a sequence of length kernel applications.

\begin{definition}
We say a `plain architecture' is an architecture composed of $M$ macro-layers, which are themselves composed of an FC and an activation layer. Each activation layer uses the same activation function $\tau$ and each FC layer is Gaussian initialized with the same fixed multiple $\sigma^2$ of the LeCun variance.
\end{definition}

A plain architecture is a fixed width A-architecture, except for the technical conditions on the activation function. Hence, for the purpose of discussion, we apply theorem \ref{mfntPropagation} and its corollaries, in addition to the results of section \ref{meanFieldActivationSection}, to plain architectures.

In a plain architecture with $\sigma=1$, we have $\mathfrak{l}_L = \mathfrak{L}^M_\tau(\mathfrak{l}_0)$, where the exponent indicates composition. Of course, not all architectures we care about are this simple. However, this example motivates us to study the length kernel under the umbrella of function iteration. In a nutshell, because of the ZSAD guideline of scale stability given in section \ref{architectureDesignParadigmsSection}, we desire the iteration to converge to a non-zero value, which is ideally close to 1. We will further discuss this point in detail throughout chapter \ref{surveyChapter}.

Below, we give the main theorem of this subsection, which gives a range of properties for the length kernel. We follow this up with an empirical study.

\begin{theorem} \label{covkerLregular}
For any $\tau$ we have
\begin{enumerate} 
\item If $\mathfrak{L}_\tau(\lambda) \neq 0$ for some $\lambda > 0$, then for all $\lambda > 0$ we have
\begin{itemize}
\item[(a)]$\mathfrak{L}_\tau(\lambda) > 0$
\item[(b)] $\mathfrak{L}_\tau$ is differentiable at $\lambda$ with $\mathfrak{L}_\tau'(\lambda) = \frac{1}{2}\mathfrak{L}_\tau(\lambda)^{-1}\mathbb{E}_{s \sim \mathcal{N}(0, \lambda^2)}\tau(s)^2(s^2\lambda^{-3}-\lambda^{-1})$
\item[(c)] $\frac{\mathfrak{L}_\tau'(\lambda)\lambda}{\mathfrak{L}_\tau(\lambda)} > -\frac{1}{2}$
\item[(d)] $(\sqrt{\lambda}\mathfrak{L}_\tau(\lambda))' > 0$
\end{itemize}
\item If $\tau$ is continuous and not zero everywhere, then $\mathfrak{L}_\tau(\lambda) > 0$ for all $\lambda > 0$.
\item If $\tau$ is continuous at zero, $\mathfrak{L}_\tau$ is continuous at zero.
\item If $\tau$ is directionally differentiable at zero, then $\mathfrak{L}_\tau$ is differentiable at zero with $\mathfrak{L}_\tau'(0) = \sqrt{\frac{\tau^-(0)^2 + \tau^+(0)^2}{2}}$ if $\tau(0) = 0$ and $\mathfrak{L}_\tau'(0) = \frac{1}{\sqrt{2\pi}}\sign(\tau(0))(\tau^+(0)-\tau^-(0))$ if $\tau(0) \neq 0$. The + and - superscripts indicate the right and left derivative respectively.
\item If $\mathfrak{L}_\tau(\lambda) \neq 0$ for some $\lambda > 0$, then for all $\lambda > 0$ the sequence $(\mathfrak{L}_\tau^n(\lambda))_n$ either (i) is strictly increasing and diverges to infinity or (ii) converges.
\end{enumerate}
\end{theorem}

The highlights are that under very mild conditions $\mathfrak{L}_\tau$ is differentiable, iterating $\mathfrak{L}_\tau$ yields convergence or strictly increasing divergence, and statement 1(c) yields that $\mathfrak{L}_\tau$ has no unstable fixed points where it is decreasing. Going forward, we can think of the iteration convergence behavior of $\mathfrak{L}_\tau$ like that of a general differentiable function.

In tables \ref{covCurveillu1}, \ref{covCurveillu2}, \ref{covCurveillu3} and \ref{covCurveillu4}, we depict the length kernel for 24 activation functions: the 12 activation functions contained in tables \ref{actFunIllu}, \ref{AFLMillu} and \ref{basisillu}, as well as the debiased versions of those activation functions. Debiasing in the context of activation functions, as described in e.g. section \ref{studyAArchitecturesSection}, corresponds to subtracting from the activation function its expectation under the unit Gaussian distribution, i.e. $\mathbb{E}_{s\sim \mathcal{N}(0,1)}\tau(s)$. This strategy was used for generating many architectures for both study A and study B. Note that tanh, odd square and sawtooth are equal to their debiased versions, and SeLU is almost equal to it. 

We plot length kernels in three colors: blue, red and magenta. When we iterate $\mathfrak{L}_\tau$ starting from a point in a blue segment, the iteration converges to a non-zero point. When we iterate $\mathfrak{L}_\tau$ starting from a point in a red segment, the iteration either converges to zero or diverges to infinity. When we iterate $\mathfrak{L}_\tau$ starting from a point in a magenta segment, the iteration is stationary. 

We find that we obtain a range of different convergence behaviors for different activation functions. 

\begin{itemize}
\item \finding{For many activation functions, all starting values of $\lambda$ converge to the same non-zero value. Their length kernels are depicted in all blue and the limit is a bold blue point.}
\item \finding{Also for many activation functions, all starting values of $\lambda$ converge to zero. Their length kernels are depicted in all red and lie below the diagonal, which is depicted in black as a visual aid.}
\item \finding{For some activation functions, small values of $\lambda$ converge to zero whereas large values diverge. Their length kernels are depicted in red, and the curve crosses the diagonal.}
\item \finding{For one activation function (abs. val.), $\mathfrak{L}_\tau$ is the identity, so $\lambda$ is preserved. Its length kernel is depicted in magenta.}
\item \finding{Finally, for one activation function (square-debiased), all values of $\lambda$ diverge. Its length kernel is depicted in red and above the diagonal.}
\end{itemize}

Similarly, \finding{we find a range of values for the convergence rate. The majority of the time, convergence is exponential. This occurs when the derivative of $\mathfrak{L}_\tau$ is neither 0, 1 or -1 at the limit point. If the derivative is 0, then the convergence is superexponential. This happens for the square and odd square activation functions. In fact, convergence is square exponential. If the derivative is either 1 or -1, convergence is sub-exponential. For tanh and even tanh, the rate is $O(\frac{1}{M^{0.5}})$, and for sawtooth the rate is sub-polynomial, i.e. slower than any $O(\frac{1}{M^\epsilon})$ with $\epsilon > 0$}. In tables \ref{covCurveillu1} through \ref{covCurveillu4}, we specify convergence rates as precisely as we were able to determine them with simple methods.

In tables \ref{covCurveillu1} through \ref{covCurveillu4}, we also give the normalized length kernel $\frac{\mathfrak{L}_\tau(\lambda)}{\mathfrak{L}_\tau(1)}$. This is the length kernel we would obtain when normalizing the activation function by its quadratic mean under the unit Gaussian, i.e. $\sqrt{\mathbb{E}_{s\sim \mathcal{N}(0,1)}\tau(s)^2}$. This normalization is done for both study A and study B architectures. Hence, we are interested in normalized length kernels, which are themselves valid length kernels, as is any positive multiple of a valid length kernel. \finding{The range of convergence behaviors observed among our normalized length kernels is largely similar to the range we observed previously, except that there are now several activation functions that have both $\lambda$ values that converge to a non-zero value (depicted in blue) as well as $\lambda$ values that diverge (depicted in red).}

While we have observed a wide range of convergence behaviors in tables \ref{covCurveillu1} through \ref{covCurveillu4}, through deliberate design of activation functions, we could generate an even more diverse set of behaviors, though this goes beyond the scope of this work. While length kernels depicted have at most one stable and one unstable fixed point, it is easy to generate activation functions that have an infinite number of them. When $\tau(s)^2$ oscillates between small and large values, so can $\mathfrak{L}_\tau$.

In addition to studying the length kernel normalized by its value at 1, i.e. $\frac{\mathfrak{L}_\tau(\lambda)}{\mathfrak{L}_\tau(1)}$, we can study the family of all normalizations of form $\frac{\mathfrak{L}_\tau(\lambda)\lambda'}{\mathfrak{L}_\tau(\lambda')}$ for some fixed $\lambda' > 0$. $\frac{\mathfrak{L}_\tau(\lambda)\lambda'}{\mathfrak{L}_\tau(\lambda')}$ has a fixed point at $\lambda = \lambda'$ with derivative $\frac{\mathfrak{L}_\tau'(\lambda)\lambda}{\mathfrak{L}_\tau(\lambda)}$. We plot $\frac{\mathfrak{L}_\tau'(\lambda)\lambda}{\mathfrak{L}_\tau(\lambda)}$ as a function of $\lambda$ in tables \ref{covCurveillu1} through \ref{covCurveillu4}. A value between -0.5 and 1 indicates stability (depicted in blue), a value above 1 indicates instability (depicted in red) and a value of 1 indicates stationary behavior (depicted in magenta). Note that statement 1(c) of theorem \ref{covkerLregular} implies that this curve cannot dip below -0.5. \finding{For the majority of activation functions, we obtain the same regime for all values of $\lambda$, though for some we obtain stability specifically for small values of $\lambda$.}

\subsection{The covariance kernel of activation functions} \label{actFunCovKernelSection}

When calculating $\mathfrak{c}_L$ according to table \ref{tableNLCPropagation}, we repeatedly apply $\mathfrak{C}_\tau(q,c)$. So just as with the length kernel, we are interested in function iteration. 

\begin{definition}
We say an architecture is `plain stable$(q,q')$' if it is a plain architecture where $\sigma^2q=q'$ and $\mathfrak{L}_\tau(\sqrt{q'}) = \sqrt{q}$.
\end{definition}

In a plain stable$(q,1)$ architecture, when the input distribution is elem-like$(q,c)$, we have $\mathfrak{c}_L = q\tilde{\mathfrak{C}}_\tau^M(1,c)$, where the exponent indicates composition and $\tilde{\mathfrak{C}}_\tau$ denotes $\frac{\mathfrak{C}_\tau}{\mathfrak{L}_\tau(1)^2}$. While this setup might appear specific, it describes networks fulfilling the scale stability ZSAD guideline from section \ref{architectureDesignParadigmsSection} in a mean field sense. This setup was used approximately for both study A and B and was the focus of prior work \citep{correlationLimit,depthScalesMeanField,neuralTangentKernel,meanFieldNTK5}. As previously mentioned, we shorten $\mathfrak{C}_\tau(1,c)$ to $\mathfrak{C}_\tau(c)$. While we mostly focus on the case $q'=1$ in this and the next subsection, we can easily obtain equivalent results for any other value of $q'$. Again, we will further use the results from this subsection in chapter \ref{surveyChapter}.

It turns out that $\mathfrak{C}_\tau(c)$ is more regular than $\mathfrak{L}_\tau(\lambda)$. After a definition, we give our main theorem of this subsection, followed by an empirical study.

\begin{definition}
We say a function $F : \mathbb{R} \rightarrow \mathbb{R}$ is `piecewise $n$-differentiable' if there exists a partition of $\mathbb{R}$ into a finite set of intervals such that (i) $F$ is $n$ times differentiable in the interior of each interval and (ii) $F$ is continuous and $n$ times directionally differentiable everywhere. In particular, the left and right derivative at each interval endpoint does not have to be the same.
\end{definition}

All of our activation functions in table \ref{actFunIllu}, except sawtooth, are piecewise $n$-differentiable for any $n$. Sawtooth is still piecewise $n$-differentiable within any bounded interval.

\begin{theorem} \label{covkerCregular}
Assume $\tau$ is piecewise 5-differentiable. Consider $\mathfrak{C}_\tau(c)$ defined on $[0,1]$. Let $a_\tau s + b_\tau$ be the least squares linear fit to $\tau$ under $\mathcal{N}(0,1)$ and let $\tilde{\tau} = \tau - a_\tau s - b_\tau$. Then

\begin{enumerate}[label=(\alph*)]
\item $\mathfrak{C}_\tau$ is differentiable and $\mathfrak{C}_\tau' = \mathfrak{C}_{\tau'}$.
\item $\mathfrak{C}_\tau$ is increasing, convex and for all $c > 0$, $\epsilon > 0$ with $c + 3\epsilon \le 1$ we have $\mathfrak{C}_\tau(c+3\epsilon) - 3\mathfrak{C}_\tau(c+2\epsilon) + 3\mathfrak{C}_\tau(c+\epsilon) - \mathfrak{C}_\tau(c)\ge 0$.
\item (i) $\mathfrak{C}_\tau(c) = \mathfrak{C}_{\tilde{\tau}}(c) + a_\tau^2c + b_\tau^2$, (ii) $\mathfrak{C}_{\tilde{\tau}}(0) = 0$, (iii) $\mathfrak{C}_{\tau}(0) = b_\tau^2$, (iv) $\mathfrak{C}_{\tilde{\tau}}'(0) = 0$, (v) $\mathfrak{C}_{\tau}'(0) = a_\tau^2$, (vi) $CAR(\tau, \mathcal{N}(0,1)) = \frac{\mathfrak{C}_{\tau}(0)}{\mathfrak{C}_{\tau}(1)}$, (vii) $LAR(\tau, \mathcal{N}(0,1)) = \frac{\mathfrak{C}_{\tau}'(0)}{\mathfrak{C}_{\tau}(1)}$ and (viii) $\mathfrak{C}_{\tau}$ is linear if and only if $\tau$ is linear.
\end{enumerate}

Now also assume $\tau$ is not linear. Then, for the normalized covariance kernel $\tilde{\mathfrak{C}}_\tau(c) = \frac{\mathfrak{C}_\tau(c)}{\mathfrak{C}_\tau(1)}$ exactly one of three cases hold on $[0,1]$.

\begin{enumerate}
\item There is a stable fixed point at $c^\text{lim}=1$ with exponential convergence rate and no other fixed point. $0 < \tilde{\mathfrak{C}}_\tau'(1) < 1$ holds.
\item There is a stable fixed point at $c^\text{lim}=1$ with sub-exponential convergence rate and no other fixed point. $\tilde{\mathfrak{C}}_\tau'(1) = 1$ holds.
\item There is an unstable fixed point at 1 and there is exactly one other fixed point $c^\text{lim}$ in $[0,1)$, which is stable. $\tilde{\mathfrak{C}}_\tau'(1) > 1$ and $1 > \tilde{\mathfrak{C}}_\tau'(c^\text{lim}) \ge 0$ hold. The convergence rate is super-exponential or exponential depending on whether $\tilde{\mathfrak{C}}_\tau'(c^\text{lim}) = 0$ holds.
\end{enumerate}

In particular, there exists exactly one stable fixed point $c^\text{lim}$ and iterating $\tilde{\mathfrak{C}}_\tau$ will lead to convergence towards $c^\text{lim}$ from any starting point.

\end{theorem}

The third part of statement (b) is similar to saying that $\mathfrak{C}_{\tau}(c)$ has a non-negative third derivative. Note that if $\tau$ is smooth, we can apply statement (a) repeatedly to obtain that the $n$'th derivative of $\mathfrak{C}_{\tau}(c)$ is the covariance kernel of the $n$'th derivative. Hence, we immediately have a wide range of properties for the $n$'th derivative, including those in the above theorem as well as propositions \ref{covkerPositive}, \ref{covkerPositive2} and \ref{covkerPositive3}.

While we do not give a detailed interpretation of the theorem here, we make use of the different parts in various places throughout this work. We note that, to a large degree, the convergence case breakdown in the second half of the theorem was previously observed empirically by \citet{correlationLimit} for various architectures based on the tanh activation function. However, to our knowledge, it was never proven for general $\tau$.

In tables \ref{covCurveillu1}, \ref{covCurveillu2}, \ref{covCurveillu3} and \ref{covCurveillu4}, we plot the normalized covariance kernel $\tilde{\mathfrak{C}}_\tau(c)$ for our 24 activation function. Note that $\tilde{\mathfrak{C}}_\tau(c)$ is itself a valid covariance kernel of the activation function $\frac{\tau}{\mathfrak{L}_\tau(1)}$. Indeed, \finding{each of the three convergence cases from the theorem is observed}. 

The LAR, CAR and NAR metrics were defined in section \ref{nlcLinearApproximationSection}. In table \ref{basisillu}, we gave LAR, CAR and NAR values for our activation functions as well as the $\tilde{\tau}$ corresponding to our $\tau$. More generally, the analysis from section \ref{nlcLinearApproximationSection} applies to activation functions. For example, we have $\mathfrak{n}_{\tilde{\tau}}(0) \ge \sqrt{2}$ and $\mathfrak{n}_{\tau}(0)^2 = \frac{LAR}{LAR + NAR} + \frac{NAR}{LAR + NAR}\mathfrak{n}_{\tilde{\tau}}(0)^2$.

\subsection{The activation function NLC} \label{actFunNLCsection}

\paragraph{Properties of $\mathfrak{n}_\tau(c)$} Using theorem \ref{covkerCregular}, we straightforwardly obtain the following about $\mathfrak{n}_\tau(c)$.

\begin{proposition} \label{covkernregular}
Assume $\tau$ is piecewise 5-differentiable. Consider $\mathfrak{n}_\tau(c)$ defined on $[0, 1)$. Then

\begin{enumerate}
\item $$\mathfrak{n}_\tau(c) = \sqrt{\frac{\mathfrak{C}_\tau'(1)(1-c)}{\mathfrak{C}_\tau(1)-\mathfrak{C}_\tau(c)}} = \sqrt{\frac{\tilde{\mathfrak{C}}_\tau'(1)(1-c)}{1-\tilde{\mathfrak{C}}_\tau(c)}}$$
\item $\mathfrak{n}_\tau$ is decreasing
\item $\lim_{c \rightarrow 1}\mathfrak{n}_\tau(c) = 1$
\item $\mathfrak{n}_\tau(c) \ge 1$
\end{enumerate}

\end{proposition}

Hence, $\mathfrak{n}_\tau(c)$ is determined entirely by the covariance kernel of $\tau$. Note the contrast of statement 1 to statement (c) from theorem \ref{covkerCregular}. It turns out that the nonlinearity of an activation function as measured by the linear approximation error is related to the derivative of the covariance kernel at 0. The nonlinearity of an activation function as measured by the NLC is related to the derivative of the covariance kernel at 1. Statement 2 makes sense in light of proposition \ref{covkerBisC}. The smaller the neuron variance under the meta-Gaussian, the more linear $\tau$ becomes across its effective domain.

\paragraph{Iterating $\mathfrak{n}_\tau(c)$} For plain stable$(q,1)$ A-architectures, the nonlinearity path equation yields

$$\mathfrak{n} = \prod_{m=1}^M\mathfrak{n}_\tau(\tilde{\mathfrak{C}}_\tau^m(\mathfrak{c}_0))$$

We are interested in the behavior of $\mathfrak{n}$ as $M$ converges to infinity. By theorem \ref{covkerCregular}, $\tilde{\mathfrak{C}}_\tau^m(\mathfrak{c}_0)$ converges to some value $c^\text{lim}$ as $m$ increases, where $c^\text{lim}$ is the single stable fixed point of $\tilde{\mathfrak{C}}_\tau$. If $c^\text{lim} \neq 1$, then clearly $\mathfrak{n}$ diverges exponentially with rate $\mathfrak{n}_\tau(c^\text{lim})$. This corresponds to case 3 in theorem \ref{covkerCregular}. Two cases remain.

Case 2 from theorem \ref{covkerCregular}: $\mathfrak{n}$ is a telescopic product that equals $\sqrt{\frac{1 - \mathfrak{c}_0}{1 - \tilde{\mathfrak{C}}^M_\tau(\mathfrak{c}_0)}}$. This also diverges to infinity, where the rate is related to the convergence rate of $\tilde{\mathfrak{C}}^M_\tau$ to 1. For ReLU and abs. val., for example, that rate is $O(\frac{1}{M^2})$, and therefore $\mathfrak{n}$ grows linearly with depth. If $\tilde{\mathfrak{C}}_\tau$ is twice differentiable at 1, that rate is $O(\frac{1}{M})$ and so $\mathfrak{n}$ grows as the square root. By theorem \ref{covkerCregular}, $\tilde{\mathfrak{C}}_\tau$ is twice differentiable if $\tau$ is differentiable and $\tau'$ is piecewise 5-differentiable. This is fulfilled by neither ReLU nor abs. val.. Because $\lim_{c \rightarrow 1}\mathfrak{n}_\tau(c) = 1$, divergence is always sub-exponential.

Case 1 from theorem \ref{covkerCregular}: The limit of $\mathfrak{n}$ also depends on the analytical properties of $\tilde{\mathfrak{C}}_\tau$ at 1. If $\tilde{\mathfrak{C}}_\tau$ is twice differentiable at 1, then it is easy to show that $\mathfrak{n}$ converges exponentially with depth. By theorem \ref{covkerCregular}, this holds, for example, for softplus, sigmoid and Gaussian. These are the three activation functions in tables \ref{covCurveillu1} to \ref{covCurveillu4} that fall under case 1. Unfortunately, we do not know what can happen when $\tilde{\mathfrak{C}}_\tau$ is not twice differentiable.

\paragraph{Properties of $\mathfrak{n}_\tau(q, 0)$} We are also interested in how $\mathfrak{n}_\tau$ responds to changes in $q$, though for the opposite reason that we are interested in $\mathfrak{n}_\tau(1, c)$. While we generally design architectures to achieve $\mathfrak{q}_k \approx 1$ for dependencies of activation layers, we also want to understand what happens if this condition fails.

In tables \ref{covCurveillu1}, \ref{covCurveillu2}, \ref{covCurveillu3} and \ref{covCurveillu4}, we plot $\mathfrak{n}_\tau(\lambda^2, 0)$, where $\lambda = \sqrt{q}$ as in section \ref{actFunLengthKernelSection}. We find that different activation functions exhibit drastically different behavior. Some activation functions, such as ReLU, abs. val. and square, have a stable $\mathfrak{n}_\tau$, whereas other activation functions, such as even tanh and sawtooth, have a rapidly increasing one. Hence, different activation functions respond very differently to e.g. changes in the variance of weight matrix entries. Here are some general patterns. \finding{

\begin{itemize}
\item $\mathfrak{n}_\tau(\lambda^2, 0)$ is invariant to debiasing.
\item SeLU, softplus and Swish are all ReLU-like for large $\lambda$ and so $\lim_{\lambda \rightarrow \infty}\mathfrak{n}_\tau(\lambda^2, 0) = \mathfrak{n}_\text{ReLU}(1, 0) = 1.21$.
\item ReLU, abs. val., square and odd square have the property $\tau(\lambda s) = \lambda\tau(s)$ for arbitrary $s$ and $\lambda \ge 0$, so $\mathfrak{n}_\tau(\lambda^2, 0)$ is constant.
\item Tanh and sigmoid ``converge to the step function'' so $\mathfrak{n}_\tau(\lambda^2, 0)$ diverges as $O(\sqrt{\lambda})$.
\item Even tanh and Gaussian ``converge to the delta function'' so $\mathfrak{n}_\tau(\lambda^2, 0)$ diverges as $O(\lambda)$.
\item Sawtooth is periodic so $\mathfrak{n}_\tau(\lambda^2, 0)$ diverges as $O(\lambda)$.
\end{itemize}

}

\subsection{The covariance kernel of A-architectures} \label{networkCovKerSection}

Now we bring things back to the architecture level. The covariance kernel of a mean field architecture with a single finite input and readout layer with respect to finite sets of inputs was defined in section \ref{covarianceKernelSection} in the context of background corollary \ref{backgroundKernel}. In the context of A-architectures and elem-like input distributions, we define analogously.

\begin{definition}
For an A-architecture $f$ without batch normalization, the `covariance kernel' $\mathfrak{C}(q,c)$, $q \ge c \ge 0$, is the function that returns $\mathfrak{c}_L$ defined according to table \ref{tableNLCPropagation}.
\end{definition}

It is easy to check that $\mathfrak{q}_L = \mathfrak{C}(q,q)$ holds. To see that $\mathfrak{C}$ is valid for the case $q=c$, which was excluded in e.g. theorem \ref{mfntPropagation} and proposition \ref{mfntPositive}, we note that each operation other than BN preserves $\mathfrak{q} \ge \mathfrak{c} \ge 0$ in table \ref{tableNLCPropagation}. For the activation operation, see propositions \ref{covkerPositive} and \ref{covkerPositive2}.

\begin{proposition}\label{mfntCNLC}

Let $f$ be an A-architecture where each activation function used is piecewise 5-differentiable and let $q > c \ge 0$. If $f$ does not contain batch normalization but can contain layer normalization layers, we have 

$$\mathfrak{n}(f,q,c) = \sqrt{\frac{\frac{d}{dq'}\mathfrak{C}(q,q')|_{q'=q}(q - c)}{\mathfrak{C}(q,q) - \mathfrak{C}(q,c)}}$$

If $f$ does not contain layer normalization but can contain batch normalization layers, an analogous statement holds. See section \ref{meanFieldBNsection} for details.
\end{proposition}

Like the nonlinearity path equation, this can be proven simply with induction. Crucially, we also use statement (a) from theorem \ref{covkerCregular}. Like for theorem \ref{mfntPropagation}, we defer the BN case to section \ref{meanFieldBNsection}. This proposition explicitly requires that the activation functions used by $f$ are piecewise 5-differentiable. Further, because $f$ is an A-architecture, its activation functions are also required to be controlled by $\mathcal{C}^\text{E2}$, non-constant and twice differentiable with a derivative controlled by $\mathcal{C}^\text{E2}$.

We can interpret proposition \ref{mfntCNLC} just as we interpreted the NLC in section \ref{nlcDefinitionSection} and gradient-based local linear approximability in section \ref{nlcSensiSection}. If we view $\mathcal{D}$ as an elementwise distribution in the spirit of proposition \ref{mfntElemLike}, then the dimensionality-normalized radius of the domain $\sqrt{\frac{1}{d_\text{in}}\Tr(\Cov_x)}$ becomes $\sqrt{q - c}$. Similarly, if the codomain is generated by $\mathcal{MN}(\mathfrak{C}(q,q), \mathfrak{C}(q,c))$, then $\sqrt{\frac{1}{d_\text{out}}\Tr(\Cov_f)}$ becomes $\sqrt{\mathfrak{C}(q,q) - \mathfrak{C}(q,c)}$. Finally, $\frac{d}{dq'}\mathfrak{C}(q,q')|_{q'=q}$ measures the sensitivity of the output co-mean with respect to small changes in input co-mean. So we can informally say the following. If $\mathbb{E}_ix[i]^2 = \mathbb{E}_ix'[i]^2 = q$ and $\mathbb{E}_ix[i]x'[i] = q - \epsilon$, then $\epsilon = \frac{1}{2d_\text{in}}||x - x'||_2^2$ and $\frac{d}{dq'}\mathfrak{C}(q,q')|_{q'=q}\epsilon + O(\epsilon^2) = \frac{1}{2d_\text{out}}||f(x) - f(x')||_2^2$. So $\sqrt{\frac{d_\text{out}}{d_\text{in}}\frac{d}{dq'}\mathfrak{C}(q,q')|_{q'=q}}$ measures the output change induced by a small input change as defined by Euclidean distance. But $\frac{1}{\sqrt{d_\text{in}}}||\mathcal{J}||_F$ also measures this change, so $\frac{d}{dq'}\mathfrak{C}(q,q')|_{q'=q}$ represents $\frac{1}{\sqrt{d_\text{out}}}||\mathcal{J}||_F^2$. Putting it all together, $\mathfrak{n}(f,q,c)$ resembles $NLCFROB$ from section \ref{nlcSimpleMetricsSection}, which has the same mean field limit as the NLC.

``Wide networks are Gaussian processes'' \citep{meanFieldNetsorGP}, as we outlined in section \ref{covarianceKernelSection}. The associated kernel function for an A-architecture with elem-like input is $\mathfrak{C}(\mathbb{E}_ix[i]^2,\mathbb{E}_ix[i]x'[i])$. Therefore, we can view $\frac{d}{dq'}\mathfrak{C}(q,q')|_{q'=q}$ as the first-order approximation of the bandwidth of the kernel. $\mathfrak{n}(f,q,c)$ then becomes the (square root of the) bandwidth normalized by the radius of domain and codomain. This normalization is desirable for all the reasons it is desirable to employ this normalization in the NLC as discussed in e.g. sections \ref{whatIsNonlinearitySection}, \ref{nlcDefinitionSection} and \ref{nlcNoiseSection}. For a fixed unnormalized bandwidth value, a large domain implies that a small fraction of input pairs in that domain receive a significant kernel value, whereas a small domain implies that a large fraction of input pairs in that domain receive a significant kernel value. Similarly, a large codomain implies that a large kernel value, and hence a large bandwidth is required for outputs to be relatively close together, whereas a small codomain implies that a small bandwidth is sufficient. We will further study the importance of this normalization in chapter \ref{relatedWorkChapter}.

Via mean field theory, we therefore find that the NLC is a measure of model complexity for the same reason that kernel bandwidth is a measure of model complexity in the field of kernel methods. In the infinite width limit of the initial state, which this, as well as prior, work has shown is extremely practically predictive, the network function is determined by a kernel. To the degree to which an effectively normalized scalar bandwidth can be defined for that kernel, the mean field NLC captures it. In turn, the mean field NLC is highly practically predictive of the NLC. While proposition \ref{mfntCNLC} is unfortunately limited to A-architectures, we suspect an analogous result holds much more generally for mean field architectures under e.g. the setup of background corollary \ref{backgroundKernel}, or even for general Gaussian processes. Proving such an extension is future work.

\newpage

\begin{table}[H]
{
\centering \small
\begin{tabular}{lcccccc}
Act. fun. &ReLU&SELU&softplus&Swish&abs. val.&tanh\\ \hline\hline
\\
$\tau(s)$&\includegraphics[scale=0.13,valign=c]{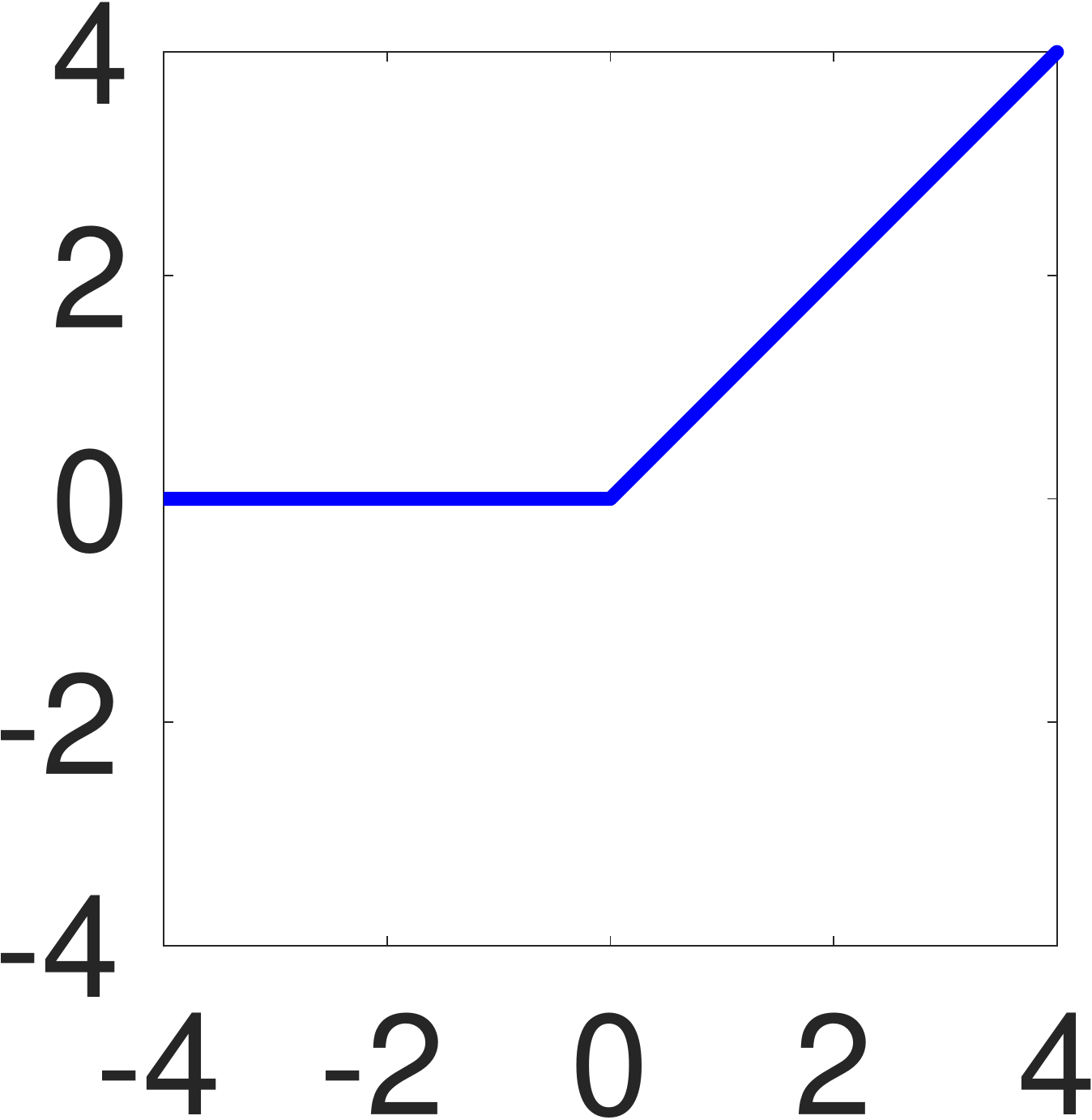}&\includegraphics[scale=0.13,valign=c]{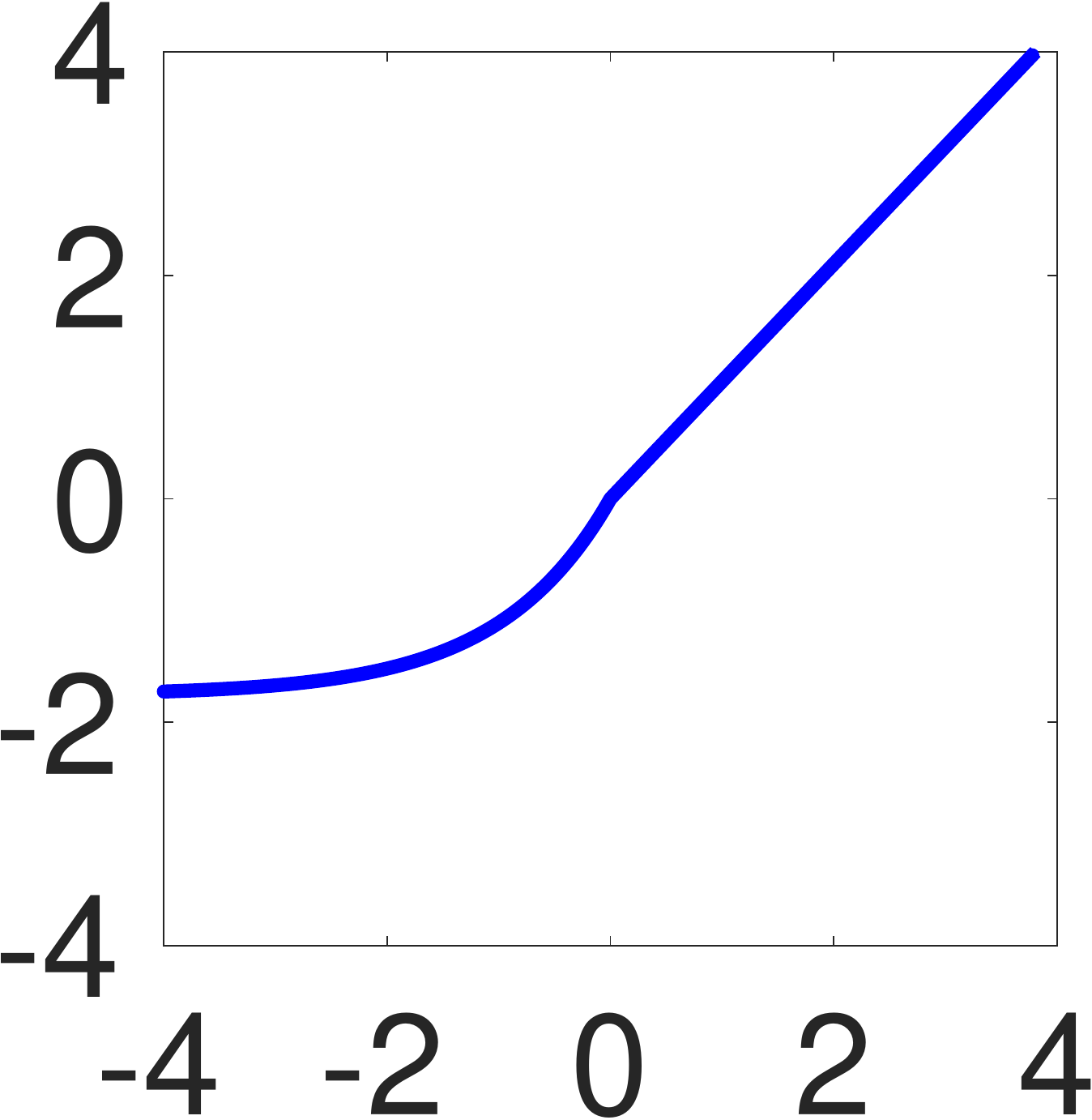}&\includegraphics[scale=0.13,valign=c]{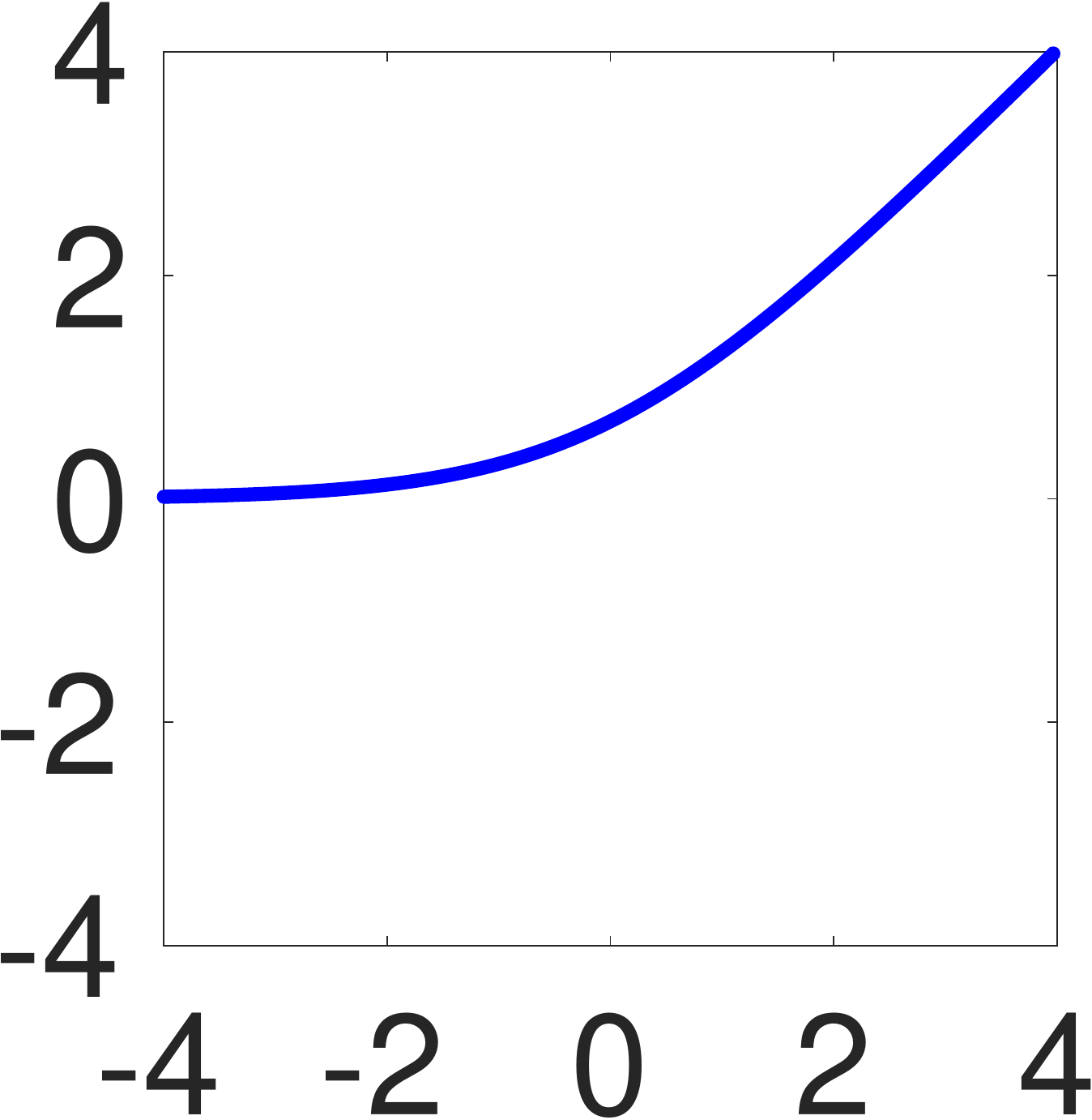}&\includegraphics[scale=0.13,valign=c]{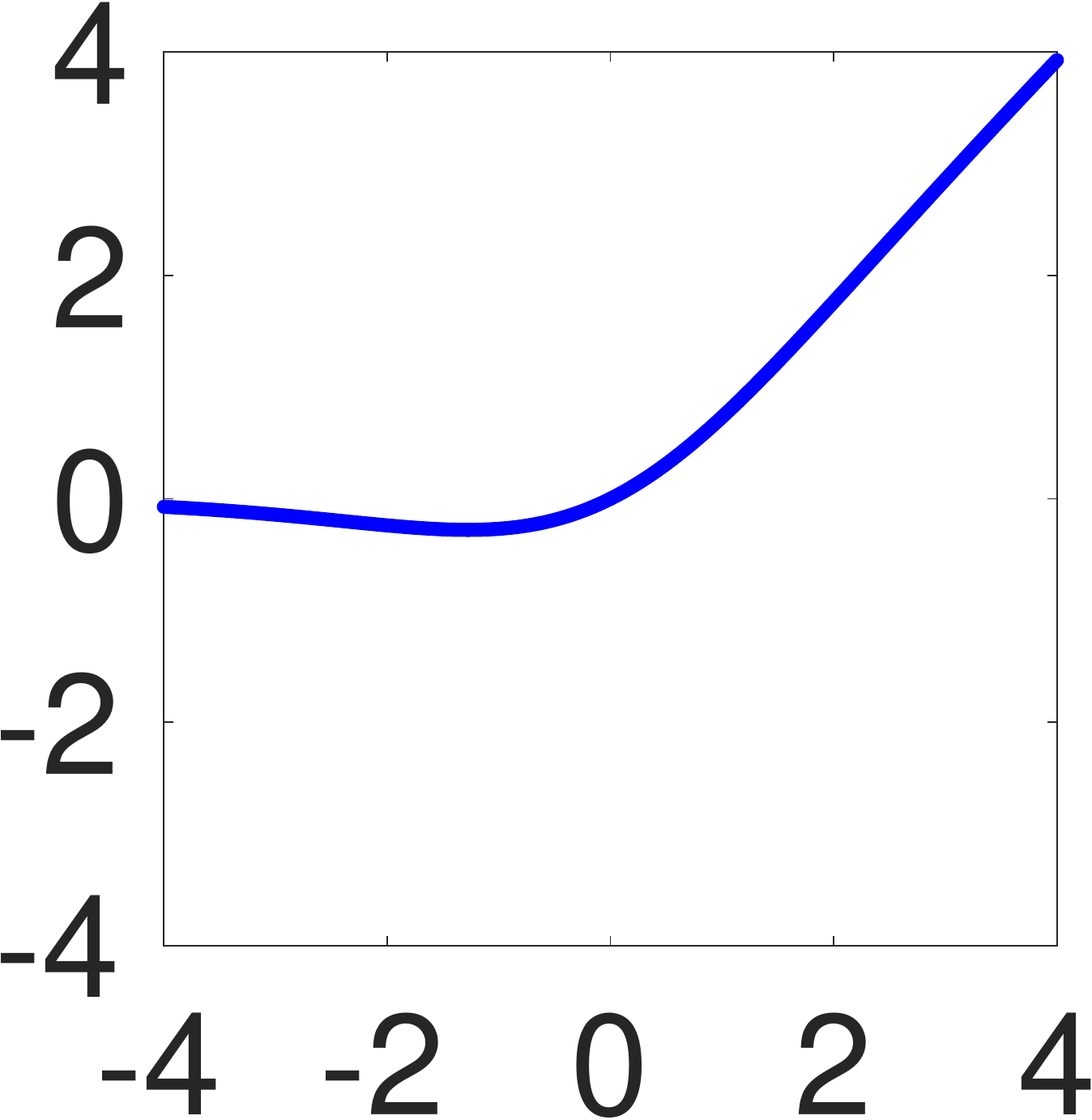}&\includegraphics[scale=0.13,valign=c]{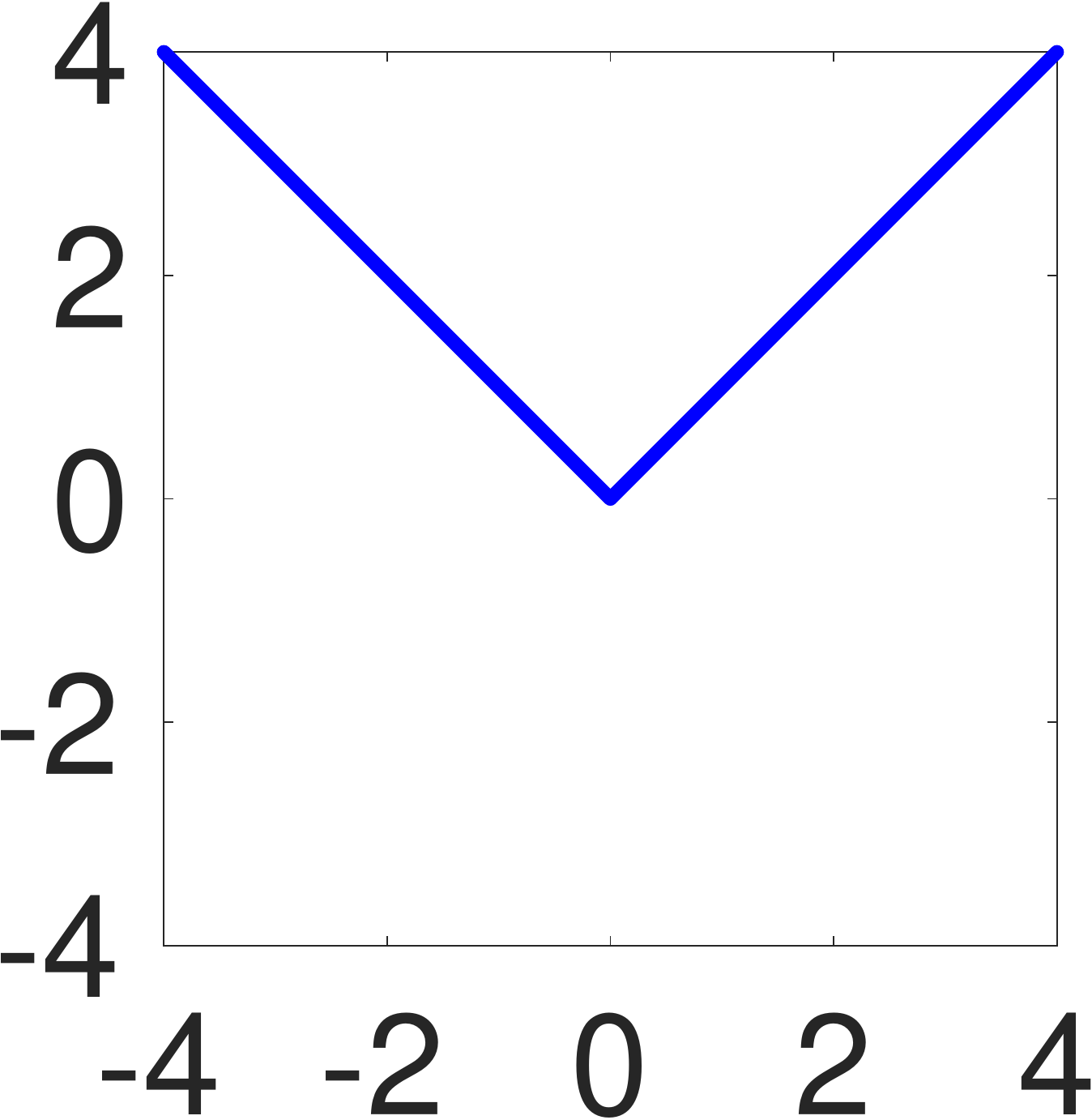}&\includegraphics[scale=0.13,valign=c]{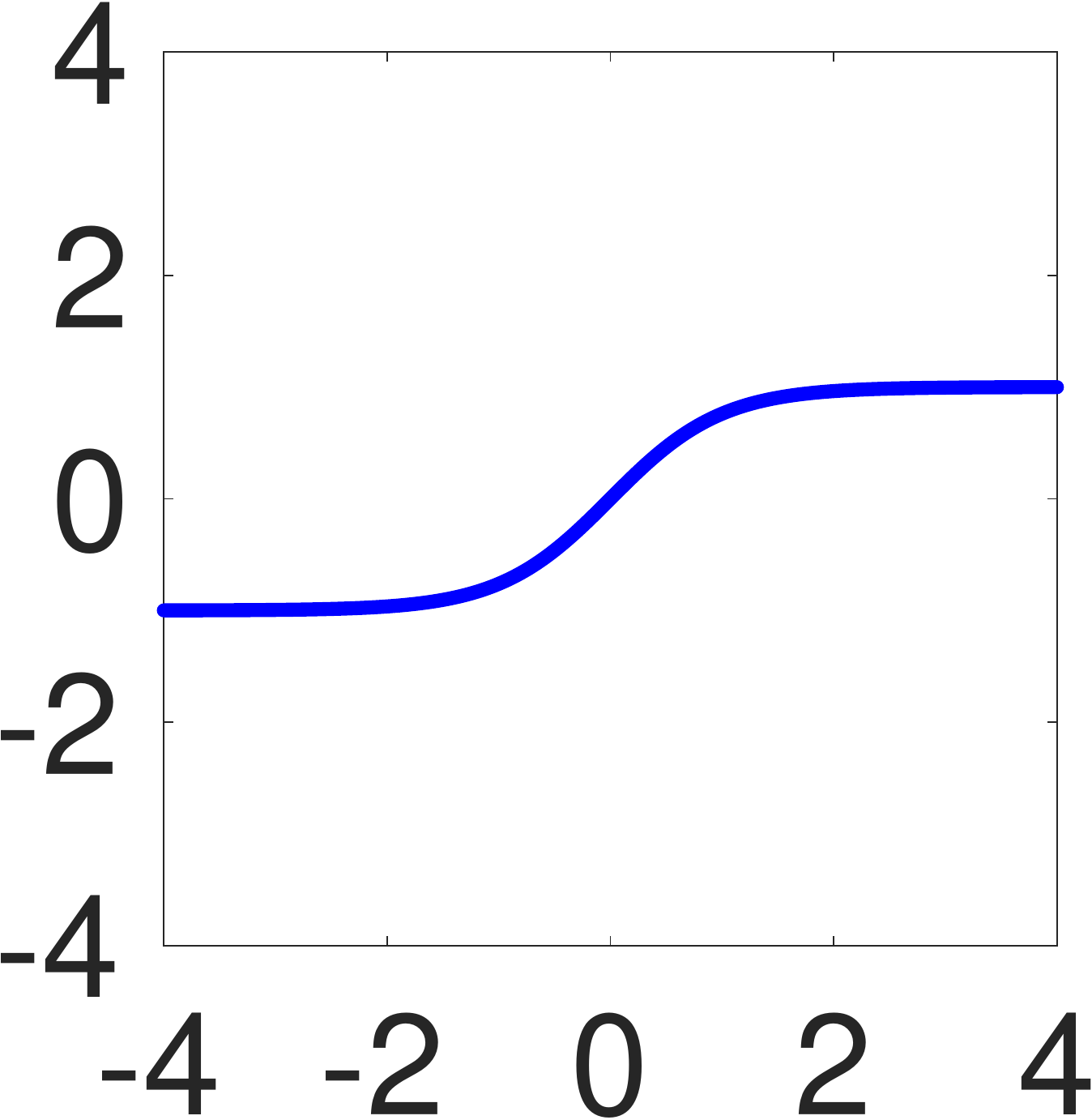}\\
$\mathfrak{L}_\tau(\lambda)$&\includegraphics[scale=0.13,valign=c]{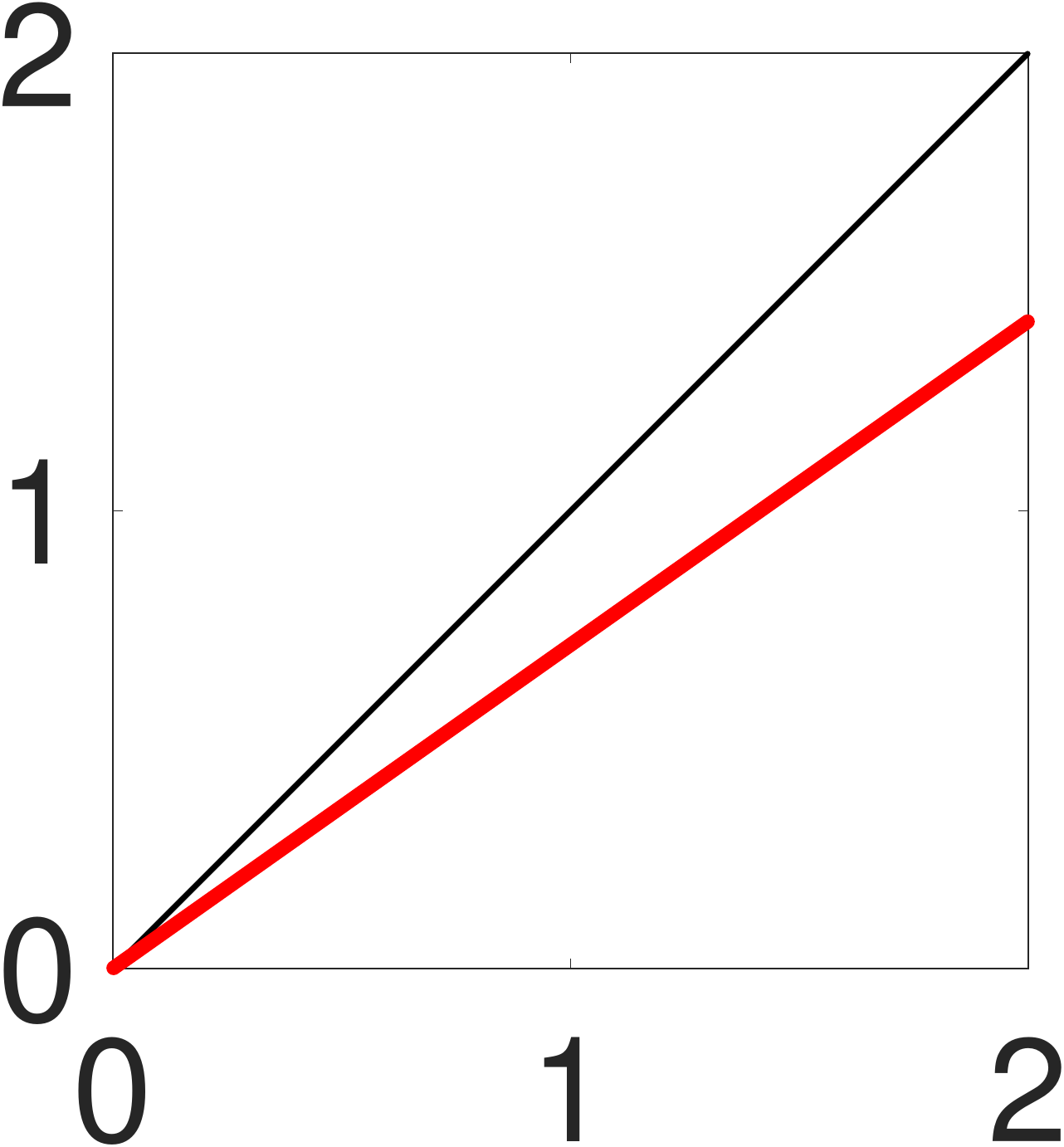}&\includegraphics[scale=0.13,valign=c]{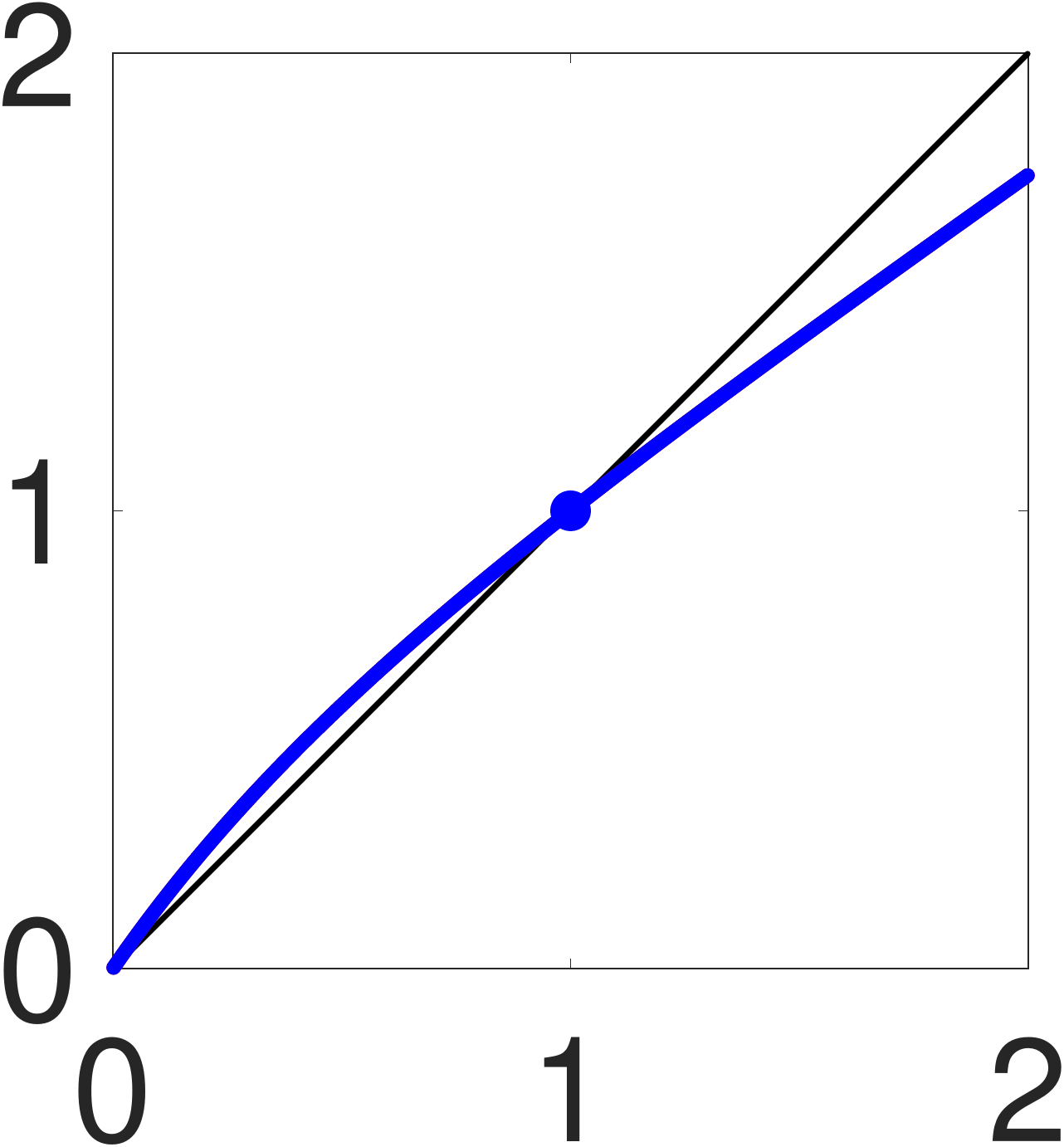}&\includegraphics[scale=0.13,valign=c]{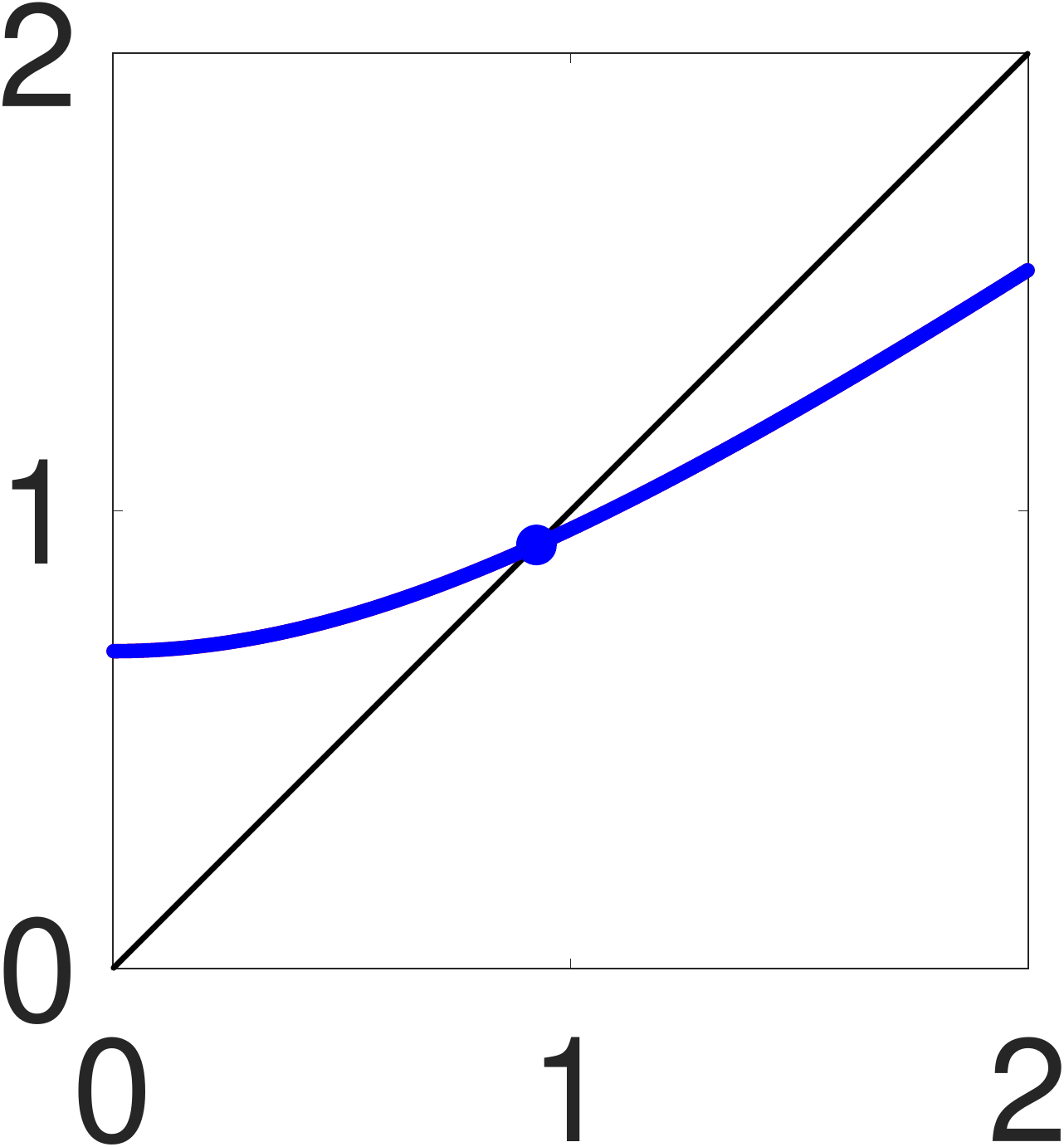}&\includegraphics[scale=0.13,valign=c]{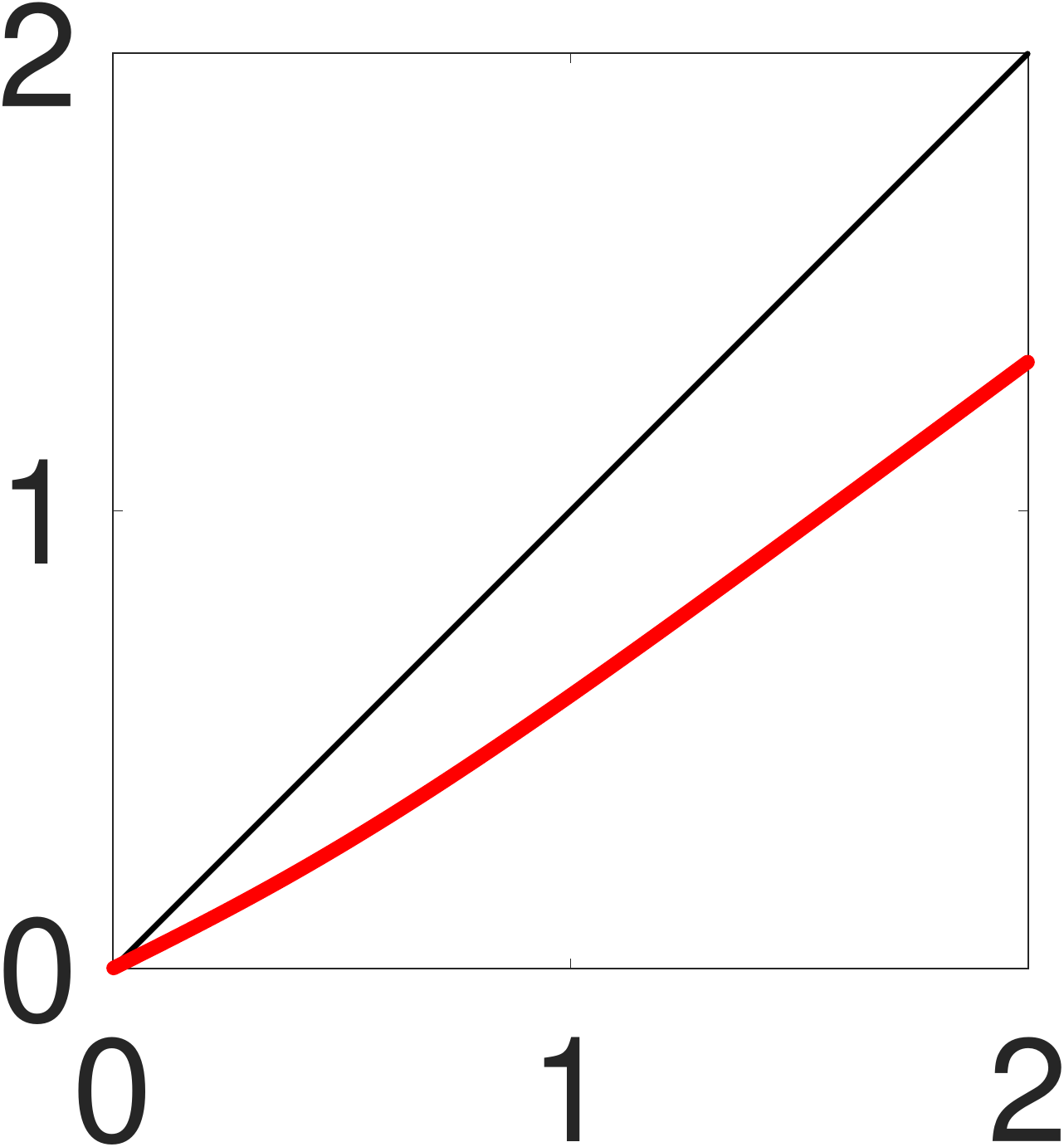}&\includegraphics[scale=0.13,valign=c]{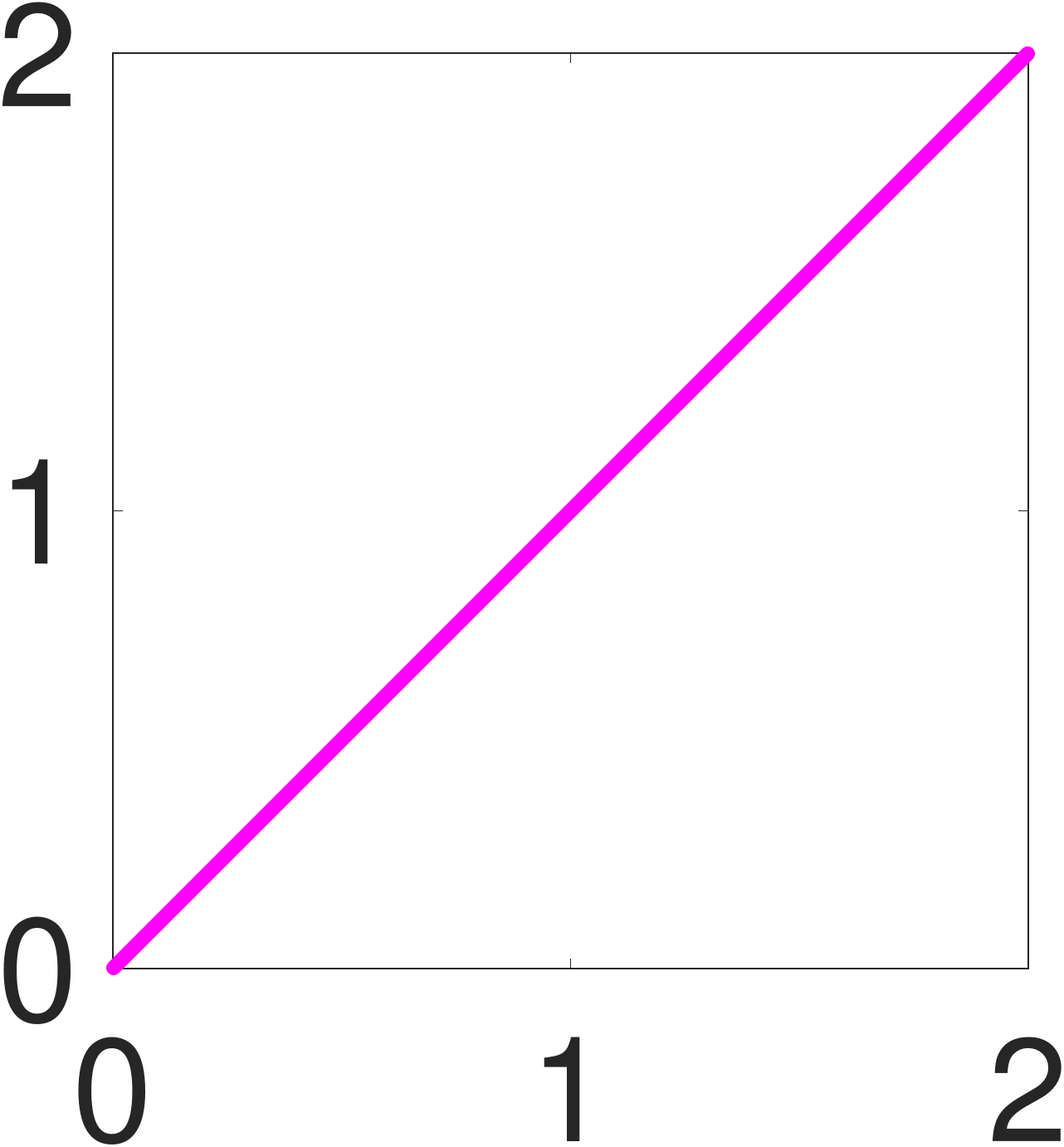}&\includegraphics[scale=0.13,valign=c]{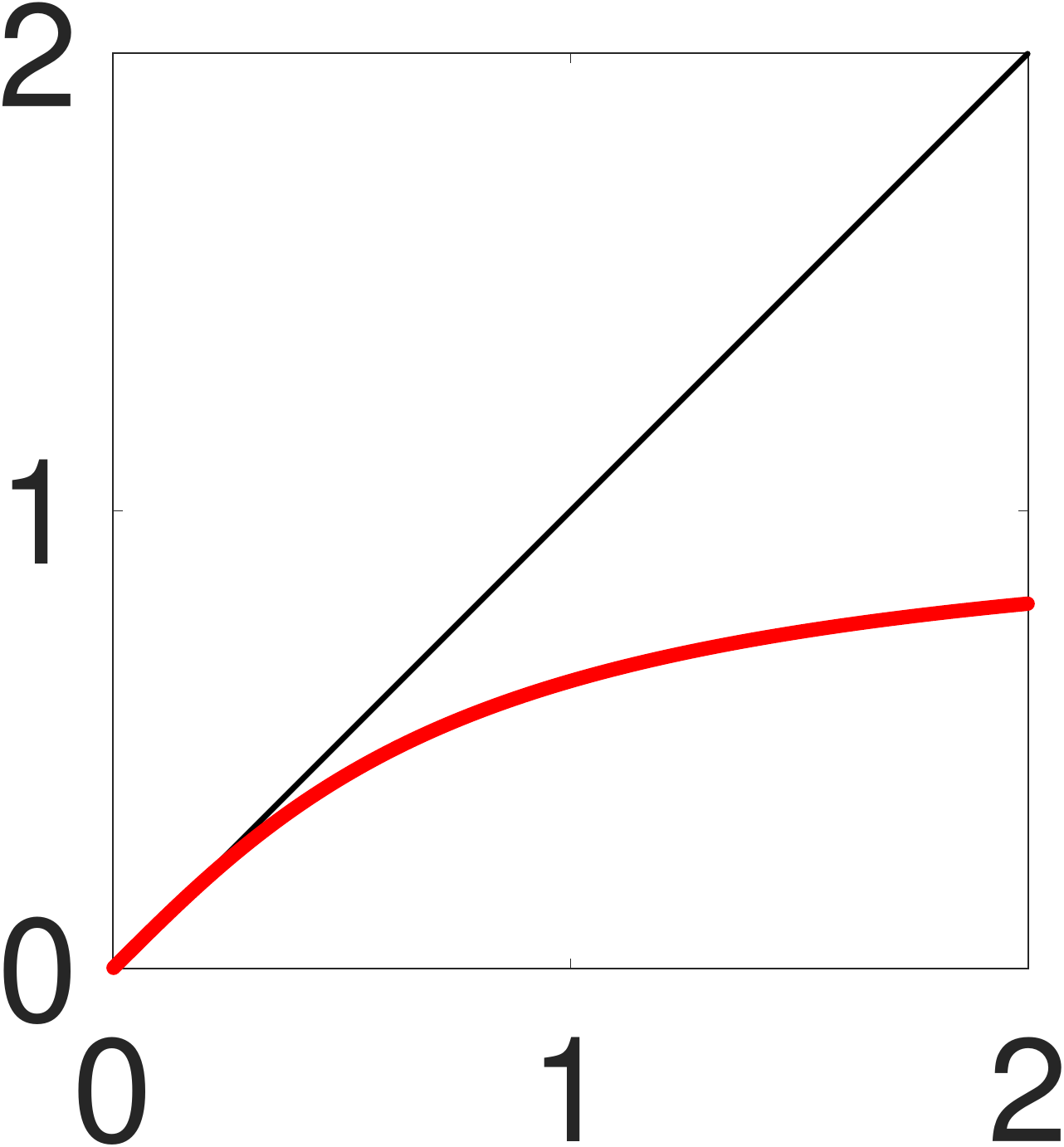}\\
Iter. limit&0&1.00&0.93&0&-&0\\
Conv. rate&$O(0.71^M)$&$O(0.78^M)$&$O(0.45^M)$&$O(0.5^M)$&-&$O(\frac{1}{M^{0.5}})$\\
$\frac{\mathfrak{L}_\tau(\lambda)}{\mathfrak{L}_\tau(1)}$&\includegraphics[scale=0.13,valign=c]{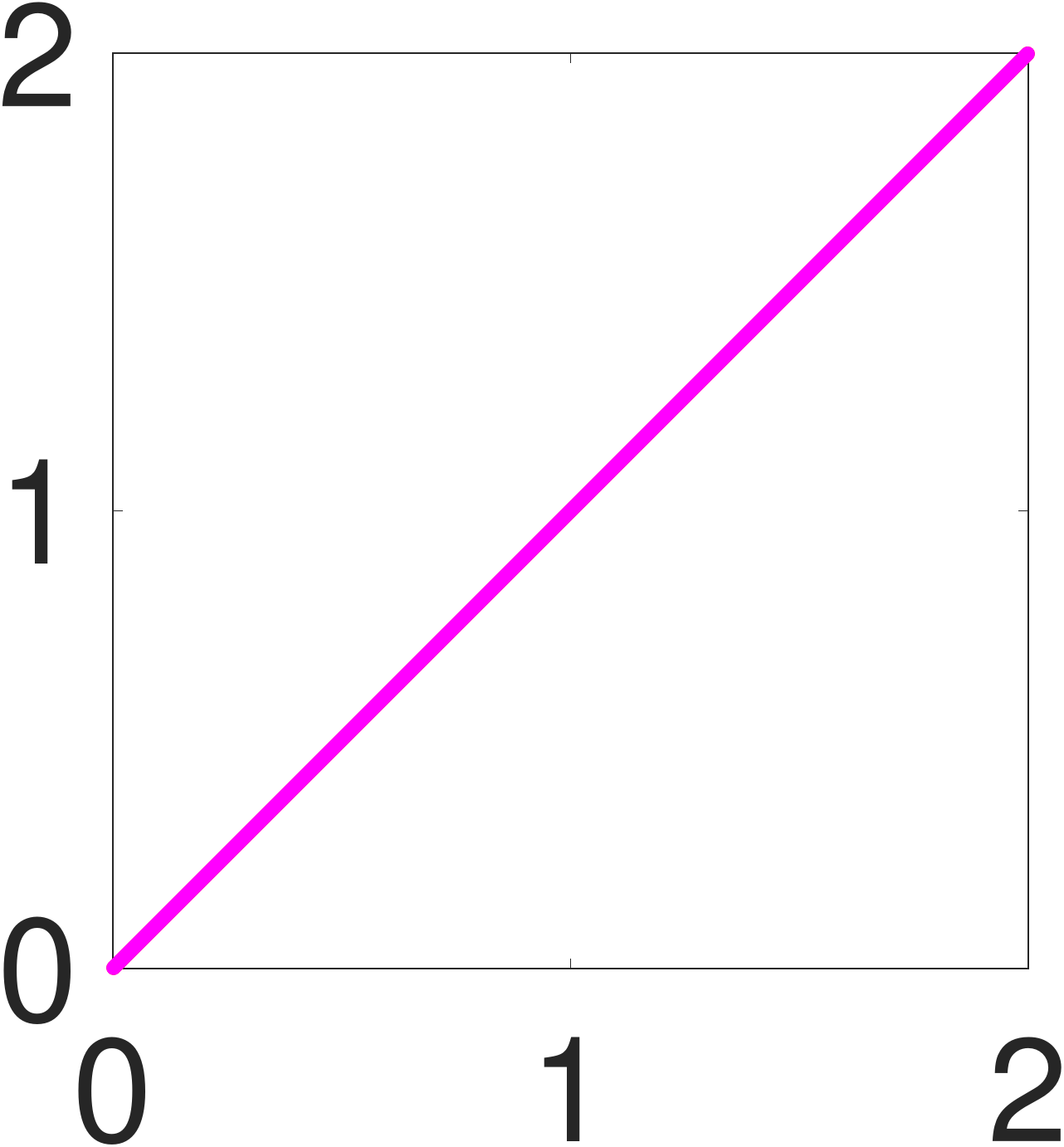}&\includegraphics[scale=0.13,valign=c]{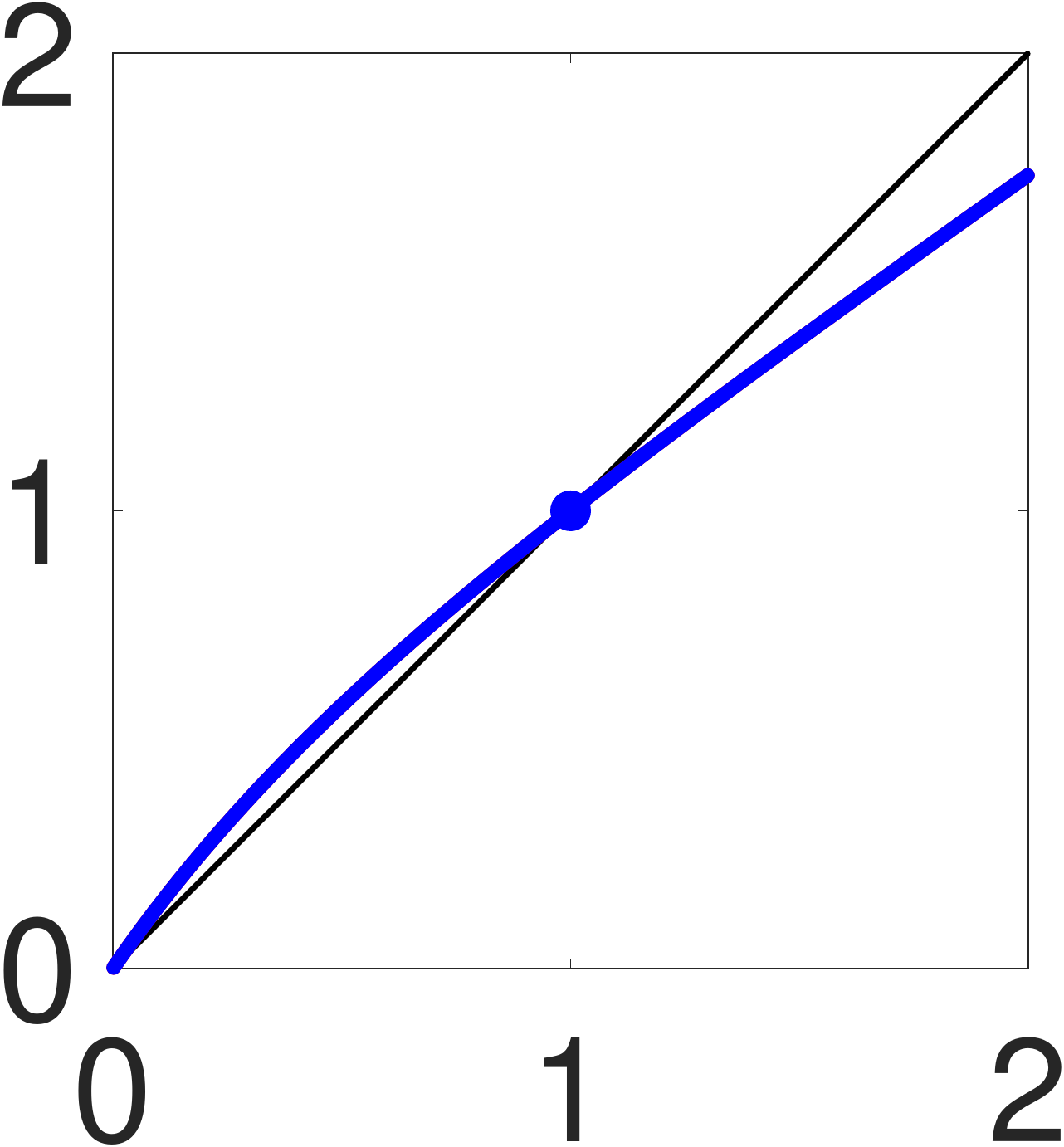}&\includegraphics[scale=0.13,valign=c]{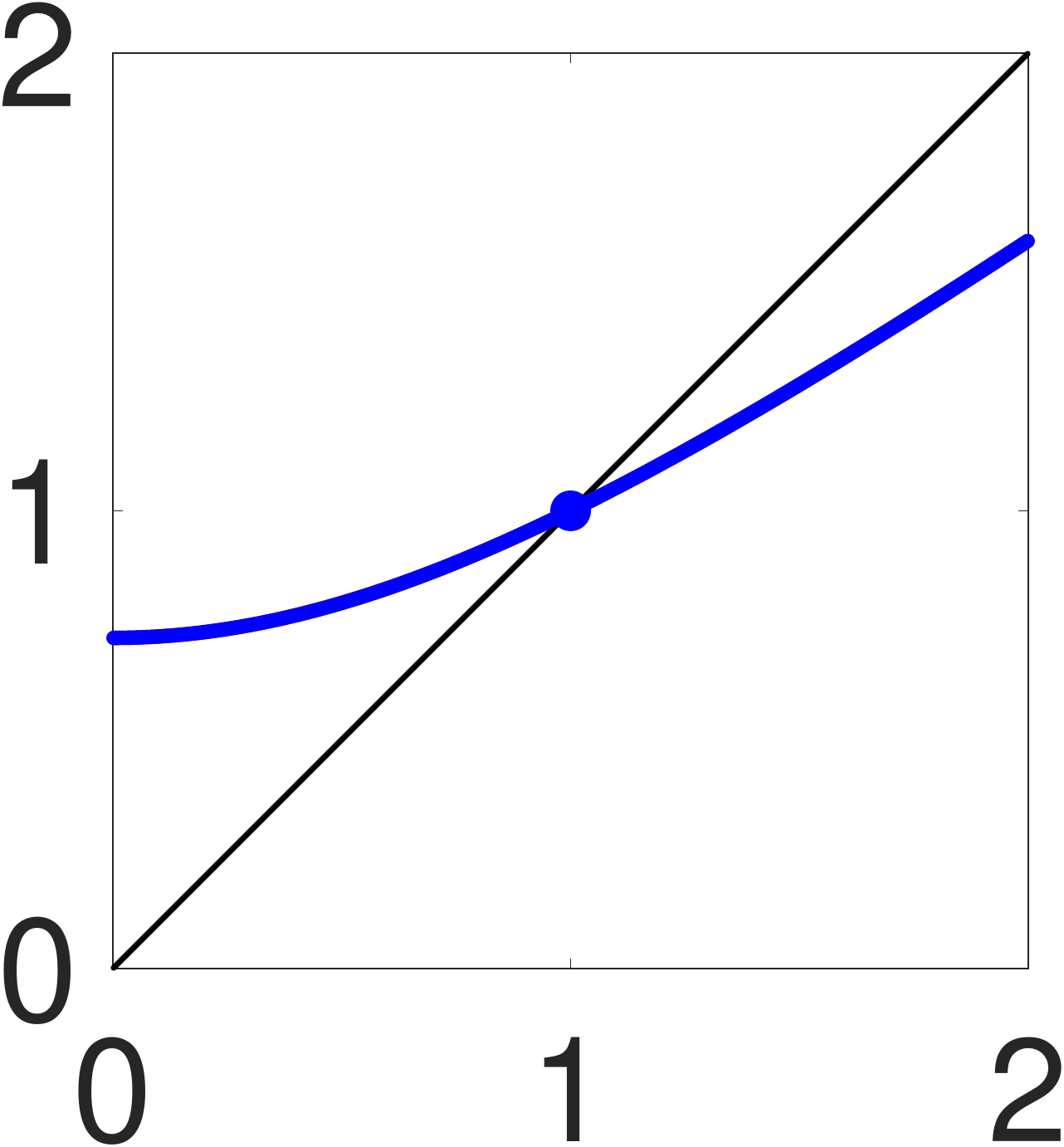}&\includegraphics[scale=0.13,valign=c]{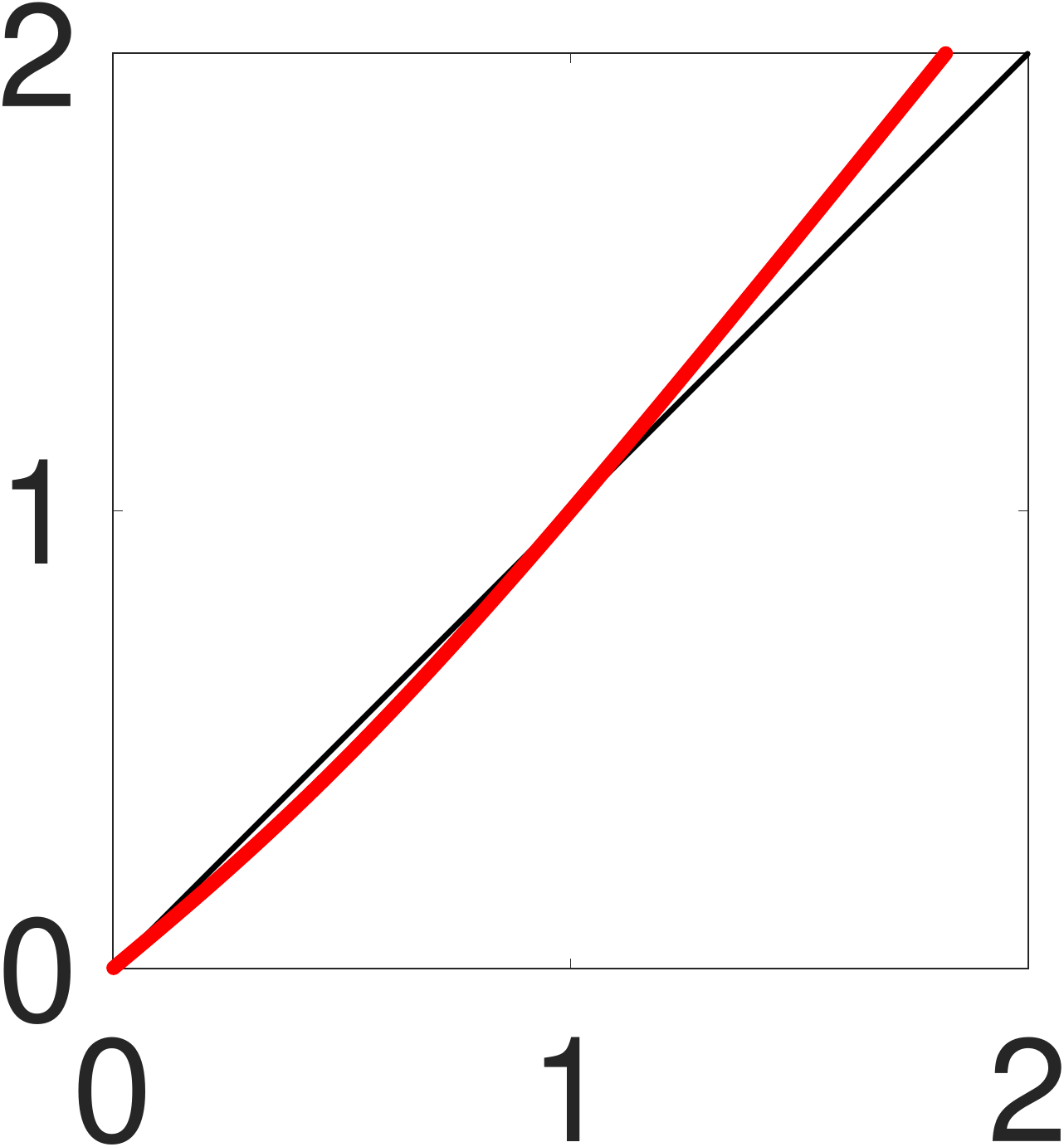}&\includegraphics[scale=0.13,valign=c]{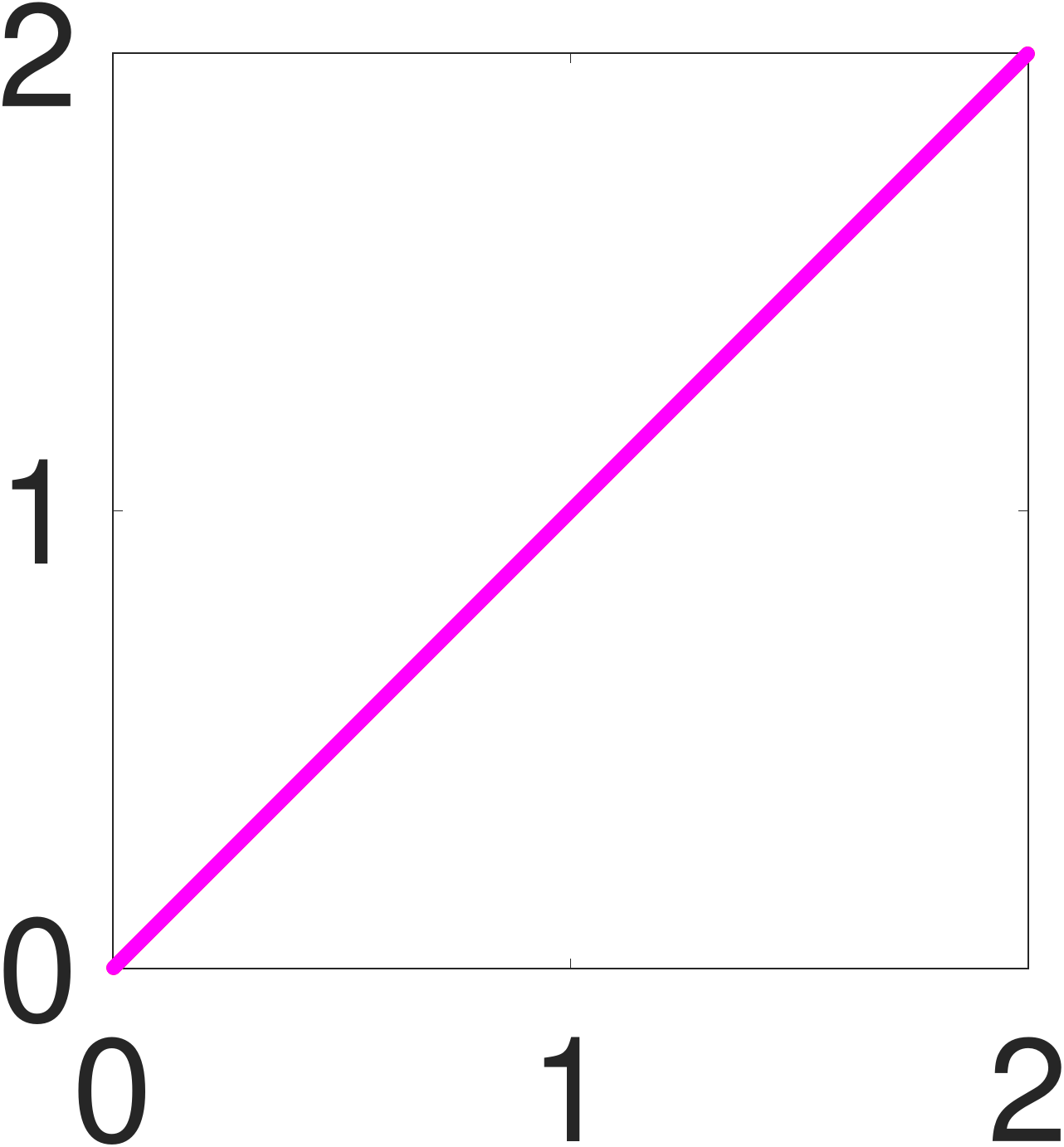}&\includegraphics[scale=0.13,valign=c]{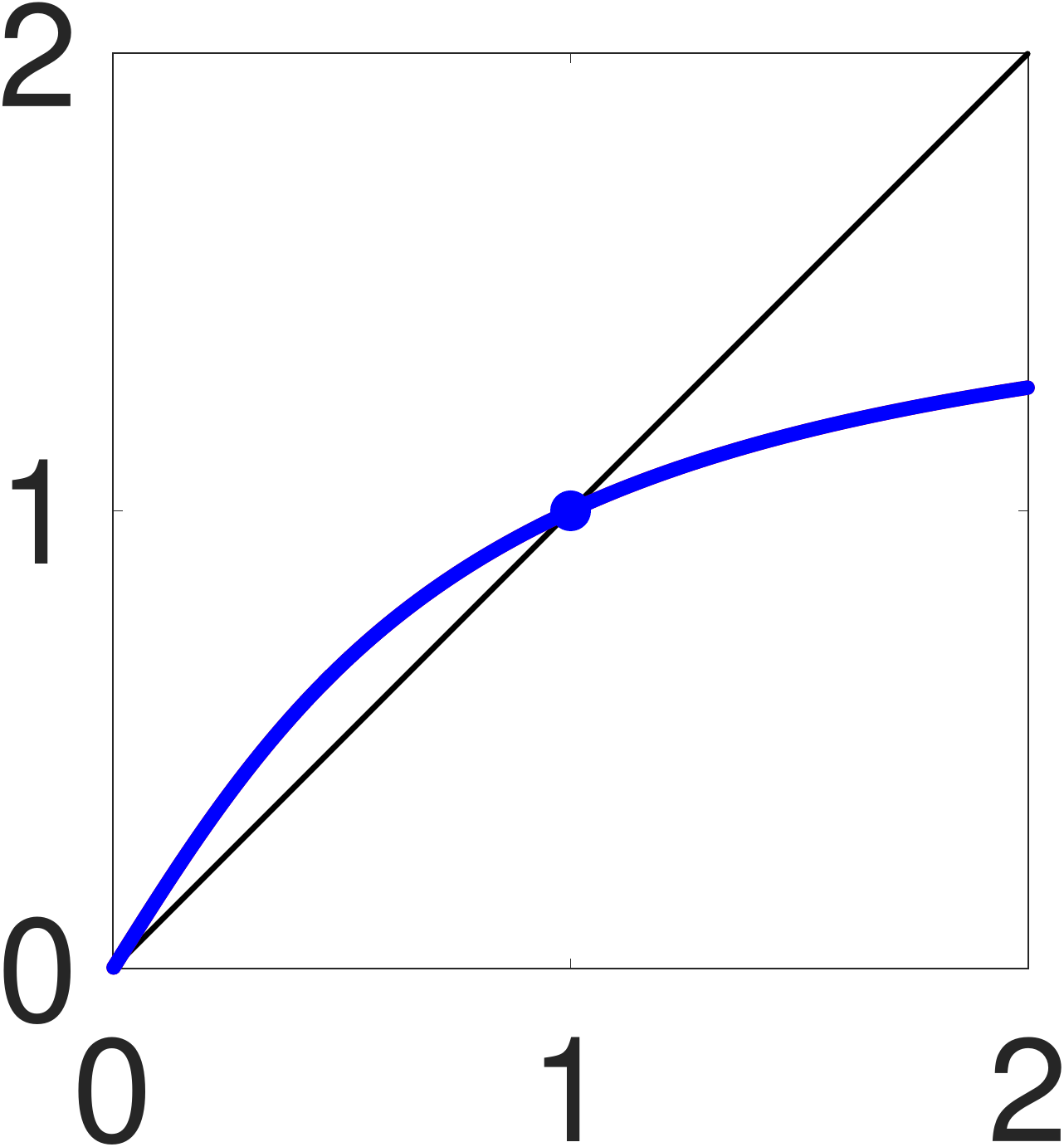}\\
Iter. limit&-&1&1&0&-&1\\
Conv. rate&-&$O(0.78^M)$&$O(0.49^M)$&$O(0.84^M)$&-&$O(0.46^M)$\\
$\frac{\mathfrak{L}_\tau'(\lambda)\lambda}{\mathfrak{L}_\tau(\lambda)}$ &\includegraphics[scale=0.13,valign=c]{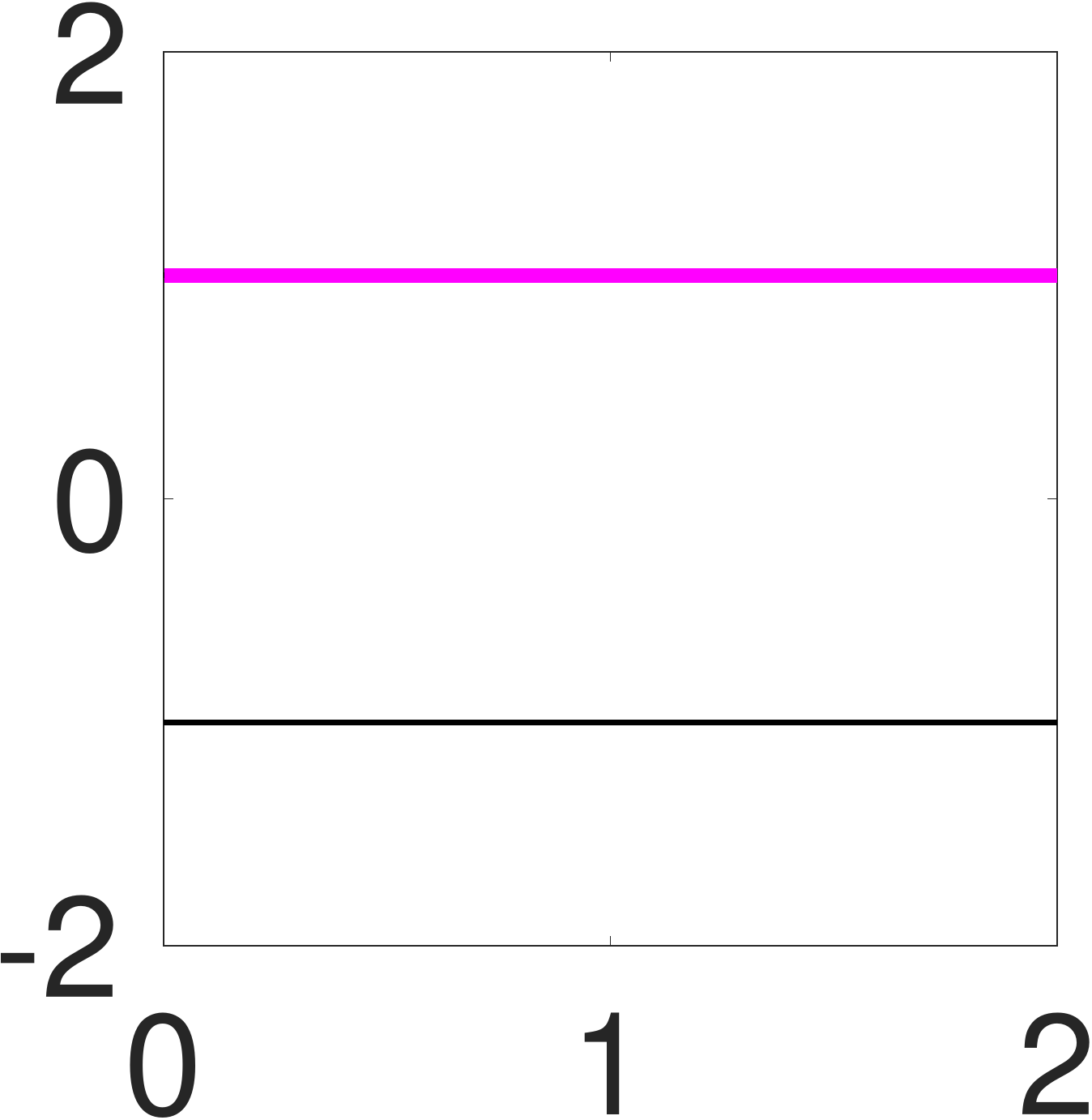}&\includegraphics[scale=0.13,valign=c]{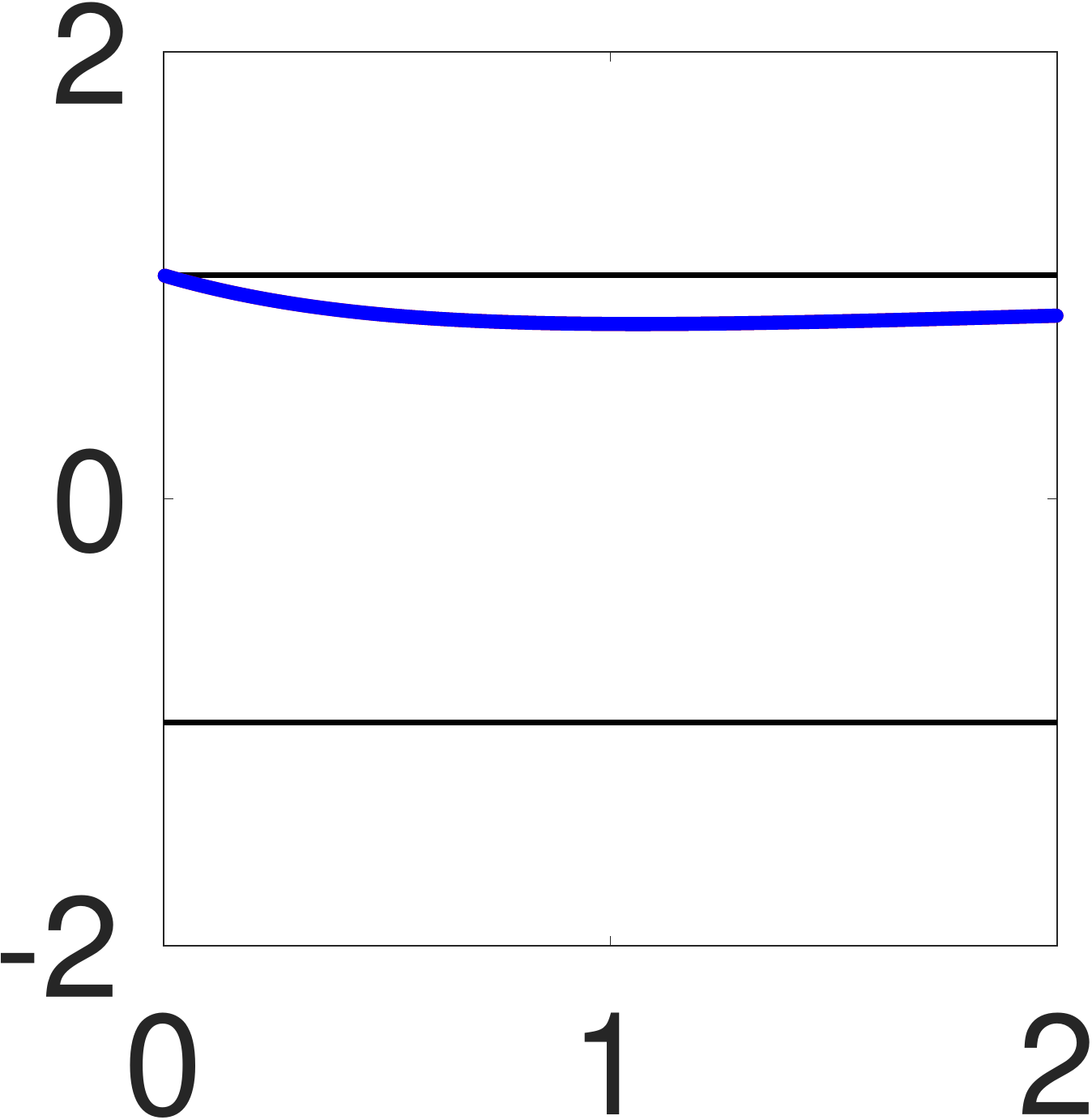}&\includegraphics[scale=0.13,valign=c]{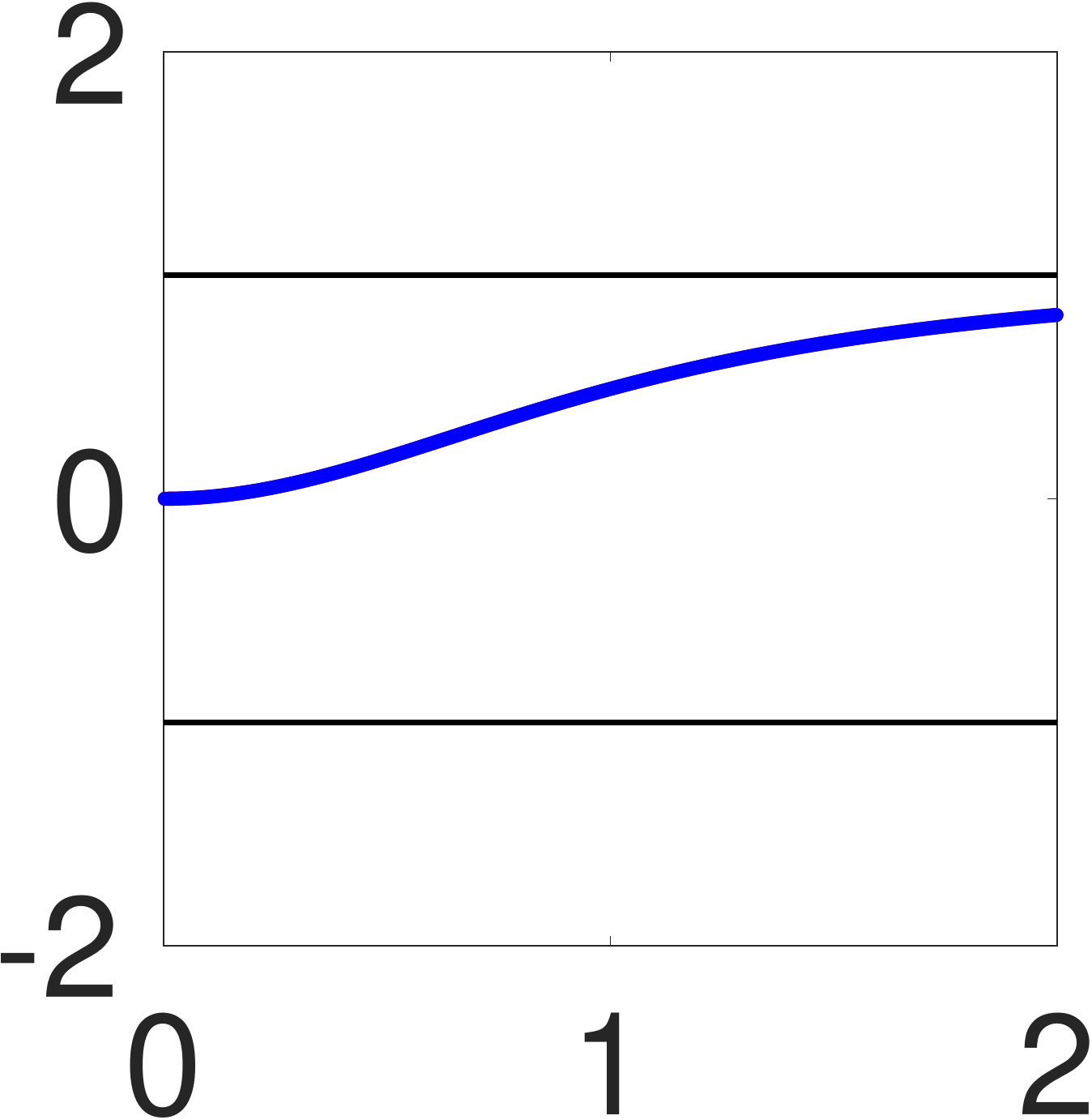}&\includegraphics[scale=0.13,valign=c]{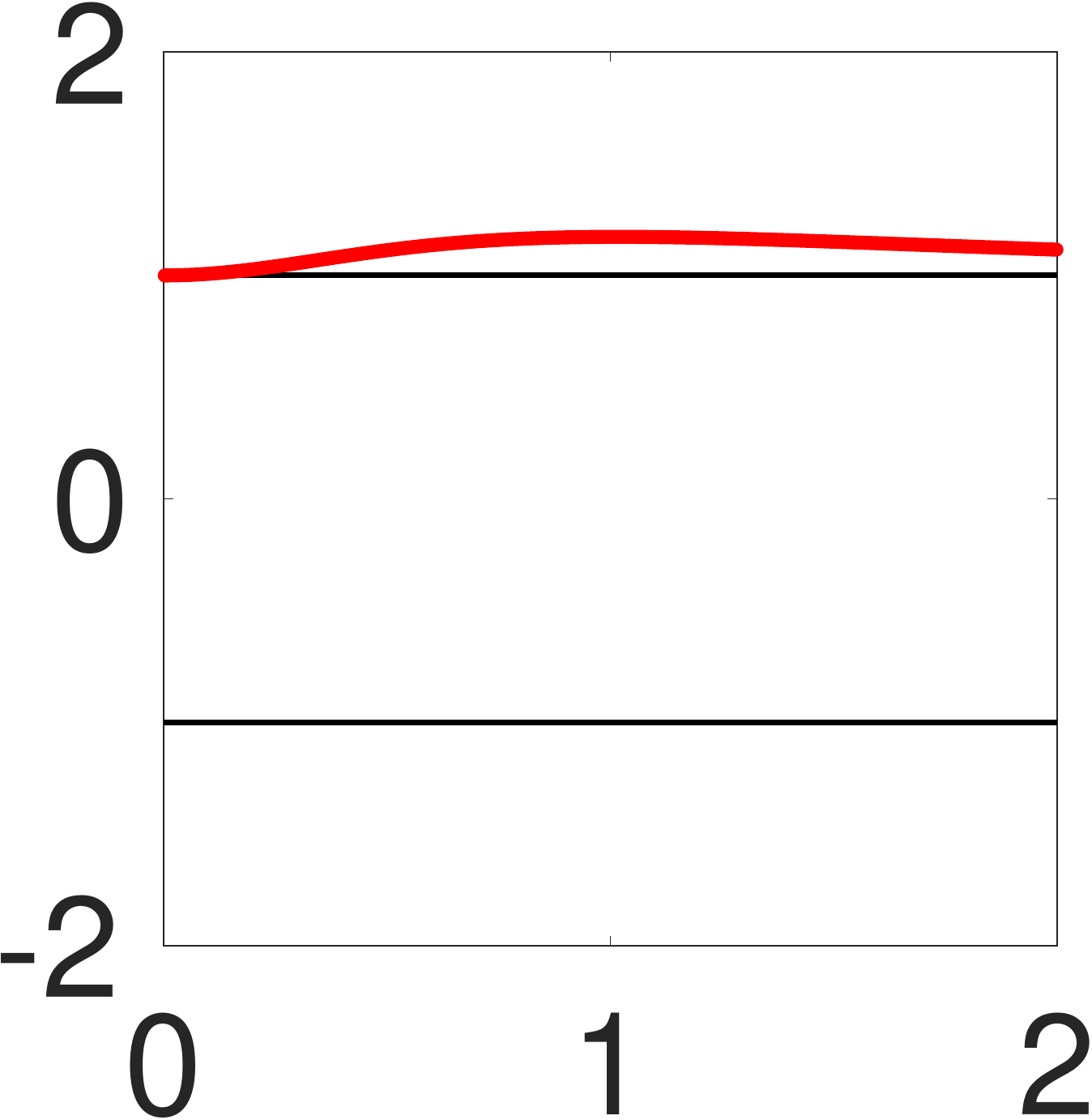}&\includegraphics[scale=0.13,valign=c]{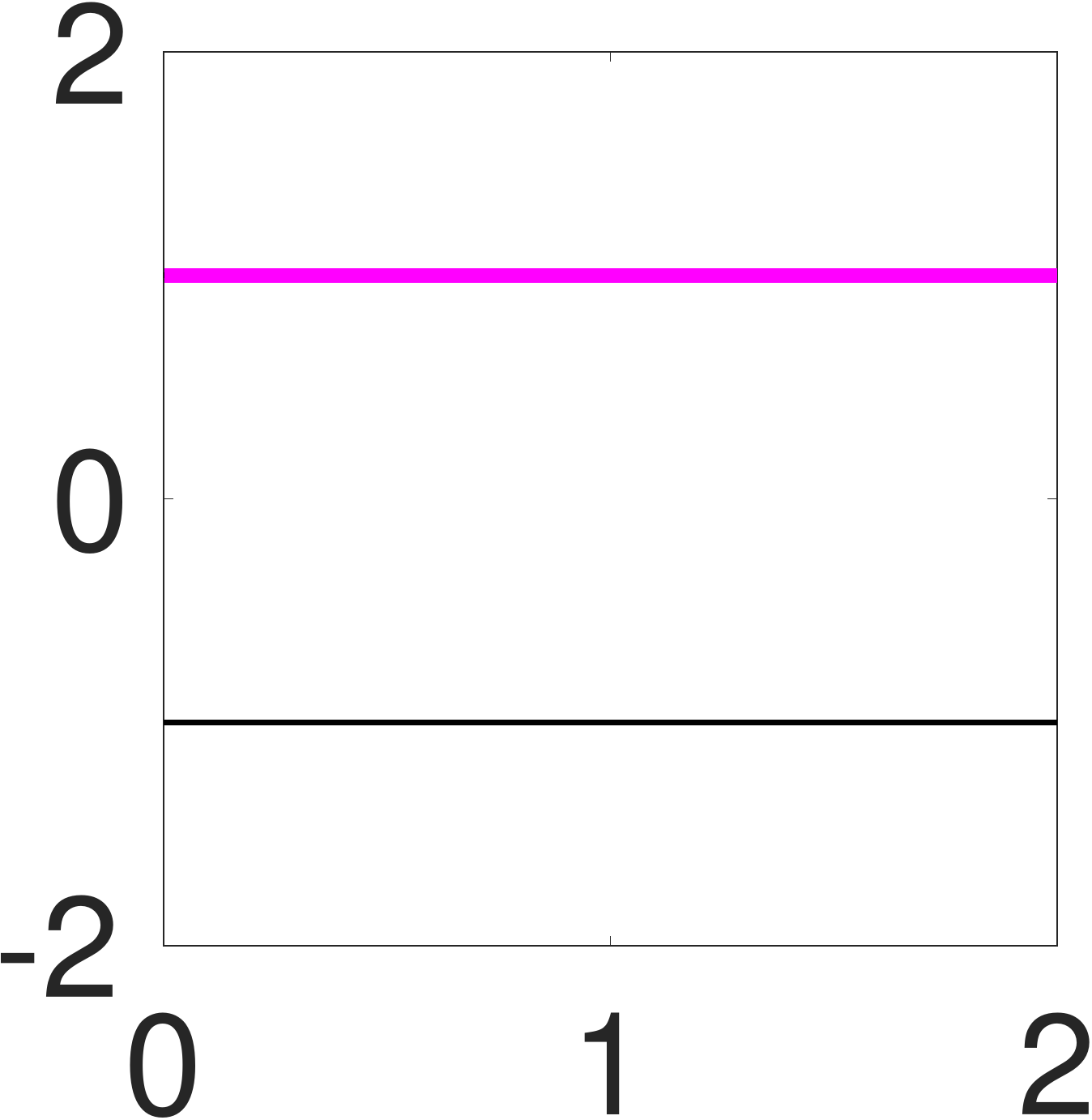}&\includegraphics[scale=0.13,valign=c]{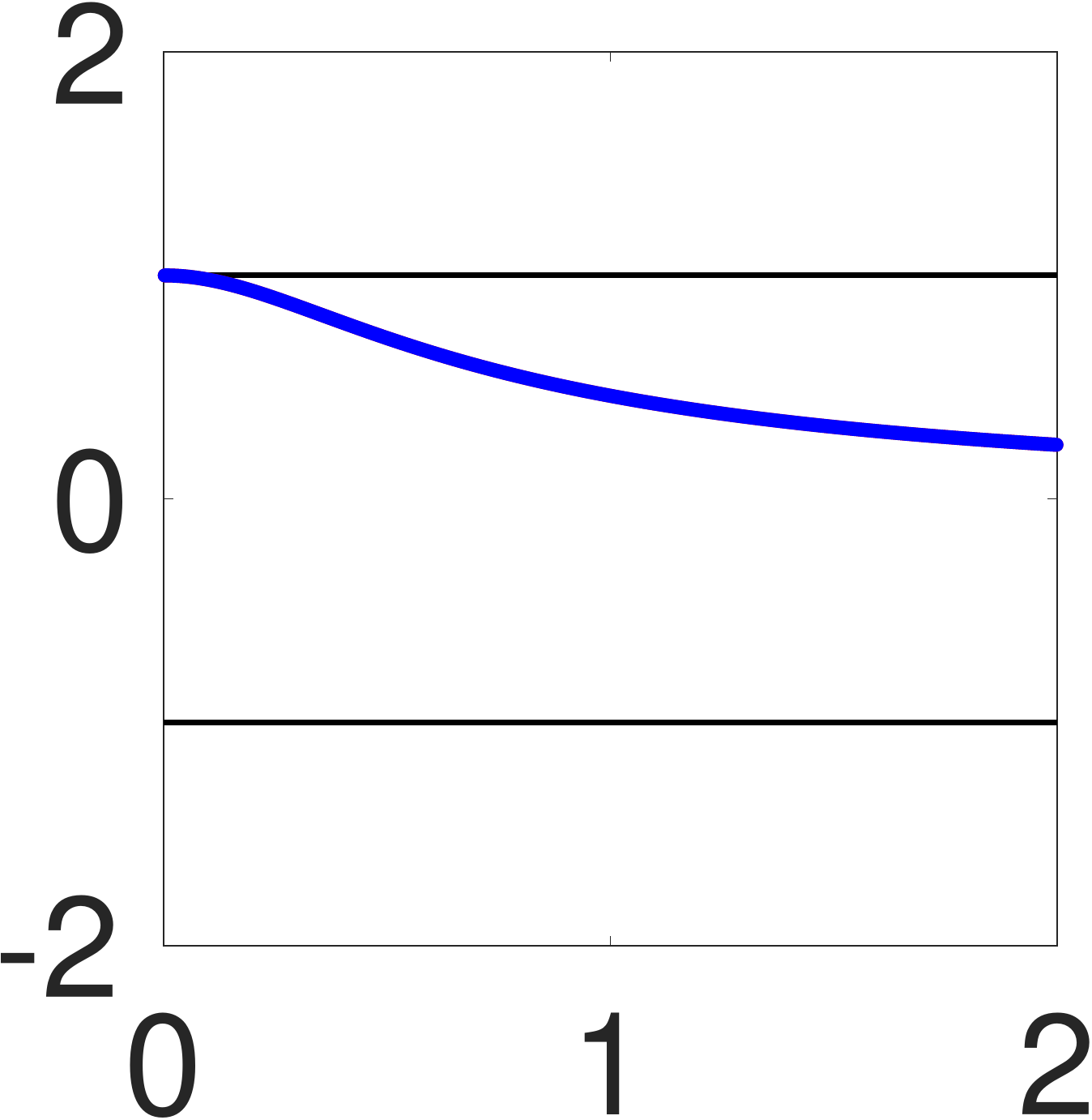}\\
$\tilde{\mathfrak{C}}_\tau(c)$&\includegraphics[scale=0.13,valign=c]{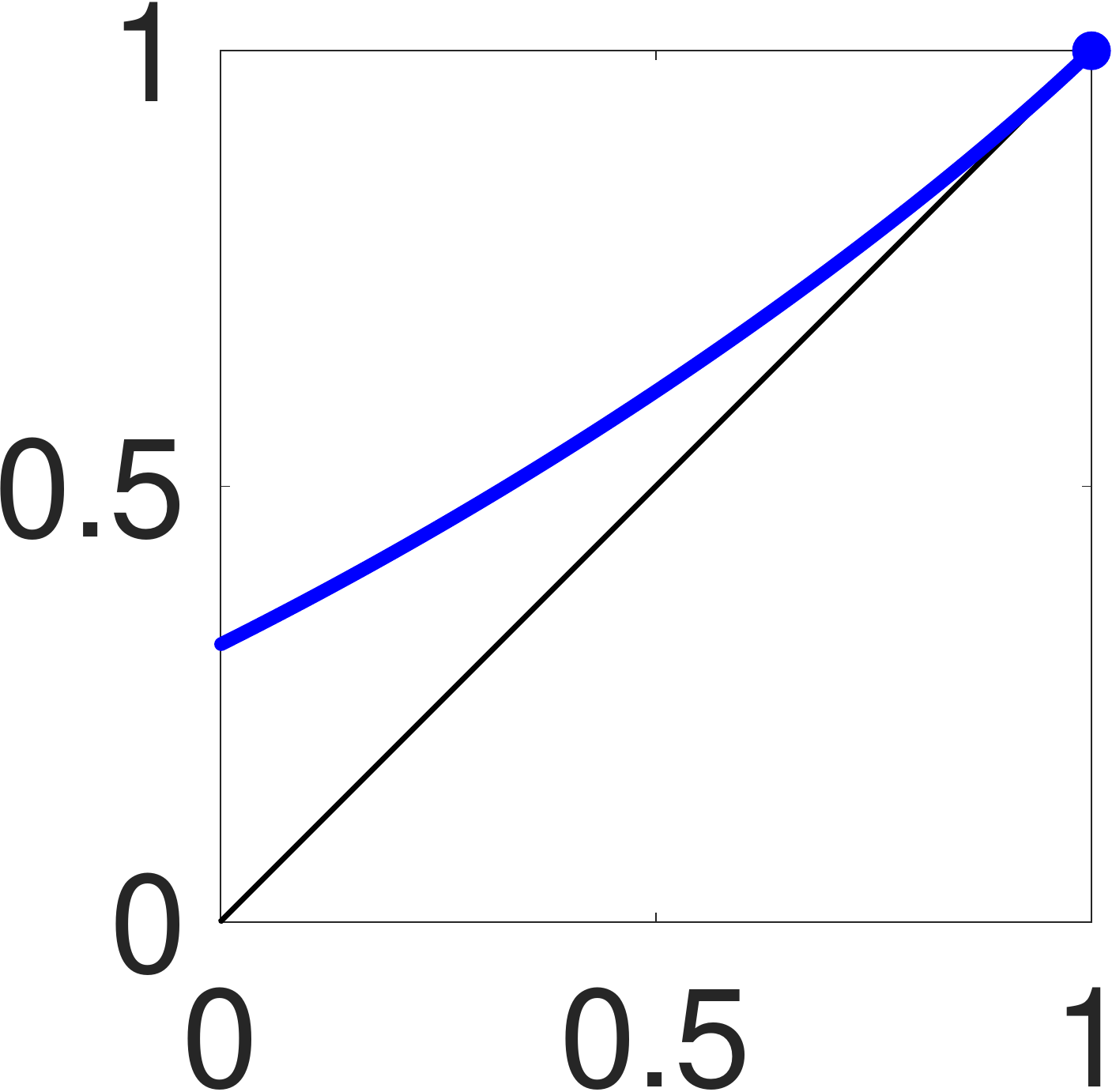}&\includegraphics[scale=0.13,valign=c]{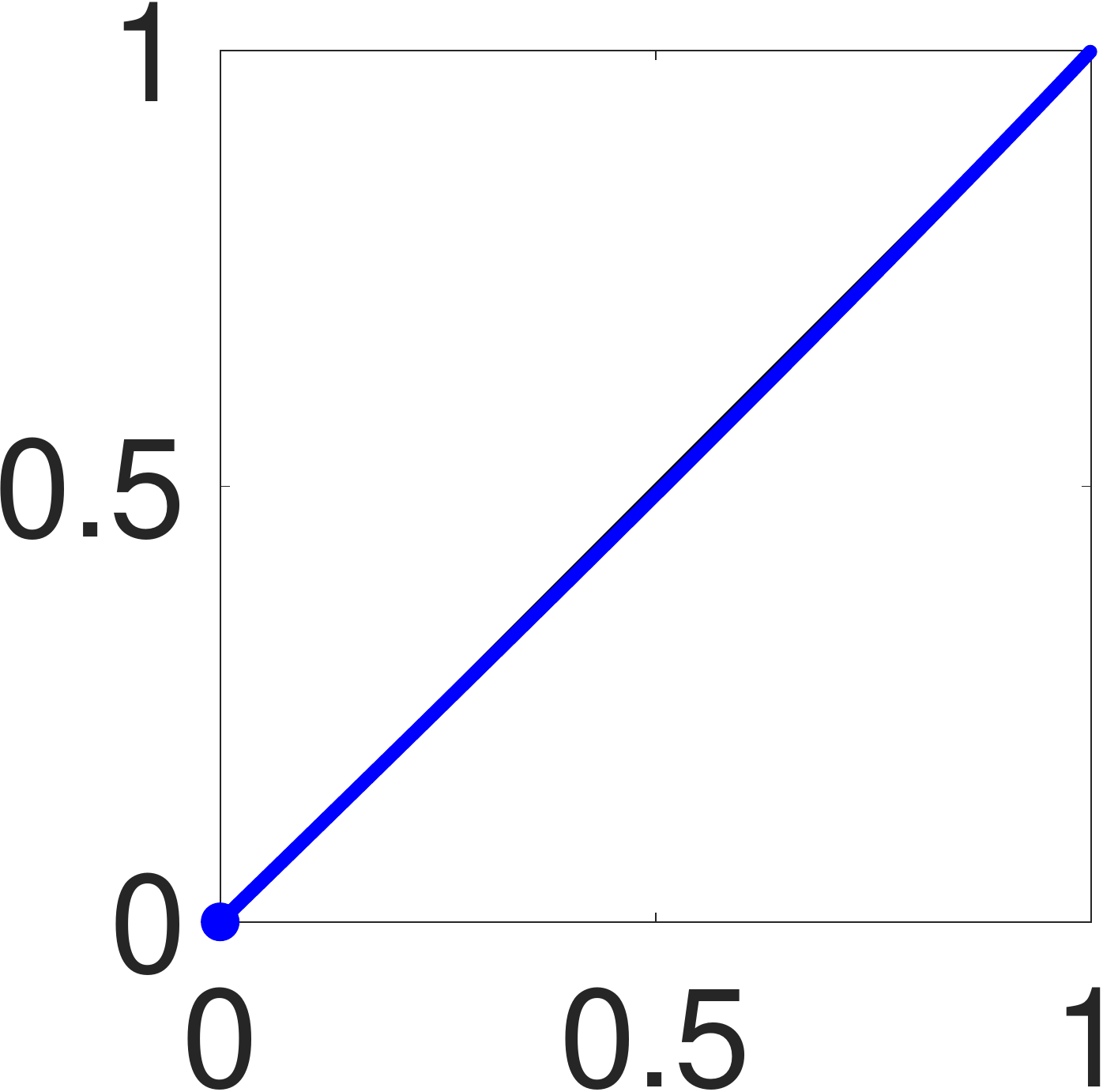}&\includegraphics[scale=0.13,valign=c]{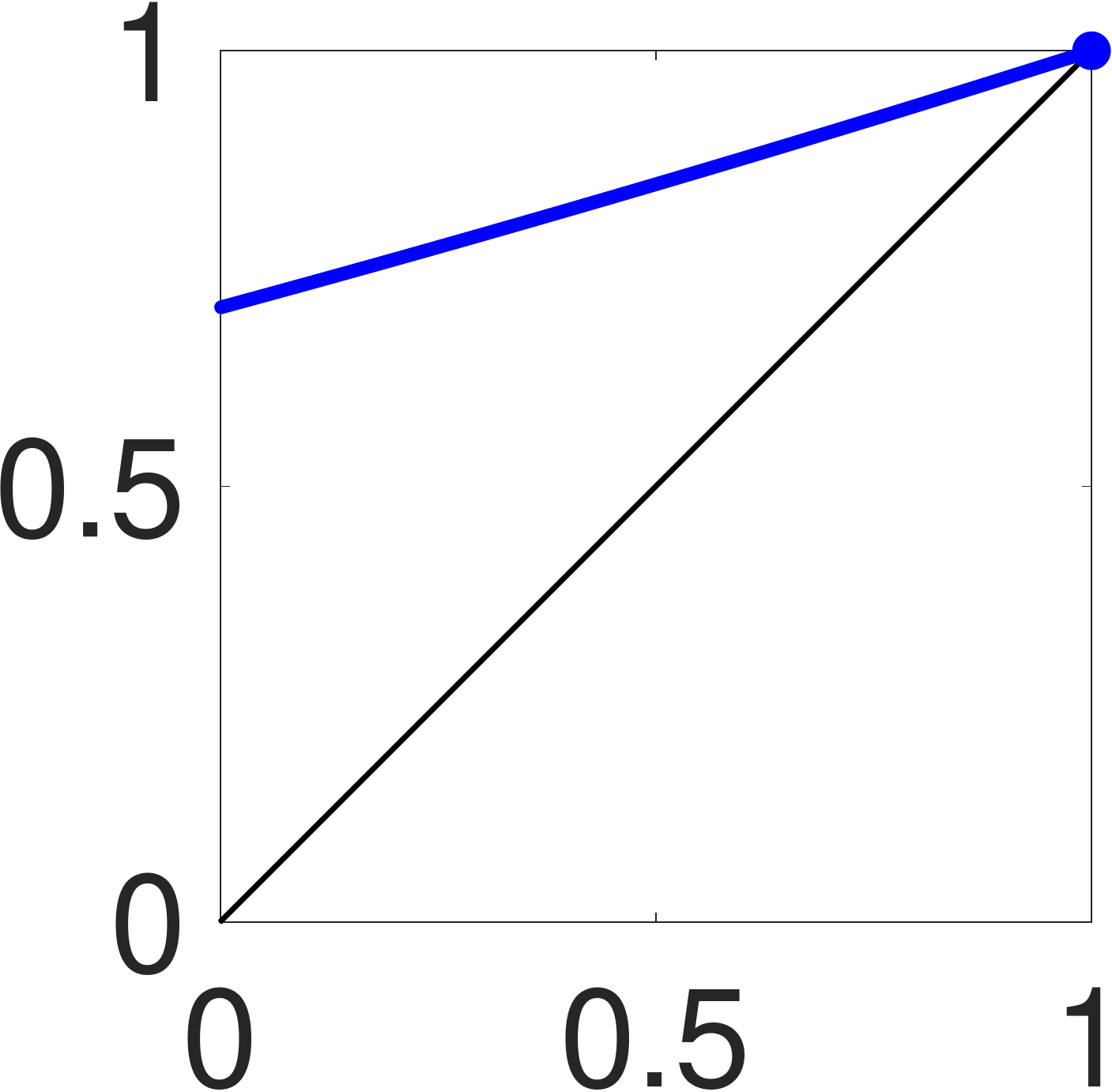}&\includegraphics[scale=0.13,valign=c]{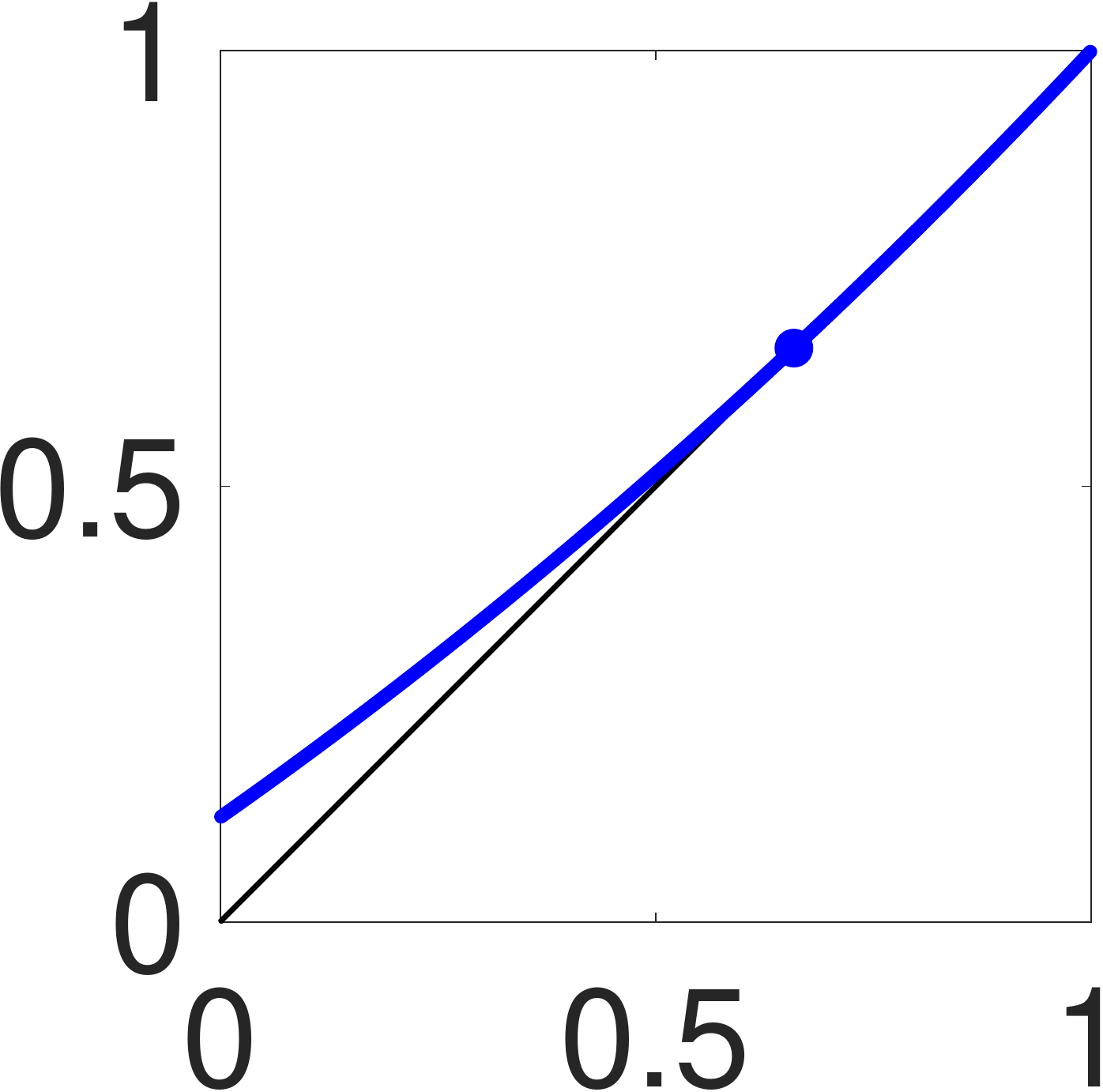}&\includegraphics[scale=0.13,valign=c]{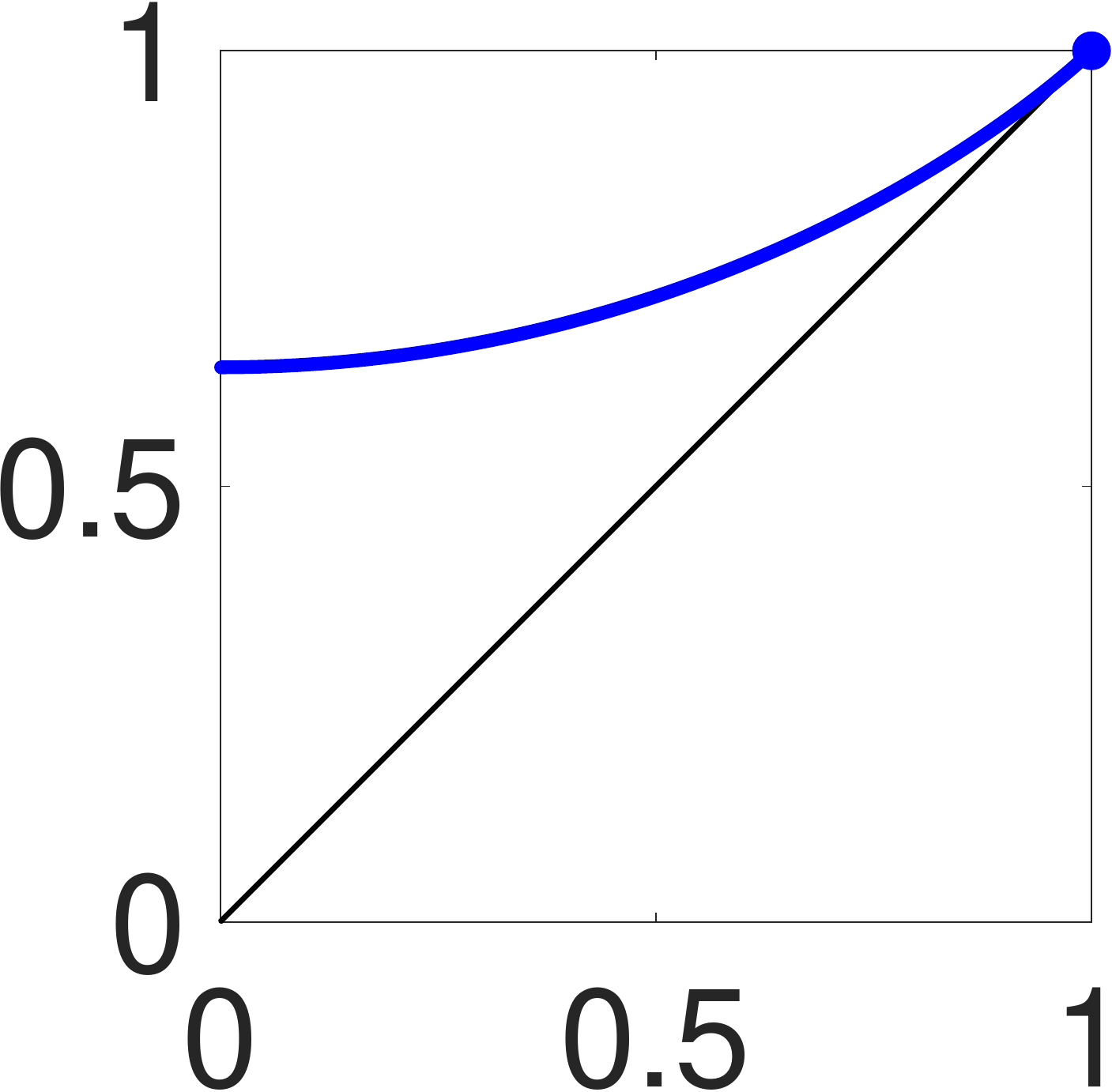}&\includegraphics[scale=0.13,valign=c]{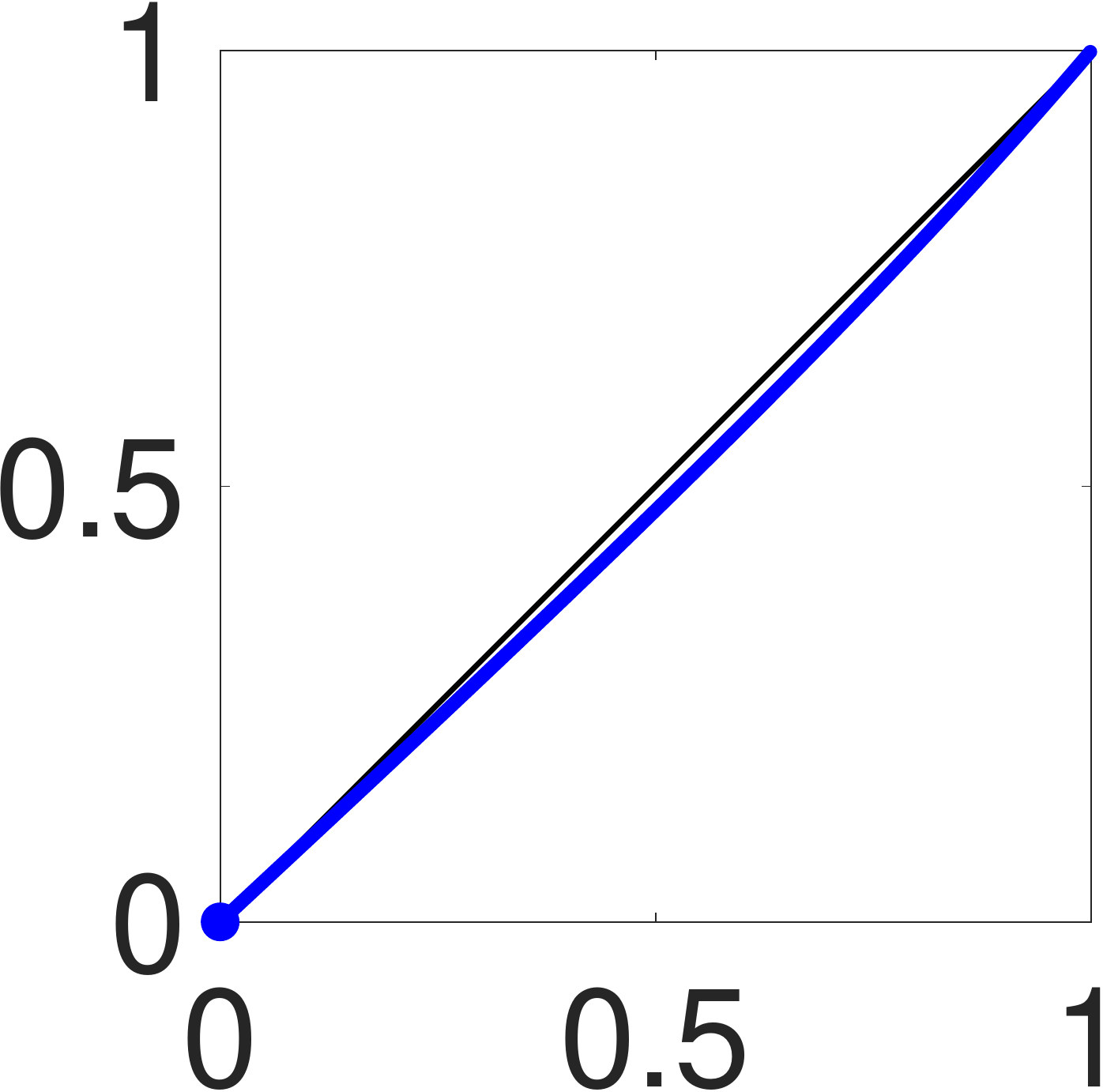}\\
Iter. limit &1&0&1&0.66&1&0\\
Conv. rate&$O(\frac{1}{M^2})$&$O(0.97^M)$&$O(0.32^M)$&$O(0.94^M)$&$O(\frac{1}{M^2})$&$O(0.93^M)$\\
$\mathfrak{n}_\tau(\lambda^2, 0)$&\includegraphics[scale=0.13,valign=c]{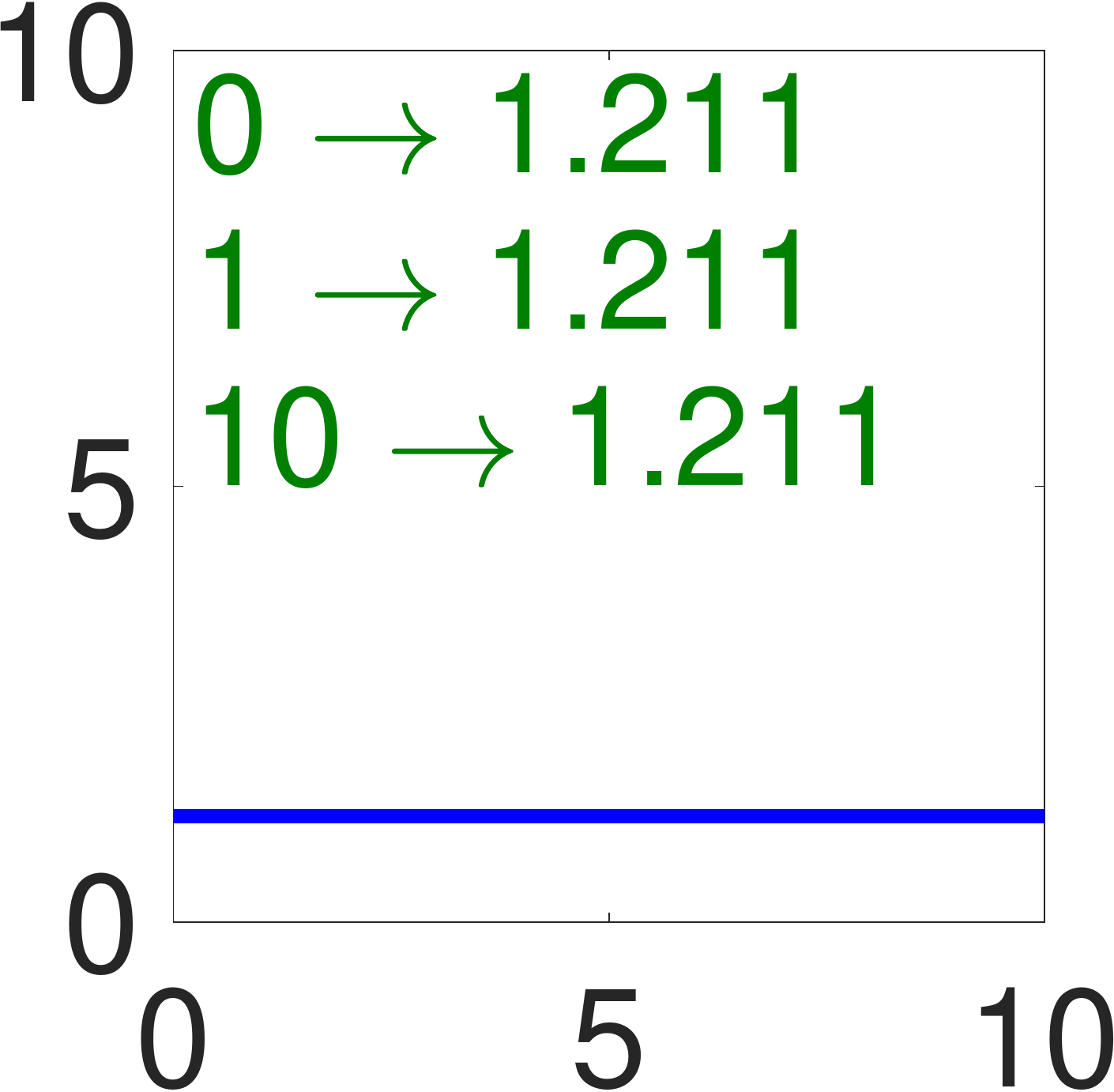}&\includegraphics[scale=0.13,valign=c]{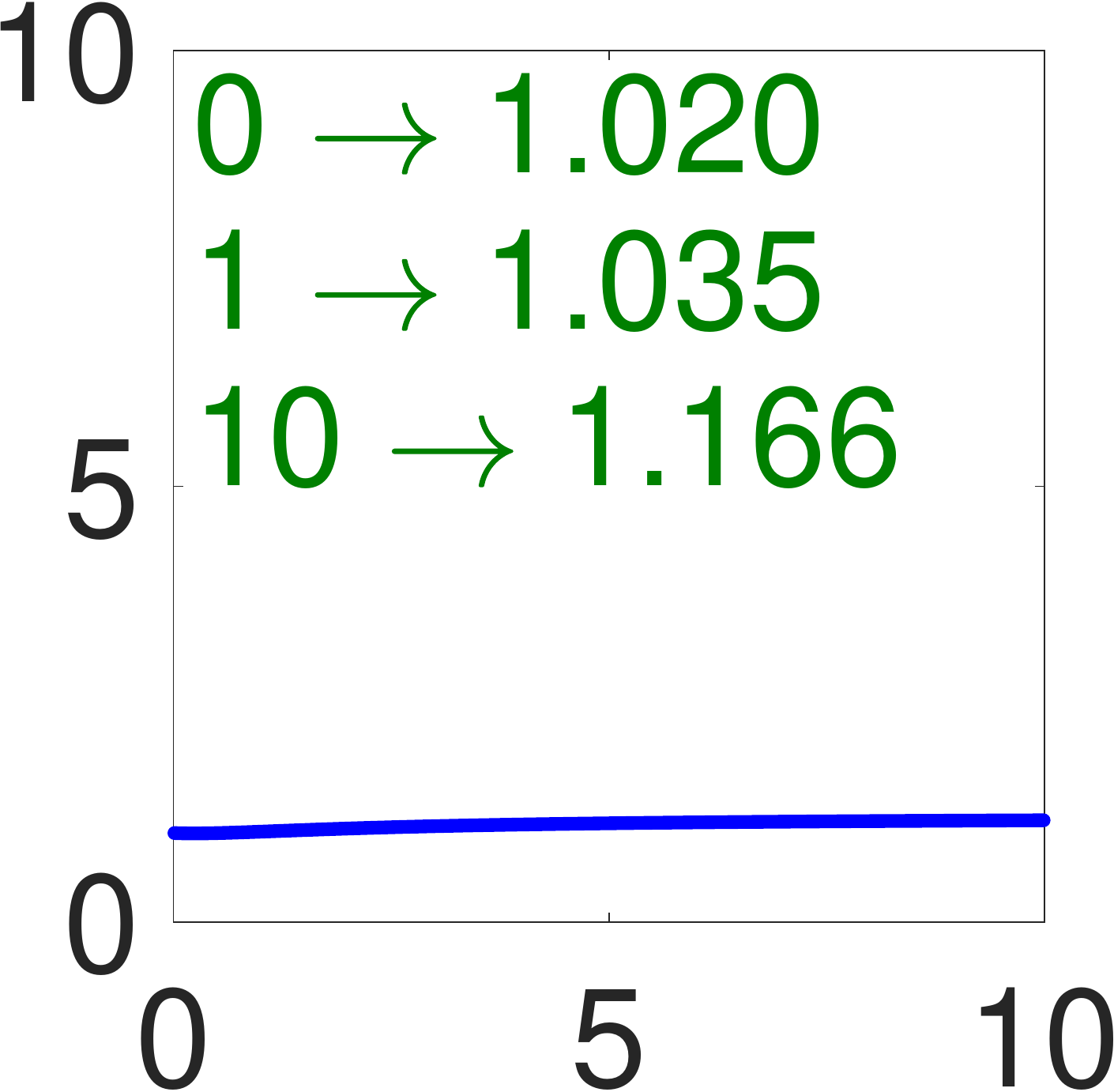}&\includegraphics[scale=0.13,valign=c]{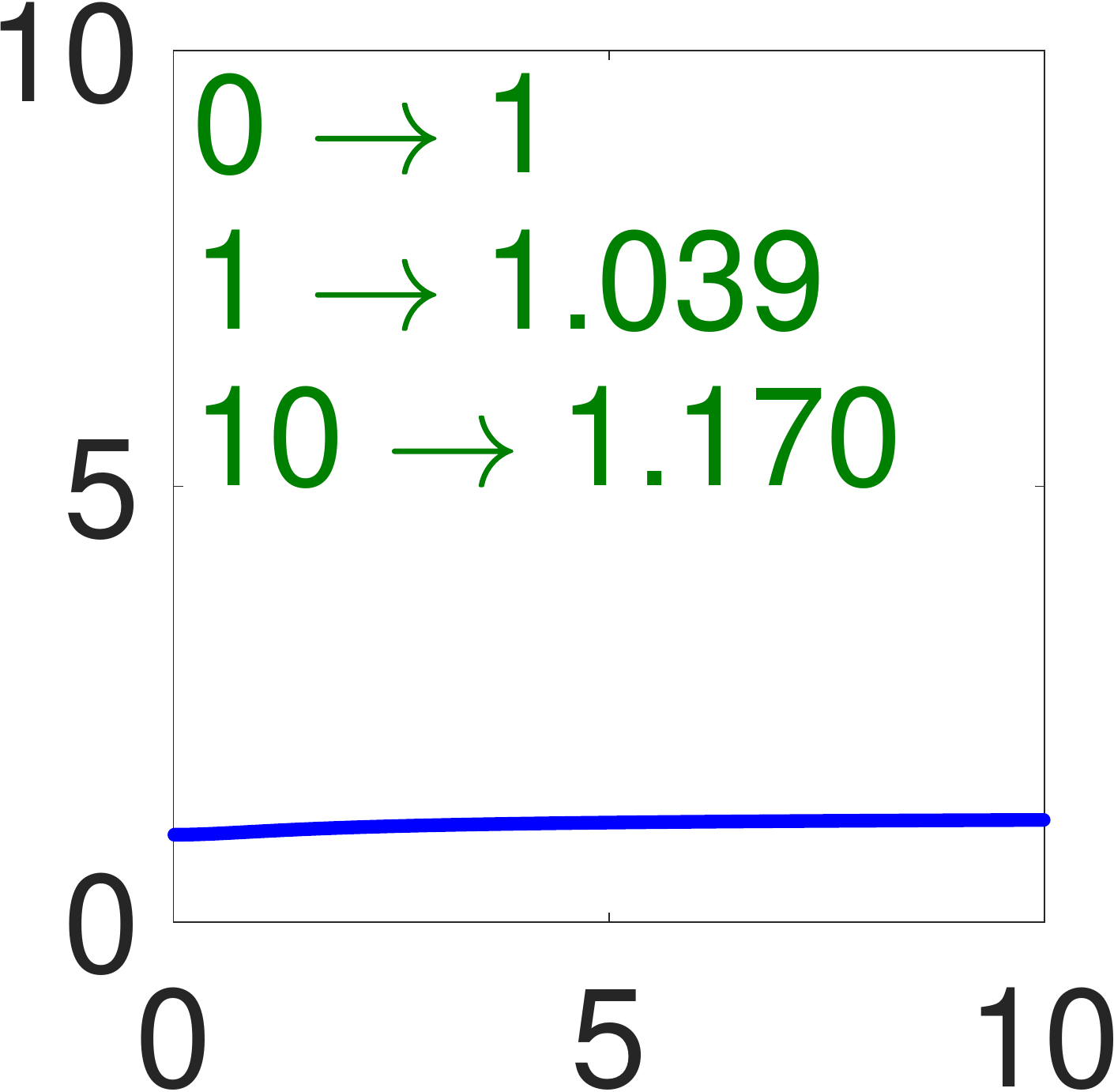}&\includegraphics[scale=0.13,valign=c]{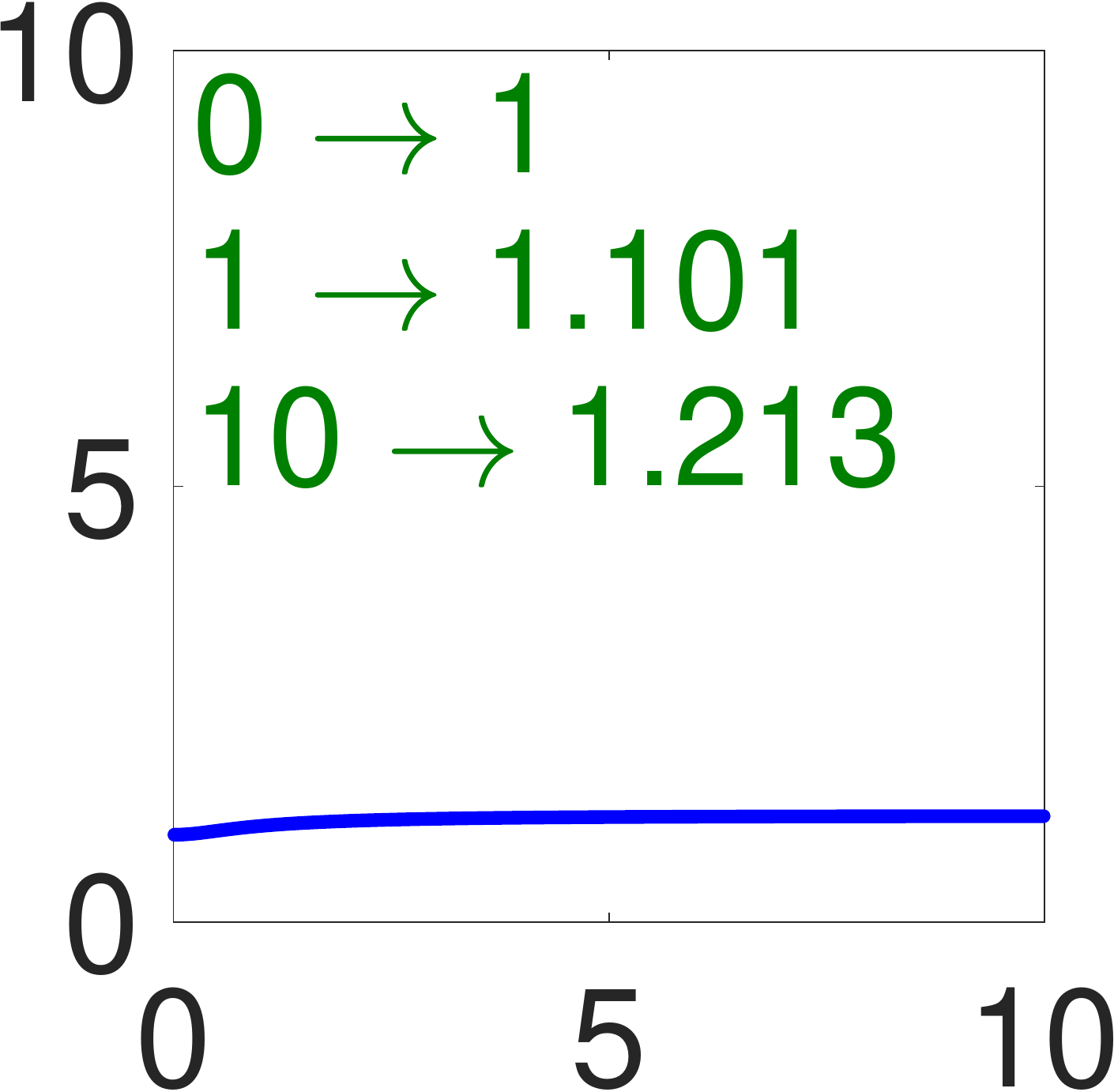}&\includegraphics[scale=0.13,valign=c]{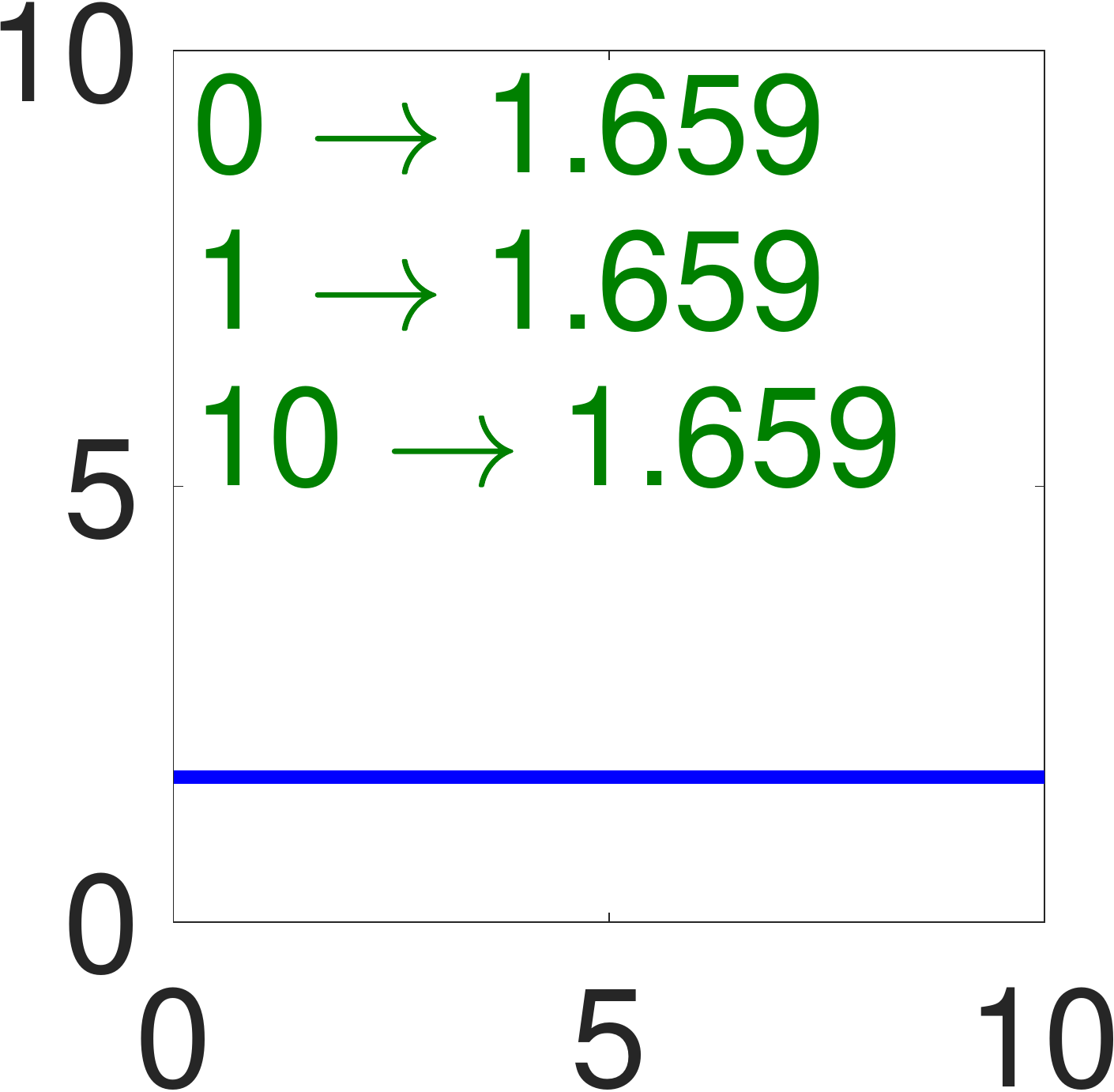}&\includegraphics[scale=0.13,valign=c]{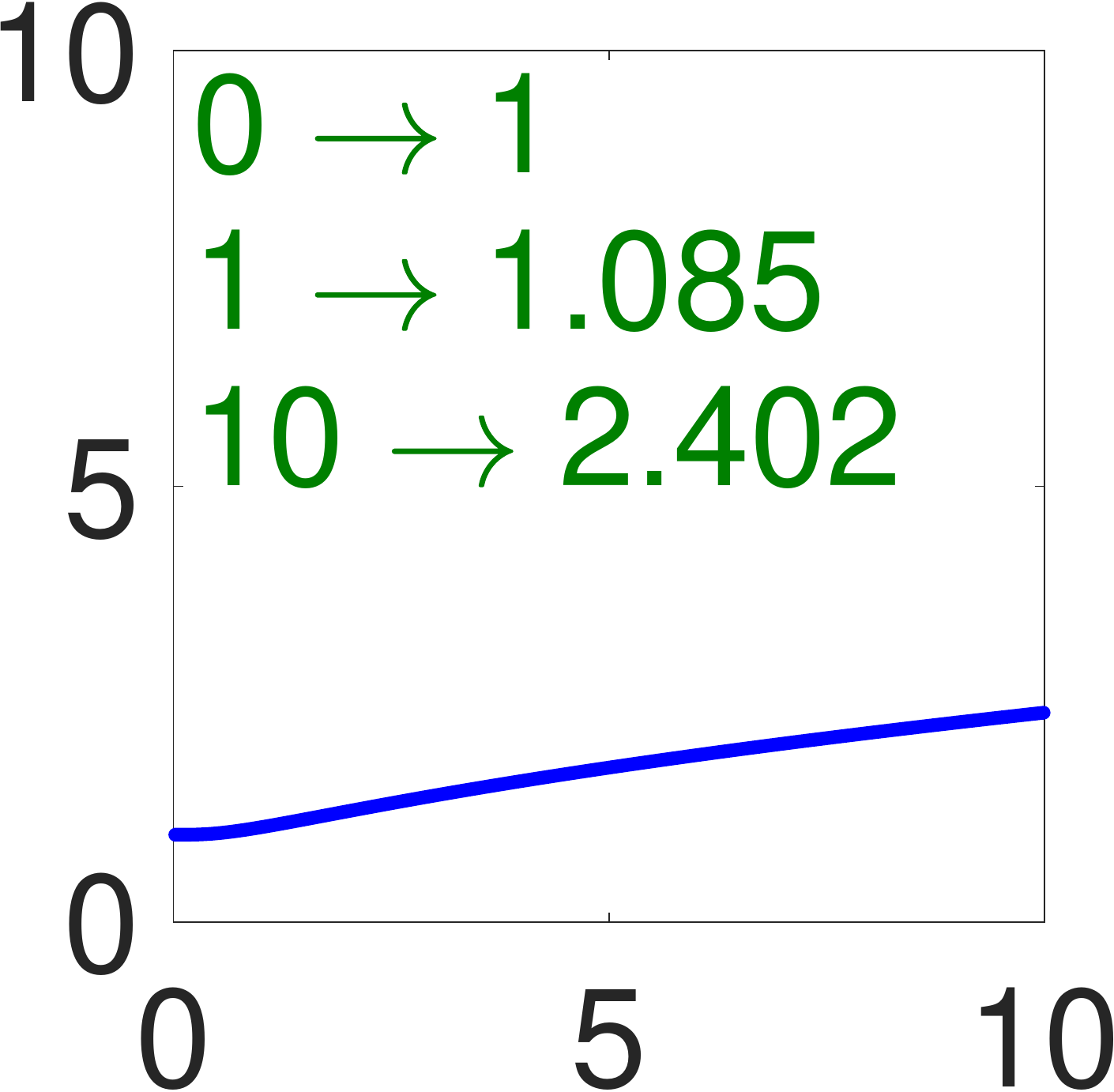}\\
$\lambda \rightarrow \infty$&1.21&1.21&1.21&1.21&1.66&$O(\sqrt{\lambda})$\\
\\
\end{tabular}
}
\caption{(This table is continued on the next three pages.) Activation function metrics for the activation functions used in this work, given in table \ref{actFunIllu}, as well as their debiased versions. Segments of $\mathfrak{L}_\tau$ and $\frac{\mathfrak{L}_\tau(\lambda)}{\mathfrak{L}_\tau(1)}$ depicted in blue correspond to x-coordinate values that converge upon iteration to a non-zero value. Segments depicted in red correspond to x-coordinate values that diverge or converge to zero upon iteration. Segments depicted in magenta correspond to x-coordinate values that are stationary upon iteration. Segments of $\frac{\mathfrak{L}_\tau'(\lambda)\lambda}{\mathfrak{L}_\tau(\lambda)}$ are depicted in red if the y-coordinate is greater than 1, blue in $(-\frac{1}{2},1)$ and magenta at 1. Non-zero iteration limits are depicted as bold blue dots. Limits and convergence rates are given below the respective graph where applicable. Values of $\mathfrak{n}_\tau(\lambda^2,0)$ at $\lambda=0$, $\lambda=1$ and $\lambda=10$ are given in green. Diagonals are given as thin black lines for visual orientation. Each convergence rate is given as precisely as we were able to determine it with basic techniques. The values are discussed in sections \ref{actFunLengthKernelSection}, \ref{actFunCovKernelSection} and \ref{actFunNLCsection}. There is no single, overarching conclusion.}
\label{covCurveillu1}
\end{table}

\begin{table}[H]
{
\centering \small
\begin{tabular}{lcccccc}
Act. fun. &sigmoid&even tanh&Gaussian&odd square&square&sawtooth\\ \hline\hline
\\
$\tau(s)$&\includegraphics[scale=0.13,valign=c]{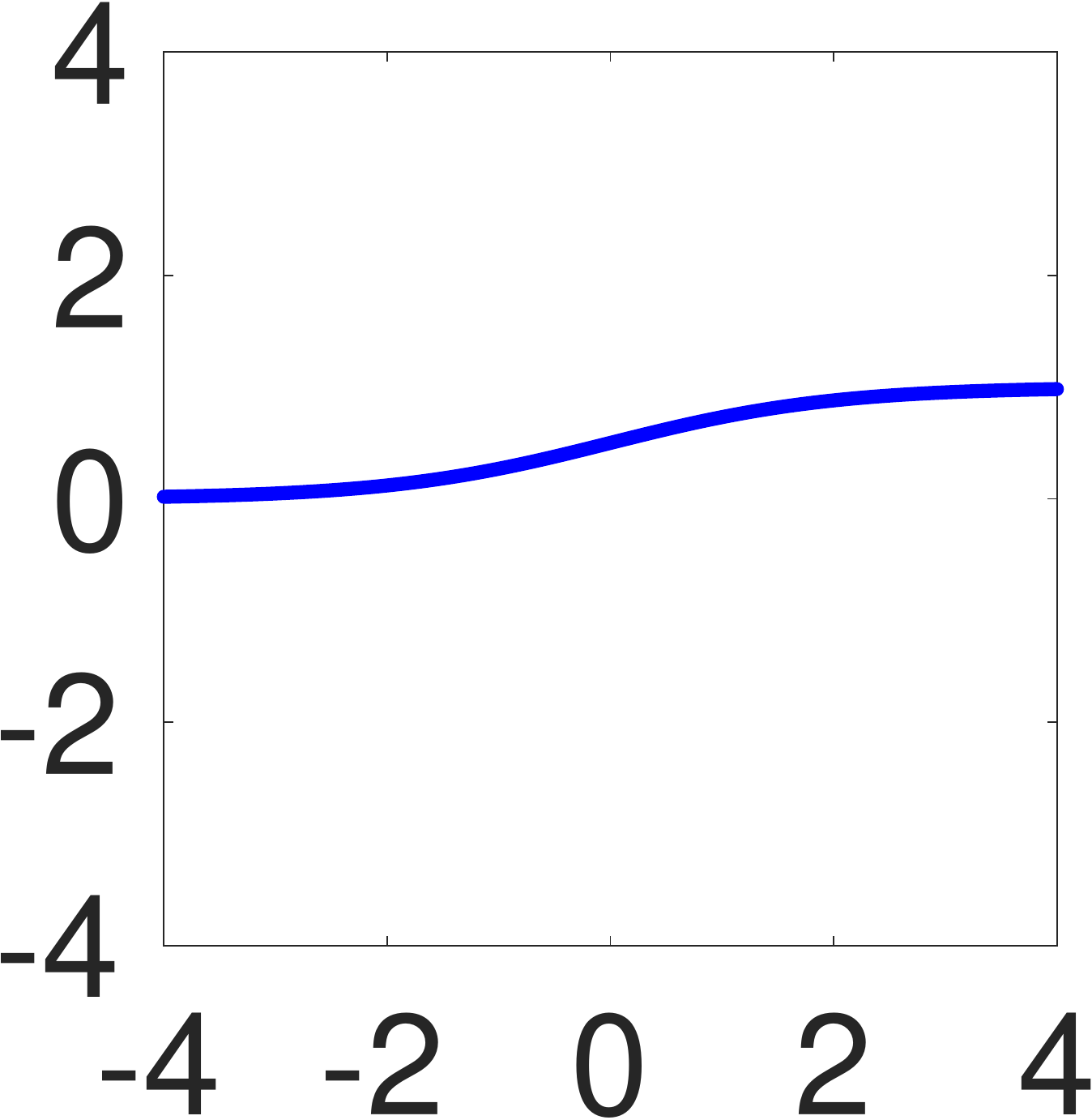}&\includegraphics[scale=0.13,valign=c]{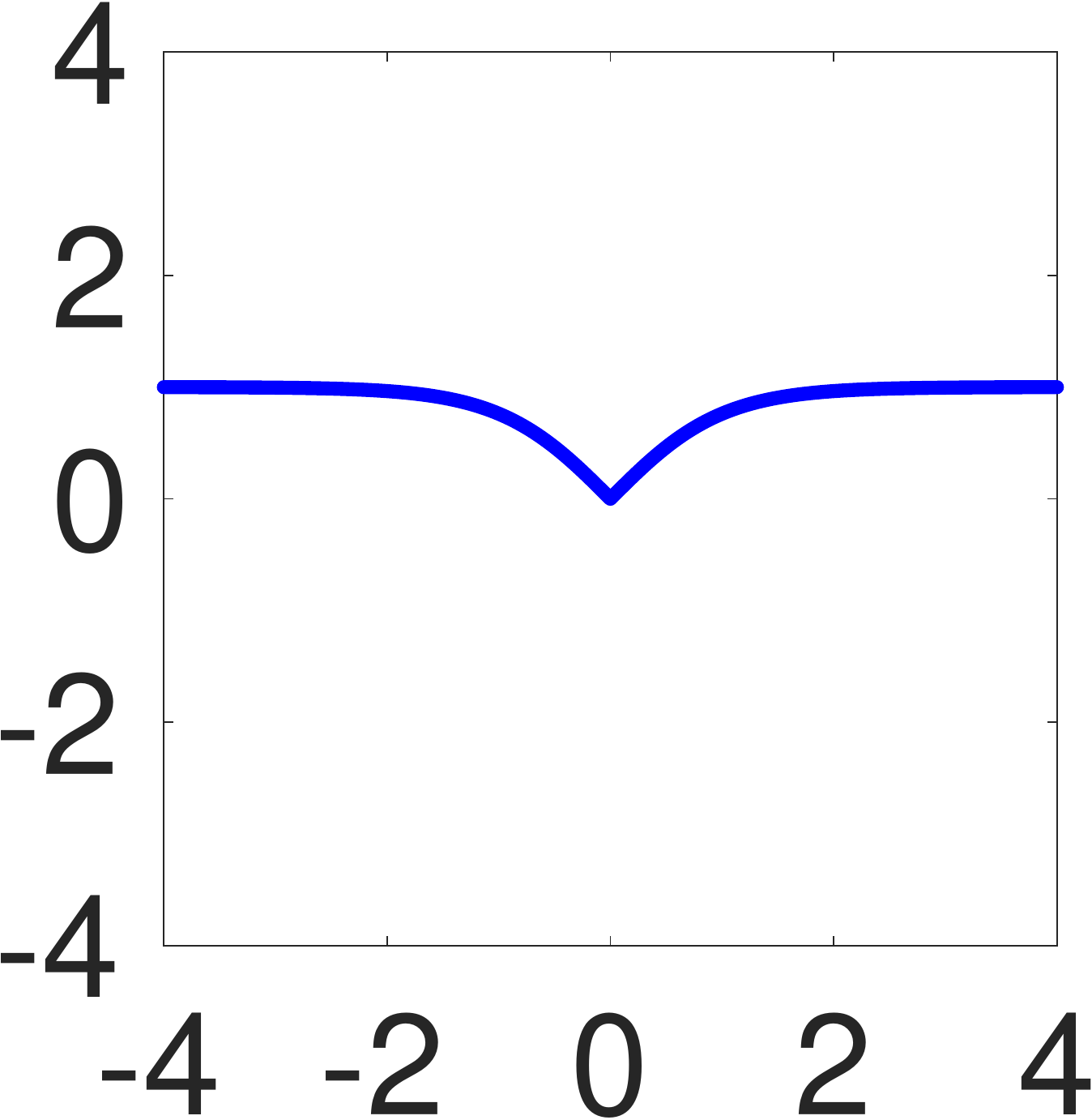}&\includegraphics[scale=0.13,valign=c]{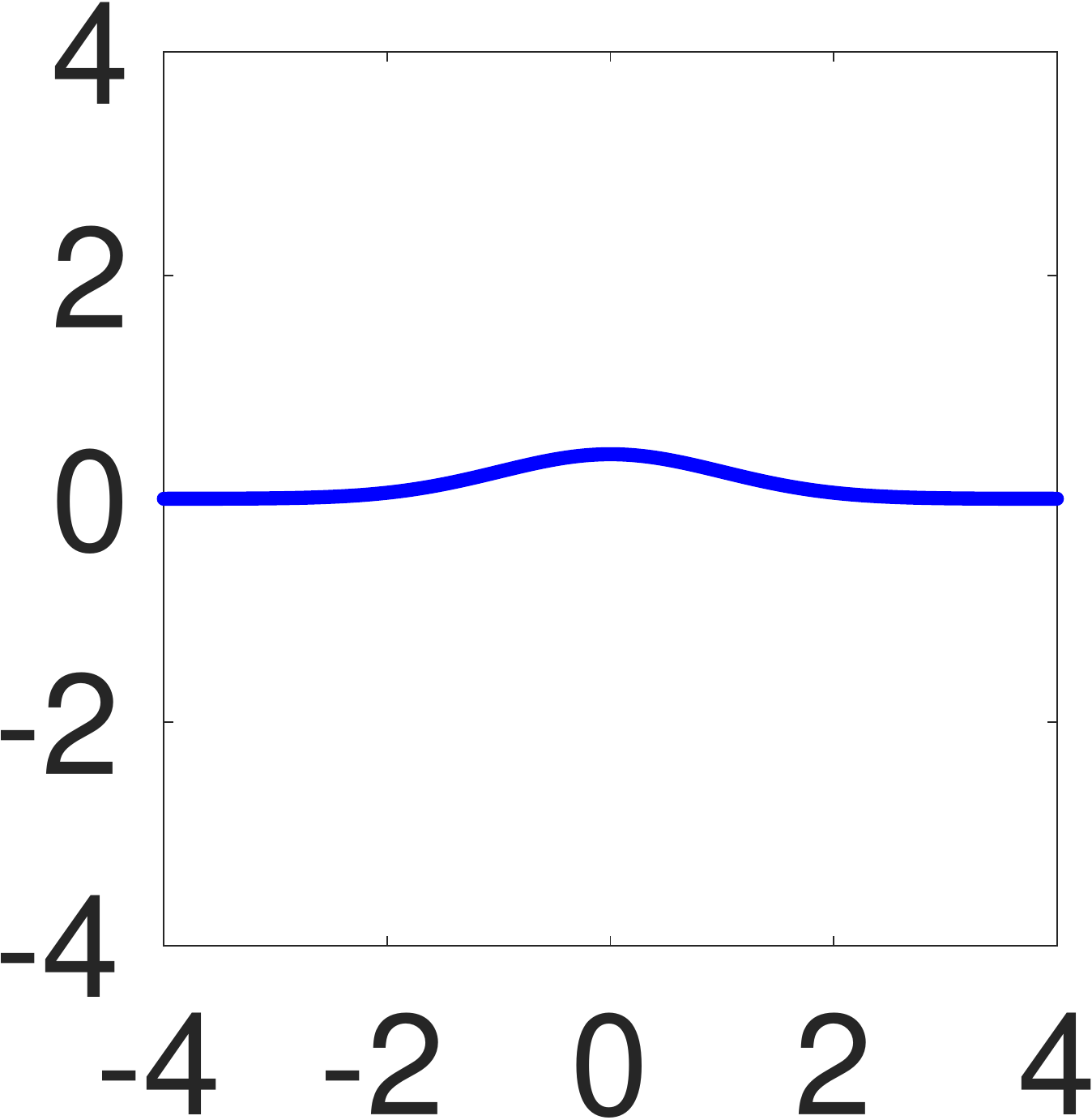}&\includegraphics[scale=0.13,valign=c]{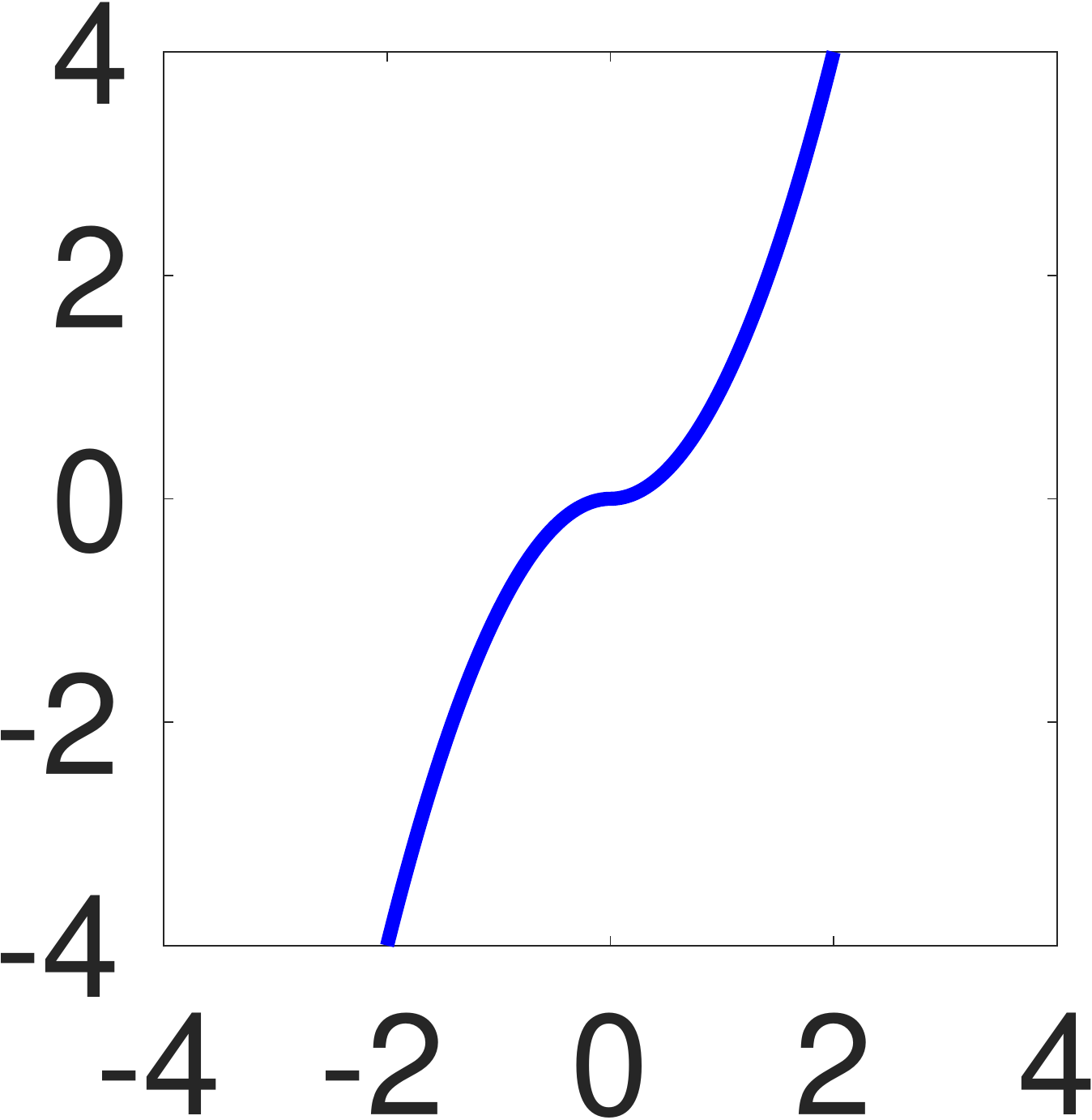}&\includegraphics[scale=0.13,valign=c]{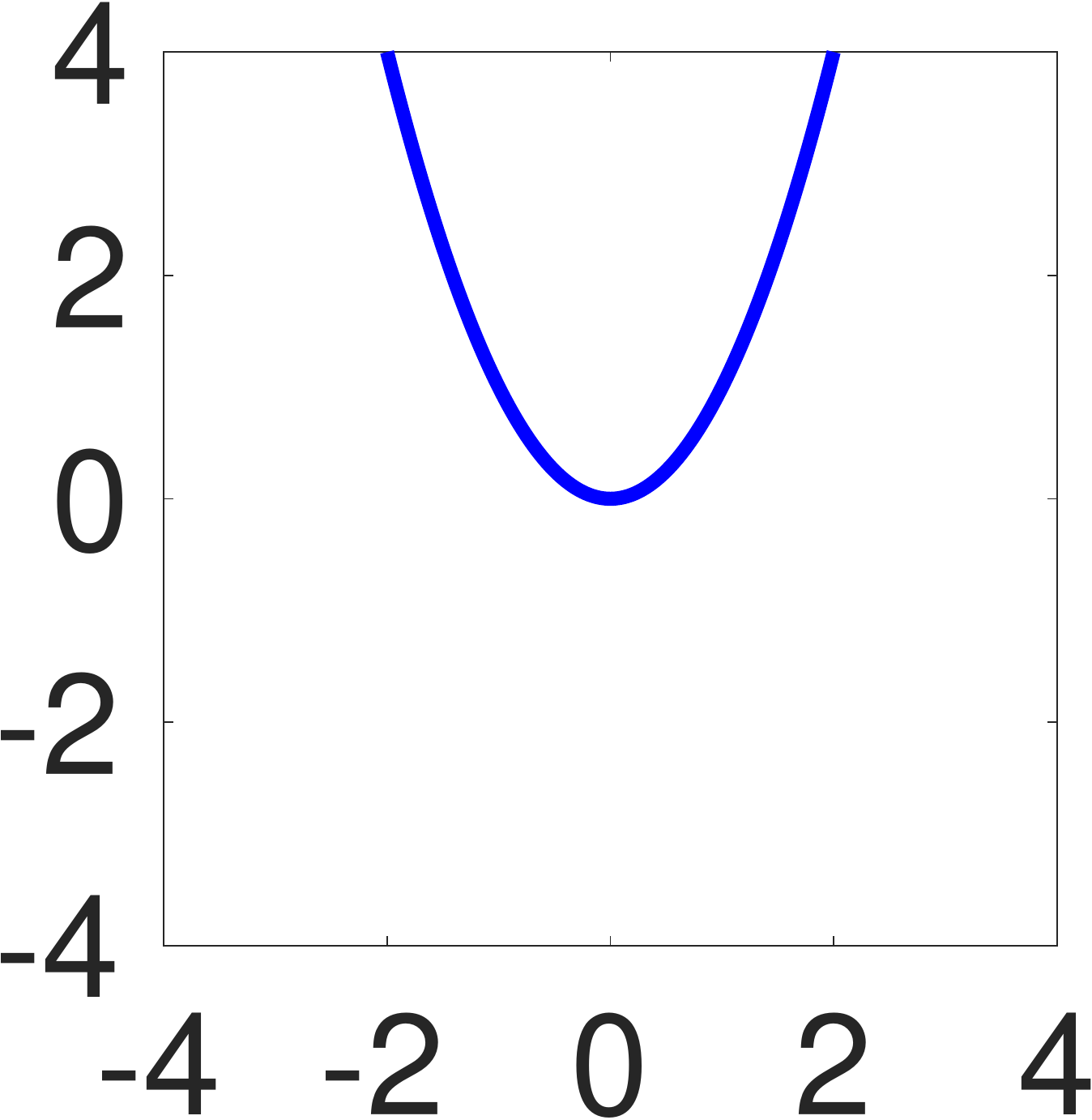}&\includegraphics[scale=0.13,valign=c]{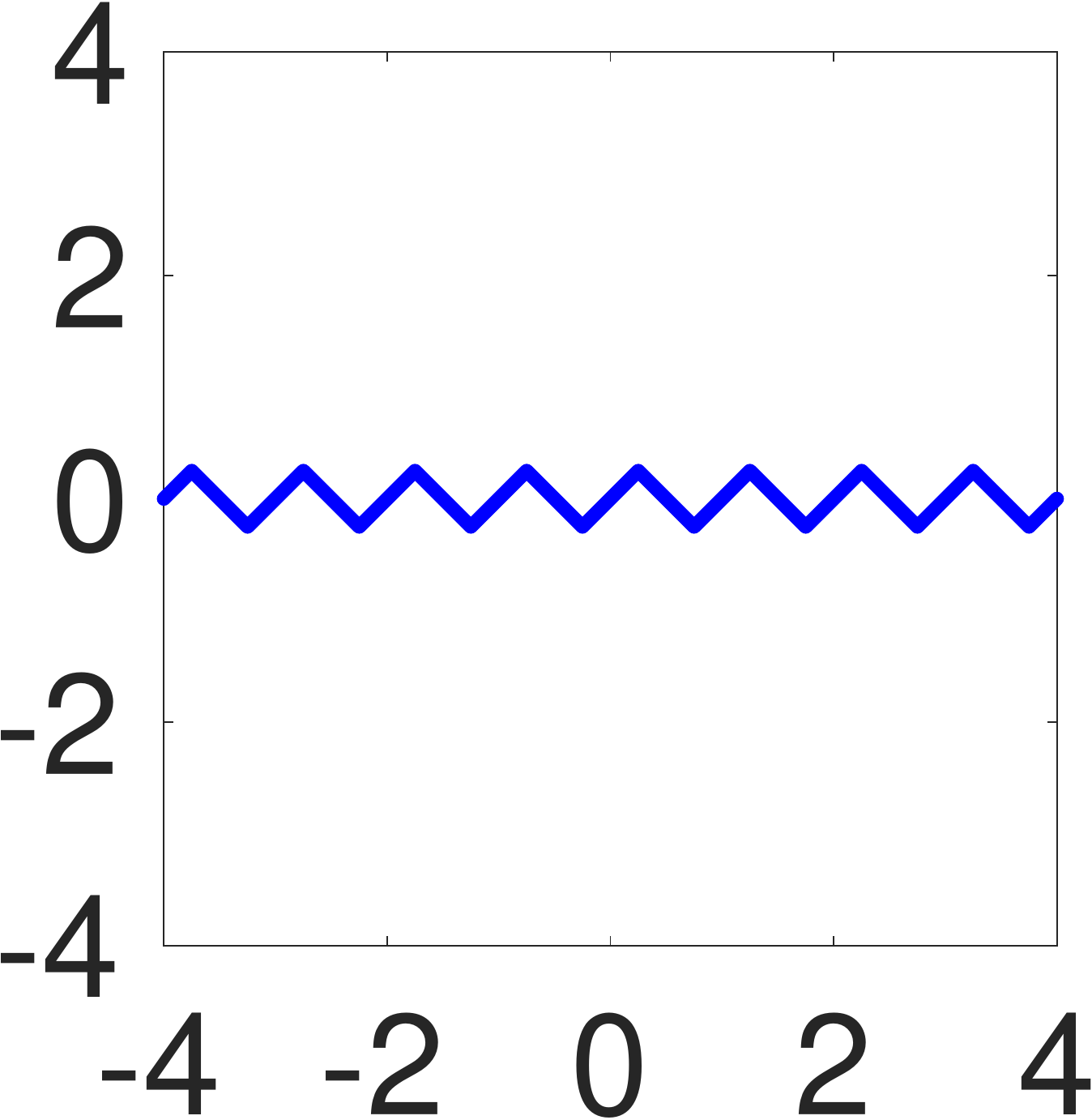}\\
$\mathfrak{L}_\tau(\lambda)$&\includegraphics[scale=0.13,valign=c]{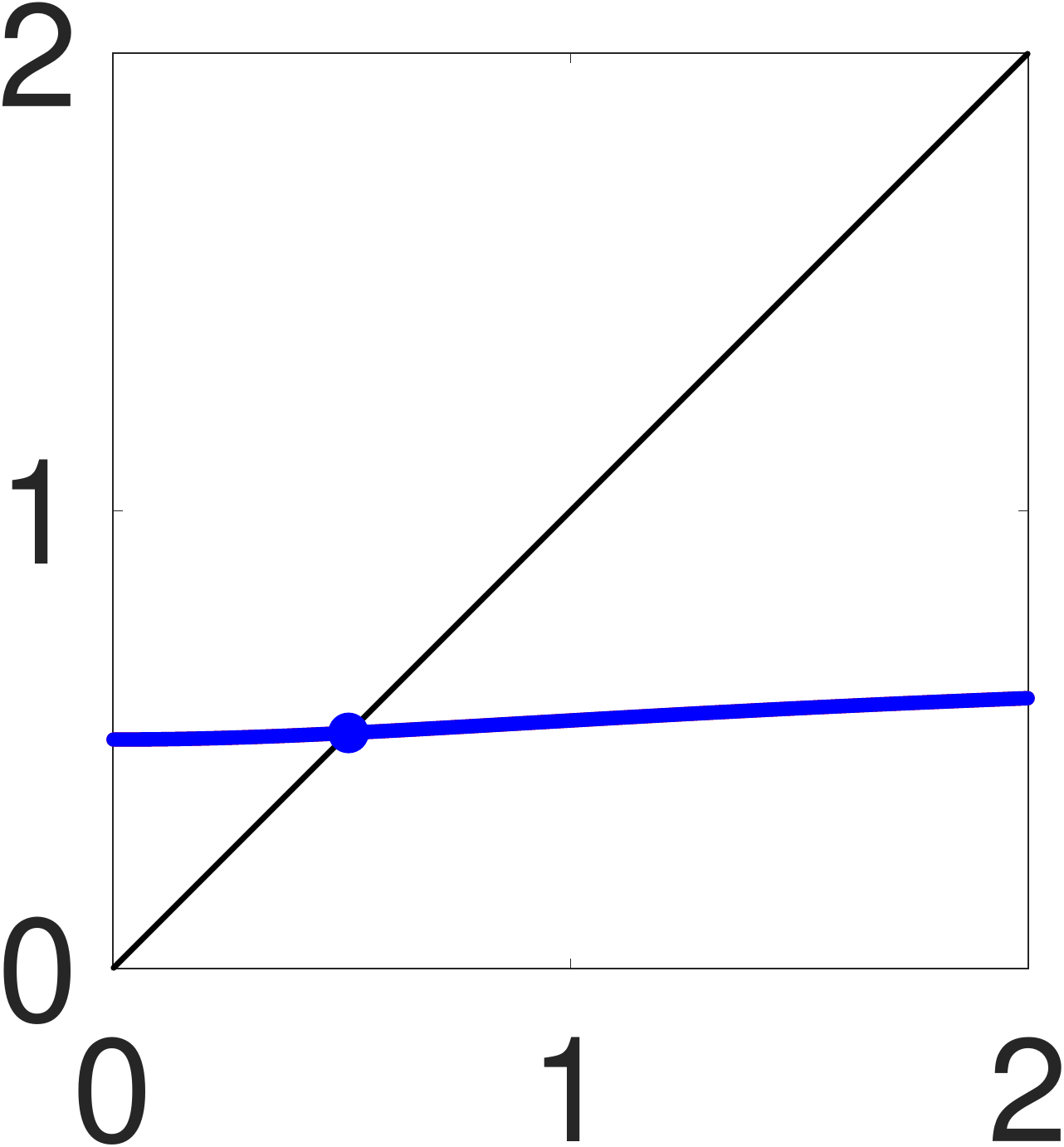}&\includegraphics[scale=0.13,valign=c]{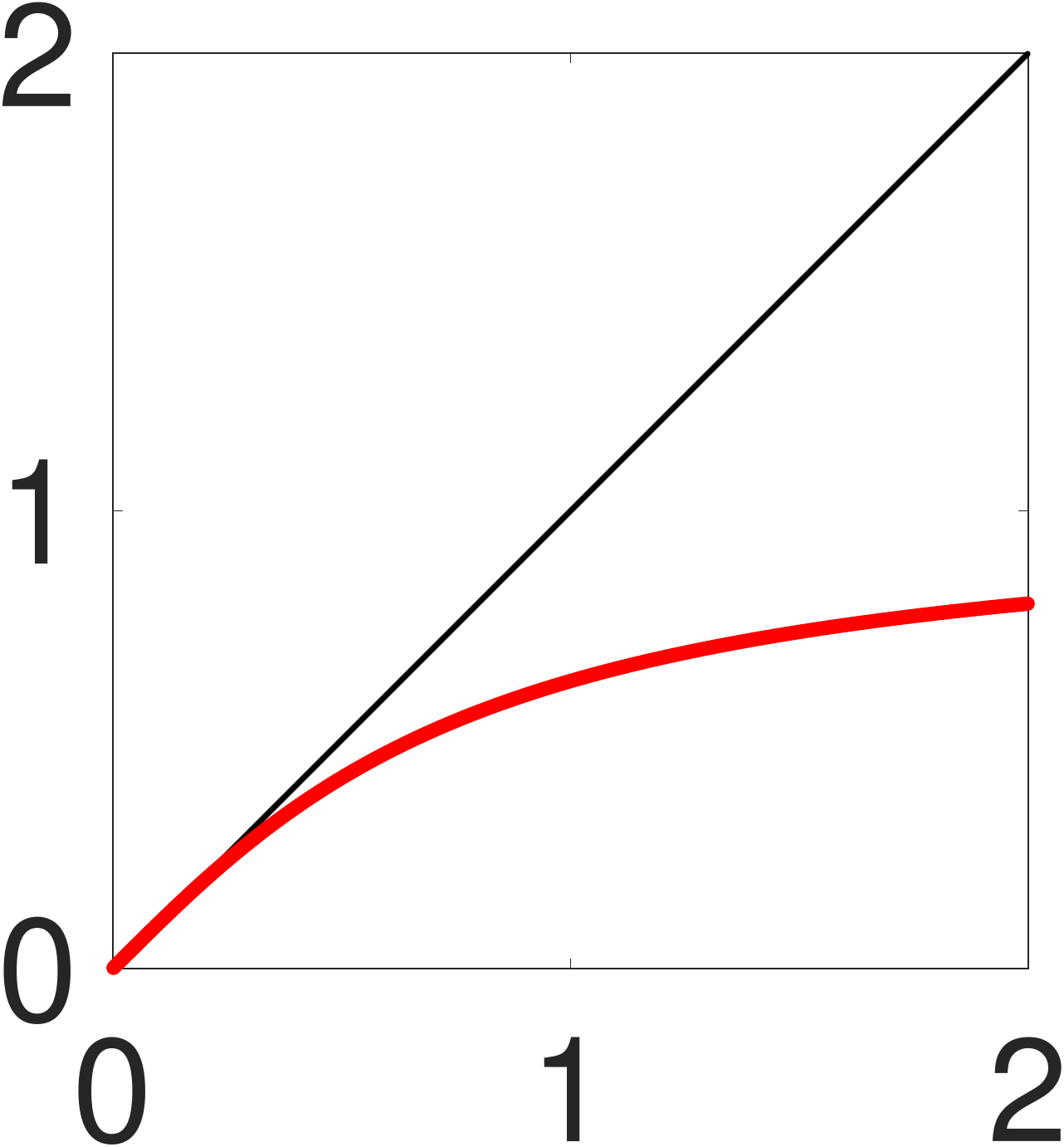}&\includegraphics[scale=0.13,valign=c]{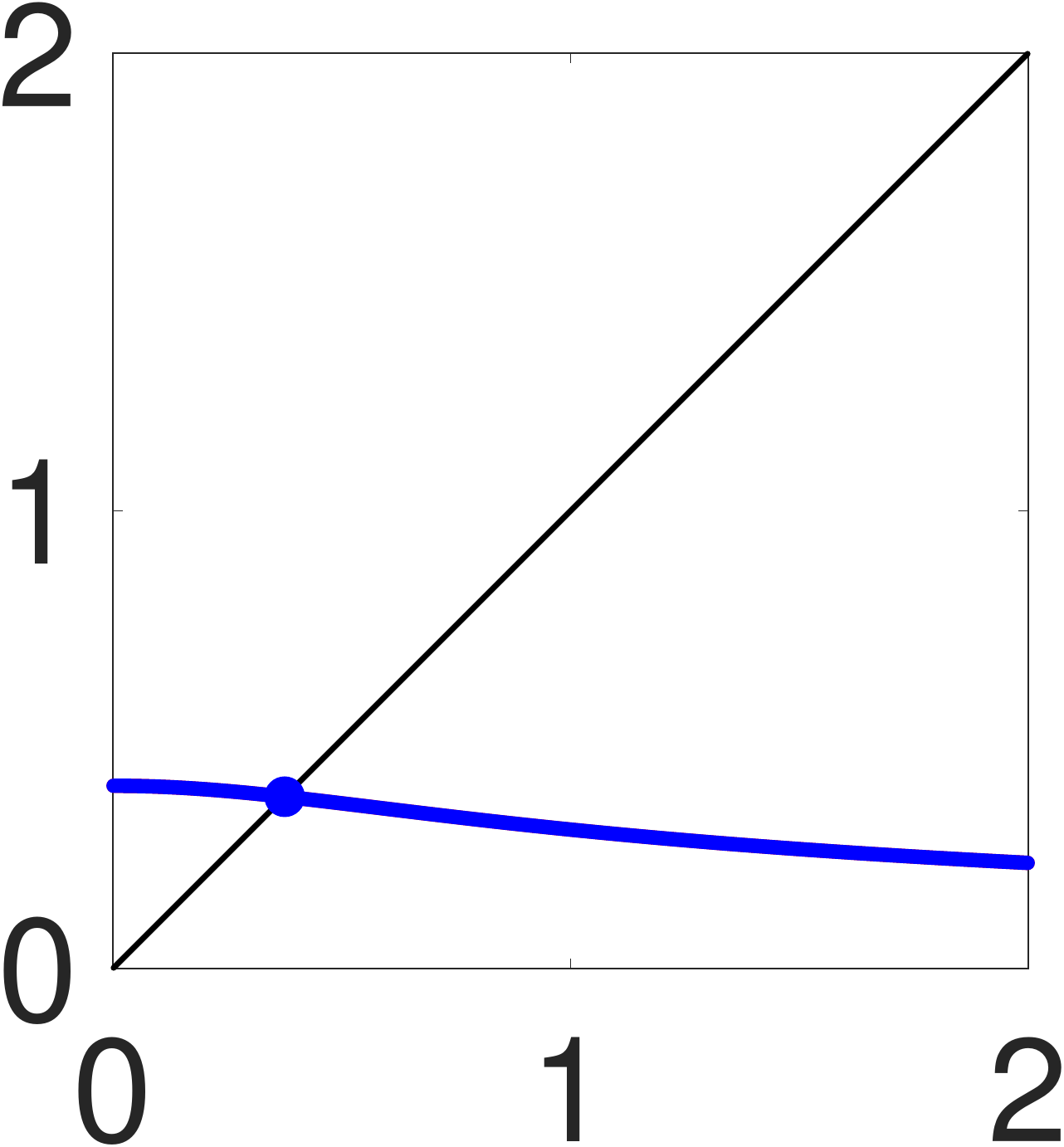}&\includegraphics[scale=0.13,valign=c]{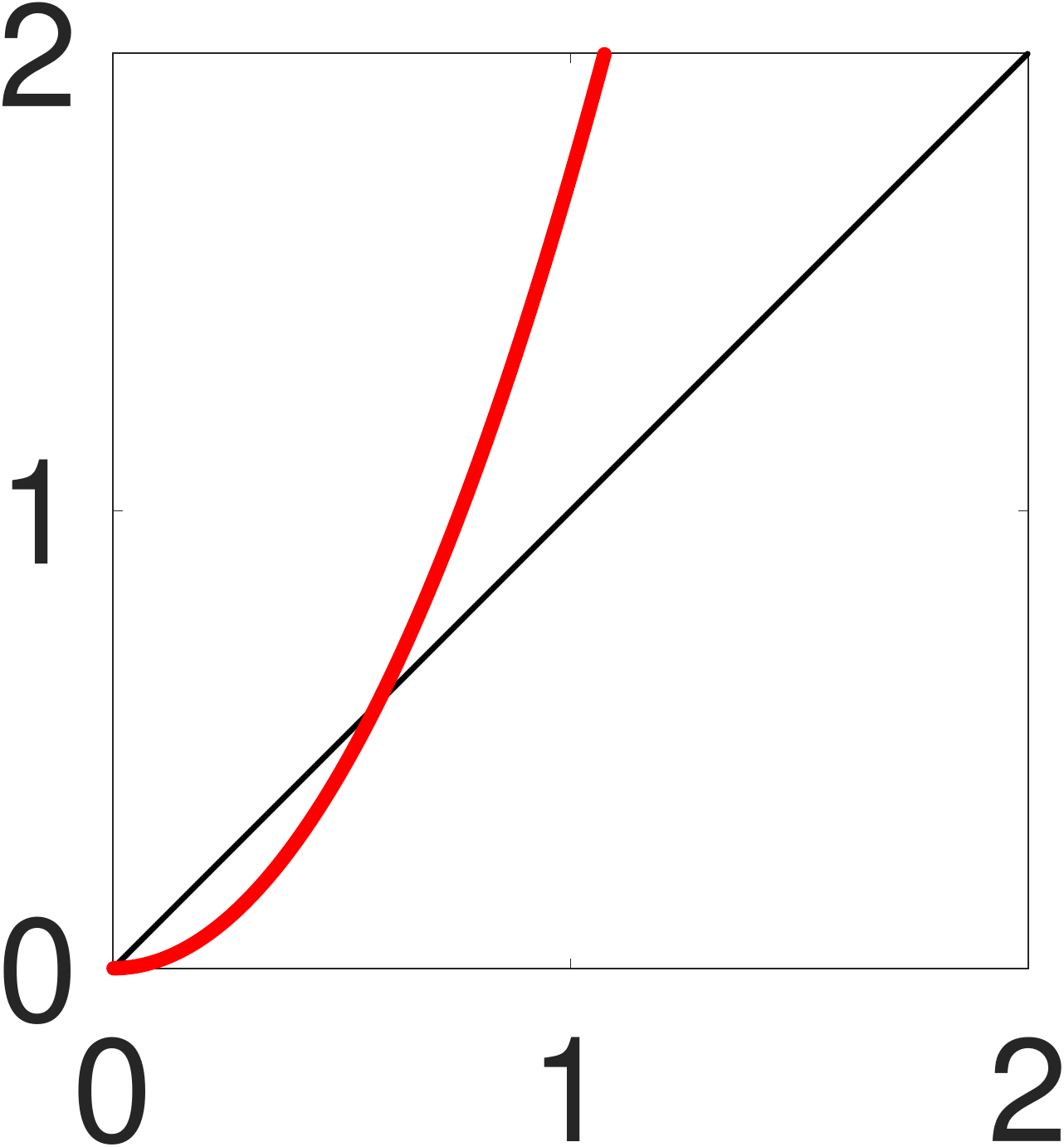}&\includegraphics[scale=0.13,valign=c]{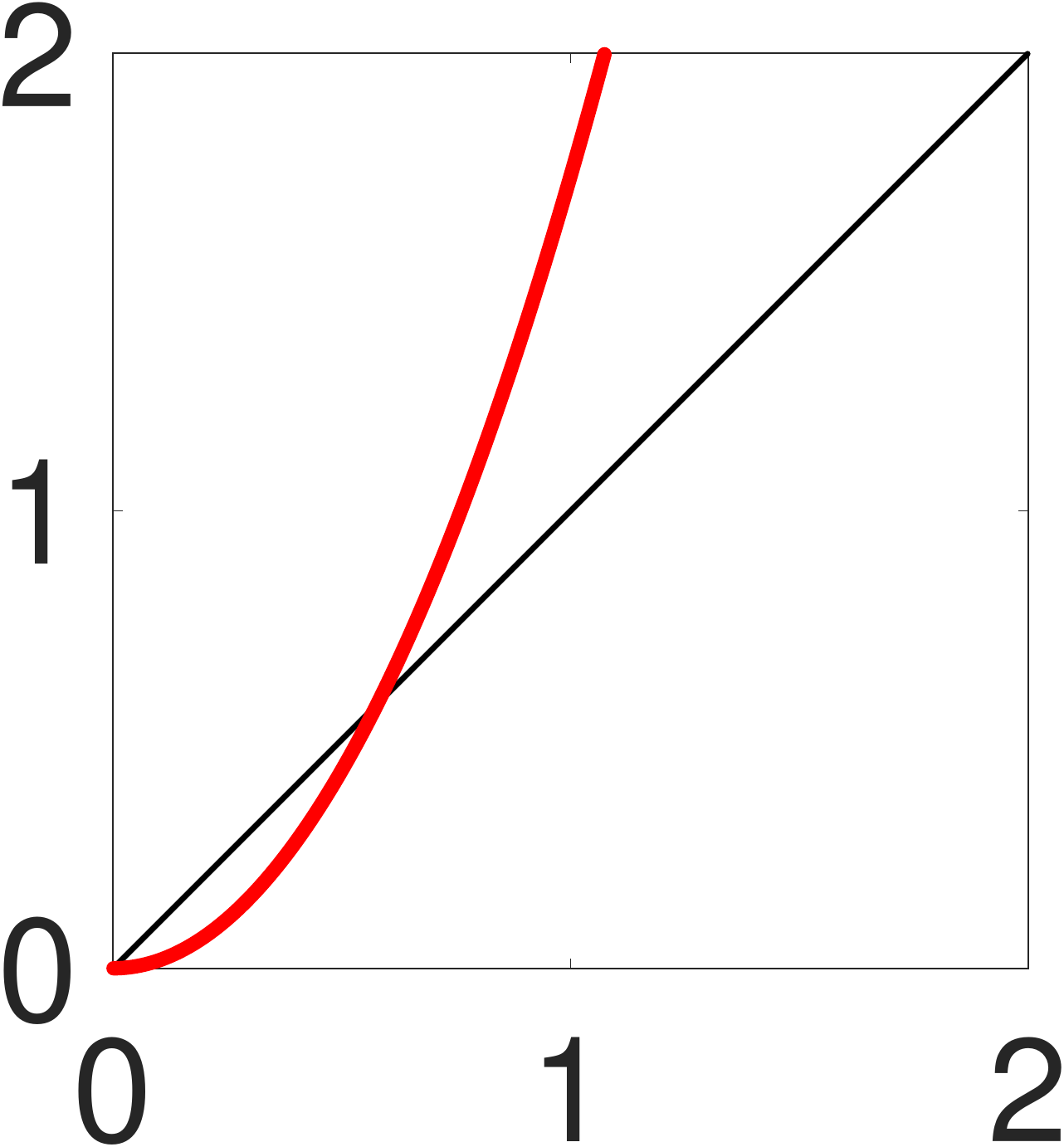}&\includegraphics[scale=0.13,valign=c]{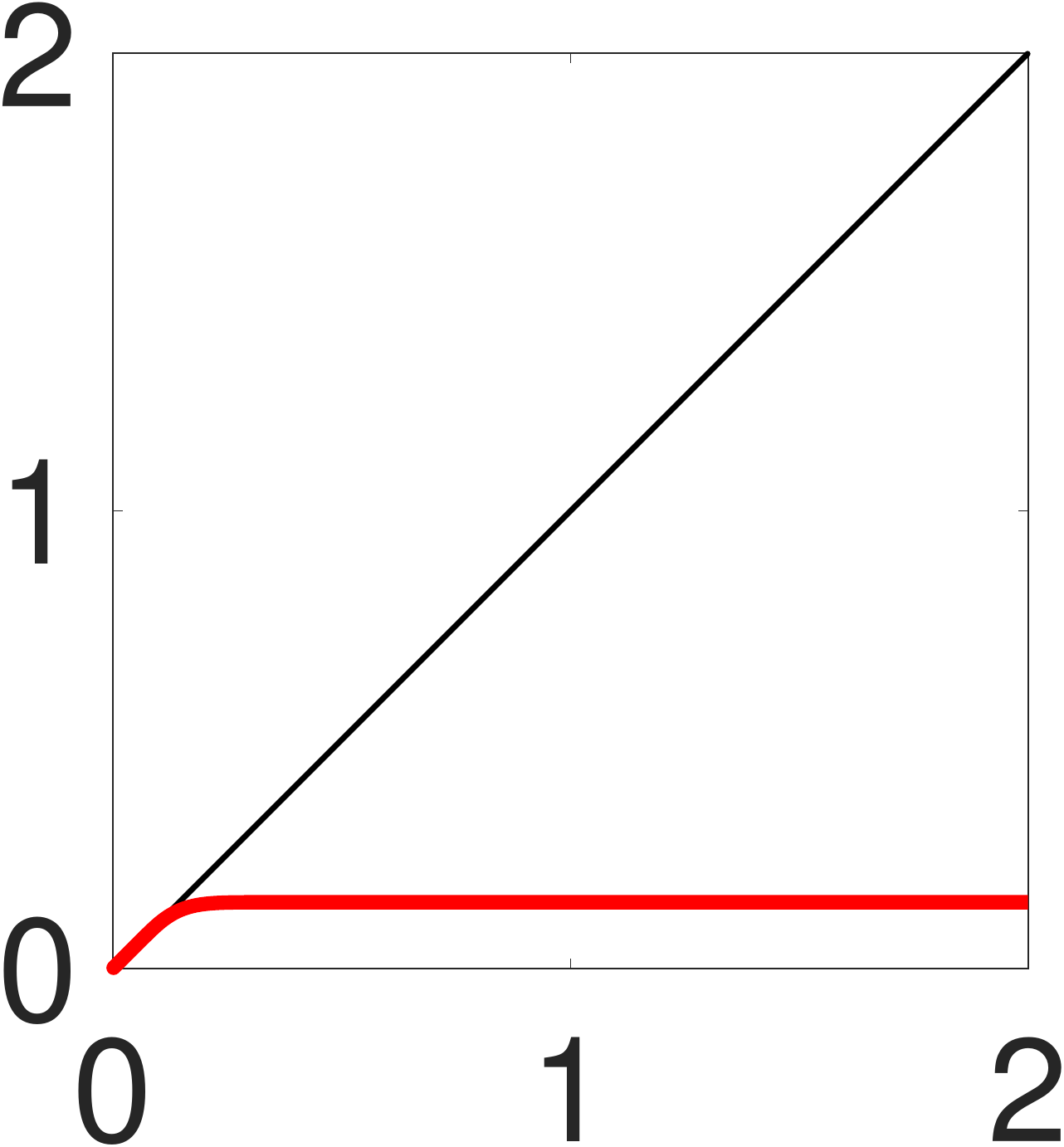}\\
Iter. limit &0.51&0&0.37&0&0&0\\
Conv. rate&$O(0.05^M)$&$O(\frac{1}{M^{0.5}})$&$O(0.11^M)$&$\frac{1}{\sqrt{3}}(\sqrt{3}\lambda)^{2^M}$&$\frac{1}{\sqrt{3}}(\sqrt{3}\lambda)^{2^M}$&$\Omega(\frac{1}{M^\epsilon}) \forall \epsilon > 0$\\
$\frac{\mathfrak{L}_\tau(\lambda)}{\mathfrak{L}_\tau(1)}$&\includegraphics[scale=0.13,valign=c]{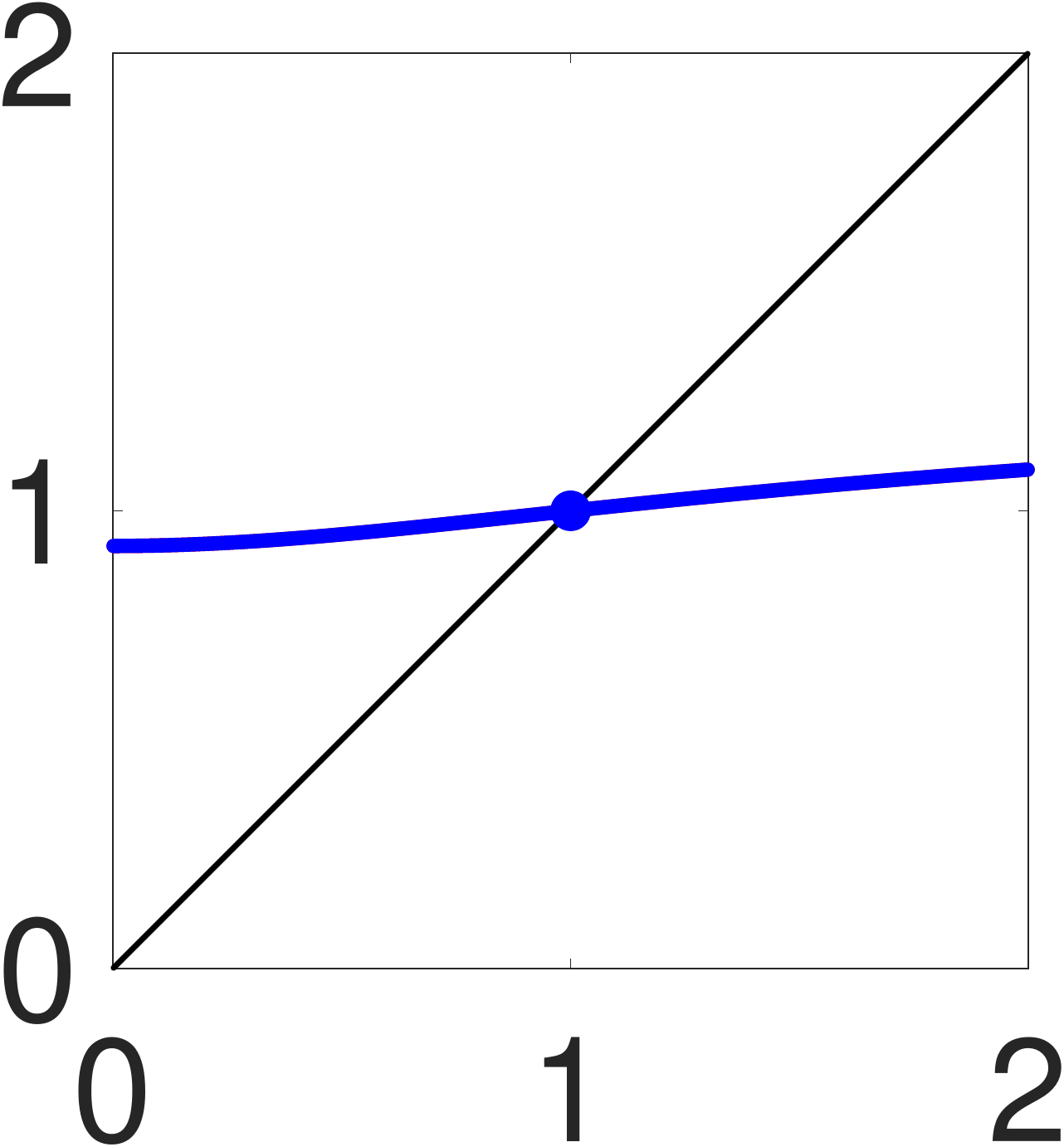}&\includegraphics[scale=0.13,valign=c]{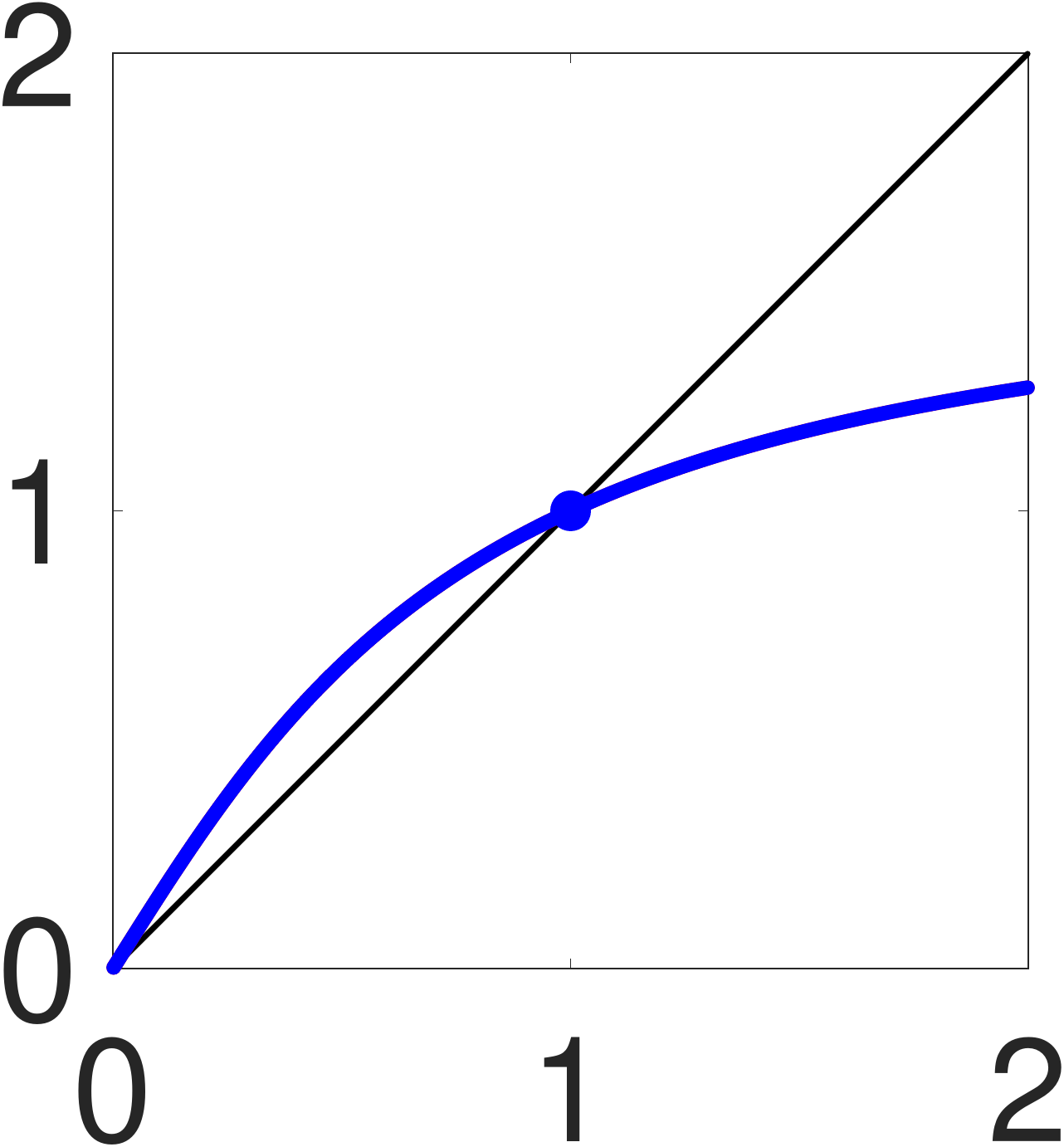}&\includegraphics[scale=0.13,valign=c]{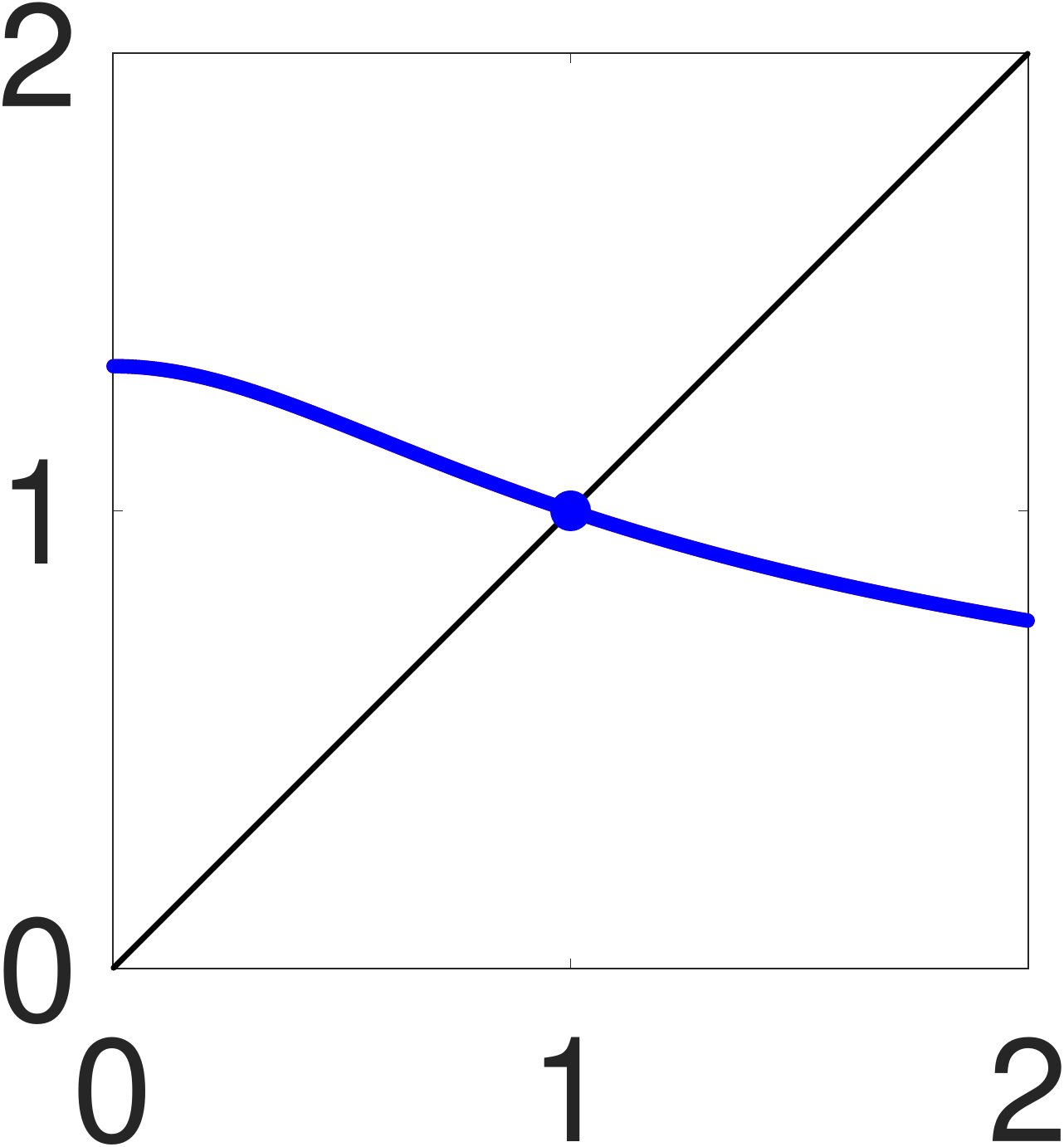}&\includegraphics[scale=0.13,valign=c]{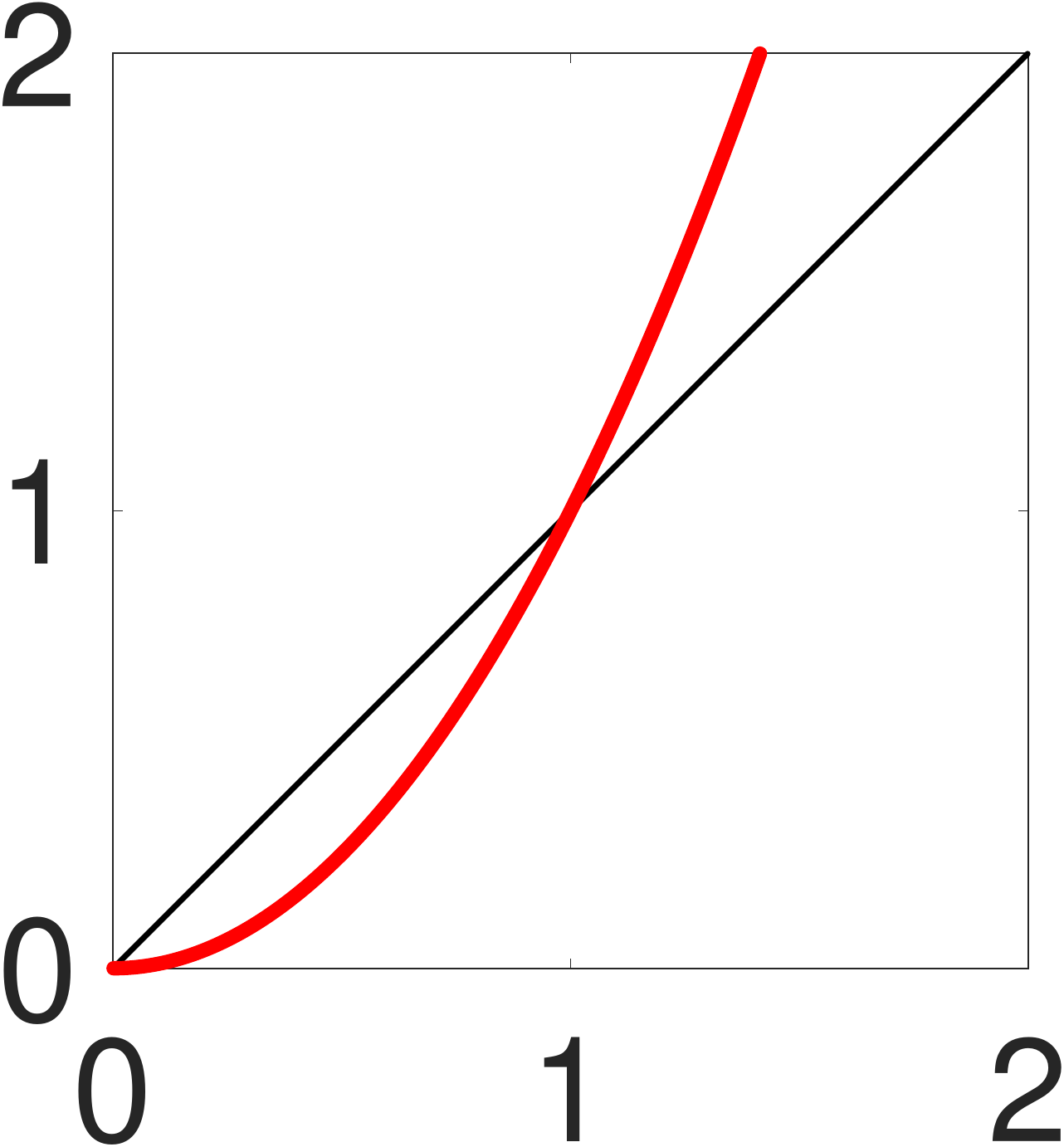}&\includegraphics[scale=0.13,valign=c]{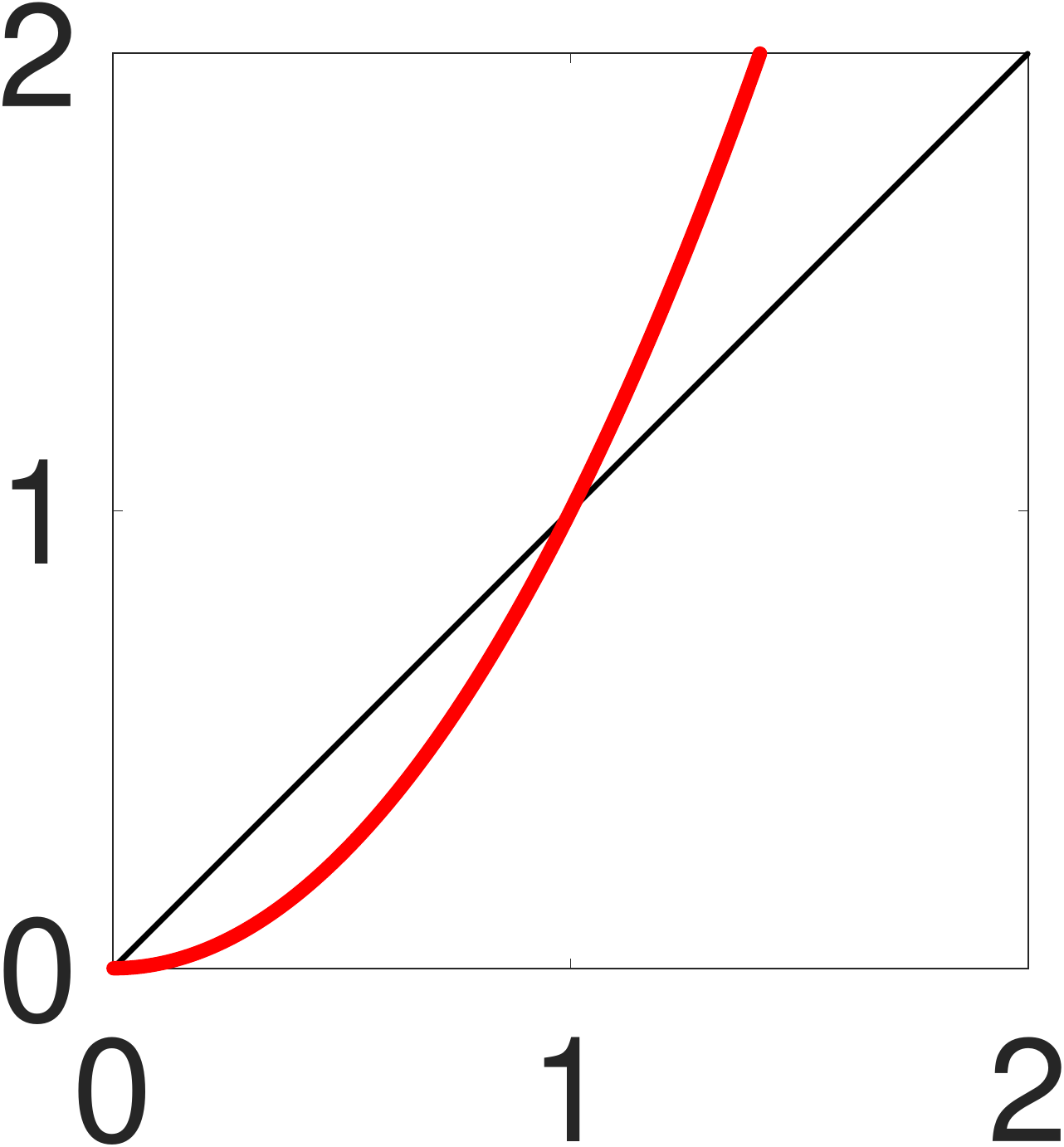}&\includegraphics[scale=0.13,valign=c]{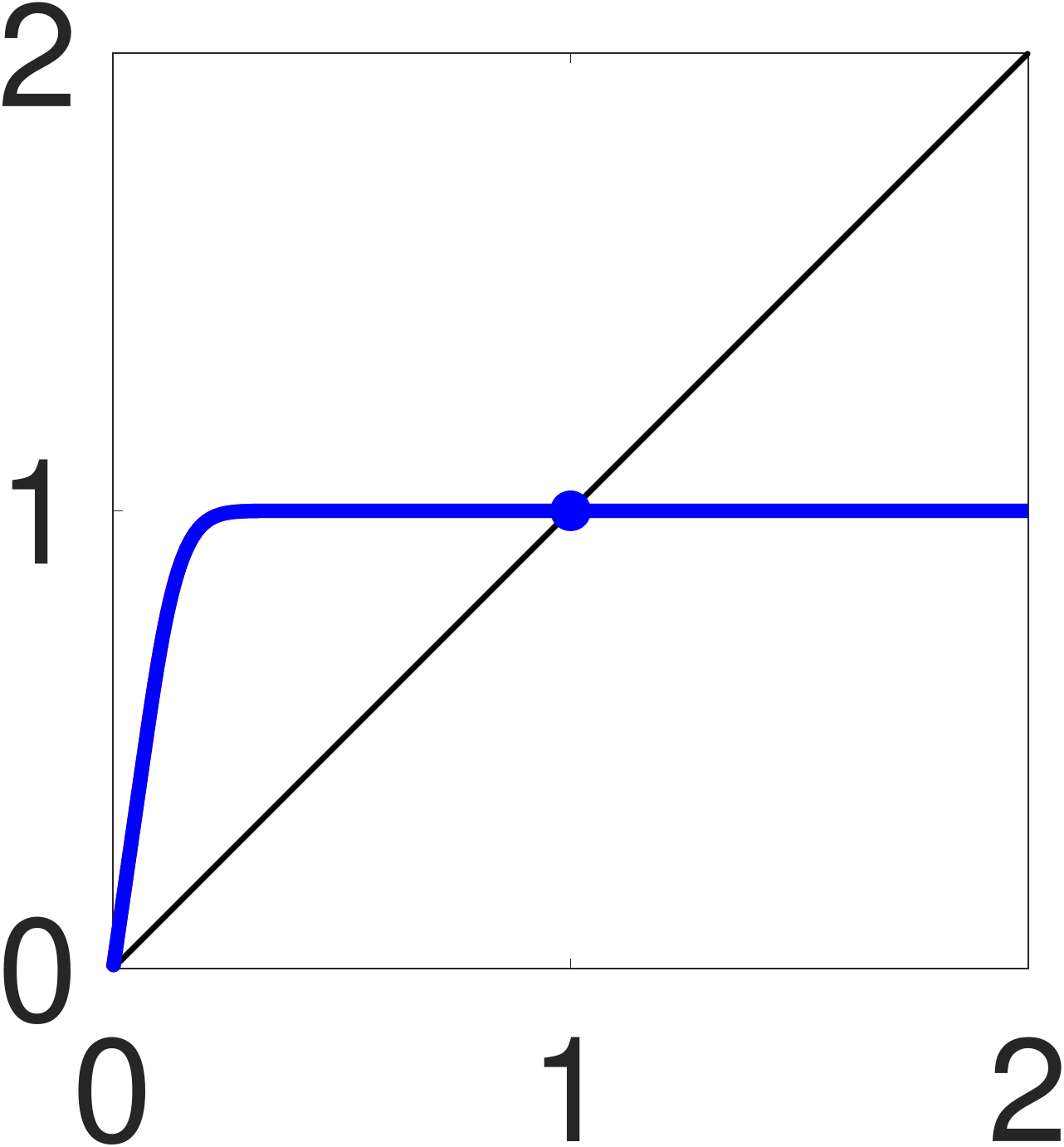}\\
Iter. limit &1&1&1&0&0&1\\
Conv. rate&$O(0.11^M)$&$O(0.46^M)$&$O(0.33^M)$&$\lambda^{2^M}$&$\lambda^{2^M}$&$\approx O(10^{-10M})$\\
$\frac{\mathfrak{L}_\tau'(\lambda)\lambda}{\mathfrak{L}_\tau(\lambda)}$ &\includegraphics[scale=0.13,valign=c]{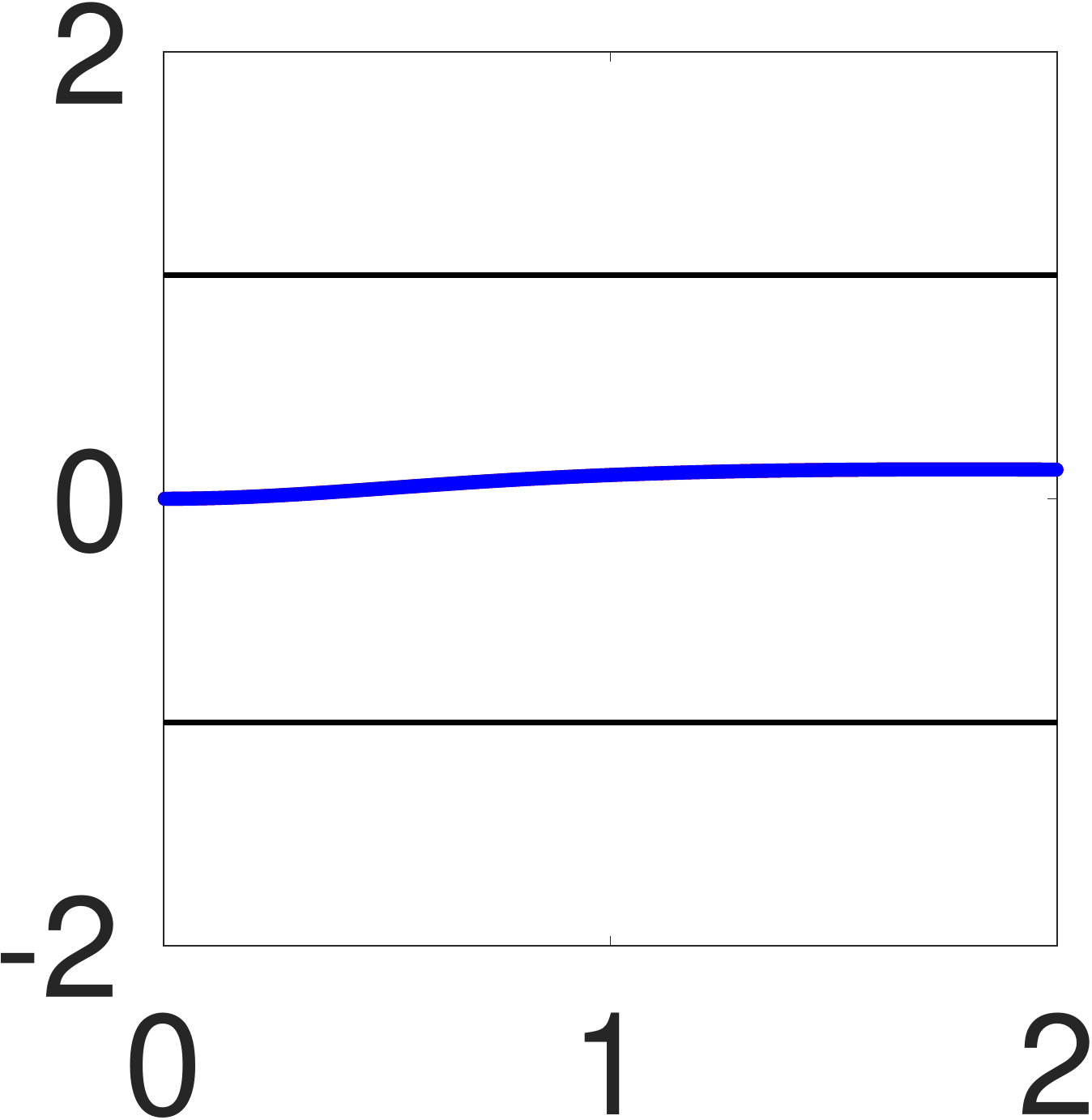}&\includegraphics[scale=0.13,valign=c]{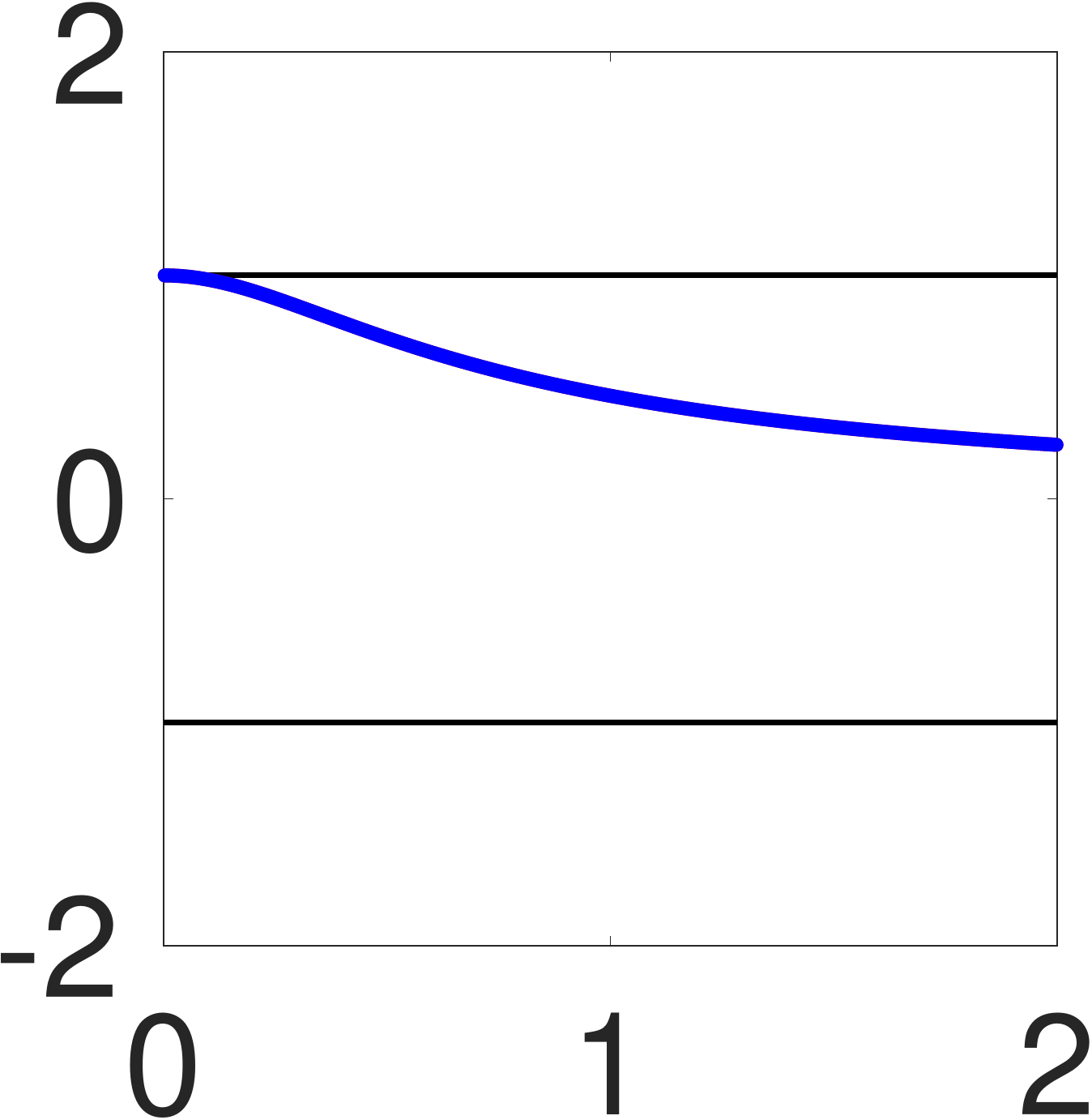}&\includegraphics[scale=0.13,valign=c]{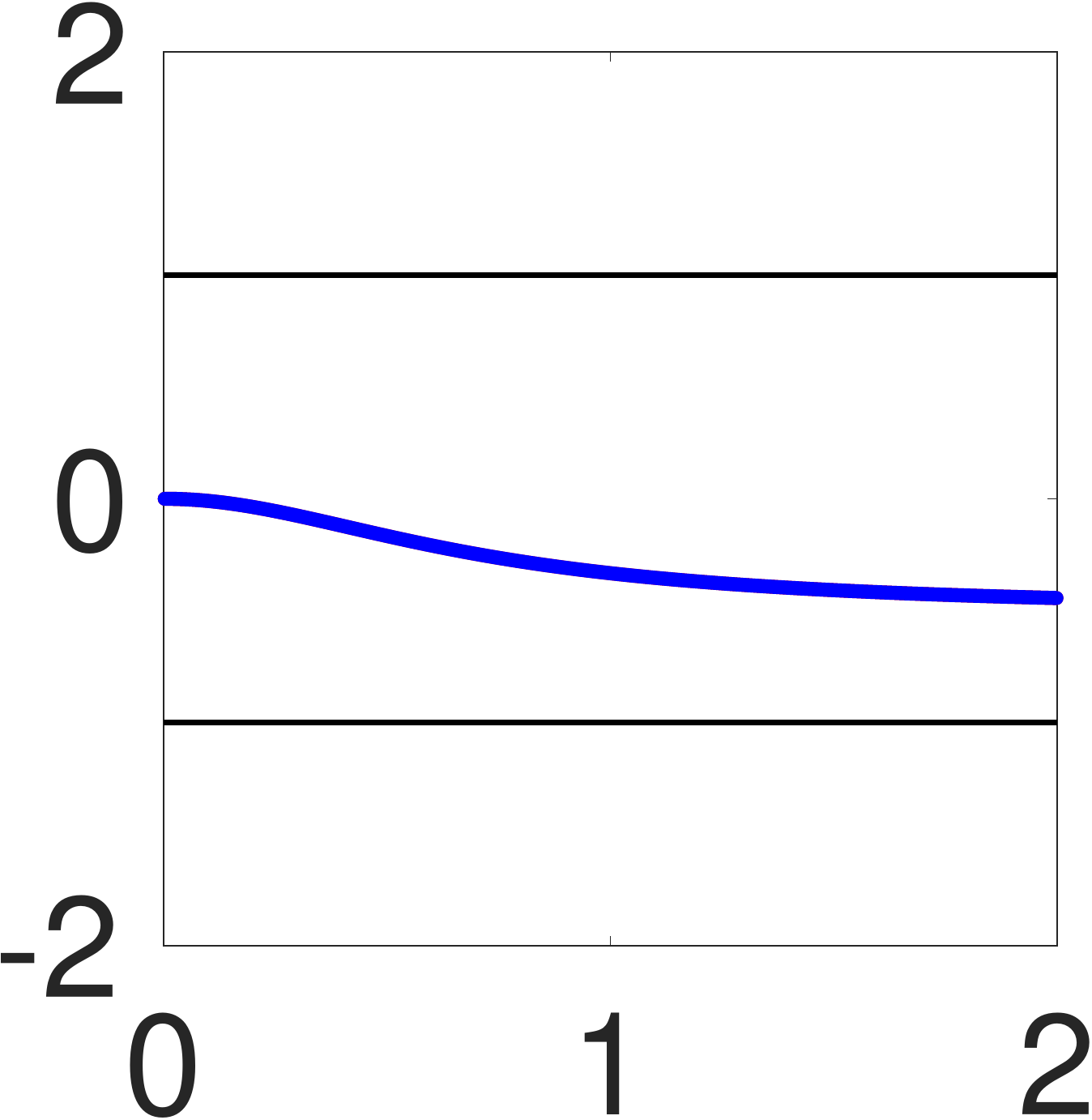}&\includegraphics[scale=0.13,valign=c]{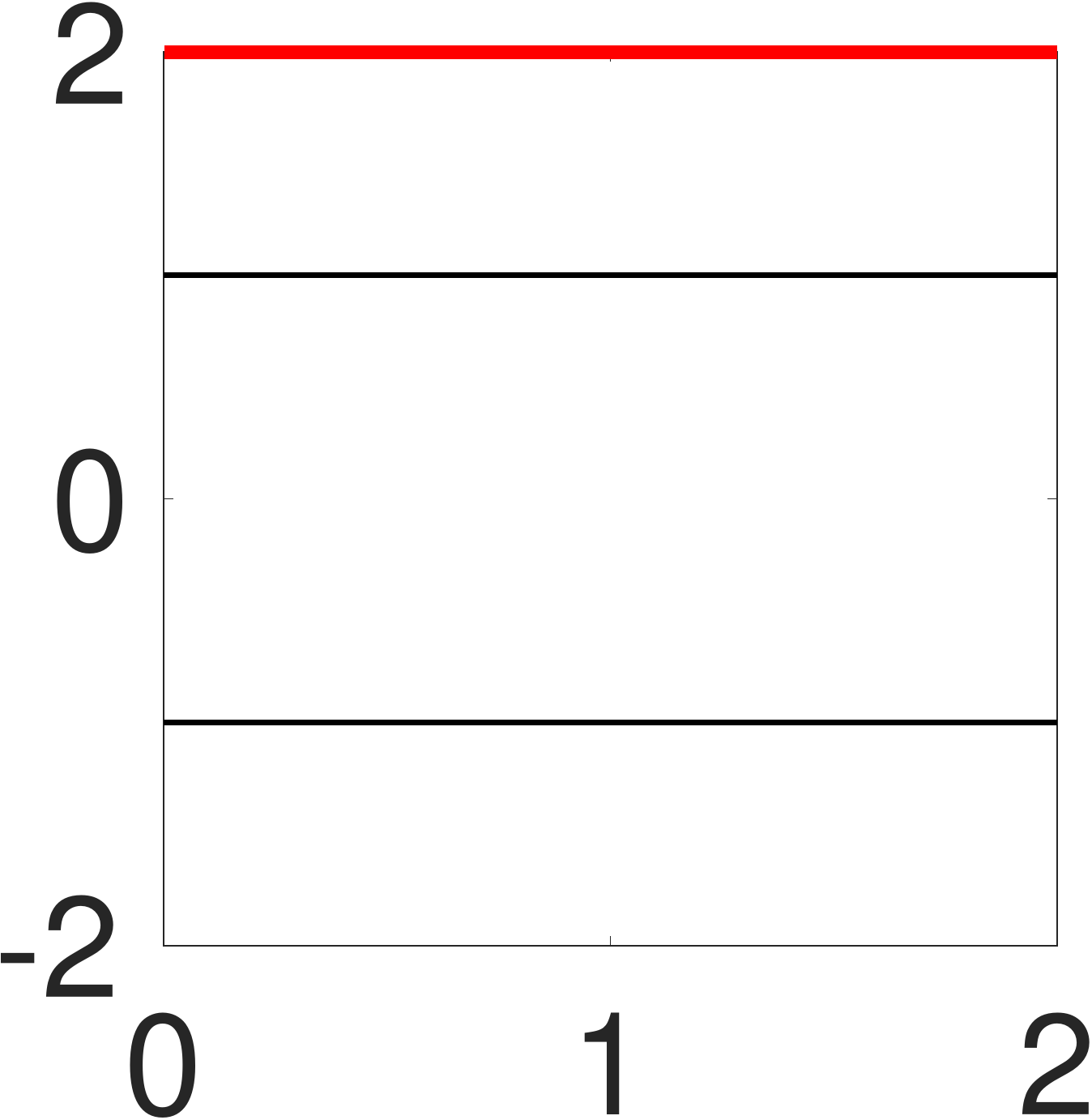}&\includegraphics[scale=0.13,valign=c]{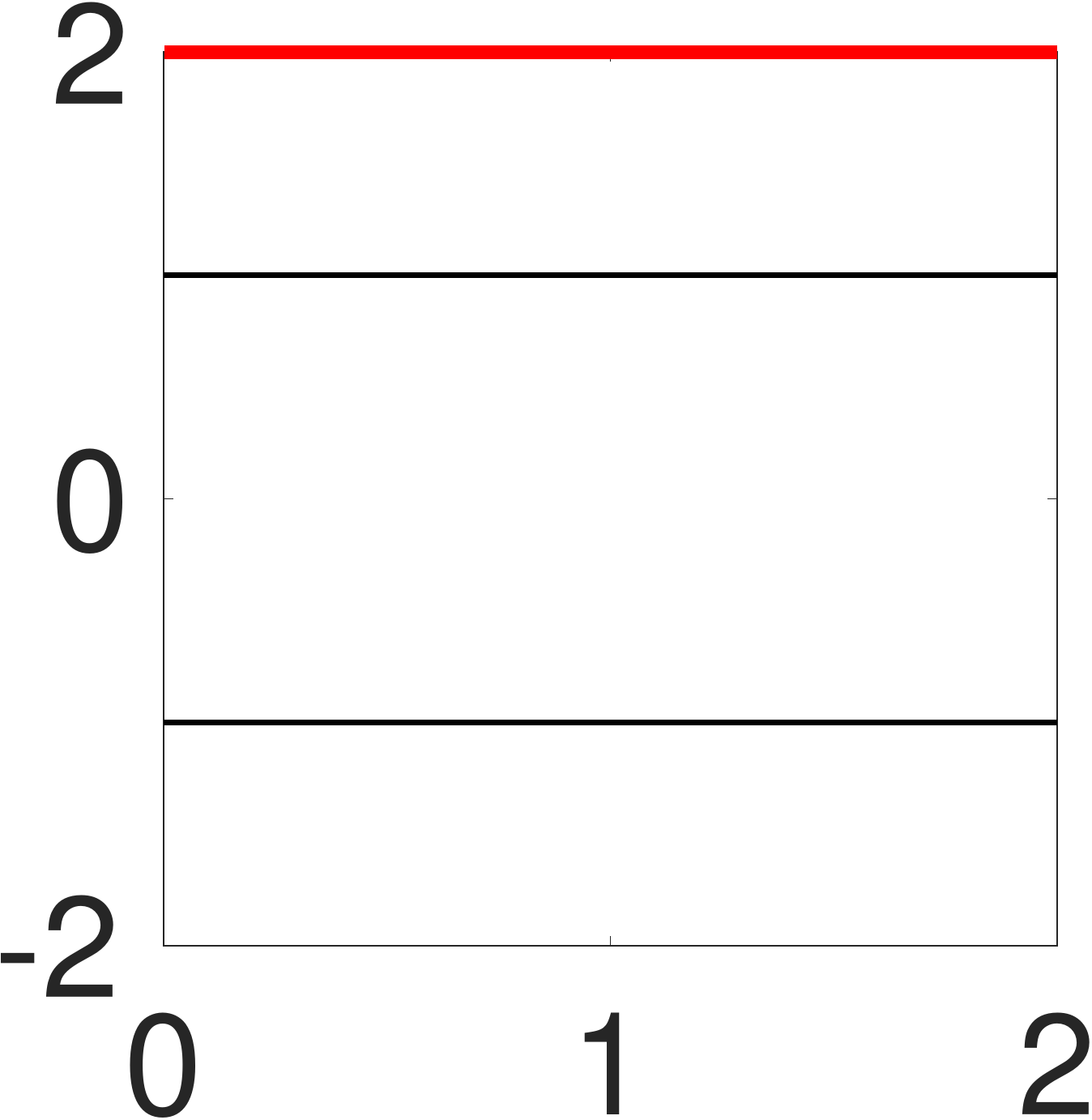}&\includegraphics[scale=0.13,valign=c]{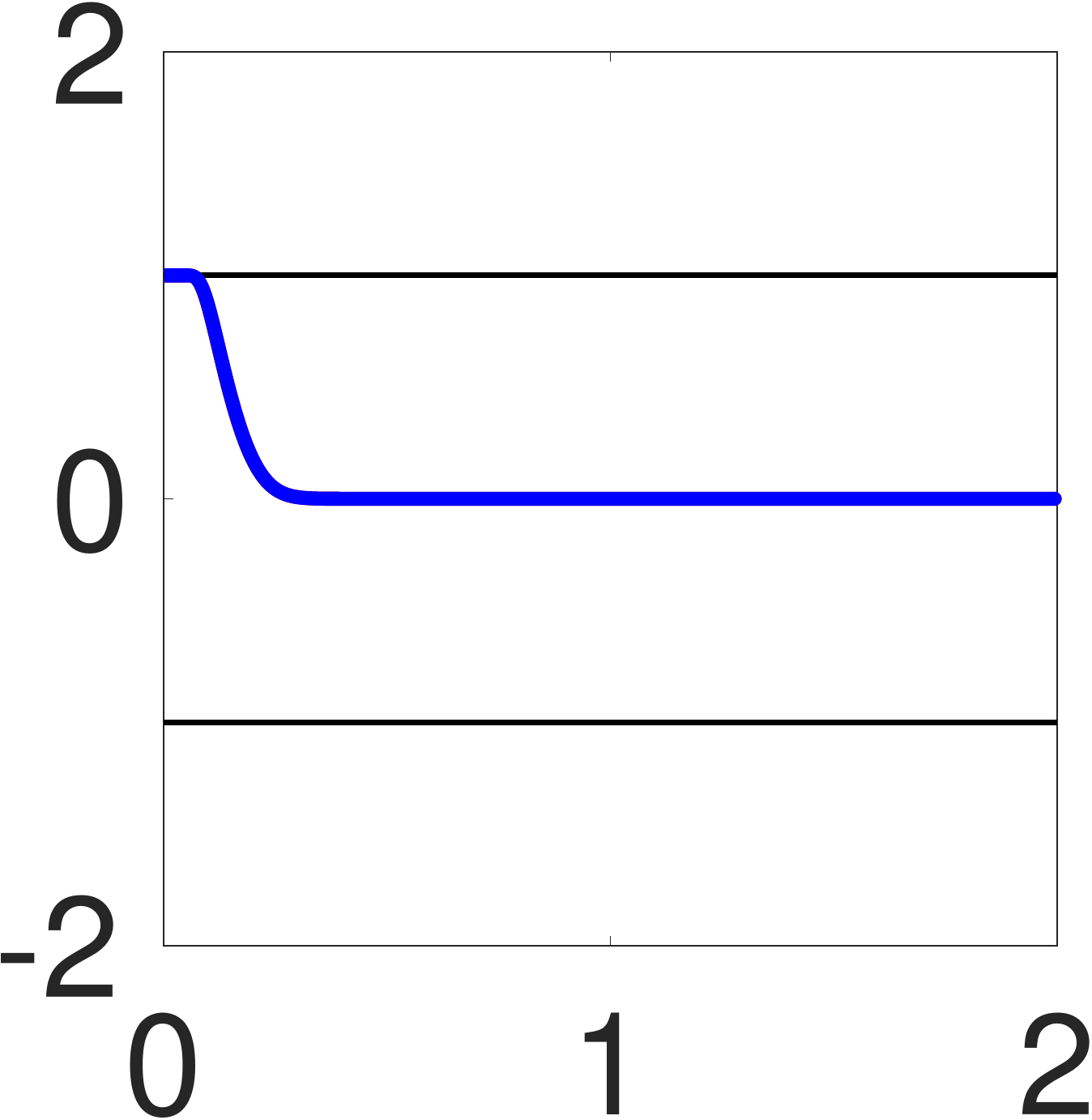}\\
$\tilde{\mathfrak{C}}_\tau(c)$&\includegraphics[scale=0.13,valign=c]{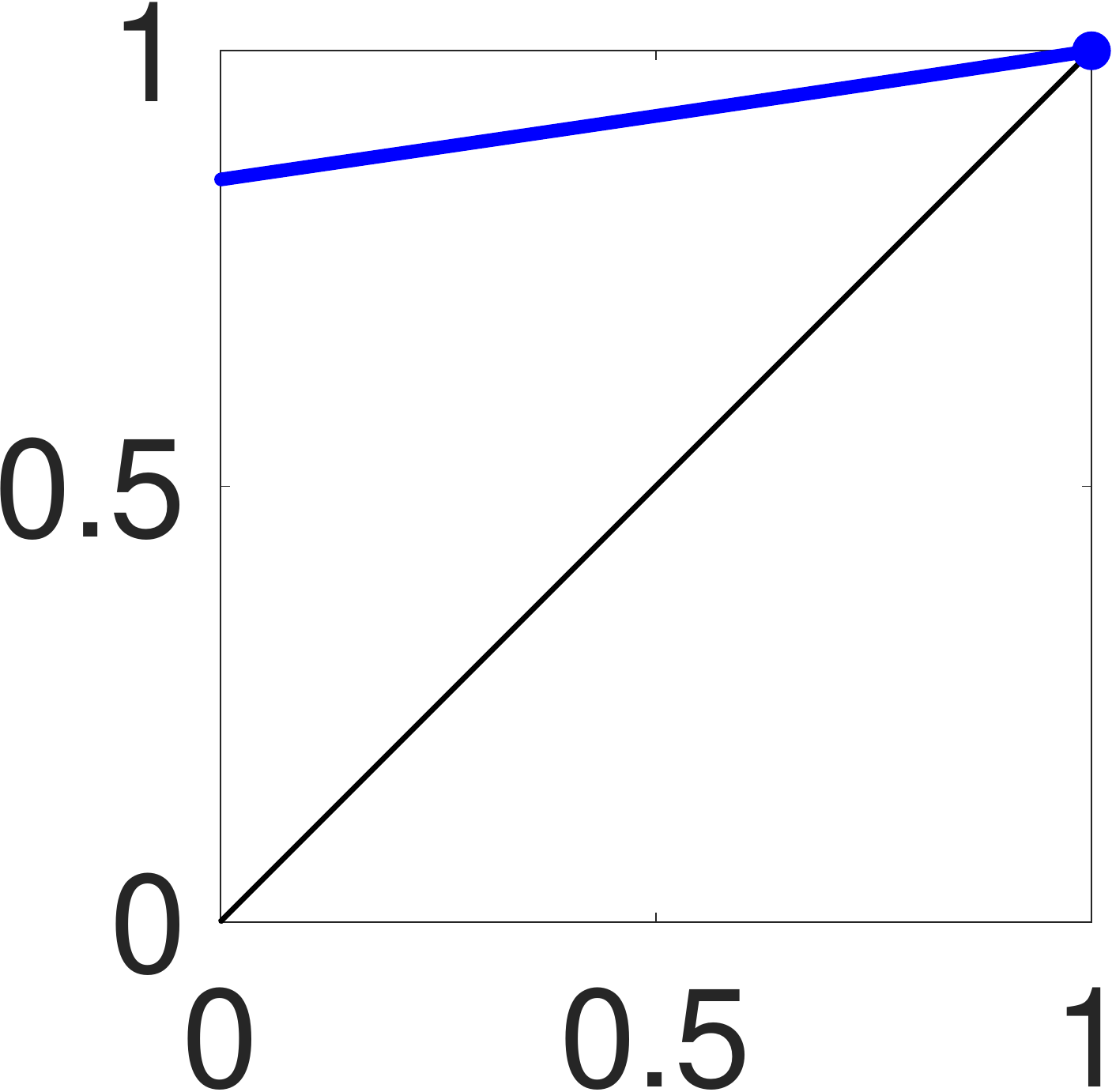}&\includegraphics[scale=0.13,valign=c]{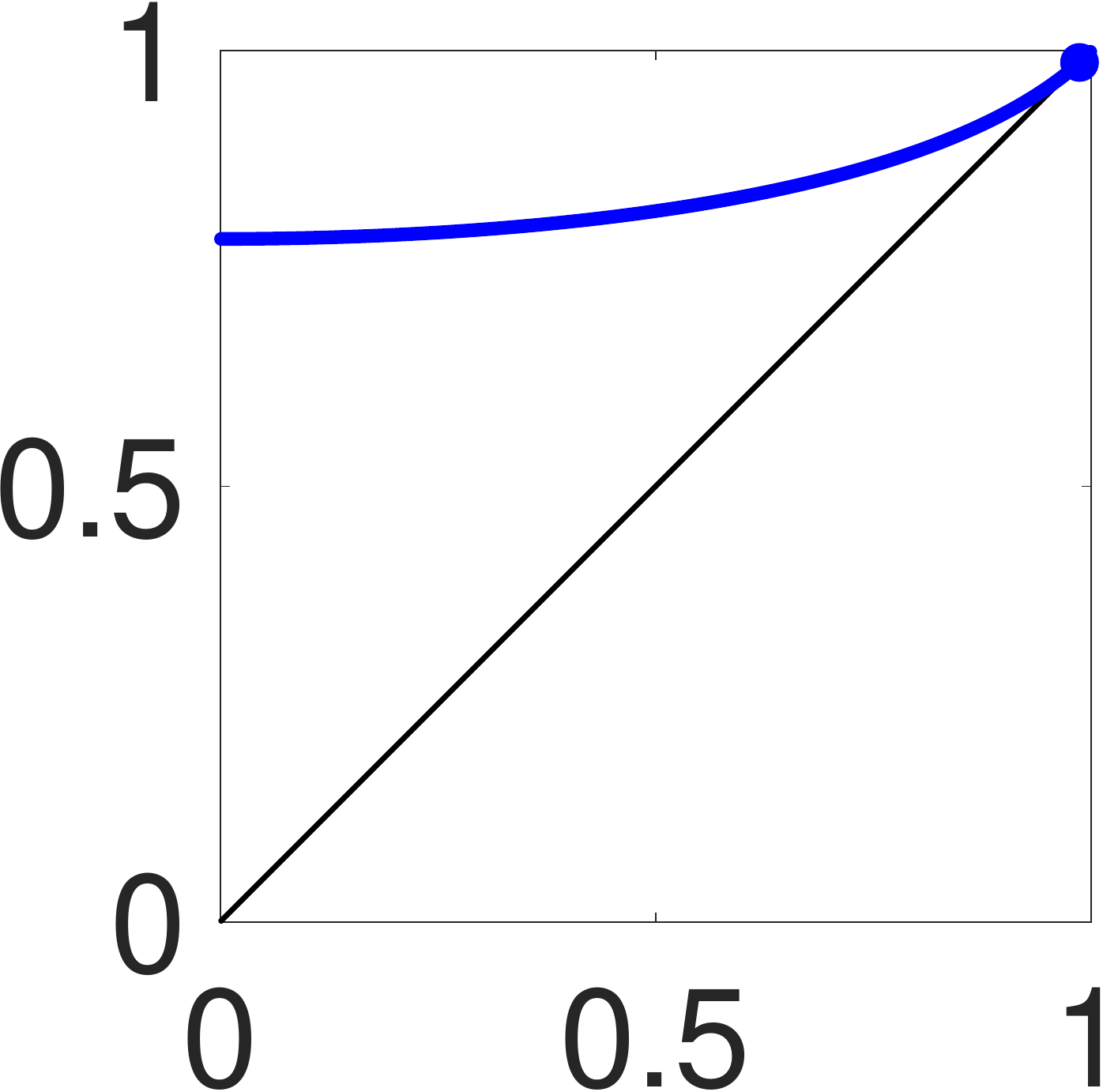}&\includegraphics[scale=0.13,valign=c]{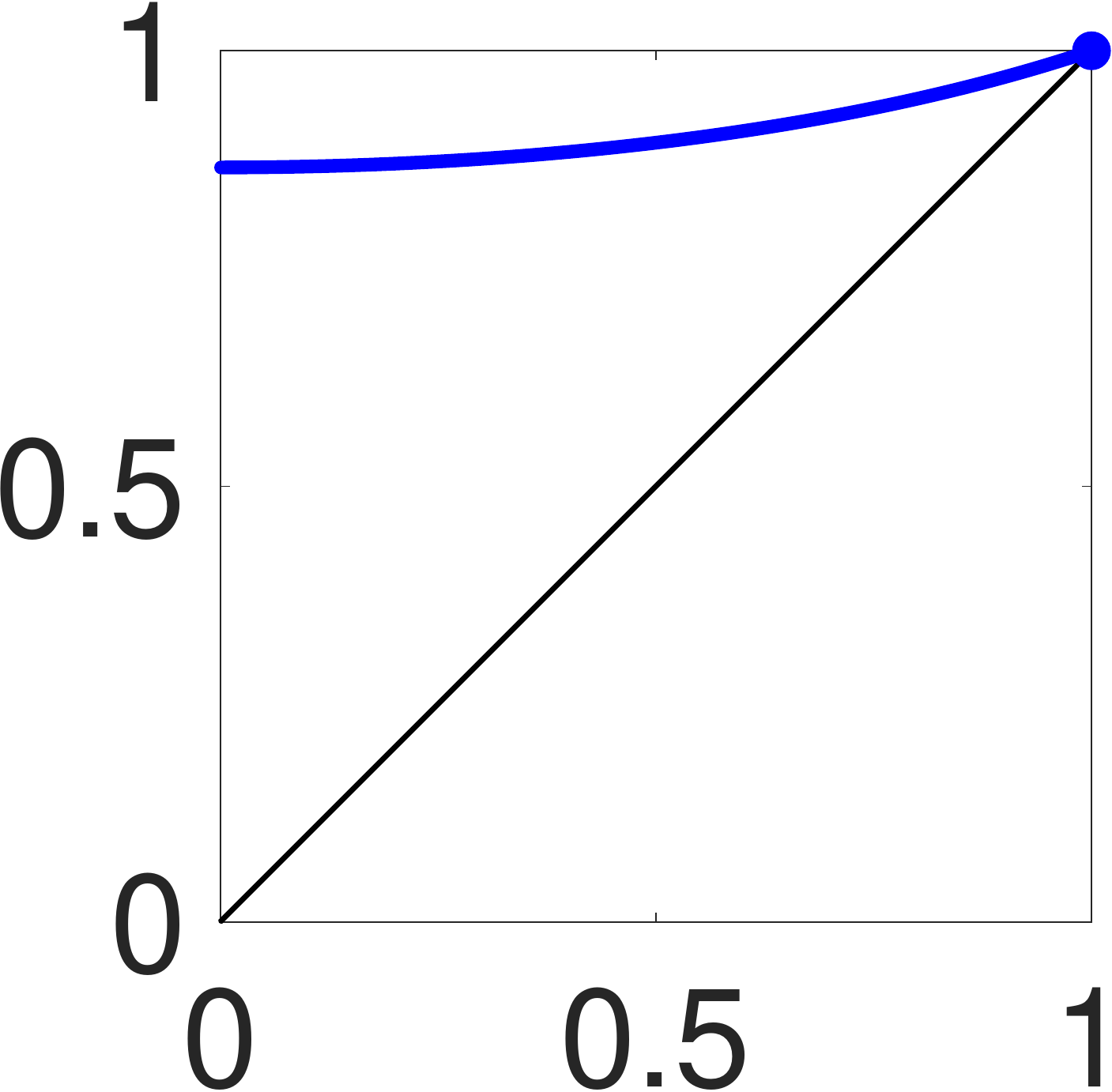}&\includegraphics[scale=0.13,valign=c]{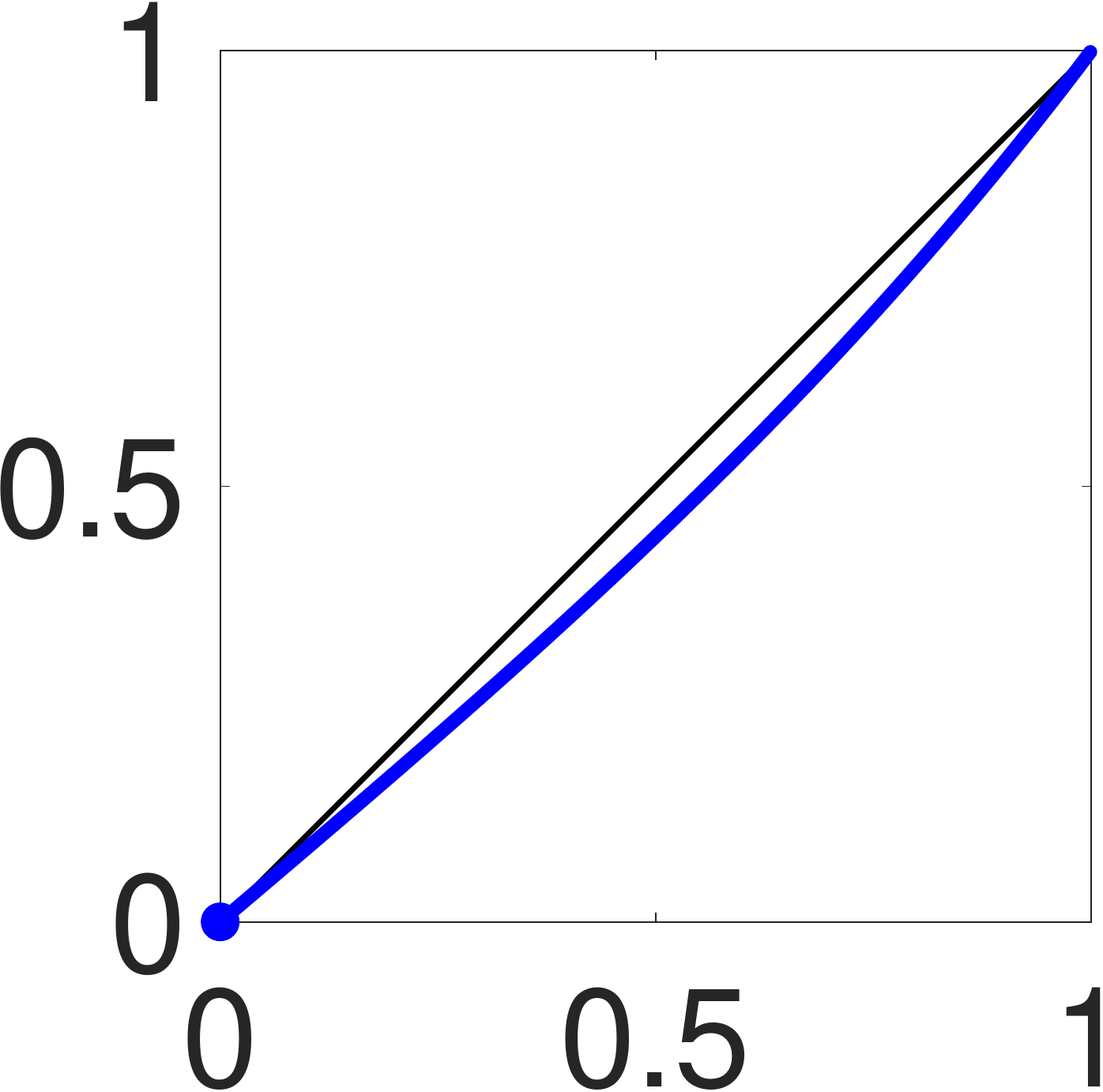}&\includegraphics[scale=0.13,valign=c]{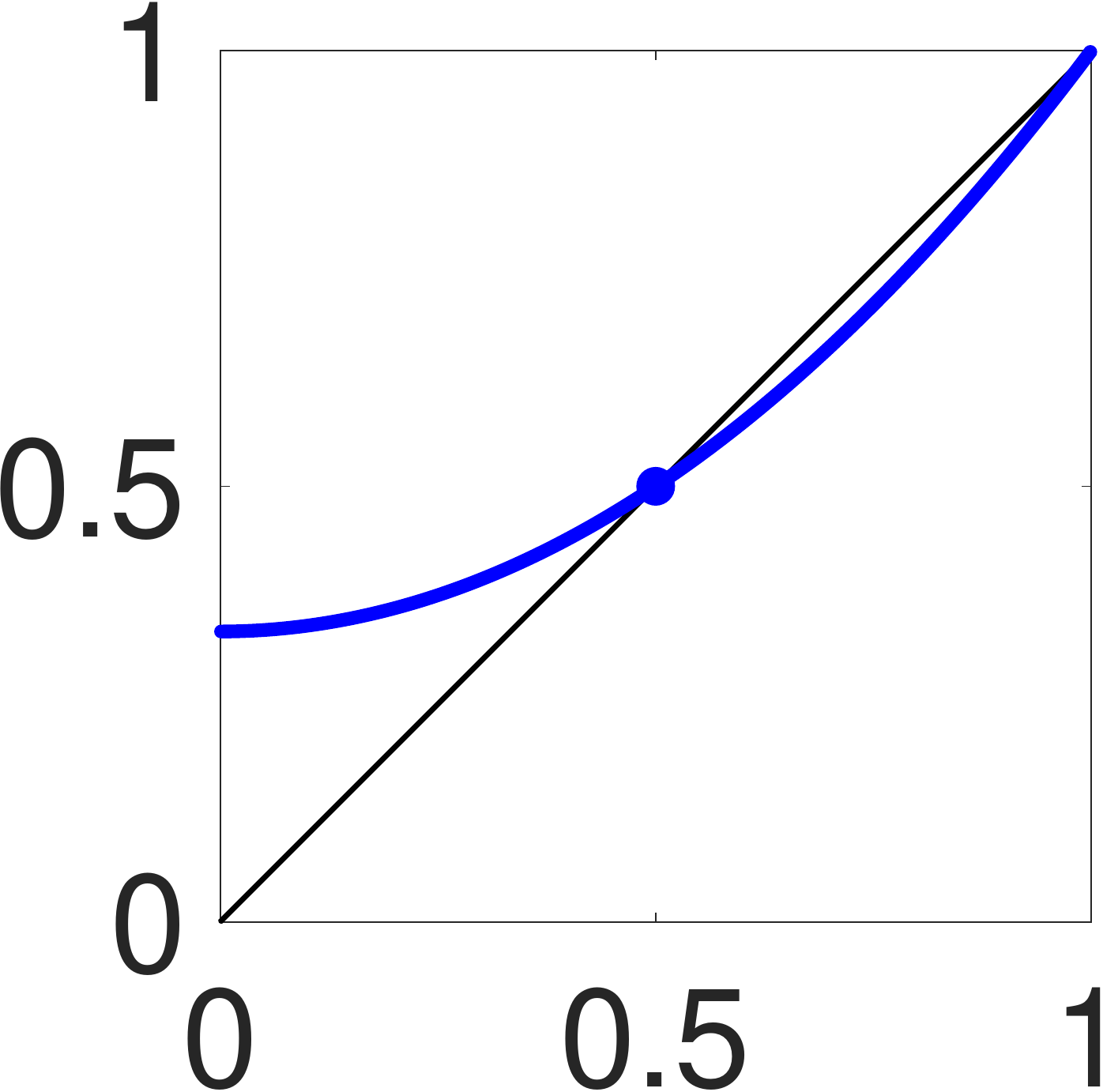}&\includegraphics[scale=0.13,valign=c]{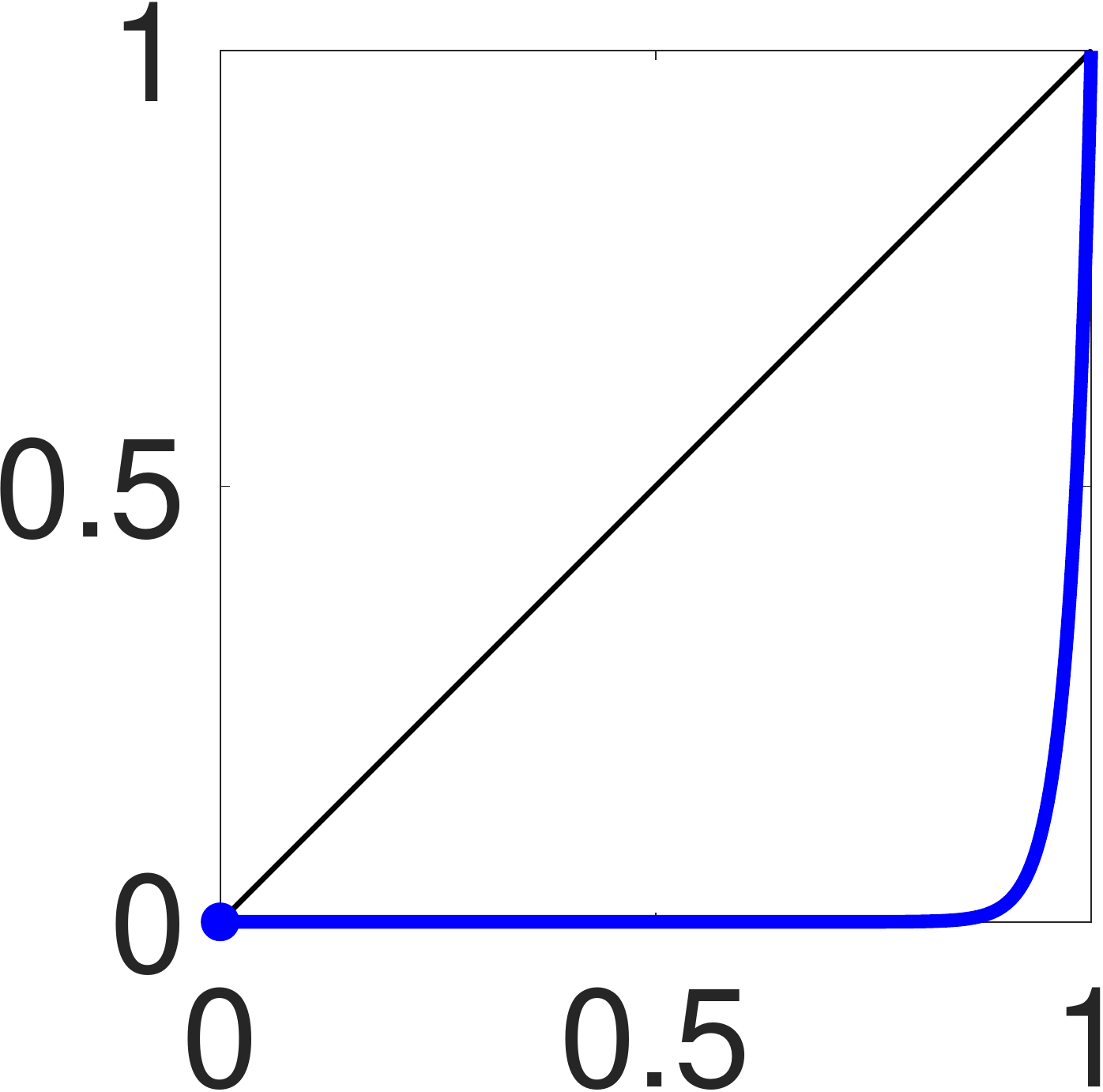}\\
Iter. limit &1&0.99&1&0&0.5&0\\
Conv. rate&$O(0.15^M)$&$O(0.91^M)$&$O(0.33^M)$&$O(0.85^M)$&$O(0.67^M)$&super-exp.\\
$\mathfrak{n}_\tau(\lambda^2, 0)$&\includegraphics[scale=0.13,valign=c]{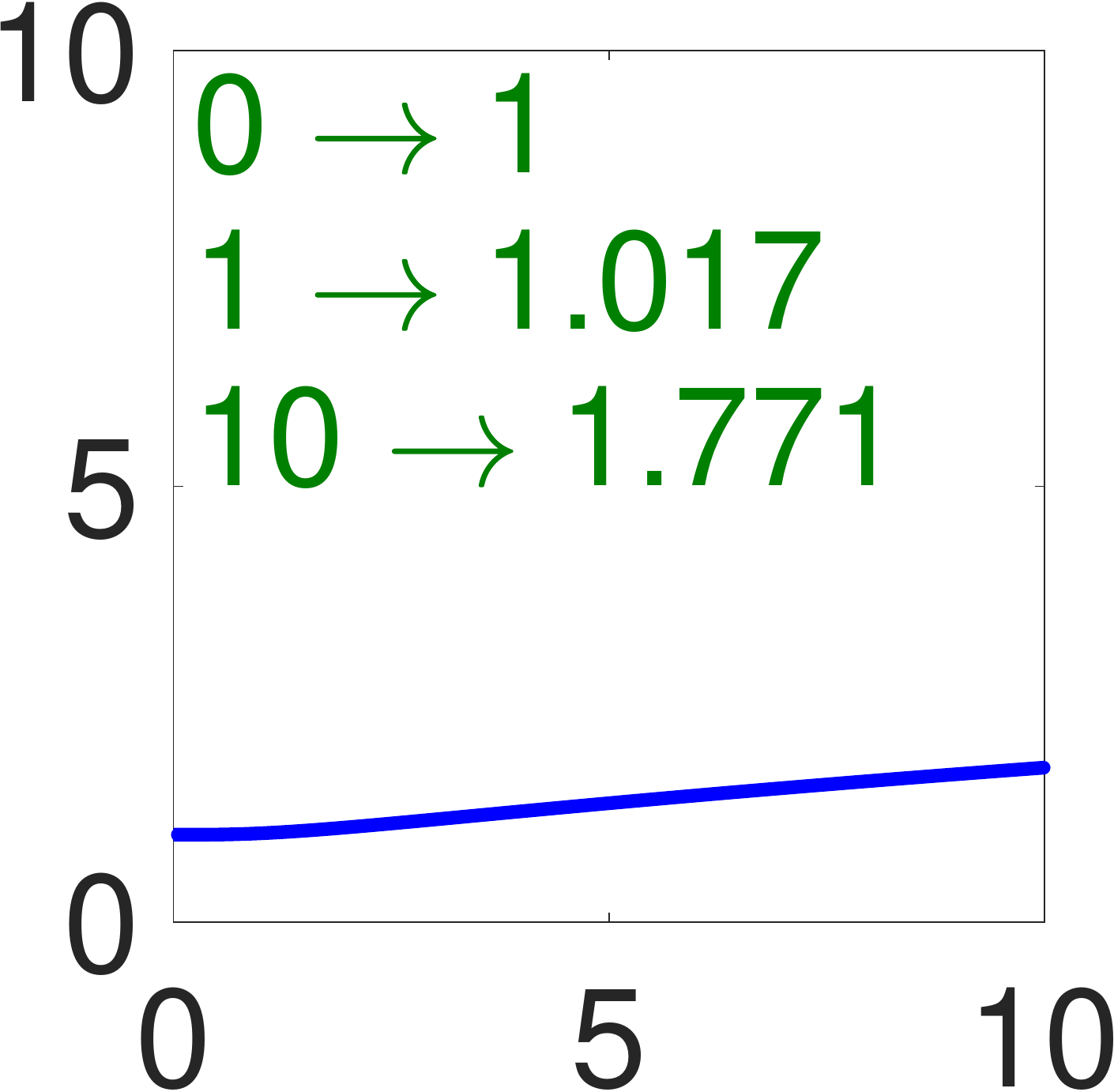}&\includegraphics[scale=0.13,valign=c]{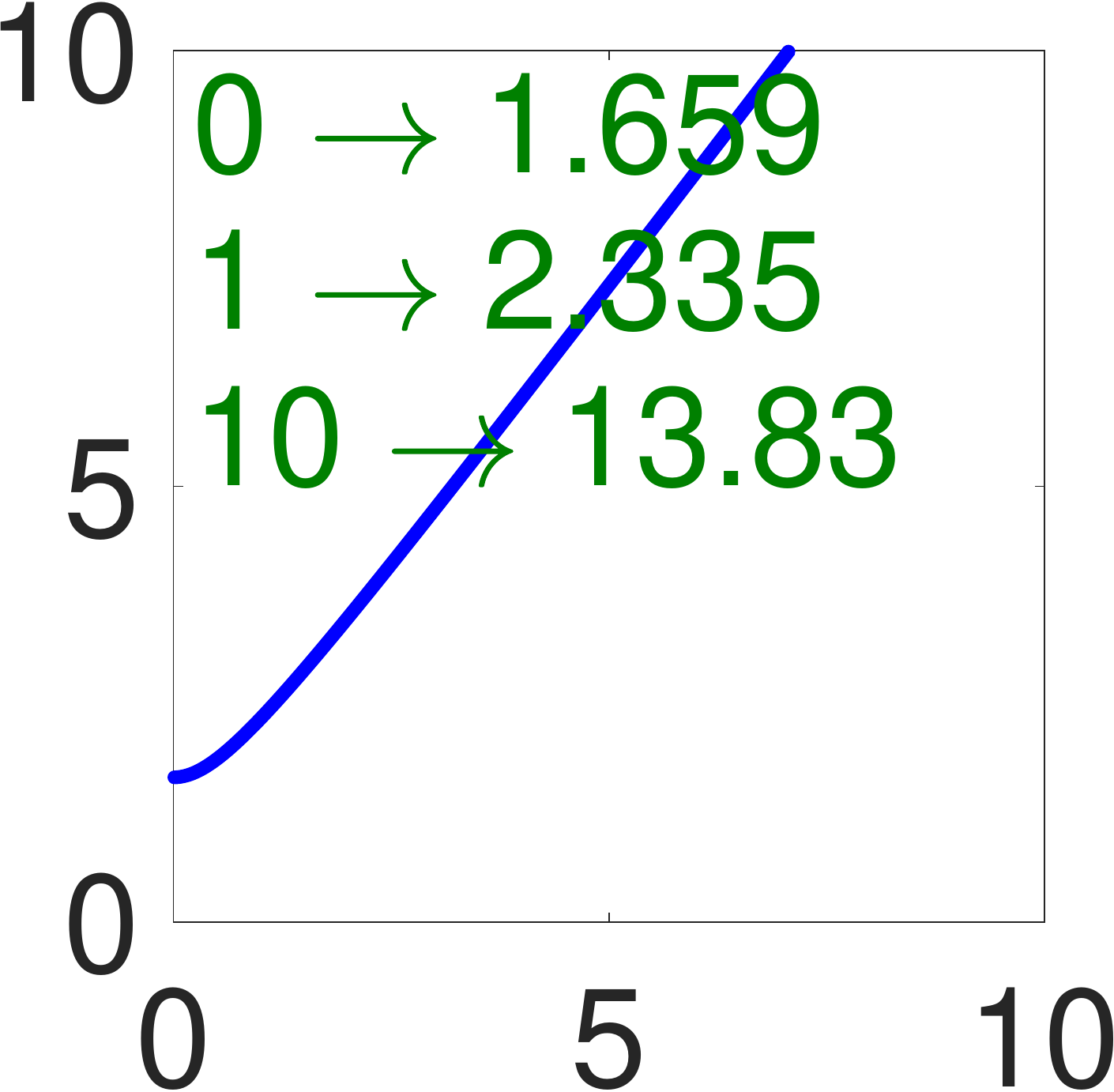}&\includegraphics[scale=0.13,valign=c]{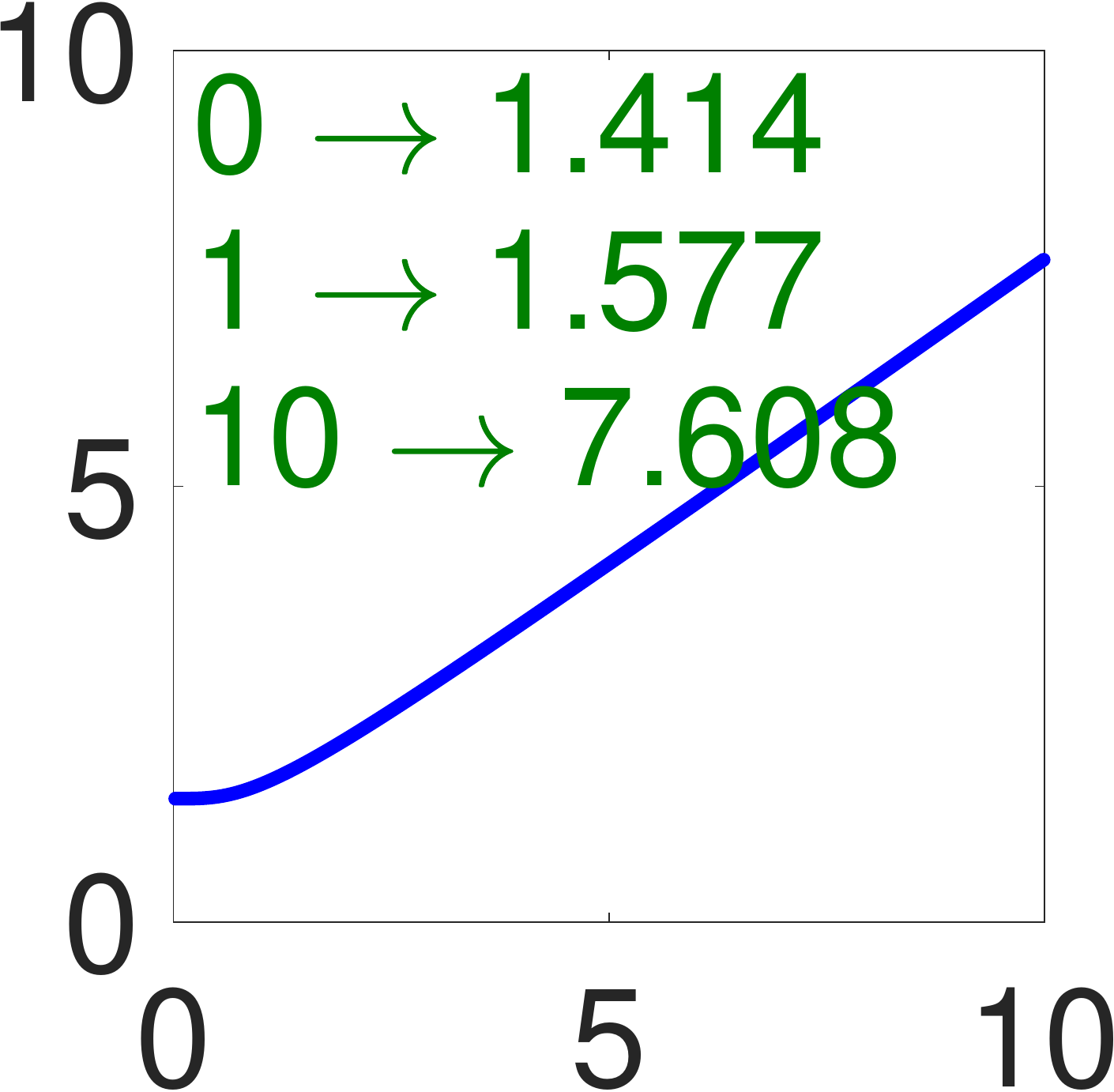}&\includegraphics[scale=0.13,valign=c]{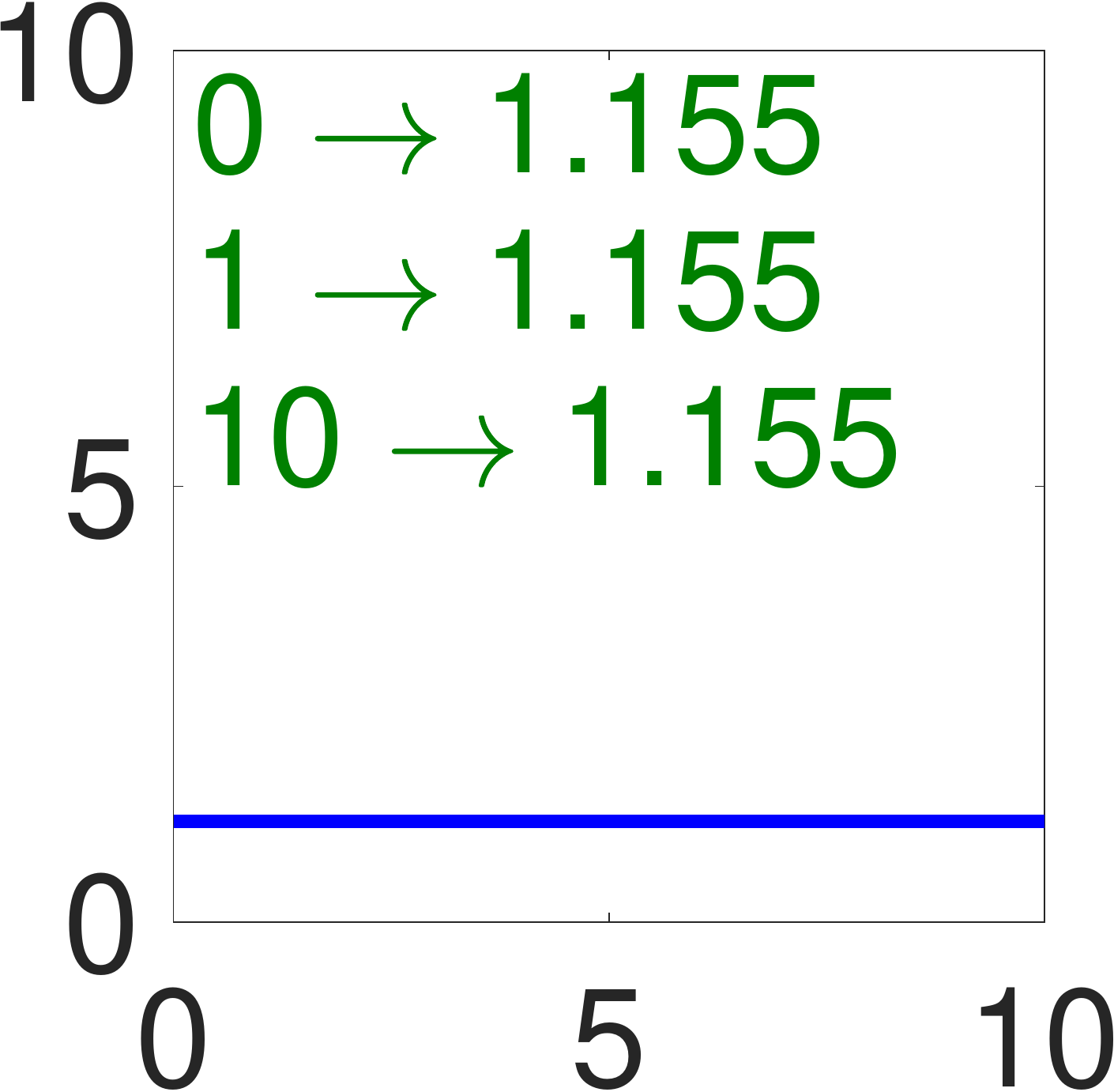}&\includegraphics[scale=0.13,valign=c]{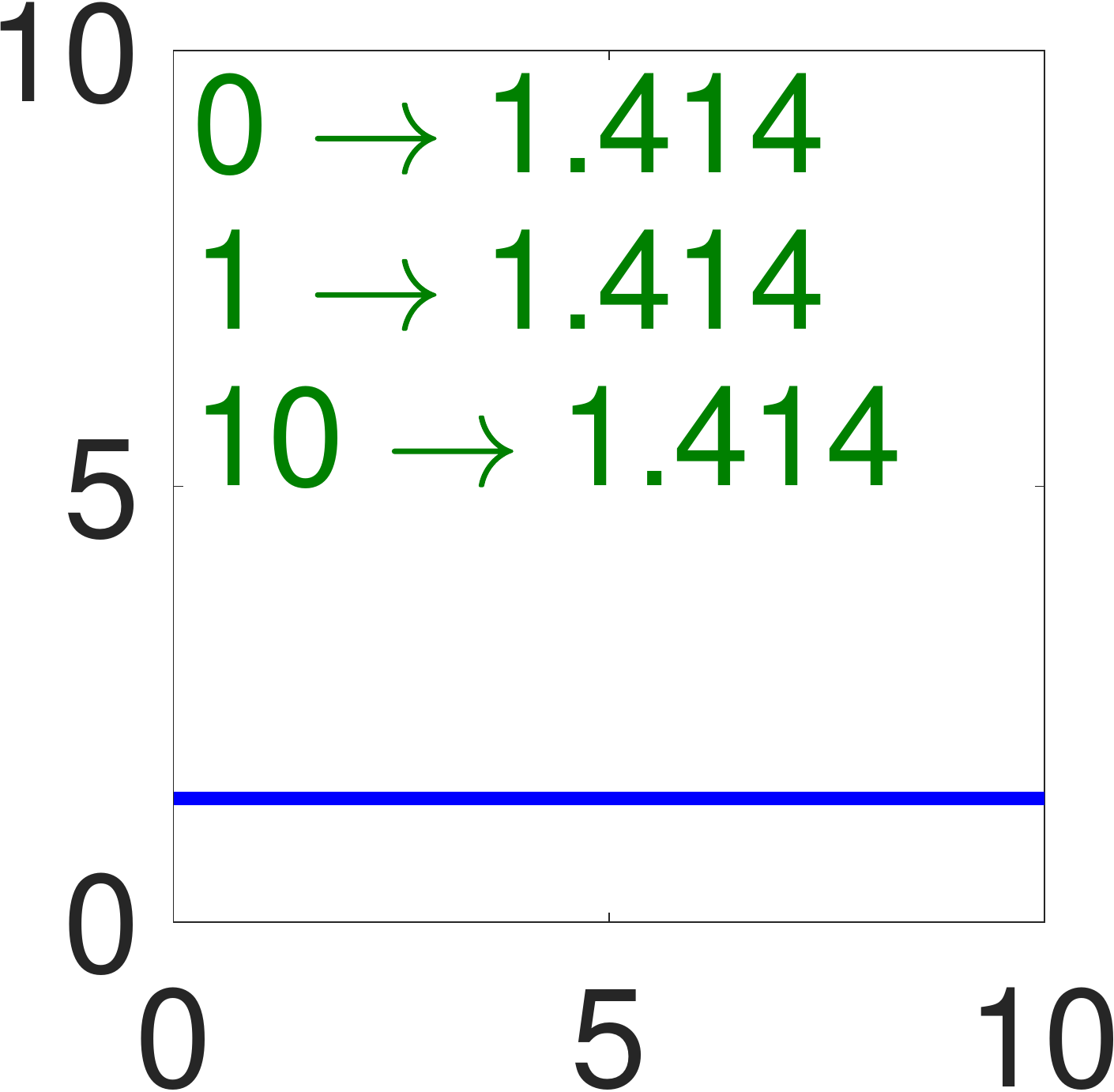}&\includegraphics[scale=0.13,valign=c]{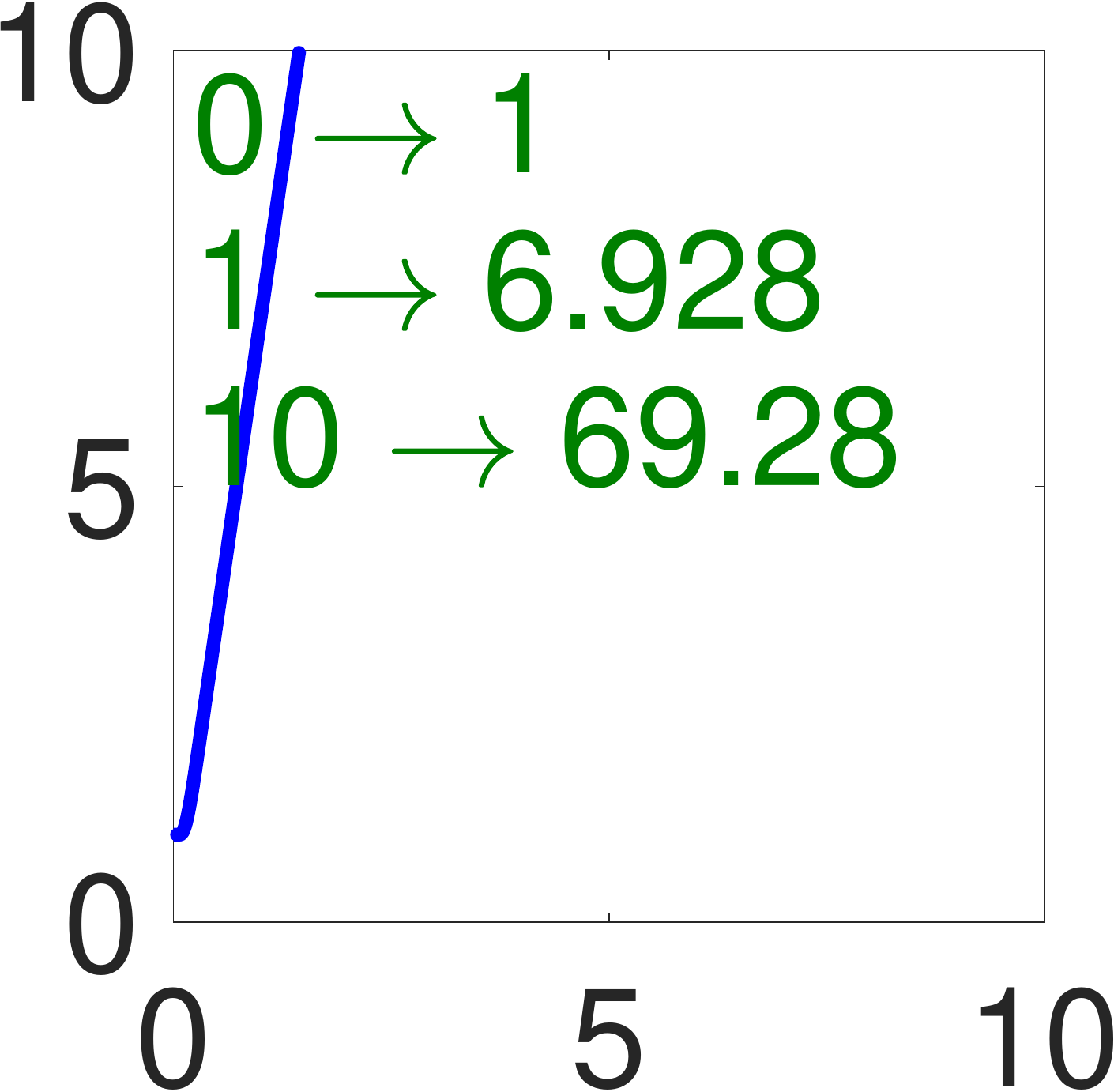}\\
$\lambda \rightarrow \infty$&$O(\sqrt{\lambda})$&$O(\lambda)$&$O(\lambda)$&1.15&1.41&$O(\lambda)$\\
\\
\end{tabular}
}
\caption{(This table is a continuation of table \ref{covCurveillu1}.)}
\label{covCurveillu2}
\end{table}

\begin{table}[H]
{
\centering \small
\begin{tabular}{lcccccc}
Act. fun. &ReLU-deb.&SELU-deb.&softpl.-deb.&Swish-deb.&abs. v.-deb.&tanh-deb.\\ \hline\hline
\\
$\tau(s)$&\includegraphics[scale=0.13,valign=c]{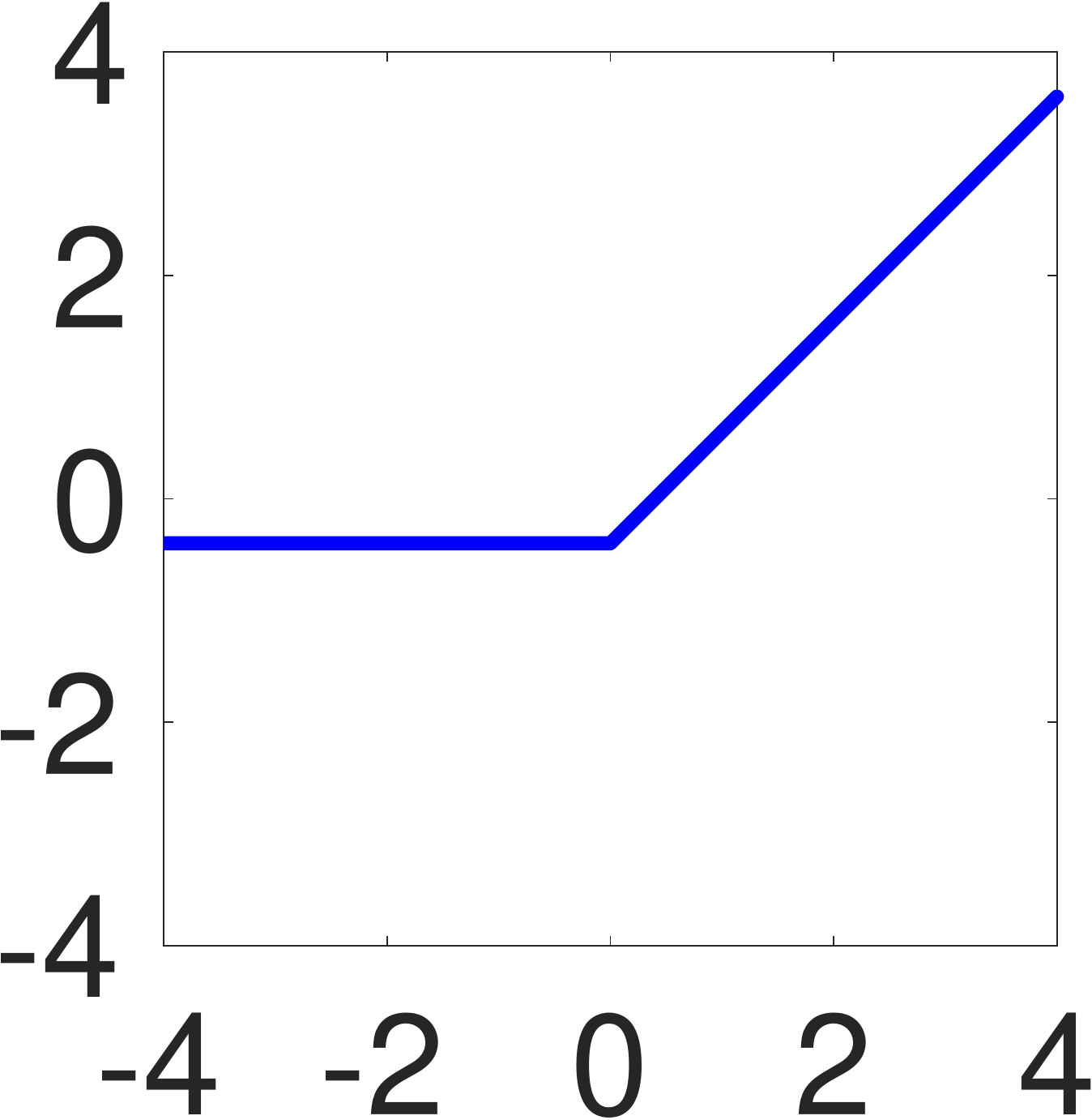}&\includegraphics[scale=0.13,valign=c]{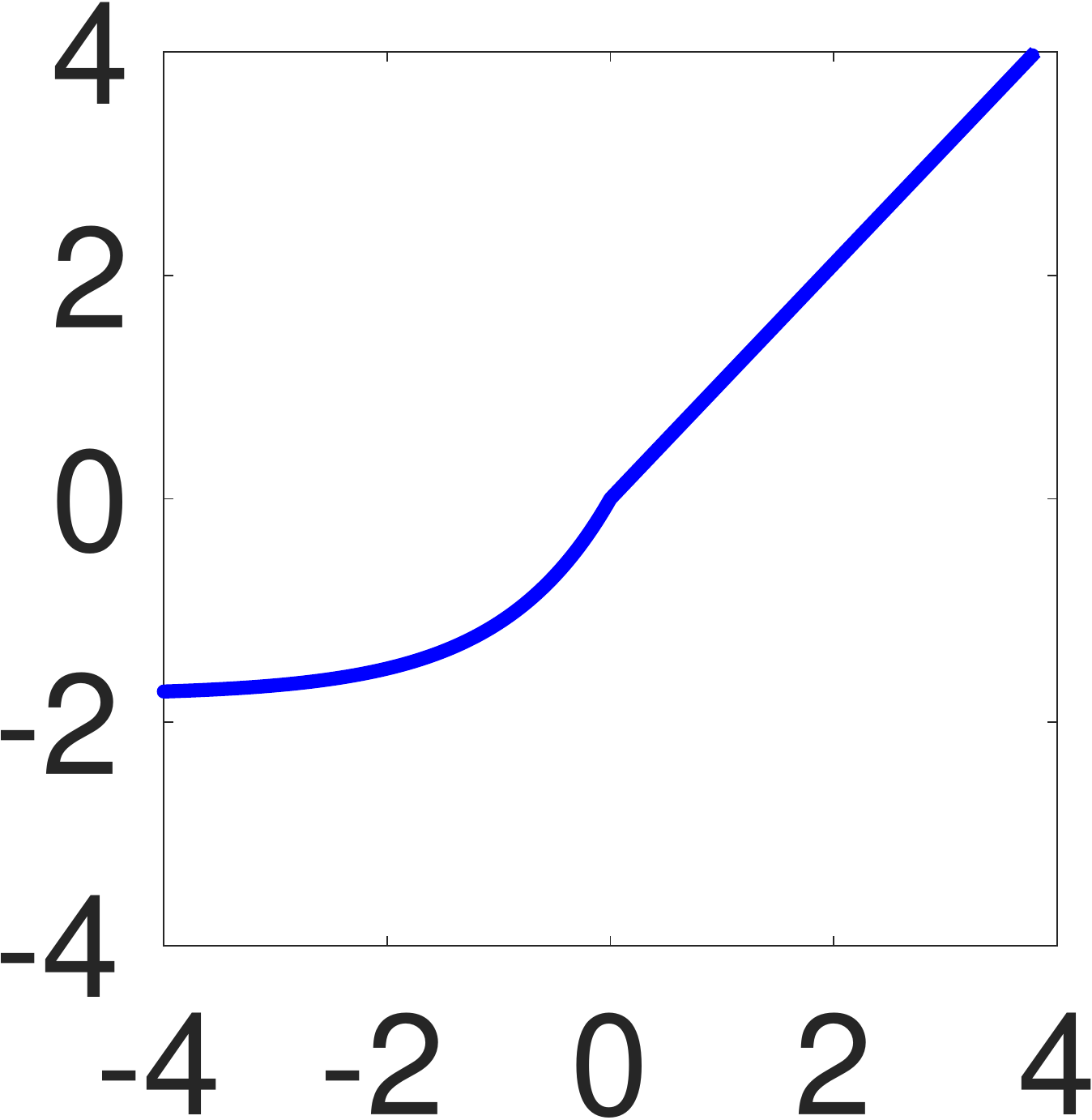}&\includegraphics[scale=0.13,valign=c]{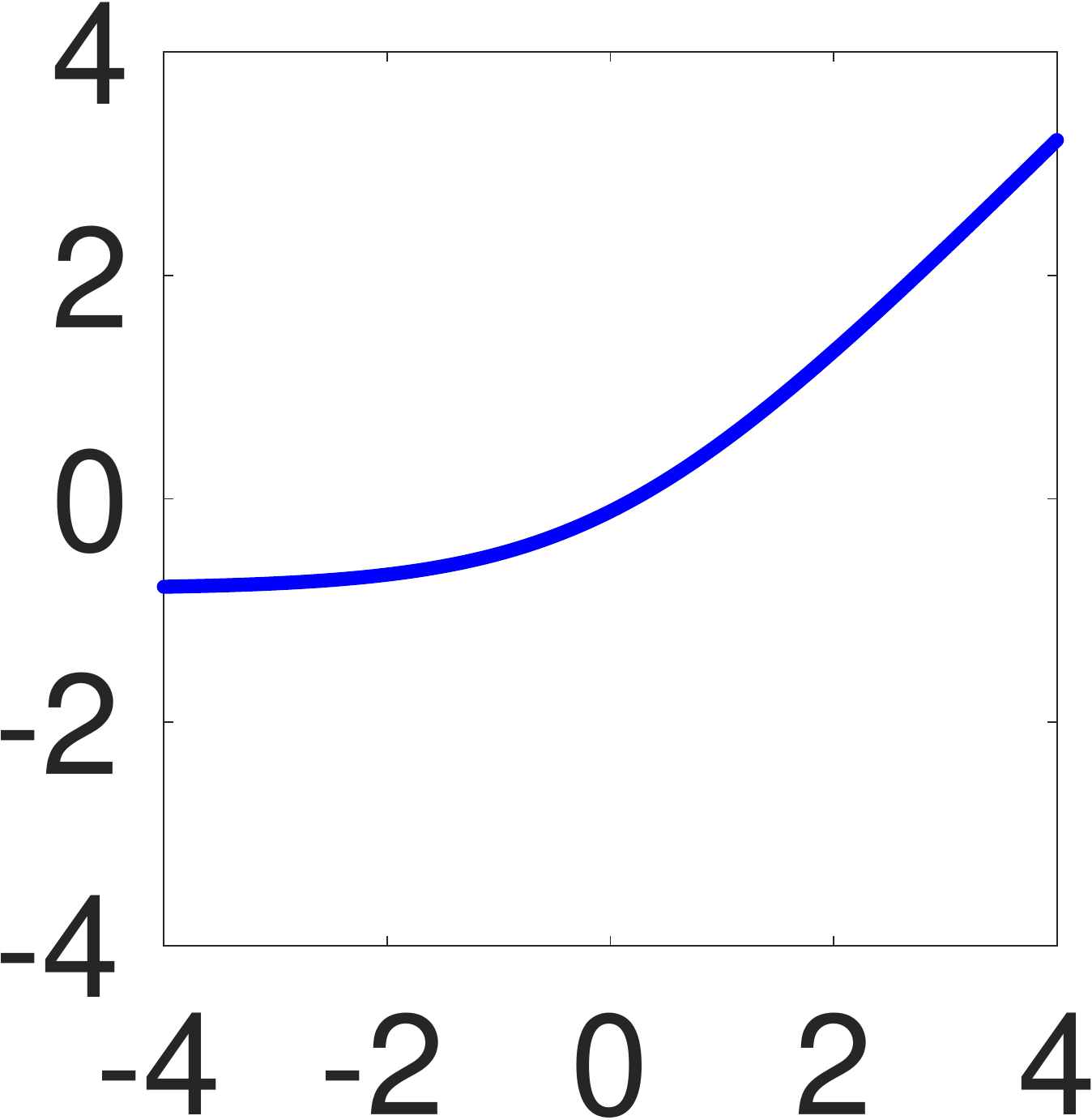}&\includegraphics[scale=0.13,valign=c]{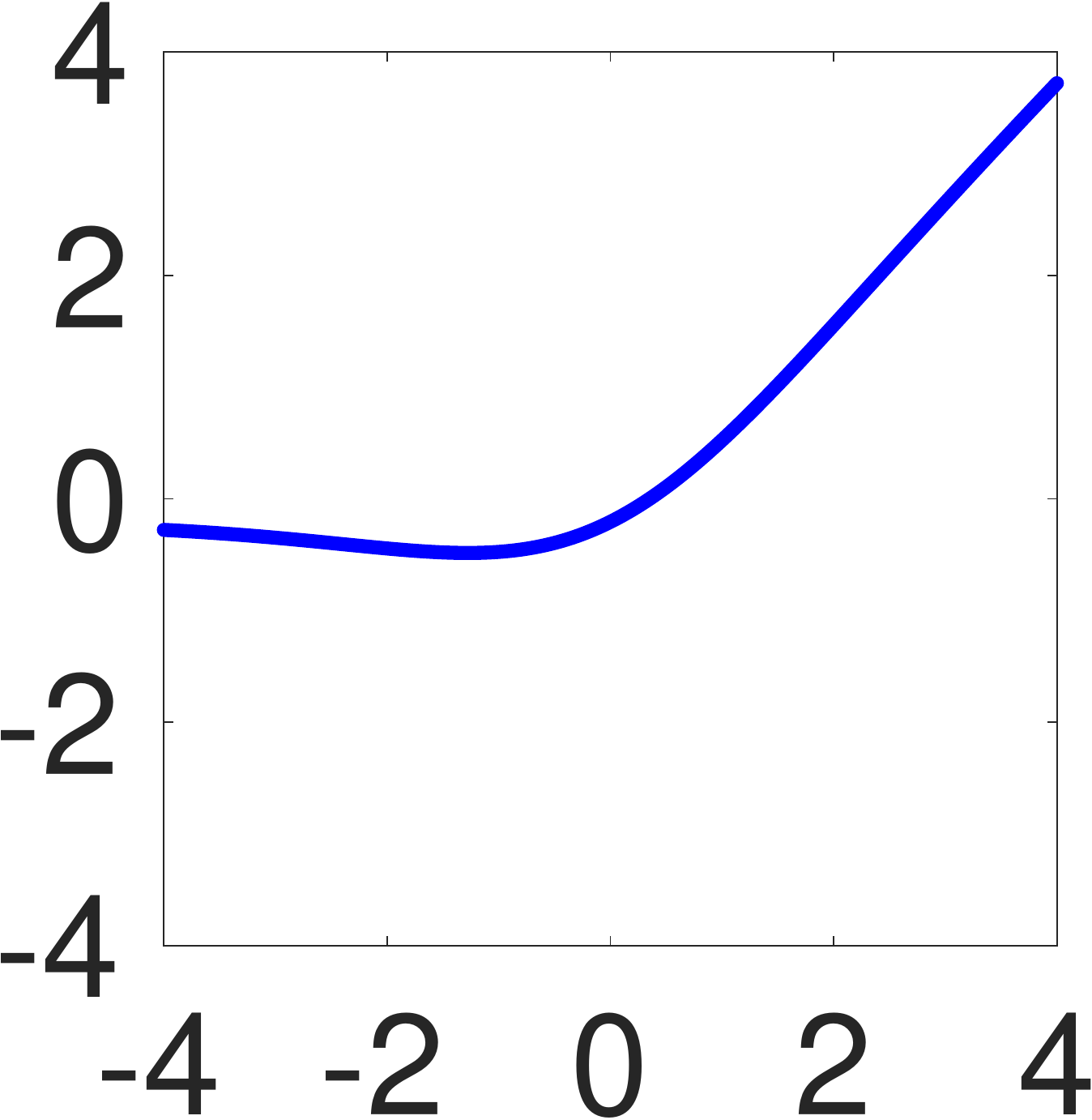}&\includegraphics[scale=0.13,valign=c]{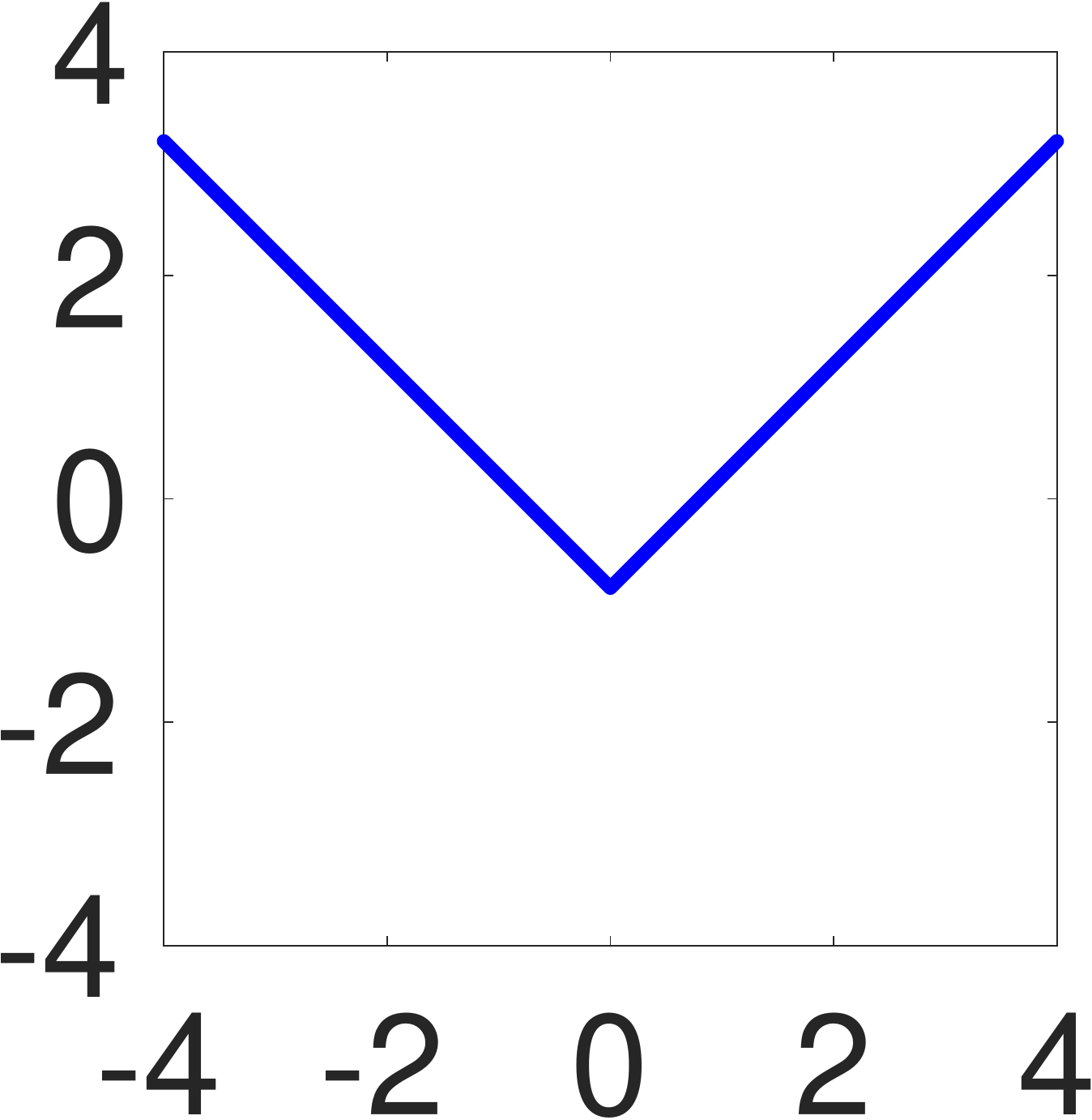}&\includegraphics[scale=0.13,valign=c]{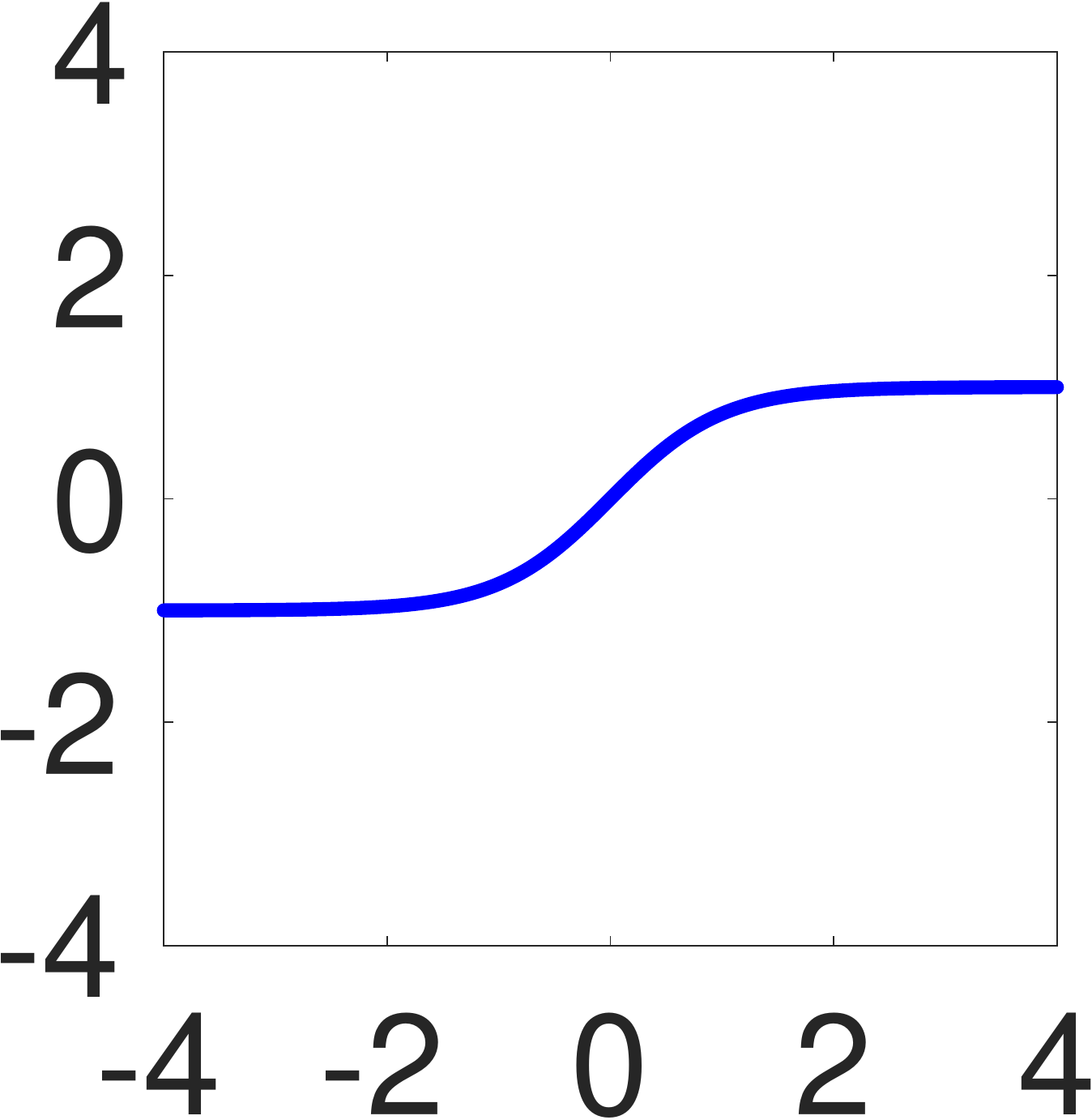}\\
$\mathfrak{L}_\tau(\lambda)$&\includegraphics[scale=0.13,valign=c]{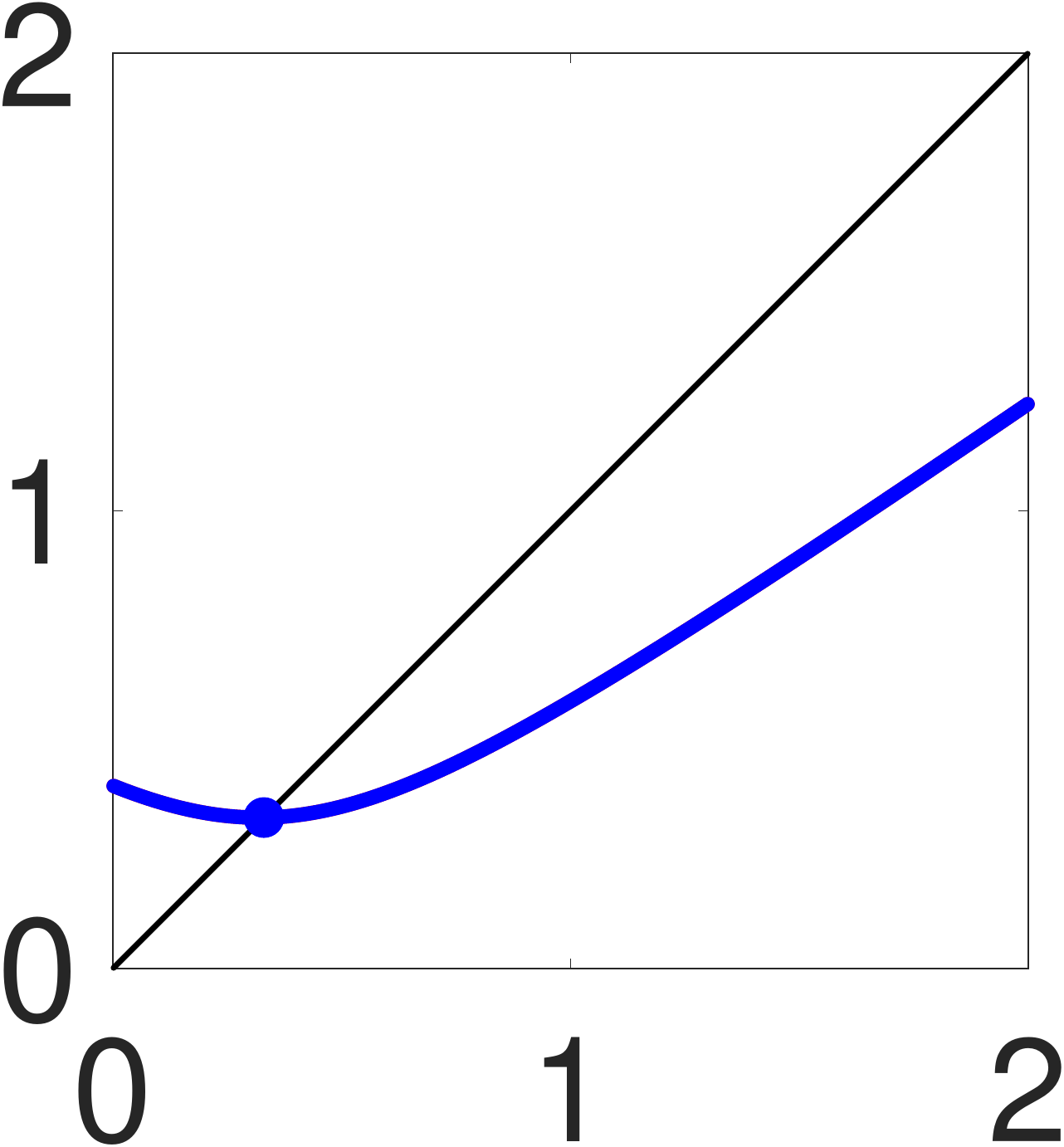}&\includegraphics[scale=0.13,valign=c]{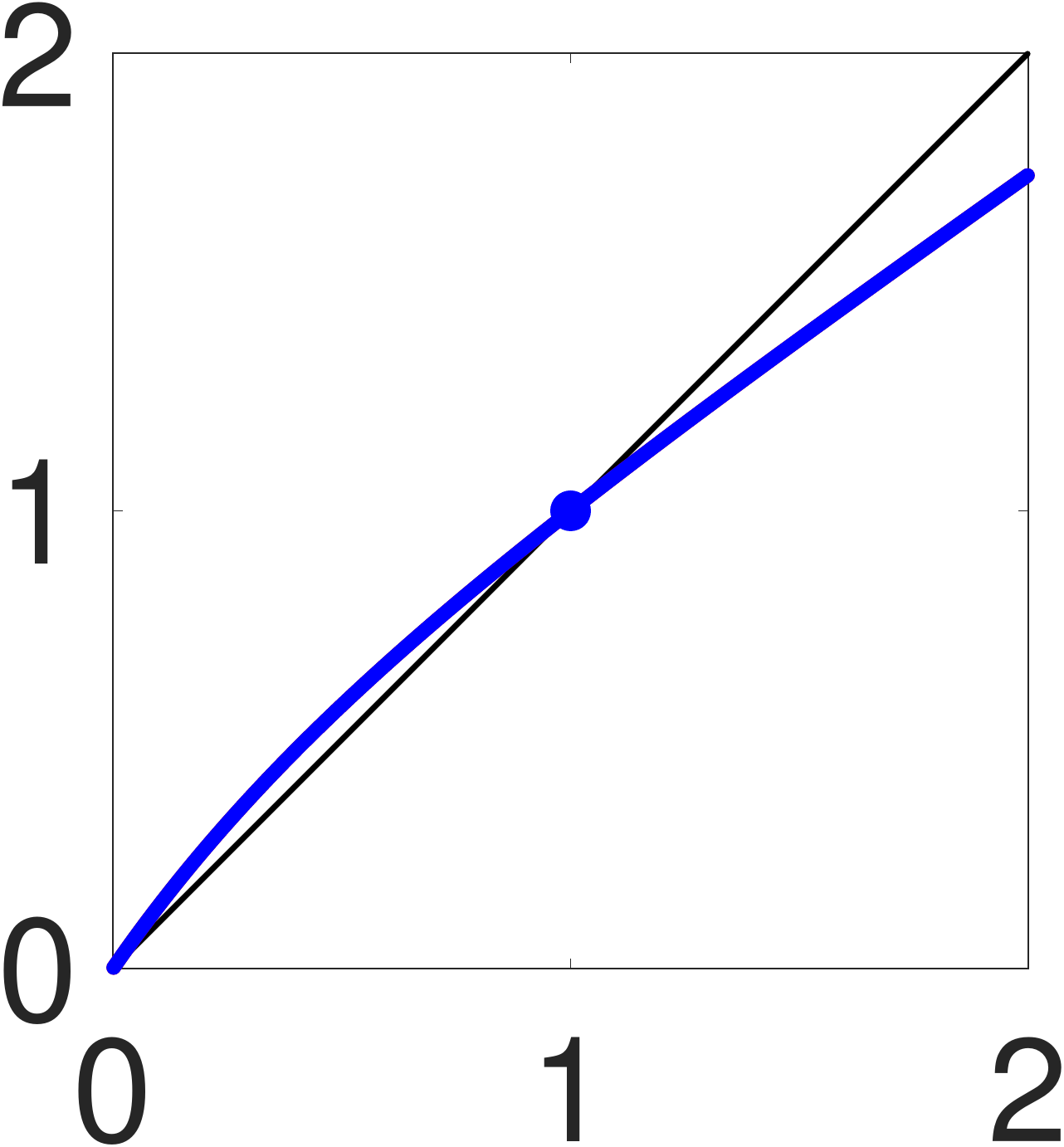}&\includegraphics[scale=0.13,valign=c]{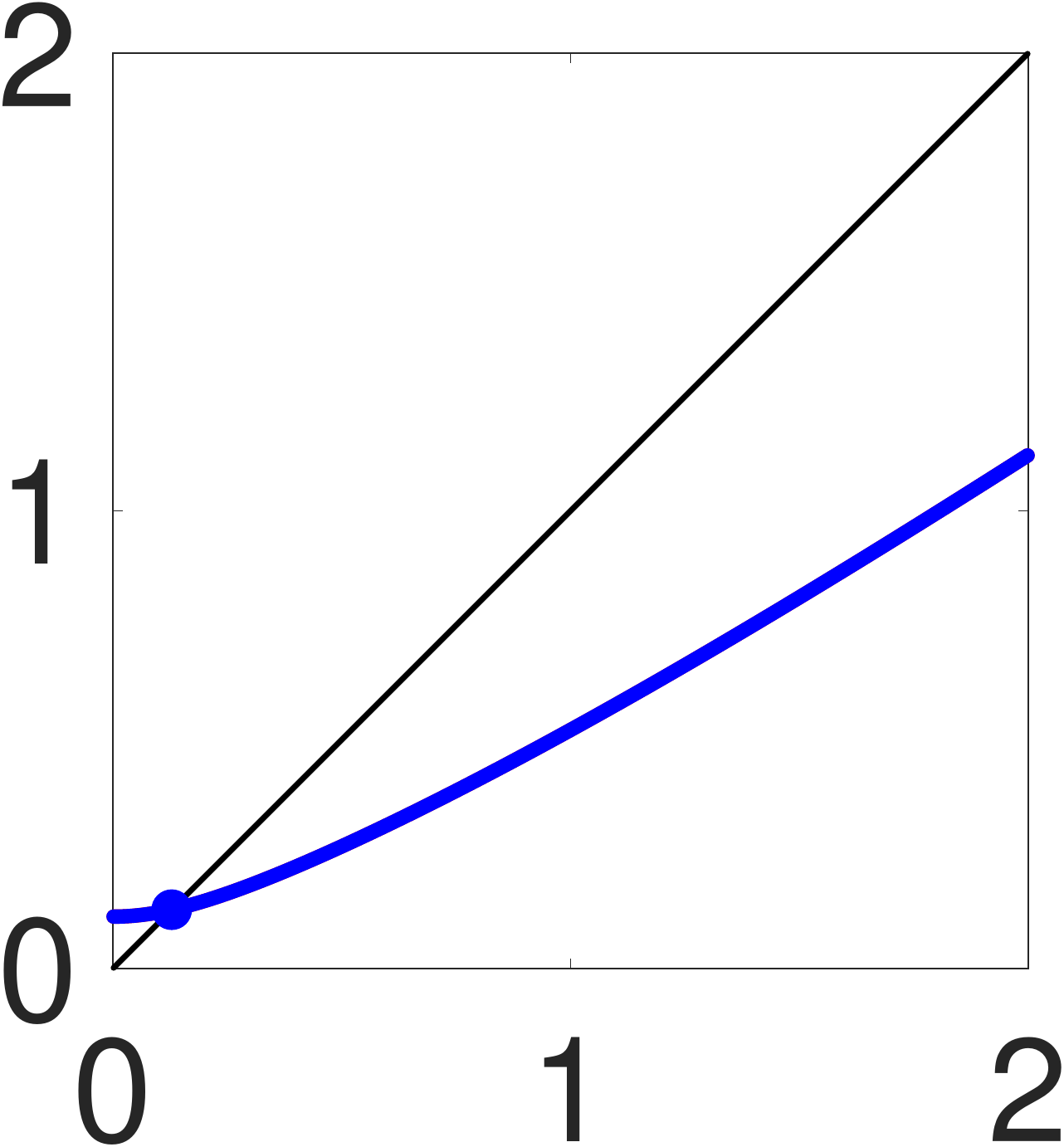}&\includegraphics[scale=0.13,valign=c]{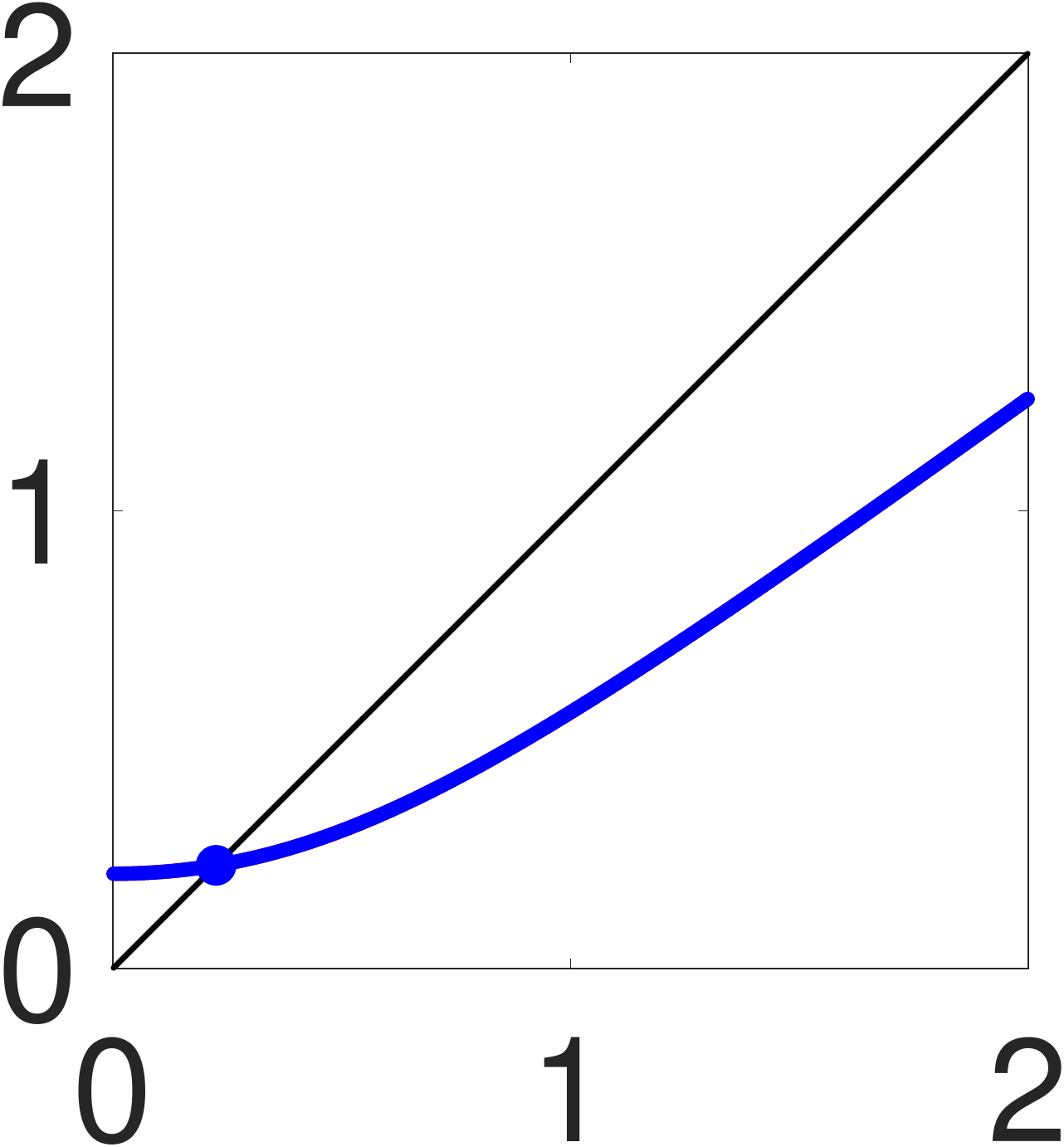}&\includegraphics[scale=0.13,valign=c]{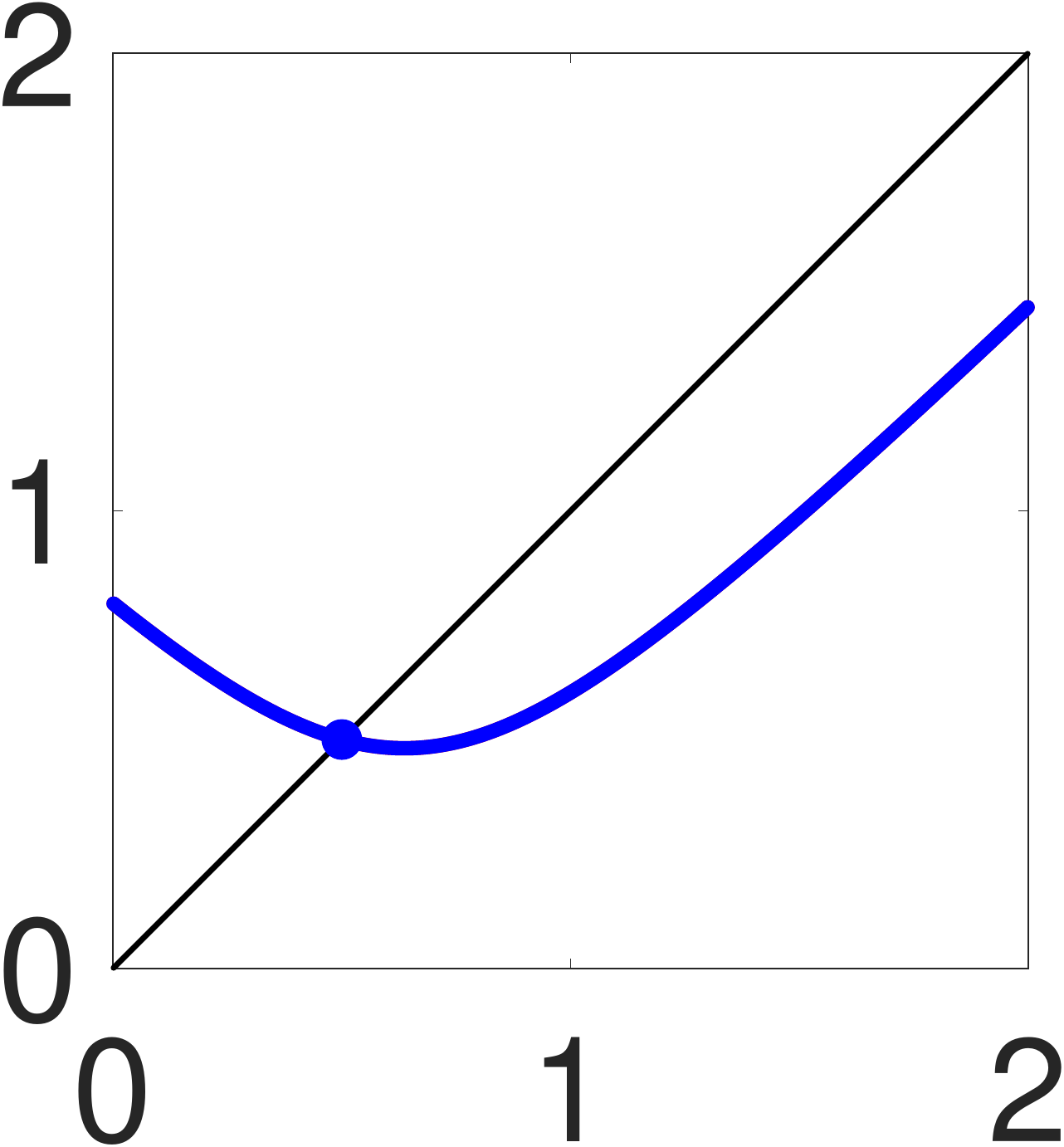}&\includegraphics[scale=0.13,valign=c]{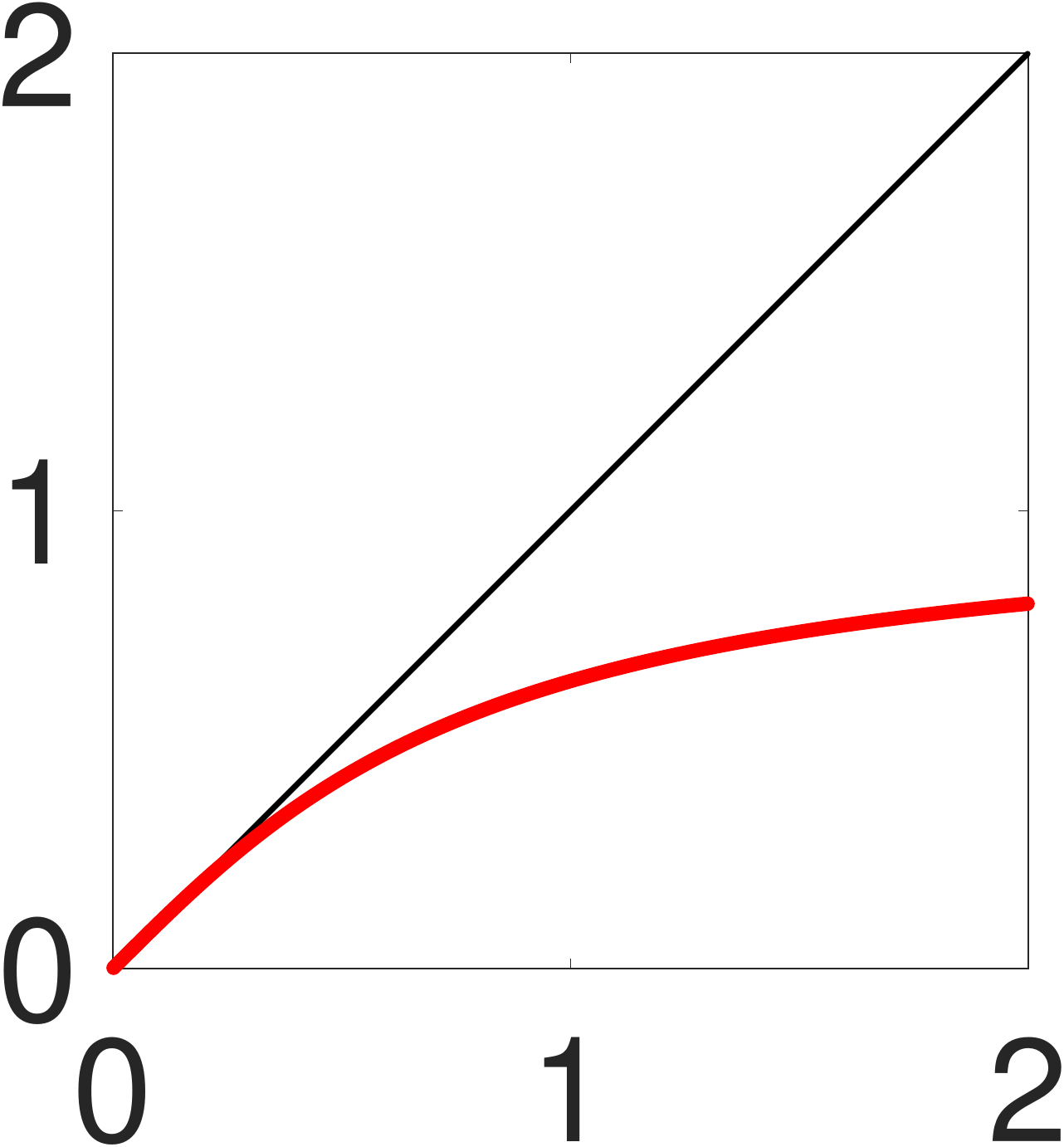}\\
Iter. limit &0.33&1.00&0.13&0.23&0.5&0\\
Conv. rate&$O(0.02^M)$&$O(0.78^M)$&$O(0.22^M)$&$O(0.17^M)$&$O(0.27^M)$&$O(\frac{1}{M^{0.5}})$\\
$\frac{\mathfrak{L}_\tau(\lambda)}{\mathfrak{L}_\tau(1)}$&\includegraphics[scale=0.13,valign=c]{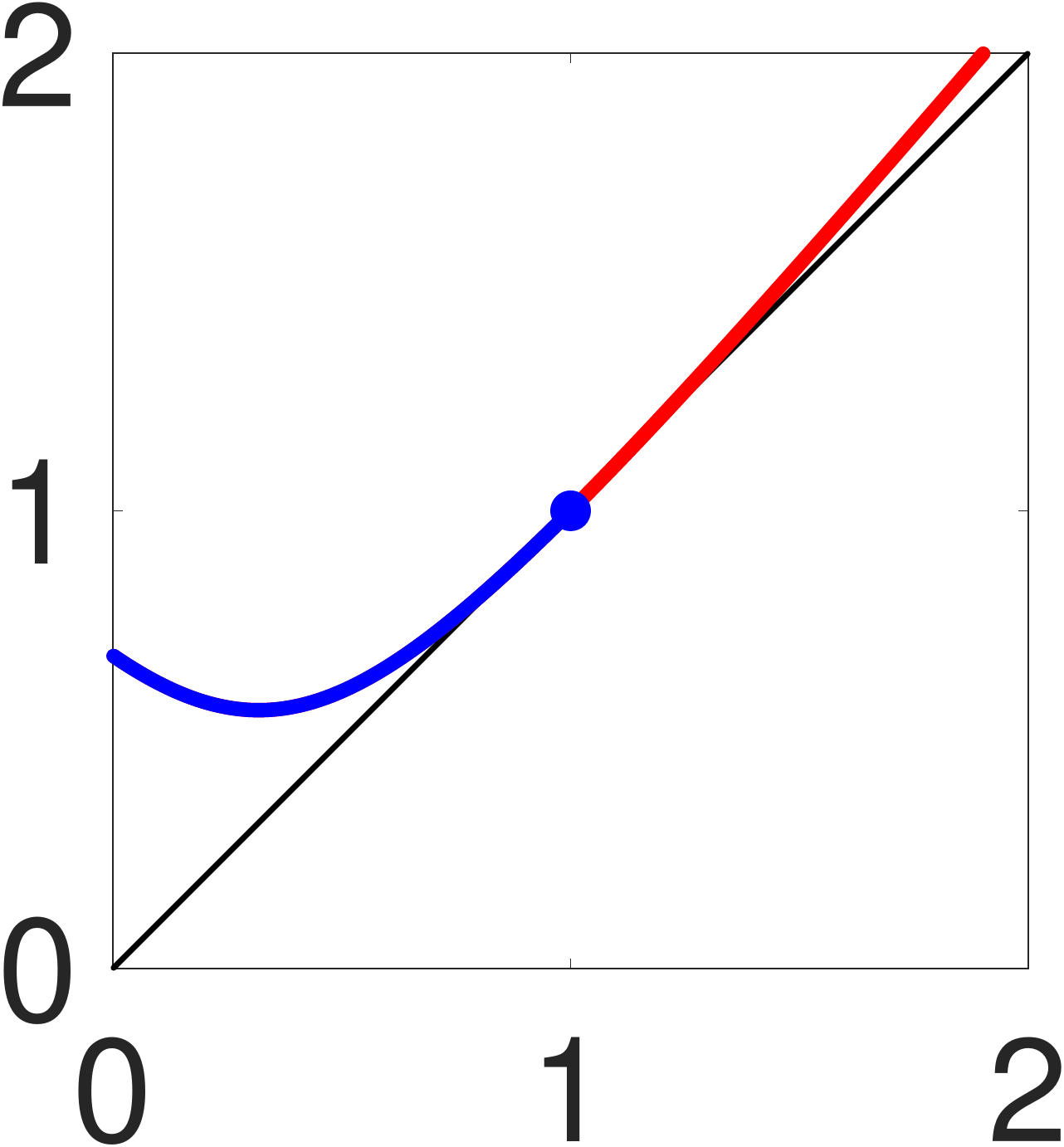}&\includegraphics[scale=0.13,valign=c]{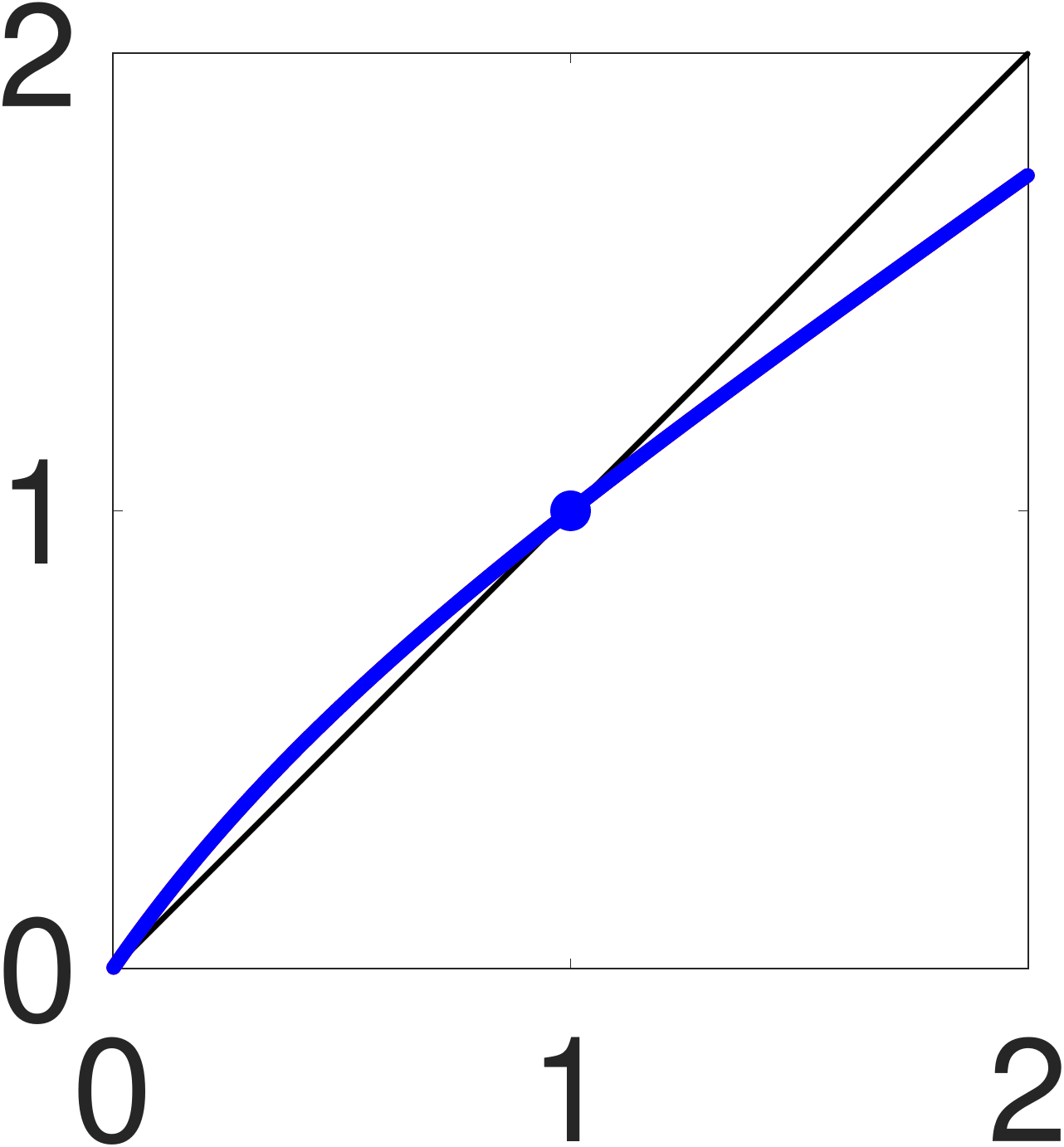}&\includegraphics[scale=0.13,valign=c]{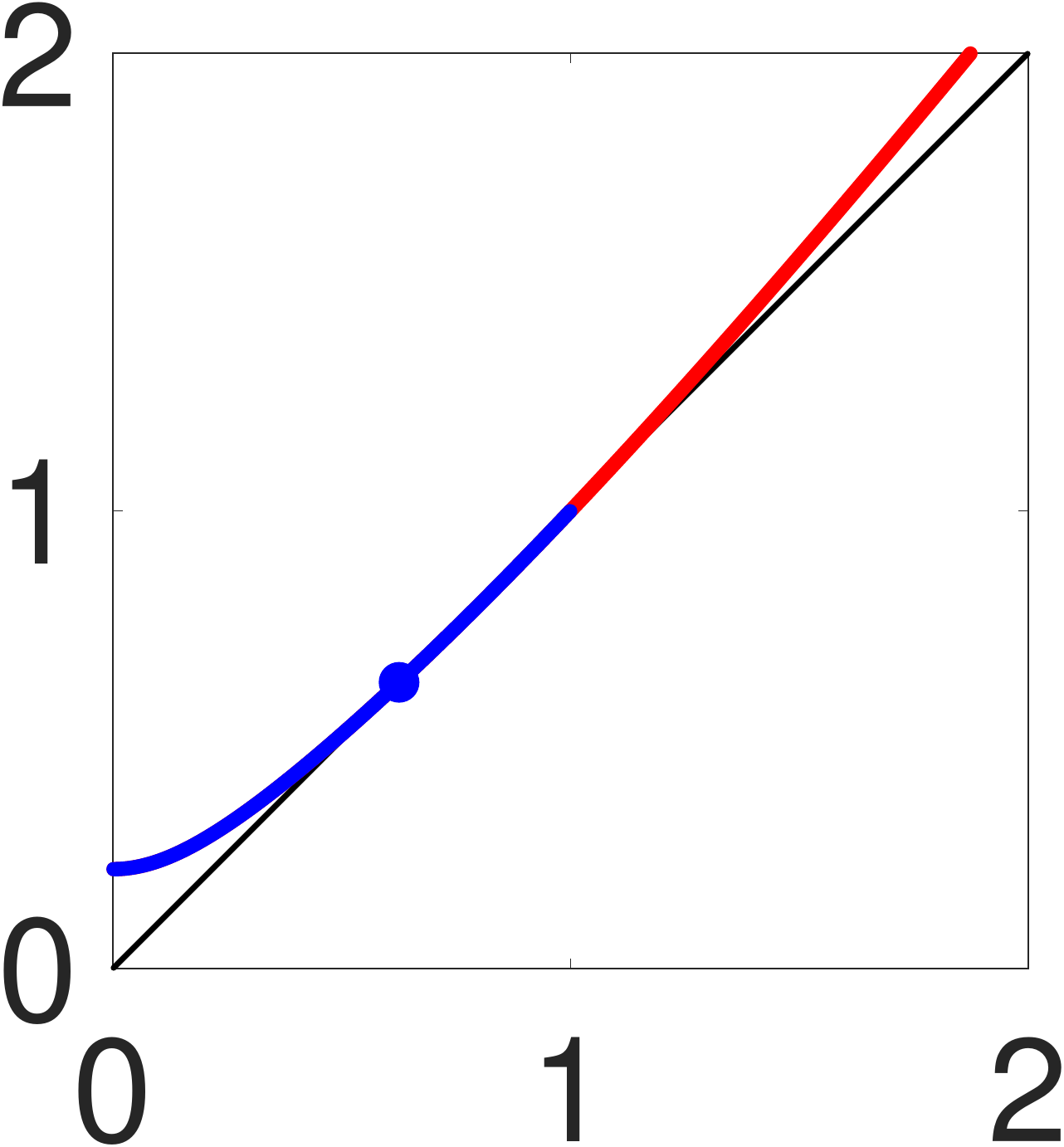}&\includegraphics[scale=0.13,valign=c]{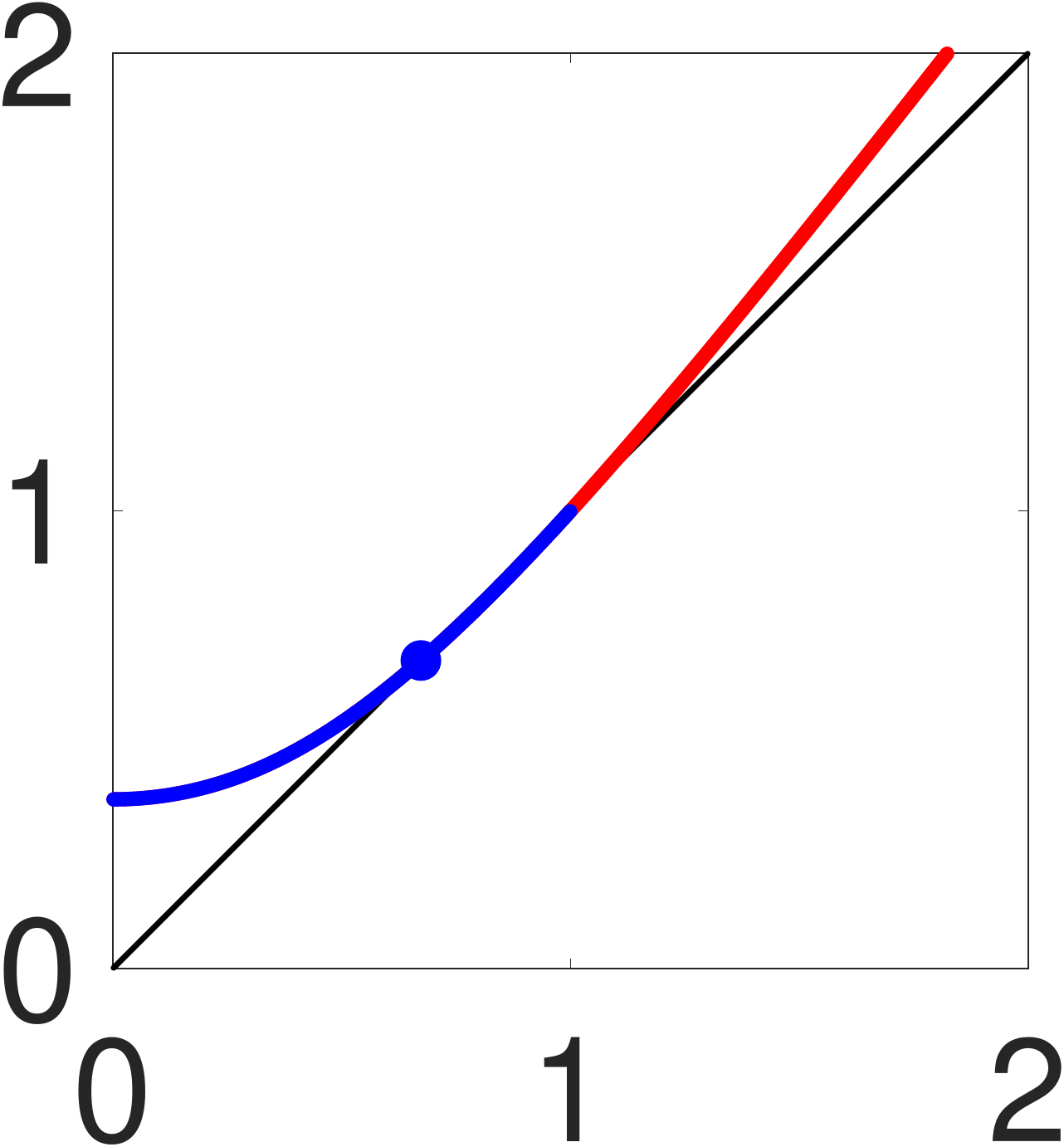}&\includegraphics[scale=0.13,valign=c]{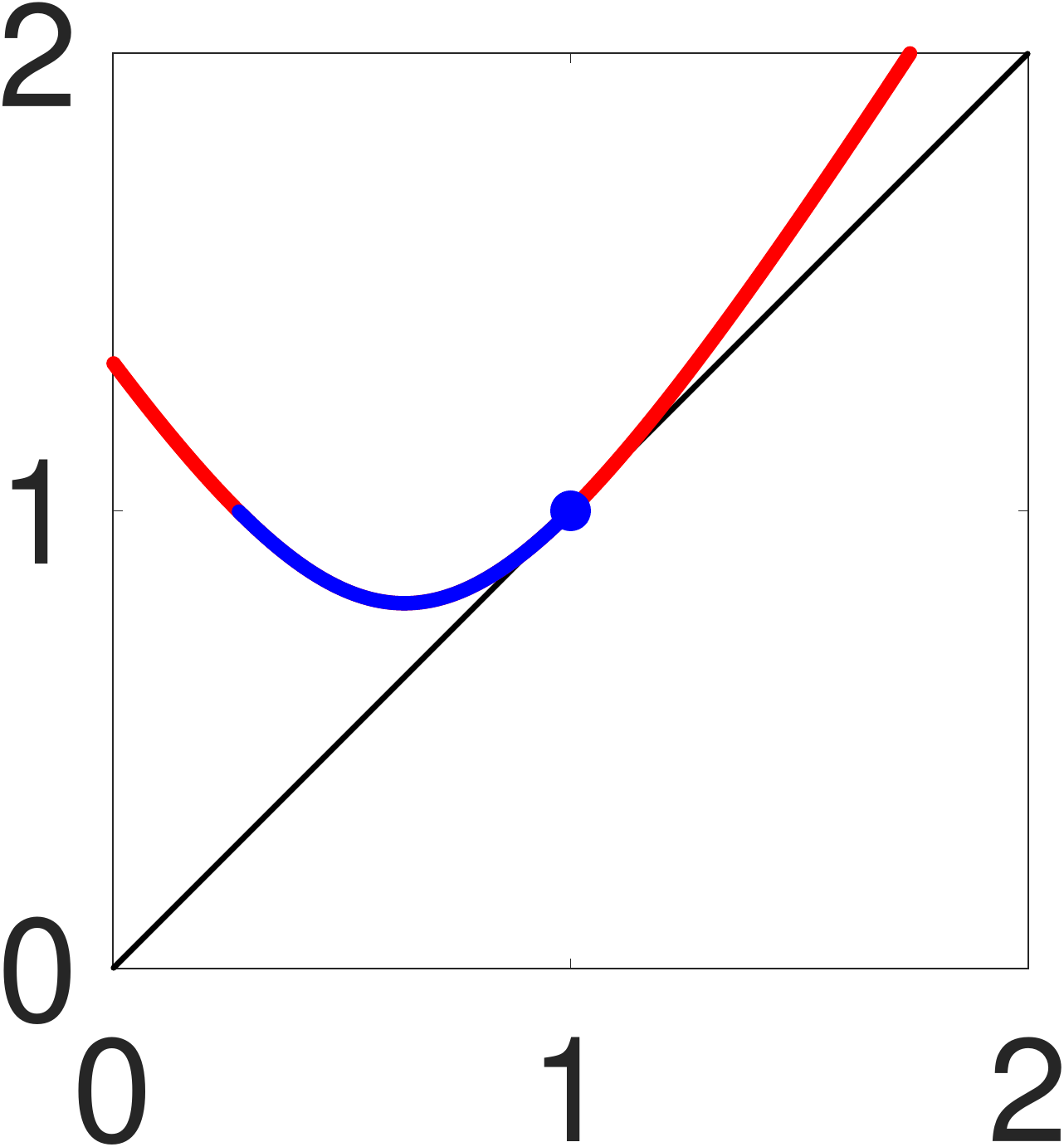}&\includegraphics[scale=0.13,valign=c]{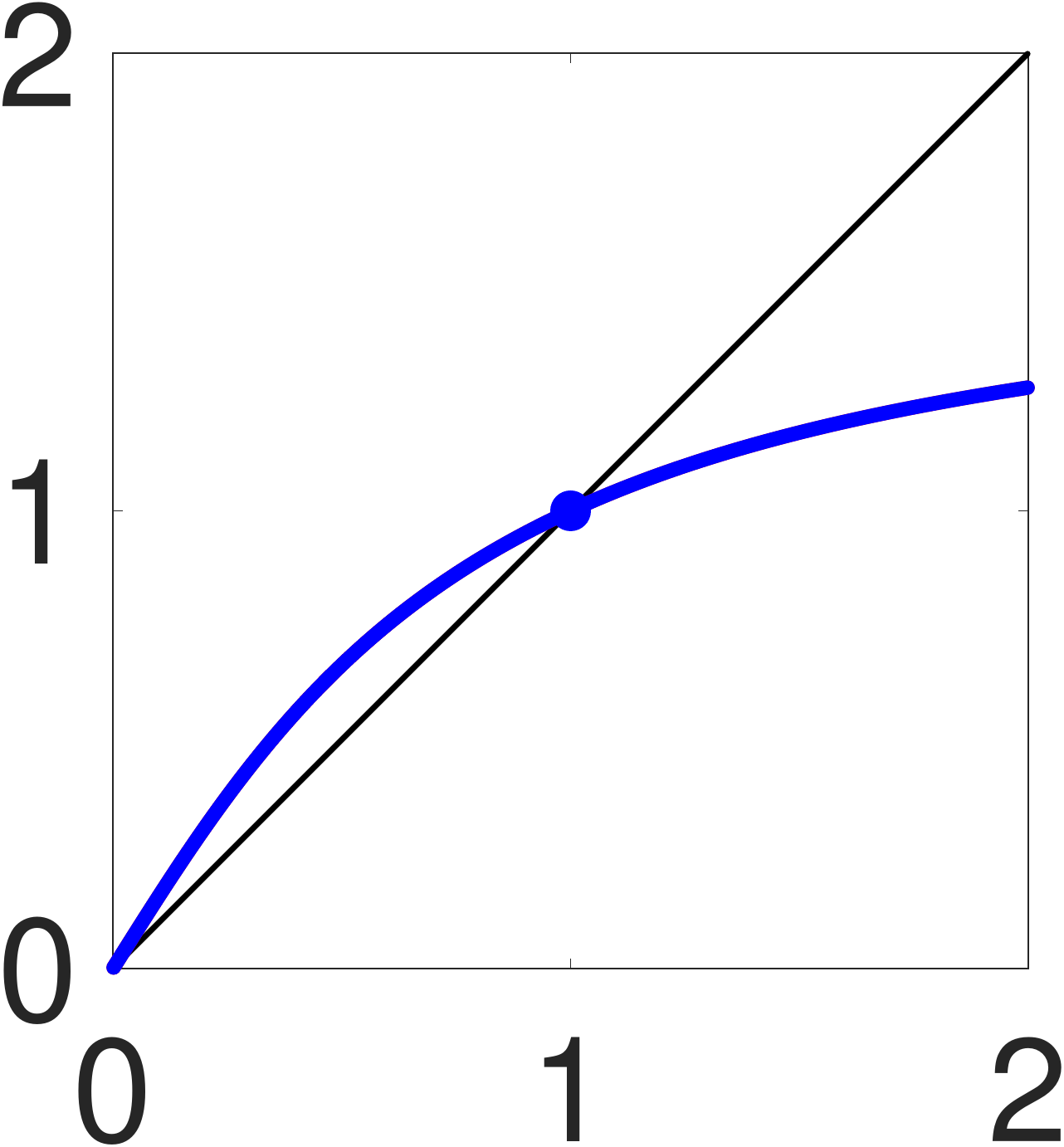}\\
Iter. limit &1&1&0.63&0.67&1&1\\
Conv. rate&sub-exp.&$O(0.78^M)$&$O(0.93^M)$&$O(0.87^M)$&sub-exp.&$O(0.46^M)$\\
$\frac{\mathfrak{L}_\tau'(\lambda)\lambda}{\mathfrak{L}_\tau(\lambda)}$ &\includegraphics[scale=0.13,valign=c]{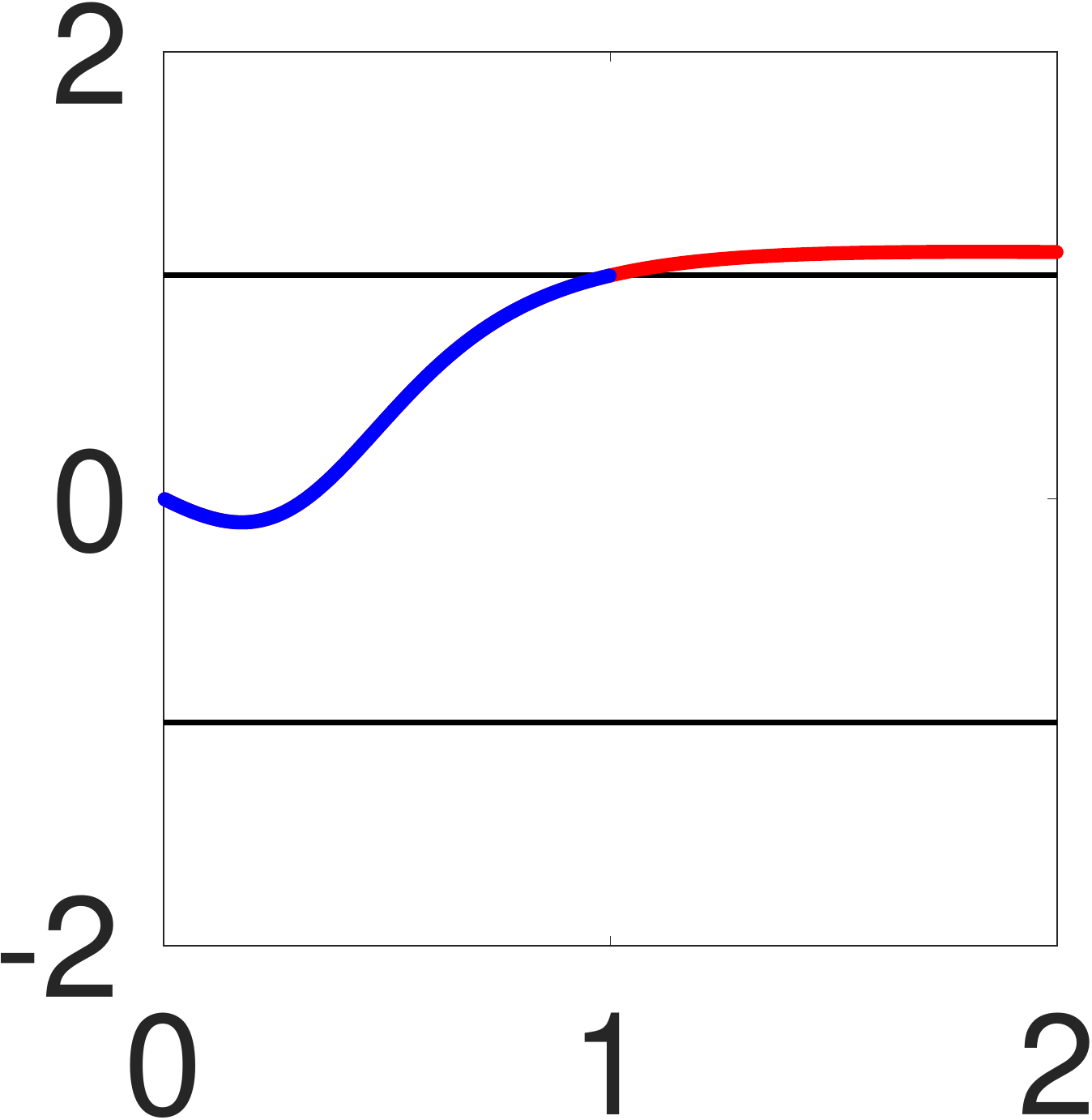}&\includegraphics[scale=0.13,valign=c]{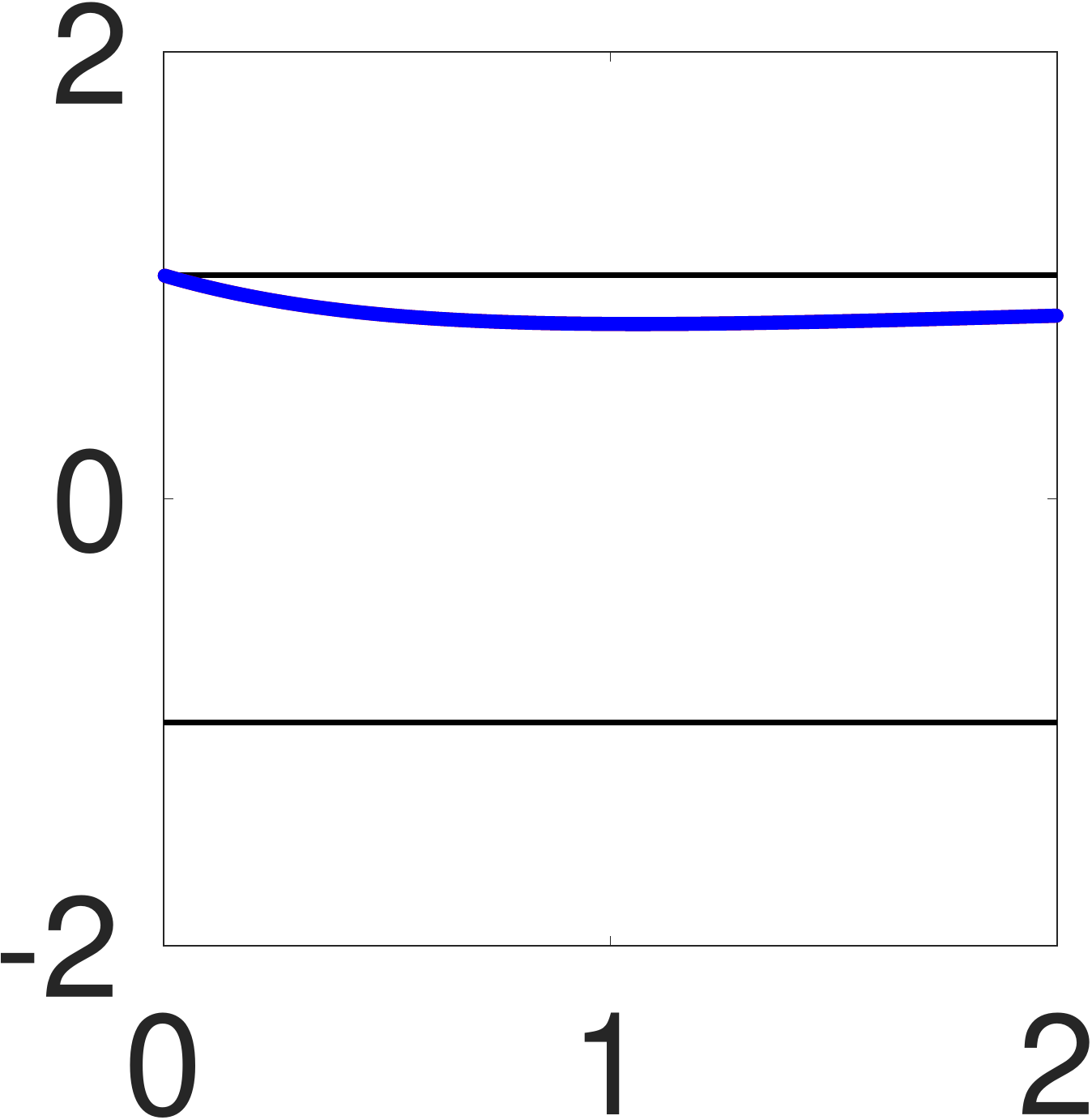}&\includegraphics[scale=0.13,valign=c]{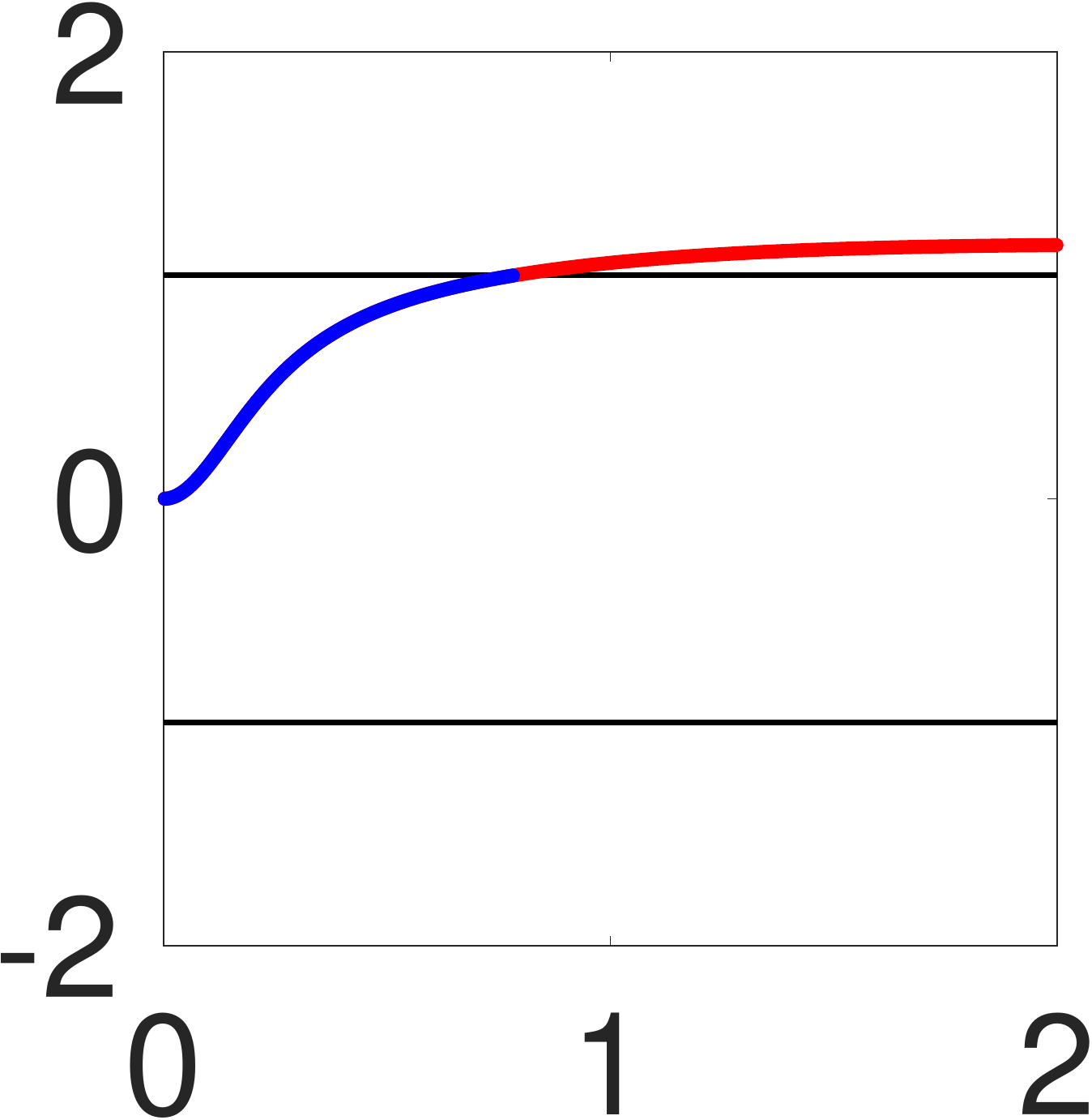}&\includegraphics[scale=0.13,valign=c]{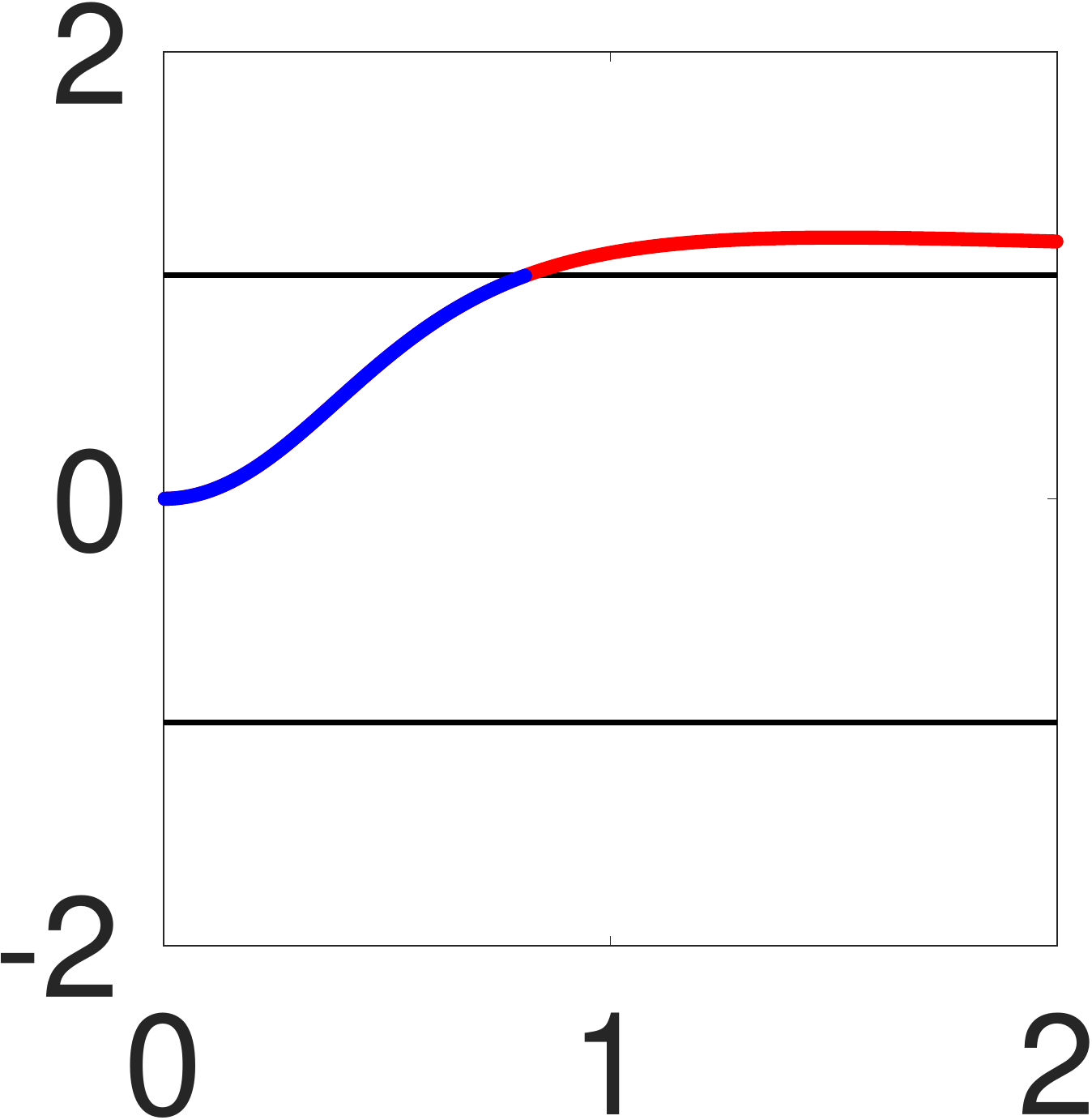}&\includegraphics[scale=0.13,valign=c]{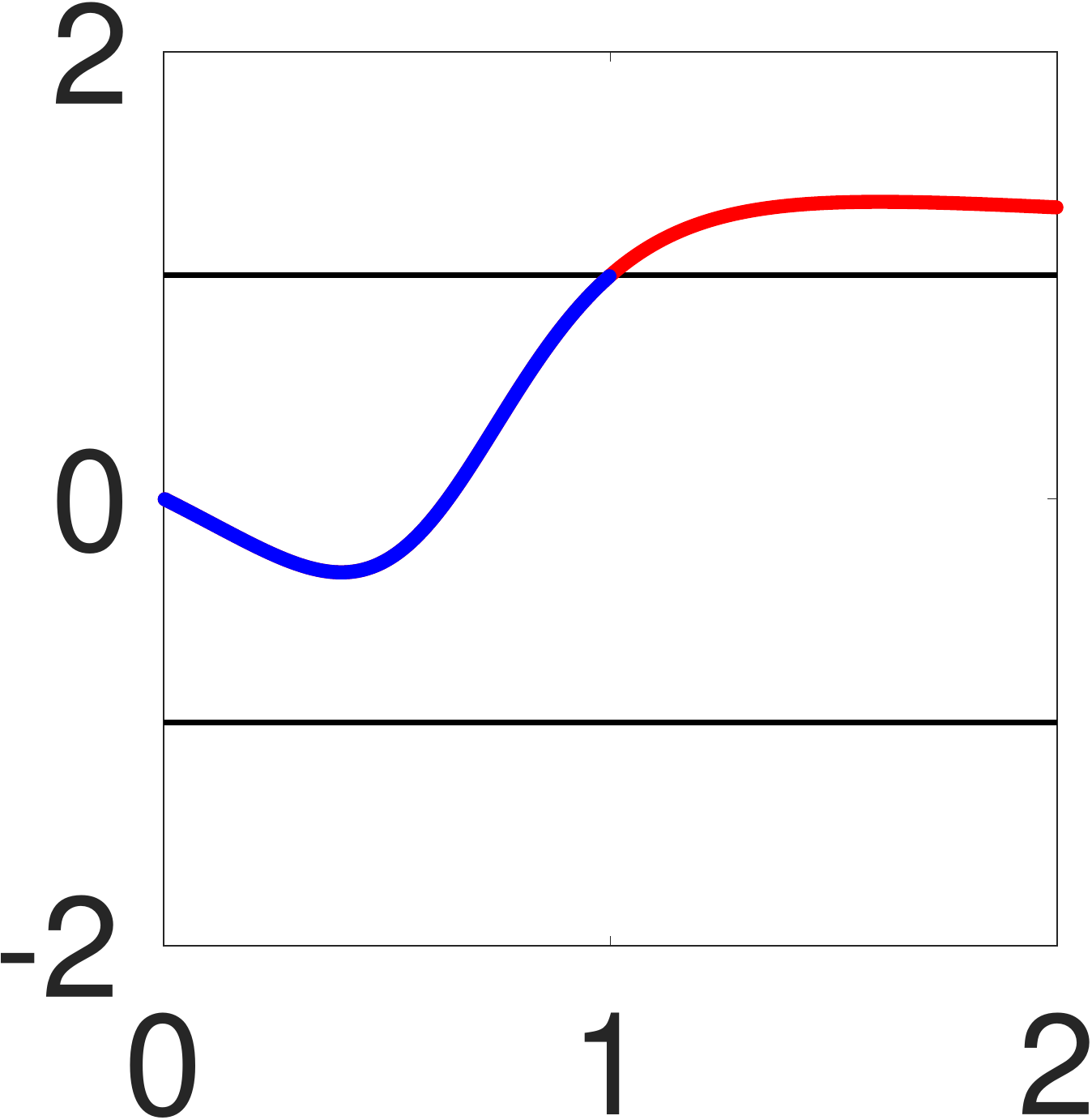}&\includegraphics[scale=0.13,valign=c]{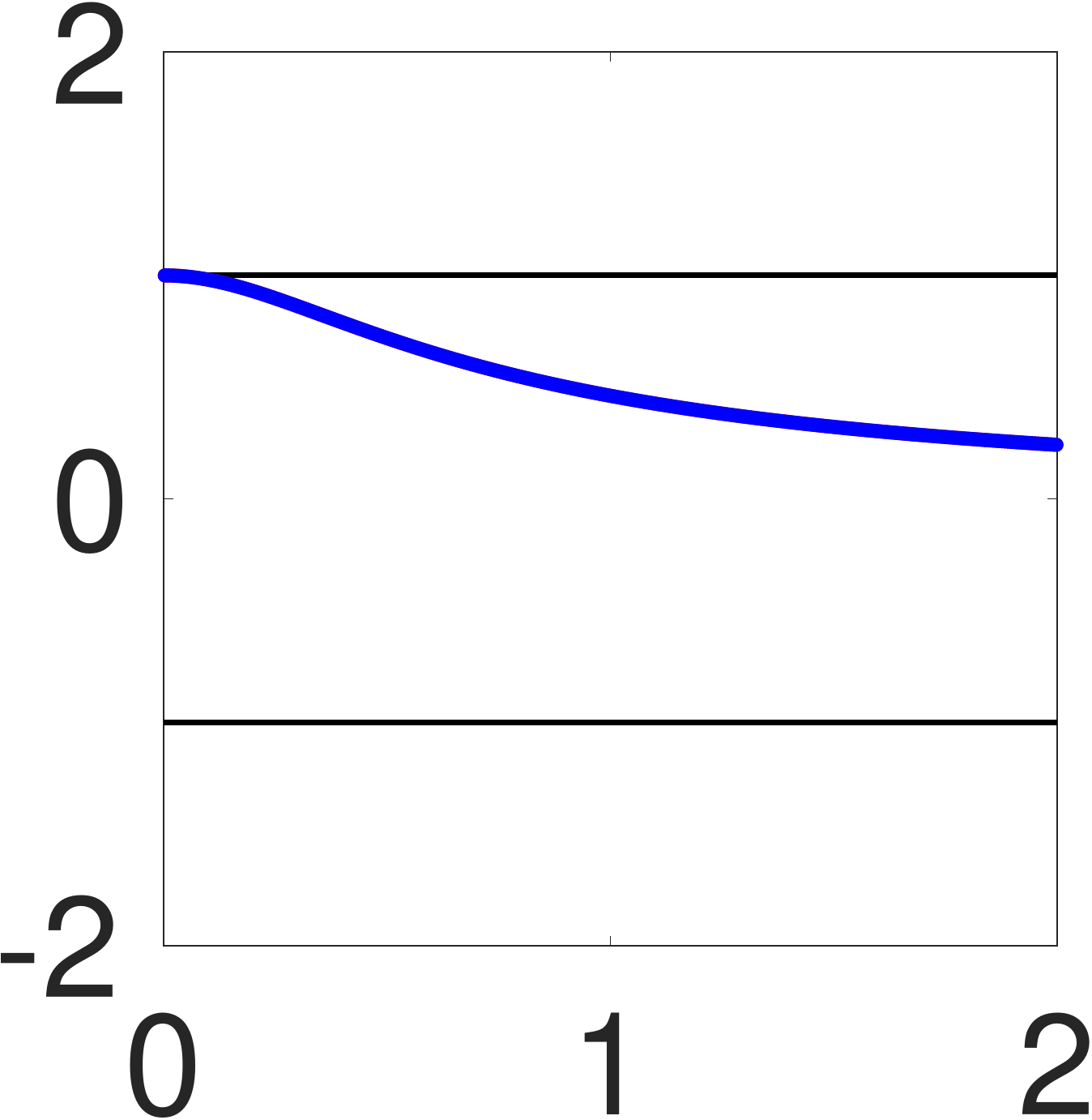}\\
$\tilde{\mathfrak{C}}_\tau(c)$&\includegraphics[scale=0.13,valign=c]{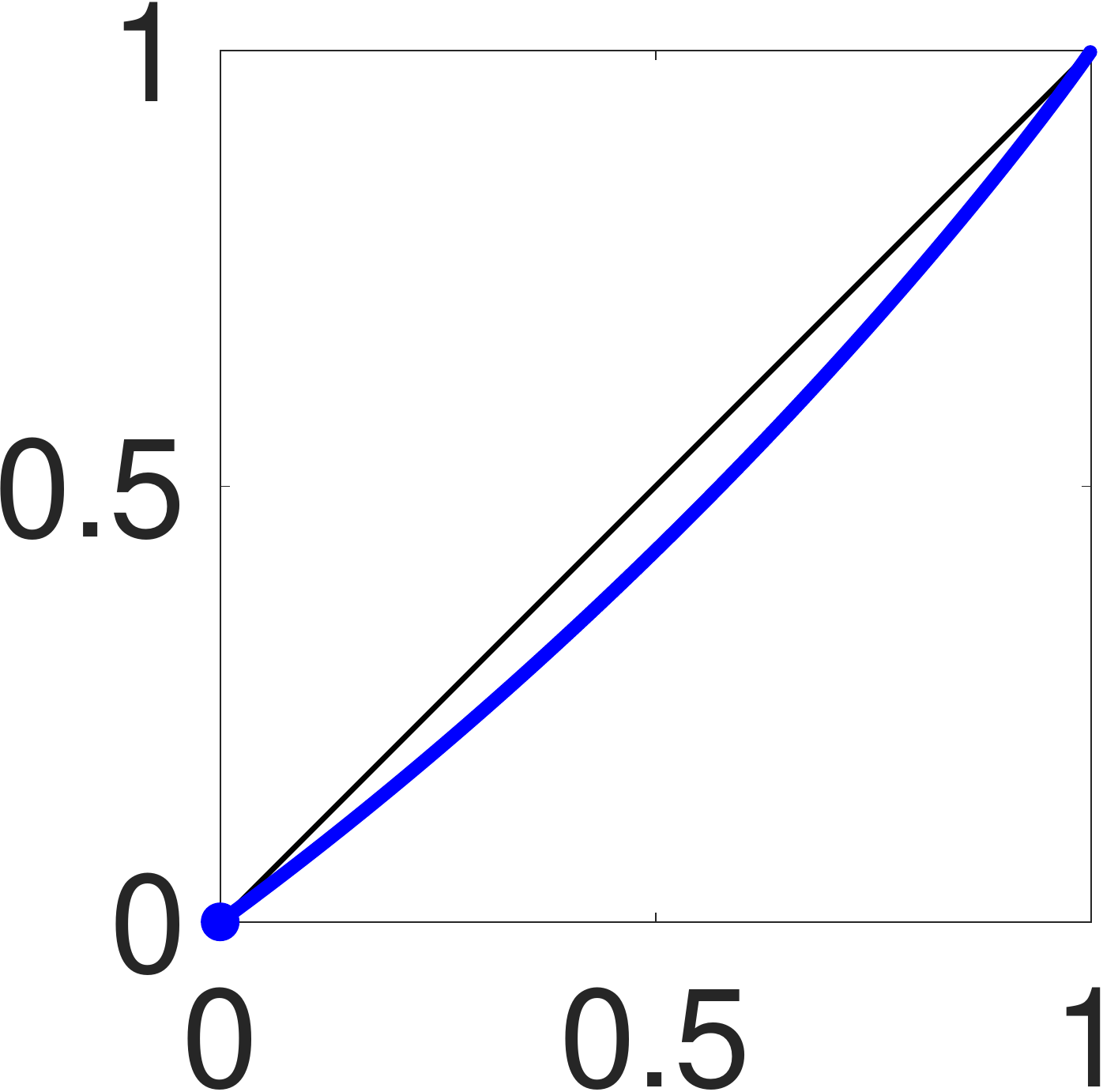}&\includegraphics[scale=0.13,valign=c]{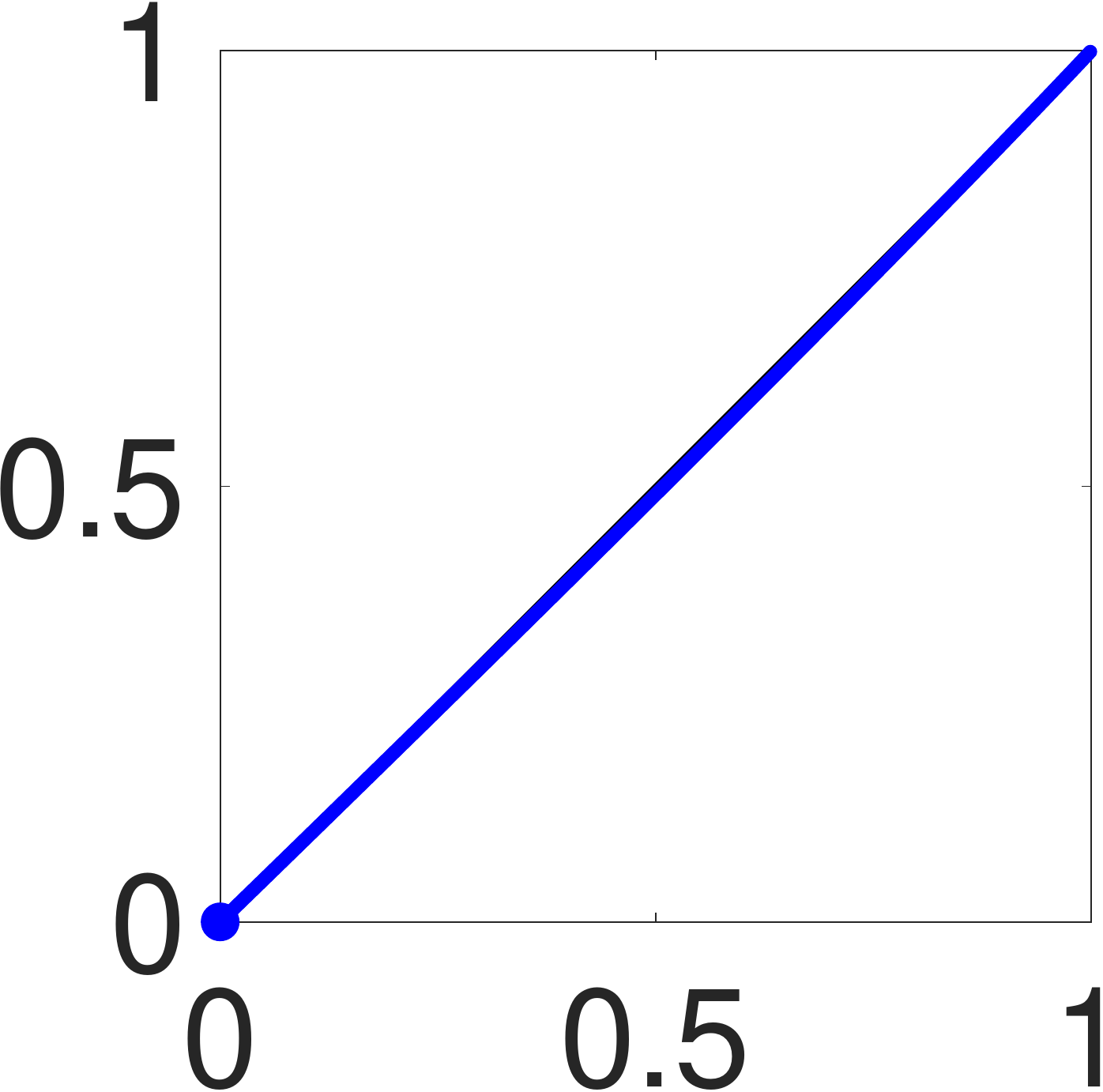}&\includegraphics[scale=0.13,valign=c]{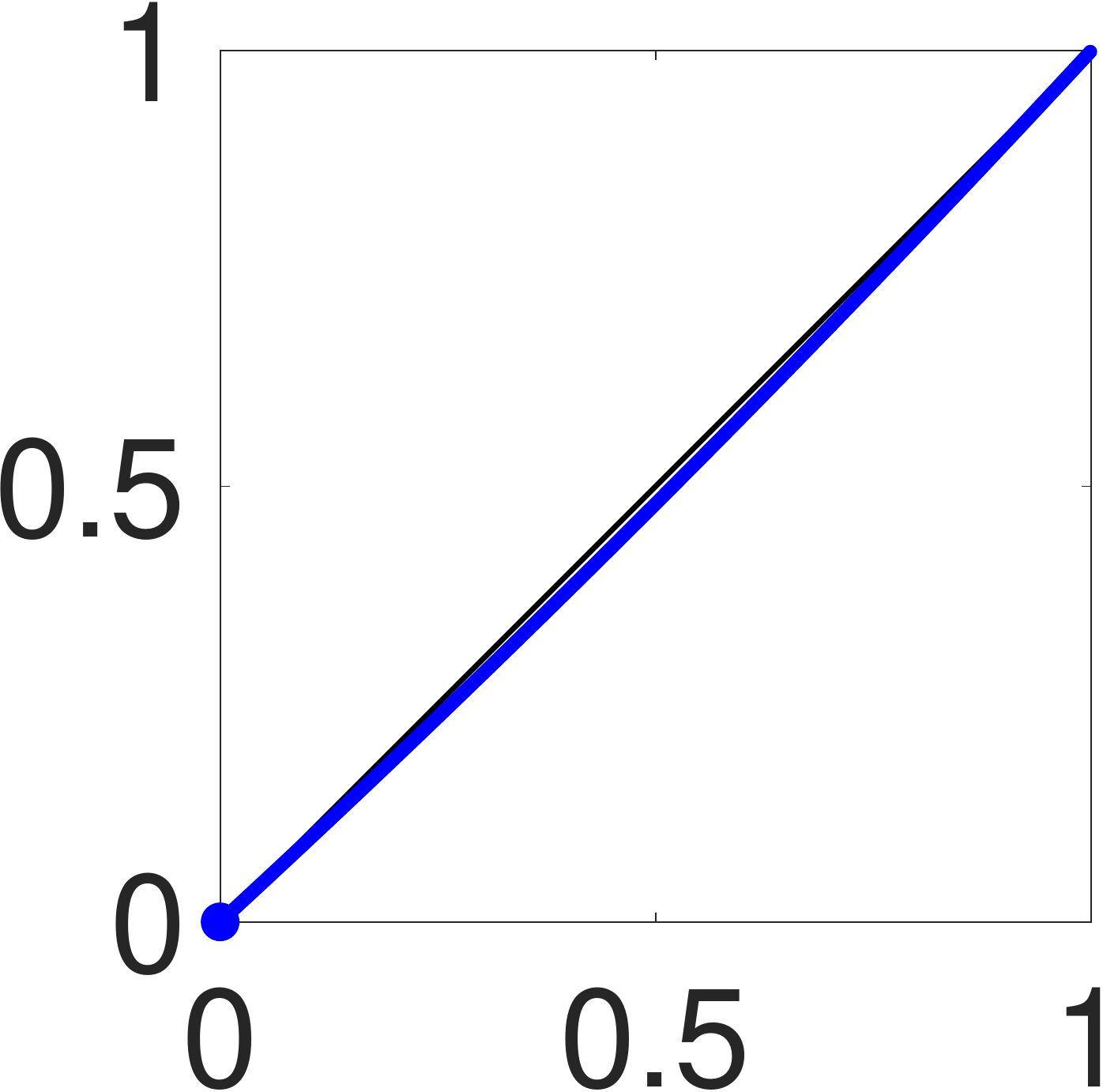}&\includegraphics[scale=0.13,valign=c]{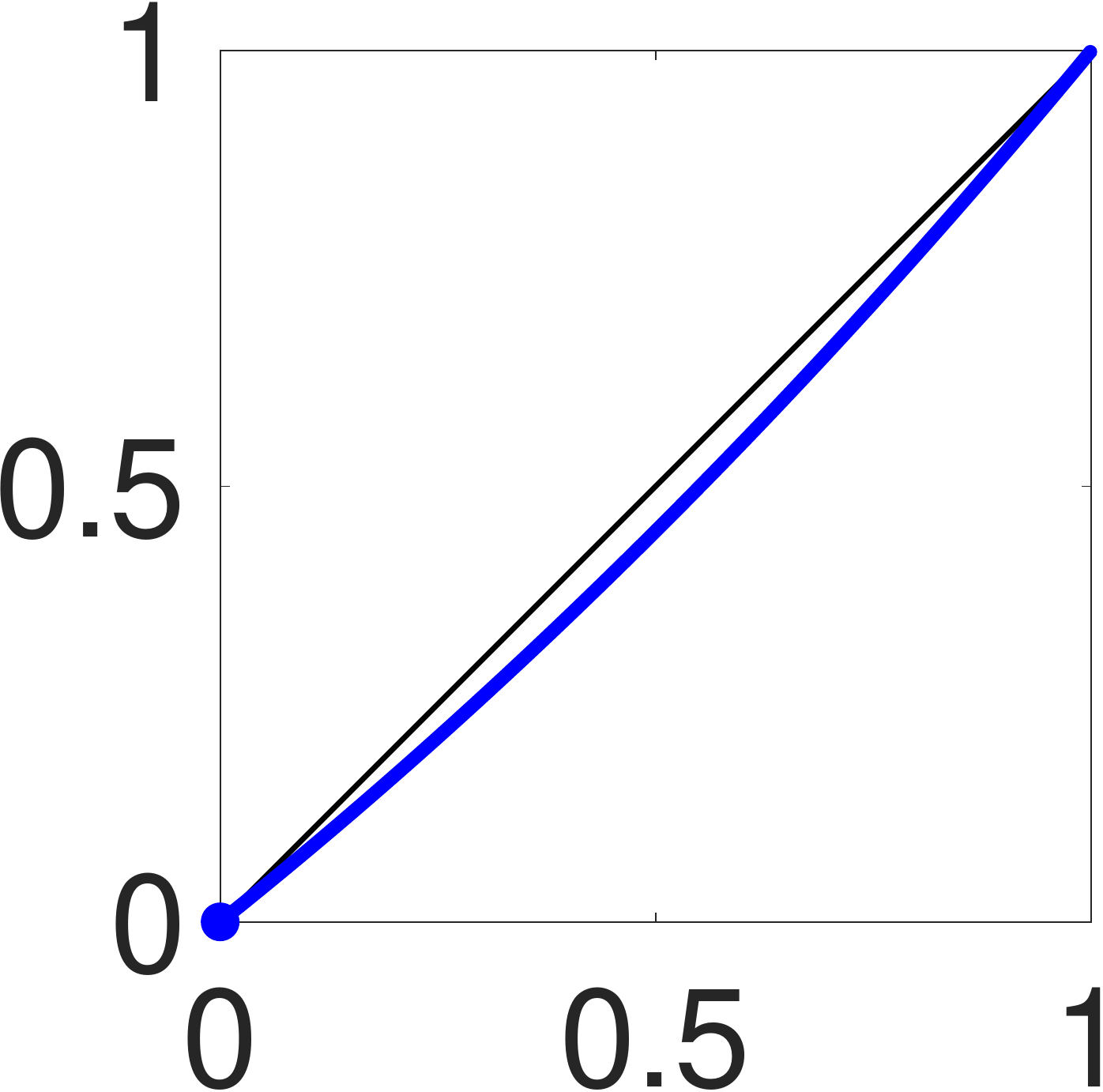}&\includegraphics[scale=0.13,valign=c]{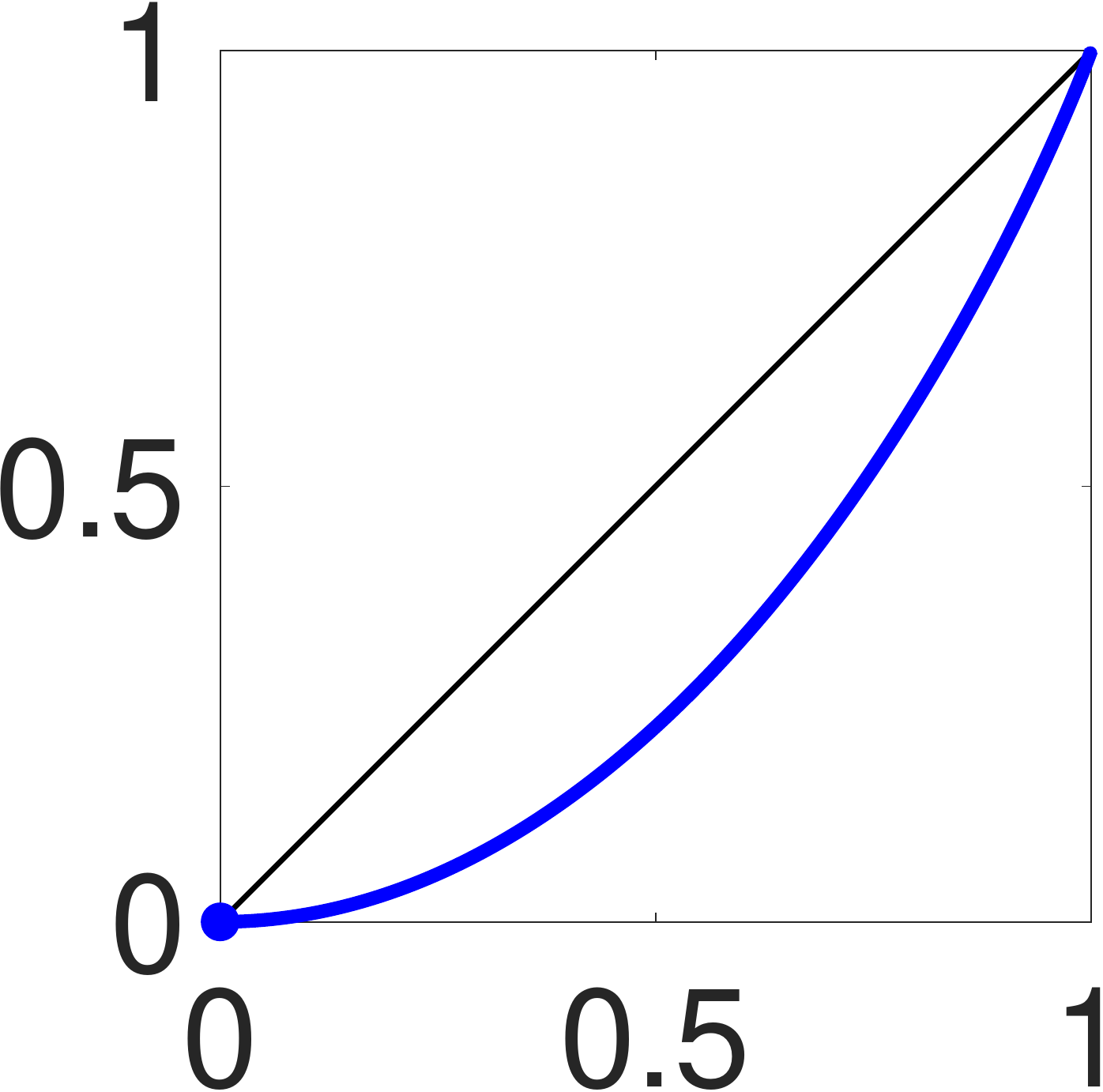}&\includegraphics[scale=0.13,valign=c]{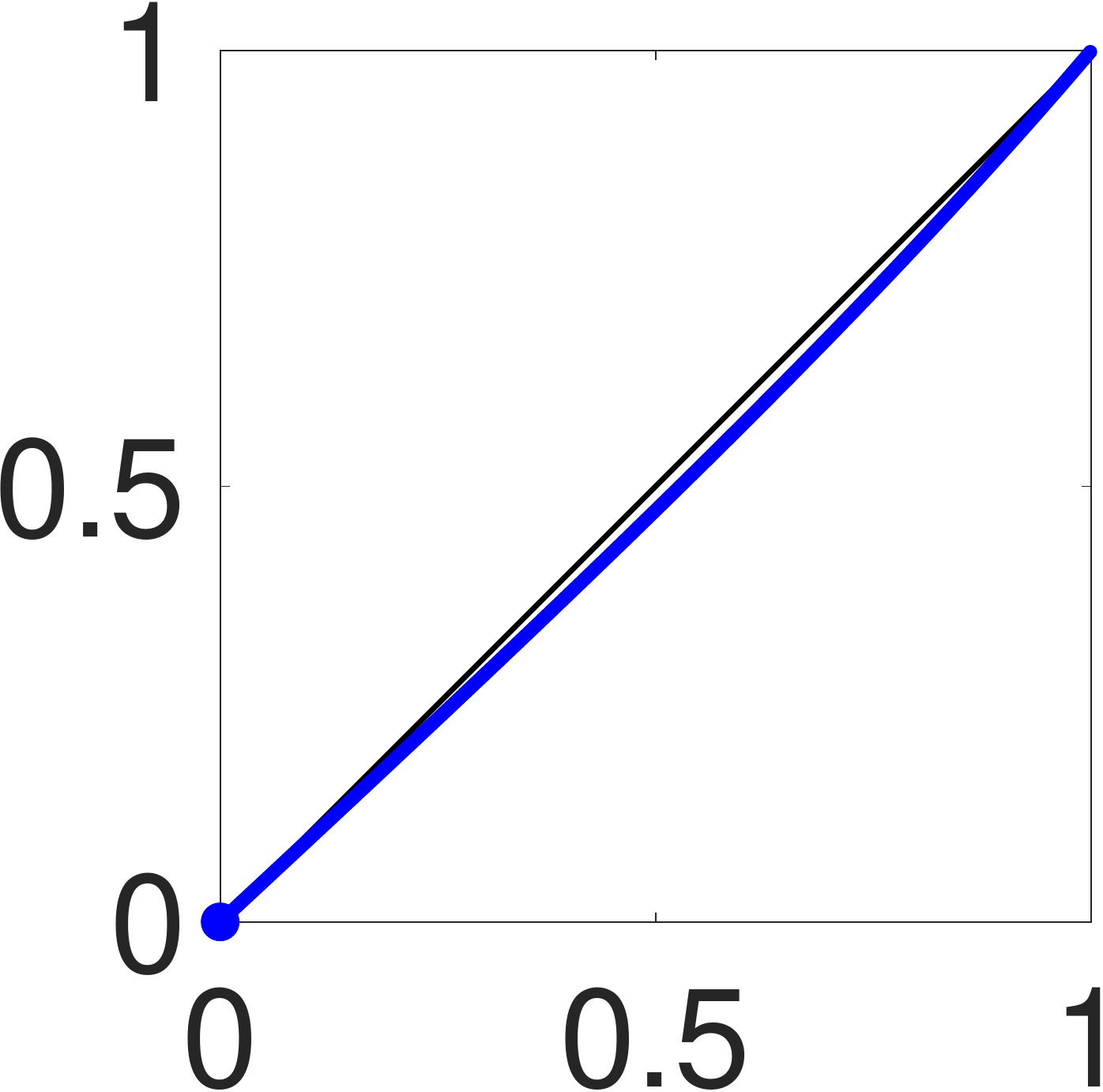}\\
Iter. limit &0&0&0&0&0&0\\
Conv. rate&$O(0.73^M)$&$O(0.97^M)$&$O(0.92^M)$&$O(0.80^M)$&super-exp.&$O(0.93^M)$\\
$\mathfrak{n}_\tau(\lambda^2, 0)$&\includegraphics[scale=0.13,valign=c]{graphsFinal/covCurveillu/nlc_relu.pdf}&\includegraphics[scale=0.13,valign=c]{graphsFinal/covCurveillu/nlc_selu.pdf}&\includegraphics[scale=0.13,valign=c]{graphsFinal/covCurveillu/nlc_softplus.pdf}&\includegraphics[scale=0.13,valign=c]{graphsFinal/covCurveillu/nlc_swish.pdf}&\includegraphics[scale=0.13,valign=c]{graphsFinal/covCurveillu/nlc_abs.pdf}&\includegraphics[scale=0.13,valign=c]{graphsFinal/covCurveillu/nlc_tanh.pdf}\\
$\lambda \rightarrow \infty$&1.21&1.21&1.21&1.21&1.66&$O(\sqrt{\lambda})$\\
\\
\end{tabular}
}
\caption{(This table is a continuation of table \ref{covCurveillu1}.)}
\label{covCurveillu3}
\end{table}

\begin{table}[H]
{
\centering \small
\begin{tabular}{lcccccc}
Act. fun. &sigm.-deb.&even t.-deb.&Gauss.-deb.&odd sq.-deb.&sq.-deb.&sawtooth-deb.\\ \hline\hline
\\
$\tau(s)$&\includegraphics[scale=0.13,valign=c]{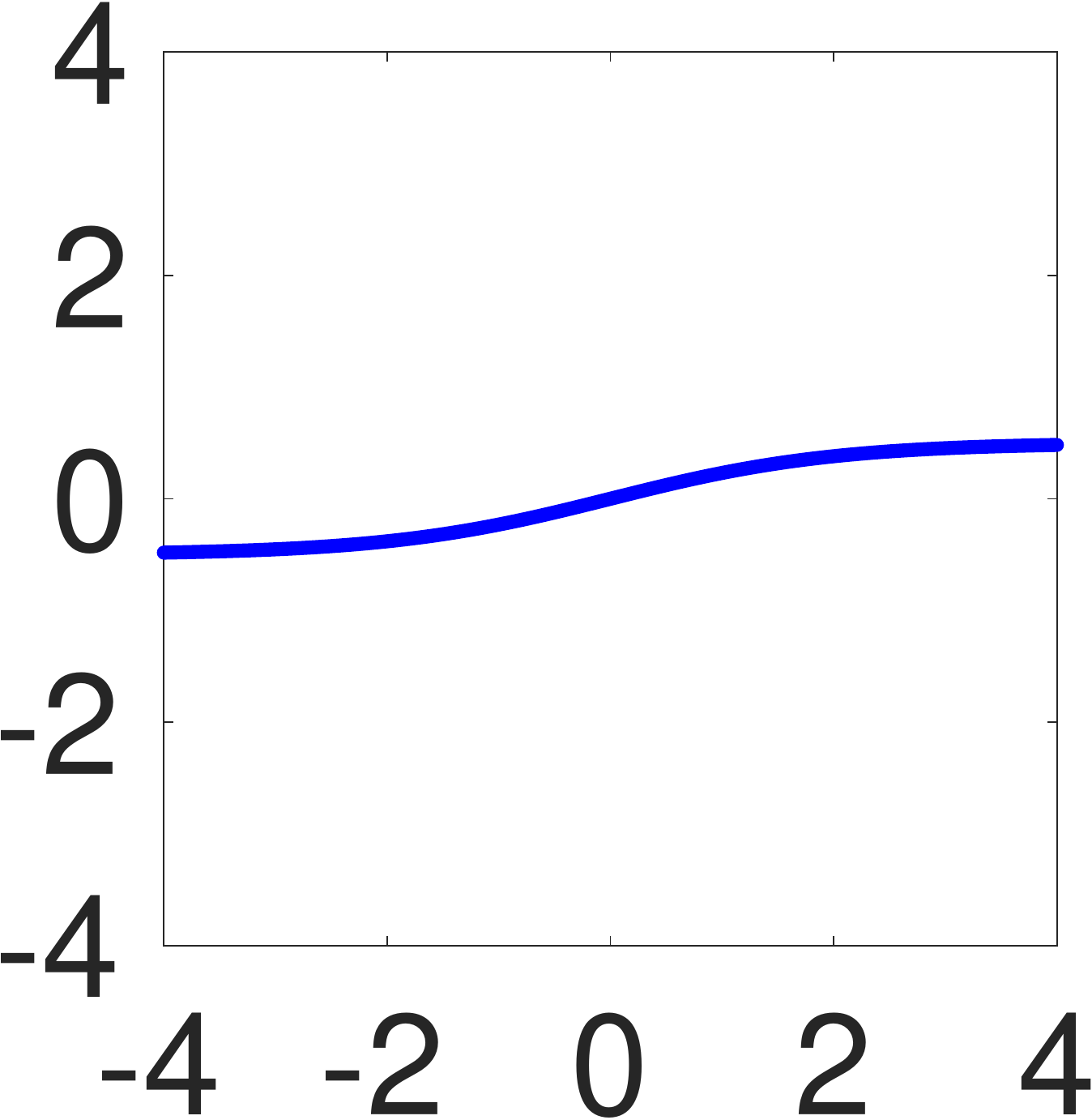}&\includegraphics[scale=0.13,valign=c]{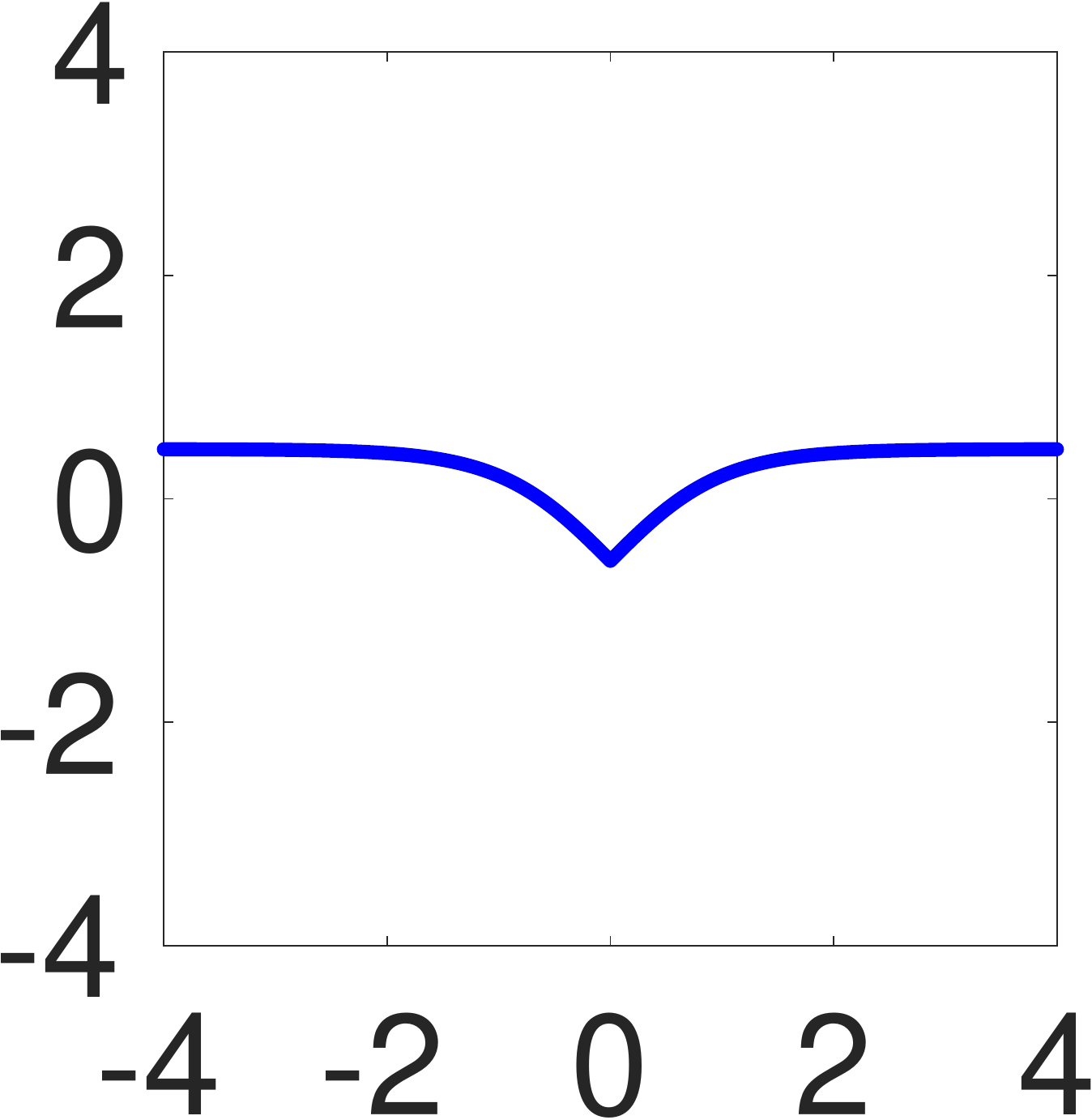}&\includegraphics[scale=0.13,valign=c]{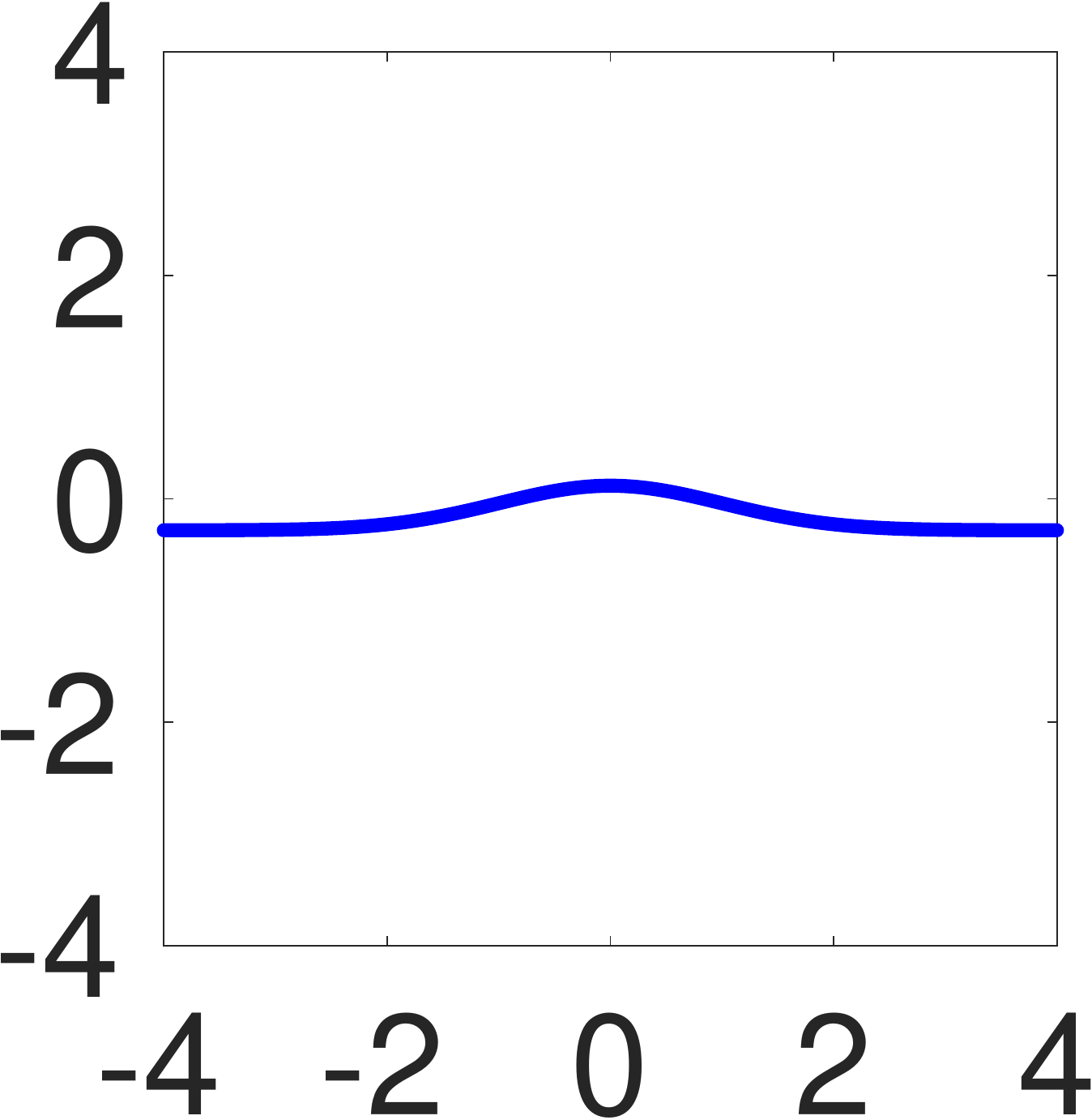}&\includegraphics[scale=0.13,valign=c]{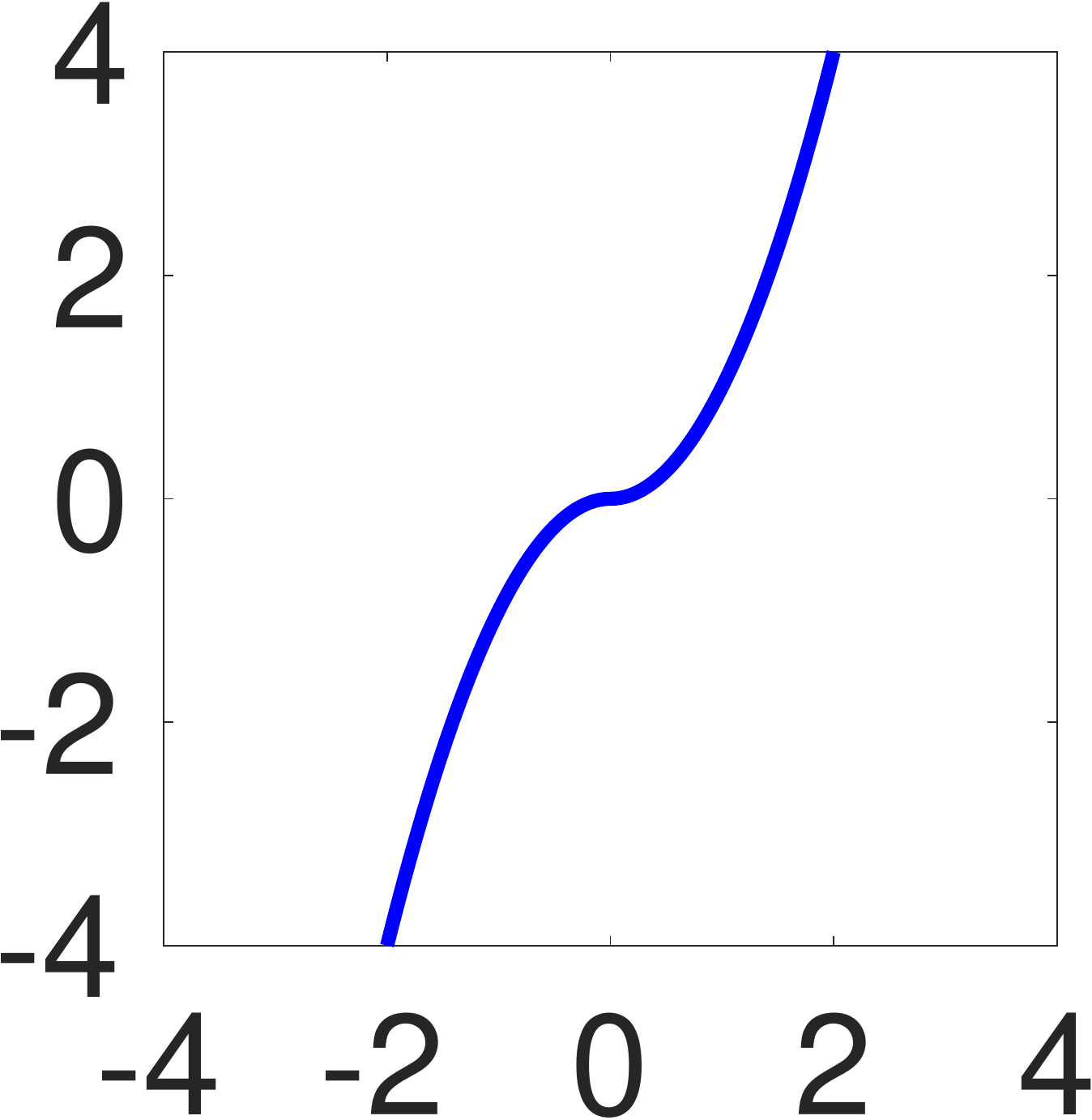}&\includegraphics[scale=0.13,valign=c]{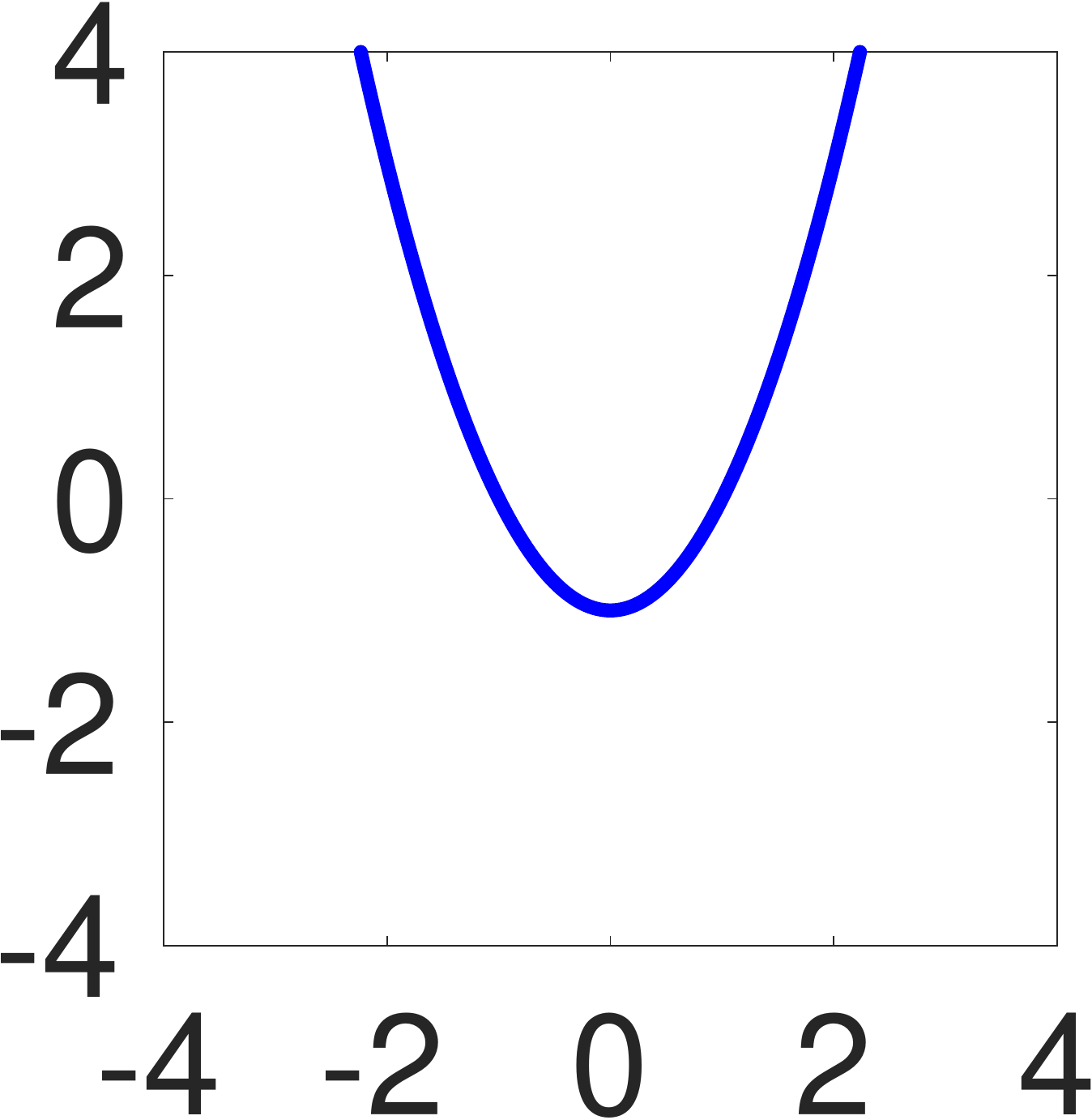}&\includegraphics[scale=0.13,valign=c]{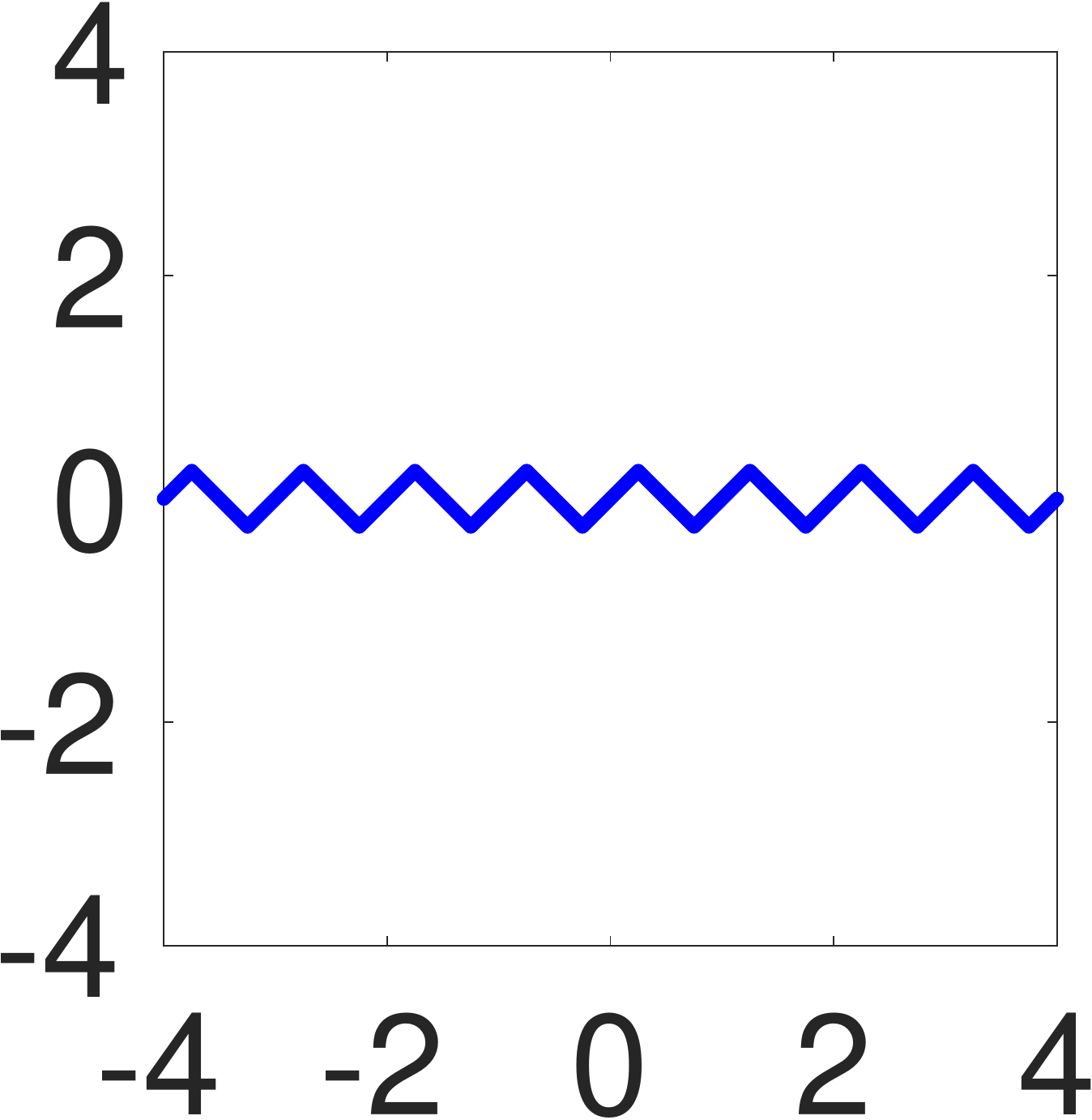}\\
$\mathfrak{L}_\tau(\lambda)$&\includegraphics[scale=0.13,valign=c]{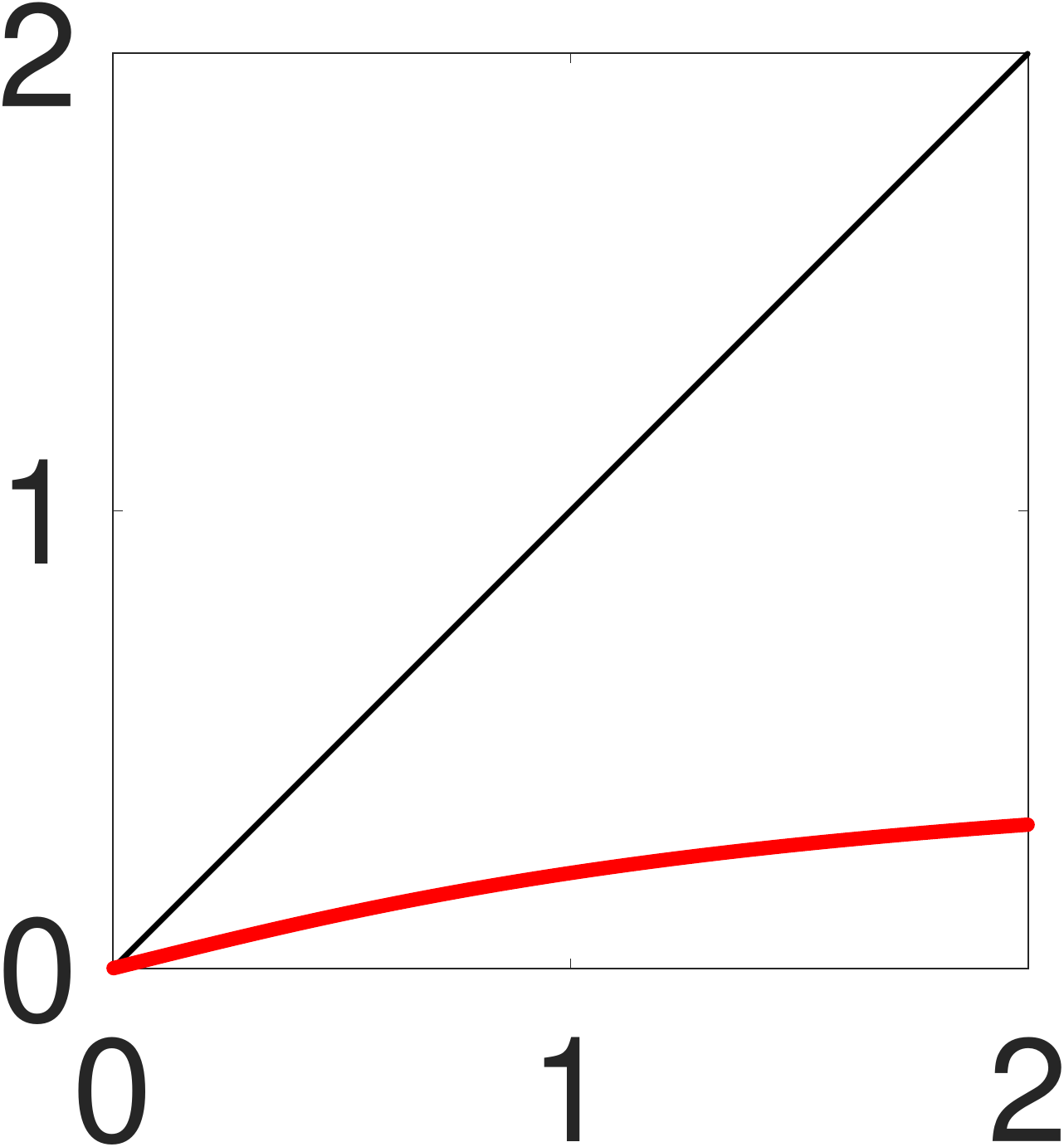}&\includegraphics[scale=0.13,valign=c]{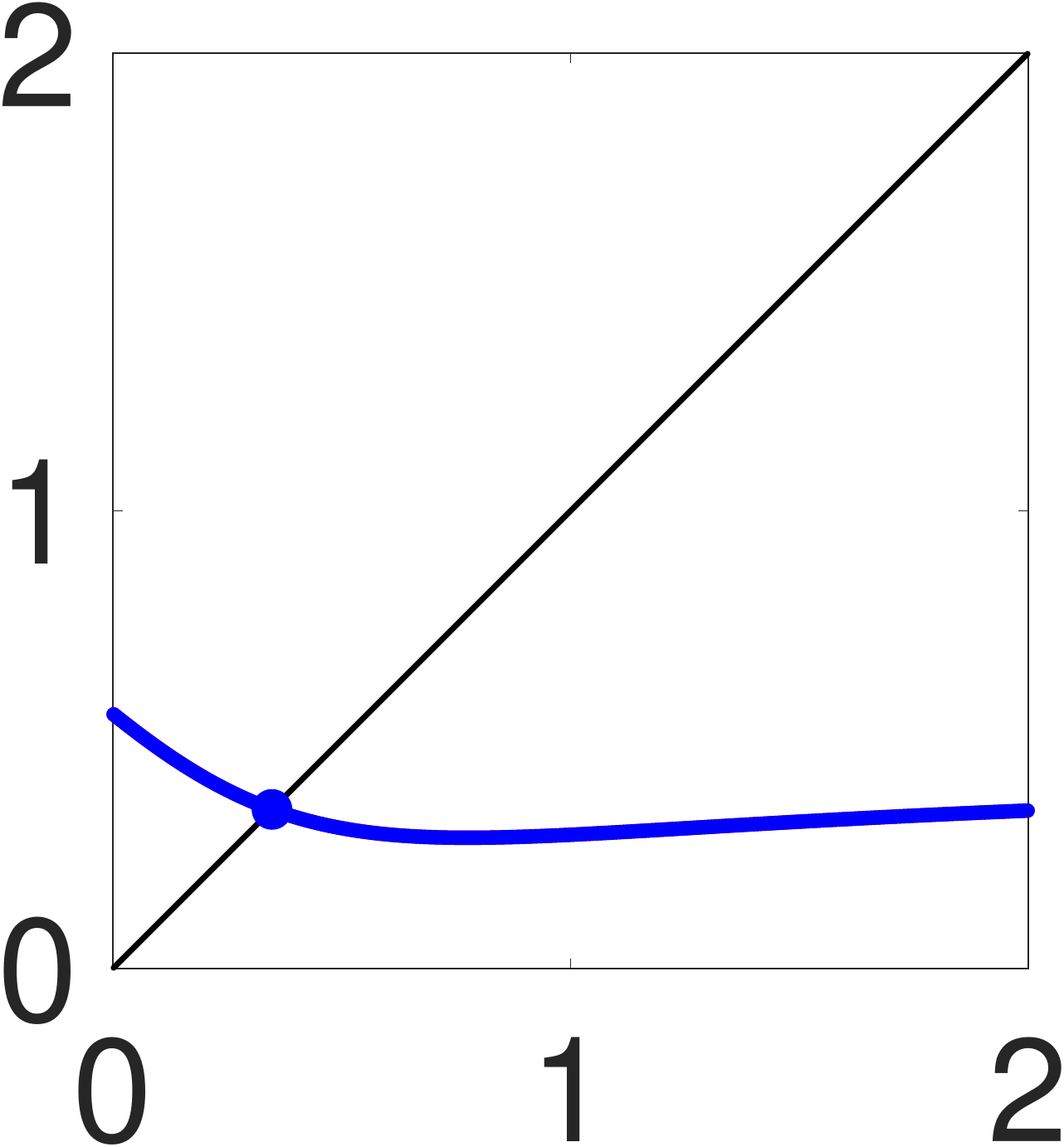}&\includegraphics[scale=0.13,valign=c]{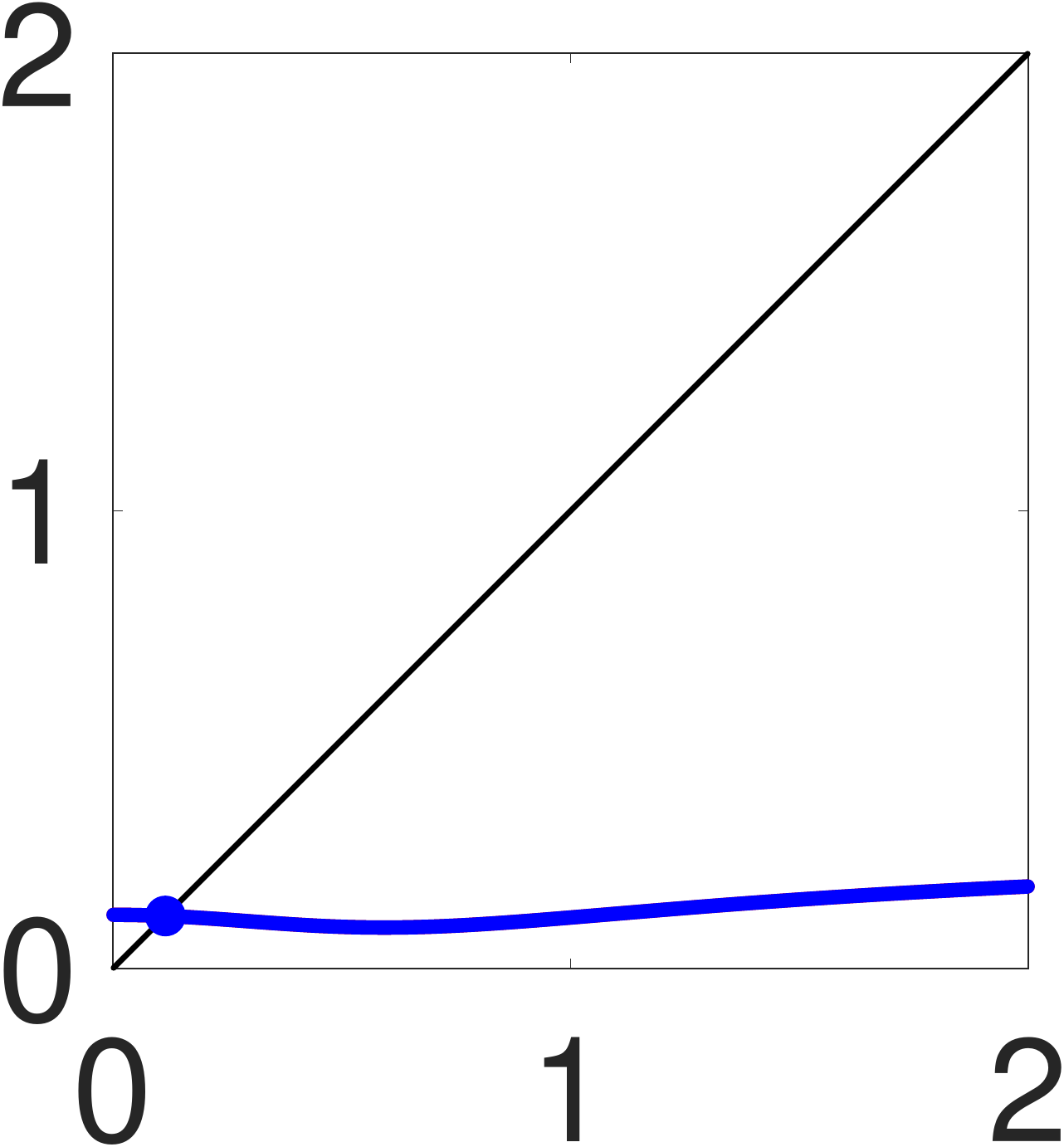}&\includegraphics[scale=0.13,valign=c]{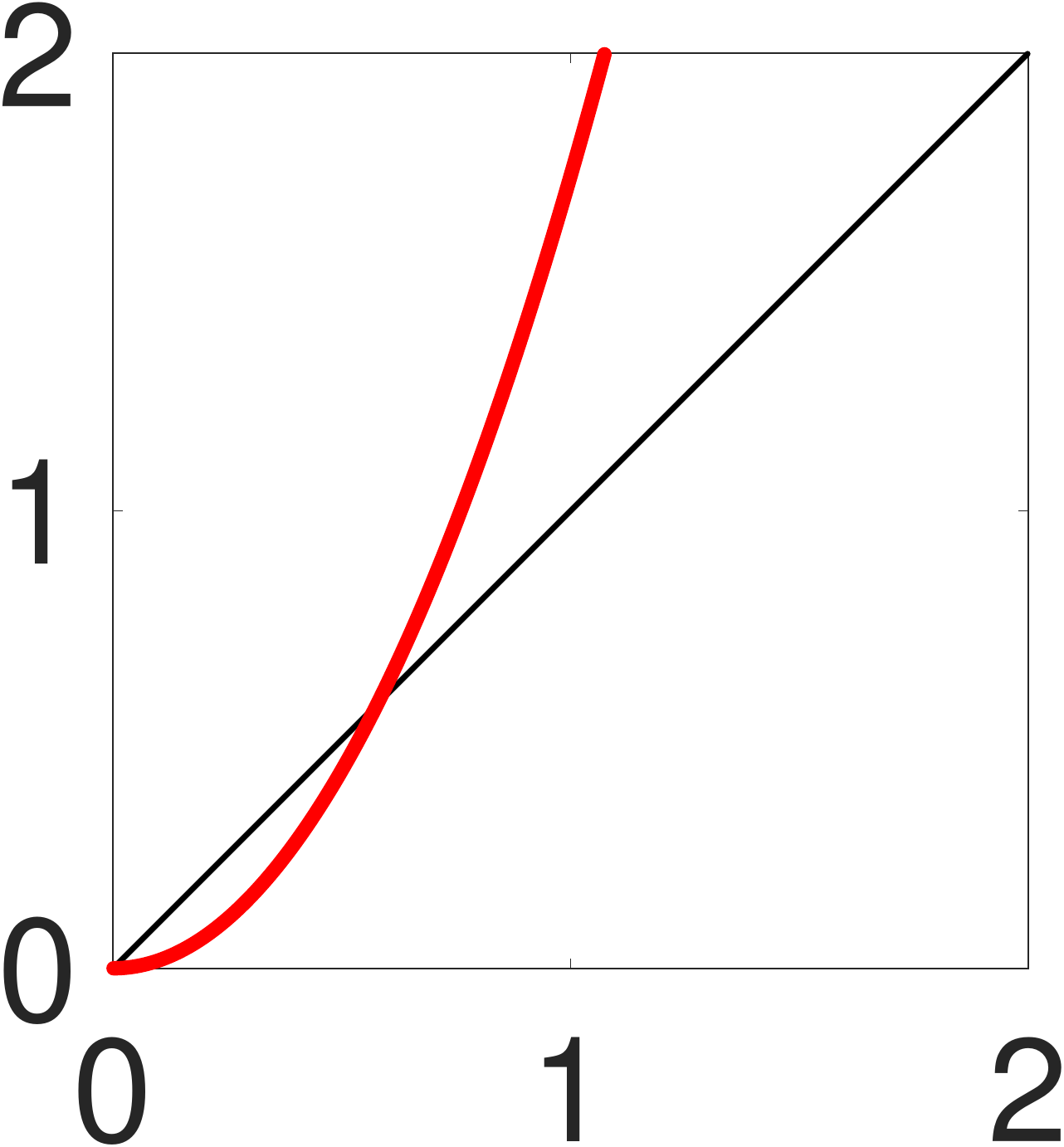}&\includegraphics[scale=0.13,valign=c]{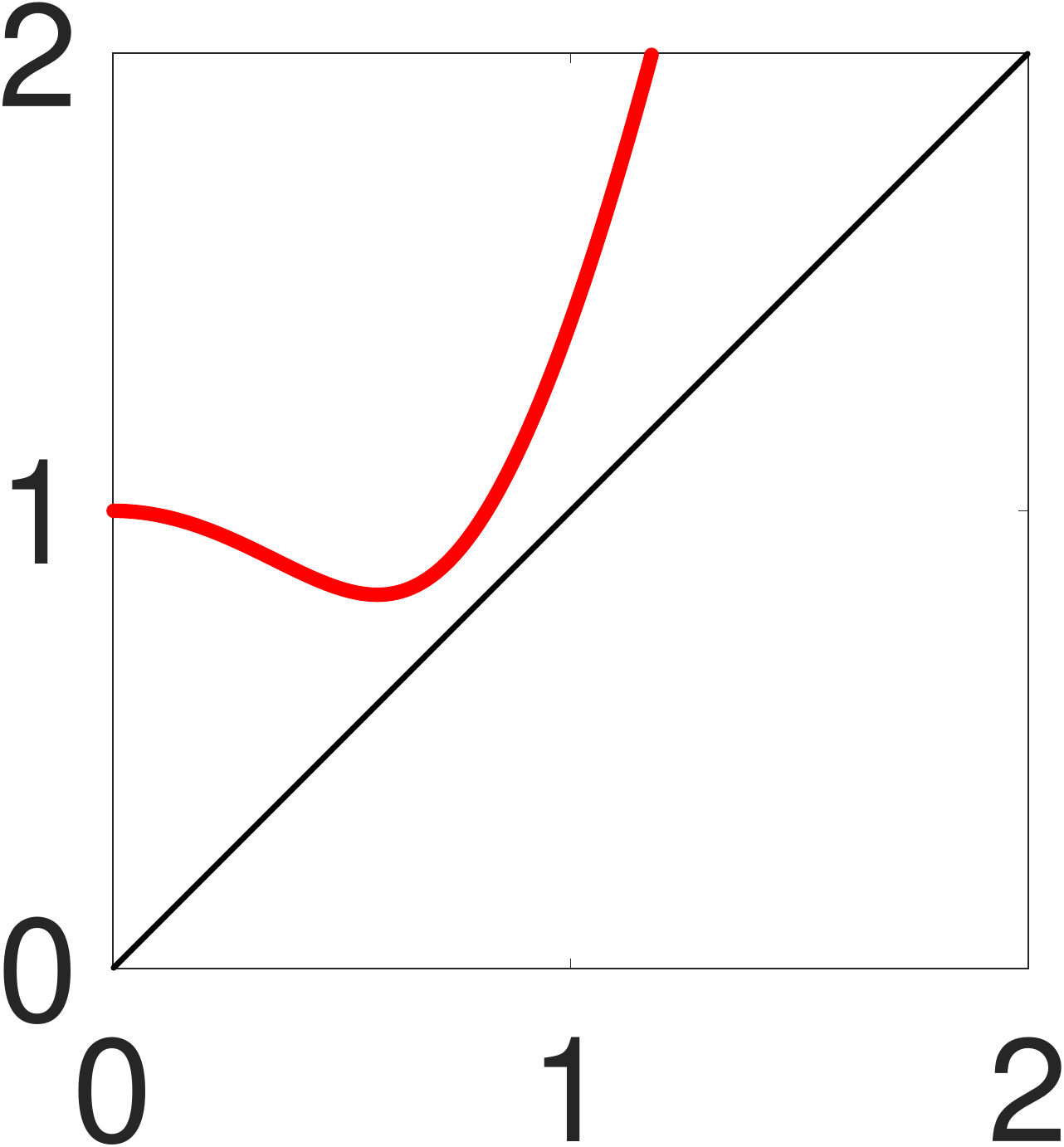}&\includegraphics[scale=0.13,valign=c]{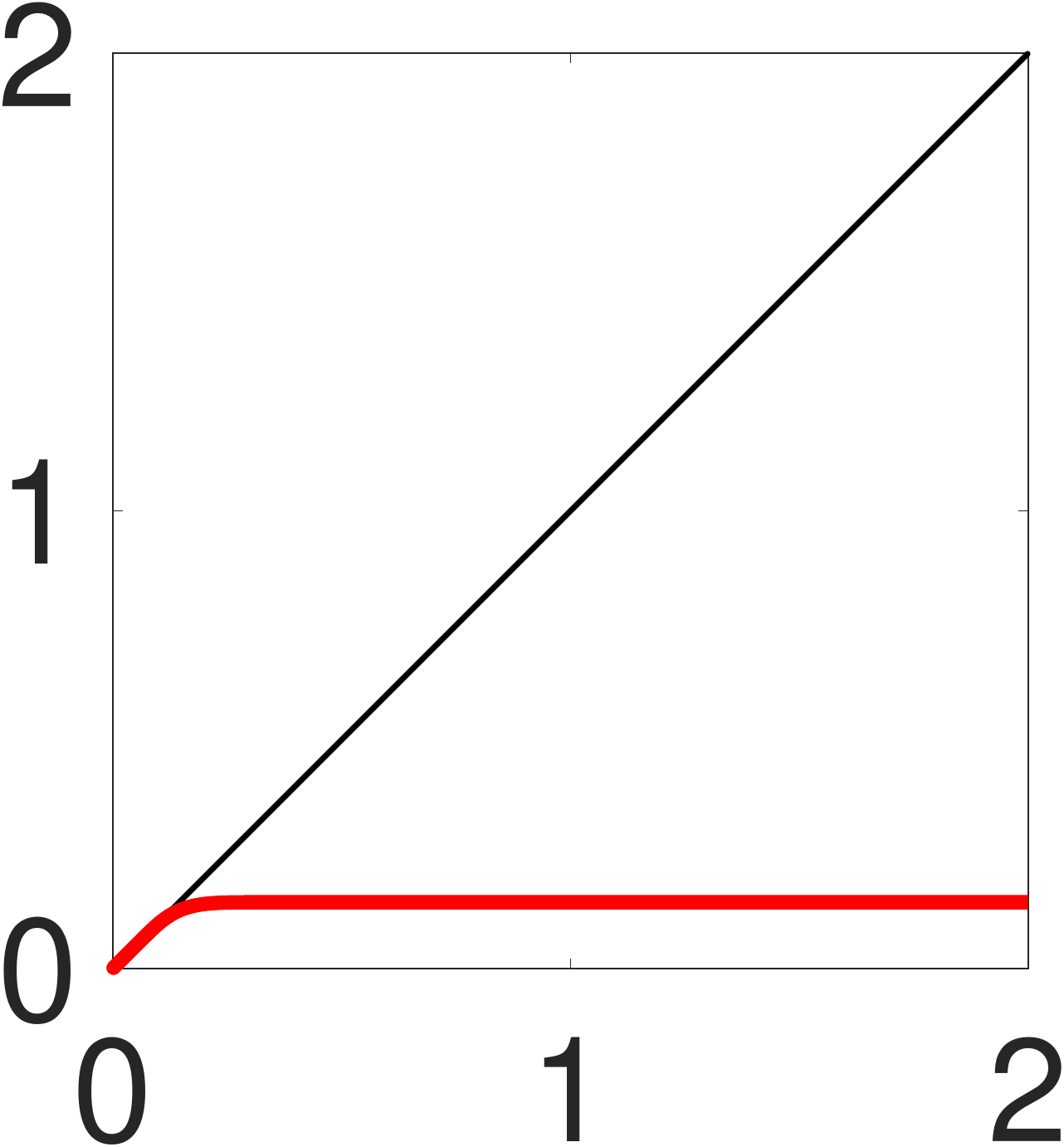}\\
Iter. limit &0&0.35&0.11&0&-&0\\
Conv. rate&$O(0.5^M)$&$O(0.36^M)$&$O(0.04^M)$&$\frac{1}{\sqrt{3}}(\sqrt{3}\lambda)^{2^M}$&-&$\Omega(\frac{1}{M^\epsilon}) \forall \epsilon > 0$\\
$\frac{\mathfrak{L}_\tau(\lambda)}{\mathfrak{L}_\tau(1)}$&\includegraphics[scale=0.13,valign=c]{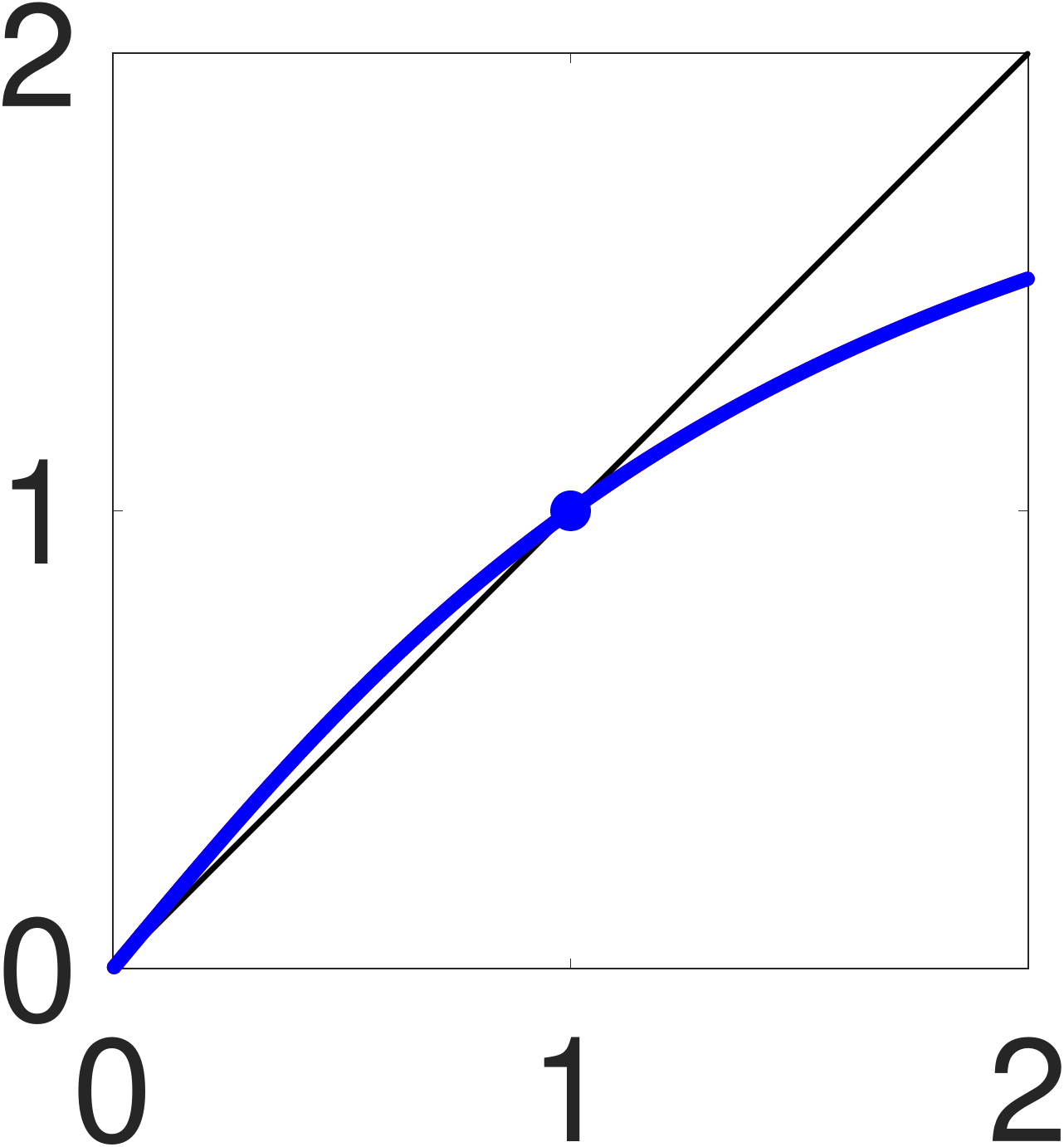}&\includegraphics[scale=0.13,valign=c]{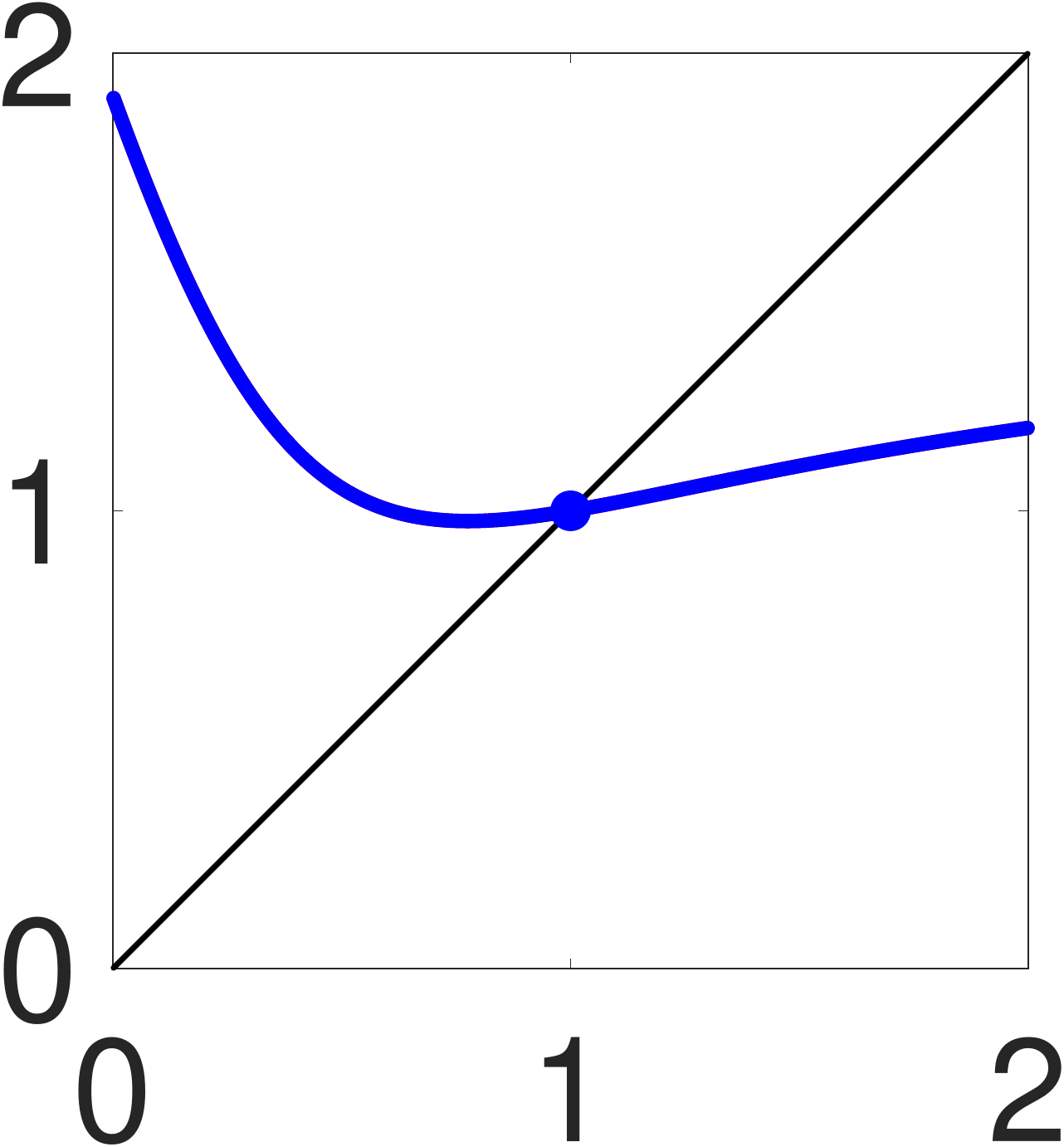}&\includegraphics[scale=0.13,valign=c]{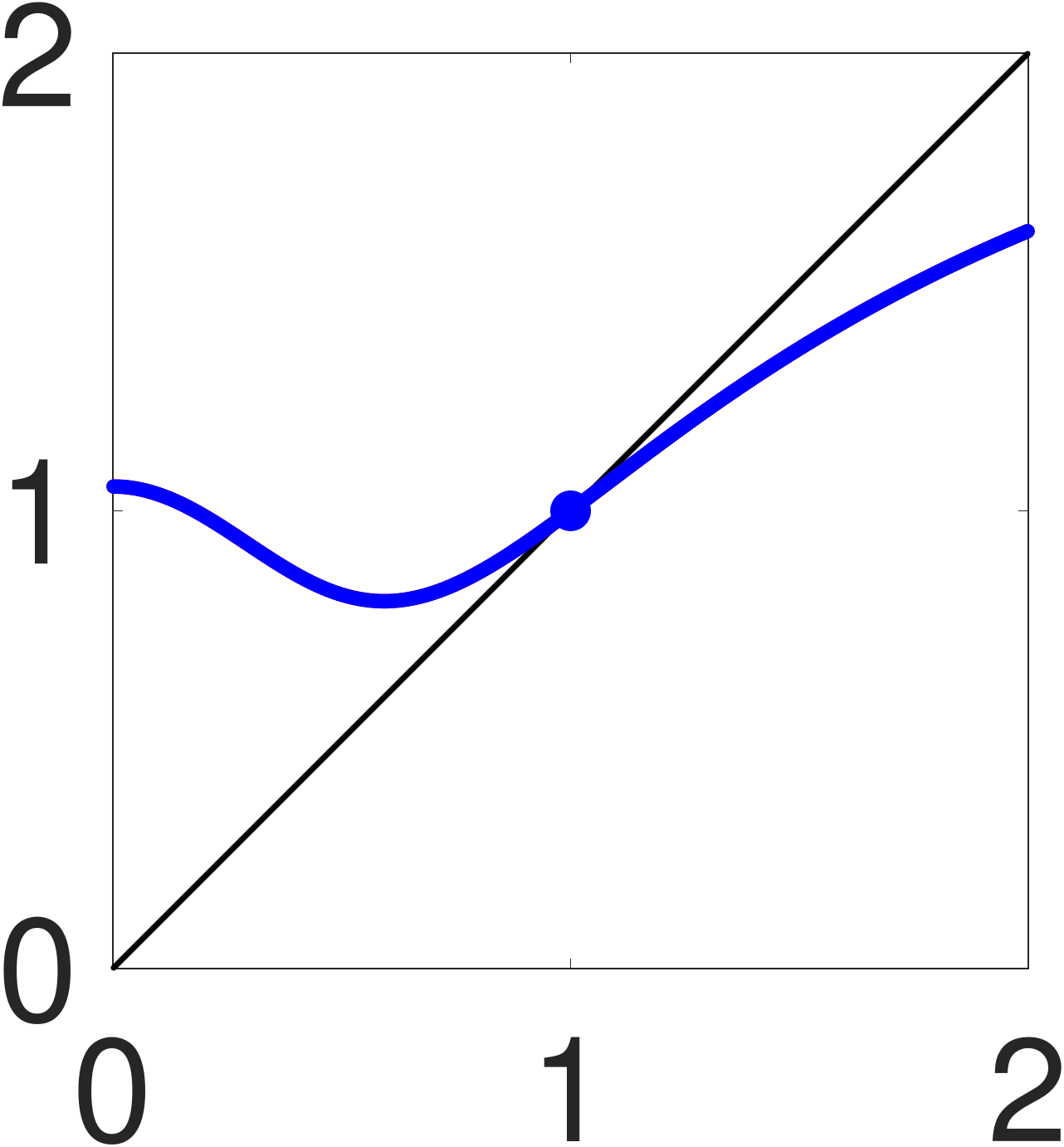}&\includegraphics[scale=0.13,valign=c]{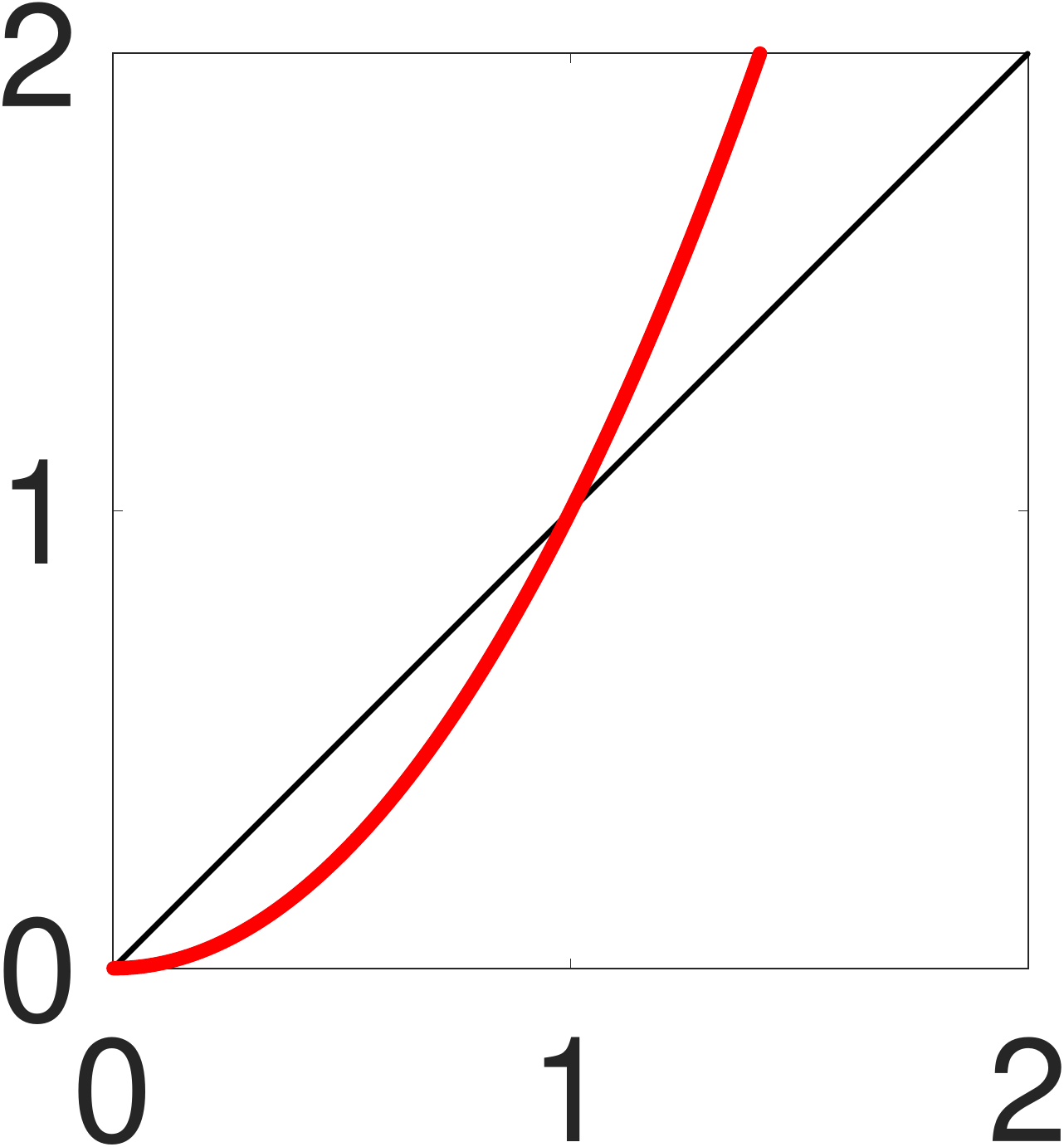}&\includegraphics[scale=0.13,valign=c]{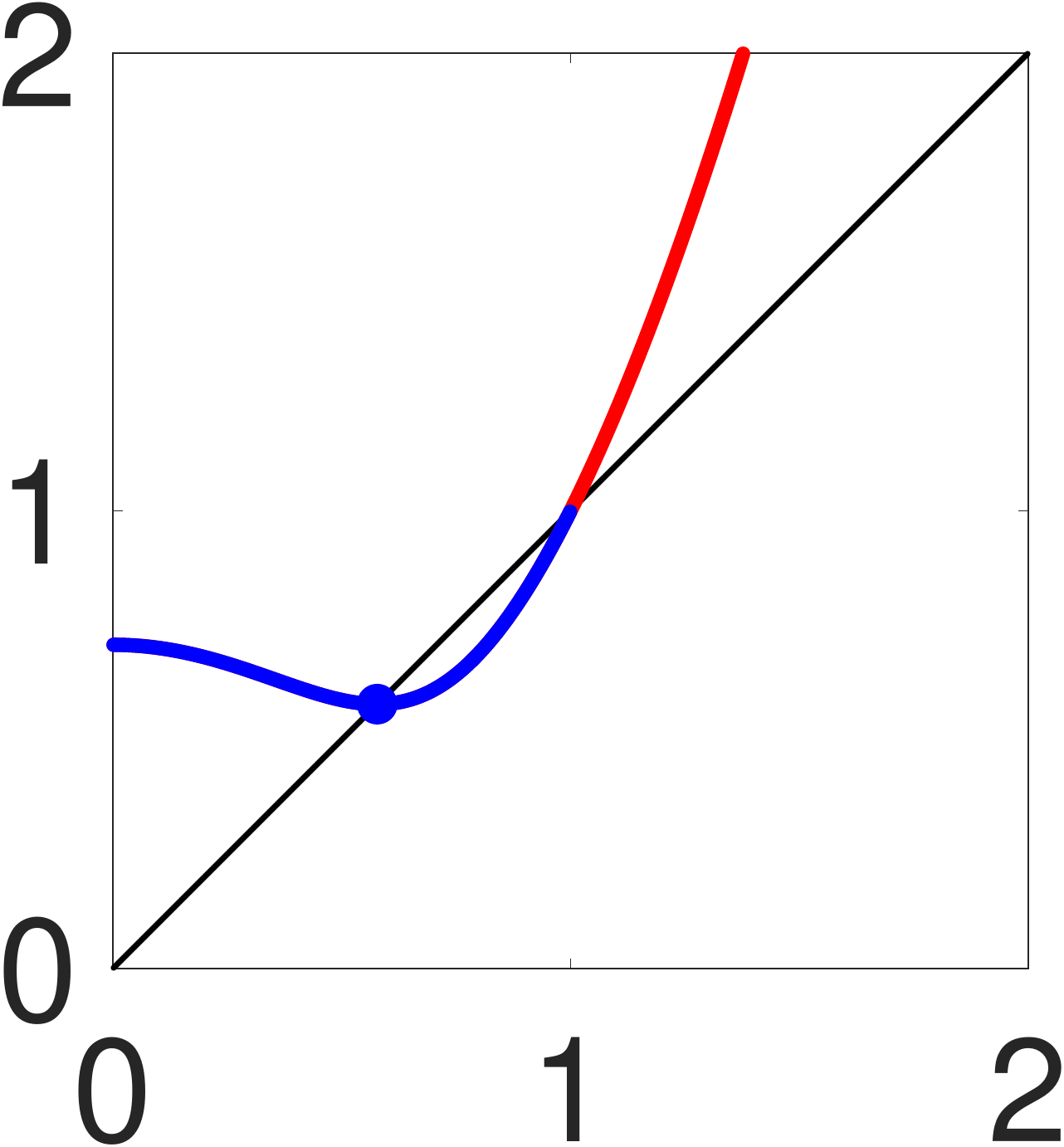}&\includegraphics[scale=0.13,valign=c]{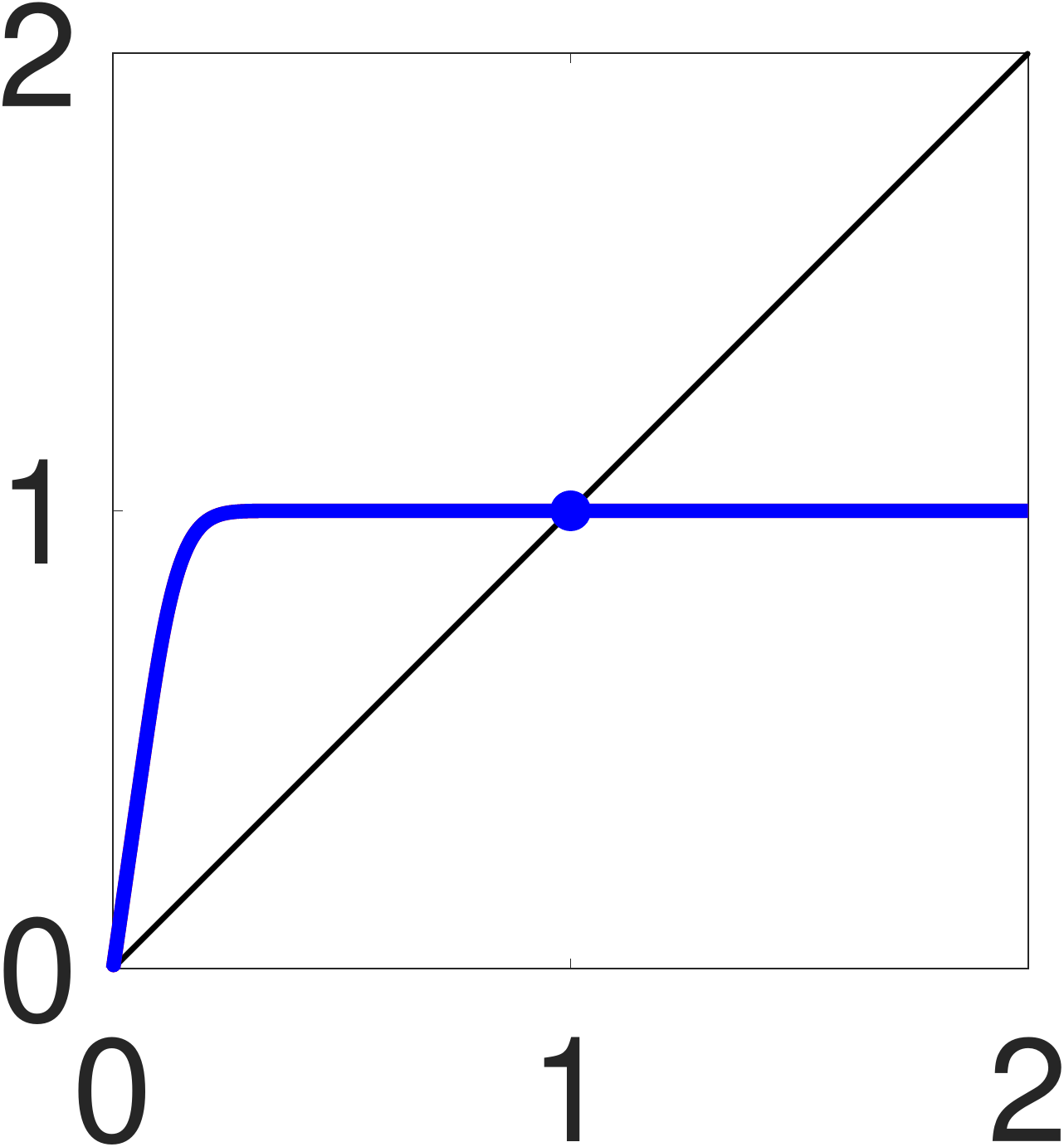}\\
Iter. limit &1&1&1&0&0.58&1\\
Conv. rate&$O(0.72^M)$&$O(0.17^M)$&$O(0.74^M)$&$\lambda^{2^M}$&super-exp.&$\approx O(10^{-10M})$\\
$\frac{\mathfrak{L}_\tau'(\lambda)\lambda}{\mathfrak{L}_\tau(\lambda)}$&\includegraphics[scale=0.13,valign=c]{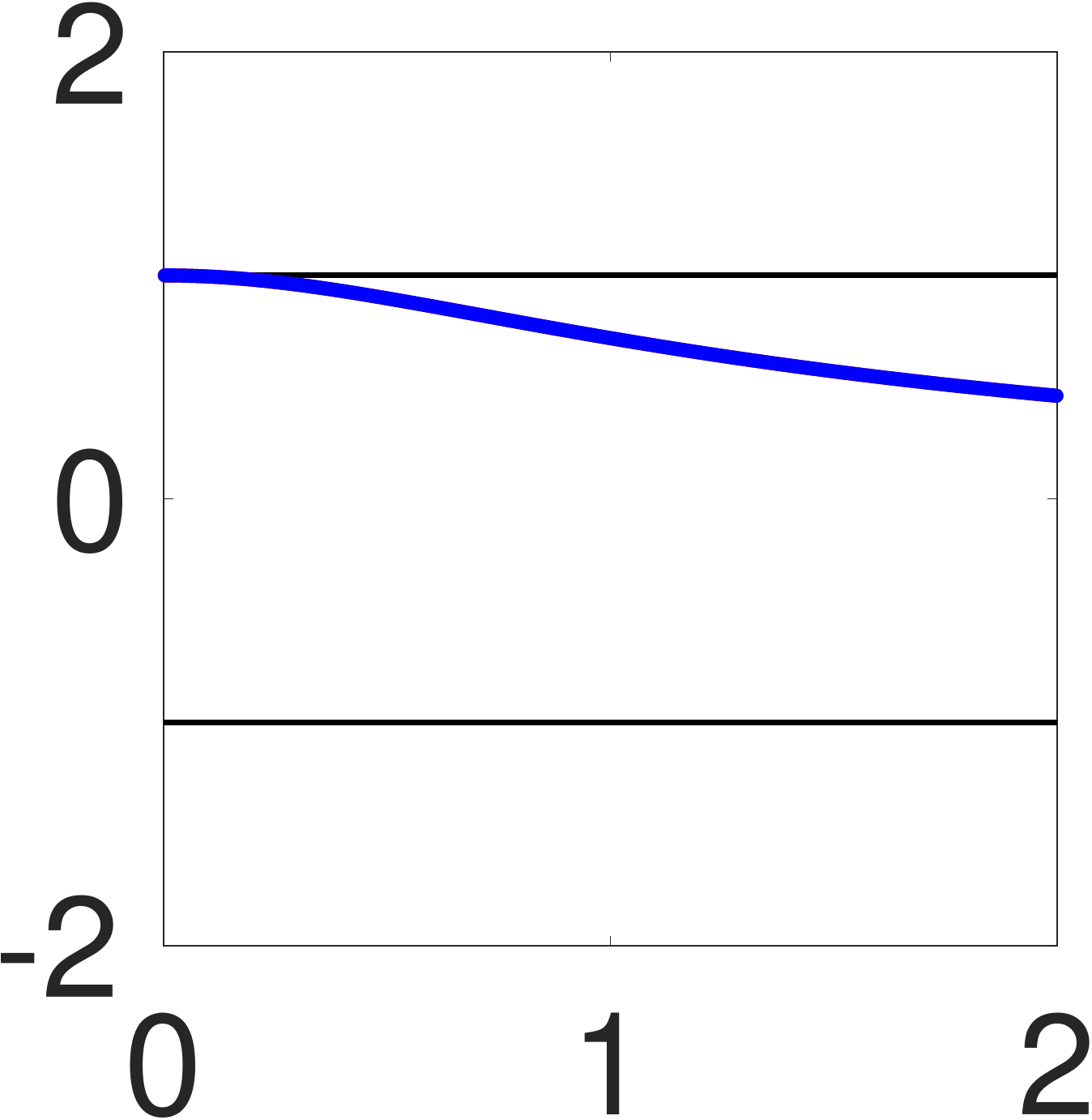}&\includegraphics[scale=0.13,valign=c]{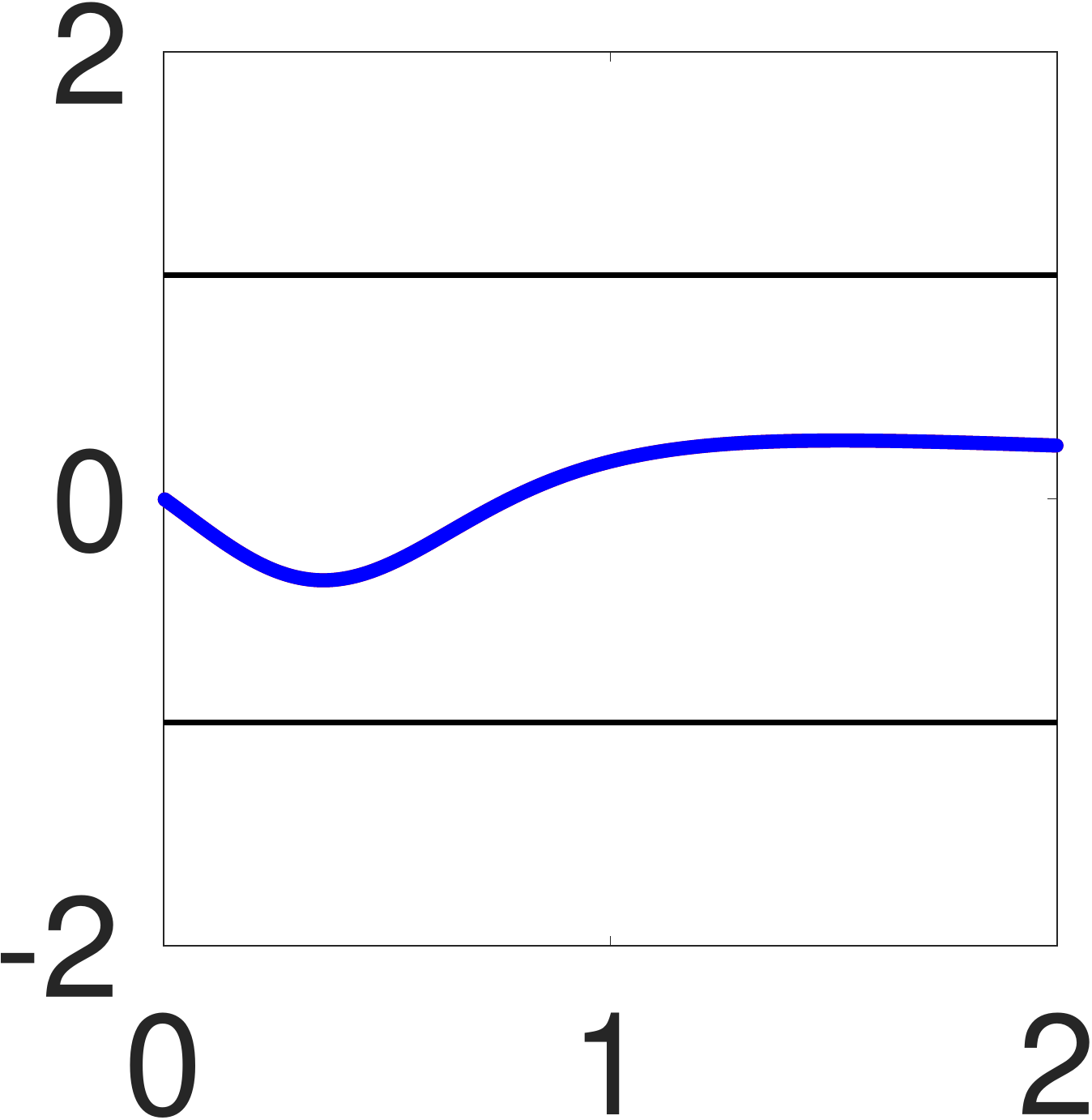}&\includegraphics[scale=0.13,valign=c]{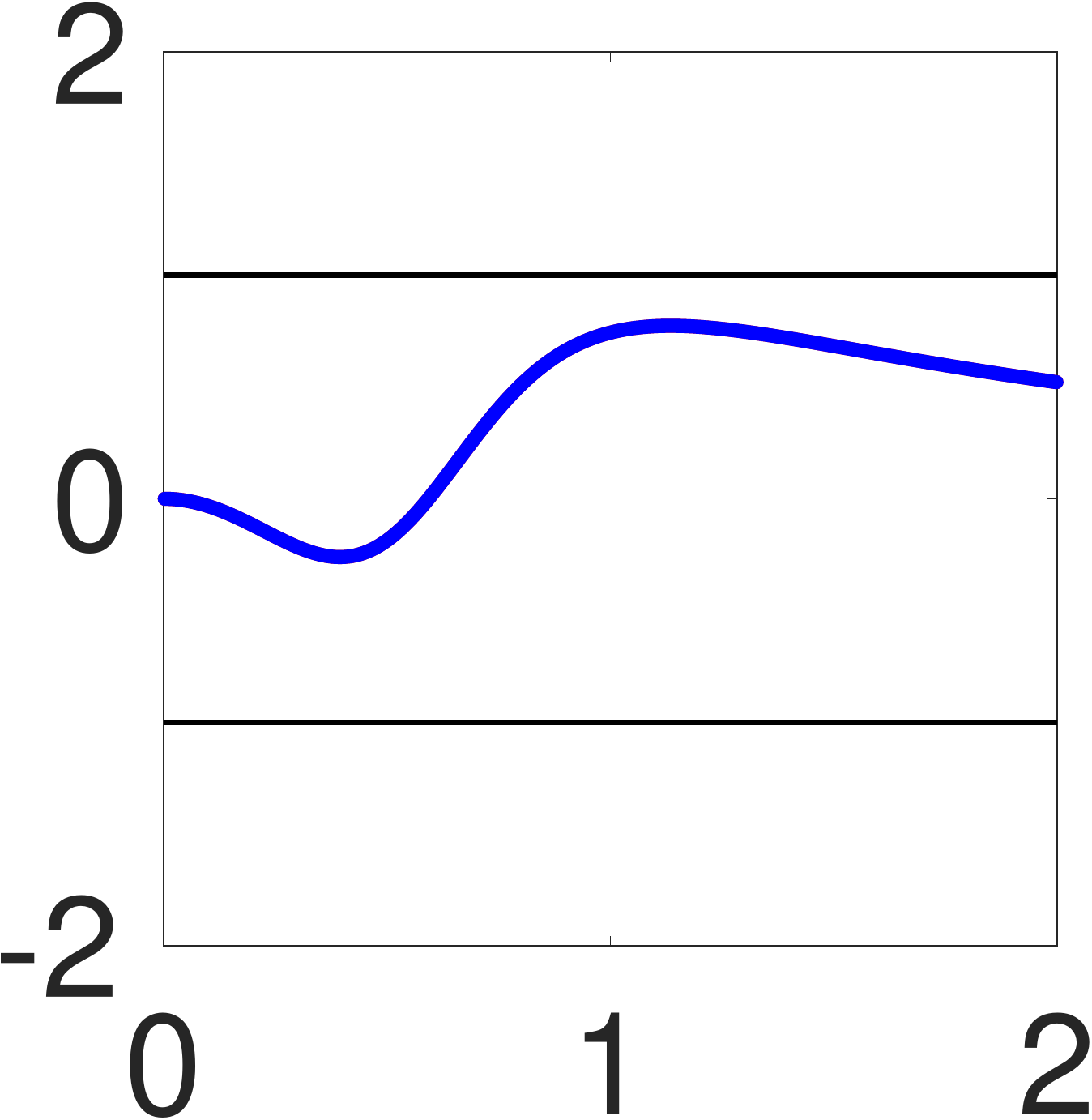}&\includegraphics[scale=0.13,valign=c]{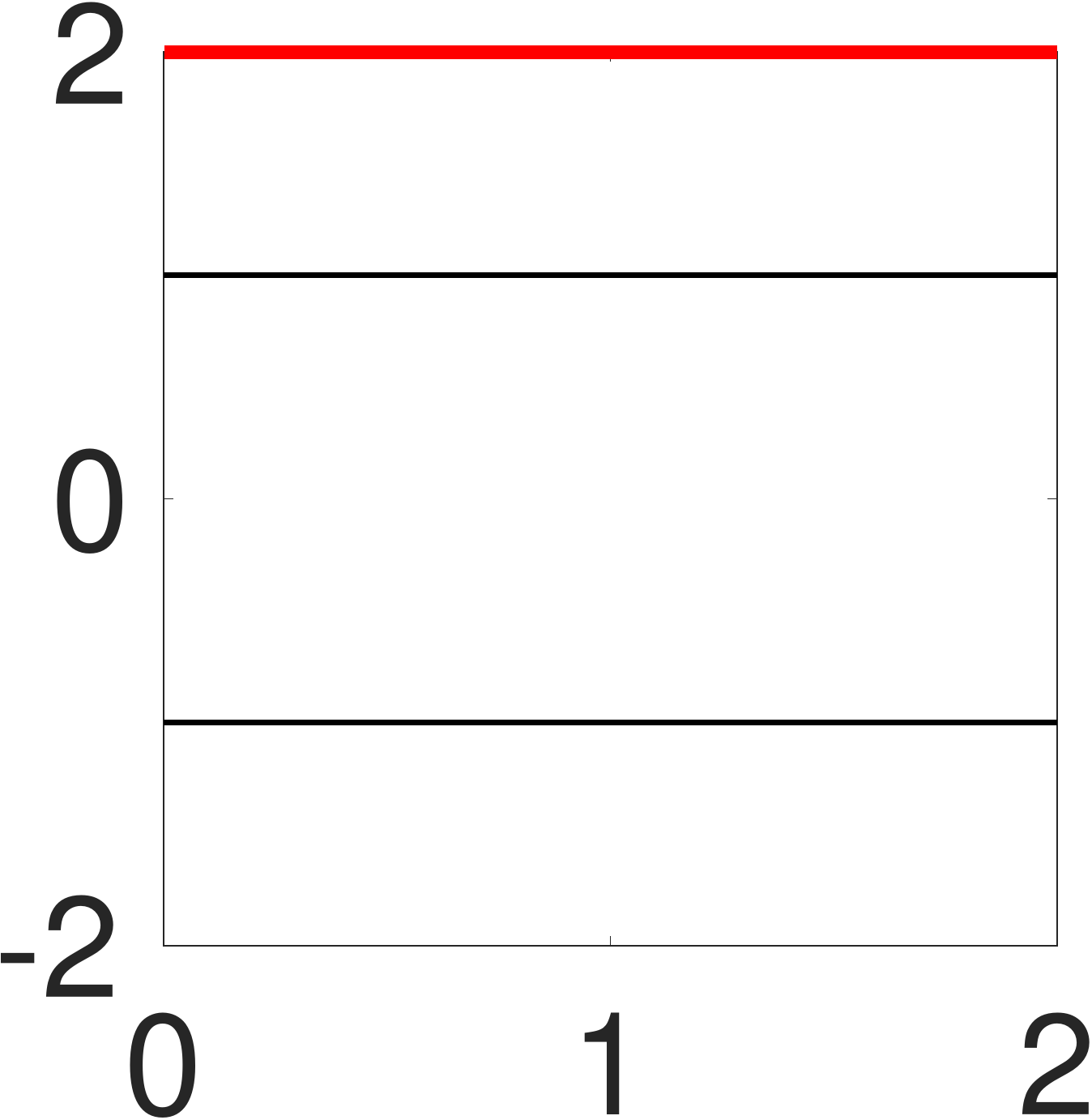}&\includegraphics[scale=0.13,valign=c]{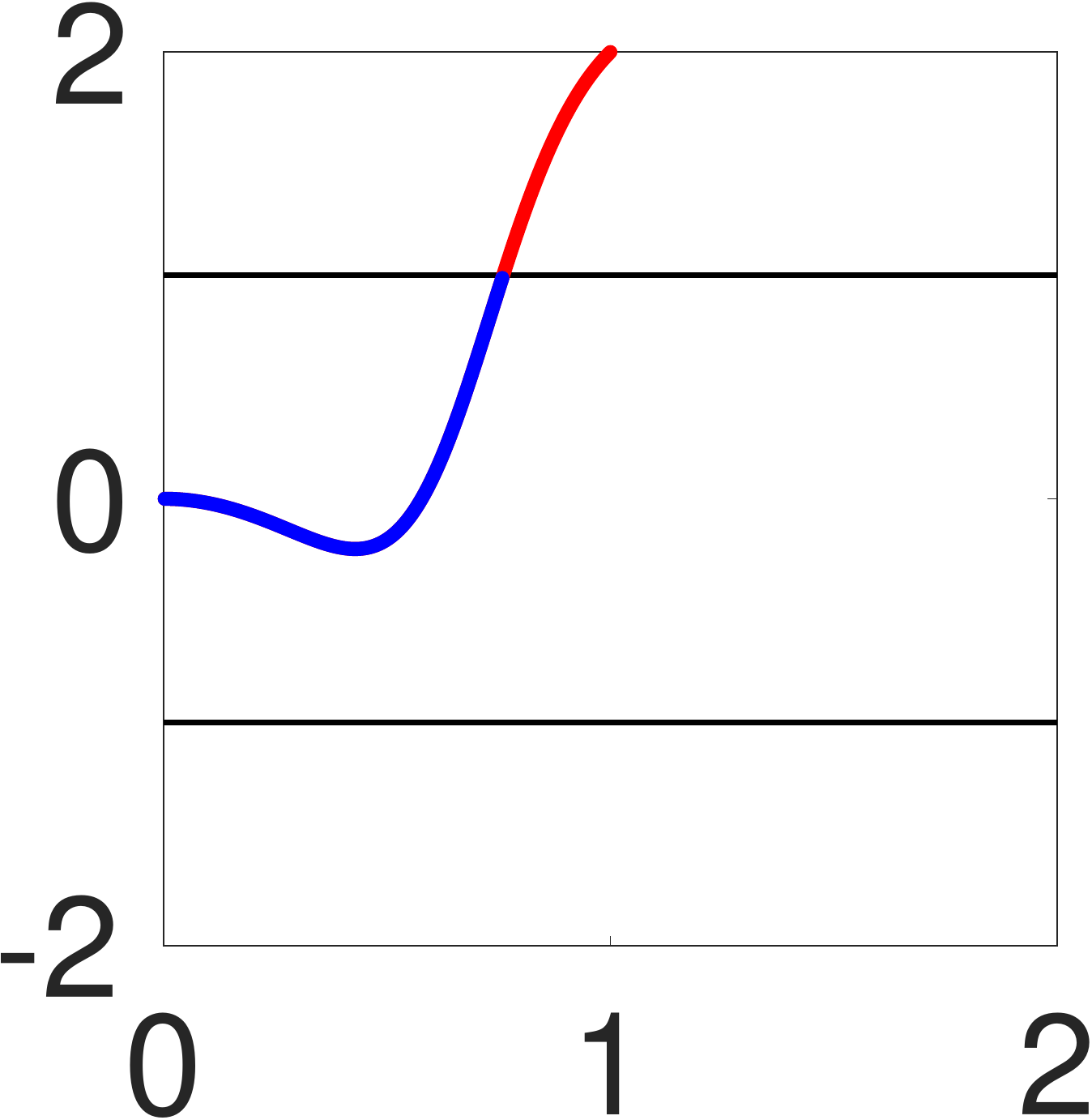}&\includegraphics[scale=0.13,valign=c]{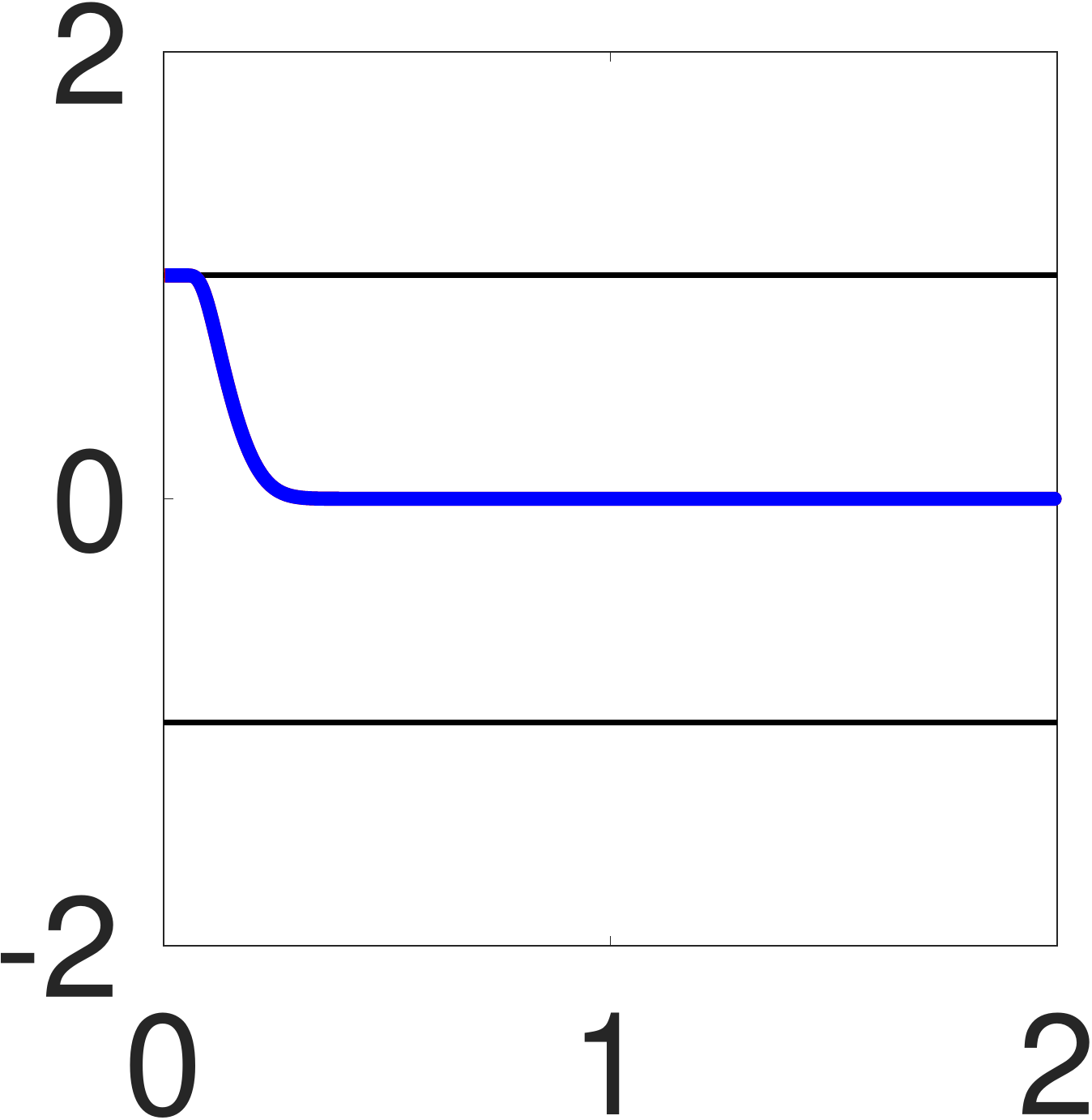}\\
$\tilde{\mathfrak{C}}_\tau(c)$&\includegraphics[scale=0.13,valign=c]{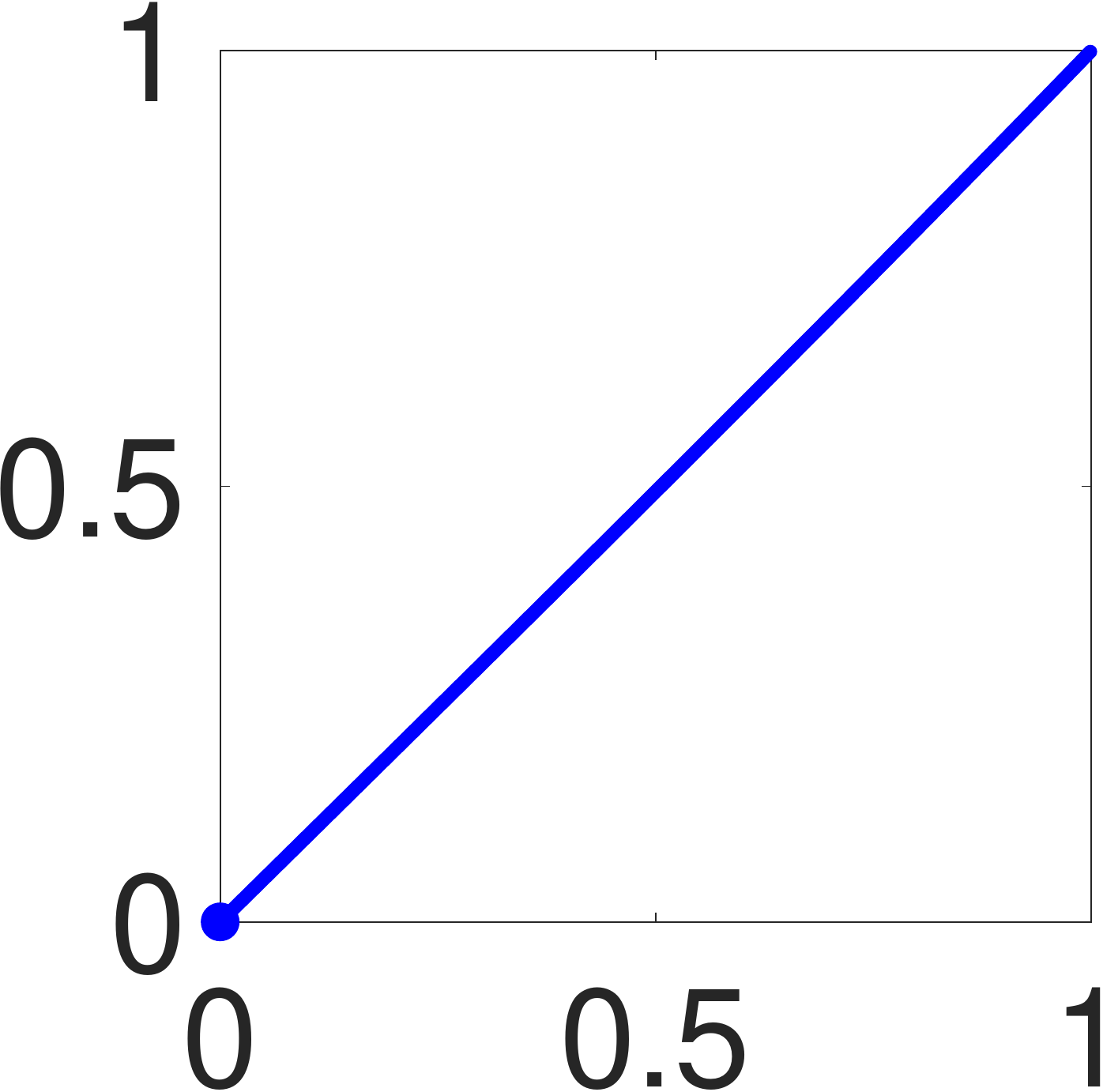}&\includegraphics[scale=0.13,valign=c]{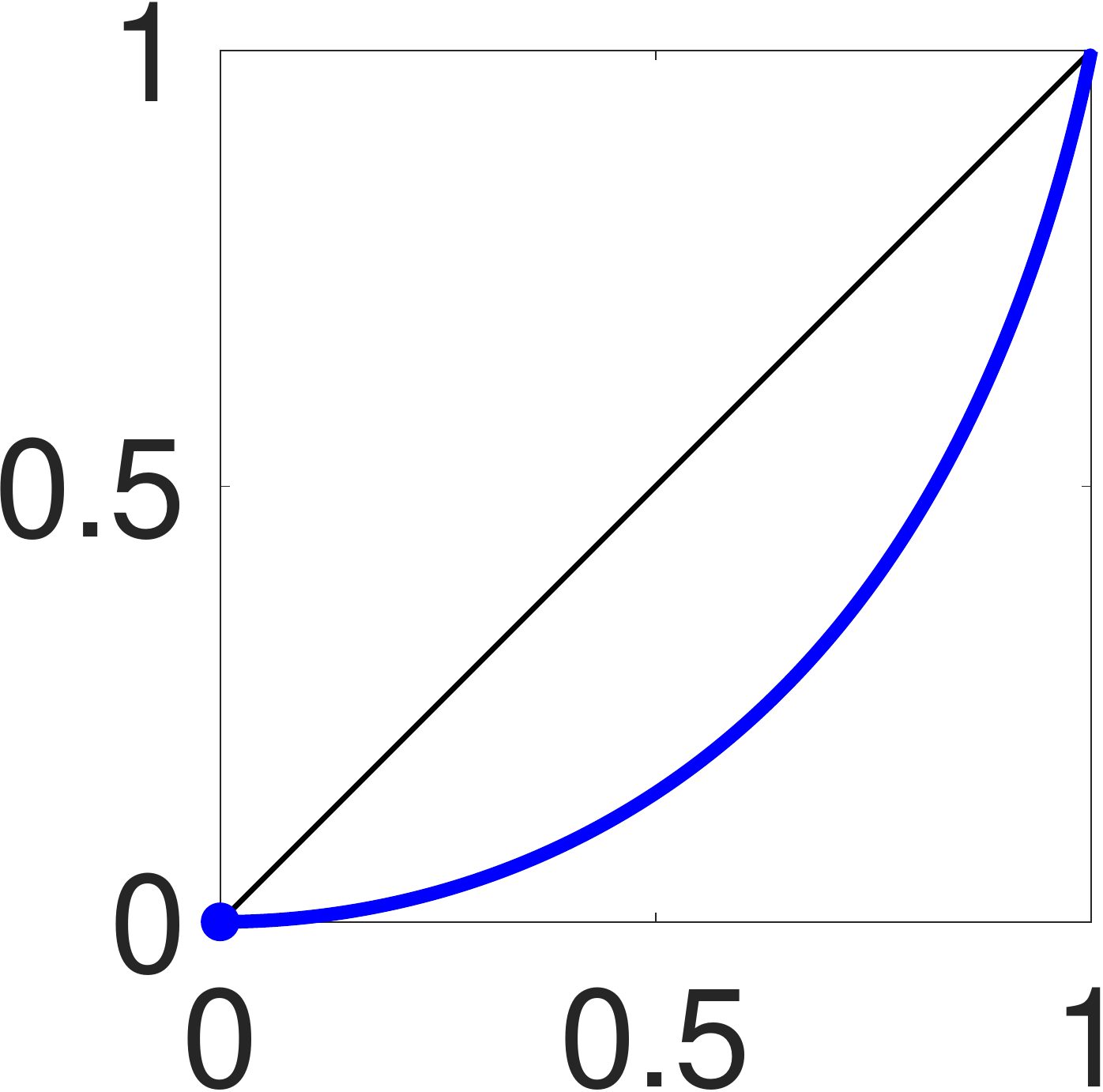}&\includegraphics[scale=0.13,valign=c]{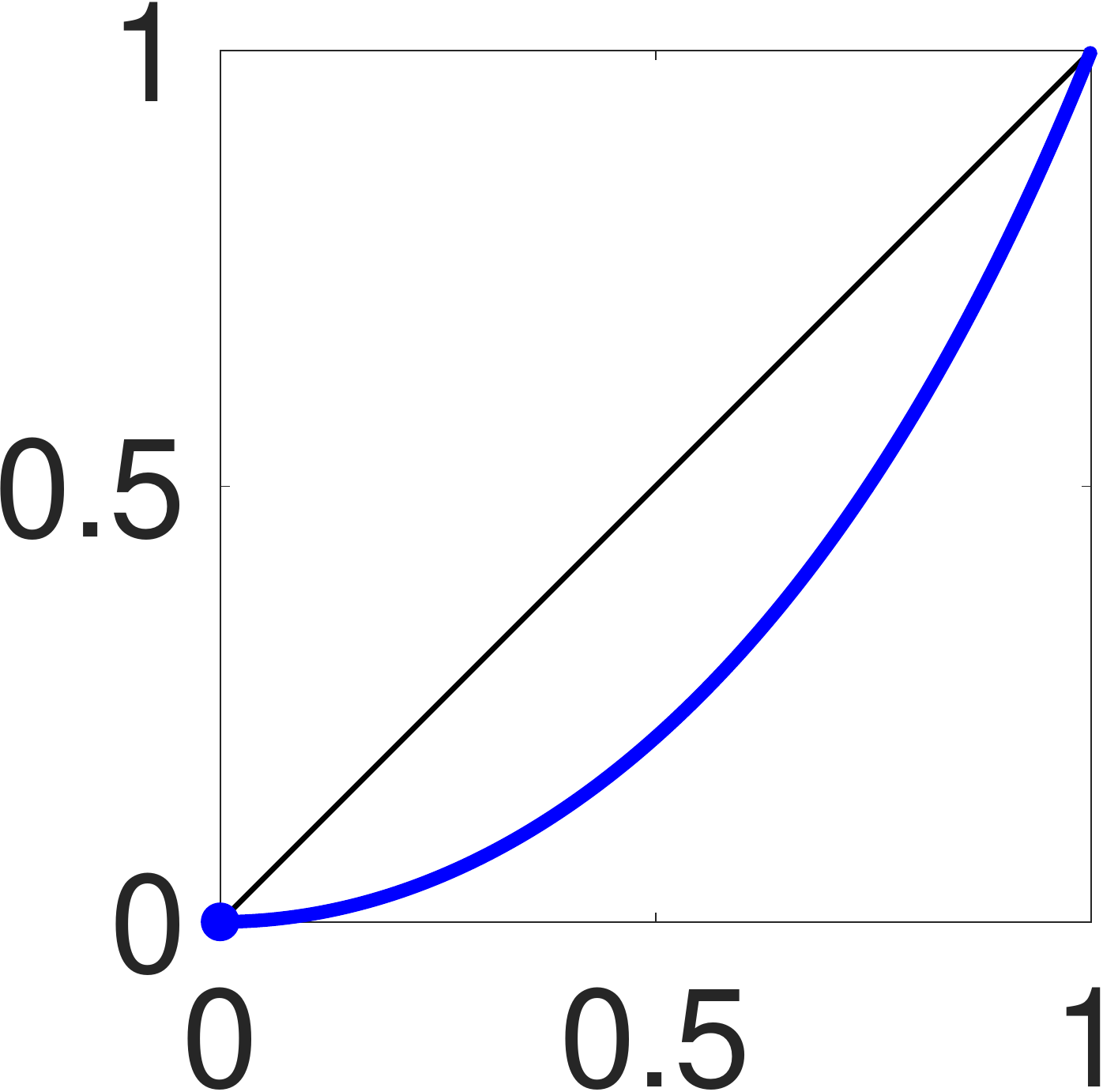}&\includegraphics[scale=0.13,valign=c]{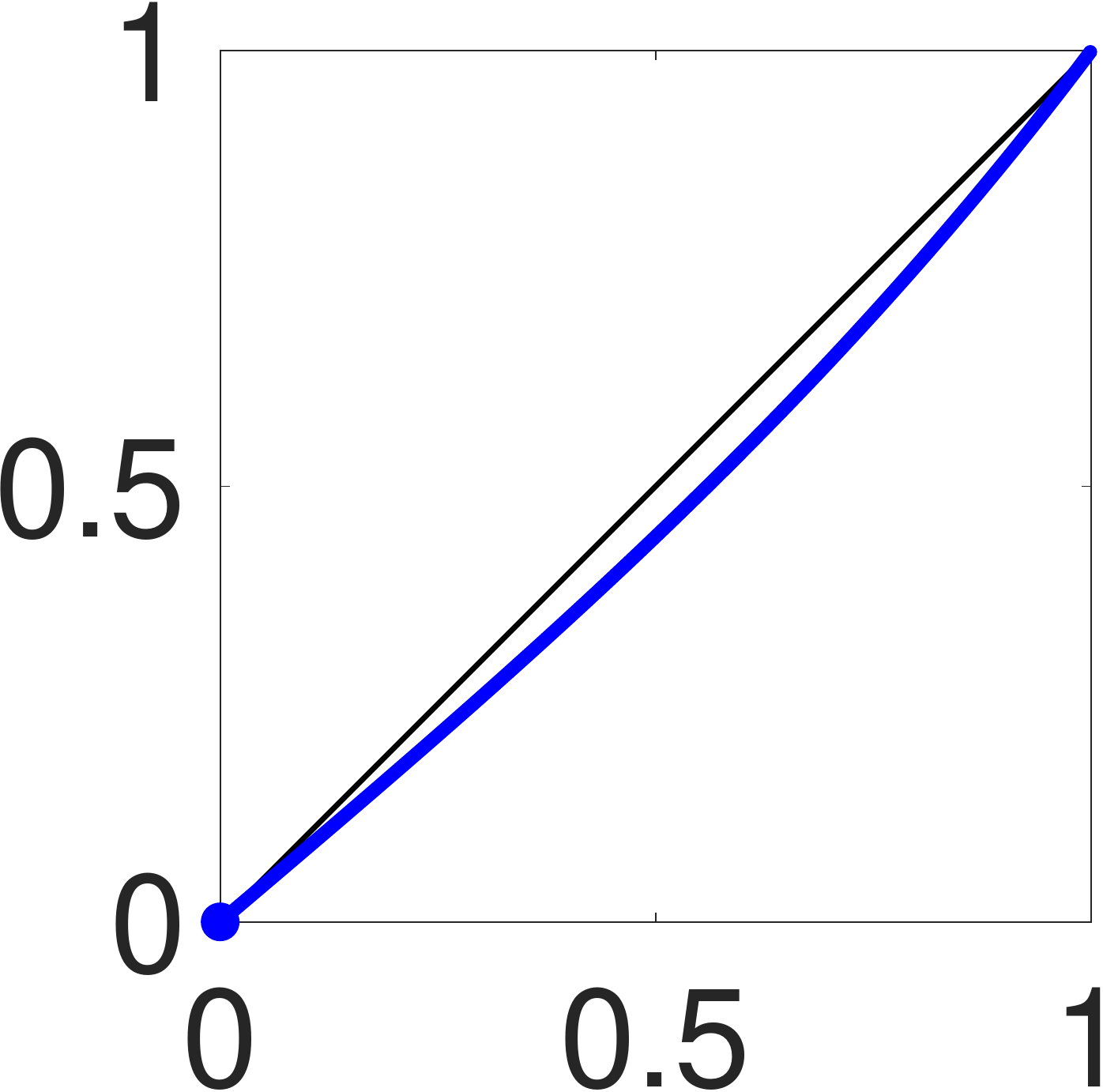}&\includegraphics[scale=0.13,valign=c]{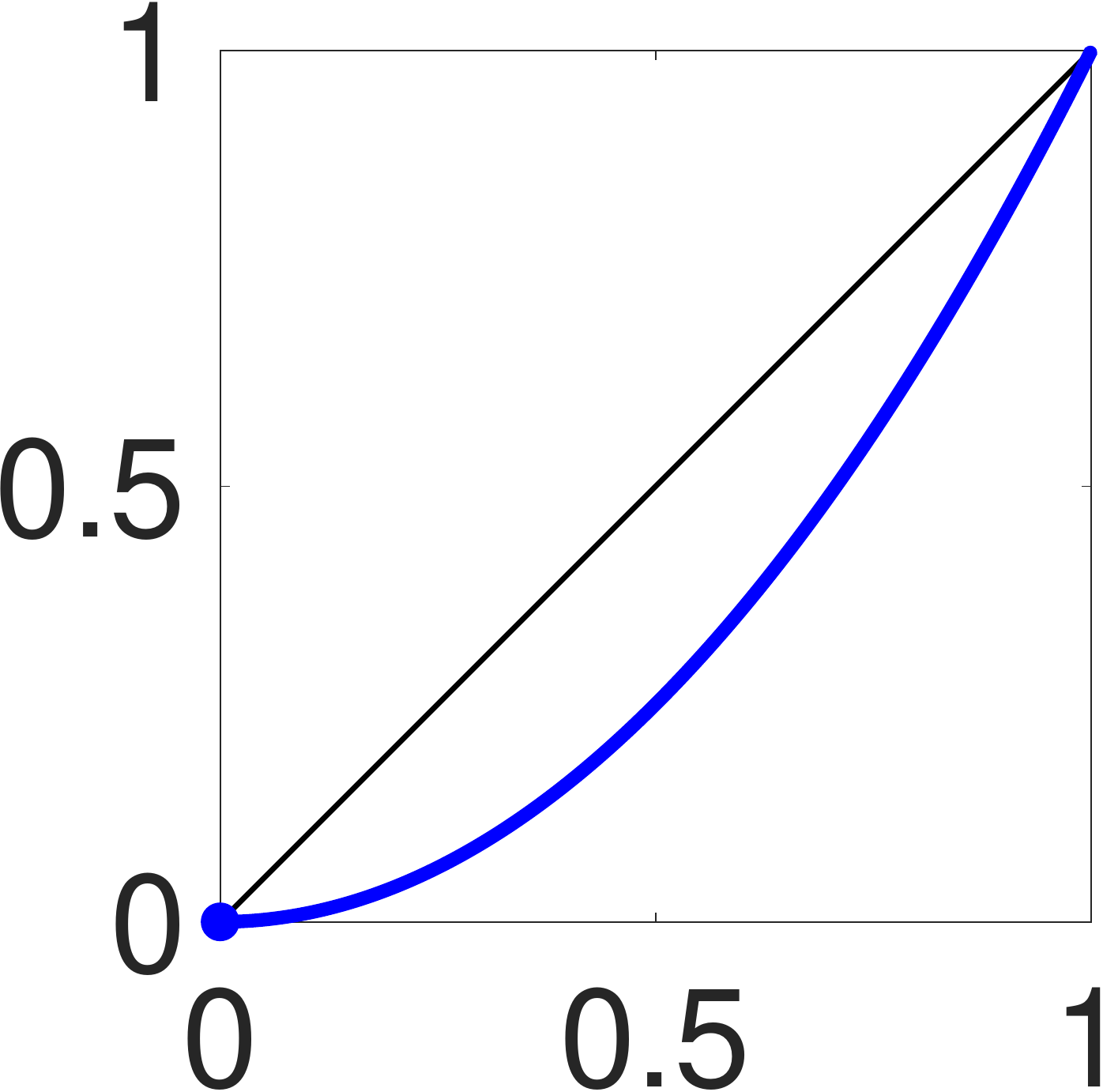}&\includegraphics[scale=0.13,valign=c]{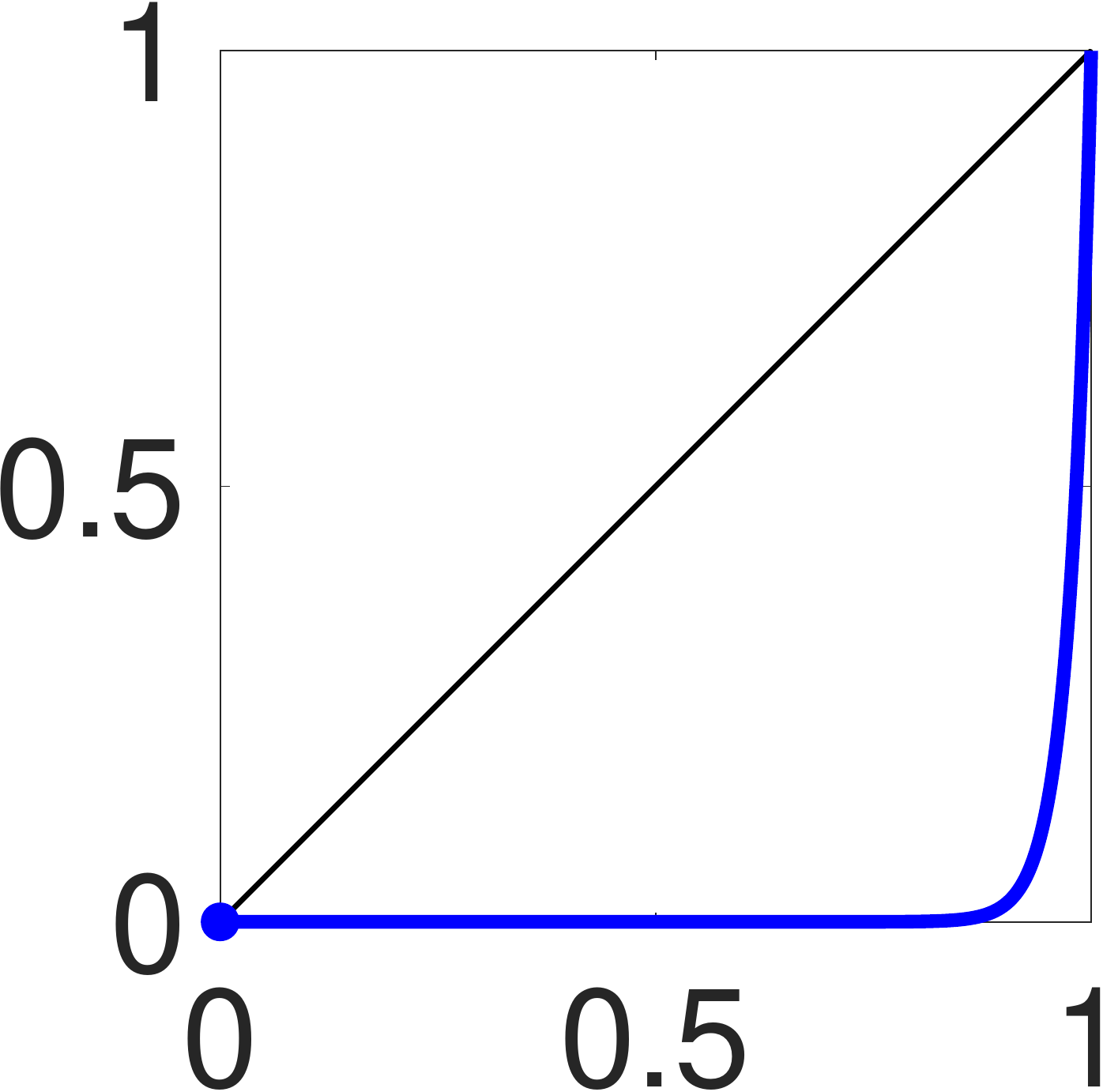}\\
Iter. limit &0&0&0&0&0&0\\
Conv. rate&$O(0.98^M)$&super-exp.&super-exp.&$O(0.85^M)$&$c^{2^M}$&super-exp.\\
$\mathfrak{n}_\tau(\lambda^2, 0)$&\includegraphics[scale=0.13,valign=c]{graphsFinal/covCurveillu/nlc_sigmoid.pdf}&\includegraphics[scale=0.13,valign=c]{graphsFinal/covCurveillu/nlc_abstanh.pdf}&\includegraphics[scale=0.13,valign=c]{graphsFinal/covCurveillu/nlc_gaussian.pdf}&\includegraphics[scale=0.13,valign=c]{graphsFinal/covCurveillu/nlc_square.pdf}&\includegraphics[scale=0.13,valign=c]{graphsFinal/covCurveillu/nlc_abssquare.pdf}&\includegraphics[scale=0.13,valign=c]{graphsFinal/covCurveillu/nlc_saw.pdf}\\
$\lambda \rightarrow \infty$&$O(\sqrt{\lambda})$&$O(\lambda)$&$O(\lambda)$&1.15&1.41&$O(\lambda)$\\
\\
\end{tabular}
}
\caption{(This table is a continuation of table \ref{covCurveillu1}.)}
\label{covCurveillu4}
\end{table}

\newpage

\section{Explaining the NLC's properties with mean field theory} \label{nlcExplainSection}

In section \ref{nlcPropertiesSection}, we demonstrated that a range of properties holds for the NLC empirically in the architectures' initial state. It turns out that many of them can be explained via mean field theory as follows. (i) The property holds exactly for $\mathfrak{n}$. (ii) $\mathfrak{n}$ closely approximates the NLC as shown in section \ref{meanFieldPracticalSection}. Below, we go through the properties one by one. We list further properties of $\mathfrak{n}$ at the end of section \ref{pathEquationSection}.

\paragraph{The NLC is robust to data distribution (section \ref{nlcRobustDataSection})} $\mathfrak{n}$ depends on the distribution only through the scalar $q$ and $c$ parameters. We showed in section \ref{elemLikeSection} that the three datasets we study in this work have characteristic $q$ and $c$ values which lead to accurate mean field estimates. $\mathfrak{n}$ will have the same value for any dataset with the same characteristic $q$ and $c$ values. In section \ref{elemLikeSection}, we also discussed the general class of such datasets. Note that when mean field theory is extended to cover convolutional layers, it utilizes the entire covariance structure of the input. See section \ref{meanFieldCNNsection}.

\paragraph{The NLC is robust to random initialization (section \ref{nlcRandomInitSection})} $\mathfrak{n}$ does not depend on $\theta$ directly, but only on the initialization scheme. Hence, it is invariant to the actual initial value of $\theta$ drawn from that scheme. When $\theta$ is random, the NLC converges almost surely as width converges to infinity.

\paragraph{The NLC can (sometimes) be proxied by even simpler metrics (section \ref{nlcSimpleMetricsSection})} Using statements (\ref{eqn5p4}), (\ref{eqn5p7}) and (\ref{eqn5p10}) from theorem \ref{mfntPropagation} and the relation $||\mathbb{S}_xf_l||_2^2 = \Tr(\Cov_{f_l})$, we obtain that the NLC and NLCFROB as defined in section \ref{nlcSimpleMetricsSection} have the same mean field limit.

\paragraph{The NLC is related to kernel methods and model complexity (section \ref{nlcKernelSection})} See section \ref{networkCovKerSection}.

\paragraph{The NLC is decomposable into NLCs of individual layers (section \ref{nlcDecomposableSection})} The mean field limits of the NLC and LNLC as defined in section \ref{nlcDecomposableSection} are identical.

\paragraph{The NLC is robust to width change (section \ref{widthInvarianceSection})} The essence of mean field theory is convergence with increasing width. Hence, a change of width will not alter the NLC significantly as long as the width remains sufficiently large and the initial weight variance co-varies according to the LeCun initialization / variance parameter. In section \ref{widthInvarianceSection}, we show that even a width of 50, which is considered small in practice, is sufficient to obtain stability.

\paragraph{The initial NLC predicts the final NLC (section \ref{nlcEvolutionSection})} In section \ref{meanFieldPracticalEmpiricalSection}, we measured our theoretical limit quantities with mean field metrics that use the magnitude of the actual parameter value. In that section, we also showed that the parameter magnitude, and hence the value of the mean field metrics, often does not change significantly during training. 

\paragraph{The NLC is continuous from layer to layer (section \ref{nlcBackpropSection})} According to the nonlinearity path equation, $\mathfrak{n}$ increases gradually and multiplicatively as we include more and more activation layers.

\section{Defining and explaining Gaussian stability} \label{gaussianStabilityExplanationSection}

We have repeatedly encountered the notion of Gaussian stability and Gaussian instability throughout this work. (Gaussian instability simply denotes the absence of Gaussian stability.) In the context of our empirical studies, we defined Gaussian unstable architectures (GUAs) and Gaussian edge architectures (GEAs) in terms of activation functions and normalization methods as given at the end of section \ref{metricTerminologySection} / \ref{metricsSummarySection}, with a full list of GUAs and GEAs available in the appendix in chapter \ref{fullListChapter}. GUAs have shown distinct empirical behavior in sections \ref{nlcRobustDataSection}, \ref{nlcRandomInitSection}, \ref{nlcSimpleMetricsSection}, \ref{nlcDecomposableSection}, \ref{widthInvarianceSection}, \ref{nlcBackpropSection}, \ref{meanFieldDistributionSection} and \ref{meanFieldPracticalSection}. The aforementioned subsections of section \ref{nlcPropertiesSection} demonstrate properties of the NLC that are explained by mean field theory in section \ref{nlcExplainSection}. Hence, as long as an architecture's behavior in the limit of infinite width is predictive of its behavior at a practical width, we expect these properties of the NLC to hold. In section \ref{meanFieldDistributionSection} and \ref{meanFieldPracticalSection}, we found that the limit is often not predictive for GUAs. This explains why these properties of the NLC hold to a lesser degree for GUAs.

While Gaussian instability causes mean field theory to break down, it can nonetheless be explained using mean field theory, as we do below. The explanation will lead to the precise definition of the property of `mean field Gaussian stability', which then informs the more practical, ``fuzzy'' definition of Gaussian stability.

We begin by considering a setup like the one used in section \ref{actFunLengthKernelSection}. Let an architecture $f$ be composed of $M+1$ macro-layers which are themselves composed of an FC and an activation layer. Let the FC layers be LeCun Gaussian initialized and let all activation layers use activation function $\tau$. Then $\mathfrak{l}_L = \mathfrak{L}_\tau^{M+1}(\mathfrak{l}_0)$. Further, assume that $\tau$ is normalized to have $\mathfrak{L}_\tau(1) = 1$ and that $\mathbb{E}_ix[i]^2 = 1$ for all inputs $x$, so that $\mathfrak{l}_0=1$ and $\mathfrak{l}_L = 1$. Let's see what happens when we forward-propagate some input $x$. After the first macro-layer, we obtain $\mathfrak{l}_2=1$. (The 2 subscript arises because the macro-layer contains two layers.) If the layer is sufficiently wide, we also obtain $\mathbb{E}_if_2(x)[i]^2 = 1 + \epsilon$ for some small $\epsilon$. However, because the macro-layer still has finite width and thus cannot follow mean field theory exactly, we generally do not obtain $\epsilon = 0$. For the remainder of the architecture, we can now take two approaches for our mean field recursion. Either, we continue to recurse on $\mathfrak{l}_2=1$ to obtain $\mathfrak{l}_L=1$, or we recurse on $1+\epsilon$, which we denote by $\tilde{\mathfrak{l}}_2$, to obtain $\tilde{\mathfrak{l}}_L = \mathfrak{L}_\tau^M(1+\epsilon)$. For example, we can obtain $\tilde{\mathfrak{l}}_L$ by applying theorem \ref{mfntPropagation} with $f_3$ as the readin layer. If $\epsilon$ is sufficiently small, we further obtain $\tilde{\mathfrak{l}}_L\approx 1 + \epsilon\frac{d\mathfrak{L}_\tau^M(1)}{d\lambda} = 1 + \epsilon\mathfrak{L}_\tau'(1)^M$.

In section \ref{actFunLengthKernelSection}, we described the possible behaviors of $\mathfrak{L}_\tau$ upon iteration near a fixed point. If $\mathfrak{L}_\tau'(1) < 1$, the fixed point at 1 is stable and $\tilde{\mathfrak{l}}_l$ converges to 1 as $l$ increases. If $\mathfrak{L}_\tau'(1) = 1$, the fixed point is stationary and $\tilde{\mathfrak{l}}_l$ remains at $1 + \epsilon + O(\epsilon^2)$. If $\mathfrak{L}_\tau'(1) > 1$, the fixed point is unstable and $\tilde{\mathfrak{l}}_l$ diverges from 1 as $l$ increases. However, divergence also implies that $\tilde{\mathfrak{l}}_l$ diverges from $\mathfrak{l}_l$. Hence, if the activation function in $f$ has $\mathfrak{L}_\tau'(1) > 1$, mean field theory makes conflicting estimates for $\mathbb{E}_if_l(x)[i]^2$ for sufficiently large $l$, and hence mean field theory cannot be predictive for sufficiently deep $f$. In plain terms, mean field theory predicts that small deviations in the layer quadratic means (LQMs) of $f$ grow exponentially during forward propagation. However, since such deviations are always introduced in a practical network due to finite width, the LQMs of $f$ diverge from their mean field limit during forward propagation, and the mean field estimate becomes inaccurate. This constitutes a ``breakdown'' of the practical predictiveness of not just theorem \ref{mfntPropagation}, but also of mean field theory given in prior work as described in e.g. section \ref{meanFieldBackgroundSection}. To our knowledge, this breakdown has not been described in literature.

Let's apply this reasoning to a more general setup. Let a network $f$ be composed of $M+1$ macro-layers composed of an FC and an activation layer, where the $\sigma_l^2$ and $\tau_l$ are arbitrary. Then $\tilde{\mathfrak{l}}_L \approx \mathfrak{l}_L + \epsilon\prod_{\tau_l}\mathfrak{L}_{\tau_l}'(\mathfrak{l}_k)$, where the product is taken over the last $M$ activation layers in $f$. We could argue that divergence happens if the value of the product is above a certain level. However, as is a recurring theme in this work, we must consider the deviation {\it relative} to the value from which it deviates. Say we have $\mathfrak{l}_L = 1$. Then we would consider $\tilde{\mathfrak{l}}_L = 1 + 1 = 2$ to be a significant deviation. However, if $\mathfrak{l}_L = 10^5$, then we would consider $\tilde{\mathfrak{l}}_L = 10^5 + 1$ to be a small deviation. Also consider that the value of both $\mathfrak{l}_L$ and $\tilde{\mathfrak{l}}_L$ can be arbitrarily multiplicatively scaled by appending an FC layer with an arbitrary initial weight variance to the architecture. Appending such an FC layer clearly should not affect our judgment of whether ``mean field theory has broken down''. The same argument applies at layer $f_2$. If $\mathfrak{l}_2$ is small, then a given deviation $\epsilon$ is more significant than if $\mathfrak{l}_2$ is large, because our judgment of how severely an LQM deviates from the mean field estimate should not depend on e.g. whether $\tau_2$ is scaled by a constant.

Hence, we care about the relative deviation at the output $\frac{\epsilon\prod_{\tau_l}\mathfrak{L}_{\tau_l}'(\mathfrak{l}_k)}{\mathfrak{l}_L}$ {\it relative} to the relative deviation at the input $\frac{\epsilon}{\mathfrak{l}_2}$. Table \ref{tableNLCPropagation} yields that the ratio of those ratios is equal to the telescopic product $\prod_{\tau_l}\frac{\mathfrak{l}_k\mathfrak{L}_{\tau_l}'(\mathfrak{l}_k)}{\mathfrak{L}_{\tau_l}(\mathfrak{l}_k)}$. So, we can say that the deviation relatively grows at some activation layer $f_l$ if $\frac{\mathfrak{l}_k\mathfrak{L}_{\tau_l}'(\mathfrak{l}_k)}{\mathfrak{L}_{\tau_l}(\mathfrak{l}_k)} > 1$, and relatively shrinks if $\frac{\mathfrak{l}_k\mathfrak{L}_{\tau_l}'(\mathfrak{l}_k)}{\mathfrak{L}_{\tau_l}(\mathfrak{l}_k)} < 1$. This idea is now formalized.

\begin{definition}
We say an activation function $\tau$ exhibits `mean field Gaussian stability' at $\lambda > 0$ if $\frac{\lambda\mathfrak{L}_{\tau}'(\lambda)}{\mathfrak{L}_{\tau}(\lambda)} < 1$, `mean field Gaussian instability' if $\frac{\lambda\mathfrak{L}_{\tau}'(\lambda)}{\mathfrak{L}_{\tau}(\lambda)} > 1$ and is `mean field Gaussian edge' if $\frac{\lambda\mathfrak{L}_{\tau}'(\lambda)}{\mathfrak{L}_{\tau}(\lambda)} = 1$. (Note that $\frac{\lambda\mathfrak{L}_{\tau}'(\lambda)}{\mathfrak{L}_{\tau}(\lambda)} \le -1$ is excluded by theorem \ref{covkerLregular}.)
\end{definition}

If we substitute $\sqrt{q} = \lambda$ and $\sqrt{\mathfrak{C}_\tau(q,q)} = \mathfrak{L}_{\tau}(\lambda)$, we obtain $\frac{\lambda\mathfrak{L}_{\tau}'(\lambda)}{\mathfrak{L}_{\tau}(\lambda)} = \frac{q\mathfrak{C}_{\tau}'(q,q)}{\mathfrak{C}_{\tau}(q,q)}$, where the derivative is with respect to the double argument $q$. Hence, mean field Gaussian stability can be defined equivalently in terms of the length kernel or covariance kernel.

We already analyzed $\frac{\lambda\mathfrak{L}_{\tau}'(\lambda)}{\mathfrak{L}_{\tau}(\lambda)}$ as a function of $\lambda$ in section \ref{actFunLengthKernelSection} and plotted that function in tables \ref{covCurveillu1} through \ref{covCurveillu4}. \finding{We find that for activation functions used in our study A architectures at $\lambda=1$, stability occurs for SELU, tanh, sigmoid, even tanh and Gaussian. Instability occurs for square and odd square. ReLU is mean field Gaussian edge. The same thing holds for the debiased activation functions.} This is the motivation behind which activation functions we include in our definition of GUAs and GEAs.

Mean field Gaussian instability does not only lead to the divergence of practical LQMs from their limit. By the same argument, it also leads to the divergence of practical LQMs corresponding to different inputs from each other. This observation provides the desired link to our empirical results. In section \ref{meanFieldDistributionEmpiricalSection}, we investigated the LCV metric, which measures the coefficient of variation of LQM values. Out of all the metrics we investigated in section \ref{meanFieldDistributionEmpiricalSection}, LCV provided the clearest signal. Especially before training, stable architectures all had extremely low LCV values, whereas many GUAs and GEAs had large ones. In proposition \ref{mfntElemLike}, we showed that layer values generated by a meta-Gaussian have small LCV. Hence, large LCVs imply that layers are not meta-Gaussian, and we found this to be the case for our GUAs throughout section \ref{meanFieldDistributionEmpiricalSection}. As stated at the beginning of this section, the breakdown of mean field theory implied by non-meta-Gaussian fully-connected layers then explains our observations in section \ref{nlcPropertiesSection}. We stress that mean field Gaussian instability does not cause our theorems \ref{mfntMetaGaussian} and \ref{mfntPropagation} to become false. However, it can cause the width that is required for a practical architecture to behave as its limit to become enormous.

It is also clear how a large LCV value would interfere with the accuracy of mean field predictions of e.g. the NLC. The strongest condition of theorem \ref{mfntPropagation} is that the input distribution is elem-like, which implies that all inputs have the same length. Mean field theory is an iterative theory, i.e. results for child layers are built from results for their parents. Hence, layer lengths remaining approximately constant throughout this recursion is a necessity.

While Gaussian instability can be reduced to the behavior of the length kernel in our setup above, the complexity of the phenomenon increases with the complexity of the architecture definition. Consider for example an architecture that also contains normalization layers between fully-connected and activation layers. If layer normalization is used, then LCV becomes 0 at the normalization layers, i.e. the deviation $\epsilon$ is eliminated. This prevents the exponential compounding of the instability at each activation layer, and hence the architecture as a whole behaves as if it contains mean field Gaussian stable activation functions. Hence, when a fully-connected architecture contains LN, we do not consider it a GUA or GEA in our studies. However, the story is different when batch normalization is used. BN only causes the layer quadratic mean across the entire batch to be 1, but not for individual inputs. This means the LQMs corresponding to different inputs in the same batch can still deviate, and the architecture can exhibit unstable behavior. In fact, this phenomenon is one of the key differences between LN and BN.

Providing an exhaustive analysis of Gaussian stability across architecture types goes beyond the scope of this work. Hence, we must settle for an imprecise definition of Gaussian stability in terms of the empirical observations we have made.

\begin{definition}
We say a neural network $f$ exhibits `Gaussian stability' if values of its fully-connected or convolutional layers are distributed as a draw from an elementwise meta-distribution generated by a meta-Gaussian when the data distribution is neural regular as defined in section \ref{elemLikeSection}. Gaussian stability can be measured by e.g. the NKURT, NGHI, CVNSTD, NCORR and especially the LCV metric given in section \ref{meanFieldDistributionEmpiricalSection}.
\end{definition}

This definition makes no direct reference to mean field theory. We also include convolutional networks in the definition. The theoretical justification for meta-Gaussian convolutional layers is similar to that for meta-Gaussian fully-connected layers. We briefly discuss the mean field theory of convolutional architectures in the next section. Among activation functions used for our study B convolutional architectures, not only are square and odd square mean field Gaussian unstable at $\lambda=1$, but also Swish, softplus-debiased and Swish-debiased. However, their instability is much less severe than that of square and odd square. In practice, we find that only some architectures based on a linearized version of abs. val.-debiased exhibit clear Gaussian instability as captured by the above definition. We specify those architectures at the end of section \ref{metricTerminologySection} / \ref{metricsSummarySection} and in section \ref{fullListB}. They are considered GUAs in our empirical studies. Further, layer normalization is not sufficient to prevent Gaussian instability in convolutional architectures, because it does not prevent the quadratic mean across some spatial regions of the layer value to diverge from the quadratic mean across other spatial regions. Hence, there are GUAs using LN in study B.

Gaussian instability can have extreme effects. Let's return one more time to our setup with alternating LeCun Gaussian initialized FC and activation layers. Let's say the square activation function is used, normalized to achieve $\mathfrak{L}_\tau(1) = 1$. When computing the recursive calculation rules of table \ref{tableNLCPropagation}, even if we compute the length kernel as accurately as possible under 32-bit floating-point precision, i.e. with a rounding error of around $10^{-7}$, at around macro-layer 30 this rounding error explodes and induces numerical underflow or overflow. Of course, this explosion happens much more quickly when we compute the length kernel inaccurately or during actual forward propagation in the finite-width network.

When we designed empirical studies A and B, we always observed that architectures based on square or odd square and no normalization layers exhibited numerical underflow or overflow. Hence, we decided to always combine square and odd square activation functions with normalization layers. We made this decision before we understood the phenomenon of Gaussian stability.

Gaussian stability plays an important role when computing metrics via statistical estimation, as we explained in sections \ref{metricEstimationSection} and \ref{nlcComputeSection}. Our metrics are often based on operators such as expectation and standard deviation applied to layer quantities.

Gaussian stability bears similarity to scale stability as given in section \ref{architectureDesignParadigmsSection}. We further distinguish the two in section \ref{forwardStabilitySection}.

\section{Mean field theory of CNNs: an outlook} \label{meanFieldCNNsection}

Due to time and space limitations, we do not extend our analysis in this chapter to architectures with convolutional layers. \citet{meanFieldNetsorGP,meanFieldCNN} show how to apply mean field theory to convolutional architectures. To apply mean field theory to A-architectures, which can contain a range of layer operations, we re-cast them as mean field architecture as defined in section \ref{meanFieldArchitectureSection}, which can contain only two layer operations, one of which can be considered a ``template''. For convolutional architectures, we can follow the same steps to derive recursive calculation rules for e.g. the square mean and co-mean of layers that use operations such as the convolutional and pooling operation in the style of table \ref{tableNLCPropagation}. However, there are a number of additional complexities.

Consider a tensor layer as defined in section \ref{tensorLayerSection} with spatial dimensions $S_{l,1}, S_{l,1}, .., S_{l,T_l-1}$ and $C_l$ channels. When re-casting the convolutional architecture to a mean field architecture, in general, that tensor layer is cast not as one layer in the mean field architecture, but as $S_{l,1}S_{l,1}..S_{l,T_l-1}$ layers. For each tensor layer in the convolutional architecture, we require one layer in the mean field architecture per spatial location, each of which has width $C_ld_\text{MF}$. Consider, then, a convolutional layer with a weight tensor that has spatial dimensions $H_{l,1}, H_{l,2}, .., H_{l,T_l-1}$. In the mean field architecture, the convolutional layer corresponds to $H_{l,1}H_{l,1}..H_{l,T_l-1}S_{l,1}S_{l,1}..S_{l,T_l-1}$ fully-connected layers, each with a weight matrix of size $C_kd_\text{MF} \times C_ld_\text{MF}$. (Here, we assume the dependency is padded.) Each fully-connected layer is responsible for the contribution of one spatial location in the dependency to one spatial location in the convolutional layer. Finally, we add the contributions to each spatial location in the convolutional layer together with a single elementwise or addition layer per spatial location.

At the end of this process, we obtain for one tensor layer not one value of $\mathfrak{m}$, $\mathfrak{c}$, $\mathfrak{q}$ and $\mathfrak{g}$, but one for each spatial location. Of course, this is significantly more unwieldy than the fully-connected case. It is consequently somewhat more difficult to extract simple yet practically meaningful patterns from the recursion.

The technical conditions of mean field theory are also more difficult to (approximately) fulfill for convolutional architectures. If the input $x$ comes as a tensor, then in order to make theorems \ref{mfntMetaGaussian} and \ref{mfntPropagation} work, we would require an elem-like condition that states that for all spatial locations and inputs, the square mean across channels is approximately equal to some globally fixed $q$. Since tensor-shaped inputs often have a small number of channels (e.g. 3 for color images), this condition is more severe than the equivalent condition for non-tensor inputs. Further, the co-mean for all pairs of inputs and spatial locations would have to be approximately equal to some globally fixed $c$.

In section \ref{nlcRobustDataSection}, we showed that the covariance structure of the input is important for the value of the NLC in convolutional architectures, but not for fully-connected architectures. This is explained by the mean field theory of CNNs viewing different spatial locations in a single input as distinct inputs.

In future work, we plan to apply heuristics to simplify the mean field recursion for convolutional architectures to obtain practically relevant insights. Now, we present a first attempt at this which establishes that mean field theory is indeed applicable to the analysis of the nonlinearity of convolutional architectures. We greatly simplify the nonlinearity path equation to

$$\mathfrak{n} \approx  \prod_{\tau_l \in f} \mathfrak{n}_{\tau_l}(1,0)$$

\begin{figure}
\centering
\includegraphics[scale=0.8]{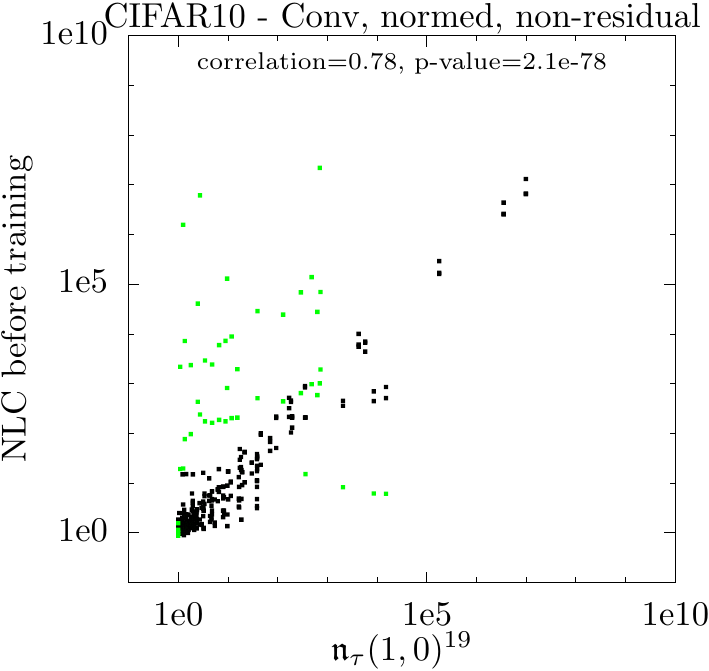}
\caption{Initial NLC vs the product of activation function NLCs with respect to the unit Gaussian, for non-residual study B architectures with debiased activation functions. GUAs are displayed in green and in the foreground. {\it Conclusion:} Despite the gross simplification of the mean field NLC, this estimate is predictive of the NLC.} \label{mfConv}
\end{figure}

We estimate the mean field NLC of the network as the product of activation function NLCs with respect to the unit Gaussian. In figure \ref{mfConv}, we plot $\prod_{\tau_l \in f} \mathfrak{n}_{\tau_l}(1,0)$ vs the initial NLC for some study B architectures in the initial state. Note that all our study B architectures contain 19 instances of the same activation operation. We find \finding{that our rough estimate of $\mathfrak{n}$ is still strongly predictive of the NLC}. This is remarkable as this estimate does not take into account varying $\mathfrak{q}_k$ and $\mathfrak{c}_k$ values, tensor structure, pooling, normalization layers or data augmentation, among other things. Note that we exclude residual architectures in figure \ref{mfConv} as our estimate of $\mathfrak{n}$ cannot account for multiple paths in the layer graph. We also exclude architectures that do not use a debiased activation function so that our simplifying assumption $\mathfrak{c}_k=0$ has a greater chance of working.

Of course, the full nonlinearity path equation still holds for convolutional architectures under the proper technical conditions, though there is now one $\mathfrak{n}_\tau$ value per spatial location and activation layer. The number of paths is exponential in the number of sequentially composed convolutional layers. However, one key property of the nonlinearity path equation still holds. Replacing an activation function with another that has a higher $\mathfrak{n}_{\tau_l}(\mathfrak{q}_k,\mathfrak{c}_k)$ value increases $\mathfrak{n}$, assuming that there are no knock-on effects on downstream $\mathfrak{q}$ and $\mathfrak{c}$ values. We can eliminate knock-on effects by ensuring that $\mathfrak{q}_l$ and $\mathfrak{c}_l$ are not altered by the substitution. This is the insight behind the nonlinearity normalization algorithm which we present in chapter \ref{nlnormChapter}.

While convolutional mean field theory is more cumbersome, it also explains the great performance gains that a convolutional architecture can induce relative to a fully-connected architecture. In figure \ref{nlcPredTestInit}, we find that \finding{CNNs greatly outperform FCNs}. In section \ref{covarianceKernelSection}, we explained how FCNs are determined by their covariance kernel, which is based on the input co-mean. CNNs can have much richer covariance kernels, which tend to map semantically similar images to large kernel values {\it even in the initial state}. CNNs encode information about images before training even begins. \citet{meanFieldCNNGP} showed that fairly high performance can be attained by applying Bayesian inference to the Gaussian process to which a CNN converges. \citet{DNNRandomWeights,DNNRandomWeights1} showed that fairly high performance can be attained by training only the last linear layer of a CNN and retaining the random weights in other layers. \citet{meanFieldCNN,cnnKernel1} investigated the feature representations of CNNs in terms of their frequency spectrum. \citet{cnnKernel2} showed that fully-connected architectures can learn high frequencies present in data much more easily when hard-coded Fourier features are added to the input.

\chapter{Beyond the NLC - explaining the performance of neural architectures} \label{beyondNlcChapter}

In the previous two chapters, we have gone a long way towards establishing ``$1 \le NLC \le 5$'' (sections \ref{nlcDefinitionSection}, \ref{nlcPredictiveSection}) as a ZSAD guideline that fulfills the utility criteria of figure \ref{boxNPM}. We continue this task in this and later chapters. In this chapter, we focus on other ZSAD guidelines, and therefore on other things that can go wrong when designing an architecture. While a large fraction of this work explores the NLC in-depth, the purpose of this chapter is to add breadth. 

How should we choose which guidelines to investigate? Of course, we do not simply want to present a disjointed list of minimally validated concepts. Instead, we use the following criteria to decide what to cover.

\begin{itemize}
\item The guideline has a significant explanatory influence on the test or training error of our study A and B architectures.
\item The guideline influenced the design of the architectures in our studies.
\item The guideline is explainable based on results from prior chapters. For example, we have derived insights from mean field theory to explain the NLC theoretically. In this chapter, we will apply these insights to other guidelines.
\item The guideline is related to the NLC on a conceptual and / or quantitative level.
\item The guideline can be used to explain popular building blocks and design strategies.
\end{itemize}

Many of our ZSAD guidelines fulfill multiple of these criteria.

Of course, we cannot investigate each guideline we cover in this chapter to the same depth as the NLC, due to space and time limitations. To mitigate the limitations of brevity, we draw heavily from analysis presented in prior chapters. To a significant degree, this chapter simply re-interprets prior analysis from the viewpoint of a different guideline. For example, robustness to input distribution and other properties follow from mean field theory like for the NLC in section \ref{nlcExplainSection}. Nonetheless, some of the insights we generate here will have to be fleshed out in future work. Basing some of our guidelines on concrete, well-defined properties is also future work.

Each section from \ref{gaussianStabilitySection} to \ref{noiseStabilitySection} is dedicated to one guideline which is named in the section title. In section \ref{learningRateSection}, we document the importance of exhaustive learning rate tuning and present some intriguing results which might lead to fleshed-out learning rate selection strategies. In section \ref{beyondNlcSummarySection}, we bring the ZSAD guidelines of this chapter and the NLC together to explain the majority of the test and training error variation of our study A and B architectures. By then also relating those guidelines to popular building blocks and design strategies in chapter \ref{surveyChapter}, we hope to make at least designing fully-connected architectures relatively foolproof. For a more detailed guideline-by-guideline overview of this chapter, see section \ref{practicalGuideSection}.

\paragraph{Background from prior chapters} Like sections \ref{nlcPropertiesSection}, \ref{meanFieldDistributionEmpiricalSection} and \ref{meanFieldPracticalEmpiricalSection}, this chapter is based on the empirical studies detailed in chapter \ref{empiricalStudiesChapter}. We recommend reading at least summary section \ref{metricsSummarySection} before proceeding. Throughout this chapter, we use the terminology, notation and conventions of section \ref{notationSummarySection}.

We continue to analyze the NLC in this chapter, which is motivated by chapter \ref{nlcChapter}. A complete understanding of chapter \ref{nlcChapter} beyond section \ref{nlcPredictiveSection} is not required. Throughout this chapter, we make references to mean field theory as studied in chapter \ref{meanFieldNnaChapter}. A high-level understanding of that chapter and an awareness of its key definitions are required to follow such references.

\paragraph{Technical considerations} As discussed in section \ref{nlcDefinitionSection}, we implicitly assume things like integrability (section \ref{integrabilitySection}), differentiability (section \ref{nonDifferentiableSection}) and non-zero denominators when necessary.

\section{Ensure Gaussian stability} \label{gaussianStabilitySection}

\begin{figure}
\centering
\includegraphics[width=0.98\textwidth]{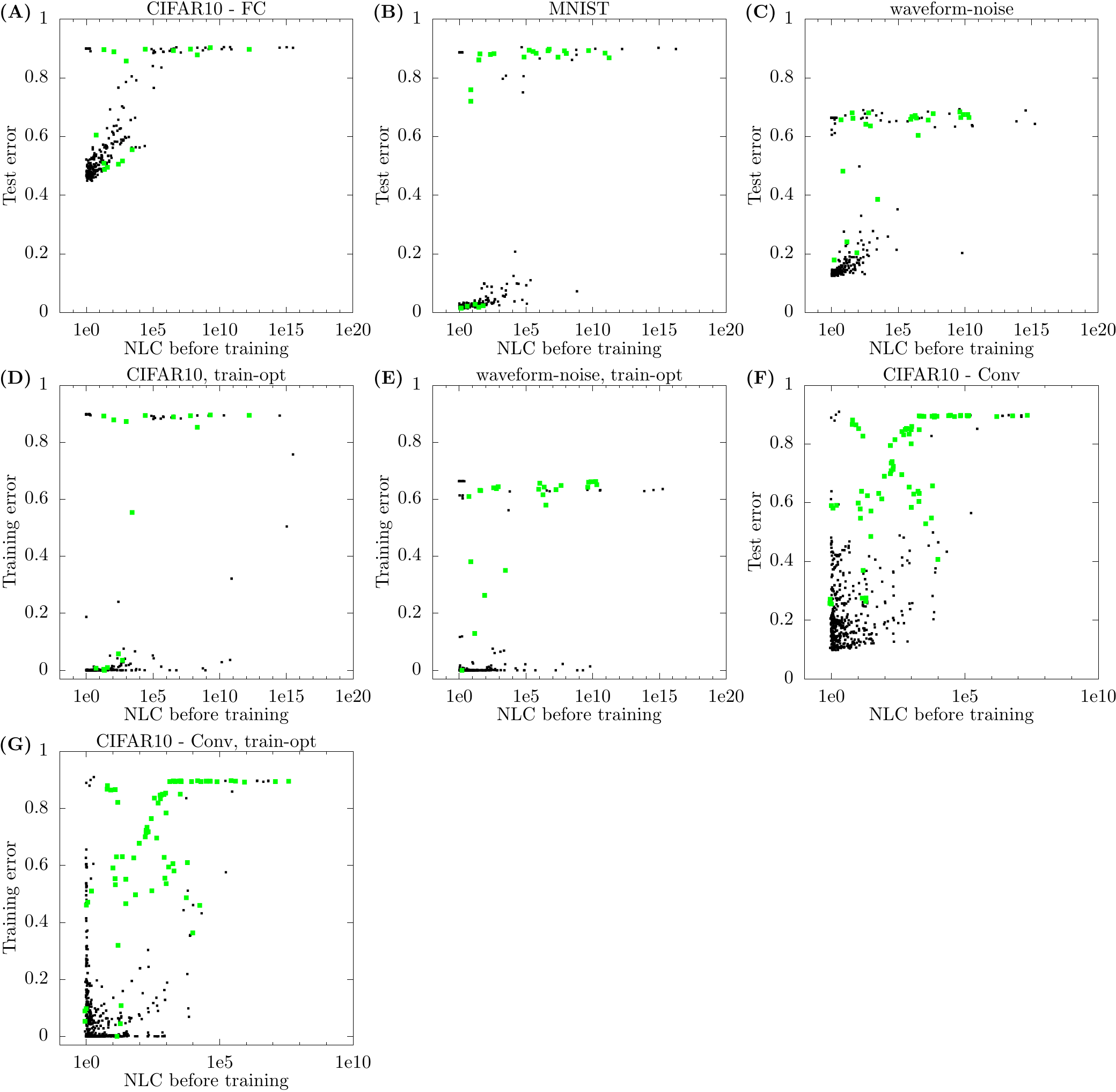}
\caption{Initial NLC vs test error and vs training error after training error minimization for study A and B architectures. Graphs are identical to graphs in figures \ref{nlcPredTestInit} and \ref{nlcPredTrainInit}, except that GUAs are displayed in green and in the foreground. Note that the x-axis range differs significantly between graphs A-E and F-G. Keep in mind that throughout this work, axis ranges can vary when related but not identical metric values are depicted. However, we also strive to keep axis ranges the same for comparability when possible. {\it Conclusion:} In general, GUAs greatly underperform other architectures with a similar NLC.} \label{beyondGaussian}
\end{figure}

Gaussian stability has been important throughout this work. We defined and discussed it in detail in section \ref{gaussianStabilityExplanationSection}. In this section, we add the observation that Gaussian stability is also highly desirable for network performance. In figure \ref{beyondGaussian}A-G, we repeat the following graphs in order: \ref{nlcPredTestInit}A, \ref{nlcPredTestInit}B, \ref{nlcPredTestInit}C, \ref{nlcPredTrainInit}A, \ref{nlcPredTrainInit}B, \ref{nlcPredTestInit}D, \ref{nlcPredTrainInit}B. These depict NLC before training vs test and training error after training for our study A and B architectures. The difference is that this time we display some points in green. Like in many previous figures and all figures throughout this chapter, green markers represent Gaussian unstable architectures (GUAs) as defined at the end of section \ref{metricTerminologySection} / \ref{metricsSummarySection}. Green markers are displayed in the foreground, and can therefore partially or fully occlude other markers. We find that \finding{a large fraction of architectures that perform significantly below what their initial NLC level would suggest are in fact GUAs}. \finding{This holds especially for our convolutional architectures.} Hence, Gaussian stability is not just necessary for accurate mean field estimates, but also for performance. 

\begin{npc} \label{npcGaussian}
`Ensure Gaussian stability' requires that a network exhibits Gaussian stability as defined in section \ref{gaussianStabilityExplanationSection}.
\end{npc}

More work is needed to uncover the mechanisms behind the performance impact of Gaussian instability. A crude interpretation is as follows. Gaussian instability is associated with high LCV values (section \ref{meanFieldDistributionEmpiricalSection}). The gradient with respect to a weight tensor is the product (for FC layers) or convolution (for convolutional layers) of the layer value of the dependency with the gradient with respect to the linear layer. If the layer quadratic means of the dependency differ drastically from one datapoint to another for reasons that are ``random'', i.e. not related to information present in the inputs, then the weight gradient of an entire batch will be dominated by the weight gradients of a small number of datapoints in that batch, leading the learning process to focus entirely on those few datapoints.

Gaussian stability, at least in simple architectures, stems from mean field Gaussian stability, as explained in section \ref{gaussianStabilityExplanationSection}. Being a mean field property, it is robust to data distribution, random initialization and width change as is the NLC, as explained in section \ref{nlcExplainSection}. Hence, the guideline of Gaussian stability is as data-agnostic as the NLC (sections \ref{guidelineDataSection}, \ref{nlcRobustDataSection}) and can be fundamentally regarded as an architecture property (section \ref{nlcRandomInitSection}).

\section{Ensure scale stability, and specifically $LSCALE_l \approx 1$} \label{forwardStabilitySection}

We defined and outlined the ZSAD guideline of scale stability in section \ref{architectureDesignParadigmsSection}. In this section, we analyze it in detail. Note that scale instability denotes simply the absence of scale stability.

Scale stability and Gaussian stability are somewhat similar. Both properties ultimately deal with neuron values. The difference is that scale stability requires that the overall magnitude of neuron values across inputs and neurons is stable, which implies that layer quadratic mean (LQM; table \ref{tableMetricsMfEmpirical}) values are not too large or small overall. On the other hand, Gaussian stability requires that LQM values do not diverge {\it relative to each other}, as this would ``break'' meta-Gaussianity. Gaussian stability is (partially) captured by the LCV metric, whereas scale stability could be captured by e.g. the LSCALE metric (table \ref{tableMetricsMfEmpirical}). In mean field terms, scale instability is associated with small or large $\mathfrak{q}$ values but not necessarily non-Gaussian neuron distributions in linear layers. Conversely, Gaussian instability is associated with inaccurate $\mathfrak{q}$ values and non-Gaussian neuron distributions in linear layers, but $\mathfrak{q}$ itself does not necessarily become large or small. Of course, in the end, the distinction between both concepts is somewhat fuzzy and subjective. 

Whereas Gaussian instability represents a breakdown of mean field theory, scale instability has a natural equivalent within mean field theory.

\begin{figure}
\centering
\includegraphics[width=0.98\textwidth]{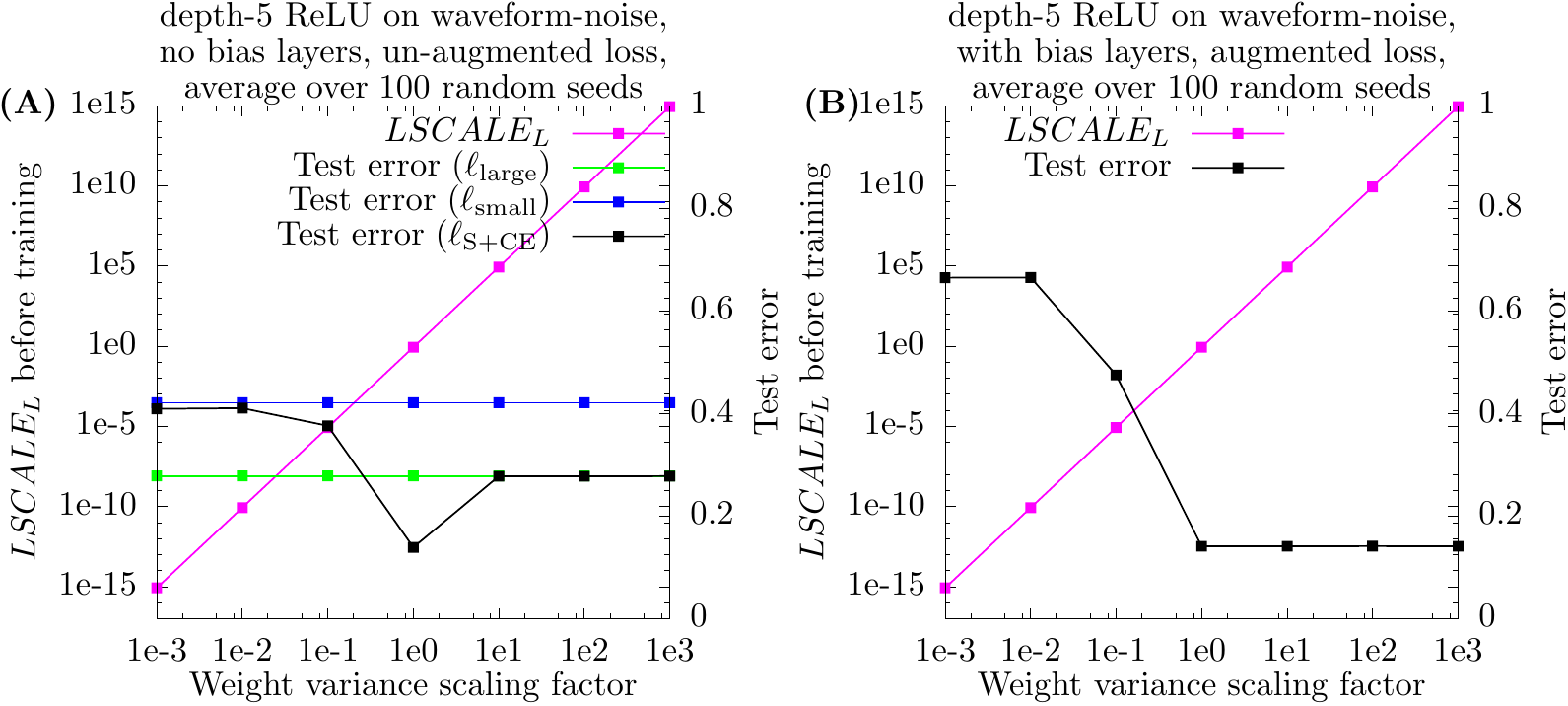}
\caption{Initial $LSCALE_L$ and test error as the initial weight variance and the loss function varies, for depth-5 fully-connected ReLU architectures on waveform-noise. A weight variance scaling factor of 1 corresponds to He initialization. Results are obtained by averaging 100 independent runs corresponding to 100 random seeds. (Each of the 100 runs conducts 40 independent training runs with different starting learning rates.) $LSCALE_L$ is averaged in log space. {\it Conclusion:} Graph A: Softmax+cross-entropy performs optimally when network outputs have approximately unit quadratic mean, and then outperforms the other loss functions. Graph B: The introduction of bias layers causes basic gradient descent to fail for small initial weights.} \label{beyondForward}
\end{figure}

\begin{definition}
We say an A-architecture $f$ exhibits `mean field scale stability' if at all layers $f_l$, $\mathfrak{l}_l = \sqrt{\mathfrak{q}_l}$ is close to 1. $\mathfrak{q}_l$ is defined according to table \ref{tableNLCPropagation}.
\end{definition}

Because scale stability ``has a mean field limit'', like Gaussian stability, it is relatively data-agnostic and can be fundamentally regarded as an architecture property, especially if we also assume that $\mathfrak{q}_0$ is close to 1.

The importance of scale stability stems from deep learning building blocks being designed to work best with LQMs around 1. Below, we discuss three different manifestations of this phenomenon. In section \ref{linearEquivarianceSection}, we show how scale stability becomes irrelevant when the neuron value magnitude is deliberately taken into account during architecture design.

\paragraph{Activation functions tend to work best under scale stability} Consider the $\mathfrak{n}_\tau(\lambda^2, 0)$ curves from tables \ref{covCurveillu1} through \ref{covCurveillu4}. The activation function NLC of many activation functions changes drastically as the $q$ / $\lambda^2$ parameter changes. Hence, the mean field NLC $\hat{\mathfrak{n}}_{l,k}$ of an activation layer can change drastically as the parameter and hence the magnitude of the input to the activation layer $\hat{\mathfrak{q}}_k$ grows. Of course, $\hat{\mathfrak{n}}_\tau\approx \mathfrak{n}_\tau$ is related to the mean field NLC of the architecture via the nonlinearity path equation. The mean field NLC is related to the NLC as shown in section \ref{meanFieldPracticalSection}. Therefore, uncontrolled growth or decay of layer values often corresponds to uncontrolled change in network NLC, and therefore to uncontrolled change in performance.

Many activation functions are designed with the assumption in mind that the LQMs of the dependency are around 1. For example, the sigmoid and tanh activation functions are meant to be continuous versions of the step function (section \ref{historySection}). However, if absolute neuron values in the dependency are much less than 1, they become equivalent to linear functions, i.e. they become `pseudo-linear' (section \ref{pseudoLinearitySection}). If absolute neuron values in the dependency tend to be much greater than 1, they become equivalent to the discrete step function. Similarly, the softplus and Swish activation functions are smooth versions of ReLU. For large inputs, they become ReLU. For small inputs, they become pseudo-linear. Their $\mathfrak{n}_\tau$ values change accordingly, as discussed in section \ref{actFunNLCsection}. It is worth noting that ReLU itself stands out in this regard. $\mathfrak{n}_\text{ReLU}(\lambda^2, 0)$ is independent of $\lambda$ because $\tau_\text{ReLU}(cs) = c\tau_\text{ReLU}(s)$ for all $c > 0$. In this way, ReLU networks are less reliant on scale stability and thus more forgiving of suboptimal design choices.

\paragraph{The softmax+cross-entropy loss function tends to work best under scale stability} Consider the following two alternative loss functions, which we denote by $\ell_\text{large}$ and $\ell_\text{small}$.

\begin{eqnarray*}
\ell_\text{large}(f,y) &=& \max_{i_L}f[i_L] - f[y]\\
\ell_\text{small}(f,y) &=& \mathbb{E}_{i_L}f[i_L] - f[y]
\end{eqnarray*}

Denote the softmax+cross-entropy loss function by $\ell_\text{S+CE}(f,y)$ (section \ref{lossFunctionSection}). It is easy to check that $\ell_\text{S+CE}(cf,y) \approx \ell_\text{large}(cf,y)$ for large $c$ and that $\ell_\text{S+CE}(cf,y) \approx \ell_\text{small}(cf,y)$ for small $c$. A priori, it is not clear whether $\ell_\text{S+CE}$, $\ell_\text{large}$ or $\ell_\text{small}$ is best for any given task or architecture. To gain some empirical understanding, we trained depth-5 ReLU architectures on waveform-noise. Specifically, we trained 7 architectures with different initial weight variances. The first architecture is He initialized, which leads to $LQM_L$ values around 1, as explained in section \ref{architectureDesignParadigmsSection}. We then created 6 additional architectures by multiplying the initial weight variances jointly by $10^{-3}$, $10^{-2}$, $10^{-1}$, $10^1$, $10^2$ and $10^3$ respectively. We ran our full training protocol for each architecture-loss function pair 100 times with different random seeds and averaged results across random seeds.

The results are given in figure \ref{beyondForward}. We find that \finding{for $\ell_\text{large}$ and $\ell_\text{small}$, test error is independent of the weight variance scaling factor}, even though the magnitude of the output, as measured by $LSCALE_L$, differs by many orders of magnitude. Because $\ell_\text{large}(cf,y) = c\ell_\text{large}(f,y)$, $\ell_\text{small}(cf,y) = c\ell_\text{small}(f,y)$ and $\tau_\text{ReLU}(cs) = c\tau_\text{ReLU}(s)$ for $c > 0$, a ReLU network with one of our alternative loss functions is in some sense scale-invariant, which causes the learning dynamics to be preserved as long as the learning rate is adjusted to compensate for the scaling factor. We explore scale invariance further in section \ref{linearEquivarianceSection}. The learning rate adjustment is performed automatically by our training protocol (section \ref{additionalExperimentsSection}). In contrast, we find that for $\ell_\text{S+CE}$ the following happens. \finding{(i) For large scaling factors, the test error is equivalent to $\ell_\text{large}$. (ii) For small scaling factors, the test error is equivalent to $\ell_\text{small}$. (iii) The test error under $\ell_\text{S+CE}$ is minimized by $LSCALE_L\approx 1$.} The first two findings are expected based on the construction of our loss functions. The third finding implies that $\ell_\text{S+CE}$ achieves the lowest test error out of the three loss functions. Again, while this is not clear a priori, it is somewhat expected given that $\ell_\text{S+CE}$ is universally popular but $\ell_\text{large}$ and $\ell_\text{small}$ are not. However, the gap in performance is surprising. The error under $\ell_\text{small}$ is more than twice as high as under $\ell_\text{S+CE}$! Hence, scale stability at the output layer can be necessary for high performance under $\ell_\text{S+CE}$.

\paragraph{Having a single learning rate for all layers tends to work best under scale stability} One of the key properties of gradient descent is that the same learning rate is applied to the gradient of each parameter component for a given update. While this choice is in some sense natural, it is not necessarily optimal. Classic optimization algorithms such as Newton's algorithm and natural gradient descent can be viewed as modulating learning in different directions in parameter space based on the local shape of the objective function, which can greatly speed up learning. In the deep learning field, parameter sub-vectors of different layers can have very different meanings and properties. The Adam algorithm, which we used in section \ref{nlcPredictiveSection} and defined in section \ref{trainingAlgorithmsSection}, is the most popular way of adapting the learning rate of individual parameter components.

Scale instability can impede learning when a single learning rate is used for all parameter components. To show this empirically, again, we train depth-5 ReLU architectures on waveform-noise with different initial weight variances, using SGD as usual. This time, we use the augmented loss function from study A (section \ref{studyATrainingSection}) which normalizes the network output before applying softmax+cross-entropy, thereby eliminating the influence of the weight variance scaling factor on performance via the loss function. Additionally, we now introduce bias layers into the architecture, so that a macro-layer is composed of a fully-connected layer, a bias layer and a ReLU layer.

In figure \ref{beyondForward}B, we give the results. Since the bias vectors are initialized to zero, we find that \finding{$LSCALE_L$ before training is the same as in graph A}. Also, when straight He initialization is used, \finding{the test error is similar to graph A}. In contrast to graph A, when the weight variance scaling factor is larger than 1, \finding{we find that the test error does not increase}. This is because we use the augmented loss function. When the scaling factor is less than 1, \finding{we find that the error increases to random levels}. This is because of the following effect. As the weight variance decreases, so does the length of the layer values. The gradient with respect to the weight matrices is multiplicatively dependent on layer values, whereas the gradient with respect to the bias vectors is not. Hence, as the scaling factor decreases, the gradient with respect to bias vectors increases relative to the gradient with respect to weight matrices. In fact, basic arithmetic yields that the gradient with respect to the bias vector in the last macro-layer is multiplied by $c^{-5}$ relative to the weight matrix gradients, where $c$ is the scaling factor. Hence, as $c$ decreases, only the bias vectors are trained. Since the majority of the architecture's modeling capacity is contained in the weight matrices, the error increases.

Again, using basic arithmetic, we find that it is trivial to circumvent this pathology by employing layer-specific learning rates. We formalize this in section \ref{linearEquivarianceSection}. For now, we note that gradient descent can have a preference for layer values of non-extreme magnitude.

We end by formalizing the ZSAD guideline of scale stability by extending our definition from section \ref{architectureDesignParadigmsSection}. Based on (i) the analysis of this section, (ii) the mean field limit given in theorem \ref{mfntPropagation} and (iii) simplicity, we recommend that scale stability be measured by the LSCALE metric. Since scale stability is not as deep and meaningful a property as nonlinearity, there is likely no ``single correct'' metric for determining its presence.

\begin{npc} \label{npcForward}
`Ensure scale stability' requires that at each layer in a network $f$, the overall magnitude of neuron values across inputs and neurons is not excessively large or small. Specifically, we advocate measuring this via $LSCALE_l \approx 1$ at each layer $f_l$, where LSCALE is the quadratic expectation of layer quadratic means as defined in table \ref{tableMetricsMfEmpirical}.
\end{npc}

\section{Ensure training stability} \label{covariateShiftSection} 

In this work, we focus primarily on studying the impact of properties of the initial state on performance. However, we have shown that properties of the final state can, in a sense, be even more important. For example, in figure \ref{nlcPredTestFinal}, we showed that the range of NLC values that an architecture can attain and still achieve high or even non-random performance is narrower after than before training. We show a similar property for LBIAS in the next section. This makes sense given that the ultimate goal is to obtain a network function that approximates the true input-label function, and therefore its properties, in the final state. If we ensure that desirable properties hold in the initial state, we need to ensure they do not disappear during training. This leads to the following ZSAD guideline.

\begin{npc} \label{npcTraining}
`Ensure training stability' requires that the value of metrics or the presence of properties that either directly or indirectly influence architecture performance, such as LSCALE, LCV, NLC, LBIAS, Gaussian stability or noise stability, are not prone to change in an uncontrolled or harmful way during training for an architecture $f$.
\end{npc}

Of course, the specificity of this guideline for architecture choice in particular is limited, because the extent of the change during training does depend greatly on training algorithm and learning rate. However, some architectures are indeed more resilient to such changes than others.

Consider again the impact of LQMs on the NLC. In section \ref{meanFieldPracticalEmpiricalSection}, we investigated how the mean field NLC $\hat{\mathfrak{n}}$ changes during training in study A architectures. Its change is dependent on the change of the parameter magnitude. We found that the parameter vector usually grows during training because of the updates that are added. This growth can be minuscule or enormous depending on the learning rate used among other factors. The $\mathfrak{n}_\tau(\lambda^2, 0)$ curves from tables \ref{covCurveillu1} through \ref{covCurveillu4} show the potential impact of parameter growth on the network NLC as discussed in the previous section. Again, we find that \finding{tanh and sigmoid are inherently susceptible whereas ReLU is not}. From table \ref{tableNLCPropagation}, we also find that \finding{using BN before every activation layer eliminates the possibility of $\hat{\mathfrak{q}}_k$ or $\hat{\mathfrak{c}}_k$ changing, and therefore keeps the mean field NLC constant}. \citet{batchNormalization} advocated BN especially in the presence of tanh or sigmoid to eliminate what they term `covariate shift'. We further discuss this in section \ref{surveyBNsection}. Our concept of training stability certainly builds upon and is similar to covariate shift. The main difference is that while training stability requires the preservation of properties that have been shown to influence performance, we are not aware of a general argument as to why preserving the entire layer distribution is important, which is implied by covariate shift.

The harmful impact of parameter growth in sigmoid and tanh networks is widely known as the ``exploding / vanishing gradient problem in sigmoid / tanh networks'', which we further discuss in section \ref{vanishTanhSection}. Finally, we note that the most extreme cases of training instability, like the most extreme cases of scale instability and Gaussian instability, can induce numerical overflow and hence training failure.

\section{Avoid neuron bias, and specifically ensure $LBIAS_l \approx 0$} \label{outputBiasSection}

We outlined the ZSAD guideline of avoiding neuron bias in section \ref{architectureDesignParadigmsSection}. In this section, we analyze it in detail. We measure it via the LBIAS metric. Like the NLC, LBIAS is highly predictive of performance. However, we show that in contrast to the NLC, it is possible to achieve significantly better-than-random test error with high LBIAS if the loss function, error function and training algorithm are somewhat modified. In section \ref{plainArchitectureSection}, we further show that NLC and LBIAS form a ``dichotomy'' when certain architecture design strategies are used. In chapter \ref{relatedWorkChapter}, we compare the NLC / LBIAS dichotomy to the exploding gradient / vanishing gradient and order / chaos dichotomies.

Like many other ZSAD guidelines, `avoid neuron bias' is not well-defined. Specifically, (i) it is not clear at what level neuron biases are pathological and (ii) it is not clear how to aggregate the biases of individual neurons to determine the ``overall bias level'' of a network.

With regards to (i), as usual, we must consider the bias {\it relative} to the overall magnitude of that which is biased, i.e. relative to the overall magnitude of neuron values. The meaning of having, say, $\mathbb{E}_xf_l[i_l] = 1$ for some $(l, i_l)$ pair is very different if we also have $\mathbb{S}_xf_l[i_l] = 0.001$ vs if we also have $\mathbb{S}_xf_l[i_l] = 1000$. ($\mathbb{S}$ denotes the standard deviation operator as defined in section \ref{metricEstimationSection}.) The absolute value of neuron expectations can be manipulated via e.g. the initial weight variance of linear layers. In a vacuum, changing this weight variance clearly should not affect our judgment of whether a network is biased.

With regards to (ii), we believe that there are multiple ways of aggregating neuron biases that may highlight different underlying pathologies. One of the key explanations for the predictiveness of the NLC is that we need to learn a network that has a nonlinearity level comparable to that of the true input-label function. We argued in section \ref{whatIsNonlinearitySection} that most practical true input-label functions have a relatively low nonlinearity level. If we apply the same reasoning to bias, then we should care about the biasedness of the network function, and therefore specifically about the biasedness of the output neurons as opposed to other neurons in the network.

These arguments lead to the following definitions.

\begin{metricDefinition}
The `layer bias' (LBIAS) is
$$LBIAS_l(f, \mathcal{D}) = \frac{\sqrt{\mathbb{E}_x||f_l(x)||_2^2}}{||\mathbb{S}_xf_l(x)||_2}$$

Further, we shorten $LBIAS_L$ to $LBIAS$. This is equivalent to how $NLC$ is the short form of $NLC_{L,0}$ defined in table \ref{tableMetricsMfEmpirical}.
\end{metricDefinition}

Further, we refer to neuron biases at the output layer as `output bias'.

$LBIAS_l$ measures the magnitude of neuron values relative to the variation of neuron values in a layer. An alternative definition is $\frac{\sqrt{||\mathbb{E}_xf_l||_2^2 + ||\mathbb{S}_xf_l||_2^2}}{||\mathbb{S}_xf_l||_2}$. So if neurons in a layer are unbiased, i.e. $\mathbb{E}_xf_l=0$, then $LBIAS_l = 1$. In contrast, if neurons in a layer are highly biased, i.e. the coefficients of variation of neuron values are small, then $LBIAS_l$ is large. By definition, $LBIAS_l \ge 1$ always holds. Note that the denominators of LBIAS and NLC are identical. LBIAS is also the ratio of the LSCALE and QMNSTD metrics defined in table \ref{tableMetricsMfEmpirical}. Like the NLC, LBIAS does not require significant assumptions beyond those inherent in the functional-gradient paradigm, though, unlike the NLC, it does not even require a gradient or local linear approximation. See section \ref{nlcFunctionalGradientSection}.

\begin{figure}[H]
\centering
\includegraphics[width=0.98\textwidth]{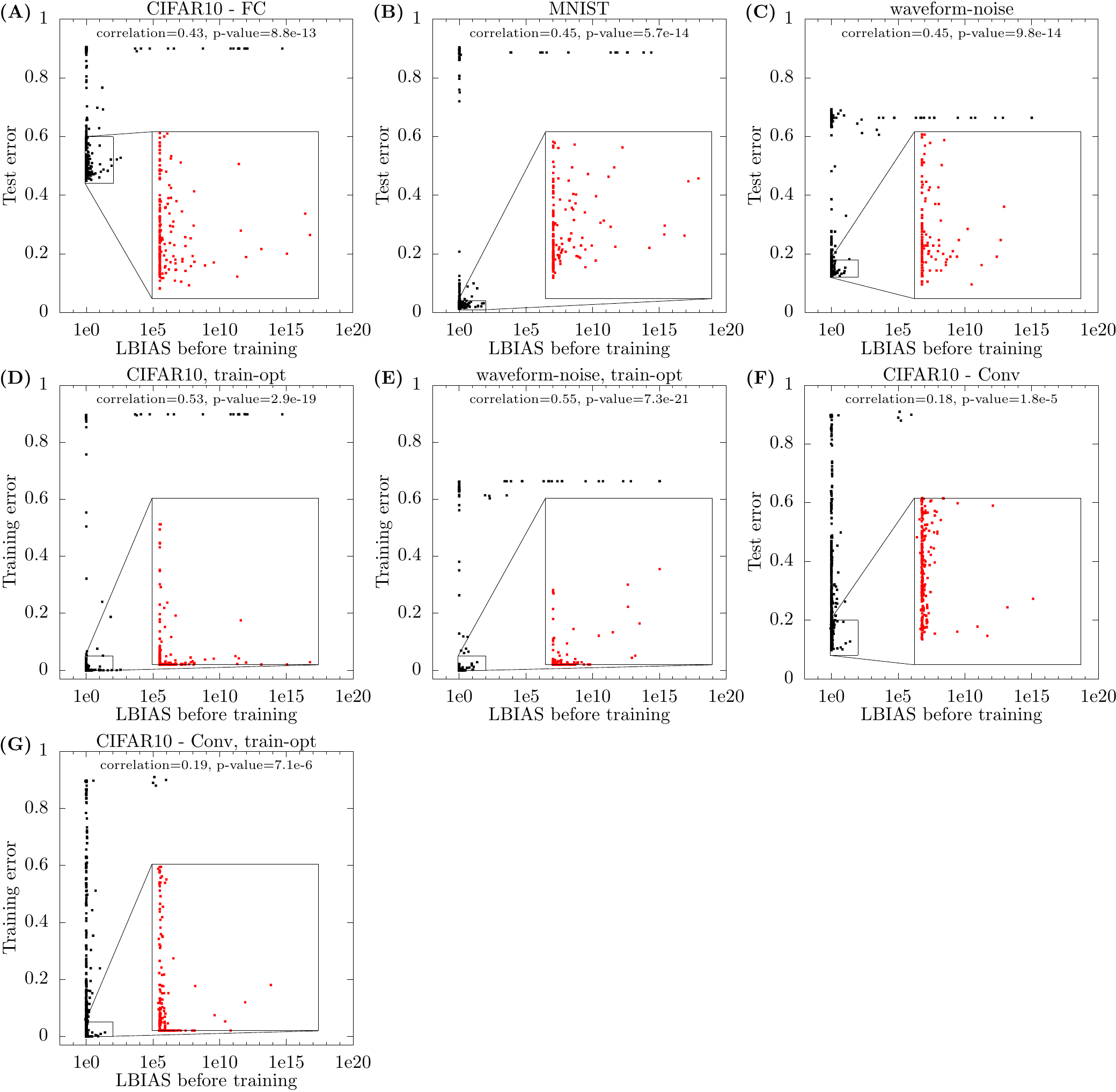}
\caption{Initial LBIAS vs test and training error for study A and B architectures. Inset graphs in the bottom right are magnifications of the region $0.8 < LBIAS < 100$. {\it Conclusion:} The LBIAS of an architecture, when evaluated in the initial state before training, is a powerful predictor of test and training error after training and attaining a small LBIAS value is essential for attaining an optimal test or training error.} \label{beyondBiasInit}
\end{figure}

\begin{figure}[H]
\centering
\includegraphics[width=0.98\textwidth]{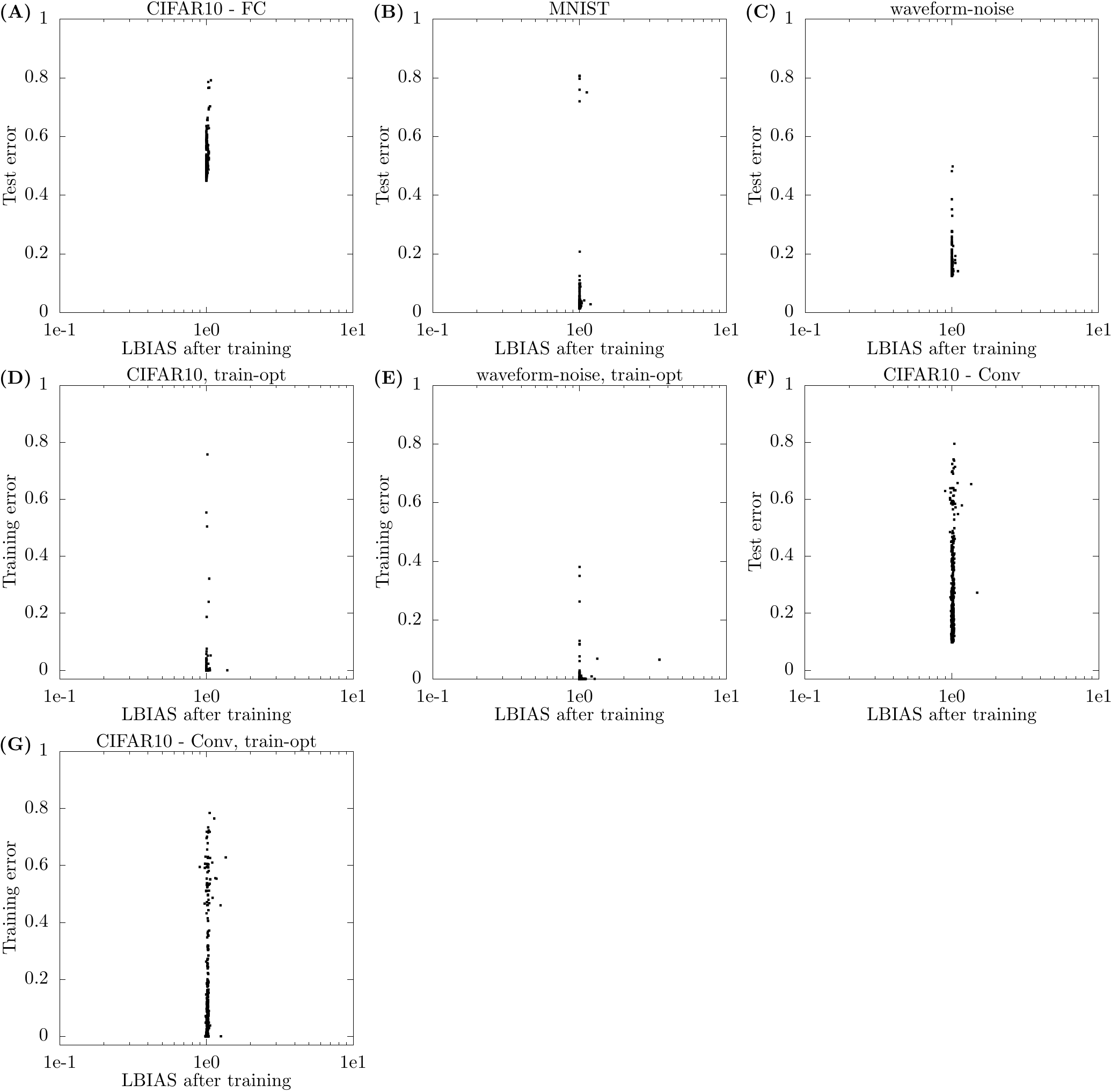}
\caption{Final LBIAS vs test and training error for study A and B architectures. {\it Conclusion:} All architectures that attain a better-than-random test or training error also attain a near-1 LBIAS after training.} \label{beyondBiasFinal}
\end{figure}

\paragraph{LBIAS is predictive of performance} In figure \ref{beyondBiasInit}, we plot the LBIAS of our study A and B architectures before training vs test error and training error. The picture obtained is similar in many ways to the one from figures \ref{nlcPredTestInit} and \ref{nlcPredTrainInit}, where we plotted the NLC vs error. \finding{We find that a large initial LBIAS leads architectures to attain a random test error in all cases. The cutoff point differs between datasets, but is always below 1000. To attain an optimal test error, an initial LBIAS of less than 10 is required. In contrast to the NLC, there are no high-LBIAS architectures that attain a better-than-random training error}. In figure \ref{beyondBiasFinal}, we plot LBIAS after training vs test error and training error. \finding{All architectures that attain a better-than-random error attain an LBIAS that is very close to 1.} In figure \ref{beyondBiasFinal} as throughout this work, whenever we plot values for metrics not based on error for architectures in the final state, we omit all architectures that did not attain a better-than-random error.

We can interpret this as follows. A high LBIAS indicates that all network outputs are relatively similar, i.e. close to some fixed non-zero expectation. If the network is randomly initialized, the output expectation would also be random. By our analysis from section \ref{meanFieldDistributionSection}, if the outputs are returned by a linear layer, the expectation would be approximately elementwise Gaussian. In general, we would expect the expectation to be non-constant across output neurons. Hence, the neuron with the highest activation is likely to be the same for all outputs, and so all inputs are assigned to the same class by the network. So at least in the initial state, networks with high LBIAS tend to assign all inputs to the same class. As with the NLC, in order for a network to have high performance, its LBIAS value should not differ too much from the true input-label function. All our datasets had roughly equal class frequency. Hence, to attain low test or training error, our architectures must assign a high value to each output neuron roughly equally often, and this roughly corresponds to low LBIAS. Hence, architectures must either ``unlearn'' a high initial LBIAS or will have random test and training error after training. This is what we find in figure \ref{beyondBiasFinal}. In figure \ref{beyondBiasInit}, we find that this happens in practice only when the initial LBIAS is not too high. This behavior is very similar to the behavior found for the NLC in sections \ref{nlcPredictiveSection} and \ref{nlcEvolutionSection}.

The initial LBIAS, along with the test and training error, for our individual architectures is given in the appendix in chapter \ref{fullListChapter}.

\paragraph{$LBIAS_l$ and mean field theory} The mean field limit of $LBIAS_l$ according to theorem \ref{mfntPropagation} is $\sqrt{\frac{\mathfrak{q}_l}{\mathfrak{q}_l - \mathfrak{c}_l}}$. We have shown that the mean field estimate of both the numerator and denominator are accurate for non-GUAs in figures \ref{mfPredqf} and \ref{mfPredqsigma}. Hence, the estimate of the ratio is also accurate. Note that not only $LBIAS_l$ has such a simple mean field limit, but also other metrics that could be reasonably used to measure neuron bias. Hence, neuron bias is relatively data-agnostic and can be fundamentally regarded as an architecture property, especially if we also assume that $\frac{\mathfrak{c}_0}{\mathfrak{q}_0}$ is close to 0.

In section \ref{actFunNLCsection}, we studied the mean field NLC of plain architectures as depth increases. We can do the same for the `mean field LBIAS'. Consider a plain stable$(q,1)$ architecture. There, $\mathfrak{c}_L = q\tilde{\mathfrak{C}}_\tau^M(\mathfrak{c}_0)$ and $\mathfrak{q}_L = q$. $\tilde{\mathfrak{C}}_\tau = \frac{\mathfrak{C}_\tau}{\mathfrak{C}_\tau(1)}$ is depicted in tables \ref{covCurveillu1} through \ref{covCurveillu4}. Hence, as $M$ increases, mean field LBIAS converges to $\sqrt{\frac{1}{1-c^\text{lim}}}$, where $c^\text{lim}$ is the stable fixed point of $\tilde{\mathfrak{C}}_\tau$. The convergence rate directly follows from the convergence rate of $\tilde{\mathfrak{C}}_\tau$ towards $c^\text{lim}$.

\begin{figure}[H]
\centering
\includegraphics[width=0.98\textwidth]{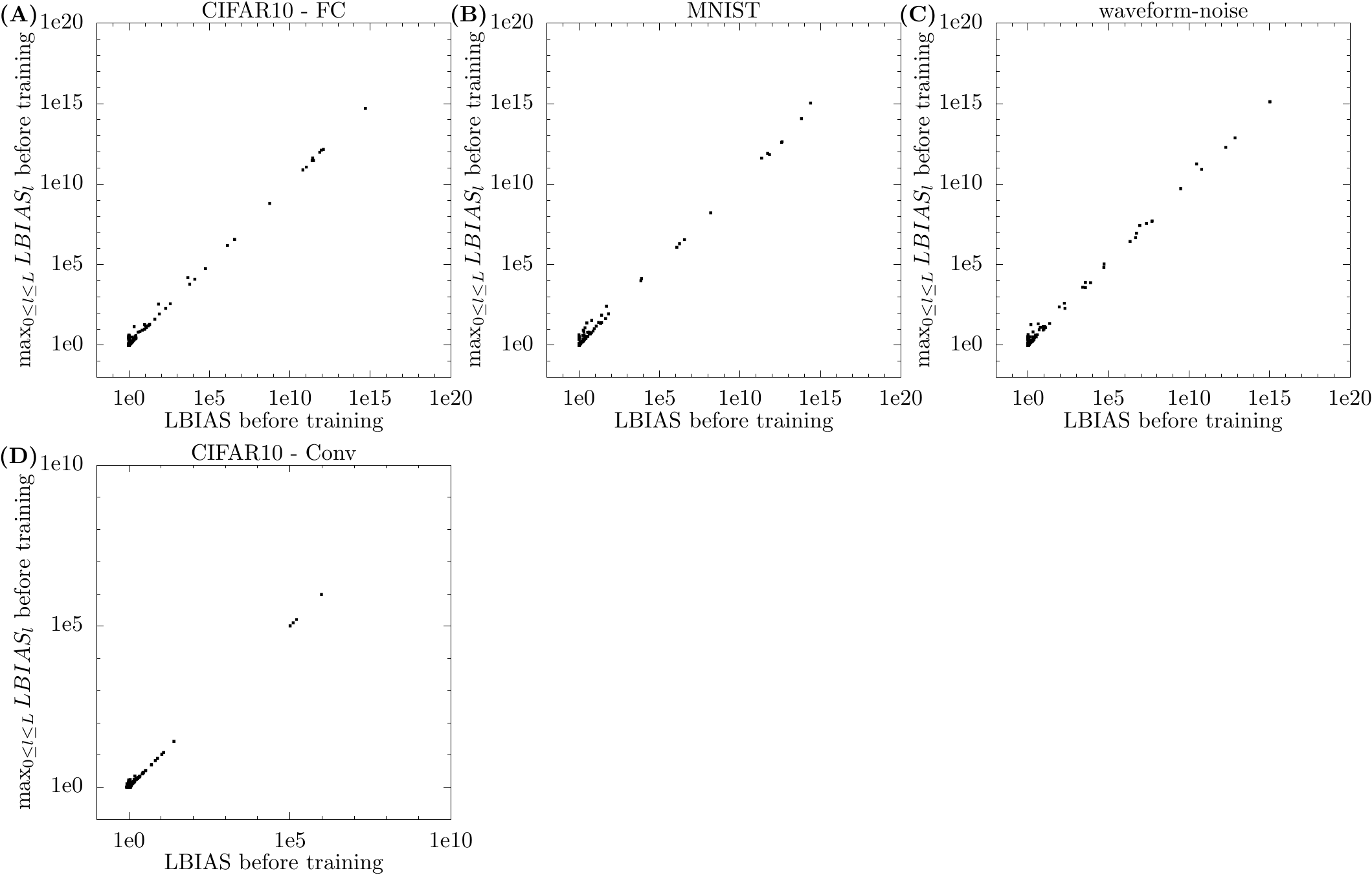}
\caption{LBIAS vs the maximum $LBIAS_l$ value across all layers for study A and B architectures in the initial state. Correlations are close to 1. {\it Conclusion:} The two values are functionally equivalent for our purposes.} \label{beyondBiasMaxInit}
\end{figure}

\begin{figure}[H]
\centering
\includegraphics[width=0.98\textwidth]{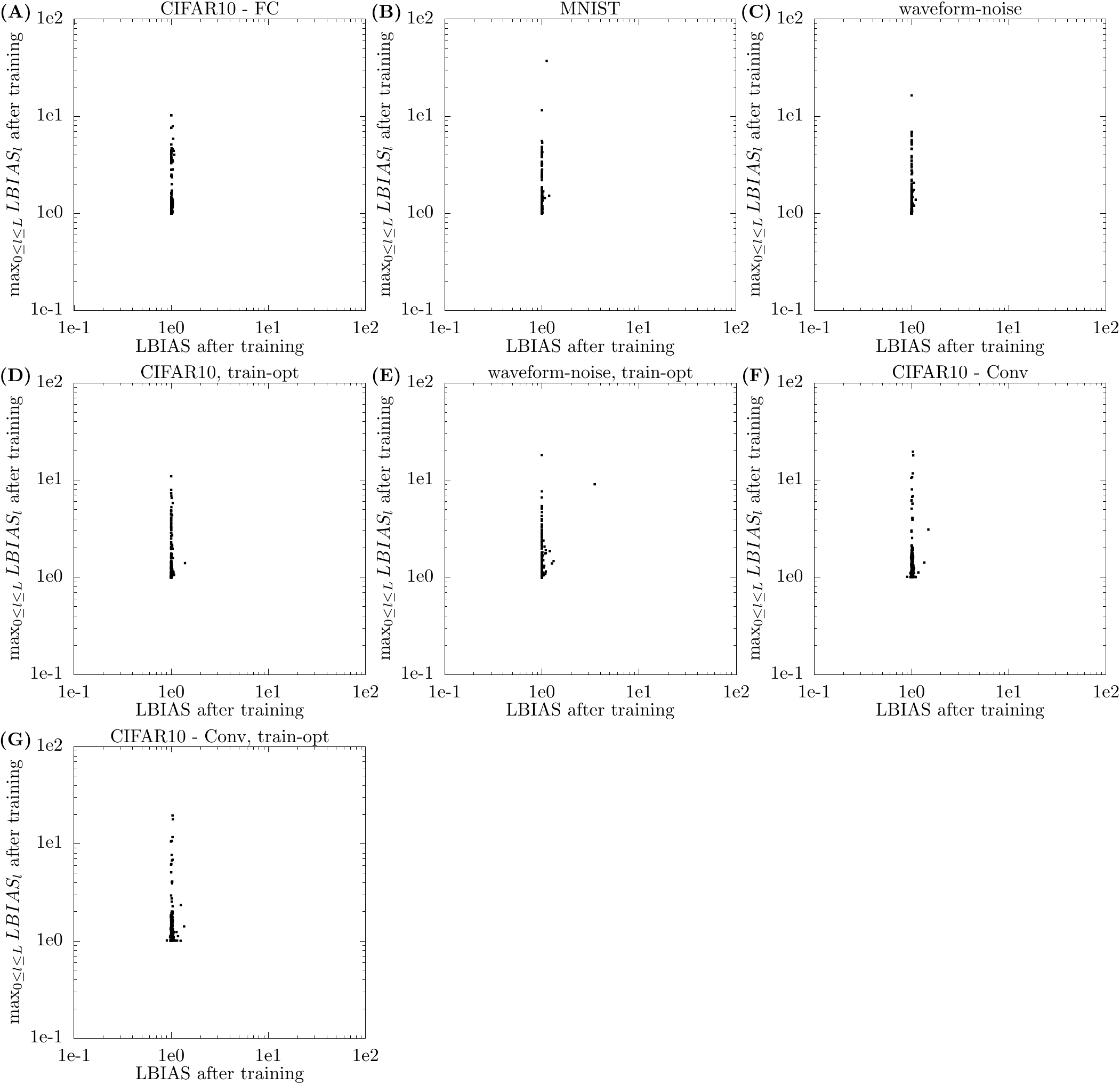}
\caption{LBIAS vs the maximum $LBIAS_l$ value across all layers for study A and B architectures in the final state. {\it Conclusion:} While LBIAS is always close to 1 for architectures that achieve a better-than-random error, we observe $\max_l LBIAS_l$ values up to around 50.} \label{beyondBiasMaxFinal}
\end{figure}

One interesting consequence of this is that at least in very simple architectures, if the input is processed to have expectation zero and hence $\mathfrak{c}_0=0$, mean field $LBIAS_l$ tends to either remain at one or increase from macro-layer to macro-layer. In figure \ref{beyondBiasMaxInit}, we plot $LBIAS=LBIAS_L$ vs the highest $LBIAS_l$ value across all layers in the initial state. Indeed, we find that \finding{across all our study A and B architectures, $LBIAS_l$ never exceeds LBIAS for any $l$ in the initial state by more than around an order of magnitude}. Hence, LBIAS can, at least in simple architectures, be regarded as a meaningful ``upper bound'' of the biasedness of the whole initial state network. So for the purpose of this work, we will always use LBIAS as a measure for the `avoid neuron bias' guideline in general, not just for output bias. In figure \ref{beyondBiasMaxFinal}, we make the same comparison in the final state. \finding{While LBIAS is always close to 1, $\max_l LBIAS_l$ can range up to 50.}

\paragraph{Architectures with high $LBIAS_l$ values can be trained if we adjust the training protocol} If the only meaningful difference between our architectures with output bias and our other architectures is that, well, their output is biased, then we should be able to eliminate the performance hit by simply normalizing the output expectation. For study A architectures, we used an augmented loss function that normalizes the magnitude of the network output. It evaluates $LSCALE_L$ before training on the training set and then divides all network outputs by that scalar value before applying softmax+cross-entropy, as described in section \ref{studyATrainingSection}. We could apply the same strategy for the output expectation.

\begin{definition}
An architecture is trained and evaluated with `output debiasing' if the following augmented loss and error functions are used. Before training, $\mathbb{E}_{x \in D_\text{train}}f(\theta^{(0)},x)$ is evaluated. Then, loss and error function subtract that fixed value from the output before applying softmax+cross-entropy loss / classification error respectively.
\end{definition}

\begin{figure}[H]
\centering
\includegraphics[width=0.98\textwidth]{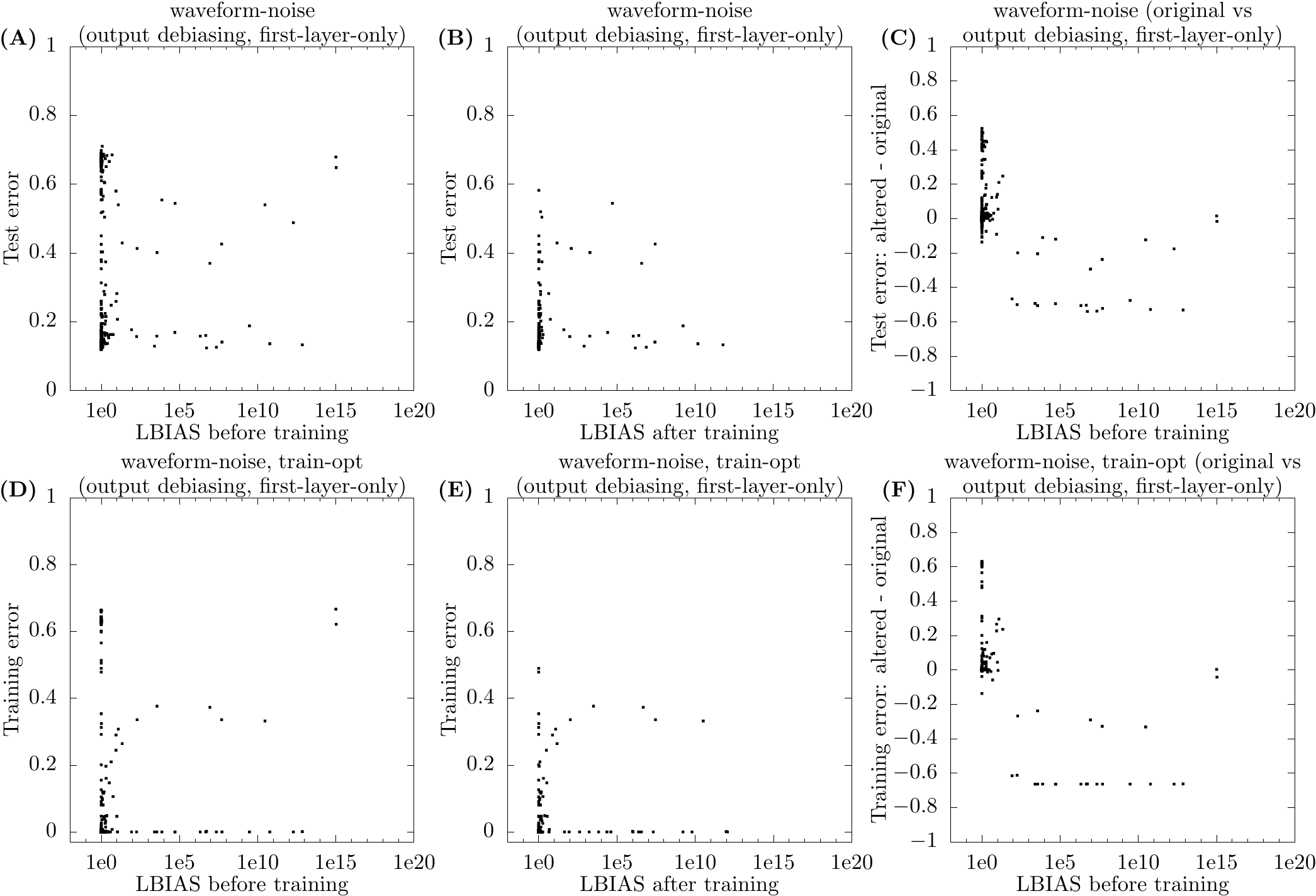}
\caption{LBIAS vs error for study A waveform-noise architectures trained with output debiasing and first-layer-only training. In graphs C and F, we plot LBIAS vs the error difference between the modified training protocol and training with the original study A protocol. The ``original'' error values are the same as in figure \ref{beyondBiasInit}. {\it Conclusion:} Output debiasing and first-layer-only training enable training and generalization for architectures with high initial LBIAS values and do not force those architectures to attain low LBIAS values in the final state.} \label{beyondFirst}
\end{figure}

Using our study A protocol plus output debiasing, we find that \finding{there is no significant improvement in the test or training error of biased architectures}. We do not depict the results. This suggests that while LBIAS is predictive of performance, it does not tell the whole story. What if bias at intermediate layers disrupts the learning process? One way to test this hypothesis is to train our architectures by only applying gradient updates to the first fully-connected layer, which is not biased, and keeping all other parameter sub-vectors fixed. We trained our waveform-noise architectures with a modified study A protocol that includes this `first-layer-only training' and output debiasing. We give the results in figure \ref{beyondFirst}. In graphs A-C, we plot results for test error. In graphs D-F, we plot results for training error after training error minimization. In graphs A and D, we plot LBIAS before training vs the error achieved. In graphs B and E, we plot LBIAS after training vs the error achieved. In all cases, we find that \finding{architectures with high LBIAS, and thus high $LBIAS_l$ values overall, can now achieve error levels that are somewhat comparable to other architectures}. In graphs D and F, we plot LBIAS before training vs the reduction in test error achieved by training with our modified protocol vs our original protocol. We find that \finding{while the error of high-bias architectures is almost always reduced, there are some relatively unbiased architectures that experience a significant to large error increase}. This is unsurprising when only one instead of all layers is trained.

Is there a way to train all layers despite bias? Let's consider a ``unit'' of two layers consisting of a linear layer followed by a bias layer: $f_l = f_kW_l + \beta_l$. This unit can be rewritten as follows: $f_l = (f_k - \bar{f}_k^{(0)})W_l + (\beta_l + \bar{f}_k^{(0)}W_l)$, where $\bar{f}_k^{(0)}$ is short for $\mathbb{E}_xf_k(\theta^{(0)},x)$. The first of the two terms has expectation zero, at least in the initial state. The idea is now to only train the first of the two terms while leaving the second term unchanged, thereby eliminating the influence of the bias on training. We can apply this idea to our study A architectures because all trainable parameter sub-vectors are contained in units of an FC and a bias layer. Let's see what this looks like in practice. We have $\frac{d\ell}{dW_l} = f_k^Tg_l$. If we write $f_l = (f_k - \bar{f}_k^{(0)})W_l' + (\beta_l + \bar{f}_k^{(0)}W_l'')$, then $\frac{d\ell}{dW'_l} = (f_k-\bar{f}_k^{(0)})^Tg_l$. If we apply an update based on this gradient to the single copy of the weight matrix stored in memory, this update would affect both the value of $W_l'$ and $W_l''$. However, we can compensate for this by applying a surrogate gradient $\bar{f}_k^{(0)}(f_k-\bar{f}_k^{(0)})^Tg_l$ to $\beta_l$. The resulting update to $\beta_l$ would cancel the change to $W_l''$, at least under SGD. Hence, we define the following algorithm.

\begin{definition}
`Debiased gradient descent' (DGD) is the training algorithm that performs the following updates for each unit $f_l = f_kW_l+\beta_l$ at each iteration.

\begin{eqnarray*}
\mathsf{grad} &=& \mathbb{E}_{(x,y)\in |B|}(f_k(\theta^{(t-1)},x)-\bar{f}_k^{(0)})^T\frac{d\ell(f(\theta^{(t-1)},x),y)}{df_l}\\
W_l^{(t)} &=& W_l^{(t-1)} - \alpha^{(t)}\mathsf{grad}\\
\beta_l^{(t)} &=& \beta_l^{(t-1)} + \alpha^{(t)}\bar{f}_k^{(0)}\mathsf{grad}\\
\end{eqnarray*}

$\alpha^{(t)}$ is the learning rate. DGD is not defined if non-empty parameter sub-vectors arise outside of FC-bias units. 
\end{definition}

\begin{figure}[H]
\centering
\includegraphics[width=0.98\textwidth]{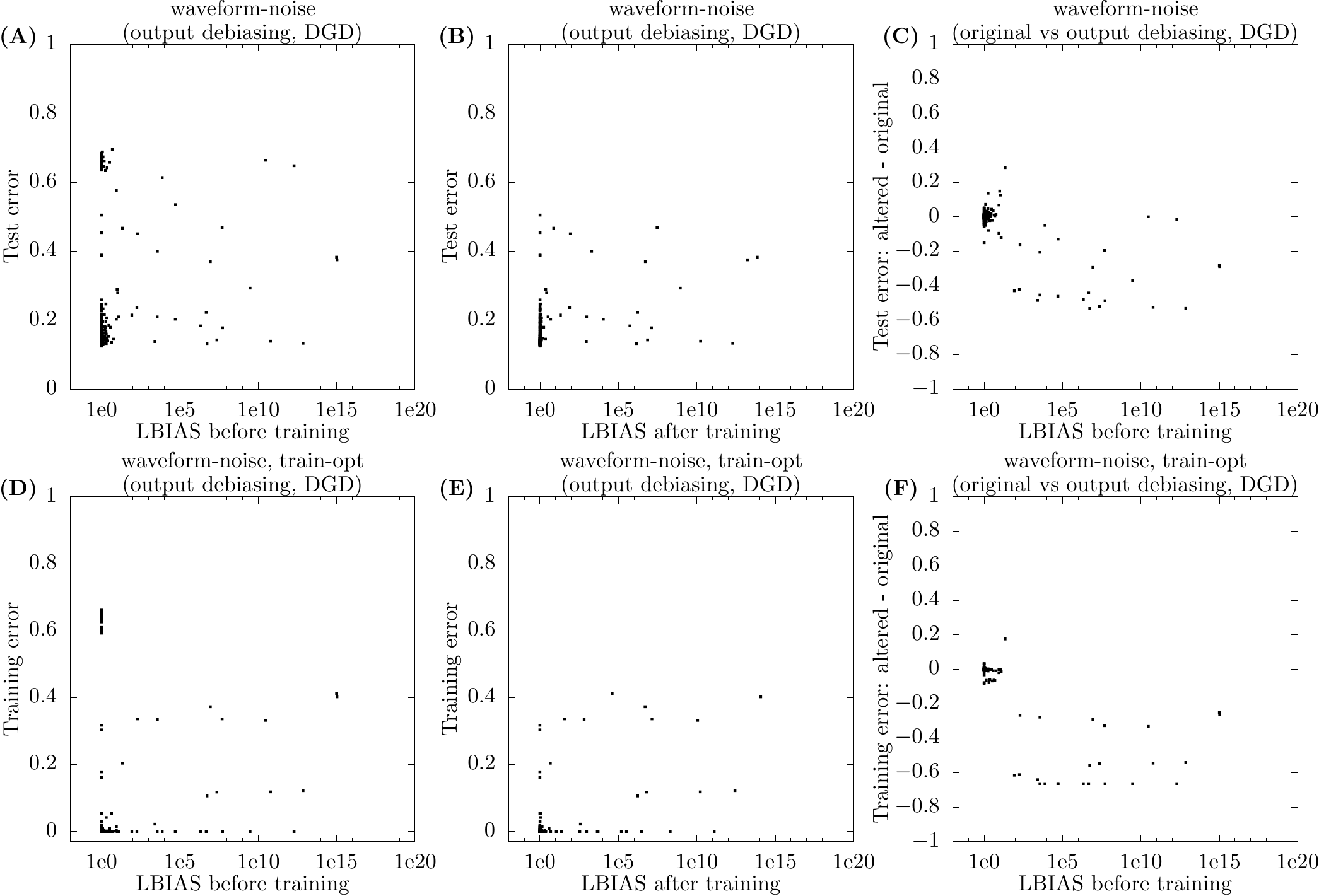}
\caption{LBIAS vs error for study A waveform-noise architectures trained with output debiasing and DGD. Graphs are equivalent to figure \ref{beyondFirst}. {\it Conclusion:} Output debiasing and DGD enable training and generalization for architectures with high initial LBIAS values and do not force those architectures to attain low LBIAS values in the final state. Additionally, they do not greatly harm the performance of unbiased architectures.} \label{beyondDGD}
\end{figure}

We give the results of training with our study A protocol where SGD was swapped for DGD and output debiasing was used in figure \ref{beyondDGD}. Individual graphs are equivalent to figure \ref{beyondFirst}. \finding{For architectures with high LBIAS, we obtain similar results compared to first-layer-only training. However, for architectures with low LBIAS, we now very rarely see a large increase in error, as desired.}

Why does DGD work? While investigating this point in detail goes beyond the scope of this work, we speculate that the negative impact of neuron bias arises because the gradient $f_k^Tg_l$ is dominated by the expectation of $f_k$ if it has greater magnitude than the variation of $f_k$. However, the expectation is the same for each input, and hence the learning itself is not responsive to the value of the input. In the DGD update, the term $(f_k^{(t)}-\bar{f}_k^{(0)})^T$ does differ significantly from input to input, at least in the initial state.

The effectiveness of DGD reveals a fundamental difference between NLC and $LBIAS_l$. While nonlinearity is an indicator of fundamental compatibility (or lack thereof) between the true input-label function and the network function, neuron bias is more of a nuisance factor that can be greatly mitigated by proper conditioning of the loss function, error function and training algorithm. In this sense, `avoid neuron bias' is similar to the `ensure scale stability' guideline, as we show in section \ref{linearEquivarianceSection}. Of course, there is a caveat in that we cannot show that architectures with high NLC can never achieve high performance under any reasonable training protocol. However, it seems that such a protocol would have to at least be capable of drastically altering the network function. If this is the case, it might just be easier to choose an architecture with a good NLC to begin with.

\paragraph{Summary} While DGD and output debiasing can mitigate the effects of large $LBIAS_l$, we still conclude that avoiding excessive $LBIAS_l$ values is desirable, at least in the context of current designs. Another reason for this is the relationship between neuron bias and pseudo-linearity discussed in section \ref{pseudoLinearitySection}. However, more work is necessary to fully understand the nature of the pathologies involved. We formalize the ZSAD guideline below.

\begin{npc} \label{npcBias}
`Avoid neuron bias' requires that absolute neuron expectations are not excessively large. Specifically, we advocate measuring this via $LBIAS_l \approx 0$ at each layer $f_l$, where $LBIAS_l$ is the layer bias as defined above. We especially recommend ensuring $LBIAS_L \approx 0$.
\end{npc}

This guideline contributes to the nonlinearity normalization algorithm in chapter \ref{nlnormChapter}.

\paragraph{Computing $LBIAS_l$} Since $LBIAS_l$ and $NLC_{l,0}$ have the same denominator, our discussion of the computation of the NLC denominator in section \ref{nlcComputeSection} applies here as well. With a good implementation, the computation does not suffer from catastrophic rounding error unless the rounding error induced by the network evaluation exceeds the (mathematical) difference of outputs. Among our study A architectures, there were four architectures for which this happened. The computed LBIAS value was around $10^{15}$, which is around 1 over the spacing of the floating-point grid. While the true LBIAS values would have been somewhat larger, we decided that the computed value was an acceptable proxy and used it for plotting results in e.g. the figures of this section. Computing the $LBIAS_l$ numerator is trivial compared to the denominator.

In light of this section, we can say that architectures with initial $LBIAS_l$ values large enough to cause computational issues are likely to be suboptimal from a performance standpoint.

\newpage

\newpage

\section{Ensure noise stability, and specifically sufficient floating-point precision} \label{noiseStabilitySection}

Many deep learning building blocks purposefully make the prediction noisy. In the case of batch normalization, the prediction for an individual input depends on the other inputs with which it is forward-propagated. In the case of dropout, it depends on which neuron values are dropped. In the case of data augmentation, it depends on choices made by the data augmentation function. Some activation functions are explicitly noisy \citep{stochasticNonlinearity}. All these choices are random and thus induce noise. While it is not the topic of this work to discuss the potential benefits of noise, we investigate one of the pitfalls in this section. In order to obtain a reliable prediction, any noise must not lead the output to vary too drastically. We can imagine that excessive noise caused by e.g. a high dropout rate or a small batch size would adversely impact performance. In this section, we confirm this hypothesis empirically for two examples: floating-point rounding error and batch normalization. Then we formulate the `ensure noise stability' guideline.

In section \ref{nlcNoiseSection}, we showed how the NLC is related to the degree to which random noise added to the network input affects the network output. In a nutshell, we found the NLC is proportional to the magnitude of the induced output noise {\it relative} to the diameter of the codomain, when adding a certain magnitude of input noise {\it relative} to the diameter of the domain. In that section, we empirically investigated the impact of artificial noise added to the input. In this section, we show that the impact of ``natural'' sources of noise is also modulated by the NLC, though not necessarily as straightforwardly.

\paragraph{Ensure that floating-point precision is sufficient} Evaluating the mathematical functions represented by e.g. the network or the training algorithm using floating-point computation introduces error. This error is not technically random, as floating-point computation is generally implemented as a deterministic system. In practice, this error is often viewed as random noise, as the exact position of the floating-point grid relative to mathematical function values is viewed as random. This assumption is very practically reasonable for our purposes, as well as most other purposes. Note that floating-point rounding error is of prime importance in quantized, low-precision networks (e.g. \citet{quantization2,quantization3,quantization4}).

\begin{figure}[H]
\centering
\includegraphics[width=0.98\textwidth]{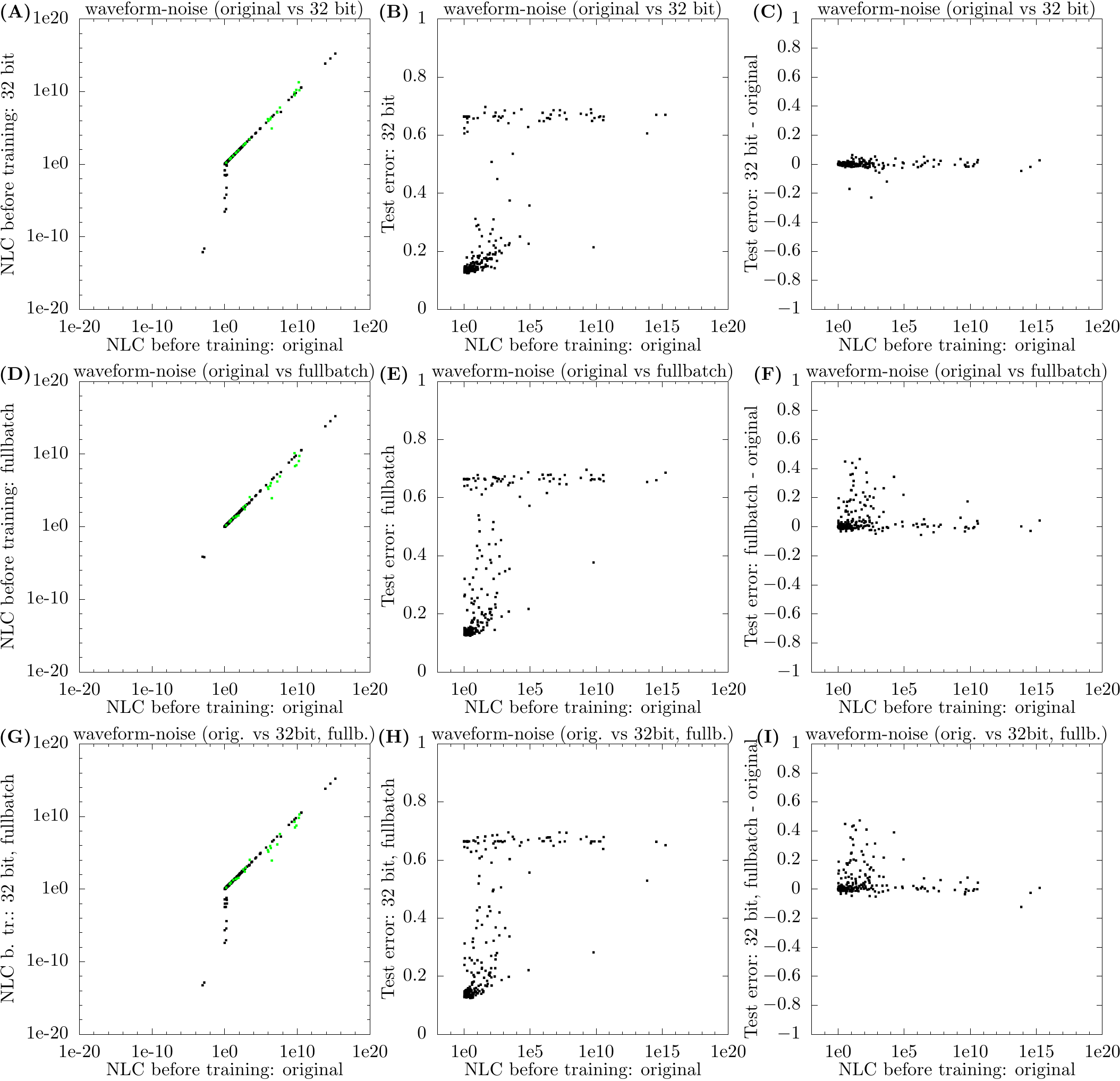}
\caption{Computed metric values for study A waveform-noise architectures. In graphs A/B/C, we depict results from performing computation using 32-bit floating-point precision. In graphs D/E/F, we depict results from using the entire training / validation / test set as a single batch during training and metric computation. In graphs G/H/I, we depict results from using the entire training / validation / test set as a batch and 32-bit precision. In graphs A/D/G, we plot the initial NLC obtained from our original regime vs the initial NLC obtained in the modified regime. Note that these values are computed estimates and do not necessarily reflect the mathematical values of the NLC. GUAs are depicted in green in the foreground. In graphs B/E/H, we plot the initial NLC from the original regime vs the test error from the modified regime. In graphs C/F/I, we plot the initial NLC from the original regime vs the difference between the test error from the modified and original regime. {\it Conclusion:} Lowered precision compromises NLC computation for some architectures. Changing batch size from 250 to 3000 does not impact the NLC. The modified regimes do not yield drastically different test error values overall.} \label{beyondNoiseTest}
\end{figure}

In the previous section, as well as section \ref{nlcComputeSection}, we encountered the importance of floating-point precision for obtaining accurate NLC and LBIAS values. We begin our empirical investigation with this issue. In figure \ref{beyondNoiseTest}A, we plot the value of the NLC computed in the initial state when using our regular 64-bit setup vs the value computed using effectively the same program under 32-bit precision for our study A waveform-noise architectures. We find that \finding{for a large majority of architectures, the value is effectively unchanged}. \finding{For some GUAs, the value changes somewhat but not drastically}. This is explained by the instability of the estimator and has nothing directly to do with the change of precision. However, \finding{for some architectures, the NLC value does change drastically}. These architectures have $LBIAS \gtrapprox 10^7$. For those architectures, the NLC denominator takes very small values. Under 32-bit, that value is effectively rounded up to the floating-point grid spacing, which leads to an overestimate of the denominator and thus to an underestimate of the NLC. \finding{For two waveform-noise architectures, which are in the lower left quadrant of figure \ref{beyondNoiseTest}A, even 64-bit precision is not enough for computing the NLC accurately}, as mentioned in section \ref{nlcComputeSection}.

Now let's look at architecture performance. We re-trained and re-evaluated our waveform-noise architectures using our study A protocol, except that all computation was conducted with 32-bit floating-point precision. The results are given in figure \ref{beyondNoiseTest}B. In figure \ref{beyondNoiseTest}C, we give the difference between the test error values obtained under 32- and 64-bit. We find that \finding{changing the precision does not have a significant impact on test error overall, though the test error of individual architectures does change to a small degree}. These kinds of small changes are due to the sensitivity of the architecture's final state to minuscule levels of noise under gradient-based training \citep{initialFinalGradient}, and are not inherently related to the change in precision. See also section \ref{sharpValleySection}.

In figure \ref{beyondNoiseTest}A and \ref{beyondNoiseTest}B, we give the results obtained from training error minimization under 32-bit. Here, \finding{there are three architectures where the training error increases massively relative to 64-bit}. Specifically, \finding{we observe that these architectures all have an initial NLC larger than $10^7$. There is only one other architecture with such a large NLC that is still trainable under 32-bit}, and it is exactly the architecture that changes its NLC significantly after the first gradient update, as described in section \ref{nlcEvolutionSection}. Hence, as long as the NLC remains larger than around $10^7$, training fails under 32-bit precision. 

This suggests that rounding error indeed behaves somewhat like uniformly random noise. In section \ref{nlcNoiseSection}, we showed that noise of relative magnitude around $\frac{1}{NLC}$ was needed to increase error. If we view the rounding error as noise of relative magnitude equal to the grid spacing, we would expect architectures to become untrainable when the NLC exceeds 1 over the grid spacing. This is exactly what we observe. Note that there can be other factors that create a need for high floating-point precision other than the NLC. If a network has, say, an intermediate layer that is extremely biased so that the first several significant digits are constant across inputs, the need for floating-point precision might be significantly higher than $\frac{1}{NLC}$.

For our study B architectures, unfortunately, we had to conduct all computation in 32-bit precision and were unable to try out 64-bit precision. Since some of our architectures had an initial NLC around $10^7$, it is possible that our experimental results were impacted, at least when it comes to evaluating the trainability of high-NLC architectures in e.g. figure \ref{nlcPredTrainInit}.

\paragraph{Ensure sufficient batch size} To investigate the impact of noise induced by batch selection under batch normalization, we re-trained our waveform-noise architectures with our study A protocol, except that we used a batch size of 3000, which corresponds to the entire training set. When computing validation and test error, we use batches of size 1000, which corresponds to the entire validation / test set. Therefore, the noise inherent in propagating the same input forward inside different batches is eliminated.

\begin{figure}[H]
\centering
\includegraphics[scale=0.8]{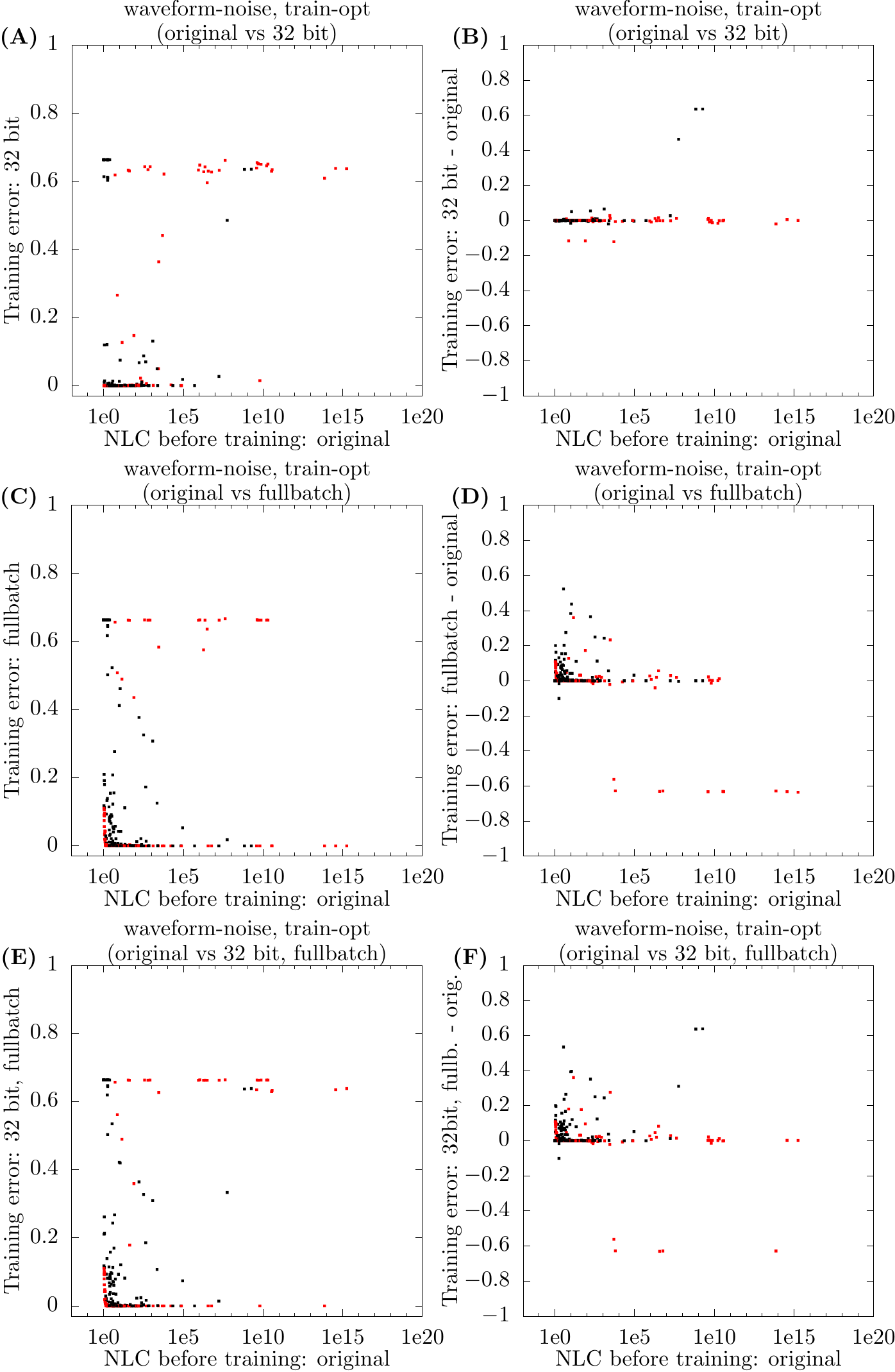}
\caption{Metric values for study A waveform-noise architectures. In graphs A/B, we depict results from performing computation using 32-bit floating-point precision. In graphs C/D, we depict results from using the entire training / validation / test set as a single batch during training and error computation. In graphs E/F, we depict results from using the entire training / validation / test set as a batch and 32-bit precision. In graphs A/C/E, we plot the initial NLC obtained from our original regime vs the training error obtained in the modified regime after training error minimization. In graphs B/D/F, we plot the initial NLC from the original regime vs the difference between the training error from the modified and original regime. Red points correspond to architectures with BN. {\it Conclusion:} Lowered precision compromises trainability for high-NLC architectures. Using full batches enables trainability for high-NLC architectures with BN.} \label{beyondNoiseTrain}
\end{figure}

In figures \ref{beyondNoiseTrain}E-F and \ref{beyondNoiseTrain}C-D, we give the results of training and evaluating architectures in ``fullbatch'' mode with 64-bit precision. In figure \ref{beyondNoiseTrain}H-I and \ref{beyondNoiseTrain}E-F, we give the results of training and evaluating in fullbatch mode with 32-bit precision. We find that \finding{test error suffers for some architectures, and never improves significantly}. Possible explanations for this are that larger batch sizes are inherently inferior, as is generally believed, or that training ``overfits'' to the single training batch and does not generalize as much to the test batch. For the training error, we obtain different results. \finding{A significant number of architectures with very high NLCs that were untrainable with small batches are trainable with large batches. The red color indicates that all those newly trainable architectures use BN. When restricted to 32-bit, fullbatch mode only reduces training error when the NLC is less than around $10^7$. The only exception to this, which is visible in the lower right quadrant of figure \ref{beyondNoiseTrain}F, is an architecture that reduces its NLC massively after the first update, just as the architecture we discussed above.} All of this is expected. Removing batch selection noise helps high-NLC architectures train, but introducing additional rounding error can negate that effect.

Based on these observations, we formulate the ZSAD guideline of noise stability.

\begin{npc} \label{npcNoise}
`Ensure noise stability' requires that the random choices made when evaluating the network function do not induce output variation of a magnitude greater than the range of predictions that can be considered accurate for a given task. In this context, floating-point rounding error can also be viewed as noise.
\end{npc}

We note that, while we did not observe a systematic increase in test error due to noise in this section, this is because our sources of noise were not large enough to impact architectures with an NLC small enough for generalization. If trainability were impacted for low-NLC architectures, then test error would be as well.

How much noise does BN induce? A rough estimate of the relative estimation error of the neuron mean and standard deviation from a sample of size $|B|$ is $\frac{1}{\sqrt{|B|}}$. If $|B| = 250$, then this value is around 0.06. However, in figure \ref{beyondNoiseTrain}D, we find that \finding{the benefit of fullbatch training only kicks in for an NLC around 5000}. Hence, noise from BN has a much lesser impact on trainability than the noise from floating-point rounding relative to its magnitude, and a much lesser impact than proposition \ref{finiteNetNoiseSensitivity} would suggest. This fact is worth further study.

Finally, in figure \ref{beyondNoiseTest}D/G, we compare the ``original'' value of the NLC vs the NLC computed under fullbatch mode and 64-bit or 32-bit precision. We find that \finding{using full batches makes no difference to the NLC value for non-GUAs}. This insight may confer flexibility when computing the NLC in practice in terms of the batch size used.

\newpage

\newpage

\begin{figure}
\centering
\includegraphics[width=0.98\textwidth]{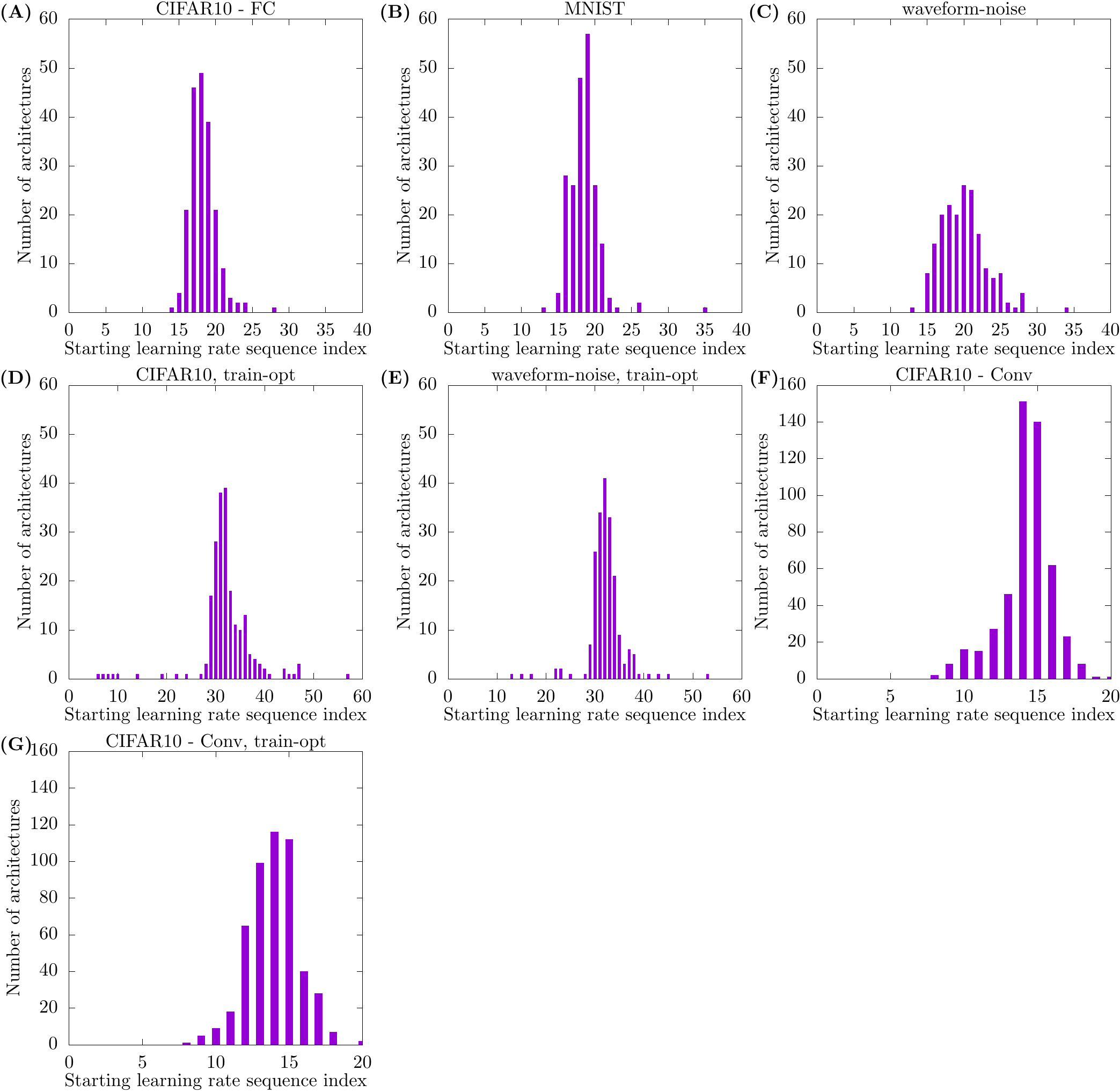}
\caption{Count of the number of architectures for which the $n$'th element in the starting learning rate sequence was chosen, for study A and B architectures. In graphs A/B/C, the choice was based on validation error. In graph F, the choice was based on test error. In graphs D/E/F, the choice was based on training error for the purpose of training error minimization. {\it Conclusion:} We were largely successful in ensuring that, for each architecture, the best SLR considered lies within the wide valley of the SLR-to-error function.} \label{beyondLRcounts}
\end{figure}

\section{Perform exhaustive learning rate tuning} \label{learningRateSection}

This section differs from the previous sections of this chapter in that we do not discuss a specific guideline for zero-shot architecture design. Throughout this work, we investigate the question of how to build neural architectures to maximize performance. In order to speak meaningfully about the ``performance of an architecture'', we need to ensure that our training protocol enables a fair comparison. As we explained in section \ref{moduloAlgorithmSection}, the most important aspect of this is to conduct an independent, exhaustive learning rate search for each architecture. In this section, we verify the importance of learning rate tuning for both scientific validity and practical performance maximization. Then, we go on to uncover patterns that might enable us to predict which learning rates work best for an architecture or task. This might then enable us to reduce the computational cost involved in learning rate tuning.

The SLR-to-error function has a characteristic shape, as depicted on page 425 of \citet{dnnBook}. There is almost always a single ``wide valley'' that has a relatively robust minimum which can be found by considering SLR grids of spacing factor around 3. This is the minimum we look for in this section. However, the SLR-to-error function also has an unknown number of valleys that can be extremely sharp, and can have minima far below the minimum of the wide valley. For example, if an SLR of 0.7465 induces an error of 0.103 and an SLR of 0.7466 induces an error of 0.104, an SLR of 0.74655 may induce an error of 0.07, for example. We term this the sharp valley problem, which we further discuss in section \ref{sharpValleySection}. Throughout this section and this work, we ignore the presence of these sharp valleys and focus on finding an SLR close to the minimum of the wide valley, as is done in practice.

Throughout this work, whenever we train an architecture, we select the best starting learning rate by conducting a large number of independent training runs. It is important to keep in mind that when we refer to the ``best starting learning rate'' in this work, we refer to the best SLR that is found via this procedure. The full list of starting learning rates chosen for each of our architectures for both validation / test error and training error minimization is given in the appendix in chapter \ref{fullListChapter} along with key metrics.

\paragraph{The range of starting learning rates we considered contains the wide error valley for almost all architectures} We selected the best SLR independently for each architecture by independently training with 40 different SLRs in study A and determining the best validation error. We re-trained study A architectures to minimize training error without early stopping based on validation error, where we used 60 different SLRs and selected based on best training error. In study B, we used 20 different SLRs and selected based on best test error or best training error for training error minimization. See sections \ref{studyATrainingSection} and \ref{studyBTrainingSection} for details about the training protocol.

\begin{figure}
\centering
\includegraphics[width=0.98\textwidth]{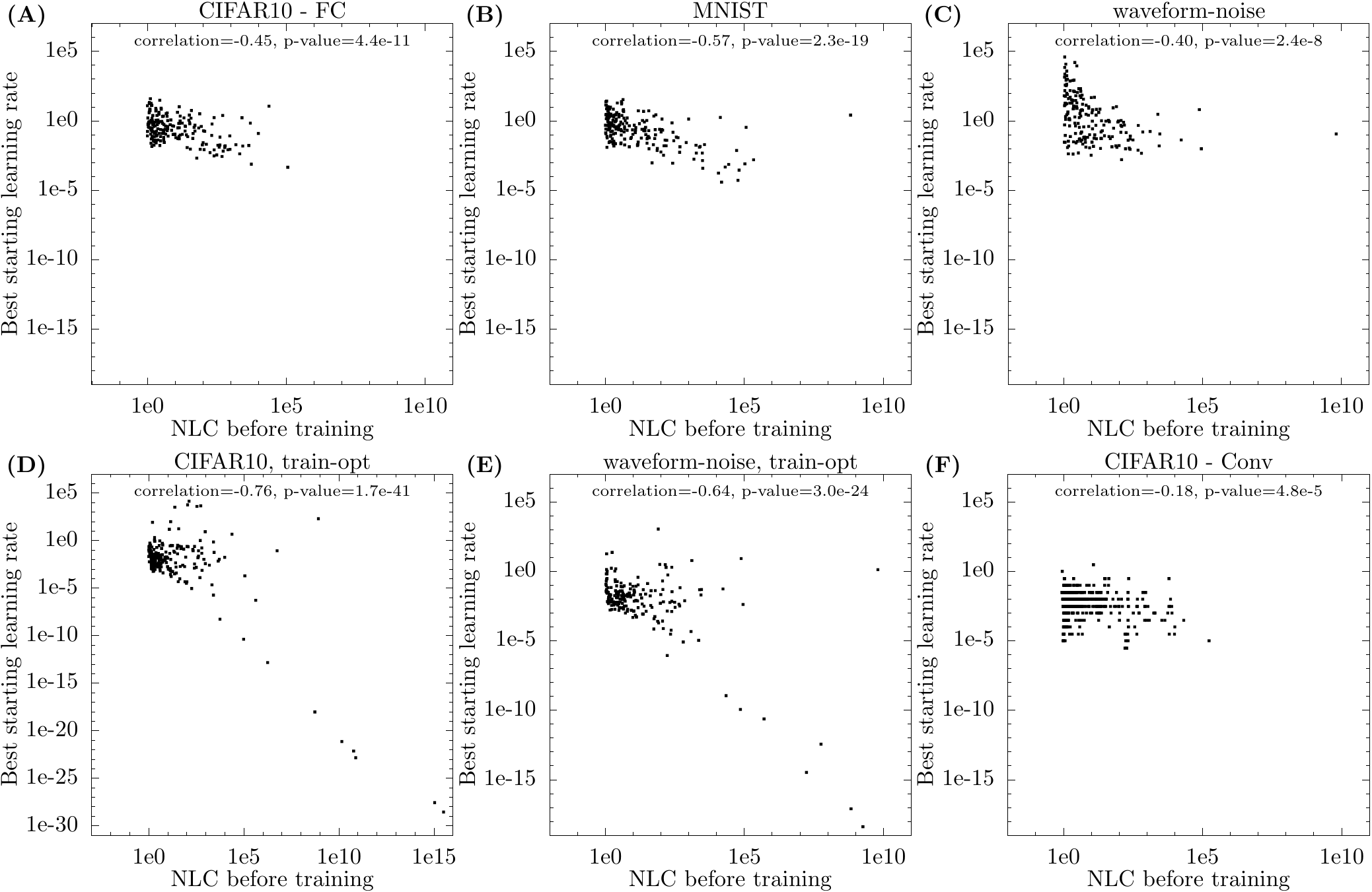}
\caption{Initial NLC vs best SLR for study A and B architectures. {\it Conclusion:} The NLC is somewhat negatively related to the best SLR. However, only in the case of a large NLC and training error minimization is this relationship strong. In that case, the best SLR scales as the inverse square of the NLC.} \label{beyondLRvalsNLC}
\end{figure}

\begin{figure}
\centering
\includegraphics[width=0.98\textwidth]{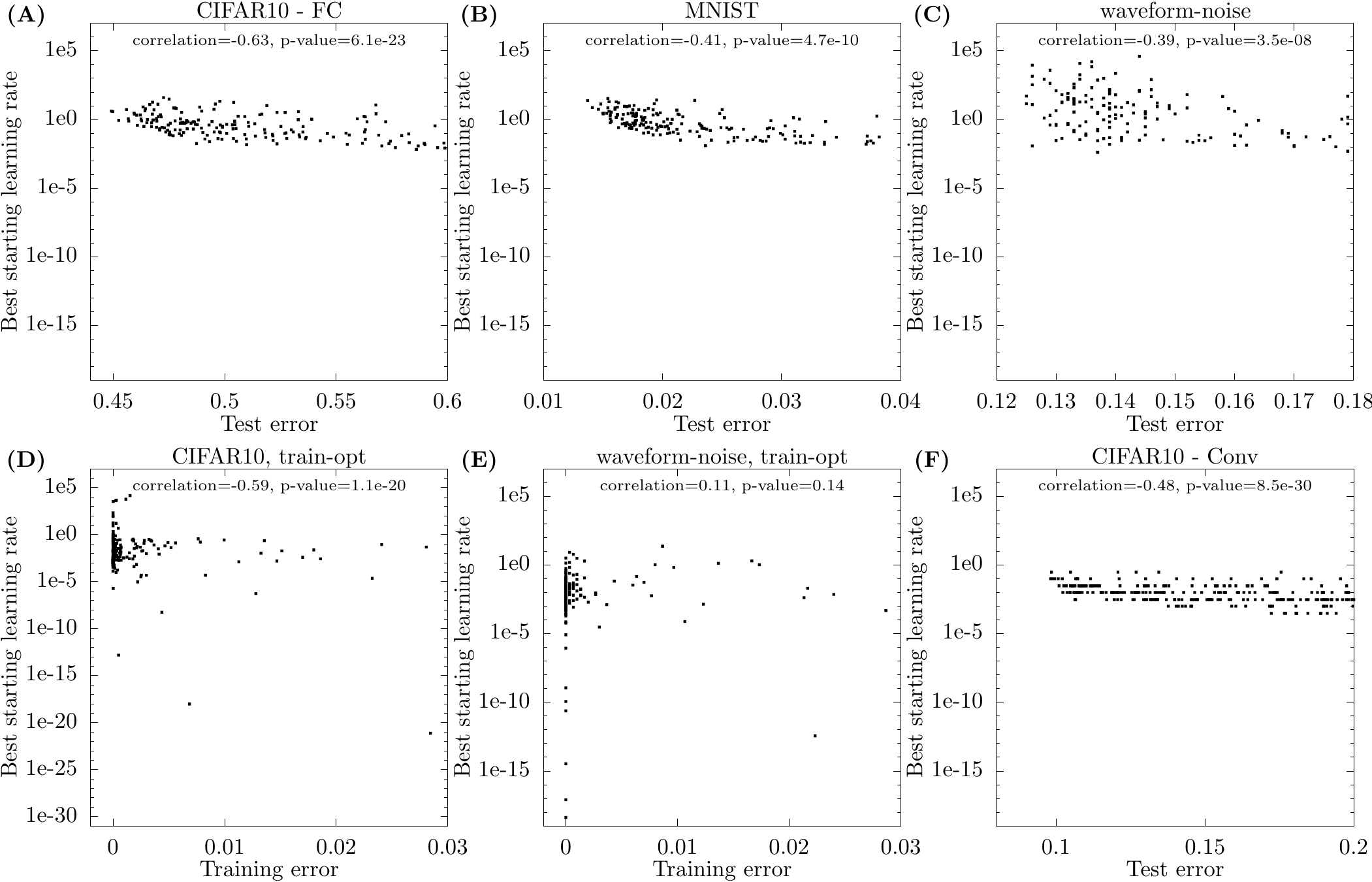}
\caption{Error vs best SLR for study A and B architectures. {\it Conclusion:} For CIFAR10 and MNIST, only a small range of SLRs can lead to close-to-optimal test error. For waveform-noise and training error, this range is much wider.} \label{beyondLRvalserror}
\end{figure}

\begin{figure}[H]
\centering
\includegraphics[width=0.98\textwidth]{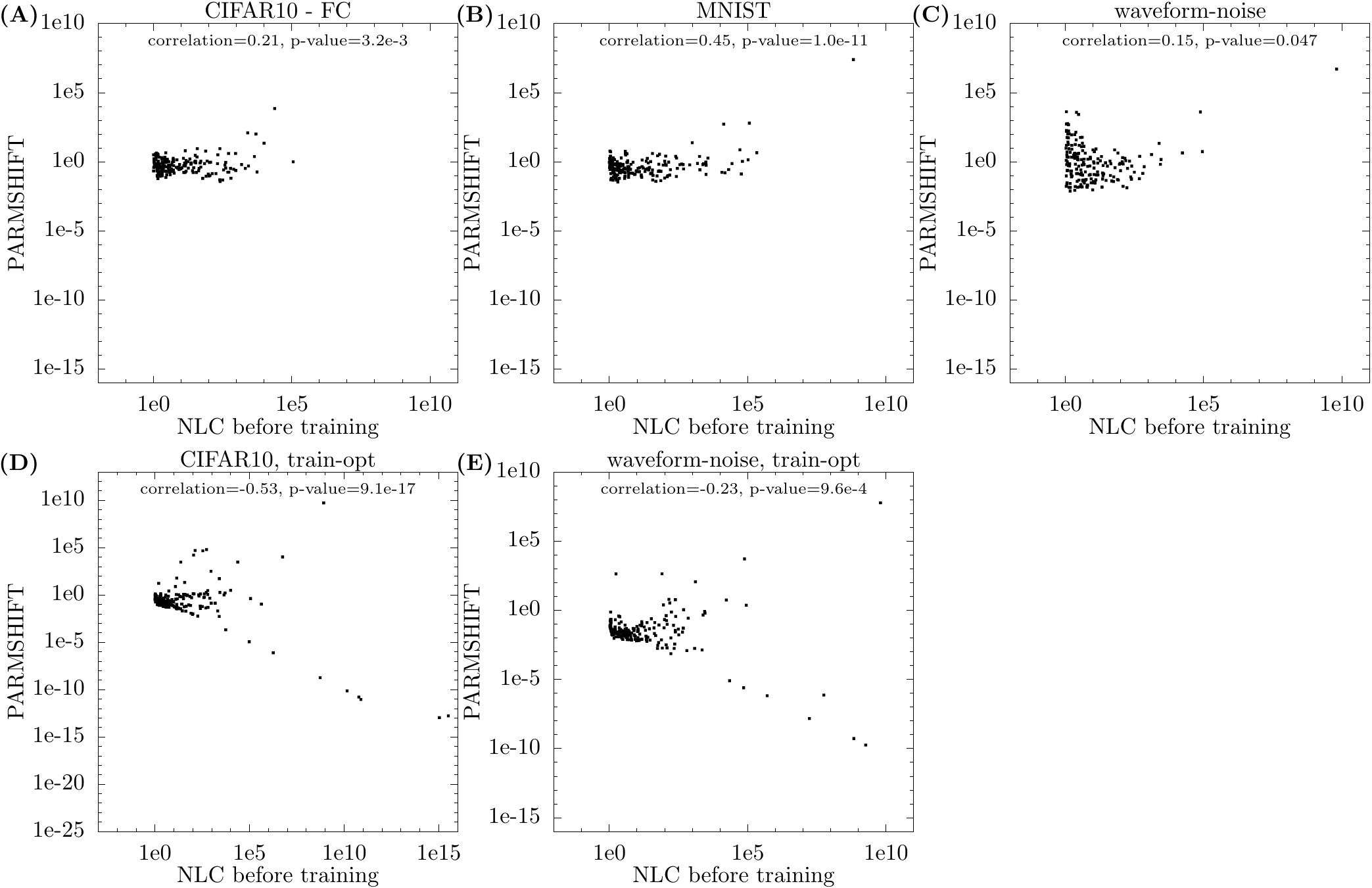}
\caption{Initial NLC vs PARMSHIFT corresponding to the best SLR, for study A architectures. {\it Conclusion:} PARMSHIFT values lie in a narrower range than SLR values. In the case of a large NLC and training error minimization, PARMSHIFT scales inversely with NLC.} \label{beyondLRvalsNLCadj}
\end{figure}

\begin{figure}[H]
\centering
\includegraphics[width=0.98\textwidth]{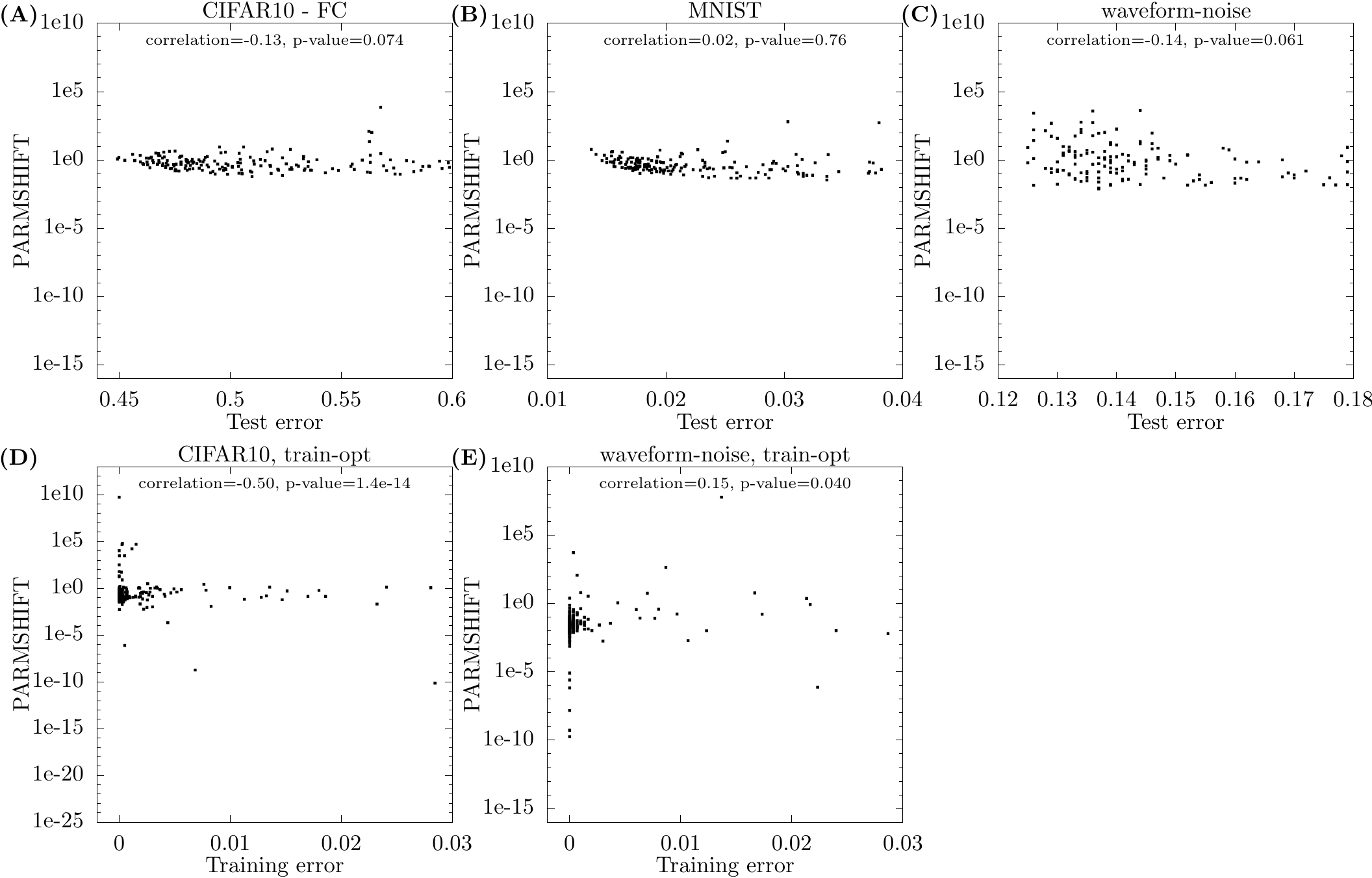}
\caption{Error vs PARMSHIFT corresponding to the best SLR, for study A architectures. {\it Conclusion:} PARMSHIFT values lie in a narrower range than SLR values. For CIFAR10 and MNIST, only a small range of PARMSHIFTs can lead to close-to-optimal test error.} \label{beyondLRvalserroradj}
\end{figure}

\begin{figure}[H]
\centering
\includegraphics[width=0.98\textwidth]{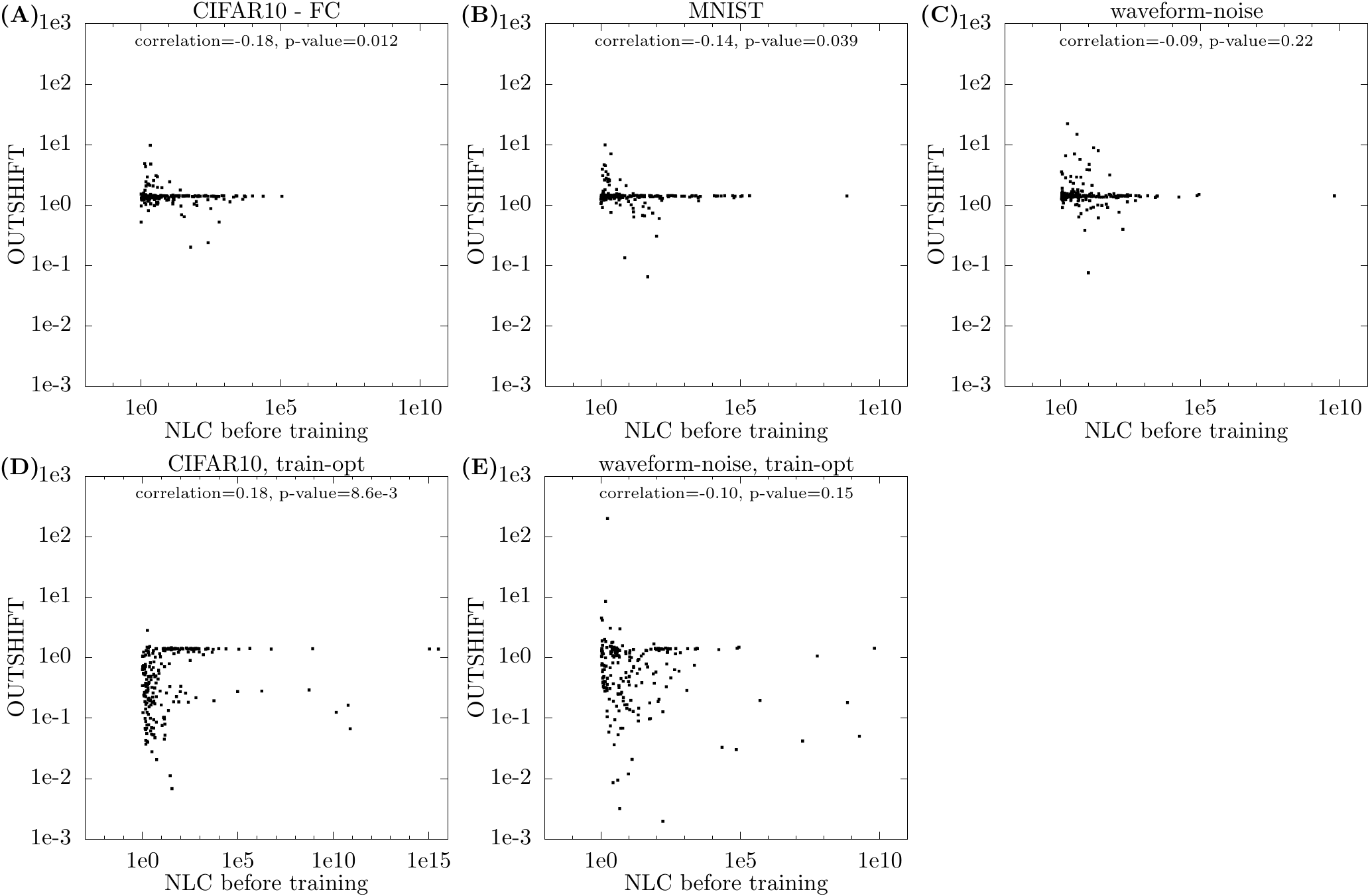}
\caption{Initial NLC vs OUTSHIFT corresponding to the best SLR, for study A architectures. {\it Conclusion:} OUTSHIFT values have a very narrow range relative to PARMSHIFT and SLR, especially for waveform-noise. When considering test error, most architectures have $OUTSHIFT \approx \sqrt{2}$.} \label{beyondLRvalsNLCdiff}
\end{figure}

\begin{figure}[H]
\centering
\includegraphics[width=0.98\textwidth]{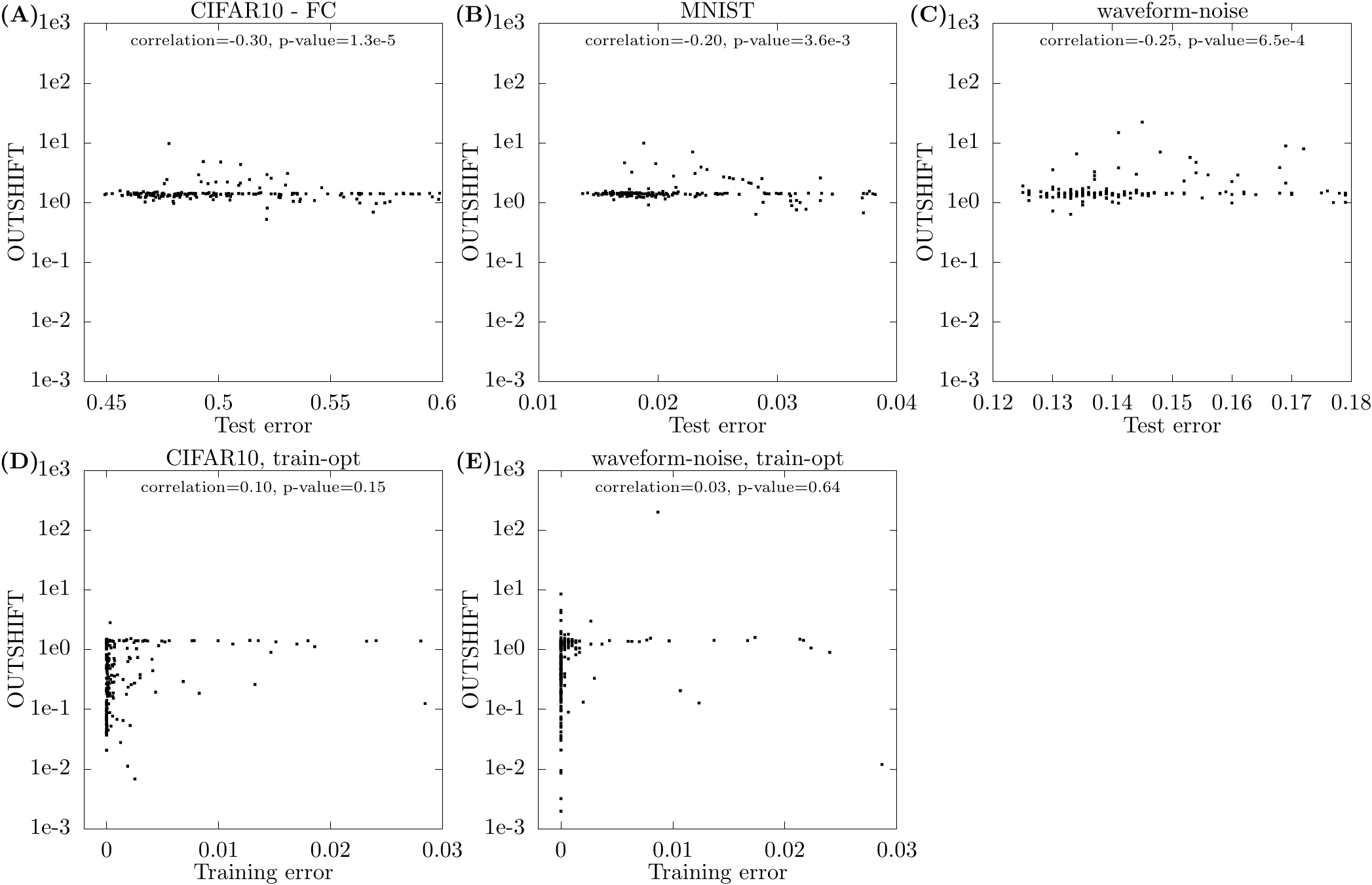}
\caption{Error vs OUTSHIFT corresponding to the best SLR, for study A architectures. {\it Conclusion:} OUTSHIFT values have a very narrow range relative to PARMSHIFT and SLR, especially for waveform-noise. All architectures that attain close-to-optimal test error have $OUTSHIFT \approx \sqrt{2}$. This is also necessary for close-to-optimal test error.} \label{beyondLRvalserrordiff}
\end{figure}

In study A, our SLRs formed a geometric sequence with spacing factor 3. The beginning of this sequence was different for each architecture. The reason behind this customization was to increase the chance that the range covered by the grid would contain the wide error valley and robust minimum described above. In a nutshell, the lowest SLR we considered for an architecture scales inversely with the magnitude of the parameter gradient. In figure \ref{beyondLRcounts}A-C, we count the number of architectures for which the $n$'th member of the SLR sequence yielded the lowest validation error, for each $n$. We find that \finding{for almost all architectures, the best SLR lies somewhere in the middle of the sequence}. Hence, the sequence positioning was successful. \finding{There were no architectures whose best SLR was either among the first 5 or last 5 elements of the sequence}. We view this as confirmation that our ranges indeed cover the wide error valley. Note that in figure \ref{beyondLRcounts} and figures \ref{beyondLRvalsNLC} through \ref{beyondLRvalserrordiff}, architectures for which the error used to select SLR was not better than random for any SLR are not depicted in our graphs, as no meaningful best SLR could be chosen for them.

\begin{figure}
\centering
\includegraphics[width=0.98\textwidth]{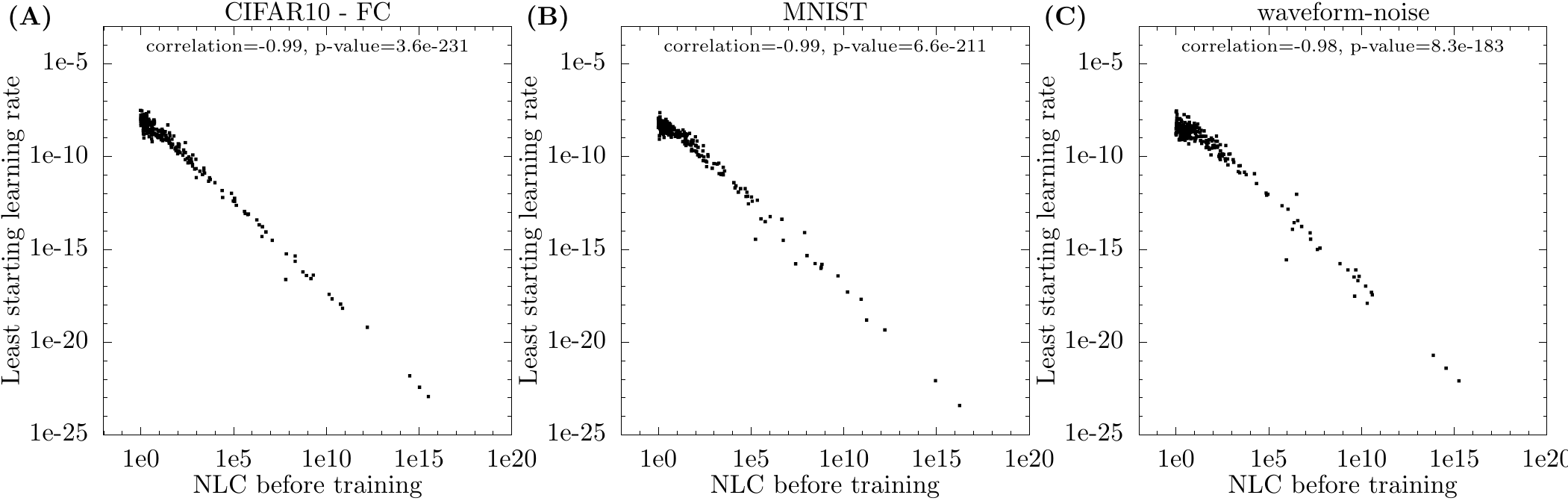}
\caption{Initial NLC vs the smallest SLR considered for study A architectures. {\it Conclusion:} The smallest SLR decreases proportionally as the NLC increases.} \label{beyondLRvalsfirst}
\end{figure}

In figure \ref{beyondLRcounts}D/E, we depict how often each element of the sequence was chosen for training error minimization in study A. Again, \finding{most chosen SLRs lie in the middle of their sequence. However, the total range of sequence elements chosen is much larger compared to when we select based on validation error.} This was the reason behind expanding our sequence to 60 elements. 40 elements would not have been enough to cover all architectures with the starting point heuristic we used. We will uncover the reason for this increased breadth later in this section. \finding{With 60 elements, we ensure that the first 5 or last 5 elements are never chosen, except for one architecture where the fourth largest element is chosen.}

In study B, we considered 20 fixed starting learning rates. We were unable to adjust the sequence on a per-architecture basis, due to code base limitations as outlined in section \ref{codeLimitationsSection}. In figure \ref{beyondLRcounts}F/G, we find that, \finding{while there is a buffer at the lower end of the sequence, the largest SLRs considered were chosen for some architectures}. Hence, we cannot guarantee that considering even larger SLRs would not have increased performance further for some architectures. As we discussed in the previous section, the trainability of some high-NLC architectures might have been impacted by using only 32-bit floating-point precision. Under 64-bit precision, the range of optimal SLRs might have been larger. Because of this uncertainty, we also exclude results from training error minimization of study B architectures from figures \ref{beyondLRvalsNLC} through \ref{beyondLRvalserrordiff}.

\paragraph{The smallest SLR scales inversely with the NLC} In figure \ref{beyondLRvalsfirst}, we plot the initial NLC vs the least SLR considered, i.e. the smallest member of our SLR sequence. \finding{We find that there is an inversely proportional relationship}. Across the architectures we consider, the NLC is roughly proportional to the magnitude of the parameter gradient, which in turn is inversely proportional to the lowest SLR we consider by construction. In general, the strength of the relationship between NLC and parameter gradient magnitude that is behind the trend of figure \ref{beyondLRvalsfirst} does depend on various architecture properties, which we do not investigate here. In general, we focus this work on the input and layer gradients rather than the parameter gradient. An analysis along the lines of section \ref{meanFieldPracticalSection}, chapter \ref{surveyChapter} or \citet{meanFieldNetsorNTK} can shed further light on the parameter gradient.

As stated in section \ref{studyATrainingSection}, the smallest SLR we consider for training error minimization of study A architectures is exactly $10^{-8}$ times the smallest SLR given in figure \ref{beyondLRvalsfirst}.

\paragraph{The best SLR has a wide range across architectures} In figure \ref{beyondLRvalsNLC}A/B/C/F, we plot the initial NLC vs the best (=selected) SLR. We find that \finding{the range of best SLRs has width around $10^7$ in log space within each graph, and around $10^{10}$ across all graphs}. In figure \ref{beyondLRvalsNLC}D/E, we plot the initial NLC vs the best SLR for training error minimization. Now, \finding{the range spans 30 orders of magnitude!} This strongly underscores the need for exhaustive learning rate tuning in order to compare architectures fairly for the purpose of empirical study, and in order to maximize performance. Yet, very few studies conduct such tuning.

\paragraph{The NLC is not predictive of the best SLR when minimizing validation / test error but somewhat predictive when minimizing training error} In figure \ref{beyondLRvalsNLC}A/B/C/F, we find that \finding{for any NLC level, the range of optimal SLRs has width around $10^4$ and $10^7$ in log space. This range seems to shift slightly as the NLC increases}. Overall, the NLC is not much help in predicting the best SLR. In figure \ref{beyondLRvalsNLC}D/E, we find that \finding{there exists a small group of architectures with high NLCs for which the best SLR scales as the inverse square of the NLC}. These are precisely the architectures that train but do not generalize, which we have investigated e.g. in the previous section. There are a few outliers to this trend. \finding{For waveform-noise, there is a single outlier with initial NLC around $10^{10}$.} This is the same outlier we discussed before, which changes its NLC after the first update. \finding{For CIFAR10, there is an outlier with initial NLC around $10^7$ and one with initial NLC around $10^9$.} These do not achieve a low final NLC. However, the magnitude of the parameter still changes drastically, which is confirmed by their PARMSHIFT values below.

\paragraph{Only a narrow range of SLRs can lead to optimal test error for CIFAR10 and MNIST, but a wide range can lead to optimal error for waveform-noise} In figure \ref{beyondLRvalserror}, we plot the test and training error achieved vs the best SLR. \finding{For CIFAR10 and MNIST, in order to attain a test error close to the optimal value for the dataset, we find that the SLR has to lie within a range of width around 10 in log space. Architectures whose best SLR lies outside that range do not attain a close-to-optimal test error. For waveform-noise, a much wider range of SLRs can lead to optimal or close-to-optimal test error}. This is one of the few situations where we observe different behavior for our datasets. We hypothesize that waveform-noise is more forgiving because it is a smaller dataset and hence represents an easier task. \finding{Minimizing training error is also easier compared to minimizing test error.}

\paragraph{The best parameter update size has a narrower range than the best SLR} In addition to looking at the SLR directly, we can study it indirectly. As the learning rate increases, the update that is applied to the parameter during the first iteration also increases. In general, we would expect the magnitude of the sum of all updates to increase as well.

\begin{metricDefinition}
The `parameter shift' (PARMSHIFT) of a final parameter value $\theta^{(T)}$ relative to an initial parameter value $\theta^{(0)}$ is

$$PARMSHIFT(\theta^{(0)}, \theta^{(T)}) = \frac{||\theta^{(T)} - \theta^{(0)}||_2}{||\theta^{(0)}||_2}$$

\end{metricDefinition}

We plot the initial NLC as well as error vs PARMSHIFT obtained from the best SLR for study A architectures in figures \ref{beyondLRvalsNLCadj} and \ref{beyondLRvalserroradj}. We find results similar to figures \ref{beyondLRvalsNLC} and \ref{beyondLRvalserror}. Again, \finding{for CIFAR10 and MNIST, only a narrow range of PARMSHIFT values can lead to close-to-optimal test error. Compared to SLR, the range of best PARMSHIFT values across architectures is narrower for at least, say, the top 50\% of architectures. In figure \ref{beyondLRvalsNLC}D/E, we now find that when minimizing training error of our high-NLC architectures, PARMSHIFT scales inversely with the NLC}. This can be interpreted in light of our earlier findings. In figure \ref{beyondLRvalsfirst}, we found that the NLC is roughly proportional to the parameter gradient magnitude. In figure \ref{beyondLRvalsNLC}, we found that the best SLR for training error minimization scales as the inverse square of the NLC. We might expect PARMSHIFT to be roughly proportional to the product of parameter gradient magnitude and SLR, thus inversely proportional to the NLC.

\paragraph{The best output update size lies in a very narrow range and is highly related to test error} Finally, in addition to investigating the magnitude of the change to the parameter, we investigate the magnitude of the change to the network output. For the same reason as for PARMSHIFT, we expect it to be closely related to the SLR.

\begin{metricDefinition}
The `output shift' (OUTSHIFT) of an architecture $f$ during the first update is 

$$OUTSHIFT(f,\mathcal{D}, \theta^{(0)}, \theta^{(1)}) = \sqrt{\frac{\mathbb{E}_x||f(\theta^{(0)},x) - f(\theta^{(1)},x)||_2^2}{\mathbb{E}_x||f(\theta^{(0)},x)||_2^2}}$$

\end{metricDefinition}

Here, we consider specifically the change of the output during the first update, as we suspect that updates during consecutive iterations will cancel each other out to a significant degree. In figures \ref{beyondLRvalsNLCdiff} and \ref{beyondLRvalserrordiff}, we repeat figures \ref{beyondLRvalsNLCadj} and \ref{beyondLRvalserroradj} with the OUTSHIFT instead of PARMSHIFT metric, which is evaluated on the test set. The results are highly noteworthy. In graphs A/B/C, \finding{the majority of architectures have an OUTSHIFT value extremely close to a single fixed value, which turns out to be $\sqrt{2}$ for reasons described below. Even architectures that deviate from this value are generally not more than a factor of 10 away. The range of OUTSHIFT values is much narrower than the range of SLR values in figure \ref{beyondLRvalsNLC}. Further, in figure \ref{beyondLRvalserrordiff}, we find that architectures with close-to-optimal test error all have $OUTSHIFT \approx \sqrt{2}$, even for waveform-noise}. Note that the range covered by the y-axis in figures \ref{beyondLRvalsNLCdiff} and \ref{beyondLRvalserrordiff} is much smaller compared to that of figures \ref{beyondLRvalsNLCadj} and \ref{beyondLRvalserroradj}.

$OUTSHIFT \approx \sqrt{2}$ arises when (i) $\mathbb{E}_x||f(\theta^{(0)},x)||_2^2 \approx \mathbb{E}_x||f(\theta^{(1)},x)||_2^2$ and (ii) $f(\theta^{(0)},x)$ is approximately independent of $f(\theta^{(1)},x)$. In other words, we need the first update to reset the outputs completely without increasing their magnitude significantly. If the architecture uses neither BN nor LN, only a very small number of SLRs achieve this exactly. If BN or LN is used, all SLRs above a certain level achieve this, because OUTSHIFT is bounded above by the fact that $\mathbb{E}_x||f(\theta^{(0)},x)||_2^2 = \mathbb{E}_x||f(\theta^{(1)},x)||_2^2 = 1$ holds.

The OUTSHIFT requirement is reminiscent of line search. Because of their large capacity, neural networks are often able to move all outputs for the first batch close to the label in the first update. An ``output reset'' would correspond to doing this and therefore to minimizing loss on the batch.

In figure \ref{beyondLRvalsNLCdiff}D/E, we find that \finding{OUTSHIFT does not decrease significantly for large NLC's under training error minimization, at least relative to SLR and PARMSHIFT}. Again, there is a similar interpretation. If the parameter gradient magnitude is proportional to the NLC and the SLR scales with the inverse square of the NLC, then the magnitude of the first parameter update scales with the inverse of the NLC. Finally, OUTSHIFT is roughly the product of update size and parameter gradient magnitude.

\paragraph{Learning rate prediction: discussion} We have presented some intriguing relationships between best SLR and other metrics. These findings are preliminary and more work is needed to develop fleshed-out learning rate selection strategies. Since we are not deeply familiar with the literature on learning rate selection, we are also not able to place these results in a wider context. However, given the breadth of our studies, we do believe there is value in mining them for insights about learning rate. We note that tying parameter to update length as in PARMSHIFT was recently done by \citet{layerwiseLearningRate} and line search was recently used for deep networks by \citet{lineSearch}.

While we found it difficult to predict the best SLR, or its surrogates PARMSHIFT and OUTSHIFT, for a given architecture, we found that the architectures that achieve close-to-optimal test error for a dataset have predictable behavior. The most significant pattern is $OUTSHIFT \approx \sqrt{2}$. This might enable us to greatly reduce our learning rate search while only attaining suboptimal error values for architectures that are suboptimal to begin with. This might be sufficient in some practical situations.

Finally, we note that using early stopping was crucial in reining in the computational expense.

\newpage

\newpage

\section{Summary: explaining the performance of our architectures} \label{beyondNlcSummarySection}

In this chapter, we have introduced several novel ZSAD guidelines, and have analyzed them as well as previously existing ZSAD guidelines. We are now in a position to explain, to a large degree, the performance of our architectures in terms of these guidelines. In chapter \ref{surveyChapter}, we further explain these guidelines in terms of the architecture definition itself.

\begin{figure}[H]
\centering
\includegraphics[width=0.98\textwidth]{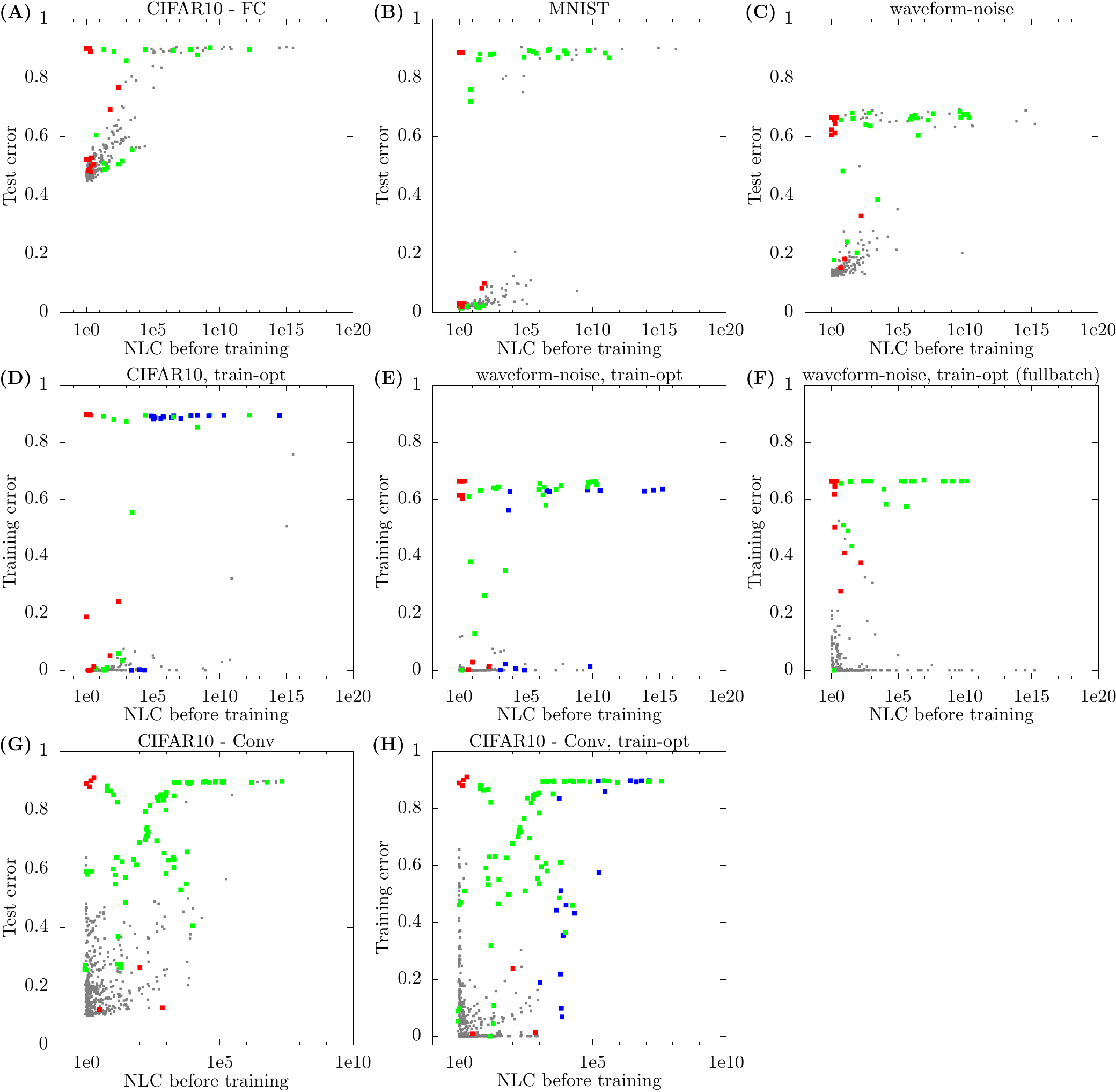}
\caption{Initial NLC vs test and training error for study A and B architectures. In graph F, we plot initial NLC vs training error when the entire training set is used as a single batch for all computations. Architectures depicted in green are GUAs. Architectures depicted in red have $LBIAS > 10$. Architectures depicted in blue have $NLC > 1000$, use BN or data augmentation, and are not GUAs. Other architectures are depicted in grey. All colored markers are displayed in the foreground relative to grey markers. We verified that no grey markers in distinctive positions were occluded. {\it Conclusion:} Gaussian instability, neuron bias, noise instability and the NLC explain the majority of performance variation.} \label{beyondSummary}
\end{figure}

In figure \ref{beyondSummary}, we plot the initial NLC vs test and training error for our study A and B architectures. Graphs A/B/C/D/E/G/H are equivalent to the graphs in figure \ref{beyondGaussian}, except for colors. GUAs are displayed in green. Architectures with an initial LBIAS greater than 10 are displayed in red. In graphs D and H, architectures with an initial NLC greater than 1000 that use BN or data augmentation and are not GUAs are depicted in blue. Data augmentation is the third source of noise that occurs in study B (but not in study A), next to noise from BN and floating-point computation. While we did not address it in section \ref{noiseStabilitySection}, our data augmentation induces noise of significant magnitude via cropping and especially horizontal flipping (sections \ref{dataAugmentationSection}, \ref{studyBTrainingSection}).

In graphs A through C, \finding{we find that almost all low-NLC study A architectures that do not generalize are either GUAs or suffer from neuron bias. The relationship between the NLC and test error among the remaining architectures, depicted in grey, is very strong}. In graph G, again, \finding{we find that among study B architectures with NLC less than around 1000, architectures that have the highest errors are GUAs or suffer from neuron bias}.

In graphs D, E and H, \finding{we find that all untrainable architectures are either GUAs or suffer from neuron bias or noise instability}. Unfortunately, study B did not end up containing architectures with an initial NLC greater than 3000 that did not also suffer from noise or Gaussian instability. Hence, our study did not verify that high-NLC convolutional architectures are trainable. (Note that we also would have been limited by 32-bit floating-point precision.) \finding{There are three CIFAR10 architectures with very high NLC, depicted in grey in graph D, which have elevated, though not random, training error.} Two of them have $NLC > 10^{15}$. Hence, they likely suffer from noise instability induced by 64-bit floating-point precision. An even higher level of precision would have to be used to determine whether these architectures are untrainable in a mathematical sense. This leaves only a single high-NLC architecture with elevated training error, which still has training error less than 0.4. Hence, we consistently show that ultra-high complexity architectures are trainable. To our knowledge, we are the first to explicitly demonstrate this. For example, \citet{depthScalesMeanField} and \citet{meanFieldCNN} previously argued this was impossible. We uncover that the requirements are a well-tuned small learning rate, noise stability and an absence of neuron bias. The latter requirement can be circumvented by e.g. debiased gradient descent. We further discuss this point in section \ref{expressivitySection}.

In graphs G and H, \finding{we find that some architectures with NLC close to 1 attain a high, though not random, training and test error}. This indicates underfitting as discussed in section \ref{nlcLinearApproximationSection}. Some of our study B architectures are very close to linear functions due to the construction of our study B activation functions.

In graph F, we give results from conducting training and error computation using the entire training set as a single batch, as done in section \ref{noiseStabilitySection}. The graph is equivalent to figure \ref{beyondNoiseTrain}C except for color. By eliminating the noise induced by small-batch BN, \finding{all architectures that are not GUAs and do not suffer from neuron bias attain better-than-random training error and all such architectures with $NLC > 10^5$ attain a training error close to zero}. 

Overall, an all-or-nothing pattern emerges. Even if there is a single ZSAD guideline that an architecture does not follow, the architecture is likely to attain suboptimal or even random performance. For low training error, we should avoid neuron bias and ensure Gaussian and noise stability. For low test error, we should also ensure $1 \le NLC \le 5$, or a different optimal NLC when the dataset is not neural regular. Scale stability was controlled in our empirical studies. Training stability and exhaustive learning rate tuning are prerequisites for our results in figure \ref{beyondSummary}. Finally, we refer the reader again to the full list of architectures in the appendix in chapter \ref{fullListChapter}, where it is possible to compare individual architecture definitions with metric values and properties.

This completes the explanation of the performance of our architectures. However, there is a larger lesson. The design of our empirical studies was essential for the discoveries we made. Without including the square and odd square activation functions, we would not have discovered Gaussian instability. Without studying very deep networks and activation functions with high $\mathfrak{n}_\tau$ values, we would not have observed such a wide range of NLC values. Without observing such a wide range of NLC values, we would not have discovered e.g. noise stability. Without conducting 64-bit precision computation, we would not be able to study high-NLC architectures in the same way. Without studying activation functions that are biased, i.e. that have $\mathbb{E}_{s\sim \mathcal{N}(0,1)}\tau(s) \neq 0$, we would not have been able to study neuron bias. Our ultimate results are dependent on choices made in study design. We were largely unaware of the outcomes of these choices when they were made. Of course, this obliviousness is essential for making scientifically and statistically sound discoveries. On the other hand, this means that our ZSAD guidelines are at least somewhat incidental and may not represent ``the most important'' guidelines, period. We also discuss this point in section \ref{architectureSensitivitySection}. We would argue that any guideline that is of paramount importance in empirical studies as broad and careful as our studies, and that is explained in terms of general principles, is likely to have significance beyond a specific set of experiments.

We did make some deliberate choices when designing our studies in order to exclude some factors that are known to impact performance. The variance of weight tensor entries in the random initialization schemes we used never deviated significantly from the LeCun variance in order to ensure scale stability. For study A, we augmented the softmax+cross-entropy loss function to remove the influence of output magnitude on performance (section \ref{forwardStabilitySection}). In many architectures, we used activation function debiasing. We ensured that the parameter dimensionality was approximately constant across architectures within each study. Without these restrictions, the signals we found in this chapter might have been less clear. We believe that one of the key insights of this work is that, in order to make progress in understanding deep learning, we must control phenomena we already understand. There exist too many possible architectures to spend significant effort on known pathologies.

\newpage

\chapter{Nonlinearity Normalization (nlnorm)} \label{nlnormChapter}

Throughout this work, we have explored the importance of attaining a right-sized NLC, especially in the initial state. In chapter \ref{meanFieldNnaChapter}, we showed how to predict the NLC from the architecture definition. However, so far, all we can do with an architecture that has a suboptimal initial NLC is to discard it. In this chapter, we show how to {\it control} the NLC of an architecture by modifying the activation functions used, effectively turning the NLC into a tunable hyperparameter. This corresponds to utility criterion \ref{criterionControllable}. Control is achieved via the simple nonlinearity normalization algorithm. Our analysis shows that the NLC can be considered not just predictive of performance, but {\it causal} for performance. Nonlinearity normalization can also eliminate neuron bias and scale instability. We find that applying nonlinearity normalization to an architecture often induces massive performance gains if that architecture's initial NLC or LBIAS is suboptimal.

We present evidence that nlnorm may significantly reduce the need for a range of building blocks like skip connections (section \ref{nlnormvsbnsection}), normalization layers (section \ref{nlnormvsbnsection}) or specific activation functions (section \ref{nlnormActFunDesignSection}). While these building blocks have a large impact on performance when nonlinearity is not considered, this impact can significantly diminish when nlnorm is employed. This means that tuning the NLC when comparing deep learning pipelines and building blocks is essential to obtain a fair comparison, especially given that no single NLC value leads to optimal or even adequate performance for every type of architecture (section \ref{nlnormBestNLCSection}).

\paragraph{Background from prior chapters} Like prior chapters, this chapter is based on the empirical studies detailed in chapter \ref{empiricalStudiesChapter}, specifically study B (section \ref{studyBSection}). We recommend reading at least summary section \ref{metricsSummarySection} before proceeding. Throughout this chapter, we use the terminology, notation and conventions of section \ref{notationSummarySection}. We use results and definitions from sections \ref{nlcPredictiveSection}, \ref{forwardStabilitySection}, \ref{outputBiasSection} and \ref{meanFieldPracticalSection} to motivate nonlinearity normalization. While we reference additional prior sections when relevant for discussion, a complete understanding of chapters \ref{nlcChapter} through \ref{beyondNlcChapter} is not required.

\paragraph{Technical considerations} As discussed in section \ref{nlcDefinitionSection}, we implicitly assume things like integrability (section \ref{integrabilitySection}), differentiability (section \ref{nonDifferentiableSection}) and non-zero denominators when necessary.

\paragraph{Limitations} We designed study B specifically to validate nlnorm. However, we did not design study A in this way, as it was conducted before we understood nonlinearity. Hence, the empirical analysis of this chapter is restricted to convolutional architectures.

\section{Definition and discussion} \label{nlnormDefinitionSection}

The idea of `nonlinearity normalization' (nlnorm) is based on the nonlinearity path equation from section \ref{pathEquationSection}. One of its implications is that when an activation function $\tau_l^{(1)}$ is replaced in an architecture with another activation function $\tau_l^{(2)}$ such that $\mathfrak{n}_{\tau_l^{(2)}}(\mathfrak{q}_k,\mathfrak{c}_k) < \mathfrak{n}_{\tau_l^{(1)}}(\mathfrak{q}_k,\mathfrak{c}_k)$, then the mean field NLC of the architecture decreases, as long as there are no knock-on effects on downstream $\mathfrak{q}$ and $\mathfrak{c}$ values. In plain words, the NPE suggests that we can control the nonlinearity of the network by controlling the nonlinearity of activation functions.

Assume for now that all activation layers in an architecture use the same activation function $\tau(s)$. nlnorm modifies this `base architecture' by replacing all instances of $\tau(s)$ with $c\ddot{\tau}(l,s) + b$, where $l$, $c$ and $b$ are parameters. (In the context of nlnorm, $l$ does not refer to a layer index.) The values of those parameters are shared between all occurrences of the activation function. We choose the three values to jointly achieve three objectives: right-sized NLC, low neuron bias, and scale stability. We begin by choosing the function $\ddot{\tau}(l,s)$ such that we obtain a range of $\mathfrak{n}_{\ddot{\tau}(l,s)}$ values, specifically $\mathfrak{n}_{\ddot{\tau}(l,s)}(1,0)$ values, as the `linearization parameter' $l$ varies. Note that we have $\mathfrak{n}_{\ddot{\tau}(l,s)} = \mathfrak{n}_{c\ddot{\tau}(l,s) + b}$ for all $c \neq 0$ and $b$. For a given value of $l$, we jointly set $b$ and $c$ to attain $\mathbb{E}_{s \sim \mathcal{N}(0,1)}c\ddot{\tau}(l,s)+b=0$ and $\mathbb{E}_{s \sim \mathcal{N}(0,1)}(c\ddot{\tau}(l,s)+b)^2=1$. Hence, we ensure that $\mathfrak{C}_{c\ddot{\tau}(l,s) + b}(1,1) = 1$ and $\mathfrak{C}_{c\ddot{\tau}(l,s) + b}(1,0) = 0$. So, in the context of table \ref{tableNLCPropagation}, if $(\mathfrak{q}_k, \mathfrak{c}_k) = (1,0)$ holds for the dependency of an activation layer, $(\mathfrak{q}_l, \mathfrak{c}_l) = (1,0)$ holds for the activation layer itself. The idea here is to ensure $(\mathfrak{q}_l, \mathfrak{c}_l) = (1,0)$ throughout the network. If this ``state of normalization'' is preserved by activation layers as well as other layers by e.g. using LeCun initialization, then it holds at all layers and specifically at the output. If mean field theory is predictive, this ensures both near-zero neuron bias and scale stability. By ensuring $(\mathfrak{q}_k, \mathfrak{c}_k) = (1,0)$ for activation layers, we also ensure that the nonlinearity path equation is made up of $\mathfrak{n}_{\ddot{\tau}(l,s)}(1,0)$ terms. Therefore, we can control the mean field NLC and, ultimately, the NLC of the architecture with $l$. The NLC indirectly becomes a tunable hyperparameter via $l$.  If mean field theory is predictive, values of $l$ that induce a high $\mathfrak{n}_{\ddot{\tau}(l,s)}(1,0)$ value correspond to a large architecture NLC. Values of $l$ that induce a small $\mathfrak{n}_{\ddot{\tau}(l,s)}(1,0)$ value correspond to a small architecture NLC. The last step is now to choose the desired NLC by choosing $l$. We summarize nlnorm for architectures with a single activation function in figure \ref{boxnlnorm}.

\begin{figure}
\makebox[\textwidth][c]{
   \fbox{\begin{minipage}{0.8\textwidth}
\centerline{\bf The nonlinearity normalization algorithm}
\centerline{\bf for architectures with a single activation function.}  
\bigskip
\begin{itemize}
\item If the single activation function is $\tau(s)$, choose a parametrized version $\ddot{\tau}(l,s)$ such that varying $l$ yields a range of values for the activation function NLC (section \ref{pathEquationSection}):

$$\mathfrak{n}_{\ddot{\tau}(l,s)}(1,0) = \sqrt{\frac{\mathbb{E}_{s\sim \mathcal{N}(0,1)}\tau'(s)^2}{\mathbb{E}_{s\sim \mathcal{N}(0,1)}\tau(s)^2 - (\mathbb{E}_{s\sim \mathcal{N}(0,1)}\tau(s))^2}}$$

See figure \ref{AFLMillu} for examples, and section \ref{linearizationSection} for an analysis of the universal choice $\ddot{\tau}(l,s) = \tau(s) + ls$.
\item Replace each occurrence of $\tau(s)$ with $c\ddot{\tau}(l,s) + b$, where $c$ and $b$ are chosen so that $\mathbb{E}_{s\sim\mathcal{N}(0,1)}c\ddot{\tau}(l,s)+b=0$ and $\mathbb{E}_{s\sim\mathcal{N}(0,1)}(c\ddot{\tau}(l,s)+b)^2=1$ and $l$ is a free parameter.
\item Choose a value of $l$ to obtain the desired network NLC in the initial state.
\end{itemize}
\end{minipage}}}
\caption{The nonlinearity normalization algorithm for architectures with a single activation function.} \label{boxnlnorm}
\end{figure}

The simple form of nlnorm given in figure \ref{boxnlnorm} may not be sufficient for a given base architecture. A base architecture may not use only a single activation function. Weight tensors may not be LeCun initialized, which may lead to scale instability if the activation functions do not compensate. Layer operations that we have not discussed in this work may be present, which may induce different kinds of nonlinearity or partially invalidate table \ref{tableNLCPropagation}. With the understanding of the material presented in this work, especially figure \ref{tableNLCPropagation}, it should be somewhat straightforward to adapt nlnorm to more complex situations. For example, if an architecture contains multiple activation functions, we can simply choose a linearization method for each of them and then either choose a single $l$ value for all activation layers or separate $l$ values. Alternatively, we may choose not to linearize activation layers that play a ``gating'' role as they occur in e.g. LSTM \citep{LSTM} or GRU \citep{GRU}. \citet{selu,saneInit,normProp} also provide examples of how to conduct ``normalization by recursion'', i.e. normalizing layers closer to the output based on assumptions derived from the normalization of layers closer to the input. For this work, the form in figure \ref{boxnlnorm} is sufficient. Providing explicit, well-defined generalizations is future work.

While $\mathfrak{n}_{c\ddot{\tau}(l,s) + b}$ is invariant to $b$ and $c$, replacing $\ddot{\tau}(l,s)$ with $c\ddot{\tau}(l,s)+b$ can and often does change the mean field NLC. If the replacement changes any $\mathfrak{q}$ or $\mathfrak{c}$ value, there can be knock-on effects throughout the architecture. For example, from our discussion in section \ref{actFunNLCsection} we can see that debiasing an activation function can cause the mean field NLC to diverge exponentially from one macro-layer to the next when it was originally converging. In general, we would expect activation function debiasing not to decrease the mean field NLC, because $\mathfrak{n}_{\tau}(c)$ is a decreasing function of $c$, as discovered in section \ref{actFunNLCsection}. We also find this to be true empirically in section \ref{nlnormControlSection}.

As usual, and as described in section \ref{metricEstimationSection}, we can compute the Gaussian expectations involved in the calculation of $b$ and $c$ accurately using standard numerical integration techniques.

\subsection{Linearization methods} \label{linearizationSection} 

We term a method for converting an activation function $\tau(s)$ into a parametrized version $\ddot{\tau}(l,s)$ a `linearization method'. $\ddot{\tau}(l,s)$ itself becomes an (activation function, linearization method) pair or `AFLM'. We introduced these terms in section \ref{studyATrainingSection}. See figure \ref{AFLMillu} for examples of AFLMs used in this work. The simplest linearization method, which features most prominently, is `linear interpolation'.

$$\tau(s) \rightarrow \tau(s) + ls$$

What values of $\mathfrak{n}_{\ddot{\tau}(l,s)}(1,0)$ can be obtained from this? Consider the theory of section \ref{nlcLinearApproximationSection}. Let $a_\tau s + b_\tau$ be the least squares linear fit to $\tau$ under unit Gaussian input and let $\tilde{\tau} = \tau - a_\tau s - b_\tau$ be the nonlinear component of $\tau$. Then 

$$\mathfrak{n}_{\ddot{\tau}(l,s)}(1,0) = \sqrt{\frac{\mathfrak{n}_{\tilde{\tau}}(1,0)^2\mathbb{E}_{s\sim\mathcal{N}(1,0)}\tilde{\tau}(s)^2 + (a_\tau + l)^2}{\mathbb{E}_{s\sim\mathcal{N}(1,0)}\tilde{\tau}(s)^2 + (a_\tau + l)^2}}$$

As $l$ converges to infinity or minus infinity, $\mathfrak{n}_{\ddot{\tau}(l,s)}(1,0)$ converges to 1. When $l = -a_\tau$, $\mathfrak{n}_{\ddot{\tau}(l,s)}(1,0) = \mathfrak{n}_{\tilde{\tau}}(1,0) \ge \sqrt{2}$ by theorem \ref{finiteNetTilde} and proposition \ref{finiteNetTheoremConnector}. Hence, as $l$ varies, $\mathfrak{n}_{\ddot{\tau}(l,s)}(1,0)$ attains at least all values between 1 and $\sqrt{2}$. In table \ref{basisillu}, we find that for popular activation functions $\tau$, $\mathfrak{n}_\tau(1,0)$ is significantly smaller than $\sqrt{2}$. So in practical situations, the desired $\mathfrak{n}_{\ddot{\tau}(l,s)}(1,0)$ value should always be attainable by linear interpolation. For popular activation functions, we also have $a_\tau \ge 0$. Hence, setting $l \ge 0$ ensures that $\mathfrak{n}_{\ddot{\tau}(l,s)}(1,0)$ varies monotonically between $\mathfrak{n}_\tau(1,0)$ and 1.

\section{Related Work} \label{nlnormRelatedWorkSection}

Mean and variance normalization is certainly a staple theme of deep learning. It is the foundation of normalization layers like BN and LN as well as methods that modify activation functions (e.g. normalization propagation \citep{normProp}), modify weight initialization (e.g. LeCun initialization, He initialization or LSUV \citep{saneInit}) or normalize weights directly (e.g. weight normalization \citep{weightNormalization}), as well as of novel activation functions (e.g. SELU \citep{selu}). As mentioned above, several of these methods normalize layers recursively. The key innovation of nlnorm is that nonlinearity is normalized in addition to mean and variance. Of course, this is not as straightforward in the sense that we must also choose an $l$ value, which nlnorm deliberately leaves open. In section \ref{bestNlcSection}, we already found that there cannot be a universal best NLC across tasks. In section \ref{nlnormBestNLCSection}, we will show that there is also variation in terms of which NLC is best for a given base architecture. Predicting the exact best NLC based on the base architecture definition is future work.

In the context of mean field theory, several studies have controlled nonlinearity indirectly by controlling weight initialization (e.g. \citet{eigenspectrum,depthScalesMeanField,meanFieldCNN}). Consider the $\mathfrak{n}_\tau(\lambda^2,0)$ curves from tables \ref{covCurveillu1} through \ref{covCurveillu4}. For many activation functions, the activation function NLC converges to 1 as $\lambda$ converges to zero. Specifically, this happens for activation functions that have a valid non-zero derivative at zero. If these activation functions are applied to small inputs, they mimic linear functions. The aforementioned studies generate small inputs by carefully reducing initial weight variance. This allows those studies to successfully train arbitrarily deep networks.

\section{Properties of nlnorm} \label{nlnormPropertiesSection}

In each subsection of this section, we establish a property of nlnorm empirically. This is similar to section \ref{nlcPropertiesSection}, where we establish properties of the NLC. We will use study B for this. nlnorm is based on theory that we did not explicitly present for convolutional architectures. The fact that it still works for such architectures is an indicator of robustness.

The architecture design and training protocol for study B has been detailed in section \ref{studyBSection} and summarized in section \ref{metricsSummarySection}. Study B contains a total of 552 architectures. Each of them uses a single activation function, which is based on one of 12 AFLMs given in table \ref{AFLMillu}. In this work, we found that the square and odd square activation functions induce Gaussian instability, which is harmful to performance, as shown in section \ref{gaussianStabilitySection}. Since nlnorm (unfortunately) does not eliminate Gaussian instability in general, we consider only architectures based on the other 10 AFLMs in this chapter, i.e. ReLU-interpolation through Gaussian. This leaves a total of 480 architectures. For each AFLM, $l$ has a default value that makes the AFLM revert to the basic activation function from table \ref{actFunIllu}. Study B contains 12 different activation functions per AFLM, corresponding to 12 different $(l,c,b)$ triplets for the expression $c\ddot{\tau}(l,s) + b$. In the first triplet, $l$ is set to the default value, $b$ is set to zero and $c$ is set to achieve $\mathbb{E}_{s \in \mathcal{N}(0,1)}(c\ddot{\tau}(l_\text{default},s))^2 = 1$. This corresponds to ensuring scale stability only. We consider this to be the ``base case'' that an architecture without nlnorm would use. As in our design of study A, we treat scale stability as a pre-established ZSAD guideline that is respected even in ``baseline architectures''. Hence, we consider architectures using this first triplet as our base architectures as defined above. Across the other 11 triplets, $l$ varies. Roughly, these 11 triplets correspond to 0\% linearization, 10\% linearization, etc. up to 100\% linearization. $(b,c)$ are set jointly to achieve $\mathbb{E}_{s \in \mathcal{N}(0,1)}(c\ddot{\tau}(l,s) + b)^2 = 1$ and $\mathbb{E}_{s \in \mathcal{N}(0,1)}c\ddot{\tau}(l,s) + b = 0$. Hence, architectures using one of these 11 triplets use nlnorm and are thus termed `normed architectures'. For each AFLM we considered in this chapter, for each triplet, study B contains 4 architectures. One of them uses BN, one uses LN, one uses BN and skip connections, and one uses neither normalization layers nor skip connections. In this chapter, we refer to the last of the four specifically as a ``vanilla architecture''. Each group of 12 architectures that share all properties except the $(l,c,b)$ triplet we term a `base group'. Finally, note that architectures vary in other properties, such as the presence of pooling or bias layers, between AFLMs. However, architectures that are based on the same AFLM do not vary except in the ways specified in this paragraph.

\subsection{nlnorm controls the NLC} \label{nlnormControlSection}

\begin{figure}
\centering
\includegraphics[width=0.98\textwidth]{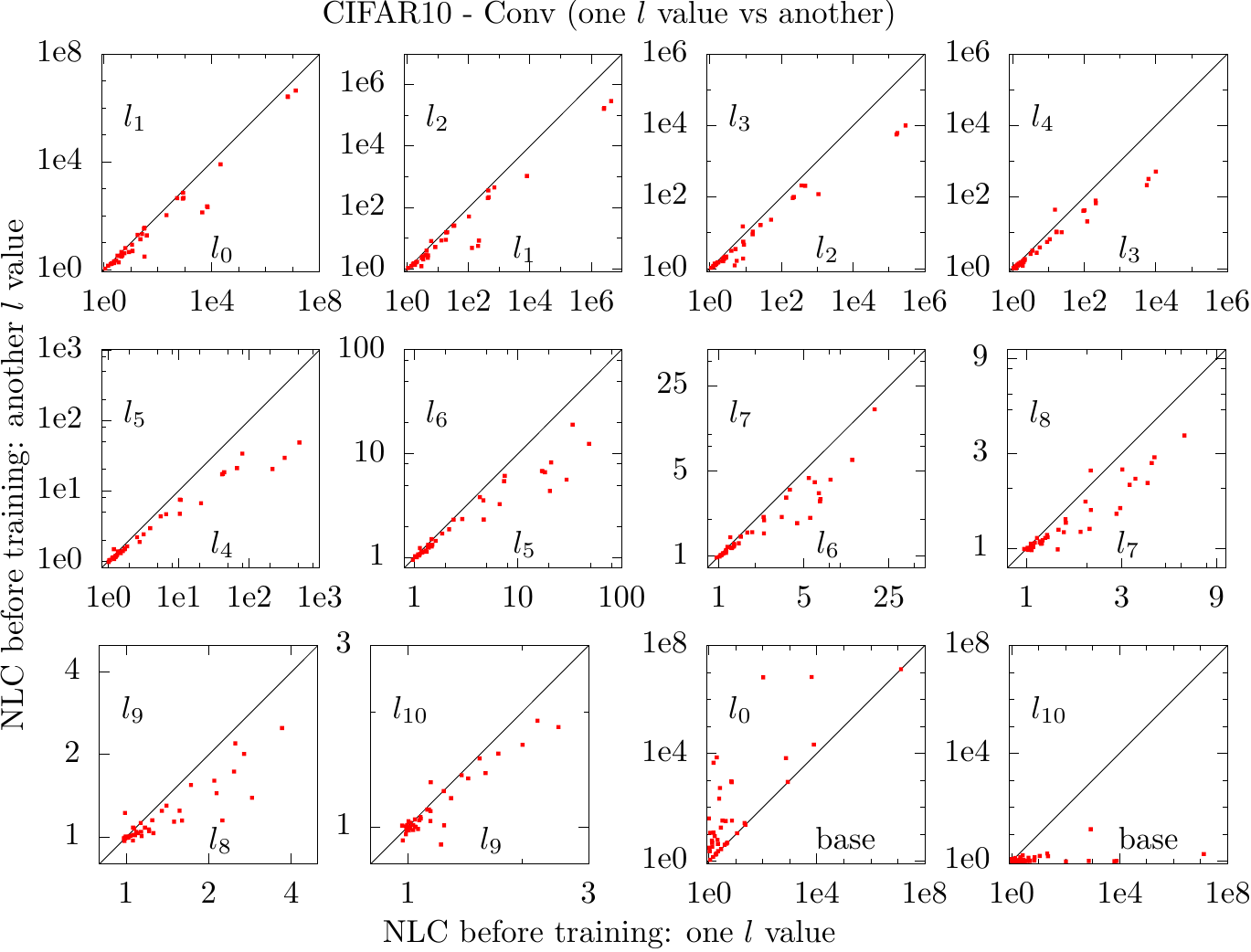}
\caption{Comparing the initial NLCs of pairs of architectures. The initial NLC of one architecture is plotted on the x-axis and the initial NLC of the other architecture is plotted on the y-axis. Both architectures in a pair are from the same base group. Each graph contains one marker per base group. The labels denote which members of the base group are used for each axis. {\it Conclusion:} $l$ values that lead to larger / smaller $\mathfrak{n}_{\ddot{\tau}(l,s)}(1,0)$ values generally lead to larger / smaller network NLCs in the initial state. $l_{10}$ induces an NLC close to 1. The base architecture NLC is approximately contained within the range of normed architecture NLCs.} \label{nlnormNlcControl}
\end{figure}

A core requirement for nlnorm to work is that it is effective at controlling the NLC of practical, finite-width architectures. In figure \ref{nlnormNlcControl}, we verify this. As described above, we consider a total of 40 sequences of 11 normed architectures that differ only in the value of $l$ and the induced values of $c$ and $b$. For each sequence, let the 11 $l$ values be ordered so that $l_0$ is the $l$ value that induces the highest $\mathfrak{n}_{\ddot{\tau}(l,s)}(1,0)$ value, and $l_{10}$ is the $l$ value that induces the lowest $\mathfrak{n}_{\ddot{\tau}(l,s)}(1,0)$ value. Note that the actual values taken by $l_0$ through $l_{10}$ differ between AFLMs. See section \ref{fullListB} for details.

For nlnorm to be effective, we require that for each base group, the architecture using $l_0$ has the highest NLC, the architecture using $l_1$ has the second highest, etc. The first 10 graphs of figure \ref{nlnormNlcControl} correspond to the 10 pairs $(l_0, l_1)$, $(l_1, l_2)$, .., $(l_9, l_{10})$. For each pair, we plot the initial NLC of the 40 architectures using the first $l$ value in the pair vs the NLC of the 40 architectures using the second $l$ value in the pair. In almost all cases, we find that \finding{the second architecture has a lower NLC than the first}, as desired. We also find that \finding{architectures using $l_{10}$ have an NLC close to 1}. This is also desired as we expect the ideal NLC to be somewhat close to 1 as well. We want the range of NLCs covered by nlnorm to contain the ideal NLC.

For nlnorm to be effective, we also require that the NLC of the base architecture is within the range of NLCs covered by the corresponding normed architectures. If this is not the case, then nlnorm may force us into an inferior NLC. In the bottom right of figure \ref{nlnormNlcControl}, we plot the initial NLCs of the 40 base architectures vs the initial NLCs of the corresponding normed architectures using $l_0$ and $l_{10}$ respectively. We find that \finding{the NLC of the base architecture is either between or very close to being between the other two NLCs}, as desired. Each of our AFLMs allows us to choose $l$ values so that $\mathfrak{n}_{\ddot{\tau}(l,s)}(1,0)$ is either exactly or arbitrarily close to 1. With such an $l$ value, we can reduce the NLC relative to the base architecture. Note that we cannot always get the NLC exactly to 1, because some nonlinearity is contained in normalization and max pooling layers that arise in some architectures. Each of our AFLMs also allows us to choose the default $l$ value so that $\ddot{\tau}(l,s)$ reverts to the basic activation function from table \ref{actFunIllu}. As mentioned above, we expect the introduction of debiasing not to decrease the NLC. Hence, by choosing $l_\text{default}$ within nlnorm, we expect an NLC approximately at least as large as the base NLC.

More generally, ensuring scale stability and low neuron bias in normed architectures causes the modeling of activation function inputs as unit Gaussians to be accurate, and therefore the simple version of nlnorm given in figure \ref{boxnlnorm} to be effective. We found that \finding{setting $b$ and $c$ always more or less ensures scale stability and low neuron bias in our architectures, with the exception of vanilla architectures based on abs. val. with a low degree of linearization}. We found those architectures to also exhibit Gaussian instability, and they are considered GUAs in section \ref{metricTerminologySection}. We also find in tables \ref{covCurveillu1} through \ref{covCurveillu4} that \finding{debiased abs. val. exhibits the highest degree of mean field Gaussian instability among activation functions that do not involve squaring}, at least as soon as layer quadratic means grow slightly larger than 1.

\subsection{nlnorm finds good nonlinearity levels} \label{nlnormWorksSection}

\begin{figure}
\centering
\includegraphics[width=0.98\textwidth]{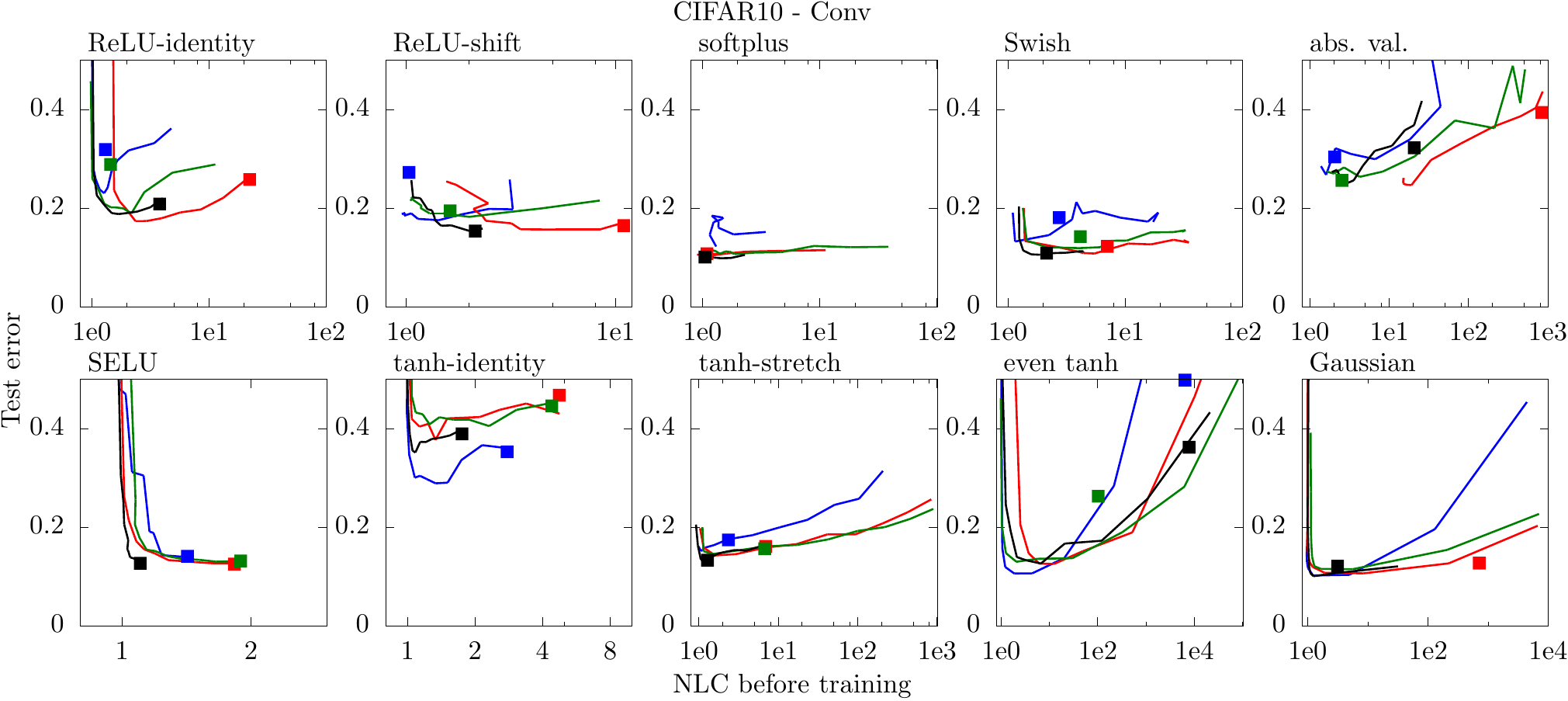}
\caption{Initial NLC vs test error, broken down by base group. Blue markers correspond to vanilla architectures. Red markers correspond to BN architectures. Green markers correspond to LN architectures. Black markers correspond to BN-ResNet architectures. Squares correspond to base architectures. Lines arise from connecting points corresponding to the normed architectures within a base group with neighboring values of $l$. {\it Conclusion:} In the majority of cases, the lowest test error is reached at an intermediate NLC value, with the smallest and largest NLC values performing significantly worse.} \label{nlnormSeries}
\end{figure}

nlnorm is based on the insight that an insufficient NLC leads to underfitting (see e.g. section \ref{nlcLinearApproximationSection}) whereas an excessive NLC leads to overfitting (see e.g. section \ref{nlcNoiseSection}). We desire that the optimal tradeoff between both phenomena is obtained by an NLC that is within the range of NLCs obtained by varying $l$. In figure \ref{nlnormSeries}, we verify this.

We plot the initial NLC vs test error for our 480 architectures. The raw x- and y-axis values are identical to figure \ref{nlcPredTestInit}D, but the visual presentation is different. Each of the 10 graphs of figure \ref{nlnormSeries} depicts results for one AFLM, and each color in a graph corresponds to one base group. Blue corresponds to the vanilla base group, red to the BN base group, green to the LN base group and black to the BN-ResNet base group. The line is generated by connecting markers corresponding to normed architectures in the base group with neighboring values of $l$. The square represents the base architecture in the base group. Indeed, we find that \finding{for most base groups, the lowest test error is attained at a point that does not correspond to the smallest or largest NLC}. Further, \finding{test error tends to increase in both directions when moving away from this optimal point}.

There are some exceptions to this pattern. \finding{For the SELU AFLM, the optimal test error is reached with the largest NLC considered}. We have $\ddot{\tau}_\text{SELU}(l,s) = \tau_\text{SELU}(s) + ls$. We only considered non-negative values of $l$. Hence, the largest $\mathfrak{n}_{\ddot{\tau}(l,s)}$ value we obtained was around 1.035 by table \ref{basisillu}. A larger value might have been ideal, especially for SELU-BN-ResNet. This shows that it can be important to choose linearization methods that can make an activation function less linear, rather than only more linear. In contrast to SELU, we find that \finding{for softplus and abs. val., the least NLC considered attains the lowest or close to the lowest test error}. This points to an important difference between linearization methods. Consider $\ddot{\tau}_\text{softplus}(l,s) = \frac{1}{l}\log(1 + e^{ls})$ and $\ddot{\tau}_\text{abs. val.}(l,s) = |s + l|$. When $l$ is small / large respectively, then $\mathfrak{n}_{\ddot{\tau}(l,s)}(1,0) \approx 1$. However, $\mathfrak{n}_{\ddot{\tau}(l,s)}(\lambda^2,0)$ can still be significantly larger than 1 for larger $q$. In fact, no matter the value of $l$, $\mathfrak{n}_{\ddot{\tau}(l,s)}(\lambda^2,0)$ converges to 1.21 for softplus and to 1.65 for abs. val. as $\lambda$ converges to infinity. Hence, growth of the parameter value during training, as studied in section \ref{meanFieldPracticalEmpiricalSection} and discussed in section \ref{forwardStabilitySection}, can cause the activation function NLC with respect to $\hat{\mathfrak{q}}_k$ to increase and hence counterbalance the effect of excessive nonlinearity normalization. This is not true to the same degree for activation functions linearized with linear interpolation. 

In figure \ref{nlnormSeries}, we also find that \finding{for most base groups, the performance of the base architecture is close to the performance of normed architectures with comparable NLC. For three AFLMs, there is significant deviation: softplus, even tanh and Gaussian. For softplus and Gaussian, while the BN and BN-ResNet base architectures perform comparably to their normed counterparts, the vanilla and LN base architectures attain a random error of 0.9}, and hence are not depicted in figure \ref{nlnormSeries}. These four base architectures are the ones that attain a very high initial LBIAS value, larger than $10^5$, as depicted in e.g. figure \ref{beyondBiasInit}F/G. This demonstrates the importance of eliminating bias with nlnorm. \finding{For even tanh, similarly, the vanilla and LN base architectures have significant LBIAS}. We have not investigated why the vanilla base architecture nevertheless outperforms normed architectures with comparable initial NLC. We note that the test error of that base architecture is still highly suboptimal overall. The even tanh-BN base architecture has an NLC greater $10^4$ and a test error greater than 0.5, and hence does not show up in the figure.

We end with two remarks. First, the non-smoothness of many curves in figure \ref{nlnormSeries} is most likely due to the fact that we only conducted a single training run per starting learning rate and architecture. If we had averaged over multiple runs corresponding to different random number sequences / initializations, we would expect results to be much less noisy. Our already high computational budget was not sufficient to do this for every architecture. Since searching for the best $l$ value and starting learning rate in a 2D grid is expensive, we expect that this will generally have to be done without multiple runs per grid point in practical situations. Hence, we were content to give results in figure \ref{nlnormSeries} that depict what one can expect from a single run. However, we would recommend that in practice, at least the same initial trainable parameter value is used for all values of $l$ and starting learning rates to reduce noise. We did not have the chance to do this. See section \ref{codeLimitationsSection}.

Second, we note that in figure \ref{nlnormSeries}, \finding{the lowest error value achieved differs significantly between AFLMs}. Since architectures corresponding to different AFLMs differ in ways other than activation function, we cannot draw a conclusion about the relative strength of AFLMs ``in a vacuum''.

\subsection{nlnorm improves performance} \label{nlnormPerformanceSection}

\begin{figure}
\centering
\includegraphics[width=0.98\textwidth]{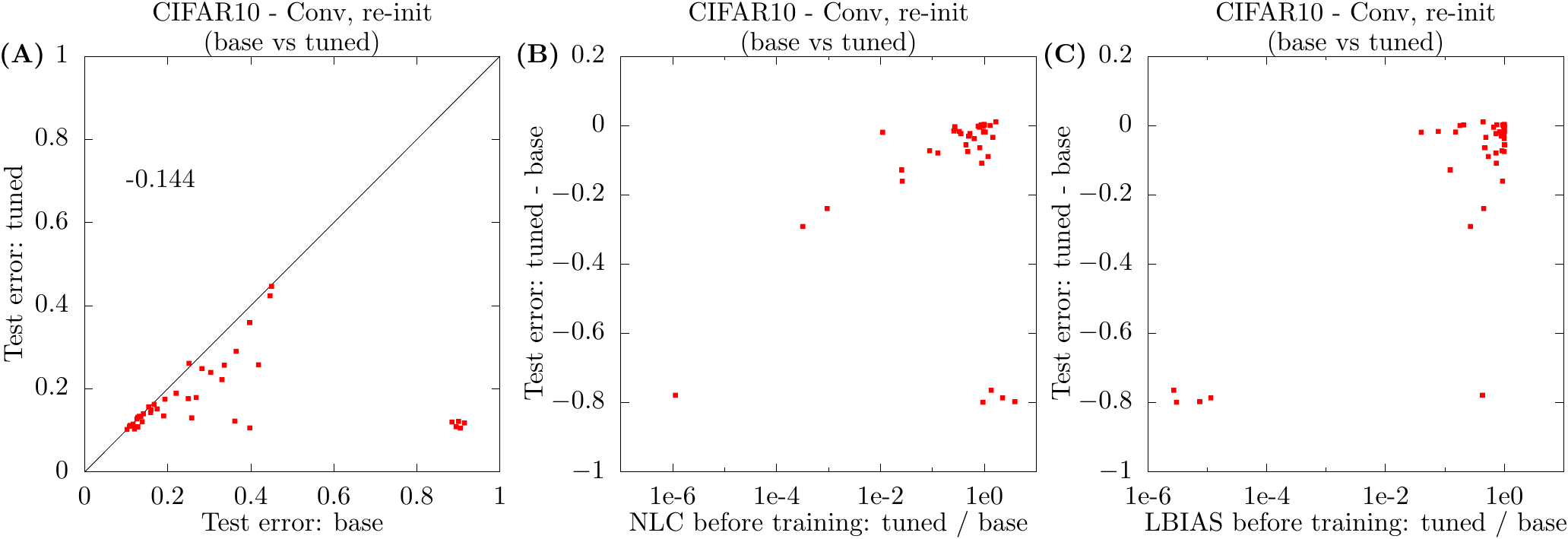}
\caption{Comparison of error, initial NLC and initial LBIAS between base and tuned architectures. All results are averaged over 10 re-training runs. Quantities depicted in log scale are averaged in log space. {\it Conclusion:} nlnorm drastically improves test error for many base architectures and never significantly increases test error. Larger NLC reduction corresponds to larger error reduction. Very large LBIAS reduction corresponds to very large error reduction.} \label{nlnormPerformance}
\end{figure}

In the previous subsection, we found that there were often values of $l$ that induced a more ideal NLC and reduced test error relative to the base architecture. In this subsection, we explicitly investigate the test error gain induced by nlnorm. To quantify test error gain, we need to quantify baseline test error and ``test error with nlnorm'', then take the difference. We would like to define the ``test error with nlnorm'' as the minimal test error attained by a normed architecture for any value of $l$. Of course, this sort of minimum suffers from the sharp valley problem, as discussed in section \ref{sharpValleySection}, and as it arises for e.g. starting learning rate in section \ref{learningRateSection}. In light of this, we define ``test error with nlnorm'' as follows. We choose the $l$ value out of the 11 values we considered for each base group that yielded the lowest test error. We call the architecture using this $l$ value the `tuned architecture'. Then we re-train the tuned architecture with 10 different random number sequences, which also induce 10 different initial parameter values, with the SLR that was best for the tuned architecture during the original runs. Finally, the test error gain is defined as the difference between the average test error attained from those 10 reruns minus the average test error obtained from 10 reruns of the base architecture, again using the best SLR from the original training of the base architecture. This re-training is also described in section \ref{studyBTrainingSection} / \ref{metricsSummarySection}.

A caveat of this procedure is that the computational expense of determining the performance with nlnorm is much greater than that of determining the baseline performance, because we need to train architectures for many $l$ values, which gives nlnorm an inherent advantage. When we chose our $l$ values for each AFLM, we tried to infuse as little prior knowledge as possible so as to not compromise the scientific validity of our results. We included $l$ values that we very strongly suspected were suboptimal based on prior analysis. This means that the number of $l$ values that need to be investigated in practice to observe the kind of performance gains we observe should be significantly less than 11. We further investigate this question in section \ref{nlnormBestNLCSection}.

In figure \ref{nlnormPerformance}A, we plot the average error from the base architecture reruns vs the average error from the tuned architecture reruns. The {\bf re-init} label above a graph such as in figure \ref{nlnormPerformance} indicates that the results depicted are averages from our 10 reruns. Metrics depicted in log scale are averaged in log space. We find that \finding{nlnorm leads to massive test error gains for many base architectures, and never leads to significant losses. The average test error gain was 0.144.} This is perhaps the second most important empirical finding in this work after section \ref{nlcPredictiveSection}. In that section, we showed the NLC is predictive of performance. Now we show that the NLC, along with $LBIAS_l$, are {\it causal} for performance, i.e. making a minimal modification in the nonlinear layers of the architecture leads to significant test error changes.

A caveat is that, while nlnorm improved the test error of architectures that did not previously perform well, it did not significantly improve the test error of architectures that already performed close-to-optimal. The lowest test error achieved by any base architecture was very close to the lowest test error achieved by any tuned architecture.

\begin{table}
\centering
\begin{tabular}{p{4.2cm}ccc}
Base / tuned architecture & Rank & Test error: re-init, base &Test error: re-init, tuned\\ \hline\hline
softplus-BN-ResNet&1&0.102&0.102\\
Gaussian-BN-ResNet&2&0.120&0.103\\
Gaussian&3&0.900&0.105\\
even tanh&4&0.398&0.106\\
Gaussian-BN&5&0.128&0.108\\
softplus-LN&6&0.900&0.108\\
softplus-BN&7&0.108&0.109\\
Swish-BN-ResNet&8&0.110&0.111\\
Swish-BN&9&0.117&0.115\\
Gaussian-LN&10&0.900&0.117
\end{tabular}
\caption{Tuned architectures that achieve the 10 lowest test error values in CIFAR10 - Conv. The test error of the corresponding base architectures is given for comparison. Values are averaged over re-training runs. {\it Conclusion:} nlnorm can help activation functions perform competitively that would otherwise perform suboptimally.} \label{nlnormActFun}
\end{table}

The \finding{largest test error gains, which correspond to points depicted in the bottom right of figure \ref{nlnormPerformance}A, correspond to softplus, softplus-LN, Gaussian, Gaussian-LN and even tanh-BN}. We discussed those architectures in the previous subsection.

In figure \ref{nlnormPerformance}B, we plot the test error gain on the y-axis vs the ratio of tuned NLC over base NLC across the 10 reruns on the x-axis. Disregarding our 5 outliers, we find that \finding{the larger the NLC reduction induced by nlnorm, the greater the test error gain tends to be}. In figure \ref{nlnormPerformance}C, we plot the ratio of LBIAS values vs test error gain. \finding{The points in the bottom left corner indicate that softplus, softplus-LN, Gaussian and Gaussian-LN exhibit random performance without the debiasing of nlnorm. Disregarding those outliers, we do not see a relationship between LBIAS reduction and error reduction}.

\subsection{nlnorm opens up activation function design} \label{nlnormActFunDesignSection}

\begin{figure}
\centering
\includegraphics[scale=0.8]{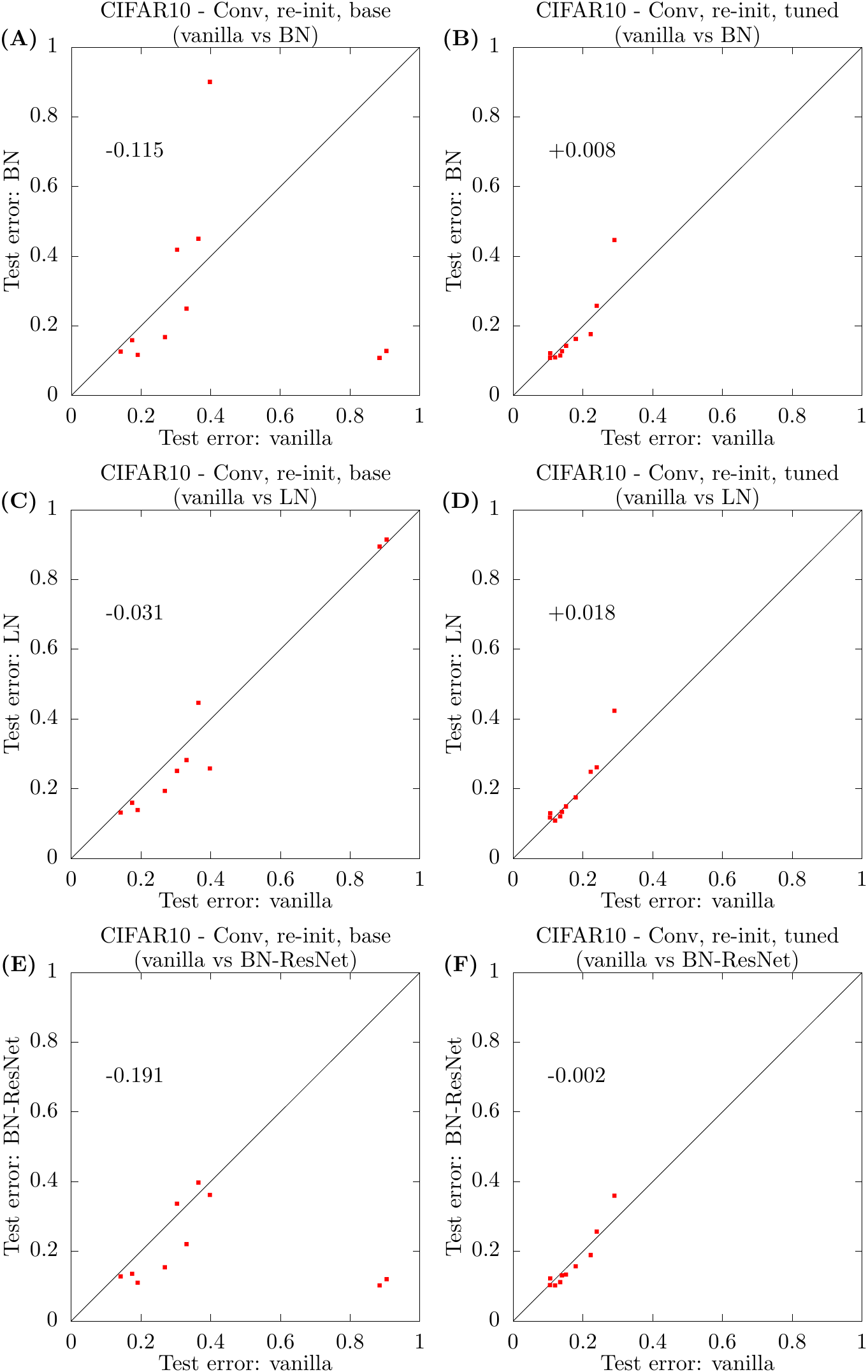}
\caption{Test error of vanilla architectures vs test error of architectures using normalization layers / skip connections, by normalization layer / skip connection type. Base architectures are depicted on the left-hand side, tuned architectures on the right-hand side. Results are averaged over re-training runs. The test error difference, averaged over AFLMs, is given as a number in each graph. We omit correlation values due to the impact of outliers. {\it Conclusion:} Normalization layers and skip connections boost performance without nlnorm, but have a small impact with nlnorm.} \label{nlnormActNorm}
\end{figure}

As mentioned above, we cannot use study B to obtain a representative performance ranking of AFLMs because architectures using different AFLMs differ in ways unrelated to the AFLM. However, it is worth pointing out that different activation functions are affected differently by nlnorm. In table \ref{nlnormActFun}, we rank the tuned architectures that attain the lowest test error. We find that \finding{some of them have a corresponding base architecture error that is very close, such as softplus-BN-ResNet}. However, \finding{for some tuned architectures, the corresponding base architecture has high error, or even random error}. Hence, the set of activation functions that can yield competitive performance with nlnorm is much greater than the set of activation functions that can yield competitive performance without nlnorm. Our results indicate that, along with mean field Gaussian instability, a suboptimal degree of nonlinearity is the number one obstacle for activation function design. nlnorm can take care of this issue automatically.

When we originally designed our studies, we invented the Gaussian and even tanh activation function because they seemed different from popular activation functions. If we only used them outright in our architectures, we would have to conclude that they are vastly inferior to popular activation functions. However, nlnorm paints a rather different picture. By controlling bias and nonlinearity, we are capable of even using ad hoc activation functions in high-performing architectures. This somewhat calls into question the myriad rationalizations that exist for carefully designed activation functions proposed over the years. Investigating the value of these activation functions modulo nonlinearity is an interesting topic for future work.

\section{nlnorm vs batchnorm vs layernorm} \label{nlnormvsbnsection}

It is natural to compare nlnorm with BN and LN, which feature prominently in this work and in popular architectures. It is also worth comparing nlnorm to skip connections because, as we argued in section \ref{pathEquationSection} and further show in section \ref{skipConnectionsSection}, skip connections also tend to reduce the NLC. In this section, we perform this comparison from a performance standpoint. In chapter \ref{surveyChapter}, we perform a qualitative comparison.

In figure \ref{nlnormActNorm}, we plot the test error values obtained with various architecture types. In graphs A/C/E, we plot the test error of vanilla base architectures vs the test error of corresponding base architectures with BN, LN and BN-ResNet respectively. There is one point in each graph for each AFLM. The two architectures in each pair differ only in their use of normalization layers / skip connections. We designed study B specifically to enable a clean comparison. In each graph, we give the average test error gain obtained from introducing normalization layers and / or skip connections across AFLMs as a number. In graphs B/D/F, we make the equivalent comparison between pairs of corresponding tuned architectures. We find that \finding{BN, LN and especially BN-ResNet boost performance significantly when nlnorm is not used, but have a much lesser impact when nlnorm is used}. Hence, at least in our experiments, and at least from a test error standpoint, nlnorm greatly reduces the need for normalization layers and skip connections. Of course, there are situations where normalization layers confer critical benefits not conferred by nlnorm, as will become clear in sections \ref{surveynlnormSection}. However, some of the benefits of activation layers clearly overlap with the benefits of nlnorm. For example, both nlnorm and BN have a debiasing effect.

This analysis underscores the lesson from section \ref{beyondNlcSummarySection}. The perceived performance impact of architecture building blocks like BN / LN / skip connections greatly depends on the way we design our baseline architectures. If we consider nonlinearity normalization as a crucial step in the design of any architecture and that the NLC needs to be tuned as a hyperparameter like learning rate, then we may perceive these normalization layers in a very different light than if we were oblivious to the importance of nonlinearity and neuron bias. There are many other building blocks and strategies that impact NLC and / or $LBIAS_l$ beyond those discussed in this work. It is interesting to consider what fraction of the performance gains of e.g. DenseNet \citep{denseNet}, LSUV \citep{saneInit}, ELU \citep{elu} or other activation functions can be explained by their impact on NLC / $LBIAS_l$. Leaky ReLU \citep{leakyRelu} and PReLU \citep{heInit} are similar to our ReLU-interpolation AFLM.

\section{What is the best NLC for an architecture?} \label{nlnormBestNLCSection}

In section \ref{bestNlcSection}, we investigated how to determine a range of good NLC values for a given task. When using nlnorm, it is possible to choose an $l$ value that immediately yields an architecture with an initial NLC inside this range. If the initial NLC of the base architecture lies significantly outside the optimal range, we should see a performance boost from this $l$ value. However, ideally, we do not simply want to choose a good $l$ value, but the best $l$ value for that base architecture. (Again, we disregard sharp valleys. See section \ref{sharpValleySection}.) In this section, we investigate whether there is a single best NLC across base groups for the task on which study B is based, which is CIFAR10 classification.

\begin{figure}
\centering
\includegraphics[width=0.49\textwidth]{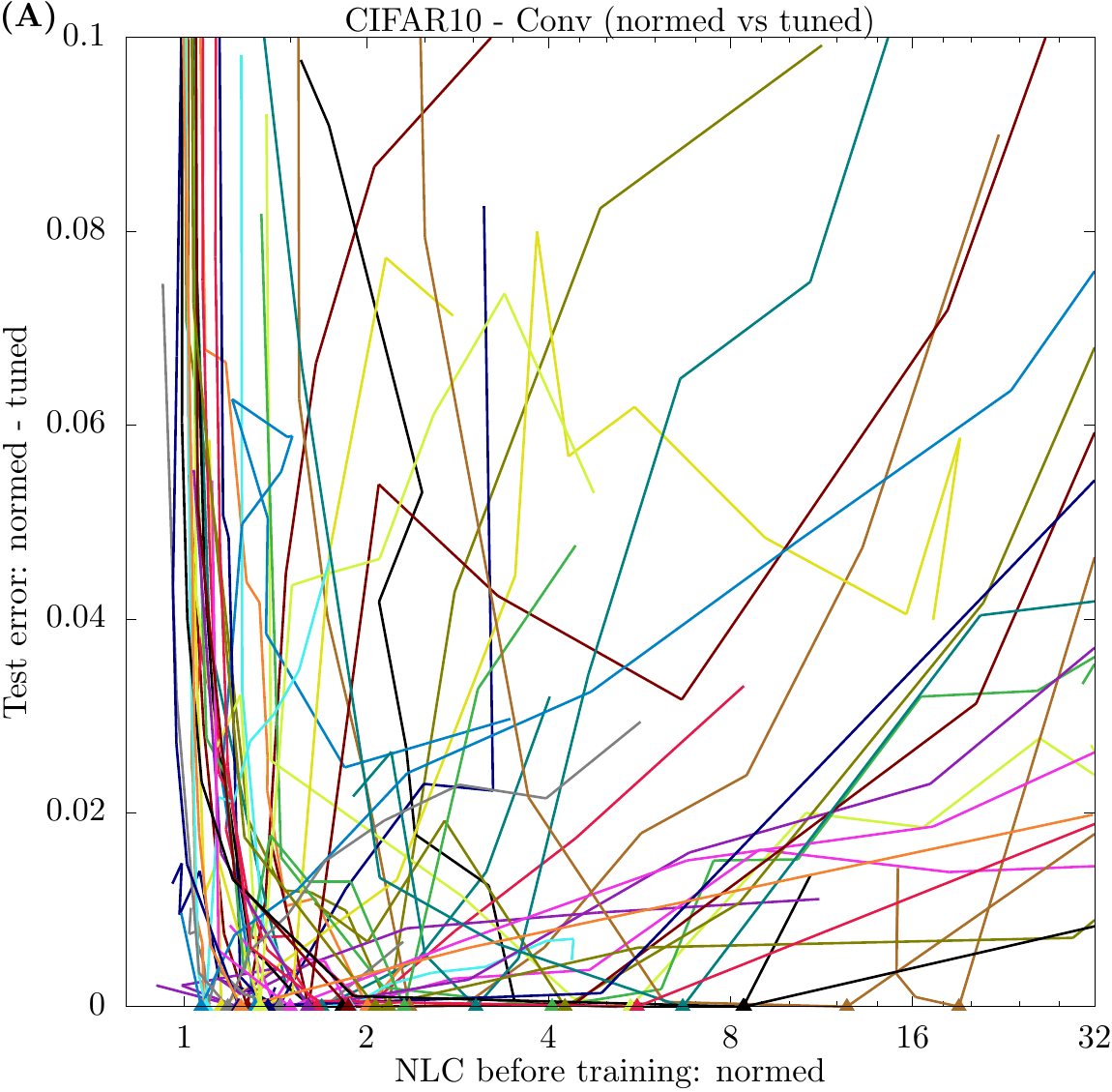}
\includegraphics[width=0.49\textwidth]{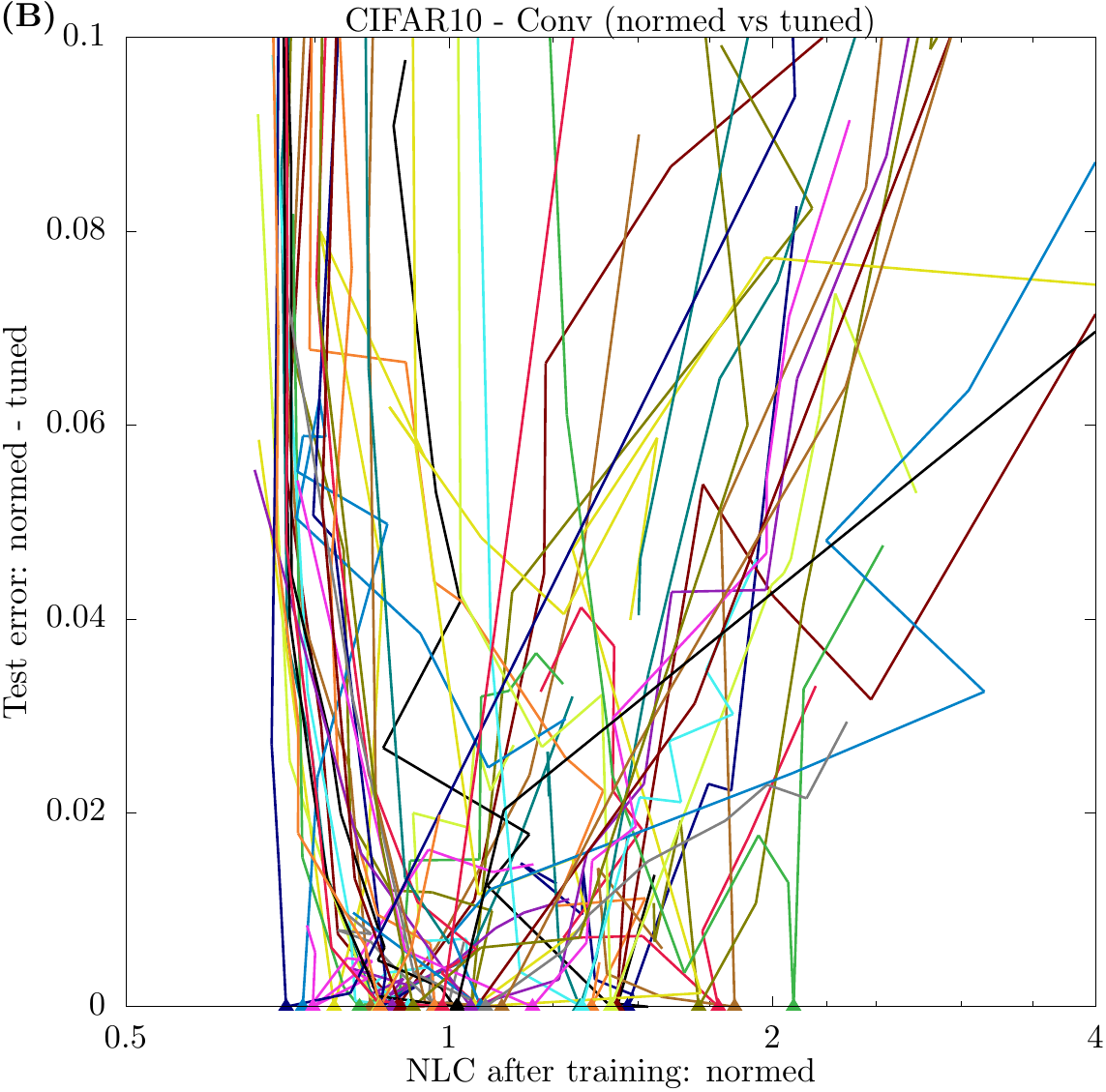}
\caption{Initial NLC (graph A) and final NLC (graph B) vs test error. Each curve is obtained by connecting points corresponding to normed architectures from the same base group with neighboring $l$ values. The test error values are normalized by subtracting the test error of the corresponding tuned architecture. Triangles placed on the x-axis depict tuned architectures. {\it Conclusion:} While there is a definite range that contains the best NLC value for each base group, both before and after training, there is no NLC that universally leads to optimal or even close-to-optimal performance relative to other members of the same base group.} \label{nlnormChaos}
\end{figure}

\begin{wrapfigure}[20]{r}{5.5cm}
\includegraphics[width=5.4cm]{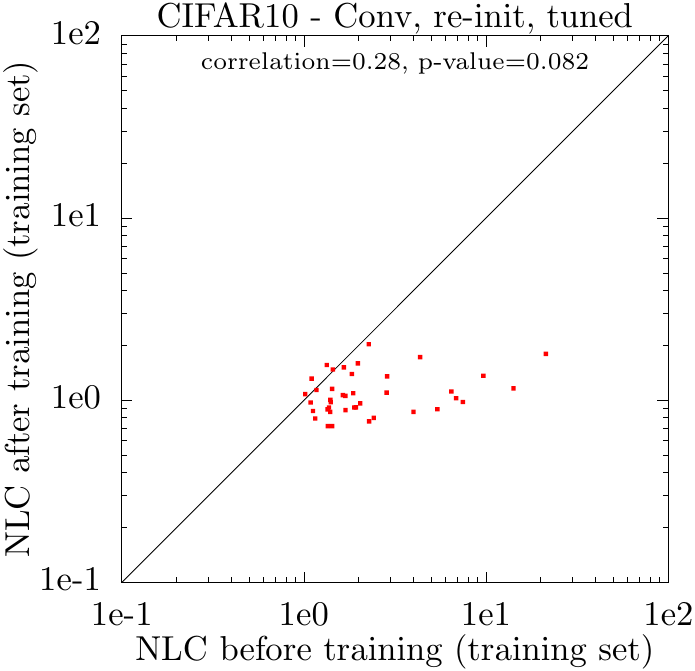}
\caption{NLC before training vs after training for tuned architectures, averaged over re-raining runs. {\it Conclusion:} There is no significant relationship.}\label{nlnormChaosTrans}
\end{wrapfigure}

In figure \ref{nlnormChaos}A, we plot curves similar to the curves of figure \ref{nlnormSeries}. The initial NLC is on the x-axis and the test error is on the y-axis. Each curve corresponds to the normed architectures in a base group. The difference is that all curves are in the same graph, and that we normalize the test error of each normed architecture by subtracting the test error of the corresponding tuned architecture. This means that the test error plotted for tuned architectures is zero. The location of tuned architectures in the graph is marked by a little triangle of the same color as the corresponding curve, placed on the x-axis. In essence, we depict the test error penalty incurred by applying nlnorm with a suboptimal $l$ value. The results are not encouraging, though not necessarily unexpected. We find that \finding{there is no best initial NLC across base groups. There is not even a single initial NLC value for which all base groups perform reasonably close to their optimum}. On the one hand, this underscores the need for nonlinearity tuning. On the other hand, it means that some trial-and-error is likely necessary to find a close-to-optimal $l$ value. To get the most out of nlnorm, we have to be willing to make a computational investment.

In figure \ref{nlcPredictiveSection}, we found that the range of good initial NLC values for CIFAR10 is between 1 and 5. In figure \ref{nlnormChaos}A, we indeed find that \finding{for some base architectures, the best initial NLC is very close to 1}. This is perhaps unfortunate, as other architectures severely underfit for such small initial NLCs. We also find that for \finding{7 base groups, the best initial NLC is larger than 5.} This suggests that the base architecture was in some way suboptimal to begin with. Further investigation is required.

We were interested in whether there are patterns in terms of NLC after training. In figure \ref{nlnormChaos}B, we plot the final NLC vs test error, in the same way as in figure \ref{nlnormChaos}A. The range of best NLCs is different. It begins as low as 0.7 and ends just above 2. Again, \finding{there is no final NLC that works well for all base groups}. In figure \ref{nlnormChaosTrans}, we plot the initial vs final NLC for tuned architectures, averaged over our 10 re-training runs. We find no significant relationship between the two. In summary, at least to the degree to which we have studied the phenomenon, the precise best NLC value for a base architecture appears to be a chaotic property.

Starting learning rate is a key hyperparameter. While it is tunable, the best value is difficult to predict. While nlnorm turns the initial NLC into a tunable hyperparameter, its best value is also difficult to predict.

\section{Impact of lack of validation set} \label{nlnormOverfitSection}

\begin{figure}
\centering
\includegraphics[width=5.4cm]{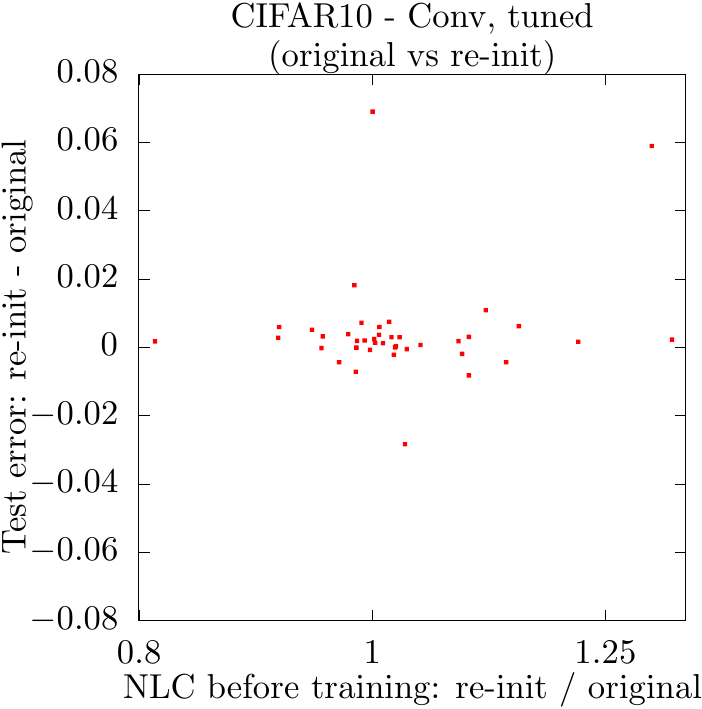}
\caption{The impact of re-training on the NLC and test error of tuned architectures. x-axis: Ratio of the initial NLC averaged across re-training runs over the initial NLC from the original run. y-axis: Difference of test error averaged across re-training runs minus the test error from the original run. {\it Conclusion:} Overall, there is little deviation between the original run and the re-training runs. There also appears to be no bias in any direction, and no trend between the two plotted values.}\label{nlnormOverfit}
\end{figure}

In study B, we did not have access to a validation set, as explained in section \ref{codeLimitationsSection}. Therefore, we had to choose the best value of $l$ and SLR by considering the least test error. This procedure could overfit on the test set. While we cannot determine the extent of this problem exactly, we can determine the extent to which the test error and NLC change from one random number sequence to the next. If our original sweep over $l$ and SLR overfits on the test set, we would expect a different random number sequence to yield a somewhat higher test error and / or a different NLC when applied to the chosen $l$ and SLR.

In figure \ref{nlnormOverfit}, on the x-axis, for each tuned architecture, we plot the ratio of the initial NLC averaged across the 10 re-training runs over the initial NLC from the original run. The original run was the one that was used to choose the tuned architecture among normed architectures in the base group. We find that \finding{the ratio is close to 1, and that there exists no systematic bias in terms of the tuning procedure selecting an $l$ or SLR value that yields a particularly large or small NLC with the original random number sequence relative to other random number sequences}. On the y-axis, for each tuned architecture, we plot the difference of the test error averaged across the 10 re-training runs minus the test error from the original run. We find that \finding{the difference is generally close to zero, and that there exists no systematic bias in terms of the tuning procedure selecting an $l$ or SLR value that, when paired with the random number sequence of the original run, yields a particularly large or small error relative to other random number sequences}. Further, we \finding{find no association between the x- and y-axis values.} Through further analysis, we also found that \finding{the largest error differences correspond to high absolute errors and the NLC ratios furthest from 1 in log space correspond to the largest absolute NLCs.}

While this analysis is not proof that our $l$ and SLR tuning procedure did not overfit on the test set, it provides significant evidence to the contrary. 

\section{Learning rates and nlnorm} \label{nlnormLearningRateSection}

In section \ref{learningRateSection}, we found that it is difficult to predict the best SLR for a given architecture. We end this chapter by investigating whether nlnorm can be helpful in this process.

\begin{figure}
\centering
\includegraphics[width=0.98\textwidth]{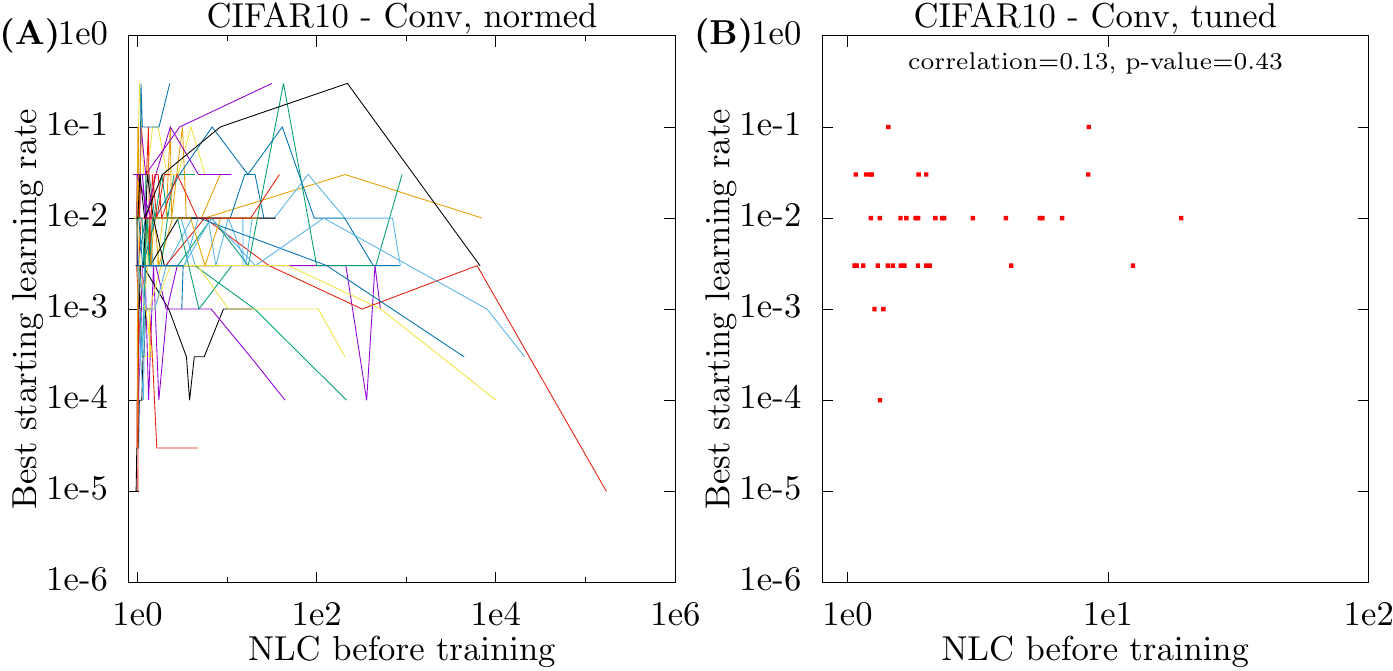}
\caption{Initial NLC vs best SLR for normed architectures. Axis values are equivalent to figure \ref{beyondLRvalsNLC}F. In graph A, we connect points corresponding to neighboring architectures within the same base group, as in e.g. figure \ref{nlnormSeries}. Also as in figure \ref{beyondLRvalsNLC}F, we omit architectures that did not achieve a better-than-random test error. Those architectures are simply skipped over when forming the curve for each base group. Graph B is equivalent to figure \ref{beyondLRvalsNLC}F restricted to tuned architectures. {\it Conclusion:} The best SLR can vary wildly even between architectures of the same base group with similar $l$ values. The range of best SLRs for tuned architectures may be somewhat narrower than for other architectures.} \label{nlnormlrate}
\end{figure}

In figure \ref{nlnormlrate}A, we plot the initial NLC vs best starting learning rate for all normed architectures. The x- and y-axis values are the same as in figure \ref{beyondLRvalsNLC}F. However, now we form a curve for each base group as in figures \ref{nlnormSeries} and \ref{nlnormChaos}. The results are surprising. We find that \finding{even between architectures from the same base group with similar initial NLCs, the best SLR can vary wildly}. Hence, performing exhaustive tuning of the SLR for one value of $l$ does not necessarily help for another value of $l$. A caveat to this analysis is that, as discussed in sections \ref{nlnormWorksSection} and \ref{codeLimitationsSection}, we had to use a different random number sequence / initial parameter value for each value of $l$ and SLR. Without this source of noise, the results may be somewhat different. 

In figure \ref{nlnormlrate}B, we again plot the same values as in figure \ref{beyondLRvalsNLC}F, but we restrict ourselves to tuned architectures. We find that \finding{while there is no single SLR that is optimal for all tuned architectures, all but one tuned architecture has its best SLR take one of five values}. For comparison, in figure \ref{beyondLRvalsNLC}F, the best SLR takes 13 different values. While this difference may partially be due to the smaller sample size in figure \ref{nlnormlrate}B, it corroborates our finding from section \ref{learningRateSection} that if we are content with finding the best SLR for the best architectures, we might not have to search as extensively.

\chapter{A survey and explanation of architecture design} \label{surveyChapter}

In this chapter, we examine and explain popular architecture building blocks and design strategies as given in section \ref{statusQuoSection} in light of the results of prior chapters. We provide an overview of the range of architecture behaviors that they induce. Each section in this chapter corresponds to a building block or design strategy. In section \ref{surveySummarySection}, we summarize this chapter by mapping strategies to guidelines in table \ref{surveySummary}.

In section \ref{statusQuoSection}, we stated that the key drivers behind popular neural architecture design strategies are (i) historical momentum as outlined in section \ref{historicalBiasSection}, (ii) computational efficiency as described in section \ref{programFunctionSection} and (iii) trial-and-error as described in section \ref{darkArtSection}. So, how is it possible to explain these strategies in terms of ZSAD guidelines that do not fall under those three umbrellas, that might even have been unknown at the time these strategies became popular? For a strategy to emerge as popular from the trial-and-error gauntlet of the deep learning community, it needs to be both simple and consistent. It needs to yield high performance from task to task without requiring a large amount of expertise or tuning. Strategies that have this property naturally tend to be explainable. In a largely empirical field like deep learning, it is common to build understanding through post-hoc explanation. In this chapter, we {\it rationalize} popular architecture designs.

\paragraph{Background from prior chapters} Throughout this chapter, we use the terminology, notation and conventions of section \ref{notationSummarySection}. Specifically, we repeatedly use and contrast the terms `architecture design strategy' and `ZSAD guideline' which are defined in section \ref{statusQuoSection} / \ref{notationSummarySection} and further explained in section \ref{statusQuoSection}. In places, we use the empirical studies laid out in chapter \ref{empiricalStudiesChapter} for validation. We recommend reading at least summary section \ref{metricsSummarySection}.

We extensively use the analysis and definitions of chapters \ref{meanFieldNnaChapter} and \ref{beyondNlcChapter} in our discussion. We recommend a strong familiarity with those chapters. Chapter \ref{nlcChapter} is largely not required as background, except for motivating the desire to achieve ZSAD guideline \ref{npcNLC}. Chapter \ref{nlnormChapter} is relevant in that nlnorm is one of the design strategies discussed in this chapter.

\paragraph{Technical considerations} As discussed in section \ref{nlcDefinitionSection}, we implicitly assume things like integrability (section \ref{integrabilitySection}), differentiability (section \ref{nonDifferentiableSection}) and non-zero denominators when necessary.

\paragraph{Limitations} In this chapter, we often focus on the behavior of building blocks and design strategies in the context of simple architectures. Design strategies have a complex interplay with each other which makes it often difficult to analyze them in a vacuum. Our objective is to be just general enough that the explanations we give for the design strategies actually reflect many of the reasons why those strategies became popular in the real world. Explaining specific designs beyond that level is not a focus of this work, as the goal of zero-shot architecture design is to establish principles that go beyond specific designs. We hope that this chapter can serve as a guide for the reader to explain any specific design they are interested in or to come up with their own designs.

We make extensive use of the results of section \ref{meanFieldPracticalSection}, including the nonlinearity path equation. As discussed at the end of the introduction of chapter \ref{meanFieldNnaChapter}, those results are technically restricted to A-architectures as defined in section \ref{aArchitectureSection} though we expect them to have much more general conceptual validity. We showed the nonlinearity path equation to be highly predictive for study A architectures and to inform NLC predictions and nlnorm for study B architectures in section \ref{meanFieldCNNsection} and chapter \ref{nlnormChapter} respectively. Architectures in neither study are technically A-architectures. Our definition of A-architectures reflects simple, popular designs including some of the strategies covered in this chapter. The fact that the nonlinearity path equation holds for architectures that use e.g. LeCun initialized linear layers, normalization layers placed before activation layers and addition layers that add a linear and a nonlinear path is another reason for the success of those strategies, as the nonlinearity path equation implies consistent nonlinearity levels and thus consistent performance.

\section{Layers and width} \label{layerSurveySection}

Arguably the most foundational design strategy for neural architectures is the use of layers, as we explained in sections \ref{layersSection} and \ref{architectureDesignParadigmsSection}. Layers come with obvious benefits. Defining an architecture in terms of layers is simpler than defining it in terms of neurons. Evaluating large numbers of identical computational units in parallel is computationally efficient on specialized hardware such as GPUs.

Large portions of this work, such as chapter \ref{nlcChapter} and parts of chapter \ref{beyondNlcChapter}, are deliberately layer-agnostic and build directly upon the functional-gradient paradigm as outlined in section \ref{functionalGradientSection}. In contrast, layers are critical to our analysis in chapter \ref{meanFieldNnaChapter}. Mean field theory is the leading framework for making specific, quantitative predictions of an architecture's behavior from its definition. It is based on taking the width of layers to their theoretical limit at infinity. Of course, without layers, there is no such thing as width. The wider a network is, the more ``layered'' we can consider it. Hence, layers underpin the predictability of architecture behavior, as well as the performance consistency required for designs to become popular as described above. The layer concept also underpins this very chapter. We discuss the ZSAD guideline of ``using an appropriate width'' further in section \ref{widthSection}.

\section{Linear layers and random initialization} \label{linearLayerSurveySection}

If layers in general are necessary for predictability with mean field theory, randomly initialized linear layers are necessary specifically. In chapter \ref{meanFieldNnaChapter}, we largely restrict ourselves to fully-connected layers and only briefly discuss convolutional layers in section \ref{meanFieldCNNsection}. The FC operation is one of two operations on which background theorem \ref{backgroundMaster} is based. The key property that is shared by FC and convolutional layers is that neurons in the initial state are the dot product of (near-)independent random weights and a large number of neurons in the dependency. This enables the application of the central limit theorem, which implies for certain input distributions that linear layers are meta-Gaussian meta-distributed (section \ref{meanFieldDistributionSection}), and that the random initial state yields effectively deterministic behavior. The practical predictiveness of mean field theory has been shown in this work and a large number of prior works.

In mean field theory, as the width of the dependency of an FC layer is taken to infinity, the initial weight variance of the FC layer must scale inversely with that width in order to yield convergent behavior. This implies that the FC layer must use a fixed multiple of the LeCun variance, which explains the importance of LeCun initialization. In section \ref{architectureDesignParadigmsSection}, we provide a basic explanation that does not reference mean field theory as to how LeCun and He initialization can induce scale stability.

Another important benefit of (wide) linear layers is that they have a high-dimensional parameter sub-vector. Parameter dimensionality is a key statistical measure of model complexity, as we further discuss in section \ref{widthSection}. For example, the dimensionality of the parameter sub-vector of fully-connected layers is $d_ld_k$, i.e. it scales as the square of architecture width.

Throughout the rest of this chapter, we largely do not discuss convolutional layers explicitly. They behave largely equivalently to fully-connected layers within the context of this chapter, except when paired with normalization layers. Of course, additional issues arise with convolutional layers, as discussed in e.g. \citet{meanFieldCNN}, that go beyond the scope of this work.

\section{Activation layers}

Neural networks require layers that make them non-linear. Activation layers apply the same nonlinear functions to each component of the dependency and usually do not have a trainable parameter sub-vector. Hence, activation layers can be viewed as the simplest possible nonlinear layer. We are not aware of any fundamental reason why nonlinearity should be introduced specifically in elementwise layers. For example, consider background theorem \ref{backgroundMaster}. While it is stated in terms of an elementwise operation, it could be generalized to many other nonlinear operations that behave in a regular and reasonable way as width converges to infinity. By reducing the complexity of nonlinear layers down to activation layers, we can derive comprehensive results, such as expressing mean field properties of architectures in terms of 1D and 2D Gaussian expectations of activation functions, as is done in theorem \ref{mfntPropagation} / table \ref{tableNLCPropagation}. Focusing on activation layers allows us to enumerate the types of architecture behaviors that can be obtained for ``arbitrary nonlinear layers'', as we do e.g. in this chapter.

\section{Macro-layers}

Dependency chains of consecutive linear layers, especially fully-connected layers, are undesirable, barring specific design objectives such as dimensionality reduction, as the space of functions that can be represented by two consecutive linear layers can generally also be represented by a single linear layer. Using consecutive linear layers can complicate training with gradient methods \citep{orthogonalInitialization}. We further discuss the pathology of not separating linear layers with (sufficiently) nonlinear layers, termed `pseudo-linearity', in section \ref{pseudoLinearitySection}.

Similarly, composing an activation layer with activation function $\tau^{(1)}(s)$ with another activation layer with activation function $\tau^{(2)}(s)$ is equivalent to a single activation layer with activation function $\tau^{(2)}(\tau^{(1)}(s))$. Hence, we generally wish to alternate linear layers and activation layers, which leads to macro-layers.

\section{Plain architectures} \label{plainArchitectureSection}

In this section, we study plain architectures as defined in section \ref{actFunLengthKernelSection}. In practical simple fully-connected architectures, all macro-layers indeed tend to be identical. We are not aware of any benefits of this convention besides simplicity.

\paragraph{Gaussian stability} In plain architectures, Gaussian stability is closely related to mean field Gaussian stability as explained in section \ref{gaussianStabilityExplanationSection}. Layer quadratic means diverge from the mean field limit $\mathfrak{l}$ approximately if $\frac{\mathfrak{L}_\tau'(\mathfrak{l})\mathfrak{l}}{\mathfrak{L}_\tau(\mathfrak{l})} > 1$. For popular activation functions given in the top row of figure \ref{actFunIllu}, as well as their debiased variants, we indeed find in figures \ref{covCurveillu1} through \ref{covCurveillu4} that $\frac{\mathfrak{L}_\tau'(\lambda)\lambda}{\mathfrak{L}_\tau(\lambda)}$ never significantly exceeds 1 for any $0 < \lambda < 2$, i.e. when there is scale stability. We have not observed significant Gaussian instability in any of our experiments with these activation functions.

\paragraph{Scale stability} From table \ref{tableNLCPropagation}, we obtain $\mathfrak{l}_L = \mathfrak{L}_{\tau\sigma}^M(\mathfrak{l}_0)$, where $\tau\sigma$ is the activation function that first multiplies its input with $\sigma$ and then applies $\tau$ and $M$ is the number of macro-layers. This is precisely the scenario we studied in section \ref{actFunLengthKernelSection}. To achieve mean field scale stability, $\mathfrak{L}_{\tau\sigma}$ must stay close to 1 upon iteration. This is necessarily the case when $\mathfrak{L}_{\tau\sigma}(\mathfrak{l}_0)=\mathfrak{l}_0$ and $\mathfrak{l}_0$ is itself close to 1. When we approximate $\mathfrak{L}_\tau$ with the identity, this yields $\sigma^2=1$, i.e. the LeCun variance itself. Setting $\tau$ to ReLU, we obtain $\sigma^2 = 2$, which corresponds to the He variance. Table \ref{covCurveillu1} shows that this is also the only value of $\sigma$ that works for ReLU. SeLU was designed specifically to achieve $\mathfrak{L}_\tau(1) = 1$ so that the LeCun variance and unit scale inputs would induce mean field scale stability. For tanh, $\mathfrak{L}_\tau$ converges to zero upon iteration when $\sigma^2 = 1$, but only slowly. It converges exponentially to zero when $\sigma^2 < 1$. When $\sigma$ is somewhat larger than 1, we have stability.

Theorem \ref{covkerLregular} states that the sequence $(\mathfrak{L}_{\tau\sigma}^M(\mathfrak{l}_0))_M$ (i) is strictly increasing and diverges to infinity or (ii) converges. It should be easy to determine into which camp a given $(\tau, \sigma)$ pair falls. If convergence is achieved, modifying both $\tau$ and $\sigma$ to achieve convergence at or near 1 is also easy.

\begin{figure}
\centering
\includegraphics[width=0.98\textwidth]{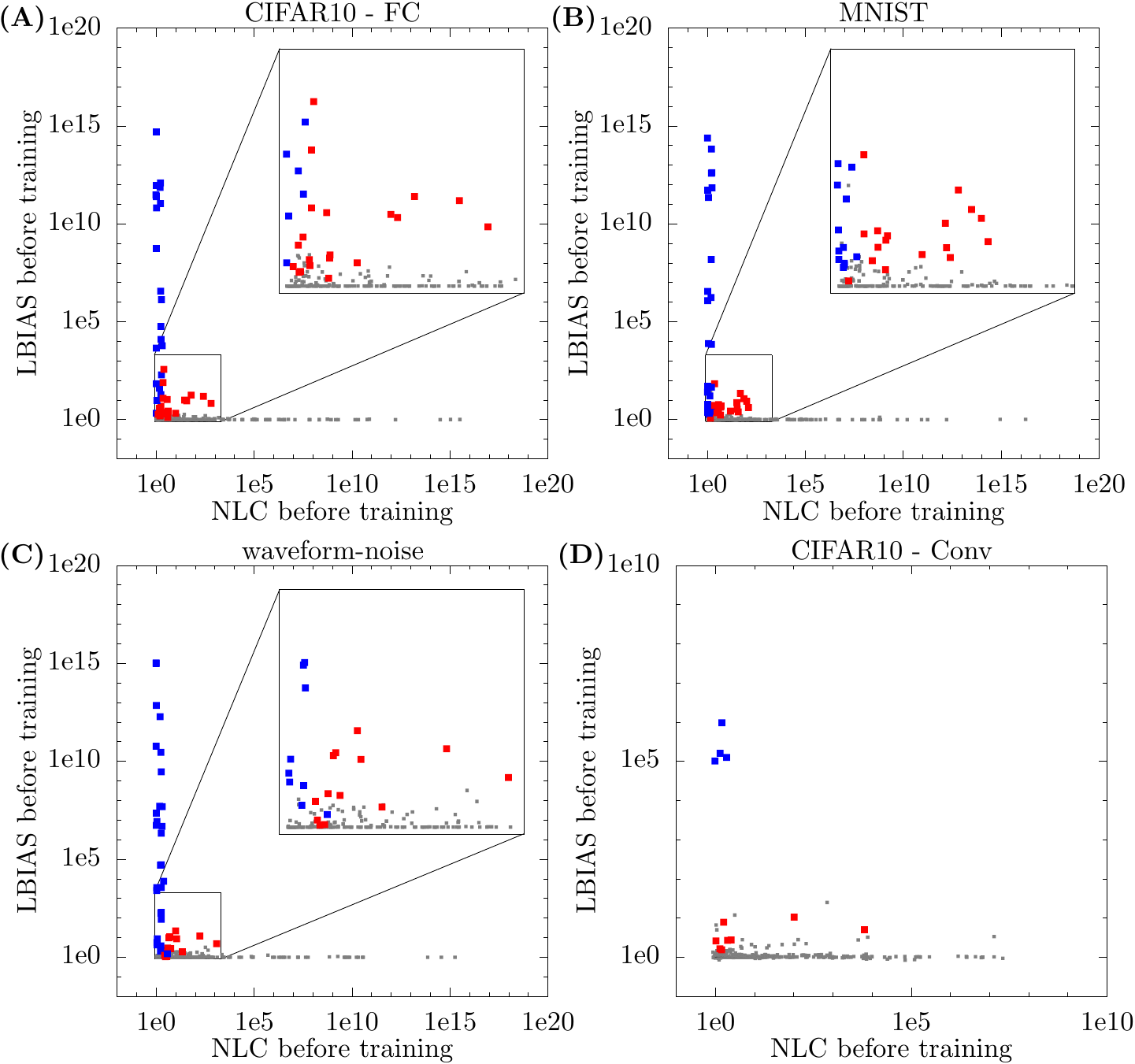}
\caption{Initial NLC vs LBIAS for study A and B architectures. Type 1 architectures are depicted in blue, type 2 architectures are depicted in red and type 3 architectures are depicted in grey. (Types are prominently defined in section \ref{plainArchitectureSection}.) Red and blue markers are displayed in the foreground relative to grey markers and thus may fully or partially occlude them. Inset graphs in the top right are magnifications of the region $0.8 < NLC,LBIAS < 2000$. {\it Conclusion:} The three different architecture types show very distinct behavior.} \label{surveyBias}
\end{figure}

\paragraph{NLC and neuron bias} Now, we assume mean field scale stability. Specifically, we consider plain stable$(q, 1)$ architectures as considered in section \ref{actFunCovKernelSection}, where we have $\sigma=\frac{1}{\mathfrak{l}_0}=\frac{1}{\sqrt{q}}$ and $\mathfrak{L}_\tau(1) = \mathfrak{l}_0 = \sqrt{q}$. As in sections \ref{actFunCovKernelSection} and \ref{actFunNLCsection}, we can easily obtain equivalent results for plain stable$(q, q')$ architectures with $q' \neq 1$.

From table \ref{tableNLCPropagation}, as in section \ref{actFunCovKernelSection}, we obtain $\mathfrak{c}_L = q\tilde{\mathfrak{C}}_{\tau}^M(\mathfrak{c}_0)$ from the above assumptions, where $\tilde{\mathfrak{C}}_{\tau} = \frac{\mathfrak{C}_{\tau}}{\mathfrak{C}_{\tau}(1)}$. Let us revisit the three convergence cases of theorem \ref{covkerCregular}. In section \ref{actFunNLCsection}, we explained that in case 3 the mean field NLC diverges exponentially with $M$. On the other hand, in case 2 we have sub-exponential divergence and in case 1 we have exponential convergence as long as $\tilde{\mathfrak{C}}_{\tau}$ is twice differentiable at 1. We can similarly apply this case breakdown to LBIAS, which has mean field limit $\sqrt{\frac{1}{1 - \mathfrak{c}_L}}$ in our scenario. In case 3, mean field LBIAS converges to $\sqrt{\frac{1}{1 - c^\text{lim}}}$. In case 2, mean field LBIAS diverges sub-exponentially. In case 1, LBIAS diverges exponentially.

This analysis reveals an intriguing ``trichotomy''. Either (1) mean field NLC converges and mean field LBIAS diverges exponentially; or (2) mean field LBIAS and mean field NLC diverge sub-exponentially; or (3) mean field LBIAS converges and mean field NLC diverges exponentially. (Going forward, we leave open the case where $\tilde{\mathfrak{C}}_{\tau}$ is not twice differentiable at 1 and $\tilde{\mathfrak{C}}_{\tau}'(1) < 1$. We also leave open the case where $\tau$ is not piecewise 5-differentiable as assumed in theorem \ref{covkerCregular}.) Notably, this trichotomy was first observed by \citet{correlationLimit} in the context of simple tanh architectures, though of course it was not phrased in terms of NLC / LBIAS. See also section \ref{orderChaosSection}.

It turns out that this trichotomy is very prominent in our empirical studies. While most of our architectures are not quite as simple as the plain architectures we consider in this section, it is nonetheless possible to associate each of them with one of the three cases of the trichotomy. Let's first consider our activation functions in tables \ref{covCurveillu1} through \ref{covCurveillu4}. For (undebiased) softplus, sigmoid and Gaussian, the stable fixed point of $\tilde{\mathfrak{C}}_{\tau}$ is at 1 and the convergence rate exponential. Hence, those activation functions suggest case 1. For (undebiased) ReLU and abs. val., the stable fixed point of $\tilde{\mathfrak{C}}_{\tau}$ is at 1 and the convergence rate is sub-exponential, suggesting case 2. For (undebiased) even tanh, the stable fixed point of $\tilde{\mathfrak{C}}_{\tau}$ is close to 1 and the convergence rate, while exponential, is relatively slow. This still suggests case 2. For all debiased activation functions as well as SELU, Swish, tanh, square and odd square, the stable fixed point of $\tilde{\mathfrak{C}}_{\tau}$ is significantly different from 1, suggesting case 3. (We do not consider sawtooth here, as it was not used in study A or B.)

Another crucial factor with respect to the trichotomy is batch normalization. We discuss this in detail in the next section. For now, we note that a BN layer always ``resets'' $\mathfrak{c}$ to zero. In a mean field sense, this is similar to activation function debiasing. Hence, the presence of BN suggests case 3. We thus divide our study A and B architectures into three types.

\begin{itemize}[leftmargin=2cm]
\item[Type 1] The architecture does not use BN, does not use activation function debiasing and is based on one of softplus, sigmoid or Gaussian.
\item[Type 2] The architecture does not use BN, does not use activation function debiasing and is based on one of ReLU, abs. val. or even tanh.
\item[Type 3] The architecture uses BN, or uses activation function debiasing, or is based on one of SELU, Swish, tanh, square or odd square.
\end{itemize}

In figure \ref{surveyBias}, we plot the initial NLC vs initial LBIAS for our study A and B architectures. Type 1 architectures are depicted in blue, type 2 architectures are depicted in red and type 3 architectures are depicted in grey. Indeed, we find a very strong pattern. \finding{Type 1 architectures always have small NLC but can have large to very large LBIAS. Type 2 architectures can simultaneously have large NLC and LBIAS, but never have very large values. Type 3 architectures always have small LBIAS, but can have large to very large NLC}. Note that small LBIAS / NLC values can be attained by shallow architectures of any type. The strength of this pattern is noteworthy given that we did not take into account architecture properties like activation function scaling / shifting / linearization, skip connections, layer normalization, convolution, pooling, etc. that arise in our architectures as described in sections \ref{studyAArchitecturesSection} and \ref{studyBArchitecturesSection}. Also see our full list of architectures with definition and metric values in the appendix in chapter \ref{fullListChapter}.

Which of the three cases is preferable? Assuming the true input-label function has no special properties that need to be imitated by the network, we are not aware of any benefits of a non-zero LBIAS in the initial state. In contrast, a certain amount of nonlinearity is desirable. Hence, we argue case 3 is best. For example, the biasedness of sigmoid explains the increase in popularity of tanh relative to sigmoid.

To build deep networks, we need relatively small $\mathfrak{n}_\tau$ values. Consider table \ref{actFunIllu}. If we rank the activation functions by their $\mathfrak{n}_\tau(1,0)$ values given in e.g. table \ref{basisillu}, the popular activation functions in the top row attain five out of the lowest six $\mathfrak{n}_\tau(1,0)$ values. Hence, a relatively small activation function NLC is a prerequisite for a popular activation function. In fact, many of the ``ReLU spin-offs'', such as SELU, ELU, Swish and softplus, have significantly lower $\mathfrak{n}_\tau(1,0)$ than ReLU. Interestingly, this fact was largely unknown to the designers of those activation functions. Their superiority over ReLU, especially in deep networks, was often rationalized in somewhat ad hoc ways. Many of these ReLU style activation functions also require an additional debiasing method, such as BN or explicit activation function debiasing, in order to not induce an excessive LBIAS.

As we argued in section \ref{outputBiasSection}, in simple architectures, LBIAS is a good proxy for neuron bias in general. Hence, our discussion of LBIAS above and in the sections below applies to neuron bias in general.

\paragraph{Training stability} As we explained in sections \ref{forwardStabilitySection}, \ref{covariateShiftSection} and at the end of section \ref{meanFieldPracticalEmpiricalSection}, an uncontrolled growth of the overall weight magnitude during training can lead to an increase of $\hat{\mathfrak{q}}_k$ / $\hat{\mathfrak{c}}_k$ at the dependencies of activation layers $f_l$. This can drastically change $\mathfrak{n}_\tau(\hat{\mathfrak{q}}_k,\hat{\mathfrak{c}}_k)$ and hence $\hat{\mathfrak{n}}$. The $\mathfrak{n}_\tau(\lambda^2,0)$ curves of tables \ref{covCurveillu1} through \ref{covCurveillu4} indicate the susceptibility of different activation functions to this effect. We specifically note that this curve is constant for ReLU (as well as abs. val.) because we have $\tau_\text{ReLU}(cs) = c\tau_\text{ReLU}(s)$ for all $c > 0$. Hence, ReLU mitigates the harm of training instability and is more forgiving to an improper initial weight variance. This was one of the major reasons for the rise of ReLU, as it addressed what was known as the ``exploding / vanishing gradient problem in sigmoid / tanh networks'' which we detail in section \ref{vanishTanhSection}.

\paragraph{SELU} \citet{selu} gave the following design criteria for SELU. (Of course, they did not phrase them in the same way.)

\begin{itemize}
\item similar to ELU
\item $\mathfrak{L}_\tau(1) = 1$
\item $\mathfrak{C}_\tau(0) = 0$
\item $\mathfrak{L}_\tau'(1) < 1$
\end{itemize}

These criteria correspond to scale stability in LeCun initialized plain architectures with unit scale inputs, activation function debiasing and mean field Gaussian stability, respectively. However, SELU's low activation function NLC, arguably its most important property, remained ``undiscovered''.

\section{Depth} \label{surveyDepthSection}

Assuming scale stability, the nonlinearity path equation implies that $\mathfrak{n}$ is bounded above approximately by $\mathfrak{n}_\tau(1,0)^M$. The largest $\mathfrak{n}_\tau(1,0)$ value attained by a popular activation function in table \ref{actFunIllu} is 1.21, and it is attained by ReLU. Hence, if we want a plain stable architecture to have a higher degree of nonlinearity than 1.21 as measured by mean field NLC using a popular activation function, we must have multiple macro-layers. We suspect that in complex tasks such as ImageNet, an NLC as low as 1.21 is not ideal. To achieve a mean field NLC of e.g. 2 with ReLU, we need at least 5 macro-layers. This is a key reason behind the success of deep networks. However, it is not clear what the practical downsides of using activation functions more nonlinear than ReLU in a shallow network are relative to using ReLU in a deeper network. This is an interesting topic for future work. We discuss the ZSAD guideline of ``using an appropriate depth'' further in section \ref{depthSection}.

\section{Normalization layers} \label{surveyBNsection}

When a macro-layer uses a normalization layer, that normalization layer is generally placed directly before the activation layer. Training stability explains this choice. As discussed above, as the weights in linear layers grow, the mean field NLC of the network can change drastically. This can be mitigated by normalization layers, which explicitly control the magnitude of neuron values. If normalization layers are placed after the activation layer, weight growth in the linear layer can still affect the magnitude of inputs to the activation layer. It is generally said that one of the key benefits of normalization layers is that they ``enable larger learning rates'', which directly corresponds to training stability.

While there are commonalities between the behavior of different normalization layers, there are also important differences. As always in this work, we compare and contrast BN and LN.

At a BN layer $f_l$, we have $\mathbb{E}_xf_l = 0$ by definition. Hence, BN removes neuron bias. In table \ref{tableNLCPropagation}, we find that at an LN layer $f_l$ we have $\frac{\mathfrak{c}_l}{\mathfrak{q}_l} = \frac{\mathfrak{c}_k}{\mathfrak{q}_k}$. Hence, LN preserves bias in a mean field sense. This is the most important reason for the greater popularity of BN relative to LN. In the previous section, all our BN architectures fell under type 3, which we argue is best, whereas the type of architectures with LN depended on the activation function. Note that when convolutional layers are used, the definition of BN changes slightly, as explained in section \ref{tensorLayerSection}. In that case, the degree to which BN removes neuron bias is less clear-cut.

When it comes to Gaussian stability, though, LN has the advantage. A core feature of Gaussian instability is the divergence of layer quadratic means, as explained in e.g. section \ref{gaussianStabilityExplanationSection}. LN sets the quadratic mean for each individual layer value to exactly 1, independently of the layer width or batch. Therefore, Gaussian instability is eliminated. In contrast, while BN sets the quadratic mean across the entire batch to 1, it fails to do so for individual layer values. Therefore, Gaussian instability is not eliminated. Throughout this work, our fully-connected GUAs used BN or no normalization. Fully-connected architectures with LN never exhibited Gaussian instability. This benefit is severely impaired if convolutional layers are used. The problem is that some spatial regions of layers can grow or shrink relative to other regions, which can still lead to low performance. Throughout the work, convolutional architectures with LN were still considered GUAs depending on activation function.

Both BN and LN ensure scale stability. Neither BN nor LN directly control the NLC, but can have drastic indirect effects by e.g. changing convergence type.

BN leads the output returned for an individual input to depend on other inputs in the batch, which are generally selected randomly. In practical situations, the batch size can be as small as 1 due to computational resource constraints. The smaller the batch size, the greater the chance of harmful noise instability (section \ref{noiseStabilitySection}). This has motivated a search for BN replacements for small-batch situations \citep{batchReNorm}.

\section{nlnorm} \label{surveynlnormSection}

Unlike the other design strategies covered in this chapter, nlnorm, which we introduced in the previous chapter, is, of course, not currently a popular design strategy. We cover it here for comparability.

nlnorm has a three-fold objective: ensure scale stability, eliminate neuron bias, control nonlinearity. To our knowledge, no normalization method has thus far granted explicit control over nonlinearity. A disadvantage of nlnorm relative to normalization layers is that nlnorm does not mitigate training instability induced by weight growth. In fact, the ``normalization by recursion'' principle, which normalizes layers closer to the output by assuming layers closer to the input are already normalized, is especially susceptible to weight growth. Despite the fact that there was not a great performance boost observed by normalization layers in figure \ref{nlnormActNorm} when nlnorm was used, normalization layers should be able to provide significant value on top of nlnorm to reduce e.g. training instability. Also, LN can eliminate Gaussian instability, whereas nlnorm generally does not. Note that nlnorm, as given in figure \ref{boxnlnorm}, applies directly only to simple architectures.

\section{Skip connections} \label{skipConnectionsSection}

\begin{figure}
\centering
\includegraphics[width=0.98\textwidth]{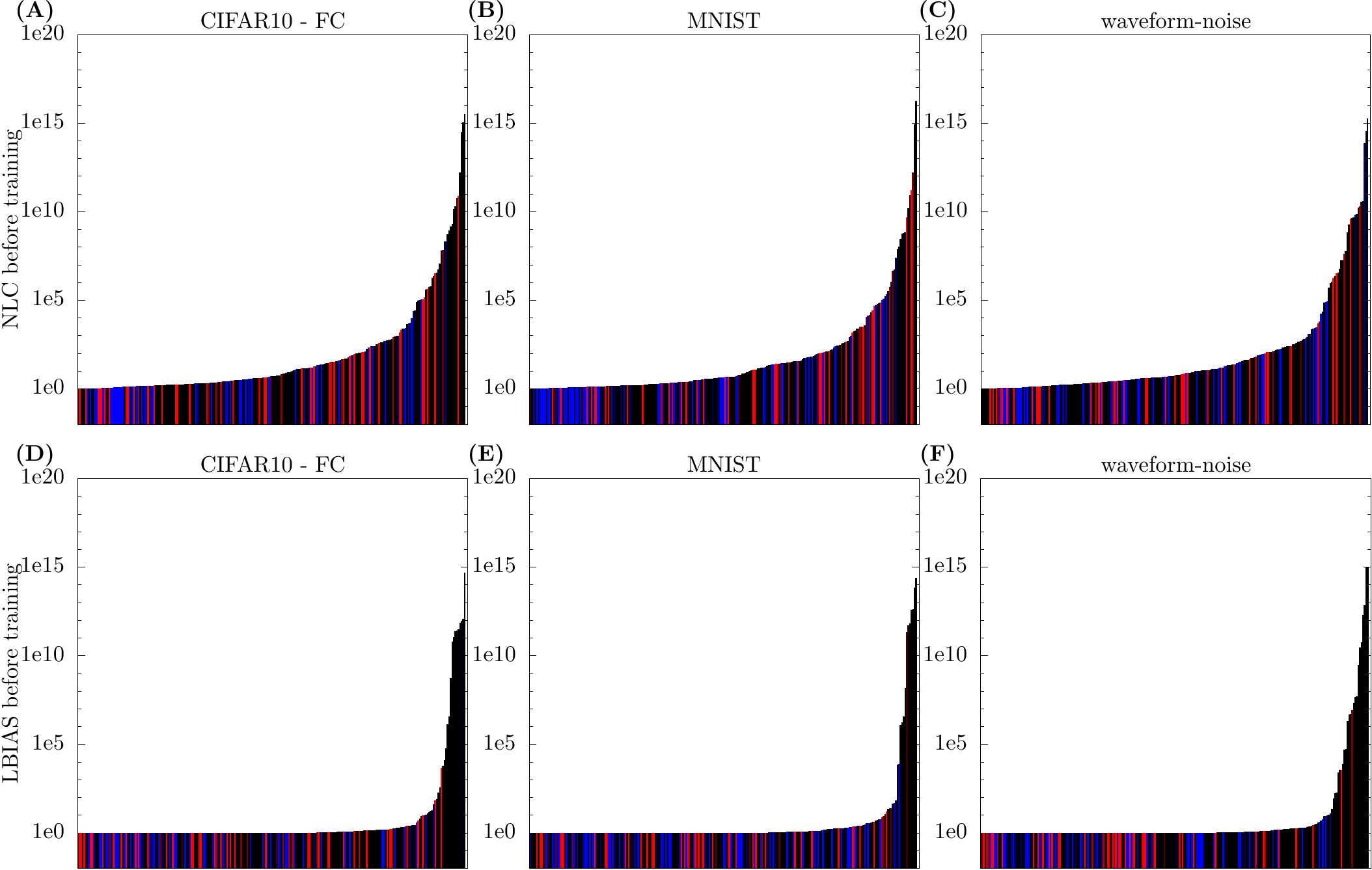}
\caption{Initial NLC and LBIAS for study A architectures. Architectures are placed on the x-axis in the order of their y-axis value. Architectures depicted in red are residual, and normalization layers are placed between residual units. Architectures depicted in blue are residual, and normalization layers are placed only within residual blocks. Architectures depicted in black are not residual. {\it Conclusion:} Skip connections significantly reduce both NLC and LBIAS. This effect is stronger when normalization layers are placed only within residual blocks.} \label{surveySkip}
\end{figure}

\paragraph{NLC and LBIAS} The primary reason why skip connections increase performance is that they reduce nonlinearity. In figure \ref{surveySkip}A-C, we plot the initial NLC of our study A architectures on the y-axis. On the x-axis, we arrange all architectures by NLC magnitude from left to right. Red and blue bars correspond to residual architectures. We find that \finding{residual architectures indeed tend to have lower NLCs}.

The nonlinearity path equation explains this. Using the notation from the end of section \ref{architectureDesignParadigmsSection} and assuming that the mean field NLC of the skip connections is 1, we obtain

\begin{eqnarray*}
&&\mathfrak{n}(f_+(f_s(f_a),f_r(f_a)))^2\\
&=& \frac{(\mathfrak{q}_s - \mathfrak{c}_s)\mathfrak{n}(f_s(f_a))^2 +(\mathfrak{q}_r - \mathfrak{c}_r)\mathfrak{n}(f_r(f_a))^2}{\mathfrak{q}_s - \mathfrak{c}_s + \mathfrak{q}_r - \mathfrak{c}_r}\\
&=& 1 + \frac{\mathfrak{q}_r - \mathfrak{c}_r}{\mathfrak{q}_s - \mathfrak{c}_s + \mathfrak{q}_r - \mathfrak{c}_r}(\mathfrak{n}(f_r(f_a))^2 - 1)
\end{eqnarray*}

The square of the mean field NLC of a residual unit is the weighted average of the square of the mean field NLC of the residual block and the square of the mean field NLC of the skip connection. Both branches are weighted by their respective ``output variance'' $\mathfrak{q}_r - \mathfrak{c}_r$ / $\mathfrak{q}_s - \mathfrak{c}_s$. Depending on the value of these variances, the mean field NLC of the unit can be far closer to 1 than the mean field NLC of the block by itself. When the mean field NLCs of residual units are multiplied in the nonlinearity path equation, the effect on the NLC of the entire network can be drastic. In one of our prior works \citep{expl}, we give a detailed, visual explanation of this effect, which we term `$k$-dilution'.

Let's look at a specific example. Consider an architecture composed of macro-layers composed of a BN layer, an activation layer with activation function $\tau$, a fully-connected layer and an addition layer. The addition layer adds the output of the fully-connected layer together with the output of the previous addition layer. The mean field NLC of each residual block is then equal to $\mathfrak{n}_\tau(1,0)$. $\mathfrak{q}_r - \mathfrak{c}_r$ is fixed across macro-layers whereas $\mathfrak{q}_s - \mathfrak{c}_s$ grows linearly as the signal accumulates. Hence, we have

$$\mathfrak{n}(f_+(f_s(f_a),f_r(f_a))) = \sqrt{1 + \frac{\mathfrak{q}_r - \mathfrak{c}_r}{\mathfrak{q}_\text{start} - \mathfrak{c}_\text{start} + m(\mathfrak{q}_r - \mathfrak{c}_r)}\mathfrak{n}_\tau(1,0)^2}$$

where $m$ is the index of the macro-layer, $\mathfrak{q}_\text{start}$ / $\mathfrak{c}_\text{start}$ are taken at the start of the first residual unit and the addition layers are assumed to use addition weights equal to 1. Hence, the mean field NLC of the residual units behaves as $1 + O(\frac{1}{m})$ as the macro-layer index increases. In plain language, the deeper the network goes, the less nonlinear the residual units become.

This simple example actually explains and is representative of the type of residual architecture that became popular starting from \citet{resNetTrueIdentity}, where skip connections bypass two macro-layers and convolutional layers are used in addition to BN. The mean field NLC of units still decreases as $1 + O(\frac{1}{m})$. This is the reason why these residual architectures work at enormous depths. However, the product of a series that behaves as $1 + O(\frac{1}{m})$ still diverges. Hence, there usually does come a depth at which one observes increased error even with this type of residual architecture.

\begin{figure}
\centering
\includegraphics[width=0.95\textwidth]{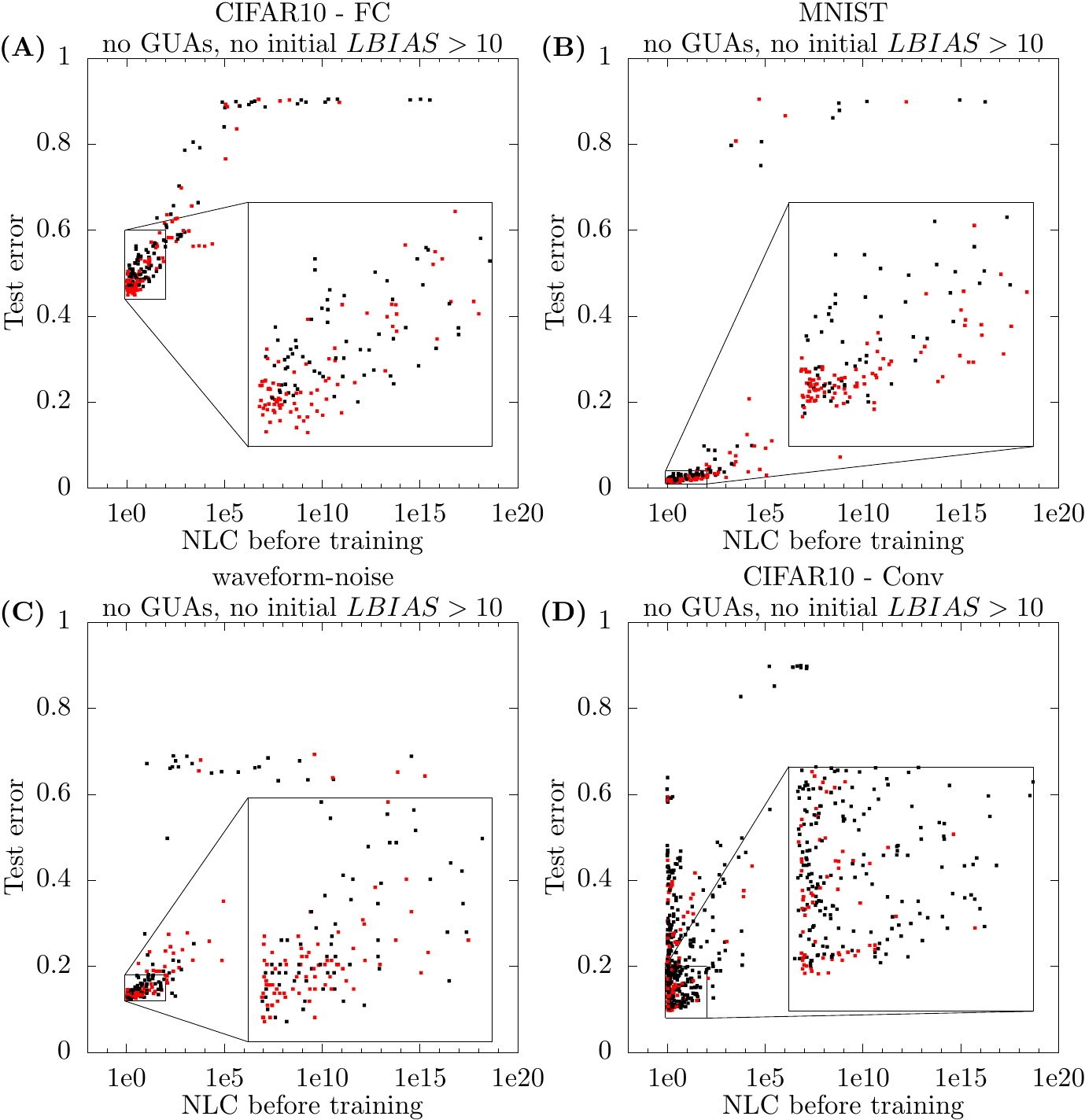}
\caption{Initial NLC vs test error for study A architectures. Graphs mirror those in figure \ref{nlcPredTestInit}. Architectures depicted in red are residual. Architectures depicted in black are non-residual. Architectures with an initial LBIAS greater than 10 and GUAs are omitted. {\it Conclusion:} Residual architectures tend to have lower NLCs than non-residual architectures and tend to outperform non-residual architectures of comparable NLC, at least when $1 \le NLC \le 100$.} \label{beyondSkip}
\end{figure}

Let's consider a slightly altered example. Instead of having the addition layer add the fully-connected layer together with the last addition layer, we have it add the fully-connected layer to the previous BN layer. This means that normalization occurs between residual units, not within residual blocks. Now we have 

$$\mathfrak{n}(f_+(f_s(f_a),f_r(f_a)) = \sqrt{1 + \frac{\mathfrak{q}_r - \mathfrak{c}_r}{1 + \mathfrak{q}_r - \mathfrak{c}_r}\mathfrak{n}_\tau(1,0)^2}$$

Hence, the mean field NLC of the residual units behaves as $O(1)$ as the macro-layer index increases. So $\mathfrak{n}$ is exponential in the number of macro-layers, which mirrors the behavior of non-residual architectures with BN. Adding this type of skip connection ``only'' reduces $\mathfrak{n}$ by a constant factor in log space. Because this between-unit normalization type is less effective at reducing the NLC, the within-block normalization type became the popular standard.

In study A, we considered both types of residual architecture, which we now compare empirically. In figure \ref{surveySkip}A-C, residual architectures that use within-block normalization are depicted in blue. Residual architectures that use between-unit normalization are depicted in red. Non-residual architectures are depicted in black. Indeed, we find that \finding{blue architectures tend to have lower NLCs than red architectures, which tend to have lower NLCs than black architectures}. Note that in 50\% of our study A residual architectures, we apply an addition weight less than 1 to the skip connection. This tempers their NLC-reducing effect. Without this, the pattern in figure \ref{surveySkip} would be more pronounced. See section \ref{studyAArchitecturesSection} for details.

Just like skip connections tend to significantly reduce NLC, they tend to significantly reduce LBIAS, as shown in figure \ref{surveySkip}D-F. Again, within-block normalization is more effective at this. Of course, the ideal outcome of eliminating neuron bias is not achieved, not even in the mean field sense.

\paragraph{Scale stability} In terms of scale stability, skip connections can actually be harmful. In the within-block normalization example above, we found that $\mathfrak{q}$ grows linearly from macro-layer to macro-layer. If the normalization layer is removed entirely, things become more problematic. As the $\mathfrak{q}$ value grows, so do the inputs to the activation function. If the activation function is e.g. tanh, then $\mathfrak{n}_\tau(\mathfrak{q}_k, \mathfrak{c}_k)$ values increase with depth, which defeats the purpose of skip connections. If the activation function is e.g. ReLU, then $\mathfrak{q}$ increases exponentially from unit to unit. This is why skip connections are always paired with normalization layers in practice. \citet{zeroInit} constructed normalization-free residual architectures. However, these architectures require a specialized and cumbersome weight initialization scheme, and may still be susceptible to training instability. \citet{zeroInit2} use trainable addition weights initialized to zero instead of normalization layers. As they point out, they need to use smaller learning rates due to training instability.

\paragraph{ResNet vs nlnorm} We introduced nlnorm as the first method that explicitly turns the NLC into a tunable hyperparameter. It would be interesting to consider to what extent the addition layers of ResNets can be used in a similar fashion. An addition layer can apply addition weights to the skip connection and residual block that are unequal to 1. We then have

$$\mathfrak{n}(f_+(f_s(f_a),f_r(f_a))^2 = 1 + \frac{w_r^2(\mathfrak{q}_r - \mathfrak{c}_r)}{w_s^2(\mathfrak{q}_s - \mathfrak{c}_s) + w_r^2(\mathfrak{q}_r - \mathfrak{c}_r)}(\mathfrak{n}(f_r(f_a))^2 - 1)$$

By controlling $w_r$ and $w_s$, we can control the mean field NLC of the residual unit and hence of the architecture itself. The similarity between this and nlnorm becomes even more clear when we realize that nlnorm using the `linear interpolation' linearization method given in section \ref{nlnormDefinitionSection} is equivalent to using identity skip connections that bypass only the activation layers. It is interesting to consider to what extent tuning the $l$ parameter of nlnorm yields similar outcomes compared to tuning the addition weights in a residual architecture where skip connections bypass many layers.

\paragraph{Related work} Our finding that popular types of ResNet exhibit ``polynomial behavior'' rather than ``exponential behavior'' with increasing depth is mirrored by e.g. \citet{resNetMeanField,reluFiniteWidthEffects,zeroInit2}. \citet{resNetRefinementOrig,resNetRefinement} showed that with depth, residual units have a diminishing impact on the network output. However, our work goes beyond this. Via e.g. the nonlinearity path equation, it is possible to place skip connections in the network and choose their strength deliberately to achieve target values for properties like the NLC, just like in nlnorm. Our analysis is also not limited to ResNet, but is fundamentally applicable to all the myriad neural architectures that have been proposed in the wake of ResNet that additively combine paths which significantly differ in their nonlinearity (e.g. FractalNet \citep{fractalNet}, DenseNet \citep{denseNet}, Residual networks of Residual networks \citep{resNetInResNet}, Inception \citep{inceptionv4}).

\paragraph{``Unexplained'' performance gains} It has been reported that residual architectures outperform non-residual architectures even when controlling for a range of factors. We find that this is true in our experiments as well. In figure \ref{beyondSkip}, we repeat the results of figure \ref{nlcPredTestInit}, but this time (i) we depict all residual architectures in red, (ii) we omit architectures with initial LBIAS greater than 10 and (iii) we omit GUAs. (In other words, we omit architectures that were plotted in red or green in figure \ref{beyondSummary} due to a known pathology.) Among architectures depicted in figure \ref{beyondSkip}, we find that \finding{residual architectures do not just tend to exhibit lower NLC values}, as also shown in figure \ref{surveySkip}, but \finding{overall attain significantly lower test error values at any given NLC level within the critical range $1 \le NLC \le 100$}.

\section{Orthogonal initialization}

Orthogonal initialization contributes to improving the overall `orthogonality' of the architecture. We do not investigate the ZSAD guideline of orthogonality in detail in this work, but define it and discuss related work in section \ref{orthogonalitySection}.

\section{Bias vector initialization} \label{surveyBiasSection}

Setting components of bias vectors that arise in bias layers to zero in the initial state prevents those bias layers from inducing neuron bias or scale instability. Of course, the picture is less clear during and after training. While it is possible for a bias layer to learn to remove neuron bias that arises in its dependency, it often ends up learning to introduce significant biases. The key is that those biases are learned explicitly and are not random or a byproduct of other learning processes. The reasons behind the potential benefits of learned biases are an interesting topic for future work.

\section{Scaling vector initialization} \label{surveyScalingSection}

Setting components of scaling vectors that arise in elementwise multiplication layers to 1 in the initial state prevents those elementwise multiplication layers from inducing scale instability. Similar to the bias vector, the scaling vector can induce significant scale changes during and after training. The benefit of learning such explicit changes is an interesting topic for future work.

\section{Summary} \label{surveySummarySection}

\begin{table}
\centering
\begin{tabular}{p{3.2cm}cccccc}
Design strategy & NLC & no neuron bias & scale st. & Gaussian st. & training st. & noise st.\\ \hline\hline
LeCun / He var.&-&-&yes&-&-\\
Activation layers&enables&-&-&-&-\\
Bias vector init&-&not harmful&not harmful&-&-\\
Scaling vector init&-&-&not harmful&-&-\\
ReLU& - & - & - & Gauss. edge &improves\\
softplus& reduces & - & - & yes & - \\
SELU&reduces&yes& - &yes& - \\
tanh& - &yes& - & yes & - \\
Depth&enables&-&-&-&-\\
BN& - & yes, esp. FC & yes & - & improves &can harm\\
LN& - & - &yes&FC only& improves\\
nlnorm&yes&yes&yes&-&-\\
Skip connections&reduces&improves&can harm& - & - \\
\\
\end{tabular}
\begin{tabular}{p{3.2cm}cccc}
Design strategy & Gaussian stability & parameter dim.& orthogonality & no pseudo-linearity\\ \hline\hline
Linear layers / random init&prerequisite&contributes&-&-\\
Macro-layers&-&-&-&enables\\
Orthogonal init&-&-&contributes&-
\end{tabular}
\caption{Very high-level summary of how building blocks and design strategies relate to ZSAD guidelines in the context of the simple architecture designs discussed in this chapter. ``yes'' means that the design strategy is consistently effective at causing the architecture to follow the guideline. See earlier sections of this chapter for details.} \label{surveySummary}
\end{table}

We summarize the insights of this chapter in table \ref{surveySummary}. Of course, this table is very high-level and relies on prior sections for context. See chapter \ref{fullListChapter} in the appendix for detail on how building blocks influenced properties and metric values of each of our architectures.

We do not claim that we have given a complete and exhaustive explanation of the benefits of the discussed strategies in this chapter, if such a thing is even possible. For example, we showed that skip connections confer performance benefits that are not explained in this work and, to our knowledge, not adequately explained in literature in general. Batch normalization has recently been linked to orthogonality \citep{bnOrthogonality}.

\chapter{The NLC versus related work} \label{relatedWorkChapter}

Determining what is ``related work'' in the context of this work is challenging because of the generality and breadth of the topics contained herein. The idea of architecture design can be understood broadly to include anything from neural architecture search to the theory of depth-2 ReLU networks; anything from an empirical survey of architecture types to a proposed new layer operation. In section \ref{zeroShotDesignSection}, we defined the term `zero-shot architecture design' to describe and distinguish our approach to architecture design and to place our work within the spectrum of deep learning research. Hence, we will focus this chapter on ZSAD guidelines that have been proposed previously. Of course, since we introduce the concept of ZSAD in this work, no prior work has explicitly identified a strategy as being a ZSAD guideline. Hence, we must make the determination of ``what counts'', which is, unfortunately, inevitably, a subjective process. Non-ZSAD related work is discussed e.g. in sections \ref{designOverviewSection}, \ref{generalizationMeasuresSection}, \ref{meanFieldBackgroundSection}, \ref{nlnormRelatedWorkSection} and \ref{skipConnectionsSection}.

Even within the ZSAD sub-field, like in any other sub-field of deep learning research in the year 2020, there is a very large number of ideas that have been worked on in a small number of studies. Some relatively widely known examples of such ideas are ``avoid dying ReLUs'' \citep{dyingRelu}, ``maximize trajectory transitions'' \citep{trajectoryTransitions} and ``avoid covariate shift'', as discussed in section \ref{covariateShiftSection}. We made the decision to focus this chapter on examining the most prominent and well-developed ZSAD guidelines in detail. Many of the critiques we have for those guidelines can be applied directly to a large fraction of the ZSAD landscape. If we can show that our work in many ways improves upon those guidelines which have been fleshed out most, then we believe there is a good chance this work advances the state-of-the-art as a whole. We hope this chapter can serve as a blueprint for the reader for analyzing guidelines they are interested in, and for comparing them to the NLC.

This chapter explains the genesis of the NLC. It was developed to build upon valuable insights gleaned from ZSAD guidelines presented in this chapter while addressing their shortcomings. Our analysis of the NLC is based on the 10 utility criteria given in figure \ref{boxNPM}. In the introduction in section \ref{darkArtSection}, we stated that existing guidelines ``remain vague and circumstantial because (i) there are no well-defined, agreed-upon metrics that measure concepts like exploding gradients and overall width in practical situations; and (ii) while prior research presented evidence that these guidelines lead to success in certain situations, their generality is unclear''. Hence, while we show that the NLC is strong along all 10 criteria, we saw the need to develop it primarily because of criteria \ref{criterionWellDefined}, \ref{criterionMeaningful} and \ref{criterionGeneral}. While neural architecture design had long been viewed as a ``dark art'', we think this work can be a step in the direction of a more scientific approach to architecture design research based on well-defined metrics and general principles. We argue that such an approach could greatly accelerate progress in architecture design research.

Like in chapter \ref{beyondNlcChapter}, each section of this chapter is dedicated to one ZSAD guideline. Whereas chapter \ref{beyondNlcChapter} introduced novel guidelines and fleshed out existing ones via novel analysis, this chapter focuses on existing guidelines through the lens of prior work, and through the lens of the deep learning community as a whole, although some novel analysis is presented as well. Our overall argument is that the guidelines we discuss here are either (i) improved upon / (partially) superseded by our guidelines or (ii) naturally complementary to them. Guidelines that fall more into category (i) are presented with a ``vs'' in the section title. Guidelines that fall more into category (ii) are presented with an ``and'' in the section title, though there is no hard distinction.

\paragraph{Background from prior chapters} Throughout this chapter, we use the terminology, notation and conventions of section \ref{notationSummarySection}. In places, we use the empirical studies laid out in chapter \ref{empiricalStudiesChapter} for validation. We recommend reading at least summary section \ref{metricsSummarySection}.

We compare and contrast our guidelines in this chapter primarily against the NLC, scale stability and neuron bias. Key sections for those guidelines include \ref{nlcDefinitionSection}, \ref{nlcPredictiveSection}, \ref{forwardStabilitySection} and \ref{outputBiasSection}, though we argue the advantageousness of the NLC based on all results from prior chapters. We also repeatedly reference the convergence behavior of the mean field limit quantities with increasing depth as discussed in sections \ref{actFunLengthKernelSection}, \ref{actFunCovKernelSection}, \ref{outputBiasSection} and \ref{plainArchitectureSection}. We reference a range of other sections when relevant for discussion.

\paragraph{Technical considerations} As discussed in section \ref{nlcDefinitionSection}, we implicitly assume things like integrability (section \ref{integrabilitySection}), differentiability (section \ref{nonDifferentiableSection}) and non-zero denominators when necessary.

\section{NLC vs exploding / vanishing gradients} \label{vanishingGradientSection}

One of the most well-known ZSAD guidelines, which dates back many decades, is the avoidance of exploding and vanishing gradients, which we shorten to EVG. Many of the foundational papers of modern architecture design \citep{normalizedInitialization,heInit,batchNormalization,resNet} as well as mean field theory \citep{correlationLimit,meanFieldCNN} reference the concept, though as we will show, not always without issues.

There are many quantitative and conceptual similarities between EVG and NLC, as well as many crucial differences. If we were to make a single argument for the importance of the NLC based on its relationship to prior work, it would be the degree to which it improves upon the EVG concept. We argue that the NLC, together with LBIAS, at least in the context of feedforward networks, is the natural evolution of the EVG concept. While the NLC can be viewed as merely a measure of EVG, we argue that the NLC stands on its own and does not need to be viewed through this lens.

In each of the following subsections, we present one criticism of the EVG concept.

In one of our prior works \citep{expl}, we introduced the `gradient scale coefficient' (GSC) as a measure of EVG. The NLC is a further evolution of GSC.

\subsection{Exploding / vanishing gradients are not well-defined} \label{vanishWellDefinedSection}

When trying to avoid EVG, the question ``When does a network have EVG?'' immediately arises. There is no consensus answer to this question. Many studies concern themselves with stabilizing gradients in some form, and not all of them use the terms ``exploding gradients'' and / or ``vanishing gradients''. Oftentimes, a measurement for EVG is chosen ad hoc and without justification.

Which metric should be used for determining the presence of EVG? Should we care about the quadratic mean of the gradient vector \citep{depthScalesMeanField,resNetMeanField}, or about the variance of gradient vector components as induced by the parameter initialization scheme \citep{heInit,normalizedInitialization}, or about the singular values of the Jacobian \citep{orthogonalInitialization,RecurrentNetsPascanu,eigenspectrum}? Depending on the metric used, different strategies arise for combating EVG, as we will show in section \ref{vanishWidthSection}.

Even if we decide on a metric which can be applied to a specific object like a gradient vector or Jacobian, it is still unclear which object to apply it to. Should we apply it to the gradient with respect to the input? Should we apply it to the gradient with respect to intermediate layers, or the gradient with respect to the parameter? The terminology of EVG suggests that the gradient ``is first of moderate size, but then becomes enormous / tiny''. It is derived from the backpropagation algorithm and the chain rule. The gradient with respect to layers further away from the output is the (sum-)product of many layerwise gradients. If the factors in that product are consistently larger / smaller than 1 according to some norm, then the product ``explodes'' / ``vanishes'' as additional factors are added as we consider layers further and further away from the output. Detecting this specific behavior is even harder than merely measuring gradients. What happens if gradient magnitude does not change monotonically, or in varying increments, or converges rather than diverges? What if the network is not a single dependency chain and there is no clear notion of ``closer to'' or ``further from'' the network output? 

The NLC is well-defined in terms of the network Jacobian, input covariance and output covariance. Hence, in contrast to EVG, it explicitly does not depend on the loss function or intermediate network layers. Because the effectiveness of the NLC is based on matching the network function to the true input-label function, it does not need to concern itself explicitly with whether there is ``explosion'' or ``vanishing'' from layer to layer. The definition of the NLC, as opposed to other popular measures of EVG, is theoretically justified.

\subsection{Exploding / vanishing gradients are confounded by re-scaling} \label{linearEquivarianceSection}

\begin{figure}
\centering
\includegraphics[width=0.98\textwidth]{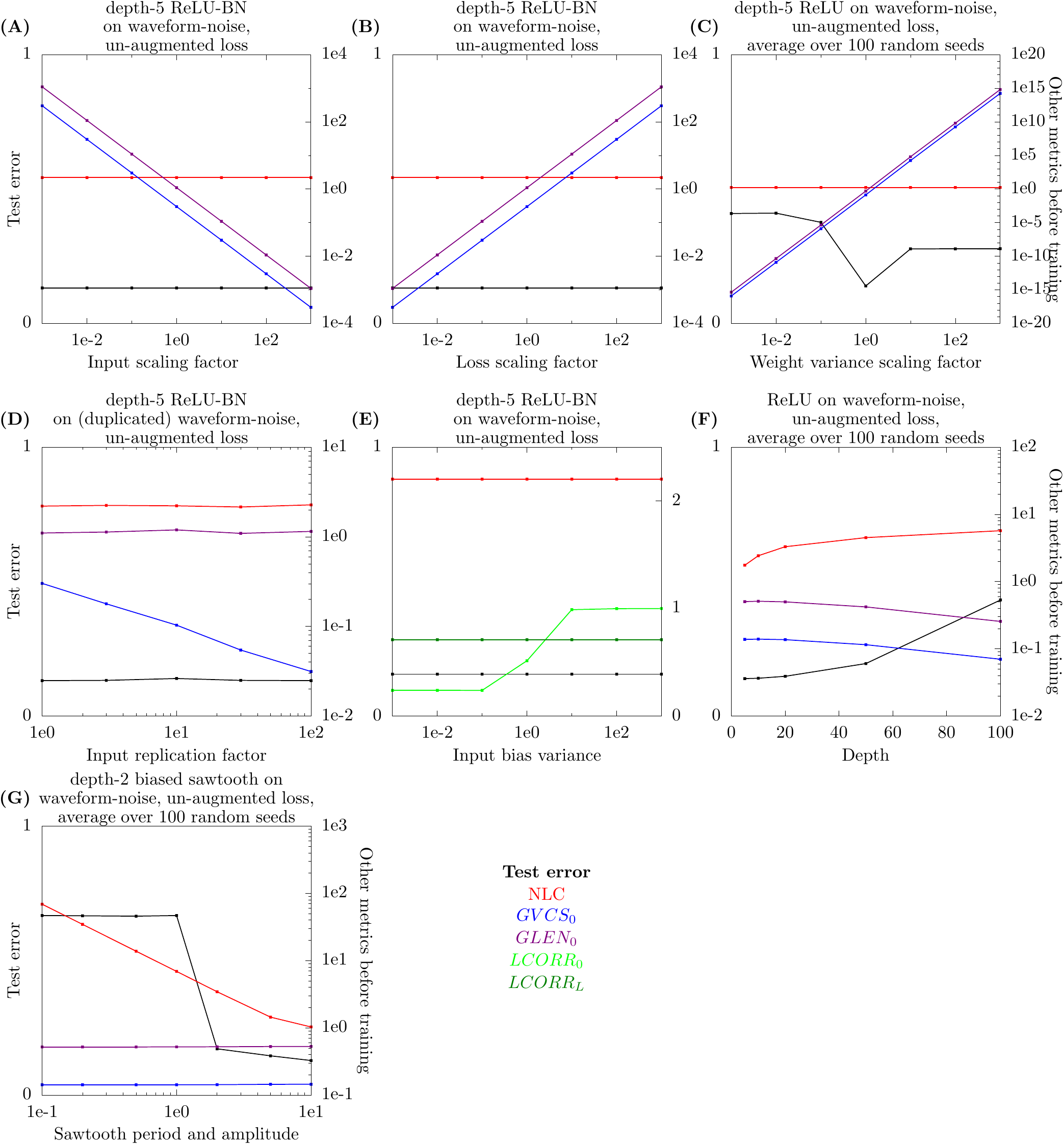}
\caption{Metric values for various simple fully-connected architectures on waveform-noise, in the final state (test error) or initial state (other metrics). We use regular, un-augmented softmax+cross-entropy as the loss function. See the top of each graph and section \ref{additionalExperimentsSection} for details. Test error is plotted on the left y-axis and all other metrics are plotted on the right y-axis. In each graph, we vary a single hyperparameter and depict how metric values vary. Averages across random seeds are taken in log space for metrics depicted in log scale (NLC, GVCS and GLEN). {\it Conclusion:} NLC correctly predicts performance in the presence of confounding hyperparameter changes, whereas other metrics do not.} \label{relatedConf}
\end{figure}

When evaluated before training, the NLC is predictive of performance after training. In certain contexts, the same is true for EVG as we show in section \ref{vanishMultipleSection}. Of course, there likely would not be a large amount of momentum behind EVG if this was not the case. However, EVG lacks general predictiveness, because there exist network transformations based on re-scaling that do not alter architecture performance, but are capable of arbitrarily changing the presence of EVG. To make this concrete, we require a metric to measure EVG for the purpose of our analysis. In fact, we will consider two metrics as defined below.

\begin{metricDefinition}
The `gradient vector component size (GVCS)' is

$$GVCS_l(f,\ell,\mathcal{D}) = \sqrt{\mathbb{E}_{(x,y)}\mathbb{E}_ig_l(x,y)[i]^2}$$
\end{metricDefinition}

\begin{metricDefinition}
The `gradient vector length (GLEN)' is

$$GLEN_l(f,\ell,\mathcal{D}) = \sqrt{\frac{1}{d_L}\mathbb{E}_{(x,y)}||g_l(x,y)||_2^2}$$
\end{metricDefinition}

GVCS captures the approach of measuring EVG from \citet{depthScalesMeanField,resNetMeanField}. While GLEN and GVCS are trivially related via $\sqrt{\frac{d_L}{d_l}}GLEN_l = GVCS_l$, important differences arise when measuring EVG via one or the other, as we show below. While we will frame a lot of our analysis of EVG in terms of those two metrics, our analysis is not specific to those metrics and would largely apply to any other metric that is tied strictly to layer gradient magnitude, as we also mention in section \ref{vanishLipschitzSection}.

We begin our analysis in this subsection by discussing three concrete examples of re-scaling transformations that confound EVG and then explain the general principle behind them. Consider an architecture that starts with a linear layer and then a BN or LN layer. We transform the dataset by multiplying all inputs with some constant $c > 0$. Clearly, this does not alter the output returned by the network for those inputs, nor does it change the parameter gradient. Therefore, the final parameter and error values are also invariant when gradient-based training is used. However, multiplying the inputs by $c$  does multiply the values of $GLEN_0$ and $GVCS_0$ by $\frac{1}{c}$. In figure \ref{relatedConf}A, we give results obtained from training a depth-5 ReLU-BN architecture where the input was scaled with different values of $c$. We find that \finding{test error is unaffected, but $GLEN_0$ and $GVCS_0$ vary}.

We obtain equivalent results when scaling the loss function with a constant $c$. This leads $GLEN_0$ and $GVCS_0$ to be multiplied by $c$. As long as the learning rate is multiplied by $\frac{1}{c}$, the final parameter and error values are unchanged for e.g. SGD or momentum. In figure \ref{relatedConf}B, we give results obtained from training a depth-5 ReLU-BN architecture when scaling the loss function with different values of $c$. We find that \finding{test error is unaffected, but $GLEN_0$ and $GVCS_0$ vary.} Note that the required learning rate adjustment happens automatically in our training protocol, which independently considers 40 different starting learning rates which are tied to parameter gradient magnitude (section \ref{additionalExperimentsSection} / \ref{metricsSummarySection}). This is yet another example where exhaustive and consistent learning rate tuning is crucial for a fair and insightful comparison (section \ref{learningRateSection}).

Yet another transformation we can consider is the scaling of the initial weight variance. Consider a He-initialized ReLU architecture. In section \ref{forwardStabilitySection} and figure \ref{beyondForward}A, we gave results from training such an architecture as well as architectures where the initial weight variance was scaled by some constant $c$. Because we have $\tau_\text{ReLU}(cs) = c\tau_{ReLU}(s)$ for all $c > 0$, error is only affected by $c$ because the behavior of the softmax+cross-entropy loss function changes with output magnitude. We repeat these results in figure \ref{relatedConf}C, and also depict how $GLEN_0$ and $GVCS_0$ vary with $c$. We find that \finding{both change by a factor of $c^5$}! The value 5 stems from the depth of the architecture.

When changing the input or loss scale, we are not changing the architecture at all. When changing the weight variance, we are changing the architecture only superficially in the case of ReLU. (Technically, it is debatable whether the loss function is part of the architecture.) Loss scaling does not affect test error at all. Input scaling does not affect test error for certain architectures. Initial weight variance scaling only affects test error in ReLU architectures based on the loss function. Hence, a measure for a ZSAD guideline should ideally be invariant to these transformations for the respective architectures. Indeed, we find in figure \ref{relatedConf}A-C that \finding{the value of the NLC does not change}. Hence, we can specify an ideal range for the NLC in $[1, 5]$ (sections \ref{nlcPredictiveSection},\ref{bestNlcSection}), which would be impossible for GLEN or GVCS without additional assumptions.

All three transformations correspond to variations that occur among practical deep learning situations. Practitioners may choose to process and normalize their data in different ways, which may correspond to a change in input scale. Not all loss functions have an agreed-upon scale. For example, the $L2$ loss is sometimes defined to be $\frac{1}{2}||f - y||_2^2$ and sometimes defined to be $||f - y||_2^2$. When a custom loss function is created, it does not even have a ``correct scale''. Finally, the initial weight variance is so non-standardized that neither TensorFlow nor PyTorch even has LeCun initialization as the default setting! Hence, all three confounders we describe are of significant practical relevance.

All three transformations are linear. In propositions \ref{finiteNetNlcOrthoMatrix} and \ref{finiteNetNlcAnyMatrix}, we already gave two examples of how the NLC is not affected by linear transformations. It turns out that there is a general class of linear transformations which can manipulate GLEN / GVCS but not NLC or error. Let $f_l(\theta_l,f_{k_l[1]}, .., f_{k_l[K_l]})$ be the $l$'th non-input layer as a function of its parameter sub-vector and its dependencies using the notation of section \ref{neuralNetworkNotationSection}. Then for any $c \neq 0$, if we simultaneously replace all non-input layers in the network with $\frac{1}{c}f_l(\theta_l,cf_{k_l[1]}, .., cf_{k_l[K_l]})$ as well as replace $\ell(f, y)$ with $\ell(cf, y)$ and multiply the inputs with $\frac{1}{c}$, the following happens. (i) The network function is unaffected. (ii) The final parameter value and error is unaffected. (iii) $NLC_{l,m}$ is unaffected for all $0 \le m \le l \le L$ with $f_m$ a bottleneck for $f_l$. (iv) $GLEN_l$ and $GVCS_l$ are multiplied by $c$ for all $0 \le l \le L$. Hence, gradient magnitude is susceptible to superficial linear re-scaling at all layers. The scaling of e.g. linear layers and activation functions is common in deep learning and throughout this work, so this susceptibility is practically harmful.

We can further extend the scope of this transformation class. If we replace $f_l(\theta_l,f_{k_l[1]}, .., f_{k_l[K_l]})$ with $\frac{1}{c}f_l(c\theta_l,cf_{k_l[1]}, .., cf_{k_l[K_l]})$ and multiply the parameter by $\frac{1}{c}$ by e.g. changing the variance of its initial distribution, then parameter gradients can also be scaled arbitrarily. Again, changes in initial parameter magnitude are common in deep learning. This transformation requires the adjustment of the learning rate for SGD by a factor of $\frac{1}{c^2}$ to preserve error. This adjustment touches upon the deeper question of ``What is architecture performance?''. As we argued in section \ref{architecturePerformanceSection}, and as has become clear throughout this work, in order to discuss architecture performance in a vacuum, we have to equate it to the best attainable performance as training algorithm and training hyperparameters such as learning rate vary. (Caveat: sharp valley problem. See section \ref{sharpValleySection}.) Therefore, if the scaling of layers and parameter initialization scheme requires a change in learning rate to compensate, then we still consider the scaled architecture to have exactly the same performance as the un-scaled architecture. Since we do not consider there to be a ``special learning rate'', we can neither prefer the original nor the scaled architecture on the basis that it ``works with that special learning rate''.

Finally, we can generalize our transformation class further by allowing different constants $c$ at each layer, which would allow us to simulate gradual gradient explosion or vanishing when choosing increasingly large / small constants respectively in the direction away from the output layer. If we scale the initial parameter sub-vectors by different constants, we would have to allow SGD to use layer-wise learning rates to compensate. We would argue that the scaled architecture still has the same performance as the un-scaled architecture and that layerwise learning rate adjustments are ``allowed'' to preserve performance. Finding and using parameter- or layer-specific learning rates is a common feature of deep learning, and part of both popular training algorithms like Adam and less popular advanced training algorithms like vSGD \citep{vSGD}, pathSGD \citep{pathSGD} and Fromage \citep{layerwiseLearningRate}.

In summary, it is possible to induce or eliminate gradient explosion or vanishing to arbitrary degrees without affecting performance or NLC, using scalings common to deep learning pipelines. Hence, gradient magnitude by itself is not robustly predictive of architecture performance.

The transformations discussed here also have important implications for the `ensure scale stability' guideline from section \ref{forwardStabilitySection}. It is clear that they are not only able to manipulate gradient magnitude, but also neuron value magnitude. Hence, they are able to induce or eliminate scale instability to arbitrary degrees without affecting performance. Hence, the scale stability guideline is conditional on the use of e.g. un-scaled popular activation and loss functions, as we stressed in section \ref{forwardStabilitySection}. Once we understand the impact of neuron value magnitude on all the building blocks in a given deep learning pipeline, and we understand how to choose learning rates, scale stability becomes irrelevant. The `avoid neuron bias' guideline may behave similarly, as indicated by the success of debiased gradient descent (section \ref{outputBiasSection}).

We note that the susceptibility of not just GLEN and GVCS, but many other metrics to scaling has recently received more attention in the community. For example, \citet{marginPrediction,lipschitz2} recognized the importance of scale-invariance when relating metrics to performance and designed metrics accordingly.

\subsection{Exploding / vanishing gradients are often entangled with layer width} \label{vanishWidthSection}

Let's look at $GLEN_l$ and $GVCS_l$ from a mean field perspective. We have $GLEN_l = \sqrt{\frac{1}{d_L}\mathbb{E}_{(x,y)}||g_L\mathcal{J}_{L,l}||_2^2}$. Assume $f_l$ is a bottleneck. If we model $g_L$ as unit Gaussian noise, we have $GLEN_l \approx \sqrt{\frac{1}{d_L}} ||\mathcal{J}_{L,l}||_F$. As we showed in section \ref{meanFieldPracticalSection} for our study A architectures, we have $\sqrt{\frac{1}{d_L}} ||\mathcal{J}_{L,l}||_F \approx \sqrt{\frac{\mathfrak{g}_L}{\mathfrak{g}_l}}$. Modeling the gradient of the loss function as Gaussian noise, or at least as independent of the network Jacobian, is a common feature of mean field theory which has been shown to lead to accurate estimates (e.g. \citet{meanFieldNetsorNTK}). So we obtain $GLEN_l \approx \sqrt{\frac{\mathfrak{g}_L}{\mathfrak{g}_l}}$ and $GVCS_l \approx \sqrt{\frac{d_L\mathfrak{g}_L}{d_l\mathfrak{g}_l}}$. While we omit the results, we verified that these approximations indeed hold empirically for our study A architectures in the manner of figure \ref{mfPredjac}. (Note that if the scale of the loss function changes, we would also have to adjust the scale of the Gaussian noise and hence the mean field estimate.)

So, mean field theory reveals that GLEN is the more meaningful metric for capturing EVG as GVCS contains a ``nuisance factor'' of $\sqrt{\frac{d_L}{d_l}}$. Whatever EVG is trying to capture, its mean field estimate should not explicitly depend on layer width.

We can demonstrate the advantageousness of GLEN by transforming an architecture by changing its width. In the previous subsection, we re-scaled elements of the learning pipeline and found that both GVCS and GLEN were confounded. Now we show that width manipulation confounds GVCS, but not GLEN. We train depth-5 ReLU-BN architectures where the input dimensionality varies. We achieve this by taking the individual features of the waveform-noise dataset and replicating them $c$ times for some positive integer $c$. Of course, since we use He initialization, the initial weight variance in the first fully-connected layer varies accordingly. We also multiply the learning rate applied to the first fully-connected (FC) layer by $\frac{1}{c}$ in order to approximately preserve the length of the SGD update of the first FC layer, thereby approximately preserving OUTSHIFT as defined in section \ref{learningRateSection}. In figure \ref{relatedConf}D, we find that \finding{this joint transformation does not significantly affect test error, NLC or $GLEN_0$, but $GVCS_0$ is multiplied by $\frac{1}{\sqrt{c}}$}.

Changes in input dimensionality are common in deep learning. Many practical datasets have an enormous number of features, from which only a small number may be sub-selected for training. Changing the number of selected features changes input dimensionality. Consider image data as an example. Neighboring pixels tend to be highly correlated. Hence, we may choose to reduce the resolution prior to training. A measure for a ZSAD guideline would ideally be invariant to this, especially when convolutional layers are not used. (If convolutional layers are used, we might have to adjust e.g. the weight tensor size to compensate \citep{meanFieldCNN}.)

It is even easier to manipulate $GVCS_l$ for $l \neq 0$ via the width of intermediate layers, as we do not have to change properties of the data. Again, GLEN is largely not susceptible to this.

The superiority of GLEN exposes an interesting and important difference between layer gradients and layer values. Throughout this work, we have built concepts like length kernel (section \ref{actFunLengthKernelSection}), covariance kernel (section \ref{networkCovKerSection}), scale stability (section \ref{forwardStabilitySection}) and Gaussian stability (section \ref{gaussianStabilityExplanationSection}) on layer quadratic means, NOT layer lengths. In the forward pass, the quadratic mean is the characteristic magnitude measure, but in the backward pass, the length is.

Despite this, GVCS is more representative of how EVG is popularly measured. Hence, many studies carry the $\sqrt{\frac{d_L}{d_l}}$ nuisance factor, which often complicates or confounds their analysis. This occurs even in some of the most prominent architecture design papers. In section \ref{architectureDesignParadigmsSection}, we showed how the LeCun initialization preserves overall neuron value magnitude / layer quadratic mean. When a gradient vector is backpropagated through a not-unreasonably-narrow LeCun-initialized FC layer $f_l$, GLEN is preserved, but GVCS is multiplied by $\sqrt{\frac{d_l}{d_k}}$. Thus, when considering GVCS, it appears as if LeCun-initialization fails. \citet{normalizedInitialization} addressed this point by advocating for what they term `normalized initialization'. Specifically, they suggest an initial weight variance of $\frac{2}{d_k+d_l}$ for FC layers. Their justification is as follows. LeCun initialization causes a ``magnitude shift'' of 1 in the forward pass, but $\sqrt{\frac{d_l}{d_k}}$ in the backward pass. Normalized initialization causes a shift of $\sqrt{\frac{2d_k}{d_k+d_l}}$ in the forward pass and $\sqrt{\frac{2d_l}{d_k+d_l}}$ in the backward pass. Since $\sqrt{\frac{2d_l}{d_k+d_l}}$ and $\sqrt{\frac{2d_k}{d_k+d_l}}$ are both closer to 1 in log space than $\sqrt{\frac{d_l}{d_k}}$ when $d_l \neq d_k$, normalized initialization reduces the ``severity of the more severe of the two shifts''. Of course, the supposed advantage of normalized initialization is not an advantage at all, because there is no reason to preserve GVCS during backpropagation. However, the disadvantage can be severe, especially in the context of tanh networks, which \citet{normalizedInitialization} used to justify their initialization scheme. As depicted in the bottom right of table \ref{covCurveillu1} and discussed in e.g. sections \ref{actFunNLCsection}, \ref{forwardStabilitySection},  \ref{covariateShiftSection} and \ref{plainArchitectureSection}, the mean field NLC of tanh layers $\hat{\mathfrak{n}}_{l,k}=\mathfrak{n}_{\tau_\text{tanh}}(\hat{\mathfrak{q}}_k,\hat{\mathfrak{c}}_k)$ is highly susceptible to $\hat{\mathfrak{q}}_k$. Multiplying $\hat{\mathfrak{q}}_k$ by $\sqrt{\frac{2d_k}{d_k+d_l}}$ can lead the mean field NLC to be arbitrarily close to 1 when the width is expanding and arbitrarily large when the width is contracting, both of which are undesirable.

Bizarrely, the LeCun initialization, which was proposed by LeCun at least as early as 1989, has since 2010 become known as the `Xavier initialization', after Xavier Glorot of \citet{normalizedInitialization}, {\it even though Glorot argued against the LeCun initialization} and advocated an arguably inferior alternative based on a meaningless utility criterion. We are not aware of another case in the deep learning field where a strategy's primary name is the first name of an author, much less of an author who did not propose the strategy.

Several years later, \citet{heInit} applied Glorot's analysis to ReLU architectures, and concluded that the initial weight variance would have to be multiplied by 2 relative to an architecture without ReLU layers. They left open the question of whether $\frac{2}{d_k}$, $\frac{4}{d_k+d_l}$ or $\frac{2}{d_l}$ should be chosen, but also pointed out that this choice is not critical because of the $\tau_\text{ReLU}(cs) = c\tau_\text{ReLU}(s)$ property which has shown up many times throughout this work. Like Glorot, Kaiming He ended up having his name associated with the correct choice of $\frac{2}{d_k}$, which became the `He initialization' or, less frequently, the `Kaiming initialization' or `MSRA initialization'.

Another example of the conflict between GVCS and GLEN is \citep{resNetMeanFieldHacking}. There, the authors identified an expression that corresponds to $\sqrt{\frac{d_L\mathfrak{g}_L}{d_l\mathfrak{g}_l}}$ as a mean field estimate of GVCS. Instead of concluding, however, that GVCS is not a meaningful metric, they conducted experiments to demonstrate that if the $\mathfrak{g}_l$ values are undesirable, this can be counteracted by manipulating layer width. No evidence was given that this would then improve performance.

\subsection{Exploding / vanishing gradients lack meaning} \label{vanishWhyBadSection}

In the last two subsections as well as subsection \ref{vanishDichotomySection}, we outline situations where EVG fails to predict test error. However, we do not consider this the most important argument against EVG and in favor of the NLC. Undoubtedly, a scenario that confounds the NLC can be constructed. For example, consider an architecture $f$ that behaves like a practical, low-gradient architecture almost everywhere except that it has an enormous Jacobian on some inconsequentially tiny set of inputs. Such an architecture could perform well yet have an enormous initial and / or final NLC. At that point, the question would arise whether this confounder is more practically relevant than the confounders discussed for EVG above. While we would argue that it is much easier to confound EVG in practice, this would ultimately be a somewhat subjective judgment.

It is more important to consider that the NLC is {\it meaningful}, i.e. it represents a fundamental and deep property of a neural network, as demonstrated in chapter \ref{nlcChapter} and throughout this work. While the explanation provided by the NLC for architecture performance can fail, for EVG there is no explanation to begin with. In the context of feedforward networks, large or small gradients are not conceptually tied to performance in general. In certain specific contexts, gradient magnitude can be related to an actual pathology, as we detail in the next subsection.

\begin{wrapfigure}[26]{r}{7cm}
\includegraphics[width=7cm]{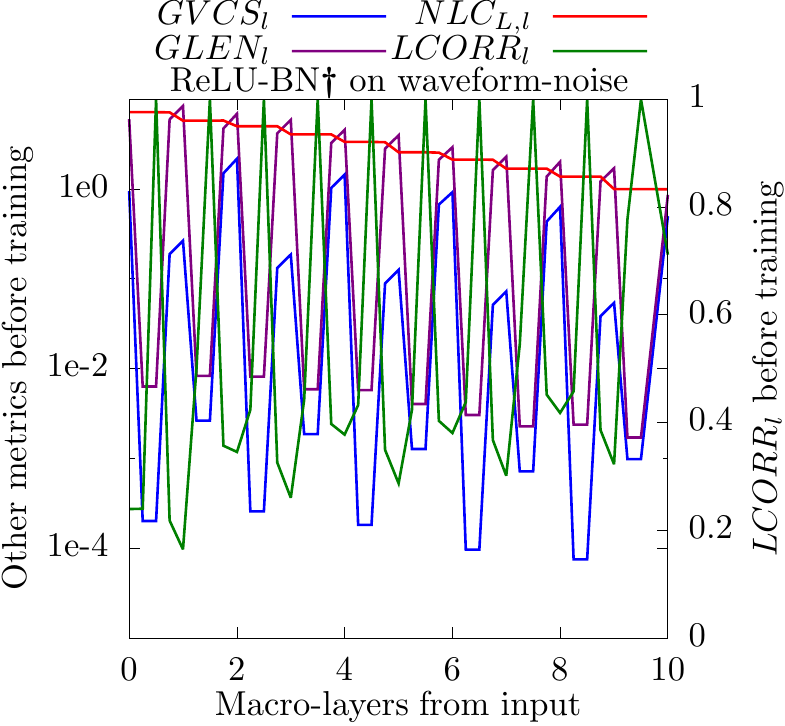}
\caption{Metric values for a depth-10 fully-connected ReLU-BN architecture with large weight and bias variance and width oscillation in the initial state. {\it Conclusion:} The NLC increases smoothly away from the output layer, but other metrics oscillate wildly. \textdagger See section \ref{vanishWhyBadSection} and \ref{additionalExperimentsSection} for architectural details.}\label{relatedOscillation}
\end{wrapfigure}

To further demonstrate the fickle nature of EVG, in figure \ref{relatedOscillation}, we give results obtained from a depth-10 ReLU-BN architecture with changing scale and width. Each macro-layer is made up of an FC layer, a bias layer, a BN layer and a ReLU layer, except the last macro-layer which does not contain a ReLU layer. The input layer has width 40 and the macro-layers have widths 1000, 10, 1000, 10, 1000, 10, 1000, 10, 1000, 3, in that order. FC layers have an initial weight variance of $10^6$ times the LeCun variance and components of the bias vectors are Gaussian initialized with mean zero and variance $10^{12}$. We find that \finding{both GLEN and GVCS oscillate wildly from layer to layer, but the NLC increases smoothly away from the output}. Clearly, there is no layer $f_l$ at which $GLEN_l$ or $GVCS_l$ represent some meaningful property of the network. Figure \ref{relatedOscillation} also calls into question the idea of a gradually growing or shrinking gradient as the exploding / vanishing terminology implies. Such a gradual change depends on a range of conditions, as we further discuss in the next subsection.

There are a myriad of methods that, at face value, ``deal with'' incorrectly sized gradients: regularization, Adam (section \ref{trainingAlgorithmsSection}), learning rates, etc. At first glance, these methods should simply be able to ``handle'' EVG. Given these methods, EVG should be merely a numerical inconvenience. We do not know what the community believes regarding whether those methods ``solve'' EVG.

\subsection{Exploding / vanishing gradients have multiple meanings} \label{vanishMultipleSection}

\begin{figure}
\centering
\includegraphics[width=0.98\textwidth]{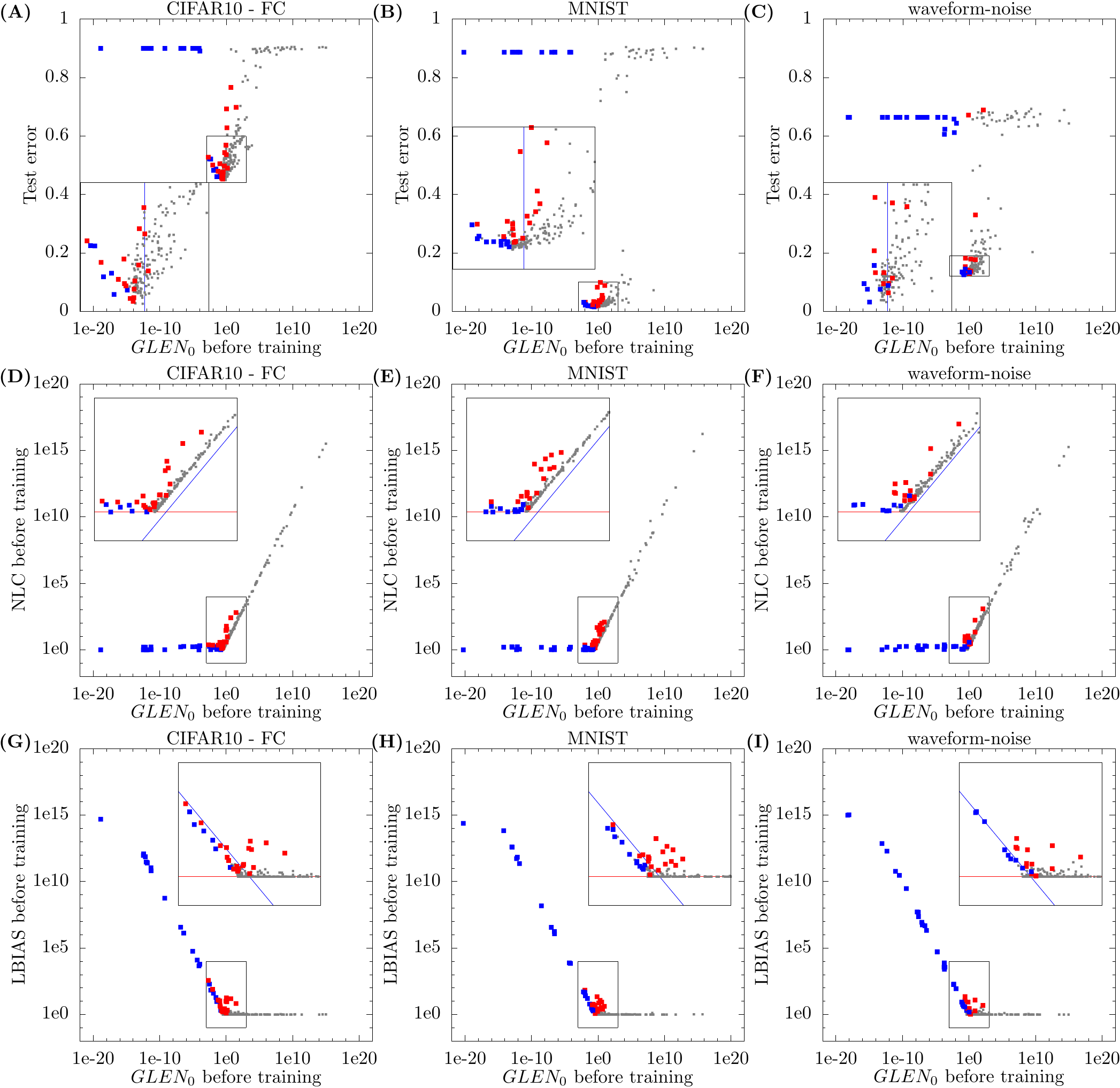}
\caption{Initial $GLEN_0$ vs test error, initial NLC and initial LBIAS for study A architectures. Colors are identical to figure \ref{surveyBias}. Inset graphs are magnifications of the region $0.001 < GLEN_0 < 1000$. Lines indicate $GLEN_0=1$, $NLC=1$, $NLC=GLEN$, $LBIAS=1$ and $GLEN*LBIAS=1$ respectively. Line colors are chosen for visual convenience. {\it Conclusion:} In the context of our study A architectures, gradient explosion / vanishing as measured by $GLEN_0$ is predictive of performance and associated with excessive NLC and LBIAS.} \label{relatedDichotomy}
\end{figure}

While EVG has no inherent meaning, in certain contexts, metrics like GLEN can be associated with a real pathology. It turns out that there is not just one such pathology, but many. We are aware of at least four distinct scenarios based on different underlying mechanisms that are lumped together under the EVG umbrella. Because of this ambiguity, many publications that discuss EVG make statements that can be considered, at the very least, misleading. This is true for even some of the most prominent architecture design papers, as we will show at the end of this subsection.

\subsubsection{In debiased networks, EVG can mean nonlinearity}

Let's compare the mean field estimates of NLC and $GLEN_0$ for A-architectures. The former is equal to $\sqrt{\frac{\mathfrak{g}_L(\mathfrak{q}_0 - \mathfrak{c}_0)}{\mathfrak{g}_0(\mathfrak{q}_L - \mathfrak{c}_L)}}$ (section \ref{meanFieldPracticalSection}). The latter is equal to $\sqrt{\frac{\mathfrak{g}_L}{\mathfrak{g}_0}}$ (see above). Under certain conditions, the two can be viewed as proportional. $\mathfrak{q}_0 \approx 1$ and $\mathfrak{c}_0 \approx 0$ is often achieved via data processing (section \ref{dataProcessingSection}). $\mathfrak{q}_L \approx 1$ corresponds to scale stability (section \ref{forwardStabilitySection}). $\mathfrak{c}_L \approx 0$ corresponds to a lack of output bias (section \ref{outputBiasSection}). $\mathfrak{g}_0=1$ holds by definition. That leaves the $\mathfrak{g}_L$ term to vary between architectures. The estimate of $GLEN_0$ also depends on the gradient of the loss function having unit scale. Hence, at least in practical fully-connected networks, the NLC is approximately proportional to $GLEN_0$ modulo data processing, scale stability, output bias and loss function. As long as these four factors are controlled for, large gradients may indicate excessive nonlinearity.

\begin{figure}
\centering
\includegraphics[width=0.98\textwidth]{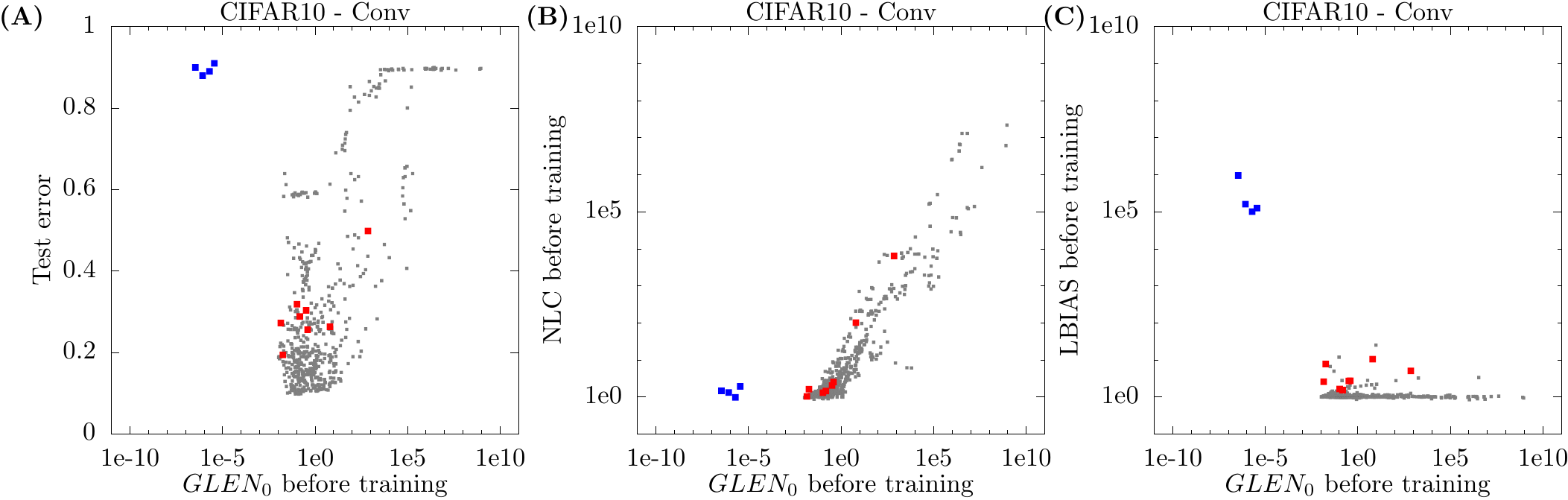}
\caption{Initial $GLEN_0$ vs test error, initial NLC and initial LBIAS for study B architectures. Colors are identical to figure \ref{surveyBias}. {\it Conclusion:} In the context of our study B architectures, gradient explosion / vanishing as measured by $GLEN_0$ is predictive of performance and associated with excessive NLC and LBIAS, though not as much as for study A architectures.}\label{relatedDichotomyConv}
\end{figure}

In figures \ref{relatedDichotomy}D-F and \ref{relatedDichotomyConv}B, we plot initial $GLEN_0$ vs NLC. The colors are the same as in figure \ref{surveyBias}, and correspond to the three convergence cases from theorem \ref{covkerCregular} discussed in section \ref{plainArchitectureSection}. We find that \finding{for type 3 architectures, depicted in grey, there is a strong linear relationship between $GLEN_0$ and NLC}. Type 3 architectures have small output bias. \finding{For type 3 FC architectures, depicted in figure \ref{relatedDichotomy}D-F, it appears we have $GLEN_0 \approx cNLC$ for some constant $c$ that is fixed across architectures}. It turns out that this constant arises from the scale of the gradient of the loss function. \finding{For convolutional architectures depicted in figure \ref{relatedDichotomyConv}B, the relationship is less strong}. This is because of the interaction of the Jacobian with the input covariance matrix in the numerator of the NLC, which we discussed e.g. in section \ref{nlcSimpleMetricsSection}.

The relationship between NLC and $GLEN_0$ for type 3 architectures also translates to a predictiveness of performance. In figures \ref{relatedDichotomy}A-C and \ref{relatedDichotomyConv}A, we plot initial $GLEN_0$ vs test error. \finding{Among architectures depicted in grey, the relationship between both metrics is clear}. As expected, \finding{the predictiveness is degraded for convolutional architectures}.

In section \ref{plainArchitectureSection}, we explained that for convergence case 3, the mean field NLC grows exponentially from macro-layer to macro-layer. Now, we find that the NLC is associated with $GLEN_0$. Hence, ``exploding gradients'' is related to ``exponentially growing NLC'' for plain stable architectures of convergence case 3. Also see our full list of architectures with definition and metric values in the appendix in chapter \ref{fullListChapter} for detailed information.

\subsubsection{In almost-linear networks, EVG can mean output bias}

Let's now compare the mean field estimates of LBIAS and $GLEN_0$ for A-architectures. The former is equal to $\sqrt{\frac{\mathfrak{q}_L}{\mathfrak{q}_L - \mathfrak{c}_L}}$ (section \ref{meanFieldPracticalSection}, \ref{outputBiasSection}). The latter is equal to $\sqrt{\frac{\mathfrak{g}_L}{\mathfrak{g}_0}}$ (see above). Again, assume $\mathfrak{q}_0 \approx 1$, $\mathfrak{c}_0 \approx 0$ and $\mathfrak{q}_L \approx 1$. Now also assume that the mean field NLC is roughly 1, i.e. $\sqrt{\frac{\mathfrak{g}_L(\mathfrak{q}_0 - \mathfrak{c}_0)}{\mathfrak{g}_0(\mathfrak{q}_L - \mathfrak{c}_L)}} \approx 1$. Then we obtain $\sqrt{\frac{\mathfrak{q}_L}{\mathfrak{q}_L - \mathfrak{c}_L}} \approx \Big(\sqrt{\frac{\mathfrak{g}_L}{\mathfrak{g}_0}}\Big)^{-1}$ and hence that LBIAS is approximately inversely proportional to $GLEN_0$ modulo data processing, scale stability, nonlinearity and loss function, at least in practical FC networks. As long as these four factors are controlled for, small gradients may indicate excessive output bias.

In figures \ref{relatedDichotomy}G-I and \ref{relatedDichotomyConv}C, we plot initial $GLEN_0$ vs LBIAS. Again, the colors are the same as in figure \ref{surveyBias}, and correspond to the three convergence cases from theorem \ref{covkerCregular} discussed in section \ref{plainArchitectureSection}. We find that \finding{for type 1 architectures, depicted in blue, there is a strong linear relationship between $GLEN_0$ and LBIAS}. These architectures have small NLC. \finding{For FC architectures depicted in figure \ref{relatedDichotomy}D-F, it appears we have $GLEN_0LBIAS \approx c$ for some constant $c$ that is fixed across architectures}. Again, it turns out that this constant arises from the scale of the gradient of the loss function. Note that the few architectures in the top left corner of figures \ref{relatedDichotomy}D-F that seemingly buck this trend slightly do so because numerical underflow affects the computation of LBIAS as described in section \ref{outputBiasSection} and \ref{nlcComputeSection}. \finding{For convolutional architectures depicted in figure \ref{relatedDichotomyConv}B, the relationship is difficult to evaluate, as there are only four type 1 architectures}.

The relationship between LBIAS and $GLEN_0$ for type 1 architectures translates to a predictiveness of performance. In figures \ref{relatedDichotomy}A-C, we find that \finding{among architectures depicted in blue, the relationship between $GLEN_0$ and test error is clear}. In figure \ref{relatedDichotomyConv}A, all four convolutional architectures with very small $GLEN_0$ exhibit random performance.

In section \ref{plainArchitectureSection}, we explained that for convergence case 1, the mean field LBIAS grows exponentially from macro-layer to macro-layer. Now, we find that LBIAS is inversely associated with $GLEN_0$. Hence, ``vanishing gradients'' is associated with ``exponentially growing LBIAS'' for plain stable architectures of convergence case 1.

\subsubsection{In RNNs, EVG can mean sensitivity imbalance}

As far as we know, the origin of the vanishing gradient concept lies in recurrent neural networks \citep{RNNvanishingGradient,hochreiterThesis}. Classical RNNs behave essentially like exponential moving averages. At each time step, the accumulated signal from past inputs is linearly combined with the signal stemming from the current input. Hence, the proportion of the signal that stems from a given time step decreases exponentially into the past. Hence, the final output is exponentially less sensitive to these inputs, and hence the gradient with respect to these older inputs decreases exponentially. This phenomenon has nothing to do with either nonlinearity or output bias as it arises even in RNNs exclusively composed of fully-connected and addition layers. Hence, the problem is of a different nature. While we have not studied RNNs in this work, this ``sensitivity imbalance'' pathology has been widely studied, as we further detail in section \ref{orthogonalitySection}.

\subsubsection{In tanh networks, EVG can mean gradient instability} \label{vanishTanhSection}

Again, we revisit the issue of the susceptibility of tanh networks to a change in weight magnitude, as we discussed in e.g. sections \ref{actFunNLCsection}, \ref{forwardStabilitySection}, \ref{covariateShiftSection}, \ref{plainArchitectureSection} and above. Large weights correspond to high mean field NLCs in plain tanh networks, whereas small weights yield a network that is close to linear. However, this does not tell the full story. The mean field regime studies the limit of increasing width. However, what happens when the weight magnitude increases but the width stays large but finite? If the weight variance is $\sigma^2$ times the LeCun variance and there are $M$ macro-layers, we have that the probability of having a gradient close to zero for a uniformly random input is $1 - O(\sigma^{-M})$ as $\sigma$ increases. For the remaining inputs, the gradient grows as $O(\sigma^M)$. Hence, the quadratic expectation of gradients across inputs grows as $O(\sqrt{\sigma}^M)$. Hence, the gradient {\it explodes in expectation} but {\it vanishes in probability}. We also obtain the $O(\sqrt{\sigma}^M)$ term when combining the $O(\sqrt{\lambda})$ growth rate in the bottom right corner of table \ref{covCurveillu1} with the $M$ exponent from the nonlinearity path equation (section \ref{pathEquationSection}).

As far as we know, the community generally understands the vanishing gradient problem in tanh networks as follows. ``If weights are initialized too large or grow because of an excessive learning rate, the gradient vanishes, which slows down training.'' The fact that gradients explode in expectation is usually omitted. In fact, for the exploding gradients to become ``invisible'', the $1 - O(\sigma^{-M})$ probability must be less than about 1 over the number of distinct inputs that are fed into the network. Of course, this requires that $\sigma$ is quite large. Throughout this work, we never encountered an architecture for which we had to choose a learning rate that leads to weights so large that this kind of vanishing was an issue, no matter the activation function. In section \ref{learningRateSection}, we did find some evidence that our very best architectures prefer starting learning rates that are on the larger side. If weight magnitude does become very large, it is not sufficient to say that ``gradients vanish'' or ``gradients explode'' in order to describe the whole phenomenon. The core issue is rather that gradient magnitudes are unstable from one input to another. This bears some similarity to the Gaussian instability phenomenon for layer values, and is an interesting topic of further study.

\subsubsection{Ambiguity in literature} \label{vanishPapersSection}

The fact that EVG amalgamates different pathologies leads to a great deal of confusion. Even the most prominent architecture design publications make statements about EVG that are, at least, misleading.

The paper that introduced SELU \citep{selu} writes:

{\it SNNs are based on scaled exponential linear units ``SELUs'' which induce self-normalizing properties like variance stabilization which in turn avoids exploding and vanishing gradients.}

SELUs do not avoid exploding gradients. When plugged into a plain architecture with LeCun initialized FC layers, the NLC grows exponentially with depth as discussed in section \ref{plainArchitectureSection}. This exploding NLC goes along with exploding $GLEN_0$, as described above. So replacing ReLU with SELU in a plain architecture actually causes exploding gradients, rather than eliminating them, as He-initialized ReLU architectures have a stable $\mathfrak{g}$ value. We suspect that the intent of the authors was to say that SELU avoids gradient instability because it is not tanh-like. Ironically, SELU can in fact reduce gradient growth because of its small activation function NLC (section \ref{pathEquationSection}, table \ref{basisillu}), but the authors do not seem to have been aware of that fact.

The paper that introduced batch normalization \citep{batchNormalization} writes:

{\it In traditional deep networks, too-high learning rate may result in the gradients that explode or vanish, as well as getting stuck in poor local minima. Batch Normalization helps address these issues.}

This is accurate in that gradient instability in tanh and sigmoid networks, which can be considered the ``traditional'' networks as explained in section \ref{historySection}, is alleviated by BN, because if layer values are ``blown up'' by large weights, they are ``scaled back down'' by BN. However, BN can actually induce exploding gradients in the nonlinearity sense described above. In section \ref{nlnormDefinitionSection}, we explained how debiasing increases nonlinearity. In section \ref{plainArchitectureSection}, we explained how a zero (converging) LBIAS is associated with an exponentially growing NLC. For example, inserting BN into a plain He-initialized ReLU architecture brings about exploding gradients \citep{expl}.

The paper that introduced ResNet \citep{batchNormalization} writes:

{\it Is learning better networks as easy as stacking more layers? An obstacle to answering this question was the notorious problem of vanishing/exploding gradients, which hamper convergence from the beginning. This problem, however, has been largely addressed by normalized initialization and intermediate normalization layers, ...}

Neither normalized initialization (section \ref{vanishWidthSection} above) nor intermediate normalization layers fundamentally deal with exploding gradients in the nonlinearity sense. As we showed in chapter \ref{nlnormChapter}, we need to normalize nonlinearity either directly or indirectly. It is notable that by far the most popular strategy for reducing the NLC, ResNet, came out of a paper that did not even recognize that exploding gradients were still an issue, much less recognized that ResNet has some ability to alleviate them.

\subsection{Exploding and vanishing gradients can cancel each other out} \label{vanishDichotomySection}

In the previous subsection, we showed how plain stable architectures of convergence case 1 may exhibit vanishing gradients inversely proportional to their LBIAS, and how plain stable architectures of convergence case 3 may exhibit exploding gradients proportional to their NLC. But what about convergence case 2? The problem is that in plain stable architectures, a stable gradient does not correspond to an absence of pathology, but to a potential presence of {\it two} pathologies: excessive NLC and excessive LBIAS. In section \ref{plainArchitectureSection}, we argued that, overall, case 3 is preferable to case 2. Indeed, in figures \ref{relatedDichotomy}A-C and \ref{relatedDichotomyConv}A, we find that \finding{many type 2 architectures underperform type 3 architectures of comparable $GLEN_0$ value}. In figure \ref{relatedConf}G, we give results obtained from training depth-2 architectures with the following activation function: $1 + c\tau_\text{sawtooth}(\frac{1}{c}s)$, which we term `biased sawtooth'. Results are averaged across 100 random seeds. $\tau_\text{sawtooth}$ is defined as in table \ref{actFunIllu}. The parameter $c$ controls both the period and amplitude of the oscillations of the activation function. Using the notation of section \ref{actFunCovKernelSection}, we have $\tilde{\mathfrak{C}}_{\tau_\text{biased sawtooth}}'(1) = 1$, so biased sawtooth falls under case 2. In fact, NLC and LBIAS both increase linearly with $c$. However, the absolute derivative of $\tau_\text{biased sawtooth}$ is always 1, independently of $c$. Hence, $GLEN_0$ is approximately constant as $c$ varies. Because NLC and LBIAS ``balance each other out'', neither is reflected in the gradient magnitude. Indeed, in figure \ref{relatedConf}G, we find that \finding{test error increases as NLC increases and $c$ decreases, but $GLEN_0$ is stable}. As in previous subsections, EVG is not predictive of performance.

The same issue can arise even in an architecture as popular as the plain He-initialized ReLU architecture, which also falls under case 2. Specifically, as explained in section \ref{actFunNLCsection}, mean field NLC and LBIAS grow linearly with depth. Mean field $GLEN_0$, on the other hand, is constant. Again, both pathologies mask each other in the value of $GLEN_0$. In figure \ref{relatedConf}F, we give results from training plain ReLU architectures of various depths. As expected, we find that \finding{NLC increases and performance decreases with depth}. It turns out that with depth, both \finding{$GLEN_0$ and $GVCS_0$ decrease slightly as they diverge from their mean field estimate}. This happens because ReLU is also on the Gaussian edge (section \ref{gaussianStabilityExplanationSection}, table \ref{covCurveillu1}), so with great depth there is some instability in both layer value and gradient magnitude.

\citet{heInit} believed that their initialization would ``solve'' the gradient issue in deep plain ReLU architectures. It turns out it solves neither of the two underlying pathologies.

\subsection{Exploding gradients can be a good thing} \label{vanishGoodSection}

The idea behind exploding gradients is that they are supposed to be harmful. However, depending on how we define ``explosion'', the opposite may be true. Throughout this section, we have mostly equated ``exploding gradients'' with ``large gradients''. However, we can also simply interpret ``explosion'' as ``exponential growth'', no matter to what level the gradient grows.

We have demonstrated that an excessive NLC is associated with high test error. However, exponential growth of the NLC is not associated with high test error, as long as the total network NLC is not excessive. Consider a plain SELU architecture. We have $\mathfrak{n}_{\tau_\text{SELU}}(1,0) = 1.035$ (table \ref{basisillu}). Hence, a depth-50 plain LeCun initialized SELU architecture still only has a mean field NLC of 5.58. This allowed \citet{selu} to train very deep SELU architectures. According to the nonlinearity path equation, an exponentially growing NLC in an architecture composed of a single dependency chain simply implies that nonlinearity is distributed evenly over the activation layers. In section \ref{pseudoLinearitySection}, we argue that this is actually a positive, not a negative, as it avoids pseudo-linearity. As mentioned above, in section \ref{plainArchitectureSection}, we argue that convergence case 3, which tends to exhibit exponentially growing gradients, is overall preferable to convergence case 2, which does not.

\subsection{Summary} \label{vanishSummarySection}

EVG is a microcosm of ZSAD guidelines. At first glance, the notion of tiny or enormous gradients ``just doesn't seem right''. Therefore, researchers ``buy into'' the idea that EVG is in fact a pathology without there being an underlying argument, and without a way to quantify and measure the phenomenon. At second glance, one may realize that methods like Adam scale gradients, and therefore EVG, may in fact be completely irrelevant. When one digs deeper, one finds that there are specific contexts where EVG does predict, and to a limited degree explain, performance. However, these contexts rely on assumptions that are generally unspecified. Over time, different researchers develop different and potentially conflicting ideas about what EVG is and means and when it arises, and the number of unsubstantiated assumptions and objectives grows. Finally, progress in architecture design is negatively affected. Had the investigation of EVG been centered around a concrete metric like $GLEN_0$, we believe these problems might have been less severe. Hence, this work is concerned first and foremost with the NLC, and only secondarily with the abstract notion of nonlinearity.

\subsection{A note on the Jacobian and Lipschitz constant} \label{vanishLipschitzSection}

In this section, we based our analysis on the GVCS and especially GLEN metrics. EVG has also been measured by metrics applied to the Jacobian or its square, e.g. Frobenius norm, mean eigenvalue and maximum eigenvalue (e.g. \citet{explosionGeneralization,eigenspectrum,RecurrentNetsPascanu, orthogonalInitialization}). A particularly prominent example of this is the Lipschitz constant. For a differentiable network, the Lipschitz constant across some input space is the maximum absolute singular value of any Jacobian across that space. While the Lipschitz constant is essentially gradient-based, the literature on the Lipschitz constant is somewhat separate from the EVG literature. \citet{lipschitz,lipschitz2,lipschitz3} are recent papers that provide an overview.

Metrics defined in terms of the Jacobian ultimately suffer from the same or similar drawbacks as those discussed in this section for GLEN and GVCS, including being confounded by re-scaling and width change. Therefore, we will not give a separate in-depth discussion.

\section{NLC vs order / chaos / edge of chaos, depth scale, correlation preservation and signal propagation} \label{orderChaosSection}

In chapter \ref{meanFieldNnaChapter}, we discussed the mean field theory of neural networks. In the context of this work, we use mean field theory primarily to predict architecture properties in the initial state based on the architecture definition. These properties are then used to further predict performance after training. It turns out that the same two steps are present in the line of works that popularized mean field theory to begin with. For the purpose of this section, we primarily consider the following works: \citet{correlationLimit,depthScalesMeanField,resNetMeanField, meanFieldCNN,meanFieldRNN,meanFieldDeepField,meanFieldBN}. Our work benefits substantially from these studies by building upon them in chapter \ref{meanFieldNnaChapter} and beyond. These authors propose several interrelated concepts for the purpose of performance prediction. The five most prominent concepts are as follows: exploding / vanishing gradient, order / chaos / edge of chaos, depth scale, correlation preservation and signal propagation. These correspond to the following ZSAD guidelines.

\begin{itemize}
\item avoid exploding / vanishing gradients
\item choose an architecture on the edge of chaos
\item maximize depth scale
\item preserve input correlation during forward propagation
\item maximize signal propagation
\end{itemize}

Of course, we will not repeat our analysis of exploding / vanishing gradients (EVG) from the last section though we will refer back to it frequently. Ultimately, the other guidelines are related to EVG, and so our discussion and criticism are similar. As in section \ref{vanishingGradientSection}, each of the following subsections covers one criticism.

\subsection{Order / chaos / edge of chaos is limited to simple, homogeneous architectures} \label{orderChaosLimitedSection}

The order / chaos / edge of chaos concept corresponds directly to the three convergence cases of theorem \ref{covkerCregular}, and hence to the trichotomy we described in section \ref{plainArchitectureSection}, which also featured prominently in section \ref{vanishingGradientSection}. In a plain stable architecture, we obtain one of three behaviors. Our case 1, which we identify with exponential LBIAS, related works term `order'. Our case 2, which we identify with sub-exponential NLC and LBIAS, they term `edge of chaos'. Our case 3, which we identify with exponential NLC, they term `chaos'. (As expected based on theorem \ref{covkerCregular}, the sub-case we exclude in section \ref{plainArchitectureSection} based on the existence of the second derivative of $\tilde{\mathfrak{C}}$ at 1 is part of `order'.) They argue in favor of edge of chaos primarily on the basis of a stable mean field estimate of gradient magnitude, which arises as described in sections \ref{vanishMultipleSection} and \ref{vanishDichotomySection}, and sub-exponential correlation convergence, as explored later in this section. Throughout the rest of this section, we shorten `order / chaos / edge of chaos' to OCE.

The first and most obvious limitation of OCE is that it is well-defined only for a small set of architectures. While related work extends the trichotomy to cover convolutional layers, bias layers, gated recurrent layers, dropout and skip connections, the restrictiveness of the assumptions that are required to obtain the highly regular convergence behavior implied by the trichotomy is still considerable. In particular, there must be no variation in weight initialization or activation function from macro-layer to macro-layer. Also, the definition of OCE ends up being multiplicitous and specific to the architecture type. Another requirement for OCE is that mean field estimates have to be predictive for the actual network at high depth, which requires at least Gaussian stability. 

\subsection{Order / chaos / edge of chaos is not predictive if the network does not behave like its infinite depth limit} \label{chaosLimitSection}

OCE is concerned with the convergence behavior as depth tends to infinity. Hence, it does not capture the behavior of architectures that are too shallow to behave ``according to their limit''.

Consider a plain SELU architecture. Because the activation function NLC (section \ref{pathEquationSection}, table \ref{basisillu}) of SELU is small, very deep architectures can still have an acceptably small NLC. Hence, while it is technically chaotic, at practical depth, the exponential growth of the NLC in SELU architectures is not harmful because the base of the exponent is so close to 1. To determine whether the value of a power is excessive, we need to take into account both base and exponent, not simply observe that ``it is a power''.

Conversely, consider a plain architecture using the `biased sawtooth' activation function as in section \ref{vanishDichotomySection} / figure \ref{relatedConf}G. This architecture type is on the edge of chaos for any value of the sawtooth period, which is supposed to imply that it can attain a low test error at considerable depth. However, for small periods, even with 2 macro-layers / 1 activation layer, the NLC is very high, which is reminiscent of a ``highly chaotic'' architecture. This is because, while $\tilde{\mathfrak{C}}$ converges sub-exponentially to 1 upon iteration, the first application of $\tilde{\mathfrak{C}}$ already returns a value close to 1 if the sawtooth period is small.

\subsection{Actually, chaos is good, edge of chaos can be bad, and order is not that bad} \label{chaosGoodSection}

In section \ref{plainArchitectureSection} and again in section \ref{vanishWhyBadSection}, we argued that chaos / convergence case 3 is actually the preferred behavior, as nonlinearity is spread evenly between macro-layers and neuron bias is avoided. We simply have to calibrate the nonlinearity of individual macro-layers with depth. \citet{eigenspectrum,meanFieldDeepField} actually do this and \citet{meanFieldBN} acknowledge the possibility of ``succeeding with chaos''. As we showed, edge of chaos can go along with both excessive NLC and LBIAS. Case in point, switching from a deep ReLU architecture (edge of chaos) to a deep SELU architecture (chaos) tends to reduce test error significantly. In figures \ref{relatedDichotomy}A-C and \ref{relatedDichotomyConv}A, we find that \finding{marker color is not the primary driver of performance}.

\citet{depthScalesMeanField} states:

{\it Thus arbitrarily deep networks may be trained only sufficiently close to criticality.}

This statement is not true for architectures that exhibit a high degree of order if one goes beyond basic training algorithms. In section \ref{outputBiasSection}, we showed that architectures that exhibit a high degree of order, i.e. have tiny gradients / enormous LBIAS, can be trained and can generalize as long as we use (i) output debiasing via the loss function, (ii) e.g. debiased gradient descent or first-layer-only training and (iii) sufficient floating-point precision. Further, as we showed in section \ref{nlcPredictiveSection}, the above statement is also not true for architectures that exhibit a high degree of chaos as long as we consider very small learning rates and have sufficient floating-point precision. Architectures with enormous NLC / gradients can train with those learning rates, though they cannot generalize. Through correspondence with the authors, we confirmed that they did not consider very small learning rates. This was a key motivation for us to design the extensive experimental protocol we used in our empirical studies (chapter \ref{empiricalStudiesChapter}).

\subsection{Order is not the opposite of chaos} \label{chaosOppositeSection}

The discussion of OCE in related work implies that order and chaos are two manifestations of the same pathology. Their experiments appear to show that both order and chaos impact performance equally. As we described in the previous subsection, ordered and chaotic architectures can in fact perform to certain degrees, but under very different circumstances. In this work, we argue that the actual pathologies that correspond to order and chaos are excessive bias and nonlinearity respectively. There is a wide range of qualitative differences, in addition to quantitative differences between the two phenomena, as we have explored throughout this work.

\subsection{Order / chaos / edge of chaos is not fine-grained, but depth scale is even more limited in scope} \label{chaosDepthScaleSection}

Using OCE itself has limitations in terms of performance prediction, because it can only assign an architecture to one of three discrete classes, but cannot distinguish within a class. Related work proposes two real-valued metrics for providing a more fine-grained ranking: `depth scale' (called `timescale' in \citet{meanFieldRNN}) and gradient vector quadratic mean, which we capture by GVCS. We have discussed the latter in detail in section \ref{vanishingGradientSection}. Using our terminology from section \ref{actFunCovKernelSection}, depth scale in plain stable$(q,1)$ architectures is defined to be $|\log\tilde{\mathfrak{C}}_\tau'(c^\text{lim})|^{-1}$, where $c^\text{lim}$ is the iteration limit of $\tilde{\mathfrak{C}}_\tau$. The idea is that the larger this value, the slower the convergence of $\mathfrak{c}$ and therefore the greater the degree to which correlations are preserved.

To start, depth scale inherits all the drawbacks of OCE discussed in previous subsections. Beyond this, it fails to rank architectures when they are on the edge of chaos. For those architectures, depth scale is equal to infinity, and hence all architectures appear ``equally good or bad''. Depth scale is also even more limited than OCE in that it depends on the architecture only via the activation function. Therefore, it is unable to capture other architecture properties. Hence, related work does not apply depth scale to e.g. residual or BN architectures.

\subsection{Correlation is confounded by shifting} \label{chaosCorrelationConfoundingSection}

Related work justifies the use of depth scale and the consideration of OCE in the first place via correlation. In plain stable architectures, as depth increases, the mean field co-mean after the FC layer converges to $c^\text{lim}$ independently of the input co-means. If means and square means are also stable, then mean field correlation also converges. Related work argues that a change of correlation between different inputs as they are forward propagated is a pathology. One reason why this is inaccurate is that correlation can be manipulated easily via shifting, just as EVG can be manipulated via scaling as discussed in subsection \ref{linearEquivarianceSection}. To demonstrate this effect empirically, as always, we define a metric.

\begin{metricDefinition}
The `layer correlation' (LCORR) is

$$LCORR_l(f, \mathcal{D}) = \sqrt{\mathbb{E}_{x,x'}(\mathbb{C}_{i,i'}(f_l(x)[i], f_l(x')[i']))^2}$$
\end{metricDefinition}

The correlation operator $\mathbb{C}$ is defined and estimated as in section \ref{metricEstimationSection}.

Consider, for example, a network that begins with a linear layer followed by BN. If we choose an arbitrary constant vector and add it to all inputs, the outputs of the network are preserved. Hence, given any high-performing architecture that starts as described, we can manipulate the input-output correlation map and still maintain high performance. 

In figure \ref{relatedConf}E, we give results from training a depth-5 ReLU-BN architecture where the input was shifted by a vector where each component was drawn from a Gaussian with mean zero and a variance value that was the same across components. We train the architecture with 7 different variance values. We train with SGD, except that we set the learning rate for the first FC layer to zero in order to eliminate the impact of the bias on training. Once we do that, the final parameter value is independent of the input bias. We find that \finding{test error, NLC and the output correlation $LCORR_L$ are independent of the input bias, but the input correlation $LCORR_0$ is not}. So, regardless of whether $LCORR_0$ is less than $LCORR_L$, similar to $LCORR_L$ or larger than $LCORR_L$, the performance is the same. Differing levels of $LCORR_0$ are not uncommon in practice as input bias depends on what type of data processing is used and whether data processing is used at all.

Shifts and biases can also manipulate correlation at intermediate layers. In figure \ref{relatedOscillation}, we gave results from the initial state of a depth-10 ReLU-BN architecture with oscillating widths, FC layers that use $10^6$ times the LeCun variance and bias vectors that have components which are initialized as independent Gaussians with mean zero and variance $10^{12}$. We find that in that architecture, \finding{LCORR moves wildly up and down from one layer to another}. If changing correlation is indeed a problem, then should this be interpreted as an extreme case of pathology? It turns out that introducing large biases that are immediately removed by a BN layer does not impact training or performance.

\subsection{Correlation preservation lacks general meaning} \label{chaosCorrelationMeaningSection}

Like EVG, correlation preservation can be related to performance in specific contexts. The mean field limit of $LCORR_l$ in A-architectures is $\sqrt{\frac{\mathfrak{c}_l - \mathfrak{m}_l^2}{\mathfrak{q}_l - \mathfrak{m}_l^2}}$, which is monotonically related to mean field $LBIAS_l$ when $\mathfrak{m}_l=0$, which happens e.g. at FC layers. Hence, correlation is related to neuron bias and output correlation is related to output bias. Hence, if a change in correlation goes along with high $LBIAS_l$ and a preservation of correlation goes along with low $LBIAS_l$, the degree of correlation preservation can be related to performance in the same vein as neuron bias.

While correlation can capture bias, there is no general argument in related work to suggest that specifically preserving correlation is a meaningful thing to do. Even when it comes to bias, there are many open questions. In the previous subsection, we found that certain enormous biases can be entirely harmless. In section \ref{outputBiasSection}, we found that it is possible to mitigate enormous biases that are present throughout the network by modifying the training protocol. It does not appear that bias is a ``deep enough'' pathology to support an entire ZSAD theory.

\subsection{Signal propagation is not well-defined and lacks meaning} \label{chaosSignalSection}

The final step used in related work for arguing for the importance of OCE, depth scale and correlation is a concept they term `signal propagation'. 

\citet{depthScalesMeanField} writes:

{\it We show the existence of depth scales that naturally limit the maximum depth of signal propagation through these random networks. ... We hypothesize that a necessary condition for a random neural network to be trainable is that
information should be able to pass through it.}

\citet{meanFieldCNN} writes:

{\it These studies revealed a maximum depth through which signals can propagate at initialization, and verified empirically that networks are trainable precisely when signals can travel all the way through them.}

\citet{meanFieldRNN} writes:

{\it Afterwards, we move on to a simple gated architecture
to explain the role of gating in facilitating signal propagation in RNNs.}

The implication in all of these statements is that ``signals propagate if and only if the correlation of inputs is roughly preserved from layer to layer''. While the phrase ``signals do not propagate'' sounds alarming, ``signal propagation'' is never defined by related work. There is not even a meaningful definition of signal propagation such that it would be related to correlation. Consider some matrix $A$ where each row corresponds to a layer value corresponding to one input. Let $A'$ be derived from $A$ by adding a bias vector to each row. Then $A$ and $A'$ can have wildly different row correlations for various bias vectors. But it makes no sense to say that either $A$ or $A'$ therefore has no signal. Many datasets have naturally biased input components. In many cases, those biases are removed by data processing. Does that mean that ``signal cannot propagate through the data processing stage''?

\section{NLC vs Hessian magnitude} \label{hessianSection}

\begin{figure}
\centering
\includegraphics[width=0.98\textwidth]{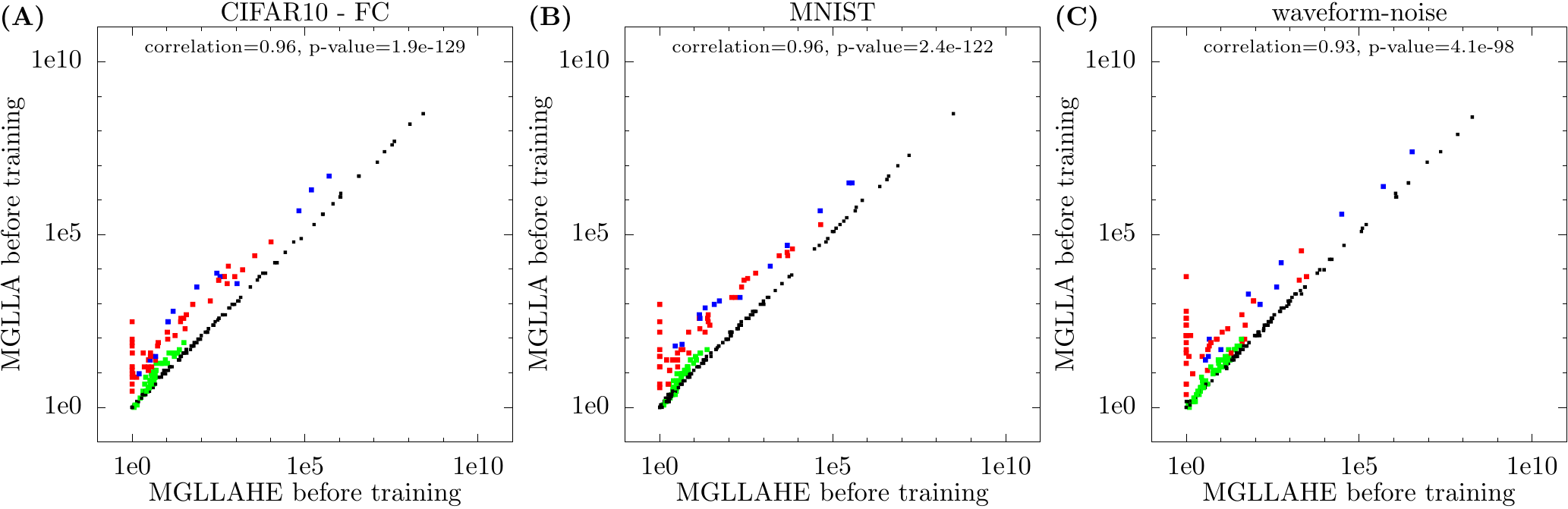}
\caption{MGLLA as defined in section \ref{nlcSensiSection} vs its counterpart defined via the local quadratic approximation, MGLLAHE, for study A architectures in the initial state. As in figures \ref{sensiDist} and \ref{nlcSensiGrad}, we set the tolerance $T$ to 2. Red points correspond to architectures based on the ReLU activation function. Blue points correspond to architectures based on the even tanh activation function. Green points correspond to architectures based on the SELU activation function. Only architectures for which $MGLLAHE < 10^9$  and $MGLLA < 10^9$ are depicted, due to limitations associated with floating-point computation. {\it Conclusion:} While MGLLAHE is a highly accurate estimate of MGLLA when the second derivative of an activation function captures the error of its first derivative well, MGLLAHE fails when this is not the case.} \label{sensiHessian}
\end{figure}

The Hessian, and second-order information in general, has not been an especially popular concept in the deep learning field, as evidenced by the popular training algorithms listed in section \ref{trainingAlgorithmsSection} which eschew explicit second-order information. One of the reasons for this is that, historically, second-order information has been difficult to compute for deep networks. \citet{hessianCompute1,hessianCompute2} have recently made progress in this area.

For general functions, the Hessian is commonly linked to the concept of nonlinearity. Let's say some function $F : \mathbb{R}^{d_\text{in}} \rightarrow \mathbb{R}$ has local quadratic approximation $F(\chi) + (\chi'-\chi)\frac{dF}{d\chi}^T + (\chi' - \chi)\frac{d^2F}{d\chi^2}(\chi' - \chi)^T$. If the quadratic term takes on large values relative to the linear term, then the local linear approximation is an inaccurate proxy for the local quadratic approximation. Since the local quadratic approximation itself differs from $F$, the local linear approximation will be far from $F$. Conversely, if the quadratic term takes on small values relative to the linear term, then the local linear approximation is an accurate proxy for the local quadratic approximation. If the local quadratic approximation also happens to be close to $F$, then so will the local linear approximation.

In essence, the magnitude of the quadratic term can be used as the basis for an estimate of the gradient-based local linear approximability. Of course, this estimate may itself be more or less accurate. In section \ref{nlcSensiSection}, we studied the gradient-based local linear approximability of neural networks, and study A architectures in particular. Concretely, we defined and computed the MGLLA metric and presented results in figure \ref{nlcSensiGrad}. To estimate gradient-based local linear approximability via the Hessian, we can simply plug the Hessian into the definition of MGLLA.

\begin{metricDefinition}
The `median gradient-based local linear approximability Hessian estimate' (MGLLAHE) is analogous to MGLLA in section \ref{nlcSensiSection}, except that the value of $f(x^{(b)} + c\delta_\text{in}^{(b)})$ is replaced by the value of the local quadratic approximation of $f$ around $x^{(b)}$ evaluated at $x^{(b)} + c\delta_\text{in}^{(b)}$, i.e. $$f(x^{(b)}) + \frac{df(x^{(b)})}{dx}\delta_\text{in}^{(b)} + \sum_{i,i',j=0}^{d_\text{in}-1,d_\text{in}-1,d_\text{out}-1}\frac{d^2f(x^{(b)})}{dx^2}[i,i',j]\delta_\text{in}[i]\delta_\text{in}[i']$$
\end{metricDefinition}

For the Hessian to yield an accurate estimate of gradient-based local linear approximability, MGLLAHE must be close to MGLLA. In figure \ref{sensiHessian}, we plot the two metrics against each other for study A architectures in the initial state. We find that \finding{while for many architectures both metric values are very close, there is a significant to extreme difference for some architectures}. To further explain the different behaviors, we color architectures based on ReLU in red, architectures based on even tanh in blue, architectures based on SELU in green, and architectures based on tanh, sigmoid, Gaussian, square or odd square in black. We find that \finding{ReLU-based architectures exhibit the largest difference between both metrics, followed by even tanh, followed by SELU, and both metrics are virtually equal for architectures depicted in black}.

Perhaps unsurprisingly, this trend is entirely based on the local quadratic approximability of the activation function. For example, in the quadratic approximation of ReLU, the quadratic term is equal to zero. Hence, it fails completely at measuring the error of the linear approximation. In figure \ref{sensiHessian}, \finding{there are architectures based on ReLU that have an MGLLA value of up to $10^4$ but have an MGLLAHE value of exactly 1}! These are architectures that do not use BN or LN and therefore the quadratic term of the entire network is zero. Hence, MGLLAHE ``believes'' that the network is a linear function. MGLLAHE can be greater than 1 when ReLU is used with BN or LN. However, we find that \finding{the match with MGLLA is still relatively poor.}

Even tanh is similar to ReLU in that a ``significant proportion of its nonlinearity'' is contained in the non-differentiable point at zero. Again, the quadratic term cannot capture this source of nonlinearity, and hence MGLLAHE significantly underestimates MGLLA for architectures based on even tanh. The same effect happens for SELU, though it is less severe because the change in direction of $\tau_\text{SELU}$ at zero is less drastic.

The fact that Hessian-based metrics fail completely for the most popular activation function should be sufficient reason to prefer the NLC over a Hessian-based metric for capturing nonlinearity. However, the difference between both approaches goes deeper. Of course, practical neural networks are often neither once nor twice differentiable everywhere, including all networks based on ReLU. So, neither NLC nor MGLLAHE is ultimately well-defined in the strictest sense. However, neural architectures must have a meaningful, computable local linear approximation in order for them to be trainable by gradient methods. We explain this in section \ref{nonDifferentiableSection}. Having at least such a surrogate gradient is baked into the functional-gradient paradigm (section \ref{functionalGradientSection}) and hence into architecture design. The NLC only requires a surrogate gradient, as explained in sections \ref{nlcFunctionalGradientSection} and \ref{nlcComputeSection}. We argue for the NLC over MGLLAHE and other Hessian-based metrics because the NLC does not place additional requirements on the architecture beyond what is already placed on it by the functional-gradient paradigm, whereas the Hessian does.

If one is seeking a measure of nonlinearity in addition to the NLC, or a measure of nonlinearity that is explicitly based on local linear approximability, we do not see a reason to prefer a Hessian-based metric over MGLLA, except possibly computational considerations. We discuss the computational issues around MGLLA in section \ref{nlcSensiSection}. We used numerical differentiation to compute MGLLAHE, and so, just like for MGLLA, we were unable to compute values larger than $10^9$.

\section{NLC vs ``use an appropriate depth''} \label{depthSection}

\begin{figure}
\centering
\includegraphics[width=0.98\textwidth]{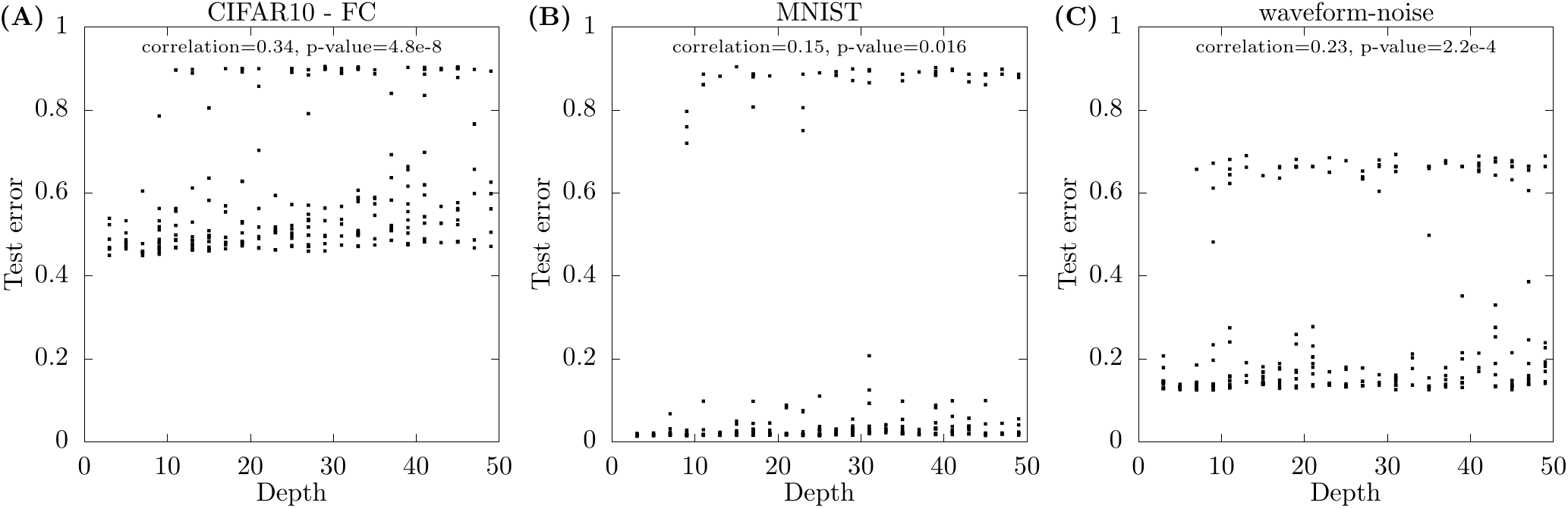}
\caption{Depth vs test error for study A architectures. {\it Conclusion:} Increasing depth tends to increase test error and there is no identifiable benefit to increasing depth beyond, say, 10.} \label{relatedDepth}
\end{figure}

We discussed the notion of depth in detail in sections \ref{neuralNetworksSection}, \ref{architectureDesignParadigmsSection} and \ref{surveyDepthSection}, and throughout this work. Mostly, we have discussed depth as a basic architecture property as well as a practical design strategy that has been empirically successful. In contrast, when we reference ``use an appropriate depth'' such as in section \ref{darkArtSection}, we refer to the ZSAD guideline that postulates that, for principled reasons, a certain depth is highly desirable, or even necessary, for achieving success with neural networks. It encapsulates the belief that depth holds the key to neural network success, which is exemplified in the term `deep learning', which elevates the notion of depth above all other neural network concepts, and even machine learning concepts. In this section, we question this common belief.

We have repeatedly criticized the ill-defined nature of depth. Breaking down a network function into layers is a subjective process. In this work, for example, we distinguished `layers' based on a single operation, like the fully-connected operation or activation operation, from `macro-layers', which are composed of e.g. a linear operation and an activation operation and a normalization operation. Depending on how we define the layer graph, the length of the longest directed path in that graph can change drastically. Even if we can decide upon a layer graph, it is not clear whether considering only the length of the longest directed path in it is sufficient. What if there are many paths from input to output layer? Is it better to consider e.g. the average length of paths? What happens if the longest path contains only linear layers and is therefore equivalent to a length 1 path (sections \ref{pseudoLinearitySection}, \ref{effectiveDepthSection})? Ultimately, like with EVG and OCE, assessing architectures by their depth value only makes sense in specific, limited contexts where we can control e.g. the overall layout of the layer graph, the composition of macro-layers, etc. Throughout this work, we have only used architectures for empirical evaluation that were simple enough such that a depth value could be specified for them based only on the ``popular consensus'' of the deep learning community.

Over the years, the primary argument that has emerged for the necessity of depth is that there are functions that can be represented by architectures of moderate depth and width but that cannot be represented by shallow (e.g. depth-2) architectures unless they have enormous width. \citet{depth9} provides an overview and \citet{depth8,depth10} are recent entries in this line of work. The functions considered by these works that supposedly require high depths are generally ``complex'' in the sense that a model that learns those functions would have to have high model complexity. In e.g. sections \ref{nlcKernelSection} and \ref{networkCovKerSection}, we established the NLC as a measure of model complexity. The core relationship between the NLC and depth is established by the nonlinearity path equation in section \ref{pathEquationSection}. Multiplying a larger number of $\mathfrak{n}_\tau$ terms can lead to a larger mean field NLC. However, the NLC is determined by an enormous number of factors besides depth, and we discussed a significant number of them in this work. Hence, to obtain a strong relationship between depth and model complexity, we must control a large number of factors, such as layer operations used, macro-layer composition, activation functions, presence of skip connections, etc. Indeed, many works that argue for the ability of deeper architectures to represent more complex functions severely restrain the types of layers and overall layout that architectures they consider are allowed to have. This is a severe drawback. This work clearly shows that model complexity can be increased e.g. by changing activation functions and hence $\mathfrak{n}_\tau$ values. Hence, architectures of any depth can be arbitrarily complex. Hence, the statement ``deeper networks are more complex'' is, at the very least, highly simplistic. ``Depth is needed because complexity is needed'' is simply inaccurate.

To make a case for the importance of depth, one would have to show that depth has value {\it even when controlling for model complexity}. In fact, we do believe that this is true and we believe that this independent contribution of depth is an important driver behind the success of deep networks. However, we are not aware of any studies showing this. This is a highly interesting topic for future work.

In section \ref{surveyDepthSection}, we pointed out that depth is necessary to achieve certain NLC levels with popular activation functions. However, this turns out to be a chicken-and-egg problem. Are low $\mathfrak{n}_\tau$ values necessary because of the prevalence of high depths or are high depths needed because of the prevalence of low $\mathfrak{n}_\tau$ values? More work is needed to answer this question.

Another point that is central to this work but is often not discussed in the ``depth literature'' is that complexity is just as likely to be harmful as it is to be helpful. As we found in section \ref{nlcPredictiveSection}, the majority of our study A architectures overfit, many of them drastically. However, we were unable to find a trend that indicates underfitting. We were able to find such a trend in study B, but only because we modify our activation functions to make them more linear, as described in sections \ref{studyBArchitecturesSection} and \ref{nlnormDefinitionSection}. Without deliberate design, the supposed benefit of depth turns out to be a curse. In figure \ref{relatedDepth}, we confirm this by plotting the depth of our study A architectures vs test error. We find \finding{a significant positive correlation, and we do not observe any benefit from increasing depth beyond, say, 10}. Increasing depth has the potential to make all pathologies we discuss in this work more severe, including excessive NLC, excessive neuron bias and scale / Gaussian / noise instability.

Another argument against the importance of depth is the fact that shallow networks can often perform just as well as practical deep networks as long as they are given side information during training. The line of work investigating this phenomenon is referred to as `information distillation' \citep{distillationOriginal}. \citet{distillation} provides a recent overview. Hence, some of the practical benefits of depth may come from a better ``fit'' with gradient-based training algorithms rather than a greater ability to represent complex functions.

While the benefits of depth may be somewhat tenuous, the relationship between model complexity and depth, and specifically between NLC and depth, allows us to construct arbitrarily deep networks with high performance. By using e.g. nlnorm (chapter \ref{nlnormChapter}), we can endow arbitrarily deep networks with a right-sized NLC. At least for fully-connected networks, controlling nonlinearity along with orthogonality (see below) is often sufficient for high performance. \citet{eigenspectrum,meanFieldCNN,meanFieldDeepField} construct arbitrarily deep, high-performing networks by controlling model complexity, and indirectly nonlinearity, by manipulating the weight initialization. \citet{meanFieldCNN} discuss additional requirements for achieving high performance with arbitrarily deep convolutional networks. The existence of these additional requirements is why we did not vary depth across our study B architectures.

\section{NLC and orthogonality} \label{orthogonalitySection}

In the final three sections of this chapter, we cover ZSAD guidelines that we argue are complementary to the guidelines we covered in earlier chapters, rather than more or less superseded by them. Hence, we also define them formally. In this section, we discuss `orthogonality', a term that was recently employed by \citet{orthogonalInitialization2}. Like them, we use it as a catch-all term that applies to various matrices.

\begin{npc} \label{npcOrthogonality}
(Background) `Ensure orthogonality' requires that the absolute singular values of characteristic matrices of the network should be relatively similar. These matrices include the Jacobian, Gram matrix, covariance matrix, Hessian and Fisher matrix.
\end{npc}

\citet{orthogonalInitialization2} focuses on the Jacobian, though the guideline has been investigated in terms of other matrices (Gram / covariance matrix: \citet{eigenspectrumGram,orthogonality1}; Hessian: \citet{eigenspectrumHessian}; Fisher matrix: \citet{eigenspectrumFisher,eigenspectrumFisher2}; weight tensor: \citet{orthogonality2}; activation matrix: \citet{bnOrthogonality}; neural tangent kernel: \citet{meanFieldNTK1}). It is important here to distinguish the {\it variation} of singular values from their {\it overall magnitude}. The latter falls into the realm of NLC, exploding / vanishing gradients and similar concepts, and we have discussed those above and throughout this work. Like for exploding / vanishing gradients, there is no agreed-upon metric for capturing eigenvalue variation. The concept of `dynamical isometry' introduced by \citet{orthogonalInitialization} comes closest. As far as we know, there is no concrete metric for the dynamical isometry of a network outside of mean field theory. However, taking the ratio of the quadratic mean of eigenvalues of $\mathcal{J}\mathcal{J}^T$ over the mean of eigenvalues of $\mathcal{J}\mathcal{J}^T$ comes close to how dynamical isometry is used e.g. in \citet{eigenspectrum}.

The primary method for achieving orthogonality at the network level is to ensure orthogonality for the linear transformations applied to the input during forward propagation. The original success story for this principle is the LSTM architecture \citep{LSTM}. In classical recurrent networks, to obtain the neuron values at a given time step, a full matrix multiplication is applied to some neuron values of the previous time step. Over multiple time steps, matrix exponentiation and hence de-orthogonalization occurs. LSTM's replace the full matrix multiplication with an elementwise multiplication, an operation known as `gating' in the context of recurrent networks. These elementwise multiplications are ``more orthogonal''. When weights are initialized advantageously \citep{orthLSTM}, the default behavior of gates is to leave the value of recurrent neurons almost unchanged from one time step to the next. Having the transformation of recurrent neurons be close to the identity prevents both de-orthogonalization and the accumulation of excessive nonlinearity. A large number of works following LSTM proposed different methods that can ensure (approximately) orthogonal time step transitions (e.g. \citet{fastfoodUnitaryRNN,strongTypedRNN,GRU,orthRNN} and many more), though not all make this desideratum explicit.

In the context of feedforward networks, many works on orthogonality come under the aforementioned umbrella of dynamical isometry, which focuses on the spectrum of the Jacobian. See \citet{orthogonalInitialization2} for an overview. Those works provide a large amount of evidence for the superiority of orthogonal initialization for fully-connected networks, and hence we use this initialization in our study A architectures (section \ref{studyAArchitecturesSection}). \citet{meanFieldCNN,fourierConvInit} investigated different types of orthogonal initialization for CNNs. Another line of work optimizes the spectrum of the covariance matrix directly with normalization layers that build upon batch normalization. This strategy is known as `whitening'. \citet{whitening1} is a seminal paper and \citet{whitening2} provides a recent overview.

The ZSAD guidelines of NLC and orthogonality are natural complements. The former is based on Jacobian magnitude; the latter is based on singular value variation. The former measures how much a network differs from a linear function; the latter measures what kind of linear function it is locally similar to. A network with a right-sized NLC can fail to perform due to lack of orthogonality. The most obvious case is when very deep networks use very linear activation functions but relatively non-orthogonal linear layers, such as Gaussian initialized fully-connected or convolutional layers \citep{eigenspectrum,meanFieldCNN}. As mentioned in the previous subsection, orthogonal linear layers are the second key ingredient for high-performing arbitrarily deep feedforward networks along with controlling nonlinearity.

In section \ref{meanFieldPracticalEmpiricalSection}, we showed that mean field theory can estimate the NLC even when linear layers are orthogonally initialized. Of course, mean field theory generally makes the theoretical assumption that linear layers are Gaussian initialized. Hence, orthogonality can be controlled to a significant degree without altering the NLC. The mean field theory of characteristic matrices has been developed \citep{meanFieldNetsorMatrix,meanFieldOrtho}.

In general, we suspect that feedforward architectures (i) with a right-sized NLC, (ii) with orthogonally initialized linear transformations and (iii) built using popular design strategies and building blocks are unlikely to take a performance hit due to lack of orthogonality.

\section{NLC and pseudo-linearity} \label{pseudoLinearitySection}

In one of our prior works \citep{expl}, we formulated the ZSAD guideline of avoiding `pseudo-linearity'.

\begin{npc} \label{npcPseudoLinearity}
(Background) `Avoid pseudo-linearity' requires that the network does not contain an activation layer that is effectively identical to a linear function across the inputs it receives from its parent in the layer graph.
\end{npc}

For example, when the tanh activation function is applied to inputs of very small magnitude, it is effectively identical to the identity function. When the ReLU activation function is applied to only positive inputs, it is identical to the identity function. When ReLU is applied to only negative inputs, it is identical to the zero function. Pseudo-linearity can make an activation layer unnecessary, especially when the preceding and following layer are fully-connected layers. Then it is possible to combine all three layers into a single FC layer without changing the set of functions expressible by the architecture and also potentially improving trainability, for example due to improved orthogonality \citep{orthogonalInitialization}, as well as reducing e.g. computational complexity and memory footprint. 

We bring up pseudo-linearity in this chapter because it relates to several earlier discussions. First, the nonlinearity path equation dictates that, for the mean field NLC to be stable as depth and the number of activation layers increases, most activation function NLCs of activation functions contained in paths with significant overall weight must converge to 1. Therefore, we are faced with a large amount of pseudo-linearity in the infinite depth limit. This calls into question the efficacy of increasing depth indefinitely. Indeed, all works that have increased architecture depth in a well-regulated manner as described in section \ref{depthSection} have found that test error converges to some ``imperfect limit'', rather than converging to zero or improving indefinitely. However, as far as we know, the statement that the distribution over network functions induced by a parameter initialization scheme converges in the finite-width, finite-nonlinearity, infinite-depth limit has thus far not been proven. This is an interesting topic for future work.

Second, pseudo-linearity arises specifically in the ``exponential LBIAS'' case of the trichotomy of section \ref{plainArchitectureSection}, which corresponds to order (section \ref{orderChaosSection}) and exponentially shrinking gradients (section \ref{vanishWhyBadSection}). There, the activation function NLC converges exponentially to 1 with depth. This is a reason why an exponential LBIAS is undesirable even if the damage is mitigated as in figures \ref{beyondFirst} and \ref{beyondDGD}.

Third, we observe that the risk of pseudo-linearity is minimized when the $\mathfrak{n}_\tau$ terms are approximately constant throughout the network. This is achieved by the ``exponential NLC'' case of the trichotomy of section \ref{plainArchitectureSection}, which corresponds to chaos (section \ref{chaosGoodSection}) and exponentially growing gradients (section \ref{vanishWhyBadSection}). In popular residual architectures, while $\mathfrak{n}_\tau$ terms may be stable across residual units, the mean field NLC of the residual units themselves tends to diminish with depth, as explained in section \ref{skipConnectionsSection}. This may be an argument against using skip connections and in favor of e.g. using a single dependency chain together with nlnorm. More work is necessary to investigate this point.

Fourth, pseudo-linearity is related to scale stability. As we discussed in section \ref{forwardStabilitySection}, one of the motivations for scale stability is that activation functions are designed with inputs of a certain magnitude in mind. When this magnitude changes, one potential negative consequence is pseudo-linearity. In tables \ref{covCurveillu1} through \ref{covCurveillu4}, we find that \finding{for activation functions like tanh, sigmoid, softplus and Swish, we have $\lim_{\lambda \rightarrow 0}\mathfrak{n}_\tau(\lambda^2, 0) = 1$}. In fact, this holds true for all activation functions that have a valid, non-zero derivative at zero. In that case, small inputs induce pseudo-linearity.

\section{NLC and ``use an appropriate width / parameter dimensionality''} \label{widthSection}

We discussed the notion of width in detail in e.g. sections \ref{layersSection}, \ref{architectureDesignParadigmsSection}, \ref{layerSurveySection} and \ref{linearLayerSurveySection}, and throughout this work. Mostly, we have discussed width as a basic layer property as well as a practical design strategy that has been empirically successful. We outlined the computational and representational benefits, as well as the predictability benefits, in section \ref{layerSurveySection}. In contrast, when we reference the ZSAD guideline ``use an appropriate width'', such as in section \ref{darkArtSection}, we refer to the principled investigation of which kinds of widths are optimal for performance; and to the general belief that a certain (large) width is a key to success with neural networks.

Of course, just as with depth, the concept of width suffers from ambiguity. Should the width of an architecture be considered the width of the widest layer, the width of the narrowest layer, the average width of layers, etc.? There does not exist an agreed-upon definition. Also, what are the potential benefits of non-constant width \citep{pyramidal}? It is generally believed that if a layer is a bottleneck for the output, then a small width can lead to ``information loss'', but it is also believed that individual narrow layers can reduce overfitting.

Width as a ZSAD guideline is highly related to the notion of parameter dimensionality. Historically, this quantity was the primary measure for model complexity in the field of statistics, and the primary predictor of overfitting and underfitting. It has been extensively investigated outside of deep learning. In neural networks based on linear layers, parameter dimensionality is approximately a quadratic function of width, as we also explained in section \ref{linearLayerSurveySection}. Hence, both concepts are treated as somewhat interchangeable by us and by the community in general. Parameter dimensionality is also largely synonymous with `model size' and `capacity'. However, the role of parameter dimensionality in deep learning is very different than in classical statistical models. For example, a very high-dimensional parameter does not generally lead to overfitting. This difference in behavior has long been seen as one of the most intriguing mysteries of deep learning.

Parameter dimensionality is considered a sufficiently important driver of performance that many deep learning studies control it when comparing the performance of different architecture types. We also ensure that parameter dimensionality does not vary significantly across our study A architectures, as well as across our study B architectures.

We are not aware of any works in the deep learning field proposing principled methods that yield concrete, ideal width values for a wide range of architecture types and tasks. Usually, the deep learning community has observed that ``bigger is better'', i.e. increasing width generally reduces test error, or at least does not increase it. ``Use an appropriate width'' then becomes ``maximize width given computational constraints''. To our knowledge, the first line of investigation that explicitly advocates that increasing width can lead to reduced test error is based on the recent concept of `double descent' \citep{width1,width5} / `triple descent' \citep{width4}. At a very high level, their argument goes as follows: ``At low width, the architecture can only absorb the most salient information from the training set which is likely to generalize to the test set. At higher width, it may also absorb some of the noise in the training set which may reduce generalization.'' This is similar to the classical argument for limiting parameter dimensionality. It remains to be seen whether the community finds this observation to hold across a range of practical tasks. Another concept that may call a large width into question is the neural tangent kernel (\citet{neuralTangentKernel}; see sections \ref{nlcKernelSection} and \ref{meanFieldNTKsection}). NTK theory predicts that in the infinite width limit, only the last linear layer learns. This phenomenon is termed `lazy training'. However, practical architectures often exceed the test error obtained from lazy training even at very high width \citep{ntkDoesntWork,ntkDoesntWork2,ntkDoesntWork3, meanFieldCNNGP,meanFieldKernelPerformance1}.

Recently, there has been a very large amount of work on gradually removing neurons from architectures to improve computational properties while not hurting test error too much. This strategy is known as `pruning'. Many works suffer from the fact that they can only reduce width after or during training \citep{groupSparseCNN,groupSparseCNN2,taylorPruning, pruneAndRewind,pruneFinal1}, which means that these methods do not help significantly in architecture design, because the layer widths used to start training still have to be chosen. Recently, more works have focused on pruning neurons in the initial state \citep{pruneInit1,pruneInit2,pruneInit3,pruneInit4,pruneInit5}. One of our prior works \citep{NNN}, as well as \citet{width3}, starts with a narrow architecture and adds neurons during training.

The remarkable relationship between the NLC and width is the seeming {\it absence} of a relationship. In the mean field limit of infinite width, the NLC converges, as we prove at least for fairly general fully-connected architectures in section \ref{meanFieldPracticalSection}. We demonstrate the robustness of the NLC to width change empirically in section \ref{widthInvarianceSection}. Hence, it is almost possible to choose width and NLC independently when designing an architecture. The same is true for width and other metrics with a mean field limit, such as GLEN, LCORR, LSCALE, LBIAS and LCV.

This reveals an intriguing fact. While NLC is a measure of model complexity for neural architectures as well as a predictor of underfitting and overfitting, it can be controlled independently of width, which corresponds to the classical measure of model complexity. Hence, there emerge two relatively independent notions of model complexity in deep learning. One is based on the complexity of the kind of functions expressed by the architecture; the other is based on how many different functions it can represent. It would be very interesting to study specifically the differences, similarities and interactions of these two axes of model complexity. We suspect that width is somewhat more responsible for reducing training error, whereas nonlinearity is somewhat more responsible for reducing the training error - test error gap. We suspect that a limited NLC is the regularizer required to prevent overfitting for very large widths.

A first step in this investigation may be the recent work of \citet{widthCapacityMetric}. They show that a gradient update based on some datapoint for an architecture that is {\it both} wide {\it and} nonlinear tends to not corrupt the architecture's output for another input of a different class.

We formalize the ZSAD guideline as follows.

\begin{npc} \label{npcWidth}
(Background) `Use an appropriate width / parameter dimensionality' requires that the network (i) has sufficient capacity to absorb information from the dataset, (ii) is sufficiently wide to prevent harmful information loss and (iii) does not have so much capacity and does not retain so much information that it overfits.
\end{npc}

The vagueness of the above guideline underscores the criticisms from the beginning of this chapter. However, it does have significant value because parameter dimensionality is well-defined. Hence, when varying the width of all layers jointly, we obtain a single width / dimensionality hyperparameter that is known to be key for performance. Crucially, it can be tuned relatively independently of the NLC.

\chapter{General nonlinearity theory} \label{finiteNetChapter}

In this chapter, we prove the theorems and propositions stated in chapter \ref{nlcChapter} about neural networks and input distributions. Our theory builds upon the functional-gradient paradigm detailed in section \ref{functionalGradientSection}. We consider neural networks simply as functions from $\mathbb{R}^{d_\text{in}}$ to $\mathbb{R}^{d_\text{out}}$ for fixed positive integers $d_\text{in}$ and $d_\text{out}$ in accordance with section \ref{neuralNetworkNotationSection}. However, in line with figure \ref{boxFunctionalLearning}, we do not impose a layer structure on neural networks as in section \ref{neuralNetworkNotationSection}. $f$ is completely ``black-box'' and therefore not dependent on any specific architecture design strategy. Further, we consider input distributions $\mathcal{D}$ simply as distributions over vectors in $\mathbb{R}^{d_\text{in}}$. Assumptions made about $f$ and $\mathcal{D}$ tend to be both mild and practical. The most restrictive assumption made by the chapter's theorems is that $\mathcal{D}$ is Gaussian.

\section{Notation, terminology and conventions} \label{finiteNetDefinitionsSection}

Below, we give notation, terminology and conventions that apply throughout this chapter and only this chapter, unless otherwise specified. Anything given in earlier chapters, including chapter \ref{backgroundChapter}, that is not repeated in the list below is not applicable to this chapter. For example, the letters $g$, $k$ and $K$ are re-assigned, and $f$ with a subscript does not refer to a layer.

\begin{itemize}
\item $\mathcal{D}$: input distribution over vectors in $\mathbb{R}^{d_\text{in}}$
\item $x$ , $x'$, $x''$ or $x_1$ etc.: an input distributed according to $\mathcal{D}$
\item $\chi$: a vector of dimensionality $d_\text{in}+1$ for which the first $d_\text{in}$ components are equal to $x$ and the last component is 1
\item $\mathbb{E}_x\mathsf{expr}$ or $\mathbb{E}\mathsf{expr}$ or $\overline{\mathsf{expr}}$: short for $\mathbb{E}_{x\sim\mathcal{D}}\mathsf{expr}$
\item $\mu_x$: the measure associated with $x$ under $\mathcal{D}$
\item $d_x$: the probability density function of $x$ when / where it exists
\item $\mu_1$: the canonical Lebesgue measure
\item $\mathbb{E}_{x\in S}$: the expectation over $x$, where $x$ is drawn from $\mathcal{D}$ restricted to a set $S$
\item $f(x): \mathbb{R}^{d_\text{in}} \rightarrow \mathbb{R}^{d_\text{out}}$ : neural network
\item $n$: the probability density function of the 1-dimensional unit Gaussian
\item $N$: the cumulative distribution function of the 1-dimensional unit Gaussian
\item $\mathcal{N}(.,.)$: a Gaussian distribution where the first argument is the mean and the second argument is the covariance
\item $i$, $i'$, $i''$, etc. :  input component index with $0\le i < d_\text{in}$
\item $j$, $j'$, $j''$ etc. : output component index with $0\le j < d_\text{out}$
\item $\mathcal{J}$: the Jacobian $\frac{df}{dx}$
\item $\Cov_\mathsf{vec} =  \mathbb{E}_x(\mathsf{vec} - \overline{\mathsf{vec}})^T(\mathsf{vec} - \overline{\mathsf{vec}})$: covariance matrix with respect to $\mathcal{D}$
\item $'$ superscript following a scalar function: the derivative
\item $[]$: square brackets used for tensor indexing and denoting closed intervals
\item NLC: see section \ref{nlcDefinitionSection}
\item $A_f$, $b_f$, $\tilde{f}$, LAR, CAR, NAR, `linear component', `constant component', and `nonlinear component' are defined as in section \ref{nlcLinearApproximationSection}. In general, when $A$ or $b$ are used with a subscript that is a function, or a tilde is placed above a function, then those expressions are defined analogously to $A_f$, $b_f$, $\tilde{f}$.
\item If $d_\text{in}=d_\text{out}=1$, let $f_K$ be the piecewise linear approximation of $f$ on some interval $[S,T]$ with $K$ equally long segments that is equal to $f$ at each knot point. That is, we have $f(S+\frac{k}{K}T)=f_K(S+\frac{k}{K}T)$ for all $0 \le k \le K$ and $f_K$ is linear between $S+\frac{k}{K}T$ and $S+\frac{k+1}{K}T$ for all $0\le k < K$.
\item Vectors defined in this chapter are row vectors by default. Jacobians have size $d_\text{out} \times d_\text{in}$.
\item In general, we omit function inputs when they are clear from context.
\item Except for theorem \ref{finiteNetNlcGreater1}, the proof of a result does not depend on a result given after it.
\end{itemize}

\section{General assumptions} \label{finiteNetAssumptionsSection}

Below, we give a few general assumptions about $f$ and $\mathcal{D}$ that we make throughout this chapter. Three of them are identical to assumptions given in section \ref{nlcDefinitionSection}. The fourth is an extended version of the corresponding assumption from section \ref{nlcDefinitionSection}. These assumptions are necessary to maintain readability, to avoid getting bogged down in impractical corner cases and to keep the length of this chapter bounded. Further conditions are given in individual propositions, lemmas and theorems.

\begin{repassumption}{assumptionDifferentiable}
$f$ is differentiable everywhere.
\end{repassumption}

In practice, neural networks are often not differentiable everywhere. This assumption is nonetheless reasonable, as we argue in section \ref{nonDifferentiableSection}.

The most common type of non-everywhere-differentiable network is differentiable almost everywhere and directionally differentiable everywhere. This happens most commonly when a directionally differentiable activation function is used, e.g. ReLU. We suspect that it is somewhat easy to extend our proofs to the directionally differentiable case. In fact, we designed our proofs specifically to be extensible in this way. For example, lemma \ref{lemma7} is not actually necessary for differentiable $f$. Also, as explained in section \ref{nonDifferentiableSection}, in practice, we can generally replace a non-everywhere-differentiable activation function with an everywhere-differentiable surrogate that is practically equivalent.

\begin{repassumption}{assumptionPositive}
$\Tr(\Cov_f) > 0$
\end{repassumption}

This assumption is very mild, as shown by proposition \ref{finiteNetPositiveDenominator}. Note that this also implies $\mathbb{E}||f||_2^2 > 0$ as $\mathbb{E}||f||_2^2 = \mathbb{E}||f-\bar{f}||_2^2 + ||\bar{f}||_2^2 =  \Tr(\Cov_f) + ||\bar{f}||_2^2$

\begin{repassumption}{assumptionNonSingular}
$\Cov_x$ is non-singular.
\end{repassumption}

This assumption is also mild, as shown by proposition \ref{finiteNetNonSingular}. Given any practical input distribution with singular covariance, we can simply attain non-singularity via linear projection / change of coordinates.

\begin{assumption} \label{assumptionIntegrableBig}
Simple expressions involving $x$, $f$ and $\mathcal{J}$ are integrable with respect to $\mathcal{D}$, where we use the term ``integrable'' as defined in section \ref{integrabilitySection}. For example, expressions like $x$, $x^Tx$, $f$, $x^Tf$, $f^Tf$, $\mathcal{J}$ and $\mathcal{J}\Cov_x\mathcal{J}^T$ are assumed to be integrable. We assume integrability holds in all reasonable subspaces of $\mathbb{R}^{d_\text{in}}$ such as lines and hyperplanes and that any multi-dimensional integral can be broken down into a sequence of 1-dimensional integrals. We assume that simple expressions involving $f$ and $\mathcal{J}$, when multiplied with the density function $d_x$, converge to zero along any direction. We assume that differentiation of $f$ can be exchanged with integration of simple expressions involving $f$, which means those integrals are themselves differentiable. We assume that this assumption itself holds for functions simply derived from $f$, such as $\tilde{f}$ and the $F_i$ in stage 4 of the proof of theorem \ref{finiteNetTilde}.
\end{assumption}

This assumption is also mild, as we argue in section \ref{integrabilitySection}. We can generally consider practical input distributions to be bounded in some fashion. Gaussian distributions are very light-tailed. Since $f$ is assumed to be differentiable, it is integrable over any bounded set with respect to the canonical Lebesgue measure.

Finally, note that we do not explicitly reference the case where $f$ contains batch normalization. In that case, we can simply regard the entire batch as a single input $x$, $\mathcal{D}$ as a distribution over batches, and $f$ as a function that returns outputs jointly for all individual inputs in a batch. See also section \ref{metricBNsection} for further context on this.

\section{Lemmas}

\begin{lemma} \label{lemma1}
Let $\chi$ be defined as in section \ref{finiteNetDefinitionsSection}. Then $\mathbb{E}\chi^T\chi$ is non-singular.
\end{lemma}

\begin{proof}
Let $\nu$ be an arbitrary non-zero vector of dimensionality $d_\text{in}+1$, let $v$ be that vector restricted to its first $d_\text{in}$ components and let $e$ be the last component of $\nu$.

{\it Case 1:} $v=0$. Then we have $e \neq 0$ and $\nu(\mathbb{E}\chi^T\chi)\nu^T = e^2 > 0$. 

{\it Case 2:} $v \neq 0$. We have

\begin{eqnarray*}
&&v\Cov_xv^T\\
&=&v(\mathbb{E}(x - \bar{x})^T(x - \bar{x}))v^T\\
&=&\mathbb{E}(v(x - \bar{x})^T)^2\\
&\ge & 0
\end{eqnarray*}

So $\Cov_x$ corresponds to a positive semi-definite quadratic form, so its eigenvalues are non-negative. Since $\Cov_x$ is also non-singular by assumption \ref{assumptionNonSingular}, its eigenvalues are positive. Hence, it corresponds to a positive definite quadratic form, so $v\Cov_xv^T > 0$. We have

\begin{eqnarray*}
&&\nu(\mathbb{E}\chi^T\chi)\nu^T\\
&=&\nu(\mathbb{E}(\chi-\bar{\chi})^T(\chi-\bar{\chi}) + \bar{\chi}^T\bar{\chi})\nu^T\\
&=&\nu(\mathbb{E}(\chi-\bar{\chi})^T(\chi-\bar{\chi}))\nu^T + \nu(\bar{\chi}^T\bar{\chi})\nu^T\\
&=&v\Cov_xv^T + (\nu\bar{\chi}^T)^2\\
&\ge &v\Cov_xv^T\\
&>& 0
\end{eqnarray*}

So in either case we have $\nu(\mathbb{E}\chi^T\chi)\nu^T > 0$, so $\mathbb{E}\chi^T\chi$ corresponds to a positive definite quadratic form, so it is non-singular.

\end{proof}

\begin{lemma}\label{lemma2}
Let $\bar{x}=0$ and let $\mathcal{D}$ be Gaussian. Then $A_f^T=\mathbb{E} \mathcal{J}$ and $b_f=\mathbb{E} f$.
\end{lemma}

\begin{proof}

Denote by $d_i$ the conditional density function of $x[i]$ given $x[i'], i'\neq i$. Then we have $d_i(x[i]) = ce^{-\frac{1}{2}(x-\bar{x})\Cov_x^{-1}(x-\bar{x})^T} = ce^{-\frac{1}{2}x\Cov_x^{-1}x^T}$ for some normalization constant $c$, so we have $\frac{dd_i}{dx[i]} = -\sum_{i''}\Cov_x^{-1}[i,i'']x[i'']d_i$. $\Cov_x^{-1}$ is valid by assumption \ref{assumptionNonSingular}. We have

\begin{eqnarray*}
&&\Big(\mathbb{E} \mathcal{J}\Big)[j,i]\\
&=&\mathbb{E}_{x[i'],i'\neq i} \mathbb{E}_{x[i]} \frac{df[j]}{dx[i]}\\
&=&\mathbb{E}_{x[i'],i'\neq i}\int_{x[i]=-\infty}^\infty \frac{df[j]}{dx[i]} d\mu_x\\
&=&\mathbb{E}_{x[i'],i'\neq i}\int_{x[i]=-\infty}^\infty \frac{df[j]}{dx[i]} d_id\mu_1\\
&=&\mathbb{E}_{x[i'],i'\neq i}\Big([f[j]d_i]_{x[i]=-\infty}^\infty - \int_{x[i]=-\infty}^\infty \frac{dd_i}{dx[i]}f[j]d\mu_1\Big)\\
&=&\mathbb{E}_{x[i'],i'\neq i}\Big([f[j]d_i]_{x[i]=-\infty}^\infty + \int_{x[i]=-\infty}^\infty \sum_{i''}\Cov_x^{-1}[i,i'']x[i'']f[j] d_id\mu_1\Big)\\
&=&\mathbb{E}\Big(\sum_{i''}\Cov_x^{-1}[i,i'']x[i'']f[j]\Big)\\
&=&\mathbb{E}\Big(\Cov_x^{-1}x^Tf\Big)^T[j,i]
\end{eqnarray*}

Here, we use integration by parts on a 1-dimensional subspace. This is allowed because $d_i$ is absolutely continuous and $\frac{df[j]}{dx[i]}$ is integrable. 

The standard formula for the least squares linear fit yields $\begin{pmatrix}A_f\\b_f\end{pmatrix} = (\mathbb{E}\chi^T\chi)^{-1}\mathbb{E}\chi^Tf$, where $(\mathbb{E}\chi^T\chi)^{-1}$ is valid by lemma \ref{lemma1}. Using $\bar{x} = 0$ we have

\begin{eqnarray*}
&&\begin{pmatrix}A_f\\b_f\end{pmatrix}\\
&=&(\mathbb{E}\chi^T\chi)^{-1}(\mathbb{E}\chi^Tf)\\
&=&\begin{pmatrix}\mathbb{E}x^Tx & \mathbb{E}x^T \\ \mathbb{E}x& 1 \end{pmatrix}^{-1}\begin{pmatrix}\mathbb{E}x^Tf\\\mathbb{E}f\end{pmatrix}\\
&=&\begin{pmatrix}\Cov_x & 0 \\ 0 & 1 \end{pmatrix}^{-1}\begin{pmatrix}\mathbb{E}x^Tf\\\mathbb{E}f\end{pmatrix}\\
&=&\begin{pmatrix}\Cov_x^{-1} & 0 \\ 0 & 1 \end{pmatrix}\begin{pmatrix}\mathbb{E}x^Tf\\\mathbb{E}f\end{pmatrix}\\
&=&\begin{pmatrix}\Cov_x^{-1}\mathbb{E}x^Tf\\\mathbb{E}f\end{pmatrix}
\end{eqnarray*}

Hence $b_f=\mathbb{E}f$ and $A_f=\Cov_x^{-1}\mathbb{E}x^Tf$, and hence $A_f^T=\mathbb{E}\mathcal{J}$.

\end{proof}

\begin{lemma} \label{lemma3}
Let $\mathbb{E}x^Tf = 0$ and $\mathbb{E}f = 0$. Then $A_f = 0$ and $b_f = 0$.
\end{lemma}

\begin{proof}
The standard formula for the least squares linear fit yields $\begin{pmatrix}A_f\\b_f\end{pmatrix} = (\mathbb{E}\chi^T\chi)^{-1}\mathbb{E}\chi^Tf=(\mathbb{E}\chi^T\chi)^{-1}\begin{pmatrix}\mathbb{E}x^Tf\\\mathbb{E}f\end{pmatrix}=0$, where $(\mathbb{E}\chi^T\chi)^{-1}$ is valid by lemma \ref{lemma1}.
\end{proof}

\begin{lemma} \label{lemma4}
Let $d_\text{in}=d_\text{out}=1$ and let $\mathcal{D}$ be Gaussian. Let $f_K$ be the piecewise linear approximation of $f$ as defined in section \ref{finiteNetDefinitionsSection} with respect to some interval $[S,T]$. Then for each $\epsilon$ there exists a $K_\epsilon$ such that for all $K>K_\epsilon$ we have $|\mathbb{E}_{x\in[S,T]}f_K-\mathbb{E}_{x\in[S,T]}f|<\epsilon$, $|\mathbb{E}_{x\in[S,T]}f_K^2-\mathbb{E}_{x\in[S,T]}f^2|<\epsilon$, $\mathbb{E}_{x\in[S,T]}f'^2_K-\mathbb{E}_{x\in[S,T]}f'^2<\epsilon$ and $|\mathbb{E}_{x\in[S,T]}xf_K-\mathbb{E}_{x\in[S,T]}xf|<\epsilon$.
\end{lemma}

\begin{proof}

In this proof only, we write $\mathbb{E}$ for $\mathbb{E}_{x\in[S,T]}$. Let $\delta=\frac{T-S}{K}$, let $k$ range from $0$ to $K-1$ and let $x_k=S+k\delta$. Let $\mathcal{T}$ denote the distribution $\mathcal{D}$ restricted to $[S,T]$, let $t(x)$ denote the probability density function of $\mathcal{T}$ and let $t_k=\int_{x_k}^{x_{k+1}}t(x)d\mu_1$. (Keep in mind that expressions like $x_k$ and $t_k$ depend on $K$. The value of $K$, as opposed to $S$ and $T$, is not fixed throughout this proof.) We say that a function `covers' a value on an interval if that value is neither a strict upper bound nor a strict lower bound for the function on the interval.

Let $c_k=\mathbb{E}_{x \in [x_k,x_{k+1}]}f_K$. We have $\mathbb{E}f_K=\sum_k t_kc_k$. Since $f_K$ is linear on $[x_k,x_{k+1}]$, $c_k$ lies between $f_K(x_k) = f(x_k)$ and $f_K(x_{k+1}) = f(x_{k+1})$, and so $c_k$ is covered by $f$ on $[x_k,x_{k+1}]$. Since $f$ is integrable over $\mathcal{T}$, $\lim_{K\rightarrow \infty}\sum_k t_kc_k$ is a valid definition of $\mathbb{E}f$. Hence we have $\lim_{K\rightarrow \infty}\sum_k t_kc_k =\lim_{K\rightarrow \infty}\mathbb{E}f_K = \mathbb{E}f$ and therefore for each $\epsilon_1$ there is a $K_1$ such that for all $K > K_1$ we have $|\mathbb{E}f_K-\mathbb{E}f| < \epsilon_1$.

Let $s_k=\mathbb{E}_{x \in [x_k,x_{k+1}]}f_K^2$. We have $\mathbb{E}f_K^2=\sum_k t_ks_k$. Case 1: Assume $f_K(x_k)$ and $f_K(x_{k+1})$ have the same sign. Since $f_K$ is linear on $[x_k,x_{k+1}]$, $f_K^2$ is monotonic on $[x_k,x_{k+1}]$. So $s_k$ lies between $f_K(x_k)^2 = f(x_k)^2$ and $f_K(x_{k+1})^2 = f(x_{k+1})^2$, and so $s_k$ is covered by $f^2$ on $[x_k,x_{k+1}]$. Case 2: Assume $f_K(x_k)$ and $f_K(x_{k+1})$ don't have the same sign. Since $f_K$ is linear on $[x_k,x_{k+1}]$, $s_k$ lies between 0 and  $\max(f_K(x_k)^2,f_K(x_{k+1})^2) = \max(f(x_k)^2,f(x_{k+1})^2)$. Since $f$ is continuous, it attains all values between $f(x_k)$ and $f(x_{k+1})$ on $[x_k,x_{k+1}]$, so $f^2$ attains both $0$ and $\max(f(x_k)^2,f(x_{k+1})^2)$, so $s_k$ is covered by $f^2$ on $[x_k,x_{k+1}]$. So in both case 1 and case 2, $s_k$ is covered by $f^2$ on $[x_k,x_{k+1}]$. Since $f^2$ is integrable over $\mathcal{T}$, $\lim_{K\rightarrow \infty}\sum_k t_ks_k$ is a valid definition of $\mathbb{E}f^2$. Hence we have $\lim_{K\rightarrow \infty}\sum_k t_ks_k = \lim_{K\rightarrow \infty}\mathbb{E}f_K^2= \mathbb{E}f^2$ and therefore for each $\epsilon_2$ there is a $K_2$ such that for all $K > K_2$ we have $|\mathbb{E}f_K^2-\mathbb{E}f^2| < \epsilon_2$.

Let $g_k=\mathbb{E}_{x \in [x_k,x_{k+1}]}f'^2_K(x)=(\frac{f(x_{k+1})-f(x_k)}{\delta})^2$. We have $\mathbb{E}f'^2_K=\sum_k t_kg_k$. By the intermediate value principle, $\frac{f(x_{k+1})-f(x_k)}{\delta}$ is covered by $f'$ on $[x_k,x_{k+1}]$. Therefore $(\frac{f(x_{k+1})-f(x_k)}{\delta})^2$ is not a strict upper bound for $f'^2$ on $[x_k,x_{k+1}]$. Let $\gamma_k$ be a value attained by $f'^2$ on $[x_k,x_{k+1}]$ that is larger than or equal to $g_k$. Since $f'^2$ is integrable over $\mathcal{T}$, $\lim_{K\rightarrow \infty}\sum_k t_k\gamma_k$ is a valid definition of $\mathbb{E}f'^2$. So $\sum_k t_kg_k = \mathbb{E}f'^2_K$ as a sequence in $K$ is less than or equal to a sequence that converges to $\mathbb{E}f'^2$. So for each $\epsilon_3$ there is a $K_3$ such that for all $K > K_3$ we have $\mathbb{E}f'^2_K - \mathbb{E}f'^2< \epsilon_3$.

Let $l_k=\mathbb{E}_{x \in [x_k,x_{k+1}]}xf_K(x)$. We have $\mathbb{E}xf_K=\sum_k t_kl_k$. Let $r_k=\midOp(x_kf_K(x_k)-l_k,x_{k+1}f_K(x_{k+1})-l_k,0)$, where $\midOp(.,.,.)$ denotes the middle of three values. Since $f_K(x)$ is linear on $[x_k,x_{k+1}]$, we can denote it as $a_kx+b_k$ there. $xf_K(x)$ is then a quadratic function on $[x_k,x_{k+1}]$. Case 1: Assume $xf_K(x)$ is monotonic on $[x_k,x_{k+1}]$. Then $l_k$ lies between $x_kf_K(x_k)$ and $x_{k+1}f_K(x_{k+1})$, and so $r_k = 0$. Case 2: Assume $xf_K(x)$ is not monotonic on $[x_k,x_{k+1}]$. Then it has a single critical point $x^*$ in $[x_k,x_{k+1}]$, where $x^*=-\frac{b_k}{2a_k}$. Then $|x_kf_K(x_k) - x^*f_K(x^*)| = |a_kx_k^2 + b_kx_k - a_k(-\frac{b_k}{2a_k})^2 - b_k(-\frac{b_k}{2a_k})| = |a_k(x + \frac{b_k}{2a_k})^2| = |a_k(x - x^*)^2| \le |a_k|\delta^2$. Similarly, $|x_{k+1}f_K(x_{k+1}) - x^*f_K(x^*)| \le |a_k|\delta^2$. So the value of $xf_K(x)$ at the critical point is not further than $|a_k|\delta^2$ away from either $x_kf_K(x_k)$ or $x_{k+1}f_K(x_{k+1})$. So $l_k$ is not further than $|a_k|\delta^2$ away from either $x_kf_K(x_k)$ or $x_{k+1}f_K(x_{k+1})$. So $|r_k| < |a_k|\delta^2$. So in either case 1 or case 2, $|r_k| < |a_k|\delta^2$.

$a_k$ is the slope of $f_K(x)$ on $[x_k,x_{k+1}]$, so $|a_k| = \sqrt{g_k}$, where $g_k$ is defined as above. So $|\sum_k t_kr_k| < \delta^2 \sum_k \sqrt{g_k}t_k$. We have shown above that $\mathbb{E}f'^2_K=\sum_k g_kt_k$ is bounded for large enough $K$, so $\sqrt{\sum_k g_kt_k}$ is bounded for large enough $K$. Since $\sum_k t_k = 1$, by Jensen's inequality we have $\sqrt{\sum_k g_kt_k} \ge \sum_k \sqrt{g_k}t_k$, so $\sum_k \sqrt{g_k}t_k$ is bounded for large enough $K$, so $\delta^2 \sum_k \sqrt{g_k}t_k$ converges to zero as $K$ converges to infinity, so $|\sum_k t_kr_k|$ converges to zero, so $\sum_k t_kr_k$ converges to zero. $r_k+l_k = \midOp(x_kf_K(x_k),x_{k+1}f_K(x_{k+1}),l_k)$ is covered by $f$ on $[x_k,x_{k+1}]$, because $l_k$ is either covered or is not the middle value. Since $xf$ is integrable over $\mathcal{T}$, $\lim_{K\rightarrow \infty}\sum_k(r_k+l_k)t_k$ is a valid definition of $\mathbb{E}xf$. So $\sum_k(r_k+l_k)t_k$ as a sequence in $K$ converges to $\mathbb{E}xf$ and $\sum_kr_kt_k$ converges to zero, so $\sum_kl_kt_k=\mathbb{E}xf_K$ converges to $\mathbb{E}xf$. So for each $\epsilon_4$ there is a $K_4$ such that for all $K > K_4$ we have $|\mathbb{E}xf_K - \mathbb{E}xf|< \epsilon_4$.

Setting $\epsilon_1$, $\epsilon_2$, $\epsilon_3$ and $\epsilon_4$ to $\epsilon$ and setting $K_\epsilon=\max(K_1,K_2,K_3,K_4)$ yields the lemma.
\end{proof}

\begin{lemma}\label{lemma5}
Let $d_\text{in} = 1$ and let $h(x)=-xn(x)+(N(x)-N(0))(1-x^2)$. Then $h$ has the following properties:

\begin{enumerate}
\item $h' \le 0$ where equality holds if and only if $x=0$
\item $h$ is strictly decreasing
\item $h=0$ if $x=0$, $h<0$ if $x>0$ and $h>0$ if $x<0$
\item On any closed and bounded interval not containing zero, $h$ is bounded away from zero.
\item For all $c>0$, $h$ is bounded away from zero on $[c,\infty)$ and $(-\infty,-c]$.
\end{enumerate}

\end{lemma}

\begin{proof}
$h'(x)=-n(x)+x^2n(x)+n(x)(1-x^2)-2x(N(x)-N(0))=-2x(N(x)-N(0))$. As $N(x)-N(0)$ has the same sign as $x$, property 1 holds.

Since $h$ is differentiable and property 1 holds, $h$ is decreasing. Assume $h(y)=h(z)$ for some $y<z$. Because $h$ is decreasing, it is constant on $[y,z]$, so $h'=0$ on $[y,z]$. But by property 1, $h'=0$ holds only at one point. Contradiction. So $h(y)\neq h(z)$. So $h$ is strictly decreasing, hence property 2.

Property 3 follows directly from property 2 and $h(0)=0$. 

Consider any closed and bounded interval not containing zero. Then it is either above or below zero. If it is above zero, by property 2, $h$ attains its maximum at the lower bound of the interval and its minimum at the upper bound of the interval. By property 3, $h$ is negative at both those points. Hence, $h$ is bounded away from zero. Similarly, if the interval is below zero, $h$ is bounded in the positive reals. Hence, property 4.

By property 2, $h$ attains its maximum on $[c,\infty)$ at $c$. By property 3, that maximum is negative. Hence $h$ is bounded away from zero. Similarly, by property 2, $h$ attains its minimum on $(-\infty,-c]$ at $-c$. By property 3, that minimum is positive. Again, $h$ is bounded away from zero. Hence, property 5.
\end{proof}

\begin{lemma}\label{lemma6}
Let $d_\text{in} = 1$ and $H(x)=-3xn(x)+(N(x)-N(0))(3-x^2)$. Then $H$ has the following properties:

\begin{enumerate}
\item $H' \le 0$ where equality holds if and only if $x=0$
\item $H$ is strictly decreasing
\item $H=0$ if $x=0$, $H<0$ if $x>0$ and $H>0$ if $x<0$
\item On any closed and bounded interval not containing zero, $H$ is bounded away from zero.
\item For all $c>0$, $H$ is bounded away from zero on $[c,\infty)$ and $(-\infty,-c]$.
\end{enumerate}

\end{lemma}

\begin{proof}
We first investigate $G(x)=xn(x)-N(x)+N(0)$. We have $G'(x)=-x^2n(x)$. So $G$ is differentiable and has a non-positive derivative. Hence it is decreasing. Assume that $G(y)=G(z)$ for some $y<z$. Then because $G$ is decreasing, it is constant on $[y,z]$, so $G'=0$ on $[y,z]$. But $G'$ is only zero at one point. Contradiction. Hence $G(y)\neq G(z)$, so $G$ is strictly decreasing. Since $G(0)=0$, $G(x)$ has the opposite sign as $x$.

We have $H'(x)=-3n(x)+3x^2n(x)+n(x)(3-x^2)-2x(N(x)-N(0))=2x(xn(x)-N(x)+N(0))=2xG(x)$. Since $G(x)$ has the opposite sign as $x$, $2xG(x)\le0$ where equality holds if and only if $x=0$. We have property 1.

Since $H$ is differentiable and property 1 holds, $H$ is decreasing. Assume $H(y)=H(z)$ for some $y<z$. Because $H$ is decreasing, it is constant on $[y,z]$, so $H'=0$ on $[y,z]$. But by property 1, $H'=0$ holds only at one point. Contradiction. So $H(y)\neq H(z)$. So $H$ is strictly decreasing, hence property 2.

Property 3 follows directly from property 2 and $H(0)=0$. 

Consider any closed, bounded interval not containing zero. Then it is either above or below zero. If it is above zero, by property 2, $H$ attains its maximum at the lower bound of the interval and its minimum at the upper bound of the interval. By property 3, $H$ is negative at both those points. Hence, $H$ is bounded away from zero. Similarly, if the interval is below zero, $H$ is bounded in the positive reals. Hence, property 4.

By property 2, $H$ attains its maximum on $[c,\infty)$ at $c$. By property 3, that maximum is negative. Hence $H$ is bounded away from zero. Similarly, by property 2, $H$ attains its minimum on $(-\infty,-c]$ at $-c$. By property 3, that minimum is positive. Again, $H$ is bounded away from zero. Hence, property 5.
\end{proof}

\begin{lemma}\label{lemma7}
Let $d_\text{in}=d_\text{out}=1$ and let $\mathcal{D}$ be Gaussian. Then for all $\epsilon$ there exists a $\delta>0$ such that for all $0\le x<\delta$ we have $|f(x)-f(0)|<\epsilon\sqrt{x}$.
\end{lemma}

\begin{proof}
We argue by contradiction. Assume that there exists an $\epsilon$ for which there is no such $\delta$. Then there exists a decreasing sequence $\delta_k>0$ converging to zero such that $|f(\delta_k)-f(0)| \ge \epsilon\sqrt{\delta_k}$ for all $k$. Let $d_\text{min}$ be the minimum value that $d_x$ takes on $[0,\delta_0]$. Since $\mathcal{D}$ is Gaussian, this value is positive. We have

\begin{eqnarray*}
&&\int_0^{\delta_k}f'^2d\mu_x\\
&\ge&d_\text{min}\int_0^{\delta_k}f'^2d\mu_1\\
&\ge&d_\text{min}(\int_0^{\delta_k}f'd\mu_1)^2(\int_0^{\delta_k}d\mu_1)^{-1}\\
&=&d_\text{min}\frac{(f(\delta_k)-f(0))^2}{\delta_k}\\
&\ge&d_\text{min}\frac{(\epsilon\sqrt{\delta_k})^2}{\delta_k}\\
&=&d_\text{min}\epsilon^2
\end{eqnarray*}

Here, we use Jensen's inequality.

Because $f'^2$ is integrable with respect to $\mathcal{D}$, we have $\int_0^{\delta_0}f'^2d\mu_x=\lim_{k\rightarrow\infty}\int_{\delta_k}^{\delta_0}f'^2d\mu_x$, so $0=\lim_{k\rightarrow\infty}\int_0^{\delta_0}f'^2d\mu_x-\int_{\delta_k}^{\delta_0}f'^2d\mu_x=\lim_{k\rightarrow\infty}\int_0^{\delta_k}f'^2d\mu_x$. But $\int_0^{\delta_k}f'^2d\mu_x$ is bounded away from zero. Contradiction. Hence, we have the lemma.

\end{proof}

\section{Propositions}

\begin{repproposition}{finiteNetPositiveDenominator}
Assume there exists an open set $S$ where $\mathcal{D}$ has a continuous, positive density function and $f$ is not constant on $S$. Then $\Tr(\Cov_f)> 0$.
\end{repproposition}

\begin{proof} (For the purpose of this proof, naturally, we do not use assumption \ref{assumptionPositive}.) Since $f$ is non-constant on $S$, there are $x_1,x_2\in S$ with $f(x_1) \neq f(x_2)$. At least one of $f(x_1),f(x_2)$ is unequal to $\bar{f}$. WLOG, $f(x_1) \neq \bar{f}$. Because $d_x$ is continuous and positive on $S$, there exists an open set $S_1$ and $\epsilon_1 > 0$ such that $x_1 \in S_1$ and $d_x(x) > \epsilon_1$ for all $x \in S_1$. Because $f$ is continuous and $f(x_1) \neq \bar{f}$, there exists an open set $S_2$ and $\epsilon_2 > 0$ such that $x_1 \in S_2$ and $||f(x) - \bar{f}||_2^2  > \epsilon_2$ for all $x \in S_2$. Let $S_3 = S_1 \cap S_2$. Because $S_1$ and $S_2$ are open, so is $S_3$. Because $x_1 \in S_1$ and $x_1 \in S_2$, $S_3$ is non-empty, we have $\int_{S_3}d\mu_1 > 0$. So we have

\begin{eqnarray*}
&&\Tr(\Cov_f)\\
&=&\Tr(\mathbb{E}(f-\bar{f})^T(f-\bar{f}))\\
&=&\mathbb{E}||f-\bar{f}||_2^2\\
&=&\int_x||f-\bar{f}||_2^2d\mu_x\\
&\ge&\int_{x\in S_3}||f-\bar{f}||_2^2d\mu_x\\
&=&\int_{x\in S_3}||f-\bar{f}||_2^2d_x(x)d\mu_1\\
&\ge&\int_{x\in S_3}\epsilon_2\epsilon_1d\mu_1\\
&=&\epsilon_2\epsilon_1\int_{S_3}d\mu_1\\
&>&0
\end{eqnarray*}

\end{proof}

\begin{repproposition}{finiteNetNonSingular}
Assume there exists a non-empty open set $S$ where $\mathcal{D}$ has a continuous, positive density function. Then $\Cov_x$ is non-singular.
\end{repproposition}

\begin{proof}
(For the purpose of this proof, naturally, we do not use assumption \ref{assumptionNonSingular}.) Assume $\Cov_x$ is singular. Then there exists a non-zero vector $v \in \mathbb{R}^{d_\text{in}}$ such that $v\Cov_x = 0$. Let $F(x) = xv^T$. $F$ is not constant on any non-empty open set, so it is non-constant on $S$. Hence we can apply the proof of proposition \ref{finiteNetPositiveDenominator} to obtain $\Cov_F > 0$. We also have

\begin{eqnarray*}
&&\Cov_F\\
&=&\mathbb{E}(F-\bar{F})^T(F-\bar{F})\\
&=&\mathbb{E}F^TF - \bar{F}^T\bar{F}\\
&=&\mathbb{E}vx^Txv^T - v\bar{x}^T\bar{x}v^T\\
&=&v(\mathbb{E}x^Tx - \bar{x}^T\bar{x})v^T\\
&=&v(\mathbb{E}(x - \bar{x})^T(x - \bar{x}))v^T\\
&=&v\Cov_xv^T\\
&=&0
\end{eqnarray*}

This is a contradiction, which completes the proof.

\end{proof}

\begin{repproposition}{finiteNetNlcEquals1}
Let $f$ be linear. Then $NLC(f,\mathcal{D}) = 1$.
\end{repproposition}

\begin{proof}
If $f : \mathbb{R}^{d_\text{in}} \rightarrow \mathbb{R}^{d_\text{out}}$ is linear, we can write $f(x) = xA + b$ for a fixed matrix $A$ and vector $b$. Then we have

\begin{eqnarray*}
&&NLC(f,\mathcal{D})^2\\
&=&\frac{\mathbb{E}\Tr(\mathcal{J}(x)\Cov_x\mathcal{J}(x)^T)}{\Tr(\Cov_f)}\\
&=&\frac{\Tr(A^T\Cov_xA)}{\Tr(\mathbb{E}_x(xA+b-\mathbb{E}_{x'}(x'A+b))^T(xA+b-\mathbb{E}_{x'}(x'A+b)))}\\
&=&\frac{\Tr(A^T\Cov_xA)}{\Tr(\mathbb{E}(xA-\bar{x}A)^T(xA-\bar{x}A))}\\
&=&\frac{\Tr(A^T\Cov_xA)}{\Tr(\mathbb{E}A^T(x-\bar{x})^T(x-\bar{x})A)}\\
&=&\frac{\Tr(A^T\Cov_xA)}{\Tr(A^T\Cov_xA)}\\
&=&1
\end{eqnarray*}

\end{proof}

\begin{repproposition}{finiteNetNlcOrthoMatrix}
Let $A$ be an orthogonal matrix of size $d_\text{out} \times d_\text{out}$ and $b$ be a $d_\text{out}$-dimensional vector. Then $NLC(fA + b,\mathcal{D}) = NLC(f,\mathcal{D})$.
\end{repproposition}

\begin{proof}
\begin{eqnarray*}
&&NLC(fA+b,\mathcal{D})^2\\
&=&\frac{\mathbb{E}\Tr(A^T\mathcal{J}(x)\Cov_x\mathcal{J}^T(x)A)}{\mathbb{E}_x\Tr((f(x)A + b-\mathbb{E}_{x'}(f(x')A + b))^T(f(x)A + b-\mathbb{E}_{x'}(f(x')A + b)))}\\
&=&\frac{\mathbb{E}_{x,x'}\Tr(A^T\mathcal{J}(x)(x'-\bar{x})^T(x'-\bar{x})\mathcal{J}(x)^TA)}{\mathbb{E}\Tr(A^T(f(x) - \bar{f})^T(f(x) - \bar{f})A)}\\
&=&\frac{\mathbb{E}_{x,x'}||A^T\mathcal{J}(x)(x'-\bar{x})^T||^2_2}{\mathbb{E}||A^T(f(x) - \bar{f})||^2_2}\\
&=&\frac{\mathbb{E}_{x,x'}||\mathcal{J}(x)(x'-\bar{x})^T||^2_2}{\mathbb{E}||f(x) - \bar{f}||^2_2}\\
&=&\frac{\mathbb{E}_{x,x'}\Tr(\mathcal{J}(x)(x'-\bar{x})^T(x'-\bar{x})\mathcal{J}(x)^T)}{\mathbb{E}\Tr((f(x) - \bar{f})^T(f(x) - \bar{f}))}\\
&=&\frac{\mathbb{E}\Tr(\mathcal{J}(x)\Cov_x\mathcal{J}(x)^T)}{\Tr(\Cov_f)}\\
&=&NLC(f,\mathcal{D})^2
\end{eqnarray*}

\end{proof}

\begin{repproposition}{finiteNetNlcAnyMatrix}
Let $A$ be a matrix of size $d_\text{in} \times d_\text{in}$ and $b$ be a $d_\text{in}$-dimensional vector. Assume $\Tr(\Cov_{f(xA+b)}) > 0$. Then $NLC(f(xA+b),\mathcal{D}) = NLC(f(x),\mathcal{D}A+b)$, where drawing $x$ from $\mathcal{D}A+b$ is equivalent to drawing $x$ from $\mathcal{D}$ and applying $xA + b$.
\end{repproposition}

\begin{proof}

\begin{eqnarray*}
&&NLC(f(xA+b),\mathcal{D})^2\\
&=&\frac{\mathbb{E}_{x}\Tr(\frac{df(xA+b)}{dx}\Cov_x\frac{df(xA+b)}{dx}^T)}{\Tr(\Cov_{f(xA+b)})}\\
&=&\frac{\mathbb{E}_{x,x'}\Tr(\mathcal{J}(xA+b)A^T(x'-\bar{x})^T(x' - \bar{x})A\mathcal{J}(xA+b)^T)}{\Tr(\Cov_{f(xA+b)})}\\
&=&\frac{\mathbb{E}_{x,x'}\Tr(\mathcal{J}(xA+b)(x'A+b-\mathbb{E}_{x''}(x''A+b))^T(x'A+b - \mathbb{E}_{x''}(x''A+b))\mathcal{J}(xA+b)^T)}{\Tr(\mathbb{E}_x(f(xA+b)-\mathbb{E}_{x''}f(x''A+b))^T(f(xA+b)-\mathbb{E}_{x''}f(x''A+b)))}\\
&=&\frac{\mathbb{E}_{x,x'\sim \mathcal{D}A+b}\Tr(\mathcal{J}(x)(x'-\mathbb{E}_{x''\sim \mathcal{D}A+b}x'')^T(x'-\mathbb{E}_{x''\sim \mathcal{D}A+b}x'')\mathcal{J}(x)^T)}{\Tr(\mathbb{E}_{x\sim \mathcal{D}A+b}(f(x)-\mathbb{E}_{x''\sim \mathcal{D}A+b}f(x''))^T(f(x)-\mathbb{E}_{x''\sim \mathcal{D}A+b}f(x'')))}\\
&=&NLC(f(x),\mathcal{D}A+b)^2
\end{eqnarray*}

\end{proof}

\begin{repproposition}{finiteNetBasisOrtho}
We have
\begin{enumerate}
\item $\mathbb{E}F^T\tilde{f} = 0$ and $\mathbb{E}F\tilde{f}^T = 0$ for any linear function $F : \mathbb{R}^{d_\text{in}} \rightarrow \mathbb{R}^{d_\text{out}}$
\item $\mathbb{E}((x-\bar{x})A_f)\tilde{f}^T = 0$, $\mathbb{E}(b_f+\bar{x}A_f)\tilde{f}^T = 0$ and $\mathbb{E}((x-\bar{x})A_f)(b_f+\bar{x}A_f)^T = 0$
\item LAR + CAR + NAR = 1
\end{enumerate}
\end{repproposition}

\begin{proof}
Since $F$ is linear, it can be written as $xA + b$ for some $A$, $b$. Let $\mathcal{A}$ denote $\begin{pmatrix}A\\b\end{pmatrix}$. The standard formula for the least squares linear fit yields $\begin{pmatrix}A_f\\b_f\end{pmatrix} = (\mathbb{E}\chi^T\chi)^{-1}\mathbb{E}\chi^Tf$, where $(\mathbb{E}\chi^T\chi)^{-1}$ is valid by lemma \ref{lemma1}.  Then we have

\begin{eqnarray*}
&&\mathbb{E}F^T\tilde{f}\\
&=&\mathbb{E}(xA + b)^T(f-\chi\begin{pmatrix}A_f\\b_f\end{pmatrix})\\
&=&\mathbb{E}(\chi\mathcal{A})^T(f-\chi(\mathbb{E}\chi^T\chi)^{-1}(\mathbb{E}\chi^Tf))\\
&=&\mathbb{E}\mathcal{A}^T(\chi^Tf-\chi^T\chi(\mathbb{E}\chi^T\chi)^{-1}(\mathbb{E}\chi^Tf))\\
&=&\mathcal{A}^T((\mathbb{E}\chi^Tf)-(\mathbb{E}\chi^T\chi)(\mathbb{E}\chi^T\chi)^{-1}(\mathbb{E}\chi^Tf))\\
&=&0
\end{eqnarray*}

Since we have $F\tilde{f}^T = \Tr(F^T\tilde{f})$, we have $\mathbb{E}F\tilde{f}^T = 0$. Hence, property 1.

$(x-\bar{x})A_f$ is linear, so by property 1 we have $\mathbb{E}((x-\bar{x})A_f)\tilde{f}^T = 0$. $b_f+\bar{x}A_f$ is linear, so by property 1 we have $\mathbb{E}(b_f+\bar{x}A_f)\tilde{f}^T = 0$. Also we have $\mathbb{E}((x-\bar{x})A_f)(b_f+\bar{x}A_f)^T = ((\bar{x}-\bar{x})A_f)(b_f+\bar{x}A_f)^T = 0$. Hence, property 2. Finally, we have

\begin{eqnarray*}
&&1\\
&=&(\mathbb{E}||f||_2^2)^{-1}\mathbb{E}||f||_2^2\\
&=&(\mathbb{E}||f||_2^2)^{-1}\mathbb{E}ff^T\\
&=&(\mathbb{E}||f||_2^2)^{-1}\mathbb{E}(\tilde{f} + (x-\bar{x})A_f + (b_f + \bar{x}A_f))(\tilde{f} + (x-\bar{x})A_f + (b_f + \bar{x}A_f))^T\\
&=&(\mathbb{E}||f||_2^2)^{-1}\mathbb{E}\Big(\tilde{f}\tilde{f}^T + (x-\bar{x})A_f((x-\bar{x})A_f)^T + (b_f + \bar{x}A_f)(b_f + \bar{x}A_f)^T \\
&&+ 2((x-\bar{x})A_f)\tilde{f}^T + 2(b_f + \bar{x}A_f)\tilde{f}^T + 2(x-\bar{x})A_f(b_f + \bar{x}A_f)^T\Big)\\
&=&(\mathbb{E}||f||_2^2)^{-1}\mathbb{E}\Big(\tilde{f}\tilde{f}^T + (x-\bar{x})A_f((x-\bar{x})A_f)^T + (b_f + \bar{x}A_f)(b_f + \bar{x}A_f)^T\Big)\\
&=&(\mathbb{E}||f||_2^2)^{-1}\mathbb{E}\Big(||\tilde{f}||_2^2 + ||(x-\bar{x})A_f||_2^2 + ||b_f + \bar{x}A_f||_2^2\Big)\\
&=&\frac{\mathbb{E}||\tilde{f}||_2^2}{\mathbb{E}||f||_2^2} + \frac{\mathbb{E}||(x-\bar{x})A_f||_2^2}{\mathbb{E}||f||_2^2} + \frac{\mathbb{E}||b_f + \bar{x}A_f||_2^2}{\mathbb{E}||f||_2^2}\\
&=&LAR + CAR + NAR
\end{eqnarray*}
We have property 3.
\end{proof}

\begin{repproposition}{finiteNetPositiveBasis}
Assume there exists an open set $S$ where $\mathcal{D}$ has a continuous, positive density function and $f$ is not linear on $S$. Then $\Tr(\Cov_{\tilde{f}})> 0$.
\end{repproposition}

\begin{proof}
We have $\tilde{f} = f - xA_f - b_f$. Assume $\tilde{f}$ is linear on $S$. Then it can be written as $xA + b$. Then $f = x(A_f+A) + (b_f + b)$, so $f$ is linear on $S$. Contradiction. So $\tilde{f}$ is not linear on $S$. So $\tilde{f}$ is not constant on $S$. Hence, by proposition \ref{finiteNetPositiveDenominator}, $\Cov_{\tilde{f}} > 0$.
\end{proof}

\begin{repproposition}{finiteNetTheoremConnector}
Let $xA_{\tilde{f}}+b_{\tilde{f}}$ be the least squares linear fit to $\tilde{f}$. Then $A_{\tilde{f}}= 0$ and $b_{\tilde{f}}=0$.
\end{repproposition}

\begin{proof}
The standard formula for the least squares linear fit yields $\begin{pmatrix}A_f\\b_f\end{pmatrix} = (\mathbb{E}\chi^T\chi)^{-1}\mathbb{E}\chi^Tf$ and $\begin{pmatrix}A_{\tilde{f}}\\b_{\tilde{f}}\end{pmatrix} = (\mathbb{E}\chi^T\chi)^{-1}\mathbb{E}\chi^T\tilde{f}$, where $(\mathbb{E}\chi^T\chi)^{-1}$ is valid by lemma \ref{lemma1}. Then we have

\begin{eqnarray*}
&&\begin{pmatrix}A_{\tilde{f}}\\b_{\tilde{f}}\end{pmatrix}\\
&=&(\mathbb{E}\chi^T\chi)^{-1}\mathbb{E}\chi^T\tilde{f}\\
&=&(\mathbb{E}\chi^T\chi)^{-1}\mathbb{E}\chi^T(f-\chi\begin{pmatrix}A_f\\b_f\end{pmatrix})\\
&=&(\mathbb{E}\chi^T\chi)^{-1}\mathbb{E}\chi^T(f - \chi(\mathbb{E}\chi^T\chi)^{-1}\mathbb{E}\chi^Tf)\\
&=& (\mathbb{E}\chi^T\chi)^{-1}\Big(\mathbb{E}\chi^Tf - (\mathbb{E}\chi^T\chi)(\mathbb{E}\chi^T\chi)^{-1}\mathbb{E}\chi^Tf\Big)\\
&=& 0
\end{eqnarray*}

Hence, $A_{\tilde{f}}= 0$ and $b_{\tilde{f}}=0$ as required.

\end{proof}

\begin{repproposition}{finiteNetNoiseSensitivity}
Let $\delta_\text{in} \sim \mathcal{N}(0,\Cov_x)$ and $\delta_\text{out} \sim \mathcal{N}(0,\Cov_f)$ be row vectors. Assume $NLC > 0$. Then $$\mathbb{E}_{x,\delta_\text{in}}||\frac{\delta_\text{in}}{NLC}\mathcal{J}(x)^T||_2^2 = \mathbb{E}_{\delta_\text{out}}||\delta_\text{out}||_2^2$$
\end{repproposition}

\begin{proof}
We have

\begin{eqnarray*}
&&\mathbb{E}_{\delta_\text{out}}||\delta_\text{out}||_2^2\\
&=&\mathbb{E}_{\delta_\text{out}}\sum_j\delta_\text{out}[j]\delta_\text{out}[j]\\
&=&\mathbb{E}_{\delta_\text{out}}\sum_j(\delta_\text{out}^T\delta_\text{out})[j,j]\\
&=&\sum_j(\mathbb{E}_{\delta_\text{out}}\delta_\text{out}^T\delta_\text{out})[j,j]\\
&=&\sum_j\Cov_f[j,j]\\
&=&\Tr(\Cov_f)
\end{eqnarray*}

Further, we have

\begin{eqnarray*}
&&\mathbb{E}_{x,\delta_\text{in}}||\delta_\text{in}\mathcal{J}(x)^T||_2^2\\
&=&\mathbb{E}_{x,\delta_\text{in}}\sum_{j}\Big(\sum_i\delta_\text{in}[i]\mathcal{J}(x)[j,i]\Big)^2\\
&=&\mathbb{E}_{x,\delta_\text{in}}\sum_{j}\Big((\sum_i\delta_\text{in}[i]\mathcal{J}(x)[j,i])(\sum_{i'}\delta_\text{in}[i']\mathcal{J}(x)[j,i'])\Big)\\
&=&\mathbb{E}_{x,\delta_\text{in}}\sum_{j}\Big(\mathcal{J}(x)\delta_\text{in}^T\delta_\text{in}\mathcal{J}(x)^T\Big)[j,j]\\
&=&\mathbb{E}_{x}\sum_{j}\Big(\mathcal{J}(x)\mathbb{E}_{\delta_\text{in}}(\delta_\text{in}^T\delta_\text{in})\mathcal{J}(x)^T\Big)[j,j]\\
&=&\mathbb{E}_{x}\sum_{j}\Big(\mathcal{J}(x)\Cov_x\mathcal{J}(x)^T\Big)[j,j]\\
&=&\mathbb{E}_{x}\Tr(\mathcal{J}(x)\Cov_x\mathcal{J}(x)^T)
\end{eqnarray*}

Finally, we have

\begin{eqnarray*}
&&\mathbb{E}_{x,\delta_\text{in}}||\frac{\delta_\text{in}}{NLC}\mathcal{J}(x)^T||_2^2\\
&=&\frac{1}{NLC^2}\mathbb{E}_{x,\delta_\text{in}}||\delta_\text{in}\mathcal{J}(x)^T||_2^2\\
&=&\frac{\Tr(\Cov_f)}{\mathbb{E}_{x}\Tr(\mathcal{J}(x)\Cov_x\mathcal{J}(x)^T)}\mathbb{E}_{x}\Tr(\mathcal{J}(x)\Cov_x\mathcal{J}(x)^T)\\
&=&\Tr(\Cov_f)\\
&=&\mathbb{E}_{\delta_\text{out}}||\delta_\text{out}||_2^2
\end{eqnarray*}

as required.
\end{proof}

\section{Theorems}

\subsection{Theorem \ref{finiteNetNlcGreater1}: $NLC \ge 1$ for Gaussian inputs}

\begin{reptheorem}{finiteNetNlcGreater1}
Let $\mathcal{D}$ be Gaussian. Then $NLC(f, \mathcal{D}) \ge 1$, where equality holds if and only if $f$ is linear.
\end{reptheorem}

\begin{proof}
In this proof, we will use theorems \ref{finiteNetTilde} and \ref{finiteNetLAR}. Since the proofs of those theorems are not dependent on this theorem, there is no circular reasoning.

By proposition \ref{finiteNetNlcEquals1}, we have $NLC(f, \mathcal{D}) = 1$ if $f$ is linear. So now assume $f$ is not linear. By proposition \ref{finiteNetPositiveBasis}, $\Tr(\Cov_{\tilde{f}}) > 0$ and so $NLC(\tilde{f}, \mathcal{D})$ is valid. We also have $\mathbb{E}||\tilde{f}||_2^2 = \mathbb{E}||\tilde{f} - (\mathbb{E}\tilde{f})||_2^2 + ||\mathbb{E}\tilde{f}||_2^2= \Tr(\Cov_{\tilde{f}})  + ||\mathbb{E}\tilde{f}||_2^2> 0$ and so $NAR(f, \mathcal{D}) > 0$. We have

\begin{eqnarray*}
&&NLC(f,\mathcal{D})^2\\
&=&\frac{LAR}{NAR + LAR} + \frac{NAR}{NAR + LAR}NLC(\tilde{f},\mathcal{D})^2\\
&=&1 + \frac{NAR}{NAR + LAR}(NLC(\tilde{f},\mathcal{D})^2 - 1)\\
&\ge&1 + \frac{2NAR}{NAR + LAR}\\
&>&1
\end{eqnarray*}

\end{proof}

\paragraph{Discussion and alternative proof sketch} Theorem \ref{finiteNetNlcGreater1} can be proven directly without relying on theorem \ref{finiteNetTilde} or \ref{finiteNetLAR}. In fact, we proved theorem \ref{finiteNetNlcGreater1} first and then developed the proof of theorem \ref{finiteNetTilde} as an extended version. Due to the length that this chapter already has, we will only give a quick summary of this direct proof here, thereby avoiding the near-duplication of tedious arguments present in the proof of theorem \ref{finiteNetTilde}.

Like the proof of theorem \ref{finiteNetTilde}, the direct proof of theorem \ref{finiteNetNlcGreater1} proceeds in stages. In the first stage, we examine odd, scalar $f$ with unit Gaussian $\mathcal{D}$. Let $T > S > 0$. Using integration by parts, we have

\begin{eqnarray*}
&&\int_S^T f^2 d\mu_x\\
&=&\int_S^T \frac{f^2}{x} (xd_x) d\mu_1\\
&=&[-\frac{f^2}{x}d_x]_S^T + \int_S^T \frac{2ff'}{x} - \frac{f^2}{x^2} d_xd\mu_1\\
&=&\frac{f(S)^2}{S}n(S) - \frac{f(T)^2}{T}n(T) + \int_S^T \frac{2ff'}{x} - \frac{f^2}{x^2} d\mu_x\\
\end{eqnarray*}

and so

\begin{eqnarray*}
&&(\mathbb{E}f'^2 - f^2) - (\mathbb{E}f)^2\\
&=&\mathbb{E}f'^2 - f^2\\
&=&2\big(\int_0^\infty f'^2 - f^2 d\mu_x\big)\\
&=&2\big(\int_0^S f'^2 - f^2 d\mu_x + \int_S^T f'^2 d\mu_x - \int_S^T f^2 d\mu_x + \int_T^\infty f'^2 - f^2 d\mu_x\big)\\
&=&2\big(\int_0^S f'^2 - f^2 d\mu_x + \int_S^T f'^2 d\mu_x\\
&& - \frac{f(S)^2}{S}n(S) + \frac{f(T)^2}{T}n(T) - \int_S^T \frac{2ff'}{x} - \frac{f^2}{x^2} d\mu_x + \int_T^\infty f'^2 - f^2 d\mu_x\big)\\
&=&2\int_S^T(f' - \frac{f}{x})^2d\mu_x + \\
&=&2\big(\int_0^S f'^2 - f^2 d\mu_x+ \int_T^\infty f'^2 - f^2 d\mu_x - \frac{f(S)^2}{S}n(S) + \frac{f(T)^2}{T}n(T)\big)
\end{eqnarray*}

The first term is strictly positive if $f$ is not linear on $[S,T]$. Also, the first term increases as $S$ decreases and $T$ increases while the second term converges to zero as $S$ converges to zero and $T$ converges to infinity. Therefore, we can choose $S$ and $T$ such that the whole expression is strictly positive. This proves theorem \ref{finiteNetNlcGreater1} for odd, scalar $f$.

In the second stage, we generalize from odd, scalar $f$ to arbitrary scalar $f$ by observing that the NLC of such $f$ is at least a weighted average between the NLC of an odd function that mirrors $f$ on the negative quadrant and the NLC of an odd function that mirrors $f$ on the positive quadrant. Next, we consider multi-dimensional unit Gaussian $\mathcal{D}$ together with $d_\text{out}=1$. We use a simplified version of the argument from stage 4 of the proof of theorem \ref{finiteNetTilde} where we only need to define the $f_i$, but not the $t_{i,k}$ or $p_{i,k}$. Finally, we extend to the general case via a series of trivial stages, analogously to the proof of theorem \ref{finiteNetTilde}.

\subsection{Theorem \ref{finiteNetTilde}: $NLC \ge \sqrt{2}$ when the least squares fit is zero}

\begin{reptheorem}{finiteNetTilde}
Let $\mathcal{D}$ be Gaussian and let the linear component of $f$ be the zero function. Then we have $$NLC(f,\mathcal{D}) \ge \sqrt{2}$$
\end{reptheorem}

\begin{proof} This proof is quite long and will proceed in a number of stages. In each stage, we prove the statement $NLC(f,\mathcal{D}) \ge \sqrt{2}$ for a more general class of pairs $(f,\mathcal{D})$ while building on the results from previous stages. The notation defined in any stage is local to that stage.

\underline{Stage 1}: The statement holds when $d_\text{in}=d_\text{out}=1$, $\mathcal{D}$ has expectation zero and variance one, $f$ is an even function and the constant component of $f$ is the zero function.

We argue by contradiction. Assume $NLC(f,\mathcal{D}) < \sqrt{2}$. Denote $\int_S^T x^p d\mu_x$ by $I_p(S,T)$ for any $S \le T$ and $p$. In this stage, we have $NLC(f,\mathcal{D})^2 = \frac{\mathbb{E}f'^2}{\mathbb{E}f^2-(\mathbb{E}f)^2}=\frac{\lim_{X\rightarrow\infty}\int_{-X}^X f'^2d\mu_x}{\lim_{X\rightarrow\infty}\int_{-X}^X f^2d\mu_x-(\lim_{X\rightarrow\infty}\int_{-X}^X fd\mu_x)^2}$. We have $\lim_{X\rightarrow\infty}I_0(-X,X)=1$. So we have $NLC(f,\mathcal{D})^2=\frac{\lim_{X\rightarrow\infty}\frac{\int_{-X}^X f'^2d\mu_x}{I_0(-X,X)}}{\lim_{X\rightarrow\infty}\frac{\int_{-X}^X f^2d\mu_x}{I_0(-X,X)}-\Big(\lim_{X\rightarrow\infty}\frac{\int_{-X}^X fd\mu_x}{I_0(-X,X)}\Big)^2}$. Hence, there is an $X$ with $\frac{\frac{\int_{-X}^X f'^2d\mu_x}{I_0(-X,X)}}{\frac{\int_{-X}^X f^2d\mu_x}{I_0(-X,X)}-\Big(\frac{\int_{-X}^X fd\mu_x}{I_0(-X,X)}\Big)^2} < 2$ and also $X > 1$. For the remainder of stage 1, let $X$ be this fixed value. We shorten the expression $I_p(-X,X)$ to $I_p$. Rearranging, we obtain $\frac{1}{I_0}\int_{-X}^Xf'^2-2f^2d\mu_x + \frac{2}{I_0^2}(\int_{-X}^Xfd\mu_x)^2 < 0$. Let $\mathcal{T}$ be $\mathcal{D}$ restricted to $[-X,X]$. Then our formula becomes $(\mathbb{E}_{x\sim\mathcal{T}}f'^2)-2(\mathbb{E}_{x\sim\mathcal{T}}f^2) +2(\mathbb{E}_{x\sim\mathcal{T}}f)^2 < 0$. Let $f_K$ be the piecewise linear approximation of $f$ on the interval $[-X,X]$ with $K$ linear segments as defined in section \ref{finiteNetDefinitionsSection}. By lemma \ref{lemma4}, we can choose an $e>0$ and $K_e$ such that $(\mathbb{E}_{x\sim\mathcal{T}}f'^2_{2K})-2(\mathbb{E}_{x\sim\mathcal{T}}f^2_{2K})+2(\mathbb{E}_{x\sim\mathcal{T}}f_{2K})^2 < -e$ for all $K>K_e$. We write $obj(f_{2K})<-e$.

The next step in the proof is to show that for each $\eta>0$, there is a $K_\eta > K_e$ and an even quadratic function $q=ax^2+c$ such that $obj(q) < obj(f_{2K_\eta})+ \eta$. Because this step is quite long and tedious, we will first show how to complete stage 1 once we have shown the existence of this $K_\eta$ and $q$. We have for arbitrary $q$

\begin{eqnarray*}
&&obj(q)\\
&=&\frac{1}{I_0}\int_{-X}^X4a^2x^2-2a^2x^4-4acx^2-2c^2d\mu_x + \frac{2}{I_0^2}(\int_{-X}^Xax^2+c d\mu_x)^2\\
&=&\frac{1}{I_0}(4a^2I_2-2a^2I_4-4acI_2-2c^2I_0) + \frac{2}{I_0^2}(a^2I_2^2+2acI_2I_0+c^2I_0^2)\\
&=&\frac{2a^2}{I_0^2}(2I_2I_0-I_4I_0+I_2^2)\\
&=&\frac{2a^2}{I_0^2}\Big(2(1-2N(-X)-2Xn(X))(1-2N(-X))-\\
&&(3-6N(-X)-2X(X^2+3)n(X))(1-2N(-X))+(1-2N(-X)-2Xn(X))^2\Big)\\
&=&\frac{2a^2}{I_0^2}(-2Xn(X)+4Xn(X)N(-X)+4X^2n(X)^2+2n(X)X^3-4X^3n(X)N(X))\\
&=&\frac{2a^2}{I_0^2}(2Xn(X)(X^2-1)(1-2N(-X))+4X^2n(X)^2)\\
&\ge&0
\end{eqnarray*}

In the last step, we used $X>1$. But if we choose $\eta < e$, then from $obj(f_{2K_\eta})<-e$ and $obj(q) < obj(f_{2K_\eta})+ \eta$ we obtain $obj(q) < 0$. This is the contradiction that would complete stage 1. So we are left to show the following.

{\it Claim 1:} For each $\eta > 0$, there is a $K_\eta > K_e$ and an even quadratic function $q=ax^2+c$ such that $obj(q) < obj(f_{2K_\eta})+ \eta$.

For each $K$, we define the even quadratic function $q_K=a_Kx^2+c_K$. We choose the parameters $a_K$ and $c_K$ to satisfy the following equalities: $f_{2K}(X)=q_K(X)$ and $\mathbb{E}_\mathcal{T}f_{2K}=\mathbb{E}_\mathcal{T}q_K$. It is clear that for each pair of values of $f_{2K}(X)$ and $\mathbb{E}_\mathcal{T}f_{2K}$, there exist unique values for $a_K$ and $c_K$. Because $f$, and hence $f_{2K}$, is even, we also have $f_{2K}(-X)=q_K(-X)$.

Let $objdiff(K) = obj(f_{2K}) - obj(q_K)$. The remainder of the proof of claim 1 will proceed as follows. In step 1, we will fix $K$ and derive an alternative formula for $objdiff$. In step 2, we will break down that formula into a sum of terms where each term is either positive or arbitrarily close to zero for large $K$. This will then allow us to pick a $K > K_e$ with $objdiff(K) > -\eta$ for arbitrary $K_e$ and $\eta$, and then we will have claim 1.

Step 1: Fix $K$. Define $F_k$, $0\le k\le K$ as follows. On $[-X,-\frac{k}{K}X]$ and $[\frac{k}{K}X,X]$, $F_k$ is equal to $f_{2K}$. On $[-\frac{k}{K}X,\frac{k}{K}X]$, $F_k$ is an even quadratic that connects with $f_{2K}$ at $-\frac{k}{K}X$ and $\frac{k}{K}X$ and has $\mathbb{E}_{x\in[-\frac{k}{K}X,\frac{k}{K}X]}F_k = \mathbb{E}_{x\in[-\frac{k}{K}X,\frac{k}{K}X]}f_{2K}$, which is equivalent to $\mathbb{E}_\mathcal{T}F_k = \mathbb{E}_\mathcal{T}f_{2K}$, to $\int_{-X}^XF_kd\mu_x = \int_{-X}^Xf_{2K}d\mu_x$ and to $\int_0^{\frac{k}{K}X}F_kd\mu_x =\int_0^{\frac{k}{K}X}f_{2K}d\mu_x$ because both functions are even. Then $F_0=f_{2K}$ and $F_K=q_K$. So $objdiff=\sum_{k=0}^{K-1}obj(F_k)-obj(F_{k+1})$. As $\int_{-X}^XF_kd\mu_x$ is the same for all $k$, we have $objdiff = \frac{1}{I_0}\sum_{k=0}^{K-1}\int_{-X}^X(F'^2_k - 2F^2_k) - (F'^2_{k+1} - 2F^2_{k+1})d\mu_x$. Because all functions are even, we then have $objdiff = \frac{2}{I_0}\sum_{k=0}^{K-1}\int_0^X(F'^2_k - 2F^2_k) - (F'^2_{k+1} - 2F^2_{k+1})d\mu_x$. Finally, we can eliminate sections of the integral where consecutive $F$'s are equal. Then we obtain $objdiff=\frac{2}{I_0}\sum_{k=0}^{K-1}\int_0^{\frac{(k+1)}{K}X}(F'^2_k - 2F^2_k) - (F'^2_{k+1} - 2F^2_{k+1})d\mu_x$. Denote the $k$'th term of this sum by $D_k$.

Let's look at a specific $D_k$. For now, we will omit most $k$ and $K$ subscripts because $k$ and $K$ are considered fixed. Instead, we introduce some additional notation. Let $\epsilon=\frac{X}{K}$, $y=k\epsilon$, $z=(k+1)\epsilon$ and let $bx+d$ be the line segment in $f_{2K}$ between $k\epsilon$ and $(k+1)\epsilon$. Let $ax^2+r$ be the equation of the even square in $F_k$ between 0 and $k\epsilon$. Let $Ax^2+R$ be the even quadratic in $F_{k+1}$ between 0 and $(k+1)\epsilon$. Then the 6 parameters $a,r,A,R,b,d$ satisfy the following constraints, which arise from $F_k(y) = f_{2K}(y)$, $F_{k+1}(z) = f_{2K}(z)$ and $\int_0^zF_kd\mu_x = \int_0^zF_{k+1}d\mu_x$ respectively.

\begin{eqnarray*}
ay^2+r&=&by+d\\
Az^2+R&=&bz+d\\
aI_2(0,y)+rI_0(0,y)+bI_1(y,z)+dI_0(y,z)&=&AI_2(0,z)+RI_0(0,z)
\end{eqnarray*}

And we have

\begin{eqnarray*}
&&D_k(a,r,A,R,b,d)\\
&=&(b^2-2d^2)I_0(y,z)-2b^2I_2(y,z)-4bdI_1(y,z)+4a(a-r)I_2(0,y)-2a^2I_4(0,y)\\
&&-2r^2I_0(0,y)-4A(A-R)I_2(0,z)+2A^2I_4(0,z)+2R^2I_0(0,z)
\end{eqnarray*}

It is easy to check that simultaneously adding a constant to $F_k$ and $F_{k+1}$ neither alters $D_k$ nor affects the validity of the constraints. Hence WLOG we can subtract $d$ from $r$ and $R$ and then set $d=0$ within the constraints and $D_k$. We then use the first two constraints to substitute values for $r$ and $R$ into the third constraint and $D_k$. We thus obtain a single constraint.

\begin{equation*}
A=\frac{aI_2(0,y)+(by-ay^2)I_0(0,y)+bI_1(y,z)-bzI_0(0,z)}{I_2(0,z)-z^2I_0(0,z)}
\end{equation*}

The denominator is $h(z)$ where $h$ is defined as in lemma \ref{lemma5}. By that lemma, $z>0$ implies $h(z)<0$, so the above ratio is valid. 

Both $N$ and $n$ are continuously differentiable with a Lipschitz derivative. Therefore, we have $N(z)=N(y)+\epsilon N'(y)+\epsilon^2\nabla(y,\epsilon)=N(y)+\epsilon n(y)+\epsilon^2\nabla(y,\epsilon)$ and $n(z)=n(y)+\epsilon n'(y)+\epsilon^2\nu(y,\epsilon)=n(y)-\epsilon yn(y)+\epsilon^2\nu(y,\epsilon)$ for some $\nu$ and $\nabla$ which are bounded for $y, \epsilon \in [0,X]$. Substituting the above expressions for $A$, $n(z)$, $N(z)$ and $z=y+\epsilon$ into $D_k$ we obtain

\begin{equation*}
D_k(a,b)=n(y)(2ya-b)^2\frac{h^2(y)}{h^2(y+\epsilon)}\epsilon + \frac{P^{(2)}b^2+P^{(1)}b+P^{(0)}}{h^2(y+\epsilon)}\epsilon^2
\end{equation*}

Here $P^{(2)}$, $P^{(1)}$ and $P^{(0)}$ are polynomials composed of $\epsilon$, $y$, $n(y)$, $N(y)$, $\nu(y,\epsilon)$, $\nabla(y,\epsilon)$ and $a$ terms. Importantly, the symbolic expression of these polynomials is not dependent on $K$ or $k$.

Now, we re-introduce the $k$ subscripts as we aggregate the values of the $D_k$ for our fixed $K$. We obtain

\begin{equation*}
objdiff = \frac{2}{I_0}\sum_{k=0}^{K-1}D_k=\frac{2}{I_0}\sum_{k=0}^{K-1}n(y_k)(2y_ka_k-b_k)^2\frac{h^2(y_k)}{h^2(y_k+\epsilon)}\epsilon + \frac{P^{(2)}b_k^2 + P^{(1)}b_k + P^{(0)}}{h^2(y_k+\epsilon)}\epsilon^2
\end{equation*}

Again, we note that the three polynomials do not have $k$ subscripts because their symbolic expression does not depend on $k$. This is the alternative expression of $objdiff$ we were looking for in step 1.

Step 2: Now $K$ is no longer considered fixed. Let $C$ be a positive integer and assume that $K$ is a multiple of $C$. Then we break down $objdiff(K)$ into $\frac{2}{I_0}\sum_{k=0}^{\frac{K}{C}-1}D_k$ and $\frac{2}{I_0}\sum_{k=\frac{K}{C}}^{K-1}D_k$.

Let's look at $\frac{2}{I_0}\sum_{k=\frac{K}{C}}^{K-1}D_k$ first, and let's do so for a fixed value of $C$. We have $y_k>\frac{X}{C}$, so by lemma \ref{lemma5}, $h(y_k+\epsilon)$ is bounded away from zero across all $K$ and $k\ge\frac{K}{C}$. All terms that make up the polynomials $P^{(0)},P^{(1)},P^{(2)}$ are bounded, except possibly $a_k$. Hence, if we can show that $a_k$ is also bounded, then the polynomials are bounded. $a_k$ and $r_k$ are determined via the system

\begin{equation*}
\begin{pmatrix}I_2(0,y_k)&I_0(0,y_k)\\y_k^2&1\end{pmatrix}\begin{pmatrix}a_k\\r_k \end{pmatrix}=\begin{pmatrix}\int_0^{y_k}f_{2K}(x)d\mu_x\\f_{2K}(y_k) \end{pmatrix}
\end{equation*}

Since $f$ is continuous, $f$ is bounded on $[0,X]$, so the right-hand side is bounded. Since $y_k$ is bounded, all entries in the 2 by 2 matrix are bounded. So, if the determinant of the matrix is bounded away from zero, then $a_k$ and $r_k$ are bounded. The determinant is $I_2(0,y_k)-y_k^2I_0(0,y_k)=h(y_k)$. Since $y_k\ge\frac{X}{C}$ is bounded away from zero, so is $h(y_k)$ by lemma \ref{lemma5}. Hence $a_k$ is indeed bounded. So we have for some constants $B_2,B_1,B_0$

\begin{eqnarray*}
&&\Big|\frac{2}{I_0}\sum_{k=\frac{K}{C}}^{K-1}\frac{P^{(2)}b_k^2 + P^{(1)}b_k + P^{(0)}}{h^2(y_k+\epsilon)}\epsilon^2\Big|\\
&\le&\frac{2}{I_0}\sum_{k=\frac{K}{C}}^{K-1}(b_k^2B_2+|b_k|B_1+B_0)\epsilon^2\\
&=&\frac{2B_2\epsilon}{I_0}\sum_{k=\frac{K}{C}}^{K-1}b_k^2\epsilon + \frac{2B_1\epsilon}{I_0}\sum_{k=\frac{K}{C}}^{K-1}|b_k|\epsilon + \frac{2B_0\epsilon}{I_0}\sum_{k=\frac{K}{C}}^{K-1}\epsilon
\end{eqnarray*}

We will now investigate each of the 3 terms in the above sum one after the other.

\begin{enumerate}
\item Remember that $b_k$ is defined to be the slope of $f_{2K}$ on $[\frac{k}{K}X,\frac{k+1}{K}X]$. By the intermediate value principle, $f'$ takes a value at least as large as $b_k$ and a value at least as small as $b_k$ on $[\frac{k}{K}X,\frac{k+1}{K}X]$. Hence, $f'^2$ takes a value at least as large as $b_k^2$ in $[\frac{k}{K}X,\frac{k+1}{K}X]$. Let such a value be $\beta_k^2$. Because $f'^2$ is integrable over the unit Gaussian measure over any finite interval, it is also integrable over the canonical Lebesgue measure over any finite interval. Hence we have $\int_\frac{X}{C}^Xf'^2d\mu_1=\lim_{K\rightarrow\infty}\sum_{k=\frac{K}{C}}^{K-1}\beta_k^2\epsilon$. So $\sum_{k=\frac{K}{C}}^{K-1}\beta_k^2\epsilon$ is bounded for sufficiently large $K$, so $\sum_{k=\frac{K}{C}}^{K-1}b_k^2\epsilon$ is bounded for sufficiently large $K$. And since $\epsilon$ converges to zero as $K$ converges to infinity, $\frac{2B_2\epsilon}{I_0}\sum_{k=\frac{K}{C}}^{K-1}b_k^2\epsilon$ converges to zero.
\item Since $\sum_{k=\frac{K}{C}}^{K-1}b_k^2\epsilon$ is bounded for sufficiently large $K$, so is $\sqrt{\sum_{k=\frac{K}{C}}^{K-1}b_k^2\epsilon}$, so is $(\sum_{k=\frac{K}{C}}^{K-1}|b_k|\epsilon)(\sum_{k=\frac{K}{C}}^{K-1}\epsilon)^{-\frac{1}{2}}$ by Jensen's inequality, so is $\sum_{k=\frac{K}{C}}^{K-1}|b_k|\epsilon$. And since $\epsilon$ converges to zero as $K$ converges to infinity, $\frac{2B_1\epsilon}{I_0}\sum_{k=\frac{K}{C}}^{K-1}|b_k|\epsilon$ converges to zero.
\item $\frac{2B_0\epsilon}{I_0}\sum_{k=\frac{K}{C}}^{K-1}\epsilon=\frac{2B_0\epsilon(X-\frac{X}{C})}{I_0}$ also converges to zero.
\end{enumerate}

So in summary, we have that, for a fixed $C$ and $K$ restricted to multiples of $C$, $\Big|\frac{2}{I_0}\sum_{k=\frac{K}{C}}^{K-1}\frac{P^{(2)}b_k^2 + P^{(1)}b_k + P^{(0)}}{h^2(y_k+\epsilon)}\epsilon^2\Big|$ converges to zero as $K$ converges to infinity.

Now let's look at $\frac{2}{I_0}\sum_{k=0}^{\frac{K}{C}-1}D_k$. We have $\frac{2}{I_0}\sum_{k=0}^{\frac{K}{C}-1}D_k=\frac{2}{I_0}\int_0^\frac{X}{C}(F'^2_0-2F^2_0) - (F'^2_\frac{K}{C}-2F^2_\frac{K}{C})d\mu_x=\frac{2}{I_0}\int_0^\frac{X}{C}F'^2_0+2F^2_\frac{K}{C}-2F^2_0 - F'^2_\frac{K}{C}d\mu_x$. The first two out of those four terms are non-negative. Since $f$ is bounded on $[0,X]$, so is $F_0$. So $\int_0^\frac{X}{C}-2F^2_0d\mu_x$ converges to zero as $C$ converges to infinity. Finally, we turn to the $\int_0^\frac{X}{C}- F'^2_\frac{K}{C}d\mu_x$ term. On $[0,\frac{X}{C}]$, $F_\frac{K}{C}$ is an even quadratic. Denote it by $ax^2+r$. Setting $\delta=\frac{X}{C}$, $a$ and $r$ are determined via the system 

\begin{equation*}
\begin{pmatrix}I_2(0,\delta)&I_0(0,\delta)\\\delta^2&1\end{pmatrix}\begin{pmatrix}a\\r \end{pmatrix}=\begin{pmatrix}\int_0^\delta f_{2K}(x)d\mu_x\\f_{2K}(\delta) \end{pmatrix}
\end{equation*}

Thus we obtain

\begin{eqnarray*}
&&\int_0^\delta- F'^2_\frac{K}{C} d\mu_x\\
&=&-4a^2I_2(0,\delta)\\
&=&-4I_2(0,\delta)\Big(\frac{\int_0^\delta f_{2K}(x)d\mu_x-I_0(0,\delta)f_{2K}(\delta)}{I_2(0,\delta)-\delta^2I_0(0,\delta)}\Big)^2\\
&=&-4I_2(0,\delta)I_0(0,\delta)^2\Big(\frac{\mathbb{E}_{x\in[0,\delta]} f_{2K}(x)-f_{2K}(\delta)}{I_2(0,\delta)-\delta^2I_0(0,\delta)}\Big)^2
\end{eqnarray*}

The denominator is clearly non-zero. By lemma \ref{lemma7}, we have that all values of $f$ are within $o(\delta^\frac{1}{2})$ of $f(0)$ on the interval $[0,\delta]$. Since $K$ is a multiple of $C$, we have $\frac{X}{K} \le \frac{X}{C} =\delta$. Therefore, if the domain of a line segment in $f_{2K}$ intersects $[0,\delta]$, that domain is contained in $[0,2\delta]$. Therefore, $f_{2K}$ can only take values in $[0,\delta]$ that $f$ takes in $[0,2\delta]$. Therefore, all values of $f_{2K}$ are also within $o(\delta^\frac{1}{2})$ of $f(0)$ on the interval $[0,\delta]$, uniformly over $K$. Hence, $\mathbb{E}_{x\in[0,\delta]} f_{2K}(x)-f_{2K}(\delta)=o(\delta^\frac{1}{2})$. It is easy to check that $I_2(0,\delta)=\Theta(\delta^3)$, $I_0(0,\delta)=\Theta(\delta)$ and $I_2(0,\delta)-\delta^2I_0(0,\delta)=\Theta(\delta^3)$. This yields that $\int_0^\delta- F'^2_\frac{K}{C} d\mu_x=o(1)$, so $\int_0^\delta- F'^2_\frac{K}{C} d\mu_x$ converges to zero as $\delta$ converges to zero, i.e. as $C$ converges to infinity.

This was the last piece of the puzzle. Now we put it all together. For some $C$ and $K$ a multiple of $C$ we have

\begin{eqnarray*}
&&objdiff(K)\\
&=&\frac{2}{I_0}\sum_{k=0}^{K-1}\int_0^{\frac{(k+1)}{K}X}(F'^2_k - 2F^2_k) - (F'^2_{k+1} - 2F^2_{k+1})d\mu_x\\
&=&\frac{2}{I_0}\Bigg(\Big(\int_0^{\frac{X}{C}}F'^2_0d\mu_x\Big) - 2\Big(\int_0^{\frac{X}{C}}F^2_0d\mu_x\Big) - \Big(\int_0^{\frac{X}{C}}F'^2_\frac{K}{C}d\mu_x\Big) + 2\Big(\int_0^{\frac{X}{C}}F^2_\frac{K}{C}d\mu_x\Big)\\
&&+\sum_{k=\frac{K}{C}}^{K-1}n(y_k)(2y_ka_k-b_k)^2\frac{h^2(y_k)}{h^2(y_k+\epsilon)}\epsilon+\sum_{k=\frac{K}{C}}^{K-1}\frac{P^{(2)}b_k^2 + P^{(1)}b_k + P^{(0)}}{h^2(y_k+\epsilon)}\epsilon^2\Bigg)
\end{eqnarray*}

Out of the six terms, the first, fourth and fifth are non-negative. We have shown that as $C$ converges to infinity, the second and third terms converge to zero. And we have shown that for a fixed $C$ as $K$ converges to infinity, the sixth term converges to zero. Therefore, no matter the value of $\eta$, we can first choose a $C$ and then a $K > K_e$ which is a multiple of $C$ such that $objdiff(K)$ is greater than $-\eta$, as required. The constant $\frac{2}{I_0}$ has no influence on the analysis. Step 2 is complete. Hence, we have claim 1. Hence, stage 1 is complete.

\underline{Stage 2}: The statement holds when $d_\text{in}=d_\text{out}=1$, $\mathcal{D}$ has expectation zero and variance one, $f$ is an odd function and the constant component of $f$ is the zero function.

We proceed largely analogously to stage 1.

\sloppy Again, we argue by contradiction. However, as opposed to stage 1, we assume $NLC(f,\mathcal{D}) < \sqrt{3}$. If we can use this to derive a contradiction, we will have proven the strictly stronger statement $NLC(f,\mathcal{D}) \ge \sqrt{3}$. Denote $\int_S^T x^p d\mu_x$ by $I_p(S,T)$ for any $S \le T$ and $p$. We have $NLC(f,\mathcal{D})^2 = \frac{\mathbb{E}f'^2}{\mathbb{E}f^2-(\mathbb{E}f)^2}$. Because $f$ is odd, we have $\mathbb{E}f=0$. Since the linear and constant component of $f$ are both the zero function, we have $f = \tilde{f}$. By proposition \ref{finiteNetBasisOrtho} part 1, we have $\mathbb{E}x\tilde{f} = \mathbb{E}xf = 0$ and so $NLC(f,\mathcal{D})^2 = \frac{\mathbb{E}f'^2}{\mathbb{E}f^2-\frac{2}{3}(\mathbb{E}xf)^2} = \frac{\lim_{X\rightarrow\infty}\int_{-X}^X f'^2d\mu_x}{\lim_{X\rightarrow\infty}\int_{-X}^X f^2d\mu_x-\frac{2}{3}(\lim_{X\rightarrow\infty}\int_{-X}^X xfd\mu_x)^2}$. We have $\lim_{X\rightarrow\infty}I_2(-X,X)=1$. So we have $NLC(f,\mathcal{D})^2=\frac{\lim_{X\rightarrow\infty}\frac{\int_{-X}^X f'^2d\mu_x}{I_2(-X,X)}}{\lim_{X\rightarrow\infty}\frac{\int_{-X}^X f^2d\mu_x}{I_2(-X,X)}-\frac{2}{3}\Big(\lim_{X\rightarrow\infty}\frac{\int_{-X}^X xfd\mu_x}{I_2(-X,X)}\Big)^2}$. Hence, there is an $X$ with $\frac{\frac{\int_{-X}^X f'^2d\mu_x}{I_2(-X,X)}}{\frac{\int_{-X}^X f^2d\mu_x}{I_2(-X,X)}-\frac{2}{3}\Big(\frac{\int_{-X}^X fd\mu_x}{I_2(-X,X)}\Big)^2} < 3$ and also $X > 100$. For the remainder of stage 2, let $X$ be this fixed value. We shorten the expression $I_p(-X,X)$ to $I_p$. Rearranging, we obtain $\frac{1}{I_2}\int_{-X}^Xf'^2-3f^2d\mu_x + \frac{2}{I_2^2}(\int_{-X}^Xxfd\mu_x)^2 < 0$. Let $\mathcal{T}$ be $\mathcal{D}$ restricted to $[-X,X]$. Then our formula becomes $\frac{I_0}{I_2}(\mathbb{E}_{x\sim\mathcal{T}}f'^2)-\frac{3I_0}{I_2}(\mathbb{E}_{x\sim\mathcal{T}}f^2) +\frac{2I_0^2}{I_2^2}(\mathbb{E}_{x\sim\mathcal{T}}xf)^2 < 0$. Let $f_K$ be the piecewise linear approximation of $f$ on the interval $[-X,X]$ with $K$ linear segments as defined in section \ref{finiteNetDefinitionsSection}. By lemma \ref{lemma4}, we can choose an $e>0$ and $K_e$ such that $\frac{I_0}{I_2}(\mathbb{E}_{x\sim\mathcal{T}}f'^2_{2K})-\frac{3I_0}{I_2}(\mathbb{E}_{x\sim\mathcal{T}}f^2_{2K})+\frac{2I_0^2}{I_2^2}(\mathbb{E}_{x\sim\mathcal{T}}xf_{2K})^2 < -e$ for all $K>K_e$. We write $obj(f_{2K})<-e$.

The next step in the proof is to show that for each $\eta>0$, there is a $K_\eta > K_e$ and an odd cubic function $q=ax^3+cx$ such that $obj(q) < obj(f_{2K_\eta})+ \eta$. Because this step is quite long and tedious, we will first show how to complete stage 2 once we have shown the existence of this $K_\eta$ and $q$. We have for arbitrary $q$

\begin{eqnarray*}
&&obj(q)\\
&=&\frac{1}{I_2}\int_{-X}^X9a^2x^4+6acx^2+c^2-3a^2x^6-6acx^4-3c^2x^2d\mu_x + \frac{2}{I_2^2}(\int_{-X}^Xax^4+cx^2 d\mu_x)^2\\
&=&\frac{1}{I_2}(9a^2I_4+6acI_2+c^2I_0-3a^2I_6-6acI_4-3c^2I_2) + \frac{2}{I_2^2}(a^2I_4^2+2acI_4I_2+c^2I_2^2)\\
&=&\frac{1}{I_2}((9I_4-3I_6+2\frac{I_4^2}{I_2})a^2+(6I_2-2I_4)ac+(I_0-I_2)c^2)\\
\end{eqnarray*}

If $a=0$, this becomes $\frac{(I_0-I_2)c^2}{I_2}$, which is non-negative so $obj(q) \ge 0$. If $a\neq 0$, then 

\begin{equation*}
obj(q)=\frac{a^2}{I_2}\Big((9I_4-3I_6+2\frac{I_4^2}{I_2})+(6I_2-2I_4)(\frac{c}{a})+(I_0-I_2)(\frac{c}{a})^2\Big)
\end{equation*}

Inside the large parentheses, we have a quadratic in $\frac{c}{a}$. The coefficient of the quadratic term is positive. The discriminant is $\frac{32n(X)^2X^4(-3+6n(X)X+X^2-2N(-X)(X^2-3))}{-1+2N(-X)+2n(X)X}$. Since $X>100$, $X^2$ dominates the sum in the numerator and $-1$ dominates the sum in the denominator. Hence, the discriminant is negative. Therefore, the quadratic is positive everywhere, so $obj(q)>0$.

So no matter the value of $a$ we have $obj(q)\ge 0$. But if we choose $\eta < e$, then from $obj(f_{2K_\eta})<-e$ and $obj(q) < obj(f_{2K_\eta})+ \eta$ we obtain $obj(q) < 0$. This is the contradiction that would complete stage 2. So we are left to show the following.

{\it Claim 1}: For each $\eta > 0$, there is a $K_\eta > K_e$ and an odd cubic function $q=ax^3+cx$ such that $obj(q)< obj(f_{2K_\eta})+\eta$.

We define for each $K$ the odd cubic function $q_K=a_Kx^3+c_Kx$. We choose the parameters $a_K$ and $c_K$ to satisfy the following two constraints: $f_{2K}(X)=q_K(X)$ and $\mathbb{E}_\mathcal{T}xf_{2K}=\mathbb{E}_\mathcal{T}xq_K$. It is clear that for each pair of values of $f_{2K}(X)$ and $\mathbb{E}_\mathcal{T}xf_{2K}$, there exist unique values for $a_K$ and $c_K$. Because $f$ and hence $f_{2K}$ is odd, we also have $f_{2K}(-X)=q_K(-X)$.

Let $objdiff(K) = obj(f_{2K}) - obj(q_K)$. The remainder of the proof of claim 1 will proceed as follows. In step 1, we will fix $K$ and derive an alternative formula for $objdiff$. In step 2, we will break down that formula into a sum of terms where each term is either positive or arbitrarily close to zero for large $K$. This will then allow us to pick a $K > K_e$ with $objdiff(K) > -\eta$ for arbitrary $K_e$ and $\eta$, and then we will have claim 1.

Step 1: Fix $K$. Define $F_k$, $0\le k\le K$ as follows. On $[-X,-\frac{k}{K}X]$ and $[\frac{k}{K}X,X]$, $F_k$ equals $f_{2K}$. On $[-\frac{k}{K}X,\frac{k}{K}X]$, $F_k$ is an odd cubic that connects with $f_{2K}$ at $-\frac{k}{K}X$ and $\frac{k}{K}X$ and has $\mathbb{E}_{x\in[-\frac{k}{K}X,\frac{k}{K}X]}xF_k = \mathbb{E}_{x\in[-\frac{k}{K}X,\frac{k}{K}X]}xf_{2K}$, which is equivalent to $\mathbb{E}_\mathcal{T}xF_k = \mathbb{E}_\mathcal{T}xf_{2K}$, to $\int_{-X}^XxF_kd\mu_x = \int_{-X}^Xxf_{2K}d\mu_x$ and to $\int_0^{\frac{k}{K}X}xF_kd\mu_x = \int_0^{\frac{k}{K}X}xf_{2K}d\mu_x$ because both functions are odd. Then $F_0=f_{2K}$ and $F_K=q_K$. So $objdiff=\sum_{k=0}^{K-1}obj(F_k)-obj(F_{k+1})$. As $\int_{-X}^XxF_kd\mu_x$ is the same for all $k$, we have $objdiff = \frac{1}{I_2}\sum_{k=0}^{K-1}\int_{-X}^X(F'^2_k - 3F^2_k) - (F'^2_{k+1} - 3F^2_{k+1})d\mu_x$. Because all functions are odd, we then have $objdiff = \frac{2}{I_2}\sum_{k=0}^{K-1}\int_0^X(F'^2_k - 3F^2_k) - (F'^2_{k+1} - 3F^2_{k+1})d\mu_x$. Finally, we can eliminate sections of the integral where consecutive $F$'s are equal. Then we obtain $objdiff=\frac{2}{I_2}\sum_{k=0}^{K-1}\int_0^{\frac{(k+1)}{K}X}(F'^2_k - 3F^2_k) - (F'^2_{k+1} - 3F^2_{k+1})d\mu_x$. Denote the $k$'th term of this sum by $D_k$.

Let's look at a specific $D_k$ where $k\ge 1$. For now, we will omit most $k$ and $K$ subscripts because $k$ and $K$ are considered fixed. Instead, we introduce some additional notation. Let $\epsilon=\frac{X}{K}$, $y=k\epsilon$, $z=(k+1)\epsilon$ and let $bx+d$ be the line segment in $f_{2K}$ between $k\epsilon$ and $(k+1)\epsilon$. Let $ax^3+rx$ be the odd cubic in $F_k$ between 0 and $k\epsilon$. Let $Ax^3+Rx$ be the equation of the odd cubic in $F_{k+1}$ between 0 and $(k+1)\epsilon$.

Then the 6 parameters $a,r,A,R,b,d$ satisfy the following constraints, which arise from $F_k(y) = f_{2K}(y)$, $F_{k+1}(z) = f_{2K}(z)$ and $\int_0^zxF_kd\mu_x = \int_0^zxF_{k+1}d\mu_x$ respectively.

\begin{eqnarray*}
ay^3+ry&=&by+d\\
Az^3+Rz&=&bz+d\\
aI_4(0,y)+rI_2(0,y)+bI_2(y,z)+dI_1(y,z)&=&AI_4(0,z)+RI_2(0,z)
\end{eqnarray*}

And we have

\begin{eqnarray*}
&&D_k(a,r,A,R,b,d)\\
&=&(b^2-3d^2)I_0(y,z)-3b^2I_2(y,z)-6bdI_1(y,z)-3a^2I_6(0,y)+(9a^2-6ar)I_4(0,y)\\
&&+(6ar-3r^2)I_2(0,y)+r^2I_0(0,y)+3A^2I_6(0,z)-(9A^2-6AR)I_4(0,z)\\
&&-(6AR-3R^2)I_2(0,z)-R^2I_0(0,z)
\end{eqnarray*}

We also have

\begin{eqnarray*}
&&\int_0^zF'_k-F'_{k+1}d\mu_x\\
&=&\int_0^z(F'_k-F'_{k+1})d_xd\mu_1\\
&=&[(F_k-F_{k+1})d_x]_0^z - \int_0^z(F_k-F_{k+1})\frac{dd_x}{dx}d\mu_1\\
&=&[(F_k-F_{k+1})d_x]_0^z + \int_0^zx(F_k-F_{k+1})d_xd\mu_1\\
&=&0 + \int_0^zx(F_k-F_{k+1})d\mu_x\\
&=&0
\end{eqnarray*}

Here, we use that $F_k$ and $F_{k+1}$ are odd and hence $F_k(0)=F_{k+1}(0)=0$. We use that $F_k$ and $F_{k+1}$ are equal to $f_{2K}$ on $[z,X]$ and hence $F_k(z)=F_{k+1}(z)$. We use that $\int_0^zxF_kd\mu_x = \int_0^zxF_{k+1}d\mu_x$. We also use integration by parts. This is allowed because the probability density function $d_x$ is absolutely continuous and $F'_k-F'_{k+1}$ is integrable.

Let $c$ be an arbitrary constant. Then we have

\begin{eqnarray*}
&&D_k\\
&=&\int_0^z F'^2_k - 3F^2_k - (F'^2_{k+1} - 3F^2_{k+1})d\mu_x\\
&=&\int_0^z (F_k-cx+cx)'^2 - 3(F_k-cx+cx)^2\\
&&- ((F_{k+1}-cx+cx)'^2 - 3(F_{k+1}-c x+c x)^2)d\mu_x\\
&=&\int_0^z (F_k-c x)'^2 + 2c (F'_k-c) + c^2 - 3(F_k-c x)^2 -6(F_k-c x)c x - 3c^2x^2\\
&&- ((F_{k+1}-c x)'^2 + 2c (F'_{k+1}-c) + c^2 - 3(F_{k+1}-c x)^2 -6(F_{k+1}-c x)c x - 3c^2x^2)d\mu_x\\
&=&\int_0^z (F_k-c x)'^2 - 3(F_k-c x)^2 - ((F_{k+1}-c x)'^2  - 3(F_{k+1}-c x)^2)\\
&& + 2c (F'_k-c) -6(F_k-c x)c x - 2c (F'_{k+1}-c)+6(F_{k+1}-c x)c xd\mu_x\\
&=&\int_0^z (F_k-c x)'^2 - 3(F_k-c x)^2 - ((F_{k+1}-c x)'^2  - 3(F_{k+1}-c x)^2)d\mu_x\\
&&+2c\int_0^zF'_k-F'_{k+1}d\mu_x - 6c\int_0^zF_kx-F_{k+1}xd\mu_x\\
&=&\int_0^z (F_k-c x)'^2 - 3(F_k-c x)^2 - ((F_{k+1}-c x)'^2  - 3(F_{k+1}-c x)^2)d\mu_x\\
\end{eqnarray*}

So subtracting a constant multiple of $x$ from $F_k$ and $F_{k+1}$ simultaneously does not alter $D_k$. It also does not affect the validity of the constraints. Hence, WLOG, we may subtract $b$ from $r$ and $R$ and then set $b=0$ within the constraints and $D_k$. We then use the first two constraints to substitute values for $r$ and $R$ into the third constraint and $D_k$. We thus obtain a single constraint.

\begin{equation*}
A=\frac{aI_4(0,y)+(\frac{d}{y}-ay^2)I_2(0,y)+dI_1(y,z)-\frac{d}{z}I_2(0,z)}{I_4(0,z)-z^2I_2(0,z)}
\end{equation*}

The large denominator is $H(z)$ where $H$ is defined as in lemma \ref{lemma6}. By that lemma, $z>0$ implies $H(z)<0$, so the large ratio is valid. Since $z>0$, $\frac{d}{z}$ is valid. Since we set $k \ge 1$, we have $y>0$ and so $\frac{d}{y}$ is valid.

Both $N$ and $n$ are continuously differentiable with a Lipschitz derivative. Therefore, we have $N(z)=N(y)+\epsilon N'(y)+\epsilon^2\nabla(y,\epsilon)=N(y)+\epsilon n(y)+\epsilon^2\nabla(y,\epsilon)$ and $n(z)=n(y)+\epsilon n'(y)+\epsilon^2\nu(y,\epsilon)=n(y)-\epsilon yn(y)+\epsilon^2\nu(y,\epsilon)$ for some $\nu$ and $\nabla$ which are bounded for $y, \epsilon \in [0,X]$. Substituting the above expressions for $A$, $n(z)$, $N(z)$ and $z=y+\epsilon$ into $D_k$ we obtain

\begin{equation*}
D_k(a,d)=n(y)\frac{(2y^3a+d)^2}{(y+\epsilon)^2}\frac{H^2(y)}{H^2(y+\epsilon)}\epsilon + \frac{P^{(2)}d^2+P^{(1)}d+P^{(0)}}{H^2(y+\epsilon)y^2(y+\epsilon)^2}\epsilon^2
\end{equation*}

Here $P^{(2)}$, $P^{(1)}$ and $P^{(0)}$ are polynomials composed of $\epsilon$, $y$, $n(y)$, $N(y)$, $\nu(y,\epsilon)$, $\nabla(y,\epsilon)$ and $a$ terms. Importantly, the symbolic expression of these polynomials is not dependent on $K$ or $k$.

Now, we re-introduce the $k$ subscripts as we aggregate the values of the $D_k$ for our fixed $K$. We obtain

{\small
\begin{equation*}
objdiff = \frac{2}{I_2}\sum_{k=0}^{K-1}D_k=\frac{2}{I_2}\Big(D_0 + \sum_{k=1}^{K-1}n(y_k)\frac{(2y_k^3a_k+d_k)^2}{(y_k+\epsilon)^2}\frac{H^2(y_k)}{H^2(y_k+\epsilon)}\epsilon + \frac{P^{(2)}d_k^2+P^{(1)}d_k+P^{(0)}}{H^2(y_k+\epsilon)y_k^2(y_k+\epsilon)^2}\epsilon^2\Big)
\end{equation*}
}

Again, we note that the three polynomials do not have $k$ subscripts because their symbolic expression does not depend on $k$. This is the alternative expression of $objdiff$ we were looking for in step 1.

Step 2: Now $K$ is no longer considered fixed. Let $C$ be a positive integer and assume that $K$ is a multiple of $C$. Then we break down $objdiff(K)$ into $\frac{2}{I_2}\sum_{k=0}^{\frac{K}{C}-1}D_k$ and $\frac{2}{I_2}\sum_{k=\frac{K}{C}}^{K-1}D_k$. 

Let's look at $\frac{2}{I_2}\sum_{k=\frac{K}{C}}^{K-1}D_k$ first, and let's do so for a fixed value of $C$. We have $y_k>\frac{X}{C}$, so by lemma \ref{lemma6}, $H(y_k+\epsilon)$ is bounded away from zero across all $K$ and $k \ge \frac{K}{C}$, and so is $y_k$ and $y_k+\epsilon$. All terms that make up the polynomials $P^{(0)},P^{(1)},P^{(2)}$ are bounded except possibly $a_k$. Hence, if we can show that $a_k$ is also bounded, then the polynomials are bounded. $a_k$ and $r_k$ are determined via the system

\begin{equation*}
\begin{pmatrix}I_4(0,y_k)&I_2(0,y_k)\\y_k^3&y_k\end{pmatrix}\begin{pmatrix}a_k\\r_k \end{pmatrix}=\begin{pmatrix}\int_0^{y_k}xf_{2K}(x)d\mu_x\\f_{2K}(y_k) \end{pmatrix}
\end{equation*}

Since $f$ is continuous, $f$ is bounded on $[0,X]$, so the right-hand side is bounded. Since $y_k$ is bounded, all entries in the 2 by 2 matrix are bounded. So if the determinant of the matrix is bounded away from zero, then $a_k$ and $r_k$ are bounded. The determinant is $y_kI_4(0,y_k)-y_k^3I_2(0,y_k)=y_kH(y_k)$. Since $y_k\ge\frac{X}{C}$ is bounded away from zero, so are both terms in this product by lemma \ref{lemma6}. Hence $a_k$ is indeed bounded. So we have for some constants $B_2,B_1,B_0$

\begin{eqnarray*}
&&\Big|\frac{2}{I_2}\sum_{k=\frac{K}{C}}^{K-1}\frac{P^{(2)}d_k^2+P^{(1)}d_k+P^{(0)}}{H^2(y_k+\epsilon)y_k^2(y_k+\epsilon)^2}\epsilon^2\Big|\\
&\le&\frac{2}{I_2}\sum_{k=\frac{K}{C}}^{K-1}(d_k^2B_2+|d_k|B_1+B_0)\epsilon^2\\
&=&\frac{2B_2\epsilon}{I_2}\sum_{k=\frac{K}{C}}^{K-1}d_k^2\epsilon + \frac{2B_1\epsilon}{I_2}\sum_{k=\frac{K}{C}}^{K-1}|d_k|\epsilon + \frac{2B_0\epsilon}{I_2}\sum_{k=\frac{K}{C}}^{K-1}\epsilon\\
&=&\frac{2B_2\epsilon}{I_2}\sum_{k=\frac{K}{C}}^{K-1}(f(x_k)-b_kx_k)^2\epsilon + \frac{2B_1\epsilon}{I_2}\sum_{k=\frac{K}{C}}^{K-1}|f(x_k)-b_kx_k|\epsilon + \frac{2B_0\epsilon}{I_2}\sum_{k=\frac{K}{C}}^{K-1}\epsilon\\
&\le&\frac{2\epsilon}{I_2}\sum_{k=\frac{K}{C}}^{K-1}B_2x_k^2b_k^2\epsilon + \frac{2\epsilon}{I_2}\sum_{k=\frac{K}{C}}^{K-1}(2B_2|f(x_k)|x_k + B_1x_k)|b_k|\epsilon \\
&&+\frac{2\epsilon}{I_2}\sum_{k=\frac{K}{C}}^{K-1}(B_2f(x_k)^2+B_1|f(x_k)|+B_0)\epsilon
\end{eqnarray*}

Here we use $f(x_k) = f_{2K}(x_k) = b_kx_k+d_k$. We will now investigate each of the 3 terms in the above sum one after the other.

\begin{enumerate}
\item Remember that $b_k$ is defined to be the slope of $f_{2K}$ on $[\frac{k}{K}X,\frac{k+1}{K}X]$. By the intermediate value principle, $f'$ takes a value at least as large as $b_k$ and a value at least as small as $b_k$ on $[\frac{k}{K}X,\frac{k+1}{K}X]$. Hence, $f'^2$ takes a value at least as large as $b_k^2$ in $[\frac{k}{K}X,\frac{k+1}{K}X]$. Let such a value be $\beta_k^2$. Because $f'^2$ is integrable over the unit Gaussian measure over any finite interval, it is also integrable over the canonical Lebesgue measure over any finite interval. Hence we have $\int_\frac{X}{C}^Xf'^2d\mu_1=\lim_{K\rightarrow\infty}\sum_{k=\frac{K}{C}}^{K-1}\beta_k^2\epsilon$. So $\sum_{k=\frac{K}{C}}^{K-1}\beta_k^2\epsilon$ is bounded for sufficiently large $K$, so $\sum_{k=\frac{K}{C}}^{K-1}b_k^2\epsilon$ is, so $\sum_{k=\frac{K}{C}}^{K-1}B_2x_k^2b_k^2\epsilon$ is as $x_k^2<X^2$. And since $\epsilon$ converges to zero as $K$ converges to infinity, $\frac{2\epsilon}{I_2}\sum_{k=\frac{K}{C}}^{K-1}B_2x_k^2b_k^2\epsilon$ converges to zero.
\item Since $\sum_{k=\frac{K}{C}}^{K-1}b_k^2\epsilon$ is bounded for sufficiently large $K$, so is $\sqrt{\sum_{k=\frac{K}{C}}^{K-1}b_k^2\epsilon}$, so is $(\sum_{k=\frac{K}{C}}^{K-1}|b_k|\epsilon)(\sum_{k=\frac{K}{C}}^{K-1}\epsilon)^{-\frac{1}{2}}$ by Jensen's inequality, so is $\sum_{k=\frac{K}{C}}^{K-1}|b_k|\epsilon$. $f$ is bounded on $[0,X]$ as $f$ is continuous. Since also $x_k < X$, $\sum_{k=\frac{K}{C}}^{K-1}(2B_2|f(x_k)|x_k + B_1x_k)|b_k|\epsilon$ is bounded for sufficiently large $K$. And since $\epsilon$ converges to zero as $K$ converges to infinity, $\frac{2\epsilon}{I_2}\sum_{k=\frac{K}{C}}^{K-1}(2B_2|f(x_k)|x_k + B_1x_k)|b_k|\epsilon$ converges to zero.
\item Again, $f$ is bounded on $[0,X]$ and $x_k < X$. So $\frac{2\epsilon}{I_2}\sum_{k=\frac{K}{C}}^{K-1}(B_2f(x_k)^2+B_1|f(x_k)|+B_0)\epsilon<\frac{2\epsilon}{I_2}\sum_{k=\frac{K}{C}}^{K-1}B\epsilon=\frac{2\epsilon}{I_2}B(X-\frac{X}{C})$ for some $B$, which converges to zero.
\end{enumerate}

\sloppy So in summary, we have that for a fixed $C$ and $K$ restricted to multiples of $C$, $\Big|\frac{2}{I_2}\sum_{k=\frac{K}{C}}^{K-1}\frac{P^{(2)}d_k^2+P^{(1)}d_k+P^{(0)}}{H^2(y_k+\epsilon)y_k^2(y_k+\epsilon)^2}\epsilon^2\Big|$ converges to zero as $K$ converges to infinity.

Now let's look at $\frac{2}{I_2}\sum_{k=0}^{\frac{K}{C}-1}D_k$. We have $\frac{2}{I_2}\sum_{k=0}^{\frac{K}{C}-1}D_k=\frac{2}{I_2}\int_0^\frac{X}{C}(F'^2_0-3F^2_0) - (F'^2_\frac{K}{C}-3F^2_\frac{K}{C})d\mu_x=\frac{2}{I_2}\int_0^\frac{X}{C}F'^2_0+3F^2_\frac{K}{C}-3F^2_0 - F'^2_\frac{K}{C}d\mu_x$. The first two out of these four terms are non-negative. Since $f$ is bounded on $[0,X]$, so is $F_0$. So $\int_0^\frac{X}{C}-3F^2_0d\mu_x$ converges to zero as $C$ converges to infinity. Finally, we turn to the $\int_0^\frac{X}{C}- F'^2_\frac{K}{C}d\mu_x$ term. On $[0,\frac{X}{C}]$, $F_\frac{K}{C}$ is an odd cubic. Denote it by $ax^3+rx$. Setting $\delta=\frac{X}{C}$, $a$ and $r$ are determined via the system 

\begin{equation*}
\begin{pmatrix}I_4(0,\delta)&I_2(0,\delta)\\\delta^3&\delta\end{pmatrix}\begin{pmatrix}a\\r \end{pmatrix}=\begin{pmatrix}\int_0^\delta xf_{2K}(x)d\mu_x\\f_{2K}(\delta) \end{pmatrix}
\end{equation*}

So we have

\begin{eqnarray*}
&&\int_0^\delta- F'^2_\frac{K}{C} d\mu_x\\
&=&-9a^2I_4(0,\delta)-6acI_2(0,\delta)-c^2I_0(0,\delta)\\
&=&\frac{-9(\delta\int_0^\delta xf_{2K}d\mu_x-I_2(0,\delta)f_{2K}(\delta))^2I_4(0,\delta)...}{(I_4(0,\delta)\delta-I_2(0,\delta)\delta^3)^2}\\
&&\frac{...-(-\delta^3\int_0^\delta xf_{2K}d\mu_x+I_4(0,\delta)f_{2K}(\delta))^2I_0(0,\delta)...}{(I_4(0,\delta)\delta-I_2(0,\delta)\delta^3)^2}\\
&&\frac{...-6(\delta\int_0^\delta xf_{2K}d\mu_x-I_2(0,\delta)f_{2K}(\delta))(-\delta^3\int_0^\delta xf_{2K}d\mu_x+I_4(0,\delta)f_{2K}(\delta))I_2(0,\delta)}{(I_4(0,\delta)\delta-I_2(0,\delta)\delta^3)^2}
\end{eqnarray*}

The denominator is clearly non-zero. By lemma \ref{lemma7}, we have that all values of $f$ are within $o(\delta^\frac{1}{2})$ of $f(0)$ on the interval $[0,\delta]$. Since $K$ is a multiple of $C$, we have $\frac{X}{K} \le \frac{X}{C} =\delta$. Therefore, if the domain of a line segment in $f_{2K}$ intersects $[0,\delta]$, that domain is contained in $[0,2\delta]$. Therefore, $f_{2K}$ can only take values in $[0,\delta]$ that $f$ takes in $[0,2\delta]$. Therefore, all values of $f_{2K}$ are also within $o(\delta^\frac{1}{2})$ of $f(0)$ on the interval $[0,\delta]$, uniformly over $K$. Because $f$ is odd, $f(0)=0$ and hence $f_{2K}(0)=0$. Hence, $f_{2K}(\delta)=o(\delta^\frac{1}{2})$ and $\int_0^\delta xf_{2K}(x)d\mu_x=o(\delta^\frac{5}{2})$. We can write $I_0(0,\delta)$, $I_2(0,\delta)$ and $I_4(0,\delta)$ in terms of $\delta$, $n(\delta)$ and $N(\delta)$ and then replace $n$ and $N$ by their Taylor expansion about zero including terms up to $\delta^7$, using that both $n$ and $N$ are Lipschitz differentiable on $[0,X]$ arbitrarily many times. This yields that the numerator of the big ratio is $o(\delta^{12})$ and the denominator $\Theta(\delta^{12})$. So $\int_0^\delta- F'^2_\frac{K}{C} d\mu_x$ converges to zero as $\delta$ converges to zero, i.e. as $C$ converges to infinity.

This is the last piece of the puzzle. Now we put it all together. For some $C$ and $K$ a multiple of $C$ we have

\begin{eqnarray*}
&&objdiff(K)\\
&=&\frac{2}{I_2}\sum_{k=0}^{K-1}\int_0^{\frac{(k+1)}{K}X}(F'^2_k - 3F^2_k) - (F'^2_{k+1} - 3F^2_{k+1})d\mu_x\\
&=&\frac{2}{I_2}\Bigg(\Big(\int_0^{\frac{X}{C}}F'^2_0d\mu_x\Big) - 3\Big(\int_0^{\frac{X}{C}}F^2_0d\mu_x\Big) - \Big(\int_0^{\frac{X}{C}}F'^2_\frac{K}{C}d\mu_x\Big) + 3\Big(\int_0^{\frac{X}{C}}F^2_\frac{K}{C}d\mu_x\Big)\\
&&+n(y_k)\frac{(2y_k^3a_k+d_k)^2}{(y_k+\epsilon)^2}\frac{H^2(y_k)}{H^2(y_k+\epsilon)}\epsilon+\frac{P^{(2)}d_k^2+P^{(1)}d_k+P^{(0)}}{H^2(y_k+\epsilon)y_k^2(y_k+\epsilon)^2}\epsilon^2\Bigg)
\end{eqnarray*}

Out of the six terms, the first, fourth and fifth are non-negative. We have shown that as $C$ converges to infinity the second and third terms converge to zero. And we have shown that for a fixed $C$ as $K$ converges to infinity, the sixth term converges to zero. Therefore, no matter the value of $\eta$, we can first choose a $C$ and then a $K > K_e$ which is a multiple of $C$ such that $objdiff(K)$ is greater than $-\eta$, as required. The constant $\frac{2}{I_2}$ has no influence on the analysis. Step 2 is complete. Hence, we have claim 1. Hence, stage 2 is complete.

\underline{Stage 3}: The statement holds when $d_\text{in}=d_\text{out}=1$, $\mathcal{D}$ has expectation zero and variance one and the constant component of $f$ is the zero function.

Since the linear and constant component of $f$ are both the zero function, we have $f = \tilde{f}$. By proposition \ref{finiteNetBasisOrtho} part 1, we have $\mathbb{E}f=0$ and $\mathbb{E}xf=0$.

Let $g(x)=\frac{f(x)+f(-x)}{2}$ and $h(x)=\frac{f(x)-f(-x)}{2}$. Then $g$ is even, $h$ is odd and $f=g+h$. Because $h$ is odd, $\mathbb{E}h=0$, so also $\mathbb{E}g=0$. Because $g$ is even, $\mathbb{E}xg=0$, so also $\mathbb{E}xh=0$. So by lemma \ref{lemma3}, we have $A_g = A_h = 0$ and $b_g=b_h=0$.

Assume $g$ is not the zero function. Because $\mathbb{E}g = 0$, $g$ is not the constant function. So by proposition \ref{finiteNetPositiveDenominator}, $\Tr(\Cov_g) > 0$. So we can apply stage 1 to $g$ and obtain $NLC(g,\mathcal{D}) \ge \sqrt{2}$. Similarly, if $h$ is not the zero function, $\Tr(\Cov_h) > 0$ and $NLC(h,\mathcal{D}) \ge \sqrt{2}$. 

If either $g$ or $h$ is the zero function, $f$ is equal to $g$ or $h$, so we have $NLC(f,\mathcal{D}) \ge \sqrt{2}$ as desired. So now assume neither $g$ nor $h$ is the zero function. Because $g$ is even and $h$ is odd, we have $\mathbb{E}gh=\mathbb{E}g'h'=0$. So we have

\begin{eqnarray*}
&&NLC(f,\mathcal{D})^2\\
&=&\frac{\mathbb{E}f'^2}{\mathbb{E}f^2 + (\mathbb{E}f)^2}\\
&=&\frac{\mathbb{E}f'^2}{\mathbb{E}f^2}\\
&=&\frac{\mathbb{E}(g+h)'^2}{\mathbb{E}(g+h)^2}\\
&=&\frac{\mathbb{E}g'^2+2\mathbb{E}g'h'+\mathbb{E}h'^2}{\mathbb{E}g^2+2\mathbb{E}gh+\mathbb{E}h^2}\\
&=&\frac{\mathbb{E}g'^2+\mathbb{E}h'^2}{\mathbb{E}g^2+\mathbb{E}h^2}\\
&=&\frac{\mathbb{E}g'^2+\mathbb{E}h'^2}{\mathbb{E}g^2-(\mathbb{E}g)^2+\mathbb{E}h^2-(\mathbb{E}h)^2}\\
&=&\frac{NLC(g,\mathcal{D})^2\Tr(\Cov_g)+NLC(h,\mathcal{D})^2\Tr(\Cov_h)}{\Tr(\Cov_g)+\Tr(\Cov_h)}
\end{eqnarray*}

So $NLC(f,\mathcal{D})^2$ is a weighted average of two values which are greater or equal to 2. Therefore $NLC(f,\mathcal{D})\ge\sqrt{2}$.

\underline{Stage 4}: The statement holds when $d_\text{out}=1$, $\mathcal{D}$ has expectation zero and identity covariance and the constant component of $f$ is the zero function.

We will proceed by induction on $d_\text{in}$. The case $d_\text{in}=1$ is equivalent to stage 3. Now fix $d_\text{in}>1$ and assume that the statement holds for all lower input dimensionalities. In this stage only, the letters $i$, $j$, $k$, $r$ and $s$ are all used to index input dimensions and range from 0 to $d_\text{in} - 1$.

Since the linear and constant component of $f$ are both the zero function, we have $f = \tilde{f}$. By proposition \ref{finiteNetBasisOrtho} part 1, we have $\mathbb{E}f=0$ and $\mathbb{E}x[i]f=0$ for all $i$. Because $\mathcal{D}$ is the unit Gaussian, we have $\mathbb{E}x[i] = 0$ and $\mathbb{E}x[i]^2 = 1$ for all $i$. These formulas will be used repeatedly throughout this stage.

Define $f_i(x[i]) = \mathbb{E}_{x[i'], i' \neq i}f$, $T_{i,k}(x[i]) = (\mathbb{E}_{x[i'], i' \neq i}fx[k]) - (\mathbb{E}fx[i]x[k])x[i]$, $t_{i,k}(x[i],x[k]) = T_{i,k}x[k]$, $P_{i,k}=\mathbb{E}fx[i]x[k]$, $p_{i,k}(x[i],x[k]) = P_{i,k}x[i]x[k]$, $\hat{f} = f - \sum_i f_i - \sum_{i \neq k}t_{i,k} - \sum_{i > k}p_{i,k}$ and $F_i = \hat{f} + \sum_{j \neq i}f_j + \sum_{j \neq i, k \neq j}t_{j,k}$.

{\it Claim 1}: $\mathbb{E}T_{i,k}x[j]=0$

If $i \neq j$, we have $\mathbb{E}T_{i,k}x[j] = (\mathbb{E}_{x[i]}T_{i,k})(\mathbb{E}_{x[j]}x[j]) = 0$. So now assume $i=j$. We have

\begin{eqnarray*}
&&\mathbb{E}T_{i,k}x[j]\\
&=&\mathbb{E}((\mathbb{E}_{x[i'], i' \neq i}fx[k]) - (\mathbb{E}fx[i]x[k])x[i])x[i]\\
&=&(\mathbb{E}\mathbb{E}_{x[i'], i' \neq i}fx[k]x[i]) - (\mathbb{E}fx[i]x[k])\mathbb{E}x[i]^2\\
&=&\mathbb{E}fx[i]x[k] - \mathbb{E}fx[i]x[k]\\
&=&0
\end{eqnarray*}

{\it Claim 2:} $\hat{f}$, the $f_i$, the $t_{i,k}$ with $i \neq k$ and the $p_{i,k}$ with $i > k$ form an orthogonal decomposition of $f$ under $\mathcal{D}$. 

We will verify that each pair of distinct components is indeed orthogonal.

{\it Claim 2a}: $\mathbb{E}f_if_k = 0$ when $i \neq k$

\begin{eqnarray*}
&&\mathbb{E}f_if_k\\
&=&\mathbb{E}(\mathbb{E}_{x[i'], i' \neq i}f)(\mathbb{E}_{x[i'], i' \neq k}f)\\
&=&(\mathbb{E}f)(\mathbb{E}f)\\
&=&0
\end{eqnarray*}

{\it Claim 2b}: $\mathbb{E}t_{i,k}t_{j,l} = 0$ when $i \neq k$, $j \neq l$ and $(i,k) \neq (j,l)$

Assume $k \neq j$ and $k \neq l$. Then we have

\begin{eqnarray*}
&&\mathbb{E}t_{i,k}t_{j,l}\\
&=&\mathbb{E}T_{i,k}x[k]T_{j,l}x[l]\\
&=&(\mathbb{E}_{x[i'],i'\neq k}T_{i,k}T_{j,l}x[l])(\mathbb{E}_{x[k]}x[k])\\
&=&0
\end{eqnarray*}

We obtain the same result with $l \neq i$ and $l \neq k$.

Now assume $k=l$ and $i\neq j$. We have

\begin{eqnarray*}
&&\mathbb{E}t_{i,k}t_{j,l}\\
&=&\mathbb{E}T_{i,k}x[k]T_{j,k}x[k]\\
&=&(\mathbb{E}_{x[i]}T_{i,k})(\mathbb{E}_{x[j]}T_{j,k})(\mathbb{E}_{x[k]}x[k]^2)\\
&=&(\mathbb{E}_{x[i]}((\mathbb{E}_{x[i'],i'\neq i}fx[k]) - (\mathbb{E}fx[i]x[k])x[i]))(\mathbb{E}_{x[j]}T_{j,k})\\
&=&((\mathbb{E}fx[k]) - (\mathbb{E}fx[i]x[k])(\mathbb{E}_{x[i]}x[i]))(\mathbb{E}_{x[j]}T_{j,k})\\
&=&0
\end{eqnarray*}

Finally, assume $k=j$ and $i=l$. We have

\begin{eqnarray*}
&&\mathbb{E}t_{i,k}t_{j,l}\\
&=&\mathbb{E}T_{i,k}T_{k,i}x[i]x[k]\\
&=&(\mathbb{E}_{x[i]}T_{i,k}x[i])(\mathbb{E}_{x[k]}T_{k,i}x[k])\\
&=&0
\end{eqnarray*}

using claim 1.

{\it Claim 2c}: $\mathbb{E}p_{i,k}p_{j,l} = 0$ when $i > k$, $l > j$ and $(i,k) \neq (j,l)$

Because of the constraints on $i$, $j$, $k$ and $l$, at least one of the four indices must have a value unequal to the other 3. Assume $i$ has a unique value.

\begin{eqnarray*}
&&\mathbb{E}p_{i,k}p_{j,l}\\
&=&P_{i,k}P_{j,l}\mathbb{E}x[i]x[j]x[k]x[l]\\
&=&P_{i,k}P_{j,l}(\mathbb{E}_{x[i'], i' \neq i}x[j]x[k]x[l])(\mathbb{E}_{x[i]}x[i])\\
&=&0
\end{eqnarray*}

If $j$, $k$ or $l$ has a unique value, the derivation is analogous.

{\it Claim 2d}: $\mathbb{E}f_jt_{i,k} = 0$ when $i \neq k$

If $j\neq k$, we have $\mathbb{E}f_jt_{i,k} = (\mathbb{E}_{x[i'], i' \neq k}f_jT_{i,k})(\mathbb{E}_{x[k]}x[k]) = 0$. If $j=k$, we have $\mathbb{E}f_jt_{i,k} = \mathbb{E}f_kT_{i,k}x[k] = (\mathbb{E}_{x[k]}f_kx[k])(\mathbb{E}_{x[i]}T_{i,k}) = (\mathbb{E}fx[k])(\mathbb{E}_{x[i]}T_{i,k}) = 0$.

{\it Claim 2e}: $\mathbb{E}f_jp_{i,k} = 0$ when $i \neq k$

We have either $j \neq i$ or $j \neq k$. By symmetry, WLOG, $j \neq k$. Then $\mathbb{E}f_jp_{i,k} = P_{i,k}\mathbb{E}f_jx[i]x[k] = P_{i,k}(\mathbb{E}_{x[k]}x[k])(\mathbb{E}_{x[i'],i' \neq k}f_jx[i]) = 0$.

{\it Claim 2f}: $\mathbb{E}t_{i,k}p_{j,l} = 0$ when $i \neq k$, $j \neq l$.

Assume $k \neq j$ and $k \neq l$. Then $\mathbb{E}t_{i,k}p_{j,l} = P_{j,l}(\mathbb{E}_{x[i'],i' \neq k}T_{i,k}x[j]x[l])(\mathbb{E}_{x[k]}x[k]) = 0$. Now assume $k = j$ or $k = l$. By symmetry, WLOG, $k = l$. Then $\mathbb{E}t_{i,k}p_{j,l} = P_{j,k}(\mathbb{E}_{x[k]}x[k]^2)(\mathbb{E}_{x[i'],i' \neq k}T_{i,k}x[j]) =P_{j,k} \mathbb{E}T_{i,k}x[j] = 0$ by claim 1.

{\it Claim 2g}: $\mathbb{E}\hat{f}f_j = 0$

Going forward, we utilize claims 2a through 2f.

\begin{eqnarray*}
&&\mathbb{E}\hat{f}f_j\\
&=&\mathbb{E}(f - \sum_i f_i - \sum_{i \neq k}t_{i,k} - \sum_{i > k}p_{i,k})f_j\\
&=&\mathbb{E}ff_j - \sum_{i \neq j}\mathbb{E}f_if_j - \mathbb{E}f_j^2 - \sum_{i \neq k}\mathbb{E}t_{i,k}f_j - \sum_{i > k}\mathbb{E}p_{i,k}f_j\\
&=&\mathbb{E}ff_j - \mathbb{E}f_j^2\\
&=&\mathbb{E}_{x[j]}(\mathbb{E}_{x[i'], i' \neq j}f)f_j - \mathbb{E}_{x[j]}f_j^2\\
&=&\mathbb{E}_{x[j]}f_j^2 - \mathbb{E}_{x[j]}f_j^2\\
&=&0
\end{eqnarray*}

{\it Claim 2h}: $\mathbb{E}\hat{f}t_{j,l} = 0$ when $j\neq l$

Claim 1 yields $\mathbb{E}_{x[j]}T_{j,l}x[j] = 0$. So we have

\begin{eqnarray*}
&&\mathbb{E}\hat{f}t_{j,l}\\
&=&\mathbb{E}(f - \sum_i f_i - \sum_{i \neq k}t_{i,k} - \sum_{i > k}p_{i,k})t_{j,l}\\
&=&\mathbb{E}ft_{j,l} - \sum_i\mathbb{E}f_it_{j,l} - \sum_{i \neq k, (i,k)\neq(j,l)}\mathbb{E}t_{i,k}t_{j,l} - \mathbb{E}t_{j,l}^2 - \sum_{i > k}\mathbb{E}p_{i,k}t_{j,l}\\
&=&\mathbb{E}ft_{j,l} - \mathbb{E}t_{j,l}^2\\
&=&\mathbb{E}ft_{j,l} - (\mathbb{E}fx[l]x[j])(\mathbb{E}_{x[j]}T_{j,l}x[j]) - \mathbb{E}t_{j,l}^2\\
&=&\mathbb{E}fT_{j,l}x[l] - \mathbb{E}_{x[j]}T_{j,l}(\mathbb{E}fx[l]x[j])x[j] - \mathbb{E}T_{j,l}^2x[l]^2\\
&=&\mathbb{E}_{x[j]}T_{j,l}(\mathbb{E}_{x[i'], i' \neq j}fx[l]) - \mathbb{E}_{x[j]}T_{j,l}(\mathbb{E}fx[l]x[j])x[j] - (\mathbb{E}_{x[j]}T_{j,l}^2)(\mathbb{E}_{x[l]}x[l]^2)\\
&=&\mathbb{E}_{x[j]}T_{j,l}((\mathbb{E}_{x[i'], i' \neq j}fx[l]) - (\mathbb{E}fx[l]x[j])x[j]) - (\mathbb{E}_{x[j]}T_{j,l}^2)\\
&=&(\mathbb{E}_{x[j]}T_{j,l}^2) - (\mathbb{E}_{x[j]}T_{j,l}^2)\\
&=&0
\end{eqnarray*}

{\it Claim 2i}: $\mathbb{E}\hat{f}p_{j,l} = 0$ when $j > l$

\begin{eqnarray*}
&&\mathbb{E}\hat{f}p_{j,l}\\
&=&\mathbb{E}(f - \sum_i f_i - \sum_{i \neq k}t_{i,k} - \sum_{i > k}p_{i,k})p_{j,l}\\
&=&\mathbb{E}fp_{j,l} - \sum_i\mathbb{E}f_ip_{j,l} - \sum_{i \neq k}\mathbb{E}t_{i,k}p_{j,l} - \sum_{i > k, (i,k) \neq (j,l)}\mathbb{E}p_{i,k}p_{j,l} - \mathbb{E}p_{j,l}^2\\
&=&\mathbb{E}fp_{j,l} - \mathbb{E}p_{j,l}^2\\
&=&P_{j,l}(\mathbb{E}fx[j]x[l]) - P_{j,l}^2\mathbb{E}x[j]^2x[l]^2\\
&=&P_{j,l}^2 - P_{j,l}^2\\
&=&0
\end{eqnarray*}

Claims 2a through 2i together yield Claim 2.

Let $d_r$ be the conditional density function of $x[r]$ given $x[i']$, $i' \neq r$. Because $\mathcal{D}$ is the unit Gaussian, $d_r$ is the density function of the one-dimensional unit Gaussian.

{\it Claim 3}: $\mathbb{E}\frac{dF_i}{dx[r]}x[s]=0$ when $r\neq s$

\begin{eqnarray*}
&&\mathbb{E}\frac{dF_i}{dx[r]}x[s]\\
&=&\mathbb{E}_{x[i'], i'\neq r}\Big(\int_{x[r]=-\infty}^\infty\frac{dF_i}{dx[r]}d_rd\mu_1\Big)x[s]\\
&=&\mathbb{E}_{x[i'], i'\neq r}\Big([F_id_r]_{x[r]=-\infty}^\infty - \int_{x[r]=-\infty}^\infty F_i\frac{dd_r}{dx[r]}d\mu_1\Big)x[s]\\
&=&\mathbb{E}_{x[i'], i'\neq r}\Big(\int_{x[r]=-\infty}^\infty F_ix[r]d_rd\mu_1\Big)x[s]\\
&=&\mathbb{E}F_ix[r]x[s]\\
\end{eqnarray*}

Here, we use integration by parts on a 1-dimensional subspace. This is allowed because $d_r$ is absolutely continuous and $\frac{dF_i}{dx[r]}$ is integrable. Because $r\neq s$, either $i\neq r$ or $i \neq s$. Assume $i\neq r$. Continuing the above chain of equations, we have

\begin{eqnarray*}
&=&\mathbb{E}F_ix[r]x[s]\\
&=&\mathbb{E}(f - f_i - \sum_{k \neq i}T_{i,k}x[k] - \sum_{j>k}P_{j,k}x[j]x[k])x[r]x[s]\\
&=&\mathbb{E}fx[r]x[s] - (\mathbb{E}_{x[i'],i'\neq r}f_ix[s])(\mathbb{E}_{x[r]}x[r]) - \sum_{k \neq i, k\neq r}(\mathbb{E}_{x[i'], i' \neq r}T_{i,k}x[k]x[s])(\mathbb{E}_{x[r]}x[r])\\
&& - (\mathbb{E}_{x[i'], i' \neq r}T_{i,r}x[s])(\mathbb{E}_{x[r]}x[r]^2) \\
&& - \sum_{j>k,\{j,k\}\neq\{r,s\}}P_{j,k}\mathbb{E}x[j]x[k]x[r]x[s] - P_{r,s}(\mathbb{E}_{x[r]}x[r]^2)(\mathbb{E}_{x[s]}x[s]^2)\\
&=&\mathbb{E}fx[r]x[s] - \mathbb{E}_{x[i'], i' \neq r}T_{i,r}x[s] - P_{r,s}\\
&=&- \mathbb{E}T_{i,r}x[s]\\
&=&0
\end{eqnarray*}

The last step uses claim 1. If $i\neq s$, the derivation is analogous.

{\it Claim 4}: $\mathbb{E}_{x[i'], i' \neq i}F_i$ is the zero function of $x[i]$

\begin{eqnarray*}
&&\mathbb{E}_{x[i'], i' \neq i}F_i\\
&=&\mathbb{E}_{x[i'], i' \neq i}f - \mathbb{E}_{x[i'], i' \neq i}f_i - \mathbb{E}_{x[i'], i' \neq i}\sum_{k \neq i} T_{i,k}x[k] - \mathbb{E}_{x[i'], i' \neq i}\sum_{j>k}P_{j,k}x[j]x[k]\\
&=&\mathbb{E}_{x[i'], i' \neq i}f - \mathbb{E}_{x[i'], i' \neq i}f - \sum_{k \neq i} (\mathbb{E}_{x[i'], i' \neq i,k}T_{i,k})(\mathbb{E}_{x[k]}x[k]) - \sum_{j>k}P_{i,k}(\mathbb{E}_{x[i'], i' \neq i}x[j]x[k])\\
&=&0
\end{eqnarray*}

{\it Claim 5:}  $\mathbb{E}_{x[i'], i' \neq i}F_ix[r]$ is the zero function of $x[i]$ when $r\neq i$

\begin{eqnarray*}
&&\mathbb{E}_{x[i'], i' \neq i}F_ix[r]\\
&=&\mathbb{E}_{x[i'], i' \neq i}fx[r] - \mathbb{E}_{x[i'], i' \neq i}f_ix[r]\\
&& - \mathbb{E}_{x[i'], i' \neq i}\sum_{k \neq i} T_{i,k}x[k]x[r] - \mathbb{E}_{x[i'], i' \neq i}\sum_{j>k}P_{j,k}x[j]x[k]x[r]\\
&=&\mathbb{E}_{x[i'], i' \neq i}fx[r] - (\mathbb{E}_{x[i'], i' \neq i}f)(\mathbb{E}_{x[r]}x[r]) \\
&&- \mathbb{E}_{x[i'], i' \neq i}\sum_{k \neq i,k \neq r} T_{i,k}x[k]x[r]  - \mathbb{E}_{x[i'], i' \neq i} T_{i,r}x[r]^2 \\
&&- \mathbb{E}_{x[i'], i' \neq i}\sum_{j>k,\{j,k\}\neq\{i,r\}}P_{j,k}x[j]x[k]x[r] - \mathbb{E}_{x[i'], i' \neq i}P_{i,r}x[i]x[r]^2\\
&=&\mathbb{E}_{x[i'], i' \neq i}fx[r] - \sum_{k \neq i,k \neq r} (\mathbb{E}_{x[i'], i' \neq i,k}T_{i,k}x[r])(\mathbb{E}_{x[k]}x[k]) - T_{i,r}\\
&&- \sum_{j>k,\{j,k\}\neq\{i,r\}}P_{j,k}(\mathbb{E}_{x[i'], i' \neq i}x[j]x[k]x[r]) - P_{i,r}x[i]\\
&=&\mathbb{E}_{x[i'], i' \neq i}fx[r] - T_{i,r} - P_{i,r}x[i]\\
&=&\mathbb{E}_{x[i'], i' \neq i}fx[r] - \mathbb{E}_{x[i'], i' \neq i}fx[r] + (\mathbb{E}fx[i]x[r])x[i] - (\mathbb{E}fx[i]x[r])x[i]\\
&=& 0
\end{eqnarray*}

Let $F_i|x[i]$  and $\mathcal{D}|x[i]$ denote the function $F_i$ and distribution $\mathcal{D}$ restricted to the subspace corresponding to a specific value of $x[i]$ respectively. Since $\mathcal{D}$ is the unit Gaussian of dimensionality $d_\text{in}$, $\mathcal{D}|x[i]$ is the unit Gaussian of dimensionality $d_\text{in} - 1$. Claim 4 and 5, together with lemma \ref{lemma3}, yield that the least squares linear fit of $F_i|x[i]$ under $\mathcal{D}|x[i]$ is the zero function. Then proposition \ref{finiteNetPositiveDenominator} and the induction hypothesis yield $NLC(F_i|x[i], \mathcal{D}|x[i]) \ge \sqrt{2}$ and hence $\mathbb{E}_{x[i'], i' \neq i}F_i^2 \le \frac{1}{2}\mathbb{E}_{x[i'], i' \neq i}\sum_{i' \neq i}\Big(\frac{dF_i}{dx[i']}\Big)^2$ for all $x[i]$ where $F_i|x[i]$ is not the zero function. Of course, this inequality holds trivially for any $x[i]$ where $F_i|x[i]$ is the zero function, so the inequality holds regardless.

Because $\mathbb{E}_{x[i'], i' \neq i}x[j]=0$ for all $j\neq i$ and values of $x[i]$, we can further apply lemma \ref{lemma2} to $F_i|x[i]$ and $\mathcal{D}|x[i]$ to obtain $\mathbb{E}_{x[i'],i'\neq i}\frac{dF_i}{dx[j]}=0$ for all $j\neq i$.

Now we put all our results from throughout this stage together. Because $\mathcal{D}$ is the unit Gaussian, we have $NLC(f,\mathcal{D})^2 = \frac{\sum_i\mathbb{E}(\frac{df}{dx[i]})^2}{\mathbb{E}f^2}$. For the numerator, we have

\begin{eqnarray*}
&&\sum_i\mathbb{E}(\frac{df}{dx[i]})^2\\
&=&\sum_i\mathbb{E}(\frac{d\hat{f}}{dx[i]} + \sum_j\frac{df_j}{dx[i]} + \sum_{j\neq k}\frac{dt_{j,k}}{dx[i]} + \sum_{j> k}\frac{dp_{j,k}}{dx[i]})^2\\
&=&\frac{1}{d_\text{in}-1}\sum_i\sum_{i'\neq i}\mathbb{E}(\frac{d\hat{f}}{dx[i']} + \sum_j\frac{df_j}{dx[i']} + \sum_{j\neq k}\frac{dt_{j,k}}{dx[i']} + \sum_{j> k}\frac{dp_{j,k}}{dx[i']})^2\\
&=&\frac{1}{d_\text{in}-1}\sum_i\sum_{i'\neq i}\mathbb{E}(\frac{d\hat{f}}{dx[i']} + \sum_{j\neq i}\frac{df_j}{dx[i']} + \sum_{j\neq k}\frac{dt_{j,k}}{dx[i']} + \sum_{j> k}\frac{dp_{j,k}}{dx[i']})^2\\
&=&\frac{1}{d_\text{in}-1}\sum_i\sum_{i'\neq i}\mathbb{E}(\frac{dF_i}{dx[i']} + \sum_{k\neq i}\frac{dt_{i,k}}{dx[i']} + \sum_{j> k}\frac{dp_{j,k}}{dx[i']})^2\\
&=&\frac{1}{d_\text{in}-1}\sum_i\sum_{i'\neq i}\mathbb{E}(\frac{dF_i}{dx[i']} + \frac{dt_{i,i'}}{dx[i']} + \sum_{j \neq i'}\frac{dp_{j,i'}}{dx[i']})^2\\
&=&\frac{1}{d_\text{in}-1}\sum_i\sum_{i'\neq i}\mathbb{E}(\frac{dF_i}{dx[i']} + T_{i,i'} + \sum_{j \neq i'}P_{j,i'}x[j])^2\\
&=&\frac{1}{d_\text{in}-1}\sum_i\sum_{i'\neq i}\mathbb{E}\Big(\frac{dF_i}{dx[i']}^2 + T_{i,i'}^2 + \sum_{j \neq i'}P_{j,i'}^2x[j]^2 + 2\frac{dF_i}{dx[i']}T_{i,i'}\\
&&+ 2\frac{dF_i}{dx[i']}\sum_{j \neq i'}P_{j,i'}x[j] + 2T_{i,i'}\sum_{j \neq i'}P_{j,i'}x[j] + 2\sum_{j\neq i',k\neq i',j \neq k}P_{j,i'}x[j]P_{k,i'}x[k]\Big)\\
&=&\frac{1}{d_\text{in}-1}\sum_i\sum_{i'\neq i}\Big(\mathbb{E}\frac{dF_i}{dx[i']}^2 + \mathbb{E}T_{i,i'}^2 + \sum_{j \neq i'}P_{j,i'}^2\mathbb{E}x[j]^2 + 2\mathbb{E}\frac{dF_i}{dx[i']}T_{i,i'}\\
&& + 2\sum_{j \neq i'}P_{j,i'}\mathbb{E}\frac{dF_i}{dx[i']}x[j] + 2\sum_{j \neq i'}P_{j,i'}\mathbb{E}T_{i,i'}x[j] + 2\sum_{j\neq i',k\neq i',j \neq k}P_{j,i'}P_{k,i'}\mathbb{E}x[j]x[k]\Big)\\
&=&\frac{1}{d_\text{in}-1}\sum_i\sum_{i'\neq i}\Big(\mathbb{E}\frac{dF_i}{dx[i']}^2 + \mathbb{E}T_{i,i'}^2 + \sum_{j \neq i'}P_{j,i'}^2+ 2\mathbb{E}_{x[i]}(\mathbb{E}_{x[i''],i''\neq i}\frac{dF_i}{dx[i']})T_{i,i'}\Big)\\
&=&\frac{1}{d_\text{in}-1}\sum_i\sum_{i'\neq i}\Big(\mathbb{E}\frac{dF_i}{dx[i']}^2 + \mathbb{E}T_{i,i'}^2 + \sum_{j \neq i'}P_{j,i'}^2\Big)\\
&=&\frac{1}{d_\text{in}-1}\sum_i\sum_{i'\neq i}\mathbb{E}(\frac{dF_i}{dx[i']}^2 + T_{i,i'}^2) + 2\sum_{j > k}P_{j,k}^2\\
&\ge&\frac{1}{d_\text{in}-1}\sum_i\sum_{i'\neq i}\mathbb{E}\frac{dF_i}{dx[i']}^2 + 2\sum_{j > k}P^2_{j,k}
\end{eqnarray*}

For the denominator, we have

\begin{eqnarray*}
&&\mathbb{E}f^2\\
&=&\mathbb{E}(\hat{f}+\sum_jf_j + \sum_{j\neq k}t_{j,k} + \sum_{j>k}p_{j,k})^2\\
&=&\mathbb{E}(\hat{f}^2+\sum_jf_j^2 + \sum_{j\neq k}t_{j,k}^2 + \sum_{j>k}p_{j,k}^2)\\
&\le&\mathbb{E}(\hat{f}^2+\sum_jf_j^2 + \sum_{j\neq k}t_{j,k}^2 + \sum_{j>k}p_{j,k}^2 + \frac{1}{d_\text{in}-1}\hat{f}^2)\\
&=&\frac{1}{d_\text{in}-1}\sum_i\mathbb{E}(\hat{f}^2+\sum_{j\neq i}f_j^2 + \sum_{j\neq k, j \neq i}t_{j,k}^2) + \mathbb{E}\sum_{j>k}p_{j,k}^2\\
&=&\frac{1}{d_\text{in}-1}\sum_i\mathbb{E}(\hat{f}+\sum_{j\neq i}f_j + \sum_{j\neq k, j \neq i}t_{j,k})^2 + \sum_{j>k}P_{j,k}^2\\
&=&\frac{1}{d_\text{in}-1}\sum_i\mathbb{E}F_i^2 + \sum_{j>k}P_{j,k}^2\\
&=&\frac{1}{d_\text{in}-1}\sum_i\mathbb{E}_{x[i]}(\mathbb{E}_{x[i'],i'\neq i}F_i^2) + \sum_{j>k}P_{j,k}^2\\
&\le&\frac{1}{d_\text{in}-1}\sum_i\mathbb{E}_{x[i]}\Big(\mathbb{E}_{x[i'],i'\neq i}\frac{1}{2}\sum_{i'\neq i}\frac{dF_i}{dx[i']}^2\Big) + \sum_{j>k}P_{j,k}^2\\
&=&\frac{1}{2(d_\text{in}-1)}\sum_i\sum_{i'\neq i}\mathbb{E}\frac{dF_i}{dx[i']}^2 + \sum_{j>k}P_{j,k}^2
\end{eqnarray*}

Hence the numerator is at least twice the denominator. Hence $NLC(f,\mathcal{D})^2 \ge 2$ and $NLC(f,\mathcal{D})\ge \sqrt{2}$ as required.

\underline{Stage 5}: The statement holds when $\mathcal{D}$ has expectation zero and identity covariance, and the constant component of $f$ is the zero function.

As in previous stages, we have $f=\tilde{f}$ and $\mathbb{E}f = 0$. Because the least squares linear fit to $f$ is the zero function, by definition, for each $j$, the least squares linear fit to $f[j]$ is the zero function. Also, by proposition \ref{finiteNetPositiveDenominator}, $f[j]$ is either the zero function or $\Tr(\Cov_{f[j]})>0$ and hence $\mathbb{E}f[j]^2 > 0$. If $f[j]$ is not the zero function, we can apply stage 4 to $f[j]$ to obtain $NLC(f[j],\mathcal{D})^2 = \frac{\sum_{i}\mathbb{E}\mathcal{J}[j,i]^2}{\mathbb{E}f[j]^2} \ge 2$. If $f[j]$ is the zero function, then $\mathcal{J}[j,i]$ is also the zero function for all $i$.

We have

\begin{eqnarray*}
&&NLC(f, \mathcal{D})^2\\
&=&\frac{\sum_{i,j}\mathbb{E}\mathcal{J}[j,i]^2}{\sum_j\mathbb{E}f[j]^2}\\
&=&\frac{\sum_{j : \text{$f[j]$ not the zero function}}(\sum_i\mathbb{E}\mathcal{J}[j,i]^2)}{\sum_{j : \text{$f[j]$ not the zero function}}\mathbb{E}f[j]^2}\\
&=&\frac{\sum_{j : \text{$f[j]$ not the zero function}}NLC(f[j],\mathcal{D})^2\mathbb{E}f[j]^2}{\sum_{j : \text{$f[j]$ not the zero function}}\mathbb{E}f[j]^2}
\end{eqnarray*}

So $NLC(f, \mathcal{D})^2$ is simply the weighted average of the $NLC(f[j],\mathcal{D})^2$ values corresponding to non-zero $f[j]$, all of which are at least 2. Note that at least one $f[j]$ is non-zero because we have $\Tr(\Cov_f)>0$ by assumption \ref{assumptionPositive}. Hence, $NLC(f, \mathcal{D})^2\ge 2$, and so $NLC(f, \mathcal{D})\ge \sqrt{2}$.

\underline{Stage 6}: The statement holds when $\mathcal{D}$ has expectation zero and the constant component of $f$ is the zero function.

Since the linear and constant component of $f$ are both the zero function, we have $f = \tilde{f}$. By proposition \ref{finiteNetBasisOrtho} part 1, we have $\mathbb{E}f=0$ and $\mathbb{E}x^Tf=0$. Let $C=\Cov_x^\frac{1}{2}$, which is valid by assumption \ref{assumptionNonSingular}. Consider $(f',\mathcal{D}')$ where $f'(x)=f(xC^T)$ and drawing $x$ from $\mathcal{D}'$ is equivalent to drawing $x$ from $\mathcal{D}$ and then post-multiplying $C^{-T}$. We have

\begin{eqnarray*}
&&\mathbb{E}_{x\sim\mathcal{D'}}f'(x)\\
&=&\mathbb{E}_{x\sim\mathcal{D}}f'(xC^{-T})\\
&=&\mathbb{E}_{x\sim\mathcal{D}}f(xC^{-T}C^T)\\
&=&\mathbb{E}_{x\sim\mathcal{D}}f(x)\\
&=&0
\end{eqnarray*}

and

\begin{eqnarray*}
&&\mathbb{E}_{x\sim\mathcal{D}'}x^Tf'(x)\\
&=&\mathbb{E}_{x\sim\mathcal{D}}C^{-1}x^Tf'(xC^{-T})\\
&=&\mathbb{E}_{x\sim\mathcal{D}}C^{-1}x^Tf(xC^{-T}C^T)\\
&=&C^{-1}\mathbb{E}_{x\sim\mathcal{D}}x^Tf(x)\\
&=&0
\end{eqnarray*}

and

\begin{eqnarray*}
&&\mathbb{E}_{x\sim\mathcal{D'}}x\\
&=&\mathbb{E}_{x\sim\mathcal{D}}xC^{-T}\\
&=&(\mathbb{E}_{x\sim\mathcal{D}}x)C^{-T}\\
&=&0
\end{eqnarray*}

and

\begin{eqnarray*}
&&\Cov_{x\sim\mathcal{D}'}\\
&=&\mathbb{E}_{x\sim\mathcal{D'}}x^Tx\\
&=&\mathbb{E}_{x\sim\mathcal{D}}(xC^{-T})^T(xC^{-T})\\
&=&\mathbb{E}_{x\sim\mathcal{D}}C^{-1}x^TxC^{-T}\\
&=&C^{-1}\Cov_xC^{-T}\\
&=&I\\
\end{eqnarray*}

and

\begin{eqnarray*}
&&\Cov_{f'(x),x\sim\mathcal{D}'}\\
&=&\mathbb{E}_{x\sim\mathcal{D'}}(f'(x)-\mathbb{E}_{x'\sim\mathcal{D'}} f'(x'))^T(f'(x)-\mathbb{E}_{x'\sim\mathcal{D'}} f'(x'))\\
&=&\mathbb{E}_{x\sim\mathcal{D'}}f'(x)^Tf'(x)\\
&=&\mathbb{E}_{x\sim\mathcal{D}}f(xC^{-T}C^T)^Tf(xC^{-T}C^T)\\
&=&\mathbb{E}_{x\sim\mathcal{D}}f(x)^Tf(x)\\
&=&\Cov_f
\end{eqnarray*}

Further, we have $\mathcal{J}'(x)=\mathcal{J}(xC^T)C$. Then we have

\begin{eqnarray*}
&&NLC(f',\mathcal{D}')^2\\
&=&\frac{\mathbb{E}_{x\sim\mathcal{D}'}\Tr(\mathcal{J}'(x)\Cov_{x\sim\mathcal{D}'}\mathcal{J}'(x)^T)}{\Tr(\Cov_{f'(x),x\sim\mathcal{D}'})}\\
&=&\frac{\mathbb{E}_{x\sim\mathcal{D}'}\Tr(\mathcal{J}'(x)\mathcal{J}'(x)^T)}{\Tr(\Cov_f)}\\
&=&\frac{\mathbb{E}_{x\sim\mathcal{D}}\Tr(\mathcal{J}(xC^{-T}C^T)CC^T\mathcal{J}(xC^{-T}C^T)^T)}{\Tr(\Cov_f)}\\
&=&\frac{\mathbb{E}_{x\sim\mathcal{D}}\Tr(\mathcal{J}(x)\Cov_x\mathcal{J}(x)^T)}{\Tr(\Cov_f)}\\
&=&NLC(f,\mathcal{D})^2
\end{eqnarray*}

Because $\mathbb{E}_{x\sim\mathcal{D}'}f'=0$, $\mathbb{E}_{x\sim\mathcal{D}'}x^Tf'=0$ and lemma \ref{lemma3}, the least squares linear fit to $f'$ under $\mathcal{D}'$ is the zero function. Also, we have $\Tr(\Cov_{f'(x),x\sim\mathcal{D}'}) = \Tr(\Cov_f) > 0$, $\mathbb{E}_{x\sim\mathcal{D}'}x=0$ and $\Cov_{x\sim\mathcal{D}'}=I$. So we can use stage 5 to obtain $NLC(f',\mathcal{D}')\ge\sqrt{2}$, and thus $NLC(f,\mathcal{D})\ge\sqrt{2}$.

\underline{Stage 7}: The statement holds when the constant component of $f$ is the zero function.

Consider $(f',\mathcal{D}')$ where $f'(x)=f(x+\bar{x})$ and drawing $x$ from $\mathcal{D}'$ is equivalent to drawing $x$ from $\mathcal{D}$ and then subtracting $\bar{x}$. As in the previous stage, we have $\mathbb{E}f = 0$ and $\mathbb{E}x^Tf = 0$. We have

\begin{eqnarray*}
&&\mathbb{E}_{x\sim\mathcal{D}'}f'(x)\\
&=&\mathbb{E}_{x\sim\mathcal{D}}f'(x-\bar{x})\\
&=&\mathbb{E}_{x\sim\mathcal{D}}f(x-\bar{x}+\bar{x})\\
&=&\mathbb{E}_{x\sim\mathcal{D}}f(x)\\
&=&0
\end{eqnarray*}

and

\begin{eqnarray*}
&&\mathbb{E}_{x\sim\mathcal{D}'}x^Tf'(x)\\
&=&\mathbb{E}_{x\sim\mathcal{D}}(x-\bar{x})^Tf(x)\\
&=&\mathbb{E}_{x\sim\mathcal{D}}x^Tf(x) - \bar{x}^T\mathbb{E}_{x\sim\mathcal{D}}f(x)\\
&=&0
\end{eqnarray*}

and

\begin{eqnarray*}
&&\mathbb{E}_{x\sim\mathcal{D}'}x\\
&=&\mathbb{E}_{x\sim\mathcal{D}}(x-\bar{x})\\
&=&0
\end{eqnarray*}

and

\begin{eqnarray*}
&&\Cov_{x\sim\mathcal{D}'}\\
&=&\mathbb{E}_{x\sim\mathcal{D}'}(x-\mathbb{E}_{x'\sim\mathcal{D}'}x')^T(x-\mathbb{E}_{x'\sim\mathcal{D}'}x')\\
&=&\mathbb{E}_{x\sim\mathcal{D}'}x^Tx\\
&=&\mathbb{E}_{x\sim\mathcal{D}}(x-\bar{x})^T(x-\bar{x})\\
&=&\Cov_x\\
\end{eqnarray*}

and

\begin{eqnarray*}
&&\Cov_{f'(x),x\sim\mathcal{D}'}\\
&=&\mathbb{E}_{x\sim\mathcal{D'}}(f'(x)-\mathbb{E}_{x'\sim\mathcal{D'}} f'(x'))^T(f'(x)-\mathbb{E}_{x'\sim\mathcal{D'}} f'(x'))\\
&=&\mathbb{E}_{x\sim\mathcal{D'}}f'(x)^Tf'(x)\\
&=&\mathbb{E}_{x\sim\mathcal{D}}f(x-\bar{x} + \bar{x})^Tf(x-\bar{x} + \bar{x})\\
&=&\mathbb{E}_{x\sim\mathcal{D}}f(x)^Tf(x)\\
&=&\Cov_f
\end{eqnarray*}

Further, we have $\mathcal{J}'(x)=\mathcal{J}(x+\bar{x})$. Then we have

\begin{eqnarray*}
&&NLC(f',\mathcal{D}')^2\\
&=&\frac{\mathbb{E}_{x\sim\mathcal{D}'}\Tr(\mathcal{J}'(x)\Cov_{x\sim\mathcal{D}'}\mathcal{J}'(x)^T)}{\Tr(\Cov_{f'(x),x\sim\mathcal{D}'})}\\
&=&\frac{\mathbb{E}_{x\sim\mathcal{D}}\Tr(\mathcal{J}(x-\bar{x}+\bar{x})\Cov_x\mathcal{J}(x-\bar{x}+\bar{x})^T)}{\Tr(\Cov_f)}\\
&=&\frac{\mathbb{E}_{x\sim\mathcal{D}}\Tr(\mathcal{J}(x)\Cov_x\mathcal{J}(x)^T)}{\Tr(\Cov_f)}\\
&=&NLC(f,\mathcal{D})^2
\end{eqnarray*}

Because $\mathbb{E}_{x\sim\mathcal{D'}}f'=0$, $\mathbb{E}_{x\sim\mathcal{D'}}x^Tf'=0$ and lemma \ref{lemma3}, the least squares linear fit to $f'$ under $\mathcal{D'}$ is the zero function. Also, we have $\Tr(\Cov_{f'(x),x\sim\mathcal{D}'}) = \Tr(\Cov_f) > 0$ and $\mathbb{E}_{x\sim\mathcal{D'}}x=0$. So we can use stage 6 to obtain $NLC(f',\mathcal{D}')\ge\sqrt{2}$, and thus $NLC(f,\mathcal{D})\ge\sqrt{2}$.

\underline{Stage 8}: The statement holds under the general conditions of theorem \ref{finiteNetTilde}.

We can write $f = \tilde{f} + b_f$. By lemma \ref{finiteNetBasisOrtho}, we have $\mathbb{E}\tilde{f}=\mathbb{E}x^T\tilde{f}=0$. By lemma \ref{lemma3}, the least squares linear fit to $\tilde{f}$ is the zero function. We also have

\begin{eqnarray*}
&&\Cov_f\\
&=&\mathbb{E}_x(\tilde{f}(x) + b_f - \mathbb{E}_{x'}(\tilde{f}(x') + b_f))^T(\tilde{f}(x) + b_f - \mathbb{E}_{x'}(\tilde{f}(x') + b_f))\\
&=&\mathbb{E}_x(\tilde{f}(x) - \mathbb{E}_{x'}\tilde{f}(x'))^T(\tilde{f}(x) - \mathbb{E}_{x'}\tilde{f}(x'))\\
&=&\Cov_{\tilde{f}}
\end{eqnarray*}

and hence $\Tr(\Cov_{\tilde{f}}) > 0$. So we can apply stage 7 to $\tilde{f}$ and obtain $NLC(\tilde{f},\mathcal{D}) \ge \sqrt{2}$. Because the Jacobian of $f$ and $\tilde{f}$ are equal and we have $\Cov_f=\Cov_{\tilde{f}}$ as shown above, $NLC(f,\mathcal{D}) = NLC(\tilde{f},\mathcal{D})$ and so $NLC(f,\mathcal{D}) \ge \sqrt{2}$.

\end{proof}

{\bf Remark.} We used the Wolfram Mathematica Software to obtain the formula for $objdiff$ given at the end of step 1 in claim 1 in stages 1 and 2.

\subsection{Theorem \ref{finiteNetLAR}: NLC depends on the least squares fit}

\begin{reptheorem}{finiteNetLAR}
Let $\mathcal{D}$ be Gaussian and assume $f$ is not linear. Then we have $$NLC(f,\mathcal{D})^2 = \frac{LAR}{NAR + LAR} + \frac{NAR}{NAR + LAR}NLC(\tilde{f},\mathcal{D})^2$$
\end{reptheorem}

\begin{proof}
Because $f$ is not linear, by proposition \ref{finiteNetPositiveBasis}, $\Tr(\Cov_{\tilde{f}}) > 0$, and so $NLC(\tilde{f},\mathcal{D})$ is valid. By proposition \ref{finiteNetBasisOrtho} part 1, we have $\mathbb{E}x^T\tilde{f} = 0$ and $\mathbb{E}\tilde{f} = 0$, and so $\mathbb{E}||\tilde{f}||_2^2 = \mathbb{E}||\tilde{f}||_2^2 - ||\mathbb{E}\tilde{f}||_2^2 = \Tr(\Cov_{\tilde{f}})$. Assume $\bar{x} = 0$.  Then proposition \ref{finiteNetTheoremConnector} followed by lemma \ref{lemma2} yields $\mathbb{E}\tilde{\mathcal{J}} = 0$, where $\tilde{\mathcal{J}}$ is the Jacobian of $\tilde{f}$. Using these statements we have

\begin{eqnarray*}
&&NLC(f,\mathcal{D})^2\\
&=&\frac{\mathbb{E}\Tr(\mathcal{J}\Cov_x\mathcal{J}^T)}{\Tr(\Cov_f)}\\
&=&\frac{\mathbb{E}\Tr(\frac{d}{dx}(\tilde{f} + xA_f + b_f)\Cov_x\frac{d}{dx}(\tilde{f} + xA_f + b_f)^T)}{\text{\scalebox{0.85}{$\Tr(\mathbb{E}_x(\tilde{f}(x) + xA_f + b_f - \mathbb{E}_{x'}(\tilde{f}(x') + x'A_f + b_f))^T(\tilde{f}(x) + xA_f + b_f - \mathbb{E}_{x'}(\tilde{f}(x') + x'A_f + b_f))$}}}\\
&=&\frac{\mathbb{E}\Tr((\tilde{\mathcal{J}} + A_f^T)\Cov_x(\tilde{\mathcal{J}} + A_f^T)^T)}{\Tr(\mathbb{E}(\tilde{f} + xA_f)^T(\tilde{f} + xA_f))}\\
&=&\frac{\Tr(\mathbb{E}\tilde{\mathcal{J}}\Cov_x\tilde{\mathcal{J}}^T + \mathbb{E}\tilde{\mathcal{J}}\Cov_xA_f + \mathbb{E}A_f^T\Cov_x\tilde{\mathcal{J}}^T + \mathbb{E}A_f^T\Cov_xA_f)}{\Tr(\mathbb{E}\tilde{f}^T\tilde{f} + \mathbb{E}(xA_f)^T\tilde{f} + \mathbb{E}\tilde{f}^TxA_f + \mathbb{E}(xA_f)^TxA_f)}\\
&=&\frac{\mathbb{E}\Tr(\tilde{\mathcal{J}}\Cov_x\tilde{\mathcal{J}}^T) + \Tr((\mathbb{E}\tilde{\mathcal{J}})\Cov_xA_f) + \Tr(A_f^T\Cov_x\mathbb{E}(\tilde{\mathcal{J}})^T) + \Tr(A_f^T\Cov_xA_f)}{\mathbb{E}\Tr(\tilde{f}^T\tilde{f}) + \Tr(A_f^T(\mathbb{E}x^T\tilde{f})) + \Tr((A_f^T(\mathbb{E}x^T\tilde{f}))^T) + \mathbb{E}\Tr((xA_f)^TxA_f)}\\
&=&\frac{NLC(\tilde{f},\mathcal{D})^2\Tr(\Cov_{\tilde{f}}) + \mathbb{E}||(x-\bar{x})A_f||_2^2}{\mathbb{E}||\tilde{f}||_2^2 + \mathbb{E}||xA_f||_2^2}\\
&=&\frac{NLC(\tilde{f},\mathcal{D})^2\mathbb{E}||\tilde{f}||_2^2 + \mathbb{E}||(x-\bar{x})A_f||_2^2}{\mathbb{E}||\tilde{f}||_2^2 + \mathbb{E}||(x-\bar{x})A_f||_2^2}\\
&=&\frac{NLC(\tilde{f},\mathcal{D})^2(\mathbb{E}||f||_2^2)NAR + (\mathbb{E}||f||_2^2)LAR}{(\mathbb{E}||f||_2^2)NAR + (\mathbb{E}||f||_2^2)LAR}\\
&=&\frac{LAR}{NAR + LAR} + \frac{NAR}{NAR + LAR}NLC(\tilde{f},\mathcal{D})^2
\end{eqnarray*}

So now assume $\bar{x} \neq 0$. Define $f'$ and $\mathcal{D}'$, where $f'(x) = f(x + \bar{x})$ and drawing $x$ from $\mathcal{D}'$ is equivalent to drawing $x$ from $\mathcal{D}$ and then subtracting $\bar{x}$. Then we have

\begin{eqnarray*}
&&\mathbb{E}_{x\sim\mathcal{D}'}f'(x)\\
&=&\mathbb{E}_{x\sim\mathcal{D}}f'(x-\bar{x})\\
&=&\mathbb{E}_{x\sim\mathcal{D}}f(x-\bar{x}+\bar{x})\\
&=&\mathbb{E}_{x\sim\mathcal{D}}f(x)\\
&=&\bar{f}
\end{eqnarray*}

and

\begin{eqnarray*}
&&\mathbb{E}_{x\sim\mathcal{D}'}x\\
&=&\mathbb{E}_{x\sim\mathcal{D}}(x-\bar{x})\\
&=&0
\end{eqnarray*}

and

\begin{eqnarray*}
&&\Cov_{x\sim\mathcal{D}'}\\
&=&\mathbb{E}_{x\sim\mathcal{D}'}(x-\mathbb{E}_{x'\sim\mathcal{D}'}x')^T(x-\mathbb{E}_{x'\sim\mathcal{D}'}x')\\
&=&\mathbb{E}_{x\sim\mathcal{D}'}x^Tx\\
&=&\mathbb{E}_{x\sim\mathcal{D}}(x-\bar{x})^T(x-\bar{x})\\
&=&\Cov_x\\
\end{eqnarray*}

and

\begin{eqnarray*}
&&\Cov_{f'(x),x\sim\mathcal{D}'}\\
&=&\mathbb{E}_{x\sim\mathcal{D'}}(f'(x)-\mathbb{E}_{x'\sim\mathcal{D'}} f'(x'))^T(f'(x)-\mathbb{E}_{x'\sim\mathcal{D'}} f'(x'))\\
&=&\mathbb{E}_{x\sim\mathcal{D'}}(f'(x)-\bar{f})^T(f'(x)-\bar{f})\\
&=&\mathbb{E}_{x\sim\mathcal{D}}(f(x-\bar{x} + \bar{x})-\bar{f})^T(f(x-\bar{x} + \bar{x})-\bar{f})\\
&=&\mathbb{E}_{x\sim\mathcal{D}}(f(x)-\bar{f})^T(f(x)-\bar{f})\\
&=&\Cov_f
\end{eqnarray*}

Further, we have $\mathcal{J}'(x)=\mathcal{J}(x+\bar{x})$, where $\mathcal{J}'$ is the Jacobian of $f'$. Then we have

\begin{eqnarray*}
&&NLC(f',\mathcal{D}')^2\\
&=&\frac{\mathbb{E}_{x\sim\mathcal{D}'}\Tr(\mathcal{J}'(x)\Cov_{x\sim\mathcal{D}'}\mathcal{J}'(x)^T)}{\Tr(\Cov_{f'(x),x\sim\mathcal{D}'})}\\
&=&\frac{\mathbb{E}_{x\sim\mathcal{D}'}\Tr(\mathcal{J}'(x)\Cov_x\mathcal{J}'(x)^T)}{\Tr(\Cov_f)}\\
&=&\frac{\mathbb{E}_{x\sim\mathcal{D}}\Tr(\mathcal{J}(x-\bar{x}+\bar{x})\Cov_x\mathcal{J}(x-\bar{x}+\bar{x})}{\Tr(\Cov_f)}\\
&=&\frac{\mathbb{E}_{x\sim\mathcal{D}}\Tr(\mathcal{J}(x)\Cov_x\mathcal{J}(x)^T)}{\Tr(\Cov_f)}\\
&=&NLC(f,\mathcal{D})^2
\end{eqnarray*}

Let $xA_{f'} + b_{f'}$ be the least squares linear fit to $f'$ and let $\tilde{f}' = f' - xA_{f'} + b_{f'}$. Then we have 

\begin{eqnarray*}
\mathbb{E}_{x\sim\mathcal{D}}||f - (xA_f + b_f)||_2^2 &=& \mathbb{E}_{x\sim\mathcal{D}'}||f' - ((x+\bar{x})A_f + b_f)||_2^2\\
\mathbb{E}_{x\sim\mathcal{D}'}||f' - (xA_{f'} + b_{f'})||_2^2 &=& \mathbb{E}_{x\sim\mathcal{D}}||f - ((x-\bar{x})A_{f'} + b_{f'})||_2^2
\end{eqnarray*}

But by definition of the least squares linear fit, we have

\begin{eqnarray*}
\mathbb{E}_{x\sim\mathcal{D}}||f - (xA_f + b_f)||_2^2 &\ge& \mathbb{E}_{x\sim\mathcal{D}}||f - ((x-\bar{x})A_{f'} + b_{f'})||_2^2\\
\mathbb{E}_{x\sim\mathcal{D}'}||f' - (xA_{f'} + b_{f'})||_2^2 &\ge&\mathbb{E}_{x\sim\mathcal{D}'}||f' - ((x+\bar{x})A_f + b_f)||_2^2 
\end{eqnarray*}

Therefore, all four terms are equal. We obtain $A_{f'} = A_f$, $b_{f'} = b_f + \bar{x}A_f$ and 

\begin{eqnarray*}
&&\tilde{f}'\\
&=&f' - (xA_{f'} + b_{f'})\\
&=&f' - (xA_f  + b_f + \bar{x}A_f)\\
&=&f(x+\bar{x}) - ((x+\bar{x})A_f  + b_f)\\
&=&\tilde{f}(x+\bar{x})
\end{eqnarray*}

and hence $\tilde{\mathcal{J}}' = \tilde{\mathcal{J}}(x + \bar{x})$, where $\tilde{\mathcal{J}}'$ is the Jacobian of $\tilde{f}'$. This yields

\begin{eqnarray*}
&&\mathbb{E}_{x\sim\mathcal{D}'}||\tilde{f}'(x)||_2^2\\
&=&\mathbb{E}_{x\sim\mathcal{D}}||\tilde{f}(x-\bar{x} + \bar{x})||_2^2\\
&=&\mathbb{E}_{x\sim\mathcal{D}}||\tilde{f}(x)||_2^2\\
\end{eqnarray*}

and

\begin{eqnarray*}
&&\mathbb{E}_{x\sim\mathcal{D}'}||f'(x)||_2^2\\
&=&\mathbb{E}_{x\sim\mathcal{D}}||f(x-\bar{x} + \bar{x})||_2^2\\
&=&\mathbb{E}_{x\sim\mathcal{D}}||f(x)||_2^2\\
\end{eqnarray*}

and

\begin{eqnarray*}
&&\mathbb{E}_{x\sim\mathcal{D}'}||(x-\mathbb{E}_{x'\sim\mathcal{D}'}x')A_{f'}||_2^2\\
&=&\mathbb{E}_{x\sim\mathcal{D}'}||xA_{f'}||_2^2\\
&=&\mathbb{E}_{x\sim\mathcal{D}'}||xA_f||_2^2\\
&=&\mathbb{E}_{x\sim\mathcal{D}}||(x-\bar{x})A_f||_2^2
\end{eqnarray*}

and

\begin{eqnarray*}
&&\Cov_{\tilde{f}'(x),x\sim\mathcal{D}'}\\
&=&\mathbb{E}_{x\sim\mathcal{D'}}(\tilde{f}'(x)-\mathbb{E}_{x'\sim\mathcal{D'}} \tilde{f}'(x'))^T(\tilde{f}'(x)-\mathbb{E}_{x'\sim\mathcal{D'}} \tilde{f}'(x'))\\
&=&\mathbb{E}_{x\sim\mathcal{D}}(\tilde{f}(x-\bar{x}+\bar{x})-\mathbb{E}_{x'\sim\mathcal{D}} \tilde{f}(x'-\bar{x}+\bar{x}))^T(\tilde{f}(x-\bar{x}+\bar{x})-\mathbb{E}_{x'\sim\mathcal{D}} \tilde{f}(x'-\bar{x}+\bar{x}))\\
&=&\mathbb{E}_{x\sim\mathcal{D}}(\tilde{f}(x)-\mathbb{E}_{x'\sim\mathcal{D}} \tilde{f}(x'))^T(\tilde{f}(x)-\mathbb{E}_{x'\sim\mathcal{D}} \tilde{f}(x'))\\
&=&\Cov_{\tilde{f}}
\end{eqnarray*}

and

\begin{eqnarray*}
&&\mathbb{E}_{x\sim\mathcal{D}'}\Tr(\tilde{\mathcal{J}}'(x)\Cov_{x\sim\mathcal{D}'}\tilde{\mathcal{J}}'(x)^T)\\
&=&\mathbb{E}_{x\sim\mathcal{D}}\Tr(\tilde{\mathcal{J}}(x-\bar{x}+\bar{x})\Cov_x\tilde{\mathcal{J}}(x-\bar{x}+\bar{x})^T)\\
&=&\mathbb{E}_{x\sim\mathcal{D}}\Tr(\tilde{\mathcal{J}}(x)\Cov_x\tilde{\mathcal{J}}(x)^T)\\
\end{eqnarray*}

\sloppy and hence $LAR(f,\mathcal{D}) = LAR(f',\mathcal{D}')$, $NAR(f,\mathcal{D}) = NAR(f',\mathcal{D}')$ and $NLC(f,\mathcal{D}) = NLC(f',\mathcal{D}')$.

Putting it all together, using $\mathbb{E}_{x\sim\mathcal{D}'}x = 0$ and the fact that we proved the theorem for the case $\bar{x} = 0$, we have

\begin{eqnarray*}
&&NLC(f,\mathcal{D})^2\\
&=&NLC(f',\mathcal{D'})^2\\
&=&\frac{LAR(f',\mathcal{D}')}{NAR(f',\mathcal{D}') + LAR(f',\mathcal{D}')} + \frac{NAR(f',\mathcal{D}')}{NAR(f',\mathcal{D}') + LAR(f',\mathcal{D}')}NLC(\tilde{f}',\mathcal{D}')^2\\
&=&\frac{LAR(f,\mathcal{D})}{NAR(f,\mathcal{D}) + LAR(f,\mathcal{D})} + \frac{NAR(f,\mathcal{D})}{NAR(f,\mathcal{D}) + LAR(f,\mathcal{D})}NLC(\tilde{f},\mathcal{D})^2
\end{eqnarray*}

\end{proof}

\chapter{Mean field nonlinearity theory} \label{mfntChapter}

In this chapter, we prove the theorems and propositions stated in chapter \ref{meanFieldNnaChapter} concerning constructs such as activation functions, mean field architectures and A-architectures. As opposed to chapter \ref{finiteNetChapter}, a neural architecture in this chapter is a directed, acyclic graph where each node corresponds to a layer that uses one of a small number of operations. Each layer is a function from one real vector space to another, and therefore can be viewed as a ``mini-architecture'' or ``mini-network'' in the sense of chapter \ref{finiteNetChapter}. Of course, the space of functions representable by a layer graph of specific operations is much more constrained than the space of all differentiable functions considered in chapter \ref{finiteNetChapter}. On the other side of this tradeoff, we get to make specific predictions about metric values and layer (meta-)distributions in this chapter, and we get to enumerate all possible behaviors across activation functions.

\section{Notation, terminology and conventions} \label{covkerNotationSection}

From a notation, terminology and convention (NTC) standpoint, this chapter is a continuation of chapter \ref{meanFieldNnaChapter} and therefore uses the NTCs of chapter \ref{meanFieldNnaChapter} as well as chapter \ref{backgroundChapter}, which are summarized in section \ref{notationSummarySection}. Of course, we do not make use of all NTCs from those chapters. A basic understanding of prior material of this thesis should make it possible to read this chapter while referring back to section \ref{notationSummarySection} and the {\bf definition} environments of chapter \ref{meanFieldNnaChapter} as necessary. We outline the most important constructs and give definitions that are unique to this chapter below.

\subsection{Neural architectures}

A `neural architecture' $f$ is a function of a trainable real `parameter' vector $\theta$ and a real `input' vector. It is also a directed, acyclic graph of `layers' $f_l$, $0 \le l \le L$. Each layer $f_l$ takes as input the outputs of its dependencies, which are its parents in the layer graph, as well as the parameter sub-vector $\theta_l$ belonging to $f_l$. $\theta_1$ through $\theta_L$ make up $\theta$. A source in the graph is an `input layer' and a sink is an `output layer'. We write $f_{k_l[1]}$, .., $f_{k_l[\kappa_l]}$, .. $f_{k_l[K_l]}$ for the dependencies of a layer $f_l$ with $K_l \ge 1$ dependencies, through we usually write $f_{k[1]}$, .., $f_{k[\kappa]}$, .. $f_{k[K]}$ as $l$ is clear from context. We also write $f_{k_l}$ for the dependency of a layer $f_l$ with a single dependency, though we usually write $f_k$ as $l$ is clear from context. 
  
We detail our NTCs surrounding neural architectures in section \ref{neuralNetworkNotationSection}. In this chapter, we consider specifically two types of architectures that we also consider in chapter \ref{meanFieldNnaChapter}: A-architectures and mean field architectures. A-architectures defined in section \ref{aArchitectureSection} represent fully-connected architectures built from popular building blocks and design strategies, and mean field architectures defined in section \ref{meanFieldArchitectureSection} represent abstract templates which are used to prove results for practical architectures like A-architectures. A layer in a mean field architecture corresponds to a line in a NETSOR program in \citet{meanFieldNetsorGP,meanFieldNetsorNTK, meanFieldNetsorMatrix}. Mean field architectures can have `finite input layers', `readin layers' and `readout layers' as defined in sections \ref{meanFieldGPsection} and \ref{covarianceKernelSection}.

Both mean field and A-architectures differ slightly from the framework of section \ref{neuralNetworkNotationSection}, as described in their definitions. Both types have layers whose width can vary via the parameter $d_\text{MF}$, which is used to take the limit of infinite width. Mean field architectures can have multiple input and output layers. An A-architecture has a single input layer $f_0$ to which the input $x$ is assigned and a single output layer $f_L$ which returns the `output' $f(\theta, x)$.

\subsection{Multi-activation functions} \label{multiActivationSection}

Mean field architectures use `elementwise layers' as defined in section \ref{meanFieldArchitectureSection}. An elementwise layer $f_l$ uses a `multi-activation function' $\rho_l$, which is applied elementwise to the `elementwise dependencies' $f_{\dot{k}_l[1]}$, .., $f_{\dot{k}_l[\dot{\kappa}_l]}$, .., $f_{\dot{k}_l[\dot{K}_l]}$ of $f_l$ and applied to the `layer means' of the `mean dependencies' $f_{\hat{k}_l[1]}$, .., $f_{\hat{k}_l[\hat{\kappa}_l]}$, .., $f_{\hat{k}_l[\hat{K}_l]}$ of $f_l$. This yields

$$f_l = \rho_l.(f_{\dot{k}_l[1]}, .., f_{\dot{k}_l[\dot{K}_l]},\mathbb{E}_if_{\hat{k}_l[1]}[i], .., \mathbb{E}_if_{\hat{k}_l[\hat{K}_l]}[i])$$

When considering $\rho_l$ as a function of its inputs, we term the inputs corresponding to elementwise dependencies the `elementwise inputs' and the inputs corresponding to mean dependencies the `mean inputs'. We denote an elementwise input corresponding to a dependency on layer $f_{l'}$ by $e_{l'}$ and a mean input corresponding to a dependency on layer $f_{l'}$ by $m_{l'}$. This yields

$$\rho_l(e_{\dot{k}_l[1]},..,e_{\dot{k}_l[\dot{K}_l]},m_{\hat{k}_l[1]},..,m_{\hat{k}_l[\hat{K}_l]})$$

We usually omit the $l$ subscript as it is clear from context. Note that in this chapter as in chapter \ref{meanFieldNnaChapter}, when the letter $m$ is used as a subscript as in $\mathcal{J}_{l,m}$, it denotes a layer index instead of a mean input.

As in table \ref{tableBackgroundPropagation} and the definition of `parameter-control' in section \ref{masterTheoremSection}, we usually write a multi-activation function simply as $\rho_l(e,m)$, where $e$ and $m$ are placeholders that refer to the tuples $(e_{\dot{k}_l[1]},..,e_{\dot{k}_l[\dot{K}_l]})$ and $(m_{\hat{k}_l[1]},..,m_{\hat{k}_l[\hat{K}_l]})$ respectively. When we use $\mathfrak{m}$ or $m^*$ as specific assignments to $m$, those terms also function as placeholders. So we write e.g. $\rho(e, m^*)$ for $\rho(e_{\dot{k}_l[1]},..,e_{\dot{k}_l[\dot{K}_l]}, m^*_{\hat{k}_l[1]},..,m^*_{\hat{k}_l[\hat{K}_l]})$.

In addition to denoting the $\dot{\kappa}_l$'th elementwise input by $e_{\dot{k}_l[\dot{\kappa}_l]}$, we also use the vector notation $e[\dot{\kappa}_l]$, and similarly $m[\hat{\kappa}_l]$. We do this especially when we consider a multi-activation function $\rho$ ``in a vacuum'' without reference to an architecture or layer, such as in lemmas \ref{lemma12}, \ref{lemma13} and \ref{lemma14}. In that case, we write $\dot{K}$ for the number of components of $e$ ad $\hat{K}$ for the number of components of $m$. However, even then $e$ and $m$ still behave as placeholders. For example, in lemma \ref{lemma14}, the $\rho^{(n)}$ do not necessarily have to have identical inputs. The components of $e$ and $m$ for some $\rho^{(n)}$ can match up in any way with the components of $e$ and $m$ for any other $\rho^{(n')}$, though there is no crossover between elementwise and mean inputs.

We sometimes reference the distribution of one or more $e$ values. In that case, we consider $(e_0, e_1, .., e_L)$ to form a Gaussian vector with moments given by $\mathbb{E}e_l = \mathfrak{m}_l$ and $\mathbb{E}e_le_m = \mathfrak{c}_{l;m}$ as defined in table \ref{tableBackgroundPropagation}. Note that $\mathfrak{c}_{l;m}$ is only defined when $f_l$ and $f_m$ have the same width. At any point where we reference the distribution of $e$ values, the well-definedness of the $\mathfrak{m}_l$ and $\mathfrak{c}_{l;m}$ values required to specify the distribution has already been guaranteed by applying background theorem \ref{backgroundMaster} to a sub-architecture in the context of an induction. Usually, the distribution of the value of a multi-activation function $\rho_l$ is induced by its elementwise inputs.

\subsection{Other NTCs} \label{otherNTCsection}

\begin{itemize}
\item Throughout this work, we consider an activation function $\tau : \mathbb{R} \rightarrow \mathbb{R}$ to be an arbitrary scalar function. Conditions on $\tau$ are discussed below.
\item $\mu_1$ denotes the canonical Lebesgue measure.
\item \sloppy $n(\mu,\sigma^2,s)$ denotes the probability density function of the Gaussian with mean $\mu$ and (co)variance $\sigma^2$ evaluated at $s$. $n(\sigma^2, s)$ is short for $n(0, \sigma^2, s)$ and $n(s)$ is short for $n(0, 1, s)$. Note that the letter $n$ can also refer to a sequence or tuple index when it is not used as a function. For example, $n(s_n)$ would refer to the unit Gaussian density evaluated at the $n$'th scalar in some sequence.
\item $a_\tau s+b_\tau$ denotes the least squares linear fit to $\tau(s)$ under $\mathcal{N}(0,1)$ and $\tilde{\tau}(s) = \tau(s) - a_\tau s-b_\tau$. LAR, CAR and NAR are defined as in section \ref{nlcLinearApproximationSection}.
\item $c$ is overloaded to refer to both the co-mean parameter of e.g. an elem-like$(q,c)$ input distribution as well as a generic constant.
\item $\mathcal{C}^d$ for a positive integer $d$ is the class of functions $F(\chi) = c||\chi||_2^d+c'$ for $c,c' > 0$ and $\chi$ of any dimensionality.
\item We say a function class $\mathcal{C}$ is `linearly closed' if any finite linear combination of members of $\mathcal{C}$ is controlled by $\mathcal{C}$.
\item We say a multi-activation function $\rho(e,m)$ is `elem-poly($d$)' if ...
\begin{itemize}
\item ...it is a weighted sum of products of expressions that are of form (i) $e[\dot{\kappa}]^c$ for some positive integer constant $c$, (ii) $m[\hat{\kappa}]^c$ for some positive integer constant $c$ or (iii) $(m[\hat{\kappa}]+c)^{-c'}$ for some positive real constants $c$,$c'$.
\item ... the total power of $e[\kappa]$ terms in any product is at most $d$.
\end{itemize}
For example, consider

\begin{eqnarray*}
\rho(e,m) &=&e[1]^2e[4]^3m[2]m[0]m[5]^2(m[1] + c_1)^{-\frac{1}{4}}(m[2] + c_2)^{-4}\\
&&- e[3]^2e[0]e[1](m[6] + c_3)^{-\frac{1}{2}}(m[9] + c_4)^{-\frac{5}{2}} + e[3]m[4]
\end{eqnarray*}

This $\rho$ would be elem-poly($d$) for any $d \ge \max(2+3,2+1+1,1) = 5$
\item $T_2(s)=\max_{|s'|\le |s|}|\tau''(s')|$ and $T_3(s)=\max_{|s'|\le |s|}|\tau'''(s')|$
\item We often shorten $\lim_{\mathsf{quantity}\rightarrow\mathsf{value}}$ to $\lim_{\mathsf{quantity}}$ when $\mathsf{value}$ is clear from context.
\item We use the notation and terminology of section \ref{layerTypesSection} for our layer operations.
\item A valid logical ordering of results in this chapter is as follows: lemmas \ref{lemma8} through \ref{lemma17}, propositions \ref{mfntElemLike}, \ref{mfntPNLCD}, \ref{covkerBisC}, \ref{covkerPositive} through \ref{mfntNPE}, theorems \ref{mfntMetaGaussian} through \ref{covkerCregular}, and propositions \ref{covkernregular}, \ref{mfntCNLC}. The proof of a result may depend on another result that comes before it but not on a result that comes after it in this ordering.
\end{itemize}

\section{General assumptions} \label{covkerAssumptionsSection}

In contrast to chapter \ref{finiteNetChapter}, we do not assume differentiability for all results in this chapter. Differentiability and continuity conditions are always stated explicitly in the results of this chapter or come along with the definition of an A-architecture given in section \ref{aArchitectureSection}.

We only make the following assumption throughout.

\begin{assumption} \label{assumptionIntegrableMeanField}
Simple expressions involving $\tau$, its derivatives, its left and right derivative, as well as $\max$ operators such as those involved in $T_2$ and $T_3$ defined above or those used in lemma \ref{lemma18}, are integrable with respect to any Gaussian measure, where we use the term ``integrable'' as defined in section \ref{integrabilitySection}. For example, expressions like $\tau(s)$, $\tau'(s)^2$ and $sT_3(s)\tau'''(t)$ are assumed to be integrable, assuming the higher derivatives are valid. We assume that any integral of such an expression over a multi-dimensional Gaussian measure can be broken down into a sequence of 1-dimensional integrals. We assume that such simple expressions, when multiplied with a Gaussian density, converge to zero along any direction.
\end{assumption}

This assumption is mild, as we argue in section \ref{integrabilitySection}. Gaussian distributions are very light-tailed. When $\tau$ is at least continuous, which we assume in most results, it is integrable over any bounded set with respect to the canonical Lebesgue measure or any Gaussian measure.

Whenever we consider an expectation or integral involving a multi-activation function $\rho$, the validity of that integral is guaranteed by background theorem \ref{backgroundMaster} or background corollary \ref{backgroundKernel}. We did not check which implicit integrability assumptions were present in \citet{meanFieldNetsorGP}. The parameter-control assumption is clearly related to integrability, though it is also clearly not enough as the function that assigns 1 to rational numbers and 0 to non-rational numbers is controlled by $\mathcal{C}^\text{E2}$ but not integrable. It is possible that continuous $\tau$ and $\rho$ that are (parameter-)controlled by $\mathcal{C}^\text{E2}$ fulfill all integrability requirements of this chapter and \citet{meanFieldNetsorGP}, though we did not check this. Note that if $\tau$ is controlled by $\mathcal{C}^\text{E2}$, so is e.g. $T_2$ and $T_3$.

\section{Lemmas}

\begin{lemma} \label{lemma8}
Let $F: \mathbb{R}^{d_\text{in}} \rightarrow \mathbb{R}$ be a continuous, non-negative function that is integrable with respect to $ d\mu_1$ and not zero everywhere. Then $\int_{\mathbb{R}^{d_\text{in}}} F d\mu_1 > 0$.
\end{lemma}

\begin{proof}
Let $\chi \in \mathbb{R}^{d_\text{in}}$ be such that $F(\chi) > 0$. Because $F$ is continuous, there exists an open set $S$ containing $\chi$ and $\epsilon > 0$ such that $F > \epsilon$ in $S$. Becasue $S$ is open, $\mu_1(S) > 0$. Since $F$ is non-negative, $\int_{\mathbb{R}^{d_\text{in}}} F d\mu_1 \ge \int_S F d\mu_1 \ge \mu_1(S)\epsilon > 0$ as required.
\end{proof}

\begin{lemma} \label{lemma9}
Let $p \ge 0$ be an integer, $\lambda > 0$ and let $F: \mathbb{R} \rightarrow \mathbb{R}$ be a function that is right-continuous at zero. Assume $s^pF(s)n(\lambda^2,s)$ is integrable with respect to $d\mu_1$ on $[0,\infty)$ for all $\lambda > 0$. Then $$\lim_{\lambda\rightarrow 0}\lambda^{-p}\int_0^\infty s^pF(s)n(\lambda^2,s)d\mu_1 = F(0)\int_0^\infty s^pn(1,s)d\mu_1$$
\end{lemma}

\begin{proof}

Let $D(s) = F(s) - F(0)$. Because $\int_0^\infty s^pn(\lambda^2,s)d\mu_1 < \infty$ for all $\lambda > 0$, we have $\int_0^\infty |D(s)|s^pn(\lambda^2,s)d\mu_1 \le \int_0^\infty(|F(s)| + |F(0)|)s^pn(\lambda^2,s)d\mu_1 < \infty$. Let $D_\text{int} = \int_0^\infty |D(s)|s^pn(1,s)d\mu_1$ and $P_\text{int} = \int_0^\infty s^pn(1,s)d\mu_1$. Because $F$ is right-continuous at 0, $F$ is bounded in $[0, \delta]$ for some $\delta > 0$. Let $\lambda < \min(\delta^2, 1)$ and $B(\lambda) = \max_{0\le s \le \sqrt{\lambda}}|D(s)|$. Then we have

\begin{eqnarray*}
&&|\lambda^{-p}\int_0^\infty s^pF(s)n(\lambda^2,s)d\mu_1-F(0)\int_0^\infty s^pn(1,s)d\mu_1|\\
&=&|\lambda^{-p}\int_0^\infty s^pF(s)n(\lambda^2,s)d\mu_1-F(0)\int_0^\infty \lambda^{-p}s^pn(\lambda^2,s)d\mu_1|\\
&=&|\lambda^{-p}\int_0^\infty s^pD(s)n(\lambda^2,s)d\mu_1|\\
&\le&\lambda^{-p}\int_0^\infty s^p|D(s)|n(\lambda^2,s)d\mu_1\\
&=&\lambda^{-p}\int_0^{\sqrt{\lambda}} s^p|D(s)|n(\lambda^2,s)d\mu_1 + \lambda^{-p}\int_{\sqrt{\lambda}}^{\infty} s^p|D(s)|n(\lambda^2,s)d\mu_1\\
&\le&B(\lambda)\lambda^{-p}\int_0^{\sqrt{\lambda}} s^pn(\lambda^2,s)d\mu_1 + \lambda^{-p}\int_{\sqrt{\lambda}}^{\infty} s^p|D(s)|\frac{n(\lambda^2,\sqrt{\lambda})}{n(1,\sqrt{\lambda})}n(1,s)d\mu_1\\
&\le&B(\lambda)\lambda^{-p}\int_0^{\infty} s^pn(\lambda^2,s)d\mu_1 + \lambda^{-p}\frac{n(\lambda^2,\sqrt{\lambda})}{n(1,\sqrt{\lambda})}D_\text{int}\\
&\le&B(\lambda)\int_0^{\infty} s^pn(1,s)d\mu_1 + \lambda^{-p}\frac{n(\lambda^2,\sqrt{\lambda})}{n(1,\sqrt{\lambda})}D_\text{int}\\
&\le&B(\lambda)P_\text{int} + \lambda^{-p}\frac{n(\lambda^2,\sqrt{\lambda})}{n(1,\sqrt{\lambda})}D_\text{int}\\
&=&B(\lambda)P_\text{int} + \lambda^{-p}\frac{\frac{1}{\sqrt{2\pi\lambda^2}}e^{-\frac{\lambda}{2\lambda^2}}}{\frac{1}{\sqrt{2\pi}}e^{-\frac{\lambda}{2}}}D_\text{int}\\
&=&B(\lambda)P_\text{int} + \lambda^{-p-1}e^{-\frac{1}{2\lambda}+\frac{\lambda}{2}}D_\text{int}
\end{eqnarray*}

$B(\lambda)$ converges to 0 as $\lambda$ converges to 0 from the right due to the right-continuity of $F$. $\lambda^{-p-1}e^{-\frac{1}{2\lambda}+\frac{\lambda}{2}}$ converges to zero exponentially as $\lambda$ converges to zero.

\end{proof}

\begin{lemma} \label{lemma10}
Let $F(s) : \mathbb{R} \rightarrow \mathbb{R}$ be a non-negative function which is bounded on any bounded interval and integrable with respect to any Gaussian measure. Let $\mu_\text{max}$ and $\sigma_\text{max}$ be positive constants. Then $\mathbb{E}_{s\sim \mathcal{N}(\mu, \sigma^2)}F(s)$ is bounded over $-\mu_\text{max} \le \mu \le \mu_\text{max}$ and $0 \le \sigma \le \sigma_\text{max}$, where $\mathbb{E}_{s\sim \mathcal{N}(\mu, 0)}F(s)$ is defined as $F(\mu)$.
\end{lemma}

\begin{proof}

Since $F$ is non-negative, $\mathbb{E}_{s\sim \mathcal{N}(\mu, \sigma^2)}F(s)$ is bounded below by zero for all $\mu$, $\sigma$. Hence, for the rest of this proof we will concern ourselves only with upper bounds. Throughout this proof, we assume $-\mu_\text{max} \le \mu \le \mu_\text{max}$ and $0 \le \sigma \le \sigma_\text{max}$.

First consider the case $\sigma=0$. Then $\mathbb{E}_{s\sim \mathcal{N}(\mu, \sigma^2)}F(s)$ is equal to a value taken by $F$ in $[-\mu_\text{max},\mu_\text{max}]$. $F$ is bounded there, so $\mathbb{E}_{s\sim \mathcal{N}(\mu, \sigma^2)}F(s)$ is bounded as well. Let this bound be $B_4$.

Going forward, we assume $0 < \sigma \le \sigma_\text{max}$ holds.

Consider some $s\le -\mu_\text{max}-\sigma_\text{max}$. Then $n(\mu,\sigma^2,s) \le n(-\mu_\text{max},\sigma^2,s)$. On the interval $(0, \sigma_\text{max}]$, $n(-\mu_\text{max},\sigma^2,s)$ is differentiable as a function of $\sigma$. We have

\begin{eqnarray*}
&&\frac{dn(-\mu_\text{max},\sigma^2,s)}{d\sigma}\\
&=&\frac{1}{\sqrt{2\pi}}(-\sigma^{-2}+(s+\mu_\text{max})^2\sigma^{-4})e^{-\frac{(s+\mu_\text{max})^2}{2\sigma^2}}\\
&\ge&\frac{1}{\sqrt{2\pi}}(-\sigma^{-2}+(-\mu_\text{max}-\sigma_\text{max}+\mu_\text{max})^2\sigma^{-4})e^{-\frac{(s+\mu_\text{max})^2}{2\sigma^2}}\\
&\ge&\frac{\sigma^{-4}}{\sqrt{2\pi}}(\sigma_\text{max}^2 - \sigma^2)e^{-\frac{(s+\mu_\text{max})^2}{2\sigma^2}}\\
&\ge&0
\end{eqnarray*}

Therefore, $n(\mu,\sigma^2,s) \le n(-\mu_\text{max},\sigma^2,s) \le (-\mu_\text{max},\sigma_\text{max}^2,s)$. Similarly, when we consider $s\ge \mu_\text{max}+\sigma_\text{max}$ we obtain $n(\mu,\sigma^2,s)\le n(\mu_\text{max},\sigma_\text{max}^2,s)$. 

Because $F$ is integrable with respect to any Gaussian measure, we can set 

$$B_1=\int_{s=-\infty}^{-\mu_\text{max}-\sigma_\text{max}}F(s)n(-\mu_\text{max},\sigma_\text{max}^2,s)d\mu_1<\infty$$ 

and 

$$B_2=\int_{s=\mu_\text{max}+\sigma_\text{max}}^{\infty}F(s)n(\mu_\text{max},\sigma_\text{max}^2,s)d\mu_1<\infty$$

Also, $F$ is bounded on $[-\mu_\text{max}-\sigma_\text{max}, \mu_\text{max}+\sigma_\text{max}]$. Let this bound be $B_3$. Finally, we have 

\begin{eqnarray*}
&&\mathbb{E}_{s\sim\mathcal{N}(\mu, \sigma^2)}F(s)\\
&=&\int_{s=-\infty}^{-\mu_\text{max}-\sigma_\text{max}}F(s)n(\mu,\sigma^2,s)d\mu_1+\int_{s=-\mu_\text{max}-\sigma_\text{max}}^{\mu_\text{max}+\sigma_\text{max}}F(s)n(\mu,\sigma^2,s)d\mu_1\\
&&+\int_{s=\mu_\text{max}+\sigma_\text{max}}^{\infty}F(s)n(\mu,\sigma^2,s)d\mu_1\\
&\le&\int_{s=-\infty}^{-\mu_\text{max}-\sigma_\text{max}}F(s)n(-\mu_\text{max},\sigma_\text{max}^2,s)d\mu_1+\int_{s=-\mu_\text{max}-\sigma_\text{max}}^{\mu_\text{max}+\sigma_\text{max}}B_3n(\mu,\sigma^2,s)d\mu_1\\
&&+\int_{s=\mu_\text{max}+\sigma_\text{max}}^{\infty}F(s)n(\mu_\text{max},\sigma_\text{max}^2,s)d\mu_1\\
&=&B_1+B_2+B_3\int_{s=-\mu_\text{max}-\sigma_\text{max}}^{\mu_\text{max}+\sigma_\text{max}}n(\mu,\sigma^2,s)d\mu_1\\
&\le&B_1+B_2+B_3\int_{s=-\infty}^{\infty}n(\mu,\sigma^2,s)d\mu_1\\
&=&B_1+B_2+B_3
\end{eqnarray*}

So $\mathbb{E}_{s\sim\mathcal{N}(\mu, \sigma^2)}F(s)$ is bounded by $B_1+B_2+B_3$.

Therefore, no matter whether $\sigma=0$, we have that $\mathbb{E}_{s\sim\mathcal{N}(\mu, \sigma^2)}F(s)$ is bounded by the maximum of $B_4$ and $B_1+B_2+B_3$, as required.

\end{proof}

\begin{lemma} \label{lemma11}
Let $a_\tau s + b_\tau$ be the least squares linear fit to $\tau(s)$ under $\mathcal{N}(0,1)$. Let $\tilde{\tau} = \tau - a_\tau s - b_\tau$. Let $a_{\tilde{\tau}} s + b_{\tilde{\tau}}$ be the least squares linear fit to $\tilde{\tau}(s)$ under $\mathcal{N}(0,1)$. Let $s \sim \mathcal{N}(0,1)$. Then $a_\tau=\mathbb{E}s\tau(s)$, $b_\tau=\mathbb{E}\tau(s)$, $a_{\tilde{\tau}}=\mathbb{E}s\tilde{\tau}(s)=0$ and $b_{\tilde{\tau}}=\mathbb{E}\tilde{\tau}(s)=0$.
\end{lemma}

\begin{proof}

The standard formula for the least squares linear fit yields

\begin{eqnarray*}
&&\begin{pmatrix}a_\tau\\b_\tau\end{pmatrix}\\
&=&\begin{pmatrix}\mathbb{E}s^2 & \mathbb{E}s \\ \mathbb{E}s & 1\end{pmatrix}^{-1}\begin{pmatrix} \mathbb{E}s\tau \\  \mathbb{E}\tau\end{pmatrix}\\
&=&\begin{pmatrix}1 & 0 \\ 0 & 1\end{pmatrix}^{-1}\begin{pmatrix} \mathbb{E}s\tau \\  \mathbb{E}\tau\end{pmatrix}\\
&=&\begin{pmatrix} \mathbb{E}s\tau \\  \mathbb{E}\tau\end{pmatrix}\\
\end{eqnarray*}

Similarly, $a_{\tilde{\tau}} = \mathbb{E}s\tilde{\tau} = \mathbb{E}s(\tau - a_\tau s - b_\tau) = a_\tau - a_\tau \mathbb{E}s^2 - b_\tau \mathbb{E}s = 0$ and $\tilde{b} = \mathbb{E}\tilde{\tau} = \mathbb{E}(\tau - a_\tau s - b_\tau) = b_\tau - b_\tau = 0$.
\end{proof}

\begin{lemma} \label{lemma12}
Let $\rho(e,m)$ be a multi-activation function. Let $m^* \in \mathbb{R}^{\hat{K}}$ be a fixed assignment to its mean inputs and let $M^*$ be an open set containing $m^*$. Let $\mathcal{C}$ be some function class. Assume:

\begin{itemize}
\item $\mathcal{C}$ is linearly closed.
\item $\rho(e, m^*)$ is controlled by $\mathcal{C}$ as a function of $e$.
\item $\rho(e, m)$ is differentiable for $m \in M^*$.
\item For each $1 \le \hat{\kappa} \le \hat{K}$, $\max_{m\in M^*}\Big|\frac{d\rho(e,m)}{dm[\hat{\kappa}]}\Big|$ is controlled by $\mathcal{C}$ as a function of $e$.
\end{itemize}

Then $\rho$ is parameter-controlled by $\mathcal{C}$ at $m^*$.

\end{lemma}

\begin{proof}
We will look at the conditions of parameter-control in turn. $\rho(e,m^*)$ being controlled by $\mathcal{C}$ is a condition of this proposition. Let $\rho^\text{mean}(m)=||m-m^*||_2$ when $m \in M^*$ and $\infty$ otherwise. This yields that $\rho^\text{mean}(m^*) = 0$ and that $\rho^\text{mean}$ is continuous at $m^*$. Let $\rho^\text{elem} = \sum_{\hat{\kappa}=1}^{\hat{K}}\max_{m\in M^*}\Big|\frac{d\rho(e,m)}{dm[\hat{\kappa}]}\Big|$. By assumption, it is the sum of functions controlled by $\mathcal{C}$, which is controlled by $\mathcal{C}$ as $\mathcal{C}$ is linearly closed. If $m\not\in M^*$, clearly we have $|\rho(e,m) - \rho(e,m^*)| \le \rho^{elem}(e)\rho^{mean}(m) = \infty$. Otherwise, we have

\begin{eqnarray*}
&&\Big|\rho(e,m) - \rho(e,m^*)\Big|\\
&\le&||m-m^*||_2\max_{m \in M^*}\Big|\Big|\frac{d\rho(e,m)}{dm}\Big|\Big|_2\\
&\le&||m-m^*||_2\max_{m \in M^*}\Big|\frac{d\rho(e,m)}{dm}\Big|_1\\
&\le&||m-m^*||_2\sum_{\hat{\kappa}=1}^{\hat{K}}\max_{m \in M^*}\Big|\frac{d\rho(e,m)}{dm[\hat{\kappa}]}\Big|\\
&=&\rho^\text{mean}(m)\rho^\text{elem}(e)
\end{eqnarray*}

as required.

\end{proof}

\begin{lemma}\label{lemma13}
Let $\rho(e, m)$ be a multi-activation function that is elem-poly$(d)$. Let $m^*$ be a fixed assignment to $m$ such that $m^*[\hat{\kappa}] \ge 0$ for any $\hat{\kappa}$ for which an $(m[\hat{\kappa}]+c)^{c'}$ expression appears in $\rho$. Let $M(m^*)$ be the open hypercube with center $m^*$ and side length equal to the least $c$ among $(m[\hat{\kappa}] + c)^{-c'}$ expressions that appear in $\rho$.

Then:
\begin{enumerate}
\item For each $1 \le \hat{\kappa} \le \hat{K}$, $\max_{m\in M(m^*)}\Big|\frac{d\rho(e,m)}{dm[\hat{\kappa}]}\Big|$ is controlled by $\mathcal{C}^{d}$  as a function of $e$.
\item $\rho$ is parameter-controlled by $\mathcal{C}^{d}$ at $m^*$.
\end{enumerate} 
\end{lemma}

\begin{proof}
Polynomials of degree at most $d$ are controlled by $\mathcal{C}^{d}$, so $\rho(e,m^*)$ is controlled by $\mathcal{C}^{d}$. If $\rho$ has no mean dependencies, that is all that needs to be checked for statement 2, and statement 1 is trivial. So now assume that $\rho$ does have one or more mean dependencies. We note that, by construction, for each $m \in M(m^*)$, the base of each $(m[\hat{\kappa}] + c)^{-c'}$ expression that appears in $\rho$ is positive, so $\rho$ is both well-defined and differentiable with respect to any $m[\hat{\kappa}]$.

Let $s$ be the side length of $M(m^*)$. Let $\lceil \rho \rceil$ be a function of $e$ that is derived from $\rho$ by (i) fixing each expression $m[\hat{\kappa}]$ that appears to $|m^*[\hat{\kappa}]| + \frac{s}{2}$, (ii) fixing each $(m[\hat{\kappa}] + c)^{-c'}$ that appears to $(m^*[\hat{\kappa}] - \frac{s}{2} + c)^{-c'}$ and (iii) changing each minus in front of a term into a plus. It is easy to check that $\frac{d\rho(e,m)}{dm[\hat{\kappa}]}$ for any $\hat{\kappa}$ is also elem-poly($d$). Finally, it is easy to check that $\max_{m\in M^*}\Big|\frac{d\rho}{dm[\hat{\kappa}]}\Big| \le \Big\lceil\frac{d\rho(e,m)}{dm[\hat{\kappa}]}\Big\rceil$. But $\Big\lceil\frac{d\rho(e,m)}{dm[\hat{\kappa}]}\Big\rceil$ is controlled by $\mathcal{C}^{d}$, as it is a polynomial of degree at most $d$. This yields statement 1.

For statement 2, we use lemma \ref{lemma12}. $\mathcal{C}^{d}$ is linearly closed, and we already showed the other three conditions of the lemma.
\end{proof}

\begin{lemma}\label{lemma14}
Let all $\rho^{(1)}$, .., $\rho^{(N)}$ be multi-activation functions which do not have to have the same elementwise or mean inputs as detailed in section \ref{multiActivationSection}. Let them all be parameter-controlled by $\mathcal{C}$ for some linearly closed class $\mathcal{C}$ at some $m^*$. Then any linear combination $\rho = \sum_{n=1}^Nw^{(n)}\rho^{(n)}$ is also parameter-controlled by $\mathcal{C}$ at $m^*$.
\end{lemma}

\begin{proof}
Since the $\rho^{(n)}$ are parameter-controlled, there exist corresponding functions $\rho^{(n)\text{elem}}$ and $\rho^{(n)\text{mean}}$. WLOG $\rho^{(n)\text{elem}}\ge 0$ for all $n$. We will check the 5 conditions of the parameter-control property for $\rho$ using $\rho^\text{elem} = \sum_n\sqrt{|w^{(n)}|}\rho^{(n)\text{elem}}$ and $\rho^\text{mean} = \sum_n\sqrt{|w^{(n)}|}\rho^{(n)\text{mean}}$.

For each $n$, we have $|\rho^{(n)}(e,m^*)| \le F^{(n)}$ for some $F^{(n)} \in \mathcal{C}$, so $|\rho(e,m^*)| \le \sum_n |w^{(n)}|F^{(n)}$. $\sum_n |w^{(n)}|F^{(n)}$ is controlled by $\mathcal{C}$ because $\mathcal{C}$ is linearly closed, so $\rho(e,m^*)$ is controlled by $\mathcal{C}$. Analogously, $\rho^\text{elem}$ is controlled by $\mathcal{C}$.

If the $\rho^{(n)\text{mean}}$ are zero and continuous at $m^*$, so is any linear combination of them. Finally, we have

\begin{eqnarray*}
&&|\rho(e,m) - \rho(e,m^*)|\\
&\le&\sum_{n=1}^N|w^{(n)}||\rho^{(n)}(e,m)-\rho^{(n)}(e,m^*)|\\
&\le&\sum_{n=1}^N|w^{(n)}|\rho^{(n)\text{elem}}\rho^{(n)\text{mean}}\\
&\le&\Big(\sum_{n=1}^N\sqrt{|w^{(n)}|}\rho^{(n)\text{elem}}\Big)\Big(\sum_{n=1}^N\sqrt{|w^{(n)}|}\rho^{(n)\text{mean}}\Big)\\
&=&\rho^\text{elem}\rho^\text{mean}
\end{eqnarray*}

as required.

\end{proof}

\begin{lemma} \label{lemma15}
Let $F:\mathbb{R}^{d_\text{in}}\rightarrow \mathbb{R}$ be integrable with respect to any Gaussian measure. Let $\hat{\Sigma}$ be a positive definite symmetric matrix. Then $$\lim_{\Sigma\rightarrow\hat{\Sigma}}\mathbb{E}_{\chi\sim\mathcal{N}(0,\Sigma)}F(\chi) = \mathbb{E}_{\chi\sim\mathcal{N}(0,\hat{\Sigma})}F(\chi)$$

\end{lemma}

\begin{proof}
\sloppy We argue by contradiction. Assume there is an $\epsilon$ such that $|\mathbb{E}_{\chi\sim\mathcal{N}(0,\Sigma)}F(\chi) - \mathbb{E}_{\chi\sim\mathcal{N}(0,\hat{\Sigma})}F(\chi)| > \epsilon$ for $\Sigma$ arbitrarily close to $\hat{\Sigma}$. Let $s$ be the largest eigenvalue of $\hat{\Sigma}$. Then there exists a constant $C$ such that, for any $\chi$ with $||\chi||_2 > C$ and any positive definite symmetric matrix $\Sigma$ with eigenvalues less than $2s$, we have $n(3sI, \chi) > n(\Sigma,\chi)$. So for any such $\Sigma$ and $c > C$ we have

\begin{eqnarray*}
&&|\mathbb{E}_{\chi\sim\mathcal{N}(0,\Sigma)}F(\chi) - \mathbb{E}_{\chi\sim\mathcal{N}(0,\hat{\Sigma})}F(\chi)|\\
&=&\Big|\int_{\mathbb{R}_{d_\text{in}}}F(\chi)(n(\Sigma, \chi) - n(\hat{\Sigma},\chi))d\mu_1\Big|\\
&=&\Big|\int_{||\chi||_2 > c}F(\chi)(n(\Sigma,s) - n(\hat{\Sigma},\chi))d\mu_1 + \int_{||\chi||_2 \le c}F(\chi)(n(\Sigma,s) - n(\hat{\Sigma},\chi))d\mu_1\Big|\\
&\le&\int_{||\chi||_2 > c}|F(\chi)|n(\Sigma,X)d\mu_1 + \int_{||\chi||_2 > c}|F(\chi)|n(\hat{\Sigma},\chi)d\mu_1 \\
&& + \Big|\int_{||\chi||_2 \le c}F(\chi)(n(\Sigma,s) - n(\hat{\Sigma},\chi))d\mu_1\Big|\\
&\le&\int_{||\chi||_2 > c}|F(\chi)|n(3sI,\chi)d\mu_1 + \int_{||\chi||_2 > c}|F(\chi)|n(3sI,\chi)d\mu_1\\
&&+\Big|\int_{||\chi||_2 \le c}F(\chi)(n(\Sigma,s) - n(\hat{\Sigma},\chi))d\mu_1\Big|\\
\end{eqnarray*}

Because $F$ is integrable with respect to any Gaussian measure, the first two terms converge to zero as $c$ converges to infinity. So we can choose $c$ large enough that they sum to less than $\frac{1}{2}\epsilon$. Now consider $c$ fixed to that value. Then the third term is a product between a function that is bounded on the bounded set $||\chi||_2 \le c$ because it is controlled and a function that converges to zero uniformly as $\Sigma$ converges to $\hat{\Sigma}$. So the third term converges to zero, so it is eventually less than $\frac{1}{2}\epsilon$. Also, as $\Sigma$ converges to $\hat{\Sigma}$, its eigenvalues are eventually less than $2s$. But that means that $|\mathbb{E}_{\chi\sim\mathcal{N}(0,\Sigma)}F(\chi) - \mathbb{E}_{\chi\sim\mathcal{N}(0,\hat{\Sigma})}F(\chi)|$  is eventually less than $\epsilon$. Contradiction.
\end{proof}

\begin{lemma}\label{lemma16}
Let $A=\begin{pmatrix}a&b\\b&d\end{pmatrix}$ and $A'=\begin{pmatrix}a'&b'\\b'&d'\end{pmatrix}$ be two full rank matrices with non-negative diagonal entries and equal off-diagonal entries. Let $c>0$ and $c' > 0$. Then $cA+c'A'$ is full rank, has non-negative diagonal entries and equal off-diagonal entries.
\end{lemma}

\begin{proof}
We have

$$cA + c'A' = \begin{pmatrix}ca+c'a'&cb+c'b'\\cb+c'b'&cd+c'd'\end{pmatrix}$$

Clearly, the diagonal entries are non-negative and off-diagonal elements are equal. $A$ and $A'$ having full rank corresponds to $ad-b^2 > 0$ and $a'd'-b'^2 > 0$ respectively. We have

\begin{eqnarray*}
&&(ca+c'a')(cd+c'd') - (cb+c'b')^2\\
&=&c^2(ad - b^2) + c^{\prime 2}(a'd'-b^{\prime 2}) + cc'(ad'+a'd-2bb')\\
&>&cc'(ad'+a'd-2bb')\\
&\ge&cc'(2\sqrt{ad'a'd}-2bb')\\
&=&2cc'(\sqrt{ad}\sqrt{a'd'}-bb')\\
&>&2cc'(\sqrt{b^2}\sqrt{b^{\prime 2}}-bb')\\
&=&0
\end{eqnarray*}

as required. Here, we use that the arithmetic mean is larger than the geometric mean for non-negative reals.

\end{proof}

\begin{lemma}\label{lemma17}
Let $(A)_n$ be an infinite sequence of independent random matrices. Let $(u)_n$ be the infinite sequence of unit Gaussian random vectors where the dimensionality of $u_n$ is equal to the right dimension of $A_n$. Assume that $||A_nu_n||_2$ converges to some limit $\Lambda$ almost surely. Then $||A_n||_F$ converges to $\Lambda$ almost surely.
\end{lemma}

\begin{proof}
We argue by contradiction. Assume $||A_n||_F$ does not converge almost surely to $\Lambda$. Then $||A_n||_F^2$ does not converge almost surely to $\Lambda^2$. By the Borel-Cantelli lemma, there exists an $\epsilon$ such that $\sum_n P(\big|||A_n||_F^2 -\Lambda^2\big| > \epsilon)= \infty$, where $P(\mathsf{event})$ denotes the probability of an event.

We make the following additional definitions.

\begin{itemize}
\item $A$ is some fixed matrix of size $D^\text{left} \times D^\text{right}$ with $\big|||A||_F^2 -\Lambda^2\big| > \epsilon$.
\item $UCV^T$ is the singular value decomposition of $A$.
\item $D = \min(D^\text{left}, D^\text{right})$
\item $c_1$, .., $c_D$ are the singular values of $A$.
\item $u$ / $u^D$ is the unit Gaussian vector of dimensionality $D^\text{right}$ / $D$.
\end{itemize}

For now, we study our fixed $A$. Because vector length and the unit Gaussian are rotation invariant, we have that the distribution of $||Au||_2^2 = ||UCV^Tu||_2^2$ is the same as that of $||Cu^D||_2^2 = \sum_{d=1}^Dc_d^2u[d]^2$. So $\mathbb{E}_u||Au||_2^2=\sum_{d=1}^Dc_d^2$ and $\mathbb{V}_u||Au||_2^2=2\sum_{d=1}^Dc_d^4$ and hence $2(\mathbb{E}_u||Au||_2^2)^2\ge\mathbb{V}_u||Au||_2^2$. $\mathbb{V}$ is the variance operator. Also, it is well-known that $||A||_F^2=\sum_{d=1}^Dc_d^2$ holds.

Since $\big|||A||_F^2 -\Lambda^2\big| > \epsilon$, we have either $||A||_F^2 -\Lambda^2 > \epsilon$ or $||A||_F^2 -\Lambda^2 < -\epsilon$. Assume for now $||A||_F^2 -\Lambda^2 > \epsilon$. Let $E_1$ be the event that $\big|||Au||_2^2-\Lambda^2\big| \le \frac{\epsilon}{2}$. Let $E_2$ be the event that $||Au||_2^2>||A||_F^2$. Let $E_3$ be the complementary event. Denote the expectation of $||Au||_2^2-||A||_F^2$ under the three events also by $E_1$ through $E_3$ and denote the probabilities of the events as $P_1$ through $P_3$. Then 

$$0 = \mathbb{E}_u||Au||_2^2 - ||A||_F^2 = P_1E_1 + P_2E_2 + P_3E_3 \le P_2E_2 - P_1\Big(||A||_F^2 - \Lambda^2 - \frac{\epsilon}{2}\Big)$$

Assume $P_1 \ge \frac{1}{2}$. Then we must also have $P_2 > 0$ and so $E_2 \ge \frac{P_1}{P_2}\Big(||A||_F^2 - \Lambda^2 - \frac{\epsilon}{2}\Big)$. Also we have

$$2(||A||_F^2)^2=2(\mathbb{E}_u||Au||_2^2)^2\ge\mathbb{V}_u||Au||_2^2\ge P_1E_1^2+P_2E_2^2+P_3E_3^2\ge P_2E_2^2$$
$$\ge \frac{P_1^2}{P_2}\Big(||A||_F^2 - \Lambda^2 - \frac{\epsilon}{2}\Big)^2\ge\frac{\frac{1}{2}^2}{1-P_1}\Big(||A||_F^2 - \Lambda^2 - \frac{\epsilon}{2}\Big)^2$$

So we obtain 

$$P_2 + P_3 = 1-P_1 \ge \frac{\Big(||A||_F^2 - \Lambda^2 - \frac{\epsilon}{2}\Big)^2}{8(||A||_F^2)^2} \ge \frac{\epsilon^2}{32(\Lambda^2 + \epsilon)^2}$$

If $P_1 < \frac{1}{2}$, we obtain $P_2 + P_3 \ge \frac{1}{2}$. So across both cases we obtain $P_2 + P_3 \ge \min(\frac{1}{2},\frac{\epsilon^2}{32(\Lambda^2 + \epsilon)^2}) > 0$. Let this lower bound be $B$.

If we have $||A||_F^2 -\Lambda^2 < -\epsilon$ instead of $||A||_F^2 -\Lambda^2 > \epsilon$, there is an analogous argument. We define $E_2$ as $||Au||_2^2<||A||_F^2$ and obtain $P_2(-E_2) \ge P_1\Big(\Lambda^2 - \frac{\epsilon}{2} -||A||_F^2\Big)$ which also leads to $P_2 + P_3 \ge B$.

Now we return to our random sequence $(A)_n$. We have

\begin{eqnarray*}
&&\sum_n P\Big(\big|||A_nu_n||_2^2 -\Lambda^2\big| > \frac{\epsilon}{2}\Big)\\
&\ge&\sum_n P\Big(\big|||A_nu_n||_2^2 -\Lambda^2\big| > \frac{\epsilon}{2}\Big| \big|||A_n||_F^2 -\Lambda^2\big| > \epsilon\Big)P\Big(\big|||A_n||_F^2 -\Lambda^2\big| > \epsilon\Big)\\
&\ge&B\sum_n P\Big(\big|||A_n||_F^2 -\Lambda^2\big| > \epsilon\Big)\\
&=& \infty
\end{eqnarray*}

So by the second Borel-Cantelli lemma, $||A_nu_n||_2^2$ does not converge to $\Lambda^2$ almost surely, so $||A_nu_n||_2$ does not converge to $\Lambda$ almost surely. Contradiction, as required.

\end{proof}

\section{Propositions} \label{mfntPropositionssection}

\begin{repproposition}{mfntElemLike}
Let $\mathcal{M}_1$ be a meta-distribution over distributions $\mathcal{X}_1$ over scalars $s$. For each $d > 0$, let $\mathcal{M}_d$ be the elementwise meta-distribution over distributions $\mathcal{X}_d$ over vectors $\chi_d$ of dimensionality $d$ that is generated by $\mathcal{M}_1$. Let $q = \mathbb{E}_{\mathcal{X}_1\sim\mathcal{M}_1,s\sim\mathcal{X}_1}s^2$ and $c =  \mathbb{E}_{\mathcal{X}_1\sim\mathcal{M}_1}(\mathbb{E}_{s\sim\mathcal{X}_1}s)^2$. Let $\chi_d$ and $\chi'_d$ be drawn from the same $\mathcal{X}_d$, which is drawn from $\mathcal{M}_d$. Then 

$$\lim_{d\rightarrow \infty} \mathbb{E}_i\chi_d[i]^2 = q \text{ a.s.}$$ 
$$\lim_{d\rightarrow \infty} \mathbb{E}_i\chi_d[i]\chi'_d[i] = c \text{ a.s.}$$ 
\end{repproposition}

\begin{proof}
Because $\mathcal{M}_d$ is elementwise, for each $d$, the $x_d[i]^2$ are independent and the $x_d[i]x'_d[i]$ are independent. So the strong law of large numbers applies to the average. Hence, we have $\lim_{d\rightarrow \infty} \mathbb{E}_i\chi_d[i]^2 = \mathbb{E}_{\mathcal{X}_1\sim\mathcal{M}_1,s\sim\mathcal{X}_1}s^2 = q$ and 

\begin{eqnarray*}
&&\lim_{d\rightarrow \infty} \mathbb{E}_i\chi_d[i]\chi'_d[i]\\
&=& \mathbb{E}_{\mathcal{X}_1\sim\mathcal{M}_1,s,s'\sim\mathcal{X}_1}ss'\\
&=&\mathbb{E}_{\mathcal{X}_1\sim\mathcal{M}_1}(\mathbb{E}_{s\sim\mathcal{X}_1}s)(\mathbb{E}_{s'\sim\mathcal{X}_1}s')\\
&=&\mathbb{E}_{\mathcal{X}_1\sim\mathcal{M}_1}(\mathbb{E}_{s\sim\mathcal{X}_1}s)^2\\
&=&c
\end{eqnarray*}

\end{proof}

\begin{repproposition}{mfntPNLCD}
Let $\mathcal{D}$ be a data distribution using a finite set of labels in the form of one-hot vectors. Let $p_1, .., p_C > 0$ be the probabilities of each of the $C > 1$ classes occurring. Let the input of $\mathcal{D}$ be elem-like$(q,c)$ for some $q > c \ge 0$.

Then $PNLCD(\mathcal{D})=\frac{1}{\sqrt{1 - \sum_{\zeta=1}^Cp_\zeta^2}}$ with probability $1 - \sum_{\zeta=1}^Cp_\zeta^2$ and 0 with probability $\sum_{\zeta=1}^Cp_\zeta^2$.
\end{repproposition}

\begin{proof}
We have 

\begin{eqnarray*}
&&\Tr(\Cov_x)\\
&=&\Tr((\mathbb{E}_x(x-\mathbb{E}_{x'}x')^T(x-\mathbb{E}_{x'}x'))\\
&=&\mathbb{E}_x(x-\mathbb{E}_{x'}x')(x-\mathbb{E}_{x'}x')^T\\
&=&\mathbb{E}_xxx^T - \mathbb{E}_{x,x'}x'x^T\\
&=&d_\text{in}(q-c)
\end{eqnarray*}

Further, we have

\begin{eqnarray*}
&&\Tr(\Cov_y)\\
&=&\mathbb{E}_yyy^T - \mathbb{E}_{y,y'}y'y^T\\
&=&\sum_{\zeta=1}^Cp_\zeta - \sum_{\zeta=1}^Cp_\zeta^2\\
&=&1 - \sum_{\zeta=1}^Cp_\zeta^2\\
\end{eqnarray*}

Let $E_1$ be the event under which $y\neq y'$, $\mathbb{E}_ix[i]x[i] = q$, $\mathbb{E}_ix'[i]x'[i] = q$ and $\mathbb{E}_ix[i]x'[i] = c$. Let $E_2$ be the event under which $y=y'$, $\mathbb{E}_ix[i]x[i] = q$, $\mathbb{E}_ix'[i]x'[i] = q$ and $\mathbb{E}_ix[i]x'[i] = c$. Under $E_1$ we have

\begin{eqnarray*}
&&PNLCD\\
&=&\sqrt{\frac{||y-y'||_2^2\Tr(\Cov_x)}{||x-x'||_2^2\Tr(\Cov_y)}}\\
&=&\sqrt{\frac{2d_\text{in}(q-c)}{(xx^T + x'x'^T - 2x'x^T)(1 - \sum_{\zeta=1}^Cp_\zeta^2)}}\\
&=&\sqrt{\frac{2d_\text{in}(q-c)}{(2d_\text{in}q-2d_\text{in}c)(1 - \sum_{\zeta=1}^Cp_\zeta^2)}}\\
&=&\frac{1}{\sqrt{1 - \sum_{\zeta=1}^Cp_\zeta^2}}\\
\end{eqnarray*}

Under $E_2$, we have $PNLCD=0$. Since $C>1$ and the $p_\zeta$ are positive and sum to 1, we have $\frac{1}{\sqrt{1 - \sum_{\zeta=1}^Cp_\zeta^2}} > 0$. The probability of $E_1$ is $1 - \sum_{\zeta=1}^Cp_\zeta^2$ and the probability of $E_2$ is $\sum_{\zeta=1}^Cp_\zeta^2$. Since both probabilities sum to 1, $PNLCD=\frac{1}{\sqrt{1 - \sum_{\zeta=1}^Cp_\zeta^2}}$ indeed holds with probability $1 - \sum_{\zeta=1}^Cp_\zeta^2$ and $PNLCD=0$ holds with probability $\sum_{\zeta=1}^Cp_\zeta^2$, as required.
\end{proof}

\begin{repproposition}{covkerPositive}
Assume $\tau$ is continuous, non-constant and $q > c > -q$. Then $$\mathfrak{C}_{\tau}(q,q) > \mathfrak{C}_{\tau}(q,c)$$
\end{repproposition}

\begin{proof}

Let $\begin{pmatrix}s \\ t \end{pmatrix}\sim \mathcal{N}(0,\Sigma)$ where $\Sigma = \begin{pmatrix} q & c \\ c & q \end{pmatrix}$. Then 

\begin{eqnarray*}
&&\mathfrak{C}_\tau(q,q) - \mathfrak{C}_\tau(q,c)\\
&=& \mathbb{E}_s \tau(s)^2 -\mathbb{E}_{s,t} \tau(s)\tau(t) \\
&=& \mathbb{E}_s \tau(s)^2 - 2\mathbb{E}_{s,t} \tau(s)\tau(t) + \mathbb{E}_t \tau(t)^2 \\
&=& \mathbb{E}_{s,t} (\tau(s) - \tau(t))^2 \\
&=& \int_{s,t \in \mathbb{R}^2} n\bigg(0,\Sigma,\begin{pmatrix}s \\ t \end{pmatrix}\bigg)(\tau(s) - \tau(t))^2d\mu_1
\end{eqnarray*}

$n\bigg(0,\Sigma,\begin{pmatrix}s \\ t \end{pmatrix}\bigg)(\tau(s) - \tau(t))^2$ is continuous because $\tau$ is continuous. It is clearly non-negative. It is not zero everywhere because $\tau$ is non-constant. And by assumption \ref{assumptionIntegrableMeanField}, it is integrable. Hence, by lemma \ref{lemma8}, its integral is positive, and so $\mathfrak{C}_\tau(q,q) > \mathfrak{C}_\tau(q,c)$.
\end{proof}

\begin{repproposition}{covkerPositive2}
For any $\tau$ and $q \ge c \ge 0$, we have $$\mathfrak{C}_{\tau}(q,c) \ge 0$$
\end{repproposition}

\begin{proof}
By proposition \ref{covkerBisC}, we have $\mathfrak{C}_{\tau}(q,c) = \mathfrak{B}_\tau(q,c) = \mathbb{E}_{t\sim\mathcal{N}(0,c)}(\mathbb{E}_{s\sim\mathcal{N}(0,q-c)}\tau(t+s))^2 \ge 0$.
\end{proof}

\begin{repproposition}{covkerPositive3}
Assume $\tau$ is continuous, not the zero function and $q > 0$. Then $$\mathfrak{C}_{\tau}(q,q) > 0$$
\end{repproposition}

\begin{proof}
We have $\mathfrak{C}_{\tau}(q,q) = \mathbb{E}_{s\sim\mathcal{N}(0,q)}\tau(s)^2=\int_{\mathbb{R}}\tau(s)^2n(q,s)d\mu_1$. By lemma \ref{lemma8}, this is positive.
\end{proof}

\begin{repproposition}{mfntPositive}
Let $f$ be an A-architecture. Let $\mathfrak{q}_l$, $\mathfrak{c}_l$ and $\mathfrak{g}_l$ be defined according to table \ref{tableNLCPropagation} where $q > c \ge 0$. Then $\mathfrak{q}_l > \mathfrak{c}_l \ge 0$ and $\mathfrak{g}_l > 0$ for all $l$.
\end{repproposition}

\begin{proof}
As with theorem \ref{mfntPropagation}, we prove this proposition by induction on $l$. Hence, in each calculation rule in table \ref{tableNLCPropagation}, we can assume the statement of the proposition holds for the dependencies. Considering that $\sigma_l^2 > 0$ for fully-connected (FC) layers by property \ref{aprop5}, the only calculation rules that do not obviously yield $\mathfrak{q}_l > \mathfrak{c}_l \ge 0$ and $\mathfrak{g}_l > 0$ are those for the activation operation. For an activation layer $f_l$ we have

\begin{eqnarray*}
\mathfrak{q}_l &=& \mathfrak{C}_{\tau_l}(\mathfrak{q}_k,\mathfrak{q}_k)\\
\mathfrak{c}_l &=& \mathfrak{C}_{\tau_l}(\mathfrak{q}_k,\mathfrak{c}_k)\\
\mathfrak{g}_l &=& \mathfrak{g}_k\mathfrak{C}_{\tau_l'}(\mathfrak{q}_k,\mathfrak{q}_k)
\end{eqnarray*}

Because $\tau$ is continuous and non-constant by property \ref{aprop2}, and because $\mathfrak{q}_k > \mathfrak{c}_k \ge 0$, we have $\mathfrak{C}_{\tau_l}(\mathfrak{q}_k,\mathfrak{q}_k) > \mathfrak{C}_{\tau_l}(\mathfrak{q}_k,\mathfrak{c}_k)$ and hence $\mathfrak{q}_l > \mathfrak{c}_l$ by proposition \ref{covkerPositive}. By proposition \ref{covkerPositive2}, we have $\mathfrak{c}_l = \mathfrak{C}_{\tau_l}(\mathfrak{q}_k,\mathfrak{c}_k) \ge 0$. Because $\tau$ is non-constant, $\tau'$ is not the zero function. Because $\tau$ is twice differentiable by property \ref{aprop2}, $\tau'$ is continuous. So by proposition \ref{covkerPositive3}, we have $\mathfrak{C}_{\tau_l'}(\mathfrak{q}_k,\mathfrak{q}_k) > 0$ and hence $\mathfrak{g}_l = \mathfrak{g}_k\mathfrak{C}_{\tau_l'}(\mathfrak{q}_k,\mathfrak{q}_k) > 0$ as required.

\end{proof}

\begin{repproposition}{mfntNPE}
Let $\mathfrak{q}_l$, $\mathfrak{c}_l$ and $\mathfrak{g}_l$ be defined according to table \ref{tableNLCPropagation}. Let $0 \le m \le l \le L$ and let $f_m$ be a bottleneck for $f_l$ in the A-architecture $f$. Then 

$$\mathfrak{n}_{l,m}(f,q,c) = \sqrt{\sum_{p \in P_{m,l}} w(p) \prod_{f_{l'}\in p} \mathfrak{n}_{\tau_{l'}}(\mathfrak{q}_{k'},\mathfrak{c}_{k'})^2}$$

where $P_{m,l}$ is the set of directed paths from $f_m$ to $f_l$, the product is over all activation layers $f_{l'}$ on path $p$ excluding its starting point and $f_{k'}$ is the dependency of $f_{l'}$.
\end{repproposition}

\begin{proof}
Denote the right-hand side of the equation by $RHS_{l,m}$. All denominators that arise in this proof are non-zero by proposition \ref{mfntPositive}.

As with theorem \ref{mfntPropagation}, we prove this proposition by induction on $l$. Fix $m$ throughout the remainder of the proof. The base case of our induction is $l=m$. In that case, we have a single path from $f_m$ to $f_l$ containing only one vertex that is also its starting point. Hence, that path contains no activation or addition layers outside of its starting point and so its weight is 1 and so $RHS_{l,m}=1$. We also have $\mathfrak{n}_{l,m}=\sqrt{\frac{\mathfrak{g}_l(\mathfrak{q}_m-\mathfrak{c}_m)}{\mathfrak{g}_m(\mathfrak{q}_l-\mathfrak{c}_l)}}=\sqrt{\frac{\mathfrak{g}_m(\mathfrak{q}_m-\mathfrak{c}_m)}{\mathfrak{g}_m(\mathfrak{q}_m-\mathfrak{c}_m)}}=1$, as required.

Now we proceed based on layer operation for the induction step based on $l$. Note that we now have $l > m$, so $l \neq 0$ so $f_l$ is not an input layer.

\begin{itemize}
\item Case: $f_l$ is an FC, bias, LN or BN layer. Then $P_{m,l}$ is the same as $P_{m,k}$, except that one vertex is added to each path. Also, since this new vertex is neither an addition nor activation layer, $RHS_{l,m}=RHS_{k,m}$. Further,

$$\mathfrak{n}_{l,m} = \sqrt{\frac{\mathfrak{g}_l(\mathfrak{q}_m-\mathfrak{c}_m)}{\mathfrak{g}_m(\mathfrak{q}_l-\mathfrak{c}_l)}} = \sqrt{\frac{\mathfrak{g}_k(\mathfrak{q}_m-\mathfrak{c}_m)}{\mathfrak{g}_m(\mathfrak{q}_k-\mathfrak{c}_k)}}\sqrt{\frac{\mathfrak{g}_l(\mathfrak{q}_k-\mathfrak{c}_k)}{\mathfrak{g}_k(\mathfrak{q}_l-\mathfrak{c}_l)}}$$

By substituting the values from table \ref{tableNLCPropagation} for $\mathfrak{g}_l$, $\mathfrak{q}_l$ and $\mathfrak{c}_l$, we find that $\sqrt{\frac{\mathfrak{g}_l(\mathfrak{q}_k-\mathfrak{c}_k)}{\mathfrak{g}_k(\mathfrak{q}_l-\mathfrak{c}_l)}}=1$, and hence $\mathfrak{n}_{l,m}=\mathfrak{n}_{k,m}$. Since $\mathfrak{n}_{k,m}=RHS_{k,m}$, we have $\mathfrak{n}_{l,m}=RHS_{l,m}$.
\item Case: $f_l$ is an activation layer. Then $P_{m,l}$ is the same as $P_{m,k}$, except that one vertex is added to each path. Since this new vertex is an activation layer, one value is added to the product for each path, but the weights remain the same. Hence 

\begin{eqnarray*}
&&RHS_{l,m}\\
&=&RHS_{k,m}\mathfrak{n}_{\tau_l}(\mathfrak{q}_k,\mathfrak{c}_k)\\
&=&\mathfrak{n}_{k,m}\mathfrak{n}_{\tau_l}(\mathfrak{q}_k,\mathfrak{c}_k)\\
&=&\sqrt{\frac{\mathfrak{g}_k(\mathfrak{q}_m-\mathfrak{c}_m)}{\mathfrak{g}_m(\mathfrak{q}_k-\mathfrak{c}_k)}}\sqrt{\frac{(\mathfrak{q}_k-\mathfrak{c}_k)\mathfrak{C}_{\tau_l'}(\mathfrak{q}_k,\mathfrak{q}_k)}{\mathfrak{C}_{\tau_l}(\mathfrak{q}_k,\mathfrak{q}_k)-\mathfrak{C}_{\tau_l}(\mathfrak{q}_k,\mathfrak{c}_k)}}\\
&=&\sqrt{\frac{\mathfrak{g}_k(\mathfrak{q}_m-\mathfrak{c}_m)}{\mathfrak{g}_m(\mathfrak{q}_k-\mathfrak{c}_k)}}\sqrt{\frac{(\mathfrak{q}_k-\mathfrak{c}_k)\frac{\mathfrak{g}_l}{\mathfrak{g}_k}}{\mathfrak{q}_l-\mathfrak{c}_l}}\\
&=&\sqrt{\frac{\mathfrak{g}_l(\mathfrak{q}_m-\mathfrak{c}_m)}{\mathfrak{g}_m(\mathfrak{q}_l-\mathfrak{c}_l)}}\\
&=&\mathfrak{n}_{l,m}\\
\end{eqnarray*}

as required.
\item Case: $f_l$ is an addition layer. Then $P_{m,l}$ is the union of the distinct sets $P_{m,k[\kappa]}$ for $1 \le \kappa \le K$, where each path also has the vertex $f_l$ added. The product for each path is still the same, but the weight is multiplied by $w_{\kappa}^2\frac{\mathfrak{q}_{k[\kappa]}-\mathfrak{c}_{k[\kappa]}}{\mathfrak{q}_l-\mathfrak{c}_l}$. So we have

\begin{eqnarray*}
&&RHS_{l,m}\\
&=&\sqrt{\sum_{\kappa=1}^Kw_{\kappa}^2\frac{\mathfrak{q}_{k[\kappa]}-\mathfrak{c}_{k[\kappa]}}{\mathfrak{q}_l-\mathfrak{c}_l}\sum_{p \in P_{m,k[\kappa]}} w(p) \prod_{f_{l'}\in p} \mathfrak{n}_{\tau_{l'}}(\mathfrak{q}_{k'},\mathfrak{c}_{k'})^2}\\
&=&\sqrt{\sum_{\kappa=1}^Kw_{\kappa}^2\frac{\mathfrak{q}_{k[\kappa]}-\mathfrak{c}_{k[\kappa]}}{\mathfrak{q}_l-\mathfrak{c}_l}\frac{\mathfrak{g}_{k[\kappa]}(\mathfrak{q}_m-\mathfrak{c}_m)}{\mathfrak{g}_m(\mathfrak{q}_{k[\kappa]}-\mathfrak{c}_{k[\kappa]})}}\\
&=&\sqrt{\frac{\sum_{\kappa=1}^Kw_{\kappa}^2\mathfrak{g}_{k[\kappa]}(\mathfrak{q}_m-\mathfrak{c}_m)}{\mathfrak{g}_m(\mathfrak{q}_l-\mathfrak{c}_l)}}\\
&=&\sqrt{\frac{\mathfrak{g}_l(\mathfrak{q}_m-\mathfrak{c}_m)}{\mathfrak{g}_m(\mathfrak{q}_l-\mathfrak{c}_l)}}\\
\end{eqnarray*}
\end{itemize}

as required.

\end{proof}

\begin{repproposition}{covkerBisC}
For any $\tau$ and $q \ge c \ge 0$, we have $$\mathfrak{B}_\tau(q,c) = \mathfrak{C}_\tau(q,c)$$
\end{repproposition}

\begin{proof}

Let $u$ be a 3-dimensional unit Gaussian column vector. Then $\sqrt{q}u[0]$ and $\frac{c}{\sqrt{q}}u[0] + \sqrt{q-\frac{c^2}{q}}u[1]$ are jointly Gaussian with mean zero and covariance $\begin{pmatrix}q & c \\ c & q \end{pmatrix}$, and so $\mathfrak{C}_\tau(q,c) = \mathbb{E}_u\tau(\sqrt{q}u[0])\tau(\frac{c}{\sqrt{q}}u[0] + \sqrt{q-\frac{c^2}{q}}u[1])$. Let 

$$A = \begin{pmatrix}\sqrt{\frac{c}{q}} & \sqrt{\frac{q-c}{q}} & 0 \\ \sqrt{\frac{c(q-c)}{q(q+c)}} & -\sqrt{\frac{c^2}{q(q+c)}} & \sqrt{\frac{q}{q(q+c)}} \\ \sqrt{\frac{q-c}{q+c}} & -\sqrt{\frac{c}{q+c}} & -\sqrt{\frac{c}{q+c}} \end{pmatrix}$$

Because $A$ is orthogonal and the distribution of $u$ is spherically symmetric, $u$ and $Au$ have the same distribution. So we have

\begin{eqnarray*}
&&\mathfrak{C}_\tau(q,c)\\
&=&\mathbb{E}_u\tau\Big(\begin{pmatrix}\sqrt{q}&0&0\end{pmatrix}u\Big)\tau\Big(\begin{pmatrix}\frac{c}{\sqrt{q}}&\sqrt{q-\frac{c^2}{q}}&0\end{pmatrix}u\Big)\\
&=&\mathbb{E}_u\tau\Big(\begin{pmatrix}\sqrt{q}&0&0\end{pmatrix}Au\Big)\tau\Big(\begin{pmatrix}\frac{c}{\sqrt{q}}&\sqrt{q-\frac{c^2}{q}}&0\end{pmatrix}Au\Big)\\
&=&\mathbb{E}_u\tau\Big(\begin{pmatrix}\sqrt{c}&\sqrt{q-c}&0\end{pmatrix}u\Big)\tau\Big(\begin{pmatrix}\sqrt{c}&0&\sqrt{q-c}\end{pmatrix}u\Big)\\
&=&\mathbb{E}_{s,t,r\sim\mathcal{N}(0,1)}\tau(\sqrt{c}s+\sqrt{q-c}t)\tau(\sqrt{c}s+\sqrt{q-c}r)\\
&=&\mathbb{E}_{s\sim\mathcal{N}(0,1)}(\mathbb{E}_{t\sim\mathcal{N}(0,1)}\tau(\sqrt{c}s+\sqrt{q-c}t))(\mathbb{E}_{r\sim\mathcal{N}(0,1)}\tau(\sqrt{c}s+\sqrt{q-c}r))\\
&=&\mathbb{E}_{s\sim\mathcal{N}(0,1)}(\mathbb{E}_{t\sim\mathcal{N}(0,1)}\tau(\sqrt{c}s+\sqrt{q-c}t))^2\\
&=&\mathbb{E}_{s\sim\mathcal{N}(0,c)}(\mathbb{E}_{t\sim\mathcal{N}(0,q-c)}\tau(s+t))^2\\
&=&\mathfrak{B}_\tau(q,c)
\end{eqnarray*}

\end{proof}

\begin{repproposition}{covkernregular}
Assume $\tau$ is piecewise 5-differentiable. Consider $\mathfrak{n}_\tau(c)$ defined on $[0, 1)$. Then

\begin{enumerate}
\item $$\mathfrak{n}_\tau(c) = \sqrt{\frac{\mathfrak{C}_\tau'(1)(1-c)}{\mathfrak{C}_\tau(1)-\mathfrak{C}_\tau(c)}} = \sqrt{\frac{\tilde{\mathfrak{C}}_\tau'(1)(1-c)}{1-\tilde{\mathfrak{C}}_\tau(c)}}$$
\item $\mathfrak{n}_\tau$ is decreasing
\item $\lim_{c \rightarrow 1}\mathfrak{n}_\tau(c) = 1$
\item $\mathfrak{n}_\tau(c) \ge 1$
\end{enumerate}

\end{repproposition}

\begin{proof}
Since $\tau$ is piecewise 5-differentiable, we can use theorem \ref{covkerCregular} in this proof. By definition, $\mathfrak{n}_\tau(c) = \sqrt{\frac{\mathfrak{C}_{\tau'}(1)(1 - c)}{\mathfrak{C}_{\tau}(1) - \mathfrak{C}_{\tau}(c)}}$. By theorem \ref{covkerCregular}, we have $\mathfrak{C}_{\tau'}(1) = \mathfrak{C}_\tau'(1)$. This yields the first part of claim 1. The second part is obtained when dividing both numerator and denominator by $\mathfrak{C}_\tau(1)$.

Let $1 > c' > c \ge 0$. Then $\frac{\mathfrak{C}_\tau(1) - \mathfrak{C}_\tau(c)}{1-c}$ is the gradient of the chord from $(c, \mathfrak{C}_\tau(c))$ to $(1, \mathfrak{C}_\tau(1))$ and $\frac{\mathfrak{C}_\tau(1) - \mathfrak{C}_\tau(c')}{1-c'}$ is the gradient of the chord from $(c', \mathfrak{C}_\tau(c'))$ to $(1, \mathfrak{C}_\tau(1))$. By theorem \ref{covkerCregular}, $\mathfrak{C}_\tau$ is convex and so $(c', \mathfrak{C}_\tau(c'))$ lies on or below the chord from $(c, \mathfrak{C}_\tau(c))$ to $(1, \mathfrak{C}_\tau(1))$. Hence, the chord corresponding to $c'$ has at least as large a gradient as the chord corresponding to $c$. So $\frac{\mathfrak{C}_\tau(1) - \mathfrak{C}_\tau(c)}{1-c} \le \frac{\mathfrak{C}_\tau(1) - \mathfrak{C}_\tau(c')}{1-c'}$, so $\sqrt{\frac{\mathfrak{C}_{\tau'}(1)(1 - c)}{\mathfrak{C}_{\tau}(1) - \mathfrak{C}_{\tau}(c)}} \ge \sqrt{\frac{\mathfrak{C}_{\tau'}(1)(1 - c')}{\mathfrak{C}_{\tau}(1) - \mathfrak{C}_{\tau}(c')}}$, which yields claim 2.

$\mathfrak{C}_\tau'(1) = \lim_{c \rightarrow 1}\frac{\mathfrak{C}_\tau(1) - \mathfrak{C}_\tau(c)}{1-c}$ by the definition of the derivative. Hence, $\lim_{c \rightarrow 1}\frac{\mathfrak{C}_\tau'(1)(1-c)}{\mathfrak{C}_\tau(1) - \mathfrak{C}_\tau(c)} = 1$, which yields claim 3.

Claim 4 follows from claims 2 and 3.

\end{proof}

\begin{repproposition}{mfntCNLC}
Let $f$ be an A-architecture where each activation function used is piecewise 5-differentiable and let $q > c \ge 0$. If $f$ does not contain batch normalization but can contain layer normalization layers, we have 

$$\mathfrak{n}(f,q,c) = \sqrt{\frac{\frac{d}{dq'}\mathfrak{C}(q,q')|_{q'=q}(q - c)}{\mathfrak{C}(q,q) - \mathfrak{C}(q,c)}}$$

If $f$ does not contain layer normalization but can contain batch normalization layers, an analogous statement holds. See section \ref{meanFieldBNsection} for details.
\end{repproposition}

\begin{proof}

Since the activation functions used are piecewise 5-differentiable, we can use theorem \ref{covkerCregular} in this proof. All denominators that arise in this proof are non-zero by proposition \ref{mfntPositive}. Denote by $\mathfrak{C}_l(q,c)$ the value of $\mathfrak{c}_l$ obtained from table \ref{tableNLCPropagation} when starting the recursion with $q$ and $c$. Then $\mathfrak{C}_L(q,c)=\mathfrak{C}(q,c)$. Note that this is different from $\mathfrak{C}_{\tau_l}$. We will prove the more general statement

$$\mathfrak{n}_{l,0}(f,q,c) = \sqrt{\frac{\frac{d}{dq'}\mathfrak{C}_l(q,q')|_{q'=q}(q - c)}{\mathfrak{C}_l(q,q) - \mathfrak{C}_l(q,c)}}$$

Denote the right-hand side of this equation by $RHS_l$. It is easy to check that $\mathfrak{C}_l(q,q)=\mathfrak{q}_l$. So the left-hand side of the equation is $\sqrt{\frac{\mathfrak{g}_l(\mathfrak{q}_0-\mathfrak{c}_0)}{\mathfrak{g}_0(\mathfrak{q}_l-\mathfrak{c}_l)}}$ and the right-hand side is $\sqrt{\frac{\frac{d}{dq'}\mathfrak{C}_l(q,q')|_{q'=q}(\mathfrak{q}_0-\mathfrak{c}_0)}{\mathfrak{q}_l-\mathfrak{c}_l}}$. Also $\mathfrak{g}_0=1$. So we simply need to show

$$\mathfrak{g}_l=\frac{d}{dq'}\mathfrak{C}_l(q,q')|_{q'=q}$$

As with theorem \ref{mfntPropagation}, we prove this proposition by induction on $l$. The base case of our induction is $l=0$, where $\mathfrak{C}_l$ is the identity so both sides are equal to 1. Now we proceed based on layer operation for the induction step based on $l$. Since we now have $l > 0$, $f_l$ is not an input layer. 

\begin{itemize}
\item Case: $f_l$ is an FC, bias, LN or activation layer. We have

$$\frac{d}{dq'}\mathfrak{C}_l(q,q')|_{q'=q}=\frac{d\mathfrak{c}_l}{dc}|_{c=q}=\frac{d\mathfrak{c}_l}{d\mathfrak{c}_k}|_{\mathfrak{c}_k=\mathfrak{q}_k}\frac{d\mathfrak{c}_k}{dc}|_{c=q}$$
$$=\frac{d\mathfrak{c}_l}{d\mathfrak{c}_k}|_{\mathfrak{c}_k=\mathfrak{q}_k}\frac{d}{dq'}\mathfrak{C}_k(q,q')|_{q'=q}=\frac{d\mathfrak{c}_l}{d\mathfrak{c}_k}|_{\mathfrak{c}_k=\mathfrak{q}_k}\mathfrak{g}_k$$

So we must show that $$\frac{\mathfrak{g}_l}{\mathfrak{g}_k}=\frac{d\mathfrak{c}_l}{d\mathfrak{c}_k}|_{\mathfrak{c}_k=\mathfrak{q}_k}$$

The $\frac{d\mathfrak{c}_l}{d\mathfrak{c}_k}|_{\mathfrak{c}_k=\mathfrak{q}_k}$ term is $\sigma_l^2$ for FC, 1 for bias, and $\frac{1}{\mathfrak{q}_k}$ for LN. It is easy to check that this equals $\frac{\mathfrak{g}_l}{\mathfrak{g}_k}$. If $f_l$ is an activation layer, we have $\frac{d\mathfrak{c}_l}{d\mathfrak{c}_k}|_{\mathfrak{c}_k=\mathfrak{q}_k}=\frac{d}{dq'}\mathfrak{C}_{\tau_l}(\mathfrak{q}_k,q')|_{q'=\mathfrak{q}_k}$ and $\frac{\mathfrak{g}_l}{\mathfrak{g}_k}=\mathfrak{C}_{\tau_l'}(\mathfrak{q}_k,\mathfrak{q}_k)$. Let $\dot{\tau}_l(s) = \tau_l(\sqrt{\mathfrak{q}_k}s)$. As $\mathfrak{q}_k > 0$ by proposition \ref{mfntPositive}, we can further express $\frac{d}{dq'}\mathfrak{C}_{\tau_l}(\mathfrak{q}_k,q')|_{q'=\mathfrak{q}_k}$ as $\mathfrak{q}_k\frac{d}{dq'}\mathfrak{C}_{\dot{\tau}_l}(1,q')|_{q'=1}$ and hence $\mathfrak{q}_k\mathfrak{C}_{\dot{\tau}_l}'(1)$. Also, we can express $\mathfrak{C}_{\tau_l'}(\mathfrak{q}_k,\mathfrak{q}_k)$ as $\mathfrak{q}_k\mathfrak{C}_{\dot{\tau}_l'}(1,1)$ and hence $\mathfrak{q}_k\mathfrak{C}_{\dot{\tau}_l'}(1)$. And $\mathfrak{q}_k\mathfrak{C}_{\dot{\tau}_l}'(1)=\mathfrak{q}_k\mathfrak{C}_{\dot{\tau}_l'}(1)$ follows from theorem \ref{covkerCregular}.
\item Case: $f_l$ is an addition layer. We have

$$\frac{d}{dq'}\mathfrak{C}_l(q,q')|_{q'=q}=\frac{d\mathfrak{c}_l}{dc}|_{c=q}=\sum_{\kappa=1}^K\frac{d\mathfrak{c}_l}{d\mathfrak{c}_{k[\kappa]}}|_{\mathfrak{c}_{k[\kappa]}=\mathfrak{q}_{k[\kappa]}}\frac{d\mathfrak{c}_{k[\kappa]}}{dc}|_{c=q}$$
$$=\sum_{\kappa=1}^Kw_\kappa^2\frac{d}{dq'}\mathfrak{C}_{k[\kappa]}(q,q')|_{q'=q}=\sum_{\kappa=1}^Kw_\kappa^2\mathfrak{g}_{k[\kappa]}=\mathfrak{g}_l$$

\end{itemize}
\end{proof}

\section{Theorems}

\subsection{Theorem \ref{mfntMetaGaussian}: fully-connected layers are meta-Gaussian meta-distributed} \label{mfntMetaGaussiansection}

\begin{reptheorem}{mfntMetaGaussian}

Let $f$ be a mean field architecture with a single finite input layer and a single readout layer. Let $\mathcal{G}$ be a generator distribution associated with the non-finite input layers of $f$. Let $\mathcal{D}$ be an input distribution associated with the finite input layer of $f$. Assume:

\begin{itemize}
\item $\mathcal{D}$ is elem-like$(q,c)$.
\item $(\clipout(\clipin(f))^N, \mathcal{G}(K(N,q,c)) \times \mathcal{G}^N)$ has rank stability for all $N$.
\item $\mathcal{G}$ is Gaussian.
\item All multi-activation functions used in $f$ are parameter-controlled by $\mathcal{C}^\text{E2}$ at $(\mathfrak{m}_{\hat{k}[1]}, .., \mathfrak{m}_{\hat{k}[\hat{K}]})$.
\end{itemize}

Then as $d_\text{MF}$ converges to infinity, the meta-distribution of the readout layer expansion-converges to the elementwise meta-distribution with generator $\mathcal{MN}(\mathfrak{C}(q,q),\mathfrak{C}(q,c))$. $\mathfrak{C}$ and the $\mathfrak{m}_{\hat{k}[\hat{\kappa}]}$ are derived via table \ref{tableBackgroundPropagation} applied to $(\clipin(f)^N, \mathcal{G}(K(N,q,c)) \times \mathcal{G}^N)$. The first level of randomness is induced by $\mathcal{D}$. The second level of randomness is induced by $\theta$ and $\mathcal{G}$.
\end{reptheorem}

\begin{proof}
To prove expansion-convergence, we need to show that the $N$-expansion converges in distribution for any $N \ge 1$. Fix $N$ for the remainder of the proof. The $N$-expansion of the output meta-distribution is the distribution of $(f(x^{(1)}), .., f(x^{(N)}))$ where randomness is induced by the $N$ draws $x^{(1)}, .., x^{(N)}$ from $\mathcal{D}$ along with $\theta$ and $\mathcal{G}$. 

For now, consider a fixed sample $x^{(1)}, .., x^{(N)}$ from $\mathcal{D}$ with co-mean kernel matrix $K(N,q,c)$. Because this theorem assumes rank stability, Gaussianity and parameter-control, we can apply background corollary \ref{backgroundKernel} to obtain that $(f(x^{(1)}), .., f(x^{(N)}))$ converges in distribution to a distribution generated by $\mathcal{N}(0,K(N,\mathfrak{C}(q,q),\mathfrak{C}(q,c)))$ when randomness is induced by $\mathcal{\theta}$ and $\mathcal{G}$. We denote this generator by $\mathcal{O}$. Let $\theta^\text{in}$ refer to the components of the parameter that belong to the readin layers and let $\theta^\text{MF}$ refer to the remaining components. We have that the joint distribution of the readin layer values across the sample is generated by $\mathcal{G}(K(N,q,c))$ when randomness is induced by $\theta^\text{in}$. Since $(f(x^{(1)}), .., f(x^{(N)}))$ only depends on the sample and $\theta^\text{in}$ via the readin layer values themselves, its limit is generated by $\mathcal{O}$ when randomness is induced by the readin layer values, $\mathcal{G}$ and $\theta^\text{MF}$.

Now we let the sample $x^{(1)}, .., x^{(N)}$ be random again. Let $E$ be the event that its co-mean kernel matrix is $K(N,q,c)$. Conditioned on $E$, the joint distribution of the readin layer values across the sample is generated by $\mathcal{G}(K(N,q,c))$ when randomness is induced by $\theta^\text{in}$ for any value of the sample. So it is generated by $\mathcal{G}(K(N,q,c))$ when randomness is induced by $\theta^\text{in}$ and the sample. But above we found that this joint distribution over readin layer values then further induces our joint distribution of output values. Hence, the limit of $(f(x^{(1)}), .., f(x^{(N)}))$ is generated by $\mathcal{O}$ when randomness is induced by the sample, $\mathcal{G}$ and $\theta$. This is the limit of the $N$-expansion of $f$. Since the event $E$ has probability 1, we obtain the same convergence in distribution when not conditioning on $E$.

What remains to be shown is that $\mathcal{O}$ is the $N$-expansion of $\mathcal{MN}(\mathfrak{C}(q,q),\mathfrak{C}(q,c))$. Drawing $N$ scalars $s^{(1)}, .., s^{(N)}$ from $\mathcal{MN}(\mathfrak{C}(q,q),\mathfrak{C}(q,c))$ corresponds to drawing a mean variable $\mu$ from $\mathcal{N}(0,c)$ and residual variables $r^{(1)}, .., r^{(N)}$ from $\mathcal{N}(0,q-c)$, all independent, and setting $s^{(n)} = \mu + r^{(n)}$ for all $n$. So $(s^{(1)}, .., s^{(N)})$ is a linear transformation of the multivariate Gaussian $(\mu, r^{(1)}, .., r^{(N)})$, so it is itself a multi-variate Gaussian. Finally, it is easy to check that the covariance matrix of $(s^{(1)}, .., s^{(N)})$ is indeed $K(N,\mathfrak{C}(q,q),\mathfrak{C}(q,c))$ as required.

\end{proof}

\subsection{Theorem \ref{mfntPropagation}: mean field limits of practical metrics and architectures} \label{mfntPropagationsection}

\begin{reptheorem}{mfntPropagation}
Let ...

\begin{itemize}
\item ... $f$ be an A-architecture with an output layer of variable width.
\item ... $\mathcal{D}$ be an input distribution.
\item ... $x^{(1)},..,x^{(N)}$ be a sample from $\mathcal{D}$ of size $N \ge 2$ and let $\mathcal{D}^{(N)}$ be the discrete uniform distribution over that sample.
\item ... $\vec{\epsilon}$ be the vector of all regularizers used by normalization layers in $f$.
\end{itemize}

Assume:

\begin{itemize}
\item $\mathcal{D}$ is elem-like$(q,c)$.
\item $q > c$
\end{itemize}

If $f$ does not contain batch normalization but can contain layer normalization layers, we have

\begin{eqnarray*}
\lim\mathbb{E}_if_l(x^{(1)})[i] &=& \mathfrak{m}_l\\
\lim\mathbb{E}_if_l(x^{(1)})[i]^2 &=& \mathfrak{q}_l\\
\lim\mathbb{E}_if_l(x^{(1)})[i]f_l(x^{(2)})[i] &=& \mathfrak{c}_l \\
\lim\frac{1}{d_l^\text{MF}}||\mathcal{J}_{l,m}(x^{(1)})||_F^2 &=& \frac{\mathfrak{g}_l}{\mathfrak{g}_m}\\
\lim\frac{1}{d_l^\text{MF}}\mathbb{E}_x||f_l(x)||^2_2 &=& \mathfrak{q}_l \\
\lim\frac{1}{d_l^\text{MF}}||\mathbb{E}_xf_l(x)||^2_2 &=& \mathfrak{c}_l \\
\lim\frac{1}{d_l^\text{MF}}||\mathbb{S}_xf_l(x)||^2_2 &=& \mathfrak{q}_l - \mathfrak{c}_l \\
\lim\frac{1}{d_l^\text{MF}}\mathbb{E}_x||\mathcal{J}_{l,m}(x)||_F^2 &=& \frac{\mathfrak{g}_l}{\mathfrak{g}_m} \\
\lim\frac{1}{d_l^\text{MF}}\mathbb{E}_x\Tr(\mathcal{J}_{l,m}(x)\Cov_{f_m}\mathcal{J}_{l,m}(x)^T) &=& \frac{\mathfrak{g}_l(\mathfrak{q}_m - \mathfrak{c}_m)}{\mathfrak{g}_m}\\
\lim NLC(f_l(f_m),f_m(\mathcal{D}^{(N)})) &=& \sqrt{\frac{\mathfrak{g}_l(\mathfrak{q}_m - \mathfrak{c}_m)}{\mathfrak{g}_m(\mathfrak{q}_l - \mathfrak{c}_l)}}
\end{eqnarray*}

where

\begin{itemize}
\item ... $x \sim \mathcal{D}^{(N)}$
\item ... $\lim$ stands for $\lim_{\vec{\epsilon}\rightarrow 0,N\rightarrow\infty}\big(\lim_{d_\text{MF}\rightarrow\infty} ... \text{a.s.}\big)$. The inner limit takes $d_\text{MF}$ to infinity and the outer limit takes $N$ to infinity as well as $\vec{\epsilon}$ to zero. The inner limit is an ``almost sure'' limit.
\item ... $\mathfrak{m}_l$, $\mathfrak{q}_l$, $\mathfrak{c}_l$ and $\mathfrak{g}_l$ are calculated via table \ref{tableNLCPropagation}. 
\item ... $f_m$ is a bottleneck for $f_l$.
\item ... $0 \le m \le l \le L$
\item ... randomness is induced by $\theta$ and the $x^{(n)}$.
\end{itemize}

Further, if the output layer is instead a readout layer of fixed dimensionality $d_\text{out}$, the above limits do not necessarily hold when $l=L$. Instead, (a) as $d_\text{MF} \rightarrow \infty$, the meta-distribution of the output layer expansion-converges to the elementwise meta-distribution with generator $\mathcal{MN}(\mathsf{parm1}, \mathsf{parm2})$, where $\lim_{\vec{\epsilon}\rightarrow 0}\mathsf{parm1}=\mathfrak{q}_L$ and $\lim_{\vec{\epsilon}\rightarrow 0}\mathsf{parm2}=\mathfrak{c}_L$. The first level of randomness is induced by $\mathcal{D}$ and the second level by $\theta$. And, (b) when $N$ is fixed, for almost all samples, as $d_\text{MF} \rightarrow \infty$, $(f(x^{(1)}), .., f(x^{(N)}))$ converges in distribution to a Gaussian that is elementwise over output vectors where the generator has mean zero and covariance matrix $K(N, \mathsf{parm1}, \mathsf{parm2})$, where again $\lim_{\vec{\epsilon}\rightarrow 0}\mathsf{parm1}=\mathfrak{q}_L$ and $\lim_{\vec{\epsilon}\rightarrow 0}\mathsf{parm2}=\mathfrak{c}_L$. Randomness for each sample is induced by $\theta$.

If $f$ does not contain layer normalization but can contain batch normalization layers, analogous statements hold. See section \ref{meanFieldBNsection} for details.
\end{reptheorem}

\begin{proof}

As in theorem \ref{finiteNetTilde}, we proceed in stages. Later stages build upon earlier stages. In each stage, we prove some aspect of the theorem.

\underline{Stage 1}: statements (\ref{eqn5p0}) and (\ref{eqn5p1})

For this stage, denote $x^{(1)}$ simply by $x$. Let's begin by checking statements (\ref{eqn5p0}) and (\ref{eqn5p1}) for the case $l=0$. Since $f_0(x) = x$ is the input layer of $f$, $\mathfrak{m}_0$ is undetermined according to table \ref{tableNLCPropagation}, so there is nothing to check. Because $\mathcal{D}$ is elem-like$(q,c)$, we have $\mathbb{E}_ix[i]^2=q$ with probability 1. So we have $\lime\limd\mathbb{E}_if_0(x)[i]^2=q$, which matches $\mathfrak{q}_0=q$ from table \ref{tableNLCPropagation} as required.

Because the event $\mathbb{E}_ix[i]^2=q$ has probability 1, almost sure convergence conditioned on that event implies almost sure convergence in general. So for the remainder of this stage, we can assume $\mathbb{E}_ix[i]^2=q$ holds. Going forward, for the case $l > 1$, we want to use background theorem \ref{backgroundMaster} to prove almost sure convergence to a limit as $d_\text{MF}\rightarrow\infty$. Then we take the limit of that limit as $\vec{\epsilon}\rightarrow 0,N\rightarrow\infty$. The background theorem requires a mean field architecture and a generator distribution. We denote the former by $F$ to distinguish it from $f$ and the latter by $\mathcal{G}$. We wish to come up with such an $F$ and $\mathcal{G}$ such that for each layer $f_l$ in $f$, there exists a ``corresponding'' layer $F_l$ in $F$ such that the joint distribution of the $f_l(x)$ for $0 \le l \le L$ is equal to the joint distribution of the $F_l$. We will refer to this criterion as ``distribution equality''. In the case of the $f_l(x)$, randomness is induced by the parameter of $f$ and in the case of the $F_l$, it is induced by $\mathcal{G}$ and the parameter of $F$. As stated in section \ref{meanFieldDistributionSection}, $\mathcal{G}$ will be used to model trainable parameter sub-vectors in non-fully-connected layers of $f$. In A-architectures, only fully-connected (FC) and bias layers have a non-empty sub-vector.

We proceed by induction on $l$, starting from $l=1$ and ending at $l=L$. At the $l$'th step, we do the following. 

\begin{itemize}
\item If $l = 1$, we initialize $F$ with one or more layers including $F_1$. If $l > 1$, we add one or more layers to $F$, including $F_l$. If in this process we added input layers to $F$, we add an additional dimension to the generator $\mathcal{G}$ for each such input layer.
\item We show that $(f_0(x), .., f_l(x))$ has the same distribution as $(F_0, .., F_l)$, i.e. distribution equality holds up to $f_l$.
\item We show that $(F,\mathcal{G})$ has rank stability (RS), parameter-control (PC), and that $\mathcal{G}$ is Gaussian as required by background theorem \ref{backgroundMaster}.
\item We apply background theorem \ref{backgroundMaster} to the current state of $F$ to obtain $\limd \mathbb{E}_iF_l[i] = \mathfrak{m}_l$ and $\limd \mathbb{E}_iF_l[i]F_l[i] =  \mathfrak{c}_{l;l}$, where $\mathfrak{m}_l$ and $\mathfrak{c}_{l;l}$ are calculated according to table \ref{tableBackgroundPropagation}. By distribution equality, the same then holds for $f_l$.
\item We verify that $\lime \mathfrak{m}_l^\text{(table \ref{tableBackgroundPropagation})} = \mathfrak{m}_l^\text{(table \ref{tableNLCPropagation})}$ and $\lime \mathfrak{c}_{l;l} = \mathfrak{q}_{l}$, where $\mathfrak{m}_l^\text{(table \ref{tableNLCPropagation})}$ and $\mathfrak{q}_{l}$ are calculated according to table \ref{tableNLCPropagation}. This yields statements (\ref{eqn5p0}) and (\ref{eqn5p1}) for $f_l$ because of distribution equality.
\end{itemize}

Crucially, for the $l$'th step, the induction hypothesis yields all the above results for the first through $l-1$'st step. Using this, we perform the induction as follows.

\begin{itemize}
\item By construction, $\mathcal{G}$ is a zero mean Gaussian. Hence, we only need to specify novel entries in the covariance matrix when $\mathcal{G}$ is expanded.
\item Since the induction hypothesis yields that $(f_1(x), .., f_{l-1}(x))$ is distributed as $(F_1, .., F_{l-1})$, to obtain distribution equality at step $l$, it is enough to show that $F_l$ can be derived from $(F_1, .., F_{l-1})$ via the same transformation which can be used to derive $f_l$ from $(f_1, .., f_{l-1})$.
\item We will prove RS by showing that, for all valid sets $S$ in the definition of RS, $A_S^\infty$ is a full rank matrix. This trivially yields RS as also noted by \citet{meanFieldNetsorGP}. In this stage, we will construct an $F$ (i) that does not contain any weight sharing and (ii) where only layers corresponding to some $f_l$ are dependencies of FC layers. Hence, all valid non-empty sets $S$ in the definition of RS will be of size 1. So if we show as part of the induction that $\mathfrak{c}_{l;l} > 0$, then the induction hypothesis directly yields RS at any given step.
\item Since PC holds at step $l-1$ by the induction hypothesis, it is enough to show that multi-activation functions corresponding to newly added layers have the property.
\end{itemize}

Now, we give additional notation, terminology and conventions that apply throughout this proof.

\begin{itemize}
\item We refer to $\mathfrak{m}_l$ calculated according to table \ref{tableBackgroundPropagation} as $\tilde{\mathfrak{m}_l}$ in order to disambiguate with $\mathfrak{m}_l$ calculated according to table \ref{tableNLCPropagation}.
\item Layers of $F$ are denoted with the usual subscript, but also sometimes additionally with superscripts to cope with layers that do not directly correspond to a layer in $f$. Whereas the subscript is an integer that can take different values, the superscript is a fixed text modifier. Any valid combination of a superscript with a subscript value is considered a unique layer. We will extend this notation to quantities that derive from layers. For example, we write $\lim_{d_\text{MF}}F_l^\text{super} = \tilde{\mathfrak{m}}_l^\text{super}$.
\item As usual, we write $\rho_l$ for the multi-activation function used by $F_l$ if $F_l$ is an elementwise layer. However, if $F_l$ is an input or FC layer, we also introduce the ``dummy multi-activation function'' $\rho_l$ that denotes the identity function.
\item We write $F_k$ for the layer in $F$ that corresponds to $f_k$, where $f_k$ is the dependency of $f_l$. The same holds for $F_{k[\kappa]}$ when $f_l$ has multiple dependencies. Note that $F_k$ is not necessarily a dependency of $F_l$ in $F$.
\item As usual, we denote elementwise function application via $.()$. We also denote the elementwise square with $\mathsf{value}^{.2}$ and the elementwise product as $\mathsf{value}.\mathsf{value}$.
\item In this proof only, we use an overline to denote the mean of vector component values, not the expectation with respect to $\mathcal{D}$. For example, we write $\overline{f_l}$ for $\mathbb{E}_if_l[i]$.
\item Finally, we remind the reader of our notation regarding multi-activation functions given in section \ref{multiActivationSection} and regarding our layer operations given in section \ref{layerTypesSection}.
\end{itemize}

We will now proceed with the induction, which is based on the layer operation of $f_l$.

Case: $f_l$ is a readin layer. 

\begin{itemize}
\item Layer addition: We add a single input layer $F_l$ to $F$ with the same width as $f_l$. Note that $f_1$ is a readin layer, so this step initializes $F$ if $l=1$.
\item Generator addition: We add the diagonal element $q\sigma_l^2$ and off-diagonal elements of zero to the covariance matrix of $\mathcal{G}$. If $l=1$, our initial generator covariance is then $\begin{pmatrix}q\sigma_1^2\end{pmatrix}$.
\item Distribution equality: The distribution of $f_l(x)$ is elementwise with generator $\mathcal{N}(0,q\sigma_l^2)$ and independent of all previous layers (except $f_0$), just as $F_l$.
\item Parameter-control: NA
\item Limits: From table \ref{tableBackgroundPropagation}, we obtain $\tilde{\mathfrak{m}}_l=0$ and $\mathfrak{c}_{l;l} = q\sigma_l^2$ and thus $\lime \tilde{\mathfrak{m}}_l=0$ and $\lime\mathfrak{c}_{l;l}= q\sigma_l^2$. Since $f_l$ is fully-connected and its dependency is the input layer, table \ref{tableNLCPropagation} yields $\mathfrak{m}_l=0$ and $\mathfrak{q}_l = q\sigma_l^2$. So we have a match, as required.
\item Rank stability: Since $q > 0$ and $\sigma_l^2 > 0$ by property \ref{aprop5}, we have $c_{l;l} > 0$.
\end{itemize}

Case: $f_l$ is a fully-connected, non-readin layer.

\begin{itemize}
\item Layer addition: We add a single fully-connected layer $F_l$ to $F$ of width and $\sigma_l$ value equal to that of $f_l$ and dependency $F_k$. $F_l$ does not share weights with previous layers of $F$.
\item Generator addition: NA
\item Distribution equality: $f_l(f_k)$ and $F_l(F_k)$ are the same function and their weight matrix is not shared by previous layers.
\item Parameter-control: NA
\item Limits: We obtain $\tilde{\mathfrak{m}}_l=0$ and $\mathfrak{c}_{l;l} = \sigma_l^2\mathfrak{c}_{k;k}$ and hence $\lime \tilde{\mathfrak{m}}_l=0=\mathfrak{m}_l$ and $\lime\mathfrak{c}_{l;l}= \sigma_l^2\lim_{\vec{\epsilon}\rightarrow 0,N\rightarrow\infty} \mathfrak{c}_{k;k} = \sigma_l^2\mathfrak{q}_k=\mathfrak{q}_l $.
\item Rank stability: Since $\mathfrak{c}_{k;k} > 0$ and $\sigma_l^2 > 0$, we have $\mathfrak{c}_{l;l} > 0$.
\end{itemize}

Case: $f_l$ is a bias layer.

\begin{itemize}
\item Layer addition: We add an input layer $F_l^\text{in}$ and an elementwise layer $F_l$, both with the same width as $f_l$. We set $\rho_l = e_l^\text{in} + \rho_k$. ($\rho_k$ is the multi-activation function of the layer in $F$ corresponding to $f_k$, which may be a dummy multi-activation function.) The elementwise inputs of $\rho_l$ are $e_l^\text{in}$ and those of $\rho_k$. Its mean inputs are those of $\rho_k$.
\item Generator addition: We add the diagonal element $\sigma_l^2$ and off-diagonal elements of zero to the covariance matrix of $\mathcal{G}$.
\item Distribution equality: While $F_k$ is not necessarily a dependency of $F_l$ in the layer graph, the transformation that maps $F_k$ to $F_l$ is still the same as $f_l(f_k)$. Both $\beta_l$ and $F_l^\text{in}$ are also independent of previous layers and have the same distribution.
\item Parameter-control: It is easy to check that if $\rho_k$ is PC by $\mathcal{C}^d$ for some $d > 0$ or $\mathcal{C}^\text{E2}$ at $\tilde{\mathfrak{m}}$ with $\rho_k^\text{mean}$ and $\rho_k^\text{elem}$, then $\rho_l$ is also PC by the same class at $\tilde{\mathfrak{m}}$ with $\rho_l^\text{mean}=\rho_k^\text{mean}$ and $\rho_l^\text{elem}$ equal to $\rho_k^\text{elem}$ except that $\rho_l^\text{elem}$ has an additional unused elementwise input $e_l^\text{in}$.
\item Limits: We have $\tilde{\mathfrak{m}}_l=\mathbb{E}_e(e_l^\text{in} + \rho_k(e,\tilde{\mathfrak{m}}))=\tilde{\mathfrak{m}}_k$ and $\mathfrak{c}_{l;l}= \mathbb{E}_e(e_l^\text{in} + \rho_k(e,\tilde{\mathfrak{m}}))^2 = \mathbb{E}_e(e_l^\text{in})^2 +\mathbb{E}_e\rho_k(e,\tilde{\mathfrak{m}})^2 + 2\mathbb{E}_ee_l^\text{in}\rho_k(e,\tilde{\mathfrak{m}})=\mathfrak{c}_{k;k} + \sigma_l^2$ and hence $\lime\tilde{\mathfrak{m}}_l=\lime\tilde{\mathfrak{m}}_k=\mathfrak{m}_k= \mathfrak{m}_l$ and $\lime\mathfrak{c}_{l;l}=\mathfrak{q}_k + \sigma_l^2=\mathfrak{q}_l$. Here we use that $F_l^\text{in}$ is independent of previous layers of $F$.
\item Rank stability: Since $\mathfrak{c}_{k;k} > 0$ and $\sigma_l^2 \ge 0$, we have $\mathfrak{c}_{l;l} > 0$. $F_l^\text{in}$ will never be the dependency of an FC layer, so it does not matter for the purpose of RS.
\end{itemize}

Case: $f_l$ is an LN layer.

\begin{itemize}
\item Layer addition: We add an elementwise layer $F_l^\text{sq}$ and another elementwise layer $F_l$, both with the same width as $f_l$. We set $\rho_l^\text{sq} = (\rho_k - m_k)^2$. Its elementwise inputs are those of $\rho_k$. Its mean inputs are those of $\rho_k$ as well as $m_k$. We set $\rho_l=(\rho_k- m_k)(m_l^\text{sq} + \epsilon_l)^{-\frac{1}{2}}$. Its elementwise inputs are those of $\rho_k$. Its mean inputs are those of $\rho_k$ as well as $m_k$ and $m_l^\text{sq}$.

We require $m_l^\text{sq} + \epsilon_l$ to be positive. Since $m_l^\text{sq}$ is assigned the value $\overline{F_l^\text{sq}}$, it is assigned a mean of squares, which is non-negative. Since $\epsilon_l > 0$ by property \ref{aprop7}, $m_l^\text{sq} + \epsilon_l$ is indeed positive.
\item Generator addition: NA
\item Distribution equality: We can transform $F_k$ to $F_l$ by applying $f_l(f_k)$.
\item Parameter-control: $\rho_l$ is defined in terms of $\rho_k$. The same is true for bias layers (see above) and addition layers (see below). Hence, we can ``recursively expand'' $\rho_l$ until it is defined explicitly in terms of its elementwise and mean inputs, as well as elements of $\vec{\epsilon}$. This recursion ``flows backward'' through the layer graph of $f$. Note that the recursion cannot hit an activation layer because of property \ref{aprop9}.

Based on the explicit definition, it is easy to check that $\rho_l$ is elem-poly(1) as defined in section \ref{otherNTCsection}. The $(m[\hat{\kappa}]+c)^{-c'}$ expressions in the definition of the elem-poly property correspond to $(m_{l'}^\text{sq} + \epsilon_{l'})^{-\frac{1}{2}}$ expressions obtained via LN layers $f_{l'}$ with $l' \le l$. Because $\overline{F_{l'}^\text{sq}}$ is the mean of squares, its limit $\tilde{\mathfrak{m}}_{l'}^\text{sq}$ is non-negative. $\epsilon_{l'} > 0$ by property \ref{apropX1}. Hence, all conditions of lemma \ref{lemma13} are met with $m^*=\tilde{\mathfrak{m}}$, which yields that $\rho_l$ is PC by $\mathcal{C}^1$ at $\tilde{\mathfrak{m}}$ and hence PC by $\mathcal{C}^\text{E2}$ at $\tilde{\mathfrak{m}}$. Analogously, $\rho_l^\text{sq}$ is elem-poly(2) and hence PC by $\mathcal{C}^2$ at $\tilde{\mathfrak{m}}$ and hence PC by $\mathcal{C}^\text{E2}$ at $\tilde{\mathfrak{m}}$.

\item Limits: First, we have $\tilde{\mathfrak{m}}_l=\mathbb{E}_e(\rho_k - \tilde{\mathfrak{m}}_k)(\tilde{\mathfrak{m}}_l^\text{sq}+\epsilon_l)^{-\frac{1}{2}} = (\tilde{\mathfrak{m}}_k - \tilde{\mathfrak{m}}_k)(\tilde{\mathfrak{m}}_l^\text{sq}+\epsilon_l)^{-\frac{1}{2}} = 0$. Now consider $\tilde{\mathfrak{m}}_k$. Assume $\tilde{\mathfrak{m}}_k \neq 0$. Then $f_k$ can't be an LN, input or FC layer, as this would yield $\tilde{\mathfrak{m}}_k=0$. By property \ref{aprop9}, it can't be an activation layer. So it is a bias or addition layer. But by the calculation rules for $\tilde{\mathfrak{m}}_k$ for layers derived from a bias (see above) or addition layer (see below), we find that $f_k$ must have a dependency with a non-zero $\tilde{\mathfrak{m}}$ value itself. That dependency can only be a bias or addition layer. Rinse and repeat, we obtain a contradiction as there are only finitely many layers. Hence, $\tilde{\mathfrak{m}}_k = 0$. 

Further, $\tilde{\mathfrak{m}}_l^\text{sq} = \mathbb{E}_e(\rho_k(e,\tilde{\mathfrak{m}})-\tilde{\mathfrak{m}}_k)^2=\mathfrak{c}_{k;k}$ and $\mathfrak{c}_{l;l}=\mathbb{E}_e(\rho_k(e,\tilde{\mathfrak{m}}) - \tilde{\mathfrak{m}}_k)^2(\tilde{\mathfrak{m}}_l^\text{sq}+\epsilon_l)^{-1} =\mathbb{E}_e\rho_k(e,\tilde{\mathfrak{m}})^2(\tilde{\mathfrak{m}}_l^\text{sq}+\epsilon_l)^{-1} = \frac{\mathfrak{c}_{k;k}}{\mathfrak{c}_{k;k}+\epsilon_l}$. So we have $\lime\tilde{\mathfrak{m}}_l = 0 = \mathfrak{m}_l$ and $\lime\mathfrak{c}_{l;l}=\lime\frac{\mathfrak{c}_{k;k}}{\mathfrak{c}_{k;k}+\epsilon_l}=\frac{\lime\mathfrak{c}_{k;k}}{\lime\mathfrak{c}_{k;k}+\lime\epsilon_l} = \frac{\mathfrak{q}_k}{\mathfrak{q}_k} = 1 = \mathfrak{q}_l$. This uses that $\mathfrak{q}_k > 0$, which is obtained from proposition \ref{mfntPositive}.
\item Rank stability: Since $\mathfrak{c}_{k;k} > 0$ and $\epsilon_l > 0$, we have $\mathfrak{c}_{l;l} > 0$.
\end{itemize}

Case: $f_l$ is an addition layer.

\begin{itemize}
\item Layer addition: We add an elementwise layer $F_l$ with the width of $f_l$. We set $\rho_l = \sum_{\kappa=1}^{K}w_{\kappa}\rho_{k[\kappa]}$. Its elementwise and mean inputs are those occurring in any of the $\rho_{k[\kappa]}$.
\item Generator addition: NA
\item Distribution equality: We can transform $F_{k[1]}, ..,F_{k[K]}$ to $F_l$ by applying $f_l(f_{k[1]}, .., f_{k[K]})$.
\item Parameter-control: Assume that each $\rho_{k[\kappa]}$ is PC by some class $\mathcal{C}$ that is linearly closed. Then by lemma \ref{lemma14} $\rho_l$ is also PC by $\mathcal{C}$. Specifically, since the $\rho_{k[\kappa]}$ are all PC by $\mathcal{C}^\text{E2}$ at $\tilde{\mathfrak{m}}$ by the induction hypothesis, $\rho_l$ is PC by $\mathcal{C}^\text{E2}$ at $\tilde{\mathfrak{m}}$.
\item Limits: \sloppy $\tilde{\mathfrak{m}}_l = \mathbb{E}_e \sum_{\kappa=1}^{K}w_\kappa\rho_{k[\kappa]}(e,\tilde{\mathfrak{m}}) = \sum_{\kappa=1}^Kw_{\kappa}\tilde{\mathfrak{m}}_{k[\kappa]}$, so $\lime\tilde{\mathfrak{m}}_l=\sum_{\kappa=1}^{K}w_{\kappa}\mathfrak{m}_{k[\kappa]}=\mathfrak{m}_l$. Further, $\mathfrak{c}_{l;l} = \mathbb{E}_e(\sum_{\kappa=1}^Kw_{\kappa}\rho_{k[\kappa]}(e,\tilde{\mathfrak{m}}))^2 = \sum_{\kappa=1}^Kw_{\kappa}^2\mathfrak{c}_{k[\kappa];k[\kappa]} + \sum_{\kappa\neq\kappa'}w_{\kappa}w_{\kappa'}\mathbb{E}_e\rho_{k[\kappa]}(e,\tilde{\mathfrak{m}})\rho_{k[\kappa']}(e,\tilde{\mathfrak{m}})$. Consider some specific $\kappa\neq\kappa'$. Just like in the case where $f_l$ is an LN layer above, we can ``recursively expand'' both $\rho_{k[\kappa]}$ and $\rho_{k[\kappa']}$ by applying the definitions of $\rho$ for LN, addition and bias layers as well as activation layers (see below). Due to property \ref{aprop10}, there cannot be a layer in the layer graph of $f$ that is touched by both the expansion of $\rho_{k[\kappa]}$ and $\rho_{k[\kappa']}$. Hence, the dependencies of $F_{k[\kappa]}$ and $F_{k[\kappa']}$ are distinct. Since there is no weight sharing in $f$ and the off-diagonal elements of the covariance matrix of $\mathcal{G}$ corresponding to different layers in $f$ are zero by construction, $\mathfrak{c}_{l';l''}=0$ when $F_{l'}$ and $F_{l''}$ are distinct FC or input layers of equal width. ($F_{l'}$ or $F_{l''}$ may also refer to $F_{l'}^\text{in}$ or $F_{l''}^\text{in}$ here.) Hence, we have $\mathbb{E}_e\rho_{k[\kappa]}(e,\tilde{\mathfrak{m}})\rho_{k[\kappa']}(e,\tilde{\mathfrak{m}}) = \mathbb{E}_e\rho_{k[\kappa]}(e,\tilde{\mathfrak{m}})\mathbb{E}_e\rho_{k[\kappa']}(e,\tilde{\mathfrak{m}}) = \tilde{\mathfrak{m}}_{k[\kappa]} \tilde{\mathfrak{m}}_{k[\kappa']}$. Further, by property \ref{aprop8}, the expansion of at least one of $\rho_{k[\kappa]}$ and $\rho_{k[\kappa']}$ does not touch an activation layer. WLOG let it be $\rho_{k[\kappa]}$. Then we obtain $\tilde{\mathfrak{m}}_{k[\kappa]} = 0$ just as we obtained $\tilde{\mathfrak{m}}_k = 0$ for the case where $f_l$ is an LN layer above. So $\mathbb{E}_e\rho_{k[\kappa]}(e,\tilde{\mathfrak{m}})\rho_{k[\kappa']}(e,\tilde{\mathfrak{m}})=0$ for any $\kappa \neq \kappa'$, so $\mathfrak{c}_{l;l} = \sum_{\kappa=1}^Kw_{\kappa}^2\mathfrak{c}_{k[\kappa];k[\kappa]}$, so $\lime \mathfrak{c}_{l;l}= \sum_{\kappa=1}^Kw_{\kappa}^2\mathfrak{q}_{k[\kappa]}=\mathfrak{q}_{l}$.
\item Rank stability: Since all the $w_{\kappa}$ are non-zero by property \ref{aprop3} and all the $\mathfrak{c}_{k[\kappa];k[\kappa]}$ are positive, so is $\mathfrak{c}_{l;l}$.
\end{itemize}

Case: $f_l$ is an activation layer.

\begin{itemize}
\item Layer addition: We add an elementwise layer $F_l$ with the width of $f_l$ and set $\rho_l = \tau_l(\rho_k)$. Its elementwise and mean inputs are those of $\rho_k$.
\item Generator addition: NA
\item Distribution equality: We can transform $F_k$ to $F_l$ by applying $f_l(f_k)$.
\item Parameter-control: By property \ref{aprop9}, the recursive expansion of $\rho_l$ does not hit an activation layer. So $\rho_k$ is elem-poly(1) and, as above, fulfills the other conditions of lemma \ref{lemma13} when $m^*=\tilde{\mathfrak{m}}$. So $\rho_k$ is PC by $\mathcal{C}^1$ at $\tilde{\mathfrak{m}}$. Because $\tau_l$ is controlled by $\mathcal{C}^\text{E2}$ by property \ref{aprop2}, we have 

\begin{eqnarray*}
&&|\tau(\rho_k(e,\tilde{\mathfrak{m}}))|\\
&\le&e^{c|\rho_k(e,\tilde{\mathfrak{m}})|^{2-c'} + c''}\\
&\le&e^{c(c'''||e||_2 + c'''')^{2-c'} + c''}\\
\end{eqnarray*}

for some $c,c',c'',c''',c''''$, all positive, where also $c' \le 2$. The latter expression is itself controlled by $\mathcal{C}^\text{E2}$, so $\rho_l(e,\tilde{\mathfrak{m}})$ is controlled by $\mathcal{C}^\text{E2}$. If $\rho_l$ has no mean inputs, we directly obtain that it is PC by $\mathcal{C}^\text{E2}$ at $\tilde{\mathfrak{m}}$. If it has mean inputs, we apply lemma \ref{lemma12} with $m^*=\tilde{\mathfrak{m}}$. Let's check the remaining conditions. Clearly, $\mathcal{C}^\text{E2}$ is linearly closed. Let $M^*$ be the $M(m^*)$ that arises in the statement of lemma \ref{lemma13} applied to $\rho_k$. Then the differentiability of $\rho_l$ follows from the differentiability of $\tau_l$. Finally, for an arbitrary component $m[\hat{\kappa}]$, we have 

\begin{eqnarray*}
&&\max_{m \in M^*}\Big|\frac{d\rho_l(e,m)}{dm[\hat{\kappa}]}\Big|\\
&=&\max_{m \in M^*}|\tau_l'(\rho_k(e,m))|\Big|\frac{d\rho_k(e,m)}{dm[\hat{\kappa}]}\Big|\\
&\le&\max_{m \in M^*}|\tau_l'(\rho_k(e,m))|\max_{m \in M^*}\Big|\frac{d\rho_k(e,m)}{dm[\hat{\kappa}]}\Big|\\
&\le&\max_{m \in M^*}e^{c|\rho_k(e,m)|^{2-c'}+c''}\max_{m \in M^*}\Big|\frac{d\rho_k(e,m)}{dm[\hat{\kappa}]}\Big|\\
&\le&e^{c|c'''''||e||_2+c''''''|^{2-c'}+c''}(c'''||e||_2+c'''')
\end{eqnarray*}

for some $c,c',c'',c''',c'''',c''''',c''''''$, all positive, where also $c' \le 2$. Here, we use lemma \ref{lemma13} on both $\rho_k$ and $\frac{d\rho_k(e,m)}{dm[\hat{\kappa}]}$. This works because $\frac{d\rho_k(e,m)}{dm[\hat{\kappa}]}$ is itself elem-poly(1) and $M(m^*)$ is the same for $\frac{d\rho_k(e,m)}{dm[\hat{\kappa}]}$ and $\rho_k$. The last expression is controlled by $\mathcal{C}^\text{E2}$, which is the last condition of lemma \ref{lemma12}.
\item Limits: Since $\rho_k$ is elem-poly(1), the distribution of $\rho_k(e,\tilde{\mathfrak{m}})$ induced by $e$ is a linear combination of Gaussians and constants, and hence itself Gaussian. By property \ref{aprop9}, the recursive expansion of $\rho_k$ does not hit an activation layer, so again $\tilde{\mathfrak{m}}_k=0$. So we have $\tilde{\mathfrak{m}}_l = \mathbb{E}_e \tau_l(\rho_k(e,\tilde{\mathfrak{m}})) = \mathbb{E}_{s\sim\mathcal{N}(0,\mathfrak{c}_{k;k})}\tau_l(s)$ and $\mathfrak{c}_{l;l}=\mathbb{E}_e \tau_l(\rho_k(e,\tilde{\mathfrak{m}}))^2=\mathbb{E}_{s\sim\mathcal{N}(0,\mathfrak{c}_{k;k})}\tau_l(s)^2$. We know that $\lime\mathfrak{c}_{k;k}=\mathfrak{q}_k$. By proposition \ref{mfntPositive}, we have $\mathfrak{q}_k > 0$. So we can apply lemma \ref{lemma15} to $\tau_l$ and $\tau_l^2$ and obtain $\lime\tilde{\mathfrak{m}}=\mathbb{E}_{s\sim\mathcal{N}(0,\mathfrak{q}_k)}\tau(s)=\mathfrak{m}_l$, which matches table \ref{tableNLCPropagation}, as well as

\begin{eqnarray*}
&&\lime\mathfrak{c}_{l;l}\\
&=&\mathbb{E}_{s\sim\mathcal{N}(0,\mathfrak{q}_k)}\tau_l(s)^2\\
&=&\mathbb{E}_{\begin{pmatrix}s\\t\end{pmatrix}\sim\mathcal{N}\bigg(0,\begin{pmatrix}\mathfrak{q}_k&\mathfrak{q}_k\\\mathfrak{q}_k&\mathfrak{q}_k\end{pmatrix}\bigg)}\tau_l(s)\tau_l(t)\\
&=&\mathfrak{C}_{\tau_l}(\mathfrak{q}_k, \mathfrak{q}_k)\\
&=&\mathfrak{q}_l
\end{eqnarray*}

as required.
\item Rank stability: Analogously to above, we have $\mathfrak{c}_{l;l} = \mathfrak{C}_{\tau_l}(\mathfrak{c}_{k;k}, \mathfrak{c}_{k;k})$. Because $\tau$ is continuous and not the zero function by property \ref{aprop2} and $\mathfrak{c}_{k;k}>0$, by proposition \ref{covkerPositive3}, we have $\mathfrak{c}_{l;l} > 0$ as required.
\end{itemize}

This completes stage 1.

\underline{Stage 2}: statement (\ref{eqn5p2})

For this stage, denote $x^{(1)}$ simply by $x'$ and $x^{(2)}$ simply by $x''$. This stage, like many of the following stages, proceeds analogously to stage 1. As before, statement (\ref{eqn5p2}) trivially holds for $l=0$. As before, we can assume $\overline{x'.x'}=q$, $\overline{x''.x''}=q$ and $\overline{x'.x''}=c$. As before, to handle the case $l > 0$, we build $F$ and $\mathcal{G}$ by induction and apply background theorem \ref{backgroundMaster} at each step. In fact, our $F$ in this stage is simply the 2-duplex of the $F$ from the previous stage. (See the definition and discussion in section \ref{covarianceKernelSection}.) In plain words, $F$ consists of two sub-architectures $F'$ and $F''$ that do not have interdependencies. Both $F'$ and $F''$ are constructed as $F$ was constructed in the previous stage. We use the same notation for layers in $F'$ and $F''$ as in stage 1, where $'$ and $''$ act as fixed modifiers.

$\mathcal{G}$ is a Gaussian with zero mean and the following covariance matrix. A diagonal entry corresponding to a readin layer $f_l$ is $q\sigma_l^2$ and a diagonal entry corresponding to a bias layer $f_l$ is $\sigma_l^2$. An off-diagonal entry is $c\sigma_l^2$ when it corresponds to a pair of layers in $F$ that stem from the same readin layer $f_l$, i.e. $F'_l$ and $F''_l$. An off-diagonal entry is $\sigma_l^2$ when it corresponds to a pair of input layers in $F$ that stem from the same bias layer $f_l$, i.e. $F_l^{\prime\text{in}}$ and $F_l^{\prime\prime\text{in}}$. Other off-diagonal entries are zero.

\sloppy If $f_l$ is a readin layer, the joint distribution of $F'_l$ and $F''_l$ is elementwise with generator $\mathcal{N}\Big(0,\sigma_l^2\begin{pmatrix}q&c\\c&q\end{pmatrix}\Big)$. This is also the joint distribution of $f_l(x')$ and $f_l(x'')$ as induced by $\theta_l$. Because $F'$ and $F''$ have no interdependencies, but share their weights and each $F_l^{\prime\text{in}}$ has the same value as the corresponding $F_l^{\prime\prime\text{in}}$, we have that $F'_1, .., F'_L, F''_1, .., F''_L$ are jointly distributed as $f_1(x'), .., f_L(x'), f_1(x''), .., f_L(x'')$. So as in the previous stage, we have distribution equality. Also, if PC holds for $F$ in stage 1, it holds for $F$ in stage 2, as the multi-activation functions are identical ``as functions''. We maintain the notation used in the previous stage. Specifically, we continue to use $\tilde{\mathfrak{m}}_l$ and $\mathfrak{c}_{l;l}$ for the corresponding values from stage 1. This yields $\tilde{\mathfrak{m}}'_l = \tilde{\mathfrak{m}}''_l = \tilde{\mathfrak{m}}_l$ and $\mathfrak{c}^{\prime;\prime}_{l;l}=\mathfrak{c}^{\prime\prime;\prime\prime}_{l;l} = \mathfrak{c}_{l;l}$

What remains is to investigate the limits and show RS for $F$. We will do the latter by again showing that all $A^\infty_S$ are full rank. Because there is weight sharing between $F'$ and $F''$, for each FC layer $f_l$, a valid $S$ may consist of either $F'_k$, $F''_k$ or both. Hence, we need to show that the matrix $\begin{pmatrix}\mathfrak{c}^{\prime;\prime}_{k;k}&\mathfrak{c}^{\prime;\prime\prime}_{k;k}\\\mathfrak{c}^{\prime;\prime\prime}_{k;k}&\mathfrak{c}^{\prime\prime;\prime\prime}_{k;k}\end{pmatrix}$ is full rank, which in turn is implied by $\mathfrak{c}_{k;k} > \mathfrak{c}^{\prime;\prime\prime}_{k;k}  \ge 0$. We proceed by induction.

Case: $f_l$ is a readin layer.

\begin{itemize}
\item Limits: From table \ref{tableBackgroundPropagation}, we obtain $\mathfrak{c}^{\prime;\prime\prime}_{l;l}=c\sigma_l^2$ and so $\lim_{\vec{\epsilon},N}\mathfrak{c}^{\prime;\prime\prime}_{l;l}=c\sigma_l^2$. Since $f_l$ is FC and its dependency is the input layer, this matches $\mathfrak{c}_l$ from table \ref{tableNLCPropagation}.
\item Rank stability: Since $q > c \ge 0$ and $\sigma_l^2 > 0$, we have $\mathfrak{c}_{l;l} - \mathfrak{c}^{\prime;\prime\prime}_{l;l} = (q-c)\sigma_l^2 > 0$ and $\mathfrak{c}^{\prime;\prime\prime}_{l;l} = c\sigma_l^2 \ge 0$.
\end{itemize}

Case: $f_l$ is a fully-connected, non-readin layer.

\begin{itemize}
\item Limits: $\mathfrak{c}^{\prime;\prime\prime}_{l;l}=\mathfrak{c}^{\prime;\prime\prime}_{k;k}\sigma_l^2$, so $\lime\mathfrak{c}^{\prime;\prime\prime}_{l;l}=\mathfrak{c}_k\sigma_l^2=\mathfrak{c}_l$.
\item Rank stability: Since $\mathfrak{c}_{l;l} = \sigma_l^2\mathfrak{c}_{k;k}$, $\mathfrak{c}_{k;k} > \mathfrak{c}^{\prime;\prime\prime}_{k;k} \ge 0$ and $\sigma_l^2 > 0$, we have $\mathfrak{c}_{l;l} - \mathfrak{c}^{\prime;\prime\prime}_{l;l} = (\mathfrak{c}_{k;k} - \mathfrak{c}^{\prime;\prime\prime}_{k;k})\sigma_k^2 > 0$ and $\mathfrak{c}^{\prime;\prime\prime}_{l;l} =\mathfrak{c}^{\prime;\prime\prime}_{k;k}\sigma_k^2 \ge 0$.
\end{itemize}

Case: $f_l$ is a bias layer.

\begin{itemize}
\item \sloppy Limits: We have $\mathfrak{c}^{\prime;\prime\prime}_{l;l} = \mathbb{E}_e(e_l^{\prime\text{in}} + \rho_k'(e,\tilde{\mathfrak{m}}))(e_l^{\prime\prime\text{in}} + \rho_k''(e,\tilde{\mathfrak{m}})) = \mathbb{E}_ee_l^{\prime\text{in}}e_l^{\prime\prime\text{in}} + \mathbb{E}_ee_l^{\prime\text{in}}\rho_k''(e,\tilde{\mathfrak{m}}) + \mathbb{E}_ee_l^{\prime\prime\text{in}}\rho_k'(e,\tilde{\mathfrak{m}}) + \mathbb{E}_e\rho_k'(e,\tilde{\mathfrak{m}})\rho_k''(e,\tilde{\mathfrak{m}}) = \mathfrak{c}^{\prime;\prime\prime}_{k;k} + \sigma_l^2$ and so $\lime \mathfrak{c}^{\prime;\prime\prime}_{l;l}= \mathfrak{c}_k + \sigma_l^2 = \mathfrak{c}_l$. Here we use that $e_l^{\prime\text{in}}$ and $e_l^{\prime\prime\text{in}}$ are independent of everything except each other.
\item Rank stability: Since $\mathfrak{c}_{l;l} = \mathfrak{c}_{k;k} + \sigma_l^2$ and $\mathfrak{c}_{k;k} > \mathfrak{c}^{\prime;\prime\prime}_{k;k} \ge 0$, we have $\mathfrak{c}_{l;l} - \mathfrak{c}^{\prime;\prime\prime}_{l;l} = \mathfrak{c}_{k;k} - \mathfrak{c}^{\prime;\prime\prime}_{k;k} > 0$ and $\mathfrak{c}^{\prime;\prime\prime}_{l;l} = \mathfrak{c}^{\prime;\prime\prime}_{k;k} + \sigma_l^2 \ge 0$. $F_l^{\prime\text{in}}$ and $F_l^{\prime\prime\text{in}}$ will never be the dependency of an FC layer, so they do not matter for the purpose of RS.
\end{itemize}

Case: $f_l$ is an LN layer.

\begin{itemize}
\item Limits: $\mathfrak{c}^{\prime;\prime\prime}_{l;l} = \mathbb{E}_e\big(\frac{\rho'_k(e,\mathfrak{m}) - \tilde{\mathfrak{m}}_k}{\sqrt{\mathfrak{c}_{k;k} + \epsilon_l}}\big)\big(\frac{\rho''_k(e,\mathfrak{m}) - \tilde{\mathfrak{m}}_k}{\sqrt{\mathfrak{c}_{k;k} + \epsilon_l}}\big) = \frac{\mathfrak{c}^{\prime;\prime\prime}_{k;k}}{\mathfrak{c}_{k;k} + \epsilon_l}$ and so $\lime\mathfrak{c}^{\prime;\prime\prime}_{l;l}=\lime \frac{\mathfrak{c}^{\prime;\prime\prime}_{k;k}}{\mathfrak{c}_{k;k} + \epsilon_l} = \frac{\lime\mathfrak{c}^{\prime;\prime\prime}_{k;k}}{\lime\mathfrak{c}_{k;k} + \lim_{\vec{\epsilon},N}\epsilon_l} = \frac{\mathfrak{c}_k}{\mathfrak{q}_k}=\mathfrak{c}_l$. Note that we have $\mathfrak{q}_k > 0$ from proposition \ref{mfntPositive}.
\item Rank stability: Since $\mathfrak{c}_{l;l} = \frac{\mathfrak{c}_{k;k}}{\mathfrak{c}_{k;k} + \epsilon_l}$, and $\mathfrak{c}_{k;k} > \mathfrak{c}^{\prime;\prime\prime}_{k;k} \ge 0$, we have $\mathfrak{c}_{l;l} - \mathfrak{c}^{\prime;\prime\prime}_{l;l} = \frac{\mathfrak{c}_{k;k} - \mathfrak{c}^{\prime;\prime\prime}_{k;k}}{\mathfrak{c}_{k;k}+\epsilon_l} > 0$ and $\mathfrak{c}^{\prime;\prime\prime}_{l;l} = \frac{\mathfrak{c}^{\prime;\prime\prime}_{k;k}}{\mathfrak{c}_{k;k}+\epsilon_l} \ge 0$.
\end{itemize}

Case: $f_l$ is an addition layer.

\begin{itemize}
\item \sloppy Limits: We have $\mathfrak{c}^{\prime;\prime\prime}_{l;l} = \mathbb{E}_e(\sum_{\kappa=1}^Kw_{\kappa}\rho'_{k[\kappa]}(e,\tilde{\mathfrak{m}}))(\sum_{\kappa=1}^Kw_{\kappa}\rho''_{k[\kappa]}(e,\tilde{\mathfrak{m}})) = \sum_{\kappa=1}^Kw_{\kappa}^2\mathfrak{c}_{k[\kappa];k[\kappa]}^{\prime;\prime\prime} + \sum_{\kappa\neq\kappa'}w_{\kappa}w_{\kappa'}\mathbb{E}_e\rho'_{k[\kappa]}(e,\tilde{\mathfrak{m}})\rho''_{k[\kappa']}(e,\tilde{\mathfrak{m}})$. Analogously to stage 1, $\rho'_{k[\kappa]}$ and $\rho''_{k[\kappa]}$ have elementwise inputs that originate from different layers in $f$ and only one of the two recursive expansions can hit an activation layer, so the latter sum is again zero. Hence, $\mathfrak{c}^{\prime;\prime\prime}_{l;l}=\sum_{\kappa=1}^Kw_{\kappa}^2\mathfrak{c}_{k[\kappa];k[\kappa]}^{\prime;\prime\prime}$ and so $\lime\mathfrak{c}^{\prime;\prime\prime}_{l;l}=\sum_{\kappa=1}^Kw_{\kappa}^2\mathfrak{c}_{k[\kappa]}=\mathfrak{c}_l$.
\item Rank stability: Since $\mathfrak{c}_{l;l}=\sum_{\kappa=1}^Kw_{\kappa}^2\mathfrak{c}_{k[\kappa];k[\kappa]}$, all the $w_\kappa$ are non-zero and $\mathfrak{c}_{k[\kappa];k[\kappa]} > \mathfrak{c}_{k[\kappa];k[\kappa]}^{\prime;\prime\prime} \ge 0$ for all $\kappa$, we have $\mathfrak{c}_{l;l} - \mathfrak{c}_{l;l}^{\prime;\prime\prime} = \sum_{\kappa=1}^{K}w_{\kappa}^2(\mathfrak{c}_{k[\kappa];k[\kappa]} - \mathfrak{c}_{k[\kappa];k[\kappa]}^{\prime;\prime\prime}) > 0$ and $\mathfrak{c}_{l;l}^{\prime;\prime\prime} = \sum_{\kappa=1}^{K}w_{\kappa}^2\mathfrak{c}_{k[\kappa];k[\kappa]}^{\prime;\prime\prime} \ge 0$.
\end{itemize}

Case: $f_l$ is an activation layer.

\begin{itemize}
\item Limits: As in stage 1, $\rho'_k(e,\tilde{\mathfrak{m}})$ and $\rho''_k(e,\tilde{\mathfrak{m}})$ are each a linear combination of Gaussians and constants, and hence are themselves Gaussian. Further, all the underlying Gaussians are jointly Gaussian, so $\rho'_k(e,\tilde{\mathfrak{m}})$ and $\rho''_k(e,\tilde{\mathfrak{m}})$ are jointly Gaussian. From stage 1, we have that their mean $\tilde{\mathfrak{m}}_k$ is zero. They have variance $\mathfrak{c}_{k;k}$ and covariance $\mathfrak{c}^{\prime;\prime\prime}_{k;k}$. So 

\begin{eqnarray*}
&&\lime\mathfrak{c}_l^{\prime;\prime\prime}\\
&=&\lime\mathbb{E}_e \tau_l(\rho'_k(e,\tilde{\mathfrak{m}}))\tau_l(\rho''_k(e,\tilde{\mathfrak{m}}))\\
&=&\lime\mathbb{E}_{\begin{pmatrix}s\\t\end{pmatrix}\sim\mathcal{N}\bigg(0,\begin{pmatrix}\mathfrak{c}_{k;k}&\mathfrak{c}^{\prime;\prime\prime}_{k;k}\\\mathfrak{c}_{k;k}^{\prime;\prime\prime}&\mathfrak{c}_{k;k}\end{pmatrix}\bigg)}\tau_l(s)\tau_l(t)\\
&=&\mathbb{E}_{\begin{pmatrix}s\\t\end{pmatrix}\sim\mathcal{N}\bigg(0,\begin{pmatrix}\mathfrak{q}_{k}&\mathfrak{c}_{k}\\\mathfrak{c}_{k}&\mathfrak{q}_{k}\end{pmatrix}\bigg)}\tau_l(s)\tau_l(t)\\
&=&\mathfrak{C}_{\tau_l}(\mathfrak{q}_k, \mathfrak{c}_k)\\
&=&\mathfrak{c}_l
\end{eqnarray*}
Again, we use lemma \ref{lemma15} and $\mathfrak{q}_k > \mathfrak{c}_k$ from proposition \ref{mfntPositive}.
\item Rank stability: Because $\tau_l$ is continuous and non-constant and $\mathfrak{c}_{k;k} > \mathfrak{c}^{\prime;\prime\prime}_{k;k} \ge 0$, by proposition \ref{covkerPositive}, we have $\mathfrak{c}_{l;l} - \mathfrak{c}^{\prime;\prime\prime}_{l;l}= \mathfrak{C}_{\tau_l}( \mathfrak{c}_{k;k},  \mathfrak{c}_{k;k}) - \mathfrak{C}_{\tau_l}( \mathfrak{c}_{k;k}, \mathfrak{c}^{\prime;\prime\prime}_{k;k}) > 0$ and by proposition \ref{covkerPositive2}, we have $\mathfrak{c}^{\prime;\prime\prime}_{l;l} = \mathfrak{C}_{\tau_l}( \mathfrak{c}_{k;k}, \mathfrak{c}^{\prime;\prime\prime}_{k;k}) \ge 0$ as required.
\end{itemize}

This completes stage 2.

\underline{Stage 3}: statements (\ref{eqn5p5}), (\ref{eqn5p6}) and (\ref{eqn5p7})

We have 

\begin{eqnarray*}
&&\lime\limd\frac{1}{d_l^\text{MF}}\mathbb{E}_x||f_l||^2_2\\
&=&\lim_{\vec{\epsilon},N}\limd\frac{1}{N}\sum_{n=1}^N\overline{f_l(x^{(n)})^{.2}}\\
&=&\lim_{\vec{\epsilon},N}\frac{1}{N}\sum_{n=1}^N \limd\overline{f_l(x^{(n)})^{.2}}\\
&=&\lim_{\vec{\epsilon},N}\frac{1}{N}\sum_{n=1}^N\begin{cases}\mathfrak{c}_{l;l} \text{ if } l>0\\q \text{ else}\end{cases}\\
&=&\lim_{\vec{\epsilon},N}\begin{cases}\mathfrak{c}_{l;l} \text{ if } l>0\\q \text{ else}\end{cases}\\
&=&\mathfrak{q}_l
\end{eqnarray*}

as required. We have

\begin{eqnarray*}
&&\lim_{\vec{\epsilon},N}\limd\frac{1}{d_l^\text{MF}}||\mathbb{E}_xf_l||^2_2\\
&=&\lim_{\vec{\epsilon},N}\limd\frac{1}{d_l^\text{MF}}||\frac{1}{N}\sum_{n=1}^Nf_l(x^{(n)})||^2_2\\
&=&\lim_{\vec{\epsilon},N}\limd\frac{1}{N^2}\sum_{n,n'=1}^N\overline{f_l(x^{(n)}).f_l(x^{(n')})}\\
&=&\lim_{\vec{\epsilon},N}\frac{1}{N^2}\sum_{n,n'=1}^N\limd\overline{f_l(x^{(n)}).f_l(x^{(n')})}\\
&=&\lim_{\vec{\epsilon},N}\begin{cases}\frac{1}{N^2}(N\mathfrak{c}_{l;l} + N(N-1)\mathfrak{c}^{\prime;\prime\prime}_{l;l}) \text{ if } l>0\\\frac{1}{N^2}(Nq + N(N-1)c) \text{ else}\end{cases}\\
&=&\mathfrak{c}_l
\end{eqnarray*}

as required. We have

\begin{eqnarray*}
&&\lim_{\vec{\epsilon},N}\limd\frac{1}{d_l^\text{MF}}||\mathbb{S}_xf_l||^2_2\\
&=&\lim_{\vec{\epsilon},N}\limd\frac{1}{d_l^\text{MF}}\mathbb{E}_x||f_l||^2_2 - \frac{1}{d_l^\text{MF}}||\mathbb{E}_xf_l||^2_2\\
&=&\lim_{\vec{\epsilon},N}\limd\frac{1}{d_l^\text{MF}}\mathbb{E}_x||f_l||^2_2 - \lim_{\vec{\epsilon},N}\limd\frac{1}{d_l^\text{MF}}||\mathbb{E}_xf_l||^2_2\\
&=&\mathfrak{q}_l - \mathfrak{c}_l
\end{eqnarray*}

as required.

\underline{Stage 4}: statement (\ref{eqn5p9})

We begin with the case $l=m$. We have $\frac{1}{d_l^\text{MF}}\mathbb{E}_x\Tr(\mathcal{J}_{l,m}\Cov_{f_m}\mathcal{J}_{l,m}^T)=\frac{1}{d_m^\text{MF}}\mathbb{E}_x\Tr(\Cov_{f_m})=\frac{1}{d_m^\text{MF}}||\mathbb{S}_xf_m||^2_2$. So by stage 3, we have $\lime \limd \frac{1}{d_l^\text{MF}}\mathbb{E}_x\Tr(\mathcal{J}_{l,m}\Cov_{f_m}\mathcal{J}_{l,m}^T)=\mathfrak{q}_m - \mathfrak{c}_m = \frac{\mathfrak{g}_l(\mathfrak{q}_m - \mathfrak{c}_m)}{\mathfrak{g}_m}$ as required. We use $\mathfrak{g}_m > 0$, which we have by proposition \ref{mfntPositive}.

For the remainder of this stage, assume $l>m$. Without loss of generality, we can also assume that $f_m$ is a bottleneck for $f_L$ in addition to being a bottleneck for $f_l$. Otherwise, we can simply apply the argument from this stage to the sub-architecture composed of $f_l$ and its ancestors, which is a valid A-architecture, to obtain statement (\ref{eqn5p9}) for the pair $(l,m)$.

We have 

\begin{eqnarray*}
&&\frac{1}{d_l^\text{MF}}\mathbb{E}_x\Tr(\mathcal{J}_{l,m}\Cov_{f_m}\mathcal{J}_{l,m}^T)\\
&=&\frac{1}{d_l^\text{MF}}\frac{1}{N^2}\sum_{n^-,n^+=1}^N\Tr\Big(\mathcal{J}_{l,m}(x^{(n^+)})(f_m(x^{(n^-)})-\frac{1}{N}\sum_{n=1}^Nf_m(x^{(n)}))^T...\\
&&...(f_m(x^{(n^-)})-\frac{1}{N}\sum_{n=1}^Nf_m(x^{(n)}))\mathcal{J}_{l,m}(x^{(n^+)})^T\Big)\\
&=&\frac{1}{d_l^\text{MF}}\frac{1}{N^2}\sum_{n^-,n^+=1}^N||(f_m(x^{(n^-)})-\frac{1}{N}\sum_{n=1}^Nf_m(x^{(n)}))\mathcal{J}_{l,m}(x^{(n^+)})^T||_2^2\\
&=&\frac{1}{N^2}\sum_{n^-,n^+=1}^N\overline{\Big((f_m(x^{(n^-)})-\frac{1}{N}\sum_{n=1}^Nf_m(x^{(n)}))\mathcal{J}_{l,m}(x^{(n^+)})^T\Big)^{.2}}\\
\end{eqnarray*}

Denote $(f_m(x^{(n^-)})-\frac{1}{N}\sum_{n=1}^Nf_m(x^{(n)}))\mathcal{J}_{l,m}(x^{(n^+)})^T$ by $p_l^{(n^-,n^+)}$. Then we have

$$\lime \limd \frac{1}{d_l^\text{MF}}\mathbb{E}_x\Tr(\mathcal{J}_{l,m}\Cov_{f_m}\mathcal{J}_{l,m}^T) = \lime \frac{1}{N^2}\sum_{n^-,n^+=1}^N\limd \overline{p_l^{(n^-,n^+)}.p_l^{(n^-,n^+)}}$$

Again, we want to build $F$ and $\mathcal{G}$ by induction and use background theorem \ref{backgroundMaster} to obtain $\limd \overline{p_l^{(n^-,n^+)}.p_l^{(n^-,n^+)}}$ for any $n^-$, $n^+$. Specifically, we conduct one induction per value of $m$. Now fix $m$ for the rest of the stage. The induction proceeds slightly differently for the cases $m=0$ and $m>0$. We will keep track of those differences. For now, fix $n^-$ and $n^+$.

For now, let's look at $m>0$. We begin by building a mean field architecture as in stage 1 by running the induction from stage 1 up to layer $f_m$. Then we create an $N$-duplex of the resultant mean field architecture like we created the 2-duplex in stage 2. Denote this $N$-duplex by $F = (F^{(1)},F^{(2)},..,F^{(N)})$. Throughout this stage, we use $\lambda$ as a layer index that ranges from 1 to $m$. The $\mathcal{G}$ for this $F$ is as in stage 2. It is a mean zero Gaussian with a covariance matrix where diagonal elements corresponding to some $F^{(n)\text{in}}_\lambda$ with $f_\lambda$ a bias layer are $\sigma_\lambda^2$ and diagonal elements corresponding to some $F^{(n)}_\lambda$ with $f_\lambda$ a readin layer are $q\sigma_\lambda^2$. Off-diagonal elements are $\sigma_\lambda^2$ if they correspond to a pair of layers $F^{(n)\text{in}}_\lambda$ and $F^{(n')\text{in}}_\lambda$ with $n\neq n'$, are $c\sigma_\lambda^2$ if they correspond to a pair of layers $F^{(n)}_\lambda$ and $F^{(n')}_\lambda$ with $n\neq n'$ where $f_\lambda$ is a readin layer, and are zero otherwise.

As in stage 2, the joint distribution of all layers $F_\lambda^{(n)}$ as induced by the parameter of $F$ and $\mathcal{G}$ is equal to the joint distribution of the $f_\lambda(x^{(n)})$ as induced by the parameter of $f$. To apply background theorem \ref{backgroundMaster} to the $N$-duplex, like in stage 2, all we need is to demonstrate rank stability and investigate limits. The induction proceeds analogously to stage 2 due to symmetry. On the limit side, we obtain $\limd \overline{F^{(n)}_\lambda} = \tilde{\mathfrak{m}}_\lambda$, $\limd \overline{F^{(n)}_\lambda.F^{(n)}_\lambda} = \mathfrak{c}_{\lambda;\lambda}$ and $\limd \overline{F^{(n)}_\lambda.F^{(n')}_\lambda} = \mathfrak{c}_{\lambda;\lambda}^{\prime;\prime\prime}$ when $n \neq n'$. For rank stability, we notice that each weight matrix is shared by $N$ FC layers in the $N$-duplex. Hence, any valid set $S$ corresponds to a subset of a set of layers $F_\lambda^{(1)},F_\lambda^{(2)},..,F_\lambda^{(N)}$. Any resulting $A_S^\infty$ has diagonal elements $\mathfrak{c}_{\lambda;\lambda}$ and off-diagonal elements $\mathfrak{c}_{\lambda;\lambda}^{\prime;\prime\prime}$. Since we showed $\mathfrak{c}_{\lambda;\lambda} > \mathfrak{c}_{\lambda;\lambda}^{\prime;\prime\prime} \ge 0$ in stage 2, all limits $A_S^\infty$ are full rank as required.

Since we now have layers in $F$ that represent the $f_m(x^{(n)})$, the next step is to add a layer that represents $f_m(x^{(n^-)})-\frac{1}{N}\sum_{n=1}^Nf_m(x^{(n)})$. To do this, we make the following extension.

\begin{itemize}
\item Layer addition: We add an elementwise layer $F^\text{MN}$ (from ``mean normalization'') with the width of $f_m$ and set $\rho^\text{MN} = \rho_m^{(n^-)} - \frac{1}{N}\sum_{n=1}^N\rho^{(n)}_m$. Its elementwise and mean inputs are those occurring in any of the $\rho^{(n)}_m$.
\item Generator addition: NA
\item Distribution equality: We can transform $F_m^{(1)},F_m^{(2)},..,F_m^{(N)}$ to $F^\text{MN}$ by applying $F_m^{(n^-)} - \frac{1}{N}\sum_{n=1}^NF_m^{(n)}$, as required.
\item Parameter-control: The argument is as for the addition layer in stage 1.
\item Limits: $\tilde{\mathfrak{m}}^\text{MN}=\mathbb{E}_e\Big(\rho_m^{(n^-)}(e,\mathfrak{m}) - \frac{1}{N}\sum_{n=1}^N\rho^{(n)}_m(e,\mathfrak{m}))\Big)=\tilde{\mathfrak{m}}_m - \frac{1}{N}\sum_{n=1}^N\tilde{\mathfrak{m}}_m = 0$ and 

\begin{eqnarray*}
&&\mathfrak{c}^{\text{MN};\text{MN}}\\
&=&\mathbb{E}_e\Big(\rho_m^{(n^-)}(e,\tilde{\mathfrak{m}}) - \frac{1}{N}\sum_{n=1}^N\rho^{(n)}_m(e,\tilde{\mathfrak{m}}))\Big)^2\\
&=&\mathbb{E}_e\rho_m^{(n^-)}(e,\mathfrak{m})^2 - \frac{2}{N}\mathbb{E}_e\rho_m^{(n^-)}(e,\mathfrak{m})^2 - \frac{2}{N}\sum_{n\neq n^-}\mathbb{E}_e\rho_m^{(n^-)}(e,\mathfrak{m})\rho_m^{(n)}(e,\mathfrak{m})\\
&&+\frac{1}{N^2}\sum_n\mathbb{E}_e\rho_m^{(n)}(e,\mathfrak{m})^2 + \frac{1}{N^2}\sum_{n\neq n'}\mathbb{E}_e\rho_m^{(n)}(e,\mathfrak{m})\rho_m^{(n')}(e,\mathfrak{m})\\
&=&\mathfrak{c}_{m;m} - \frac{2}{N}\mathfrak{c}_{m;m} - \frac{2(N-1)}{N}\mathfrak{c}_{m;m}^{\prime;\prime\prime} + \frac{1}{N}\mathfrak{c}_{m;m} + \frac{N(N-1)}{N^2}\mathfrak{c}_{m;m}^{\prime;\prime\prime}\\
&=&(1-\frac{1}{N})(\mathfrak{c}_{m;m} - \mathfrak{c}_{m;m}^{\prime;\prime\prime})
\end{eqnarray*}

and similarly $\mathfrak{c}^{\text{MF};(n)}_{;m} = \limd \overline{F^\text{MN}.F^{(n)}_m} = (\mathbbm{1}_{n=n^-}-\frac{1}{N})(\mathfrak{c}_{m;m} - \mathfrak{c}_{m;m}^{\prime;\prime\prime})$. Here, $\mathbbm{1}_{n=n^-}$ denotes the indicator function of $n$ being equal to $n^-$.
\item Rank stability: In the current state of $F$, $F^\text{MN}$ is not the dependency of an FC layer.
\end{itemize}

After adding $F^\text{MN}$, we also extend $F^{(n^+)}$ beyond layer $F^{(n^+)}_m$ as we did in stage 1. This corresponds to forward-propagating $x^{(n^+)}$ beyond $f_m$. We denote all of $F^{(n^+)}$ also simply by $F^+$. Parameter-control and rank stability follows as in stage 1.

Now let's turn to the case $m=0$. In that scenario, we initialize our mean field architecture $F$ to be $F^{(n^+)}$, usually shortened to $F^+$, as obtained by applying stage 1 to $f$ and $x^{(n^+)}$. 

Going forward, we treat the general case where $m$ can be zero or non-zero. For each $l$, we want to add a layer to $F$ that represents $p_l^{(n^-,n^+)}$. We denote this layer by $F_l^-$. If $f_l$ has a single dependency (e.g. is not an addition layer), we have 

$$p_l^{(n^-,n^+)}=p_k^{(n^-,n^+)}\mathcal{J}_{l,k}(x^{(n^+)})^T$$

Hence, to obtain distribution equality, we ensure that $F_l^-$ can be obtained by right-multiplying $F_k^-$ by the transpose of the matrix obtained by evaluating $\mathcal{J}_{l,k}(f_k)$ at $F_k^+$. In terms of rank stability, the novel sets $S$ that will arise will contain either just $F_l^-$ or both $F_l^-$ and $F_l^+$. Denote $\begin{pmatrix}\mathfrak{c}_{l;l}^{-;-}&\mathfrak{c}_{l;l}^{-;+}\\\mathfrak{c}_{l;l}^{-;+}&\mathfrak{c}_{l;l}^{+;+}\end
{pmatrix}$ by $\Sigma_l$. Again, we will show that $A_S^\infty$ is full rank, which corresponds to $\det \Sigma_l = \mathfrak{c}_{l;l}^{-;-}\mathfrak{c}_{l;l}^{+;+}- \mathfrak{c}_{l;l}^{-;+}\mathfrak{c}_{l;l}^{-;+} > 0$, which also implies $\mathfrak{c}_{l;l}^{-;-} > 0$. Note that $\mathfrak{c}_{l;l}^{+;+}$ is just $\mathfrak{c}_{l;l}$ from stage 1 and $\tilde{\mathfrak{m}}_l^+$ is just $\tilde{\mathfrak{m}}_l$. To start off our induction when $m>0$, we must show full-rankness when $S$ contains either just $F^\text{MN}$ or both of $F^\text{MN}$ and $F^+_m$. This corresponds to $\det \begin{pmatrix}\mathfrak{c}^{\text{MN};\text{MN}}&\mathfrak{c}_{;m}^{\text{MN};+}\\\mathfrak{c}_{;m}^{\text{MN};+}&\mathfrak{c}_{m;m}\end{pmatrix} > 0$. Because we have $\mathfrak{c}_{m;m} > \mathfrak{c}_{m;m}^{\prime;\prime\prime} \ge 0$ and $N \ge 2$ we have

\begin{eqnarray*}
&&\det  \begin{pmatrix}\mathfrak{c}^{\text{MN};\text{MN}}&\mathfrak{c}_{;m}^{\text{MN};+}\\\mathfrak{c}_{;m}^{\text{MN};+}&\mathfrak{c}_{m;m}\end{pmatrix}\\
&=&(1-\frac{1}{N})(\mathfrak{c}_{m;m} - \mathfrak{c}_{m;m}^{\prime;\prime\prime})\mathfrak{c}_{m;m} - (\mathbbm{1}_{n^-=n^+}-\frac{1}{N})^2(\mathfrak{c}_{m;m} - \mathfrak{c}_{m;m}^{\prime;\prime\prime})^2\\
&\ge&(1-\frac{1}{N})(\mathfrak{c}_{m;m} - \mathfrak{c}_{m;m}^{\prime;\prime\prime})^2 - (\mathbbm{1}_{n^-=n^+}-\frac{1}{N})^2(\mathfrak{c}_{m;m} - \mathfrak{c}_{m;m}^{\prime;\prime\prime})^2\\
&=&(\mathfrak{c}_{m;m} - \mathfrak{c}_{m;m}^{\prime;\prime\prime})^2(1-\mathbbm{1}_{n^-=n^+}(1-\frac{2}{N})+\frac{1}{N}-\frac{1}{N^2})\\
&>&0
\end{eqnarray*}

Now we proceed again by layer type. The induction step has the same components as in stage 1, except we do not investigate the limit with respect to $N$ and $\vece$ for now.

Case: $f_l$ is a readin layer.

\begin{itemize}
\item Layer addition: We add a single input layer $F_l^-$ to $F$ with the same width as $f_l$.
\item Generator addition: We add the diagonal entry $(1-\frac{1}{N})(q-c)\sigma_l^2$ to the covariance matrix of $\mathcal{G}$. We add off-diagonal entries of zero, except entries corresponding to $F_l^-$ and $F_l^+$ are $(\mathbbm{1}_{n^-=n^+}-\frac{1}{N})(q-c)\sigma_l^2$.
\item Distribution equality: Since $f_m$ is a bottleneck for $f_L$, we can only have $f_l$ with $l>m$ be a readin layer if $m=0$. So we need the distribution of $F_l^-$ to be that of $p_l^{(n^-,n^+)}=(x^{(n^-)} - \frac{1}{N}\sum_nx^{(n)})\mathcal{J}_{l,0}(x^{(n^+)})=(x^{(n^-)} - \frac{1}{N}\sum_nx^{(n)})W_l$. Because $W_l$ has IID Gaussian elements, $p_l^{(n^-,n^+)}$ is elementwise Gaussian with mean zero and variance

$$\sigma_l^2\overline{(x^{(n^-)} - \frac{1}{N}\sum_nx^{(n)}).(x^{(n^-)} - \frac{1}{N}\sum_nx^{(n)})}=(1-\frac{1}{N})(q-c)\sigma_l^2$$

This matches our choice for the diagonal entry of the generator. In terms of the joint distribution of $(F^-_1, .., F^-_l, F^+_1, .., F^+_L)$, we note that $F^-_l$ is independent of all layers with index less than $l$ in both $F^-$ and $F^+$. This is as required as the randomness of $p_l^{(n^-,n^+)}$ is induced only by $W_l$, which is independent of layers $f_1$ through $f_{l-1}$. The joint distribution of $p_l^{(n^-,n^+)}$ and $f_l(x^{(n^+)})$ is that of $(x^{(n^-)} - \frac{1}{N}\sum_nx^{(n)})W_l$ and $x^{(n^+)}W_l$. Again, this is elementwise Gaussian with covariance 

$$\sigma_l^2\overline{(x^{(n^-)} - \frac{1}{N}\sum_nx^{(n)}).x^{(n^+)}}=(\mathbbm{1}_{n^-=n^+}-\frac{1}{N})(q-c)\sigma_l^2$$

This matches our choice for the off-diagonal entry of the generator corresponding to $F_l^-$ and $F_l^+$. Finally, layers $F^+_{l+1}$ through $F^+_L$ are either input layers corresponding to readin layers in $f$, which are correctly independent of $F^-_l$, or transformations of previous layers. Hence, we do obtain the correct joint distribution for $(F^-_1, .., F^-_l, F^+_1, .., F^+_L)$.
\item Parameter-control: NA
\item Limits: From table \ref{tableBackgroundPropagation}, we obtain $\tilde{\mathfrak{m}}_l^-=0$, $\mathfrak{c}_{l;l}^{-;-}=(1-\frac{1}{N})(q-c)\sigma_l^2$ and $\mathfrak{c}_{l;l}^{-;+}=(\mathbbm{1}_{n^-=n^+}-\frac{1}{N})(q-c)\sigma_l^2$.
\item Rank stability: Since $q > c$, $\mathfrak{c}_{l;l}=q\sigma_l^2$ and $N \ge 2$, we have 
\begin{eqnarray*}
&&\det \Sigma_l\\
&=&(1-\frac{1}{N})(q-c)\sigma_l^2q\sigma_l^2 - ((\mathbbm{1}_{n^-=n^+}-\frac{1}{N})(q-c)\sigma_l^2)^2\\
&\ge&(1-\frac{1}{N})(q-c)^2\sigma_l^4 - (\mathbbm{1}_{n^-=n^+}-\frac{1}{N})^2(q-c)^2\sigma_l^4\\
&=&(q-c)^2\sigma_l^4(1-\mathbbm{1}_{n^-=n^+}(1-\frac{2}{N})+\frac{1}{N}-\frac{1}{N^2})\\
&>&0
\end{eqnarray*}

as required.
\end{itemize}

Case: $f_l$ is a fully-connected, non-readin layer.

\begin{itemize}
\item Layer addition: We add a single fully-connected layer $F^-_l$ with width and $\sigma_l$ value equal to that of $f_l$ and dependency $F^-_k$. In a slight abuse of notation, if $f_m$ is a dependency of $f_l$, we use $F^-_k$ to refer to $F^\text{MN}$. Note that because $f_m$ is a bottleneck, the layer index of a dependency of $f_l$ is at least $m$. $F^-_l$ shares weight matrix with $F^+_l$.
\item Generator addition: NA
\item Distribution equality: We have $\mathcal{J}_{l,k}^T=W_l$, so
multiplying by the transpose of the Jacobian is the same as multiplying by the weight matrix. $W_l$ has the same distribution in both $f$ and $F$. Also $F^-_l$ and $F^+_l$ use the same $W_l$, as are $f_l$ and $\mathcal{J}_{l,k}$. $W_l$ is not used by any other layer in $F$, as it is not used by another layer or layer Jacobian in $f$.
\item Parameter-control: NA
\item Limits: From table \ref{tableBackgroundPropagation}, we obtain $\tilde{\mathfrak{m}}_l^-=0$, $\mathfrak{c}_{l;l}^{-;-}=\sigma_l^2\mathfrak{c}_{k;k}^{-;-}$ and $\mathfrak{c}_{l;l}^{-;+}=\sigma_l^2\mathfrak{c}_{k;k}^{-;+}$.
\item Rank stability: We have $\mathfrak{c}_{l;l} = \mathfrak{c}_{k;k}\sigma_l^2$ and $\sigma_l^2 > 0$, so $\det\Sigma_l = \sigma_l^2\det\Sigma_k > 0$.
\end{itemize}

Case: $f_l$ is a bias layer.

\begin{itemize}
\item Layer addition: We add an elementwise layer $F_l^-$ with the width of $f_l$ and $\rho_l^-=\rho_k^-$. Its elementwise and mean inputs are those of $\rho_k^-$.
\item Generator addition: NA
\item Distribution equality: The Jacobian of a bias layer is the identity, and hence we require that $F^-_l$ and $F^-_k$ have the same value.
\item Parameter-control: Since $\rho_k$ is PC by $\mathcal{C}^\text{E2}$ at $\tilde{\mathfrak{m}}$, so is $\rho_l$.
\item Limits: $\tilde{\mathfrak{m}}_l^-=\tilde{\mathfrak{m}}_k^-$, $\mathfrak{c}_{l;l}^{-;-}=\mathfrak{c}_{k;k}^{-;-}$ and $\mathfrak{c}_{l;l}^{-;+} = \mathbb{E}_e(\rho_k^-(\rho_k^+ + e_l^{+\text{in}}))= \mathbb{E}_e\rho_k^-\rho_k^+ + \mathbb{E}_e\rho_k^- e_l^{+\text{in}}=\mathfrak{c}_{k;k}^{-;+} + 0 = \mathfrak{c}_{k;k}^{-;+}$.
\item Rank stability: Since $\mathfrak{c}_{l;l}=\mathfrak{c}_{k;k} + \sigma_l^2$, we have $\det\Sigma_l = \mathfrak{c}_{k;k}^{-;-}(\mathfrak{c}_{k;k} + \sigma_l^2) - (\mathfrak{c}_{k;k}^{-;+})^2 \ge \det\Sigma_k > 0$.
\end{itemize}

Case: $f_l$ is an LN layer.

\begin{itemize}
\item Layer addition: We add an elementwise layer $F_l^{\text{dot}}$ and another elementwise layer $F_l^-$, both with the same width as $f_l$. We set $\rho_l^{\text{dot}} = \rho_k^-\rho_k^+$ and $$\rho_l^-=(m_l^{+\text{sq}} + \epsilon_l)^{-\frac{1}{2}}(\rho_k^--m_k^-) - (m_l^{+\text{sq}} + \epsilon_l)^{-\frac{3}{2}}(m^\text{dot}_l - m^-_lm^+_l)(\rho^+_k-m^+_l)$$ Their elementwise inputs are those occurring in $\rho_k^-$ or $\rho_k^+$. Their mean inputs are those occurring in $\rho_k^-$ or $\rho_k^+$, and $\rho_l^-$ also has mean inputs $m_k^-$, $m_k^+$, $m_l^{+\text{sq}}$ and $m_l^\text{dot}$. As in stage 1, $m_l^{+\text{sq}} + \epsilon_l$ is positive.
\item Generator addition: NA
\item Distribution equality: Substituting the formula for $F_l^\text{dot}$ into that of $F_l$, it is easy to check that what we obtain is the result of multiplying $F^-_k$ by the Jacobian of the LN operation applied to $F^+_k$.
\item Parameter-control: Just as in stage 1, it is easy to check that $\rho_l^-$ is elem-poly(1) and $\rho_l^\text{dot}$ is elem-poly(2) by recursively expanding the definition of $\rho_l^-$. The expansion itself is a bit more complex than in stage 1. As it flows backward through the layer graph of $f$, each layer $f_{l'}$ in $f$ with $l' \le l$ may be represented in $F$ by either $F^-_{l'}$ and $F^+_{l'}$ or by $F_{l'}^{(1)}, .., F_{l'}^{(N)}$. Hence, we must apply the recursive definitions corresponding to those layers. If the expansion hits $f_m$, we must also apply the recursive definition of $\rho^\text{MN}$. However, all our arguments from earlier stages based on recursive expansion still go through in the same way. We just have to consider a somewhat wider range of possible recursion steps.

As before, we can apply lemma \ref{lemma13} to obtain that $\rho_l^-$ is PC by $\mathcal{C}^1$ at $\tilde{\mathfrak{m}}$ and $\rho_l^\text{dot}$ is PC by $\mathcal{C}^2$ at $\tilde{\mathfrak{m}}$, and hence both are PC by $\mathcal{C}^\text{E2}$ at $\tilde{\mathfrak{m}}$.
\item Limits: $\tilde{\mathfrak{m}}^\text{dot}_l=\mathbb{E}_e\rho_k^-\rho_k^+=\mathfrak{c}^{-;+}_{k;k}$ and $\tilde{\mathfrak{m}}^-_l=...(\mathbb{E}_e\rho_k^- -\tilde{\mathfrak{m}}^-_k) - ...(\mathbb{E}_e\rho_k^+ -\tilde{\mathfrak{m}}_k)=0$ As in stage 1, we find that $\tilde{\mathfrak{m}}^-_k$ can only be non-zero if there exists a directed path in $f$ from an activation layer to $f_l$ that does not include an FC layer. This is excluded by property \ref{aprop9}. $\mathfrak{c}_{k;k} > 0$ from stage 1. Hence

\begin{eqnarray*}
&&\mathfrak{c}_{l;l}^{-;-}\\
&=&\mathbb{E}_e\Big((\mathfrak{c}_{k;k} + \epsilon_l)^{-\frac{1}{2}}(\rho_k^--\tilde{\mathfrak{m}}^-_k) - (\mathfrak{c}_{k;k} + \epsilon_l)^{-\frac{3}{2}}(\mathfrak{c}_{k;k}^{-;+} - \tilde{\mathfrak{m}}^-_k\tilde{\mathfrak{m}}_k)(\rho^+_k-\tilde{\mathfrak{m}}_k)\Big)^2\\
&=&(\mathfrak{c}_{k;k} + \epsilon_l)^{-1}\mathbb{E}_e\rho_k^-\rho_k^- -2\mathfrak{c}_{k;k}^{-;+}(\mathfrak{c}_{k;k} + \epsilon_l)^{-2}\mathbb{E}_e\rho_k^-\rho_k^+ \\
&&+ \mathfrak{c}_{k;k}^{-;+}\mathfrak{c}_{k;k}^{-;+}(\mathfrak{c}_{k;k} + \epsilon_l)^{-3}\mathbb{E}_e\rho_k^+\rho_k^+\\
&=&\frac{\mathfrak{c}_{k;k}^{-;-}}{\mathfrak{c}_{k;k} + \epsilon_l} -2\frac{\mathfrak{c}_{k;k}^{-;+}\mathfrak{c}_{k;k}^{-;+}}{(\mathfrak{c}_{k;k} + \epsilon_l)^2} + \frac{\mathfrak{c}_{k;k}^{-;+}\mathfrak{c}_{k;k}^{-;+}\mathfrak{c}_{k;k}}{(\mathfrak{c}_{k;k} + \epsilon_l)^3}
\end{eqnarray*}

and 

\begin{eqnarray*}
&&\mathfrak{c}_{l;l}^{-;+}\\
&=&\mathbb{E}_e\Big((\mathfrak{c}_{k;k} + \epsilon_l)^{-\frac{1}{2}}(\rho_k^--\tilde{\mathfrak{m}}^-_k)\\
&&- (\mathfrak{c}_{k;k} + \epsilon_l)^{-\frac{3}{2}}(\mathfrak{c}_{k;k}^{-;+} - \tilde{\mathfrak{m}}^-_k\tilde{\mathfrak{m}}_k)(\rho^+_k-\tilde{\mathfrak{m}}_k)\Big)(\mathfrak{c}_{k;k} + \epsilon_l)^{-\frac{1}{2}}(\rho_k^+- \tilde{\mathfrak{m}}_k)\\
&=&(\mathfrak{c}_{k;k} + \epsilon_l)^{-1}\mathbb{E}_e\rho_k^-\rho_k^+ -\mathfrak{c}_{k;k}^{-;+}(\mathfrak{c}_{k;k} + \epsilon_l)^{-2}\mathbb{E}_e\rho_k^+\rho_k^+\\
&=&\frac{\mathfrak{c}_{k;k}^{-;+}\epsilon_l}{(\mathfrak{c}_{k;k} + \epsilon_l)^2}
\end{eqnarray*}
\item Rank stability: Since $\mathfrak{c}_{l;l} = \frac{\mathfrak{c}_{k;k}}{\mathfrak{c}_{k;k} + \epsilon_l}$, we have
\begin{eqnarray*}
&&\det\Sigma_l\\
&=&\Big(\frac{\mathfrak{c}_{k;k}^{-;-}}{\mathfrak{c}_{k;k} + \epsilon_l} -2\frac{\mathfrak{c}_{k;k}^{-;+}\mathfrak{c}_{k;k}^{-;+}}{(\mathfrak{c}_{k;k} + \epsilon_l)^2} + \frac{\mathfrak{c}_{k;k}^{-;+}\mathfrak{c}_{k;k}^{-;+}\mathfrak{c}_{k;k}}{(\mathfrak{c}_{k;k} + \epsilon_l)^3}\Big)\frac{\mathfrak{c}_{k;k}}{\mathfrak{c}_{k;k} + \epsilon_l} - \Big(\frac{\mathfrak{c}_{k;k}^{-;+}\epsilon_l}{(\mathfrak{c}_{k;k} + \epsilon_l)^2}\Big)^2\\
&=&\frac{\det\Sigma_k}{(\mathfrak{c}_{k;k} + \epsilon_l)^2}\\
&>& 0
\end{eqnarray*}

as required.

\end{itemize}

Case: $f_l$ is an addition layer.

\begin{itemize}
\item Layer addition: We add an elementwise layer $F_l^-$ with the width of $f_l$. We set $\rho_l^- = \sum_{\kappa=1}^{K}w_{\kappa}\rho_{k[\kappa]}^-$. Its elementwise and mean inputs are those occurring in any of the $\rho_{k[\kappa]}$.
\item Generator addition: NA
\item Distribution equality: We have 
\begin{eqnarray*}
&&p_l^{(n^-,n^+)}\\
&=&(f_m(x^{(n^-)})-\frac{1}{N}\sum_{n=1}^Nf_m(x^{(n)}))\mathcal{J}_{l,m}(x^{(n^+)})^T\\
&=&(f_m(x^{(n^-)})-\frac{1}{N}\sum_{n=1}^Nf_m(x^{(n)}))\sum_\kappa\mathcal{J}_{k[\kappa],m}(x^{(n^+)})^T\mathcal{J}_{l,k[\kappa]}(x^{(n^+)})^T\\
&=&\sum_\kappa w_\kappa(f_m(x^{(n^-)})-\frac{1}{N}\sum_{n=1}^Nf_m(x^{(n)}))\mathcal{J}_{k[\kappa],m}(x^{(n^+)})^T\\
&=&\sum_\kappa w_\kappa p_{k[\kappa]}^{(n^-,n^+)}
\end{eqnarray*}

This corresponds to our definition of $\rho_l$, as required.
\item Parameter-control: As in stage 1.
\item \sloppy Limits: $\tilde{\mathfrak{m}}_l^- = \mathbb{E}_e \sum_{\kappa=1}^{K}w_{\kappa}\rho_{k[\kappa]}^-(e,\tilde{\mathfrak{m}}) = \sum_{\kappa=1}^Kw_{\kappa}\tilde{\mathfrak{m}}_{k[\kappa]}^-$. We also have $\mathfrak{c}_{l;l}^{-;-}=\sum_{\kappa=1}^Kw_{\kappa}^2\mathfrak{c}_{k[\kappa];k[\kappa]}^{-;-}$ and $\mathfrak{c}_{l;l}^{-;+}=\sum_{\kappa=1}^Kw_{\kappa}^2\mathfrak{c}_{k[\kappa];k[\kappa]}^{-;+}$. We use the same recursion argument as in stages 1 and 2. As noted above under the LN case, the recursion is a bit more complex than in previous stages, but ultimately analogous.
\item Rank stability: Since $\mathfrak{c}_{l;l}=\sum_{\kappa}w_{\kappa}^2\mathfrak{c}_{k[\kappa];k[\kappa]}$, we have $\det\Sigma_l=\det\sum_\kappa w_\kappa^2 \Sigma_{k[\kappa]}$. The diagonal elements of the $\Sigma_{k[\kappa]}$ are limits of means of squares, so they are non-negative. The $\det \Sigma_{k[\kappa]}$ have full rank. So we can apply lemma \ref{lemma16} $K-1$ times to obtain that the determinant of the weighted sum is positive.
\end{itemize}

Case: $f_l$ is an activation layer.

\begin{itemize}
\item Layer addition: We add an elementwise layer $F_l$ with the width of $f_l$ and set $\rho_l^- = \rho_k^-\tau'_l(\rho^+_k)$. Its elementwise and mean inputs are those occurring in either $\rho_k^-$ or $\rho^+_k$.
\item Generator addition: NA
\item Distribution equality: We have

\begin{eqnarray*}
&&p_l^{(n^-,n^+)}\\
&=&p_k^{(n^-,n^+)}\mathcal{J}_{l,k}(x^{(n^+)})\\
&=&p_k^{(n^-,n^+)}.\tau_l'.(f_k(x^{(n^+)}))
\end{eqnarray*}

This corresponds to our choice of $\rho_l^-$.
\item Parameter-control: As in stage 1, $\rho_k^-$ and $\rho_k^+$ are elem-poly(1) and fulfill the other conditions of lemma \ref{lemma13} with $m^*=\tilde{\mathfrak{m}}$. So $\rho_k^-$ and $\rho_k^+$ are PC by $\mathcal{C}^1$ at $\tilde{\mathfrak{m}}$. As in stage 1, because $\tau'_l$ is controlled by $\mathcal{C}^\text{E2}$, $\tau'_l(\rho_k^+(e,\tilde{\mathfrak{m}}))$ is controlled by $\mathcal{C}^\text{E2}$, and hence also $\rho_l^-(e,\tilde{\mathfrak{m}})=\rho_k^-(e,\tilde{\mathfrak{m}})\tau'_l(\rho_k^+(e,\tilde{\mathfrak{m}}))$. If $\rho_l$ has no mean inputs, we directly obtain that it is PC by $\mathcal{C}^\text{E2}$ at $\tilde{\mathfrak{m}}$. If $\rho_l$ has mean inputs, we apply lemma \ref{lemma12} with $m^*=\tilde{\mathfrak{m}}$. Let's check the remaining conditions. Clearly, $\mathcal{C}^\text{E2}$ is linearly closed. Let $M$ be the intersection of the $M(m^*)$ arising from the statement of lemma \ref{lemma13} applied to $\rho^+$ and the $M(m^*)$ arising from the statement of lemma \ref{lemma13} applied to $\rho^-$. That intersection is open as both hypercubes are open. It is non-empty because both contain $m^*=\tilde{\mathfrak{m}}$. In $M$, the differentiability of $\rho_l$ follows from the differentiability of $\tau'_l$. Finally, for an arbitrary component $m[\hat{\kappa}]$, we have

\begin{eqnarray*}
&&\max_{m \in M^*}\Big|\frac{d\rho_l^-(e,m)}{dm[\hat{\kappa}]}\Big|\\
&=&\max_{m \in M^*}\Big|\tau_l''(\rho_k^+(e,m))\frac{d\rho_k^+(e,m)}{dm[\hat{\kappa}]}\rho_k^-(e,m) + \tau_l'(\rho_k^+(e,m))\frac{d\rho_k^-(e,m)}{dm[\hat{\kappa}]}\Big|\\
&\le&\max_{m \in M^*}\Big|\tau_l''(\rho_k^+(e,m))\Big|\max_{m \in M^*}\Big|\frac{d\rho_k^+(e,m)}{dm[\hat{\kappa}]}\Big|\max_{m \in M^*}\Big|\rho_k^-(e,m)\Big| \\
&&+ \max_{m \in M^*}\Big|\tau_l'(\rho_k^+(e,m))\Big|\max_{m \in M^*}\Big|\frac{d\rho_k^-(e,m)}{dm[\hat{\kappa}]}\Big|\\
&\le&\max_{m \in M^*}e^{c|\rho_k^+(e,m)|^{2-c'}+c}\max_{m \in M^*}\Big|\frac{d\rho_k^+(e,m)}{dm[\hat{\kappa}]}\Big|\max_{m \in M^*}\Big|\rho_k^-(e,m)\Big| \\
&&+ \max_{m \in M^*}e^{c|\rho_k^+(e,m)|^{2-c'}+c}\max_{m \in M^*}\Big|\frac{d\rho_k^-(e,m)}{dm[\hat{\kappa}]}\Big|\\
&\le&e^{c|c||e||_2+c|^{2-c'}+c}(c||e||_2+c)(c||e||_2+c) + e^{c|c||e||_2+c|^{2-c'}+c}(c||e||_2+c)\\
\end{eqnarray*}

We slightly abuse notation here by having each occurrence of $c$ and $c'$ refer to a different positive constant. Each occurrence of $c'$ is also less than or equal to 2. Here, we use lemma \ref{lemma13} repeatedly. This works because each derivative with respect to $m[\hat{\kappa}]$ is itself elem-poly(1) and has the same $M(m^*)$ hypercube as the corresponding multi-activation function itself. The last expression is controlled by $\mathcal{C}^\text{E2}$, which is the last condition of lemma \ref{lemma12}.
\item Limits: Since $\rho_k^-$ and $\rho_k^+$ are elem-poly(1), $\rho_k^-(e,\tilde{\mathfrak{m}})$ and $\rho_k^+(e,\tilde{\mathfrak{m}})$ are linear combinations of Gaussians and constants, and hence are jointly Gaussian. The recursive expansions of $\rho_k^-$ and $\rho_k^+$ do not hit an activation layer, so again $\tilde{\mathfrak{m}}_k^-=\tilde{\mathfrak{m}}_k^+=0$. So we have $\tilde{\mathfrak{m}}_l^-=\mathbb{E}_{t,s\sim\mathcal{N}(0,\Sigma_k)}t\tau'_l(s)$, $\mathfrak{c}_{l;l}^{-;-}=\mathbb{E}_{t,s\sim\mathcal{N}(0,\Sigma_k)}t^2\tau'_l(s)^2$ and $\mathfrak{c}_{l;l}^{-;+}=\mathbb{E}_{t,s\sim\mathcal{N}(0,\Sigma_k)}t\tau'_l(s)\tau_l(s)$.
\item Rank stability: Let $t,s\sim\mathcal{N}(0,\Sigma_k)$, $u\sim\mathcal{N}(0, \mathfrak{c}_{k;k})$ and $v\sim\mathcal{N}(0, \frac{\det\Sigma_k}{\mathfrak{c}_{k;k}})$. $\mathfrak{c}_{k;k} > 0$ from stage 1. Since $\mathfrak{c}_{l;l}=\mathbb{E}_{t,s\sim\mathcal{N}(0,\Sigma_k)}\tau(s)^2$, we have 
\begin{eqnarray*}
&&\det \Sigma_l\\
&=&\mathbb{E}_{t,s}t^2\tau'_l(s)^2\mathbb{E}_{t,s}\tau(s)^2-(\mathbb{E}_{t,s}t\tau'_l(s)\tau_l(s))^2\\
&=&\mathbb{E}_{v,u}\Big(v+\frac{\mathfrak{c}_{k;k}^{-;+}}{\mathfrak{c}_{k;k}}u\Big)^2\tau'_l(u)^2\mathbb{E}_{v,u}\tau(u)^2-\Big(\mathbb{E}_{v,u}\Big(v+\frac{\mathfrak{c}_{k;k}^{-;+}}{\mathfrak{c}_{k;k}}u\Big)\tau'_l(u)\tau_l(u)\Big)^2\\
&=&\mathbb{E}_vv^2\mathbb{E}_uu^2\tau'_l(u)^2\mathbb{E}_u\tau(u)^2 + \Big(\frac{\mathfrak{c}_{k;k}^{-;+}}{\mathfrak{c}_{k;k}}\Big)^2\mathbb{E}_uu^2\tau'_l(u)^2\mathbb{E}_u\tau_l(u)^2\\
&& - \Big(\frac{\mathfrak{c}_{k;k}^{-;+}}{\mathfrak{c}_{k;k}}\Big)^2\Big(\mathbb{E}_uu\tau'_l(u)\tau_l(u)\Big)^2\\
&\ge&\mathbb{E}_vv^2\mathbb{E}_u\tau'_l(u)^2\mathbb{E}_u\tau_l(u)^2\\
&=&\frac{\det\Sigma_k}{\mathfrak{c}_{k;k}}\int_{\mathbb{R}}n(\mathfrak{c}_{k;k}, u)\tau'_l(u)^2d\mu_1\int_{\mathbb{R}}n(\mathfrak{c}_{k;k}, u)\tau_l(u)^2d\mu_1
\end{eqnarray*}

The penultimate step uses the Cauchy-Schwartz inequality. We have $\det\Sigma_k > 0$ and $\mathfrak{c}_{k;k} > 0$. Since $\tau_l$ is twice differentiable and non-constant, both $n(\mathfrak{c}_{k;k}, u)\tau'_l(u)^2$ and $n(\mathfrak{c}_{k;k}, u)\tau_l(u)^2$ are continuous and not the zero function. Hence, we can apply lemma \ref{lemma8} to both expressions to obtain that both integrals are positive, and hence that $\det \Sigma_l$ is positive.
\end{itemize}

Throughout the induction, we establish recursive calculation rules for the $\mathfrak{c}_{l;l}^{-;-}$ and $\mathfrak{c}_{l;l}^{-;+}$. The former represents $\limd\overline{p_l^{(n^-,n^+)}.p_l^{(n^-,n^+)}}$ as desired. If we recursively expand the rules for $\mathfrak{c}_{l;l}^{-;-}$ and $\mathfrak{c}_{l;l}^{-;+}$ as well as the rules for $\mathfrak{c}^{\text{MN};\text{MN}}$ and $\mathfrak{c}^{\text{MN};+}_{;m}$, we obtain expressions in terms of $q$ and $c$ (case $m=0$), $\mathfrak{c}_{m;m}^{\prime;\prime\prime}$ and $\mathfrak{c}_{m;m}$ (case $m>0$), the architecture definition, the $\mathfrak{c}_{l';l'}$ for $l' \le l$, $N$, and $\mathbbm{1}_{n^-=n^+}$. Importantly, these expressions depend only on  $n^+$ and $n^-$ through $\mathbbm{1}_{n^-=n^+}$. Hence, as we now let $n^+$ and $n^-$ vary again, $\mathfrak{c}_{l;l}^{-;-}$ and $\mathfrak{c}_{l;l}^{-;+}$ only obtain two different values corresponding to $n^-=n^+$ and $n^-\neq n^+$. Assuming that $\lime \mathfrak{c}_{l;l}^{-;-}$ is valid in both cases, we have

\begin{eqnarray*}
&&\lime \limd \frac{1}{d_l^\text{MF}}\mathbb{E}_x\Tr(\mathcal{J}_{l,m}\Cov_{f_m}\mathcal{J}_{l,m}^T)\\
&=&\lime  \frac{1}{N^2}\sum_{n^-,n^+=1}^N\limd\overline{p_l^{(n^-,n^+)}.p_l^{(n^-,n^+)}}\\
&=&\lime \frac{1}{N}\mathfrak{c}_{l;l}^{-;-}|_{n^-=n^+} + \frac{N-1}{N}\mathfrak{c}_{l;l}^{-;-}|_{n^-\neq n^+}\\
&=&\lime \mathfrak{c}_{l;l}^{-;-}|_{n^-\neq n^+}
\end{eqnarray*}

So to establish statement (\ref{eqn5p9}), we must show (i) $\lime \mathfrak{c}_{l;l}^{-;-}|_{n^-\neq n^+} = \frac{\mathfrak{g}_l(\mathfrak{q}_m-\mathfrak{c}_m)}{\mathfrak{g}_m}$ and (ii) $\lime \mathfrak{c}_{l;l}^{-;-}|_{n^-=n^+}$ is valid. We conduct one last induction to show this. As part of the induction, we also show $\lime \mathfrak{c}_{l;l}^{-;+}|_{n^-\neq n^+} = 0$. $\mathfrak{g}_m>0$ and $\mathfrak{q}_k>0$ by proposition \ref{mfntPositive} as usual. 

\begin{itemize}
\item Case: $F^\text{MF}$. $\lime\mathfrak{c}^{\text{MN};\text{MN}}=\lime(1-\frac{1}{N})(\mathfrak{c}_{m;m}-\mathfrak{c}_{m;m}^{\prime;\prime\prime})=\mathfrak{q}_m-\mathfrak{c}_m$. Further, $\lime\mathfrak{c}^{\text{MF};+}_{;m}=\lime(\mathbbm{1}_{n^-=n^+}-\frac{1}{N})(\mathfrak{c}_{m;m} - \mathfrak{c}_{m;m}^{\prime;\prime\prime})=\mathbbm{1}_{n^-=n^+}(\mathfrak{q}_m-\mathfrak{c}_m)$. This case starts off our induction when $m>0$, i.e. $F^\text{MF}$ corresponds to $p^{(n^-,n^+)}_m$ and $\mathfrak{c}^{\text{MN};\text{MN}}$ can be considered $\mathfrak{c}^{-;-}_{m;m}$. So we require $\lime\mathfrak{c}^{\text{MN};\text{MN}}=\frac{\mathfrak{g}_m(\mathfrak{q}_m-\mathfrak{c}_m)}{\mathfrak{g}_m}$, which is indeed the case. Further, $\mathfrak{c}^{\text{MF};+}_{;m}$ can be considered $\mathfrak{c}^{-;+}_{m;m}$, so when $n^-\neq n^+$ we must have $\lime \mathfrak{c}^{\text{MF};+}_{;m}=0$, which is indeed the case.
\item Case: $f_l$ is a readin layer. If $f_l$ is a readin layer, we have $m=0$. So $\frac{\mathfrak{g}_l(\mathfrak{q}_m-\mathfrak{c}_m)}{\mathfrak{g}_m}=\frac{\sigma_l^2(q-c)}{1}=\sigma_l^2(q-c)$. Also, $\lime\mathfrak{c}_{l;l}^{-;-}=\lime(1-\frac{1}{N})(q-c)\sigma_l^2=\sigma_l^2(q-c)$ as required. Also, $\lime\mathfrak{c}_{l;l}^{-;+}=\lime(\mathbbm{1}_{n^-=n^+}-\frac{1}{N})(q-c)\sigma_l^2=\mathbbm{1}_{n^-=n^+}(q-c)\sigma_l^2$. So when $n^-\neq n^+$, we have $\lime\mathfrak{c}_{l;l}^{-;+}=0$ as required.
\item Case: $f_l$ is a fully-connected, non-readin layer. $\lime\mathfrak{c}_{l;l}^{-;-}=\sigma_l^2\lime\mathfrak{c}_{k;k}^{-;-}=\sigma_l^2\frac{\mathfrak{g}_k(\mathfrak{q}_m-\mathfrak{c}_m)}{\mathfrak{g}_m}=\frac{\mathfrak{g}_l(\mathfrak{q}_m-\mathfrak{c}_m)}{\mathfrak{g}_m}$ as required. $\lime\mathfrak{c}_{l;l}^{-;+}=\sigma_l^2\lime\mathfrak{c}_{k;k}^{-;+}$. If $n^-\neq n^+$, $\lime\mathfrak{c}_{k;k}^{-;+}=0$, so $\lime\mathfrak{c}_{l;l}^{-;+}=0$ as required.
\item Case: $f_l$ is a bias layer. $\lime\mathfrak{c}_{l;l}^{-;-}=\lime\mathfrak{c}_{k;k}^{-;-}=\frac{\mathfrak{g}_k(\mathfrak{q}_m-\mathfrak{c}_m)}{\mathfrak{g}_m}=\frac{\mathfrak{g}_l(\mathfrak{q}_m-\mathfrak{c}_m)}{\mathfrak{g}_m}$ as required. $\lime\mathfrak{c}_{l;l}^{-;+}=\lime\mathfrak{c}_{k;k}^{-;+}$. If $n^-\neq n^+$, $\lime\mathfrak{c}_{k;k}^{-;+}=0$, so $\lime\mathfrak{c}_{l;l}^{-;+}=0$ as required.
\item Case: $f_l$ is an LN. 
\begin{eqnarray*}
&&\lime\mathfrak{c}_{l;l}^{-;-}\\
&=&\lime\frac{\mathfrak{c}_{k;k}^{-;-}}{\mathfrak{c}_{k;k} + \epsilon_l} -2\frac{\mathfrak{c}_{k;k}^{-;+}\mathfrak{c}_{k;k}^{-;+}}{(\mathfrak{c}_{k;k} + \epsilon_l)^2} + \frac{\mathfrak{c}_{k;k}^{-;+}\mathfrak{c}_{k;k}^{-;+}\mathfrak{c}_{k;k}}{(\mathfrak{c}_{k;k} + \epsilon_l)^3}\\
&=&\frac{\lime\mathfrak{c}_{k;k}^{-;-}}{\lime\mathfrak{c}_{k;k} + \lime\epsilon_l} -2\frac{\lime\mathfrak{c}_{k;k}^{-;+}\lime\mathfrak{c}_{k;k}^{-;+}}{(\lime\mathfrak{c}_{k;k} + \lime\epsilon_l)^2}\\
&& + \frac{\lime\mathfrak{c}_{k;k}^{-;+}\lime\mathfrak{c}_{k;k}^{-;+}\lime\mathfrak{c}_{k;k}}{(\lime\mathfrak{c}_{k;k} + \lime\epsilon_l)^3}\\
&=&\frac{\mathfrak{g}_k(\mathfrak{q}_m-\mathfrak{c}_m)}{\mathfrak{g}_m\mathfrak{q}_k} -2\frac{\lime\mathfrak{c}_{k;k}^{-;+}\lime\mathfrak{c}_{k;k}^{-;+}}{\mathfrak{q}_k^2} + \frac{\mathfrak{q}_k\lime\mathfrak{c}_{k;k}^{-;+}\lime\mathfrak{c}_{k;k}^{-;+}}{\mathfrak{q}_k^3}\\
&=&\frac{\mathfrak{g}_l(\mathfrak{q}_m-\mathfrak{c}_m)}{\mathfrak{g}_m} -\frac{\lime\mathfrak{c}_{k;k}^{-;+}\lime\mathfrak{c}_{k;k}^{-;+}}{\mathfrak{q}_k^2}\\
\end{eqnarray*}

If $n^-\neq n^+$, $\lime\mathfrak{c}_{k;k}^{-;+}=0$, so $\lime\mathfrak{c}_{l;l}^{-;-}=\frac{\mathfrak{g}_l(\mathfrak{q}_m-\mathfrak{c}_m)}{\mathfrak{g}_m}$ as required. $\lime\mathfrak{c}_{l;l}^{-;+}=\lime\frac{\mathfrak{c}_{k;k}^{-;+}\epsilon_l}{(\mathfrak{c}_{k;k} + \epsilon_l)^2}=0$ as required.
\item Case: $f_l$ is an addition layer.

$$\lime\mathfrak{c}_{l;l}^{-;-}=\lime\sum_{\kappa=1}^Kw_\kappa^2\mathfrak{c}_{k[\kappa];k[\kappa]}^{-;-}=\sum_{\kappa=1}^Kw_\kappa^2\frac{\mathfrak{g}_{k[\kappa]}(\mathfrak{q}_m-\mathfrak{c}_m)}{\mathfrak{g}_m}=\frac{\mathfrak{g}_l(\mathfrak{q}_m-\mathfrak{c}_m)}{\mathfrak{g}_m}$$

as required. $\lime\mathfrak{c}_{l;l}^{-;+}=\lime\sum_{\kappa=1}^Kw_\kappa^2\mathfrak{c}_{k[\kappa];k[\kappa]}^{-;+}=\sum_{\kappa=1}^Kw_\kappa^2\lime\mathfrak{c}_{k[\kappa];k[\kappa]}^{-;+}$ If $n^-\neq n^+$, the $\lime\mathfrak{c}_{k[\kappa];k[\kappa]}^{-;+}$ are zero, so $\lime\mathfrak{c}_{l;l}^{-;+}=0$ as required.

\item Case: $f_l$ is an activation layer. As before, let $u\sim\mathcal{N}(0, \mathfrak{c}_{k;k})$ and $v\sim\mathcal{N}(0, \frac{\det\Sigma_k}{\mathfrak{c}_{k;k}})$. Then 

\begin{eqnarray*}
&&\lime\mathfrak{c}_{l;l}^{-;-}\\
&=&\lime\mathbb{E}_{t,s\sim\mathcal{N}(0,\Sigma_k)}t^2\tau'_l(s)^2\\
&=&\lime\mathbb{E}_{v,u}\Big(v+\frac{\mathfrak{c}_{k;k}^{-;+}}{\mathfrak{c}_{k;k}}u\Big)^2\tau'_l(u)^2\\
&=&\lime\Big(\mathbb{E}_vv^2\mathbb{E}_u \tau'_l(u)^2+ \frac{\mathfrak{c}_{k;k}^{-;+}}{\mathfrak{c}_{k;k}}\mathbb{E}_vv\mathbb{E}_uu\tau'_l(u)^2 + \frac{\mathfrak{c}_{k;k}^{-;+}\mathfrak{c}_{k;k}^{-;+}}{\mathfrak{c}_{k;k}\mathfrak{c}_{k;k}}\mathbb{E}_uu^2\tau'_l(u)^2\Big)\\
&=&\lime\Big( \frac{\det\Sigma_k}{\mathfrak{c}_{k;k}}\mathbb{E}_u \tau'_l(u)^2 + \frac{\mathfrak{c}_{k;k}^{-;+}\mathfrak{c}_{k;k}^{-;+}}{\mathfrak{c}_{k;k}\mathfrak{c}_{k;k}}\mathbb{E}_uu^2\tau'_l(u)^2\Big)\\
&=&\lime\Big( \frac{\mathfrak{c}_{k;k}^{-;-}\mathfrak{c}_{k;k}-\mathfrak{c}_{k;k}^{-;+}\mathfrak{c}_{k;k}^{-;+}}{\mathfrak{c}_{k;k}}\mathbb{E}_u \tau'_l(u)^2+ \frac{\mathfrak{c}_{k;k}^{-;+}\mathfrak{c}_{k;k}^{-;+}}{\mathfrak{c}_{k;k}\mathfrak{c}_{k;k}}\mathbb{E}_uu^2\tau'_l(u)^2\Big)\\
\end{eqnarray*}

$\lime \mathfrak{c}_{k;k} = \mathfrak{q}_k > 0$. Also, since $\tau'_l$ is controlled by $\mathcal{C}^\text{E2}$, so is $\tau'_l(u)^2$ and $u^2\tau'_l(u)^2$. So we can apply lemma \ref{lemma15} and obtain

\begin{eqnarray*}
&&\lime\mathfrak{c}_{l;l}^{-;-}\\
&=&\frac{\lime\mathfrak{c}_{k;k}^{-;-}\mathfrak{q}_k-\lime\mathfrak{c}_{k;k}^{-;+}\lime\mathfrak{c}_{k;k}^{-;+}}{\mathfrak{q}_k}\mathbb{E}_{s\sim\mathcal{N}(0,\mathfrak{q}_k)} \tau'_l(s)^2\\
&&+ \frac{\lime\mathfrak{c}_{k;k}^{-;+}\lime\mathfrak{c}_{k;k}^{-;+}}{\mathfrak{q}_k\mathfrak{q}_k}\mathbb{E}_{s\sim\mathcal{N}(0,\mathfrak{q}_k)}s^2\tau'_l(s)^2\\
\end{eqnarray*}

Is $n^-\neq n^+$, we further have $\lime\mathfrak{c}_{k;k}^{-;+}=0$, so then 

\begin{eqnarray*}
&&\lime\mathfrak{c}_{l;l}^{-;-}\\
&=&\lime\mathfrak{c}_{k;k}^{-;-}\mathbb{E}_{s\sim\mathcal{N}(0,\mathfrak{q}_k)}\tau'_l(s)^2\\
&=&\frac{\mathfrak{g}_k(\mathfrak{q}_m-\mathfrak{c}_m)}{\mathfrak{g}_m}\mathbb{E}_{s,t\sim\mathcal{N}\Big(0,\begin{pmatrix}
\mathfrak{q}_k&\mathfrak{q}_k\\\mathfrak{q}_k&\mathfrak{q}_k\end{pmatrix}\Big)}\tau'_l(s)\tau'_l(t)\\
&=&\mathfrak{C}_{\tau'_l}(\mathfrak{q}_k,\mathfrak{q}_k)\frac{\mathfrak{g}_k(\mathfrak{q}_m-\mathfrak{c}_m)}{\mathfrak{g}_m}\\
&=&\frac{\mathfrak{g}_l(\mathfrak{q}_m-\mathfrak{c}_m)}{\mathfrak{g}_m}
\end{eqnarray*}

as required. We also have

\begin{eqnarray*}
&&\lime\mathfrak{c}_{l;l}^{-;+}\\
&=&\lime\mathbb{E}_{t,s\sim\mathcal{N}(0,\Sigma_k)}t\tau'_l(s)\tau_l(s)\\
&=&\lime\mathbb{E}_{v,u}\Big(v+\frac{\mathfrak{c}_{k;k}^{-;+}}{\mathfrak{c}_{k;k}}u\Big)\tau'_l(u)\tau_l(u)\\
&=&\lime \frac{\mathfrak{c}_{k;k}^{-;+}}{\mathfrak{c}_{k;k}}\mathbb{E}_uu\tau'_l(u)\tau_l(u)\\
\end{eqnarray*}

As before, $\lime \mathfrak{c}_{k;k} = \mathfrak{q}_k > 0$ and $u\tau'_l(u)\tau_l(u)$ is controlled by $\mathcal{C}^\text{E2}$, so $\lime\mathfrak{c}_{l;l}^{-;+}=\frac{\lime \mathfrak{c}_{k;k}^{-;+}}{\mathfrak{q}_k}\mathbb{E}_{s\sim\mathcal{N}(0,\mathfrak{q}_k)}s\tau'_l(s)\tau_l(s)$. When $n^-\neq n^+$, $\lime \mathfrak{c}_{k;k}^{-;+}=0$, so $\lime\mathfrak{c}_{l;l}^{-;+}=0$ as required.
\end{itemize}

This completes the induction and hence stage 4.

\underline{Stage 5}: statement (\ref{eqn5p10})

\begin{eqnarray*}
&&\lime \limd NLC(f_l(f_m),f_m(\mathcal{D}^{(N)}))\\
&=&\lime \limd \sqrt{\frac{\mathbb{E}_x\Tr(\mathcal{J}_{l,m}(x)\Cov_{f_m}\mathcal{J}_{l,m}^T(x))}{\Tr(\Cov_{f_l})}}\\
&=&\lime \limd \sqrt{\frac{\mathbb{E}_x\Tr(\mathcal{J}_{l,m}(x)\Cov_{f_m}\mathcal{J}_{l,m}^T(x))}{||\mathbb{S}_xf_l(x)||_2^2}}\\
&=&\lime \limd \sqrt{\frac{\frac{1}{d_l^\text{MF}}\mathbb{E}_x\Tr(\mathcal{J}_{l,m}(x)\Cov_{f_m}\mathcal{J}_{l,m}^T(x))}{\frac{1}{d_l^\text{MF}}||\mathbb{S}_xf_l(x)||_2^2}}\\
&=&\sqrt{\frac{\lime \limd \frac{1}{d_l^\text{MF}}\mathbb{E}_x\Tr(\mathcal{J}_{l,m}(x)\Cov_{f_m}\mathcal{J}_{l,m}^T(x))}{\lime \limd \frac{1}{d_l^\text{MF}}||\mathbb{S}_xf_l(x)||_2^2}}\\
&=&\sqrt{\frac{\mathfrak{g}_l(\mathfrak{q}_m-\mathfrak{c}_m)}{\mathfrak{g}_m(\mathfrak{q}_l-\mathfrak{c}_l)}}
\end{eqnarray*}

The denominator is positive by proposition \ref{mfntPositive}.

\underline{Stage 6}: statement (\ref{eqn5p4})

The case $l=m$ is trivial. For $l > m$, as in stage 4, we will treat the cases $m>0$ and $m=0$ somewhat separately.

Case: $m > 0$. We will show $\lime \limd \frac{1}{d_l^\text{MF}}||\mathcal{J}_{l,m}(x^{(1)})u||_2^2 = \frac{\mathfrak{g}_l}{\mathfrak{g}_m}$, where randomness is induced by $\theta$, $x^{(1)}$ and the unit Gaussian vector $u$. From lemma \ref{lemma17}, we then obtain $\limd \frac{1}{d_l^\text{MF}}||\mathcal{J}_{l,m}(x^{(1)})u||_2^2=\limd \frac{1}{d_l^\text{MF}}||\mathcal{J}_{l,m}(x^{(1)})||_F^2$, and hence $\lime \limd \frac{1}{d_l^\text{MF}}||\mathcal{J}_{l,m}(x^{(1)})||_F^2 = \frac{\mathfrak{g}_l}{\mathfrak{g}_m}$ as required.

We have $\limd \frac{1}{d_l^\text{MF}}||\mathcal{J}_{l,m}(x^{(1)})u||_2^2 = \limd \overline{(u\mathcal{J}_{l,m}(x^{(1)})^T)^{.2}}$. This is the same expression we investigated in stage 4 as $p_l^{(n^-,n^+)}$ with $n^+=1$, except that $f_m(x^{(n^-)})-\frac{1}{N}\sum_{n=1}^Nf_m(x^{(n)})$ is replaced by $u$. In fact, if we let $p^{(n^-,1)}_l$ refer to $u\mathcal{J}_{l,m}(x^{(1)})^T$, we can proceed almost exactly as in stage 4 until the point where we take the limit with respect to $N$ and $\vece$. The following differences arise.

\begin{itemize}
\item $F^\text{MN}$ is an input layer. It corresponds to $u$. Because $u$ is unit Gaussian, we add a diagonal entry of 1 and off-diagonal entries of zero to the covariance matrix of $\mathcal{G}$. Distribution equality is obvious. The limits become $\tilde{\mathfrak{m}}^\text{MN} = 0$, $\mathfrak{c}^\text{MN;MN} = 1$ and $\mathfrak{c}^\text{MN;(n)}_{;m} = 0$. In terms of rank stability to start off the induction, we have $\det \begin{pmatrix}\mathfrak{c}^\text{MN;MN}& \mathfrak{c}^\text{MN;(n)}_{;m}\\ \mathfrak{c}^\text{MN;(n)}_{;m}& \mathfrak{c}_{m;m}\end{pmatrix} = \mathfrak{c}_{m;m} > 0$.
\item The recursive expansion of $\rho_l^-$ halts at $\rho^\text{MN}$, as this is now a dummy multi-activation function. All relevant properties that follow from the expansion are preserved. Specifically, we use that $e^\text{MN}$ is independent of all other $e$ terms for input and FC layers in $F$ and that $\tilde{\mathfrak{m}}^\text{MN}=0$.
\item The case where $f_l$ is a readin layer does not appear in the induction step as $l > m > 0$ and $f_m$ is a bottleneck for $f_l$.
\end{itemize}

Case: $m=0$. We have

\begin{eqnarray*}
&&\lime \limd \frac{1}{d_l^\text{MF}}||\mathcal{J}_{l,m}(x^{(1)})||_F^2\\
&=&\lime \limd \frac{1}{d_l^\text{MF}}\Big|\Big|\sum_{f_{l'} \text{ readin}}\mathcal{J}_{l',0}(x^{(1)})^T\mathcal{J}_{l,l'}(x^{(1)})^T\Big|\Big|_F^2\\
&=&\lime \limd \frac{1}{d_l^\text{MF}}\Big|\Big|\sum_{f_{l'} \text{ readin}}W_{l'}\mathcal{J}_{l,l'}(x^{(1)})^T\Big|\Big|_F^2\\
&=&\lime \limd \frac{1}{d_l^\text{MF}}\sum_{i=0}^{d_\text{in}-1}\Big|\Big|\sum_{f_{l'} \text{ readin}}W_{l'}[i, :]\mathcal{J}_{l,l'}(x^{(1)})^T\Big|\Big|_2^2\\
&=&\lime \limd \frac{1}{d_l^\text{MF}}d_\text{in}\Big|\Big|\sum_{f_{l'} \text{ readin}}W_{l'}[0, :]\mathcal{J}_{l,l'}(x^{(1)})^T\Big|\Big|_2^2\\
&=&\lime \limd \overline{\Big(\sum_{f_{l'} \text{ readin}}\sqrt{d_\text{in}}W_{l'}[0, :]\mathcal{J}_{l,l'}(x^{(1)})^T\Big)^{.2}}\\
\end{eqnarray*}

Here, $W_{l'}[i, 0]$ denotes the $i$'th row of $W_{l'}$ and $\sum_{f_{l'} \text{ readin}}$ is over all readin layers. We use that, by construction, the rows of each $W_{l'}$ have the same distribution. 

Again, we proceed almost exactly as in stage 4 until the point where we take the limit with respect to $N$ and $\vece$, where we now let $p_l^{(n^-, 1)}$ refer to $\sum_{f_{l'} \text{ readin}}\sqrt{d_\text{in}}W_{l'}[0, :]\mathcal{J}_{l,l'}(x^{(1)})^T$. The induction step for non-readin layers $f_l$ is unaltered, as we still have $p_l^{(n^-, 1)} = p_k^{(n^-, 1)}\mathcal{J}_{l,k}(x^{(1)})^T$ when $f_l$ is not an addition layer and $p_l^{(n^-, 1)} = \sum_{\kappa=1}^Kw_\kappa p_{k[\kappa]}^{(n^-, 1)}$ when $f_l$ is an addition layer. Having $\sum_{f_{l'} \text{ readin}}$ be part of the definition of $p_l^{(n^-, 1)}$ does not affect our arguments. When $f_l$ is a readin layer, the distribution of $p_l^{(n^-, 1)}$ is that of $\sqrt{d_\text{in}}W_{l}[0, :]$, which is elementwise, has a Gaussian generator with mean zero and variance $\sigma_l^2$, and is independent of the analogous distributions for other readin layers as there is no weight sharing in $f$. Hence, we add a diagonal entry of $\sigma_l^2$ and off-diagonal entries of zero to $\mathcal{G}$ for $F_l$. The limits become $\tilde{\mathfrak{m}}_l^-=0$, $\mathfrak{c}_{l;l}^{-;-}=\sigma_l^2$ and $\mathfrak{c}_{l;l}^{-;+}=0$. Rank stability holds as $\det \Sigma_l = \mathfrak{c}_{l;l}^{-;-}\mathfrak{c}_{l;l} - \mathfrak{c}_{l;l}^{-;+}\mathfrak{c}_{l;l}^{-;+}=q\sigma_l^4 > 0$.

Now we return again to the general case where $m$ can be zero or non-zero. We are left with taking the limit with respect to $N$ and $\vece$. To do this, we use that we obtained the same recursive calculation rules for $\mathfrak{c}_{l;l}^{-;-}$ and $\mathfrak{c}_{l;l}^{-;+}$ as in stage 4, except for the readin layer case. Since $\mathfrak{c}_{l;l}^{-;-}$ now represents either $\limd \overline{(u\mathcal{J}_{l,m}(x^{(1)})^T)^{.2}}$ or $\limd \overline{\Big(\sum_{f_{l'} \text{ readin}}\sqrt{d_\text{in}}W_{l'}[0, :]\mathcal{J}_{l,l'}(x^{(1)})^T\Big)^{.2}}$ depending on the value of $m$, showing $\lime \mathfrak{c}_{l;l}^{-;-} = \frac{\mathfrak{g}_l}{\mathfrak{g}_m}$ completes this stage. As part of the induction, we also show $\lime \mathfrak{c}_{l;l}^{-;+} = 0$. Denominators are positive as before. Compared to stage 4, all limit quantities are now also independent of $n^-$ and $N$.

\begin{itemize}
\item Case: $F^\text{MF}$. $\lime\mathfrak{c}^{\text{MN};\text{MN}}=1=\frac{\mathfrak{g}_m}{\mathfrak{g}_m}$ and $\lime\mathfrak{c}^{\text{MF};+}_{;m}=0$. This starts off our induction for the case $m > 0$ in the required manner.
\item Case: $f_l$ is a readin layer. If $f_l$ is a readin layer, we have $m=0$, so $\frac{\mathfrak{g}_l}{\mathfrak{g}_m}=\frac{\sigma_l^2}{1}=\sigma_l^2$. Also, $\lime\mathfrak{c}_{l;l}^{-;-}=\sigma_l^2$ and $\lime\mathfrak{c}_{l;l}^{-;+}=0$ as required due to the generator addition described above.
\item Case: $f_l$ is a fully-connected, non-readin layer. $\lime\mathfrak{c}_{l;l}^{-;-}=\sigma_l^2\lime\mathfrak{c}_{k;k}^{-;-}=\sigma_l^2\frac{\mathfrak{g}_k}{\mathfrak{g}_m}=\frac{\mathfrak{g}_l}{\mathfrak{g}_m}$ as required. $\lime\mathfrak{c}_{l;l}^{-;+}=\sigma_l^2\lime\mathfrak{c}_{k;k}^{-;+}=0$ as required.
\item Case: $f_l$ is a bias layer. $\lime\mathfrak{c}_{l;l}^{-;-}=\lime\mathfrak{c}_{k;k}^{-;-}=\frac{\mathfrak{g}_k}{\mathfrak{g}_m}=\frac{\mathfrak{g}_l}{\mathfrak{g}_m}$ as required. $\lime\mathfrak{c}_{l;l}^{-;+}=\lime\mathfrak{c}_{k;k}^{-;+}=0$ as required.
\item Case: $f_l$ is an LN layer. 
\begin{eqnarray*}
&&\lime\mathfrak{c}_{l;l}^{-;-}\\
&=&\lime\frac{\mathfrak{c}_{k;k}^{-;-}}{\mathfrak{c}_{k;k} + \epsilon_l} -2\frac{\mathfrak{c}_{k;k}^{-;+}\mathfrak{c}_{k;k}^{-;+}}{(\mathfrak{c}_{k;k} + \epsilon_l)^2} + \frac{\mathfrak{c}_{k;k}^{-;+}\mathfrak{c}_{k;k}^{-;+}\mathfrak{c}_{k;k}}{(\mathfrak{c}_{k;k} + \epsilon_l)^3}\\
&=&\frac{\lime\mathfrak{c}_{k;k}^{-;-}}{\lime\mathfrak{c}_{k;k} + \lime \epsilon_l}\\
&=&\frac{\mathfrak{g}_k}{\mathfrak{g}_m\mathfrak{q}_k}\\
&=&\frac{\mathfrak{g}_l}{\mathfrak{g}_m}
\end{eqnarray*}

as required. $\lime\mathfrak{c}_{l;l}^{-;+}=\lime\frac{\mathfrak{c}_{k;k}^{-;+}\epsilon_l}{(\mathfrak{c}_{k;k} + \epsilon_l)^2}=0$ as required.
\item Case: $f_l$ is an addition layer.

$$\lime\mathfrak{c}_{l;l}^{-;-}=\lime\sum_{\kappa=1}^Kw_\kappa^2\mathfrak{c}_{k[\kappa];k[\kappa]}^{-;-}=\sum_{\kappa=1}^Kw_\kappa^2\frac{\mathfrak{g}_{k[\kappa]}}{\mathfrak{g}_m}=\frac{\mathfrak{g}_l}{\mathfrak{g}_m}$$

as required. $\lime\mathfrak{c}_{l;l}^{-;+}=\lime\sum_{\kappa=1}^Kw_\kappa^2\mathfrak{c}_{k[\kappa];k[\kappa]}^{-;+}=\sum_{\kappa=1}^Kw_\kappa^2\lime\mathfrak{c}_{k[\kappa];k[\kappa]}^{-;+} = 0$ as required.

\item Case: $f_l$ is an activation layer. Analogous to stage 4, we have 

\begin{eqnarray*}
&&\lime\mathfrak{c}_{l;l}^{-;-}\\
&=&\lime\Big( \frac{\mathfrak{c}_{k;k}^{-;-}\mathfrak{c}_{k;k}-\mathfrak{c}_{k;k}^{-;+}\mathfrak{c}_{k;k}^{-;+}}{\mathfrak{c}_{k;k}}\mathbb{E}_u \tau'_l(u)^2+ \frac{\mathfrak{c}_{k;k}^{-;+}\mathfrak{c}_{k;k}^{-;+}}{\mathfrak{c}_{k;k}\mathfrak{c}_{k;k}}\mathbb{E}_uu^2\tau'_l(u)^2\Big)\\
&=&\lime \mathfrak{c}_{k;k}^{-;-} \mathbb{E}_u \tau'_l(u)^2 \\
&=&\frac{\mathfrak{g}_k}{\mathfrak{g}_m}\mathbb{E}_{s,t\sim\mathcal{N}\Big(0,\begin{pmatrix}
\mathfrak{q}_k&\mathfrak{q}_k\\\mathfrak{q}_k&\mathfrak{q}_k\end{pmatrix}\Big)}\tau'_l(s)\tau'_l(t)\\
&=&\mathfrak{C}_{\tau'_l}(\mathfrak{q}_k,\mathfrak{q}_k)\frac{\mathfrak{g}_k}{\mathfrak{g}_m}\\
&=&\frac{\mathfrak{g}_l}{\mathfrak{g}_m}
\end{eqnarray*}

as required.

We also have

\begin{eqnarray*}
&&\lime\mathfrak{c}_{l;l}^{-;+}\\
&=&\lime \frac{\mathfrak{c}_{k;k}^{-;+}}{\mathfrak{c}_{k;k}}\mathbb{E}_uu\tau'_l(u)\tau_l(u)\\
&=&\frac{\lime \mathfrak{c}_{k;k}^{-;+}}{\lime \mathfrak{c}_{k;k}}\mathbb{E}_{s\sim\mathcal{N}(0,\mathfrak{q}_k)}u\tau'_l(u)\tau_l(u)\\
&=&0
\end{eqnarray*}

as required.
\end{itemize}

\underline{Stage 7}: statement (\ref{eqn5p8})

Defining $\mathfrak{c}_{l;l}^{-;-}$ as in stage 6, we have

\begin{eqnarray*}
&&\lim_{\vec{\epsilon},N}\limd\frac{1}{d_l^\text{MF}}\mathbb{E}_x||\mathcal{J}_{l,m}||^2_F\\
&=&\lim_{\vec{\epsilon},N}\limd\frac{1}{N}\sum_{n=1}^N\frac{1}{d_l^\text{MF}}||\mathcal{J}_{l,m}(x^{(n)})||^2_F\\
&=&\lim_{\vec{\epsilon},N}\frac{1}{N}\sum_{n=1}^N\limd\frac{1}{d_l^\text{MF}}||\mathcal{J}_{l,m}(x^{(n)})||^2_F\\
&=&\lim_{\vec{\epsilon},N}\frac{1}{N}\sum_{n=1}^N\begin{cases}\mathfrak{c}_{l;l}^{-;-} \text{ if } l>m\\1 \text{ else}\end{cases}\\
&=&\lim_{\vec{\epsilon},N}\begin{cases}\mathfrak{c}_{l;l}^{-;-} \text{ if } l>m\\1 \text{ else}\end{cases}\\
&=&\frac{\mathfrak{g}_l}{\mathfrak{g}_m}
\end{eqnarray*}

\underline{Stage 8}: statements (a) and (b)

Assume now that $f_L$ is a readout layer. Fix $N$. Throughout this proof, we have used background theorem \ref{backgroundMaster}. Now, we use theorem \ref{mfntMetaGaussian} and background corollary \ref{backgroundKernel} for statements (a) and (b) respectively.

To apply those results, we construct a mean field architecture $F$ with a single finite input layer and readout layer. We proceed as in stage 1 with the following differences.

\begin{itemize}
\item Like $f_L$, $F_L$ is now a readout layer of dimensionality $d_L$.
\item $F$ now has a finite input layer $F_0$ that is associated with $\mathcal{D}$ as well as individual inputs $x^{(n)}$.
\item For a readin layer $f_l$ in $f$, instead of adding an input layer to $F$ and extending $\mathcal{G}$, we add a readin layer $F_l$ to $F$ with the same width and variance parameter as $f_l$ that does not share weights and has dependency $F_0$.
\item $\mathcal{G}$ now only covers layers derived from bias layers.
\end{itemize}

It is easy to see that, as in stage 1, we have distribution equality. For (a), this means that $f_L$ and $F_L$ have the same meta-distribution. For (b), this means that $(f_L(x^{(1)}), .., f_L(x^{(N)}))$ has the same distribution as $(F_L(x^{(1)}), .., F_L(x^{(N)}))$ for any fixed sample $x^{(1)}, .., x^{(N)}$. Note that, as in the statement of background corollary \ref{backgroundKernel}, the surrogate input generated by $\mathcal{G}$ does not vary between inputs $x^{(n)}$, as it represents the random initial bias vectors.

Let's verify the conditions of theorem \ref{mfntMetaGaussian} and background corollary \ref{backgroundKernel}. For (b), we only care about almost all samples, i.e. we can restrict ourselves to a set of samples that has probability 1. So as before, we can assume $\overline{x^{(n)}.x^{(n)}}=\overline{x^{(n')}.x^{(n')}}=q$ and $\overline{x^{(n)}.x^{(n')}}=c$ for any $1 \le n, n' \le N$, $n \neq n'$. Then, the conditions of theorem \ref{mfntMetaGaussian} and background corollary \ref{backgroundKernel} become equivalent as $K_\text{in}$ becomes $K(N, q, c)$. Further, it turns out that $(\clipout(\clipin(F)), \mathcal{G}(K(N,q,c)) \times \mathcal{G}^N)$ is the mean field architecture and generator we constructed early in stage 4 denoted as $(F^{(1)}, .., F^{(N)})$ for the case $m=L-1$. We denoted the mean field architecture by $(F^{(1)}, .., F^{(N)})$. Hence, we showed that the conditions of background theorem \ref{backgroundMaster} hold for $(\clipout(\clipin(F)), \mathcal{G}(K(N,q,c)) \times \mathcal{G}^N)$ in stage 4. Hence, the conditions of theorem \ref{mfntMetaGaussian} and background corollary \ref{backgroundKernel} hold.

So we have that (i) the meta-distribution of $F$ expansion-converges as $d_\text{MF} \rightarrow \infty$ to the elementwise meta-distribution with generator $\mathcal{MN}(\mathfrak{c}_{L;L}, \mathfrak{c}_{L;L}^{\prime;\prime\prime})$ and (ii) for almost all samples $(F(x^{(1)}), .., F(x^{(N)}))$ converges in distribution as $d_\text{MF} \rightarrow \infty$ to a Gaussian that is elementwise over output vectors where the generator has mean zero and covariance matrix $K(N, \mathfrak{c}_{L;L}, \mathfrak{c}_{L;L}^{\prime;\prime\prime})$. By distribution equality, the same holds for $f$. And, as we showed in stage 1 and 2, $\lim_{\vec{\epsilon}} \mathfrak{c}_{L;L} = \mathfrak{q}_L$ and $\lim_{\vec{\epsilon}} \mathfrak{c}_{L;L}^{\prime;\prime\prime} = \mathfrak{c}_L$. (The limit over $N$ was not used in stages 1 and 2.)

Finally, we note that when $f_L$ is a readout layer, statements (\ref{eqn5p0}) through (\ref{eqn5p10}) still hold when $l < L$ as $(f_0, .., f_{L-1})$ is a valid A-architecture without finite output layer.

\end{proof}

\subsection{Theorem \ref{covkerLregular}: properties of the length kernel $\mathfrak{L}_\tau(\lambda)$} \label{covkerLregularsection}

\begin{reptheorem}{covkerLregular}
For any $\tau$ we have
\begin{enumerate} 
\item If $\mathfrak{L}_\tau(\lambda) \neq 0$ for some $\lambda > 0$, then for all $\lambda > 0$ we have
\begin{itemize}
\item[(a)]$\mathfrak{L}_\tau(\lambda) > 0$
\item[(b)] $\mathfrak{L}_\tau$ is differentiable at $\lambda$ with $\mathfrak{L}_\tau'(\lambda) = \frac{1}{2}\mathfrak{L}_\tau(\lambda)^{-1}\mathbb{E}_{s \sim \mathcal{N}(0, \lambda^2)}\tau(s)^2(s^2\lambda^{-3}-\lambda^{-1})$
\item[(c)] $\frac{\mathfrak{L}_\tau'(\lambda)\lambda}{\mathfrak{L}_\tau(\lambda)} > -\frac{1}{2}$
\item[(d)] $(\sqrt{\lambda}\mathfrak{L}_\tau(\lambda))' > 0$
\end{itemize} \label{covkerLregular1}
\item If $\tau$ is continuous and not zero everywhere, then $\mathfrak{L}_\tau(\lambda) > 0$ for all $\lambda > 0$. \label{covkerLregular2}
\item If $\tau$ is continuous at zero, $\mathfrak{L}_\tau$ is continuous at zero. \label{covkerLregular3}
\item If $\tau$ is directionally differentiable at zero, then $\mathfrak{L}_\tau$ is differentiable at zero with $\mathfrak{L}_\tau'(0) = \sqrt{\frac{\tau^-(0)^2 + \tau^+(0)^2}{2}}$ if $\tau(0) = 0$ and $\mathfrak{L}_\tau'(0) = \frac{1}{\sqrt{2\pi}}\sign(\tau(0))(\tau^+(0)-\tau^-(0))$ if $\tau(0) \neq 0$. The + and - superscripts indicate the right and left derivative respectively. \label{covkerLregular4}
\item If $\mathfrak{L}_\tau(\lambda) \neq 0$ for some $\lambda > 0$, then for all $\lambda > 0$ the sequence $(\mathfrak{L}_\tau^n(\lambda))_n$ either (i) is strictly increasing and diverges to infinity or (ii) converges. \label{covkerLregular5}
\end{enumerate}
\end{reptheorem}

\begin{proof}
We prove the claims in turn.

Claim \ref{covkerLregular1}(a): We argue by contradiction. Assume we have a $\lambda > 0$ such that $\mathfrak{L}_\tau(\lambda) \neq 0$ and a $\lambda' > 0$ such that $\mathfrak{L}_\tau(\lambda') \le 0$. Then

\begin{eqnarray*}
&&0\\
&<&\mathfrak{L}_\tau(\lambda)^2\\
&=&\mathbb{E}_{s\sim\mathcal{N}(0,\lambda^2)}\tau(s)^2\\
&=&\int_{\mathbb{R}}\tau(s)^2n(\lambda^2,s)d\mu_1
\end{eqnarray*}

We write $\nu(\lambda,s)$ for $n(\lambda^2,s)$. By the definition of the integral, there exists $S > 0$ such that $0 < \int_{-S}^S\tau(s)^2\nu(\lambda,s)d\mu_1$. Hence

\begin{eqnarray*}
&&0\\
&<&\Big(\max\Big(\frac{\nu(\lambda,0)}{\nu(\lambda',0)},\frac{\nu(\lambda,S)}{\nu(\lambda',S)}\Big)\Big)^{-1}\int_{-S}^S\tau(s)^2\nu(\lambda,s)d\mu_1\\
&\le&\int_{-S}^S\tau(s)^2\nu(\lambda',s)d\mu_1\\
&\le&\mathbb{E}_{s\sim\mathcal{N}(0,\lambda'^2)}\tau(s)^2
\end{eqnarray*}

Since $\mathfrak{L}_\tau(\lambda') = \sqrt{\mathbb{E}_{s\sim\mathcal{N}(0,\lambda'^2)}\tau(s)^2}$, we have $\mathfrak{L}_\tau(\lambda') > 0$. Contradiction. Hence, claim \ref{covkerLregular1}(a) holds.

Claim \ref{covkerLregular1}(b): Fix $\lambda > 0$. $\nu(\lambda,s)$ is twice differentiable with respect to $\lambda$ for $s \in \mathbb{R}$. Using the Taylor expansion with the Lagrange form of the remainder, for some $\delta$ with $\lambda > \delta > 0$, we have $\nu(\lambda + \epsilon, s) = \nu(\lambda,s) + \epsilon \frac{d}{d\lambda}\nu(\lambda,s) + \frac{1}{2}\epsilon^2 \frac{d^2}{d\lambda^2}\nu(\lambda^+(s),s)$ for all $s \in \mathbb{R}$, $|\epsilon| < \delta$ where $\lambda^+(s) \in [\lambda - \delta, \lambda + \delta]$. We have

\begin{eqnarray*}
&&\Big|\int_{s\in\mathbb{R}}\tau(s)^2\frac{d^2}{d\lambda^2}\nu(\lambda^+(s),s)d\mu_1\Big|\\
&=&\Big|\int_{s\in\mathbb{R}}\tau(s)^2\nu(\lambda^+,s)(s^4(\lambda^+)^{-6}-5s^2(\lambda^+)^{-4}+2(\lambda^+)^{-2})d\mu_1\Big|\\
&\le&\int_{s\in\mathbb{R}}\tau(s)^2\nu(\lambda^+,s)s^4(\lambda^+)^{-6}d\mu_1+5\int_{s\in\mathbb{R}}\tau(s)^2\nu(\lambda^+,s)s^2(\lambda^+)^{-4}d\mu_1\\
&&+2\int_{s\in\mathbb{R}}\tau(s)^2\nu(\lambda^+,s)(\lambda^+)^{-2}d\mu_1\\
&\le&\int_{s\in\mathbb{R}}\tau(s)^2\frac{\lambda+\delta}{\lambda-\delta}\nu(\lambda+\delta,s)s^4(\lambda-\delta)^{-6}d\mu_1\\
&&+5\int_{s\in\mathbb{R}}\tau(s)^2\frac{\lambda+\delta}{\lambda-\delta}\nu(\lambda+\delta,s)s^2(\lambda-\delta)^{-4}d\mu_1\\
&&+2\int_{s\in\mathbb{R}}\tau(s)^2\frac{\lambda+\delta}{\lambda-\delta}\nu(\lambda+\delta,s)(\lambda-\delta)^{-2}d\mu_1\\
&=&(\lambda+\delta)(\lambda-\delta)^{-7}\mathbb{E}_{s\sim\mathcal{N}(0,\lambda+\delta)}s^4\tau(s)^2 + 5(\lambda+\delta)(\lambda-\delta)^{-5}\mathbb{E}_{s\sim\mathcal{N}(0,\lambda+\delta)}s^2\tau(s)^2\\
&&+2(\lambda+\delta)(\lambda-\delta)^{-3}\mathbb{E}_{s\sim\mathcal{N}(0,\lambda+\delta)}\tau(s)^2\\
&<&\infty
\end{eqnarray*}

Hence

\begin{eqnarray*}
&&\lim_{\epsilon \rightarrow 0}\frac{1}{\epsilon}(\mathfrak{L}_\tau(\lambda+\epsilon)^2 - \mathfrak{L}_\tau(\lambda)^2)\\
&=&\lim_{\epsilon \rightarrow 0}\frac{1}{\epsilon}\int_{s\in\mathbb{R}}\tau(s)^2\nu(\lambda+\epsilon,s) - \nu(\lambda,s) d\mu_1\\
&=&\lim_{\epsilon \rightarrow 0}\frac{1}{\epsilon}\int_{s\in\mathbb{R}}\tau(s)^2\Big(\epsilon \nu(\lambda,s)(s^2\lambda^{-3} - \lambda^{-1}) +  \frac{1}{2}\epsilon^2 \frac{d^2}{d\lambda^2}\nu(\lambda^+(s),s)\Big) d\mu_1\\
&=&\lim_{\epsilon \rightarrow 0}\int_{s\in\mathbb{R}}\tau(s)^2\nu(\lambda,s)(s^2\lambda^{-3} - \lambda^{-1}) + \frac{1}{2}\epsilon\tau(s)^2\frac{d^2}{d\lambda^2}\nu(\lambda^+(s),s) d\mu_1\\
&=&\int_{s\in\mathbb{R}}\tau(s)^2\nu(\lambda,s)(s^2\lambda^{-3} - \lambda^{-1}) d\mu_1\\
&=&\mathbb{E}_{s \sim \mathcal{N}(0, \lambda^2)}\tau(s)^2(s^2\lambda^{-3}-\lambda^{-1})
\end{eqnarray*}

\sloppy Let $L(\lambda) = \mathfrak{L}_\tau(\lambda)^2$. We showed that $L$ is differentiable with $L'(\lambda) = \mathbb{E}_{s \sim \mathcal{N}(0, \lambda^2)}\tau(s)^2(s^2\lambda^{-3}-\lambda^{-1})$. As $\mathfrak{L}_\tau(\lambda) > 0$ by claim \ref{covkerLregular1}(a), $\mathfrak{L}_\tau$ is then differentiable with $\mathfrak{L}_\tau'(\lambda) = \frac{1}{2}\frac{L'(\lambda)}{\sqrt{L(\lambda)}} = \frac{1}{2}\mathfrak{L}_\tau(\lambda)^{-1}\mathbb{E}_{s \sim \mathcal{N}(0, \lambda^2)}\tau(s)^2(s^2\lambda^{-3}-\lambda^{-1})$ as required.

Claim \ref{covkerLregular1}(c): Fix $\lambda > 0$. By claim \ref{covkerLregular1}(a), we have $\mathfrak{L}_\tau(\lambda) > 0$. So $\int_{s\in\mathbb{R}}\tau(s)^2\nu(\lambda,s)d\mu_1 > 0$ so there exists an $\epsilon > 0$ with $\int_{s\in\mathbb{R},s\not\in[-\epsilon,\epsilon]}\tau(s)^2\nu(\lambda,s) > 0$, so 

\begin{eqnarray*}
&&0\\
&<&\epsilon^2\int_{s\in\mathbb{R},s\not\in[-\epsilon,\epsilon]}\tau(s)^2\nu(\lambda,s)d\mu_1\\
&\le&\int_{s\in\mathbb{R},s\not\in[-\epsilon,\epsilon]}\tau(s)^2\nu(\lambda,s)s^2d\mu_1\\
&\le&\int_{s\in\mathbb{R}}\tau(s)^2\nu(\lambda,s)s^2d\mu_1\\
&=&\mathbb{E}_{s \sim \mathcal{N}(0, \lambda^2)}\tau(s)^2s^2
\end{eqnarray*}

So we have

\begin{eqnarray*}
&&\mathfrak{L}_\tau'(\lambda)\lambda\\
&=&\frac{1}{2}\mathfrak{L}_\tau(\lambda)^{-1}\mathbb{E}_{s \sim \mathcal{N}(0, \lambda^2)}\tau(s)^2(s^2\lambda^{-2}-1)\\
&=&\frac{1}{2}\mathfrak{L}_\tau(\lambda)^{-1}(\lambda^{-2}\mathbb{E}_{s \sim \mathcal{N}(0, \lambda^2)}\tau(s)^2s^2-\mathbb{E}_{s \sim \mathcal{N}(0, \lambda^2)}\tau(s)^2)\\
&>&\frac{1}{2}\mathfrak{L}_\tau(\lambda)^{-1}(-\mathbb{E}_{s \sim \mathcal{N}(0, \lambda^2)}\tau(s)^2)\\
&=&-\frac{1}{2}\mathfrak{L}_\tau(\lambda)^{-1}\mathfrak{L}_\tau(\lambda)^2\\
&=&-\frac{1}{2}\mathfrak{L}_\tau(\lambda)
\end{eqnarray*}

as required.

Claim \ref{covkerLregular1}(d): Fix $\lambda > 0$. We have

\begin{eqnarray*}
&&(\mathfrak{L}_\tau(\lambda)\lambda^{\frac{1}{2}})'\\
&=&(\mathfrak{L}_\tau'(\lambda)\lambda)\lambda^{-\frac{1}{2}} + \frac{1}{2}\mathfrak{L}_\tau(\lambda)\lambda^{-\frac{1}{2}}\\
&>&\Big(-\frac{1}{2}\mathfrak{L}_\tau(\lambda)\Big)\lambda^{-\frac{1}{2}} + \frac{1}{2}\mathfrak{L}_\tau(\lambda)\lambda^{-\frac{1}{2}}\\
&=&0
\end{eqnarray*}

as required.

Claim \ref{covkerLregular2}: Fix $\lambda > 0$. We have 

$$\mathfrak{L}(\lambda)^2 = \mathbb{E}_{s\sim\mathcal{N}(0,\lambda^2)}\tau(s)^2 = \int_\mathbb{R}\nu(\lambda,s)\tau(s)^2d\mu_1$$

Since $\tau$ is not the zero function and is continuous, so is $\nu(\lambda,s)\tau(s)^2$. Hence, by lemma \ref{lemma8}, the integral is positive, so $\mathfrak{L}(\lambda)$ is as well.

Claim \ref{covkerLregular3}: Let $\lambda > 0$. We have $\mathfrak{L}_\tau(\lambda)^2 = \int_0^\infty(\tau(s)^2 + \tau(-s)^2)\nu(\lambda,s)d\mu_1$. If $\tau(s)$ is continuous at zero, so is $\tau(s)^2 + \tau(-s)^2$. Hence we can apply lemma \ref{lemma9} using $F(s) = \tau(s)^2 + \tau(-s)^2$ and $p=0$ to obtain $\lim_{\lambda\rightarrow 0}\mathfrak{L}_\tau(\lambda)^2 = 2\tau(0)^2\int_0^\infty \nu(\lambda,s)d\mu_1 = \tau(0)^2 = \mathfrak{L}_\tau(0)^2$. Since $\mathfrak{L}_\tau$ is non-negative, $\lim_{\lambda\rightarrow 0}\mathfrak{L}_\tau(\lambda) = \mathfrak{L}_\tau(0)$ as required.

Claim \ref{covkerLregular4}: Let $\lambda > 0$. Because $\tau$ is directionally differentiable at 0, we have $\tau(s) = \tau(0) + s\tau^+(0) + sh(s)$ when $s \ge 0$ and $\tau(s) = \tau(0) + s\tau^-(0) + sh(s)$ when $s \le 0$ where $\lim_{s\rightarrow 0} h(s) = h(0) = 0$.

Case 1: $\tau(0) = 0$. Then we have

\begin{eqnarray*}
&&\lambda^{-2}\mathfrak{L}_\tau(\lambda)^2\\
&=&\lambda^{-2}\int_0^\infty\big((s\tau^+(0)+sh(s))^2 + (-s\tau^-(0)-sh(-s))^2\big)\nu(\lambda,s)d\mu_1\\
&=&\lambda^{-2}\int_0^\infty\big((\tau^+(0)+h(s))^2 + (\tau^-(0)+h(-s))^2\big)s^2\nu(\lambda,s)d\mu_1\\
\end{eqnarray*}

$(\tau^+(0)+h(s))^2 + (\tau^-(0)+h(-s))^2$ is continuous at zero, so we can apply lemma \ref{lemma9} with this expression as $F$ and $p=2$ to obtain

\begin{eqnarray*}
&&\lim_{\lambda\rightarrow 0}\lambda^{-2}\mathfrak{L}_\tau(\lambda)^2\\
&=&(\tau^+(0)^2 + \tau^-(0)^2)\int_0^\infty s^2\nu(1,s)d\mu_1\\
&=&(\tau^+(0)^2 + \tau^-(0)^2)\frac{1}{2}\\
\end{eqnarray*}

As $\mathfrak{L}_\tau(0)=|\tau(0)| = 0$, we have $\lim_{\lambda\rightarrow 0}\lambda^{-1}(\mathfrak{L}_\tau(\lambda)-\mathfrak{L}_\tau(0)) = \sqrt{\frac{\tau^-(0)^2 + \tau^+(0)^2}{2}}$ as required.

Case 2: $\tau(0) \neq 0$. Then we have

\begin{eqnarray*}
&&\lambda^{-1}(\mathfrak{L}_\tau(\lambda)^2-\mathfrak{L}_\tau(0)^2)\\
&=&\lambda^{-1}\int_0^\infty\big((\tau(0) + s\tau^+(0) + sh(s))^2 + (\tau(0) - s\tau^-(0) - sh(-s))^2 - 2\tau(0)^2\big)\nu(\lambda,s)d\mu_1 \\
&=&\lambda^{-1}\int_0^\infty\Big(2\tau(0)(\tau^+(0) + h(s) - \tau^-(0) - h(-s))\\
&& + s(\tau^+(0) + h(s))^2 + s(\tau^-(0) + h(-s))^2\Big)s\nu(\lambda,s)d\mu_1\\
\end{eqnarray*}

The expression inside the large brackets is continuous at zero, so we can apply lemma \ref{lemma9} with this expression as $F$ and $p=1$ to obtain

\begin{eqnarray*}
&&\lim_{\lambda\rightarrow 0}\lambda^{-1}(\mathfrak{L}_\tau(\lambda)^2-\mathfrak{L}_\tau(0)^2)\\
&=&2\tau(0)(\tau^+(0)-\tau^-(0))\int_0^\infty s\nu(1,s)d\mu_1\\
&=&\sqrt{\frac{2}{\pi}}\tau(0)(\tau^+(0)-\tau^-(0))
\end{eqnarray*}

Let $L(\lambda) = \mathfrak{L}_\tau(\lambda)^2$. We showed $L'(0)=\sqrt{\frac{2}{\pi}}\tau(0)(\tau^+(0)-\tau^-(0))$. Hence $\mathfrak{L}_\tau'(0)=\frac{1}{2}\frac{L'(0)}{\sqrt{L(0)}}=\frac{1}{2}\frac{\sqrt{\frac{2}{\pi}}\tau(0)(\tau^+(0)-\tau^-(0))}{|\tau(0)|} = \frac{1}{\sqrt{2\pi}}\sign(\tau(0))(\tau^+(0)-\tau^-(0))$ as required.

Claim \ref{covkerLregular5}: The condition of claim \ref{covkerLregular1} holds, so claim \ref{covkerLregular1} holds. Fix $\lambda$. Denote the sequence $(\mathfrak{L}_\tau^n(\lambda))_n$ by $S_n$. If $S_n$ is increasing, then it either diverges to infinity or it converges, so (i) or (ii) hold. If there is an $n$ such that $S_n = S_{n+1}$, the sequence converges, so (ii) holds. So going forward assume there is an $n$ such that $S_n > S_{n+1}$.

Case 1: There are finitely many $n$ such that $S_n > S_{n+1}$.

Let $n$ now be the largest such value. We prove by induction that for all $n' > n$ we have $S_{n+1} \le S_{n'} < \sqrt{S_nS_{n+1}}$. The statement trivially holds for $n'=n+1$. Assume it holds for some $n' > n$. Because $n$ is the largest value with $S_n > S_{n+1}$, we have $S_{n'+1} \ge S_{n'} \ge S_{n+1}$. Further, by claim \ref{covkerLregular1}(d), we have $\mathfrak{L}_\tau(S_{n'}) < \mathfrak{L}_\tau(S_n)\sqrt{\frac{S_n}{S_{n'}}} = S_{n+1}\sqrt{\frac{S_n}{S_{n'}}} \le \sqrt{S_nS_{n+1}}$. This completes the induction. So eventually $S_n$ is increasing and bounded above, so (ii) holds.

Case 2: There are infinitely many $n$ such that $S_n > S_{n+1}$.

Let the sequence of such values $n$ be denoted by $S_{n_d}$. Assume there is some $d$ such that $S_{n_d} \le S_{n_{d+1}}$. Then we can't have $n_{d+1} = n_{d} + 1$ because $S_{n_d} > S_{n_d+1}$ by definition. So $n_{d+1} > n_{d} + 1$. Then from the induction under case 1 we obtain $S_{n_d+1} \le S_{n_{d+1}-1} < \sqrt{S_{n_d}S_{n_d+1}}$ and finally $S_{n_{d+1}} < \sqrt{S_{n_d}S_{n_d+1}}$. Contradiction. So $S_{n_d}$ is strictly decreasing, so it converges. Again, using the induction from case 1, we obtain that $\sqrt{S_{n_d}S_{n_d+1}} > S_n$ for all $n_d < n < n_{d+1}$ and hence $S_{n_d} > S_n$ for all $n > n_d$. So if $S_{n_d}$ converges to zero, so does $S_n$, so (ii) holds. So going forward assume that $S_{n_d}$ converges to a positive value. Call it $L$. Assume $S_n$ does not converge to $L$. Then there is a $\delta$ such that for arbitrarily large $n$ we have $|S_n - L| > \delta$. Since there is an $S_{n_d}$ with $S_{n_d} < L + \delta$, eventually $S_n < L + \delta$. Hence for arbitrarily large $n$ we have $S_n < L - \delta$. Because $S_{n_d+1}, .., S_{n_{d+1}}$ is strictly increasing for each $d$, there are arbitrarily large $d$ such that $S_{n_d+1} < L - \delta$. But then $S_{n_{d+1}} < \sqrt{S_{n_d}S_{n_d+1}}$ and $S_{n_{d+1}} > L$ yields $S_{n_d} > L + \delta$. So there are arbitrarily large $d$ such that $S_{n_d} > L + \delta$, so $S_{n_d}$ does not converge to $L$. Contradiction. So $S_n$ does converge to $L$, so (ii) holds, as required.

\end{proof}

\subsection{Theorem \ref{covkerCregular}: properties of the covariance kernel $\mathfrak{C}_\tau(c)$} \label{covkerCregularsection}

\begin{reptheorem}{covkerCregular}
Assume $\tau$ is piecewise 5-differentiable. Consider $\mathfrak{C}_\tau(c)$ defined on $[0,1]$. Let $a_\tau s + b_\tau$ be the least squares linear fit to $\tau$ under $\mathcal{N}(0,1)$ and let $\tilde{\tau} = \tau - a_\tau s - b_\tau$. Then

\begin{enumerate}[label=(\alph*)]
\item $\mathfrak{C}_\tau$ is differentiable and $\mathfrak{C}_\tau' = \mathfrak{C}_{\tau'}$.
\item $\mathfrak{C}_\tau$ is increasing, convex and for all $c > 0$, $\epsilon > 0$ with $c + 3\epsilon \le 1$ we have $\mathfrak{C}_\tau(c+3\epsilon) - 3\mathfrak{C}_\tau(c+2\epsilon) + 3\mathfrak{C}_\tau(c+\epsilon) - \mathfrak{C}_\tau(c)\ge 0$.
\item (i) $\mathfrak{C}_\tau(c) = \mathfrak{C}_{\tilde{\tau}}(c) + a_\tau^2c + b_\tau^2$, (ii) $\mathfrak{C}_{\tilde{\tau}}(0) = 0$, (iii) $\mathfrak{C}_{\tau}(0) = b_\tau^2$, (iv) $\mathfrak{C}_{\tilde{\tau}}'(0) = 0$, (v) $\mathfrak{C}_{\tau}'(0) = a_\tau^2$, (vi) $CAR(\tau, \mathcal{N}(0,1)) = \frac{\mathfrak{C}_{\tau}(0)}{\mathfrak{C}_{\tau}(1)}$, (vii) $LAR(\tau, \mathcal{N}(0,1)) = \frac{\mathfrak{C}_{\tau}'(0)}{\mathfrak{C}_{\tau}(1)}$ and (viii) $\mathfrak{C}_{\tau}$ is linear if and only if $\tau$ is linear.
\end{enumerate}

Now also assume $\tau$ is not linear. Then, for the normalized covariance kernel $\tilde{\mathfrak{C}}_\tau(c) = \frac{\mathfrak{C}_\tau(c)}{\mathfrak{C}_\tau(1)}$ exactly one of three cases hold on $[0,1]$.

\begin{enumerate}
\item There is a stable fixed point at $c^\text{lim}=1$ with exponential convergence rate and no other fixed point. $0 < \tilde{\mathfrak{C}}_\tau'(1) < 1$ holds.
\item There is a stable fixed point at $c^\text{lim}=1$ with sub-exponential convergence rate and no other fixed point. $\tilde{\mathfrak{C}}_\tau'(1) = 1$ holds.
\item There is an unstable fixed point at 1 and there is exactly one other fixed point $c^\text{lim}$ in $[0,1)$, which is stable. $\tilde{\mathfrak{C}}_\tau'(1) > 1$ and $1 > \tilde{\mathfrak{C}}_\tau'(c^\text{lim}) \ge 0$ hold. The convergence rate is super-exponential or exponential depending on whether $\tilde{\mathfrak{C}}_\tau'(c^\text{lim}) = 0$ holds.
\end{enumerate}

In particular, there exists exactly one stable fixed point $c^\text{lim}$ and iterating $\tilde{\mathfrak{C}}_\tau$ will lead to convergence towards $c^\text{lim}$ from any starting point.

\end{reptheorem}

\begin{proof}

Like the proof of theorem \ref{finiteNetTilde}, this proof proceeds in a number of stages, where each stage builds upon the results of previous stages. It is easy to check that an alternative definition for $\mathfrak{C}_\tau(c)$ is $\mathbb{E}_{s,t\sim\mathcal{N}(0,1)}\tau(s)\tau(cs+\sqrt{1-c^2}t)$. Throughout this proof, let $s$ and $t$ be independent unit Gaussians. 

\underline{Stage 1}: Assume $\tau$ is twice differentiable. Then for $1 > c \ge 0$, $\mathfrak{C}_\tau$ is differentiable at $c$ and $\mathfrak{C}_\tau'(c) = \mathfrak{C}_{\tau'}(c)$.

Throughout stage 1, consider $1 > c \ge 0$ fixed and consider $\mathfrak{C}_\tau$ defined on $[-1,1]$. Let $|\epsilon| < \frac{1}{2}(1-c)$ and let 

$$\gamma = \frac{2c\epsilon - \epsilon^2}{1-c^2}$$

and

$$\delta=-\epsilon s-\sqrt{1-c^2}t+\sqrt{1-(c-\epsilon)^2}t$$

Then we have 

$$\delta=-\epsilon s+\sqrt{1-c^2}t(\sqrt{1+\gamma} - 1)=-\epsilon s+\sqrt{1-c^2}t(\frac{1}{2}\gamma + r(\gamma)\gamma^2)$$

where $r(\gamma)$ is derived from the Taylor expansion of the square root about 1. All values between $1$ and $1+\gamma$ are contained in the interval $[\frac{1-\frac{1}{4}(1+c)^2}{1-c^2},\frac{1}{1-c^2}]$. Because the square root is twice differentiable in that interval with a bounded second derivative, the Lagrange form of the remainder yields that $|r(\gamma)|$ is bounded. Let the bound be $R$. Then 

\begin{eqnarray*}
&&\mathfrak{C}_\tau(c-\epsilon)-\mathfrak{C}_\tau(c)\\
&=&\mathbb{E}_{s,t}\tau(s)\tau((c-\epsilon)s+\sqrt{1-(c-\epsilon)^2}t)-\mathbb{E}_{s,t}\tau(s)\tau(cs+\sqrt{1-c^2}t)\\
&=&\mathbb{E}_{s,t}\tau(s)\Big(\tau(cs+\sqrt{1-c^2}t+\delta)-\tau(cs+\sqrt{1-c^2}t)\Big)\\
&=&\mathbb{E}_{s,t }\tau(s)\Big(\tau(cs+\sqrt{1-c^2}t)+\delta\tau'(cs+\sqrt{1-c^2}t)\\
&&+\frac{1}{2}\delta^2 h(cs+\sqrt{1-c^2}t,\delta)-\tau(cs+\sqrt{1-c^2}t)\Big)\\
&=&\mathbb{E}_{s,t }\tau(s)\Big((-\epsilon s+\sqrt{1-c^2}t(\frac{1}{2}\gamma + r(\gamma)\gamma^2))\tau'(cs+\sqrt{1-c^2}t)\\
&&+\frac{1}{2}(-\epsilon s+\sqrt{1-c^2}t(\frac{1}{2}\gamma + r(\gamma)\gamma^2))^2 h(cs+\sqrt{1-c^2}t,\delta)\Big)\\
&=&\mathbb{E}_{s,t }\Big(\epsilon(-s+\frac{c}{\sqrt{1-c^2}}t)\tau(s)\tau'(cs+\sqrt{1-c^2}t)\\
&&+\epsilon^2(P_0+P_1h(cs+\sqrt{1-c^2}t,\delta))\Big)
\end{eqnarray*}

Here, $h$ is derived from the Taylor expansion of $\tau$ about $cs+\sqrt{1-c^2}t$. Because $\tau$ is twice differentiable, the Lagrange form of the remainder yields $h(cs+\sqrt{1-c^2}t,\delta)=\tau''(s')$ for some $s'$ with $cs+\sqrt{1-c^2}t \le s' \le cs+\sqrt{1-c^2}t+\delta$. This in turn yields $|h(cs+\sqrt{1-c^2}t,\delta)|\le T_2(|s|+|t|)$, where $T_2$ is defined as in section \ref{otherNTCsection}.

Here, $P_0$ and $P_1$ can be considered low-order polynomials of $s$, $t$, $\epsilon$, $r$, $\tau(s)$ and $\tau'(cs+\sqrt{1-c^2}t)$ terms. (Remember that $c$ is a constant.)

First consider $\mathbb{E}_{s,t}(P_0+P_1h(cs+\sqrt{1-c^2}t,\delta))$, the coefficient of $\epsilon^2$. Let $P_0^+$ and $P_1^+$ be the polynomials that are equal to $P_0$ and $P_1$, except each term has a positive sign. Then

\begin{eqnarray*}
&&\Big|\mathbb{E}_{s,t}\Big(P_0(s,t,\epsilon,r,\tau(s),\tau'(cs+\sqrt{1-c^2}t))\\
&&+P_1(s,t,\epsilon,r,\tau(s),\tau'(cs+\sqrt{1-c^2}t))h(cs+\sqrt{1-c^2}t,\delta)\Big)\Big|\\
&\le&\mathbb{E}_{s,t}\Big(P_0^+(|s|,|t|,\frac{1}{2}(1-c),R,|\tau(s)|,|\tau'(cs+\sqrt{1-c^2}t)|)\\
&&+P_1^+(|s|,|t|,\frac{1}{2}(1-c),R,|\tau(s)|,|\tau'(cs+\sqrt{1-c^2}t)|)T_2(|s|+|t|)\Big)
\end{eqnarray*}

Using the ``full force'' of assumption \ref{assumptionIntegrableMeanField}, this bound is finite. Since it is also independent of $\epsilon$, we have $\lim_{\epsilon\rightarrow 0}\epsilon|\mathbb{E}_{s,t}(P_0+P_1h)| = 0$, so $\lim_{\epsilon\rightarrow 0}\epsilon\mathbb{E}_{s,t}(P_0+P_1h) = 0$. 

Now consider $\mathbb{E}_{s,t}(-s+\frac{c}{\sqrt{1-c^2}}t)\tau(s)\tau'(cs+\sqrt{1-c^2}t)$, the coefficient of $\epsilon$. Let $u$ be a 2-dimensional unit Gaussian column vector. Then we can rewrite the coefficient using vector notation as

$$\mathbb{E}_{s,t}\Big(\begin{pmatrix}-1&\frac{c}{\sqrt{1-c^2}}\end{pmatrix}u\Big)\tau\Big(\begin{pmatrix}1&0\end{pmatrix}u\Big)\tau'\Big(\begin{pmatrix}c&\sqrt{1-c^2}\end{pmatrix}u\Big)$$

Because the distribution of $u$ is spherically symmetric, it is invariant under multiplication with an orthogonal matrix. Let 

$$A=\begin{pmatrix} c & -\sqrt{1-c^2} \\ \sqrt{1-c^2} & c  \end{pmatrix}$$

It is easy to check that $A$ is orthogonal. Therefore we can continue transforming the coefficient as follows.

\begin{eqnarray*}
&=&\mathbb{E}_{s,t}\Big(\begin{pmatrix}-1&\frac{c}{\sqrt{1-c^2}}\end{pmatrix}Au\Big)\tau\Big(\begin{pmatrix}1&0\end{pmatrix}Au\Big)\tau'\Big(\begin{pmatrix}c&\sqrt{1-c^2}\end{pmatrix}Au\Big)\\
&=&\mathbb{E}_{s,t}\Big(\begin{pmatrix}0 & \frac{1}{\sqrt{1-c^2}}\end{pmatrix}u\Big)\tau\Big(\begin{pmatrix}c&-\sqrt{1-c^2}\end{pmatrix}u\Big)\tau'\Big(\begin{pmatrix}1 & 0\end{pmatrix}u\Big)\\
&=&\mathbb{E}_{s,t}\frac{1}{\sqrt{1-c^2}}t\tau(cs-\sqrt{1-c^2}t)\tau'(s)\\
&=&\frac{1}{\sqrt{1-c^2}}\mathbb{E}_s\big(\tau'(s)\mathbb{E}_t t\tau(cs-\sqrt{1-c^2}t)\big)
\end{eqnarray*}

We further have

\begin{eqnarray*}
&&\mathbb{E}_t\tau'(cs+\sqrt{1-c^2}t)\\
&=&\int_{t=-\infty}^\infty \tau'(cs-\sqrt{1-c^2}t)n(t)d\mu_1\\
&=&[-\frac{1}{\sqrt{1-c^2}}\tau(cs-\sqrt{1-c^2}t)n(t)]_{-\infty}^\infty\\
&& - \int_{t=-\infty}^\infty\Big(-\frac{1}{\sqrt{1-c^2}}\tau(cs-\sqrt{1-c^2}t)\Big)(-tn(t))d\mu_1\\
&=&-\frac{1}{\sqrt{1-c^2}}\mathbb{E}_t t\tau(cs-\sqrt{1-c^2}t)\\
\end{eqnarray*}

Here, we use integration by parts. This is allowed because $n(t)$ is absolutely continuous and $\tau'$ is integrable. Continuing the previous chain of equations, we have

\begin{eqnarray*}
&&\frac{1}{\sqrt{1-c^2}}\mathbb{E}_s\big(\tau'(s)\mathbb{E}_t t\tau(cs-\sqrt{1-c^2}t)\big)\\
&=&\frac{1}{\sqrt{1-c^2}}\mathbb{E}_s\big(\tau'(s)(-\sqrt{1-c^2}\mathbb{E}_t\tau'(cs+\sqrt{1-c^2}t))\big)\\
&=&-\mathbb{E}_{s,t}\tau'(s)\tau'(cs+\sqrt{1-c^2}t)
\end{eqnarray*}

Putting everything together, we obtain

\begin{eqnarray*}
&&\mathfrak{C}_{\tau'}(c)\\
&=&\mathbb{E}_{s,t}\tau'(s)\tau'(cs+\sqrt{1-c^2}t)\\
&=&-\mathbb{E}_{s,t}(-s+\frac{c}{\sqrt{1-c^2}}t)\tau(s)\tau'(cs+\sqrt{1-c^2}t)\\
&&-\lim_{\epsilon \rightarrow 0}\epsilon\mathbb{E}_{s,t}(P_0+P_1h(cs+\sqrt{1-c^2}t,\delta))\\
&=&-\lim_{\epsilon \rightarrow 0}\frac{1}{\epsilon}\mathbb{E}_{s,t}\Big(\epsilon(-s+\frac{c}{\sqrt{1-c^2}}t)\tau(s)\tau'(cs+\sqrt{1-c^2}t)\\
&&+\epsilon^2(P_0+P_1h(cs+\sqrt{1-c^2}t,\delta))\Big)\\
&=&-\lim_{\epsilon \rightarrow 0} \frac{\mathfrak{C}_\tau(c-\epsilon)-\mathfrak{C}_\tau(c)}{\epsilon}\\
&=&\mathfrak{C}_\tau'(c)
\end{eqnarray*}

as required. 

\underline{Stage 2}: Assume $\tau$ is thrice differentiable. Then $\mathfrak{C}_\tau$ is differentiable at 1 and $\mathfrak{C}_\tau'(1) = \mathfrak{C}_{\tau'}(1)$.

Let $1 \ge \epsilon > 0$ and let $\delta=-\epsilon s+\sqrt{\epsilon}\sqrt{2-\epsilon}t$. Then we have 

$$\delta=-\epsilon s+\sqrt{2}\epsilon^\frac{1}{2}t -\frac{1}{2\sqrt{2}}\epsilon^\frac{3}{2}t+r(\epsilon)\epsilon^\frac{5}{2}t$$

where $r(\epsilon)$ is derived from the Taylor expansion of the square root about 2. All values between $2-\epsilon$ and $2$ are contained in the interval $[1,2]$. Because the square root is twice differentiable in that interval with a bounded second derivative, the Lagrange form of the derivative yields that $|r(\epsilon)|$ is bounded. Let the bound be $R$. Then 

\begin{eqnarray*}
&&\mathfrak{C}_\tau(1-\epsilon)-\mathfrak{C}_\tau(1)\\\
&=&\mathbb{E}_{s,t}\tau(s)\Big(\tau((1-\epsilon)s+\sqrt{\epsilon}\sqrt{2-\epsilon}t)-\tau(s)\Big)\\
&=&\mathbb{E}_{s,t}\tau(s)\Big(\tau(s+\delta)-\tau(s)\Big)\\
&=&\mathbb{E}_{s,t}\tau(s)\Big(\tau(s)+\delta\tau'(s)+\frac{1}{2}\delta^2\tau''(s)+\frac{1}{6}\delta^3h(s,\delta)-\tau(s)\Big)\\
&=&\mathbb{E}_{s,t}\tau(s)\Big((-\epsilon s+\sqrt{2}\epsilon^\frac{1}{2}t -\frac{1}{2\sqrt{2}}\epsilon^\frac{3}{2}t+r(\epsilon)\epsilon^\frac{5}{2}t)\tau'(s)\\
&&+\frac{1}{2}(-\epsilon s+\sqrt{2}\epsilon^\frac{1}{2}t -\frac{1}{2\sqrt{2}}\epsilon^\frac{3}{2}t+r(\epsilon)\epsilon^\frac{5}{2}t)^2\tau''(s)\\
&&+\frac{1}{6}(-\epsilon s+\sqrt{2}\epsilon^\frac{1}{2}t -\frac{1}{2\sqrt{2}}\epsilon^\frac{3}{2}t+r(\epsilon)\epsilon^\frac{5}{2}t)^3h(s,\delta)\Big)\\
&=&\mathbb{E}_{s,t}\Big(\sqrt{2}\epsilon^\frac{1}{2}t\tau(s)\tau'(s)-\epsilon s\tau(s)\tau'(s)\\
&&+\epsilon t^2\tau(s)\tau''(s)+\epsilon^\frac{3}{2}(P_0+P_1h(s,\delta))\Big)\\
&=&\mathbb{E}_{s,t}\Big(\epsilon(-s\tau(s)\tau'(s)+\tau(s)\tau''(s))+\epsilon^\frac{3}{2}(P_0+P_1h(s,\delta))\Big)
\end{eqnarray*}

The last step uses $\mathbb{E}t=0$ and $\mathbb{E}t^2=1$.

Here, $h$ is derived from the Taylor expansion of $\tau$ about $s$. Because $\tau$ is thrice differentiable, the Lagrange form of the remainder yields $h(s,\delta)=\tau'''(s')$ for some $s'$ with $s \le s' \le s+\delta$. This in turn yields $|h(s,\delta)|\le T_3(|s|+|t|)$, where $T_3$ is defined as in section \ref{otherNTCsection}. 

Here, $P_0$ and $P_1$ are low-order polynomials of $s$, $t$, $\sqrt{\epsilon}$, $r$, $\tau(s)$, $\tau'(s)$ and $\tau''(s)$ terms. (Remember that $c$ is a constant.) 

First consider $\mathbb{E}_{s,t}(P_0+P_1h(s,\delta))$, the coefficient of $\epsilon^\frac{3}{2}$. Let $P_0^+$ and $P_1^+$ be the polynomials that are equal to $P_0$ and $P_1$, except each term has a positive sign. Then

\begin{eqnarray*}
&&\Big|\mathbb{E}_{s,t}\Big(P_0(s,t,\sqrt{\epsilon},r,\tau(s),\tau'(s),\tau''(s))+\\
&&P_1(s,t,\sqrt{\epsilon},r,\tau(s),\tau'(s),\tau''(s))h(s,\delta)\Big)\Big|\\
&\le&\mathbb{E}_{s,t}\Big(P_0^+(|s|,|t|,1,R,|\tau(s)|,|\tau'(s)|,|\tau''(s)|)\\
&&+P_1^+(|s|,|t|,1,R,|\tau(s)|,|\tau'(s)|,|\tau''(s)|)T_3(|s|+|t|)\Big)
\end{eqnarray*}

Again, by assumption \ref{assumptionIntegrableMeanField}, this bound is finite. So $\lim_{\epsilon\rightarrow 0}\sqrt{\epsilon}|\mathbb{E}_{s,t}(P_0+P_1h)| = 0$, so $\lim_{\epsilon\rightarrow 0}\sqrt{\epsilon}\mathbb{E}_{s,t}(P_0+P_1h) = 0$. 

Now consider $\mathbb{E}_{s,t}(-s\tau(s)\tau'(s)+\tau(s)\tau''(s))$, the coefficient of $\epsilon$. Using integration by parts, we have 

\begin{eqnarray*}
&&\mathbb{E}_s\tau'(s)^2\\
&=&\int_{s=-\infty}^\infty\tau'(s)\big(\tau'(s)n(s)\big)d\mu_1\\
&=&[\tau(s)\big(\tau'(s)n(s)\big)]_{-\infty}^\infty - \int_{s=-\infty}^\infty\tau(s)\big(\tau''(s)n(s)-\tau'(s)sn(s)\big)d\mu_1\\
&=&\mathbb{E}_ss\tau(s)\tau'(s)-\tau(s)\tau''(s)
\end{eqnarray*}

This is allowed because $\tau'$ is integrable and $\tau(s)n(s)$ is absolutely continuous. The latter follows from assumption \ref{assumptionIntegrableMeanField} applied to $(\tau(s)n(s))' = \tau''(s)n(s)-\tau'(s)sn(s)$. This yields that $(\tau(s)n(s))'$ converges to zero as $s \rightarrow \infty$ and $s \rightarrow -\infty$. Since $(\tau(s)n(s))'$ is also continuous, it is bounded. Hence, $\tau(s)n(s)$ is indeed absolutely continuous.

Putting everything together, we obtain

\begin{eqnarray*}
&&\mathfrak{C}_{\tau'}(1)\\
&=&\mathbb{E}_s\tau'(s)^2\\
&=&\mathbb{E}_s(s\tau(s)\tau'(s)-\tau(s)\tau''(s))-\lim_{\epsilon \rightarrow 0}\sqrt{\epsilon}\mathbb{E}_{s,t}(P_0+P_1h(s,\delta))\\
&=&-\lim_{\epsilon \rightarrow 0}\frac{1}{\epsilon}\mathbb{E}_{s,t}\Big(\epsilon(-s\tau(s)\tau'(s)+\tau(s)\tau''(s))+\epsilon^\frac{3}{2}(P_0+P_1h(s,\delta))\Big)\\
&=&-\lim_{\epsilon \rightarrow 0} \frac{\mathfrak{C}_\tau(1-\epsilon)-\mathfrak{C}_\tau(1)}{\epsilon}\\
&=&\mathfrak{C}_\tau'(1)
\end{eqnarray*}

as required.

\underline{Stage 3}: Assume $\tau$ is 5-differentiable. Then $\mathfrak{C}_\tau$ is thrice differentiable and $\mathfrak{C}_\tau' = \mathfrak{C}_{\tau'} \ge 0$, $\mathfrak{C}_\tau'' = \mathfrak{C}_{\tau''} \ge 0$ and $\mathfrak{C}_\tau''' = \mathfrak{C}_{\tau'''} \ge 0$.

Stages 1 and 2 together yield $\mathfrak{C}'_\tau = \mathfrak{C}_{\tau'}$ for all $c$. Since both $\tau'$ and $\tau''$ are thrice differentiable, we can apply those stages to these functions to obtain $ \mathfrak{C}_{\tau''} = \mathfrak{C}'_{\tau'} = (\mathfrak{C}_{\tau'})' = (\mathfrak{C}'_{\tau})'=\mathfrak{C}''_{\tau}$ and $\mathfrak{C}_{\tau'''} = \mathfrak{C}'_{\tau''} = (\mathfrak{C}_{\tau''})' = (\mathfrak{C}''_{\tau})'=\mathfrak{C}'''_{\tau}$ as required. Applying proposition \ref{covkerPositive2} to $\mathfrak{C}_{\tau'}$, $\mathfrak{C}_{\tau''}$ and $\mathfrak{C}_{\tau''}$, we obtain that they are non-negative.

\underline{Stage 4}: (a) and (b) hold under the conditions of the theorem.

Throughout the remainder of this proof, we will continue to use the $\tau'$ notation even though $\tau$ is not differentiable everywhere. The following rules apply:

\begin{itemize}
\item In the context of integration, $\tau'$ may be chosen to refer to either the left or right derivative at any point where $\tau$ is not differentiable. Since expressions involving either or both the left and right derivative are assumed to be integrable by assumption \ref{assumptionIntegrableMeanField}, and valid Lebesgue integrals are not affected when altering the value of the integrand at a finite number of points, this choice is immaterial.
\item In the context of bounding $\tau'$, any bound has to apply to both the left and right derivative.
\end{itemize}

\sloppy If $\tau$ is piecewise 5-differentiable with a single piece encompassing $\mathbb{R}$, stage 4 follows directly from stage 3. So now assume there are two or more pieces. Denote the boundaries of the intervals corresponding to the pieces by $s_1, .., s_D$ where $D \ge 1$. These intervals are then $(\infty, s_1], [s_1, s_2], .., [s_{D-1}, s_D], [s_D, \infty)$ and there are $D+1$ of them. Let $(\epsilon)_n$, $n\ge 0$ be a positive decreasing sequence that converges to zero, where also $\epsilon_0 < \frac{1}{2}(s_{d+1}-s_d)$ for all $1\le d \le D-1$. Then for any $n$ the intervals $[s_d-\epsilon_n,s_d+\epsilon_n]$, $1\le d \le D$ are all disjoint. Denote the union of those intervals by $I_n$. Then each of the $I_n$ is contained in $I_0$. Since $\tau$ is continuous and $I_0$ is bounded, there is a bound $V$ for $|\tau|$ across $I_0$. Since $\tau$ is continuously differentiable on each of $[s_d-\epsilon_0,s_d]$, $1\le d \le D$ and $[s_d,s_d+\epsilon_0]$, $1\le d \le D$, $\tau'$ is bounded on each of those intervals. Since there are a finite number of such intervals, we can let some $G$ be an upper bound for $|\tau'|$ across $I_0$.

For each $n$, we want to choose an activation function $\tau_n$ that is equal to $\tau$ everywhere except on $I_n$ and 5-differentiable everywhere. We can choose such a $\tau_n$ by simply matching the function value and first five derivatives of $\tau$ at the boundary points of $I_n$. Because $|\tau|$ is bounded by $V$ on $I_n$, we can choose a $\tau_n$ such that $|\tau_n|$ is bounded by, say, $2V$ on $I_n$. Since $|\tau'|$ is bounded by $G$ on the $[s_d-\epsilon_n,s_d]$ and the $[s_d,s_d+\epsilon_n]$, by the intermediate value principle, we have that $|\tau(s_d)-\tau(s_d-\epsilon_n)| \le G\epsilon_n$ and $|\tau(s_d+\epsilon_n)-\tau(s_d)| \le G\epsilon_n$, so the absolute value of the slope of the line segment from $(s_d-\epsilon_n,\tau(s_d-\epsilon_n))$ to $(s_d+\epsilon_n,\tau(s_d+\epsilon_n))$ is at most $G$. Therefore we can choose $\tau_n$ so that $|\tau_n'|$ is bounded by, say, $2G$ on $I_n$. Bounding $\tau_n$ also means that it is integrable with respect to any Gaussian measure.

Let $\mu_n=\int_{s\in I_n}n(s)d\mu_1$. Then we have $\mu_n<2D\epsilon_n$. We have

\begin{eqnarray*}
&&|\mathfrak{C}_\tau(c) - \mathfrak{C}_{\tau_n}(c)|\\
&=&\Big|\mathbb{E}_{s,t}\Big(\tau(s)\tau(cs+\sqrt{1-c^2}t)-\tau_n(s)\tau_n(cs+\sqrt{1-c^2}t)\Big)\Big|\\
&\le&\mathbb{E}_{s,t}\Big|\tau(s)\tau(cs+\sqrt{1-c^2}t)-\tau_n(s)\tau_n(cs+\sqrt{1-c^2}t)\Big|\\
&\le&\mathbb{E}_{s,t}\Big|\tau(s)\tau(cs+\sqrt{1-c^2}t)-\tau_n(s)\tau(cs+\sqrt{1-c^2}t)\Big|\\
&&+\mathbb{E}_{s,t}\Big|\tau_n(s)\tau(cs+\sqrt{1-c^2}t)-\tau_n(s)\tau_n(cs+\sqrt{1-c^2}t)\Big|\\
&=&\mathbb{E}_{s,t}|\tau(cs+\sqrt{1-c^2}t)||\tau(s)-\tau_n(s)|\\
&&+\mathbb{E}_{s,t}|\tau_n(s)||\tau(cs+\sqrt{1-c^2}t)-\tau_n(cs+\sqrt{1-c^2}t)|\\
&\le&\mathbb{E}_{s,t}|\tau(cs+\sqrt{1-c^2}t)||\tau(s)-\tau_n(s)|\\
&&+\mathbb{E}_{s,t}(|\tau(s)|+|\tau(s)-\tau_n(s)|)|\tau(cs+\sqrt{1-c^2}t)-\tau_n(cs+\sqrt{1-c^2}t)|\\
&=&\mathbb{E}_{s,t}\mathbbm{1}_{s \in I_n}|\tau(cs+\sqrt{1-c^2}t)||\tau(s)-\tau_n(s)|\\
&&+\mathbb{E}_{s,t}\mathbbm{1}_{cs+\sqrt{1-c^2}t \in I_n}|\tau(s)||\tau(cs+\sqrt{1-c^2}t)-\tau_n(cs+\sqrt{1-c^2}t)| \\
&&+\mathbb{E}_{s,t} \mathbbm{1}_{s\in I_n \text{ and } cs+\sqrt{1-c^2}t \in I_n}|\tau(s)-\tau_n(s)||\tau(cs+\sqrt{1-c^2}t)-\tau_n(cs+\sqrt{1-c^2}t)|\\
&\le&\mathbb{E}_{s,t}\mathbbm{1}_{s \in I_n}|\tau(cs+\sqrt{1-c^2}t)|(|\tau(s)|+|\tau_n(s)|)\\
&&+\mathbb{E}_{s,t}\mathbbm{1}_{cs+\sqrt{1-c^2}t \in I_n}|\tau(s)|(|\tau(cs+\sqrt{1-c^2}t)|+|\tau_n(cs+\sqrt{1-c^2}t)|) \\
&&+\text{\scalebox{0.98}{$\mathbb{E}_{s,t} \mathbbm{1}_{s\in I_n \text{ and } cs+\sqrt{1-c^2}t \in I_n}(|\tau(s)|+|\tau_n(s)|)(|\tau(cs+\sqrt{1-c^2}t)|+|\tau_n(cs+\sqrt{1-c^2}t)|)$}}\\
&\le&3V\mathbb{E}_{s,t}\mathbbm{1}_{s \in I_n}|\tau(cs+\sqrt{1-c^2}t)|+3V\mathbb{E}_{s,t}\mathbbm{1}_{cs+\sqrt{1-c^2}t \in I_n}|\tau(s)|+9V^2\mathbb{E}_{s,t} \mathbbm{1}_{s\in I_n}
\end{eqnarray*}

Let $u$ be a 2-dimensional unit Gaussian column vector. Because the distribution of $u$ is spherically symmetric, it is invariant under multiplication with an orthogonal matrix. Let 

$$A=\begin{pmatrix} c & -\sqrt{1-c^2} \\ \sqrt{1-c^2} & c  \end{pmatrix}$$

It is easy to check that $A$ is orthogonal. Then using vector notation we obtain

\begin{eqnarray*}
&&\mathbb{E}_{s,t}\mathbbm{1}_{cs+\sqrt{1-c^2}t\in I_n}|\tau(s)|\\
&=&\mathbb{E}_u\mathbbm{1}_{\begin{pmatrix}c &\sqrt{1-c^2}\end{pmatrix}u\in I_n}|\tau(\begin{pmatrix}1 &0\end{pmatrix}u)|\\
&=&\mathbb{E}_u\mathbbm{1}_{\begin{pmatrix}c &\sqrt{1-c^2}\end{pmatrix}Au\in I_n}|\tau(\begin{pmatrix}1 &0\end{pmatrix}Au)|\\
&=&\mathbb{E}_u\mathbbm{1}_{\begin{pmatrix}1 & 0\end{pmatrix}u\in I_n}|\tau(\begin{pmatrix}c &-\sqrt{1-c^2}\end{pmatrix}u)|\\
&=&\mathbb{E}_{s,t}\mathbbm{1}_{s \in I_n}|\tau(cs-\sqrt{1-c^2}t)|\\
&=&\mathbb{E}_{s,t}\mathbbm{1}_{s \in I_n}|\tau(cs+\sqrt{1-c^2}t)|
\end{eqnarray*}

The last step uses that the distribution of $t$ is symmetric about zero. Continuing the long chain of equations from above we have

\begin{eqnarray*}
&&3V\mathbb{E}_{s,t}\mathbbm{1}_{s \in I_n}|\tau(cs+\sqrt{1-c^2}t)|+3V\mathbb{E}_{s,t}\mathbbm{1}_{cs+\sqrt{1-c^2}t \in I_n}|\tau(s)|+9V^2\mathbb{E}_{s,t} \mathbbm{1}_{s\in I_n}\\
&=&6V\mathbb{E}_{s,t}\mathbbm{1}_{s \in I_n}|\tau(cs+\sqrt{1-c^2}t)|+ 9V^2\mu_n\\
&=&6V\int_{s\in I_n}\Big(\mathbb{E}_t|\tau(cs+\sqrt{1-c^2}t)|\Big)n(s)d\mu_1 + 9V^2\mu_n
\end{eqnarray*}

We have $\mathbb{E}_t|\tau(cs+\sqrt{1-c^2}t)|=\mathbb{E}_z|\tau(z)|$, where $z$ has mean $cs$ and standard deviation $\sqrt{1-c^2}$. So the standard deviation of $z$ is bounded by $1$. Further, $|cs|$ is bounded by $\max(|s_1-\epsilon_0|,|s_D+\epsilon_0|)$. Further, $|\tau|$ is continuous, so it is bounded on any bounded interval. So by lemma \ref{lemma10}, we have that $\mathbb{E}_z|\tau(z)|$ is bounded across $c$ and $s$. Let that bound be $B$. Then

$$6V\int_{s\in I_n}\Big(\mathbb{E}_t|\tau(cs+\sqrt{1-c^2}t)|\Big)n(s)d\mu_1 + 9V^2\mu_n \le 6VB\mu_n + 9V^2\mu_n \le 12VBD\epsilon_n + 18V^2D\epsilon_n$$

So in summary, we have $|\mathfrak{C}_\tau(c) - \mathfrak{C}_{\tau_n}(c)| \le 12VBD\epsilon_n + 18V^2D\epsilon_n$. Analogously, we obtain $|\mathfrak{C}_{\tau'}(c) - \mathfrak{C}_{\tau_n'}(c)| \le 12GBD\epsilon_n + 18G^2D\epsilon_n$. The only difference in the derivation is that $|\tau'|$ is not continuous. However, $|\tau'|$ is continuous on each of a finite set of pieces, so $|\tau'|$ is still bounded on any bounded interval via the maximum of the piecewise bounds. So lemma \ref{lemma10} still applies.

Both $12VBD\epsilon_n + 18V^2D\epsilon_n$ and $12GBD\epsilon_n + 18G^2D\epsilon_n$ are independent of $c$, so $\mathfrak{C}_{\tau_n}$ converges uniformly to $\mathfrak{C}_\tau$ and $\mathfrak{C}_{\tau_n'}$ converges uniformly to $\mathfrak{C}_{\tau'}$. Since also $\mathfrak{C}'_{\tau_n} = \mathfrak{C}_{\tau_n'}$ by stage 4, we have $\mathfrak{C}'_\tau = \mathfrak{C}_{\tau'}$. This is (a). 

By stage 3, all the $\mathfrak{C}_{\tau_n}$ have the properties in (b). It is easy to check that these properties are also preserved under uniform function convergence.

\underline{Stage 5}: (c) holds under the conditions of the theorem.

Using lemma \ref{lemma11}, we have

\begin{eqnarray*}
&&\mathfrak{C}_\tau(c)\\
&=&\mathbb{E}_{s,t}\tau(s)\tau(cs+\sqrt{1-c^2}t)\\
&=&\mathbb{E}_{s,t}(\tilde{\tau}(s)+a_\tau s+b_\tau )(\tilde{\tau}(cs+\sqrt{1-c^2}t)+a_\tau (cs+\sqrt{1-c^2}t)+b_\tau )\\
&=&\mathbb{E}_{s,t}\Big(\tilde{\tau}(s)\tilde{\tau}(cs+\sqrt{1-c^2}t)+\tilde{\tau}(s)a_\tau (cs+\sqrt{1-c^2}t)\\
&&+\tilde{\tau}(s)b_\tau +a_\tau s\tilde{\tau}(cs+\sqrt{1-c^2}t)+a_\tau^2s(cs+\sqrt{1-c^2}t)+a_\tau b_\tau s\\
&&+b_\tau \tilde{\tau}(cs+\sqrt{1-c^2}t)+a_\tau b_\tau (cs+\sqrt{1-c^2}t)+b_\tau^2\Big)\\
&=&\Big(\mathbb{E}_{s,t}\tilde{\tau}(s)\tilde{\tau}(cs+\sqrt{1-c^2}t)+a_\tau c\mathbb{E}_{s}s\tilde{\tau}(s)+a_\tau \sqrt{1-c^2}\mathbb{E}_{s}\tilde{\tau}(s)\mathbb{E}_{t}t\\
&&+b_\tau \mathbb{E}_s\tilde{\tau}(s)+a_\tau \mathbb{E}_{s,t}s\tilde{\tau}(cs+\sqrt{1-c^2}t)+a_\tau^2c\mathbb{E}_ss^2+a_\tau^2\sqrt{1-c^2}\mathbb{E}_ss\mathbb{E}_tt+a_\tau b_\tau \mathbb{E}_ss\\
&&+b_\tau \mathbb{E}_{s,t}\tilde{\tau}(cs+\sqrt{1-c^2}t)+a_\tau b_\tau c\mathbb{E}_ss+a_\tau b_\tau \sqrt{1-c^2}\mathbb{E}_tt+b_\tau^2\Big)\\
&=&\Big(\mathbb{E}_{s,t}\tilde{\tau}(s)\tilde{\tau}(cs+\sqrt{1-c^2}t)+a_\tau \mathbb{E}_{s,t}s\tilde{\tau}(cs+\sqrt{1-c^2}t)\\
&&+a_\tau^2c+b_\tau \mathbb{E}_{s,t}\tilde{\tau}(cs+\sqrt{1-c^2}t)+b_\tau^2\Big)
\end{eqnarray*}

$cs+\sqrt{1-c^2}t$ is unit Gaussian just as $s$, so $\mathbb{E}_{s,t}\tilde{\tau}(cs+\sqrt{1-c^2}t)=\mathbb{E}_s\tilde{\tau}(s)=0$.

Let $u$ be a 2-dimensional unit Gaussian column vector. Because the distribution of $u$ is spherically symmetric, it is invariant under multiplication with an orthogonal matrix. Let 

$$A=\begin{pmatrix} c & -\sqrt{1-c^2} \\ \sqrt{1-c^2} & c  \end{pmatrix}$$

It is easy to check that $A$ is orthogonal. Using vector notation, we have

\begin{eqnarray*}
&&\mathbb{E}_{s,t}s\tilde{\tau}(cs+\sqrt{1-c^2}t)\\
&=&\mathbb{E}_u\Big(\begin{pmatrix}1 & 0\end{pmatrix}u\Big)\tau\Big(\begin{pmatrix}c & \sqrt{1-c^2}\end{pmatrix}u\Big)\\
&=&\mathbb{E}_u\Big(\begin{pmatrix}1 & 0\end{pmatrix}Au\Big)\tau\Big(\begin{pmatrix}c & \sqrt{1-c^2}\end{pmatrix}Au\Big)\\
&=&\mathbb{E}_u\Big(\begin{pmatrix}c & -\sqrt{1-c^2}\end{pmatrix}u\Big)\tau\Big(\begin{pmatrix}1 & 0\end{pmatrix}u\Big)\\
&=&\mathbb{E}_{s,t}(cs-\sqrt{1-c^2}t)\tilde{\tau}(s)\\
&=&c\mathbb{E}_ss\tilde{\tau}(s) - \sqrt{1-c^2}\mathbb{E}_s\tilde{\tau}(s)\mathbb{E}_tt\\
&=&0
\end{eqnarray*}

So we obtain
 
\begin{eqnarray*}
&&\mathfrak{C}_\tau(c)\\
&=&\mathbb{E}_{s,t}\tilde{\tau}(s)\tilde{\tau}(cs+\sqrt{1-c^2}t)+a_\tau \mathbb{E}_{s,t}s\tilde{\tau}(cs+\sqrt{1-c^2}t)\\
&&+a_\tau^2c+b_\tau \mathbb{E}_{s,t}\tilde{\tau}(cs+\sqrt{1-c^2}t)+b_\tau^2\\
&=&\mathbb{E}_{s,t}\tilde{\tau}(s)\tilde{\tau}(cs+\sqrt{1-c^2}t)+a_\tau^2c+b_\tau^2\\
&=&\mathfrak{C}_{\tilde{\tau}}(c)+a_\tau^2c+b_\tau^2
\end{eqnarray*}

This is (i). Throughout the rest of this stage, we will repeatedly use lemma \ref{lemma11}. We have

\begin{eqnarray*}
&&\mathfrak{C}_{\tilde{\tau}}(0)\\
&=&\mathbb{E}_{s,t}\tilde{\tau}(s)\tilde{\tau}(t)\\
&=&(\mathbb{E}_s\tilde{\tau}(s))^2\\
&=&0
\end{eqnarray*}

This is (ii). 

\begin{eqnarray*}
&&\mathfrak{C}_\tau(0)\\
&=&\mathfrak{C}_{\tilde{\tau}}(0)+0a_\tau^2+b_\tau^2\\
&=&b_\tau^2
\end{eqnarray*}

This is (iii).

\begin{eqnarray*}
&&\mathfrak{C}_{\tilde{\tau}}'(0)\\
&=&\mathfrak{C}_{\tilde{\tau}'}(0)\\
&=&\mathbb{E}_{s,t}\tilde{\tau}'(s)\tilde{\tau}'(t)\\
&=&(\mathbb{E}_s\tilde{\tau}'(s))^2\\
&=&\Big(\int_{\mathbb{R}} \tilde{\tau}'(s)n(s)d\mu_1\Big)^2\\
&=&\Big([\tilde{\tau}(s)n(s)]_{-\infty}^\infty - \int_{\mathbb{R}} \tilde{\tau}(s)(-sn(s))d\mu_1\Big)^2\\
&=&\Big(\int_{\mathbb{R}} s\tilde{\tau}(s)n(s)d\mu_1\Big)^2\\
&=&(\mathbb{E}_ss\tilde{\tau}(s))^2\\
&=&0
\end{eqnarray*}

Here, we use integration by parts, which is allowed because $\tau'$, and hence $\tilde{\tau}'$, is integrable and $n(s)$ is uniformly continuous. We also use that $\tilde{\tau}$ is piecewise 5-differentiable since $\tau$ is. Thus, we obtain (iv).

\begin{eqnarray*}
&&\mathfrak{C}'_\tau(0)\\
&=&\mathfrak{C}_{\tau'}(0)\\
&=&(\mathbb{E}_s\tau'(s))^2\\
&=&(\mathbb{E}_ss\tau(s))^2\\
&=&a_\tau^2
\end{eqnarray*}

This is (v). We use integration by parts as above.

\begin{eqnarray*}
&&CAR(\tau, \mathcal{N}(0,1))\\
&=&\frac{(b_\tau +a_\tau\mathbb{E}_ss)^2}{\mathbb{E}_s\tau(s)^2}\\
&=&\frac{b_\tau^2}{\mathbb{E}_s\tau(s)^2}\\
&=&\frac{\mathfrak{C}_{\tau}(0)}{\mathfrak{C}_{\tau}(1)}
\end{eqnarray*}

This is (vi).

\begin{eqnarray*}
&&LAR(\tau, \mathcal{N}(0,1))\\
&=&\frac{\mathbb{E}_s((s-\mathbb{E}_ss)a_\tau)^2}{\mathbb{E}_s\tau(s)^2}\\
&=&\frac{a_\tau^2\mathbb{E}_ss^2}{\mathbb{E}_s\tau(s)^2}\\
&=&\frac{\mathfrak{C}_{\tau}'(0)}{\mathfrak{C}_{\tau}(1)}
\end{eqnarray*}

This is (vii). If $\tau$ is linear, then $\tilde{\tau}$ is the zero function and so $\mathfrak{C}_{\tilde{\tau}}$ is the zero function. Hence, $\mathfrak{C}_\tau(c) = a_\tau^2c+b_\tau^2$, which is linear. Conversely, if $\mathfrak{C}_\tau$ is linear, then so is $\mathfrak{C}_{\tilde{\tau}}$. Since $\mathfrak{C}_{\tilde{\tau}}(0) = \mathfrak{C}_{\tilde{\tau}}'(0) = 0$, $\mathfrak{C}_{\tilde{\tau}}$ is the zero function, and so $\mathfrak{C}_{\tilde{\tau}}(1) = \mathbb{E}\tilde{\tau}(s)^2 = 0$. Since $\tilde{\tau}(s)n(s)$ is continuous, by the contrapositive of lemma \ref{lemma8}, $\tilde{\tau}$ is the zero function, and so $\tau$ is linear. This yields (viii), which completes the stage.

\underline{Stage 6}: The second part of the theorem holds.

We have $\mathfrak{C}_\tau(1) = \mathbb{E}_s\tau(s)^2$. Because $\tau$ is not linear, it is not zero everywhere. Hence we can apply lemma \ref{lemma8} to $\tau(s)^2n(s)$ to obtain $\mathfrak{C}_\tau(1) > 0$. Hence, $\tilde{\mathfrak{C}}_\tau$ is valid. Also, theorem statement (b) clearly holds for $\tilde{\mathfrak{C}}_\tau$ and theorem statement (a) implies that $\tilde{\mathfrak{C}}_\tau$ is differentiable.

It is obvious that no two of the three convergence cases can hold simultaneously. We will now show that at least one of them holds. Clearly, 1 is a fixed point of $\tilde{\mathfrak{C}}_\tau$. We will now consider three distinct possibilities based on the number of other fixed points.

Possibility 1: $\tilde{\mathfrak{C}}_\tau(c)$ has no fixed points apart from $c=1$.

Because $\tilde{\mathfrak{C}}_\tau$ is increasing, we have $\tilde{\mathfrak{C}}'_\tau(1) \ge 0$. Because $\tilde{\mathfrak{C}}_\tau$ is convex and increasing, $\tilde{\mathfrak{C}}_\tau'(1) = 0$ implies that $\tilde{\mathfrak{C}}_\tau$ is constant. This would contradict proposition \ref{covkerPositive} since $\tau$ is non-constant. Hence, $\tilde{\mathfrak{C}}'_\tau(1) > 0$.

If $1 > \tilde{\mathfrak{C}}'_\tau(1) > 0$, convergence case 1 holds. If $\tilde{\mathfrak{C}}'_\tau(1) = 1$, convergence case 2 holds. Finally, assume that $\tilde{\mathfrak{C}}'_\tau(1) > 1$. Then there is a $c$ such that $\tilde{\mathfrak{C}}_\tau(c) < c$. But we also have that $\tilde{\mathfrak{C}}_\tau(0) \ge 0$ by proposition \ref{covkerPositive2}, that $\tilde{\mathfrak{C}}_\tau$ is continuous and that there is no fixed point besides 1. Contradiction.

Possibility 2: $\tilde{\mathfrak{C}}_\tau(c)$ has exactly one fixed point apart from $c=1$.

Let this fixed point be $c^\text{lim} < 1$. Because $\tilde{\mathfrak{C}}_\tau$ is continuous, it is either above or below the diagonal in $[c^\text{lim}, 1]$. It cannot lie above the diagonal, else the chord from $\tilde{\mathfrak{C}}_\tau(c^\text{lim})$ to $\tilde{\mathfrak{C}}_\tau(1)$ lies below $\tilde{\mathfrak{C}}_\tau$. So $\tilde{\mathfrak{C}}_\tau$ lies below the diagonal. Let $1 > c_2 > c^\text{lim}$. Then $\frac{\tilde{\mathfrak{C}}_\tau(1) - \tilde{\mathfrak{C}}_\tau(c_2)}{1 - c_2} > 1$ and $\frac{\tilde{\mathfrak{C}}_\tau(c_2) - \tilde{\mathfrak{C}}_\tau(c^\text{lim})}{c_2 - c^\text{lim}} < 1$. By the intermediate value principle and because $\tilde{\mathfrak{C}}_\tau$ is convex, we have $\tilde{\mathfrak{C}}_\tau'(1) > 1$ and $\tilde{\mathfrak{C}}_\tau'(c^\text{lim}) < 1$. Because $\tilde{\mathfrak{C}}_\tau$ is increasing, $\tilde{\mathfrak{C}}_\tau'(c^\text{lim}) \ge 0$, so convergence case 3 holds.

Possibility 3: $\tilde{\mathfrak{C}}_\tau(c)$ has more than one fixed point apart from $c=1$.

Let $c_1$ be the infimum of all fixed points. Because $\tilde{\mathfrak{C}}_\tau$ is continuous, $c_1$ is itself a fixed point. Under possibility 3, there exists an additional fixed point $c_2$ such that $1 > c_2 > c_1$. Consider some $c_2 > c_3 > c_1$. Because the chord from $\tilde{\mathfrak{C}}_\tau(c_1)$ to $\tilde{\mathfrak{C}}_\tau(c_2)$ cannot lie below $\tilde{\mathfrak{C}}_\tau(c_3)$, we must have $\tilde{\mathfrak{C}}_\tau(c_3) \le c_3$. Because the chord from $\tilde{\mathfrak{C}}_\tau(c_3)$ to $\tilde{\mathfrak{C}}_\tau(1)$ cannot lie below $\tilde{\mathfrak{C}}_\tau(c_2)$, we must have $\tilde{\mathfrak{C}}_\tau(c_3) \ge c_3$. Hence, $\tilde{\mathfrak{C}}_\tau(c_3) = c_3$. Now consider some $c_3 > c_2$. Because the chord from $\tilde{\mathfrak{C}}_\tau(c_2)$ to $\tilde{\mathfrak{C}}_\tau(1)$ cannot lie below $\tilde{\mathfrak{C}}_\tau(c_3)$, we must have $\tilde{\mathfrak{C}}_\tau(c_3) \le c_3$. Because the chord from $\tilde{\mathfrak{C}}_\tau(c_1)$ to $\tilde{\mathfrak{C}}_\tau(c_3)$ cannot lie below $\tilde{\mathfrak{C}}_\tau(c_2)$, we must have $\tilde{\mathfrak{C}}_\tau(c_3) \ge c_3$. Hence, $\tilde{\mathfrak{C}}_\tau(c_3) = c_3$. In summary, we have $\tilde{\mathfrak{C}}_\tau(c) = c$ for all $c \ge c_1$. Assume $c_1 = 0$. Then $\tilde{\mathfrak{C}}_\tau(c) = c$ for all $c$ and so $\mathfrak{C}$ is linear. So by stage 5, $\tau$ is linear. Contradiction. So $c_1 > 0$. Let $\epsilon < \min(\frac{1}{2}(1-c_1), c_1)$ and let $c_4 = c_1 - \epsilon$. It is easy to check that $c_4$, $c_4+\epsilon$, $c_4+2\epsilon$ and $c_4+3\epsilon$ are in $[0,1]$. Since the chord from $\tilde{\mathfrak{C}}_\tau(c_4)$ to $\tilde{\mathfrak{C}}_\tau(1)$ cannot lie below $\tilde{\mathfrak{C}}_\tau(c_1)$, we have $\tilde{\mathfrak{C}}_\tau(c_4) \ge c_4$. Because $c_4$ is not a fixed point, we further have $\tilde{\mathfrak{C}}_\tau(c_4) > c_4$. But then $\tilde{\mathfrak{C}}_\tau(c_4+3\epsilon) - 3\tilde{\mathfrak{C}}_\tau(c_4+2\epsilon) + 3\tilde{\mathfrak{C}}_\tau(c_4+\epsilon) - \tilde{\mathfrak{C}}_\tau(c_4) = c_4 - \tilde{\mathfrak{C}}_\tau(c_4) < 0$. Contradiction. So possibility 3 is invalid. This completes the proof.

\end{proof}

\section{Mean field theory of batch normalization} \label{meanFieldBNsection}

There are two theoretical results covered in this chapter that do not straightforwardly extend to the case where an A-architecture contains batch normalization layers: theorem \ref{mfntPropagation} and proposition \ref{mfntCNLC}. As throughout most of this work, as discussed in section \ref{batchWiseLayersSection}, we relegate the BN case to its own section so that we can present the BN-free case without the additional conceptual and notational complexity that BN brings with it. As usual, we conceptualize an architecture with BN as a function that takes not an individual input, but a batch of inputs in addition to a parameter. We formalize our notation, terminology and conventions for this situation in subsection \ref{bnNotationSection}. In subsection \ref{bnLemmaSection}, we give a single lemma that is used to model the limit of expectation operators as the batch size converges to infinity. Finally, we give and prove generalizations of theorem \ref{mfntPropagation} and proposition \ref{mfntCNLC} to the BN case in subsections \ref{mfntPropagationBNsection} and \ref{mfntCNLCbnsection} respectively.

Both results rely on taking the limit of batch size to infinity. To our knowledge, this is a novel approach in mean field theory. For layer normalization, taking the width of layers to infinity means that the moments used for normalization are taken over an infinite sample. This allows us to replace these random moments with their almost sure limit, which yields the very simple calculation rules in table \ref{tableNLCPropagation}. For batch normalization, the strategy is analogous. When taking the batch size to infinity, the random moments taken across the batch can also be replaced by their almost sure limit, which yields equally simple calculation rules. Conditional on these limits, the propagation of individual inputs and their gradients becomes effectively independent of the batch. Considering this similarity, the form of theorem \ref{mfntPropagation} for the BN case is ultimately analogous to the LN case, except that three limit operators are present instead of two.

The generalization of proposition \ref{mfntCNLC} is less intuitive. Table \ref{tableNLCPropagation} implies that $\frac{d}{dq'}\mathfrak{C}(q,q')|_{q'=q}$, the derivative of the covariance kernel used in section \ref{networkCovKerSection}, is zero if there is no directed path from input to output layer in the layer graph not containing a BN layer. To explain this, we must consider the elem-like$(q,c)$ condition on the input distribution. It implies that the $q$ and $c$ parameter affect not only the square mean and co-mean of one or two inputs, but of almost all inputs. Varying the $c$ parameter changes the co-mean of $f(x)$ and $f(x')$ as the co-mean of $x$ and $x'$ varies, and as the co-mean of $x$ and $x'$ with other inputs varies as well. Varying the co-mean with other inputs is immaterial if the architecture does not contain BN as those other inputs do not affect the value of $f(x)$ and $f(x')$. However, if the architecture does contain BN, varying $c$ effectively varies not just the co-mean of $x$ and $x'$, but the function $f$ itself via its dependence on the batch. In order to eliminate this side effect, we need to allow the co-mean of $x$ and $x'$ to differ from all other co-means. We call this ``special'' co-mean the $r$ parameter. Based on this insight, we prove an extension to theorem \ref{mfntPropagation} that establishes calculation rules for the mean field limit of the co-mean of $f_l(x)$ and $f_l(x')$, which we term $\mathfrak{r}_l$. Finally, we replace $\frac{d}{dq'}\mathfrak{C}(q,q')|_{q'=q}$ by a covariance kernel derivative that represents $\frac{d\mathfrak{r}_L}{dr}|_{r=q}$, and we obtain the desired result.

\subsection{Additional notation, terminology and conventions} \label{bnNotationSection}

The following notation, terminology and conventions (NTCs) apply throughout the remainder of this section, and hence throughout the remainder of this chapter. We say a `batch input' or simply `batch' $X$ is a tuple $(x^{(1)}, .., x^{(B)})$ of `individual inputs' or simply `inputs', which are vectors of dimensionality $d_\text{in}$. $B$ denotes batch size. An A-architecture $f$ is a function that given a parameter $\theta$ and batch input $X$ returns a `batch output' $f(X)$. We say it is composed of `big layers' $f_l$, $0 \le l \le L$ that form a `big layer graph'. A big layer $f_l$ is composed of `small layers' $f_{l,b}$, $1\le b \le B$. We refer to $l$ as the `big layer index' and $b$ as the `batch index'. When propagating $X$ forward, a small layer represents the value obtained for an individual input and a big layer represents the value obtained jointly for all individual inputs. We use the terms `FC layer', `BN layer', etc. to refer to both big and small layers. Sometimes we will also just use the term `layer' when we do not need to be precise or big vs small is clear from context. A BN big layer $f_l = (f_{l,1}, .., f_{l,B})$ takes in all values in $(f_{k,1}, .., f_{k,B})$ and applies the BN operation. All other big layers apply their operation batch index-wise. The small layers form a small layer graph. All small layers except BN small layers are only dependent on other small layers with the same batch index, whereas a BN small layer $f_{l,b}$ is dependent on $B$ small layers with varying batch index but only a single big layer index $k$. Quantities associated with big layers, such as weight matrices $W_l$, bias vectors $\beta_l$, regularization parameters $\epsilon_l$, activation functions $\tau_l$ and addition weights $w_{l,\kappa_l}$, are shared across batch indices.

We further overload our notation by letting $X$ refer to the vector that is obtained by concatenating the individual inputs in $X$ and letting $f_l$ refer to the function that is obtained by concatenating the outputs of the $f_{l,b}$. Based on this, we write $\mathcal{J}_{l,m} = \frac{df_l}{df_m}$ for the derivative of one big layer with respect to another, which is a matrix of size $Bd_l^\text{MF} \times Bd_m^\text{MF}$. We write $\mathcal{J}_{l,m,b} = \frac{df_{l,b}}{df_m}$ for the derivative of a small layer with respect to a big layer.

The properties of A-architectures given in section \ref{aArchitectureSection} generalize to the BN case by (i) having terms such as $d_0$ and $d_ld_\text{MF}$ refer to the width of small layers and (ii) considering that $f$ has a single input and output big layer but $B$ input and output small layers and (iii) replacing the phrase ``layer graph'' with ``big layer graph''. For example, an A-architecture may contain two consecutive BN big layers even though this would violate property \ref{aprop10} if we considered the small layer graph instead. 

Since mean field architectures have no BN layers and are naturally capable of having multiple input and output layers, we do not have to change our definition of them to accommodate the BN case. We will only tweak our notation slightly as will become clear in the proof of section \ref{mfntPropagationBNsection}.

Given these definitions, all other NTCs used throughout the remainder of this chapter are either identical to or canonical generalizations of the corresponding concepts from earlier in this chapter.

\subsection{Lemma} \label{bnLemmaSection}

\begin{lemma} \label{lemma18}
Let ...

\begin{itemize}
\item ... $G$ and $H$ be functions $\mathbb{R}^{d_G} \rightarrow \mathbb{R}$.
\item ... $(\chi)_d$ be a sequence of $d_G$-dimensional random vectors where $\chi_d \sim\mathcal{N}(0, \Sigma_d)$ for some $\Sigma_d$.
\item ... $\hat{\chi}$ be a $d_G$-dimensional random vector where $\hat{\chi} \sim \mathcal{N}(0, \hat{\Sigma})$
\item ... $(\omega^G)_d$ and $(\omega^H)_d$ be sequences of $d_G$-dimensional random vectors.
\end{itemize}

Assume:

\begin{itemize}
\item $\hat{\Sigma}$ is positive definite.
\item $\Sigma_d\rightarrow\hat{\Sigma}$ as $d \rightarrow \infty$.
\item There exists a $B > 0$ such that for all $d$, the components of $\omega_d^G$ and $\omega_d^H$ have absolute value less than $B$.
\item $\omega_d^G$ and $\omega_d^H$ converge almost surely to some vectors $\hat{\omega}^G$ and $\hat{\omega}^H$ elementwise as $d\rightarrow \infty$.
\item $G$ and $H$ are differentiable
\item\sloppy $G$, $G^\text{max}(\chi) = \max_{\chi' : ||\chi'||_2 \le ||\chi||_2}|G(\chi')|$, $G^\text{maxg}(\chi) = \max_{\chi' : ||\chi'||_2 \le ||\chi||_2}\Big|\Big|\frac{dG(\chi')}{d\chi}\Big|\Big|_2$, $H$, $H^\text{max}$, $H^\text{maxg}$, $G^\text{max}H^\text{max}$, $||\chi||_2G^\text{maxg}H^\text{max}$, $||\chi||_2G^\text{max}H^\text{maxg}$, $||\chi||_2^2G^\text{maxg}H^\text{maxg}$ and $G^\text{combo}(\chi)=G(\hat{\omega}^G.\chi)H(\hat{\omega}^H.\chi)$ are integrable with respect to any Gaussian measure.
\end{itemize}

Then $$\lim_{d\rightarrow\infty}  \mathbb{E}G(\omega^G_d.\chi_d)H(\omega^H_d.\chi_d) = \mathbb{E}G(\hat{\omega}^G.\hat{\chi})H(\hat{\omega}^H.\hat{\chi})$$

where the expectation is taken jointly over all random vectors that occur in the expression and $\mathsf{vec1}.\mathsf{vec2}$ denotes the elementwise product of two vectors as usual.
\end{lemma}

\begin{proof}

\sloppy We argue by contradiction. Assume there is an $\epsilon$ such that $|\mathbb{E}G(\omega^G_d.\chi_d)H(\omega^H_d.\chi_d) - \mathbb{E}G(\hat{\omega}^G.\hat{\chi})H(\hat{\omega}^H.\hat{\chi})| > \epsilon$ for arbitrarily large $d$. Let $B_2$ be 1 or the largest absolute value among components in either $\hat{\omega}^G$ or $\hat{\omega}^H$, whichever is larger. Then for any $d$ and $0 < \delta < B_2$ we have

\begin{eqnarray*}
&&|\mathbb{E}G(\omega^G_d.\chi_d)H(\omega^H_d.\chi_d) - \mathbb{E}G(\hat{\omega}^G.\hat{\chi})H(\hat{\omega}^H.\hat{\chi})|\\
&\le&|\mathbb{E}G(\omega^G_d.\chi_d)H(\omega^H_d.\chi_d) - \mathbb{E}G(\hat{\omega}^G.\chi_d)H(\hat{\omega}^H.\chi_d)|\\
&&+ |\mathbb{E}G(\hat{\omega}^G.\chi_d)H(\hat{\omega}^H.\chi_d) - \mathbb{E}G(\hat{\omega}^G.\hat{\chi})H(\hat{\omega}^H.\hat{\chi})|\\
&\le&\Big|\int_{|\omega^G_d[i] - \hat{\omega}^G[i]| < \delta \text{ and } |\omega^H_d[i] - \hat{\omega}^H[i]| < \delta \text{ for all $i$}}G(\omega^G_d.\chi_d)H(\omega^H_d.\chi_d) - G(\hat{\omega}^G.\chi_d)H(\hat{\omega}^H.\chi_d)d\mu\Big|\\
&&+\Big|\int_{|\omega^G_d[i] - \hat{\omega}^G[i]| \ge \delta \text{ or } |\omega^H_d[i] - \hat{\omega}^H[i]| \ge \delta \text{ for some $i$}}G(\omega^G_d.\chi_d)H(\omega^H_d.\chi_d)\\
&& - G(\hat{\omega}^G.\chi_d)H(\hat{\omega}^H.\chi_d)d\mu\Big| + |\mathbb{E}G(\hat{\omega}^G.\chi_d)H(\hat{\omega}^H.\chi_d) - \mathbb{E}G(\hat{\omega}^G.\hat{\chi})H(\hat{\omega}^H.\hat{\chi})|
\end{eqnarray*}

$d\mu$ here is the joint measure of all random vectors. $i$ ranges over all components of the $\omega$. We will address each of the three terms in turn. 

\underline{First term}: Denote the $d_G$-dimensional gradient of $G$ / $H$ by $G'$ / $H'$. Let $\chi^\text{max} \sim \mathcal{N}(0, sI)$ be a $d_G$-dimensional Gaussian vector, where $s$ is a positive constant larger than any eigenvalue of any of the $\Sigma_d$ or $\hat{\Sigma}$. $s$ exists because $\Sigma_d$ converges. Using the Taylor expansion and the Lagrange form of the remainder, we have, for some (random) $\chi^\text{inter-G}$ on the line segment between $\hat{\omega}^G.\chi_d$ and $\omega^G_d.\chi_d$ and some (random) $\chi^\text{inter-H}$ on the line segment between $\hat{\omega}^H.\chi_d$ and $\omega^H_d.\chi_d$, the following.

\begin{eqnarray*}
&&\Big|\int_{|\omega^G_d[i] - \hat{\omega}^G[i]| < \delta \text{ and } |\omega^H_d[i] - \hat{\omega}^H[i]| < \delta \text{ for all $i$}}G(\omega^G_d.\chi_d)H(\omega^H_d.\chi_d) - G(\hat{\omega}^G.\chi_d)H(\hat{\omega}^H.\chi_d)d\mu\Big|\\
&=&\Big|\int_{...}G(\hat{\omega}^G.\chi_d + (\omega^G_d-\hat{\omega}^G).\chi_d)H(\hat{\omega}^H.\chi_d + (\omega^H_d-\hat{\omega}^H).\chi_d) - G(\hat{\omega}^G.\chi_d)H(\hat{\omega}^H.\chi_d)d\mu\Big|\\
&=&\Big|\int_{...}\Big(G(\hat{\omega}^G.\chi_d) + \sum_i\big((\omega_d^G-\hat{\omega}^G).\chi_d.G'(\chi^\text{inter-G})\big)[i]\Big)...\\
&&...\Big(H(\hat{\omega}^H.\chi_d) + \sum_i\big((\omega^H_d-\hat{\omega}^H).\chi_d.H'(\chi^\text{inter-H})\big)[i]\Big)\\
&& - G(\hat{\omega}^G.\chi_d)H(\hat{\omega}^H.\chi_d)d\mu\Big|\\
&=&\Big|\int_{...}\sum_i\big((\omega_d^G-\hat{\omega}^G).\chi_d.G'(\chi^\text{inter-G})\big)[i]H(\hat{\omega}^H.\chi_d) \\
&&+  \sum_i\big((\omega^H_d-\hat{\omega}^H).\chi_d.H'(\chi^\text{inter-H})\big)[i]G(\hat{\omega}^G.\chi_d)\\
&& + \sum_i\big((\omega_d^G-\hat{\omega}^G).\chi_d.G'(\chi^\text{inter-G})\big)[i] \sum_i\big((\omega^H_d-\hat{\omega}^H).\chi_d.H'(\chi^\text{inter-H})\big)[i]d\mu\Big|\\
&\le&\int_{...}||(\omega^G_d-\hat{\omega}^G).\chi_d||_2||G'(\chi^\text{inter-G})||_2|H(\hat{\omega}^H.\chi_d)| \\
&& + ||(\omega^H_d-\hat{\omega}^H).\chi_d||_2||H'(\chi^\text{inter-H})||_2|G(\hat{\omega}^G.\chi_d)|\\
&&+||(\omega^G_d-\hat{\omega}^G).\chi_d||_2||G'(\chi^\text{inter-G})||_2||(\omega^H_d-\hat{\omega}^H).\chi_d||_2||H'(\chi^\text{inter-H})||_2d\mu\\
&\le&\int_{...}||(\omega_d^G-\hat{\omega}^G).\chi_d||_2G^\text{maxg}(\chi^\text{inter-G})H^\text{max}(\hat{\omega}^H.\chi_d)\\
&&+ ||(\omega_d^H-\hat{\omega}^H).\chi_d||_2H^\text{maxg}(\chi^\text{inter-H})G^\text{max}(\hat{\omega}^G.\chi_d)\\
&&+||(\omega_d^G-\hat{\omega}^G).\chi_d||_2G^\text{maxg}(\chi^\text{inter-G})||(\omega_d^H-\hat{\omega}^H).\chi_d||_2H^\text{maxg}(\chi^\text{inter-H})d\mu\\
&\le&\int_{...}\big|\big||\omega^G_d-\hat{\omega}^G|.|\chi_d|\big|\big|_2G^\text{maxg}(\max(|\hat{\omega}^G|,|\omega^G_d|).|\chi_d|)H^\text{max}(|\hat{\omega}^H|.|\chi_d|)\\
&&+\big|\big||\omega^H_d-\hat{\omega}^H|.|\chi_d|\big|\big|_2H^\text{maxg}(\max(|\hat{\omega}^H|,|\omega^H_d|).|\chi_d|)G^\text{max}(|\hat{\omega}^G|.|\chi_d|)\\
&&+\big|\big||\omega^G_d-\hat{\omega}^G|.|\chi_d|\big|\big|_2G^\text{maxg}(\max(|\hat{\omega}^G|,|\omega^G_d|).|\chi_d|)...\\
&&...\big|\big||\omega^H_d-\hat{\omega}^H|.|\chi_d|\big|\big|_2H^\text{maxg}(\max(|\hat{\omega}^H|,|\omega^H_d|).|\chi_d|)d\mu\\
&\le&\int_{...}\big|\big|2B_2\delta|\chi_d|\big|\big|_2G^\text{maxg}(2B_2|\chi_d|)H^\text{max}(2B_2|\chi_d|)\\
&&+\big|\big|2B_2\delta|\chi_d|\big|\big|_2H^\text{maxg}(2B_2|\chi_d|)G^\text{max}(2B_2|\chi_d|)\\
&&+\big|\big|2B_2\delta|\chi_d|\big|\big|_2G^\text{maxg}(2B_2|\chi_d|)\big|\big|2B_2\delta|\chi_d|\big|\big|_2H^\text{maxg}(2B_2|\chi_d|)d\mu\\
&\le&\delta\mathbb{E}\big|\big|2B_2\chi_d\big|\big|_2G^\text{maxg}(2B_2\chi_d)H^\text{max}(2B_2\chi_d) + \delta\mathbb{E}\big|\big|2B_2\chi_d\big|\big|_2H^\text{maxg}(2B_2\chi_d)G^\text{max}(2B_2\chi_d)\\
&& + \delta^2\mathbb{E}\big|\big|2B_2\chi_d\big|\big|_2^2G^\text{maxg}(2B_2\chi_d)H^\text{maxg}(2B_2\chi_d)\\
&\le&\delta\mathbb{E}\big|\big|2B_2\chi^\text{max}\big|\big|_2G^\text{maxg}(2B_2\chi^\text{max})H^\text{max}(2B_2\chi^\text{max})\\
&&+ \delta\mathbb{E}\big|\big|2B_2\chi^\text{max}\big|\big|_2H^\text{maxg}(2B_2\chi^\text{max})G^\text{max}(2B_2\chi^\text{max})\\
&&+ \delta^2\mathbb{E}\big|\big|2B_2\chi^\text{max}\big|\big|_2^2G^\text{maxg}(2B_2\chi^\text{max})H^\text{maxg}(2B_2\chi^\text{max})
\end{eqnarray*}

Here, all integrals are over the same set and $\max(.,.)$ is taken elementwise when applied to vector(s). Since $||\chi||_2G^\text{maxg}(\chi)H^\text{max}(\chi)$, $||\chi||_2G^\text{max}(\chi)H^\text{maxg}(\chi)$ and $||\chi||_2^2G^\text{maxg}(\chi)H^\text{maxg}(\chi)$ are integrable with respect to any Gaussian measure, each of the three expectations in the last line of the derivation above is indeed finite. Since the above holds for any $0 < \delta < B_2$, we can choose $\delta$ small enough so that the expression on the last line is less than $\frac{\epsilon}{3}$. Fix $\delta$ to this value going forward. The key to the above derivation is that $G^\text{max}$, $G^\text{maxg}$, $H^\text{max}$, $H^\text{maxg}$ and $|| . ||_2$ depend only on the length of their argument in a monotonically increasing fashion. Hence, increasing the length of their argument does not reduce their value.

\underline{Second term:} Let $(r)_d$ be a sequence of values in $\mathbb{R}_{\ge 0} \cup \{\infty\}$ such that the following holds. For each $d$, the probability that $|\omega^G_d[i] - \hat{\omega}^G[i]| \ge \delta$ or $|\omega^H_d[i] - \hat{\omega}^H[i]| \ge \delta$ for some $i$ is equal to the probability that $||B\chi^\text{max}||_2$ is larger than $r_d$. Then

\begin{eqnarray*}
&&\Big|\int_{|\omega^G_d[i] - \hat{\omega}^G[i]| \ge \delta \text{ or } |\omega^H_d[i] - \hat{\omega}^H[i]| \ge \delta \text{ for some $i$}}G(\omega^G_d.\chi_d)H(\omega^H_d.\chi_d) \\
&&- G(\hat{\omega}^G.\chi_d)H(\hat{\omega}^H.\chi_d)d\mu\Big|\\
&\le&\int_{...}|G(\omega^G_d.\chi_d)H(\omega^H_d.\chi_d)| + |G(\hat{\omega}^G.\chi_d)H(\hat{\omega}^H.\chi_d)|d\mu\\
&\le&\int_{...}G^\text{max}(\omega^G_d.\chi_d)H^\text{max}(\omega^H_d.\chi_d) + G^\text{max}(\hat{\omega}^G.\chi_d)H^\text{max}(\hat{\omega}^H.\chi_d)d\mu\\
&\le&\int_{...}G^\text{max}(B\chi_d)H^\text{max}(B\chi_d) + G^\text{max}(B\chi_d)H^\text{max}(B\chi_d)d\mu\\
&\le&\int_{...}G^\text{max}(B\chi^\text{max})H^\text{max}(B\chi^\text{max}) + G^\text{max}(B\chi^\text{max})H^\text{max}(B\chi^\text{max})d\mu\\
&=&2\int_{||B\chi^\text{max}||_2> r_d}G^\text{max}(B\chi^\text{max})H^\text{max}(B\chi^\text{max})d\mu
\end{eqnarray*}

Because the $\omega^G_d$ and $\omega^H_d$ converge almost surely elementwise, the probability that $|\omega^G_d[i] - \hat{\omega}^G[i]| \ge \delta$ or $|\omega^H_d[i] - \hat{\omega}^H[i]| \ge \delta$ for some $i$ converges to zero. So $r_d$ converges to infinity. Since $G^\text{max}H^\text{max}$ is integrable with respect to any Gaussian measure, $\mathbb{E}G^\text{max}(B\chi^\text{max})H^\text{max}(B\chi^\text{max})$ is a valid finite value. So the last line in the above derivation converges to zero.

\underline{Third term:} Both $\chi_d$ and $\hat{\chi}$ are Gaussian with mean zero and covariance matrices $\Sigma_d$ and $ \hat{\Sigma}$. The latter is positive definite. So the third term converges to zero by lemma \ref{lemma15} applied to $G^\text{combo}$ as defined in the statement of this lemma.

Hence, for $d$ beyond some value, both the second and third term are less than $\frac{\epsilon}{3}$ each, so $|\mathbb{E}G(\omega^G_d.\chi_d)H(\omega^H_d.\chi_d) - \mathbb{E}G(\hat{\omega}^G.\hat{\chi})H(\hat{\omega}^H.\hat{\chi})| < \epsilon$. Contradiction.
\end{proof}

\subsection{Theorem \ref{mfntPropagation} part 2} \label{mfntPropagationBNsection}

\begin{reptheorem}{mfntPropagation} (part 2; see section \ref{mfntPropagationsection} for part 1)

Let ...

\begin{itemize}
\item ... $f$ be an A-architecture with a big output layer of variable width.
\item ... $\mathcal{D}$ be an input distribution.
\item ... $\mathcal{D}^{B}$ be the distribution over batches of size $B \ge 3$ where in each batch, each individual input is drawn independently from $\mathcal{D}$.
\item ... $X^{(1)} = (x^{(1,1)}, .., x^{(1,B)})$, $X^{(2)} = (x^{(2,1)}, .., x^{(2,B)})$, .., $X^{(N)} = (x^{(N,1)}, .., x^{(N,B)})$ be a sample from $\mathcal{D}^B$ of size $N \ge 2$ and let $\mathcal{D}^{(N,B)}$ be the discrete uniform distribution over that sample.
\item ... $\vec{\epsilon}$ be the vector of all regularizers used by normalization layers in $f$.
\end{itemize}

Assume:

\begin{itemize}
\item $\mathcal{D}$ is elem-like$(q,c)$.
\item $q > c$
\end{itemize}

If $f$ does not contain layer normalization but can contain batch normalization layers, we have

\begin{eqnarray}
\lim\mathbb{E}_if_{l,1}(X^{(1)})[i] &=& \mathfrak{m}_l \label{eqn52p1}\\
\lim\mathbb{E}_if_{l,1}(X^{(1)})[i]^2 &=& \mathfrak{q}_l\label{eqn52p2}\\
\lim\mathbb{E}_if_{l,1}(X^{(1)})[i]f_{l,2}(X^{(1)})[i] &=& \mathfrak{c}_l \label{eqn52p3}\\
\lim\mathbb{E}_if_{l,1}(X^{(1)})[i]f_{l,1}(X^{(2)})[i] &=& \mathfrak{c}_l \label{eqn52p4}\\
\lim\frac{1}{d_l^\text{MF}}||\mathcal{J}_{l,m,1}(X^{(1)})||_F^2 &=& \frac{\mathfrak{g}_l}{\mathfrak{g}_m}\label{eqn52p5}\\
\lim\frac{1}{d_l^\text{MF}}\mathbb{E}_{X}||f_{l,1}(X)||^2_2 &=& \mathfrak{q}_l \label{eqn52p6}\\
\lim\frac{1}{d_l^\text{MF}}||\mathbb{E}_{X}f_{l,1}(X)||^2_2 &=& \mathfrak{c}_l\label{eqn52p7}\\
\lim\frac{1}{d_l^\text{MF}}||\mathbb{S}_{X}f_{l,1}(X)||^2_2 &=& \mathfrak{q}_l - \mathfrak{c}_l \label{eqn52p8}\\
\lim\frac{1}{d_l^\text{MF}}\mathbb{E}_{X}||\mathcal{J}_{l,m,1}(X)||_F^2 &=& \frac{\mathfrak{g}_l}{\mathfrak{g}_m} \label{eqn52p9}\\
\lim\frac{1}{d_l^\text{MF}}\mathbb{E}_X\Tr(\mathcal{J}_{l,m,1}(X)\Cov_{f_m}\mathcal{J}_{l,m,1}(X)^T) &=& \frac{\mathfrak{g}_l(\mathfrak{q}_m - \mathfrak{c}_m)}{\mathfrak{g}_m}\label{eqn52p10}\\
\lim NLC(f_l(f_m),f_m(\mathcal{D}^{(N,B)})) &=& \sqrt{\frac{\mathfrak{g}_l(\mathfrak{q}_m - \mathfrak{c}_m)}{\mathfrak{g}_m(\mathfrak{q}_l - \mathfrak{c}_l)}}\label{eqn52p11}
\end{eqnarray}

where

\begin{itemize}
\item ... $X \sim \mathcal{D}^{(N,B)}$
\item ... $\lim$ stands for $\lim_{\vec{\epsilon}\rightarrow 0}\Big(\lim_{B\rightarrow\infty,N\rightarrow\infty}\big(\lim_{d_\text{MF}\rightarrow\infty} ... \text{a.s.}\big)\Big)$. The inner limit takes $d_\text{MF}$ to infinity. The middle limit takes $B$ and $N$ to infinity. The outer limit takes $\vec{\epsilon}$ to zero. The inner limit is an ``almost sure'' limit.
\item ... $\mathfrak{m}_l$, $\mathfrak{q}_l$, $\mathfrak{c}_l$ and $\mathfrak{g}_l$ are calculated via table \ref{tableNLCPropagation} applied to the big layer graph of $f$.
\item ... $f_m$ is a bottleneck for $f_l$.
\item ... $0 \le m \le l \le L$
\item ... randomness is induced by $\theta$ and the $X^{(n)}$.
\end{itemize}

\sloppy Further, if $f_{L,1}$ instead has fixed dimensionality $d_\text{out}$, the above limits do not necessarily hold when $l=L$. Instead, (c) as $d_\text{MF} \rightarrow \infty$, the meta-distribution of $f_{L,1}$ expansion-converges to the elementwise meta-distribution with generator $\mathcal{MN}(\mathsf{parm1}, \mathsf{parm2})$, where $\lim_{\vec{\epsilon}\rightarrow 0}\lim_{B\rightarrow \infty}\mathsf{parm1}=\mathfrak{q}_L$ and $\lim_{\vec{\epsilon}\rightarrow 0}\lim_{B\rightarrow \infty}\mathsf{parm2}=\mathfrak{c}_L$. The first level of randomness is induced by $\mathcal{D}$ and the second level by $\theta$. And, (d) when $N$ is fixed, for almost all samples, as $d_\text{MF} \rightarrow \infty$, $(f_{L,1}(X^{(1)}), .., f_{L,1}(X^{(N)}))$ converges in distribution to a Gaussian that is elementwise over output vectors where the generator has mean zero and covariance matrix $K(N, \mathsf{parm1}, \mathsf{parm2})$, where again $\lim_{\vec{\epsilon}\rightarrow 0}\lim_{B\rightarrow \infty}\mathsf{parm1}=\mathfrak{q}_L$ and $\lim_{\vec{\epsilon}\rightarrow 0}\lim_{B\rightarrow \infty}\mathsf{parm2}=\mathfrak{c}_L$. Randomness for each sample is induced by $\theta$.
\end{reptheorem}

\paragraph{Remark} We note that by symmetry, analogous results hold when the batch index 1 is replaced by any other batch index $b$ and the batch index 2 is replaced by any $b' \neq b$. We then also need to assume $B \ge b$ and $B \ge b'$, which is immaterial as we take the limit of $B$ to infinity.

There are several ways part 2 of theorem \ref{mfntPropagation} could have been framed. Instead of taking the expectations just over the batch $X$, they could be taken over $X$ and $b$. For example, instead of $\lim\frac{1}{d_l^\text{MF}}||\mathbb{E}_{X}f_{l,1}(X)||^2_2 = \mathfrak{c}_l$, we could have used $\lim\frac{1}{d_l^\text{MF}}||\mathbb{E}_{X,b}f_{l,b}(X)||^2_2 = \mathfrak{c}_l$, where $\mathbb{E}$ applied to the discrete, finite index $b$ refers to the mean as usual. We could have also used big layers, which would result in e.g. $\lim\frac{1}{Bd_l^\text{MF}}||\mathbb{E}_{X}f_{l}(X)||^2_2 = \mathfrak{c}_l$ or in considering the meta-distribution of $f_L$. We could consider gradients with respect to small layers, which may yield e.g. $\lim\frac{1}{d_l^\text{MF}}||\frac{df_{l,b}}{df_{m,b'}}(X^{(1)})||_F^2 = \mathbbm{1}_{b=b'}\frac{\mathfrak{g}_l}{\mathfrak{g}_m}$. All these possible choices lead to analogous results. We made the choice to frame the theorem in terms of individual small layers and gradients with respect to big layers and applied the choice consistently throughout the theorem, except for the NLC which is defined in section \ref{nlcBnsection} for the BN case by considering big layers. Transforming one framing of the theorem into another is left as an exercise to the reader.

\begin{proof}

The proof is a continuation of the proof of part 1 of the theorem given in section \ref{mfntPropagationsection}. Notation, terminology and conventions continue to apply with modifications as described in section \ref{bnNotationSection}. In addition, we have the following NTCs.

\begin{itemize}
\item In this proof, we will add layers to a mean field architecture $F$ that have the same joint distribution as the small layers in $f$. Hence, we will denote the layer that corresponds to $f_{l,b}$ as $F_{l,b}$, even though $F_{l,b}$ is not a small layer but a ``regular layer'' in the mean field architecture context. As before, we will define the layer graph of $F$ simply over all legal combinations of $\mathsf{sub}$ and $\mathsf{super}$ for $F^\mathsf{super}_\mathsf{sub}$. We write e.g. $F_{k,b}$ for the layer corresponding to $f_{k,b}$ where $f_k$ is the dependency of $f_l$ in the big layer graph of $f$ but $F_{k,b}$ is not necessarily a dependency of $F_{l,b}$. We extend this notational convention as before to e.g. $\tilde{\mathfrak{m}}_{l,b}$ and $\mathfrak{c}_{l,b;l,b}$ defined via table \ref{tableBackgroundPropagation}. On the other hand, the batch index is always omitted for quantities that do not depend on it, such as $\tau_l$ or the $\mathfrak{m}_l$, $\mathfrak{q}_l$, $\mathfrak{c}_l$ and $\mathfrak{g}_l$ defined via table \ref{tableNLCPropagation}.
\item To simplify the presentation, we usually fix a batch index $b$ to 1 and a batch index $b' \neq b$ to 2. When we do this, the argument implicitly applies to all possible choices for $b$ and $b'$ due to symmetry. For example, when we show that $\rho_{l,1}$ is controlled, we generally show that all $\rho_{l,b}$ are controlled. When we show $\mathfrak{c}_{l,1;l,2}=0$, we generally implicitly show $\mathfrak{c}_{l,b;l,b'}=0$ for all $b \neq b'$.
\item We place two dots over any quantity to indicate its limit with respect to $N$ and $B$, e.g. $\ddot{\mathfrak{c}}_{l,b;l,b} = \lim_{B\rightarrow\infty,N\rightarrow\infty} \mathfrak{c}_{l,b;l,b}$. The double dot overrides a tilde, so e.g. $\ddot{\mathfrak{m}}_{l,b} = \lim_{B\rightarrow\infty,N\rightarrow\infty} \tilde{\mathfrak{m}}_{l,b}$. We provide more details on this throughout the proof.
\item $\widehat{expr}$ denotes the mean over the batch index or ``batch mean'', e.g. $\widehat{e_me_l} = \mathbb{E}_be_{m,b}e_{l,b}$. Note that we use a little hat when referencing quantities that arise in lemma \ref{lemma18}.
\item Analogous to previous stages, we can assume $\overline{x^{(n,b)}.x^{(n',b')}}$ is equal to $q$ when $n=n'$ and $b = b'$ and equal to $c$ otherwise.
\item As before, we apply proposition \ref{mfntPositive} throughout the proof, but we no longer make this explicit.
\item Going forward, no layer in $F$ has mean dependencies, so we only have to consider elementwise dependencies for elementwise layers and elementwise inputs for multi-activation functions. We write e.g. $\mathbb{E}_e\rho_{l,1}$ short for $\mathbb{E}_e\rho_{l,1}(e)$. According to lemma H.4 from \cite{meanFieldNetsorGP}, we then also have rank stability, so we no longer have to consider this property explicitly. Further, parameter-control reduces to ``simple'' control.
\end{itemize}

\underline{Stage 9}: statements (\ref{eqn52p1}), (\ref{eqn52p2}) and (\ref{eqn52p3})

The case $l=0$ is again trivial. For $l > 1$, we will construct a mean field architecture $F$ which has a layer $F_{l,b}$ corresponding to each $f_{l,b}$. We proceed by induction on $l$ as usual. For readin, FC, bias, addition and activation big layers, we add layers to $F$ and extend $\mathcal{G}$ as if we were constructing the $N$-duplex $(F^{(1)}, .., F^{(N)})$ early in stage 4 and its associated generator, except that the $n$ index is replaced by a $b$ index. Distribution equality follows as in stage 4. If $f_l$ is a BN layer, we add elementwise layers $F_{l,b}$, each with the same width as any of the $f_{l,b}$. We set

$$\rho_{l,b} = \frac{\rho_{k,b} - \widehat{\rho_k}}{\sqrt{\nu_l}} \text{ where } \nu_l = \mathbb{E}_{b'}(\rho_{k,b'} - \widehat{\rho_{k}})^2+\epsilon_l$$

The inputs of $\rho_{l,b}$ are those occurring in any of the $\rho_{k,b'}$ for any $b'$. We obtain distribution equality because $F_{k,1}, .., F_{k,B}$ can be transformed to $F_{l,b}$ by applying $f_{l,b}(f_{k,1}, .., f_{k,B})$. $\nu_l > 0$ holds as $\epsilon_l > 0$.

It is important to note that the resulting $F$ and $\mathcal{G}$ are symmetric with respect to the batch index. We exploit this in our notation by setting the batch index to 1 or 2 as described above.

Now we show limits and control. From background theorem \ref{backgroundMaster} and distribution equality, we will obtain $\limd \overline{f_{l,1}(X^{(1)})}=\tilde{\mathfrak{m}}_{l,1}$, $\limd \overline{f_{l,1}(X^{(1)}).f_{l,1}(X^{(1)})} = \mathfrak{c}_{l,1;l,1}$ and $\limd \overline{f_{l,1}(X^{(1)}).f_{l,2}(X^{(1)})}=\mathfrak{c}_{l,1;l,2}$, where $\tilde{\mathfrak{m}}_{l,1}$, $ \mathfrak{c}_{l,1;l,1}$ and $\mathfrak{c}_{l,1;l,2}$ are defined via table \ref{tableBackgroundPropagation}. As part of the induction, we will show (i) the validity of and give calculation rules for $\ddot{\mathfrak{m}}_{l,1}$, $\ddot{\mathfrak{c}}_{l,1;l,1}$ and $\ddot{\mathfrak{c}}_{l,1;l,2}$, (ii) $\mathfrak{m}_l = \limep\ddot{\mathfrak{m}}_{l,1}$, $\mathfrak{q}_l = \limep\ddot{\mathfrak{c}}_{l,1;l,1}$ and $\mathfrak{c}_l = \limep\ddot{\mathfrak{c}}_{l,1;l,2}$, where $\mathfrak{m}_l$, $\mathfrak{q}_l$ and $\mathfrak{c}_l$ are defined via table \ref{tableNLCPropagation} and (iii) if there is no directed path from an activation layer to $f_l$ not containing an FC layer, $\limbas \mathbb{E}_b (\rho_{l,b} - \widehat{\rho_l})^2 = \ddot{\mathfrak{c}}_{l,1;l,1} - \ddot{\mathfrak{c}}_{l,1;l,2}$, where randomness in induced by $e$ distributed according to table \ref{tableBackgroundPropagation}. Finally, we show $\ddot{\mathfrak{c}}_{l,1;l,1} > \ddot{\mathfrak{c}}_{l,1;l,2} \ge 0$ (``non-singularity''). Since this stage is independent of $N$, we can use $\limbn$ and $\limb$ interchangeably.

Case: $f_l$ is a readin layer. 

\begin{itemize}
\item Control: NA
\item Limits: From table \ref{tableBackgroundPropagation}, we obtain $\tilde{\mathfrak{m}}_{l,1}=0$, $\mathfrak{c}_{l,1;l,1} = q\sigma_l^2$ and $\mathfrak{c}_{l,1;l,2} = c\sigma_l^2$ and thus $\ddot{\mathfrak{m}}_{l,1}=0$, $\ddot{\mathfrak{c}}_{l,1;l,1}=q\sigma_l^2$ and $\ddot{\mathfrak{c}}_{l,1;l,2}=c\sigma_l^2$ and thus $\limep \ddot{\mathfrak{m}}_{l,1}=0$, $\limep\ddot{\mathfrak{c}}_{l,1;l,1}=q\sigma_l^2$ and $\limep\ddot{\mathfrak{c}}_{l,1;l,2}=c\sigma_l^2$. Since $f_l$ is fully-connected and its dependency is the input layer, table \ref{tableNLCPropagation} yields $\mathfrak{m}_l=0$, $\mathfrak{q}_l = q\sigma_l^2$ and $\mathfrak{c}_l = c\sigma_l^2$. So we have a match, as required.

We have

\begin{eqnarray*}
&&\limbas \mathbb{E}_b (\rho_{l,b} - \widehat{\rho_l})^2\\
&=&\limbas \mathbb{E}_b (u_{l,b} - \widehat{u_l})^2 \text{ where } (u_{l,1},..,u_{l,B}) \sim \mathcal{N}(0, K(B, \mathfrak{c}_{l,1;l,1},\mathfrak{c}_{l,1;l,2}))\\
&=&\limbas \frac{(B-1)(\mathfrak{c}_{l,1;l,1}-\mathfrak{c}_{l,1;l,2})}{B}\frac{1}{B-1}\sum_{b=1}^{B-1}u_{l,b}^2 \text{ where } (u_{l,1},..,u_{l,B-1}) \sim \mathcal{N}(0, I)\\
\end{eqnarray*}

By the strong law of large numbers, this is equal to $\ddot{\mathfrak{c}}_{l,1;l,1} - \ddot{\mathfrak{c}}_{l,1;l,2}$ as required.
\item Non-singularity: $\ddot{\mathfrak{c}}_{l,1;l,1} - \ddot{\mathfrak{c}}_{l,1;l,2} = (q-c)\sigma_l^2 > 0$ and $\ddot{\mathfrak{c}}_{l,1;l,2}=c\sigma_l^2 \ge 0$
\end{itemize}

Case: $f_l$ is a fully-connected, non-readin layer.

\begin{itemize}
\item Control: NA
\item Limits: We obtain $\tilde{\mathfrak{m}}_{l,1}=0$, $\mathfrak{c}_{l,1;l,1} = \sigma_l^2\mathfrak{c}_{k,1;k,1}$ and $\mathfrak{c}_{l,1;l,2} = \sigma_l^2\mathfrak{c}_{k,1;k,2}$ and thus $\ddot{\mathfrak{m}}_{l,1}=0$, $\ddot{\mathfrak{c}}_{l,1;l,1} = \sigma_l^2\ddot{\mathfrak{c}}_{k,1;k,1}$ and $\ddot{\mathfrak{c}}_{l,1;l,2}= \sigma_l^2\ddot{\mathfrak{c}}_{k,1;k,2}$ and thus $\limep \ddot{\mathfrak{m}}_{l,1}=0=\mathfrak{m}_l$, $\limep\ddot{\mathfrak{c}}_{l,1;l,1}= \sigma_l^2\limep\ddot{\mathfrak{c}}_{k,1;k,1}=\sigma_l^2\mathfrak{q}_k=\mathfrak{q}_l$ and $\limep\ddot{\mathfrak{c}}_{l,1;l,2}= \sigma_l^2\limep\ddot{\mathfrak{c}}_{k,1;k,2}=\sigma_l^2\mathfrak{c}_k=\mathfrak{c}_l$. Finally, $\limbas \mathbb{E}_b (\rho_{l,b} - \widehat{\rho_l})^2=\ddot{\mathfrak{c}}_{l,1;l,1} - \ddot{\mathfrak{c}}_{l,1;l,2}$ as for the readin layer case above.
\item Non-singularity: $\ddot{\mathfrak{c}}_{l,1;l,1} - \ddot{\mathfrak{c}}_{l,1;l,2} = \sigma_l^2 (\ddot{\mathfrak{c}}_{k,1;k,1} - \ddot{\mathfrak{c}}_{k,1;k,2})> 0$ and $\ddot{\mathfrak{c}}_{l,1;l,2}= \sigma_l^2\ddot{\mathfrak{c}}_{k,1;k,2} \ge 0$
\end{itemize}

Case: $f_l$ is a bias layer.

\begin{itemize}
\item Control: Since $\rho_{l,1}$ only has an additional linear term relative to $\rho_{k,1}$, if $\rho_{k,1}$ is controlled by $\mathcal{C}^\text{E2}$, then so is $\rho_{l,1}$.
\item Limits: We have $\tilde{\mathfrak{m}}_{l,1}=\mathbb{E}_e(e_{l,1}^\text{in} + \rho_{k,1})=\tilde{\mathfrak{m}}_{k,1}$, $\mathfrak{c}_{l,1;l,1}= \mathbb{E}_e(e_{l,1}^\text{in} + \rho_{k,1})^2 = \mathbb{E}_ee_{l,1}^\text{in}e_{l,1}^\text{in} +\mathbb{E}_e\rho_{k,1}^2 + 2\mathbb{E}_ee_{l,1}^\text{in}\rho_{k,1}=\mathfrak{c}_{k,1;k,1} + \sigma_l^2$ and $\mathfrak{c}_{l,1;l,2}= \mathbb{E}_e(e_{l,1}^\text{in} + \rho_{k,1})(e_{l,2}^\text{in} + \rho_{k,2}) = \mathbb{E}_ee_{l,1}^\text{in}e_{l,2}^\text{in} + \mathbb{E}_e\rho_{k,1}\rho_{k,2} + \mathbb{E}_ee_{l,1}^\text{in}\rho_{k,2} + \mathbb{E}_ee_{l,2}^\text{in}\rho_{k,1}=\mathfrak{c}_{k,1;k,2} + \sigma_l^2$ and thus $\ddot{\mathfrak{m}}_{l,1}=\ddot{\mathfrak{m}}_{k,1}$, $\ddot{\mathfrak{c}}_{l,1;l,1}=\ddot{\mathfrak{c}}_{k,1;k,1}+\sigma_l^2$ and $\ddot{\mathfrak{c}}_{l,1;l,2}=\ddot{\mathfrak{c}}_{k,1;k,2}+\sigma_l^2$ and thus $\limep \ddot{\mathfrak{m}}_{l,1}=\limep\ddot{\mathfrak{m}}_{k,1}=\mathfrak{m}_k=\mathfrak{m}_l$, $\limep\ddot{\mathfrak{c}}_{l,1;l,1}= \sigma_l^2+\limep\ddot{\mathfrak{c}}_{k,1;k,1}=\mathfrak{q}_k+\sigma_l^2=\mathfrak{q}_l$ and $\limep\ddot{\mathfrak{c}}_{l,1;l,2}= \sigma_l^2+\limep\ddot{\mathfrak{c}}_{k,1;k,2}=\mathfrak{c}_k+\sigma_l^2=\mathfrak{c}_l$. Here we use that the $F_{l,1}^\text{in}$ are independent of previous layers of $F$.

If there is no directed path from an activation layer to $f_l$ not containing an FC layer, there is no directed path from an activation layer to $f_k$ not containing an FC layer. Hence

\begin{eqnarray*}
&&\limbas \mathbb{E}_b (\rho_{l,b} - \widehat{\rho_l})^2\\
&=&\limbas \mathbb{E}_b (\rho_{k,b} + e_{l,b}^\text{in} - \mathbb{E}_{b'}(\rho_{k,b'} + e_{l,b'}^\text{in}))^2\\
&=&\limbas \mathbb{E}_b (\rho_{k,b} - \widehat{\rho_k})^2\\
&=&\ddot{\mathfrak{c}}_{k,1;k,1} - \ddot{\mathfrak{c}}_{k,1;k,2}\\
&=&\ddot{\mathfrak{c}}_{l,1;l,1} - \ddot{\mathfrak{c}}_{l,1;l,2}\\
\end{eqnarray*}

\item Non-singularity: $\ddot{\mathfrak{c}}_{l,1;l,1} - \ddot{\mathfrak{c}}_{l,1;l,2}=\ddot{\mathfrak{c}}_{k,1;k,1} - \ddot{\mathfrak{c}}_{k,1;k,2} > 0$ and $\ddot{\mathfrak{c}}_{l,1;l,2}=\ddot{\mathfrak{c}}_{k,1;k,2}+\sigma_l^2 \ge 0$.
\end{itemize}

Case: $f_l$ is a BN layer.

\begin{itemize}
\item Control: We have $\widehat{\rho_l^2} = \frac{\mathbb{E}_b(\rho_{k,b} - \widehat{\rho_k})^2}{\mathbb{E}_b(\rho_{k,b} - \widehat{\rho_k})^2+\epsilon_l} < 1$, so $\rho_{l,1} < \sqrt{B}$, so $\rho_{l,1}$ is bounded, so it is controlled by $\mathcal{C}^\text{E2}$.

\item Limits: By symmetry, we have $\tilde{\mathfrak{m}}_{l,1} = \widehat{\tilde{\mathfrak{m}}_l} = \mathbb{E}_b \mathbb{E}_e \frac{\rho_{k,b} - \widehat{\rho_k}}{\sqrt{\mathbb{E}_{b'}(\rho_{k,b'} - \widehat{\rho_k})^2+\epsilon_l}} = \mathbb{E}_e \frac{\widehat{\rho_k} - \widehat{\rho_k}}{\sqrt{\mathbb{E}_{b'}(\rho_{k,b'} - \widehat{\rho_k})^2+\epsilon_l}} = 0$ and thus $\ddot{\mathfrak{m}}_{l,1}=0$. 

As in previous stages, we consider the recursive expansion of $\rho_{l,1}$ that flows backwards through the small layer graph of $f_l$ until it hits input and / or fully-connected layers. It is easy to see that if the expansion touches a small layer via some path in the small layer graph, it touches the corresponding big layer via an analogous path in the big layer graph. Hence, by property \ref{aprop9}, the expansion cannot hit an activation layer. It yields an expression of the following form. 

$$\rho_{l,1} = \sum_{f_{l'}} c_{l'}\bigg(\prod_{f_{l''}\in P_{l',l}}\frac{1}{\sqrt{\nu_{l''}}}\bigg)(e_{l',1}-\widehat{e_{l'}})$$

Here, the sum is over input and FC big layers $f_{l'}$ from which there exists a directed path to $f_l$ that does not contain another FC layer. Because of property \ref{aprop10}, there exists at most one such path. We refer to this path including its endpoint $f_l$ but excluding its starting point $f_{l'}$ as $P_{l',l}$. $c_{l'}$ is the product of addition weights on $P_{l',l}$, i.e. it is the product of addition weights at addition layers in $P_{l',l}$ that correspond to an addition layer dependency that is also in $P_{l',l}$ or is equal to $f_{l'}$. $c_{l'}=1$ if there are no applicable addition weights. The product $\prod_{f_{l''}\in P_{l',l}}$ is over BN layers $f_{l''}$ in $P_{l',l}$.

We now use lemma \ref{lemma18} to obtain the limit of the expansion. In the following, we explain how objects from this proof map onto objects used in the lemma. $B$ maps onto $d$. The number of layers $f_{l'}$ summed over maps onto $d_G$. Each $e_{l',1}-\widehat{e_{l'}}$ maps onto one component of $\chi_d$. The corresponding $c_{l'}\prod_{f_{l''}\in P_{l',l}}\frac{1}{\sqrt{\nu_{l''}}}$ maps onto the corresponding component of $\omega^G_d$. $\Sigma_d$ is then a diagonal matrix with diagonal entries $\frac{B-1}{B}(\mathfrak{c}_{l',1;l',1}-\mathfrak{c}_{l',1;l',2})$ and $\hat{\Sigma}$ is a diagonal matrix with diagonal entries $\ddot{\mathfrak{c}}_{l',1;l',1}-\ddot{\mathfrak{c}}_{l',1;l',2}$, because there is no weight sharing between big layers in $f$. To obtain $\ddot{\mathfrak{c}}_{l,1;l,1}=\limbn\mathbb{E}_e\rho_{l,1}^2$, we set $G(\mathsf{arg})=(\sum_i\mathsf{arg}[i])^2$, where $i$ ranges over the vector components as usual. We do not need to consider $H$ and $\omega^H_d$ and set both to constant 1.

Let's check the conditions. Differentiability and integrability are trivial. $\hat{\Sigma}$ is a diagonal matrix with positive diagonal entries and is thus positive definite, due to the non-singularity aspect of the induction. $\lim_d \Sigma_d= \hat{\Sigma}$ holds by construction. Input and FC layers in $F$ have an $\tilde{\mathfrak{m}}$ value of zero, so $\chi_d$ and $\hat{\chi}$ have mean zero. As a bound on the components of $\omega^G_d$, we can use $\max_{f_{l'}}c_{l'}\prod_{f_{l''}\in P_{l',l}}\frac{1}{\sqrt{\epsilon_{l''}}}$. The elementwise almost sure convergence of $\omega^G_d$ is obtained from $\limbas \mathbb{E}_b(\rho_{k'',b}-\widehat{\rho_{k''}})^2 = \ddot{\mathfrak{c}}_{k'',1;k'',1}-\ddot{\mathfrak{c}}_{k'',1;k'',2}$, which is part of the induction because, by property \ref{aprop9}, there is no directed path from an activation layer to $f_{l''}$ not containing an FC layer. So we have $\ddot{\nu}_{l''} = \ddot{\mathfrak{c}}_{k'',1;k'',1}-\ddot{\mathfrak{c}}_{k'',1;k'',2} + \epsilon_l$, which is positive by non-singularity and as $\epsilon_l > 0$.

Applying lemma \ref{lemma18} yields

$$\ddot{\mathfrak{c}}_{l,1;l,1}=\mathbb{E}_u\Big(\sum_{f_{l'}} \frac{c_{l'}u_{l',1}}{\prod_{f_{l''}\in P_{l',l}}\sqrt{\ddot{\nu}_{l''}}}\Big)^2 \text{ where } u_{l',1} \sim \mathcal{N}(0,\ddot{\mathfrak{c}}_{l',1;l',1}-\ddot{\mathfrak{c}}_{l',1;l',2})$$

where the $u_{l',1}$ are also independent of each other. Hence we further have 

$$\ddot{\mathfrak{c}}_{l,1;l,1}=\sum_{f_{l'}} \frac{c_{l'}^2(\ddot{\mathfrak{c}}_{l',1;l',1}-\ddot{\mathfrak{c}}_{l',1;l',2})}{\prod_{f_{l''}\in P_{l',l}}\ddot{\nu}_{l''}}=\sum_{f_{l'}} \frac{c_{l'}^2(\ddot{\mathfrak{c}}_{l',1;l',1}-\ddot{\mathfrak{c}}_{l',1;l',2})}{\prod_{f_{l''}\in P_{l',l}}\ddot{\mathfrak{c}}_{k'',1;k'',1}-\ddot{\mathfrak{c}}_{k'',1;k'',2}+\epsilon_l}$$

Let's look at $\ddot{\mathfrak{c}}_{l,1;l,2}$. Again, we use lemma \ref{lemma18}. Now $d_G$ is twice the previous value. $\chi_d$ now has pairs of components composed of $e_{l',1}-\widehat{e_{l'}}$ and $e_{l',2}-\widehat{e_{l'}}$. The $\omega^G_d$ component is the same for each pair. $\Sigma_d$ becomes a block-diagonal matrix with $2\times 2$ blocks corresponding to the $f_{l'}$, its previous diagonal entries duplicated and off-diagonal entries $-\frac{1}{B}(\mathfrak{c}_{l',1;l',1}-\mathfrak{c}_{l',1;l',2})$ within each block. $\hat{\Sigma}$ remains a diagonal matrix with its previous diagonal entries duplicated. Finally, $G(\mathsf{arg}) = (\sum_{i \text{ odd}}\mathsf{arg}[i])(\sum_{i \text{ even}}\mathsf{arg}[i])$. We obtain $\ddot{\mathfrak{c}}_{l,1;l,2}=0$. Using this, as well as the recursive calculation rules for bias layers and addition layers (see below), it is easy to check that $\ddot{\mathfrak{c}}_{l,1;l,1} = \frac{\ddot{\mathfrak{c}}_{k,1;k,1}-\ddot{\mathfrak{c}}_{k,1;k,2}}{\ddot{\mathfrak{c}}_{k,1;k,1}-\ddot{\mathfrak{c}}_{k,1;k,2}+\epsilon_l}$. So $\limep\ddot{\mathfrak{m}}_{l,1}=0=\mathfrak{m}_l$, $\limep\ddot{\mathfrak{c}}_{l,1;l,2}=0=\mathfrak{c}_l$ and $\limep\ddot{\mathfrak{c}}_{l,1;l,1}=\limep \frac{\ddot{\mathfrak{c}}_{k,1;k,1}-\ddot{\mathfrak{c}}_{k,1;k,2}}{\ddot{\mathfrak{c}}_{k,1;k,1}-\ddot{\mathfrak{c}}_{k,1;k,2}+\epsilon_l}=\frac{\limep\ddot{\mathfrak{c}}_{k,1;k,1}-\limep\ddot{\mathfrak{c}}_{k,1;k,2}}{\limep\ddot{\mathfrak{c}}_{k,1;k,1}-\limep\ddot{\mathfrak{c}}_{k,1;k,2}+\limep\epsilon_l} = \frac{\mathfrak{q}_k-\mathfrak{c}_k}{\mathfrak{q}_k-\mathfrak{c}_k}=1$.

If there is no directed path from an activation layer to $f_l$ not containing an FC layer, there is no directed path from an activation layer to $f_k$ not containing an FC layer. Hence

\begin{eqnarray*}
&&\limbas \mathbb{E}_b (\rho_{l,b} - \widehat{\rho_l})^2\\
&=&\limbas \mathbb{E}_b \frac{(\rho_{k,b} -\widehat{\rho_k} - \mathbb{E}_{b'}(\rho_{k,b'} -\widehat{\rho_k}))^2}{\mathbb{E}_b(\rho_{k,b} - \widehat{\rho_k})^2 + \epsilon_l}\\
&=&\frac{\limbas\mathbb{E}_b(\rho_{k,b} - \widehat{\rho_k})^2}{\limbas\mathbb{E}_b(\rho_{k,b} - \widehat{\rho_k})^2 + \epsilon_l}\\
&=&\frac{\ddot{\mathfrak{c}}_{k,1;k,1}-\ddot{\mathfrak{c}}_{k,1;k,2}}{\ddot{\mathfrak{c}}_{k,1;k,1}-\ddot{\mathfrak{c}}_{k,1;k,2} + \epsilon_l}\\
&=&\ddot{\mathfrak{c}}_{l,1;l,1}\\
&=&\ddot{\mathfrak{c}}_{l,1;l,1}-\ddot{\mathfrak{c}}_{l,1;l,2}\\
\end{eqnarray*}

\item Non-singularity: We have $\ddot{\mathfrak{c}}_{l,1;l,1}-\ddot{\mathfrak{c}}_{l,1;l,2}=\frac{\ddot{\mathfrak{c}}_{k,1;k,1}-\ddot{\mathfrak{c}}_{k,1;k,2}}{\ddot{\mathfrak{c}}_{k,1;k,1}-\ddot{\mathfrak{c}}_{k,1;k,2}+\epsilon_l} > 0$ and $\ddot{\mathfrak{c}}_{l,1;l,2}=0\ge 0$.
\end{itemize}

Case: $f_l$ is an addition layer.

\begin{itemize}
\item Control: Since $\mathcal{C}^\text{E2}$ is linearly closed, if the $\rho_{k[\kappa],1}$ are controlled by $\mathcal{C}^\text{E2}$, so is $\rho_{l,1}$.
\item \sloppy Limits: The argument is analogous to stage 1. $\tilde{\mathfrak{m}}_{l,1} = \mathbb{E}_e \sum_{\kappa=1}^{K}w_\kappa\rho_{k[\kappa],1} = \sum_{\kappa=1}^Kw_{\kappa}\tilde{\mathfrak{m}}_{k[\kappa],1}$. Further, $\mathfrak{c}_{l,1;l,1} = \mathbb{E}_e(\sum_{\kappa=1}^Kw_{\kappa}\rho_{k[\kappa],1})^2 = \sum_{\kappa=1}^Kw_{\kappa}^2\mathfrak{c}_{k[\kappa],1;k[\kappa],1} + \sum_{\kappa\neq\kappa'}w_{\kappa}w_{\kappa'}\mathbb{E}_e\rho_{k[\kappa],1}\rho_{k[\kappa'],1}$. Consider some specific $\kappa\neq\kappa'$. Due to property \ref{aprop10}, there cannot be a big layer in the big layer graph of $f$ that is touched by both the recursive expansion of $\rho_{k[\kappa],1}$ and $\rho_{k[\kappa'],1}$. Hence, the dependencies of $F_{k[\kappa],1}$ correspond to different big layers in $f$ than the dependencies of $F_{k[\kappa'],1}$. Since there is no weight sharing between big layers and the entries of the covariance matrix of $\mathcal{G}$ corresponding to different big layers in $f$ are zero by construction, $\mathfrak{c}_{l,b;m,b'}=0$ when $F_{l,b}$ and $F_{m,b'}$ are FC or input layers of equal width and $l \neq m$. (Here, $F_{l,b}$ or $F_{m,b'}$ may also refer to $F_{l,b}^\text{in}$ or $F_{m,b'}^\text{in}$.) Hence, we have $\mathbb{E}_e\rho_{k[\kappa],1}\rho_{k[\kappa'],1}=\mathbb{E}_e\rho_{k[\kappa],1}\mathbb{E}_e\rho_{k[\kappa'],1}=\tilde{\mathfrak{m}}_{k[\kappa],1}\tilde{\mathfrak{m}}_{k[\kappa'],1}$. Further, by property \ref{aprop8}, the expansion of at least one of $\rho_{k[\kappa],1}$ and $\rho_{k[\kappa'],1}$ does not touch an activation layer. WLOG let it be $\rho_{k[\kappa],1}$. Assume $\tilde{\mathfrak{m}}_{k[\kappa],1}\neq 0$. By the induction, we must have that $f_{k[\kappa],1}$ is a bias or addition layer with a dependency that also corresponds to an $\tilde{\mathfrak{m}}$ value unequal to 1. Then that dependency must also have one such dependency. This cannot continue forever. So indeed $\tilde{\mathfrak{m}}_{k[\kappa],1}=0$. So $\mathbb{E}_e\rho_{k[\kappa],1}\rho_{k[\kappa'],1}=0$ for any $\kappa \neq \kappa'$, so $\mathfrak{c}_{l,1;l,1} = \sum_{\kappa=1}^Kw_{\kappa}^2\mathfrak{c}_{k[\kappa],1;k[\kappa],1}$. Similarly, $\mathfrak{c}_{l,1;l,2} = \sum_{\kappa=1}^Kw_{\kappa}^2\mathfrak{c}_{k[\kappa],1;k[\kappa],2}$.

Then we have $\ddot{\mathfrak{m}}_{l,1}=\sum_{\kappa=1}^Kw_{\kappa}\ddot{\mathfrak{m}}_{k[\kappa],1}$, $\ddot{\mathfrak{c}}_{l,1;l,1} = \sum_{\kappa=1}^Kw_{\kappa}^2\ddot{\mathfrak{c}}_{k[\kappa],1;k[\kappa],1}$ and $\ddot{\mathfrak{c}}_{l,1;l,2} = \sum_{\kappa=1}^Kw_{\kappa}^2\ddot{\mathfrak{c}}_{k[\kappa],1;k[\kappa],2}$ and thus $\limep\ddot{\mathfrak{m}}_{l,1}=\sum_{\kappa=1}^Kw_{\kappa}\limep\ddot{\mathfrak{m}}_{k[\kappa],1}=\sum_{\kappa=1}^Kw_{\kappa}\mathfrak{m}_k=\mathfrak{m}_l$, $\limep\ddot{\mathfrak{c}}_{l,1;l,1} = \sum_{\kappa=1}^Kw_{\kappa}^2\limep\ddot{\mathfrak{c}}_{k[\kappa],1;k[\kappa],1}=\sum_{\kappa=1}^Kw_{\kappa}^2\mathfrak{q}_{k[\kappa]}=\mathfrak{q}_l$ and $\limep\ddot{\mathfrak{c}}_{l,1;l,2} = \sum_{\kappa=1}^Kw_{\kappa}^2\limep\ddot{\mathfrak{c}}_{k[\kappa],1;k[\kappa],2}=\sum_{\kappa=1}^Kw_{\kappa}^2\mathfrak{c}_{k[\kappa]}=\mathfrak{c}_l$.

Now assume that there is no directed path from an activation layer to $f_l$ not containing an FC layer. Then there is no directed path from an activation layer to any $f_{k[\kappa]}$ not containing an FC layer. We have

\begin{eqnarray*}
&&\limbas \mathbb{E}_b (\rho_{l,b} - \widehat{\rho_l})^2\\
&=&\limbas \mathbb{E}_b \Big(\sum_{\kappa=1}^Kw_\kappa(\rho_{k[\kappa],b} - \widehat{\rho_{k[\kappa]}})\Big)^2\\
&=&\sum_{\kappa}w_\kappa^2\limbas \mathbb{E}_b (\rho_{k[\kappa],b} - \widehat{\rho_{k[\kappa]}})^2\\
&& + \sum_{\kappa\neq \kappa'}w_\kappa w_{\kappa'}\limbas \mathbb{E}_b (\rho_{k[\kappa],b} - \widehat{\rho_{k[\kappa]}})(\rho_{k[\kappa'],b} - \widehat{\rho_{k[\kappa']}})
\end{eqnarray*}

The induction hypothesis yields $\limbas \mathbb{E}_b (\rho_{k[\kappa],b} - \widehat{\rho_{k[\kappa]}})^2=\ddot{\mathfrak{c}}_{k[\kappa],1;k[\kappa],1}-\ddot{\mathfrak{c}}_{k[\kappa],1;k[\kappa],2}$. Further, the recursive expansion of $\rho_{k[\kappa],1} - \widehat{\rho_{k[\kappa]}}$ is as for the BN layer case above, and so for any $\kappa \neq \kappa'$

\begin{eqnarray*}
&&\limbas \mathbb{E}_b (\rho_{k[\kappa],b} - \widehat{\rho_{k[\kappa]}})(\rho_{k[\kappa'],b} - \widehat{\rho_{k[\kappa']}})\\
&=&\limbas \mathbb{E}_b \Big(\sum_{f_{l'}} c_{l'}\bigg(\prod_{f_{l''}\in P_{l',l}}\frac{1}{\sqrt{\nu_{l''}}}\bigg)(e_{l',b}-\widehat{e_{l'}})\Big)...\\
&&...\Big(\sum_{f_{\lambda}} c_{\lambda}\bigg(\prod_{f_{l''}\in P_{\lambda,l}}\frac{1}{\sqrt{\nu_{l''}}}\bigg)(e_{\lambda,b}-\widehat{e_{\lambda}})\Big)\\
&=&\sum_{f_{l'}}\sum_{f_{\lambda}}c_{l'}c_{\lambda}\frac{\limbas \mathbb{E}_b(e_{l',b}-\widehat{e_{l'}})(e_{\lambda,b}-\widehat{e_{\lambda}})}{\bigg(\prod_{f_{l''}\in P_{l',l}}\frac{1}{\sqrt{\ddot{\nu}_{l''}}}\bigg)\bigg(\prod_{f_{l''}\in P_{\lambda,l}}\frac{1}{\sqrt{\ddot{\nu}_{l''}}}\bigg)}
\end{eqnarray*}

Here, $\sum_{f_{l'}}$ is over the input and FC big layers from which there exists a directed path to $f_{k[\kappa]}$ that does not contain another FC layer. $\sum_{f_{\lambda}}$ is the analogous sum for $f_{k[\kappa']}$. Because of property \ref{aprop10}, there is no big layer that appears in both sums and so the $e_{l',b}$ are independent of the $e_{\lambda,b}$. Hence we continue the above derivation as follows.

\begin{eqnarray*}
&=&\sum_{f_{l'}}\sum_{f_{\lambda}}c_{l'}c_{\lambda}\frac{\limbas \mathbb{E}_b(u_{l',b}-\widehat{u_{l'}})(u_{\lambda,b}-\widehat{u_{\lambda}})}{\bigg(\prod_{f_{l''}\in P_{l',l}}\frac{1}{\sqrt{\ddot{\nu}_{l''}}}\bigg)\bigg(\prod_{f_{l''}\in P_{\lambda,l}}\frac{1}{\sqrt{\ddot{\nu}_{l''}}}\bigg)}\\
&& \text{where } u_{l',b} \sim \mathcal{N}(0, \mathfrak{c}_{l',1;l',1}-\mathfrak{c}_{l',1;l',2}), u_{\lambda,b} \sim \mathcal{N}(0, \mathfrak{c}_{\lambda,1;\lambda,1}-\mathfrak{c}_{\lambda,1;\lambda,2})\\
&&\text{for } 1 \le b \le B\\
&=&\sum_{f_{l'}}\sum_{f_{\lambda}}c_{l'}c_{\lambda}\frac{\limbas \widehat{u_{l'}u_{\lambda}}-\limbas\widehat{u_{l'}}\limbas\widehat{u_{\lambda}}}{\bigg(\prod_{f_{l''}\in P_{l',l}}\frac{1}{\sqrt{\ddot{\nu}_{l''}}}\bigg)\bigg(\prod_{f_{l''}\in P_{\lambda,l}}\frac{1}{\sqrt{\ddot{\nu}_{l''}}}\bigg)}\\
\end{eqnarray*}

By the strong law of large numbers, all three limits are zero, so $\limbas \mathbb{E}_b (\rho_{k[\kappa],b} - \widehat{\rho_{k[\kappa]}})(\rho_{k[\kappa'],b} - \widehat{\rho_{k[\kappa']}})=0$, so $\limbas \mathbb{E}_b (\rho_{l,b} - \widehat{\rho_l})^2=\sum_{\kappa=1}^Kw_\kappa^2(\ddot{\mathfrak{c}}_{k[\kappa],1;k[\kappa],1}-\ddot{\mathfrak{c}}_{k[\kappa],1;k[\kappa],2})=\ddot{\mathfrak{c}}_{l,1;l,1}-\ddot{\mathfrak{c}}_{l,1;l,2}$ as required.

\item Non-singularity: As the addition weights are non-zero, we have $\ddot{\mathfrak{c}}_{l,1;l,1}-\ddot{\mathfrak{c}}_{l,1;l,2}=\sum_{\kappa}w_\kappa^2(\ddot{\mathfrak{c}}_{k[\kappa],1;k[\kappa],1}-\ddot{\mathfrak{c}}_{k[\kappa],1;k[\kappa],2}) > 0$ and $\ddot{\mathfrak{c}}_{l,1;l,2}=\sum_{\kappa}w_\kappa^2\ddot{\mathfrak{c}}_{k[\kappa],1;k[\kappa],2} \ge 0$.
\end{itemize}

Case: $f_l$ is an activation layer.

\begin{itemize}
\item Control: The recursive expansion yields the following form for $\rho_{l,1}$.

{\fontsize{8.5}{1}
$$\rho_{l,1} = \tau_l(\rho_{k,1}) = \tau_l\Bigg(\sum_{f_{l'} \text{ normed}} c_{l'}\bigg(\prod_{f_{l''}\in P_{l',l}}\frac{1}{\sqrt{\nu_{l''}}}\bigg)(e_{l',1}-\widehat{e_{l'}})+ \sum_{f_{l'}\text{ unnormed}}c_{l'} e_{l',1} + \sum_{f_{l'}\text{ bias}}c_{l'} e_{l',1}^\text{in}\Bigg)$$
}

The first sum is the same as for the BN layer case above, except that it is restricted to FC and input big layers for which $P_{l',l}$ contains a BN big layer. The second sum is over FC and input big layers for which $P_{l',l}$ exists and does not contain a BN layer and the third sum is over bias layers for which $P_{l',l}$ exists and does not contain a BN layer and for which $\sigma_{l'}^2 > 0$. The $c_{l'}$ are the products of addition weights as before, which are equal to 1 if there is no addition layer in $P_{l',l}$.

The $\nu_{l''}$ are bounded below by $\epsilon_{l''}$. Hence, $\rho_{k,1}$ is controlled by $\mathcal{C}^1$. $\tau_l$ is controlled by $\mathcal{C}^\text{E2}$ by property \ref{aprop2}. So $\tau_l(\rho_{k,1}) \le \tau_l(c||e||_2 + c') \le e^{(c||e||_2 + c')^{2-c''} + c'''}$ for some $c, c', c'', c''' > 0$, which is controlled by $\mathcal{C}^\text{E2}$.

\item Limits: We'll apply lemma \ref{lemma18} five times: (i) on $\mathbb{E}_e\rho_{k,1}^2$, (ii) on $\mathbb{E}_e\rho_{k,1}\rho_{k,2}$, (iii) on $\mathbb{E}_e\rho_{l,1}$, (iv) on $\mathbb{E}_e\rho_{l,1}^2$ and (v) on $\mathbb{E}_e\rho_{l,1}\rho_{l,2}$. The first two are analogous to the BN layer case above and yield

$$\ddot{\mathfrak{c}}_{k,1;k,1}=\mathbb{E}_u\Big(\sum_{f_{l'}\text{ normed}} \frac{c_{l'}u_{l',1}}{\prod_{f_{l''}\in P_{l',l}}\sqrt{\ddot{\nu}_{l''}}}+\sum_{f_{l'} \text{ unnormed}}c_{l'} u_{l',1} + \sum_{f_{l'} \text{ bias}}c_{l'} u_{l'}^\text{in}\Big)^2$$

$$\text{ where } \begin{cases}u_{l',1} \sim \mathcal{N}(0,\ddot{\mathfrak{c}}_{l',1;l',1}-\ddot{\mathfrak{c}}_{l',1;l',2}) \text{ for sum 1}\\u_{l',1} \sim \mathcal{N}(0,\ddot{\mathfrak{c}}_{l',1;l',1})  \text{ for sum 2}\\u_{l'}^\text{in} \sim \mathcal{N}(0,\sigma_{l'}^2)  \text{ for sum 3} \end{cases}$$

and

$$\ddot{\mathfrak{c}}_{k,1;k,2}=\mathbb{E}_u\Bigg(\Big(\sum_{f_{l'}\text{ normed}} \frac{c_{l'}u_{l',1}}{\prod_{f_{l''}\in P_{l',l}}\sqrt{\ddot{\nu}_{l''}}}+\sum_{f_{l'} \text{ unnormed}}c_{l'} u_{l',1} + \sum_{f_{l'} \text{ bias}}c_{l'} u_{l'}^\text{in}\Big)...$$

$$...\Big(\sum_{f_{l'}\text{ normed}} \frac{c_{l'}u_{l',2}}{\prod_{f_{l''}\in P_{l',l}}\sqrt{\ddot{\nu}_{l''}}}+\sum_{f_{l'} \text{ unnormed}}c_{l'} u_{l',2} + \sum_{f_{l'} \text{ bias}}c_{l'} u_{l'}^\text{in}\Big)\Bigg)$$

$$ \text{ where } \begin{cases}u_{l',1},u_{l',2} \sim \mathcal{N}\Bigg(0,\begin{pmatrix}\ddot{\mathfrak{c}}_{l',1;l',1}-\ddot{\mathfrak{c}}_{l',1;l',2}&0\\0&\ddot{\mathfrak{c}}_{l',1;l',1}-\ddot{\mathfrak{c}}_{l',1;l',2}\end{pmatrix}\Bigg)\text{ for sum 1}\\u_{l',1},u_{l',2} \sim \mathcal{N}\Bigg(0,\begin{pmatrix}\ddot{\mathfrak{c}}_{l',1;l',1}& \ddot{\mathfrak{c}}_{l',1;l',2}\\\ddot{\mathfrak{c}}_{l',1;l',2}&\ddot{\mathfrak{c}}_{l',1;l',1}\end{pmatrix}\Bigg)  \text{ for sum 2}\\u_{l'}^\text{in}  \sim \mathcal{N}(0,\sigma_{l'}^2)  \text{ for sum 3}  \end{cases}$$

\sloppy Again, the components of $\chi_d$ and $\hat{\chi}$ correspond to the different $u$ terms. By non-singularity, $\hat{\Sigma}$ is positive definite. For (ii), note that the $u_{l'}^\text{in}$ do not depend on the batch index. Because $e_{l',b}^\text{in}$ takes the same value for each batch index, all the $e_{l',b}^\text{in}$ are represented by a single component of $\chi_d$ / $\omega_d^G$. So $G(\mathsf{arg})$ becomes $(\sum_{i \text{ odd or bias}}\mathsf{arg}[i])(\sum_{i \text{ even or bias}}\mathsf{arg}[i])$. (iii) and (vi) are analogous to the setup of (i) where $G(\mathsf{arg})=\tau_l(\sum_i\mathsf{arg}[i])$ and $G(\mathsf{arg}) = \tau_l(\sum_i\mathsf{arg}[i])^2$ respectively. Differentiability follows from property \ref{aprop2} and integrability from assumption \ref{assumptionIntegrableMeanField}. This yields $\ddot{\mathfrak{m}}_{l,1} = \mathbb{E}_u\tau_l(...)$ and $\ddot{\mathfrak{c}}_{l,1;l,1} = \mathbb{E}_u\tau_l(...)^2$ where the expression inside the parentheses is an in (i) above. Because that expression is a linear combination of zero mean Gaussians, we can combine the above results to obtain $\ddot{\mathfrak{m}}_{l,1} = \mathbb{E}_{s\sim\mathcal{N}(0,\ddot{\mathfrak{c}}_{k,1;k,1})}\tau_l(s)$ and $\ddot{\mathfrak{c}}_{l,1;l,1} = \mathbb{E}_{s\sim\mathcal{N}(0,\ddot{\mathfrak{c}}_{k,1;k,1})}\tau_l(s)^2 = \mathfrak{C}_{\tau_l}(\ddot{\mathfrak{c}}_{k,1;k,1},\ddot{\mathfrak{c}}_{k,1;k,1})$.

Finally, for (v) we use the setup of (ii) with $G(\mathsf{arg}) = \tau_l(\sum_{i \text{ odd or bias}}\mathsf{arg}[i])\tau_l(\sum_{i \text{ even or bias}}\mathsf{arg}[i])$. This yields the same result as in (ii) except that $\tau_l$ is applied to each of the two large parentheses. Combining this with previous results, we obtain $\ddot{\mathfrak{c}}_{l,1;l,2}= \mathfrak{C}_{\tau_l}(\ddot{\mathfrak{c}}_{k,1;k,1},\ddot{\mathfrak{c}}_{k,1;k,2})$. Finally, by lemma \ref{lemma15}, we have $\limep\ddot{\mathfrak{m}}_{l,1}=\mathbb{E}_{s\sim\mathcal{N}(0,\mathfrak{q}_k)}\tau_l(s)=\mathfrak{m}_l$, $\limep\ddot{\mathfrak{c}}_{l,1;l,1} = \mathfrak{C}_{\tau_l}(\mathfrak{q}_k,\mathfrak{q}_k)=\mathfrak{q}_l$ and $\limep\ddot{\mathfrak{c}}_{l,1;l,2} = \mathfrak{C}_{\tau_l}(\mathfrak{q}_k,\mathfrak{c}_k)=\mathfrak{c}_l$ as required.

Because the path containing just $f_l$ is a directed path from an activation layer to $f_l$, $\limbas \mathbb{E}_b (\rho_{l,b} - \widehat{\rho_l})^2$ is immaterial.

\item Non-singularity: We have $\ddot{\mathfrak{c}}_{l,1;l,1}-\ddot{\mathfrak{c}}_{l,1;l,2}= \mathfrak{C}_{\tau_l}(\ddot{\mathfrak{c}}_{k,1;k,1},\ddot{\mathfrak{c}}_{k,1;k,1})-\mathfrak{C}_{\tau_l}(\ddot{\mathfrak{c}}_{k,1;k,1},\ddot{\mathfrak{c}}_{k,1;k,2}) > 0$ by proposition \ref{covkerPositive} and $\ddot{\mathfrak{c}}_{l,1;l,2}=\mathfrak{C}_{\tau_l}(\ddot{\mathfrak{c}}_{k,1;k,1},\ddot{\mathfrak{c}}_{k,1;k,2}) \ge 0$ by proposition \ref{covkerPositive2}.
\end{itemize}

\underline{Stage 10}: statement (\ref{eqn52p4})

We proceed along the lines of stage 2. The statement trivially holds for $l=0$. For $l>0$, we consider the 2-duplex of the architecture constructed for stage 9, where we write $F'$ and $F''$ for the sub-architectures as in stage 2. $\mathcal{G}$ is as in stage 4, except that there are $2B$ instead of $N$ components for each $f_l$. It is Gaussian with mean zero and a covariance matrix as follows. An entry corresponding to layers $F_{l,b}^\mathsf{prime1}$ and $F_{l',b'}^\mathsf{prime2}$ deriving from readin layers is 0 if $l \neq l'$; $q\sigma_l^2$ if $l=l'$, $b=b'$ and $\mathsf{prime1}=\mathsf{prime2}$; and $c\sigma_l^2$ otherwise. An entry corresponding to $F_{l,b}^{\mathsf{prime1}\text{ in}}$ and $F_{l',b'}^{\mathsf{prime2}\text{ in}}$ deriving from bias layers is 0 if $l \neq l'$ and $\sigma_l^2$ otherwise. An entry corresponding to $F_{l,b}^{\mathsf{prime1}}$ and $F_{l',b'}^{\mathsf{prime2}\text{ in}}$ is zero. $\mathsf{prime1}$ and $\mathsf{prime2}$ stand for $'$ or $''$. Distribution equality follows as before. Control follows from stage 9. What remains is to investigate the limits. As in stage 9, $F$ and $\mathcal{G}$ are symmetric with respect to the batch index. Specifically, the joint distribution of $\rho_{l,b}'$ and $\rho_{l,b'}''$ is independent of both $b$ and $b'$, so it is enough to investigate e.g. $\mathbb{E}_e\rho_{l,1}'\rho_{l,1}''$, which is the same as e.g. $\mathbb{E}_e\rho_{l,1}'\rho_{l,2}''$. We have $\mathfrak{c}^{\prime;\prime}_{l,1;l,1}=\mathfrak{c}^{\prime\prime;\prime\prime}_{l,1;l,1}=\mathfrak{c}_{l,1;l,1}$, where the value of $\ddot{\mathfrak{c}}_{l,1;l,1}$ carries over from the mean field architecture constructed in stage 9 for the same $f$. The induction below will show $\ddot{\mathfrak{c}}^{\prime;\prime\prime}_{l,1;l,1}=\ddot{\mathfrak{c}}_{l,1;l,2}$. This automatically yields $\ddot{\mathfrak{c}}^{\prime;\prime}_{l,1;l,1} - \ddot{\mathfrak{c}}^{\prime;\prime\prime}_{l,1;l,1}=\ddot{\mathfrak{c}}^{\prime\prime;\prime\prime}_{l,1;l,1} - \ddot{\mathfrak{c}}^{\prime;\prime\prime}_{l,1;l,1} > 0$ and $\ddot{\mathfrak{c}}^{\prime;\prime\prime}_{l,1;l,1} \ge 0$, i.e. non-singularity, as well as $\limep \ddot{\mathfrak{c}}^{\prime;\prime\prime}_{l,1;l,1}=\mathfrak{c}_l$. Like stage 9, this stage is independent of $N$ so we can use $\limbn$ and $\limb$ interchangeably.

\begin{itemize}
\item Case: $f_l$ is a readin layer. From table \ref{tableBackgroundPropagation}, we obtain $\mathfrak{c}^{\prime;\prime\prime}_{l,1;l,1}=c\sigma_l^2$ and so $\ddot{\mathfrak{c}}^{\prime;\prime\prime}_{l,1;l,1}=c\sigma_l^2=\ddot{\mathfrak{c}}_{l,1;l,2}$.
\item Case: $f_l$ is a fully-connected, non-readin layer. $\mathfrak{c}^{\prime;\prime\prime}_{l,1;l,1}=\sigma_l^2\mathfrak{c}^{\prime;\prime\prime}_{k,1;k,1}$, and so $\ddot{\mathfrak{c}}^{\prime;\prime\prime}_{l,1;l,1}=\sigma_l^2\ddot{\mathfrak{c}}^{\prime;\prime\prime}_{k,1;k,1}=\sigma_l^2\ddot{\mathfrak{c}}_{k,1;k,2}=\ddot{\mathfrak{c}}_{l,1;l,2}$.
\item Case: $f_l$ is a bias layer. We have $\mathfrak{c}^{\prime;\prime\prime}_{l,1;l,1} = \mathbb{E}_e(e_{l,1}^{\prime\text{in}} + \rho_{k,1}')(e_{l,1}^{\prime\prime\text{in}} + \rho_{k,1}'') = \mathfrak{c}^{\prime;\prime\prime}_{k,1;k,1} + \sigma_l^2$ and so $\ddot{\mathfrak{c}}^{\prime;\prime\prime}_{l,1;l,1}=\ddot{\mathfrak{c}}^{\prime;\prime\prime}_{k,1;k,1} + \sigma_l^2=\ddot{\mathfrak{c}}_{k,1;k,2} + \sigma_l^2=\ddot{\mathfrak{c}}_{l,1;l,2}$.
\item Case: $f_l$ is an BN layer. $\mathfrak{c}^{\prime;\prime\prime}_{l,1;l,1}$ is obtained analogously to $\mathfrak{c}_{l,1;l,2}$ via lemma \ref{lemma18} in stage 9, except that the components of $\chi_d$ are now pairs of $e'_{l',1}-\widehat{e'_{l'}}$ and $e''_{l',1}-\widehat{e''_{l'}}$ instead of $e_{l',1}-\widehat{e_{l'}}$ and $e_{l',2}-\widehat{e_{l'}}$ (The $'$ in the subscript has nothing to do with the $'$ superscript.) This yields off-diagonal entries of zero for $\Sigma_d$ but the same diagonal entries and hence the same $\hat{\Sigma}$. So we obtain the same value for $\mathfrak{c}^{\prime;\prime\prime}_{l,1;l,1}$ as we did for $\mathfrak{c}_{l,1;l,2}$.
\item Case: $f_l$ is an addition layer. We have $\mathfrak{c}^{\prime;\prime\prime}_{l,1;l,1} = \sum_{\kappa=1}^Kw_{\kappa}^2\mathfrak{c}_{k[\kappa],1;k[\kappa],1}^{\prime;\prime\prime} + \sum_{\kappa\neq\kappa'}w_{\kappa}w_{\kappa'}\mathbb{E}_e\rho'_{k[\kappa],1}\rho''_{k[\kappa'],1}$. Analogously to stage 9, $\rho'_{k[\kappa],1}$ and $\rho''_{k[\kappa],1}$ have inputs that originate from different big layers in $f$ and only one of the two recursive expansions can hit an activation layer, so the latter sum is again zero. Hence, $\ddot{\mathfrak{c}}^{\prime;\prime\prime}_{l,1;l,1}=\sum_{\kappa=1}^Kw_{\kappa}^2\ddot{\mathfrak{c}}^{\prime;\prime\prime}_{k[\kappa],1;k[\kappa],1}=\sum_{\kappa=1}^Kw_{\kappa}^2\ddot{\mathfrak{c}}_{k[\kappa],1;k[\kappa],2}=\ddot{\mathfrak{c}}_{l,1;l,2}$ as required.
\item Case: $f_l$ is an activation layer. Again, $\mathfrak{c}^{\prime;\prime\prime}_{l,1;l,1}$ is obtained analogously to $\mathfrak{c}_{l,1;l,2}$ via lemma \ref{lemma18} in stage 9, except that the components of $\chi_d$ are now pairs of $e'_{l',1}-\widehat{e'_{l'}}$ and $e''_{l',1}-\widehat{e''_{l'}}$ for $f_{l'}$ normed, pairs of $e'_{l',1}$ and $e''_{l',1}$ for $f_{l'}$ unnormed, as well as the $e^{\prime\text{in}}_{l',1}$. Again, $\hat{\Sigma}$ is the same as before. So we obtain the same value for $\mathfrak{c}^{\prime;\prime\prime}_{l,1;l,1}$ as we did for $\mathfrak{c}_{l,1;l,2}$.
\end{itemize}

\underline{Stage 11}: statements (\ref{eqn52p6}), (\ref{eqn52p7}) and (\ref{eqn52p8})

We have 

\begin{eqnarray*}
&&\limep\limbn\limd\frac{1}{d_l^\text{MF}}\mathbb{E}_{X}||f_{l,1}(X)||^2_2
\\
&=&\limep\limbn\limd\frac{1}{N}\sum_{n=1}^N\overline{f_{l,1}(X^{(n)})^{.2}}\\
&=&\limep\limbn\frac{1}{N}\sum_{n=1}^N \limd\overline{f_{l,1}(X^{(n)})^{.2}}\\
&=&\limep\limbn\frac{1}{N}\sum_{n=1}^N\begin{cases}\mathfrak{c}_{l,1;l,1} \text{ if } l>0\\q \text{ else}\end{cases}\\
&=&\limep\limbn\begin{cases}\mathfrak{c}_{l,1;l,1} \text{ if } l>0\\q \text{ else}\end{cases}\\
&=&\mathfrak{q}_l
\end{eqnarray*}

as required. We have

\begin{eqnarray*}
&&\limep\limbn\limd\frac{1}{d_l^\text{MF}}||\mathbb{E}_{X}f_{l,1}(X)||^2_2\\
&=&\limep\limbn\limd\frac{1}{d_l^\text{MF}}||\frac{1}{N}\sum_{n=1}^Nf_{l,1}(X^{(n)})||^2_2\\
&=&\limep\limbn\limd\frac{1}{N^2}\sum_{n,n'=1}^N\overline{f_{l,1}(X^{(n)}).f_{l,1}(X^{(n')})}\\
&=&\limep\limbn\frac{1}{N^2}\sum_{n,n'=1}^N\limd\overline{f_{l,1}(X^{(n)}).f_{l,1}(X^{(n')})}\\
&=&\limep\limbn\begin{cases}\frac{1}{N^2}(N\mathfrak{c}_{l,1;l,1} + N(N-1)\mathfrak{c}_{l,1;l,1}^{\prime;\prime\prime}) \text{ if } l>0\\\frac{1}{N^2}(Nq + N(N-1)c) \text{ else}\end{cases}\\
&=&\mathfrak{c}_l
\end{eqnarray*}

as required. We have

\begin{eqnarray*}
&&\limep\limbn\limd\frac{1}{d_l^\text{MF}}||\mathbb{S}_{X}f_{l,1}(X)||^2_2\\
&=&\limep\limbn\limd\frac{1}{d_l^\text{MF}}\mathbb{E}_{X}||f_{l,1}(X)||^2_2 - \frac{1}{d_l^\text{MF}}||\mathbb{E}_{X}f_{l,1}(X)||^2_2 \\
&=&\limep\limbn\limd\frac{1}{d_l^\text{MF}}\mathbb{E}_{X}||f_{l,1}(X)||^2_2 - \limep\limbn\limd\frac{1}{d_l^\text{MF}}||\mathbb{E}_{X}f_{l,1}(X)||^2_2 \\
&=&\mathfrak{q}_l - \mathfrak{c}_l
\end{eqnarray*}

as required.

\underline{Stage 12}: statement (\ref{eqn52p10})

\sloppy We proceed along the lines of stage 4. We begin with the case $l=m$. We have $\frac{1}{d_l^\text{MF}}\mathbb{E}_X\Tr(\mathcal{J}_{l,m,1}\Cov_{f_m}\mathcal{J}_{l,m,1}^T)=\frac{1}{d_m^\text{MF}}\mathbb{E}_X\Tr((f_{m,1}(X)-\mathbb{E}_{X'}f_{m,1}(X'))(f_{m,1}(X)-\mathbb{E}_{X'}f_{m,1}(X'))^T)=\frac{1}{d_m^\text{MF}}||\mathbb{S}_Xf_{m,1}||^2_2$. So by stage 11, we have $\limep\limbn \limd \frac{1}{d_l^\text{MF}}\mathbb{E}_X\Tr(\mathcal{J}_{l,m}\Cov_{f_m}\mathcal{J}_{l,m}^T)=\mathfrak{q}_m - \mathfrak{c}_m = \frac{\mathfrak{g}_l(\mathfrak{q}_m - \mathfrak{c}_m)}{\mathfrak{g}_m}$ as required.

For the remainder of this stage, assume $l>m$. Without loss of generality, we can also assume that $f_m$ is a bottleneck for $f_L$ in addition to being a bottleneck for $f_l$. Otherwise, we can simply apply the argument from this stage to the sub-architecture composed of $f_l$ and its ancestors, which is a valid A-architecture, to obtain statement (\ref{eqn52p10}) for the pair $(m,l)$.

We have 

\begin{eqnarray*}
&&\frac{1}{d_l^\text{MF}}\mathbb{E}_X\Tr(\mathcal{J}_{l,m,1}\Cov_{f_m}\mathcal{J}_{l,m,1}^T)\\
&=&\frac{1}{d_l^\text{MF}}\frac{1}{N^2}\sum_{n^-,n^+=1}^N\Tr\Big(\mathcal{J}_{l,m,1}(X^{(n^+)})(f_m(X^{(n^-)})-\frac{1}{N}\sum_{n=1}^Nf_m(X^{(n)}))^T...\\
&&...(f_m(X^{(n^-)})-\frac{1}{N}\sum_{n=1}^Nf_m(X^{(n)}))\mathcal{J}_{l,m,1}(X^{(n^+)})^T\Big)\\
&=&\frac{1}{d_l^\text{MF}}\frac{1}{N^2}\sum_{n^-,n^+=1}^N||(f_m(X^{(n^-)})-\frac{1}{N}\sum_{n=1}^Nf_m(X^{(n)}))\mathcal{J}_{l,m,1}(X^{(n^+)})^T||_2^2\\
&=&\frac{1}{N^2}\sum_{n^-,n^+=1}^N\overline{\Big((f_m(X^{(n^-)})-\frac{1}{N}\sum_{n=1}^Nf_m(X^{(n)}))\mathcal{J}_{l,m,1}(X^{(n^+)})^T\Big)^{.2}}\\
\end{eqnarray*}

Denote $(f_m(X^{(n^-)})-\frac{1}{N}\sum_{n=1}^Nf_m(X^{(n)}))\mathcal{J}_{l,m,1}(X^{(n^+)})^T$ by $p_{l,1}^{(n^-,n^+)}$. Then we have

$$\limep \limbn \limd \frac{1}{d_l^\text{MF}}\mathbb{E}_X\Tr(\mathcal{J}_{l,m,1}\Cov_{f_m}\mathcal{J}_{l,m,1}^T) = \limep \limbn \frac{1}{N^2}\sum_{n^-,n^+=1}^N \limd \overline{p_{l,1}^{(n^-,n^+)}.p_{l,1}^{(n^-,n^+)}}$$

As in stage 4, we conduct one induction per value of $m$ to obtain $ \limd \overline{p_{l,1}^{(n^-,n^+)}.p_{l,1}^{(n^-,n^+)}}$, $\limbn \limd \overline{p_{l,1}^{(n^-,n^+)}.p_{l,1}^{(n^-,n^+)}}$ and $\limep \limbn \limd \overline{p_{l,1}^{(n^-,n^+)}.p_{l,1}^{(n^-,n^+)}}$. Fix $m$ for the rest of the stage. For now, also fix $n^-$ and $n^+$.

Let's look at the case $m>0$. We construct $F$ as in stage 10, except that we construct the $N$-duplex of the $F$ from stage 9 instead of the 2-duplex. We write $(F^{(1)}, .., F^{(N)})$ for that $N$-duplex as in stage 4, though we note that the $F^{(n)}$ now themselves resemble a $B$-duplex and carry a batch index, as opposed to stage 4. As in stage 4, we use $\lambda$ to denote a layer index that ranges from 1 to $m$. $\mathcal{G}$ is constructed as in stage 10, except that $\mathsf{super1}$ and $\mathsf{super2}$ now range from 1 to $N$. We obtain control and distribution equality as before. By symmetry, we have that $\tilde{\mathfrak{m}}_{\lambda,1}^{(n)} = \tilde{\mathfrak{m}}_{\lambda,1}$, $\mathfrak{c}_{\lambda,1;\lambda,1}^{(n);(n)} = \mathfrak{c}_{\lambda,1;\lambda,1}$, $\mathfrak{c}_{\lambda,1;\lambda,2}^{(n);(n)} = \mathfrak{c}_{\lambda,1;\lambda,2}$ and $\mathfrak{c}_{\lambda,1;\lambda,1}^{(n);(n')} = \mathfrak{c}^{\prime;\prime\prime}_{\lambda,1;\lambda,1}$ for $n \neq n'$, where the right-hand sides stem from stages 9 and 10. By symmetry, we also have that those four limit quantities capture all combinations of batch indices, similar to stage 10. The limit of the right-hand sides as $N,B \rightarrow \infty$ is as in prior stages: $\ddot{\mathfrak{m}}_{\lambda,1}$, $\ddot{\mathfrak{c}}_{\lambda,1;\lambda,1}$, $\ddot{\mathfrak{c}}_{\lambda,1;\lambda,2}$ and $\ddot{\mathfrak{c}}_{\lambda,1;\lambda,2}$ respectively.

Next, we add layers $F^\text{MN}_b$ as in stage 4 with $\rho^\text{MN}_b = \rho_{m,b}^{(n^-)} - \frac{1}{N}\sum_{n=1}^N\rho^{(n)}_{m,b}$. Distribution equality follows as in stage 4 and control follows as for the addition layer in stage 9. As in stage 4, we have $\tilde{\mathfrak{m}}^\text{MN}_1=0$ and $\mathfrak{c}^{\text{MN};\text{MN}}_{1;1}=(1-\frac{1}{N})(\mathfrak{c}_{m,1;m,1} - \mathfrak{c}_{m,1;m,1}^{\prime;\prime\prime})$. We also have

$$\mathfrak{c}^{\text{MN};\text{MN}}_{1;2}=\mathbb{E}_e(\rho_{m,1}^{(n-)}-\frac{1}{N}\sum_{n=1}^N\rho_{m,1}^{(n)})(\rho_{m,2}^{(n-)}-\frac{1}{N}\sum_{n=1}^N\rho_{m,2}^{(n)})=(1-\frac{1}{N})(\mathfrak{c}_{m,1;m,2} - \mathfrak{c}_{m,1;m,1}^{\prime;\prime\prime})$$

By symmetry, these three limit quantities cover all possible combinations of batch indices. When it comes to the joint distribution of $\rho^\text{MN}_b$ and $\rho_{m,b}^{(n)}$, the cases $b=b'$ and $b\neq b'$ are distinct because in $\rho^\text{MN}_b$ there is interaction between the $\rho^{(n)}_{m,b}$ for each $b$. Hence we consider both $(b,b') = (1,1)$ and $(b,b') = (1,2)$ here, as well as going forward when building upon the $F^\text{MN}_b$. We obtain $\mathfrak{c}^{\text{MF};(n)}_{1;m,1} = (\mathbbm{1}_{n=n^-}-\frac{1}{N})(\mathfrak{c}_{m,1;m,1} - \mathfrak{c}_{m,1;m,1}^{\prime;\prime\prime})$ as in stage 4 and

$$\mathfrak{c}^{\text{MF};(n)}_{1;m,2}=\mathbb{E}_e(\rho_{m,1}^{(n-)}-\frac{1}{N}\sum_{n'=1}^N\rho_{m,1}^{(n')})\rho_{m,2}^{(n)}=(\mathbbm{1}_{n=n^-}-\frac{1}{N})(\mathfrak{c}_{m,1;m,2} - \mathfrak{c}_{m,1;m,1}^{\prime;\prime\prime})$$

In terms of $\limbn$ quantities, we have $\ddot{\mathfrak{m}}^\text{MN}_1=0$, $\ddot{\mathfrak{c}}^{\text{MN};\text{MN}}_{1;1}=\ddot{\mathfrak{c}}_{m,1;m,1} - \ddot{\mathfrak{c}}_{m,1;m,2}$, $\ddot{\mathfrak{c}}^{\text{MF};(n)}_{1;m,1} = \mathbbm{1}_{n=n^-}(\ddot{\mathfrak{c}}_{m,1;m,1} - \ddot{\mathfrak{c}}_{m,1;m,2})$ and $\ddot{\mathfrak{c}}^{\text{MN};\text{MN}}_{1;2}=\ddot{\mathfrak{c}}^{\text{MF};(n)}_{1;m,2}=0$.

After adding the $F^\text{MN}_b$, we also extend $F^{(n^+)}$ beyond layers $F^{(n^+)}_{m,b}$ as we did in stage 9. This corresponds to forward-propagating $X^{(n^+)}$ beyond $f_m$. We denote all of $F^{(n^+)}$ also simply by $F^+$. Control follows as in stage 9.

Now let's turn to the case $m=0$. In this scenario, we initialize our mean field architecture $F$ to be $F^{(n^+)}$, usually shortened to $F^+$, as obtained by applying stage 9 to $f$ and $X^{(n^+)}$. 

Going forward, we treat the general case where $m$ can be zero or non-zero. For each $l$, we want to add a layer to $F$ for each $b$ that represents $p_{l,b}^{(n^-,n^+)}$. We denote this layer by $F_{l,b}^-$. If $f_l$ is not a BN layer, we do this as in stage 4, except that we now add $B$ layers per $f_l$. Each of them depends only on other layers in $F$ with the same batch index. 

If $f_l$ is a readin layer, the $F_{l,b}^-$ are generated independently of each other and as $F_l^-$ in stage 4. Distribution equality follows similarly to stage 4. We have $p_{l,b}^{(n^-,n^+)}=(x^{(n^-,b)}-\frac{1}{N}\sum_{n=1}^Nx^{(n,b)})W_l$ and $f_{l,b}(X^{(n^+)}) = x^{(n^+,b)}W_l$. Both are mean zero elementwise Gaussians. Further

\begin{eqnarray*}
&&\mathbb{E}_{W_l}p_{l,b}^{(n^-,n^+)}[i]p_{l,b'}^{(n^-,n^+)}[i]\\
&=&\sigma_l^2\overline{(x^{(n^-,b)}-\frac{1}{N}\sum_{n=1}^Nx^{(n,b)}).(x^{(n^-,b')}-\frac{1}{N}\sum_{n=1}^Nx^{(n,b')})} \\
&=&\mathbbm{1}_{b=b'}(1-\frac{1}{N})(q-c)\sigma_l^2
\end{eqnarray*}

and

\begin{eqnarray*}
&&\mathbb{E}_{W_l}p_{l,b}^{(n^-,n^+)}[i]f_{l,b'}(X^{(n^+)})[i]\\
&=&\sigma_l^2\overline{(x^{(n^-,b)}-\frac{1}{N}\sum_{n=1}^Nx^{(n,b)}).x^{(n^+,b')}}\\
&=&\mathbbm{1}_{b=b'}(\mathbbm{1}_{n^-=n^+}-\frac{1}{N})(q-c)\sigma_l^2
\end{eqnarray*}

for all $i$. This corresponds to our choice of generator.

If $f_l$ is a BN layer, we add elementwise layers $F_{l,b}^-$, each with the same width as any of the $f_{l,b}$. We set

$$\rho_{l,b}^- = \frac{\rho_{k,b}^- - \widehat{\rho_k^-}}{\sqrt{\nu_l^+}} - \frac{(\widehat{\rho_{k}^-\rho_{k}^+} - \widehat{\rho_{k}^-}\widehat{\rho_{k}^+})(\rho_{k,b}^+ - \widehat{\rho_{k}^+})}{\sqrt{\nu_l^+}^3}$$

where $\nu_l^+=\mathbb{E}_{b}(\rho_{k,b}^+ - \widehat{\rho_{k}^+})^2+\epsilon_l$ as in stage 9. The inputs are those occurring in any of the $\rho_{k,b}^-$ or $\rho_{k,b}^+$. It is easy to check that $F_{l,b}^-=\sum_{b'=1}^BF_{k,b'}^-J_{l,k,b,b'}^T$, where $J_{l,k,b,b'}$ is the matrix obtained by evaluating $\frac{df_{l,b}(f_{k,1}, .., f_{k,B})}{df_{k,b'}}$ at $(F_{k,1}^+, .., F_{k,B}^+)$. Hence, distribution equality holds.

So we are left to show limits and control. With regards to limits, we show that (i) $\tilde{\mathfrak{m}}^-_{l,1}$, $\mathfrak{c}^{-;-}_{l,1;l,1}$, $\mathfrak{c}^{-;-}_{l,1;l,2}$, $\mathfrak{c}^{-;+}_{l,1;l,1}$ and $\mathfrak{c}^{-;+}_{l,1;l,2}$, which are valid because of background theorem \ref{backgroundMaster}, are dependent on $n^+$ and $n^-$ only via $\mathbbm{1}_{n^-=n^+}$, which is fixed; (ii) $\ddot{\mathfrak{m}}^-_{l,1}$, $\ddot{\mathfrak{c}}^{-;-}_{l,1;l,1}$, $\ddot{\mathfrak{c}}^{-;-}_{l,1;l,2}$, $\ddot{\mathfrak{c}}^{-;+}_{l,1;l,1}$ and $\ddot{\mathfrak{c}}^{-;+}_{l,1;l,2}$ are valid for $n^-=n^+$ and $n^-\neq n^+$; (iii) $\ddot{\mathfrak{c}}^{-;+}_{l,1;l,1}|_{n^-\neq n^+}=\ddot{\mathfrak{c}}^{-;-}_{l,1;l,2}|_{n^-\neq n^+}=\ddot{\mathfrak{c}}^{-;+}_{l,1;l,2}|_{n^-\neq n^+}=0$; and (iv) $\limep\ddot{\mathfrak{c}}^{-;-}_{l,1;l,1}=\frac{\mathfrak{g}_l(\mathfrak{q}_m-\mathfrak{c}_m)}{\mathfrak{g}_m}$. As in stage 4, in a slight abuse of notation, if $f_m$ with $m>0$ is the dependency of $f_l$, we write $F_{k,b}^-$ for $F^\text{MN}_b$. Hence, to start off the induction when $m>0$, we use that $\tilde{\mathfrak{m}}^\text{MF}_{1}$, $\mathfrak{c}^{\text{MF};\text{MF}}_{1;1}$, $\mathfrak{c}^{\text{MF};\text{MF}}_{1;2}$, $\mathfrak{c}^{\text{MF};+}_{1;m,1}$ and $\mathfrak{c}^{\text{MF};+}_{1;m,2}$ are dependent on $n^+$ and $n^-$ only via $\mathbbm{1}_{n^-=n^+}$; that $\ddot{\mathfrak{m}}^\text{MF}_{1}$, $\ddot{\mathfrak{c}}^{\text{MF};\text{MF}}_{1;1}$, $\ddot{\mathfrak{c}}^{\text{MF};\text{MF}}_{1;2}$, $\ddot{\mathfrak{c}}^{\text{MF};+}_{1;m,1}$ and $\ddot{\mathfrak{c}}^{\text{MF};+}_{1;m,2}$ are valid; that $\ddot{\mathfrak{c}}^{\text{MF};+}_{1;m,1}|_{n^-\neq n^+}=\ddot{\mathfrak{c}}^{\text{MF};\text{MF}}_{1;2}=\ddot{\mathfrak{c}}^{\text{MF};+}_{1;m,2}=0$; and that $\limep\ddot{\mathfrak{c}}^{\text{MF};\text{MF}}_{1;1}=\limep\ddot{\mathfrak{c}}_{m,1;m,1} - \limep\ddot{\mathfrak{c}}_{m,1;m,2} =\mathfrak{q}_m-\mathfrak{c}_m=\frac{\mathfrak{g}_m(\mathfrak{q}_m-\mathfrak{c}_m)}{\mathfrak{g}_m}$.

Case: $f_l$ is a readin layer.

\begin{itemize}
\item Control: NA
\item Limits: From table \ref{tableBackgroundPropagation}, we obtain $\tilde{\mathfrak{m}}_{l,1}^-=0$, $\mathfrak{c}_{l,1;l,1}^{-;-}=(1-\frac{1}{N})(q-c)\sigma_l^2$, $\mathfrak{c}_{l,1;l,2}^{-;-}=0$, $\mathfrak{c}_{l,1;l,1}^{-;+}=(\mathbbm{1}_{n^-=n^+}-\frac{1}{N})(q-c)\sigma_l^2$ and $\mathfrak{c}_{l,1;l,2}^{-;+}=0$. $\ddot{\mathfrak{m}}_{l,1}$, $\ddot{\mathfrak{c}}_{l,1;l,1}^{-;-}$, $\ddot{\mathfrak{c}}_{l,1;l,2}^{-;-}$, $\ddot{\mathfrak{c}}_{l,1;l,1}^{-;+}$ and $\ddot{\mathfrak{c}}_{l,1;l,2}^{-;+}$ are identical except the $\frac{1}{N}$ term becomes zero. (i) through (iii) hold. Finally, since $f_l$ is a readin layer and $f_m$ a bottleneck, we have $m=0$. So via table \ref{tableNLCPropagation} we have $\limep \ddot{\mathfrak{c}}_{l,1;l,1}^{-;-} = (q-c)\sigma_l^2=\frac{\mathfrak{g}_l(\mathfrak{q}_0-\mathfrak{c}_0)}{\mathfrak{g}_0}$ so (iv).
\end{itemize}

Case: $f_l$ is a fully-connected, non-readin layer.

\begin{itemize}
\item Control: NA
\item Limits: From table \ref{tableBackgroundPropagation}, we obtain $\tilde{\mathfrak{m}}_{l,1}^-=0$, $\mathfrak{c}_{l,1;l,1}^{-;-}=\sigma_l^2\mathfrak{c}_{k,1;k,1}^{-;-}$, $\mathfrak{c}_{l,1;l,2}^{-;-}=\sigma_l^2\mathfrak{c}_{k,1;k,2}^{-;-}$, $\mathfrak{c}_{l,1;l,1}^{-;+}=\sigma_l^2\mathfrak{c}_{k,1;k,1}^{+;-}$ and $\mathfrak{c}_{l,1;l,2}^{-;+}=\sigma_l^2\mathfrak{c}_{k,1;k,2}^{+;-}$. $\ddot{\mathfrak{m}}_{l,1}$, $\ddot{\mathfrak{c}}_{l,1;l,1}^{-;-}$, $\ddot{\mathfrak{c}}_{l,1;l,2}^{-;-}$, $\ddot{\mathfrak{c}}_{l,1;l,1}^{-;+}$ and $\ddot{\mathfrak{c}}_{l,1;l,2}^{-;+}$ have analogous formulas. (i) through (iii) carry over from $f_k$ to $f_l$. Finally, $\limep \ddot{\mathfrak{c}}_{l,1;l,1}^{-;-}|_{n^-\neq n^+} = \sigma_l^2 \limep \ddot{\mathfrak{c}}_{k,1;k,1}^{-;-}|_{n^-\neq n^+} = \sigma_l^2\frac{\mathfrak{g}_k(\mathfrak{q}_m-\mathfrak{c}_m)}{\mathfrak{g}_m}=\frac{\mathfrak{g}_l(\mathfrak{q}_m-\mathfrak{c}_m)}{\mathfrak{g}_m}$ so (iv).
\end{itemize}

Case: $f_l$ is a bias layer.

\begin{itemize}
\item Control: Since $\rho_{k,1}$ is controlled by $\mathcal{C}^\text{E2}$, so is $\rho_{l,1}$.
\item Limits: $\tilde{\mathfrak{m}}_{l,1}^-=\tilde{\mathfrak{m}}_{k,1}^-$, $\mathfrak{c}_{l,1;l,1}^{-;-}=\mathfrak{c}_{k,1;k,1}^{-;-}$, $\mathfrak{c}_{l,1;l,2}^{-;-}=\mathfrak{c}_{k,1;k,2}^{-;-}$, $\mathfrak{c}_{l,1;l,1}^{+;-} = \mathbb{E}_e(\rho_{k,1}^-(\rho_{k,1}^+ + e_{k,1}^{+\text{in}}))= \mathbb{E}_e\rho_{k,1}^-\rho_{k,1}^+ + \mathbb{E}_e\rho_{k,1}^- e_{k,1}^{+\text{in}}=\mathfrak{c}_{k,1;k,1}^{+;-} + 0 = \mathfrak{c}_{k,1;k,1}^{+;-}$ and similarly $\mathfrak{c}_{l,1;l,2}^{+;-} = \mathfrak{c}_{k,1;k,2}^{+;-}$. The $\limbn$ quantities have analogous formulas. (i) through (iii) carry over from $f_k$ to $f_l$ and (iv) is trivial.
\end{itemize}

Case: $f_l$ is an BN layer.

\begin{itemize}
\item Control: Expanding $\rho^-_{l,1}$ recursively yields that it can be expressed as a polynomial over $e_{l',b}^-$, $e_{l',b}^+$, $e_{\lambda,b}^{(n)}$, $\frac{1}{\sqrt{\nu_{l'}^+}}$ and $\frac{1}{\sqrt{\nu_{\lambda}^{(n)}}}$ terms for various $b$, $n$, $l'$ with $l > l' > m$ and $\lambda \le m$. Throughout the BN layer case, we restrict $l'$ to $l > l' > m$. Note that $e_{l',b}^{+\text{in}}$ and $e_{\lambda,b}^{(n)\text{ in}}$ terms cancel out due to the mean normalization at BN layers and the $F^\text{MN}_b$. The $\frac{1}{\sqrt{\nu_{l'}^+}}$ and $\frac{1}{\sqrt{\nu_{\lambda}^{(n)}}}$ are bounded above in absolute value by $\frac{1}{\sqrt{\epsilon_{l'}}}$ and $\frac{1}{\sqrt{\epsilon_{\lambda}}}$ respectively. So $\rho^-_{l,1}$ is controlled by $\mathcal{C}^d$, where $d$ is the largest degree of any polynomial term counting only the $e_{l',b}^-$, $e_{l',b}^+$ and $e_{\lambda,b}^{(n)}$. Hence, it is controlled by $\mathcal{C}^\text{E2}$.
\item Limits: {\it Joint distribution of inputs.} We begin by examining the joint distribution of the inputs of $\rho^-_{l,1}$. These correspond to the $e_{l',b}^-$, $e_{l',b}^+$ and $e_{\lambda,b}^{(n)}$. Of course, they are jointly Gaussian. All $\tilde{\mathfrak{m}}$ values at input / FC layers in $F$ are zero by construction / by table \ref{tableBackgroundPropagation} respectively, so the mean of this Gaussian is zero. The entries of the covariance matrix corresponding to inputs with different big layer index are also zero, as there is no weight sharing in $f$ across big layers. This leaves the investigation of the covariance matrices corresponding to individual big layer index values as $n$, $b$ and the $+$ / $-$ superscript varies. We have two cases. In each case, we will re-express the Gaussian vector in terms of simpler Gaussian vectors and scalars that are all independent of each other.

Case: $(e_{l',1}^-, .., e_{l',B}^-, e_{l',1}^+, .., e_{l',B}^+)$. The corresponding covariance matrix has size $2B \times 2B$. The top left quadrant has diagonal entries $\mathfrak{c}^{-;-}_{l',1;l',1}$ and off-diagonal entries $\mathfrak{c}^{-;-}_{l',1;l',2}$. The lower right quadrant has diagonal entries $\mathfrak{c}_{l',1;l',1}$ and off-diagonal entries $\mathfrak{c}_{l',1;l',2}$. The other two quadrants have diagonal entries $\mathfrak{c}^{+;-}_{l',1;l',1}$ and off-diagonal entries $\mathfrak{c}^{+;-}_{l',1;l',2}$. Hence, we can decompose the second half of the vector as follows.

\begin{eqnarray*}
&&c^{\{1\}}_{l'}u_{l'}^++ c^{\{2\}}_{l'}s_{l'}^+ \text{ where}\\
&&c^{\{1\}}_{l'}=\sqrt{\mathfrak{c}_{l',1;l',1}-\mathfrak{c}_{l',1;l',2}}\\
&&c^{\{2\}}_{l'}=\sqrt{\frac{B-1}{B}\mathfrak{c}_{l',1;l',2} +  \frac{1}{B}\mathfrak{c}_{l',1;l',1}}
\end{eqnarray*}

where $u_{l'}^+$ has mean zero and a covariance matrix with diagonal entries $\frac{B-1}{B}$ and off-diagonal entries $-\frac{1}{B}$, and $s_{l'}^+$ is a unit Gaussian scalar. Then we can decompose the first half of the vector as follows.

\begin{eqnarray*}
&&c^{\{3\}}_{l'}u_{l'}^++ c^{\{4\}}_{l'}s_{l'}^+ + c^{\{5\}}_{l'}u_{l'}^-+ c^{\{6\}}_{l'}s_{l'}^- \text{ where}\\
&&c^{\{3\}}_{l'}=\frac{\mathfrak{c}_{l',1;l',1}^{+;-}-\mathfrak{c}_{l',1;l',2}^{+;-}}{\sqrt{\mathfrak{c}_{l',1;l',1}-\mathfrak{c}_{l',1;l',2}}}\\
&&c^{\{4\}}_{l'}=\frac{\frac{B-1}{B}\mathfrak{c}_{l',1;l',2}^{+;-} + \frac{1}{B}\mathfrak{c}_{l',1;l',1}^{+;-}}{\sqrt{\frac{B-1}{B}\mathfrak{c}_{l',1;l',2} + \frac{1}{B}\mathfrak{c}_{l',1;l',1}}}\\
&&c^{\{5\}}_{l'}=\sqrt{\frac{(\mathfrak{c}_{l',1;l',1}^{-;-}-\mathfrak{c}_{l',1;l',2}^{-;-})(\mathfrak{c}_{l',1;l',1}-\mathfrak{c}_{l',1;l',2})-(\mathfrak{c}_{l',1;l',1}^{+;-}-\mathfrak{c}_{l',1;l',2}^{+;-})^2}{\mathfrak{c}_{l',1;l',1}-\mathfrak{c}_{l',1;l',2}}}\\
&&c^{\{6\}}_{l'}=\sqrt{\frac{\text{\scalebox{0.85}{$(\frac{B-1}{B}\mathfrak{c}_{l',1;l',2}^{-;-} + \frac{1}{B}\mathfrak{c}_{l',1;l',1}^{-;-})(\frac{B-1}{B}\mathfrak{c}_{l',1;l',2} + \frac{1}{B}\mathfrak{c}_{l',1;l',1})-(\frac{B-1}{B}\mathfrak{c}_{l',1;l',2}^{+;-} + \frac{1}{B}\mathfrak{c}_{l',1;l',1}^{+;-})^2$}}}{\frac{B-1}{B}\mathfrak{c}_{l',1;l',2} + \frac{1}{B}\mathfrak{c}_{l',1;l',1}}}\\
\end{eqnarray*}

where $u_{l'}^-$ and $s_{l'}^-$ have the same distribution as $u_{l'}^+$ and $s_{l'}^+$. Let's look at the square roots that arise in the formulas above. Because the covariance matrix is positive semi-definite, the quadratic form defined by the covariance matrix applied to any vector is non-negative. Using this principle on $(0 \times B, 1, -1, 0 \times (B-2))$ yields $\mathfrak{c}_{l',1;l',1}-\mathfrak{c}_{l',1;l',2} \ge 0$. Here, $\times$ refers to the repetition of a vector component. Using this principle on $(0\times B, 1 \times B)$ yields $(B-1)\mathfrak{c}_{l',1;l',2} + \mathfrak{c}_{l',1;l',1} \ge 0$. The Cauchy-Schwartz inequality yields $\frac{B^2}{(B-1)^2}\Big(\mathbb{E}_e(e_{l',1}^- - \widehat{e_{l'}^-})^2\mathbb{E}_e(e_{l',1}^+ - \widehat{e_{l'}^+})^2 - (\mathbb{E}_e(e_{l',1}^- - \widehat{e_{l'}^-})(e_{l',1}^+ - \widehat{e_{l'}^+}))^2\Big) = (\mathfrak{c}_{l',1;l',1}^{-;-}-\mathfrak{c}_{l',1;l',2}^{-;-})(\mathfrak{c}_{l',1;l',1}-\mathfrak{c}_{l',1;l',2})-(\mathfrak{c}_{l',1;l',1}^{+;-}-\mathfrak{c}_{l',1;l',2}^{+;-})^2 \ge 0$ and $\mathbb{E}_e\widehat{e_{l'}^-}^2\mathbb{E}_e\widehat{e_{l'}^+}^2 - (\mathbb{E}_e\widehat{e_{l'}^-}\widehat{e_{l'}^+})^2 = (\frac{B-1}{B}\mathfrak{c}_{l',1;l',2}^{-;-} + \frac{1}{B}\mathfrak{c}_{l',1;l',1}^{-;-})(\frac{B-1}{B}\mathfrak{c}_{l',1;l',2} + \frac{1}{B}\mathfrak{c}_{l',1;l',1})-(\frac{B-1}{B}\mathfrak{c}_{l',1;l',2}^{+;-} + \frac{1}{B}\mathfrak{c}_{l',1;l',1}^{+;-})^2 \ge 0$. So expressions to which square roots are applied are indeed non-negative.

Let's look at denominators that arise in the formulas above. When $\mathfrak{c}_{l',1;l',1}-\mathfrak{c}_{l',1;l',2}=0$, the Cauchy-Schwartz-derived relationship above yields $\mathfrak{c}_{l',1;l',1}^{+;-}-\mathfrak{c}_{l',1;l',2}^{+;-}=0$ and hence we can remove $u_{l'}^+$ from our decompositions, which corresponds to setting $c_{l'}^{\{1\}}=c_{l'}^{\{3\}}=0$ and $c^{\{5\}}_{l'}=\sqrt{\mathfrak{c}_{l',1;l',1}^{-;-}-\mathfrak{c}_{l',1;l',2}^{-;-}}$. When $\frac{B-1}{B}\mathfrak{c}_{l',1;l',2} + \frac{1}{B}\mathfrak{c}_{l',1;l',1}=0$, the Cauchy-Schwartz-derived relationship above yields $\frac{B-1}{B}\mathfrak{c}_{l',1;l',2}^{+;-} + \frac{1}{B}\mathfrak{c}_{l',1;l',1}^{+;-}=0$ and hence we can remove $s_{l'}^+$ from our decompositions, which corresponds to setting $c_{l'}^{\{2\}}=c_{l'}^{\{4\}}=0$ and $c_{l'}^{\{6\}}=\sqrt{\frac{B-1}{B}\mathfrak{c}_{l',1;l',2}^{-;-} + \frac{1}{B}\mathfrak{c}_{l',1;l',1}^{-;-}}$. So we never have to divide by zero.

Finally, let's look at the limits of the $c^{\{.\}}$ with respect to $N$ and $B$. From the induction hypothesis, we have that the individual $\mathfrak{c}^{-;-}$ and $\mathfrak{c}^{+;-}$ terms have valid limits. By non-singularity from stage 9, we have $\ddot{\mathfrak{c}}_{l',1;l',1}-\ddot{\mathfrak{c}}_{l',1;l',2} > 0$. So we immediately obtain that $\ddot{c}^{\{1\}}_{l'}=\limbn c^{\{1\}}_{l'}$, $\ddot{c}^{\{3\}}_{l'}$ and $\ddot{c}^{\{5\}}_{l'}$ have identical formulas to $c^{\{1\}}_{l'}$, $c^{\{3\}}_{l'}$ and $c^{\{5\}}_{l'}$ above, except that two dots are placed above each $\mathfrak{c}$ term and $\ddot{c}^{\{2\}}_{l'}=\sqrt{\ddot{\mathfrak{c}}_{l',1;l',2}}$. From the induction hypothesis we also have $\ddot{c}^{\{3\}}_{l'}|_{n^-\neq n^+}=0$. The Cauchy-Schwartz-derived relationship above yields $c^{\{4\}}_{l'} \le \sqrt{\frac{B-1}{B}\mathfrak{c}_{l',1;l',2}^{-;-} + \frac{1}{B}\mathfrak{c}_{l',1;l',1}^{-;-}}$. Simply ignoring the negative term in the square root similarly yields $c^{\{6\}}_{l'} \le \sqrt{\frac{B-1}{B}\mathfrak{c}_{l',1;l',2}^{-;-} + \frac{1}{B}\mathfrak{c}_{l',1;l',1}^{-;-}}$. So the induction hypothesis yields $\ddot{c}^{\{4\}}_{l'}|_{n^-\neq n^+} = \ddot{c}^{\{6\}}_{l'}|_{n^-\neq n^+} = 0$. (Note that when the denominator of $c^{\{4\}}_{l'}$ and $c^{\{6\}}_{l'}$ is zero, they have ``alternate values'' 0 and $\sqrt{\frac{B-1}{B}\mathfrak{c}_{l',1;l',2}^{-;-} + \frac{1}{B}\mathfrak{c}_{l',1;l',1}^{-;-}}$ respectively. This does not affect the limit.) If $n^-=n^+$, as $\sqrt{\frac{B-1}{B}\mathfrak{c}_{l',1;l',2}^{-;-} + \frac{1}{B}\mathfrak{c}_{l',1;l',1}^{-;-}}$ converges, $c^{\{4\}}_{l'} $ and $c^{\{6\}}_{l'}$ are still bounded for sufficiently large $N$ and $B$, but it is not apparent whether they necessarily converge.

Case: $(e_{\lambda,1}^{(1)}, .., e_{\lambda,B}^{(1)}, e_{\lambda,1}^{(2)}, .., e_{\lambda,B}^{(N)})$. The corresponding covariance matrix has size $NB \times NB$. Diagonal entries are $\mathfrak{c}_{\lambda,1;\lambda,1}$. Off-diagonal entries in the block-diagonal with block size $B$ are $\mathfrak{c}_{\lambda,1;\lambda,2}$. Other entries are $\mathfrak{c}^{\prime;\prime\prime}_{\lambda,1;\lambda,1}$. We can decompose the $n$'th segment of length $B$ as follows.

\begin{eqnarray*}
&&c^{\{1\}}_{\lambda}u_{\lambda}^{(n)} + c^{\{7\}}_{\lambda}v_{\lambda}[n]  + c^{\{8\}}_{\lambda}s_{\lambda}\text{ where}\\
&&c^{\{1\}}_{\lambda}=\sqrt{\mathfrak{c}_{\lambda,1;\lambda,1}-\mathfrak{c}_{\lambda,1;\lambda,2}}\\
&&c^{\{7\}}_{\lambda}=\sqrt{\frac{1}{B}\mathfrak{c}_{\lambda,1;\lambda,1}+\frac{B-1}{B}\mathfrak{c}_{\lambda,1;\lambda,2}-\mathfrak{c}^{\prime;\prime\prime}_{\lambda,1;\lambda,1}}\\
&&c^{\{8\}}_{\lambda}=\sqrt{\frac{1}{BN}\mathfrak{c}_{\lambda,1;\lambda,1}+\frac{B-1}{BN}\mathfrak{c}_{\lambda,1;\lambda,2}+\frac{N-1}{N}\mathfrak{c}^{\prime;\prime\prime}_{\lambda,1;\lambda,1}}\\
\end{eqnarray*}

where $u_{\lambda}$ has mean zero and a covariance matrix with diagonal entries $\frac{B-1}{B}$ and off-diagonal entries $-\frac{1}{B}$, $v_{\lambda}$ is $N$-dimensional, has mean zero and a covariance matrix with diagonal entries $\frac{N-1}{N}$ and off-diagonal entries $-\frac{1}{N}$, and $s_{\lambda}$ is a unit Gaussian scalar. We reuse the $\{1\}$ superscript because the formulas for $c^{\{1\}}_{l'}$ and $c^{\{1\}}_{\lambda}$ are the same. 

Let's look at the square roots. Using the principle of the non-negative quadratic form on $(1, -1, 0 \times (NB-2))$, we obtain $\mathfrak{c}_{\lambda,1;\lambda,1}-\mathfrak{c}_{\lambda,1;\lambda,2} \ge 0$. Using it on $(1\times B, -1\times B, 0 \times B(N-2))$ yields $\frac{1}{B}\mathfrak{c}_{\lambda,1;\lambda,1}+\frac{B-1}{B}\mathfrak{c}_{\lambda,1;\lambda,2}-\mathfrak{c}^{\prime;\prime\prime}_{\lambda,1;\lambda,1} \ge 0$. Using it on $(1\times(NB))$ yields $\frac{1}{BN}\mathfrak{c}_{\lambda,1;\lambda,1}+\frac{B-1}{BN}\mathfrak{c}_{\lambda,1;\lambda,2}+\frac{N-1}{N}\mathfrak{c}^{\prime;\prime\prime}_{\lambda,1;\lambda,1} \ge 0$. The limits are then $\ddot{c}^{\{7\}}_\lambda=\sqrt{\ddot{\mathfrak{c}}_{\lambda,1;\lambda,2}-\ddot{\mathfrak{c}}^{\prime;\prime\prime}_{\lambda,1;\lambda,1}}=0$ and $\ddot{c}^{\{8\}}_\lambda=\sqrt{\ddot{\mathfrak{c}}^{\prime;\prime\prime}_{\lambda,1;\lambda,1}}=\sqrt{\ddot{\mathfrak{c}}_{\lambda,1;\lambda,2}}$ by stage 10.

Note that we consider the component index of $u$ vectors to be a batch index, and hence we consider the component mean $\overline{u}$ and batch mean $\widehat{u}$ interchangeable.

{\it Recursive expansion.} Next, we consider the recursive expansion of $\rho_{l,1}^-$. By expanding only $\rho_{l',b}^-$, $\rho_{b}^\text{MN}$ and $\rho_{\lambda,b}^{(n)}$ terms that do not occur inside a $\nu$, we obtain the following ``first expansion''.

\begin{eqnarray*}
&&\rho_{l,1}^- \\
&=&\sum_{f_{l'}} c_{l'}\Bigg(\prod_{f_{l''}\in P_{l',l}}\frac{1}{\sqrt{\nu_{l''}^+}}\Bigg)(e_{l',1}^--\widehat{e_{l'}^-})\\
&&+\sum_{f_{l'}}\sum_{\mathcal{L} \in \mathsf{indpow}(P_{l',l}), Z=|\mathcal{L}|>0} c_{l'}(-1)^{|Z|}\Bigg(\prod_{f_{l''}\in P_{l',l}}\frac{1}{\sqrt{\nu_{l''}^+}}\Bigg)...\\
&&...\frac{(\widehat{e_{l'}^-\rho_{k_{\mathcal{L}[1]}}^+}-\widehat{e_{l'}^-}\widehat{\rho_{k_{\mathcal{L}[1]}}^+})(\rho^+_{k_{\mathcal{L}[Z]},1}-\widehat{\rho^+_{k_{\mathcal{L}[Z]}}})}{\nu_{\mathcal{L}[Z]}^+}\prod_{z=1}^{|Z|-1}\frac{\widehat{\rho_{k_{\mathcal{L}[z]}}^+\rho_{k_{\mathcal{L}[z+1]}}^+}-\widehat{\rho_{k_{\mathcal{L}[z]}}^+}\widehat{\rho_{k_{\mathcal{L}[z+1]}}^+}}{\nu_{\mathcal{L}[z]}^+}\\
&&+\sum_{f_{\lambda}}\sum_{n=1}^N (\mathbbm{1}_{n=n^-}-\frac{1}{N})c_{\lambda}\Bigg(\prod_{f_{l''}\in P_{m,l}}\frac{1}{\sqrt{\nu_{l''}^+}}\Bigg)\Bigg(\prod_{f_{l''}\in P_{\lambda,m}}\frac{1}{\sqrt{\nu_{l''}^{(n)}}}\Bigg)(e_{\lambda,1}^{(n)}-\widehat{e_{\lambda}^{(n)}})\\
&&+\sum_{f_{\lambda}}\sum_{\mathcal{L} \in \mathsf{indpow}(P_{m,l}), Z=|\mathcal{L}|>0}\sum_{n=1}^N(\mathbbm{1}_{n=n^-}-\frac{1}{N})c_{\lambda}(-1)^{|Z|}\Bigg(\prod_{f_{l''}\in P_{m,l}}\frac{1}{\sqrt{\nu_{l''}^+}}\Bigg)...\\
&&...\Bigg(\prod_{f_{l''}\in P_{\lambda,m}}\frac{1}{\sqrt{\nu_{l''}^{(n)}}}\Bigg)\frac{(\widehat{e_{\lambda}^{(n)}\rho_{k_{\mathcal{L}[1]}}^+}-\widehat{e_{\lambda}^{(n)}}\widehat{\rho_{k_{\mathcal{L}[1]}}^+})(\rho^+_{k_{\mathcal{L}[Z]},1}-\widehat{\rho^+_{k_{\mathcal{L}[Z]}}})}{\nu_{\mathcal{L}[Z]}^+}...\\
&&...\prod_{z=1}^{|Z|-1}\frac{\widehat{\rho_{k_{\mathcal{L}[z]}}^+\rho_{k_{\mathcal{L}[z+1]}}^+}-\widehat{\rho_{k_{\mathcal{L}[z]}}^+}\widehat{\rho_{k_{\mathcal{L}[z+1]}}^+}}{\nu_{\mathcal{L}[z]}^+}
\end{eqnarray*}

Here, as in stage 9, $\sum_{f_{l'}}$ is over input and FC layers $f_{l'}$ from which there exists a directed path to $f_l$ that does not contain another FC layer, and $P_{l',l}$ refers to that path excluding its starting point $f_{l'}$. Now, $l'$ is restricted to $l > l' > m$. $\sum_{f_{\lambda}}$ is the equivalent sum for layer indices less than or equal to $m$. Here, as in stage 9, the $\prod_{f_{l''}\in P}$ are over BN layers in path $P$ and the $c_{l'}$ / $c_\lambda$ are the products of addition weights on $P$, which is 1 if there are no addition weights. Since $f_m$ is a bottleneck for $f_l$, $P_{\lambda,l}$ can be broken down into $P_{\lambda,m}$ and $P_{m,l}$. Since these paths do not contain their starting point, this represents a partitioning of the layers in $P_{\lambda,l}$. 

Here, $\sum_{\mathcal{L} \in \mathsf{indpow}(P), Z=|\mathcal{L}|>0}$ is the sum over the powerset of the set of all layer indices of BN layers in $P$ except for the empty set. $\mathcal{L}$ is a vector of indices that is an element of the powerset where indices are sorted in decreasing order. For example, if path $P$ contains BN layers $f_4$, $f_7$ and $f_{13}$, then $\mathcal{L}$ takes values $(4)$, $(7)$, $(13)$, $(7,4)$, $(13,4)$, $(13,7)$, and $(13,7,4)$. $Z$ is the size of $\mathcal{L}$. $k_{\mathcal{L}[z]}$ is the index of the dependency of $f_{\mathcal{L}[z]}$. $\sum_{\mathcal{L} \in \mathsf{indpow}(P), Z=|\mathcal{L}|>0}$ arises by repeatedly applying the recursive definition of $\rho_{l,b}^-$ for BN layers, which is the sum of two parts. Passing through a BN layer causes any affected term in the recursive expansion to be replaced by two terms, which yields an exponential number of terms in the end. Each of those terms has a signature that determines whether the term corresponds to the left, simple part or the right, complex part of the recursive definition at any BN layer. $\mathcal{L}$ is that signature.

To better understand the first expansion, we recommend deriving it by hand for several example layer graphs. 

We say the first expansion has four ``super terms''. Super term 1 is on the first line, super term 2 is on the second and third line, super term 3 is on the fourth line and super term 4 is on the fifth, sixth and seventh line.

We continue the expansion process. Unfortunately, we cannot give further expansions explicitly as they are simply too complex. We must be content with describing them. The next step will be to replace each $\rho^+$ term (not inside a $\nu^+$) with its expansion in terms of $e^+$ and $\frac{1}{\sqrt{\nu^+}}$ terms. That expansion takes the form of the expansion of $\rho_{k,1}$ in stage 9, activation layer case. The replacement has the following effect. Super terms 1 and 3 are unaffected. Inside the $\sum_{\mathcal{L} \in \mathsf{indpow}(P_{l',l}), Z=|\mathcal{L}|>0}$ and $\sum_{\mathcal{L} \in \mathsf{indpow}(P_{m,l}), Z=|\mathcal{L}|>0}$ sums of super terms 2 and 4, we add $2Z$ additional sums, two per value of $z$, over the big layer indices that correspond to inputs of $\rho_{k_{\mathcal{L}[z]}}^+$. The numerator of the final part of the super term 2 becomes

$$(\widehat{e_{l'}^-e_{k_{l_1}}^+}-\widehat{e_{l'}^-}\widehat{e_{k_{l_1}}^+})(e^+_{k_{l_{2Z}},1}-\widehat{e^+_{k_{l_{2Z}}}})\prod_{z=1}^{|Z|-1}\widehat{e_{k_{l_{2z}}}^+e_{k_{l_{2z+1}}}^+}-\widehat{e_{k_{l_{2z}}}^+}\widehat{e_{k_{l_{2z+1}}}^+}$$

where the $l_1$, .., $l_{2Z}$ are the indices of summation of the $2Z$ additional sums. For the super term 4, we obtain an analogous expression with $e_{\lambda}^{(n)}$ instead of $e_{l'}^-$. Also inside the additional sums, we obtain a product of addition weights / $\frac{1}{\sqrt{\nu^+}}$ terms stemming from addition layers / BN layers between the $\rho^+$ and their inputs respectively. Note that all $e_{l'}^{+\text{in}}$ and $e_{\lambda}^{+\text{in}}$ terms cancel out.

The next step in the expansion process is to replace each $e$ term (not inside a $\nu$) by the applicable $u^-$, $s^-$, $u^+$, $s^+$, $u^{(n)}$, $v$ and $s$. $e$ terms arise only inside of $(\mathsf{expr1} - \mathsf{expr2})$ constructs where $\mathsf{expr1}$ and $\mathsf{expr2}$ differ only in whether / how the batch mean is taken. Hence, the $s^-$, $s^+$, $v$ and $s$ cancel out, as they are constant across the batch. Also, $\mathsf{expr2}$ vanishes as the batch mean of individual $u$ vectors is zero. Hence super term 1 becomes

$$\Bigg(\sum_{f_{l'}} c_{l'}c^{\{5\}}\Bigg(\prod_{f_{l''}\in P_{l',l}}\frac{1}{\sqrt{\nu_{l''}^+}}\Bigg)u_{l'}^-[1]\Bigg)+ \Bigg(\sum_{f_{l'}} c_{l'}c^{\{3\}}\Bigg(\prod_{f_{l''}\in P_{l',l}}\frac{1}{\sqrt{\nu_{l''}^+}}\Bigg)u_{l'}^+[1]\Bigg)$$

We will refer to an instance of the expression inside the first $\sum_{f_{l'}}$ for one $f_{l'}$ as a ``minus term'' and to an instance of the expression inside the second $\sum_{f_{l'}}$ for one $f_{l'}$ as a ``short plus term''. Similarly, we obtain two copies for super term 2. In the first copy, $e_{l'}^-$ becomes $c^{\{5\}}u_{l'}^-$ and in the second, it becomes $c^{\{3\}}u_{l'}^+$. We will refer to an instance of the expression inside the sums of the first copy as a ``minus plus term'' and to an instance of the expression inside the sums of the second copy as a ``long plus term''. In both cases, the $e^+$ are replaced by $c^{\{1\}}u^+$. Note that going forward, we also use $u^+$ to refer to $u^{(n^+)}$ when the big layer index is less than or equal to $m$. Super term 3 becomes

$$\sum_{f_{\lambda}}\sum_{n=1}^N(\mathbbm{1}_{n=n^-}-\frac{1}{N}) c_{\lambda}c^{\{1\}}_\lambda\Bigg(\prod_{f_{l''}\in P_{m,l}}\frac{1}{\sqrt{\nu_{l''}^+}}\Bigg)\Bigg(\prod_{f_{l''}\in P_{\lambda,m}}\frac{1}{\sqrt{\nu_{l''}^{(n)}}}\Bigg)u_{\lambda}^{(n)}[1]$$

We refer to an instance of the expression inside the $\sum_{f_{\lambda}}$ as an ``n-term''. (Note that an n-term includes $\sum_{n=1}^N$.) Finally, in super term 4 the $e^{(n)}_\lambda$ become $c^{\{1\}}_\lambda u_{\lambda}^{(n)}$ and the $e^+$ become $c^{\{1\}} u^{+}$. We refer to an instance of the expression inside all sums except $\sum_{n=1}^N$ as an ``n-plus term''. We use ``pure term'' to refer to either a minus term, short plus term, long plus term, minus plus term, n-term or n-plus term. The expansion of $\rho_{l,1}$ now consists of a sum over a number of pure terms that is fixed as $N$ and $B$ vary. We refer to that sum as the ``second expansion''.

From the second expansion, it is already clear that (i) holds, i.e. that expectations involving $\rho_{l,b}^-$ and $\rho_{l,b}^+$ depend on $n^+$ and $n^-$ only through $\mathbbm{1}_{n^-=n^+}$ as the same holds for the $c^{\{.\}}$ by the induction hypothesis.

Each type of pure term has a similar form, which we call ``standard form''. Specifically, it is a product of up to five things: an ``N-sum'', an ``N-frequency'', a ``multiplier'', a ``normalizer'' and a ``polynomial''. When present, the n-sum takes form $\sum_{n=1}^N$ and comes at the beginning of the pure term. The N-frequency is a function of $N$, $\mathbbm{1}_{n^-=n^+}$, $\mathbbm{1}_{n=n^+}$ and $\mathbbm{1}_{n=n^-}$ that converges as $N$ converges to infinity for any $n$, $n^-$, $n^+$. The multiplier is a product of addition weights, the $c^{\{.\}}$, and some number of $(-1)^{|Z|}$ terms. The normalizer is a product of $\frac{1}{\sqrt{\nu^+}}$ and possibly $\frac{1}{\sqrt{\nu^{(n)}}}$ terms with various subscripts. The polynomial is a product of terms which can include $u_{l'}^-[1]$ terms (e.g. minus term), $u^+[1]$ terms (e.g. short plus term, long plus term and n-plus term), $u^{(n)}_\lambda[1]$ terms (e.g. n-term), $\widehat{u_{l'}^-u^+}$ terms (e.g. minus plus term), $\widehat{u^{(n)}_\lambda u^+}$ terms (e.g. n-plus term), and  $\widehat{u^+u^+}$ terms (e.g. long plus term and n-plus term).

{\it Expectation limits of standard form terms} Let's look at the limit of the expectation of a standard form term as $B$ and $N$ converge to infinity. Consider first the N-sum, assuming it is present. Due to the symmetry of standard form terms, the expectation for a given value of $n$ depends on $n$ only via $\mathbbm{1}_{n=n^+}$ and $\mathbbm{1}_{n=n^-}$. (Remember that when $n=n^+$, we have $u^{(n)}_\lambda = u^+_\lambda$ for $\lambda \le m$.) Hence, we can eliminate the N-sum inside an expectation as follows. If $n^-=n^+$, we can duplicate the standard form term and replace the $(n)$ superscript with $(n^+)$ in one copy and some other fixed $(n^\circ)$ in the other copy, while replacing the N-frequency by its aggregate value. The aggregate N-frequency is obtained by summing all N-frequencies that belong to a given copy. For the first copy, this is only the N-frequency for $n=n^+$. For the second copy, this is all N-frequencies for $n \neq n^+$. For example, for n-terms, the aggregate N-frequency is $\frac{1}{N} - 1$ for the first copy and $1 - \frac{1}{N}$ for the second copy. If $n^-\neq n^+$, we can triplicate the standard form term and replace the $(n)$ superscript with $(n^+)$ in one copy, $(n^-)$ in the second copy and some other fixed $(n^\circ)$ in the third copy, while again replacing the N-frequency by its aggregate value. For example, for n-terms, the aggregate N-frequency is $-\frac{1}{N}$ for the first copy, $1 - \frac{1}{N}$ for the second copy and $-\frac{N-2}{N}$ for the third copy. Assuming the aggregate N-frequencies converge with $N$, all copies have themselves standard form.

Now consider the limit of the expectation of a standard form term without an N-sum. Consider the polynomial. We can multiply it out by eliminating all parentheses and batch mean operators. This yields what we call ``chains'' of individual $u$ components multiplied together, which are also multiplied by some power of $\frac{1}{B}$ stemming from the batch mean operators. $u$ components in these chains have component indices from 1 to B, which correspond to batch indices. However, since the multiplier, normalizer and N-frequency either do not contain or are symmetric with respect to batch indices, we can aggregate chains that are equivalent up to batch index reassignment inside the expectation. For example, $u^+[1](u^+[4])^2u^+[7]$ (times multiplier, normalizer and N-frequency) has the same expectation as $(u^+[1])^2u^+[2]u^+[3]$, and $u^-[1]u^-[B]u^+[1]u^+[B]$ has the same expectation as $u^-[1]u^-[2]u^+[1]u^+[2]$. This aggregation procedure allows us to eliminate all batch index values greater than the degree of the chains, which is fixed as $B$ and $N$ vary. Call this degree $D$. Hence, as $B$ and $N$ vary, there is a finite set of possible ``aggregated chains''. The aggregation process causes each of these aggregated chains to be multiplied by the sum of powers of $\frac{1}{B}$ corresponding to the individual chains that were aggregated into it. This sum may depend on $B$ but converges as $B$ converges to infinity. Hence, we call it ``B-frequency''. So we have that the standard form term is a sum over products of N-frequency, B-frequency, multiplier, normalizer and aggregated chain. By allowing the B-frequency to be zero, we can consider the sum to be over a fixed number of products and we can consider the set of aggregated chains fixed.

Both N-frequency and B-frequency converge in the limit of $N$ and $B$. Unless the multiplier contains a $c^{\{4\}}$ or $c^{\{6\}}$ term and $n^-=n^+$, the multiplier converges as addition weights and $(-1)^{|Z|}$ terms are constant and the $c^{\{.\}}$ converge as shown above. Now we apply lemma \ref{lemma18} to each product of aggregated chain and normalizer, in a way that is similar to stage 9. $B$ maps onto $d$. Each different $u$ component from the aggregated chain maps onto a component of $\chi_d$. For example, if the aggregated chain is $(u^-[1])^2u^-[2](u^+[1])^3u^+[2]$, we obtain $d_G=4$. $\Sigma_d$ is then block-diagonal with blocks of sizes at most $D$ which have $\frac{d-1}{d}$ on the diagonal and $-\frac{1}{d}$ off the diagonal. So $\hat{\Sigma}$ is the identity matrix. $G(\mathsf{arg})$ is a product of $\mathsf{arg}$ components that mirrors the aggregated chain. The normalizer to the power of $\frac{1}{D}$ maps onto each component of $\omega_d^G$. $H$ and $\omega_d^H$ are equal to constant 1. It is easy to check that neither the distribution of an aggregated chain nor of the normalizer depends on $N$, so the limit with respect to $N$ can be ignored. The conditions of lemma \ref{lemma18} hold as for the BN layer case in stage 9. The application of lemma \ref{lemma18} yields that the product of aggregated chain and normalizer has a valid expectation limit as $B$ converges to infinity, and hence as $B$ and $N$ converge to infinity. This limit is obtained by endowing $u$ components with the unit Gaussian distribution and taking the almost sure limit of the normalizer. So finally we have that the valid limit of the expectation of the product of N-frequency, B-frequency, multiplier, normalizer and aggregated chain is equal to the product of the limit of the N-frequency, the limit of the B-frequency, the limit of the multiplier, the almost sure limit of the normalizer and the expectation of the aggregated chain with respect to the unit Gaussian, assuming the multiplier does not contain a $c^{\{4\}}$ or $c^{\{6\}}$ term.

{\it Specific expectation limits}  Now we are ready to investigate the $\limbn$ quantities: $\ddot{\mathfrak{m}}_{l,1}$, $\ddot{\mathfrak{c}}_{l,1;l,1}^{-;+}$, $\ddot{\mathfrak{c}}_{l,1;l,2}^{-;+}$, $\ddot{\mathfrak{c}}_{l,1;l,2}^{-;-}$ and $\ddot{\mathfrak{c}}_{l,1;l,1}^{-;-}$. We will break these down into expectation limits of pure terms / standard form terms. Most of the time, the expectation limit of a standard form term turns out to be zero, for one of the following reasons: the N-frequency converges to zero; the B-frequency of aggregated chains converges to zero; the multiplier converges to zero; or the aggregated chains contain exactly one copy of some $u$ component. In the latter case, because individual $u$ components are independent from other components and have mean zero under the unit Gaussian, the expectation of the entire aggregated chain is zero. 

We have $\tilde{\mathfrak{m}}_{l,1}=\widehat{\tilde{\mathfrak{m}}_l}=0$. The first half of this statement holds by symmetry. The second half uses that the batch mean of any $u$ vector is zero. Hence, $\ddot{\mathfrak{m}}_{l,1}=0$.

Let's look at $\ddot{\mathfrak{c}}_{l,1;l,1}^{-;+}=\limbn \mathbb{E}_e \rho_{l,1}^-\rho_{l,1}^+$. $\rho_{l,1}^+$ recursively expands into a sum of terms containing a multiplier made up of addition weights and a $c^{\{1\}}$ term, a normalizer and a polynomial of form $u^+[1]$. We call those terms ``co-terms'', and they have standard form. So $\rho_{l,1}^-\rho_{l,1}^+$ is the sum of products of one pure term and one co-term. Each product has standard form and does not contain a $c^{\{4\}}$ or $c^{\{6\}}$ term, so the limit of its expectation is valid if an N-sum is not present. An N-sum is only present in a product of an n-term and co-term and a product of an n-plus term and co-term. We remove the N-sum via duplication as described above. Note that because in an n-term and n-plus term only the N-frequency depends on $n^-$, we can aggregate all $n$ values unequal to $n^+$, no matter whether $n^-=n^+$. The aggregate N-frequency is $\frac{1}{N} - \mathbbm{1}_{n^-=n^+}$ for the copy corresponding to $n\neq n^+$ and $\mathbbm{1}_{n^-=n^+}-\frac{1}{N}$ for the copy corresponding to $n= n^+$. Both converge, so both copies have standard form, so the limits of their expectations are valid. Hence overall, $\ddot{\mathfrak{c}}_{l,1;l,1}^{-;+}$ itself is valid.

Consider the value of the limit for the case $n^- \neq n^+$. We break this down by pure term type. For a product of minus term and co-term or minus plus term and co-term, each aggregated chain contains a single $u^-$ component. Hence, the expectation limit is zero. For a product of short plus term and co-term or long plus term and co-term, we have that the multiplier contains $c_{l'}^{\{3\}}$. Since $\ddot{c}^{\{3\}}_{l'}|_{n^-\neq n^+}=0$, the expectation limit is zero. For a product of n-term and co-term or n-plus term and co-term, we obtain an N-frequency of $\frac{1}{N} - \mathbbm{1}_{n^-=n^+}$ / $\mathbbm{1}_{n^-=n^+}-\frac{1}{N}$ as mentioned above. Both converge to zero when $n^- \neq n^+$. So the expectation limit is again zero. So, overall $\ddot{\mathfrak{c}}_{l,1;l,1}^{-;+}|_{n^-\neq n^+}=0$ as required.

We have $\mathfrak{c}_{l,1;l,2}^{-;+}=\mathbb{E}_e\rho_{l,1}^-\rho_{l,2}^+=\mathbb{E}_e\rho_{l,1}^-(\frac{B}{B-1}\widehat{\rho_{l}^+} - \frac{1}{B-1}\rho_{l,1}^+)=- \frac{1}{B-1}\mathbb{E}_e\rho_{l,1}^-\rho_{l,1}^+=- \frac{1}{B-1}\mathfrak{c}_{l,1;l,1}^{-;+}$. Here we use symmetry and that the batch mean is zero at a big BN layer. So $\ddot{\mathfrak{c}}_{l,1;l,2}^{-;+}=\limbn -\frac{1}{B-1}\mathfrak{c}_{l,1;l,1}^{-;+}=\limbn -\frac{1}{B-1}\limbn\mathfrak{c}_{l,1;l,1}^{-;+}=0$, so specifically $\ddot{\mathfrak{c}}_{l,1;l,2}^{-;+}|_{n^-\neq n^+}=0$ as required.

Similarly, we have $\mathfrak{c}_{l,1;l,2}^{-;-}=- \frac{1}{B-1}\mathfrak{c}_{l,1;l,1}^{-;-}$, so if $\ddot{\mathfrak{c}}_{l,1;l,1}^{-;-}$ is valid as we show below, $\ddot{\mathfrak{c}}_{l,1;l,2}^{-;-}=0$ and specifically $\ddot{\mathfrak{c}}_{l,1;l,2}^{-;-}|_{n^-\neq n^+}=0$ as required.

Finally, let's look at $\ddot{\mathfrak{c}}_{l,1;l,1}^{-;-}=\limbn \mathbb{E}_e \rho_{l,1}^-\rho_{l,1}^-$. $\rho_{l,1}^-\rho_{l,1}^-$ is the sum of products of two pure terms. Such a product has standard form, except it may contain two N-sums. We use the same replication trick as before when this is the case. Let the indices of summation corresponding to the two sums be $n$ and $n'$. Then we require one copy for $n=n'=n^+$, one copy for $n=n^+ \neq n'$, one copy for $n'=n^+\neq n$, one copy for $n=n'\neq n^+$, and one copy for when all three are different. The aggregate N-frequencies when $n^-=n^+$ are $\frac{(N-1)^2}{N^2}$, $-\frac{(N-1)^2}{N^2}$, $-\frac{(N-1)^2}{N^2}$, $\frac{(N-1)}{N^2}$ and $\frac{(N-1)(N-2)}{N^2}$ in that order. When $n^-\neq n^+$, they are $\frac{1}{N^2}$, $-\frac{1}{N^2}$, $-\frac{1}{N^2}$, $\frac{N^2-N-1}{N^2}$ and $\frac{-N^2+N+2}{N^2}$. All converge. Since the pure terms do not contain $c^{\{4\}}$ or $c^{\{6\}}$ terms, $\ddot{\mathfrak{c}}_{l,1;l,1}^{-;-}$ is valid.

Consider the value of the limit for the case $n^-\neq n^+$. Again, we go by pure term type.

Case: product of short plus term or long plus term and any pure term. The multiplier contains $c^{\{3\}}$, so the expectation limit is zero.

Case: product of minus term or minus plus term and n-term or n-plus term. Each product of an aggregated chain from the minus or minus plus term and an aggregated chain from a copy of the n-term or n-plus term contains a single $u^-$ component, so the expectation limit is zero.

Case: product of minus term or minus plus term and minus term or minus plus term where the big layer indices from the leading sum $\sum_{f_{l'}}$ are different. Each product of aggregated chains contains two single $u^-$ components, one each for the two big layer indices, so the expectation limit is zero.

Case: product of minus plus term and minus plus term where the big layer indices are the same. Then the polynomial of the product is composed as follows. It has a $\widehat{u^-_{l'}u^+_{l_1}}$ term; a $\widehat{u^-_{l'}u^+_{l_2}}$ term; a $u^+_{l_3}[1]$ term; a $u^+_{l_4}[1]$ term; and some number of $\widehat{u^+u^+}$ terms with various subscripts. The $l_1$ through $l_4$ are less than $l$ and not necessarily different. Let the product of the minus plus terms be written informally as $\mathsf{other} * \mathsf{poly}$. Then

\begin{eqnarray*}
&&\mathbb{E}_u \mathsf{other} * \mathsf{poly}\\
&=&\mathbb{E}_u \mathsf{other}\widehat{u^-_{l'}u^+_{l_1}}\widehat{u^-_{l'}u^+_{l_2}}u^+_{l_3}[1]u^+_{l_4}[1]\prod\widehat{u^+u^+}\\
&=&\mathbb{E}_u \mathsf{other}\widehat{u^-_{l'}u^+_{l_1}}\widehat{u^-_{l'}u^+_{l_2}}\widehat{u^+_{l_3}u^+_{l_4}}\prod\widehat{u^+u^+}\\
&=&\mathbb{E}_u \mathsf{other}(\frac{1}{B}u^-_{l'}[1]u^+_{l_1}[1]u^-_{l'}[1]u^+_{l_2}[1] \\
&&+ \frac{B-1}{B}u^-_{l'}[1]u^+_{l_1}[1]u^-_{l'}[2]u^+_{l_2}[2])\widehat{u^+_{l_3}u^+_{l_4}}\prod\widehat{u^+u^+}\\
\end{eqnarray*}

Here we use symmetry with respect to batch indices repeatedly. Now we form aggregated chains in a 2-step process. First, we aggregate the ``partial chains'' obtained from multiplying out $\widehat{u^+_{l_3}u^+_{l_4}}\prod\widehat{u^+u^+}$, where we do not replace batch indices 1 and 2 and replace no batch index with 1 or 2. (This can lead to batch indices up to $D+2$ being included in the aggregated partial chains. However, this does not affect our ability to apply lemma \ref{lemma18} as discussed above.) Second, we multiply a copy of each aggregated partial chain with $\frac{1}{B}u^-_{l'}[1]u^+_{l_1}[1]u^-_{l'}[1]u^+_{l_2}[1]$ and a copy of each aggregated partial chain with $\frac{B-1}{B}u^-_{l'}[1]u^+_{l_1}[1]u^-_{l'}[2]u^+_{l_2}[2]$. Since the B-frequencies of the aggregated partial chains are convergent, the B-frequencies of aggregated chains stemming from $\frac{1}{B}u^-_{l'}[1]u^+_{l_1}[1]u^-_{l'}[1]u^+_{l_2}[1]$ converge to zero. Further, aggregated chains stemming from $\frac{B-1}{B}u^-_{l'}[1]u^+_{l_1}[1]u^-_{l'}[2]u^+_{l_2}[2]$ contain a single $u^-_{l'}[1]$ and a single $u^-_{l'}[2]$ component. So for either type of aggregated chain, the expectation limit is zero.

Case: product of minus plus term and minus term where the big layer indices are the same. The argument is as in the previous case where 

\begin{eqnarray*}
&&\mathbb{E}_u \mathsf{other} * \mathsf{poly}\\
&=&\mathbb{E}_u \mathsf{other}\widehat{u^-_{l'}u^+_{l_1}}u^-_{l'}[1]u^+_{l_2}[1]\prod\widehat{u^+u^+}\\
&=&\mathbb{E}_u \mathsf{other}(\frac{1}{B}u^-_{l'}[1]u^+_{l_1}[1]u^-_{l'}[1]u^+_{l_2}[1] + \frac{B-1}{B}u^-_{l'}[1]u^+_{l_1}[1]u^-_{l'}[2]u^+_{l_2}[2])\prod\widehat{u^+u^+}\\
\end{eqnarray*}

Case: product of an n-term or n-plus term and an n-term or n-plus term where the big layer indices are different. We consider the 5 copies as described above. When $n=n^+$ and / or $n'=n^+$, the N-frequency converges to zero. Otherwise, each product of aggregated chains contains a single $u^{(n)}$ and a single $u^{(n')}$ component, which are unequal to any $u^+$ component because $n\neq n^+$ and $n'\neq n^+$. The expectation limit is zero.

Case: product of an n-term or n-plus term and n-plus term where the big layer indices are the same. We consider the 5 copies again. When $n=n^+$ and / or $n'=n^+$, the N-frequency converges to zero. Otherwise, when $n \neq n'$, there are single $u^{(n)}$ and $u^{(n')}$ components. When $n=n'$, the expectation equals

$$\mathbb{E}_u \mathsf{other}(\frac{1}{B}u^{(n)}_{\lambda}[1]u^+_{l_1}[1]u^{(n)}_{\lambda}[1]u^+_{l_2}[1] + \frac{B-1}{B}u^{(n)}_{\lambda}[1]u^+_{l_1}[1]u^{(n)}_{\lambda}[2]u^+_{l_2}[2])\prod\widehat{u^+u^+}$$

So the expectations based on individual aggregated chains have limit zero either based on B-frequency or a single $u$ component as above.

Case: product of an n-term and n-term where the big layer indices are the same. We consider for now only 4 of the 5 copies: those where at least one of $n$ and $n'$ equals $n^+$ and the one where all three are unequal. For the first three, the N-frequency converges to zero. When all three are unequal, there is a single $u^{(n)}$ and $u^{(n')}$ component.

So, in summary, we are left with one copy from the case above as well as products of minus terms where the big layer index is the same. Hence, letting $n^\circ$ be some value other than $n^+$ and endowing $u$ components with unit Gaussian distributions, we have for $n^-\neq n^+$

\begin{eqnarray*}
&&\ddot{\mathfrak{c}}_{l,1;l,1}^{-;-}\\
&=&\sum_{f_{l'}} \Bigg(\frac{\limbn c_{l'}^2c^{\{5\}}_{l'}c^{\{5\}}_{l'}}{\limbas \prod_{f_{l''}\in P_{l',l}}\nu_{l''}^+}\Bigg)\mathbb{E}_uu_{l'}^-[1]^2 \\
&&+ \sum_{f_{\lambda}} \Bigg(\frac{\limbn\frac{N^2-N-1}{N^2}c_{\lambda}^2c^{\{1\}}_{\lambda}c^{\{1\}}_{\lambda}}{\limbas\Big(\prod_{f_{l''}\in P_{m,l}}\nu_{l''}^+\Big)\Big(\prod_{f_{l''}\in P_{\lambda,m}}\nu_{l''}^{(n^\circ)}\Big)}\Bigg)\mathbb{E}_uu_{\lambda}^{(n^\circ)}[1]^2\\
&=&\sum_{f_{l'}} \frac{c_{l'}^2\ddot{c}^{\{5\}}_{l'} \ddot{c}^{\{5\}}_{l'}}{\prod_{f_{l''}\in P_{l',l}}\ddot{\nu}_{l''}} + \sum_{f_{\lambda}} \frac{c_{\lambda}^2\ddot{c}^{\{1\}}_{\lambda}\ddot{c}^{\{1\}}_{\lambda}}{\Big(\prod_{f_{l''}\in P_{m,l}}\ddot{\nu}_{l''}\Big)\Big(\prod_{f_{l''}\in P_{\lambda,m}}\ddot{\nu}_{l''}\Big)}\\
&=&\sum_{f_{l'}} \frac{c_{l'}^2\ddot{\mathfrak{c}}_{l',1;l',1}^{-;-}}{\prod_{f_{l''}\in P_{l',l}}\ddot{\nu}_{l''}} + \sum_{f_{\lambda}} \frac{c_{\lambda}^2(\ddot{\mathfrak{c}}_{\lambda,1;\lambda,1}-\ddot{\mathfrak{c}}_{\lambda,1;\lambda,2})}{\prod_{f_{l''}\in P_{\lambda,l}}\ddot{\nu}_{l''}}
\end{eqnarray*}

This is very reminiscent of the corresponding expression from the BN layer case of stage 9. The value of $\ddot{\nu}_{l''}$ carries over from stage 9. Again, using the recursive calculation rules for bias, addition and BN layers in this stage, for $\ddot{\mathfrak{c}}^{\text{MN};\text{MN}}_{1;1}$, and for bias, addition and BN layers in stage 9, it is easy to check that this is equivalent to $\ddot{\mathfrak{c}}_{l,1;l,1}^{-;-} = \frac{\ddot{\mathfrak{c}}_{k,1;k,1}^{-;-}}{\ddot{\mathfrak{c}}_{k,1;k,1}-\ddot{\mathfrak{c}}_{k,1;k,2}+\epsilon_l}$. Hence $\limep \ddot{\mathfrak{c}}_{l,1;l,1}^{-;-}|_{n^-\neq n^+} = \frac{\limep\ddot{\mathfrak{c}}_{k,1;k,1}^{-;-}|_{n^-\neq n^+}}{\limep\ddot{\mathfrak{c}}_{k,1;k,1}|_{n^-\neq n^+}-\limep\ddot{\mathfrak{c}}_{k,1;k,2}|_{n^-\neq n^+}+\limep\epsilon_l}=\frac{\mathfrak{g}_k(\mathfrak{q}_m-\mathfrak{c}_m)}{(\mathfrak{q}_k-\mathfrak{c}_k)\mathfrak{g}_m}=\frac{\mathfrak{g}_l(\mathfrak{q}_m-\mathfrak{c}_m)}{\mathfrak{g}_m}$ as required.
\end{itemize}

Case: $f_l$ is an addition layer.

\begin{itemize}
\item Control: Since $\mathcal{C}^\text{E2}$ is linearly closed, if the $\rho_{k[\kappa],1}^-$ are controlled by $\mathcal{C}^\text{E2}$, so is $\rho_{l,1}^-$.
\item \sloppy Limits: $\tilde{\mathfrak{m}}_{l,1}^- = \mathbb{E}_e \sum_{\kappa=1}^{K}w_{\kappa}\rho_{k[\kappa],1}^- = \sum_{\kappa=1}^Kw_{\kappa}\tilde{\mathfrak{m}}_{k[\kappa],1}^-$. We also have $\mathfrak{c}_{l,1;l,1}^{-;+}=\sum_{\kappa=1}^Kw_{\kappa}^2\mathfrak{c}_{k[\kappa],1;k[\kappa],1}^{-;+}$, $\mathfrak{c}_{l,1;l,2}^{-;+}=\sum_{\kappa=1}^Kw_{\kappa}^2\mathfrak{c}_{k[\kappa],1;k[\kappa],2}^{-;+}$, $\mathfrak{c}_{l,1;l,1}^{-;-}=\sum_{\kappa=1}^Kw_{\kappa}^2\mathfrak{c}_{k[\kappa],1;k[\kappa],1}^{-;-}$ and $\mathfrak{c}_{l,1;l,2}^{-;-}=\sum_{\kappa=1}^Kw_{\kappa}^2\mathfrak{c}_{k[\kappa],1;k[\kappa],2}^{-;-}$, . We use the same recursion argument as in e.g. stage 9. The expansion itself is a bit more complex than in that stage. As it flows backward through the small layer graph of $f$, each $f_{l',b}$ / $f_{\lambda,b}$ is represented in $F$ by $F^-_{l',b}$ and $F^+_{l',b}$ / the $F_{\lambda,b}^{(n)}$ respectively. Hence, we must apply the recursive definitions corresponding to those layers. If the expansion touches $f_m$, we must also apply the recursive definition of the $\rho^\text{MN}_b$. However, the argument still goes through with a wider range of recursion steps.

So, $\ddot{\mathfrak{m}}_{l,1}^- = \sum_{\kappa=1}^Kw_{\kappa}\ddot{\mathfrak{m}}_{k[\kappa],1}^-$, $\ddot{\mathfrak{c}}_{l,1;l,1}^{-;-}=\sum_{\kappa=1}^Kw_{\kappa}^2\ddot{\mathfrak{c}}_{k[\kappa],1;k[\kappa],1}^{-;-}$ etc. so (i) through (iii) carry over from the dependencies to $f_l$. Finally, we have $\limep \ddot{\mathfrak{c}}_{l,1;l,1}^{-;-}|_{n^-\neq n^+}=\sum_{\kappa=1}^Kw_{\kappa}^2\limep\ddot{\mathfrak{c}}_{k[\kappa],1;k[\kappa],1}^{-;-}|_{n^-\neq n^+}=\sum_{\kappa=1}^Kw_{\kappa}^2\frac{\mathfrak{g}_{k[\kappa]}(\mathfrak{q}_m-\mathfrak{c}_m)}{\mathfrak{g}_m}=\frac{\mathfrak{g}_{l}(\mathfrak{q}_m-\mathfrak{c}_m)}{\mathfrak{g}_m}$ and so (iv).
\end{itemize}

Case: $f_l$ is an activation layer.

\begin{itemize}
\item Control: We have $\rho_{l,1}^-=\rho_{k,1}^-\tau_l'(\rho_{k,1}^+)$. $\tau_l'(\rho_{k,1}^+)$ is controlled by $\mathcal{C}^\text{E2}$ as $\tau_l(\rho_{k,1}^+)$ is controlled by $\mathcal{C}^\text{E2}$ in stage 9. The induction hypothesis yields that $\rho_{k,1}^-$ is controlled by $\mathcal{C}^\text{E2}$. The product of two functions controlled by $\mathcal{C}^\text{E2}$ is itself controlled by $\mathcal{C}^\text{E2}$.
\item Limits: We follow the same steps as for the BN layer case of this stage. 

{\it Joint distribution of inputs} The joint distribution of the inputs of $\rho_{l,1}$ is as for the BN layer case, except that $e^{+\text{in}}_{l',1}$ terms for $l > l' > m$ and $e^{+\text{in}}_{\lambda,1}$ terms for $\lambda \le m$ can arise inside of $\tau_l'()$. (Again, we restrict $l'$ to $l > l' > m$.) The $e^{+\text{in}}_{l',1}$ / $e^{+\text{in}}_{\lambda,1}$ are Gaussian scalars independent of all other inputs of $\rho_{l,1}$ with mean zero and variance $\sigma_{l'}^2$ / $\sigma_{\lambda}^2$. Hence, we set $e^{+\text{in}}_{l',1}=\sigma_{l'}^2s^\text{in}_{l'}$ and $e^{+\text{in}}_{\lambda,1}=\sigma_\lambda^2s^\text{in}_{\lambda}$ where the $s^\text{in}_{l'}$ and $s^\text{in}_{\lambda}$ are unit Gaussian scalars. Because the bias vector is shared across individual inputs, those scalars depend neither on the batch index nor on the duplex index $n$.

{\it Recursive expansion} The ``first expansion'' is

\begin{eqnarray*}
&&\rho_{l,1}^- \\
&=&\sum_{f_{l'}\text{ normed}} c_{l'}\Bigg(\prod_{f_{l''}\in P_{l',l}}\frac{1}{\sqrt{\nu_{l''}^+}}\Bigg)(e_{l',1}^--\widehat{e_{l'}^-})\tau_l'(\rho_{k,1}^+)\\
&&+\sum_{f_{l'}\text{ normed}}\sum_{\mathcal{L} \in \mathsf{indpow}(P_{l',l}), Z=|\mathcal{L}|>0} c_{l'}(-1)^{|Z|}\Bigg(\prod_{f_{l''}\in P_{l',l}}\frac{1}{\sqrt{\nu_{l''}^+}}\Bigg)...\\
&&...\frac{(\widehat{e_{l'}^-\rho_{k_{\mathcal{L}[1]}}^+}-\widehat{e_{l'}^-}\widehat{\rho_{k_{\mathcal{L}[1]}}^+})(\rho^+_{k_{\mathcal{L}[Z]},1}-\widehat{\rho^+_{k_{\mathcal{L}[Z]}}})}{\nu_{\mathcal{L}[Z]}^+}...\\
&&...\Bigg(\prod_{z=1}^{|Z|-1}\frac{\widehat{\rho_{k_{\mathcal{L}[z]}}^+\rho_{k_{\mathcal{L}[z+1]}}^+}-\widehat{\rho_{k_{\mathcal{L}[z]}}^+}\widehat{\rho_{k_{\mathcal{L}[z+1]}}^+}}{\nu_{\mathcal{L}[z]}^+}\Bigg)\tau_l'(\rho_{k,1}^+)\\
&&+\sum_{f_{\lambda}\text{ normed}}\sum_{n=1}^N (\mathbbm{1}_{n=n^-}-\frac{1}{N})c_{\lambda}\Bigg(\prod_{f_{l''}\in P_{m,l}}\frac{1}{\sqrt{\nu_{l''}^+}}\Bigg)\Bigg(\prod_{f_{l''}\in P_{\lambda,m}}\frac{1}{\sqrt{\nu_{l''}^{(n)}}}\Bigg)...\\
&&...(e_{\lambda,1}^{(n)}-\widehat{e_{\lambda}^{(n)}})\tau_l'(\rho_{k,1}^+)\\
&&+\sum_{f_{\lambda}\text{ normed}}\sum_{\mathcal{L} \in \mathsf{indpow}(P_{m,l}), Z=|\mathcal{L}|>0}\sum_{n=1}^N(\mathbbm{1}_{n=n^-}-\frac{1}{N})c_{\lambda}(-1)^{|Z|}...\\
&&...\Bigg(\prod_{f_{l''}\in P_{m,l}}\frac{1}{\sqrt{\nu_{l''}^+}}\Bigg)\Bigg(\prod_{f_{l''}\in P_{\lambda,m}}\frac{1}{\sqrt{\nu_{l''}^{(n)}}}\Bigg)...\\
&&...\frac{(\widehat{e_{\lambda}^{(n)}\rho_{k_{\mathcal{L}[1]}}^+}-\widehat{e_{\lambda}^{(n)}}\widehat{\rho_{k_{\mathcal{L}[1]}}^+})(\rho^+_{k_{\mathcal{L}[Z]},1}-\widehat{\rho^+_{k_{\mathcal{L}[Z]}}})}{\nu_{\mathcal{L}[Z]}^+}...\\
&&...\Bigg(\prod_{z=1}^{|Z|-1}\frac{\widehat{\rho_{k_{\mathcal{L}[z]}}^+\rho_{k_{\mathcal{L}[z+1]}}^+}-\widehat{\rho_{k_{\mathcal{L}[z]}}^+}\widehat{\rho_{k_{\mathcal{L}[z+1]}}^+}}{\nu_{\mathcal{L}[z]}^+}\Bigg)\tau_l'(\rho_{k,1}^+)\\
&&+\sum_{f_{l'}\text{ unnormed}} c_{l'}e_{l',1}^-\tau_l'(\rho_{k,1}^+)+\sum_{f_{\lambda}\text{ unnormed}}\sum_{n=1}^N (\mathbbm{1}_{n=n^-}-\frac{1}{N})c_{\lambda}e_{\lambda,1}^{(n)}\tau_l'(\rho_{k,1}^+)\\
\end{eqnarray*}

Here, as in the activation layer case of stage 9, the leading sums are either over input and FC big layers from which the directed path to $f_l$ not containing another FC layer does contain a BN layer, or does not contain a BN layer. The part of the first expansion corresponding to normed $f_{l'}$ and $f_\lambda$ is as for the BN layer case. The part of the first expansion corresponding to unnormed $f_{l'}$ and $f_\lambda$ is much simpler because the recursive definitions of multi-activation functions derived from BN layers are not applied.

The first expansion now has two additional super terms 5 and 6, and super terms 1 through 4 differ in that they now also contain a $\tau_l'(\rho_{k,1}^+)$ term. We continue the expansion process. Replacing $\rho^+$ terms not in a $\tau_l'()$ or $\nu^+$ with their expansion in terms of $e^+$ and $\frac{1}{\sqrt{\nu^+}}$ terms has the same effect as in the BN layer case, i.e. it adds $2Z$ additional sums and some other components to the second and fourth super term. Expanding the $\rho^+_{k,1}$ inside the $\tau_l'()$ replaces them with

$$\sum_{f_{l'} \text{ normed}}c_{l'}\Big(\prod_{f_{l''}\in P_{l',l}}\frac{1}{\sqrt{\nu_{l''}^+}}\Big)(e_{l',1}^+-\widehat{e_{l'}^+}) + \sum_{f_{l'}\text{ unnormed}}c_{l'} e_{l',1}^+ + \sum_{f_{l'}\text{ bias}}c_{l'} e_{l',1}^{+\text{in}}$$
$$+\sum_{f_{\lambda} \text{ normed}}c_{\lambda'}\Big(\prod_{f_{l''}\in P_{\lambda,l}}\frac{1}{\sqrt{\nu_{l''}^+}}\Big)(e_{\lambda,1}^+ -\widehat{e_{\lambda}^+}) + \sum_{f_{\lambda}\text{ unnormed}}c_{\lambda} e_{\lambda,1}^+ + \sum_{f_{\lambda}\text{ bias}}c_{\lambda} e_{\lambda,1}^{+\text{in}}$$

This is equivalent to the expansion of $\rho_{k,1}$ in stage 9, activation layer case, except that we break down each sum into big layers with index larger than $m$ and big layers with index at most $m$.

Now let's look at replacing the $e$ terms with $u$ and $v$ components and $s$ terms. The expression inside $\tau_l'()$ becomes $\mathsf{exp}_{l,1}$ where

\begin{eqnarray*}
\mathsf{exp}_{l,b}&=&\sum_{f_{l'} \text{ normed}}c_{l'}c_{l'}^{\{1\}}\Big(\prod_{f_{l''}\in P_{l',l}}\frac{1}{\sqrt{\nu_{l''}^+}}\Big)u_{l'}^+[b] + \sum_{f_{l'}\text{ unnormed}}c_{l'}c_{l'}^{\{1\}}u_{l'}^+[b] \\
&& + \sum_{f_{l'}\text{ unnormed}}c_{l'}c_{l'}^{\{2\}}s_{l'}^+ + \sum_{f_{l'}\text{ bias}}c_{l'}\sigma_{l'}^2 s_{l'}^\text{in}\\
&&+\sum_{f_{\lambda} \text{ normed}}c_{\lambda'}c_{\lambda}^{\{1\}}\Big(\prod_{f_{l''}\in P_{\lambda,l}}\frac{1}{\sqrt{\nu_{l''}^+}}\Big)u_{\lambda}^+[b]  \\
&&+ \sum_{f_{\lambda}\text{ unnormed}}c_{\lambda} c_{\lambda}^{\{1\}}u_{\lambda}^+[b] + \sum_{f_{\lambda}\text{ unnormed}}c_{\lambda}c_{\lambda}^{\{7\}} v_{\lambda}[n^+] \\
&&+ \sum_{f_{\lambda}\text{ unnormed}}c_{\lambda}c_{\lambda}^{\{8\}} s_{\lambda}+ \sum_{f_{\lambda}\text{ bias}}c_{\lambda} \sigma_{\lambda}^2s_{\lambda}^\text{in}
\end{eqnarray*}

Note that each $u$ or $v$ component or $s$ term that arises in this expression arises in only one sum, and only once in that sum. What is multiplied to it we term its ``coefficient''. For super terms 1 through 4, introducing $u$, $v$ and $s$ outside of the $\tau_l'(\mathsf{exp}_{l,1})$ works exactly as in the BN layer case. We obtain minus terms, minus plus terms, short plus terms, long plus terms, n-terms and n-plus terms as before, except that these terms now also contain $\tau_l'(\mathsf{exp}_{l,1})$. Our two new super terms 5 and 6 become

$$\sum_{f_{l'}\text{ unnormed}} c_{l'}c_{l'}^{\{3\}}u_{l'}^+[1]\tau_l'(\mathsf{exp}_{l,1}) + \sum_{f_{l'}\text{ unnormed}} c_{l'}c_{l'}^{\{4\}}s_{l'}^+\tau_l'(\mathsf{exp}_{l,1})$$

$$ + \sum_{f_{l'}\text{ unnormed}} c_{l'}c_{l'}^{\{5\}}u_{l'}^-[1]\tau_l'(\mathsf{exp}_{l,1}) + \sum_{f_{l'}\text{ unnormed}} c_{l'}c_{l'}^{\{6\}}s_{l'}^-\tau_l'(\mathsf{exp}_{l,1})$$

and

$$\sum_{f_{\lambda}\text{ unnormed}}\sum_{n=1}^N (\mathbbm{1}_{n=n^-}-\frac{1}{N})c_{\lambda}c_{\lambda}^{\{1\}}u_{\lambda}^{(n)}[1]\tau_l'(\mathsf{exp}_{l,1})$$

$$ + \sum_{f_{\lambda}\text{ unnormed}}\sum_{n=1}^N (\mathbbm{1}_{n=n^-}-\frac{1}{N})c_{\lambda}c_{\lambda}^{\{7\}}v_{\lambda}[n]\tau_l'(\mathsf{exp}_{l,1})$$

respectively. $s_\lambda$ cancels out. This yields 6 new kinds of pure term, which we call 3-term, 4-term, 5-term, 6-term, 1-term and 7-term, respectively, for lack of better names, based on the superscript of $c$. Each of these new pure terms refers to an instance of the expression inside the leading sum. We thus obtain the ``second expansion'' as a sum over a fixed number of 12 kinds of pure terms, instead of a sum over a fixed number of 6 kinds of pure terms as for the BN layer case. Based on the second expansion, it is clear that (i) holds.

We need to expand our definition of the ``standard form'' from the BN layer case to cover the pure terms in this second expansion. In addition to N-sum, N-frequency, multiplier, normalizer and polynomial, standard form terms now also contain a ``tau-deriv'' of form $\tau_l'(\mathsf{exp}_{l,1})$. Polynomials are now also allowed to have $s_{l'}^+$ terms (e.g. 4-term),  $s_{l'}^-$ terms (e.g. 6-term) and $v_\lambda[n]$ terms (e.g. 7-term).

{\it Expectation limits of standard form terms} Let's look at the limit of the expectation of a ``new'' standard form term as $N$ and $B$ converge to infinity. N-sums are eliminated and the polynomial is broken down into aggregated chains as for the BN layer case, except that batch index value 1, which arises in $\mathsf{exp}_{l,1}$, is excluded from aggregation. We then apply lemma \ref{lemma18} to each product of normalizer, aggregated chain and tau-deriv. As in the BN layer case, the distribution of normalizer and aggregated chain do not depend on $N$. While the distribution of some terms in $\mathsf{exp}_{l,1}$ such as $v_\lambda[n^+]$ terms do depend on $N$, the distribution of $\mathsf{exp}_{l,1}$ itself does not. Hence, we can map both $B$ and $N$ onto $d$ for lemma \ref{lemma18}, thereby setting $B=N$, and still get the general limit with respect to $B$ and $N$ out of the lemma, as $N$ does not affect the value of the expectation. Each different $u$ component, $v$ component or $s$ term that arises in the aggregated chain or tau-deriv maps onto a component of $\chi_d$. As for the BN layer case, $\hat{\Sigma}$ is the identity matrix, $G(\mathsf{arg})$ is a product of $\mathsf{arg}$ components that mirrors the aggregated chain, and the normalizer to the power of $\frac{1}{D}$ maps onto each component of $\omega_d^G$. Now, the coefficient of the $u$ or $v$ component or $s$ term in $\mathsf{exp}_{l,1}$ corresponding to $\chi_d[i]$ maps onto $\omega^H_d[i]$ for each $i$. $\omega^H_d[i]$ is zero if the component or term does not appear. Finally, $H(\mathsf{arg}) = \tau_l'(\sum_i\mathsf{arg}[i])$. Let's check the conditions. Differentiability holds by property \ref{aprop2} and integrability by assumption \ref{assumptionIntegrableMeanField}. $\hat{\Sigma}$ is positive definite. The components of $\omega_d^G$ are bounded and converge a.s. as for the BN layer case above and in stage 9. The components of $\omega_d^H$ are bounded and converge a.s. because $\frac{1}{\sqrt{\nu^+}}$ terms are bounded and converge a.s. and the $c^{\{.\}}$ represent deterministic, convergent sequences in $B=N$. Note that $c^{\{4\}}$ or $c^{\{6\}}$ terms do not appear in the tau-deriv.

So, analogously to the BN layer case above, we have that the valid limit of the expectation of the product of N-frequency, B-frequency, multiplier, normalizer, aggregated chain and tau-deriv is equal to the product of the limit of the N-frequency, the limit of the B-frequency, the limit of the multiplier, the almost sure limit of the normalizer, and the expectation of the product of aggregated chain and tau-deriv with respect to the unit Gaussian, where the coefficients of the Gaussian variables in the tau-deriv are also replaced by their almost sure limit. As discussed under the BN layer case, the limit of the multiplier may not exist when $n^- = n^+$ and it contains $c^{\{4\}}$ or $c^{\{6\}}$ terms. We consider this specific case separately below.

{\it Specific expectation limits} Again, we are ready to investigate the $\limbn$ quantities. Again, we will break these down into the expectation limits of pure terms / standard form terms, which are most often zero.

Let's look at $\ddot{\mathfrak{m}}_{l,1}^-$. $\rho_{l,1}^-$ is the sum of ``new'' standard form terms. We eliminate each N-sum and obtain aggregate N-frequencies $\frac{1}{N} - \mathbbm{1}_{n^-=n^+}$ and $\mathbbm{1}_{n^-=n^+} - \frac{1}{N}$ for the two copies respectively as for the BN layer case. Expectation limits of all standard form terms, which are now N-sum-free, are valid except possibly those of 4-terms and 6-terms. 6-terms contain a single $s^-_{l'}$ term. (Going forward, ``single'' implies that exactly one copy of the component or term appears in the polynomial and no copy appears in the tau-deriv.) So the expectation limit of the product of normalizer, polynomial and tau-deriv is zero. Since $c^{\{6\}}_{l'}$ is bounded for sufficiently large $N$ and $B$, the multiplier is eventually bounded. So multiplying the zero limit with the eventually bounded multiplier and bounded frequencies still yields a zero limit for the entire 6-term, and hence a valid limit. For 4-terms, we similarly have that $c^{\{4\}}_{l'}$ is eventually bounded. By non-singularity from stage 9, $\ddot{\mathfrak{c}}_{l',1;l',2} \ge 0$. If $\ddot{\mathfrak{c}}_{l',1;l',2} > 0$, then $\ddot{c}^{\{4\}}_{l'}$ is simply the limit of its numerator over the limit of its denominator. A valid $\ddot{c}^{\{4\}}_{l'}$ yields a valid limit of the expectation of the 4-term. On the other hand, if $\ddot{\mathfrak{c}}_{l',1;l',2} = 0$, we have $\ddot{c}^{\{2\}}_{l'}=0$. This implies that $s_{l'}^+$ has a coefficient in $\mathsf{exp}_{l,1}$ with limit zero. Hence, the $s_{l'}^+$ in the polynomial / aggregated chain of the 4-term becomes effectively a single term and we obtain a zero expectation limit as for 6-terms above. So in any case, the expectation limit of the 4-term is valid. So,  overall, $\ddot{\mathfrak{m}}_{l,1}^-$ is valid.

Consider the value of $\ddot{\mathfrak{m}}_{l,1}^-$ for $n^- \neq n^+$. Minus, minus plus terms and 5-terms become zero due to a single $u^-$ component. Short plus terms / long plus terms / 3-terms / 4-terms / 6-terms become zero as $\ddot{c}^{\{3\}}$ / $\ddot{c}^{\{3\}}$ / $\ddot{c}^{\{3\}}$ / $\ddot{c}^{\{4\}}$ / $\ddot{c}^{\{6\}}$ is zero respectively. For n-terms, n-plus terms, 1-terms and 7-terms, the N-frequency of both copies becomes zero. So $\ddot{\mathfrak{m}}_{l,1}^-|_{n^-\neq n^+} = 0$.

Let's look at $\ddot{\mathfrak{c}}_{l,1;l,1}^{-;+}$ and $\ddot{\mathfrak{c}}_{l,1;l,2}^{-;+}$. Multiplying $\rho_{l,1}^+$ to $\rho_{l,1}^-$ corresponds to additionally multiplying each pure term that already contains a $\tau_l'(\mathsf{exp}_{l,1})$ with a $\tau_l(\mathsf{exp}_{l,1})$ as well. Compared to a standard form term, in our application of lemma \ref{lemma18}, we can simply replace $H(\mathsf{arg}) = \tau_l'(\sum_i\mathsf{arg}[i])$ with $H(\mathsf{arg}) = \tau_l'(\sum_i\mathsf{arg}[i])\tau_l(\sum_i\mathsf{arg}[i])$. We obtain that $\ddot{\mathfrak{c}}_{l,1;l,1}^{-;+}$ is valid and $\ddot{\mathfrak{c}}_{l,1;l,1}^{-;+}|_{n^-\neq n^+}=0$ in the same manner as for $\ddot{\mathfrak{m}}_{l,1}^-$ above. Similarly, multiplying $\rho_{l,2}^+$ to $\rho_{l,1}^-$ corresponds to additionally multiplying each pure term that already contains a $\tau_l'(\mathsf{exp}_{l,1})$ with a $\tau_l(\mathsf{exp}_{l,2})$ as well. Every $u$ or $v$ component or $s$ term that arises in both $\mathsf{exp}_{l,1}$ and $\mathsf{exp}_{l,2}$ is paired with the same coefficient in both. Hence, compared to a standard form term, in our application of lemma \ref{lemma18}, we can simply let $H(\mathsf{arg})$ be the product of $\tau_l'$ applied to the sum of components of $\mathsf{arg}$ present in $\mathsf{exp}_{l,1}$ and $\tau_l$ applied to the sum of components of $\mathsf{arg}$ present in $\mathsf{exp}_{l,2}$. We also exclude batch index value 2 from aggregation. We obtain that $\ddot{\mathfrak{c}}_{l,1;l,2}^{-;+}$ is valid and $\ddot{\mathfrak{c}}_{l,1;l,2}^{-;+}|_{n^-\neq n^+}=0$ in the same manner as for $\ddot{\mathfrak{m}}_{l,1}^-$ above.

Let's look at $\ddot{\mathfrak{c}}_{l,1;l,1}^{-;-}$. $\rho_{l,1}^-\rho_{l,1}^-$ is the sum of products of two pure terms. Such products have standard form except that $\tau_l'(\mathsf{exp}_{l,1})^2$ is present instead of $\tau_l'(\mathsf{exp}_{l,1})$, which is a trivial change, and that two N-sums can arise, which are eliminated by creating five copies as for the BN layer case. So any product of two pure terms not containing a 4-term or 6-term has a valid expectation limit. For a product of a 4/6-term and a non-4/6-term, and for a product of two 4/6-terms where the big layer indices from the leading sum $\sum_{f_{l'}}$ are different, we obtain validity as for an individual 4/6-term above. Thus we are left for each unnormed $f_{l'}$ to confirm the validity of

$$\limbn c_{l'}c_{l'}c_{l'}^{\{4\}}c_{l'}^{\{4\}}\mathbb{E}_{u,s,v}s_{l'}^+s_{l'}^+\tau_l'(\mathsf{exp}_{l,1})^2 + 2c_{l'}c_{l'}c_{l'}^{\{4\}}c_{l'}^{\{6\}}\mathbb{E}_{u,s,v}s_{l'}^+s_{l'}^-\tau_l'(\mathsf{exp}_{l,1})^2$$

$$ + c_{l'}c_{l'}c_{l'}^{\{6\}}c_{l'}^{\{6\}}\mathbb{E}_{u,s,v}s_{l'}^-s_{l'}^-\tau_l'(\mathsf{exp}_{l,1})^2$$

If $\ddot{\mathfrak{c}}_{l',1;l',2} > 0$, then $\ddot{c}^{\{4\}}_{l'}$ and $\ddot{c}^{\{6\}}_{l'}$ are simply the limit of their respective numerators over the limit of their respective denominators, and so the above limit is valid. On the other hand, if $\ddot{\mathfrak{c}}_{l',1;l',2} = 0$, we have $\ddot{c}^{\{2\}}_{l'}=0$. This implies that $s_{l'}^+$ has a coefficient in $\mathsf{exp}_{l,1}$ with limit zero. Since $s_{l'}^-$ does not appear in $\mathsf{exp}_{l,1}$ at all, we have

$$\limbn\mathbb{E}_{u,s,v}s_{l'}^+s_{l'}^+\tau_l'(\mathsf{exp}_{l,1})^2=\limbn\mathbb{E}_{u,s,v}s_{l'}^-s_{l'}^-\tau_l'(\mathsf{exp}_{l,1})^2$$

and also

$$\limbn\mathbb{E}_{u,s,v}c_{l'}c_{l'}c_{l'}^{\{4\}}c_{l'}^{\{6\}}s_{l'}^+s_{l'}^-\tau_l'(\mathsf{exp}_{l,1})^2=0$$

due to a single $s_{l'}^+$ term and a single $s_{l'}^-$ term and both $c_{l'}^{\{4\}}$ and $c_{l'}^{\{6\}}$ being eventually bounded. So we obtain

\begin{eqnarray*}
&&\limbn c_{l'}c_{l'}c_{l'}^{\{4\}}c_{l'}^{\{4\}}\mathbb{E}_{u,s,v}s_{l'}^+s_{l'}^+\tau_l'(\mathsf{exp}_{l,1})^2 + 2c_{l'}c_{l'}c_{l'}^{\{4\}}c_{l'}^{\{6\}}\mathbb{E}_{u,s,v}s_{l'}^+s_{l'}^-\tau_l'(\mathsf{exp}_{l,1})^2\\
&& + c_{l'}c_{l'}c_{l'}^{\{6\}}c_{l'}^{\{6\}}\mathbb{E}_{u,s,v}s_{l'}^-s_{l'}^-\tau_l'(\mathsf{exp}_{l,1})^2\\
&=&c_{l'}c_{l'}\Big(\limbn c_{l'}^{\{4\}}c_{l'}^{\{4\}}+c_{l'}^{\{6\}}c_{l'}^{\{6\}}\Big)\Big(\limbn\mathbb{E}_{u,s,v}s_{l'}^-s_{l'}^-\tau_l'(\mathsf{exp}_{l,1})^2\Big)\\
&=&c_{l'}c_{l'}\ddot{\mathfrak{c}}_{l',1;l',2}^{-;-}\Big(\limbn\mathbb{E}_{u,s,v}s_{l'}^-s_{l'}^-\tau_l'(\mathsf{exp}_{l,1})^2\Big)\\
\end{eqnarray*}

The expression on the last line is valid, so the expression on the first line is valid, so overall $\ddot{\mathfrak{c}}_{l,1;l,1}^{-;-}$ is valid.

Consider the value of $\ddot{\mathfrak{c}}_{l,1;l,1}^{-;-}$ for $n^- \neq n^+$. Products of two pure terms where at least one is a short plus term, long plus term, 3-term, 4-term, 6-term or 7-term become zero due to the $c^{\{3\}}$, $c^{\{4\}}$, $c^{\{6\}}$ and $c^{\{7\}}$ becoming zero. So we are left with minus terms, minus plus terms, n-terms, n-plus terms, 1-terms and 5-terms. Any product of an n-term, n-plus term or 1-term on the one hand, and a minus term, minus plus term or 5-term on the other becomes zero as the aggregate N-frequencies are $\frac{1}{N} - \mathbbm{1}_{n^-=n^+}$ and $\mathbbm{1}_{n^-=n^+} - \frac{1}{N}$ as before. So we are left with two kinds of products. The first kind has only n-terms, n-plus terms and 1-terms and the second kind has only minus terms, minus plus terms and 5-terms.

Let's consider the first kind. For a product of an n-term, n-plus term or 1-term and another such term, we create 5 copies to eliminate the double N-sum as for the BN layer case. The aggregate N-frequency becomes zero when $n$ or $n'$ are equal to $n^+$. Further, for the copy where $n$, $n'$ and $n^+$ are all unequal, aggregated chains contain a single $u^{(n)}$ and $u^{(n')}$ component. So we are left with only the copy corresponding to $n=n'\neq n^+$. When the big layer indices from the leading sum are different, that copy also has aggregated chains with a single $u^{(n)}$ component. So we are left with one copy stemming from products where the big layer index of both pure terms is the same. In that situation, the product cannot contain a 1-term on the one hand and an n-term or n-plus term on the other as 1-terms correspond to unnormed big layers, and n-terms and n-plus terms correspond to normed big layers. For a product of an n-plus term and an n-term or n-plus term, we use the same argument as for the BN layer case to show it becomes zero due to either a single $u^{(n)}$ component or the B-frequency becoming zero, depending on the aggregated chain. The tau-deriv is immaterial to this argument. So we are left with one copy from products of 1-terms that share big layer index and with one copy from products of n-terms that share big layer index.

Let's now consider the second kind. Differing big layer indices lead to single $u^-$ components. So we are left with products of 5-terms for unnormed big layers, and minus terms and / or minus plus terms for normed big layers. For a product of a minus plus term and a minus or minus plus term, we use the same argument as for the BN layer case to show it becomes zero due to either a single $u^-$ components or the B-frequency becoming zero, depending on the aggregated chain. So we are left with products of minus terms that share big layer index and with products of 5-terms that share big layer index.

So analogously to the BN layer case, endowing $u$ and $v$ components and $s$ terms with unit Gaussian distributions, we obtain for $n^- \neq n^+$ 

\begin{eqnarray*}
&&\ddot{\mathfrak{c}}_{l,1;l,1}^{-;-}\\
&=&\sum_{f_{l'}} \Bigg(\frac{\limbn c_{l'}^2c^{\{5\}}_{l'}c^{\{5\}}_{l'}}{\limbas\prod_{f_{l''}\in P_{l',l}}\nu_{l''}^+}\Bigg)\mathbb{E}_{u,s,v}u_{l'}^-[1]^2\tau_l'(\ddot{\mathsf{exp}}_{l,1})^2 \\
&&\text{\scalebox{0.99}{$+\sum_{f_{\lambda}} \Bigg(\frac{\limbn\frac{N^2-N-1}{N^2}c_{\lambda}^2c^{\{1\}}_{\lambda}c^{\{1\}}_{\lambda}}{\limbas \Big(\prod_{f_{l''}\in P_{m,l}}\nu_{l''}^+\Big)\Big(\prod_{f_{l''}\in P_{\lambda,m}}\nu_{l''}^{(n^\circ)}\Big)}\Bigg)\mathbb{E}_{u,s,v}u_{\lambda}^{(n^\circ)}[1]^2\tau_l'(\ddot{\mathsf{exp}}_{l,1})^2$}}\\
&=&\Bigg(\sum_{f_{l'}} \frac{c_{l'}^2\ddot{\mathfrak{c}}_{l',1;l',1}^{-;-}}{\prod_{f_{l''}\in P_{l',l}}\ddot{\nu}_{l''}} + \sum_{f_{\lambda}} \frac{c_{\lambda}^2(\ddot{\mathfrak{c}}_{\lambda,1;\lambda,1}-\ddot{\mathfrak{c}}_{\lambda,1;\lambda,2})}{\prod_{f_{l''}\in P_{\lambda,l}}\ddot{\nu}_{l''}}\Bigg)\mathbb{E}_{u,s,v}\tau_l'(\ddot{\mathsf{exp}}_{l,1})^2
\end{eqnarray*}

Here, we aggregate minus terms and 5-terms as well as n-terms and 1-terms into a single leading sum each. $\ddot{\mathsf{exp}}_{l,1}$ is obtained by replacing the coefficients in $\mathsf{exp}_{l,1}$ with their limit. $\ddot{\mathsf{exp}}_{l,1}$ is a linear combination of unit Gaussians, and hence is a mean zero Gaussian itself. By stage 9, activation layer case, the variance of that Gaussian is $\ddot{\mathfrak{c}}_{k,1;k,1}$. Using the recursive calculation rules for bias, addition and BN layers in this stage, for $\ddot{\mathfrak{c}}^{\text{MN};\text{MN}}_{1;1}$, and for the bias, addition and BN layers in stage 9, it is easy to check that $\Big(\sum_{f_{l'}} \frac{c_{l'}^2\ddot{\mathfrak{c}}_{l',1;l',1}^{-;-}}{\prod_{f_{l''}\in P_{l',l}}\ddot{\nu}_{l''}} + \sum_{f_{\lambda}} \frac{c_{\lambda}^2(\ddot{\mathfrak{c}}_{\lambda,1;\lambda,1}-\ddot{\mathfrak{c}}_{\lambda,1;\lambda,2})}{\prod_{f_{l''}\in P_{\lambda,l}}\ddot{\nu}_{l''}}\Big)=\ddot{\mathfrak{c}}_{k,1;k,1}^{-;-}$. So finally, $\ddot{\mathfrak{c}}_{l,1;l,1}^{-;-} = \ddot{\mathfrak{c}}_{k,1;k,1}^{-;-}\mathbb{E}_{s\sim\mathcal{N}(0,\ddot{\mathfrak{c}}_{k,1;k,1})}\tau_l'(s)^2$ and one last application of lemma \ref{lemma15} yields $\limep \ddot{\mathfrak{c}}_{l,1;l,1}^{-;-}|_{n^-\neq n^+} =  \frac{\mathfrak{g}_{k}(\mathfrak{q}_{m}-\mathfrak{c}_{m})}{\mathfrak{g}_{m}}\mathbb{E}_{s\sim\mathcal{N}(0,\mathfrak{q}_{k})}\tau_l'(s)^2=\frac{\mathfrak{g}_{k}(\mathfrak{q}_{m}-\mathfrak{c}_{m})}{\mathfrak{g}_{m}}\mathfrak{C}_{\tau_l'}(\mathfrak{q}_{k},\mathfrak{q}_{k})=\frac{\mathfrak{g}_{l}(\mathfrak{q}_{m}-\mathfrak{c}_{m})}{\mathfrak{g}_{m}}$ as required.

Finally, let's look at $\ddot{\mathfrak{c}}_{l,1;l,2}^{-;-}$. We obtain validity as for $\ddot{\mathfrak{c}}_{l,1;l,1}^{-;-}$. In the context of lemma \ref{lemma18}, $\tau_l'(\mathsf{exp}_{l,1})\tau_l'(\mathsf{exp}_{l,2})$ is handled  just as $\tau_l'(\mathsf{exp}_{l,1})\tau_l(\mathsf{exp}_{l,2})$ is handled for $\ddot{\mathfrak{c}}_{l,1;l,2}^{-;+}$ above. To obtain the value of the limit for $n^- \neq n^+$, we also proceed along the lines of $\ddot{\mathfrak{c}}_{l,1;l,1}^{-;-}$. Again, we show that expectation limits of (copies of) products of pure terms become zero. We use the same arguments as for $\ddot{\mathfrak{c}}_{l,1;l,1}^{-;-}$, except for products of two 5-terms; products of one minus or minus plus term and another minus or minus plus term; the $n=n'\neq n^+$ copy of a product of two 1-terms; and for the $n=n'\neq n^+$ copy of a product of one n-term or n-plus term and another n-term or n-plus term. For these products, when the big layer index differs, we obtain a single $u^-$ or $u^{(n)}$ component as usual. But even when the big layer index is the same for both pure terms, there is a single $u_{l'}^-[1]$ and $u_{l'}^-[2]$ component, or a single $u_{\lambda}^{(n)}[1]$ and $u_{\lambda}^{(n)}[2]$ component. So, overall we obtain $\ddot{\mathfrak{c}}_{l,1;l,2}^{-;-}|_{n^-\neq n^+}=0$ as required.

\end{itemize}

So as in stage 4, letting $n^-$ and $n^+$ vary again, we have

\begin{eqnarray*}
&&\limep \limbn \limd \frac{1}{d_l^\text{MF}}\mathbb{E}_X\Tr(\mathcal{J}_{l,m,1}\Cov_{f_m}\mathcal{J}_{l,m,1}^T)\\
&=&\limep \limbn \frac{1}{N^2}\sum_{n^-,n^+=1}^N\limd\overline{p_{l,1}^{(n^-,n^+)}.p_{l,1}^{(n^-,n^+)}}\\
&=&\limep \limbn \frac{1}{N}\mathfrak{c}_{l,1;l,1}^{-;-}|_{n^-=n^+} + \frac{N-1}{N}\mathfrak{c}_{l,1;l,1}^{-;-}|_{n^-\neq n^+}\\
&=&\limep \limbn \mathfrak{c}_{l;l}^{-;-}|_{n^-\neq n^+}\\
&=&\frac{\mathfrak{g}_{l}(\mathfrak{q}_{m}-\mathfrak{c}_{m})}{\mathfrak{g}_{m}}
\end{eqnarray*}

as required.

\underline{Stage 13}: statement (\ref{eqn52p11})

By symmetry and writing $\mathfrak{c}_{0,1;0,1}$ / $\mathfrak{c}_{0,1;0,1}^{\prime;\prime\prime}$ for $q$ / $c$ respectively, we have

\begin{eqnarray*}
&&\limep \limbn \limd NLC(f_l(f_m),f_m(\mathcal{D}^{(N,B)}))\\
&=&\limep \limbn\limd \sqrt{\frac{\mathbb{E}_X\Tr(\mathcal{J}_{l,m}(X)\Cov_{f_m}\mathcal{J}_{l,m}^T(X))}{\Tr(\Cov_{f_l})}}\\
&=&\limep \limbn \limd \sqrt{\frac{\sum_{b=1}^B\mathbb{E}_X\Tr(\mathcal{J}_{l,m,b}(X)\Cov_{f_m}\mathcal{J}_{l,m,b}^T(X))}{\sum_{b=1}^B\mathbb{E}_X||f_{l,b}(X)||_2^2 - ||\mathbb{E}_Xf_{l,b}(X)||_2^2}}\\
&=&\limep \limbn \sqrt{\frac{\sum_{b=1}^B\frac{1}{N}\mathfrak{c}_{l,b;l,b}^{-;-}|_{n^-=n^+} + \frac{N-1}{N}\mathfrak{c}_{l,b;l,b}^{-;-}|_{n^-\neq n^+}}{\sum_{b=1}^B\frac{N-1}{N}(\mathfrak{c}_{l,b;l,b}-\mathfrak{c}^{\prime;\prime\prime}_{l,b;l,b})}}\\
&=&\limep \limbn \sqrt{\frac{\frac{1}{N}\mathfrak{c}_{l,1;l,1}^{-;-}|_{n^-=n^+} + \frac{N-1}{N}\mathfrak{c}_{l,1;l,1}^{-;-}|_{n^-\neq n^+}}{\frac{N-1}{N}(\mathfrak{c}_{l,1;l,1}-\mathfrak{c}^{\prime;\prime\prime}_{l,1;l,1})}}\\
&=&\sqrt{\frac{\mathfrak{g}_l(\mathfrak{q}_m-\mathfrak{c}_m)}{\mathfrak{g}_m(\mathfrak{q}_l-\mathfrak{c}_l)}}
\end{eqnarray*}

\underline{Stage 14}: statement (\ref{eqn52p5})

This stage resembles stage 6, where references to stage 4 are replaced with references to stage 12. The case $l=m$ is again trivial. For $l > m$, as in stage 12, we will treat the cases $m > 0$ and $m = 0$ separately. However, now, the differences are more extensive and so there is no overlap in our presentation of the two cases.

Case: $m > 0$. We will show $\limep \limbn \limd \frac{1}{d_l^\text{MF}}||\mathcal{J}_{l,m,1}(X^{(1)})u||_2^2 = \frac{\mathfrak{g}_l}{\mathfrak{g}_m}$ where randomness is induced by $\theta$, $X^{(1)}$ and the unit Gaussian vector $u$, which has dimensionality $Bd_m^\text{MF}$. From lemma \ref{lemma17}, we then obtain $\limd \frac{1}{d_l^\text{MF}}||\mathcal{J}_{l,m,1}(X^{(1)})u||_2^2=\limd \frac{1}{d_l^\text{MF}}||\mathcal{J}_{l,m,1}(X^{(1)})||_F^2$, and hence $\limep \limbn \limd \frac{1}{d_l^\text{MF}}||\mathcal{J}_{l,m,1}(X^{(1)})||_F^2 = \frac{\mathfrak{g}_l}{\mathfrak{g}_m}$ as required.

We have $\limd \frac{1}{d_l^\text{MF}}||\mathcal{J}_{l,m,1}(X^{(1)})u||_2^2 = \limd \overline{(u\mathcal{J}_{l,m,1}(X^{(1)})^T)^{.2}}$. This is the same expression we investigated in stage 12 as $p_{l,1}^{(n^-,n^+)}$ with $n^+=1$, except that $f_m(X^{(n^-)})-\frac{1}{N}\sum_{n=1}^Nf_m(X^{(n)})$ is replaced by $u$. In fact, if we let $p^{(n^-,1)}_{l,1}$ refer to $u\mathcal{J}_{l,m,1}(X^{(1)})^T$, we can proceed analogously to stage 12, which eventually yields 

\begin{eqnarray*}
&&\limep\limbn\limd \frac{1}{d_l^\text{MF}}||\mathcal{J}_{l,m,1}(X^{(1)})||_F^2\\
&=&\limep\limbn\limd \overline{p^{(n^-,1)}_{l,1}.p^{(n^-,1)}_{l,1}}\\
&=&\limep\limbn\limd \overline{F^-_{l,1}.F^-_{l,1}}\\
&=&\limep \ddot{\mathfrak{c}}_{l,1;l,1}^{-;-}\\
&=&\frac{\mathfrak{g}_l}{\mathfrak{g}_m}
\end{eqnarray*}

The major difference is the form of the recursive expansion as it arises in the BN and activation layer cases of the induction, which we dive into below. The following minor differences also arise.

\begin{itemize}
\item The $F^\text{MN}_b$ are input layers. Each of them corresponds to a segment of $u$ of dimensionality $d_m^\text{MF}$. Because $u$ is unit Gaussian, we add a diagonal entry of 1 and off-diagonal entries of zero to the covariance matrix of $\mathcal{G}$ for each $F^\text{MN}_b$. Distribution equality is obvious. The $\limd$ quantities become $\tilde{\mathfrak{m}}^\text{MN}_1 = 0$, $\mathfrak{c}^\text{MN;MN}_{1;1} = 1$ and $\mathfrak{c}^\text{MN;MN}_{1;2}=\mathfrak{c}^\text{MN;(n)}_{1;m,1} = \mathfrak{c}^\text{MN;(n)}_{1;m,2}=0$. The same holds for the $\limbn$ quantities. Finally, $\limep \ddot{\mathfrak{c}}^\text{MN;MN}_{1;1}=1=\frac{\mathfrak{g}_m}{\mathfrak{g}_m}$. This is required to start off the induction.
\item $n^-$ does not arise in the construction of $F$ and all limit quantities are independent of $n^-$ and $N$. The $n^-\neq n^+$ case becomes the general case.
\item The case where $f_l$ is a readin layer does not appear in the induction step as $l > m > 0$ and $f_m$ is a bottleneck for $f_l$. We can ignore all associated arguments.
\item When $f_l$ is an addition layer, the recursive expansion of $\rho_l^-$ halts at the $\rho^\text{MN}_b$, as these are now dummy multi-activation functions. The same arguments apply as the $e^\text{MN}_b$ are independent of all other $e$ terms for input and FC layers in $F$ and as $\tilde{\mathfrak{m}}^\text{MN}_1=0$.
\item If $f_l$ is an FC, bias or addition layer, we obtain $\limep \ddot{\mathfrak{c}}^{-;-}_{l,1;l,1}=\frac{\mathfrak{g}_l}{\mathfrak{g}_m}$ as required, as the induction hypothesis now yields $\limep \ddot{\mathfrak{c}}^{-;-}_{k,1;k,1}=\frac{\mathfrak{g}_k}{\mathfrak{g}_m}$.
\end{itemize}

Let's look at the BN layer case. The ``first expansion'' becomes

\begin{eqnarray*}
&&\rho_{l,1}^- \\
&=&\sum_{f_{l'}} c_{l'}\Bigg(\prod_{f_{l''}\in P_{l',l}}\frac{1}{\sqrt{\nu_{l''}^+}}\Bigg)(e_{l',1}^--\widehat{e_{l'}^-})\\
&&+\sum_{f_{l'}}\sum_{\mathcal{L} \in \mathsf{indpow}(P_{l',l}), Z=|\mathcal{L}|>0} c_{l'}(-1)^{|Z|}\Bigg(\prod_{f_{l''}\in P_{l',l}}\frac{1}{\sqrt{\nu_{l''}^+}}\Bigg)...\\
&&...\frac{(\widehat{e_{l'}^-\rho_{k_{\mathcal{L}[1]}}^+}-\widehat{e_{l'}^-}\widehat{\rho_{k_{\mathcal{L}[1]}}^+})(\rho^+_{k_{\mathcal{L}[Z]},1}-\widehat{\rho^+_{k_{\mathcal{L}[Z]}}})}{\nu_{\mathcal{L}[Z]}^+}\prod_{z=1}^{|Z|-1}\frac{\widehat{\rho_{k_{\mathcal{L}[z]}}^+\rho_{k_{\mathcal{L}[z+1]}}^+}-\widehat{\rho_{k_{\mathcal{L}[z]}}^+}\widehat{\rho_{k_{\mathcal{L}[z+1]}}^+}}{\nu_{\mathcal{L}[z]}^+}\\
&&+\sum_{f_m}c_{m}\Bigg(\prod_{f_{l''}\in P_{m,l}}\frac{1}{\sqrt{\nu_{l''}^+}}\Bigg)(e_1^\text{MN}-\widehat{e^\text{MN}})\\
&&+\sum_{f_m}\sum_{\mathcal{L} \in \mathsf{indpow}(P_{m,l}), Z=|\mathcal{L}|>0}c_{m}(-1)^{|Z|}\Bigg(\prod_{f_{l''}\in P_{m,l}}\frac{1}{\sqrt{\nu_{l''}^+}}\Bigg)...\\
&&...\frac{(\widehat{e^\text{MN}\rho_{k_{\mathcal{L}[1]}}^+}-\widehat{e^\text{MN}}\widehat{\rho_{k_{\mathcal{L}[1]}}^+})(\rho^+_{k_{\mathcal{L}[Z]},1}-\widehat{\rho^+_{k_{\mathcal{L}[Z]}}})}{\nu_{\mathcal{L}[Z]}^+}\prod_{z=1}^{|Z|-1}\frac{\widehat{\rho_{k_{\mathcal{L}[z]}}^+\rho_{k_{\mathcal{L}[z+1]}}^+}-\widehat{\rho_{k_{\mathcal{L}[z]}}^+}\widehat{\rho_{k_{\mathcal{L}[z+1]}}^+}}{\nu_{\mathcal{L}[z]}^+}
\end{eqnarray*} 

Here, for consistency of notation, we write $\sum_{f_m}$ for the sum over only layer $f_m$ if that layer is touched by the recursive expansion and a sum over no layers otherwise. (Note again that the recursive expansion halts at the $\rho^\text{MN}_b$.) The difference between this first expansion and that for the BN layer case of stage 12 is that super terms 3 and 4 have been replaced by copies of super terms 1 and 2 where $l'$ is replaced by $m$. Control holds as in stage 12.

We can decompose $(e_1^\text{MN}, .., e_B^\text{MN})$ as $u^\text{MN} + c^{\{9\}}s^\text{MN}$ where $u^\text{MN}$ is a Gaussian vector with mean zero and covariance matrix as our other $u$ vectors, $s^\text{MN}$ is a unit Gaussian scalar and $c^{\{9\}}=\frac{1}{B}$. We use this to derive the ``second expansion''. It contains minus terms, minus plus terms, short plus terms and long plus terms as in the BN layer case of stage 12. We no longer obtain n-terms and n-plus terms, but we obtain similar terms where some $u^{(n)}$ components are replaced by $u^\text{MN}$ components, and the N-sum, N-frequency and part of the normalizer are removed. For example, an n-term becomes

$$c_{m}\Bigg(\prod_{f_{l''}\in P_{m,l}}\frac{1}{\sqrt{\nu_{l''}^+}}\Bigg)u^\text{MN}[1]$$

Note that no such term is present if $\sum_{f_m}$ is the empty sum. Again, $\rho_{l,1}$ is a sum over a fixed number of six kinds of pure terms of standard form. Compared to the BN layer case of stage 12, standard form terms now do not contain N-sums and N-frequencies, but we allow them to contain $u^\text{MN}$ components. The application of lemma \ref{lemma18} is as in the BN layer case of stage 12. Since no N-sums are present, each product of a pure term and a pure or co-term has standard form. Since now no $c^{\{4\}}$ terms or $c^{\{6\}}$ terms are present, we immediately obtain that $\ddot{\mathfrak{m}}_{l,1}$, $\ddot{\mathfrak{c}}_{l,1;l,1}^{-;+}$, $\ddot{\mathfrak{c}}_{l,1;l,2}^{-;+}$, $\ddot{\mathfrak{c}}_{l,1;l,2}^{-;-}$ and $\ddot{\mathfrak{c}}_{l,1;l,1}^{-;-}$ are valid. $\ddot{\mathfrak{m}}_{l,1}=\ddot{\mathfrak{c}}_{l,1;l,2}^{-;+}=\ddot{\mathfrak{c}}_{l,1;l,2}^{-;-}=0$ holds again by symmetry. To determine the value of $\ddot{\mathfrak{c}}_{l,1;l,1}^{-;+}$ and $\ddot{\mathfrak{c}}_{l,1;l,1}^{-;-}$, we again consider individual products of pure terms and pure terms or co-terms and find that most of them become zero because of the multiplier converging to zero, the presence of a single $u$ component, or the B-frequency converging to zero. We obtain $\ddot{\mathfrak{c}}_{l,1;l,1}^{-;+}=0$ and when we endow $u$ components with unit Gaussian distributions, we have

\begin{eqnarray*}
&&\ddot{\mathfrak{c}}_{l,1;l,1}^{-;-}\\
&=&\sum_{f_{l'}} \Bigg(\frac{\limbn c_{l'}^2c^{\{5\}}_{l'}c^{\{5\}}_{l'}}{\limbas \prod_{f_{l''}\in P_{l',l}}\nu_{l''}^+}\Bigg)\mathbb{E}_uu_{l'}^-[1]^2 + \sum_{f_{m}} \Bigg(\frac{\limbn c_{m}^2}{\limbas \prod_{f_{l''}\in P_{m,l}}\nu_{l''}^+}\Bigg)\mathbb{E}_uu^\text{MN}[1]^2\\
&=&\sum_{f_{l'}} \frac{c_{l'}^2\ddot{\mathfrak{c}}_{l',1;l',1}^{-;-}}{\prod_{f_{l''}\in P_{l',l}}\ddot{\nu}_{l''}} + \sum_{f_{m}} \frac{c_m^2}{\prod_{f_{l''}\in P_{m,l}}\ddot{\nu}_{l''}}
\end{eqnarray*}

Again, using the recursive calculation rules for bias, addition and BN layers, it is easy to check that this is equivalent to $\ddot{\mathfrak{c}}_{l,1;l,1}^{-;-} = \frac{\ddot{\mathfrak{c}}_{k,1;k,1}^{-;-}}{\ddot{\mathfrak{c}}_{k,1;k,1}-\ddot{\mathfrak{c}}_{k,1;k,2}+\epsilon_l}$. Hence $\limep \ddot{\mathfrak{c}}_{l,1;l,1}^{-;-} = \frac{\limep\ddot{\mathfrak{c}}_{k,1;k,1}^{-;-}}{\limep\ddot{\mathfrak{c}}_{k,1;k,1}-\limep\ddot{\mathfrak{c}}_{k,1;k,2}+\limep\epsilon_l}=\frac{\mathfrak{g}_k}{(\mathfrak{q}_k-\mathfrak{c}_k)\mathfrak{g}_m}=\frac{\mathfrak{g}_l}{\mathfrak{g}_m}$ as required.

Finally, let's turn to the activation layer case. The first recursive expansion becomes

\begin{eqnarray*}
&&\rho_{l,1}^- \\
&=&\sum_{f_{l'}\text{ normed}} c_{l'}\Bigg(\prod_{f_{l''}\in P_{l',l}}\frac{1}{\sqrt{\nu_{l''}^+}}\Bigg)(e_{l',1}^--\widehat{e_{l'}^-})\tau_l'(\rho_{k,1}^+)\\
&&+\sum_{f_{l'}\text{ normed}}\sum_{\mathcal{L} \in \mathsf{indpow}(P_{l',l}), Z=|\mathcal{L}|>0} c_{l'}(-1)^{|Z|}\Bigg(\prod_{f_{l''}\in P_{l',l}}\frac{1}{\sqrt{\nu_{l''}^+}}\Bigg)...\\
&&...\frac{(\widehat{e_{l'}^-\rho_{k_{\mathcal{L}[1]}}^+}-\widehat{e_{l'}^-}\widehat{\rho_{k_{\mathcal{L}[1]}}^+})(\rho^+_{k_{\mathcal{L}[Z]},1}-\widehat{\rho^+_{k_{\mathcal{L}[Z]}}})}{\nu_{\mathcal{L}[Z]}^+}\Bigg(\prod_{z=1}^{|Z|-1}\frac{\widehat{\rho_{k_{\mathcal{L}[z]}}^+\rho_{k_{\mathcal{L}[z+1]}}^+}-\widehat{\rho_{k_{\mathcal{L}[z]}}^+}\widehat{\rho_{k_{\mathcal{L}[z+1]}}^+}}{\nu_{\mathcal{L}[z]}^+}\Bigg)\tau_l'(\rho_{k,1}^+)\\
&&+\sum_{f_m\text{ normed}}c_m\Bigg(\prod_{f_{l''}\in P_{m,l}}\frac{1}{\sqrt{\nu_{l''}^+}}\Bigg)(e_1^\text{MN}-\widehat{e^\text{MN}})\tau_l'(\rho_{k,1}^+)\\
&&+\sum_{f_m \text{ normed}}\sum_{\mathcal{L} \in \mathsf{indpow}(P_{m,l}), Z=|\mathcal{L}|>0}c_m(-1)^{|Z|}\Bigg(\prod_{f_{l''}\in P_{m,l}}\frac{1}{\sqrt{\nu_{l''}^+}}\Bigg)...\\
&&...\frac{(\widehat{e^\text{MN}\rho_{k_{\mathcal{L}[1]}}^+}-\widehat{e^\text{MN}}\widehat{\rho_{k_{\mathcal{L}[1]}}^+})(\rho^+_{k_{\mathcal{L}[Z]},1}-\widehat{\rho^+_{k_{\mathcal{L}[Z]}}})}{\nu_{\mathcal{L}[Z]}^+}...\\
&&...\Bigg(\prod_{z=1}^{|Z|-1}\frac{\widehat{\rho_{k_{\mathcal{L}[z]}}^+\rho_{k_{\mathcal{L}[z+1]}}^+}-\widehat{\rho_{k_{\mathcal{L}[z]}}^+}\widehat{\rho_{k_{\mathcal{L}[z+1]}}^+}}{\nu_{\mathcal{L}[z]}^+}\Bigg)\tau_l'(\rho_{k,1}^+)\\
&&+\sum_{f_{l'}\text{ unnormed}} c_{l'}e_{l',1}^-\tau_l'(\rho_{k,1}^+)+\sum_{f_m\text{ unnormed}}c_m e_1^\text{MN}\tau_l'(\rho_{k,1}^+)\\
\end{eqnarray*}

The difference between this first expansion and that for the activation layer case of stage 12 is that super terms 3, 4 and 6 have been replaced by copies of super terms 1, 2 and 5 where $l'$ is replaced by $m$. Control holds as in stage 12. Again, we derive the second expansion. It contains minus terms, minus plus terms, short plus terms and long plus terms as in the activation layer case of stage 12. We no longer obtain n-terms and n-plus terms, but we obtain similar terms where some $u^{(n)}$ components are replaced by $u^\text{MN}$ components, and the N-sum, N-frequency and part of the normalizer are removed. For example, an n-term becomes

$$c_{m}\Bigg(\prod_{f_{l''}\in P_{m,l}}\frac{1}{\sqrt{\nu_{l''}^+}}\Bigg)u^\text{MN}[1]\tau_l'(\mathsf{exp}_{l,1})$$

Note that no such term is present if $\sum_{f_m}$ is the empty sum. We obtain 3-terms, 4-terms, 5-terms and 6-terms as before. Instead of 1-terms we now have

$$c_m u^\text{MN}[1]\tau_l'(\mathsf{exp}_{l,1})$$

and instead of 7-terms we now have

$$c_{m}c^{\{9\}}s^\text{MN}\tau_l'(\mathsf{exp}_{l,1})$$

Again, $\rho_{l,1}$ is a sum over a fixed number of 12 kinds of pure terms of standard form. Compared to the activation layer case of stage 12, standard form terms now do not contain N-sums and N-frequencies but we allow them to contain $u^\text{MN}$ components and $s^\text{MN}$ terms. The application of lemma \ref{lemma18} is as in the activation layer case of stage 12 and depends on whether $\tau_l'(\mathsf{exp}_{l,1})$, $\tau_l'(\mathsf{exp}_{l,1})^2$, $\tau_l'(\mathsf{exp}_{l,1})\tau_l(\mathsf{exp}_{l,1})$, $\tau_l'(\mathsf{exp}_{l,1})\tau_l'(\mathsf{exp}_{l,2})$ or $\tau_l'(\mathsf{exp}_{l,1})\tau_l(\mathsf{exp}_{l,2})$ is present. Pure terms containing $c^{\{4\}}$ and $c^{\{6\}}$ are as they were in the activation layer case of stage 12, and so the arguments for the validity of the expectation limits of those terms and products involving those terms are unchanged. Since also now no N-sums are present that need to be eliminated, we obtain the validity of the $\limbn$ quantities. To determine the value of these quantities, we again consider individual pure terms and their products and find that most of them become zero because of the multiplier converging to zero, the presence of a single $u$ component, or the B-frequency converging to zero. We obtain $\ddot{\mathfrak{m}}_{l,1}^-=\ddot{\mathfrak{c}}_{l,1;l,1}^{-;+}=\ddot{\mathfrak{c}}_{l,1;l,2}^{-;+}=\ddot{\mathfrak{c}}_{l,1;l,2}^{-;-}=0$ and, when we endow $u$ and $v$ components and $s$ terms with unit Gaussian distributions, we have

\begin{eqnarray*}
&&\ddot{\mathfrak{c}}_{l,1;l,1}^{-;-}\\
&=&\sum_{f_{l'}} \Bigg(\frac{\limbn c_{l'}^2c^{\{5\}}_{l'}c^{\{5\}}_{l'}}{\limbas\prod_{f_{l''}\in P_{l',l}}\nu_{l''}^+}\Bigg)\mathbb{E}_{u,s,v}u_{l'}^-[1]^2\tau_l'(\ddot{\mathsf{exp}}_{l,1})^2\\
&& + \sum_{f_{m}} \Bigg(\frac{\limbn c_{m}^2}{\limbas\prod_{f_{l''}\in P_{m,l}}\nu_{l''}^+}\Bigg)\mathbb{E}_{u,s,v}u^\text{MN}[1]^2\tau_l'(\ddot{\mathsf{exp}}_{l,1})^2\\
&=&\Big(\sum_{f_{l'}} \frac{c_{l'}^2\ddot{\mathfrak{c}}_{l',1;l',1}^{-;-}}{\prod_{f_{l''}\in P_{l',l}}\ddot{\nu}_{l''}} + \sum_{f_m} \frac{c_m^2}{\prod_{f_{l''}\in P_{m,l}}\ddot{\nu}_{l''}}\Big)\mathbb{E}_{u,s,v}\tau_l'(\ddot{\mathsf{exp}}_{l,1})^2
\end{eqnarray*}

Again, using the recursive calculation rules for bias, addition and BN layers, it is easy to check that we have $\Big(\sum_{f_{l'}} \frac{c_{l'}^2\ddot{\mathfrak{c}}_{l,1;l,1}^{-;-}}{\prod_{f_{l''}\in P_{l',l}}\ddot{\nu}_{l''}} + \sum_{f_m} \frac{c_m^2}{\prod_{f_{l''}\in P_{m,l}}\ddot{\nu}_{l''}}\Big)=\ddot{\mathfrak{c}}_{k,1;k,1}^{-;-}$. So as in the activation layer case of stage 12, $\ddot{\mathfrak{c}}_{l,1;l,1}^{-;-} = \ddot{\mathfrak{c}}_{k,1;k,1}^{-;-}\mathbb{E}_{s\sim\mathcal{N}(0,\ddot{\mathfrak{c}}_{k,1;k,1})}\tau_l'(s)^2$ and one last application of lemma \ref{lemma15} yields $\limep \ddot{\mathfrak{c}}_{l,1;l,1}^{-;-}=  \frac{\mathfrak{g}_{k}}{\mathfrak{g}_{m}}\mathbb{E}_{s\sim\mathcal{N}(0,\mathfrak{q}_{k})}\tau_l'(s)^2=\frac{\mathfrak{g}_{k}}{\mathfrak{g}_{m}}\mathfrak{C}_{\tau_l'}(\mathfrak{q}_{k},\mathfrak{q}_{k})=\frac{\mathfrak{g}_{l}}{\mathfrak{g}_{m}}$ as required.

Case: $m=0$. We have

\begin{eqnarray*}
&&\limep \limbn \limd \frac{1}{d_l^\text{MF}}||\mathcal{J}_{l,m,1}(X^{(1)})||_F^2\\
&=&\limep \limbn \limd \frac{1}{d_l^\text{MF}}\Big|\Big|\sum_{f_{l'} \text{ readin}}\mathcal{J}_{l',0}(X^{(1)})^T\mathcal{J}_{l,l',1}(X^{(1)})^T\Big|\Big|_F^2\\
&=&\limep \limbn \limd \frac{1}{d_l^\text{MF}}\Big|\Big|\sum_{f_{l'} \text{ readin}}W_{l'}^B\mathcal{J}_{l,l',1}(X^{(1)})^T\Big|\Big|_F^2\\
&=&\limep \limbn \limd \frac{1}{d_l^\text{MF}}\sum_{i=0}^{d_\text{in}-1}\sum_{b=0}^{B-1}\Big|\Big|\sum_{f_{l'} \text{ readin}}W_{l'}^B[i+d_\text{in}b, :]\mathcal{J}_{l,l',1}(X^{(1)})^T\Big|\Big|_2^2\\
&=&\limep \limbn \limd \frac{1}{d_l^\text{MF}}\frac{1}{B}\sum_{i=0}^{d_\text{in}-1}\sum_{b=0}^{B-1}\sum_{b=1}^{B}\Big|\Big|\sum_{f_{l'} \text{ readin}}W_{l'}^B[i+d_\text{in}b, :]\mathcal{J}_{l,l',b}(X^{(1)})^T\Big|\Big|_2^2\\
&=&\limep \limbn \limd \frac{1}{d_l^\text{MF}}\frac{1}{B}\sum_{i=0}^{d_\text{in}-1}\sum_{b=0}^{B-1}\Big|\Big|\sum_{f_{l'} \text{ readin}}W_{l'}^B[i+d_\text{in}b, :]\mathcal{J}_{l,l'}(X^{(1)})^T\Big|\Big|_2^2\\
&=&\limep \limbn \limd \frac{1}{d_l^\text{MF}}d_\text{in}\Big|\Big|\sum_{f_{l'} \text{ readin}}W_{l'}^B[0, :]\mathcal{J}_{l,l'}(X^{(1)})^T\Big|\Big|_2^2\\
&=&\limep \limbn \limd \overline{\Big(\sum_{f_{l'} \text{ readin}}\sqrt{d_\text{in}}W_{l'}^B[0, :]\mathcal{J}_{l,l'}(X^{(1)})^T\Big)^{.2}}\\
\end{eqnarray*}

Here, $\sum_{f_{l'} \text{ readin}}$ is over all readin layers. $W_{l'}^B$ is the block-diagonal matrix of size $Bd_\text{in} \times Bd_{l'}^\text{MF}$ with $B$ blocks of $W_{l'}$. $W_{l'}^B[i+d_\text{in}b, :]$ indicates a row of that matrix. In the derivation, we use symmetry repeatedly.

Again, we proceed along the lines of stage 12, where we now let $p_{l,1}^{(n^-, 1)}$ refer to $\sum_{f_{l'} \text{ readin}}\sqrt{d_\text{in}}W_{l'}^B[0, :]\mathcal{J}_{l,l',1}(X^{(1)})^T$. However, we now no longer have $\frac{1}{d_l^\text{MF}}||\mathcal{J}_{l,m,1}(X^{(1)})||_F^2 = \overline{p_{l,1}^{(n^-, 1)}.p_{l,1}^{(n^-, 1)}}$ but

\begin{eqnarray*}
&&\limep\limbn\limd\frac{1}{d_l^\text{MF}}||\mathcal{J}_{l,m,1}(X^{(1)})||_F^2\\
&=&\limep\limbn\limd\sum_{b=1}^B \overline{p^{(n^-,1)}_{l,b}.p^{(n^-,1)}_{l,b}}\\
&=&\limep\limbn\limd\sum_{b=1}^B \overline{F^-_{l,b}.F^-_{l,b}}\\
&=&\limep\limbn\sum_{b=1}^B \mathfrak{c}_{l,b;l,b}^{-;-}
\end{eqnarray*}

We will have to show that the expression on the last line is $\frac{\mathfrak{g}_l}{\mathfrak{g}_m}$.

Let's start by looking at the readin layers. If $f_l$ if a readin big layer, the distribution of the $p_{l,b}^{(n^-, 1)}$ is the distribution of the segments of $\sqrt{d_\text{in}}W_l^B[0, :]$. Hence, components of $p_{l,1}^{(n^-, 1)}$ are unit Gaussian and the components of $p_{l,b}^{(n^-, 1)}$ with $b \neq 1$ are equal to zero. The unit Gaussians are independent of those corresponding to other readin big layers as there is no weight sharing in $f$. Hence, the entries added to the covariance matrix of $\mathcal{G}$ for the $F_{l,b}^-$ are zero, except the diagonal entry added for $F_{l,1}^-$, which is $\sigma_l^2$. While distribution equality is clear, the major consequence of this is that there is no symmetry between the batch index value 1 and other batch index values for any quantity that depends directly or indirectly on the distribution of the $F^-_{l,b}$ or the $\rho^-_{l,b}(e)$. Thus, we obtain the following distinct $\limd$ quantities: $\tilde{\mathfrak{m}}_{l,1}^-$, $\tilde{\mathfrak{m}}_{l,2}^-$, $\tilde{\mathfrak{m}}_{l,1}^+=\tilde{\mathfrak{m}}_{l,1}$, $\mathfrak{c}_{l,1;l,1}^{-;-}$, $\mathfrak{c}_{l,1;l,2}^{-;-}$, $\mathfrak{c}_{l,2;l,2}^{-;-}$, $\mathfrak{c}_{l,2;l,3}^{-;-}$, $\mathfrak{c}_{l,1;l,1}^{-;+}$, $\mathfrak{c}_{l,1;l,2}^{-;+}$, $\mathfrak{c}_{l,2;l,1}^{-;+}$, $\mathfrak{c}_{l,2;l,2}^{-;+}$, $\mathfrak{c}_{l,2;l,3}^{-;+}$, $\mathfrak{c}_{l,1;l,1}^{+;+}=\mathfrak{c}_{l,1;l,1}$ and $\mathfrak{c}_{l,1;l,2}^{+;+}=\mathfrak{c}_{l,1;l,2}$. The differences arise based on which batch indices are equal to 1 and based on whether they are equal. Because the distribution of the layers in $F^+$ is not affected by the distribution of the layers in $F^-$, $\tilde{\mathfrak{m}}_{l,1}$, $\mathfrak{c}_{l,1;l,1}$ and $\mathfrak{c}_{l,1;l,2}$ continue to represent all batch index combinations for $\limd$ quantities that depend only on $F^+$.

This wider range of limit quantities will unfortunately complicate our limit analysis for BN and activation big layers. Before we go ahead with this analysis, we again document ``minor'' differences to the analysis in stage 12, just as for the case $m > 0$ above.

\begin{itemize}
\item The $F^\text{MN}_b$ do not arise, as they pertain only to the case $m > 0$. Further, we can ignore all arguments that are associated specifically with $m > 0$.
\item $n^-$ does not arise in the construction of $F$ and all limit quantities are independent of $n^-$ and $N$. The $n^-\neq n^+$ case becomes the general case.
\item The induction is now used to show the following with regards to limits: (i) $\ddot{\mathfrak{c}}_{l,1;l,2}^{-;-}=\ddot{\mathfrak{c}}_{l,2;l,2}^{-;-}=\ddot{\mathfrak{c}}_{l,2;l,3}^{-;-}=\ddot{\mathfrak{c}}_{l,1;l,1}^{-;+}=\ddot{\mathfrak{c}}_{l,1;l,2}^{-;+}=\ddot{\mathfrak{c}}_{l,2;l,1}^{-;+}=\ddot{\mathfrak{c}}_{l,2;l,2}^{-;+}=\ddot{\mathfrak{c}}_{l,2;l,3}^{-;+}=0$; (ii) $\ddot{\mathfrak{c}}_{l,1;l,1}^{-;-} > 0$; (iii) $\limep \ddot{\mathfrak{c}}_{l,1;l,1}^{-;-}=\frac{\mathfrak{g}_l}{\mathfrak{g}_m}$; and (iv) $\limbn B\ddot{\mathfrak{c}}_{l,2;l,2}^{-;-} = 0$. From this we obtain $ \limep\limbn\sum_{b=1}^B \mathfrak{c}_{l,b;l,b}^{-;-} = \limep\limbn \mathfrak{c}_{l,1;l,1}^{-;-} + \limep\limbn\frac{B-1}{B} (B\mathfrak{c}_{l,2;l,2}^{-;-}) = \limep \ddot{\mathfrak{c}}_{l,1;l,1}^{-;-}=\frac{\mathfrak{g}_l}{\mathfrak{g}_m}$ as required. When $f_l$ is a readin, FC, bias or addition layer, (i) through (iv) carry over easily from the dependencies or the generator, as in stage 12. For an addition layer, as we have $m=0$, the recursive expansion cannot touch $f_m$. This makes the expansion simpler than in stage 12, as there are no recursive definitions of the $\rho^{(n)}_{\lambda,b}$ to apply. The argument still goes through.
\end{itemize}

Case: $f_l$ is a BN layer.

The first expansion becomes

\begin{eqnarray*}
&&\rho_{l,1}^- \\
&=&\sum_{f_{l'}} c_{l'}\Bigg(\prod_{f_{l''}\in P_{l',l}}\frac{1}{\sqrt{\nu_{l''}^+}}\Bigg)(e_{l',1}^--\widehat{e_{l'}^-})\\
&&+\sum_{f_{l'}}\sum_{\mathcal{L} \in \mathsf{indpow}(P_{l',l}), Z=|\mathcal{L}|>0} c_{l'}(-1)^{|Z|}\Bigg(\prod_{f_{l''}\in P_{l',l}}\frac{1}{\sqrt{\nu_{l''}^+}}\Bigg)...\\
&&...\frac{(\widehat{e_{l'}^-\rho_{k_{\mathcal{L}[1]}}^+}-\widehat{e_{l'}^-}\widehat{\rho_{k_{\mathcal{L}[1]}}^+})(\rho^+_{k_{\mathcal{L}[Z]},1}-\widehat{\rho^+_{k_{\mathcal{L}[Z]}}})}{\nu_{\mathcal{L}[Z]}^+}\prod_{z=1}^{|Z|-1}\frac{\widehat{\rho_{k_{\mathcal{L}[z]}}^+\rho_{k_{\mathcal{L}[z+1]}}^+}-\widehat{\rho_{k_{\mathcal{L}[z]}}^+}\widehat{\rho_{k_{\mathcal{L}[z+1]}}^+}}{\nu_{\mathcal{L}[z]}^+}
\end{eqnarray*}

That is, we are left with only super terms 1 and 2. Control holds as in the BN layer case of stage 12. Note that the definitions of the multi-activation functions corresponding to $F^-$ are, of course, still fully symmetric with respect to the batch index even if the distribution induced by $e$ is not.

{\it Joint distribution of inputs} Compared to the BN layer case of stage 12, the inputs of $\rho_{l,1}^-$ now come only in Gaussian vectors $(e_{l',1}^-, .., e_{l',B}^-, e_{l',1}^+, .., e_{l',B}^+)$, which are independent of each other. Again, the mean is zero. Call the covariance matrix $Q_{l'}$. We decompose the Gaussian vector as follows.

\begin{eqnarray*}
e_{l',1}^+ &=& c^{\{1\}}_{l'}z_{l'}^+\\
e_{l',1}^- &=& c^{\{2\}}_{l'}z_{l'}^+ + c^{\{3\}}_{l'}z_{l'}^-\\
e_{l',b}^+ &=& c^{\{4\}}_{l'}z_{l'}^+ + c^{\{5\}}_{l'}z_{l'}^- + c^{\{6\}}_{l'}s_{l'}^+ + c^{\{11\}}_{l'}u_{l'}^+[b] \text{ for } b \ge 2\\
e_{l',b}^- &=& c^{\{7\}}_{l'}z_{l'}^+ + c^{\{8\}}_{l'}z_{l'}^- + c^{\{9\}}_{l'}s_{l'}^+  + c^{\{10\}}_{l'}s_{l'}^- + c^{\{12\}}_{l'}u_{l'}^+[b] + c^{\{13\}}_{l'}u_{l'}^-[b] \text{ for } b \ge 2\\
\end{eqnarray*}

We override the definition of the $u$, $s$ and $c^{\{.\}}$ from stage 12. $z_{l'}^+$, $z_{l'}^-$, $s_{l'}^+$ and $s_{l'}^-$ are unit Gaussian scalars. $u_{l'}^+$ and $u_{l'}^-$ are $B-1$-dimensional Gaussian vectors with mean zero and a covariance matrix that has diagonal entries $\frac{B-2}{B-1}$ and off-diagonal entries $-\frac{1}{B-1}$. For convenience, their component index ranges from 2 through $B$. An assignment to the 13 $c^{\{.\}}_{l'}$ terms is valid if it satisfies the following 13 constraints: $\mathfrak{c}_{l',1;l',1}^{-;-}=\mathbb{E}_ee_{l',1}^-e_{l',1}^-$, $\mathfrak{c}_{l',1;l',2}^{-;-}=\mathbb{E}_ee_{l',1}^-e_{l',2}^-$, $\mathfrak{c}_{l',2;l',2}^{-;-}=\mathbb{E}_ee_{l',2}^-e_{l',2}^-$, $\mathfrak{c}_{l',2;l',3}^{-;-}=\mathbb{E}_ee_{l',2}^-e_{l',3}^-$, $\mathfrak{c}_{l',1;l',1}^{-;+}=\mathbb{E}_ee_{l',1}^-e_{l',1}^+$, $\mathfrak{c}_{l',1;l',2}^{-;+}=\mathbb{E}_ee_{l',1}^-e_{l',2}^+$, $\mathfrak{c}_{l',2;l',1}^{-;+}=\mathbb{E}_ee_{l',2}^-e_{l',1}^+$, $\mathfrak{c}_{l',2;l',2}^{-;+}=\mathbb{E}_ee_{l',2}^-e_{l',2}^+$, $\mathfrak{c}_{l',2;l',3}^{-;+}=\mathbb{E}_ee_{l',2}^-e_{l',3}^+$, $\mathfrak{c}_{l',1;l',1}=\mathbb{E}_ee_{l',1}^+e_{l',1}^+$, $\mathfrak{c}_{l',1;l',2}=\mathbb{E}_ee_{l',1}^+e_{l',2}^+$, $\mathfrak{c}_{l',1;l',1}=\mathbb{E}_ee_{l',2}^+e_{l',2}^+$ and $\mathfrak{c}_{l',1;l',2}=\mathbb{E}_ee_{l',2}^+e_{l',3}^+$. To start with, we obtain 

$$c_{l'}^{\{11\}} = \sqrt{\mathbb{E}_ee_{l',2}^+e_{l',2}^+ - \mathbb{E}_ee_{l',2}^+e_{l',3}^+} = \sqrt{\mathfrak{c}_{l',1;l',1} - \mathfrak{c}_{l',1;l',2}}$$

$$c_{l'}^{\{12\}} = \frac{\mathbb{E}_ee_{l',2}^-e_{l',2}^+ - \mathbb{E}_ee_{l',2}^-e_{l',3}^+}{c^{\{11\}}_{l'}} = \frac{\mathfrak{c}_{l',2;l',2}^{-;+} - \mathfrak{c}_{l',2;l',3}^{-;+}}{\sqrt{\mathfrak{c}_{l',1;l',1} - \mathfrak{c}_{l',1;l',2}}}$$

$$c_{l'}^{\{13\}} = \sqrt{\mathbb{E}_ee_{l',2}^-e_{l',2}^- - \mathbb{E}_ee_{l',2}^-e_{l',3}^- - c^{\{12\}}_{l'}c^{\{12\}}_{l'}}$$ $$= \sqrt{\frac{(\mathfrak{c}_{l',2;l',2}^{-;-} - \mathfrak{c}_{l',2;l',3}^{-;-})(\mathfrak{c}_{l',1;l',1} - \mathfrak{c}_{l',1;l',2}) - (\mathfrak{c}_{l',2;l',2}^{-;+} - \mathfrak{c}_{l',2;l',3}^{-;+})^2}{\mathfrak{c}_{l',1;l',1} - \mathfrak{c}_{l',1;l',2}}}$$

This is reminiscent of $c_{l'}^{\{1\}}$, $c_{l'}^{\{3\}}$ and $c_{l'}^{\{5\}}$ from the BN layer case of stage 12. Let's look at the square roots that arise in the formulas above. Because $Q_{l'}$ is positive semi-definite, the quadratic form defined by $Q_{l'}$ applied to any vector is non-negative. Using this principle on $(0 \times B, 1, -1, 0 \times (B-2))$ yields $\mathfrak{c}_{l',1;l',1}-\mathfrak{c}_{l',1;l',2} \ge 0$. Here, $\times$ refers to the repetition of a vector component. Going forward, we use $\widetilde{\mathsf{expr}}$ to refer to the mean of $\mathsf{expr}$ over batch index values 2 through $B$, i.e. $\widetilde{\mathsf{expr}}=\frac{1}{B-1}\sum_{b'=2}^B \mathsf{expr}|_{b=b'}$. The Cauchy-Schwartz inequality yields $\frac{(B-1)^2}{(B-2)^2}\Big(\mathbb{E}_e(e_{l',2}^- - \widetilde{e_{l'}^-})^2\mathbb{E}_e(e_{l',2}^+ - \widetilde{e_{l'}^+})^2 - (\mathbb{E}_e(e_{l',2}^- - \widetilde{e_{l'}^-})(e_{l',2}^+ - \widetilde{e_{l'}^+}))^2\Big) = (\mathfrak{c}_{l',2;l',2}^{-;-}-\mathfrak{c}_{l',2;l',3}^{-;-})(\mathfrak{c}_{l',1;l',1}-\mathfrak{c}_{l',1;l',2})-(\mathfrak{c}_{l',2;l',3}^{+;-}-\mathfrak{c}_{l',2;l',3}^{+;-})^2 \ge 0$. So expressions to which square roots are applied are indeed non-negative.

Let's look at denominators that arise in the formulas above. When $\mathfrak{c}_{l',1;l',1}-\mathfrak{c}_{l',1;l',2}=0$, the Cauchy-Schwartz-derived relationship above yields $\mathfrak{c}_{l',2;l',2}^{+;-}-\mathfrak{c}_{l',2;l',3}^{+;-}=0$, and hence we can remove $u_{l'}^+$ from the decomposition, which corresponds to setting $c_{l'}^{\{11\}}=c_{l'}^{\{12\}}=0$ and $c_{l'}^{\{13\}}=\sqrt{\mathfrak{c}_{l,2;l,2}^{-;-} - \mathfrak{c}_{l,2;l,3}^{-;-}}$. So we never have to divide by zero.

Let's look at the limits of $c^{\{11\}}_{l'}$, $c^{\{12\}}_{l'}$ and $c^{\{13\}}_{l'}$ with respect to $N$ and $B$. From the induction hypothesis, we have that the individual $\mathfrak{c}^{-;-}_{l',.;l',.}$ and $\mathfrak{c}^{+;-}_{l',.;l',.}$ have valid limits. By non-singularity from stage 9, we have $\ddot{\mathfrak{c}}_{l',1;l',1}-\ddot{\mathfrak{c}}_{l',1;l',2} > 0$. So we obtain that $\ddot{c}^{\{11\}}_{l'}$, $\ddot{c}^{\{12\}}_{l'}$ and $\ddot{c}^{\{13\}}_{l'}$ have identical formulas to $c^{\{11\}}_{l'}$, $c^{\{12\}}_{l'}$ and $c^{\{13\}}_{l'}$ above except that two dots are placed above each $\mathfrak{c}$ term. So, by the induction hypothesis, we further have $\ddot{c}_{l'}^{\{11\}} = \sqrt{\ddot{\mathfrak{c}}_{l',1;l',1} - \ddot{\mathfrak{c}}_{l',1;l',2}}$ and $\ddot{c}_{l'}^{\{12\}} = \ddot{c}_{l'}^{\{13\}} = 0$. Using the principle of the non-negative quadratic form on $(0, 1, -1, 0 \times (2B-3))$ yields $\mathfrak{c}_{l',2;l',2}^{-;-}-\mathfrak{c}_{l',2;l',3}^{-;-} \ge 0$.  Using the principle of the non-negative quadratic form on $(0, 1, 1, 0 \times (2B-3))$ yields $\mathfrak{c}_{l',2;l',2}^{-;-}+\mathfrak{c}_{l',2;l',3}^{-;-} \ge 0$. Combining those results yields $|\mathfrak{c}_{l',2;l',2}^{-;-}|\ge |\mathfrak{c}_{l',2;l',3}^{-;-}| \ge 0$. Since the induction hypothesis yields $\limbn B\mathfrak{c}_{l',2;l',2}^{-;-}=0$, we also have $\limbn B\mathfrak{c}_{l',2;l',3}^{-;-}=0$. Finally, since $c^{\{12\}}_{l'}c^{\{12\}}_{l'}+c^{\{13\}}_{l'}c^{\{13\}}_{l'} = \mathfrak{c}_{l',2;l',2}^{-;-}-\mathfrak{c}_{l',2;l',3}^{-;-}$, we have $\limbn \sqrt{B}c^{\{12\}}_{l'}=\limbn \sqrt{B}c^{\{13\}}_{l'}=0$.

Knowing the value of $c^{\{11\}}_{l'}$, $c^{\{12\}}_{l'}$ and $c^{\{13\}}_{l'}$, an assignment to the other 10 $c^{\{.\}}_{l'}$ terms is valid if they satisfy the following system.

$$\begin{pmatrix}
\mathfrak{c}_{l',1;l',1} & \mathfrak{c}_{l',1;l',1}^{-;+} & \mathfrak{c}_{l',1;l',2}  & \mathfrak{c}_{l',2;l',1}^{-;+}\\
\mathfrak{c}_{l',1;l',1}^{-;+} & \mathfrak{c}_{l',1;l',1}^{-;-} & \mathfrak{c}_{l',1;l',2}^{-;+} & \mathfrak{c}_{l',1;l',2}^{-;-} \\
\mathfrak{c}_{l',1;l',2} & \mathfrak{c}_{l',1;l',2}^{-;+} & \frac{B-2}{B-1}\mathfrak{c}_{l',1;l',2}+\frac{1}{B-1}\mathfrak{c}_{l',1;l',1} & \frac{B-2}{B-1}\mathfrak{c}_{l',2;l',3}^{-;+}+\frac{1}{B-1}\mathfrak{c}_{l',2;l',2}^{-;+}\\
\mathfrak{c}_{l',2;l',1}^{-;+} & \mathfrak{c}_{l',1;l',2}^{-;-} & \frac{B-2}{B-1}\mathfrak{c}_{l',2;l',3}^{-;+}+\frac{1}{B-1}\mathfrak{c}_{l',2;l',2}^{-;+} & \frac{B-2}{B-1}\mathfrak{c}_{l',2;l',3}^{-;-}+\frac{1}{B-1}\mathfrak{c}_{l',2;l',2}^{-;-}
\end{pmatrix} = A_{l'}A_{l'}^T$$

where

$$A_{l'} = \begin{pmatrix}
c^{\{1\}}_{l'} & 0 & 0 & 0 \\
c^{\{2\}}_{l'} & c^{\{3\}}_{l'} & 0 & 0 \\
c^{\{4\}}_{l'} & c^{\{5\}}_{l'} & c^{\{6\}}_{l'} & 0 \\
c^{\{7\}}_{l'} & c^{\{8\}}_{l'} & c^{\{9\}}_{l'} & c^{\{10\}}_{l'} \\
\end{pmatrix}$$

In other words, $A_{l'}A_{l'}^T$ must be the Cholesky decomposition of a matrix. Call it $R_{l'}$ . Let $v$ be an arbitrary 4-dimensional row vector. Then 

$$vR_{l'}v^T=v^*Q_{l'}(v^*)^T\ge 0 \text{ where } v^*=(v[2], \frac{v[4]}{B-1}\times (B-1), v[1], \frac{v[3]}{B-1}\times (B-1))$$

as $Q_{l'}$ is positive semi-definite. So $R_{l'}$ is itself positive semi-definite. So we can set unique valid values for $c^{\{1\}}_{l'}$ through $c^{\{10\}}_{l'}$ by letting $A_{l'}A_{l'}^T$ be the Cholesky decomposition where the number of columns that are not all zero equals the rank of $R_{l'}$.

What remains is to investigate the limits of $c_{l'}^{\{1\}}$ through $c_{l'}^{\{10\}}$. We have $c_{l'}^{\{7\}}c_{l'}^{\{7\}} + c_{l'}^{\{8\}}c_{l'}^{\{8\}} + c_{l'}^{\{9\}}c_{l'}^{\{9\}} + c_{l'}^{\{10\}}c_{l'}^{\{10\}} =  \frac{B-2}{B-1}\mathfrak{c}_{l',2;l',3}^{-;-}+\frac{1}{B-1}\mathfrak{c}_{l',2;l',2}^{-;-}$. $\ddot{\mathfrak{c}}_{l',2;l',3}^{-;-}=\ddot{\mathfrak{c}}_{l',2;l',2}^{-;-}=0$ so $\ddot{c}_{l'}^{\{7\}}=\ddot{c}_{l'}^{\{8\}}=\ddot{c}_{l'}^{\{9\}}=\ddot{c}_{l'}^{\{10\}}=0$. Since $\limbn B\ddot{\mathfrak{c}}_{l',2;l',2}^{-;-}=\limbn B\ddot{\mathfrak{c}}_{l',2;l',3}^{-;-}=0$, we also have $\limbn \sqrt{B}c_{l'}^{\{7\}}=\limbn \sqrt{B}c_{l'}^{\{8\}}=\limbn \sqrt{B}c_{l'}^{\{9\}}=\limbn \sqrt{B}c_{l'}^{\{10\}}=0$. For $c_{l'}^{\{1\}}$ through $c_{l'}^{\{6\}}$, we need to use explicit formulas. We obtain

\begin{eqnarray*}
c^{\{1\}}_{l'}&=&\sqrt{\mathfrak{c}_{l',1;l',1}}\\
c^{\{2\}}_{l'}&=&\frac{\mathfrak{c}_{l',1;l',1}^{-;+}}{\sqrt{\mathfrak{c}_{l',1;l',1}}}\\
c^{\{3\}}_{l'}&=& \sqrt{\mathfrak{c}_{l',1;l',1}^{-;-} - \frac{\mathfrak{c}_{l',1;l',1}^{-;+}\mathfrak{c}_{l',1;l',1}^{-;+}}{\mathfrak{c}_{l',1;l',1}}}\\
c^{\{4\}}_{l'}&=& \frac{\mathfrak{c}_{l',1;l',2}}{\sqrt{\mathfrak{c}_{l',1;l',1}}}\\
c^{\{5\}}_{l'}&=& \frac{\mathfrak{c}_{l',1;l',2}^{-;+}\mathfrak{c}_{l',1;l',1}-\mathfrak{c}_{l',1;l',1}^{-;+}\mathfrak{c}_{l',1;l',2}}{\mathfrak{c}_{l',1;l',1}c^{\{3\}}_{l'}}\\
c^{\{6\}}_{l'}&=&\sqrt{\frac{B-2}{B-1}\mathfrak{c}_{l',1;l',2}+\frac{1}{B-1}\mathfrak{c}_{l',1;l',1} - c^{\{4\}}_{l'}c^{\{4\}}_{l'} - c^{\{5\}}_{l'}c^{\{5\}}_{l'}}\\
\end{eqnarray*}

Stage 9 yields $\ddot{\mathfrak{c}}_{l',1;l',1} > 0$ and the induction hypothesis of this stage yields $\ddot{\mathfrak{c}}_{l',1;l',1}^{-;-} > 0$ as well as $\ddot{\mathfrak{c}}_{l',1;l',1}^{-;+}=0$. So $\ddot{c}^{\{3\}}_{l'} = \sqrt{\ddot{\mathfrak{c}}_{l',1;l',1}^{-;-}} > 0$, so all denominators in the above formulas have positive limits, so all limits are straightforward and we obtain 

\begin{eqnarray*}
\ddot{c}^{\{1\}}_{l'}&=&\sqrt{\ddot{\mathfrak{c}}_{l',1;l',1}}\\
\ddot{c}^{\{2\}}_{l'}&=&0\\
\ddot{c}^{\{3\}}_{l'}&=&\sqrt{\ddot{\mathfrak{c}}_{l',1;l',1}^{-;-}}\\
\ddot{c}^{\{4\}}_{l'}&=& \frac{\ddot{\mathfrak{c}}_{l',1;l',2}}{\sqrt{\ddot{\mathfrak{c}}_{l',1;l',1}}}\\
\ddot{c}^{\{5\}}_{l'}&=& 0\\
\ddot{c}^{\{6\}}_{l'}&=&\sqrt{\frac{\ddot{\mathfrak{c}}_{l',1;l',2}(\ddot{\mathfrak{c}}_{l',1;l',1}-\ddot{\mathfrak{c}}_{l',1;l',2})}{\ddot{\mathfrak{c}}_{l',1;l',1}}}
\end{eqnarray*}

{\it Recursive expansion} The first expansion is given above. Replacing the remaining $\rho^+$ terms (not inside a $\nu^+$) with $e^+$ and $\frac{1}{\sqrt{\nu^+}}$ terms is as for super terms 1 and 2 in the BN layer case of stage 12. However, the second expansion in terms of $u$ components and $s$ and $z$ terms is more complex. For now, we simply replace the $e$ terms (not inside a $\nu^+$) with their decomposition in terms of $u$ components and $s$ and $z$ terms. We do not further simplify the resulting expression for now. We thus obtain two kinds of pure terms. ``Short terms'' originate from super term 1 and ``long terms'' originate from super term 2. Pure terms are an instance of the expression inside the leading sum(s) as in stage 12. So the second expansion is a sum over a fixed number of 2 kinds of pure terms. Again, they contain a multiplier, normalizer and polynomial. N-sums and N-frequencies are not present. The polynomial can now also contain $s$, $z$ and $c^{\{.\}}$ terms that stem from the replacement of the $e$ terms by their decomposition. In this context, we consider a standard form term to be defined by those conditions.

{\it Expectation limits of standard form terms} Relative to the BN layer case of stage 12, multiplying out the polynomial results in a sum of ``chains'' that can now contain $s$, $z$ and $c^{\{.\}}$ terms in addition to $u$ components. Nonetheless, aggregation works as before except that chains are only considered equivalent if the same $s$, $z$ and $c^{\{.\}}$ terms are present, of which there is a bounded number. The B-frequencies converge. After aggregation, the $c^{\{.\}}$ terms in the aggregated chain can be absorbed into the multiplier, which causes the multiplier to differ from aggregated chain to aggregated chain. Nonetheless, applying lemma \ref{lemma18} works as before. $\chi_d$ has one component for each distinct $u$ component or $s$ or $z$ term that arises in the aggregated chain. $\hat{\Sigma}$ is still the identity. The valid limit of the expectation of the product of B-frequency, (chain-specific) multiplier, normalizer and ($c^{\{.\}}$-free) aggregated chain is equal to the product of the limit of the B-frequency, the limit of the multiplier, the almost sure limit of the normalizer and the expectation of the aggregated chain with respect to the unit Gaussian.

When investigating limit quantities such as $\ddot{\mathfrak{c}}_{l,1;l,1}^{-;+} =\limbn \mathbb{E}_e\rho_{l,1}^-\rho_{l,1}^+$, the recursive expansion of the $\rho_{l,b}^+$ is as in the BN layer case of stage 9 / 12, except that the decomposition of the $e$ terms differs, which yields co-terms that can contain $s$, $z$ and $c^{\{.\}}$ terms in their polynomial in addition to $u$ components. Nonetheless, they are standard form terms in this context, and products of a pure term and a pure term or co-term are standard form terms. This yields that all $\limbn$ quantities of interest are valid as long as the additional $B$ factor in $\limbn B\mathfrak{c}_{l,2;l,2}^{-;-}$ does not cause divergence, which we show below.

{\it Specific expectation limits} Let's look at $\ddot{\mathfrak{m}}_{l,1}^-$, $\ddot{\mathfrak{m}}_{l,2}^-$, $\ddot{\mathfrak{c}}_{l,1;l,1}^{-;+}$, $\ddot{\mathfrak{c}}_{l,1;l,2}^{-;+}$, $\ddot{\mathfrak{c}}_{l,2;l,1}^{-;+}$, $\ddot{\mathfrak{c}}_{l,2;l,2}^{-;+}$ and $\ddot{\mathfrak{c}}_{l,2;l,3}^{-;+}$. For each of these, each aggregated chain contains exactly one copy of exactly one of $c^{\{2\}}_{l_1}z_{l_1}^+$, $c^{\{3\}}_{l_1}z_{l_1}^-$, $c^{\{7\}}_{l_1}z_{l_1}^+$, $c^{\{8\}}_{l_1}z_{l_1}^-$, $c_{l_1}^{\{9\}}s_{l_1}^+$, $c^{\{10\}}_{l_1}s_{l_1}^-$, $c^{\{12\}}_{l_1}u_{l_1}^+[b]$ or $c^{\{13\}}_{l_1}u_{l_1}^-[b]$ for some $l_1 < l$. The expectation limit of a product of multiplier, normalizer and aggregated chain is zero if the chain contains any of the above except $c^{\{3\}}_{l_1}z_{l_1}^-$, due to the chain-specific multiplier becoming zero. If the aggregated chain contains $c^{\{3\}}_{l_1}z_{l_1}^-$ and it does not contain another $z_{l_1}^-$ term, the expectation limit is zero due to the single $z_{l_1}^-$ term. Finally, if the aggregated chain contains $c^{\{3\}}_{l_1}z_{l_1}^-$ and it does contain another $z_{l_1}^-$ term, it must also contain a $c^{\{5\}}_{l_1}$ term, which has limit zero. So overall we have $\ddot{\mathfrak{m}}_{l,1}^- = \ddot{\mathfrak{m}}_{l,2}^- = \ddot{\mathfrak{c}}_{l,1;l,1}^{-;+} = \ddot{\mathfrak{c}}_{l,1;l,2}^{-;+} = \ddot{\mathfrak{c}}_{l,2;l,1}^{-;+} = \ddot{\mathfrak{c}}_{l,2;l,2}^{-;+} = \ddot{\mathfrak{c}}_{l,2;l,3}^{-;+} = 0$ as required.

Let's look at $\ddot{\mathfrak{c}}_{l,1;l,1}^{-;-}$, $\ddot{\mathfrak{c}}_{l,1;l,2}^{-;-}$, $\ddot{\mathfrak{c}}_{l,2;l,2}^{-;-}$ and $\ddot{\mathfrak{c}}_{l,2;l,3}^{-;-}$. Each of $\rho_{l,1}^-\rho_{l,1}^-$, $\rho_{l,1}^-\rho_{l,2}^-$, $\rho_{l,2}^-\rho_{l,2}^-$ and $\rho_{l,2}^-\rho_{l,3}^-$ is a sum of products of two pure terms. As usual, we consider the expectation limit of individual products by pure term type. A product of a short term and a long term as well as a product of a long term and another long term contains in its polynomial a factor stemming from $\widehat{e_{l_1}^-e_{l_2}^+} - \widehat{e_{l_1}^-}\widehat{e_{l_2}^+}$, where $l_1 < l$ and $l_2 < l$ are not necessarily different. We have

\begin{eqnarray*}
&&\widehat{e_{l_1}^-e_{l_2}^+} - \widehat{e_{l_1}^-}\widehat{e_{l_2}^+}\\
&=&\frac{1}{B^2}\Big(B(c^{\{2\}}_{l_1}z_{l_1}^+ + c^{\{3\}}_{l_1}z_{l_1}^-)c^{\{1\}}_{l_2}z_{l_2}^+ + B\sum_{b=2}^B(c^{\{7\}}_{l_1}z_{l_1}^+ + c^{\{8\}}_{l_1}z_{l_1}^- + c^{\{9\}}_{l_1}s_{l_1}^+ + c^{\{10\}}_{l_1}s_{l_1}^- \\
&&+ c^{\{12\}}_{l_1}u_{l_1}^+[b] + c^{\{13\}}_{l_1}u_{l_1}^-[b])(c^{\{4\}}_{l_2}z_{l_2}^+ + c^{\{5\}}_{l_2}z_{l_2}^- + c^{\{6\}}_{l_2}s_{l_2}^+ + c^{\{11\}}_{l_2}u_{l_2}^+[b])\\
&&-(c^{\{2\}}_{l_1}z_{l_1}^+ + c^{\{3\}}_{l_1}z_{l_1}^- + \sum_{b=2}^B(c^{\{7\}}_{l_1}z_{l_1}^+ + c^{\{8\}}_{l_1}z_{l_1}^- + c^{\{9\}}_{l_1}s_{l_1}^+ + c^{\{10\}}_{l_1}s_{l_1}^-\\
&&+ c^{\{12\}}_{l_1}u_{l_1}^+[b] + c^{\{13\}}_{l_1}u_{l_1}^-[b]))(c^{\{1\}}_{l_2}z_{l_2}^+ + \sum_{b=2}^B(c^{\{4\}}_{l_2}z_{l_2}^+ + c^{\{5\}}_{l_2}z_{l_2}^- + c^{\{6\}}_{l_2}s_{l_2}^+ + c^{\{11\}}_{l_2}u_{l_2}^+[b]))\Big)\\
&=&\frac{B-1}{B^2}(c^{\{2\}}_{l_1}z_{l_1}^+ + c^{\{3\}}_{l_1}z_{l_1}^- - c^{\{7\}}_{l_1}z_{l_1}^+ - c^{\{8\}}_{l_1}z_{l_1}^- - c^{\{9\}}_{l_1}s_{l_1}^+ - c^{\{10\}}_{l_1}s_{l_1}^-)(c^{\{1\}}_{l_2}z_{l_2}^+ - c^{\{4\}}_{l_2}z_{l_2}^+\\
&&- c^{\{5\}}_{l_2}z_{l_2}^- - c^{\{6\}}_{l_2}s_{l_2}^+) + \frac{B-1}{B}c^{\{12\}}_{l_1}c^{\{11\}}_{l_2}\widetilde{u_{l_1}^+u_{l_2}^+} + \frac{B-1}{B}c^{\{13\}}_{l_1}c^{\{11\}}_{l_2}\widetilde{u_{l_1}^-u_{l_2}^+}
\end{eqnarray*}

Here we use that $\widetilde{u} = 0$ for all $u$ vectors. Now we form aggregated chains in a 3-step process. First, we aggregate the ``partial chains'' obtained from multiplying out all of the polynomial except for the factor stemming from $\widehat{e_{l_1}^-e_{l_2}^+} - \widehat{e_{l_1}^-}\widehat{e_{l_2}^+}$. Second, we multiply a copy of each aggregated partial chain with $\frac{B-1}{B^2}(c^{\{2\}}_{l_1}z_{l_1}^+ + c^{\{3\}}_{l_1}z_{l_1}^- - c^{\{7\}}_{l_1}z_{l_1}^+ - c^{\{8\}}_{l_1}z_{l_1}^- - c^{\{9\}}_{l_1}s_{l_1}^+ - c^{\{10\}}_{l_1}s_{l_1}^-)(c^{\{1\}}_{l_2}z_{l_2}^+ - c^{\{4\}}_{l_2}z_{l_2}^+ - c^{\{5\}}_{l_2}z_{l_2}^- - c^{\{6\}}_{l_2}s_{l_2}^+)$, a copy of each aggregated partial chain with $\frac{B-1}{B}c^{\{12\}}_{l_1}c^{\{11\}}_{l_2}\widetilde{u_{l_1}^+u_{l_2}^+}$ and a copy of each aggregated partial chain with $\frac{B-1}{B}c^{\{13\}}_{l_1}c^{\{11\}}_{l_2}\widetilde{u_{l_1}^-u_{l_2}^+}$. Third, we form aggregated chains from each copy. Since the B-frequencies of aggregated partial chains  are convergent, the B-frequencies of aggregated chains stemming from the first copy converge to zero. Aggregated chains stemming from the second and third copy contain $c^{\{12\}}_{l_1}$ and $c^{\{13\}}_{l_1}$ respectively, which converge to zero. So for any type of aggregated chain, the expectation limit is zero.

This leaves products of a short term and another short term. If the big layer indices from the leading sum are different, all aggregated chains contain two single $u$ components or $s$ or $z$ terms, one each for the two big layer indices. The expectation limit is zero. If the big layer indices are both equal to some $l'$, the polynomial stems from $(e_{l',b} - \widehat{e_{l'}})(e_{l',b'} - \widehat{e_{l'}})$, where $(b, b')$ is one of $(1,1)$, $(1,2)$, $(2,2)$ and $(2,3)$ depending on the limit quantity considered. We have

\begin{eqnarray*}
&&e_{l',b} - \widehat{e_{l'}}\\
&=&(\mathbbm{1}_{b=1} - \frac{1}{B})c^{\{2\}}_{l'}z_{l'}^+ + (\mathbbm{1}_{b=1} - \frac{1}{B})c^{\{3\}}_{l'}z_{l'}^- + (\mathbbm{1}_{b\neq 1} - \frac{B-1}{B})c^{\{7\}}_{l'}z_{l'}^+\\
&& + (\mathbbm{1}_{b\neq 1} - \frac{B-1}{B})c^{\{8\}}_{l'}z_{l'}^- + (\mathbbm{1}_{b\neq 1} - \frac{B-1}{B})c^{\{9\}}_{l'}s_{l'}^+ + (\mathbbm{1}_{b\neq 1} - \frac{B-1}{B})c^{\{10\}}_{l'}s_{l'}^-\\
&&+ \mathbbm{1}_{b\neq 1}c^{\{12\}}_{l'}u_{l'}^+[b] + \mathbbm{1}_{b\neq 1}c^{\{13\}}_{l'}u_{l'}^-[b]
\end{eqnarray*}

There are a total of eight terms in this expression. An aggregated chain stems from the product of two of those terms. If one of those terms contains $c^{\{2\}}_{l'}$, $c^{\{7\}}_{l'}$, $c^{\{8\}}_{l'}$, $c^{\{9\}}_{l'}$, $c^{\{10\}}_{l'}$, $c^{\{12\}}_{l'}$ or $c^{\{13\}}_{l'}$, the chain-specific multiplier converges to zero. Otherwise, the B-frequency converges to zero unless $b=b'=1$. So overall we have $\ddot{\mathfrak{c}}_{l,1;l,2}^{-;-} = \ddot{\mathfrak{c}}_{l,2;l,2}^{-;-} = \ddot{\mathfrak{c}}_{l,2;l,3}^{-;-} = 0$, and endowing $u$ components and $s$ and $z$ terms with unit Gaussian distributions we have

\begin{eqnarray*}
&&\ddot{\mathfrak{c}}_{l,1;l,1}^{-;-}\\
&=&\sum_{f_{l'}}\Bigg(\frac{\limbn c_{l'}^2c_{l'}^{\{3\}}c_{l'}^{\{3\}}(1 - \frac{1}{B})^2}{\limbas \prod_{f_{l''}\in P_{l',l}} \nu_{l''}^+}\Bigg)\mathbb{E}_zz_{l'}^-z_{l'}^-\\
&=&\sum_{f_{l'}}\frac{c_{l'}^2\ddot{\mathfrak{c}}_{l',1;l',1}}{\prod_{f_{l''}\in P_{l',l}} \ddot{\nu}_{l''}}
\end{eqnarray*}

This is the same expression as for the case $m > 0$ at the beginning of stage 14 except that the recursive expansion cannot hit $f_m$, so again we have $\ddot{\mathfrak{c}}_{l,1;l,1}^{-;-} = \frac{\ddot{\mathfrak{c}}_{k,1;k,1}^{-;-}}{\ddot{\mathfrak{c}}_{k,1;k,1}-\ddot{\mathfrak{c}}_{k,1;k,2}+\epsilon_l}$ and $\limep \ddot{\mathfrak{c}}_{l,1;l,1}^{-;-} =\frac{\mathfrak{g}_l}{\mathfrak{g}_m}$ as required. By the induction hypothesis, we have $\ddot{\mathfrak{c}}_{k,1;k,1}^{-;-} > 0$ and by non-singularity from stage 9 we have $\ddot{\mathfrak{c}}_{k,1;k,1}-\ddot{\mathfrak{c}}_{k,1;k,2} \ge 0$, so $\ddot{\mathfrak{c}}_{l,1;l,1}^{-;-} > 0$ as required.

Finally, let's look at $\limbn B\mathfrak{c}_{l,2;l,2}^{-;-}$. We obtain this limit just as we obtained $\ddot{\mathfrak{c}}_{l,2;l,2}^{-;-} = \limbn \mathfrak{c}_{l,2;l,2}^{-;-}$, except that we absorb the additional $B$ factor either into the B-frequency or chain-specific multiplier before taking their limit, as detailed below. We proceed by pure term type as usual.

A product of a long term and another long term contains in its polynomial two factors stemming from $\widehat{e_{l_1}^-e_{l_2}^+} - \widehat{e_{l_1}^-}\widehat{e_{l_2}^+}$ and $\widehat{e_{l_3}^-e_{l_4}^+} - \widehat{e_{l_3}^-}\widehat{e_{l_4}^+}$ respectively. Both factors expand as shown above. As above, we form aggregated chains in a 3-step process. First, we aggregate ``partial chains'' from multiplying out the remaining factors. Second, we multiply a copy of each aggregated partial chain with 
{\small
$$\frac{B-1}{B^2}(c^{\{2\}}_{l_1}z_{l_1}^+ + c^{\{3\}}_{l_1}z_{l_1}^- - c^{\{7\}}_{l_1}z_{l_1}^+ - c^{\{8\}}_{l_1}z_{l_1}^- - c^{\{9\}}_{l_1}s_{l_1}^+ - c^{\{10\}}_{l_1}s_{l_1}^-)(c^{\{1\}}_{l_2}z_{l_2}^+ - c^{\{4\}}_{l_2}z_{l_2}^+ - c^{\{5\}}_{l_2}z_{l_2}^- - c^{\{6\}}_{l_2}s_{l_2}^+)...$$

$$...\frac{B-1}{B^2}(c^{\{2\}}_{l_3}z_{l_3}^+ + c^{\{3\}}_{l_3}z_{l_3}^- - c^{\{7\}}_{l_3}z_{l_3}^+ - c^{\{8\}}_{l_3}z_{l_3}^- - c^{\{9\}}_{l_3}s_{l_3}^+ - c^{\{10\}}_{l_3}s_{l_3}^-)(c^{\{1\}}_{l_4}z_{l_4}^+ - c^{\{4\}}_{l_4}z_{l_4}^+ - c^{\{5\}}_{l_4}z_{l_4}^- - c^{\{6\}}_{l_4}s_{l_4}^+)$$
}
and a copy of each aggregated partial chain with 

$$\frac{B-1}{B^2}(c^{\{2\}}_{l_1}z_{l_1}^+ + c^{\{3\}}_{l_1}z_{l_1}^- - c^{\{7\}}_{l_1}z_{l_1}^+ - c^{\{8\}}_{l_1}z_{l_1}^- - c^{\{9\}}_{l_1}s_{l_1}^+ - c^{\{10\}}_{l_1}s_{l_1}^-)...$$

$$...(c^{\{1\}}_{l_2}z_{l_2}^+ - c^{\{4\}}_{l_2}z_{l_2}^+ - c^{\{5\}}_{l_2}z_{l_2}^- - c^{\{6\}}_{l_2}s_{l_2}^+)(\frac{B-1}{B}c^{\{12\}}_{l_3}c^{\{11\}}_{l_4}\widetilde{u_{l_3}^+u_{l_4}^+} + \frac{B-1}{B}c^{\{13\}}_{l_3}c^{\{11\}}_{l_4}\widetilde{u_{l_3}^-u_{l_4}^+})$$

and a copy of each aggregated partial chain with 

$$(\frac{B-1}{B}c^{\{12\}}_{l_1}c^{\{11\}}_{l_2}\widetilde{u_{l_1}^+u_{l_2}^+} + \frac{B-1}{B}c^{\{13\}}_{l_1}c^{\{11\}}_{l_2}\widetilde{u_{l_1}^-u_{l_2}^+})\frac{B-1}{B^2}...$$

$$...(c^{\{2\}}_{l_3}z_{l_3}^+ + c^{\{3\}}_{l_3}z_{l_3}^- - c^{\{7\}}_{l_3}z_{l_3}^+ - c^{\{8\}}_{l_3}z_{l_3}^- - c^{\{9\}}_{l_3}s_{l_3}^+ - c^{\{10\}}_{l_3}s_{l_3}^-)(c^{\{1\}}_{l_4}z_{l_4}^+ - c^{\{4\}}_{l_4}z_{l_4}^+ - c^{\{5\}}_{l_4}z_{l_4}^- - c^{\{6\}}_{l_4}s_{l_4}^+)$$

and a copy of each aggregated partial chain with 

$$(\frac{B-1}{B}c^{\{12\}}_{l_1}c^{\{11\}}_{l_2}\widetilde{u_{l_1}^+u_{l_2}^+} + \frac{B-1}{B}c^{\{13\}}_{l_1}c^{\{11\}}_{l_2}\widetilde{u_{l_1}^-u_{l_2}^+})...$$

$$...(\frac{B-1}{B}c^{\{12\}}_{l_3}c^{\{11\}}_{l_4}\widetilde{u_{l_3}^+u_{l_4}^+} + \frac{B-1}{B}c^{\{13\}}_{l_3}c^{\{11\}}_{l_4}\widetilde{u_{l_3}^-u_{l_4}^+})$$

Third, we form aggregated chains from each copy. Since aggregated partial chains have convergent B-frequency, aggregated chains stemming from the first copy have a B-frequency that when multiplied with $B$ converges to zero, because of the two $\frac{B-1}{B^2}$ factors. Aggregated chains stemming from the second or third copy have a B-frequency that when multiplied with $B$ converges, because of the $\frac{B-1}{B^2}$ factor. Further, the chain-specific multiplier for those chains contains either a $c^{\{12\}}$ or $c^{\{13\}}$ term, so it converges to zero. Aggregated chains stemming from the fourth copy contain two $c^{\{12\}}$ terms, two $c^{\{13\}}$ terms or both terms. Hence, the multiplier for those chains times $B$ converges to zero. So no matter the copy, the expectation limit of the product of B-frequency, multiplier, normalizer and aggregated chain is zero.

A product of a short term and a long term contains in its polynomial a factor stemming from $\widehat{e_{l_1}^-e_{l_2}^+} - \widehat{e_{l_1}^-}\widehat{e_{l_2}^+}$ and a factor stemming from $e_{l_3,2}^- - \widehat{e_{l_3}^-}$. The first factor expands as above. The second factor expands as

$$\frac{1}{B}(c^{\{7\}}_{l_3}z_{l_3}^+ + c^{\{8\}}_{l_3}z_{l_3}^- + c^{\{9\}}_{l_3}s_{l_3}^+ + c^{\{10\}}_{l_3}s_{l_3}^- - c^{\{2\}}_{l_3}z_{l_3}^+ - c^{\{3\}}_{l_3}z_{l_3}^-) + c^{\{12\}}_{l_3}u_{l_3}^+[2] + c^{\{13\}}_{l_3}u_{l_3}^-[2]$$

We create aggregated partial chains as above, except that we do not replace batch index values of 2 and replace no batch index value with 2, as the expansion of $e_{l_3,2}^- - \widehat{e_{l_3}^-}$ lacks symmetry with respect to that batch index value. We create four copies of aggregated partial chains as above. For each copy, the argument that the resulting products of B-frequency, multiplier, normalizer and aggregated chain have expectation limit zero is also as above.

This leaves products of a short term and a short term. The polynomial stems from $(e_{l_1,2} - \widehat{e_{l_1}})(e_{l_2,2'} - \widehat{e_{l_2}})$. Each factor expands as above. Again, we create four copies and obtain expectation limits of zero. So overall $\limbn B\mathfrak{c}_{l,2;l,2}^{-;-}=0$ as required.

Case: $f_l$ is an activation layer.

The first expansion becomes 

\begin{eqnarray*}
&&\rho_{l,1}^- \\
&=&\sum_{f_{l'}\text{ normed}} c_{l'}\Bigg(\prod_{f_{l''}\in P_{l',l}}\frac{1}{\sqrt{\nu_{l''}^+}}\Bigg)(e_{l',1}^--\widehat{e_{l'}^-})\tau_l'(\rho_{k,1}^+)\\
&&+\sum_{f_{l'}\text{ normed}}\sum_{\mathcal{L} \in \mathsf{indpow}(P_{l',l}), Z=|\mathcal{L}|>0} c_{l'}(-1)^{|Z|}\Bigg(\prod_{f_{l''}\in P_{l',l}}\frac{1}{\sqrt{\nu_{l''}^+}}\Bigg)...\\
&&...\frac{(\widehat{e_{l'}^-\rho_{k_{\mathcal{L}[1]}}^+}-\widehat{e_{l'}^-}\widehat{\rho_{k_{\mathcal{L}[1]}}^+})(\rho^+_{k_{\mathcal{L}[Z]},1}-\widehat{\rho^+_{k_{\mathcal{L}[Z]}}})}{\nu_{\mathcal{L}[Z]}^+}\Bigg(\prod_{z=1}^{|Z|-1}\frac{\widehat{\rho_{k_{\mathcal{L}[z]}}^+\rho_{k_{\mathcal{L}[z+1]}}^+}-\widehat{\rho_{k_{\mathcal{L}[z]}}^+}\widehat{\rho_{k_{\mathcal{L}[z+1]}}^+}}{\nu_{\mathcal{L}[z]}^+}\Bigg)\tau_l'(\rho_{k,1}^+)\\
&&+\sum_{f_{l'}\text{ unnormed}} c_{l'}e_{l',1}^-\tau_l'(\rho_{k,1}^+)\\
\end{eqnarray*}

That is, we are left with only super terms 1, 2 and 5. Control holds as in the activation layer case of stage 12. The joint distribution of inputs is as in the BN layer case above except that $e_{l',1}^{+\text{in}}$ terms can arise inside of $\tau_l'()$ that are decomposed as $\sigma_{l'}^2s_{l'}^\text{in}$ as in stage 12, where $s_{l'}^\text{in}$ is a unit Gaussian scalar. The second expansion contains short terms and long terms as for the BN layer case above, which now also contain a $\tau_l'(\mathsf{exp}_{l,1})$ term where now 

\begin{eqnarray*}
\mathsf{exp}_{l,b}&=&\sum_{f_{l'} \text{ normed}}c_{l'}\Big(\prod_{f_{l''}\in P_{l',l}}\frac{1}{\sqrt{\nu_{l''}^+}}\Big)\Big((\mathbbm{1}_{b=1} - \frac{1}{B})c^{\{1\}}_{l'}z_{l'}^+  + (\mathbbm{1}_{b\neq 1} - \frac{B-1}{B})c^{\{4\}}_{l'}z_{l'}^+ \\
&& + (\mathbbm{1}_{b\neq 1} - \frac{B-1}{B})c^{\{5\}}_{l'}z_{l'}^- + (\mathbbm{1}_{b\neq 1} - \frac{B-1}{B})c^{\{6\}}_{l'}s_{l'}^+ + \mathbbm{1}_{b\neq 1}c^{\{11\}}_{l'}u_{l'}^+[b]\Big)\\
&&\sum_{f_{l'} \text{ unnormed}}c_{l'}\Big(\mathbbm{1}_{b=1}c^{\{1\}}_{l'}z_{l'}^+  + \mathbbm{1}_{b\neq 1}c^{\{4\}}_{l'}z_{l'}^+ + \mathbbm{1}_{b\neq 1} c^{\{5\}}_{l'}z_{l'}^-\\
&& + \mathbbm{1}_{b\neq 1}c^{\{6\}}_{l'}s_{l'}^+ + \mathbbm{1}_{b\neq 1}c^{\{11\}}_{l'}u_{l'}^+[b]\Big) + \sum_{f_{l'}\text{ bias}}c_{l'}\sigma_{l'}^2 s_{l'}^\text{in}
\end{eqnarray*}

Note that each $u$ component or $s$ or $z$ term that arises in this expression for a specific $b$ arises only once. In addition to short terms and long terms, the second expansion now also contains ``unnormed terms'' that are instances of the expression inside $\sum_{f_{l'}\text{ unnormed}}$. An unnormed term in $\rho_{l,b}^-$ for some $f_{l'}$ is 

$$c_{l'}\Big(\mathbbm{1}_{b=1}c^{\{2\}}_{l'}z_{l'}^+ + \mathbbm{1}_{b=1}c^{\{3\}}_{l'}z_{l'}^- + \mathbbm{1}_{b\neq 1}c^{\{7\}}_{l'}z_{l'}^+ + \mathbbm{1}_{b\neq 1} c^{\{8\}}_{l'}z_{l'}^- + \mathbbm{1}_{b\neq 1}c^{\{9\}}_{l'}s_{l'}^+$$
$$ + \mathbbm{1}_{b\neq 1}c^{\{10\}}_{l'}s_{l'}^- + \mathbbm{1}_{b\neq 1}c^{\{12\}}_{l'}u_{l'}^+[b] + \mathbbm{1}_{b\neq 1}c^{\{13\}}_{l'}u_{l'}^-[b]\Big)\tau_l'(\mathsf{exp}_{l,b})$$

Standard form terms can again contain $s$, $z$ and $c^{\{.\}}$ terms in the polynomial, but now they can also contain a tau-deriv $\tau_l'(\mathsf{exp}_{l,1})$. The application of lemma \ref{lemma18} is as in the activation layer case of stage 12 except (i) different Gaussian variables are present, (ii) no dependency of terms in $\mathsf{exp}_{l,b}$ on $N$ needs to be considered, (iii) chain aggregation works as in the BN layer case above and (iv) $c^{\{.\}}$ terms from the aggregated chain are absorbed into the multiplier. So the valid limit of the expectation of a standard form term is the sum of products of the limit of a B-frequency, the limit of a chain-specific multiplier, the almost sure limit of a normalizer, and the expectation of a product of $c^{\{.\}}$-free aggregated chain and tau-deriv with respect to the unit Gaussian, where the coefficients of the Gaussian variables in the tau-deriv are also replaced by their almost sure limit.

When investigating our limit quantities, as in the activation layer case of stage 12, we consider the expectation limit of products of two pure terms as well as products of a pure term with $\tau_l(\mathsf{exp}_{l,b})$. These products can differ from standard form terms in that they contain $\tau_l'(\mathsf{exp}_{l,b})\tau_l'(\mathsf{exp}_{l,b'})$ or $\tau_l'(\mathsf{exp}_{l,b})\tau_l(\mathsf{exp}_{l,b'})$ instead of just $\tau_l'(\mathsf{exp}_{l,b})$. $b$ and $b'$ depend on the limit quantity considered. In contrast to stage 12, if one of $b$ and $b'$ is equal to 1 and the other is unequal to 1, the coefficient of some Gaussian variables (namely $z^+$ terms) differs between the two $\tau$ terms. This means that we require a generalization of lemma \ref{lemma18} where we do not just have two functions $G$ and $H$, but a third function as well that is multiplied to the first two inside the expectation that has its own $\omega$ vector. Obtaining such a generalization is straightforward. The proof essentially requires replacing products of two functions with products of three functions and replacing statements about the components of two $\omega$ vectors with statements about the components of three $\omega$ vectors. In the statement of the generalization, we need to extend the list of functions that are assumed to be integrable with respect to any Gaussian measure, but this is still covered by assumption \ref{assumptionIntegrableMeanField}. The three $\omega$ vectors can then represent the normalizer, the coefficients in the first $\tau$ term and the coefficients in the second $\tau$ term respectively. $H(\mathsf{arg})$ then remains $\tau_l'(\sum_i\mathsf{arg}[i])$ and the third function becomes either $\tau_l'(\sum_i\mathsf{arg}[i])$ or $\tau_l(\sum_i\mathsf{arg}[i])$. In the end, we obtain that all $\limbn$ quantities of interest are valid as long as the additional $B$ factor in $\limbn B\mathfrak{c}_{l,2;l,2}^{-;-}$ does not cause divergence, which we show below.

$\ddot{\mathfrak{m}}_{l,1}^- = \ddot{\mathfrak{m}}_{l,2}^- = \ddot{\mathfrak{c}}_{l,1;l,1}^{-;+} = \ddot{\mathfrak{c}}_{l,1;l,2}^{-;+} = \ddot{\mathfrak{c}}_{l,2;l,1}^{-;+} = \ddot{\mathfrak{c}}_{l,2;l,2}^{-;+} = \ddot{\mathfrak{c}}_{l,2;l,3}^{-;+} = 0$ follows as for the BN layer case above when we also note that $z_{l_1}^-$ cannot have a non-zero coefficient in $\ddot{\mathsf{exp}_{l,b}}$ as $\ddot{c}^{\{5\}}_{l_1}=0$. 

Let's look at $\ddot{\mathfrak{c}}_{l,1;l,1}^{-;-}$, $\ddot{\mathfrak{c}}_{l,1;l,2}^{-;-}$, $\ddot{\mathfrak{c}}_{l,2;l,2}^{-;-}$, $\ddot{\mathfrak{c}}_{l,2;l,3}^{-;-}$. As in the BN layer case above, products involving only short and long terms become zero due to B-frequency or chain-specific multiplier becoming zero, except for the product of two short terms when $b=b'=1$. This leaves products involving an unnormed term. For a product of an unnormed term and a short or long term, the big layer indices from the leading sum differ. If $l'$ is the big layer index corresponding to the unnormed term, each aggregated chain contains exactly one copy of exactly one of $c^{\{2\}}_{l'}z_{l'}^+$, $c^{\{3\}}_{l'}z_{l'}^-$, $c^{\{7\}}_{l'}z_{l'}^+$, $c^{\{8\}}_{l'}z_{l'}^-$, $c^{\{9\}}_{l'}s_{l'}^+$, $c^{\{10\}}_{l'}s_{l'}^-$, $c^{\{12\}}_{l'}u_{l'}^+[b]$ or $c^{\{13\}}_{l'}u_{l'}^-[b]$. The expectation limit is zero if the chain contains any of the above except $c^{\{3\}}_{l'}z_{l'}^-$, due to the chain-specific multiplier becoming zero. If the aggregated chain contains $c^{\{3\}}_{l'}z_{l'}^-$ and it does not contain another $z_{l'}^-$ term, the expectation limit is zero due to the single $z_{l'}^-$ term, as $z_{l'}^-$ cannot have a non-zero coefficient in $\ddot{\mathsf{exp}_{l,b}}$ as $\ddot{c}^{\{5\}}_{l'}=0$.  Finally, if the aggregated chain contains $c^{\{3\}}_{l'}z_{l'}^-$ and it does contain another $z_{l'}^-$ term, it must also contain another $c^{\{5\}}_{l'}$ term, which has limit zero.

For a product of an unnormed term and an unnormed term, the polynomial stems from $e_{l_1,b}\tau_l(\rho_{k,b}^+)e_{l_2,b'}\tau_l(\rho_{k,b'}^+)$, where $(b, b')$ is one of $(1,1)$, $(1,2)$, $(2,2)$ and $(2,3)$ depending on the limit quantity considered. Expanding and multiplying out $e_{l_1,b}e_{l_2,b'}$ yields $\mathbbm{1}_{b=1}c^{\{3\}}_{l_1}z_{l_1}^-\tau_l'(\mathsf{exp}_{l,b})\mathbbm{1}_{b'=1}c^{\{3\}}_{l_2}z_{l_2}^-\tau_l'(\mathsf{exp}_{l,b'})$ as the only term that does not become zero due to its multiplier. If $(b,b') \neq (1,1)$, it is zero. If $l_1 \neq l_2$, it becomes zero due to the single $z_{l_1}^-$ and $z_{l_2}^-$ terms which cannot have a non-zero coefficient in $\ddot{\mathsf{exp}}_{l,b}$ or $\ddot{\mathsf{exp}}_{l,b'}$. So, overall we have $\ddot{\mathfrak{c}}_{l,1;l,2}^{-;-} = \ddot{\mathfrak{c}}_{l,2;l,2}^{-;-} = \ddot{\mathfrak{c}}_{l,2;l,3}^{-;-} = 0$. For $\ddot{\mathfrak{c}}_{l,1;l,1}^{-;-}$, combining the non-zero ``leftover'' from products of two short terms and products of two unnormed terms, endowing $u$ components and $s$ and $z$ terms with unit Gaussian distributions, we obtain

\begin{eqnarray*}
&&\ddot{\mathfrak{c}}_{l,1;l,1}^{-;-}\\
&=&\sum_{f_{l'}}\Bigg(\frac{c_{l'}^2\ddot{c}^{\{3\}}_{l'}\ddot{c}^{\{3\}}_{l'}}{\limbas \prod_{f_{l''}\in P_{l',l}} \nu_{l''}^+}\Bigg)\mathbb{E}_{u,s,z}z_{l'}^-z_{l'}^-\tau_l'(\ddot{\mathsf{exp}}_{l,1})^2\\
&=&\sum_{f_{l'}}\frac{c_{l'}^2\ddot{\mathfrak{c}}_{l',1;l',1}}{\prod_{f_{l''}\in P_{l',l}} \ddot{\nu}_{l''}}\mathbb{E}_{u,s,z}\tau_l'(\ddot{\mathsf{exp}}_{l,1})^2
\end{eqnarray*}

Here, we use again that $z_{l'}^-$ cannot have a non-zero coefficient in $\ddot{\mathsf{exp}}_{l,1}$. This is an analogous expression as for the case $m > 0$ at the beginning of stage 14 except that the recursive expansion cannot hit $f_m$. While the recursive expansion of the content of $\tau_l'()$ differs from the case $m > 0$, its limiting distribution is still $\mathcal{N}(0,\ddot{\mathfrak{c}}_{k,1;k,1})$. So again we have $\ddot{\mathfrak{c}}_{l,1;l,1}^{-;-} = \ddot{\mathfrak{c}}_{k,1;k,1}^{-;-}\mathbb{E}_{s\sim\mathcal{N}(0,\ddot{\mathfrak{c}}_{k,1;k,1})}\tau_l'(s)^2$ and $\limep \ddot{\mathfrak{c}}_{l,1;l,1}^{-;-}=\frac{\mathfrak{g}_{l}}{\mathfrak{g}_{m}}$ as required. By the induction hypothesis, we have $\ddot{\mathfrak{c}}_{k,1;k,1}^{-;-} > 0$, and by non-singularity from stage 9, we have $\ddot{\mathfrak{c}}_{k,1;k,1} > 0$. Along with proposition \ref{covkerPositive3} and condition \ref{aprop2}, we obtain $\ddot{\mathfrak{c}}_{l,1;l,1}^{-;-} > 0$ as required.

Finally, let's look at $\limbn B\mathfrak{c}_{l,2;l,2}^{-;-}$. A product of two pure terms contains in its polynomial two factors that, depending on pure term type, stem from $e_{l_1,2}^-$, $\widehat{e_{l_1}^-e_{l_2}^+} - \widehat{e_{l_1}^-}\widehat{e_{l_2}^+}$ or $e_{l_1,2}^- - \widehat{e_{l_1}^-}$. Expanding and multiplying out those two factors yields a sum of terms where each term contains two of the following: $\frac{B-1}{B^2}$ factor, $\frac{1}{B}$ factor, $c^{\{7\}}$ term, $c^{\{8\}}$ term, $c^{\{9\}}$ term, $c^{\{10\}}$ term, $c^{\{12\}}$ term, $c^{\{13\}}$ term. (This includes the possibility of having two copies of the same factor or term.) The same 3-step process for forming aggregated chains as in the BN layer case above yields $\limbn B\mathfrak{c}_{l,2;l,2}^{-;-} = 0$ as required.

\underline{Stage 15}: statement (\ref{eqn52p9})

Defining $\mathfrak{c}_{l,1;l,1}^{-;-}$ as in stage 14, we have

\begin{eqnarray*}
&&\limep \limbn \limd \frac{1}{d_l^\text{MF}}\mathbb{E}_X||\mathcal{J}_{l,m,1}(X)||^2_F\\
&=&\limep \limbn \limd\frac{1}{N}\sum_{n=1}^N\frac{1}{d_l^\text{MF}}||\mathcal{J}_{l,m,1}(X^{(n)})||^2_F\\
&=&\limep \limbn \frac{1}{N}\sum_{n=1}^N\limd\frac{1}{d_l^\text{MF}}||\mathcal{J}_{l,m,1}(X^{(n)})||^2_F\\
&=&\limep \limbn \frac{1}{N}\sum_{n=1}^N\begin{cases}\mathfrak{c}_{l,1;l,1}^{-;-} \text{ if } l>m\\1 \text{ else}\end{cases}\\
&=&\limep \limbn\begin{cases}\mathfrak{c}_{l,1;l,1}^{-;-} \text{ if } l>m\\1 \text{ else}\end{cases}\\
&=&\frac{\mathfrak{g}_l}{\mathfrak{g}_m}
\end{eqnarray*}

\underline{Stage 16}: statements (c) and (d)

Assume now that $f_{L,1}$ is a readout layer. Fix $N$. As in stage 8, we construct a mean field architecture $F$ with a single readout layer. The difference is that now we require multiple finite input layers. We proceed as in stage 9 with the following differences.

\begin{itemize}
\item Like $f_{L,1}$, $F_{L,1}$ is now a readout layer of dimensionality $d_L$. The $F_{L,b}$ for $b \ge 2$ are removed from $F$.
\item $F$ now has finite input layers $F_{0,b}$ which are jointly associated with $\mathcal{D}^B$ and the batch inputs $X^{(n)}$.
\item For a readin layer $f_l$ in $f$, instead of adding input layers $F_{l,b}$ to $F$ and extending $\mathcal{G}$, we add readin layers $F_{l,b}$ with the same width and variance parameter as the $f_{l,b}$ that share weights with each other but not other readin layers in $F$ and have dependency $F_{0,b}$ respectively.
\item $\mathcal{G}$ now only covers layers derived from bias big layers.
\end{itemize}

It is easy to see that, as in stage 9, we have distribution equality. For (c), this means that $f_{L,1}$ and $F_{L,1}$ have the same meta-distribution. For (b), this means that $(f_{L,1}(X^{(1)}), .., f_{L,1}(X^{(N)}))$ has the same distribution as $(F_{L,1}(X^{(1)}), .., F_{L,1}(X^{(N)}))$ for any fixed sample $X^{(1)}, .., X^{(N)}$. Note that as in the statement of background corollary \ref{backgroundKernel}, the surrogate input generated by $\mathcal{G}$ does not vary between batch inputs $X^{(n)}$, as it represents the random initial bias vectors.

In stage 8, we used theorem \ref{mfntMetaGaussian} and background corollary \ref{backgroundKernel} for statements (a) and (b) respectively. Now we require generalized versions that handle the case where there is more than one finite input layer. It is easy to obtain such generalizations. For background corollary \ref{backgroundKernel}, we simply need to replace $\mathcal{G}(K_\text{in})$ with the proper generator for the wider set of readin layers. The value of the covariance kernel for a pair $(X^{(n)},X^{(n')})$ then depends on all the square means and co-means of the individual inputs in $X^{(n)}$ and $X^{(n')}$. In the context of this stage, $\clipout(\clipin(F))^N$ has $NBI$ input layers that stem from readin layers in $F$, where $I$ is the number of readin big layers in $f$. The proper generator that yields distribution equality is $\mathcal{G}(K(NB, q, c))$, i.e. a mean zero Gaussian where the covariance matrix entry corresponding to $F_{l,b}^{(n)}$ and $F_{l',b'}^{(n')}$ is

$$\sigma_l\sigma_{l'}(\mathbbm{1}_{b=b' \text{ and } n=n'}q + \mathbbm{1}_{b\neq b' \text{ or } n\neq n'}c)\mathbbm{1}_{\text{$l=m$ or $f_l$ and $f_m$ share weights}}$$

Here, we use $n$ and $n'$ to represent duplex indices. As (d) only requires considering almost all samples, we can assume that all square means of individual inputs are $q$ and all co-means of individual inputs are $c$. As opposed to the general case, we can thus define a covariance kernel over only those two parameters.

We base the generalized theorem \ref{mfntMetaGaussian} on the generalized background corollary. We associate the finite input layers with $\mathcal{D}^B$ and replace $\mathcal{G}(K(N,q,c))$ with $\mathcal{G}(K(NB,q,c))$. The proof goes through as before. The conditions of the generalizations hold because $(\clipout(\clipin(F))^N, \mathcal{G}(K(NB,q,c)) \times \mathcal{G}^N)$ is the mean field architecture and generator we constructed early in stage 12 denoted as $(F^{(1)}, .., F^{(N)})$ for the case $m=L-1$. We proved the conditions in that stage.

So we have that (i) the meta-distribution of $F_{L,1}$ expansion-converges as $d_\text{MF} \rightarrow \infty$ to the elementwise meta-distribution with generator $\mathcal{MN}(\mathfrak{c}_{L,1;L,1}, \mathfrak{c}_{L,1;L,1}^{\prime;\prime\prime})$ and (ii) for almost all samples $(F_{L,1}(X^{(1)}), .., F_{L,1}(X^{(N)}))$ converges in distribution as $d_\text{MF} \rightarrow \infty$ to a Gaussian that is elementwise over output vectors where the generator has mean zero and covariance matrix $K(N, \mathfrak{c}_{L,1;L,1}, \mathfrak{c}_{L,1;L,1}^{\prime;\prime\prime})$. By distribution equality, the same holds for $f_{L,1}$. And, as we showed in stage 9 and 10, $\limep \limb \mathfrak{c}_{L,1;L,1} = \mathfrak{q}_L$ and $\limep \limb \mathfrak{c}_{L,1;L,1}^{\prime;\prime\prime} = \mathfrak{c}_L$. (The limit over $N$ was not used in stages 9 and 10.)

Finally, we note that when $f_{L,1}$ is a readout layer, statements (\ref{eqn52p1}) through (\ref{eqn52p11}) still hold when $l < L$ as $(f_0, .., f_{L-1})$ is a valid A-architecture without finite output layer.

\end{proof}

\subsection{Proposition \ref{mfntCNLC} part 2} \label{mfntCNLCbnsection}

\begin{repproposition}{mfntCNLC} (part 2; see section \ref{mfntPropositionssection} for part 1)

Let $f$ be an A-architecture and let $X^{(1)} = (x^{(1,1)}, .., x^{(1,B)})$ and  $X^{(2)} = (x^{(2,1)}, .., x^{(2,B)})$ be two batch inputs. Assume:

\begin{itemize}
\item $\mathbb{E}_ix^{(n,b)}[i]x^{(n,b)}[i]=q$ for some $q$ and all $1 \le n \le 2$ and $1 \le b \le B$
\item $\mathbb{E}_ix^{(n,b)}[i]x^{(n',b')}[i]=c$ for some $c$ and all $1 \le n, n', \le 2$, $1 \le b, b' \le B$,  except when $(n,b) = (n',b')$ or $(n,n',b,b')=(1,2,1,1)$
\item $\mathbb{E}_ix^{(1,1)}[i]x^{(2,1)}[i]=r$ for some $r$
\item $q > r \ge c \ge 0$
\item The activation functions used in $f$ are piecewise 5-differentiable.
\end{itemize}

If $f$ does not contain layer normalization but can contain batch normalization layers, we have

$$\lim_{\vec{\epsilon}\rightarrow 0}\Big(\lim_{B\rightarrow\infty}\big(\lim_{d_\text{MF}\rightarrow\infty}\mathbb{E}_if_{l,1}(X^{(1)})[i]f_{l,1}(X^{(2)})[i]=\mathfrak{r}_l  \text{ a.s.}\big)\Big)$$

where...

\begin{itemize}
\item ... $\vec{\epsilon}$ is the vector of all regularizers used by normalization layers in $f$.
\item ... $0 \le l \le L$
\item ... $\mathfrak{r}_l = \begin{cases} r \text{ parameter of inputs if $f_l$ is an input layer} \\  \sigma_l^2\mathfrak{r}_k \text{ if $f_l$ is an FC layer} \\ \mathfrak{C}_{\tau_l}(\mathfrak{q}_k,\mathfrak{r}_k) \text{ if $f_l$ is an activation layer} \\  \mathfrak{r}_k + \sigma_l^2 \text{ if $f_l$ is a bias layer} \\ \frac{\mathfrak{r}_k - \mathfrak{c}_k}{\mathfrak{q}_k - \mathfrak{c}_k} \text{ if $f_l$ is a BN layer} \\  \sum_{\kappa_l=1}^{K_l} w_{l,\kappa_l}^2\mathfrak{r}_{k_l[\kappa_l]} \text{ if $f_l$ is an addition layer} \end{cases}$
\item ... randomness is induced by $\theta$ and the $X^{(n)}$.
\end{itemize}

Define the covariance kernel $\mathfrak{C}(q,c,r)$ as the function that returns $\mathfrak{r}_L$ given $q$, $c$ and $r$. Then we further have

$$\mathfrak{n}(f,q,c) = \sqrt{\frac{\frac{d}{dq'}\mathfrak{C}(q,c,q')|_{q'=q}(q - c)}{\mathfrak{C}(q,c,q) - \mathfrak{C}(q,c,c)}}$$

\end{repproposition}

\paragraph{Remark} While we phrase part 2 of this proposition entirely in terms of the three-argument covariance kernel $\mathfrak{C}(q,c,r)$ for simplicity, it is worth noting that $\mathfrak{C}(q,c,q)$ and $\mathfrak{C}(q,c,c)$ are identical to $\mathfrak{C}(q,q)$ and $\mathfrak{C}(q,c)$ respectively, as used in part 1 of the proposition.

It is worth noting that the co-means stipulated by the proposition are achievable. Let $v^\text{all}$, $v^\text{1}$, $v^\text{(1,1)}$, .., $v^\text{(1,B)}$, $v^\text{(2,1)}$, .., $v^\text{(2,B)}$ be a set of orthonormal vectors. Then letting $x^{(n,b)} = \sqrt{c}v^\text{all} + \sqrt{r-c}v^\text{1} + \sqrt{q-r}v^{(n,b)}$ when $b=1$ and $x^{(n,b)} = \sqrt{c}v^\text{all} +\sqrt{q-c}v^{(n,b)}$ otherwise, we obtain the desired co-means.

\begin{proof}

Like the statement of the proposition, the proof has two parts. The first part is a continuation of the proof of theorem \ref{mfntPropagation} for the BN case given in section \ref{mfntPropagationBNsection}, i.e. it can be viewed as another stage. The second part of the proof proceeds along the lines of the proof of the first part of this proposition given in section \ref{mfntPropositionssection}, and is based only on the calculation rules for the $\mathfrak{r}_l$, $\mathfrak{q}_l$, $\mathfrak{c}_l$ and $\mathfrak{g}_l$ given above and in table \ref{tableNLCPropagation}. Because activation functions used are piecewise 5-differentiable, we can use theorem \ref{covkerCregular} in this proof.

\underline{Part 1}: We proceed along the lines of stage 10. The statement trivially holds for $l=0$. $F$ is constructed as in that stage, where again $F'_{l,b}$ corresponds to $f_{l,b}(X^{(1)})$ and $F''_{l,b}$ corresponds to $f_{l,b}(X^{(2)})$. $\mathcal{G}$ is similar. It is a Gaussian with mean zero and covariance matrix as follows. An entry corresponding to layers $F_{l,b}^\mathsf{prime1}$ and $F_{l',b'}^\mathsf{prime2}$ deriving from readin layers is 0 if $l \neq l'$ and is otherwise $q\sigma_l^2$, $r\sigma_l^2$ or $c\sigma_l^2$ depending on how $(\mathsf{prime1},\mathsf{prime2},b,b')$ matches up with the conditions of this proposition. An entry corresponding to $F_{l,b}^{\mathsf{prime1}\text{ in}}$ and $F_{l',b'}^{\mathsf{prime2}\text{ in}}$ deriving from bias layers is 0 if $l \neq l'$ and $\sigma_l^2$ otherwise. An entry corresponding to $F_{l,b}^\mathsf{prime1}$ and $F_{l',b'}^{\mathsf{prime2} \text{ in}}$ is zero. As in stage 10, we have distribution equality, control, $\mathfrak{c}^{\prime;\prime}_{l,1;l,1}=\mathfrak{c}^{\prime\prime;\prime\prime}_{l,1;l,1} = \mathfrak{c}_{l,1;l,1}$ and $\mathfrak{c}^{\prime;\prime}_{l,1;l,2}=\mathfrak{c}^{\prime\prime;\prime\prime}_{l,1;l,2} = \mathfrak{c}_{l,1;l,2}$ where $\mathfrak{c}_{l,1;l,1}$ and $\mathfrak{c}_{l,1;l,2}$ stem from stage 9. A difference to stage 10 is that now there is less symmetry between $F'$ and $F''$. Hence, we now have to consider $\mathbb{E}_e\rho'_{l,1}\rho''_{l,1}$, $\mathbb{E}_e\rho'_{l,1}\rho''_{l,2}$ and $\mathbb{E}_e\rho'_{l,2}\rho''_{l,2}$ separately. The second expression accounts for the case where one batch index is equal to 1 and the third expression accounts for the case where neither batch index is equal to 1. As part of the induction, we show (i) $\ddot{\mathfrak{c}}^{\prime;\prime\prime}_{l,1;l,2}=\ddot{\mathfrak{c}}^{\prime;\prime\prime}_{l,2;l,2}=\ddot{\mathfrak{c}}_{l,1;l,2}$, (ii) $\limep\ddot{\mathfrak{c}}^{\prime;\prime\prime}_{l,1;l,1} = \mathfrak{r}_l$,  (iii) $\ddot{\mathfrak{c}}_{l,1;l,1} > \ddot{\mathfrak{c}}_{l,1;l,1}^{\prime;\prime\prime} \ge \ddot{\mathfrak{c}}_{l,1;l,2}$ and (iv) $\mathfrak{q}_l > \mathfrak{r}_l \ge \mathfrak{c}_l$. We also use proposition \ref{mfntPositive} and non-singularity from stage 9.

\begin{itemize}
\item Case: $f_l$ is a readin layer. From table \ref{tableBackgroundPropagation}, we obtain $\mathfrak{c}^{\prime;\prime\prime}_{l,1;l,1}=r\sigma_l^2$ and $\mathfrak{c}^{\prime;\prime\prime}_{l,1;l,2}=\mathfrak{c}^{\prime;\prime\prime}_{l,2;l,2}=c\sigma_l^2=\mathfrak{c}_{l,1;l,2}$. Hence, the same relationships hold for the $\limbn$ and $\limep \limbn$ quantities, which yields (i). Also $\limep \ddot{\mathfrak{c}}^{\prime;\prime\prime}_{l,1;l,1}=r\sigma_l^2=\mathfrak{r}_l$, so (ii). (iii) follows from $q > r \ge c$. (iv) follows from table \ref{tableNLCPropagation}.
\item Case: $f_l$ is a fully-connected, non-readin layer. We have $\mathfrak{c}^{\prime;\prime\prime}_{l,1;l,1}=\sigma_l^2\mathfrak{c}^{\prime;\prime\prime}_{k,1;k,1}$, $\mathfrak{c}^{\prime;\prime\prime}_{l,1;l,2}=\sigma_l^2\mathfrak{c}^{\prime;\prime\prime}_{k,1;k,2}$ and $\mathfrak{c}^{\prime;\prime\prime}_{l,2;l,2}=\sigma_l^2\mathfrak{c}^{\prime;\prime\prime}_{k,2;k,2}$. The same relationships hold for the $\limbn$ and $\limep \limbn$ quantities, so (i) and (iii) carry over from $f_k$. Also $\limep \ddot{\mathfrak{c}}^{\prime;\prime\prime}_{l,1;l,1}=\sigma_l^2\mathfrak{r}_k=\mathfrak{r}_l$, so (ii). (iv) carries over from $f_k$ using table \ref{tableNLCPropagation}.
\item Case: $f_l$ is a bias layer. We have $\mathfrak{c}^{\prime;\prime\prime}_{l,1;l,1} = \mathbb{E}_e(e_{l,1}^{\prime\text{in}} + \rho_{k,1}')(e_{l,1}^{\prime\prime\text{in}} + \rho_{k,1}'') = \mathfrak{c}^{\prime;\prime\prime}_{k,1;k,1} + \sigma_l^2$ and similarly $\mathfrak{c}^{\prime;\prime\prime}_{l,1;l,2} = \mathfrak{c}^{\prime;\prime\prime}_{k,1;k,2} + \sigma_l^2$ and $\mathfrak{c}^{\prime;\prime\prime}_{l,2;l,2} = \mathfrak{c}^{\prime;\prime\prime}_{k,2;k,2} + \sigma_l^2$. The same relationships hold for the $\limbn$ and $\limep \limbn$ quantities, so (i) and (iii) carry over from $f_k$. Also $\limep \ddot{\mathfrak{c}}^{\prime;\prime\prime}_{l,1;l,1}=\mathfrak{r}_k + \sigma_l^2=\mathfrak{r}_l$, so (ii). (iv) carries over using table \ref{tableNLCPropagation}.
\item Case: $f_l$ is an BN layer. $\ddot{\mathfrak{c}}^{\prime;\prime\prime}_{l,1;l,1}$, $\ddot{\mathfrak{c}}^{\prime;\prime\prime}_{l,1;l,2}$ and $\ddot{\mathfrak{c}}^{\prime;\prime\prime}_{l,2;l,2}$ are obtained analogously to $\ddot{\mathfrak{c}}_{l,1;l,2}$ in stage 9 via lemma \ref{lemma18}, except that the components of $\chi_d$ are now pairs of $e'_{l',1}-\widehat{e'_{l'}}$ and $e''_{l',1}-\widehat{e''_{l'}}$, $e'_{l',1}-\widehat{e'_{l'}}$ and $e''_{l',2}-\widehat{e''_{l'}}$ and $e'_{l',2}-\widehat{e'_{l'}}$ and $e''_{l',2}-\widehat{e''_{l'}}$ respectively. In the latter two cases, this again yields a diagonal $\hat{\Sigma}$ with entries $\ddot{\mathfrak{c}}_{l',1;l',1} - \ddot{\mathfrak{c}}_{l',1;l',2}$, and hence $\ddot{\mathfrak{c}}^{\prime;\prime\prime}_{l,1;l,2}=\ddot{\mathfrak{c}}^{\prime;\prime\prime}_{l,2;l,2}=0=\ddot{\mathfrak{c}}_{l,1;l,2}$, so (i). But in the first case it yields a block-diagonal $\hat{\Sigma}$ with off-diagonal entries $\ddot{\mathfrak{c}}^{\prime;\prime\prime}_{l',1;l',1} - \ddot{\mathfrak{c}}_{l',1;l',2}$ in each block. Because $\ddot{\mathfrak{c}}_{l',1;l',1} > \ddot{\mathfrak{c}}^{\prime;\prime\prime}_{l',1;l',1} \ge \ddot{\mathfrak{c}}_{l',1;l',2}$, this is positive definite. Hence

$$\ddot{\mathfrak{c}}^{\prime;\prime\prime}_{l,1;l,1}=\sum_{f_{l'}} \frac{c_{l'}^2(\ddot{\mathfrak{c}}_{l',1;l',1}^{\prime;\prime\prime}-\ddot{\mathfrak{c}}_{l',1;l',2})}{\prod_{f_{l''}\in P_{l',l}}\ddot{\mathfrak{c}}_{k'',1;k'',1}-\ddot{\mathfrak{c}}_{k'',1;k'',2}+\epsilon_{l''}}$$

Using the calculation rules or BN, bias and addition layers, it is easy to check that this is equivalent to $\ddot{\mathfrak{c}}^{\prime;\prime\prime}_{l,1;l,1} = \frac{\ddot{\mathfrak{c}}^{\prime;\prime\prime}_{k,1;k,1}-\ddot{\mathfrak{c}}_{k,1;k,2}}{\ddot{\mathfrak{c}}_{k,1;k,1}-\ddot{\mathfrak{c}}_{k,1;k,2}+\epsilon_l}$. So $\limep \ddot{\mathfrak{c}}^{\prime;\prime\prime}_{l,1;l,1} = \frac{\mathfrak{r}_k - \mathfrak{c}_k}{\mathfrak{q}_k - \mathfrak{c}_k} = \mathfrak{r}_l$, so (ii). (iii) carries over from $f_k$ using also $\ddot{\mathfrak{c}}_{l,1;l,2}=0$. (iv) carries over using table \ref{tableNLCPropagation}.
\item Case: $f_l$ is an addition layer. Analogously to stage 10, $\mathfrak{c}^{\prime;\prime\prime}_{l,1;l,1} = \sum_{\kappa=1}^Kw_{\kappa}^2\mathfrak{c}_{k[\kappa],1;k[\kappa],1}^{\prime;\prime\prime}$, $\mathfrak{c}^{\prime;\prime\prime}_{l,1;l,2} = \sum_{\kappa=1}^Kw_{\kappa}^2\mathfrak{c}_{k[\kappa],1;k[\kappa],2}^{\prime;\prime\prime}$ and $\mathfrak{c}^{\prime;\prime\prime}_{l,2;l,2} = \sum_{\kappa=1}^Kw_{\kappa}^2\mathfrak{c}_{k[\kappa],2;k[\kappa],2}^{\prime;\prime\prime}$. The relationships hold for $\limbn$ and $\limep \limbn$ quantities. (i) and (iii) carry over from the dependencies and (ii) and (iv) carry over using table \ref{tableNLCPropagation}.
\item Case: $f_l$ is an activation layer. Again, $\ddot{\mathfrak{c}}^{\prime;\prime\prime}_{l,1;l,1}$, $\ddot{\mathfrak{c}}^{\prime;\prime\prime}_{l,1;l,2}$ and $\ddot{\mathfrak{c}}^{\prime;\prime\prime}_{l,2;l,2}$ are obtained analogously to $\mathfrak{c}_{l,1;l,2}$ in stage 9 via lemma \ref{lemma18}, except that the components of $\chi_d$ are slightly different. For example, for $\ddot{\mathfrak{c}}^{\prime;\prime\prime}_{l,1;l,2}$, they are pairs of $e'_{l',1}-\widehat{e'_{l'}}$ and $e''_{l',2}-\widehat{e''_{l'}}$ for $f_{l'}$ normed, pairs of $e'_{l',1}$ and $e''_{l',2}$ for $f_{l'}$ unnormed, as well as the $e^{\prime\text{in}}_{l',1}$. For $\ddot{\mathfrak{c}}^{\prime;\prime\prime}_{l,1;l,2}$ and $\ddot{\mathfrak{c}}^{\prime;\prime\prime}_{l,2;l,2}$, we obtain the same $\hat{\Sigma}$ as in stage 9, and so $\ddot{\mathfrak{c}}^{\prime;\prime\prime}_{l,1;l,2}=\ddot{\mathfrak{c}}^{\prime;\prime\prime}_{l,2;l,2}=\ddot{\mathfrak{c}}_{l,1;l,2}$, so (i).

For $\ddot{\mathfrak{c}}^{\prime;\prime\prime}_{l,1;l,1}$, the lemma yields

$$\ddot{\mathfrak{c}}_{l,1;l,1}^{\prime;\prime\prime}=\mathbb{E}_u\Bigg(\tau_l\Big(\sum_{f_{l'}\text{ normed}} \frac{c_{l'}u'_{l',1}}{\prod_{f_{l''}\in P_{l',l}}\sqrt{\ddot{\nu}_{l''}}}+\sum_{f_{l'} \text{ unnormed}}c_{l'} u'_{l',1} + \sum_{f_{l'} \text{ bias}}c_{l'} u_{l'}^\text{in}\Big)...$$

$$...\tau_l\Big(\sum_{f_{l'}\text{ normed}} \frac{c_{l'}u''_{l',1}}{\prod_{f_{l''}\in P_{l',l}}\sqrt{\ddot{\nu}_{l''}}}+\sum_{f_{l'} \text{ unnormed}}c_{l'} u''_{l',1} + \sum_{f_{l'} \text{ bias}}c_{l'} u_{l'}^\text{in}\Big)\Bigg)$$

$u'_{l',1}$ and $u''_{l',1}$ have mean zero, variance $\ddot{\mathfrak{c}}_{l',1;l',1}$ and covariance $\ddot{\mathfrak{c}}^{\prime;\prime\prime}_{l',1;l',1}$ if $f_{l'}$ is unnormed, and variance $\ddot{\mathfrak{c}}_{l',1;l',1}-\ddot{\mathfrak{c}}_{l',1;l',2}$ and covariance $\ddot{\mathfrak{c}}^{\prime;\prime\prime}_{l',1;l',1}-\ddot{\mathfrak{c}}_{l',1;l',2}$ if $f_{l'}$ is normed. Because $\ddot{\mathfrak{c}}_{l',1;l',1} > \ddot{\mathfrak{c}}^{\prime;\prime\prime}_{l',1;l',1}\ge \ddot{\mathfrak{c}}_{l',1;l',2}$, the corresponding $\hat{\Sigma}$ is positive definite. As in stage 9, by applying lemma \ref{lemma18} to $\mathbb{E}_e\rho_{k,1}'\rho_{k,1}'$ and $\mathbb{E}_e\rho_{k,1}'\rho_{k,1}''$, we obtain that the expressions inside the $\tau_l()$ have mean zero, variance $\ddot{\mathfrak{c}}_{k,1;k,1}$ and covariance $\ddot{\mathfrak{c}}_{k,1;k,1}^{\prime;\prime\prime}$. So $\ddot{\mathfrak{c}}_{l,1;l,1}^{\prime;\prime\prime}=\mathfrak{C}_{\tau_l}(\ddot{\mathfrak{c}}_{k,1;k,1}, \ddot{\mathfrak{c}}_{k,1;k,1}^{\prime;\prime\prime})$, and applying lemma \ref{lemma15} one more time and using $\mathfrak{q}_k > \mathfrak{r}_k \ge \mathfrak{c}_k \ge 0$, we obtain $\limep \ddot{\mathfrak{c}}_{l,1;l,1}^{\prime;\prime\prime}=\mathfrak{C}_{\tau_l}(\mathfrak{q}_k, \mathfrak{r}_k)=\mathfrak{r}_l$ as required. (ii) becomes $\mathfrak{C}_{\tau_l}(\ddot{\mathfrak{c}}_{k,1;k,1}, \ddot{\mathfrak{c}}_{k,1;k,1}) > \mathfrak{C}_{\tau_l}(\ddot{\mathfrak{c}}_{k,1;k,1}, \ddot{\mathfrak{c}}_{k,1;k,1}^{\prime;\prime\prime}) \ge \mathfrak{C}_{\tau_l}(\ddot{\mathfrak{c}}_{k,1;k,1}, \ddot{\mathfrak{c}}_{k,1;k,2})$ and (iv) becomes $\mathfrak{C}_{\tau_l}(\mathfrak{q}_k, \mathfrak{q}_k) > \mathfrak{C}_{\tau_l}(\mathfrak{q}_k, \mathfrak{r}_k) \ge \mathfrak{C}_{\tau_l}(\mathfrak{q}_k,\mathfrak{c}_k)$. These carry over from $f_k$ using theorem \ref{covkerCregular}(b) and proposition \ref{covkerPositive}.
\end{itemize}

\underline{Part 2}: All denominators that arise in this part are non-zero by proposition \ref{mfntPositive}. Denote by $\mathfrak{C}_l(q,c,r)$ the value of $\mathfrak{r}_l$ obtained from table \ref{tableNLCPropagation} as well as the calculation rules from the statement of this proposition when starting the recursion with $q$, $c$ and $r$. Then $\mathfrak{C}_L(q,c,r)=\mathfrak{C}(q,c,r)$. We will prove the more general statement

$$\mathfrak{n}_{l,0}(f,q,c) = \sqrt{\frac{\frac{d}{dq'}\mathfrak{C}_l(q,c,q')|_{q'=q}(q - c)}{\mathfrak{C}_l(q,c,q) - \mathfrak{C}_l(q,c,c)}}$$

Denote the right-hand side of this equation by $RHS_l$. It is easy to check that $\mathfrak{C}_l(q,c,q)=\mathfrak{q}_l$ and that $\mathfrak{C}_l(q,c,c)=\mathfrak{c}_l$. So the left-hand side of the equation is $\sqrt{\frac{\mathfrak{g}_l(\mathfrak{q}_0-\mathfrak{c}_0)}{\mathfrak{g}_0(\mathfrak{q}_l-\mathfrak{c}_l)}}$ and the right-hand side is $\sqrt{\frac{\frac{d}{dq'}\mathfrak{C}_l(q,c,q')|_{q'=q}(\mathfrak{q}_0-\mathfrak{c}_0)}{\mathfrak{q}_l-\mathfrak{c}_l}}$. Also $\mathfrak{g}_0=1$. So we simply need to show

$$\mathfrak{g}_l=\frac{d}{dq'}\mathfrak{C}_l(q,c,q')|_{q'=q}$$

Again, we proceed by induction on $l$ based on layer operation. The base case of our induction is $l=0$, where $\mathfrak{C}_l$ is the identity so both sides are equal to 1. For the induction step, we have $l > 0$ so $f_l$ is not an input layer. 

\begin{itemize}
\item Case: $f_l$ is an FC, bias, BN or activation layer. We have

$$\frac{d}{dq'}\mathfrak{C}_l(q,c,q')|_{q'=q}=\frac{d\mathfrak{r}_l}{dr}|_{r=q}=\frac{d\mathfrak{r}_l}{d\mathfrak{r}_k}|_{\mathfrak{r}_k=\mathfrak{q}_k}\frac{d\mathfrak{r}_k}{dr}|_{r=q}$$

$$=\frac{d\mathfrak{r}_l}{d\mathfrak{r}_k}|_{\mathfrak{r}_k=\mathfrak{q}_k}\frac{d}{dq'}\mathfrak{C}_k(q,c,q')|_{q'=q}=\frac{d\mathfrak{r}_l}{d\mathfrak{r}_k}|_{\mathfrak{r}_k=\mathfrak{q}_k}\mathfrak{g}_k$$

So we must show that $$\frac{\mathfrak{g}_l}{\mathfrak{g}_k}=\frac{d\mathfrak{r}_l}{d\mathfrak{r}_k}|_{\mathfrak{r}_k=\mathfrak{q}_k}$$

The $\frac{d\mathfrak{r}_l}{d\mathfrak{r}_k}|_{\mathfrak{r}_k=\mathfrak{q}_k}$ term is $\sigma_l^2$ for FC, 1 for bias, and $\frac{1}{\mathfrak{q}_k -\mathfrak{c}_k}$ for BN. It is easy to check that this equals $\frac{\mathfrak{g}_l}{\mathfrak{g}_k}$. If $f_l$ is an activation layer, we have $\frac{d\mathfrak{r}_l}{d\mathfrak{r}_k}|_{\mathfrak{r}_k=\mathfrak{q}_k}=\frac{d}{dq'}\mathfrak{C}_{\tau_l}(\mathfrak{q}_k,q')|_{q'=\mathfrak{q}_k}$. This is the same expression we encountered in the proof of the first part of proposition \ref{mfntCNLC} and so again we have $\frac{d\mathfrak{r}_l}{d\mathfrak{r}_k}|_{\mathfrak{r}_k=\mathfrak{q}_k} = \frac{\mathfrak{g}_l}{\mathfrak{g}_k}$.
\item Case: $f_l$ is an addition layer. We have

$$\frac{d}{dq'}\mathfrak{C}_l(q,c,q')|_{q'=q}=\frac{d\mathfrak{r}_l}{dr}|_{r=q}=\sum_{\kappa=1}^K\frac{d\mathfrak{r}_l}{d\mathfrak{r}_{k[\kappa]}}|_{\mathfrak{r}_{k[\kappa]}=\mathfrak{q}_{k[\kappa]}}\frac{d\mathfrak{r}_{k[\kappa]}}{dr}|_{r=q}$$
$$=\sum_{\kappa=1}^Kw_\kappa^2\frac{d}{dq'}\mathfrak{C}_{k[\kappa]}(q,c,q')|_{q'=q}=\sum_{\kappa=1}^Kw_\kappa^2\mathfrak{g}_{k[\kappa]}=\mathfrak{g}_l$$

\end{itemize}
\end{proof}

\appendix

\chapter{Full list of study A and B architectures}\label{fullListChapter}

\section{Full list of study A architectures} \label{fullListA}
\setlength{\tabcolsep}{1pt}
{\small
\begin{longtable}{cccccccccccccccccccc}
\rot{Dataset}&\rot{Depth}&\rot{Width}&\rot{Activation function}&\rot{Activation function dilation ($d$)}&\rot{Activation function shift ($t$)}&\rot{Has activation function debiasing}&\rot{Weight matrix scaling factor}&\rot{Has nonzero initial bias vectors}&\rot{Normalization layer}&\rot{Skip connection start}&\rot{Skip connection addition weight}&\rot{NLC before training}&\rot{LBIAS before training}&\rot{Best starting learning rate}&\rot{Test error}&\rot{Best starting learning rate (train-opt)}&\rot{Training error (train-opt)}&\rot{Is GUA}&\rot{Is GEA}\\
\hline\hline\endhead 

C&25&191&SELU&0.8&0.0&\checkmark&1.0&\checkmark&none&none&&1.68e0&1.52e0&5.91e0&0.49&9.42e-2&0.00&\xmark&\xmark\\
C&3&671&Gaussian&0.8&0.2&\xmark&0.9&&BN&norm.&1.00&1.15e0&1.00e0&6.08e2&0.45&8.72e1&0.00&\xmark&\xmark\\
C&45&143&Gaussian&0.8&0.0&\xmark&1.1&&BN&norm.&1.00&3.12e1&1.00e0&1.78e2&0.56&2.55e1&0.01&\xmark&\xmark\\
C&39&154&ReLU&1.0&0.2&\checkmark&0.9&\xmark&LN&norm.&0.23&3.40e2&1.00e0&2.80e0&0.57&6.40e5&0.00&\xmark&\xmark\\
C&45&143&sigmoid&1.0&0.2&\xmark&1.0&&BN&norm.&1.00&1.12e0&1.00e0&1.31e2&0.48&1.88e1&0.00&\xmark&\xmark\\
C&19&220&even tanh&0.8&0.0&\checkmark&1.1&\xmark&none&none&&1.79e6&1.01e0&&0.89&2.41e-11&0.00&\xmark&\xmark\\
C&27&184&SELU&0.8&0.0&\checkmark&0.9&\checkmark&LN&none&&1.78e0&1.93e0&2.00e1&0.48&9.56e-1&0.00&\xmark&\xmark\\
C&39&154&SELU&1.0&0.2&\xmark&0.9&&BN&add.&1.00&1.21e0&1.00e0&1.01e3&0.48&1.61e1&0.00&\xmark&\xmark\\
\end{longtable}
}

\newpage

\section{Full list of study B architectures}\label{fullListB}
{
\begin{longtable}{cccccccccccccccccc}
\rot{AFLM}&\rot{Has act. fun. debiasing}&\rot{Linearization parameter ($l$)}&\rot{Has trainable bias / scaling vectors}&\rot{Data processing}&\rot{Uses data augmentation}&\rot{Weight initialization}&\rot{Pooling}&\rot{Uses global average pooling}&\rot{Is residual}&\rot{Normalization layer}&\rot{NLC before training}&\rot{LBIAS before training}&\rot{Best starting learning rate}&\rot{Test error}&\rot{Best starting learning rate (train-opt)}&\rot{Training error (train-opt)}&\rot{Is GUA}\\
\hline\hline\endhead

ReLU-interp.&\xmark&0&\checkmark&PC&\xmark&DG&avg&\xmark&\xmark&none&1.30e0&1.64e0&1.00e-4&0.32&1.00e-4&0.00&\xmark\\
ReLU-interp.&\xmark&0&\checkmark&PC&\xmark&DG&avg&\xmark&\xmark&BN&2.22e1&1.11e0&3.00e-3&0.26&3.00e-3&0.00&\xmark\\
ReLU-interp.&\xmark&0&\checkmark&PC&\xmark&DG&avg&\xmark&\xmark&LN&1.44e0&1.54e0&1.00e-4&0.29&3.00e-4&0.00&\xmark\\
ReLU-interp.&\xmark&0&\checkmark&PC&\xmark&DG&avg&\xmark&\checkmark&BN&3.79e0&1.16e0&1.00e-2&0.21&1.00e-3&0.00&\xmark\\
ReLU-interp.&\checkmark&0&\checkmark&PC&\xmark&DG&avg&\xmark&\xmark&none&4.76e0&1.01e0&3.00e-5&0.36&1.00e-4&0.00&\xmark\\
ReLU-interp.&\checkmark&0&\checkmark&PC&\xmark&DG&avg&\xmark&\xmark&BN&2.22e1&1.02e0&1.00e-2&0.26&1.00e-3&0.00&\xmark\\
ReLU-interp.&\checkmark&0&\checkmark&PC&\xmark&DG&avg&\xmark&\xmark&LN&1.13e1&1.06e0&3.00e-3&0.29&3.00e-4&0.00&\xmark\\
ReLU-interp.&\checkmark&0&\checkmark&PC&\xmark&DG&avg&\xmark&\checkmark&BN&4.02e0&1.01e0&1.00e-2&0.22&1.00e-3&0.00&\xmark\\
\end{longtable}
}

\chapter*{Attribution, Acknowledgments and Disclosure of Funding}
\addcontentsline{toc}{chapter}{Acknowledgments and Disclosure of Funding}

This work is based on the PhD thesis with the same name, author, year and institution \citep{thesis}. Both works may be cited interchangeably.

We thank Jaime G. Carbonell\footnote{Deceased 2020; before: School of Computer Science, Carnegie Mellon University, Pittsburgh, PA 15213, USA} and Sergey Ioffe\footnote{Google AI Perception, Mountain View, CA 94043, USA} for in-depth discussions.

This research was sponsored by the National Science Foundation under grant number DBI0546594, the National Institute of Health under grant numbers 1R01GM087694, 5P30DA03577802, and R01GM114311, the Defense Advanced Research Projects Agency under grant number FA87501220324, and Boeing under grant number A0178422015002U.  The views and conclusions contained in this document are those of the author and should not be interpreted as representing the official policies, either expressed or implied, of any sponsoring institution, the U.S. government or any other entity.

\bibliographystyle{plainnat}

\bibliography{citations}

\end{document}